# Artificial Intelligence Index Report

2026

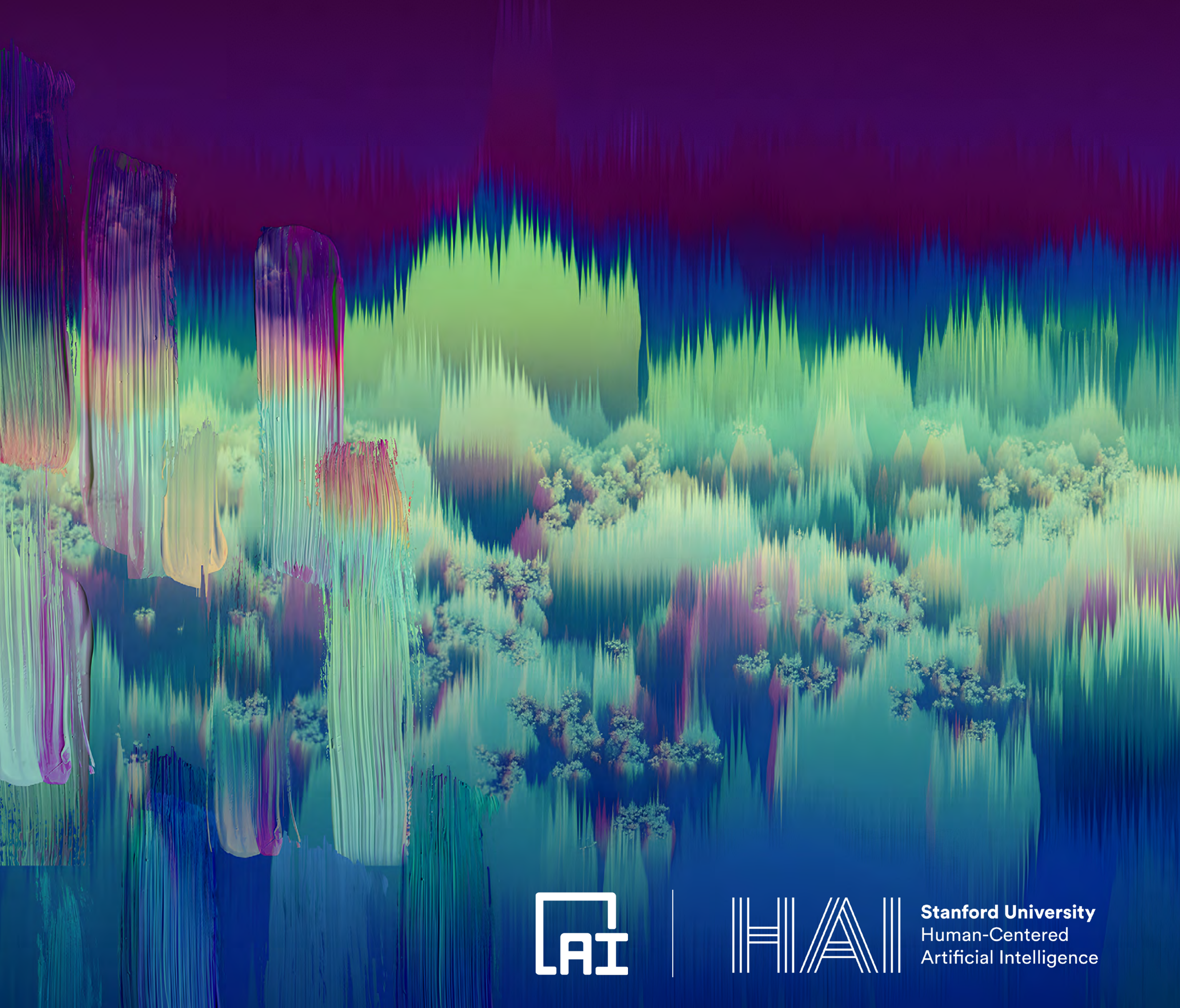

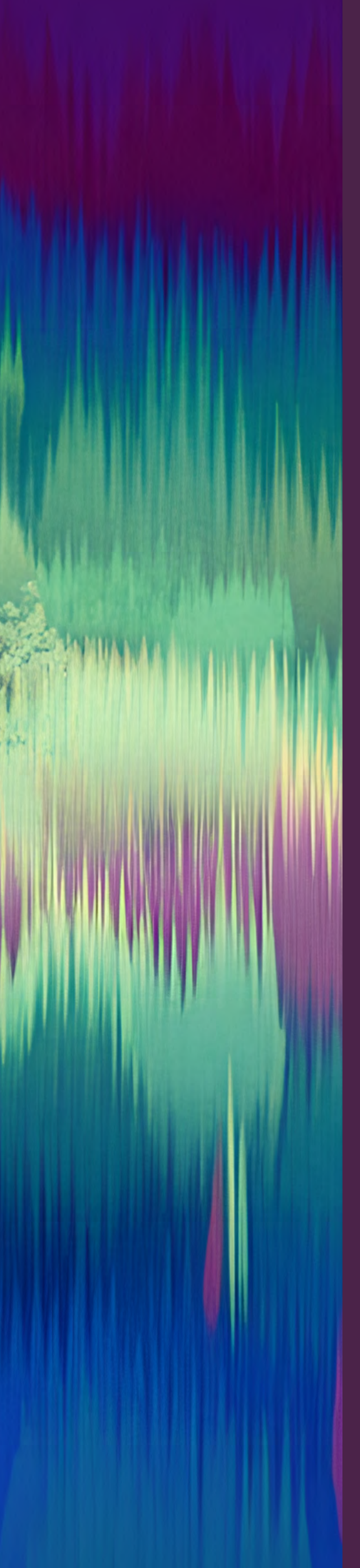

# Contents

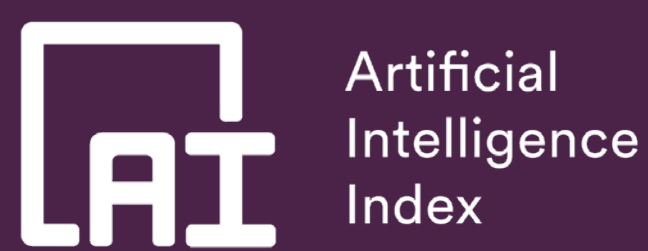
Artificial
Intelligence
Index

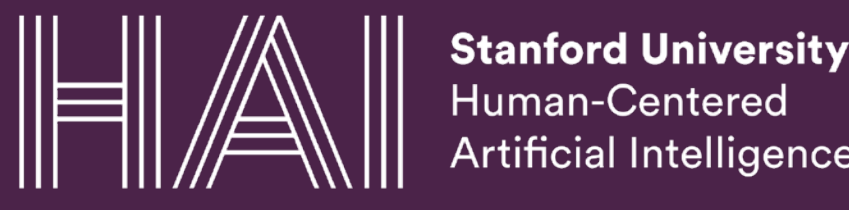
Stanford University
Human-Centered
Artificial Intelligence

# Introduction

Welcome to the ninth edition of the AI Index report. As AI continues to advance rapidly, the question becomes whether the systems built around it can keep up. Governance frameworks, evaluation methods, education systems, and the data infrastructure needed to track AI's impact are struggling to match the pace of the technology itself. That gap—between what AI can do and how prepared we are to manage it—runs through every chapter of this year's report. New in this edition, the report tracks how AI is being tested more ambitiously across reasoning, safety, and real-world task execution, and why those measurements are increasingly difficult to rely on. It also features new estimates of generative AI's economic value alongside emerging evidence of its labor market effects, an analytical framework on AI sovereignty, and a science chapter developed in collaboration with Schmidt Sciences. For the first time, the report features standalone chapters on AI in science and AI in medicine, reflecting AI's growing impact across these two domains.

For close to a decade, the AI Index has worked to bring reliable global data to a field that is evolving faster than most efforts to measure it. The report equips policymakers, researchers, executives, journalists, and the public with the necessary evidence to make informed decisions about AI. As the technology moves deeper into classrooms, clinics, and legislatures—and reshapes how people work, learn, and govern—the cost of incomplete data continues to rise.

In a field where much data is produced by organizations with a stake in the technology's success, the demand for neutral and rigorous measurement continues to grow. The AI Index remains independent and focused on revealing the long-term patterns underneath the headlines. The report is relied on by governments, research institutions, and companies around the world, and referenced by media outlets and in academic papers.

The pages that follow offer the most comprehensive, independently sourced picture of AI's trajectory that is available. They also make clear where that picture remains incomplete—because what we cannot yet measure matters just as much as what we can.

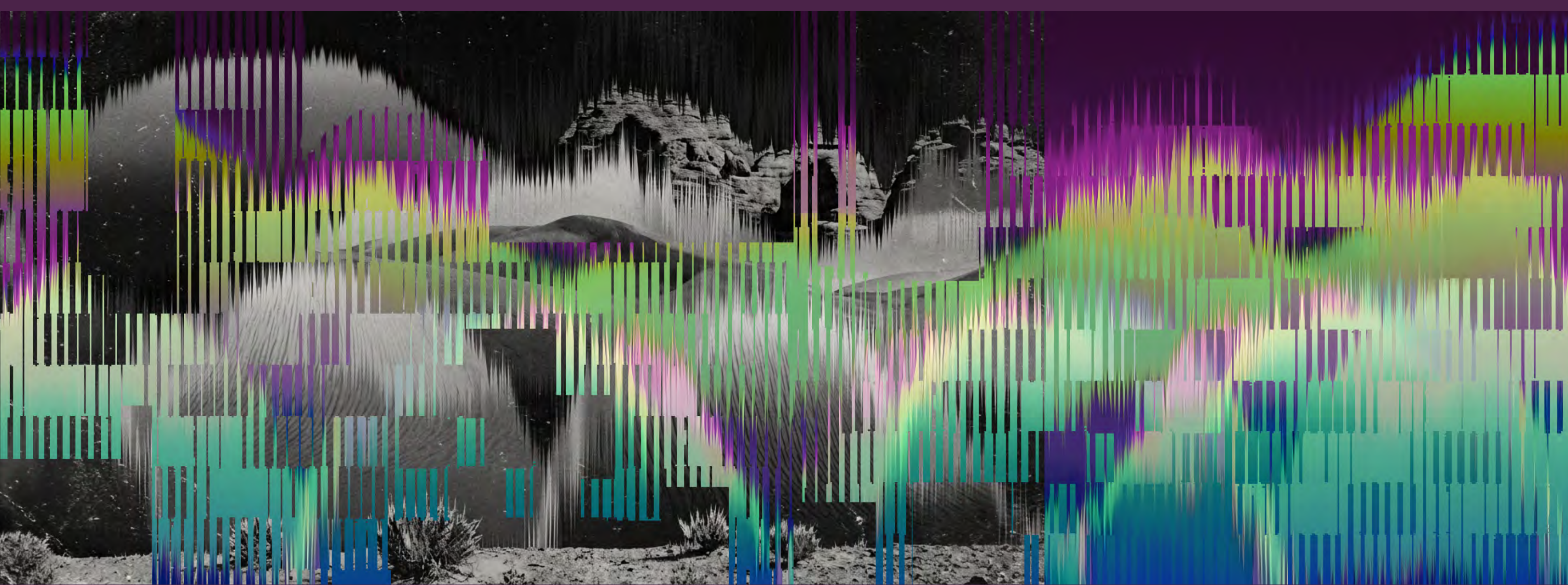

# Message from the Co-chairs

A year ago, this report documented AI's arrival as a mainstream force. This year's data shows what happens after arrival.

This is a technology that has reached mass adoption faster than the personal computer or the internet. Generative AI hit nearly 53% population-level adoption within three years. Leading AI companies are reaching meaningful revenue scale in a fraction of the time it took previous technology generations, and global corporate investment more than doubled in 2025. Organizational adoption rose to 88%, and early estimates suggest the consumer value of generative AI has grown substantially within a year.

At the technical frontier, leading models are now nearly indistinguishable from one another. Open-weight models are more competitive than ever. But as models converge, the tools used to evaluate them are struggling to stay relevant. Benchmarks are saturating, frontier labs are disclosing less, and independent testing does not always confirm what developers report.

The chapters that follow trace what this scale of activity and capability means in practice. In science, AI shifted from accelerating individual research steps to attempting full replacement of entire workflows. In medicine, clinical AI tools moved from pilot programs to broader deployment, with systems like ambient AI scribes scaling across health systems.

Governments around the world acted on AI in 2025, but not in the same direction. The EU AI Act's first prohibitions took effect, while the United States shifted toward deregulation. Japan, South Korea, and Italy each passed national AI laws, and more than half of newly adopted national AI strategies came from developing countries entering the policy landscape for the first time. AI sovereignty emerged as a central organizing principle across all of these efforts. The public is also navigating competing signals. Global optimism about AI rose in 2025, but so did nervousness.

The data does not point in a single direction. It reveals a field that is scaling faster than the systems around it can adapt. We encourage you to explore and decide for yourself.

**Yolanda Gil and Raymond Perrault**
Co-chairs, AI Index Report

# Steering Committee

## CHAIR

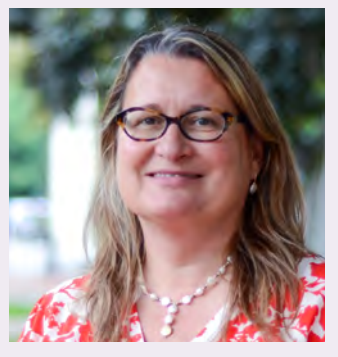

**YOLANDA GIL**
University of Southern California, Information Sciences Institute

## CO-CHAIR

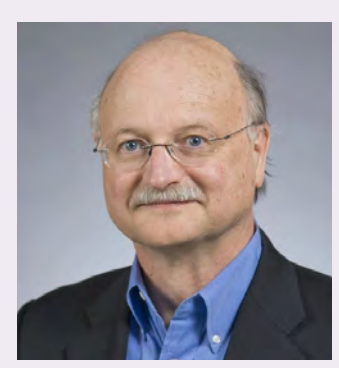

**RAYMOND PERRAULT**
SRI International

## MEMBERS

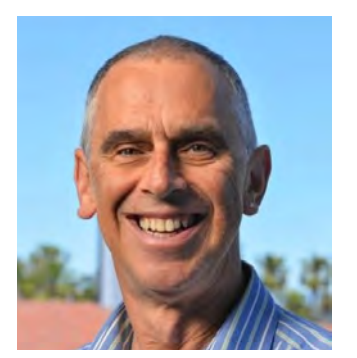

**RUSS ALTMAN**
Stanford University

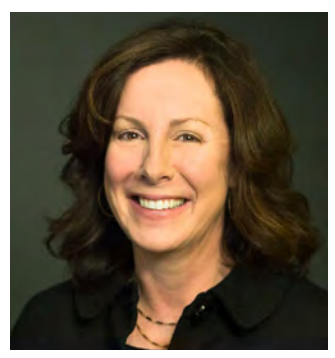

**CARLA BRODLEY**
Northeastern University

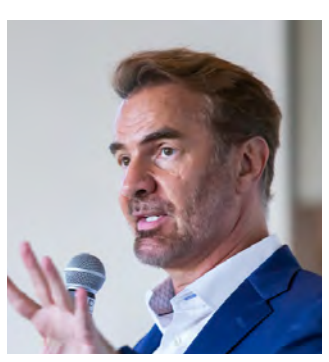

**ERIK BRYNJOLFSSON**
Stanford University

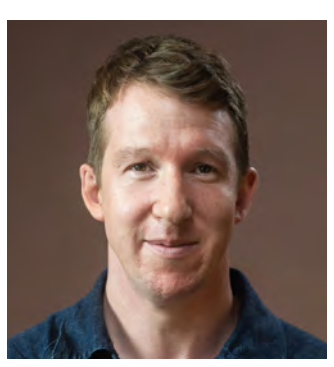

**JACK CLARK**
Anthropic, OECD

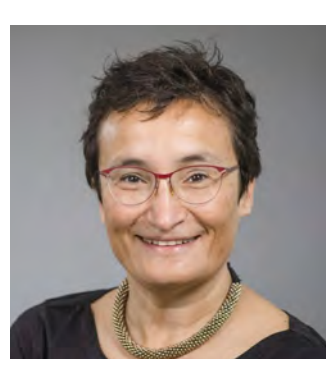

**VIRGINIA DIGNUM**
Umeå University

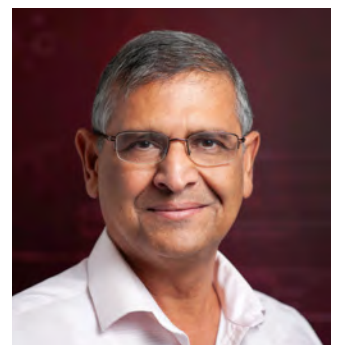

**VIPIN KUMAR**
University of Minnesota

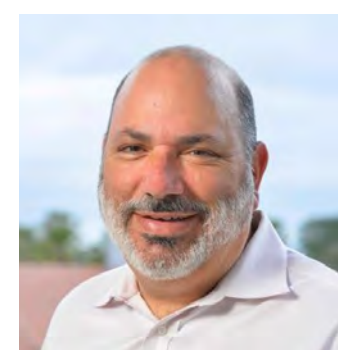

**JAMES LANDAY**
Stanford University

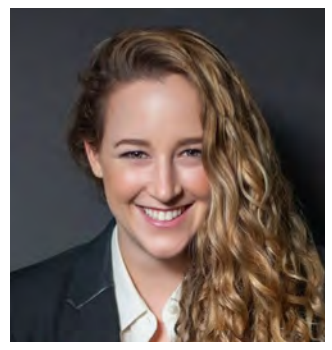

**TERAH LYONS**
JPMorgan Chase & Co.

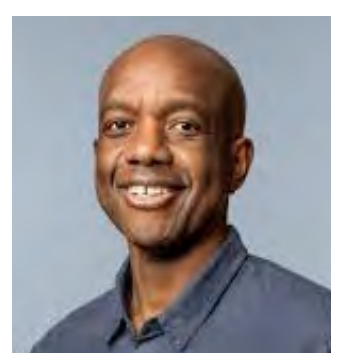

**JAMES MANYIKA**
Google, University of Oxford

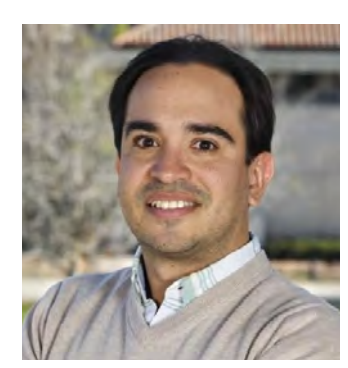

**JUAN CARLOS NIEBLES**
Stanford University, Salesforce

**VANESSA PARLI**
Stanford University

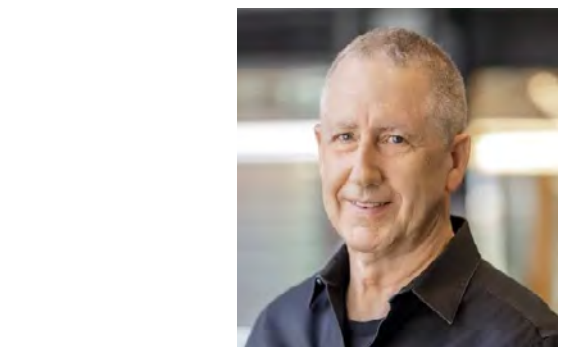

**YOAV SHOHAM**
Stanford University, AI21 Labs

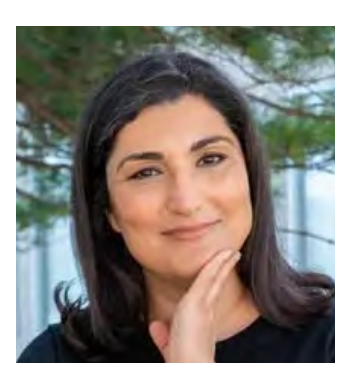

**ELHAM TABASSI**
Brookings

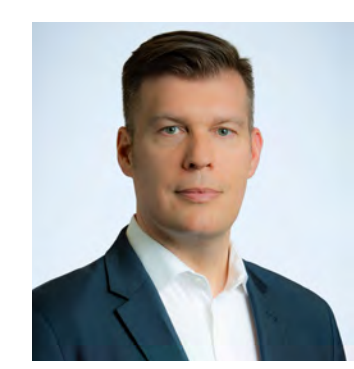

**RUSSELL WALD**
Stanford University

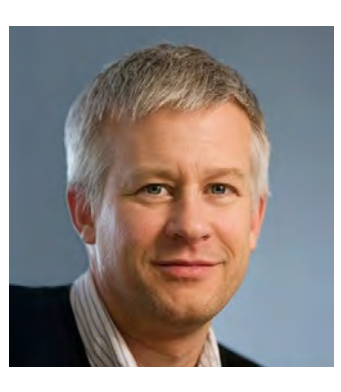

**DAN WELD**
University of Washington

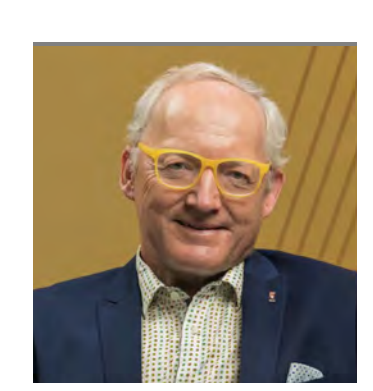

**TOBY WALSH**
UNSW Sydney

# Staff and Researchers

**LEAD AND EDITOR-IN-CHIEF**

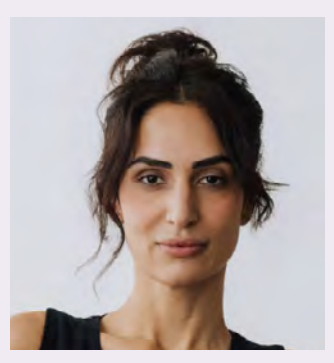

**SHA SAJADIEH**
Stanford University

**RESEARCH MANAGER**

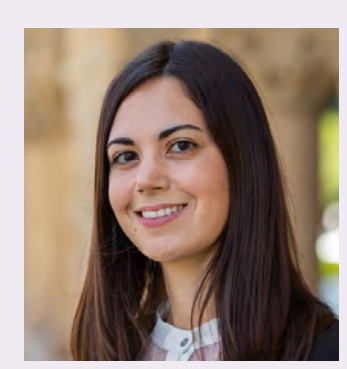

**LOREDANA FATTORINI**
Stanford University

**AFFILIATED RESEARCHERS**

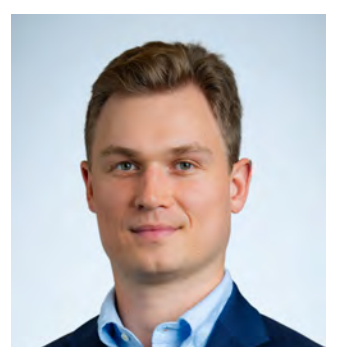

**NESTOR MASLEJ**
Stanford University

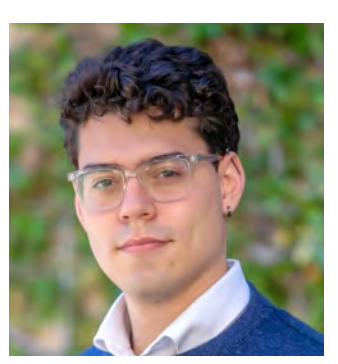

**JUAN NICOLAS PAVA**
Stanford University

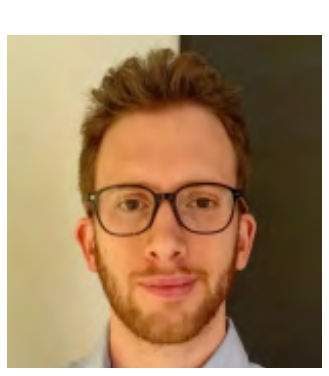

**LAPO SANTARLASCI**
IMT School for Advanced Studies Lucca

**UNDERGRADUATE RESEARCHER**

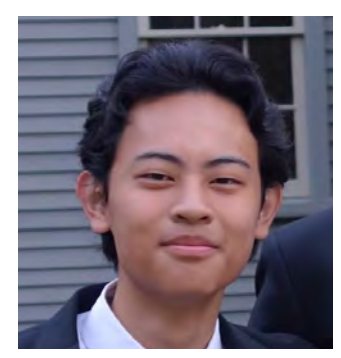

**HENRY ZHANG**
Stanford University

**GRADUATE RESEARCHER**

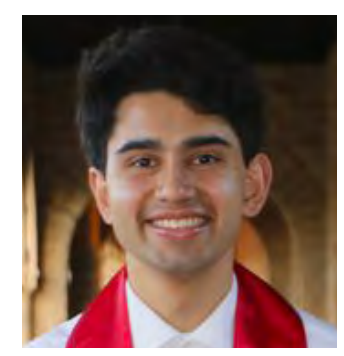

**SUKRUT OAK**
Stanford University

# How to Cite This Report

Sha Sajadieh, Loredana Fattorini, Raymond Perrault, Yolanda Gil, Vanessa Parli, Lapo Santarlasci, Juan Pava, Nestor Maslej, Russ Altman, Erik Brynjolfsson, Carla Brodley, Jack Clark, Virginia Dignum, Vipin Kumar, James Landay, Terah Lyons, James Manyika, Juan Carlos Niebles, Yoav Shoham, Elham Tabassi, Russell Wald, Toby Walsh, Dan Weld. "The AI Index 2026 Annual Report," AI Index Steering Committee, Institute for Human-Centered AI, Stanford University, Stanford, CA, April 2026.

# Public Data and Tools

The AI Index 2026 Report is supplemented by raw data and an interactive tool. We invite each reader to use the data and the tool in a way most relevant to their work and interests.

- Raw data and charts: The public data and high-resolution images of all the charts in the report are available on Google Drive.
- Global AI Vibrancy Tool: Compare the AI ecosystems of 36 countries. The Global AI Vibrancy tool will be updated by the end of 2026.

The AI Index welcomes feedback and new ideas for next year. Contact us at shahed@stanford.edu.

The AI Index is written by a team of human researchers. The authors used ChatGPT and Claude to help refine and copy edit drafts. All images in this publication were generated with AI by Johanna Friedman (2026), Gemini 3.1 (W-Nanobanana2), Gemini 3 (W-Nanobanana Pro).

## Supporting Partners

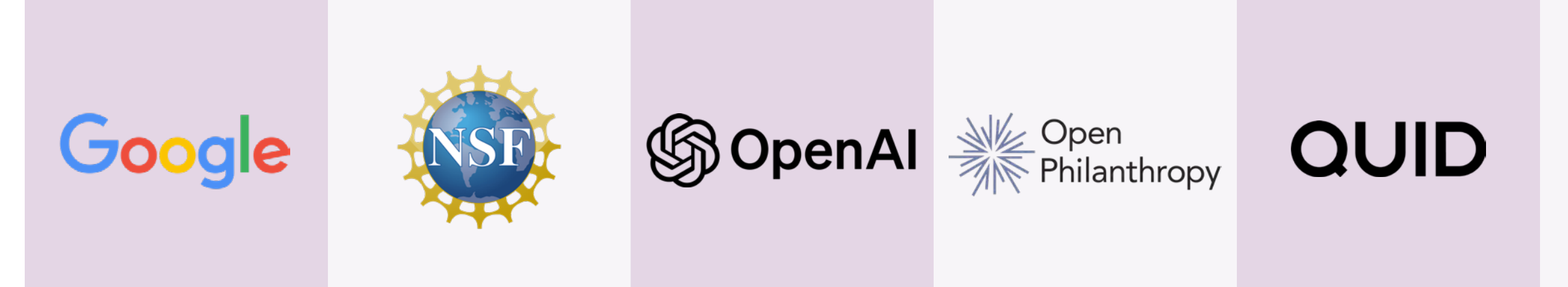

## Analytics and Research Partners

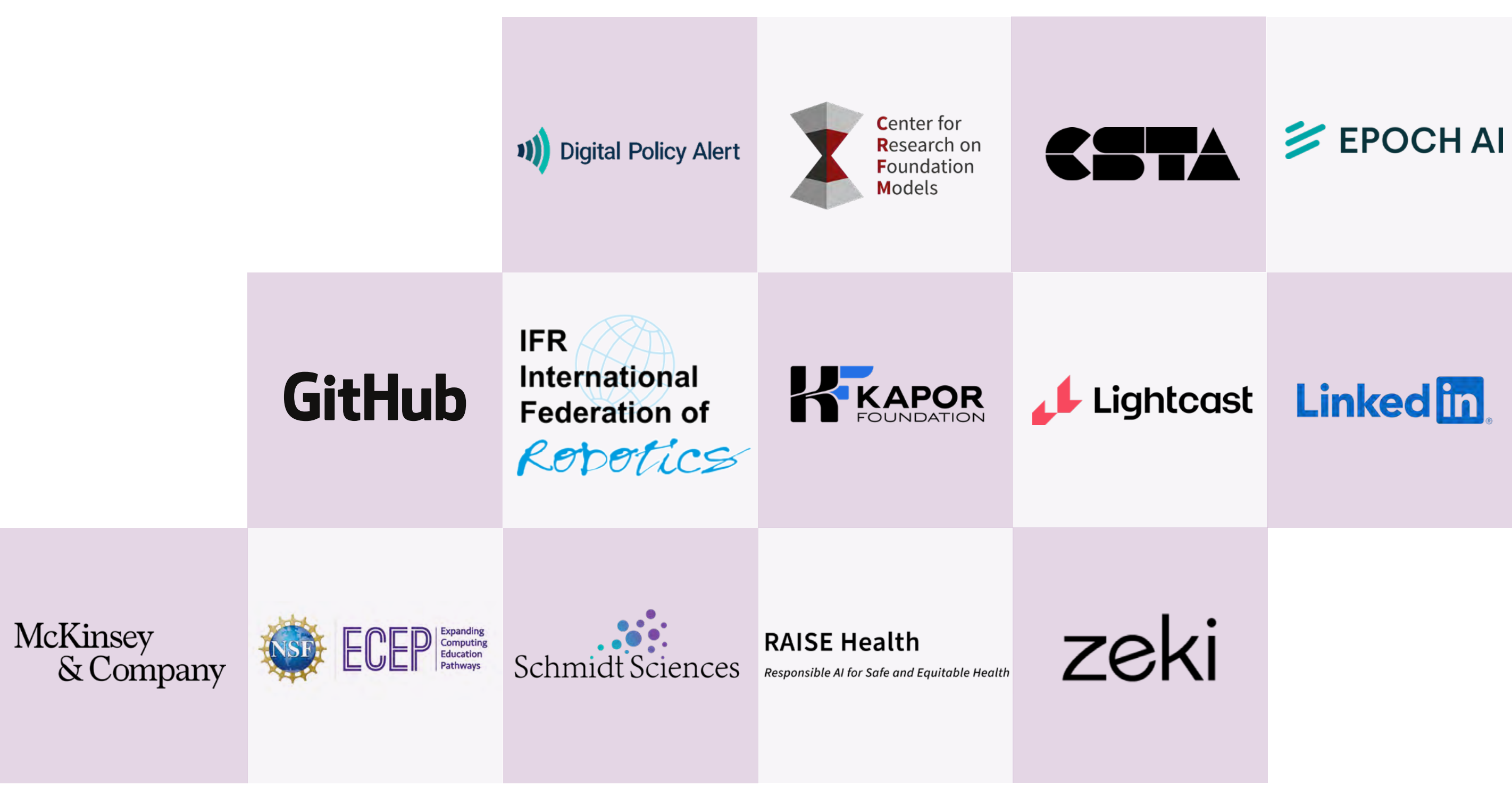

# Contributors

The AI Index would like to acknowledge the following individuals by chapter and section for their contributions of data, analysis, advice, and expert commentary included in the AI Index Report 2026:

## Introduction

Loredana Fattorini, Yolanda Gil, Vanessa Parli, Ray Perrault, Sha Sajadieh

Chapter 1

## Research and Development

Usman Anwar, Theo Burn, Emily Chen, Rachel Cook, Jean-Stanislas Denain, Meredith Ellison, Loredana Fattorini, Nicole Finn, Isabella Florez, Yolanda Gil, Tom Hurd, Nabeel Khan, James Landay, Shayne Longpre, Nestor Maslej, Magdalena Ortiz, Khalifa Oyebanji, Orestis Papakyriakopoulos, Vanessa Parli, Ray Perrault, Tom Piercey, Jennifer Rachford, Thomas Richadson, Vesna Sabljakovic-Fritz, Sha Sajadieh, Lapo Santarlasci, Sebastian Sardina, Andrew Shi, Yoav Shoham, Seth Polsley, Daniel Weld, Kevin Xu, Meg Young

Chapter 2

## Technical Performance

Erik Brynjolfsson, Loredana Fattorini, Yolanda Gil, Tasha Kim, Sanmi Koyejo, Nestor Maslej, Juan Carlos Niebles, Sukrut Oak, Vanessa Parli, Ray Perrault, Sha Sajadieh, Yoav Shoham, Toby Walsh, Daniel Weld, Henry Zhang

Chapter 3

## Responsible AI

Gabriel Morgan Asaftei, Rishi Bommasani, Virginia Dignum, Loredana Fattorini, Yolanda Gil, Nestor Maslej, Katherine Ottenbreit, Vanessa Parli, Juan Nicolas Pava, Ray Perrault, Brittany Presten, Cécile Prinsen, Roger Roberts, Sha Sajadieh, Lapo Santarlasci, Abby Sticha, Elham Tabassi, Yuanhao Zou

Chapter 4

## Economy

Tara Balakrishnan, Bharat Chandar, Erik Brynjolfsson, Ruyu Chen, Michael Chui, Heather English, Loredana Fattorini, Yolanda Gil, Bryce Hall, Heather Hanselman, Rosie Hood, Akash Kaura, Elena Magrini, Nestor Maslej, James Manyika, Rebecca Milde, David Nguyen, Katherine Ottenbreit, Vanessa Parli, Ray Perrault, Courtney Prabhakar, Brittany Presten, Roger Roberts, Sha Sajadieh, Lapo Santarlasci, Alex Singla, Alex Sukharevsky, Casey Weston, Henry Zhang

Chapter 5

## Science

Michael Clear, Steven Dillmann, Loredana Fattorini, Yolanda Gil, James Manyika, Vipin Kumar, Minjoon Kouh, Suhas Mahesh, Vanessa Parli, Ray Perrault, Sha Sajadieh

Chapter 6

## Medicine

Russ Altman, Peter Brodeur, Akshay Chaudhari, Muhammad Ahmed Tajammul Chaudhury, Jonathan Chen, Matthew DeVerna, Abdoul Jalil Djiberou Mahamadou, Loredana Fattorini, Andrea Fisher, Ethan Goh, Yolanda Gil, Jeff Hancock, Tina Hernandez-Boussard, Malte Jensen, Rohan Koodli, Arman Koul, Sanmi Koyejo, Alejandro Lozano, Danielle Luz, David Magnus, Stephen P. Ma, Bethel Mieso, Fateme Nateghi Haredasht, Madelena Ng, Natalie Pageler, Ayush Pandit, Vanessa Parli, Ray Perrault, Sean Riordan, Ronald Robertson, Austin Schoeffler, Christoph Sadée, Sha Sajadieh, Min Woo Sun, Kotoha Togami, Sang Truong, Chong Wang, Dennis Wall, David Wu

Chapter 7

## Education

Carla Brodley, Joshua Childs, Lisa Cruz Novohatski, Loredana Fattorini, Yolanda Gil, Rachel Goins, Laura Hinton, Sonia Koshy, James Landay, Kirsten Lundgren, Jacqueline McCune, Vanessa Parli, Ray Perrault, Sha Sajadieh, Bryan Twarek, Rebecca Zarch

Chapter 8

### Policy and Governance

Virginia Dignum, Loredana Fattorini, Johannes Fritz, Yolanda Gil, Nestor Maslej, Vanessa Parli, Juan Nicolas Pava, Ray Perrault, Sha Sajadieh, Lapo Santarlasci, Kamran Sattary, Tyler Lenox Smith, Elham Tabassi, Russell Wald

Chapter 9

### Public Opinion

Erik Brynjolfsson, Matt Carmichael, Zack Devlin-Foltz, Loredana Fattorini, Nadja Flechner, Yolanda Gil, Connacher Murphy, Vanessa Parli, Juan Nicolas Pava, Ray Perrault, Matt Reynolds, Sha Sajadieh, Russell Wald, Henry Zhang

The AI Index thanks the following organizations and individuals who provided data for inclusion in this year's report:

## ORGANIZATIONS

### Epoch AI

Jean-Stanislas Denain

### GitHub

Kevin Xu

### Lightcast

Elena Magrini, Rebecca Milde

### LinkedIn

Rosie Hood, Akash Kaura, Casey Weston

### McKinsey & Company

Heather Hanselman, Katherine Ottenbreit, Brittany Presten, Cécile Prinsen, Roger Roberts, Abby Sticha

### Quid

Heather English

### Zeki

Tom Hurd

The AI Index also thanks Jeanina Matias, Nancy King, Carolyn Lehman, Shana Lynch, Jonathan Mindes, and Johanna Friedman for their help in preparing this report; Christopher Ellis and Madeleine Wright for their help in maintaining the AI Index website; and Annie Benisch, Marc Gough, Caroline Meinhardt, Drew Spence, Casey Weston, and Daniel Zhang for their work in helping promote the report.

# Top Takeaways

1. **AI capability is not plateauing. It is accelerating and reaching more people than ever.** Industry produced over 90% of notable frontier models in 2025, and several of those models now meet or exceed human baselines on PhD-level science questions, multimodal reasoning, and competition mathematics. On a key coding benchmark—SWE-bench Verified—performance rose from 60% to near 100% of meeting the human baseline in a single year. Organizational adoption reached 88%, and 4 in 5 university students now use generative AI.

2. **The U.S.-China AI model performance gap has effectively closed.** U.S. and Chinese models have traded the lead multiple times since early 2025. In February 2025, DeepSeek-R1 briefly matched the top U.S. model, and as of March 2026 Anthropic's top model leads by just 2.7%. The U.S. still produces more top-tier AI models and higher-impact patents, while China leads in publication volume, citations, patent output, and industrial robot installations. South Korea stands out for its innovation density, leading the world in AI patents per capita.

3. **The United States hosts the most AI data centers, with the majority of their chips fabricated by one Taiwanese foundry.** The United States hosts 5,427 data centers, more than 10 times any other country, and it consumes more energy than any other country. A single company, TSMC, fabricates almost every leading AI chip, making the global AI hardware supply chain dependent on one foundry in Taiwan—though a TSMC-U.S. expansion began operations in 2025.

4. **AI models can win a gold medal at the International Mathematical Olympiad but cannot reliably tell time—an example of what researchers call the jagged frontier of AI.** Gemini Deep Think earned a gold medal at IMO, yet the top model reads analog clocks correctly just 50.1% of the time. AI agents made a leap from 12% to ~66% task success on OSWorld, which tests agents on real computer tasks across operating systems, though they still fail roughly 1 in 3 attempts on structured benchmarks.

5. **Robots still fail at most household tasks, even as they excel in controlled environments.** Robots succeed in only 12% of household tasks, highlighting how far AI is from mastering the physical world. On RLBench, robotic manipulation in software-based simulations has reached 89.4% success, but the gap between predictable lab settings and unpredictable household environments is wide.

6. **Responsible AI is not keeping pace with AI capability, with safety benchmarks lagging and incidents rising sharply.** Almost all leading frontier AI model developers report results on capability benchmarks, but reporting on responsible AI benchmarks remains spotty. Documented AI incidents rose to 362, up from 233 in 2024. Adding to the challenge, recent research found that improving one responsible AI dimension, such as safety, can degrade another, such as accuracy.

7 **The United States leads in AI investment, but its ability to attract global talent is declining.** U.S. private AI investment reached $285.9 billion in 2025, more than 23 times the $12.4 billion invested in China—though looking at just private investment figures likely understates China's total AI spending, given its government guidance funds. The U.S. also led in entrepreneurial activity with 1,953 newly funded AI companies in 2025, more than 10 times the next closest country. However, the number of AI researchers and developers moving to the U.S. has dropped 89% since 2017, with an 80% decline in the last year alone.

8 **AI adoption is spreading at historic speed, and consumers are deriving substantial value from tools they often access for free.** Generative AI reached 53% population adoption within three years, faster than the PC or the internet, though the pace varies by country and correlates strongly with GDP per capita. Some show higher-than-expected adoption, such as Singapore (61%) and the United Arab Emirates (64%), while the U.S. ranks 24th at 28.3%. The estimated value of generative AI tools to U.S. consumers reached $172 billion annually by early 2026, with the median value per user tripling between 2025 and 2026.

9 **Productivity gains from AI are appearing in many of the same fields where entry-level employment is starting to decline.** Studies show productivity gains of 14% to 26% in customer support and software development, with weaker or negative effects in tasks requiring more judgment. AI agent deployment remains in single digits across nearly all business functions. In software development, where AI's measured productivity gains are clearest, U.S. developers ages 22 to 25 saw employment fall nearly 20% from 2024, even as the headcount for older developers continues to grow.

10 **AI's environmental footprint is expanding alongside its capabilities.** Grok 4's estimated training emissions reached 72,816 tons of CO2 equivalent. AI data center power capacity rose to 29.6 GW, comparable to New York state at peak demand, and annual GPT-4o inference water use alone may exceed the drinking water needs of 1.2 million people.

11 **AI models for science can outperform human scientists, though bigger models do not always perform better.** Frontier models outperform human chemists on average on ChemBench, yet they score below 20% on replication in astrophysics and 33% on Earth observation questions. A 111-million-parameter protein language model, MSAPairformer, beat previous leading methods on ProteinGym, and a 200-million-parameter genomics model, GPN-Star, outperformed a model nearly 200 times larger. Most AI foundation models for science come from cross-sector collaborations, in contrast with the industry-dominated landscape of general-purpose AI.

12 **AI is transforming clinical care, but rigorous evidence remains limited.** AI tools that automatically generate clinical notes from patient visits saw substantial adoption in 2025. Across multiple hospital systems, physicians reported up to 83% less time spent writing notes and significant reductions in burnout. Beyond certain tools, however, the evidence base for clinical AI remains thin. A review of more than 500 clinical AI studies found that nearly half relied on exam-style questions rather than real patient data, with only 5% using real clinical data.

13 **Formal education is lagging behind AI, but people are learning AI skills at every stage of life.** Over 80% of U.S. high school and college students now use AI for school-related tasks, but only half of middle and high schools have AI policies in place, and just 6% of teachers say those policies are clear. Outside the classroom, AI engineering skills are accelerating fastest in the United Arab Emirates, Chile, and South Africa. The number of new AI PhDs in the U.S. and Canada increased 22% from 2022 to 2024, the PhDs that make up that increase took jobs in academia, not in industry.

14 **AI sovereignty is becoming a defining feature of national policy, but capabilities remain uneven, even as open-source development helps to redistribute who participates.** National AI strategies are expanding, particularly among developing economies, and state-backed investments in AI supercomputing are rising in parallel—a sign of growing ambitions for domestic control over AI ecosystems. Yet model production remains concentrated in the U.S. and China. Open-source development is starting to redistribute participation, with contributions from the rest of the world now outpacing Europe and approaching the United States on GitHub, fueling more linguistically diverse models and benchmarks.

15 **AI experts and the public have very different perspectives on the technology's future, and global trust in institutions to manage AI is fragmented.** When it comes to how people do their jobs, 73% of experts expect a positive impact, compared with just 23% of the public, a 50-point gap. Similar divides appear for AI's impact on the economy and medical care. Globally, trust in governments to regulate AI varies. Among surveyed countries, the United States reported the lowest level of trust in its own government to regulate AI, at 31%. Globally, the EU is trusted more than the United States or China to regulate AI effectively.

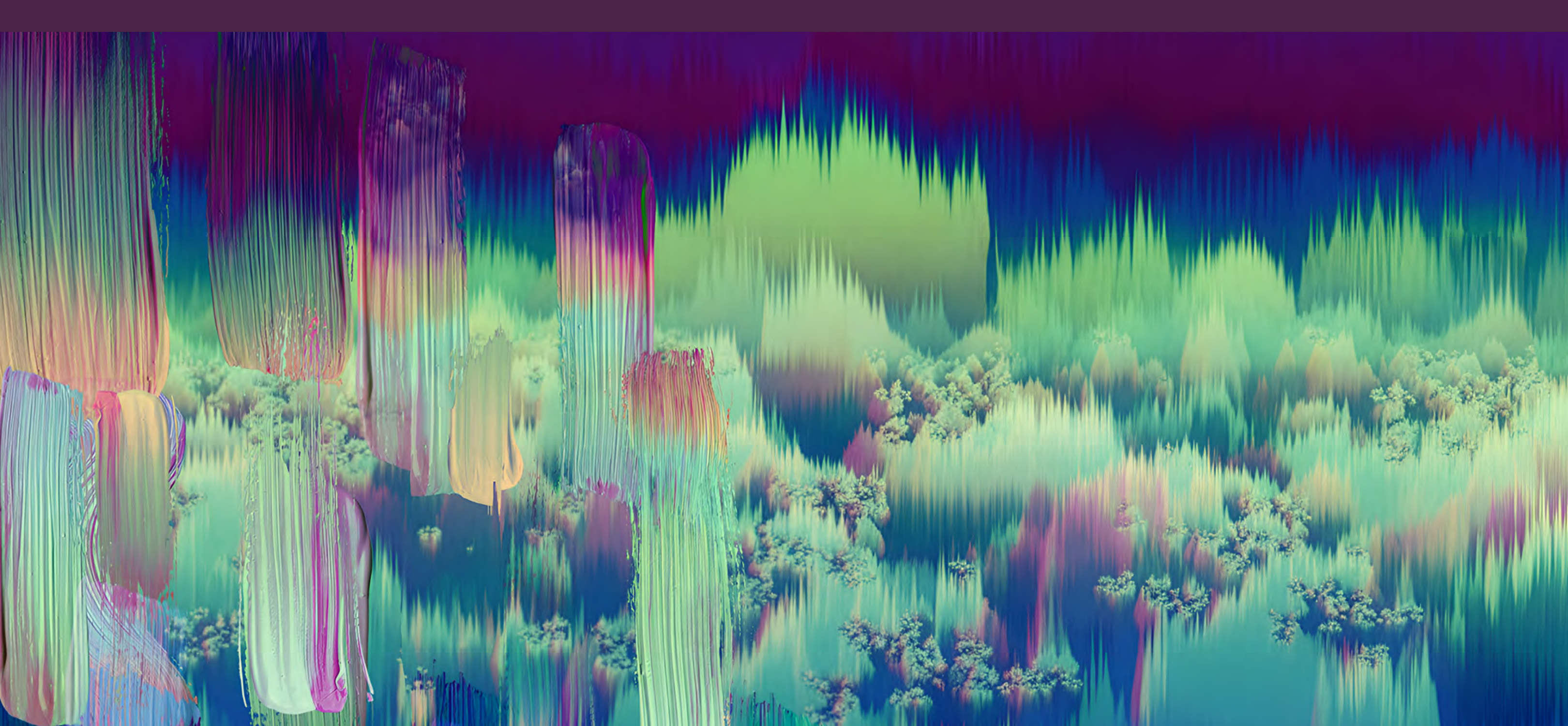

# 1

# Research and Development

## Overview

The resources powering AI development continued to grow in 2025, but fewer notable models were released than the year before, and the systems at the frontier are increasingly concentrated among a small number of organizations. Industry now accounts for over 90% of notable AI models, and the most capable systems are also the least transparent, with training code, dataset sizes, and parameter counts increasingly withheld. The computing power behind these models has grown roughly 3.3 times per year since 2022, yet almost all of it flows through a single chip foundry in Taiwan, making the global hardware supply chain fragile. Open-source development and AI publications continued to grow, and the research landscape is becoming more geographically distributed. China now leads in publication volume, citation share, and patent grants, while smaller countries like Switzerland and Singapore lead in AI researchers per capita. Yet some dimensions of the field have not changed at all. Gender gaps in AI talent remain deeply entrenched, with no meaningful progress in any country since 2010. This chapter covers the research and development pipeline, from the landscape of AI models through the compute, data centers, energy, and open-source software that support them, to the broader research ecosystem of publications, patents, and talent.

## Contents

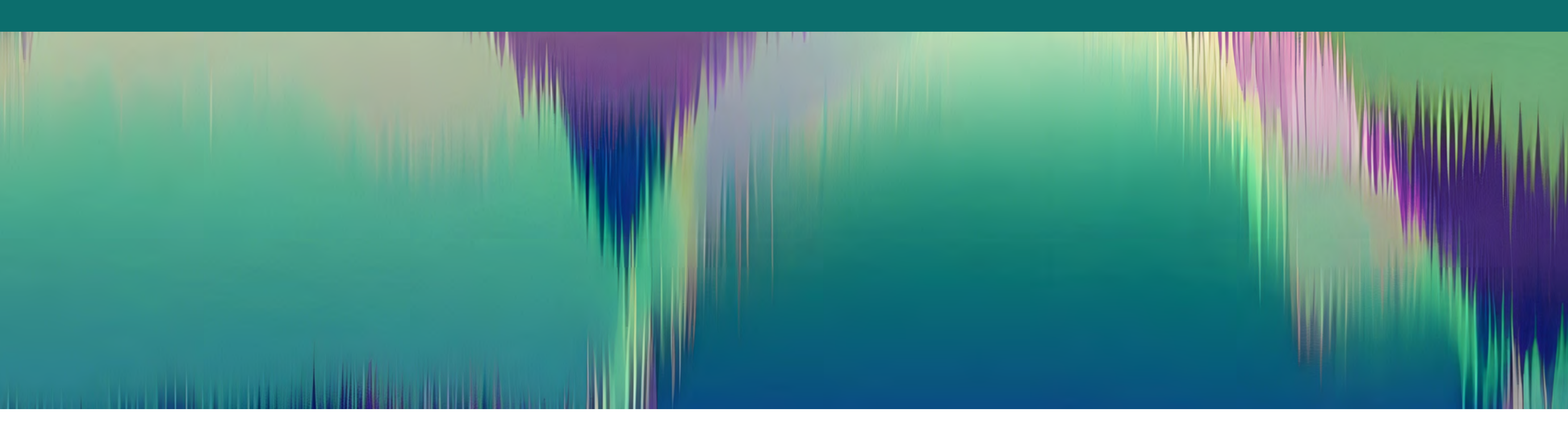

# Chapter Highlights

1. **Industry produced over 90% of notable AI models in 2025, but the most capable models are now the least transparent.** Training code, parameter counts, dataset sizes, and training duration are no longer disclosed for several of the most resource-intensive systems, including those from OpenAI, Anthropic, and Google.

2. **China leads in research, while the U.S. leads in notable model development.** China leads in publication volume, citations, and patent grants, while the U.S. retains higher-impact patents and produced 59 notable models in 2025 to China's 35. South Korea leads in AI patents per capita, and China's share of the top 100 most-cited AI papers grew from 33 in 2021 to 41 in 2024.

3. **Reported parameters held in the trillions as disclosure dropped.** Parameter counts have stayed near 1 trillion for three years, though reporting from frontier labs has stopped. Training compute, which can be estimated independently, has continued to rise.

4. **Synthetic data is still not replacing real data in pre-training, but data quality and post-training techniques are showing promise.** OLMo 3.1 Think 32B, with nearly 90 times fewer parameters than Grok 4, achieves comparable results on several benchmarks through pruning, deduplication, and curation alone.

5. **Global AI compute capacity grew 3.3x per year since 2022, reaching 17.1 million H100-equivalents.** Nvidia accounts for over 60% of total compute, with Google and Amazon supplying much of the remainder and Huawei holding a small but growing share. The buildout is being driven by hyperscaler data center expansion and sustained demand for frontier model training and inference.

6. **The United States leads in AI data centers, and one Taiwanese foundry fabricates the majority of chips inside them.** The United States hosts 5,427 data centers, more than ten times any other country, consuming more energy than any other region. A single company, TSMC, fabricates almost every leading AI chip and makes the global AI hardware supply chain dependent on one foundry in Taiwan, though a TSMC-U.S. expansion began to operate in 2025.

7. **AI's environmental footprint increases across power, water, and emissions.** In 2025, Grok 4's estimated training emissions reached 72,816 tons of $CO_2$ equivalent. AI data center power capacity rose to 29.6 GW, comparable to New York state at peak demand, and annual GPT-4o inference water use alone may exceed the drinking water needs of 1.2 million people.

8. **Open-source AI development continues to scale, with 5.6 million projects on GitHub and Hugging Face uploads tripling since 2023.** U.S.-based projects still attract the most engagement, with 30 million cumulative GitHub stars across projects that have crossed the 10-star threshold.

9 **The number of AI researchers and developers moving to the United States has dropped 89% since 2017.** The decline is accelerating, down 80% in the last year alone. The U.S. is still home to more AI talent than any other country, but it is attracting new talent at the lowest rate in over a decade.

10 **The AI talent map is shifting, but gender gaps remain deeply entrenched.** Switzerland and Singapore lead the world in AI researchers and developers per capita and some countries show relatively higher female representation, including Saudi Arabia (32.3%), Canada (29.6%), and Australia (30.1%), though no country approaches gender parity.

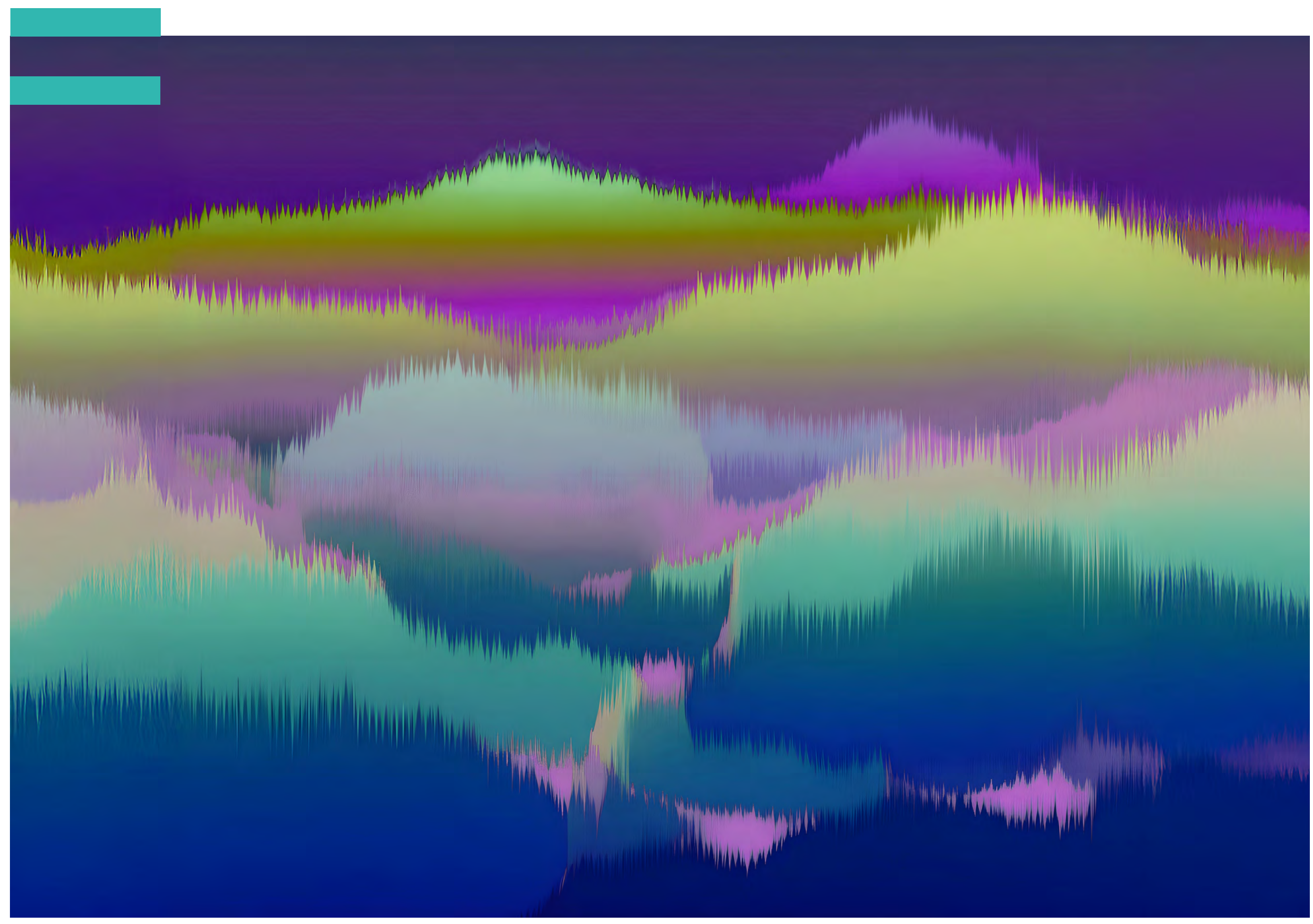

# 1.1 Notable AI Models

This section starts with the models themselves. Using Epoch AI's curated dataset of notable models, this section examines where frontier AI models are coming from, how they are deployed, and what it takes to build them. Epoch AI designates models as noteworthy based on criteria such as state-of-the-art advancements, historical significance, or high citation rates. This is a manual curation, so the dataset is not a census of all AI models or a full map of all model development activity.[1] Trends should be read as patterns within the domain. The sections that follow track the infrastructure and inputs behind these systems, including compute, data centers, energy costs, and open-source software, before looking at the broader research ecosystem through publications, patents, and talent.

This chapter focuses on the research and development pipeline and its inputs. The next chapter, Technical Performance, reviews model capabilities and benchmark performance in detail.

## By National Affiliation[2]

Notable model production remains concentrated within a small number of countries (Figures 1.1.1–1.1.3). Historically, the United States has produced the largest in total output numbers, followed by China. This pattern continued in 2025 as the United States led with the release of 59 notable AI models, China with 35, and South Korea with 8. The number of new model releases declined year over year across all major geographic areas.

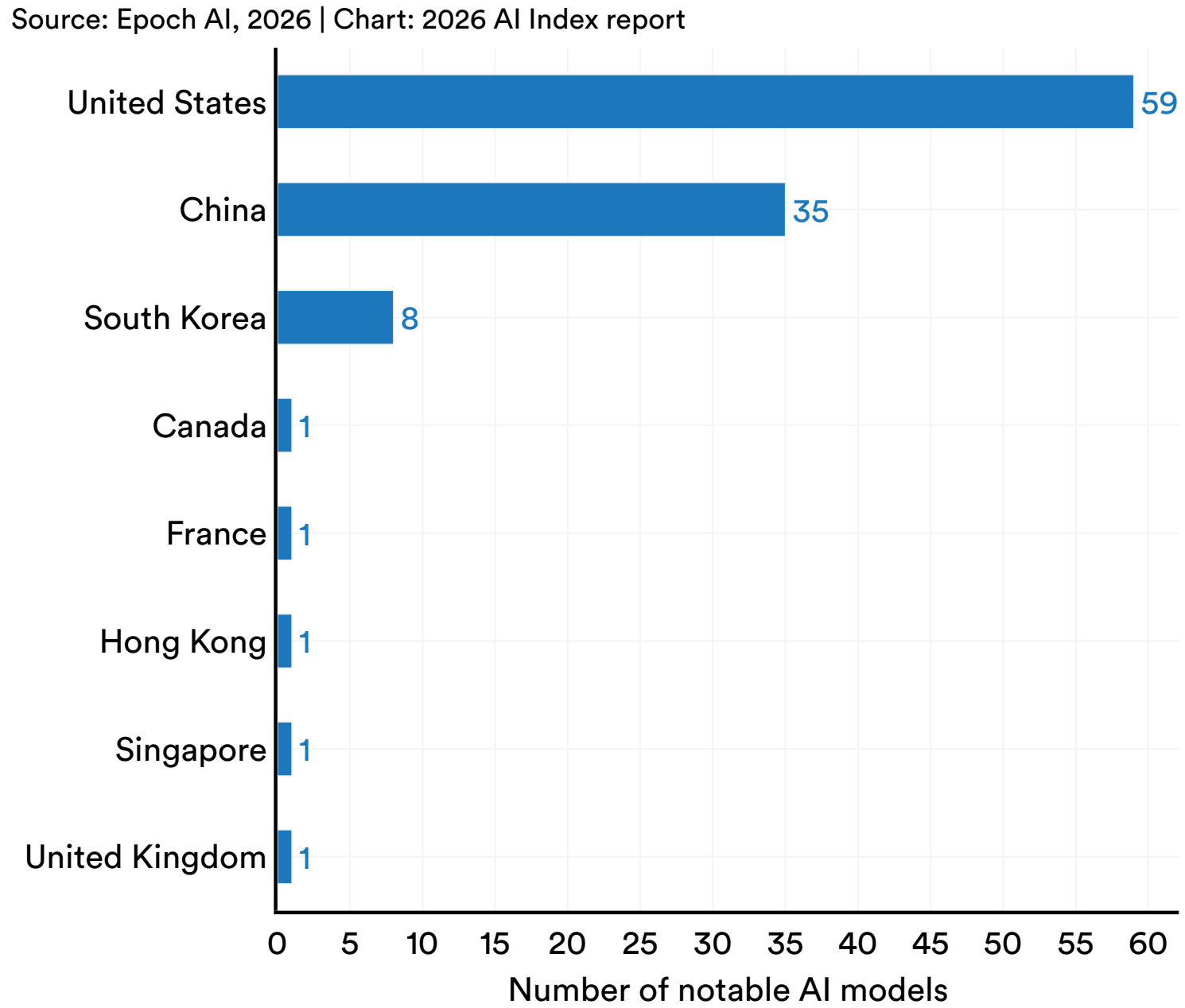
Number of notable AI models by select geographic areas, 2025
Source: Epoch AI, 2026 | Chart: 2026 AI Index report

Figure 1.1.1[3]

1 New and historic models are continually added to the Epoch AI database, so the total year-by-year counts of models included in this year's AI Index might not exactly match those published in last year's report. The data is based on a snapshot taken on April 22, 2026.

2 A machine learning model is associated with a specific country if at least one author of the paper introducing it is affiliated with an institution based in that country. In cases where a model's authors come from several countries, double-counting can occur.

3 This chart highlights model releases from a select group of geographic areas. More comprehensive data on model releases by country will be available in the upcoming AI Index Global Vibrancy Tool.

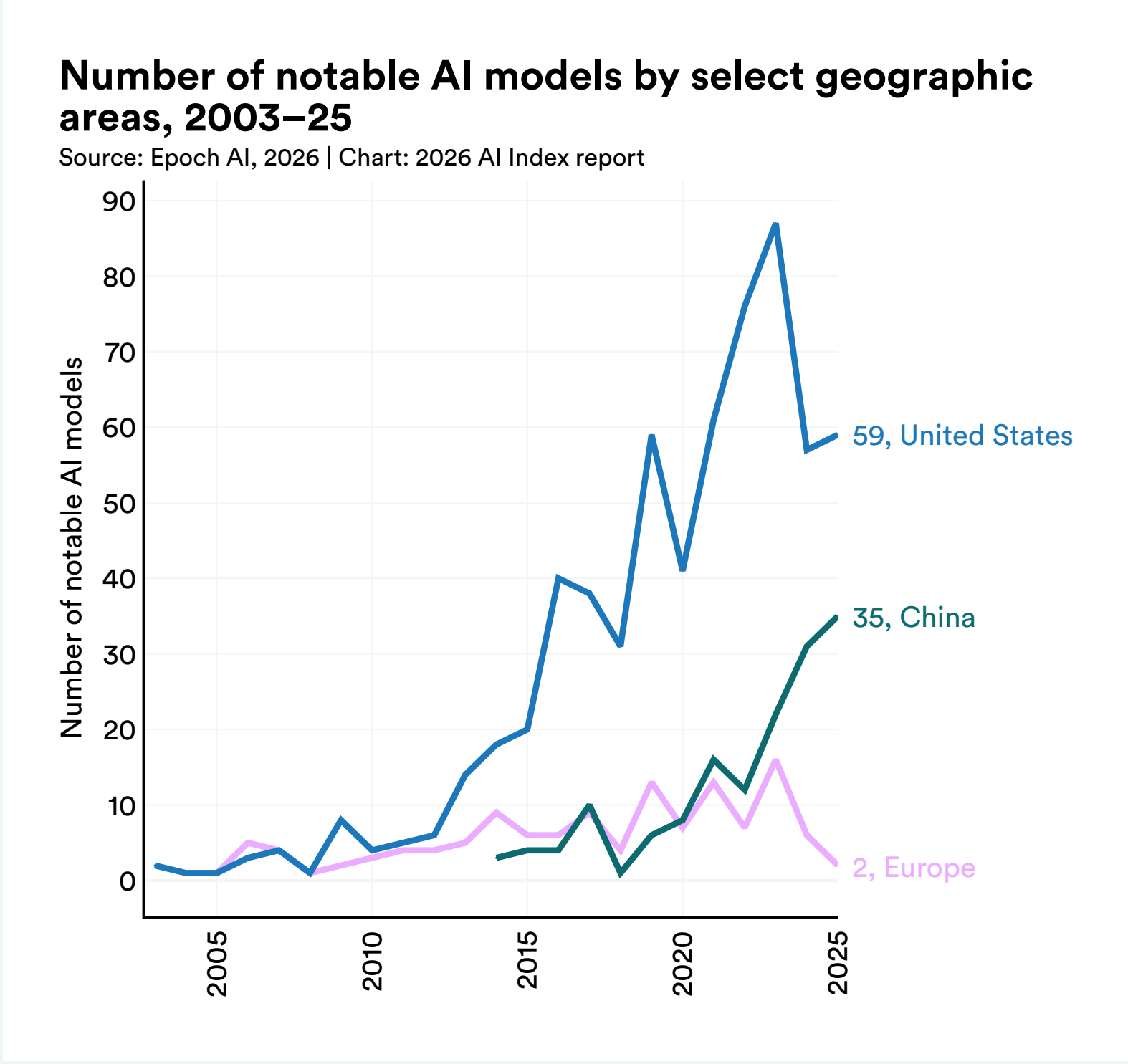

Figure 1.1.2

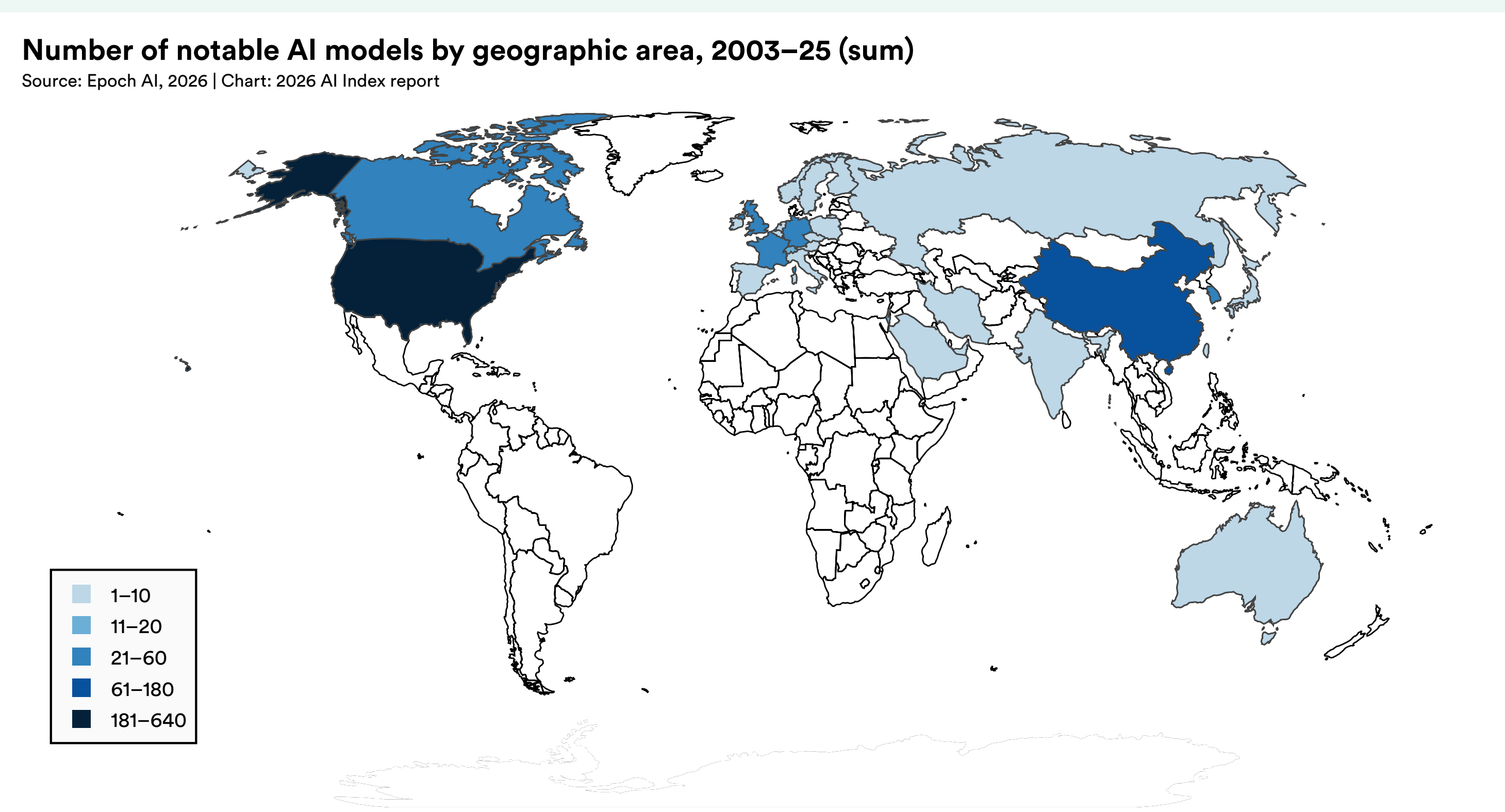

Figure 1.1.3

## By Sector and Organization

The development of notable AI models continues to be predominantly concentrated in industry (Figures 1.1.4 and 1.1.5). Over the past decade, the share produced by industry has grown steadily and now represents the largest share by a wide margin (91.2%). In 2025, Epoch AI identified two notable AI models originating from academia, compared to 93 from industry.

Within industry, a small set of organizations account for a large share of releases (Figures 1.1.6 and 1.1.7). In 2025, the top contributors were OpenAI (20), Google (14), and Alibaba (11). Since 2014, Google has produced the largest number of notable models, followed by Meta and OpenAI. Within academia, Tsinghua University (26), Stanford University (26), and Carnegie Mellon University (25) have been the most prolific over the past decade.

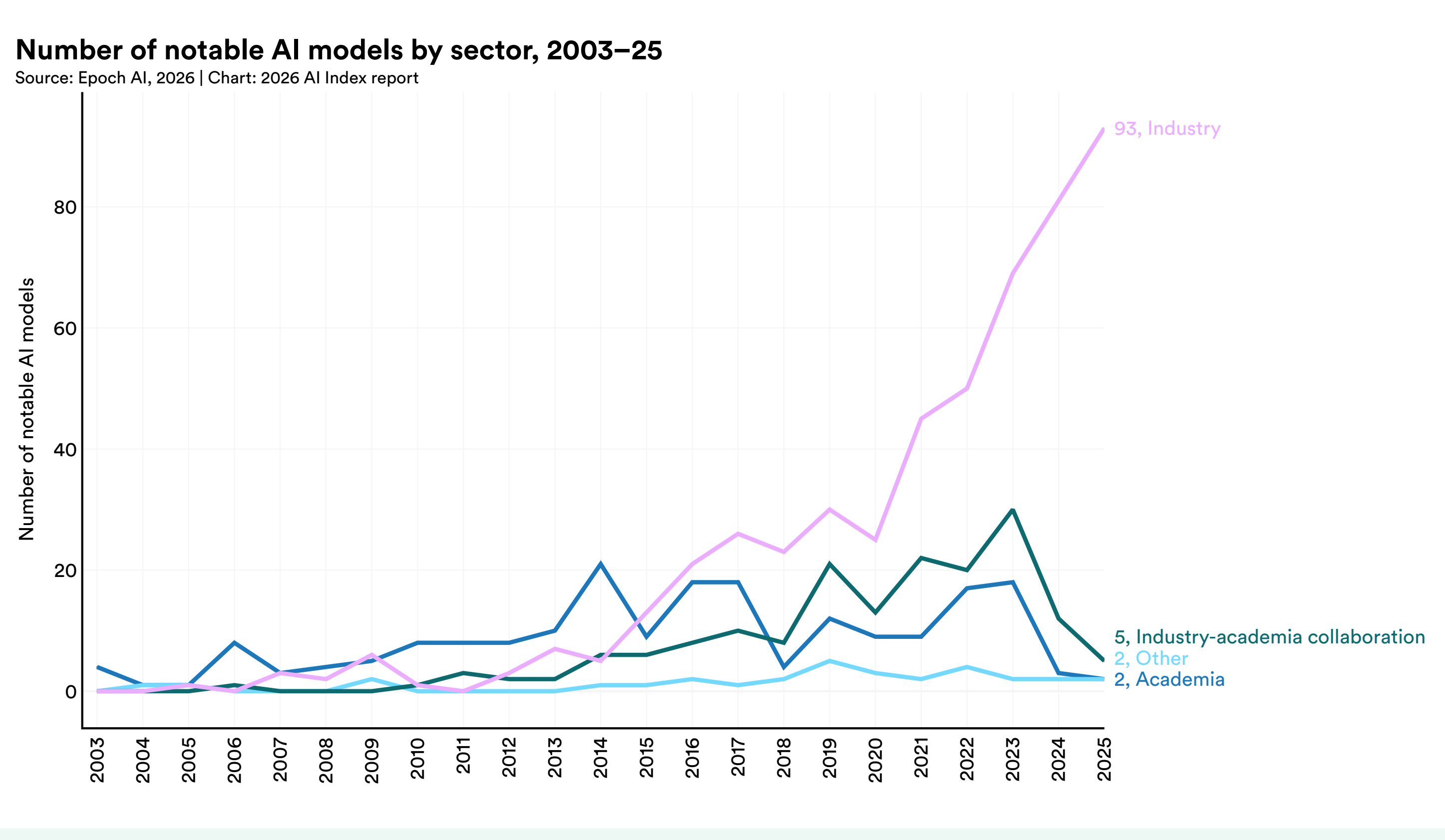

Figure 1.1.4

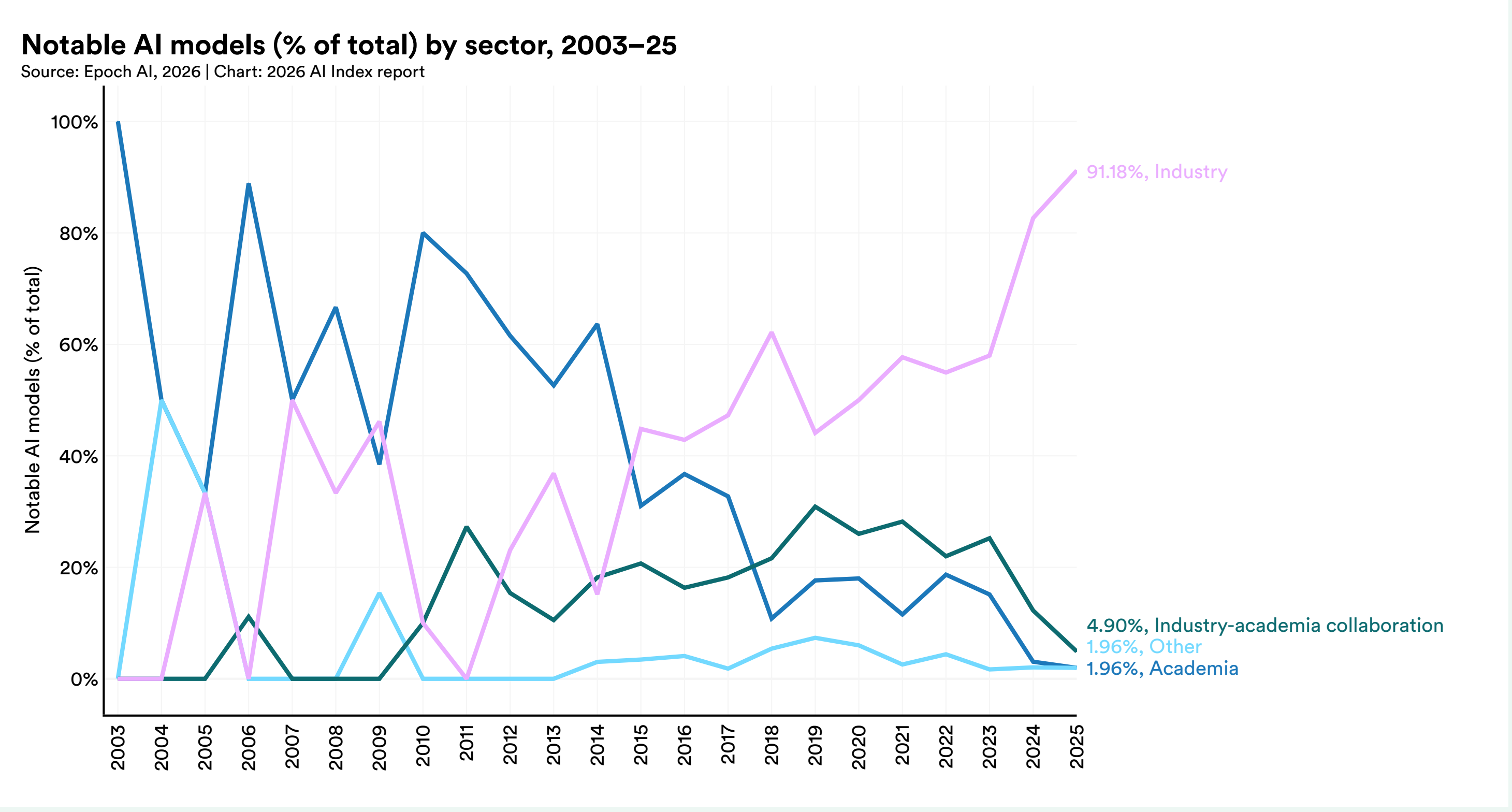

Figure 1.1.5

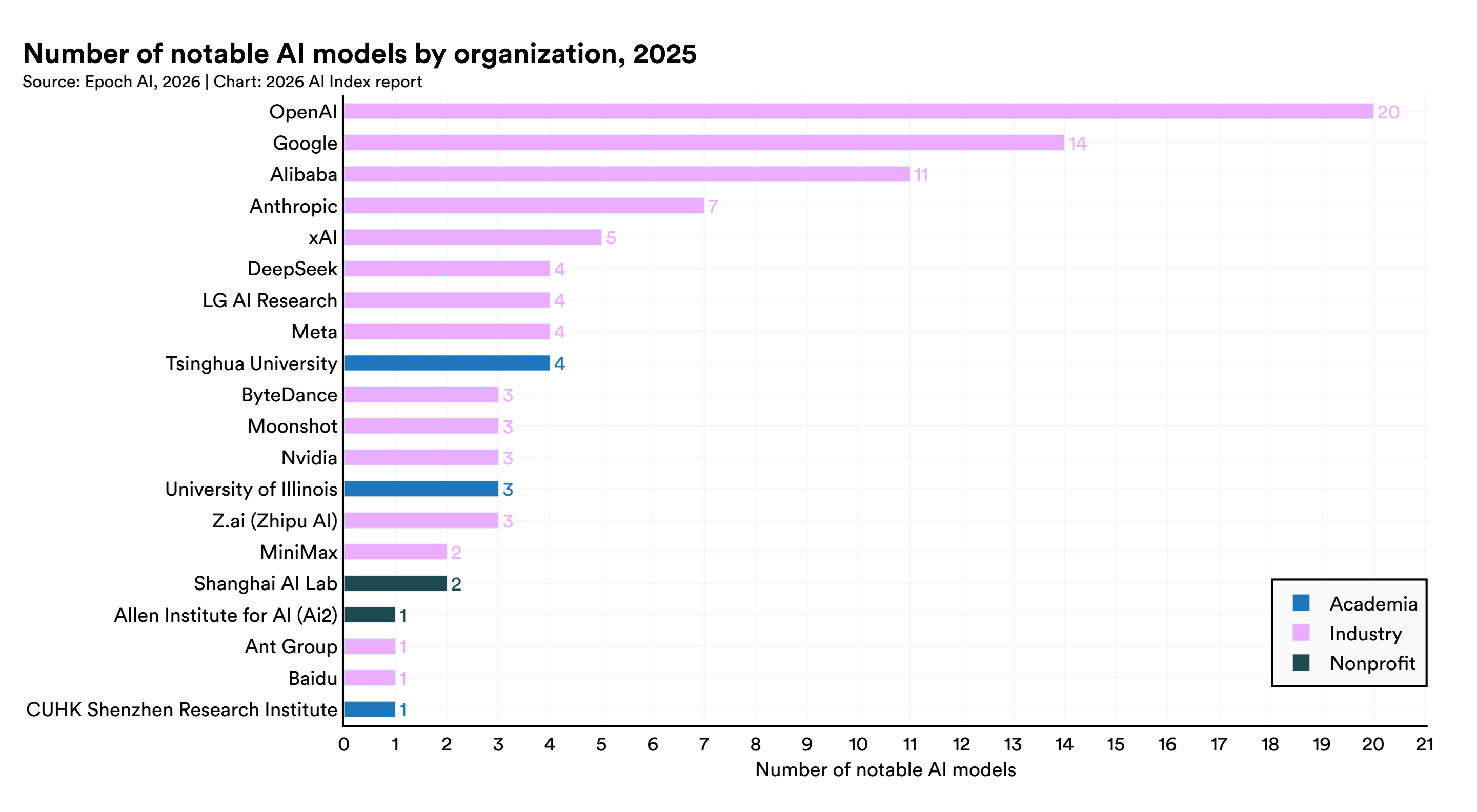

Figure 1.1.6[4]

4 In the organizational tally figures, research published by DeepMind is classified under Google.

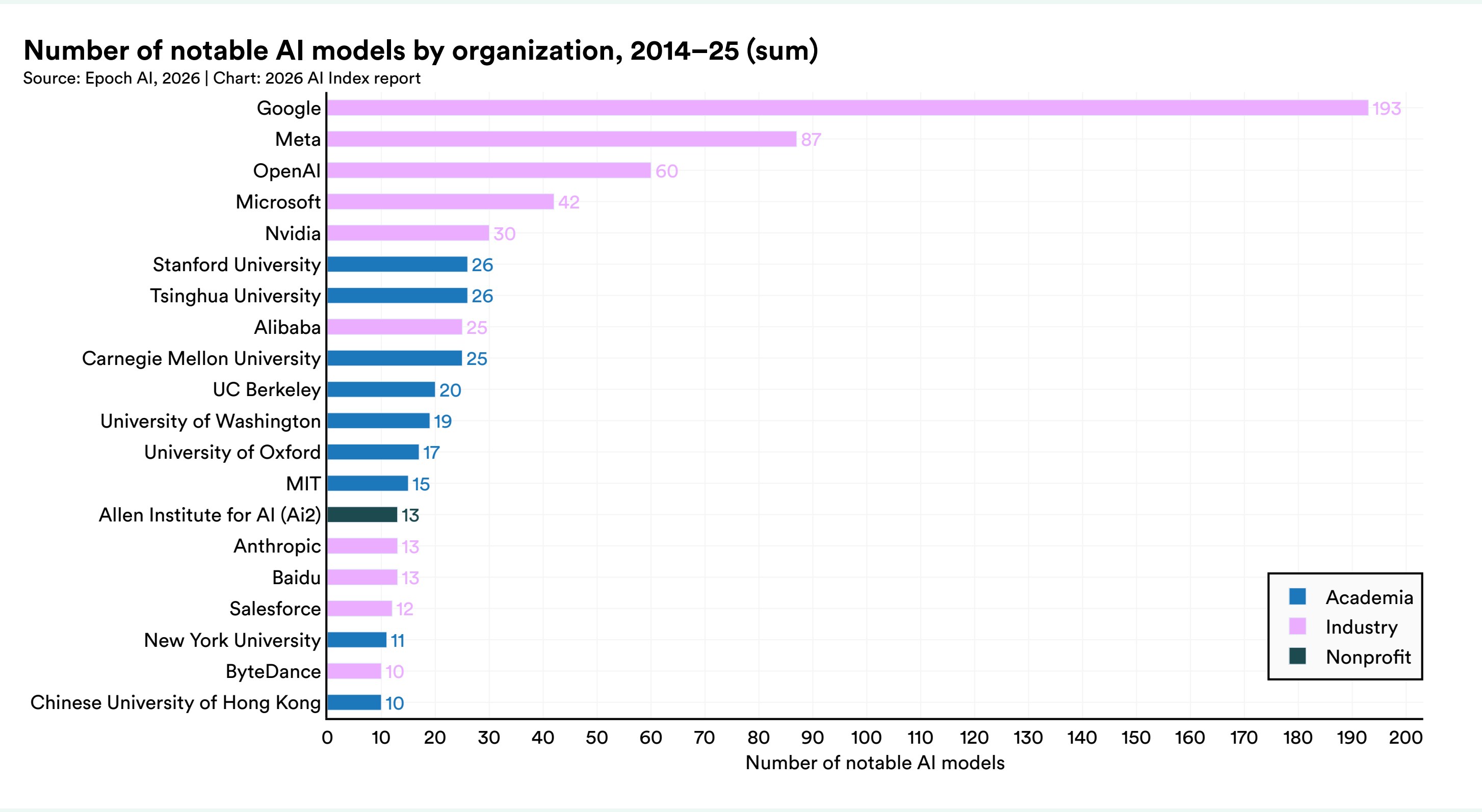

Figure 1.1.7

## Model Release

Release patterns for notable AI models have continued to shift toward controlled access (Figure 1.1.8). In 2025, API access was the most common release type, with 47 of 102 models made available this way. and API-only releases have steadily increased since 2020. The second most common release type was "open weights (unrestricted)," meaning the models are fully available for use, modification, and redistribution. The remaining models were released in a mix of access types, including "hosted access (no API),"[5] "open weights (restricted use),"[6] and "open weights (noncommercial)." The "unknown" designation refers to models that have unclear or undisclosed access types, and "unreleased" models remain proprietary, accessible only to their developers or select partners.

Training code is becoming even less accessible than model code overall (Figure 1.1.9). In 2025, 81 of 102 notable models were released without their corresponding training code, compared to 4 that made their code "open source." In 2020, models with open source and unreleased training code were about the same in number, but by 2023, the majority were unreleased and the gap has continued to widen. This growing opacity limits the ability of external researchers to reproduce results, audit development, and validate safety claims. These challenges are central to the responsible AI and governance discussions in Chapter 3 and Chapter 8.

5 Hosted access refers to using computing resources or services (such as software, hardware, or storage) provided remotely by a third party, rather than personally owning or managing them. Instead of running software or infrastructure locally, hosted access involves accessing these resources via the cloud or another remote service, typically over the internet. For example, using GPUs through platforms like AWS, Google Cloud, or Microsoft Azure—rather than running them on one's own hardware—is considered hosted access.

6 Open weights models share their architecture at varying levels of restriction, "noncommercial" limits use to research purposes, "restricted use" permits broader use with some conditions, and "unrestricted" places no limitations on use, modification, or redistribution.

## Number of notable AI models by access type, 2014–25

Source: Epoch AI, 2026 | Chart: 2026 AI Index report

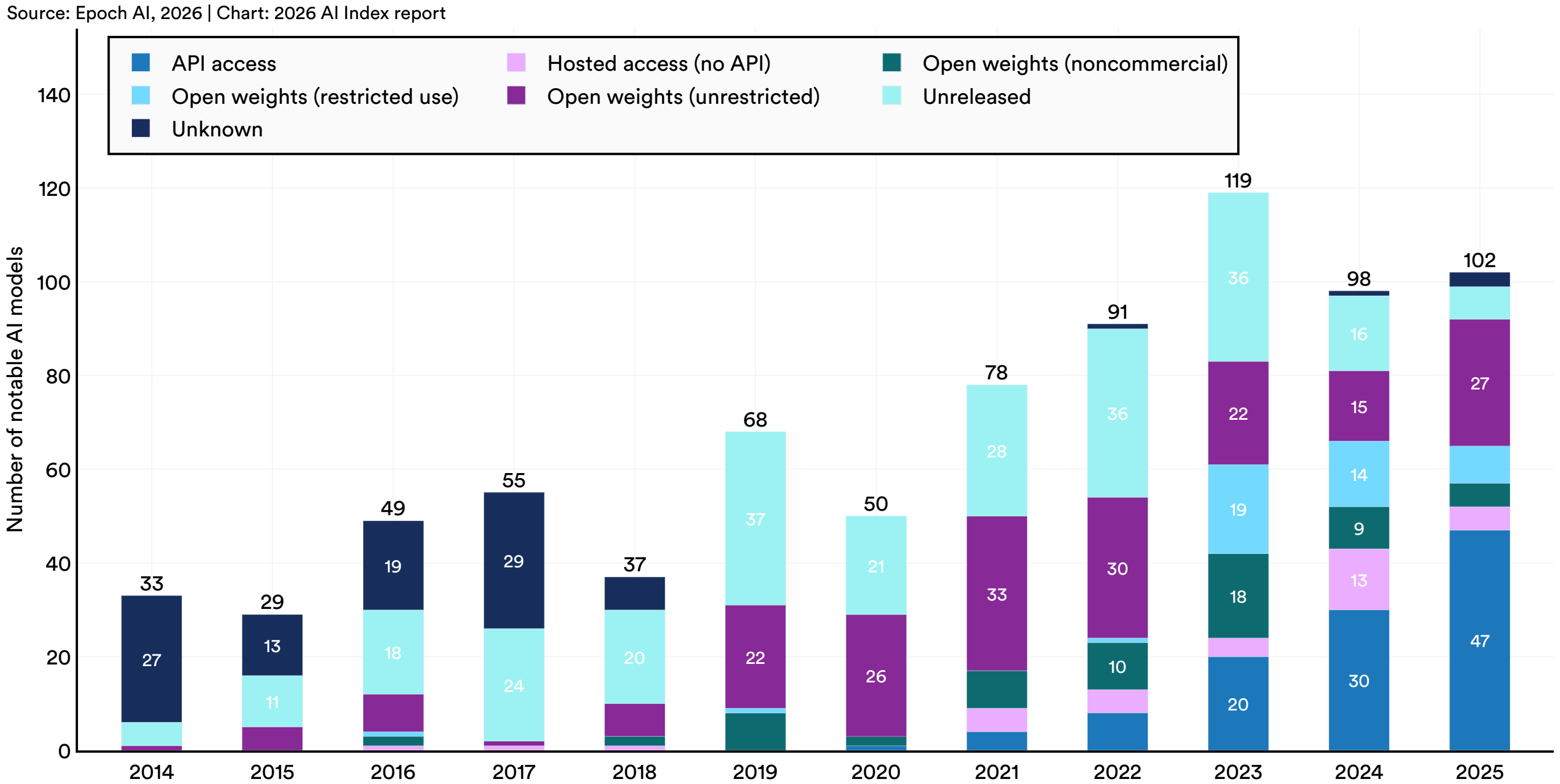

Figure 1.1.8[7]

## Number of notable AI models by training code access type, 2014–25

Source: Epoch AI, 2026 | Chart: 2026 AI Index report

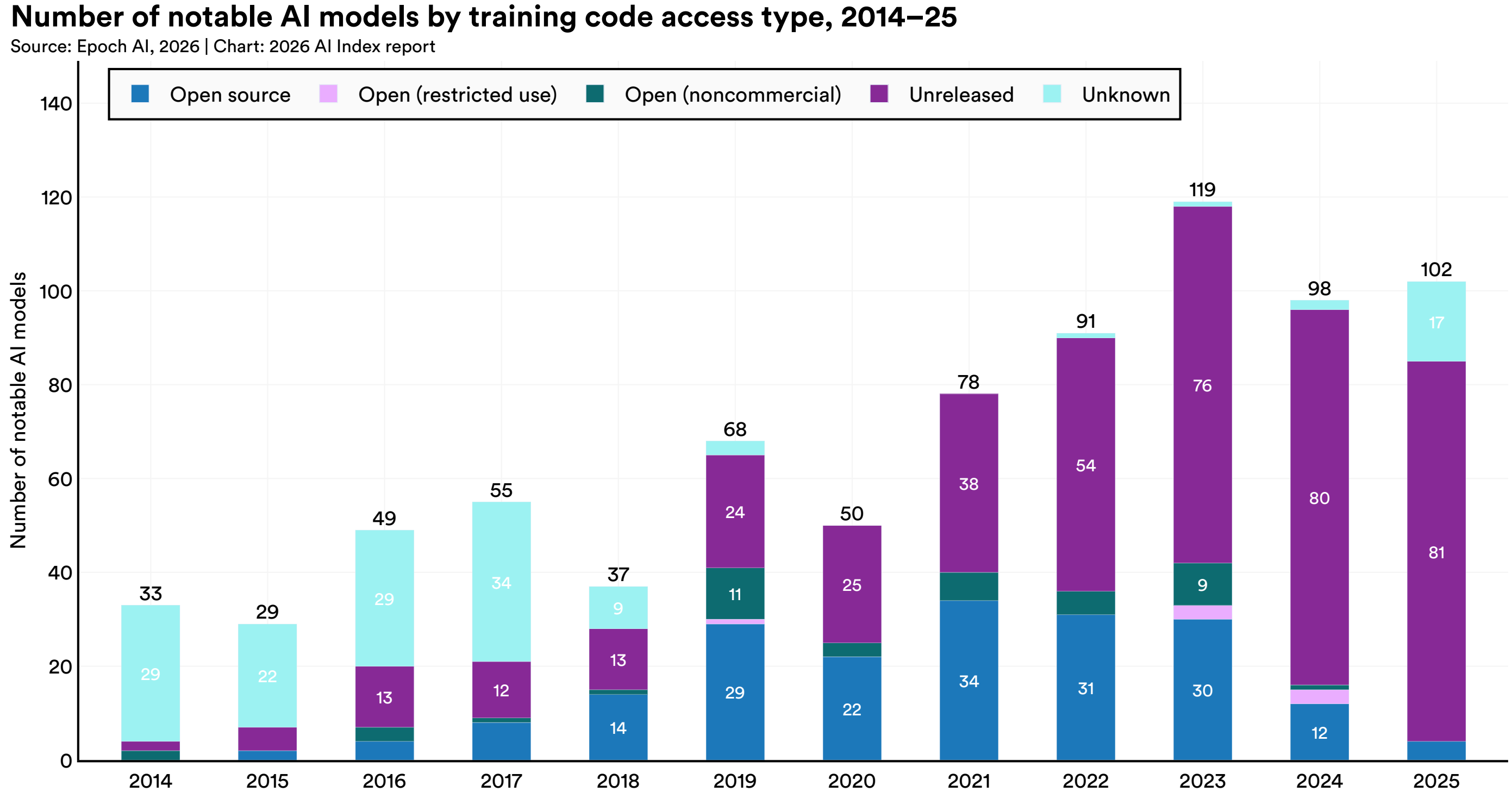

Figure 1.1.9

7 Not all models in the Epoch database are categorized by access type, so the totals in Figures 1.1.8 and 1.1.9 may not fully align with those reported elsewhere in the chapter.

## Parameter and Compute Trends

Parameter counts for notable AI models have increased significantly from the early 2010s through 2022, driven by the growing complexity of model architecture, greater data availability, improvements in hardware, and proven efficacy of larger models (Figures 1.1.10–1.1.12[8]). Since then, growth in reported parameter counts has flattened, but this is likely understating actual growth due to the absence of certain data points. Several of the most resource-intensive models released in recent years, including those from OpenAI, Anthropic, and Google, have not publicly disclosed parameter counts, training dataset sizes, or training duration.

Similarly, training dataset sizes and training duration increased through the early 2020s, with leading models training on tens of trillions of tokens over periods exceeding 100 days. Again, due to limited disclosure from major frontier labs, the more recent data is incomplete.

**Number of parameters of notable AI models by sector, 2003–25**
Source: Epoch AI, 2026 | Chart: 2026 AI Index report

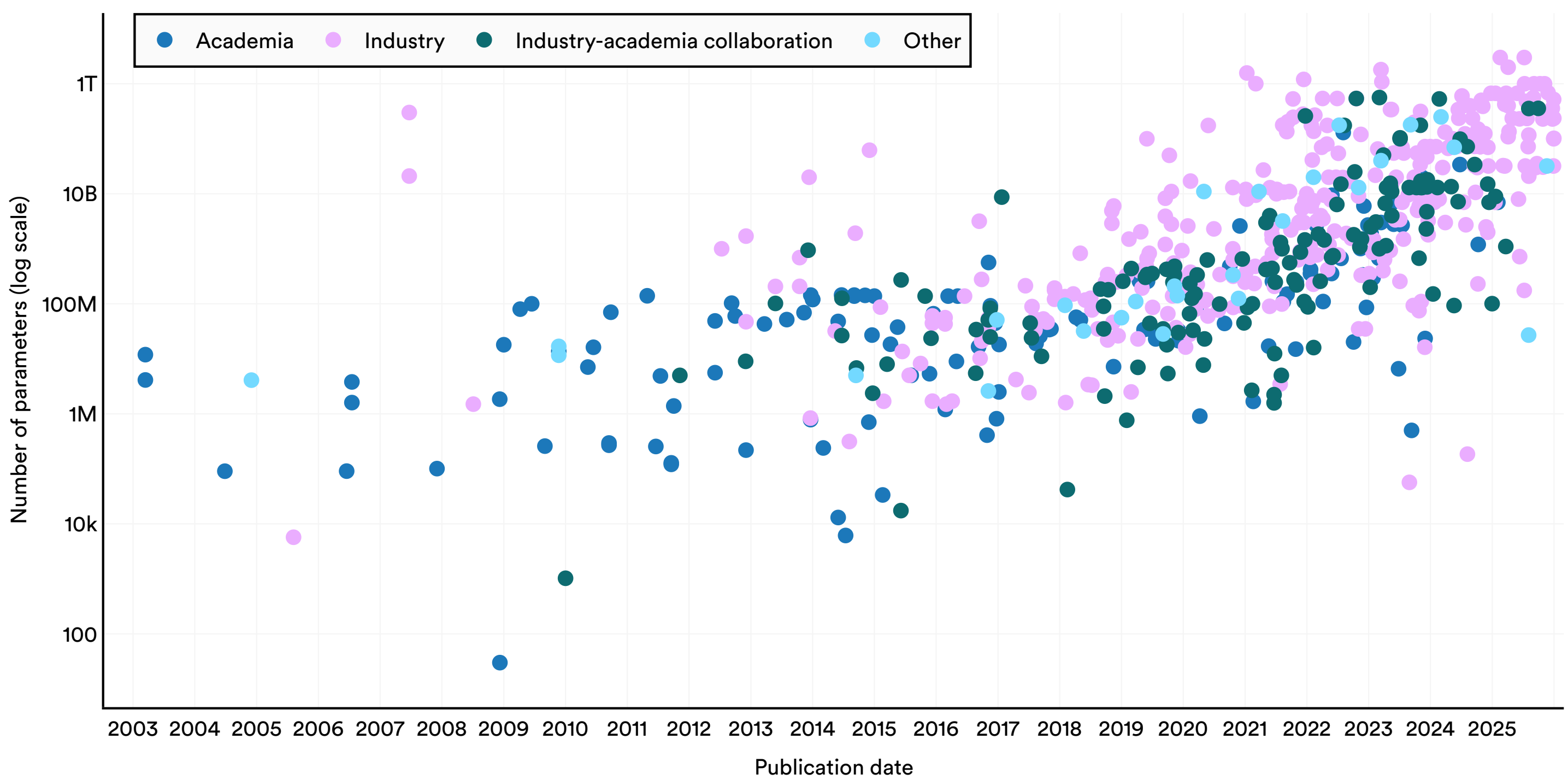

Figure 1.1.10

8 Several of the figures in this section use a log scale to reflect the exponential growth in AI model parameters and compute in recent years.

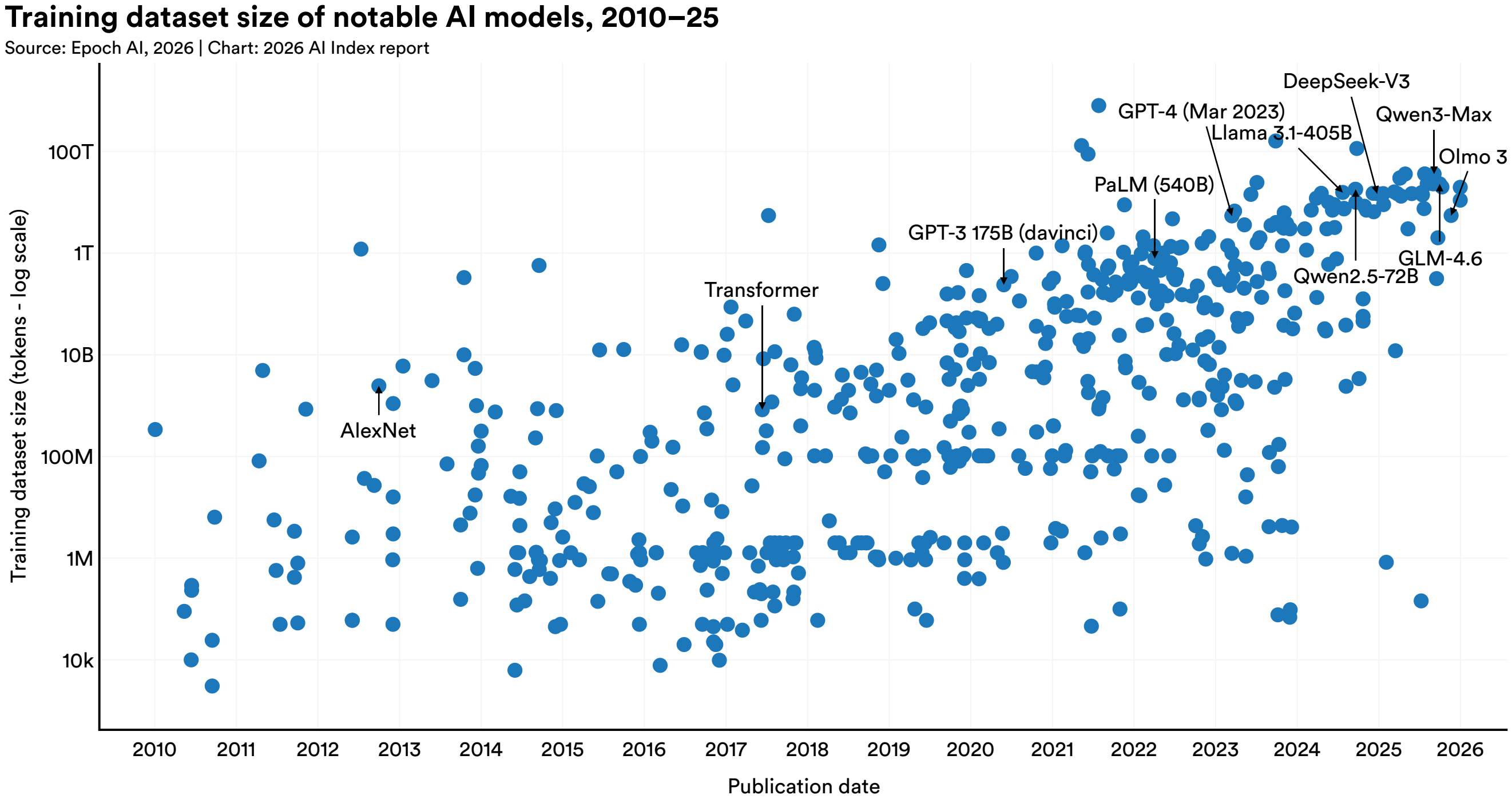

Figure 1.1.11

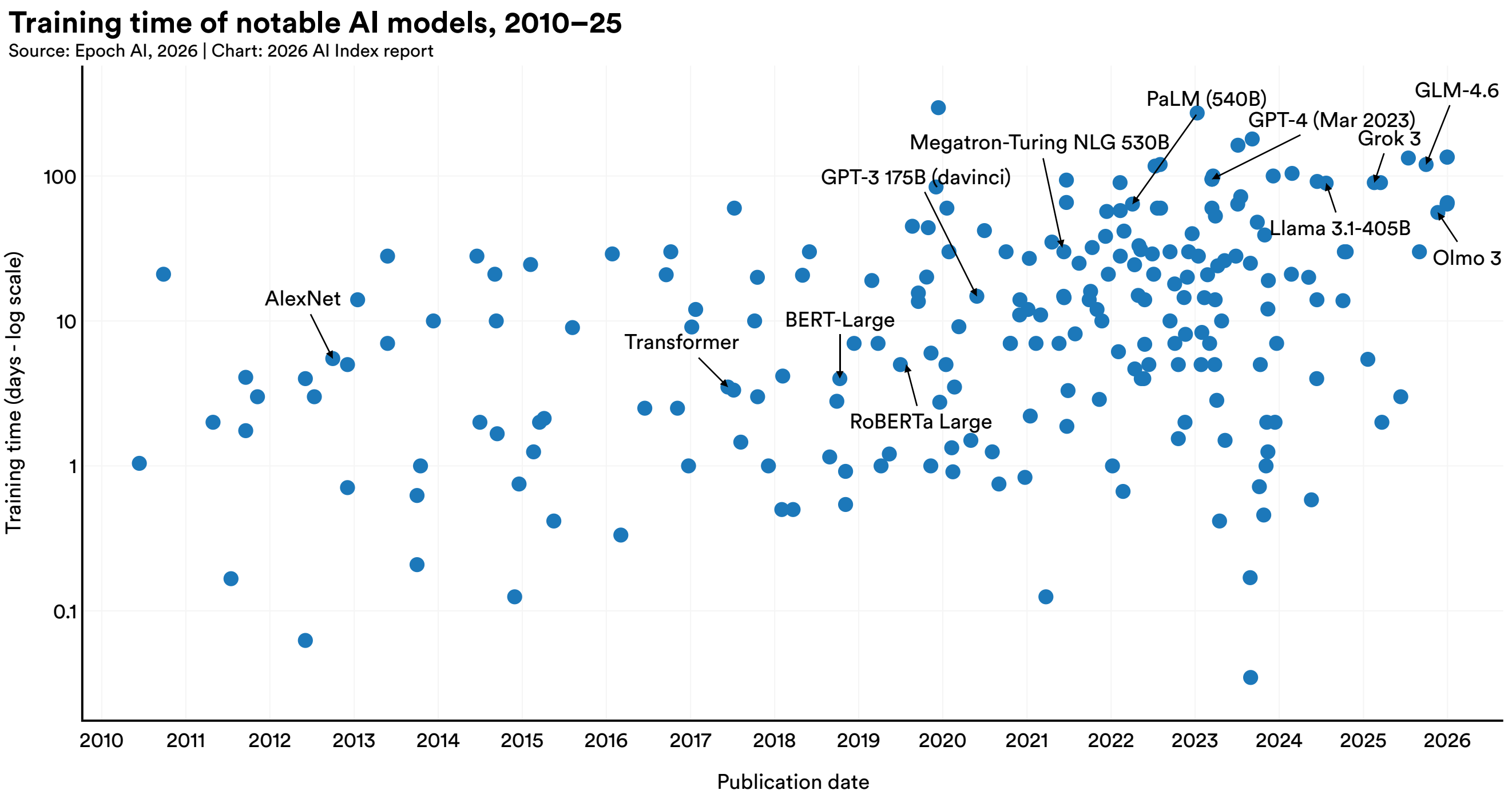

Figure 1.1.12

Since compute can be estimated even when not directly reported, training compute trends for notable models show clear growth over the same period (Figures 1.1.13 and 1.1.14). Compute requirements for notable models have risen by several orders of magnitude, with industry accounting for the highest values. When comparing the two countries with highest model output, U.S. models continue to be the most computationally intensive compared to Chinese models. However, the comparison in recent years cannot be fully substantiated because U.S. models have not directly reported their training compute.

## Training compute of notable AI models by sector, 2003–25

Source: Epoch AI, 2026 | Chart: 2026 AI Index report

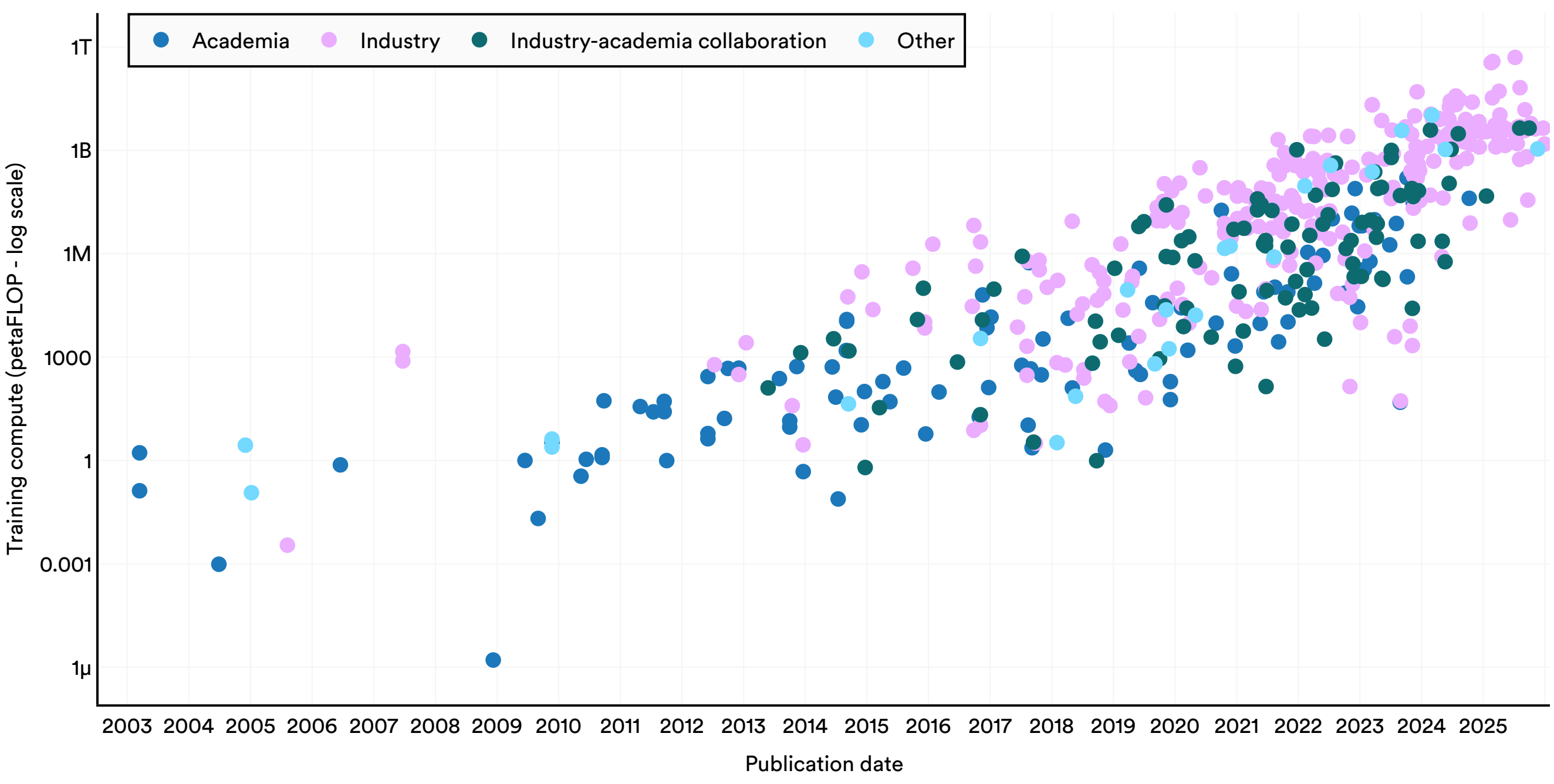

Figure 1.1.13[9]

## Training compute of select notable AI models in the United States and China, 2018–25

Source: Epoch AI, 2026 | Chart: 2026 AI Index report

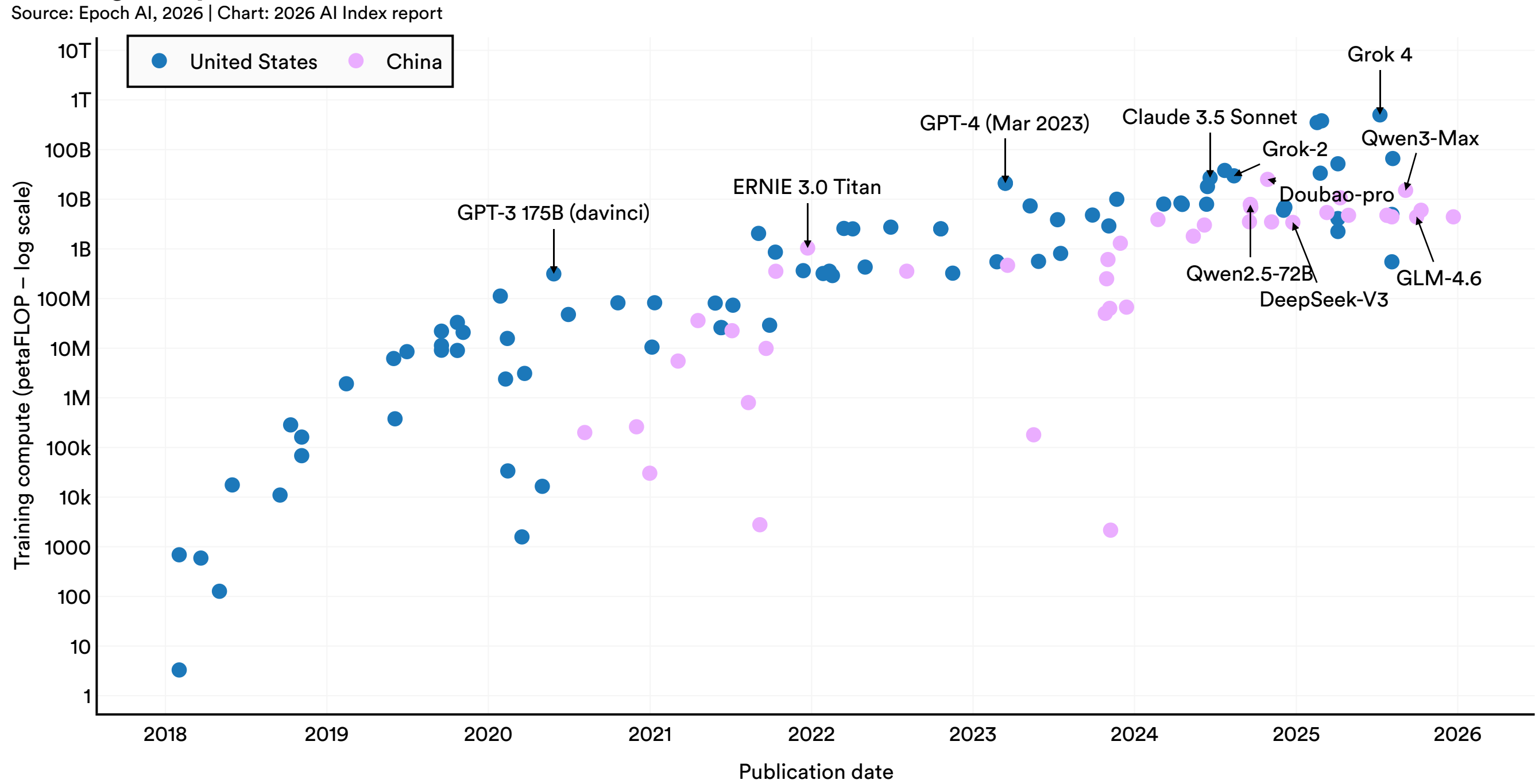

Figure 1.1.14

9 Estimating training compute is an important aspect of AI model analysis, yet it often requires indirect measurement. When direct reporting is unavailable, Epoch estimates compute by using hardware specifications and usage patterns or by counting arithmetic operations based on model architecture and training data. In cases where neither approach is feasible, benchmark performance can serve as a proxy to infer training compute by comparing models with known compute values. Full details of Epoch's methodology can be found in the documentation section of their website.

HIGHLIGHT:

# Will Models Run Out of Data?

Last year, the AI Index highlighted concerns around data bottlenecks and the sustainability of the scaling approach as it relates to training data. Leading AI researchers have publicly claimed that the available pool of high-quality human text and web data for training large models has been exhausted, a state often referred to as “peak data.” This has continued to raise industry-wide concerns about the sustainability of scaling laws, which have historically depended on ever-larger datasets. One set of projections from Epoch AI suggests that, under certain assumptions, the estimated depletion date could fall between 2026 and 2032.

## Synthetic Data in Pre-training

Limits on the availability of real-world data may be less consequential if synthetic data (data generated by AI systems) can be used to improve the performance of subsequent models. Previous editions of the AI Index found no definitive evidence that synthetic data improves model performance during the pre-training phase.[10] The 2024 report referenced research suggesting that model performance can collapse when real training data is replaced with synthetic data. The 2025 report noted more recent findings that such collapse can be avoided if real data remains part of the training set, but that simply adding more data does not necessarily lead to performance gains.

The consensus remains largely unchanged. There is still no definitive evidence that synthetic data can fully offset real-data depletion in pre-training contexts. However, recent research suggests that synthetic data may offer value in more limited settings. Hybrid training approaches, which combine real and synthetic data, can significantly accelerate training, sometimes by a factor of five to 10 at scale, without surpassing real data in final model performance. Training on purely synthetic data has shown promise for smaller models or narrowly defined tasks, such as classification, code generation, or work in low-resource languages, but these gains have not generalized to large, general-purpose language models. Where synthetic-only training has achieved performance comparable to real data, it has typically involved substantially smaller models that are not directly comparable to current state-of-the-art systems. For example, the SYNTHLLM family of models, trained entirely on synthetic data, achieves strong results yet still lags behind leading models on major benchmarks (Figure 1.1.15).

| Model | GSM8K | MATH | Minerva | Olympiad | College | Gaokao | Average |
|---|---|---|---|---|---|---|---|
| Llama-3.2-1B-Instruct | 47.2 | 28.0 | 5.9 | 5.5 | 18.8 | 25.2 | 21.8 |
| SYNTHLLM-1B (3.2M) | 45.7 | 32.9 | 6.6 | 7.6 | 27.3 | 28.8 | 24.8 |
| SYNTHLLM-1B (7.4M) | **50.4** | **37.4** | **8.1** | **8.3** | **31.7** | **30.6** | **27.4** |
| Llama-3.2-3B-Instruct | 78.0 | 47.5 | 17.3 | 14.8 | 31.8 | 37.7 | 37.9 |
| SYNTHLLM-3B (3.2M) | 78.1 | 54.3 | 16.9 | 17.5 | 38.3 | 46.5 | 41.9 |
| SYNTHLLM-3B (7.4M) | **80.7** | **60.0** | **18.8** | **21.9** | **42.3** | **50.9** | **45.8** |
| Llama-3.1-8B-Instruct | 84.2 | 48.9 | 25.7 | 13.2 | 32.1 | 43.4 | 41.2 |
| Llama-3.1-70B-Instruct | **94.5** | 66.1 | **34.2** | 29.6 | 41.4 | 56.6 | 54.2 |
| NuminaMath-CoT-7B | 75.4 | 55.2 | 19.1 | 19.9 | 36.9 | 47.5 | 42.3 |
| NuminaMath-CoT-72B | 90.8 | 66.7 | 25.0 | 32.6 | 39.7 | 54.0 | 51.5 |
| MAmmoTH2-Plus-8B (10M) | 78.4 | 41.9 | 10.7 | 11.3 | 16.1 | 31.9 | 31.7 |
| JiuZhang3.0-8B (6M) | 88.7 | 51.2 | 21.7 | 18.8 | 37.5 | 43.4 | 43.6 |
| NaturalReasoning-8B (2.8M)* | - | 55.6 | - | - | - | - | - |
| OpenMathInstruct-2-8B (14M) | 91.1 | 67.5 | 22.5 | 27.7 | 39.2 | 53.5 | 50.3 |
| SYNTHLLM-8B (3.2M) | 88.4 | 66.1 | 25.4 | 30.2 | 44.3 | 56.9 | 51.9 |
| SYNTHLLM-8B (7.4M) | 92.1 | **71.3** | 26.5 | **33.0** | **45.3** | **61.0** | **54.9** |

Source: Qin et al., 2025

Figure 1.1.15

10 Pre-training refers to the initial phase of model development in which a model is trained (typically via self-supervised learning) on large, general-purpose datasets to acquire broad linguistic or multimodal representations. Post-training refers to subsequent refinement of the base model, through techniques such as supervised fine-tuning or reinforcement learning, to specialize behavior, improve alignment, or optimize performance on particular tasks.

**HIGHLIGHT:**

### Data-centric Methods

Discussions on data availability often overlook an important shift in recent AI research. Performance gains are increasingly driven by improving the quality of existing datasets, not by acquiring more. Rather than scaling data indiscriminately, researchers are spending more effort in pruning, curating, and refining training inputs. Data-centric methods emphasize performance improvements through practices such as cleaning labels, deduplicating samples, and constructing higher-quality datasets. A growing body of research shows that training models on low-quality or polluted data can significantly degrade performance. Likewise, recent evidence illustrates that data pruning, selecting the most informative training inputs, often outperforms approaches that train on all available data indiscriminately.

Recent large-scale model development illustrates this paradigm in practice. Olmo 3 researchers prioritized large-scale deduplication, quality-aware data selection, and stage-specific training curricula rather than indiscriminate data scaling. These interventions, combined with iterative feedback loops to evaluate and refine candidate data mixes, allowed their models to achieve competitive performance despite training on substantially fewer tokens than other leading state-of-the-art models (Figure 1.1.16). Olmo 3.1's Think 32B model, for example, contains roughly 32 billion parameters, nearly 90 times fewer than Grok 4's 3 trillion, yet it achieves comparable performance on several benchmarks, including American Invitational Mathematics Examination (AIME)[11] 2025.

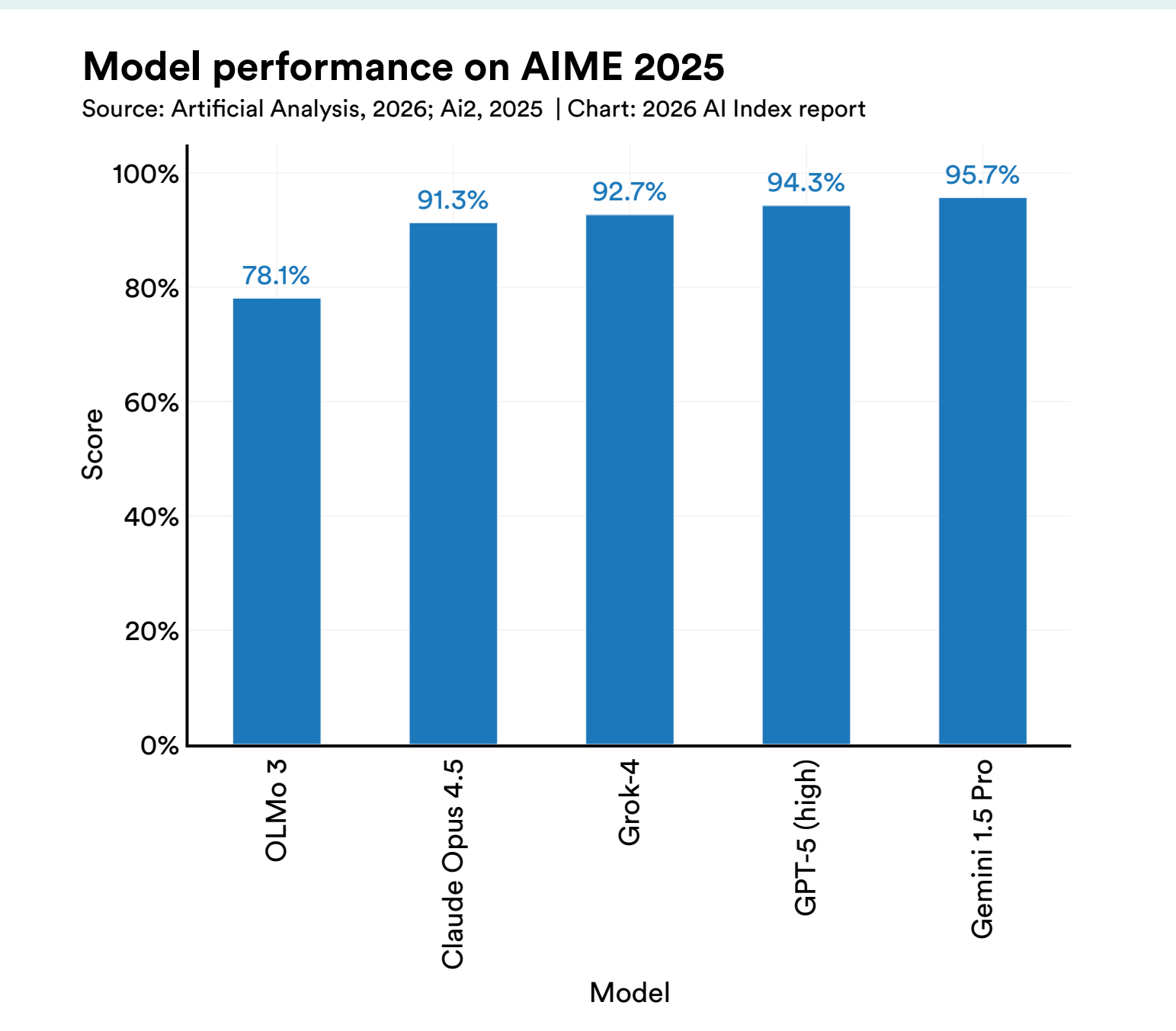

Figure 1.1.16

### Synthetic Data in Post-training

Recent research shows that synthetically generated data can be effective for improving model performance in post-training settings, including fine-tuning, alignment, instruction tuning, and reinforcement learning. A growing body of research released in 2025 supports this finding. Evidence suggests that synthetic post-training data is effective in few-shot generation settings, for improving long-context capabilities, for optimizing reinforcement learning workflows, and for strengthening reasoning more broadly.

### Prevalence of Synthetic Content

Since the launch of ChatGPT in November 2022, there have been predictions that the internet would soon become overrun by AI-generated content. Recent research from Graphite suggests that beginning in January 2025, over 50% of newly published online content was generated by AI (Figure 1.1.17). Others have projected that the share in 2026 could be even higher.

11 The American Invitational Mathematics Examination (AIME) is an annual high school math competition widely used as a benchmark for AI mathematical reasoning, with each year's exam providing a fresh test set.

**HIGHLIGHT:**

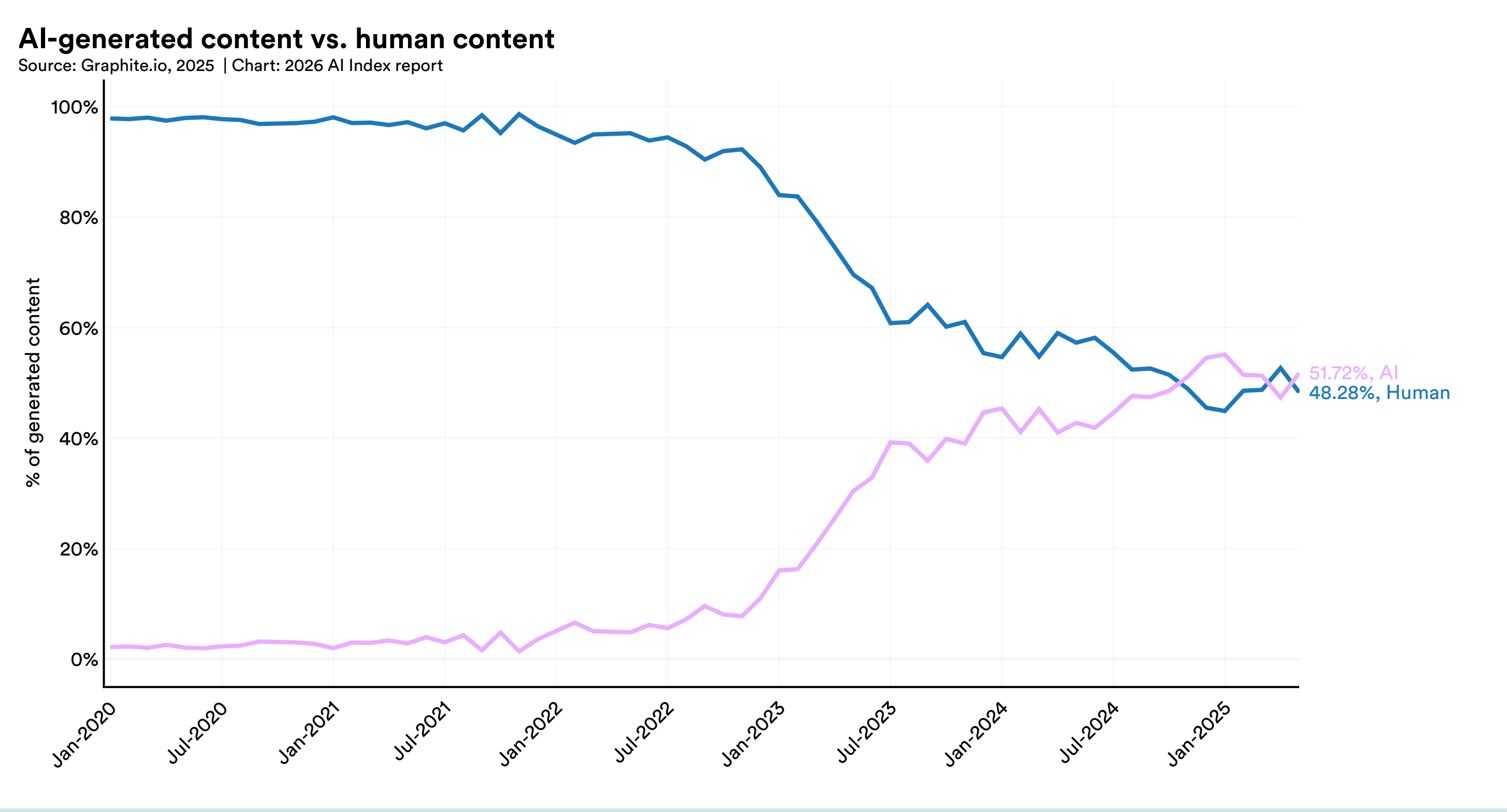

Figure 1.1.17

Given growing concerns about the suitability of synthetic data for training AI systems, this trend raises questions about the long-term reliability of current scaling trajectories. In response, many firms that depend on high-quality training data have increasingly turned to proprietary sources. In May 2025, the New York Times entered into a licensing agreement with Amazon to allow its content to be used for training purposes. By mid-2025, Meta was reportedly engaged in similar discussions with news organizations, while health and life sciences companies such as Bristol Myers Squibb have pursued comparable strategies. These developments suggest that firms training frontier AI systems are adjusting their data acquisition strategies as the volume of openly available training data continues to decline.

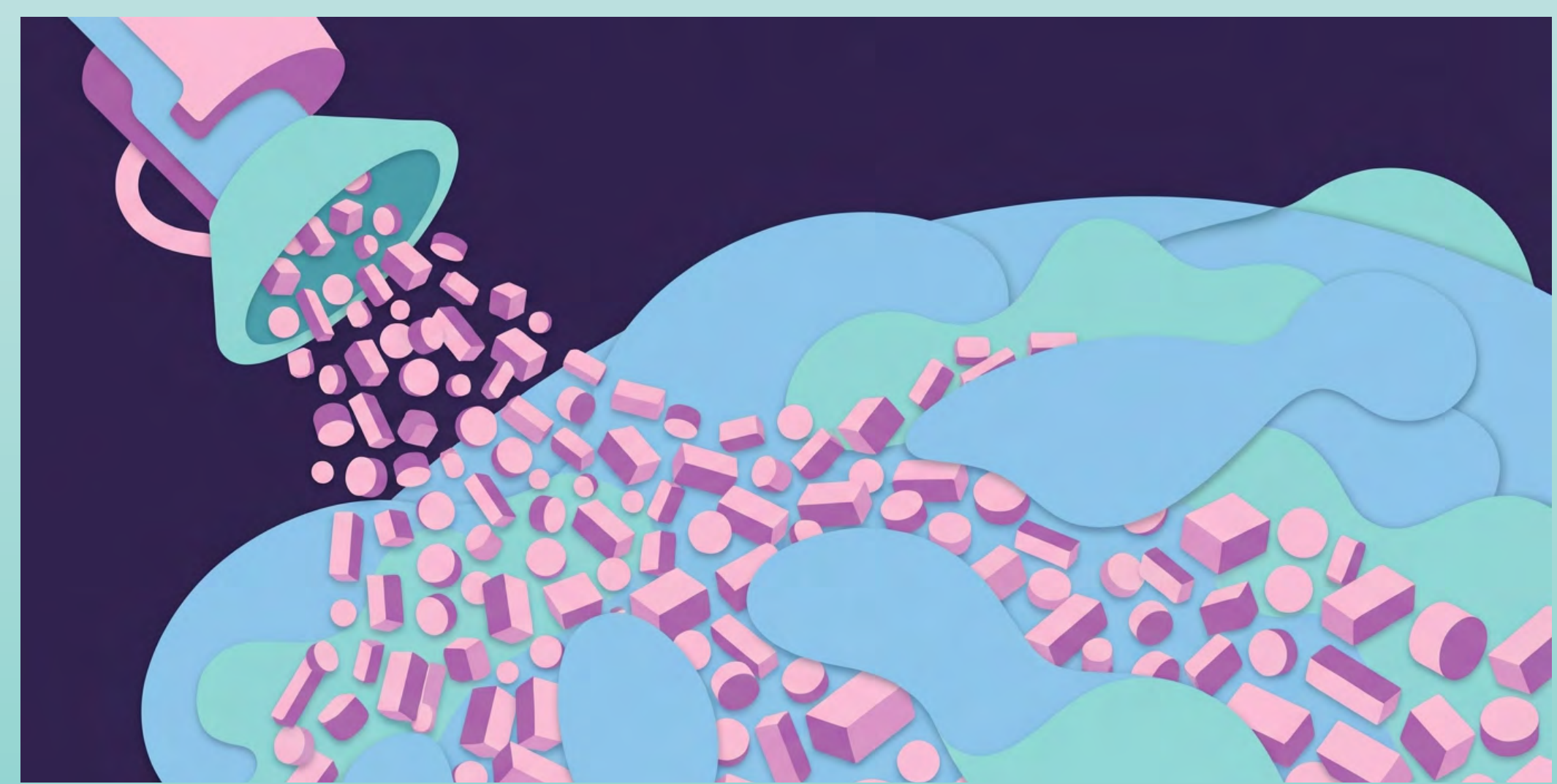

# 1.2 Compute and Infrastructure

The development of AI models requires significant infrastructure investment. As training processes have expanded in scale and complexity, the underlying hardware has also improved in both speed and efficiency. In turn, these gains shape what kinds of models researchers and labs can realistically build. The growth in training compute discussed in the previous section would not have been possible without corresponding improvements in hardware capabilities. This section leverages data from Epoch AI to track hardware performance, adoption, and aggregate computing capacity over time.

## Performance and Efficiency

Peak computational performance of machine learning hardware has increased exponentially across releases between 2008 and 2025 (Figure 1.2.1). The gains are especially visible at lower precision types, where precision refers to the number of bits used to represent numerical values. Lower precision formats such as FP16 and Tensor-FP16/BF16 now show the highest performance levels and have become standard in many training and inference settings.

**Peak computational performance of ML hardware for different precisions, 2008–25**

Source: Epoch AI, 2026 | Chart: 2026 AI Index report

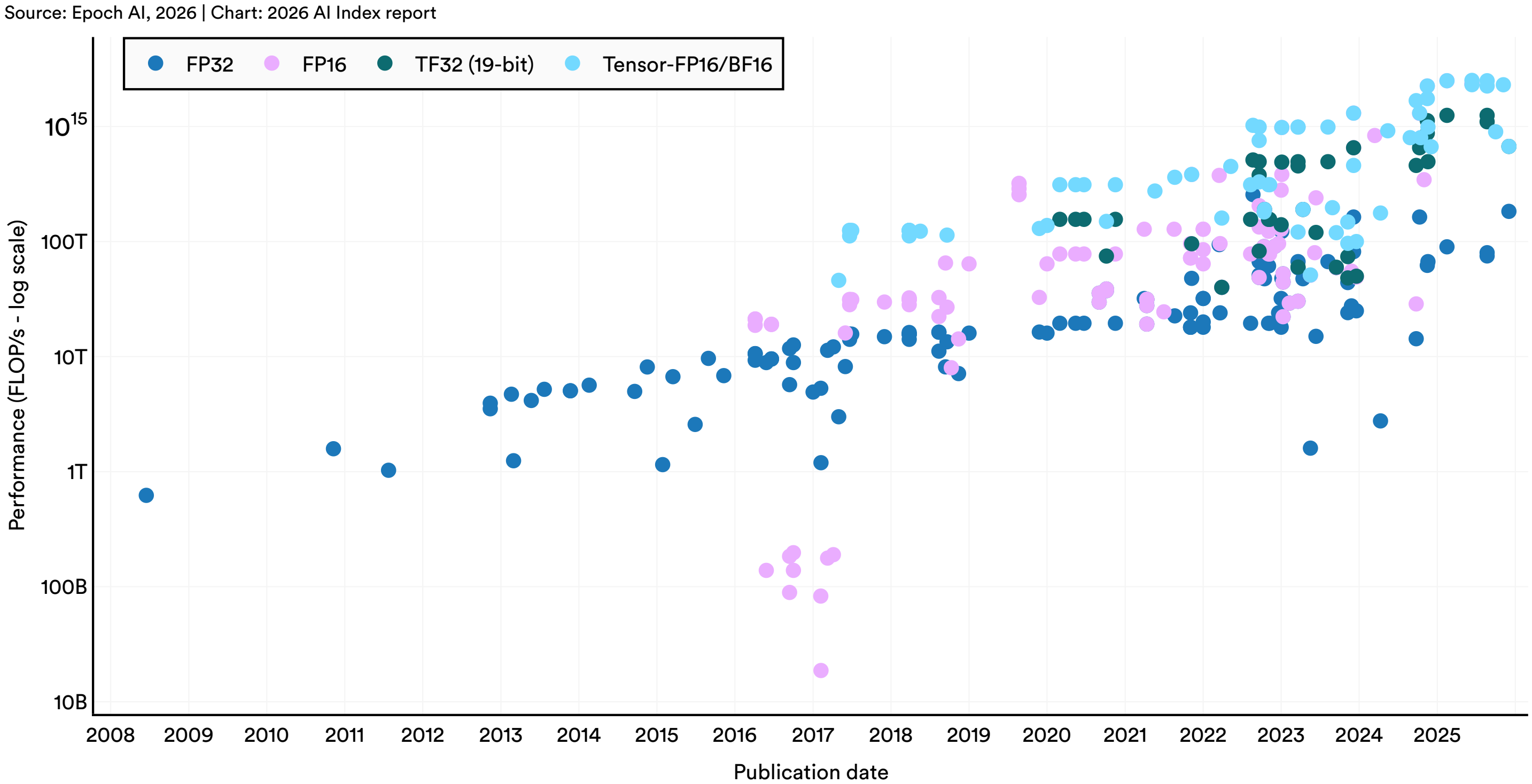

Figure 1.2.1

## Hardware for Notable Models

Hardware adoption patterns among notable AI models reflect the gains in performance and efficiency (Figure 1.2.2). Since 2017, the cumulative number of notable models trained on A100-class hardware has increased, with 84 models trained in 2025. The previous generation, V100, continues to power a sizable share (69 models). Newer hardware, such as the H100, has seen early rapid adoption (28), while other categories, such as TPU v3 and TPU v4, show stable curves.

**Cumulative number of notable AI models trained by accelerator, 2017–25**

Source: Epoch AI, 2026 | Chart: 2026 AI Index report

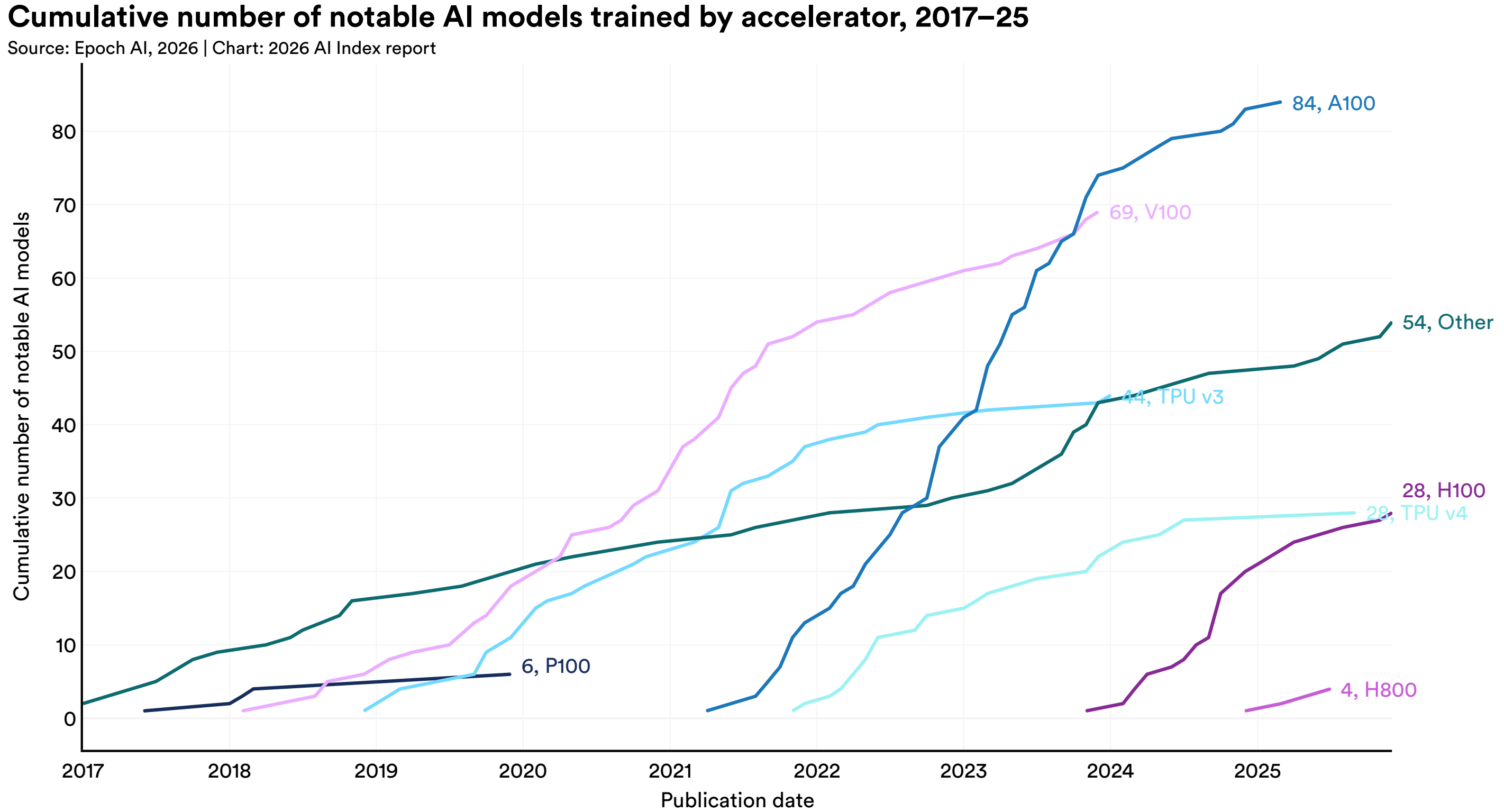

Figure 1.2.2

## Global Computing Capacity

The supply of AI computing capacity from major chip designers has continued to increase (Figure 1.2.3). Total capacity has increased by an estimated 3.3x per year since 2022, reaching approximately 17.1 million H100-equivalents.[12] Nvidia AI chips currently account for over 60% of total compute, with Google and Amazon supplying much of the remainder and Huawei holding a small but growing share. The growth in compute capacity tracks closely with investment patterns described in Chapter 4, where leading AI companies have increased their capital expenditure and infrastructure has become the fastest growing focus area of private AI funding.

12 Since these estimates are inferred from revenue data, financial disclosures and analyst reports, they reflect broader trends rather than exact counts. Data coverage also varies by manufacturer; Nvidia and Google data starts in 2022, while others start in 2024.

**Global computing capacity from AI chips across major designers, 2022–25**
Source: Epoch AI, 2026 | Chart: 2026 AI Index report

Figure 1.2.3

## Data Center Power Capacity

The expansion of computing capacity carries a direct energy cost. Total AI data center power capacity reached approximately 29.6 GW by Q4 2025, enough to power all of New York state at peak demand (Figure 1.2.4). AI chip power, measured by thermal design power, accounted for roughly 11.8 GW of the total, with the remainder attributed to cooling, networking, and other data center infrastructure. This estimate is based on the rated power capacity of leading AI chips sold over time, with a multiplier of approximately 2.5 applied to account for the additional requirements of powering infrastructure.

**Global AI data center power capacity, 2022–25**
Source: Epoch AI, 2026 | Chart: 2026 AI Index report

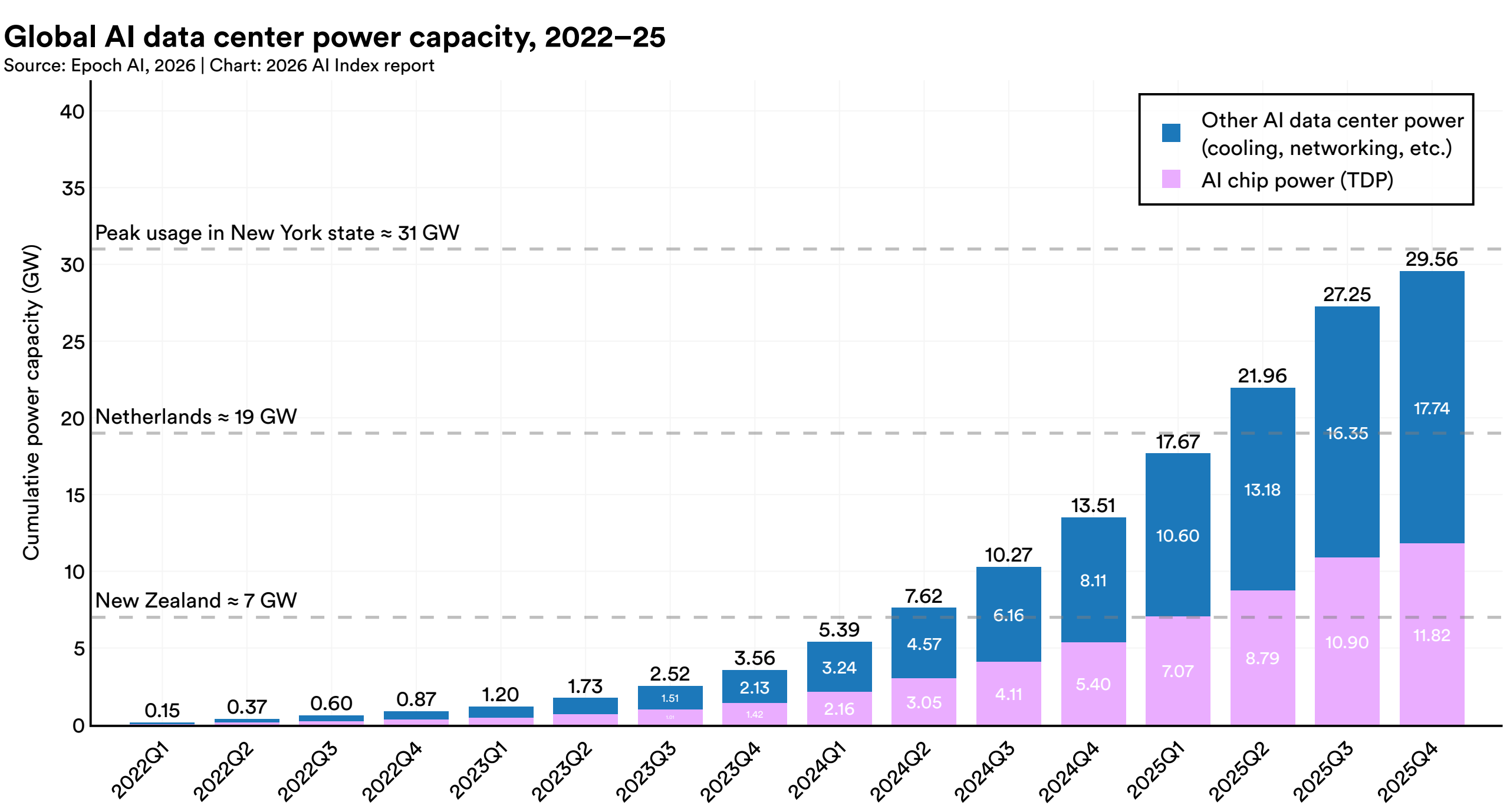

Figure 1.2.4

# 1.3 Data Centers

The physical infrastructure underlying AI development extends beyond models and compute described in the previous section. Data centers are where compute is housed, and their capacity, geographic distribution, and underlying supply chains shape what AI systems can be built and where. This section draws on data from Cloudscene to track the global distribution of data centers and introduces an overview of the broader AI infrastructure ecosystem to provide context for the geographic and supply chain dynamics.

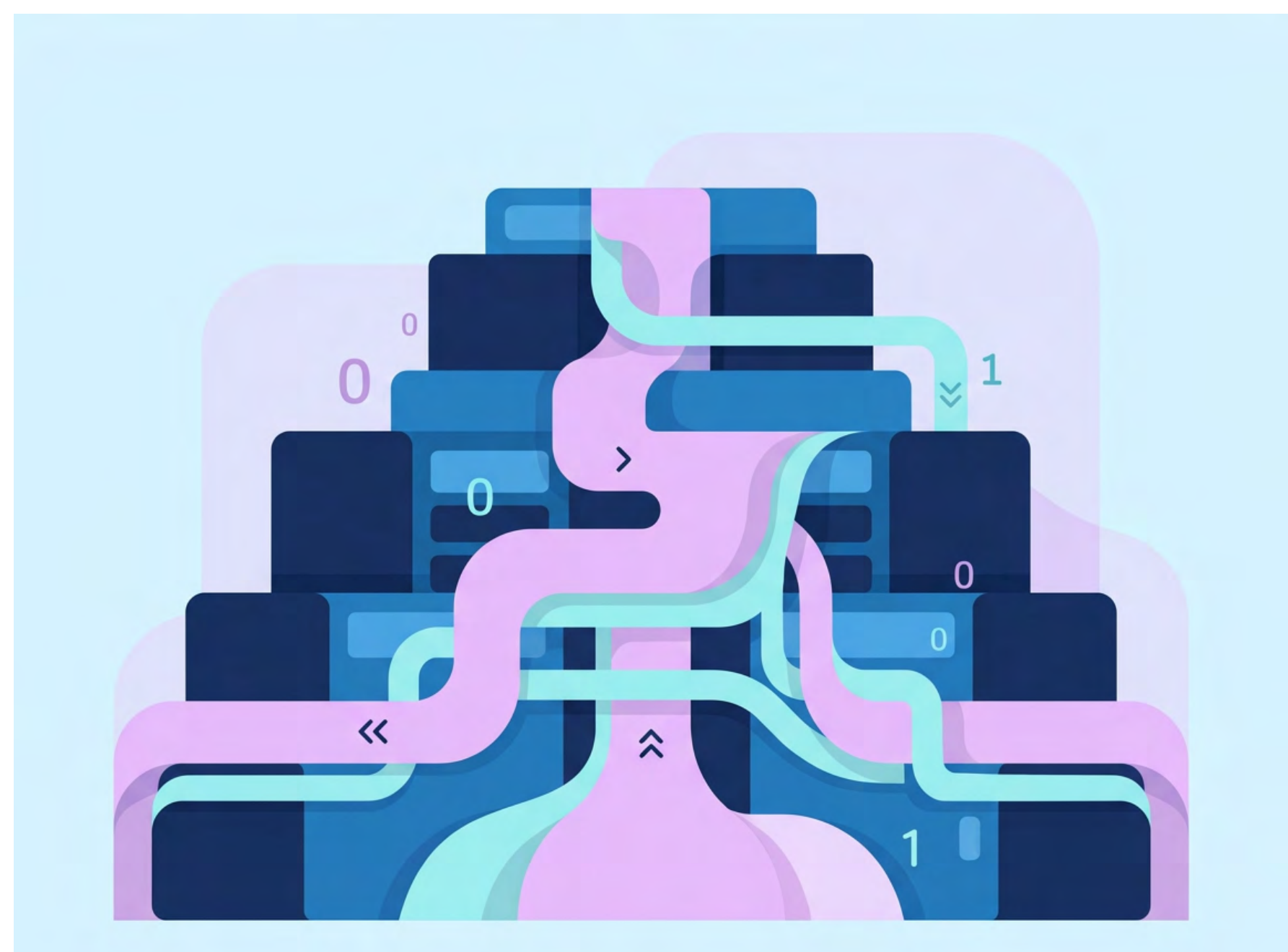

## AI Infrastructure: Beyond GPUs

Modern AI data centers depend on a combination of compute, storage, communications, and specialized hardware that enables AI systems to run at large scale. GPUs and custom accelerators such as Tensor Processing Units (TPUs) are the most widely discussed, but they are only one layer of a broader infrastructure stack. All data processed by these chips is held in high-bandwidth memory (HBM), which supports moving large volumes of data in and out efficiently. The leading manufacturers of HBM are SK Hynix (South Korea), Samsung (South Korea), and Micron (USA). During training, GPUs must continuously share data with one another, which requires fast, high throughput network connectivity achieved with fiber-optic cables running high-bandwidth networking architectures such as InfiniBand.

The supply chain behind this hardware adds another dimension. Companies like Nvidia and SK Hynix design but do not manufacture chips. Instead, they provide designs to specialized semiconductor foundries, primarily the Taiwan Semiconductor Manufacturing Company (TSMC) and Samsung Foundry, which fabricate the chips at the nanometer scales modern AI hardware requires. The fabricated chips are then packaged and tested by assembly companies such as ASE Group (Taiwan) and Amkor Technology (United States). TSMC is a single point of dependency in the global AI supply chain, as it fabricates virtually every leading AI chip, including Nvidia's Blackwell GPUs and AMD's MI300X. There are high barriers to entry at every layer—requiring decades of accumulated expertise, specialized equipment, and significant capital investment to overcome.

The infrastructure ecosystem is relevant beyond AI capabilities, as it shapes education priorities and workforce development. Chapter 7 (Education) distinguishes between AI software-related and AI hardware-related degrees. That distinction is also relevant here, where different countries play different roles across the hardware supply chain.

## Geographic Distribution

Most of the world's data center infrastructure is located in a small number of countries (Figures 1.3.1 and 1.3.2). In 2025, the United States led by a wide margin, with 5,427 data centers, more than 10 times the count of any other country. Germany (529), the United Kingdom (523), and China (449) followed, while the majority of the remaining countries each had fewer than 300 facilities. The U.S. may show a clear lead, but the other country rankings should be assessed with the understanding that data center counts do not capture differences in facility size, computing capacity, or utilization.

**Global distribution of data centers, 2025**
Source: Cloudscene, 2025 | Chart: 2026 AI Index report

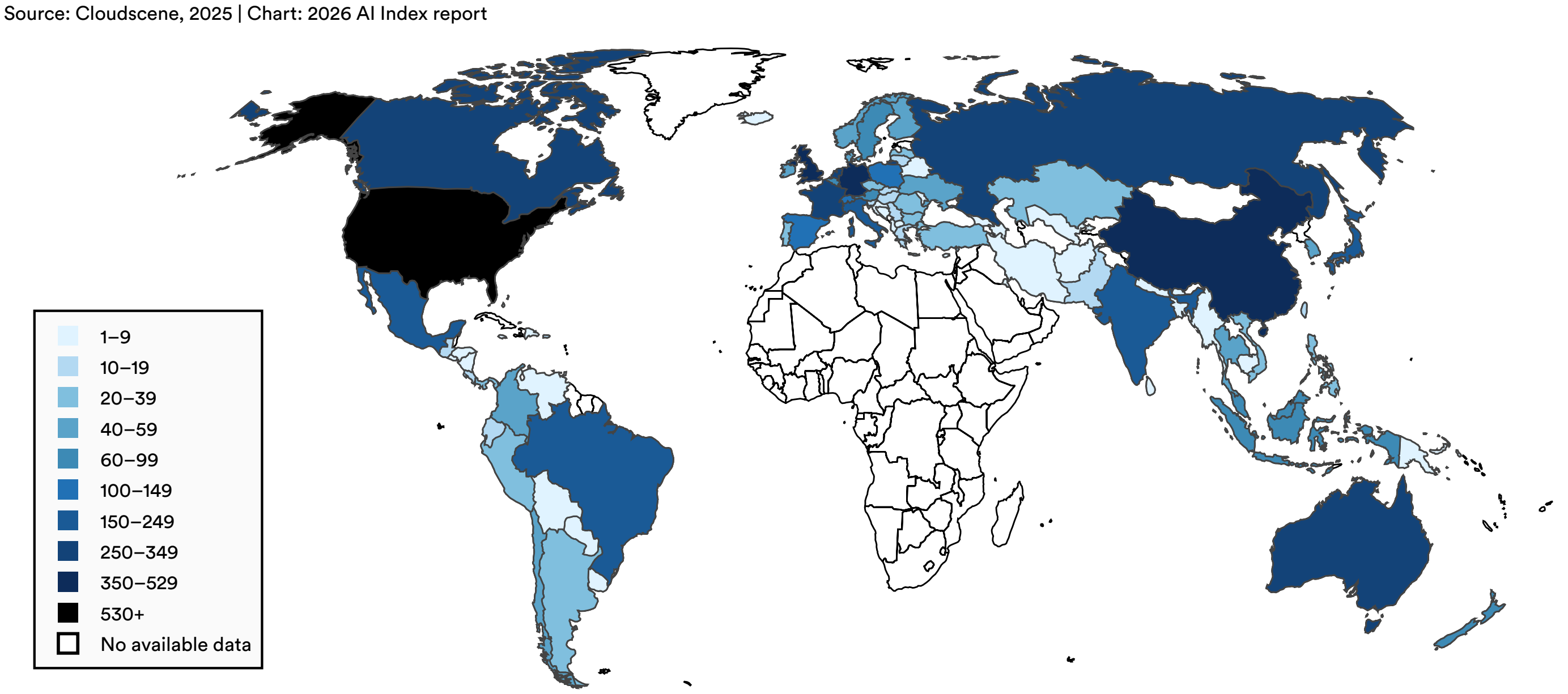

Figure 1.3.1

**Number of data centers by geographic area, 2025**
Source: Cloudscene, 2025 | Chart: 2026 AI Index report

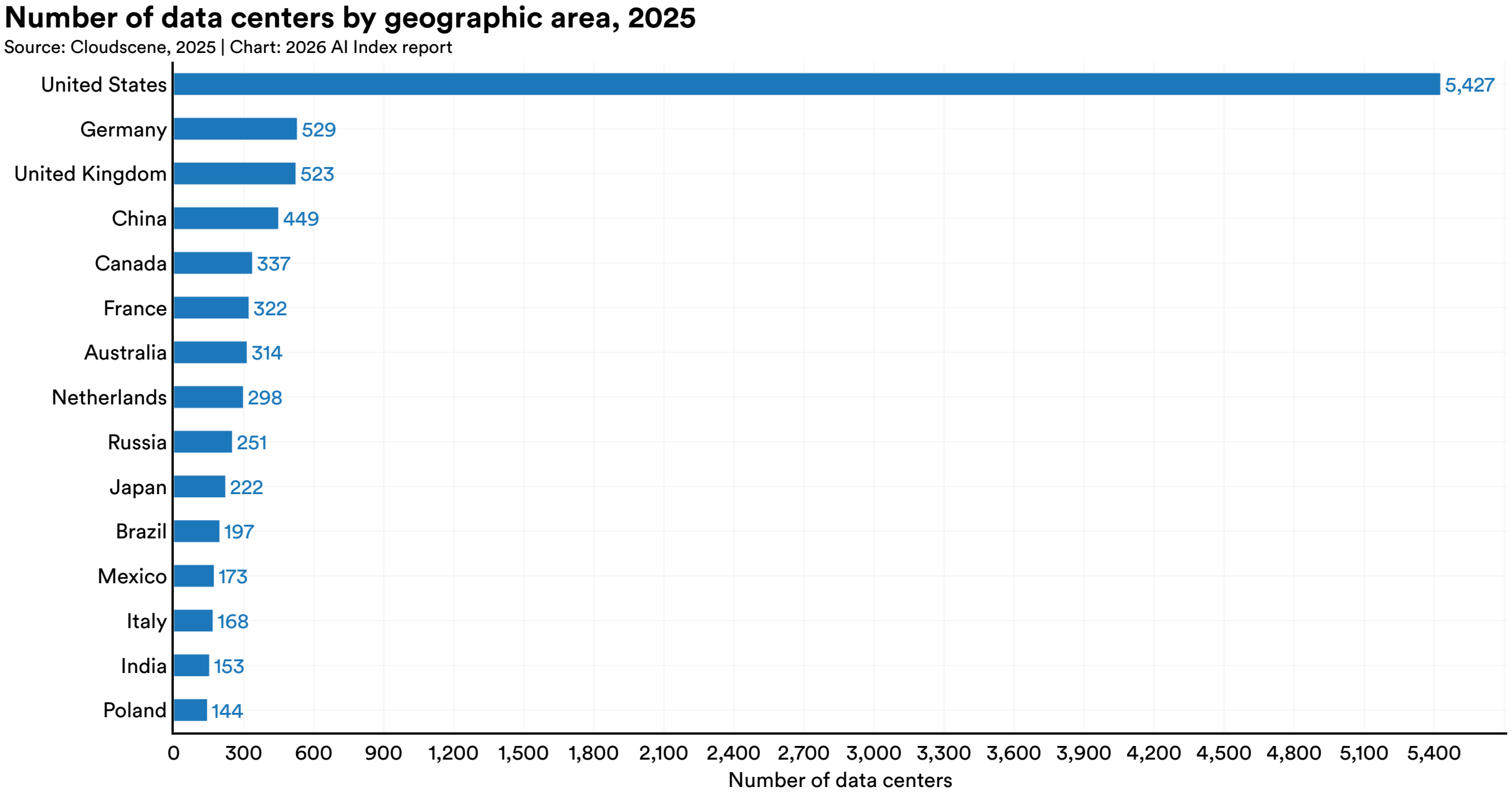

Figure 1.3.2

# 1.4 Energy and Environmental Impact

As AI systems have scaled and become more widely deployed, their energy consumption and environmental footprint have become very visible. The compute and infrastructure trends described in the preceding sections translate into heavy demands on energy, water, and carbon emissions. This section examines those costs across three areas of AI development: training, inference, and data center energy usage. The analysis draws from Epoch AI's model-level data, recent academic benchmarking research (Jegham et al., 2025), the International Energy Agency's reporting on data centers (IEA, 2025), and de Vries and Gao (2025).

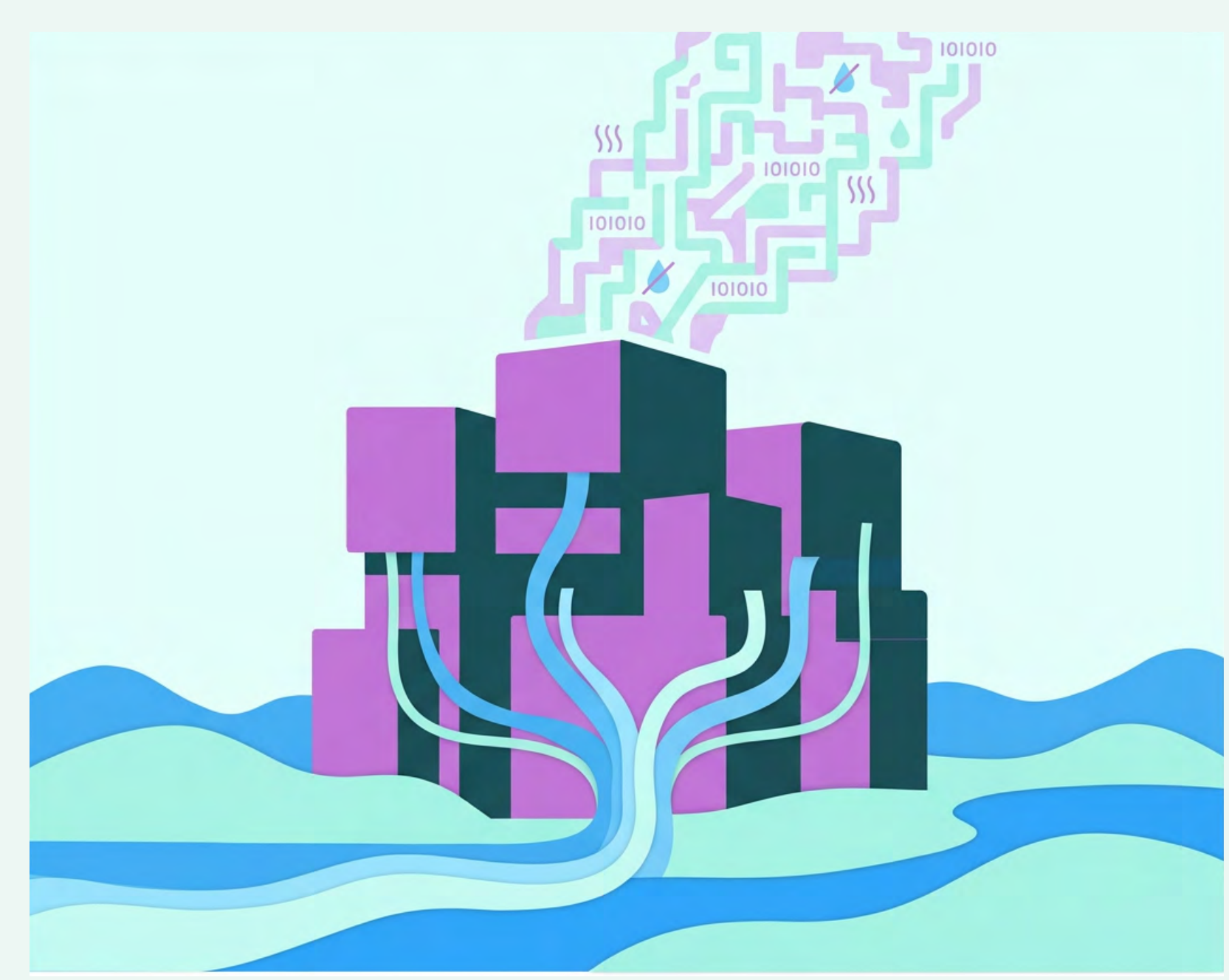

## Training

Leading machine learning hardware has grown more efficient since 2016, as measured in FLOP/s per watt (Figure 1.4.1). Leading chips deliver about 10 times more computation per watt than those available a decade ago, with Nvidia B200 and Google TPU v5e among the most efficient. However, models have scaled faster than efficiency has improved, so total power required to train frontier systems has continued to increase. Total power draw for training models has grown by several orders of magnitude since the early 2010s (Figure 1.4.2). The most compute-intensive models in the data set, such as Grok 3 and Llama 4 Behemoth, required upward of 100 million watts during training. Due to limited disclosure by their developers, power draw information is not available for many of the newest models that have been released.

Carbon emissions from training have increased even more sharply (Figure 1.4.3). Training AlexNet in 2012 produced an estimated 0.01 tons of $CO_2$ equivalent, while training Grok 4 in 2025 produced about 72,816 tons. To put this into context, that is more than the lifetime carbon emissions of an average car (63 tons). Larger models generally produce more emissions but not always, as it can also depend on hardware efficiency, training duration, and the carbon intensity of the energy sources used. DeepSeek v3, for example, produced approximately 597 tons, which is much less than models of comparable size (Figure 1.4.4).

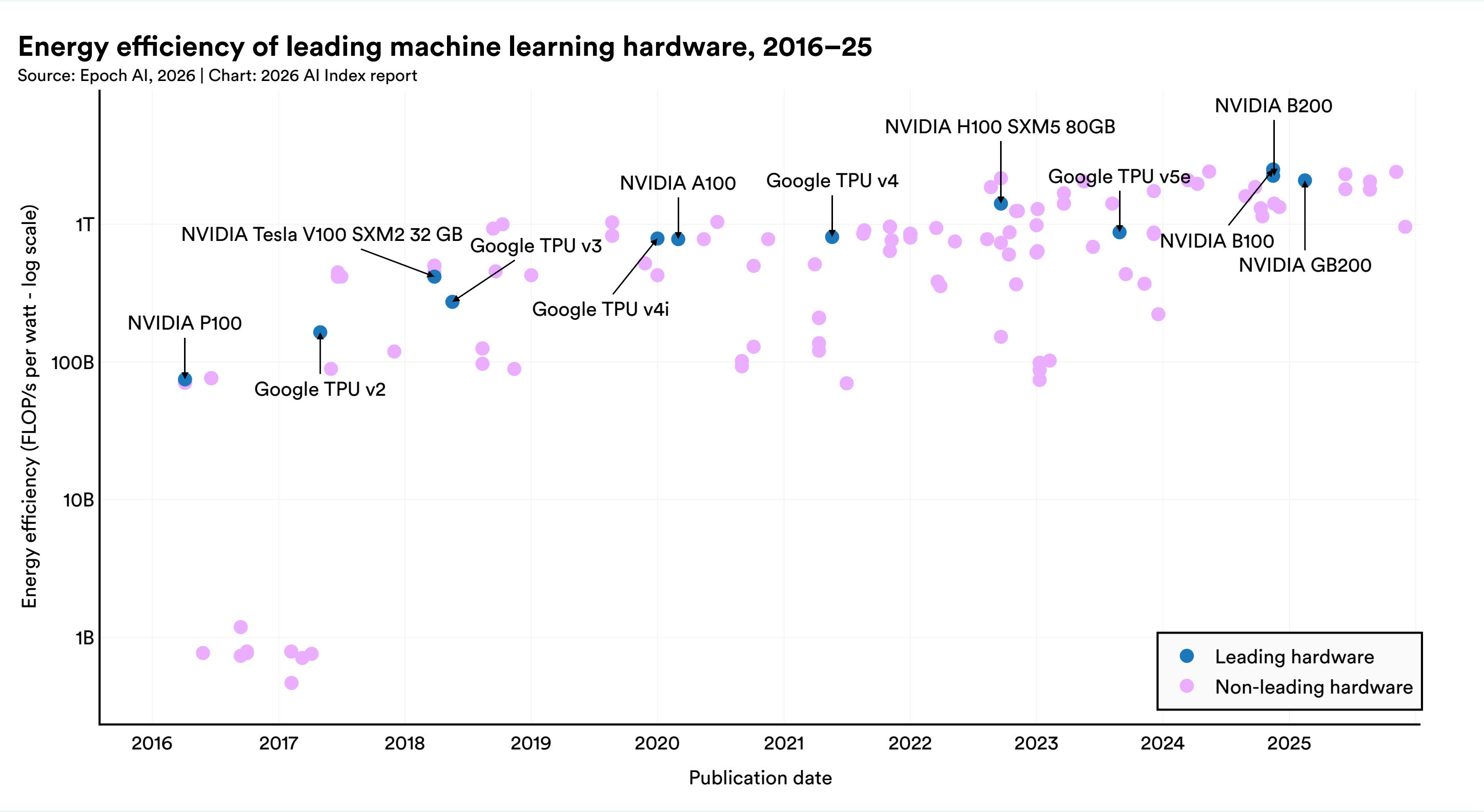

Figure 1.4.1

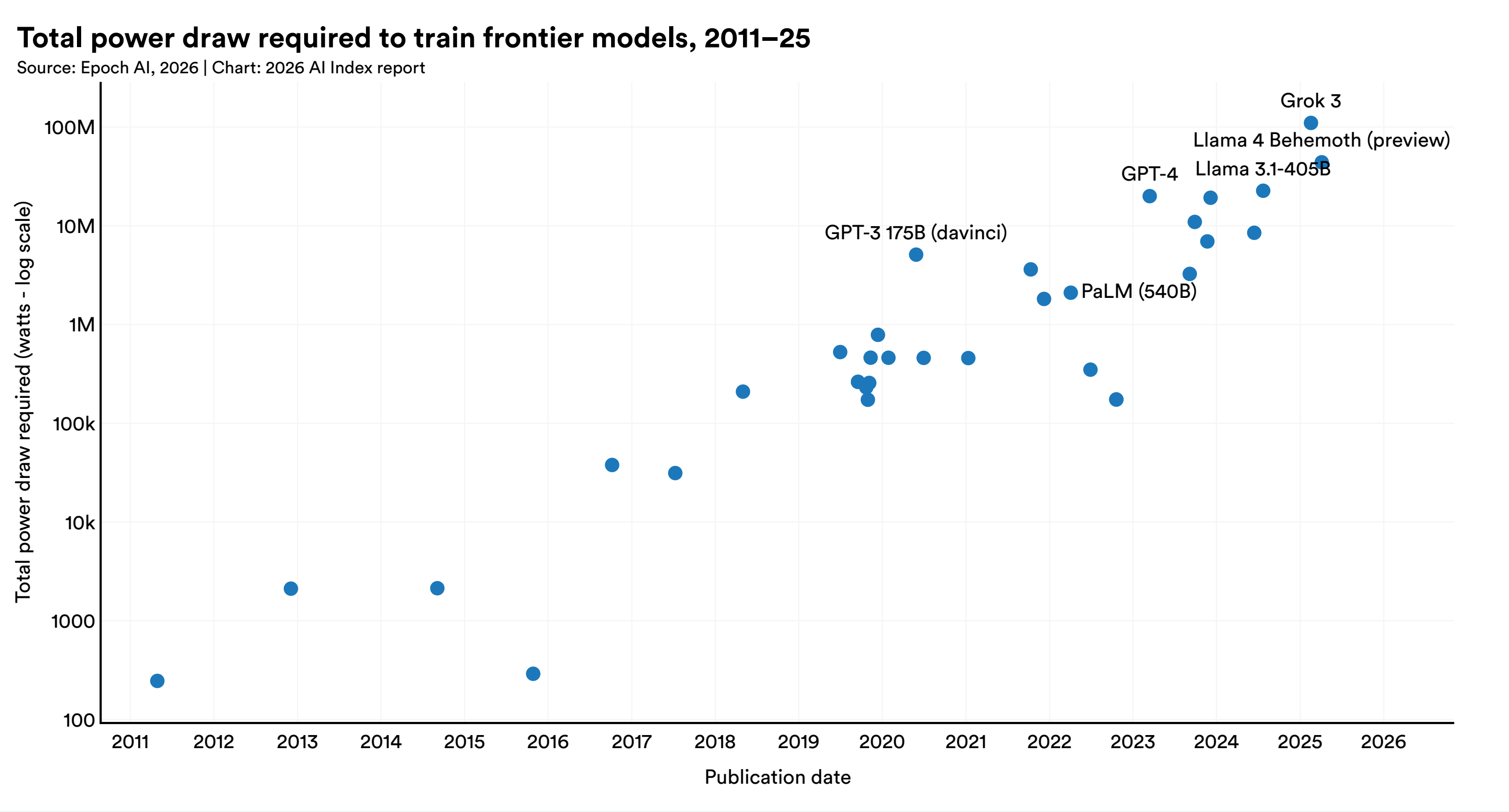

Figure 1.4.2

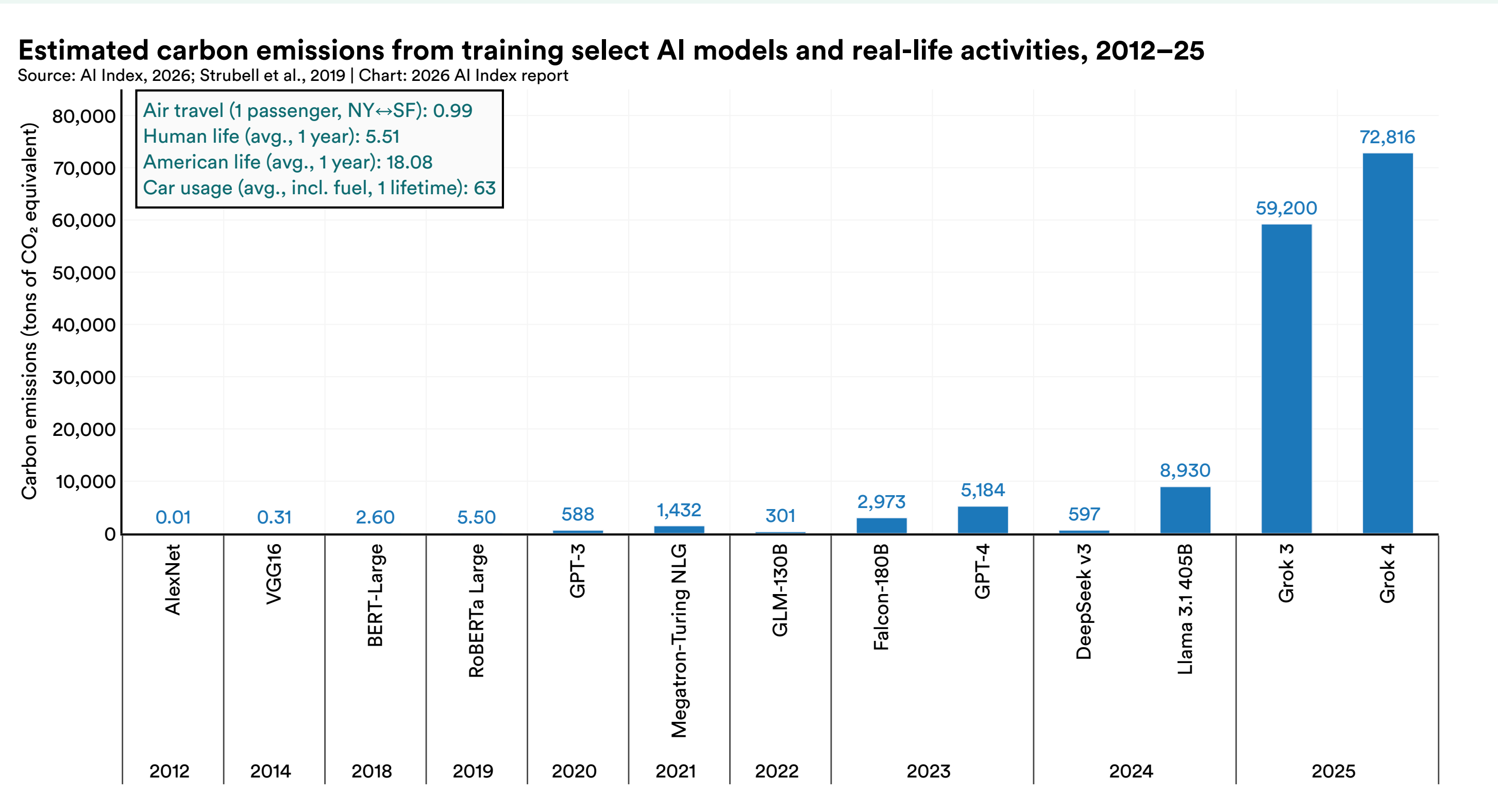

Figure 1.4.3

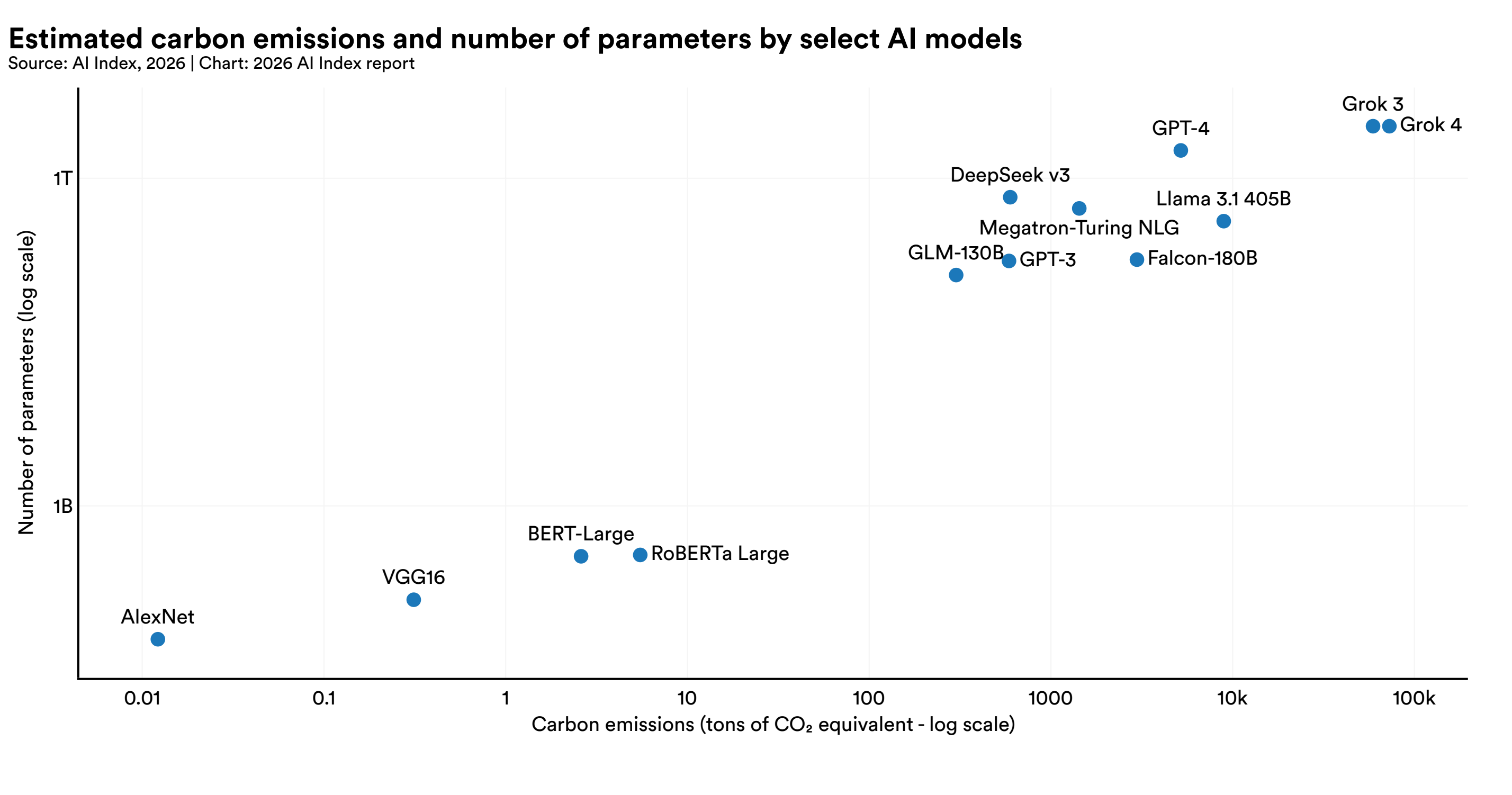

Figure 1.4.4

## Inference

Training costs have typically received the most attention, but inference represents a growing share of AI's total energy footprint. Once a model is deployed at scale, the cumulative energy required to serve queries can exceed the one-time cost of training within months.

Recent benchmarking by Jegham et al. (2025) provides per model estimates of inference energy consumption and carbon emissions for medium-length prompts (defined as approximately 1,000 input tokens and 1,000 output tokens). Among the top 15 models by energy consumption in 2025, DeepSeek V3.2 Exp and DeepSeek V3.2 consumed the most per query (23 Wh), followed by GPT-5 (high) at 21.9 Wh (Figure 1.4.5). Models such as Claude 4 Opus and GPT-5 min (medium) sit at the lower end, consuming between 5 and 6 Wh. When ranked by carbon emissions, the models also follow a similar pattern (Figure 1.4.6). DeepSeek V3.2 Exp and DeepSeek V3.2 produced the highest per medium-length prompt, approximately 14 grams of CO2 equivalent each. For comparison, Claude 4 Opus and Mistral Medium 3 were the lowest at 1.6 and 1.5 grams, respectively. There is a wide spread even among models released in the same year, showing not only that inference efficiency varies but that higher capability is not necessarily proportional to the environmental cost.

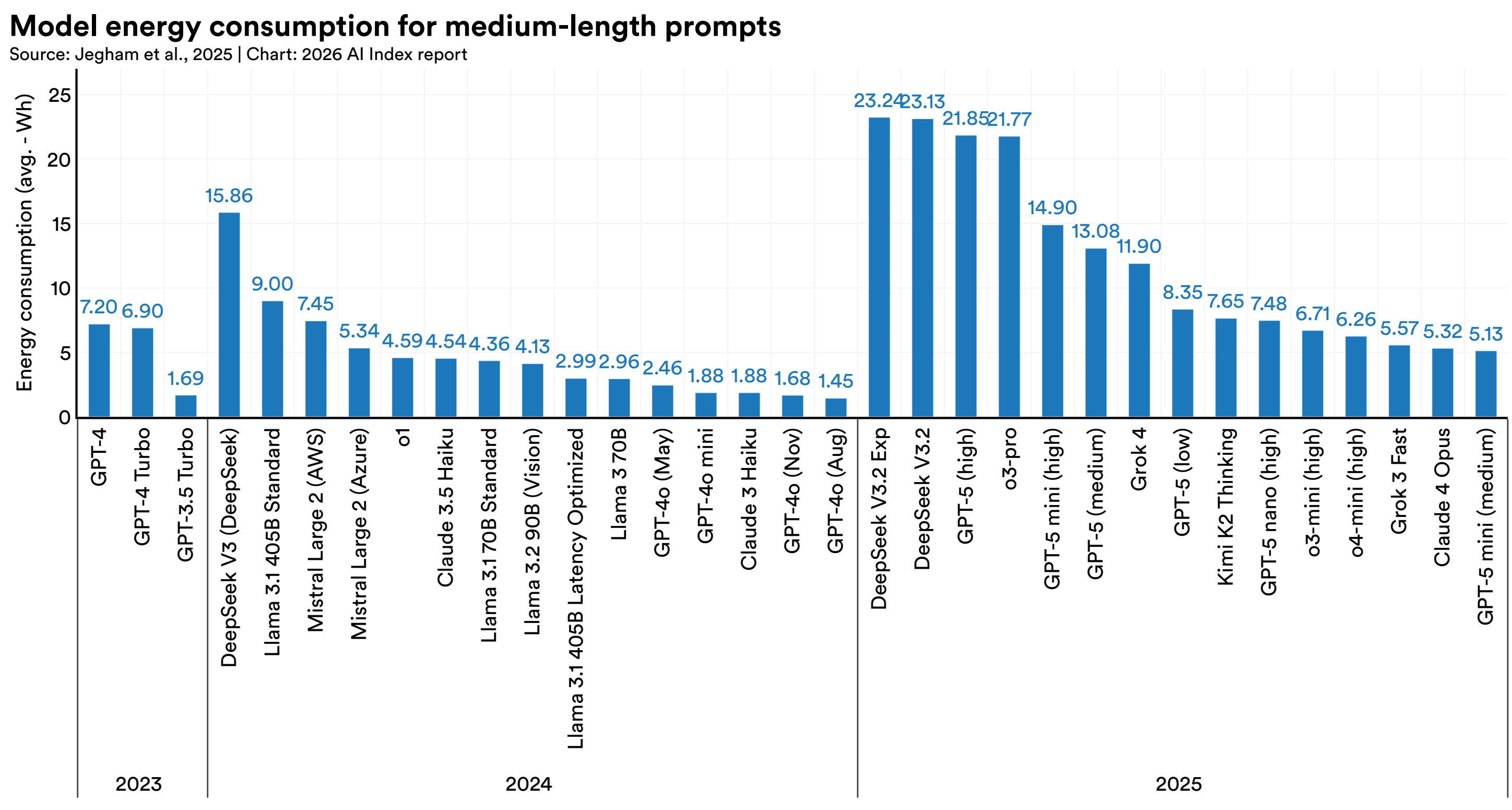

Figure 1.4.5[13]

13 This figure shows the top 15 models by energy consumption for 2024 and 2025. The full set of models is available through the source dashboard.

**Model carbon emissions for medium-length prompts**
Source: Jegham et al., 2025 | Chart: 2026 AI Index report

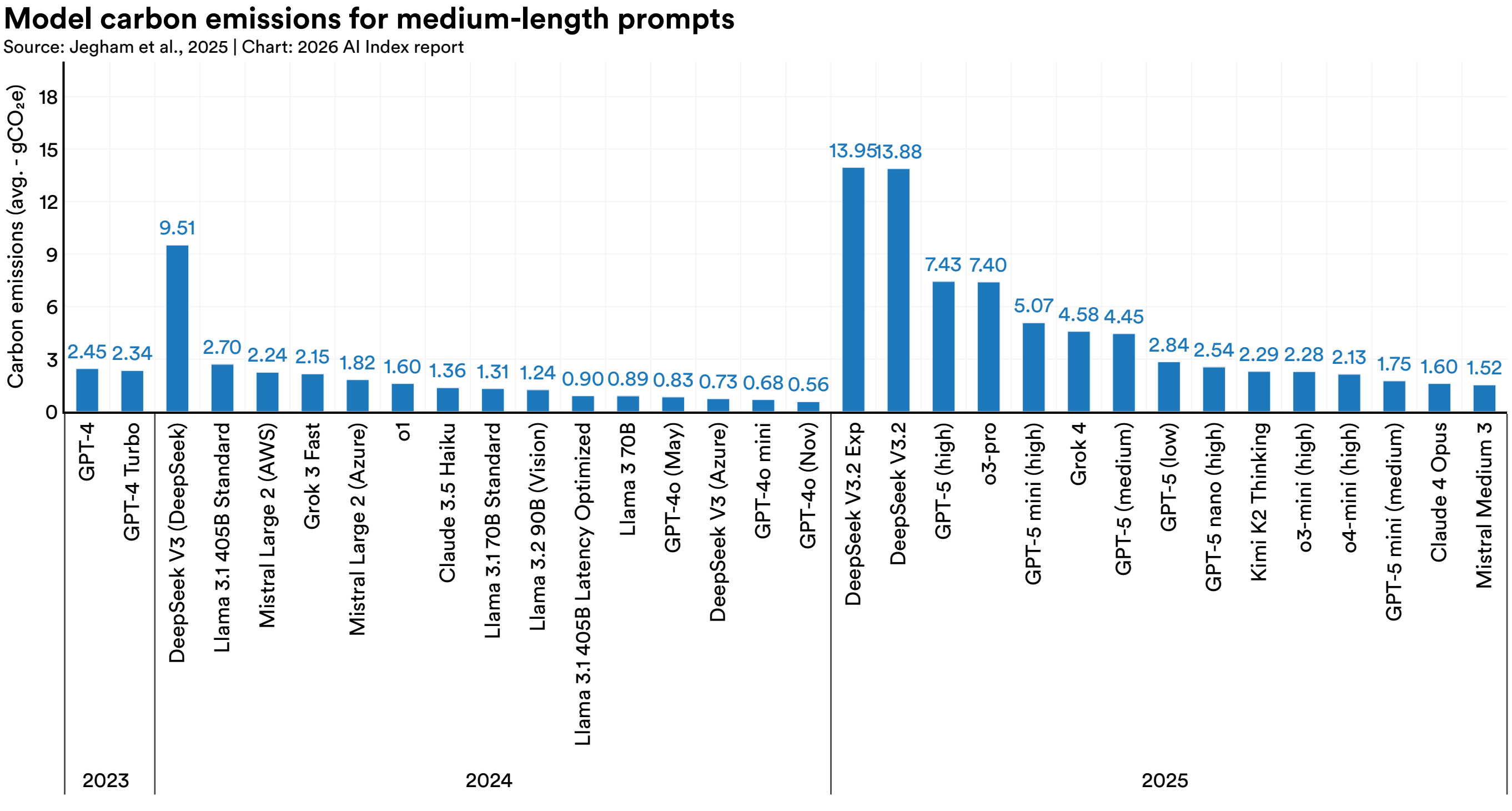

Figure 1.4.6[14]

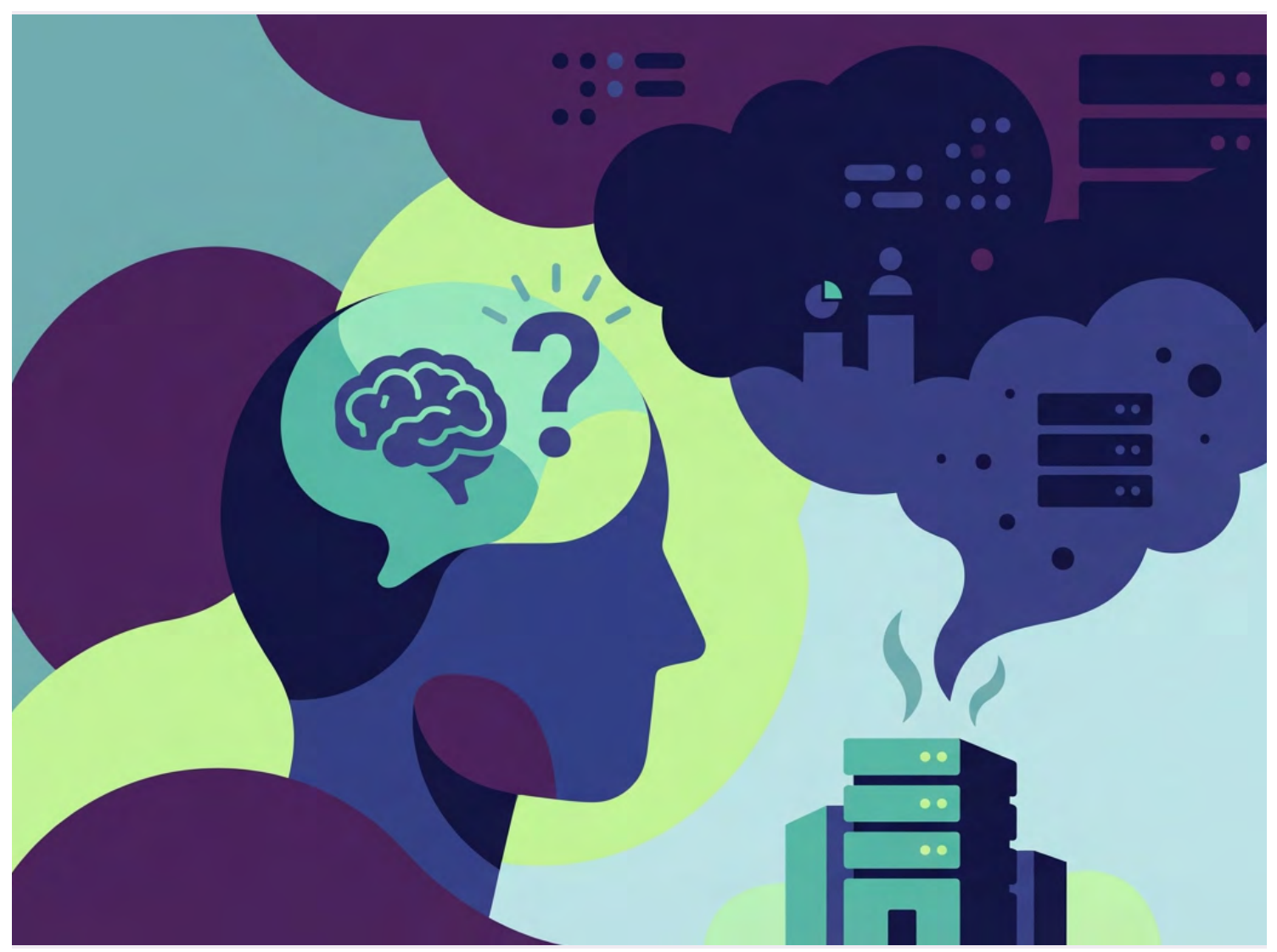

At the level of a single query, the numbers seem more modest. A short GPT-4o query consumes approximately 0.42 Wh, which is 40% more than a Google search at 0.3 Wh (Figure 1.4.7). A daily session of eight medium-length queries uses the energy comparable to charging two smartphones (9.7 Wh). But across hundreds of millions of daily queries, the consumption scales into something much larger.

The same scaling dynamic is true for water consumption (Figure 1.4.8). Annual estimates for GPT-4o inference range from about 1.3 to 1.6 million kiloliters, which, at the high end, exceeds the annual drinking water needs of 1.2 million people.

14 This figure shows the top 15 models by energy consumption for 2024 and 2025. The full set of models is available through the source dashboard.

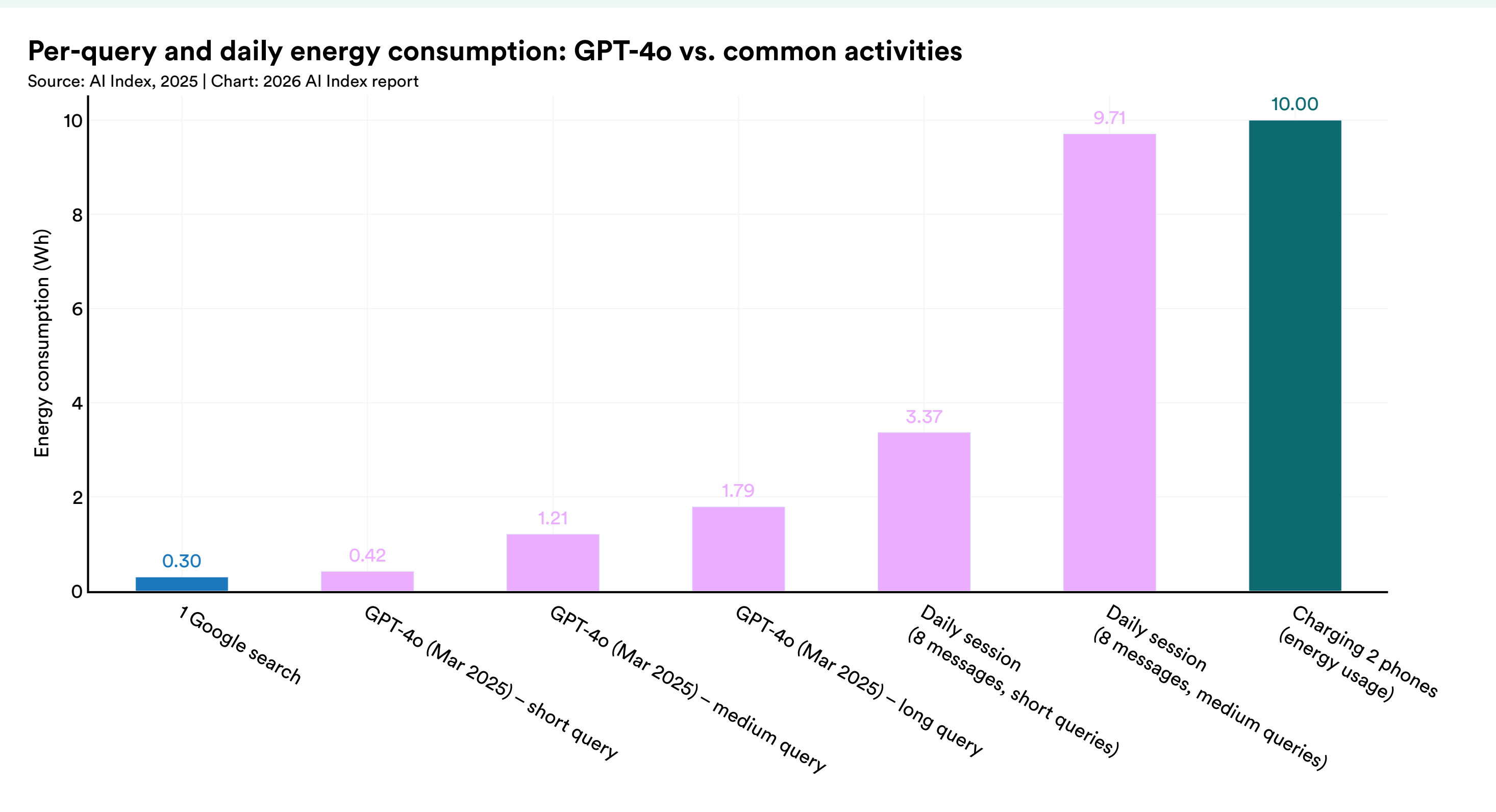

Figure 1.4.7

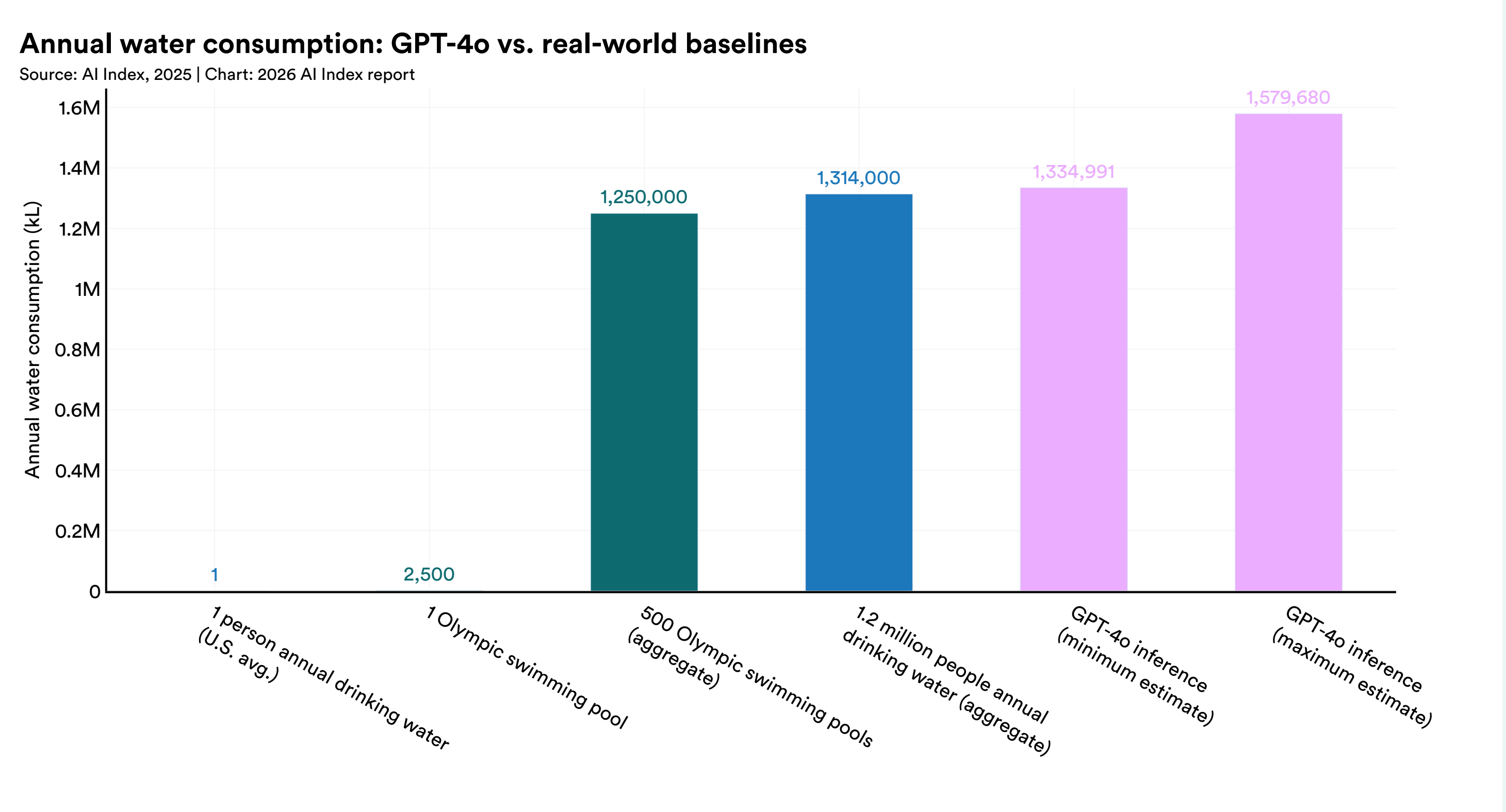

Figure 1.4.8

## Data Center Usage

The power demands of models and queries add up to a much larger infrastructure footprint. The estimated power demand from AI accelerator modules reached approximately 5,200 MW cumulatively through 2024 (Figure 1.4.9). Nvidia accounted for the largest share, which is consistent with the company's leading position in global AI chip capacity (as discussed in Section 1.2). When including the full systems supporting those accelerators (servers, cooling, networking), estimated demand reached approximately 9,400 MW (Figure 1.4.10). However, these figures from de Vries and Gao (2025) carry uncertainty from variation in utilization rates and facility-level efficiency, as reflected in the error bars on the charts.

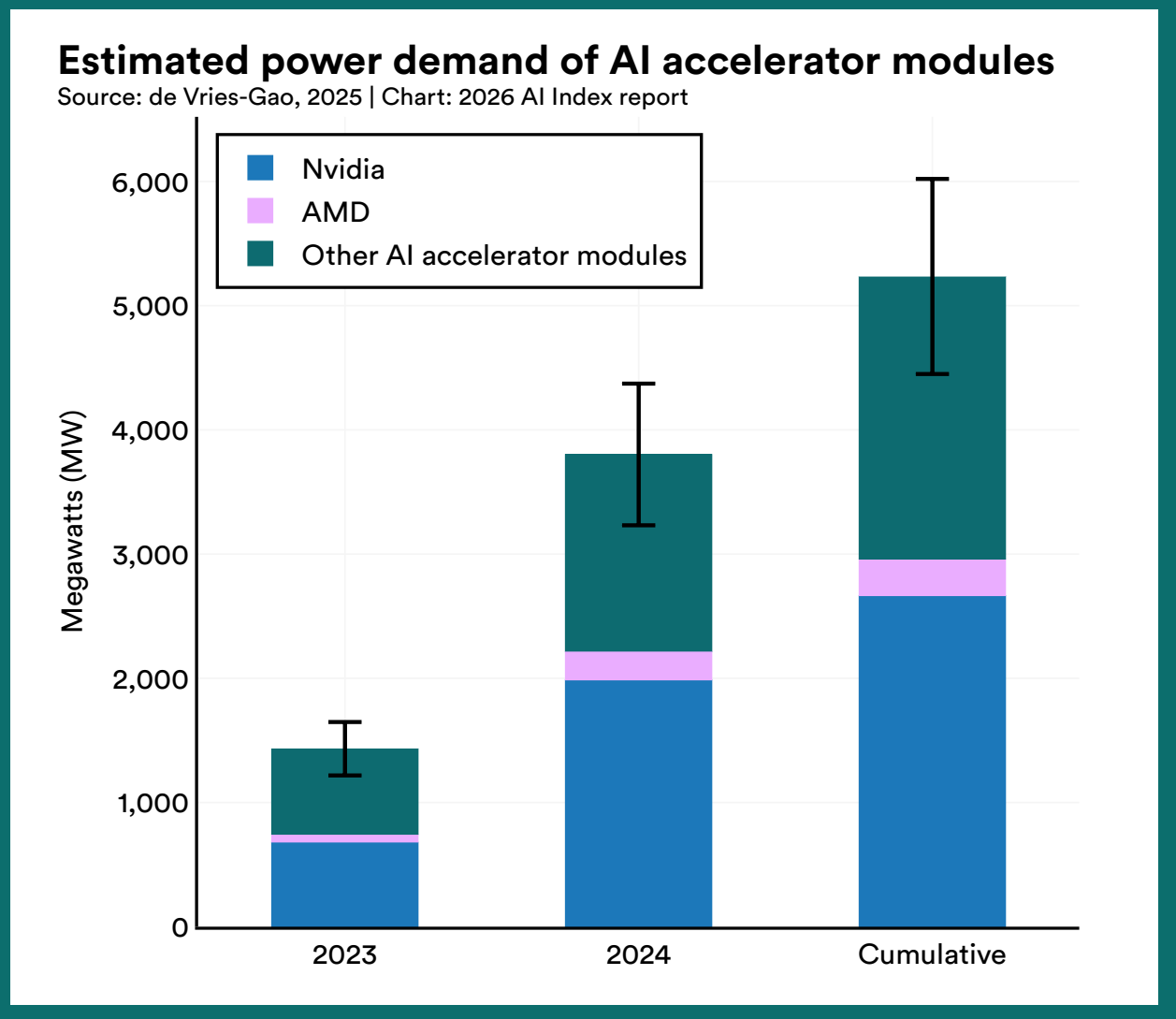

Figure 1.4.9

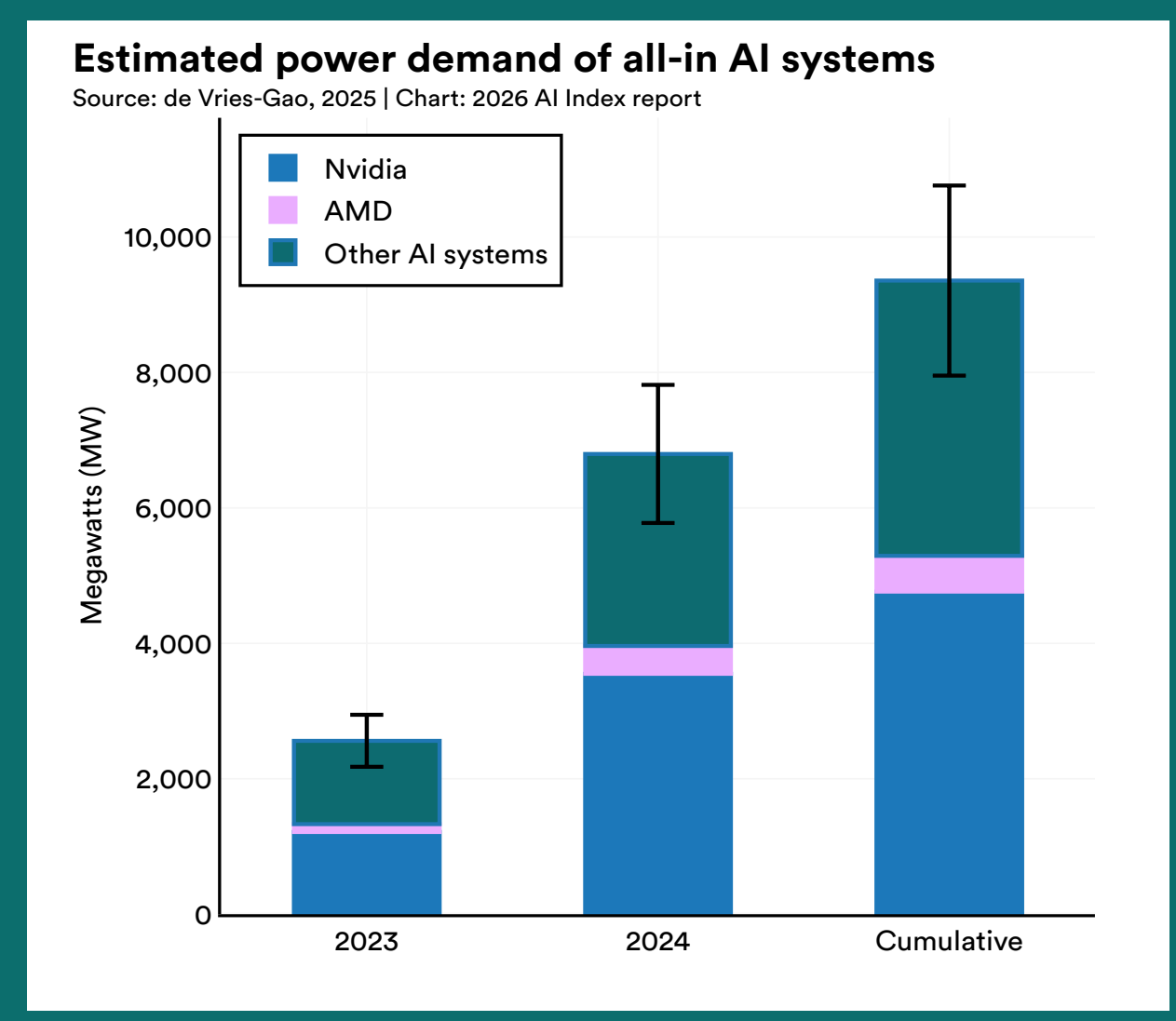

Figure 1.4.10

To put that scale in perspective, the cumulative power demand of all-in AI systems is comparable to the national electricity consumption of Switzerland or Austria, and roughly half that of Bitcoin mining (Figure 1.4.11). Excluding crypto, global data centers accounted for the highest estimated power demand at around 47,000 MW, with AI hardware making up a growing share of that total.

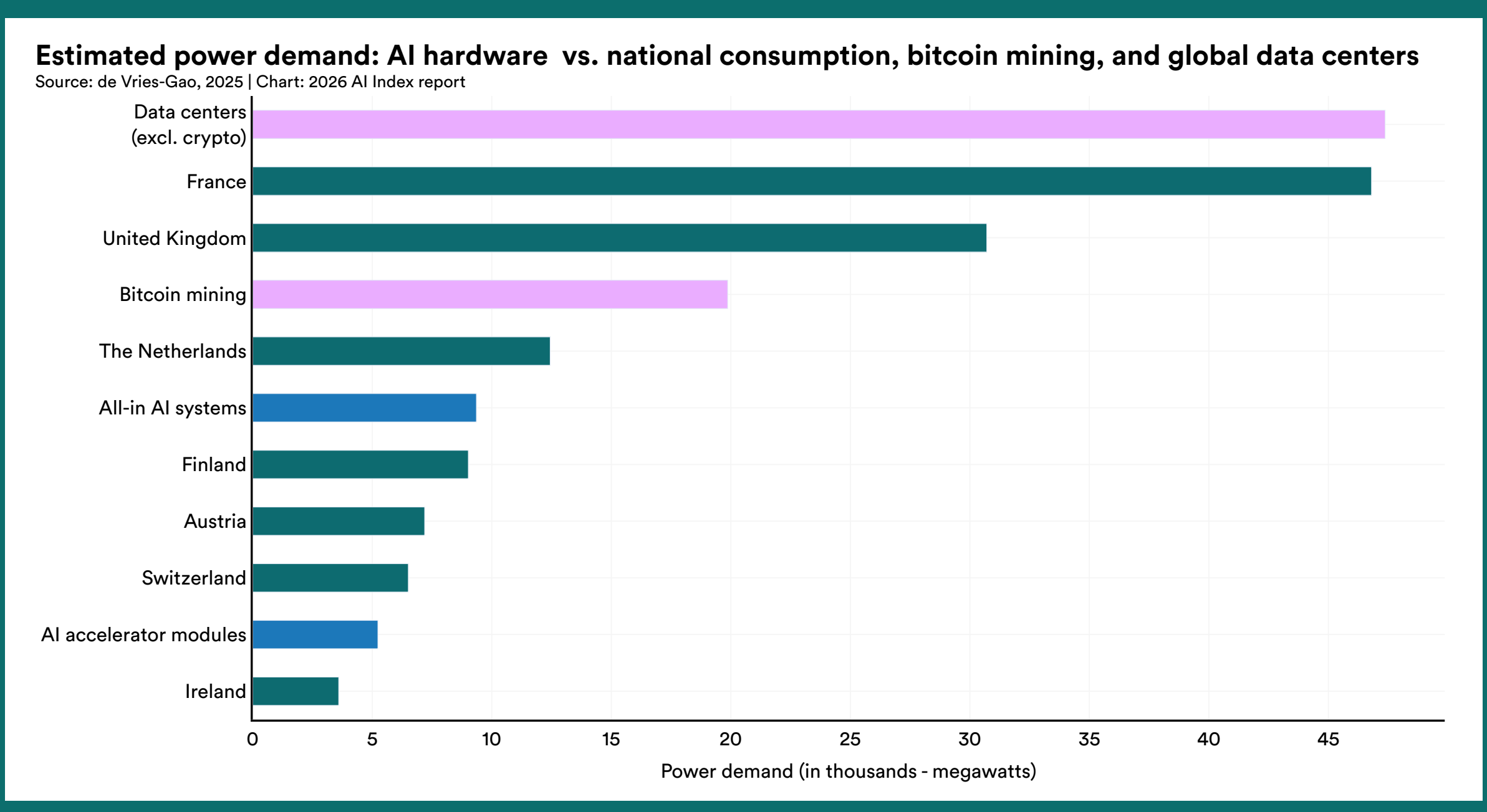

Figure 1.4.11

Cost, however, has been moving in the opposite direction. Since 2006, the cost of GPU computation has fallen by more than 99% (Figure 1.4.12). This decline has been key to enabling the scaling trends described throughout this chapter, making it economically feasible to train and deploy models at levels that would have been cost prohibitive even a decade ago. At the regional level, data center electricity consumption has increased across all major regions, and it is projected to continue to rise through 2030 (Figure 1.4.13). The United States accounts for the largest share, followed by China, Europe, and the rest of Asia.

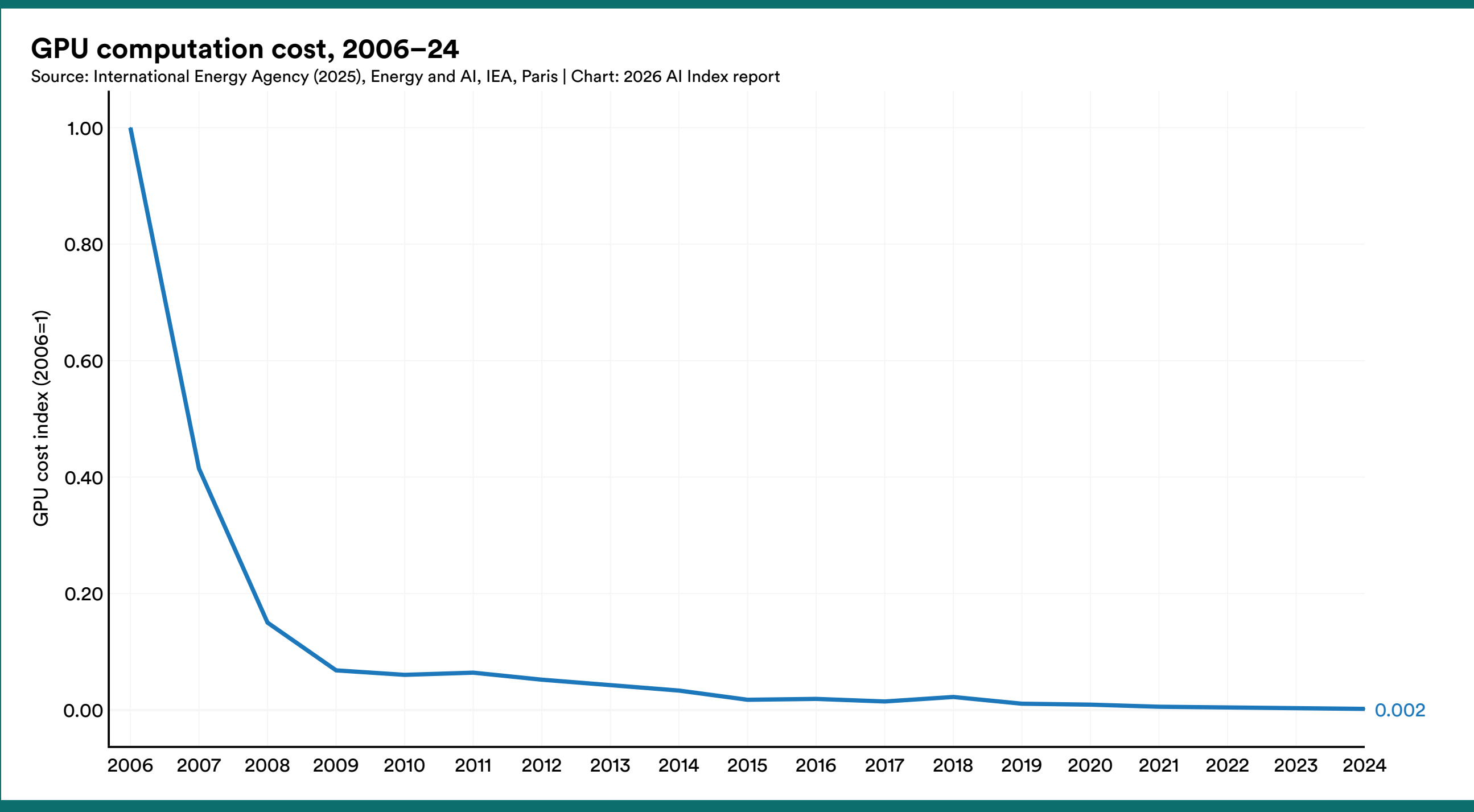

Figure 1.4.12

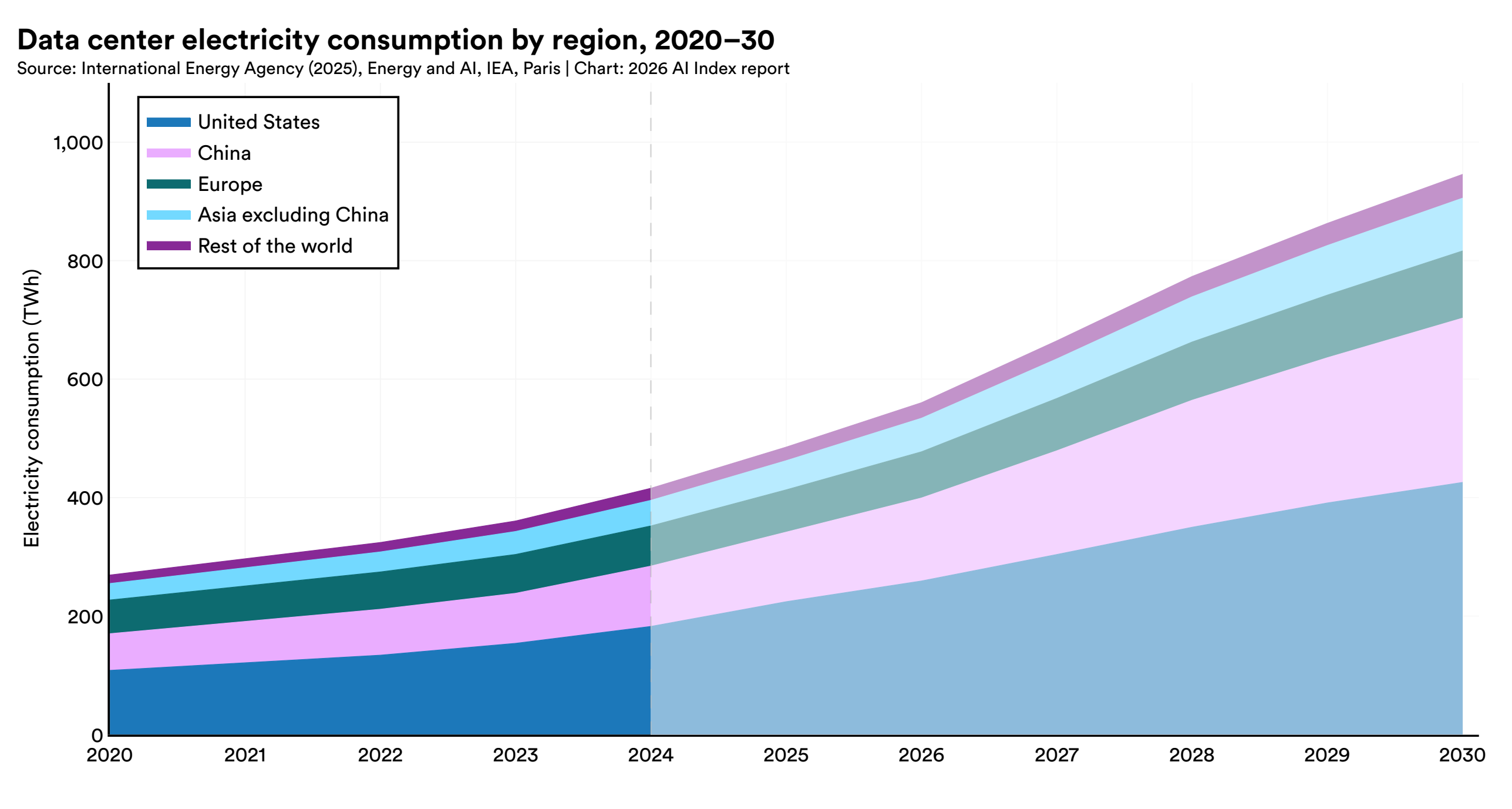

Figure 1.4.13[15]

15 Data in this chart reflects IEA projections rather than observed consumption.

# 1.5 Open-Source AI Software

## AI Development Activity Overview

The preceding sections have focused on notable frontier models and the infrastructure required to build and maintain them. Open-source platforms like GitHub and Hugging Face offer a different view that captures the developer ecosystem experimenting with and building on AI models. Much of this activity is not reflected in academic publications or frontier model releases. The AI Index analyzes data from both platforms[16] to better understand how open-source AI development is evolving over time.

### Projects

The scale of open-source development has grown steadily. The number of AI-related GitHub projects increased from 1,549 in 2011 to approximately 5.6 million in 2025, with year-over-year growth accelerating 23.7% from 2024 (Figure 1.5.1). However, most repositories often consist of personal or experimental work and receive minimal attention. When filtering for projects with at least 10 stars, a rough proxy for community engagement, the count drops to 206,880 in 2025 (Figure 1.5.2). The growth trajectory is similar for both measures.

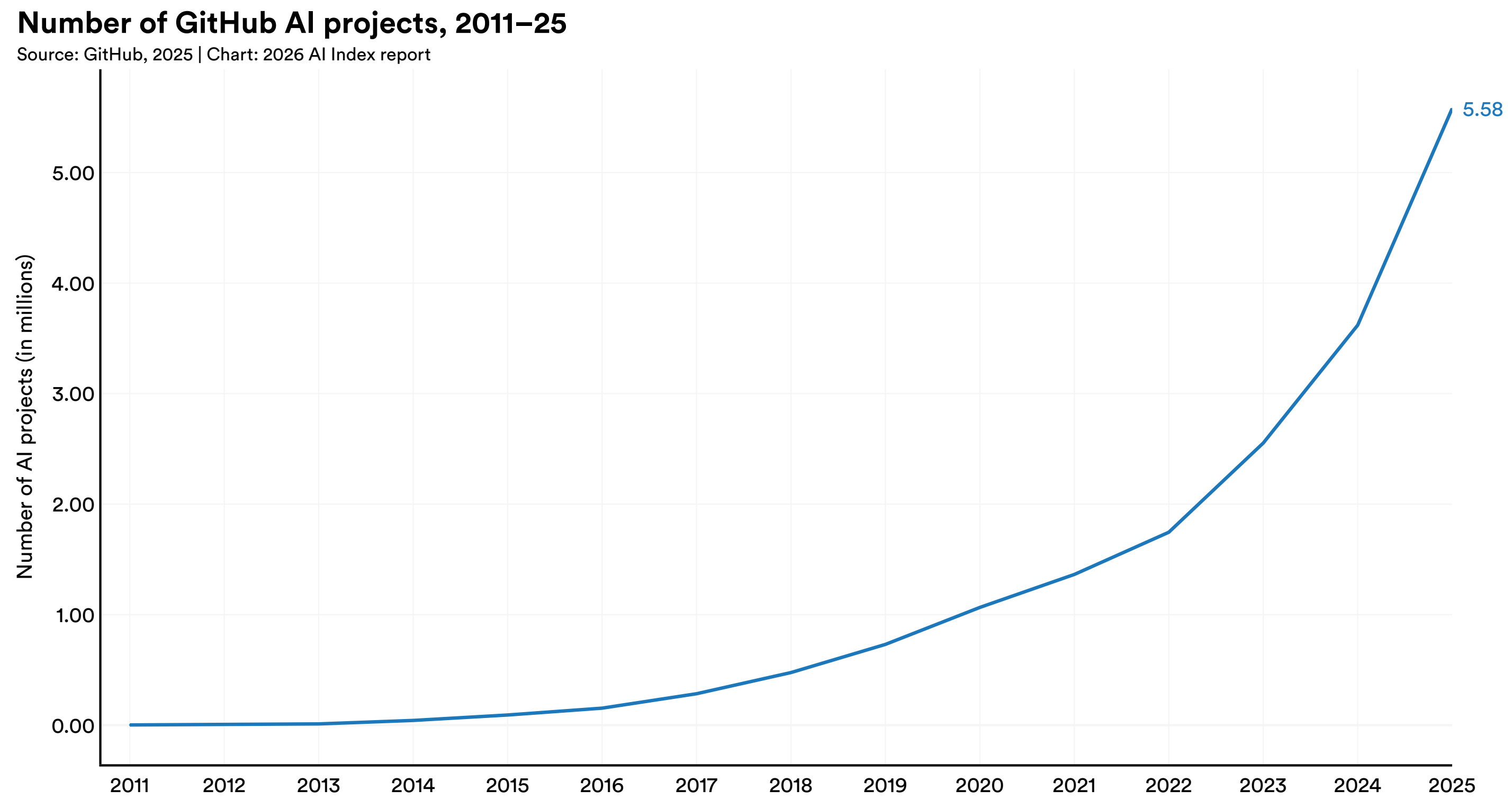

Figure 1.5.1

16 Chinese researchers often use alternatives to GitHub, such as Gitee and GitCode, for code sharing, but the data from those sites is not included in this report. A full methodological description is available in the Appendix.

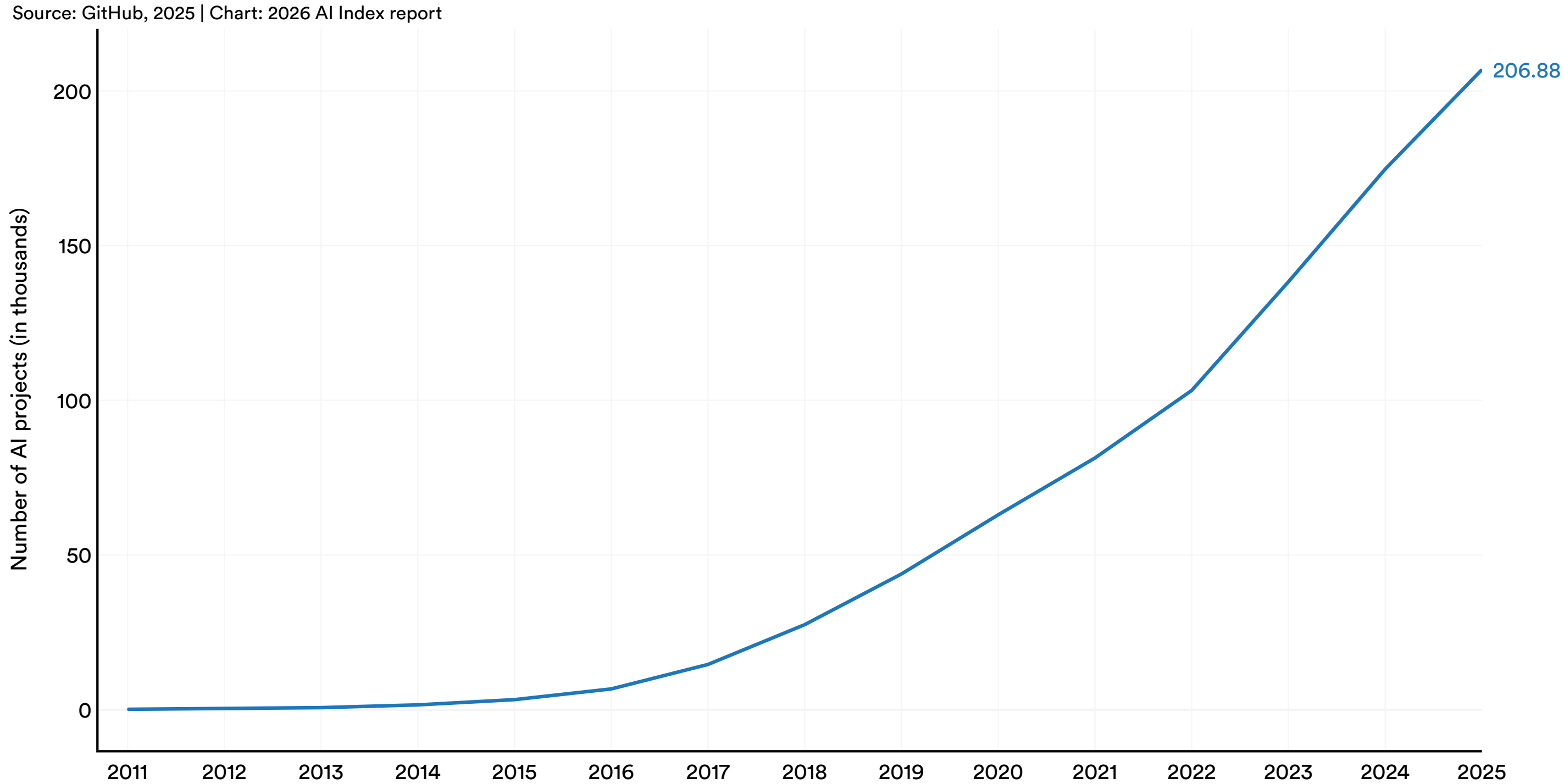

Figure 1.5.2

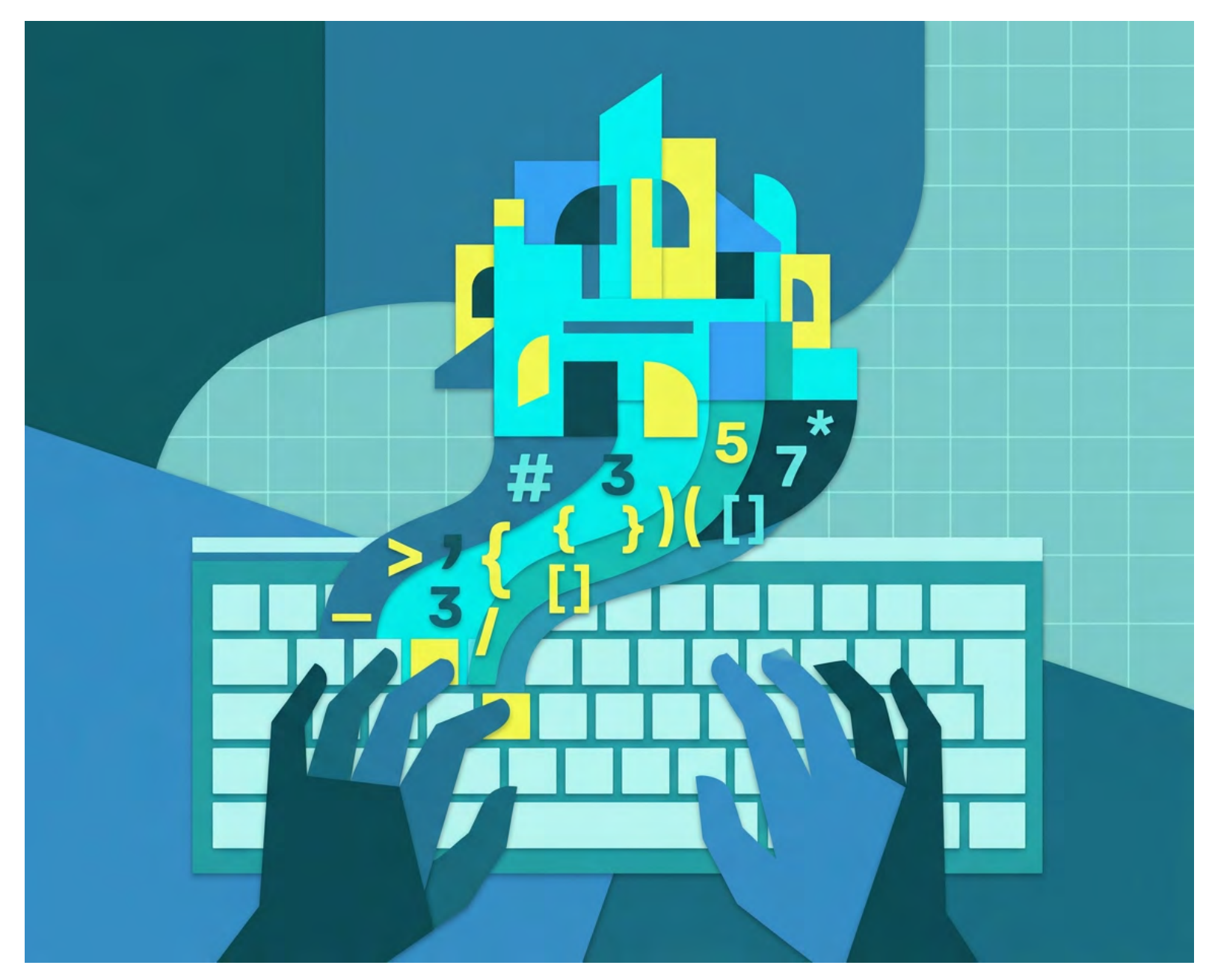

The geographic distribution of more visible open-source AI projects has shifted over time (Figure 1.5.3). Among projects with at least 10 stars, the United States accounted for the largest share in 2025 (31.7%), though that has declined steadily from nearly 80% in 2011 as developers in other regions have increased their presence on the platform. Europe and the rest of the world have grown in number of projects, while China's share has leveled off since 2019. India remains a growing contributor, representing 5.2% of projects with at least 10 stars. Because GitHub data does not capture Chinese developers who use domestic platforms such as Gitee or GitCode, China's share of global open-source AI activity is likely understated. The existing geographic attribution for China uses self-reported location rather than IP-based geolocation.

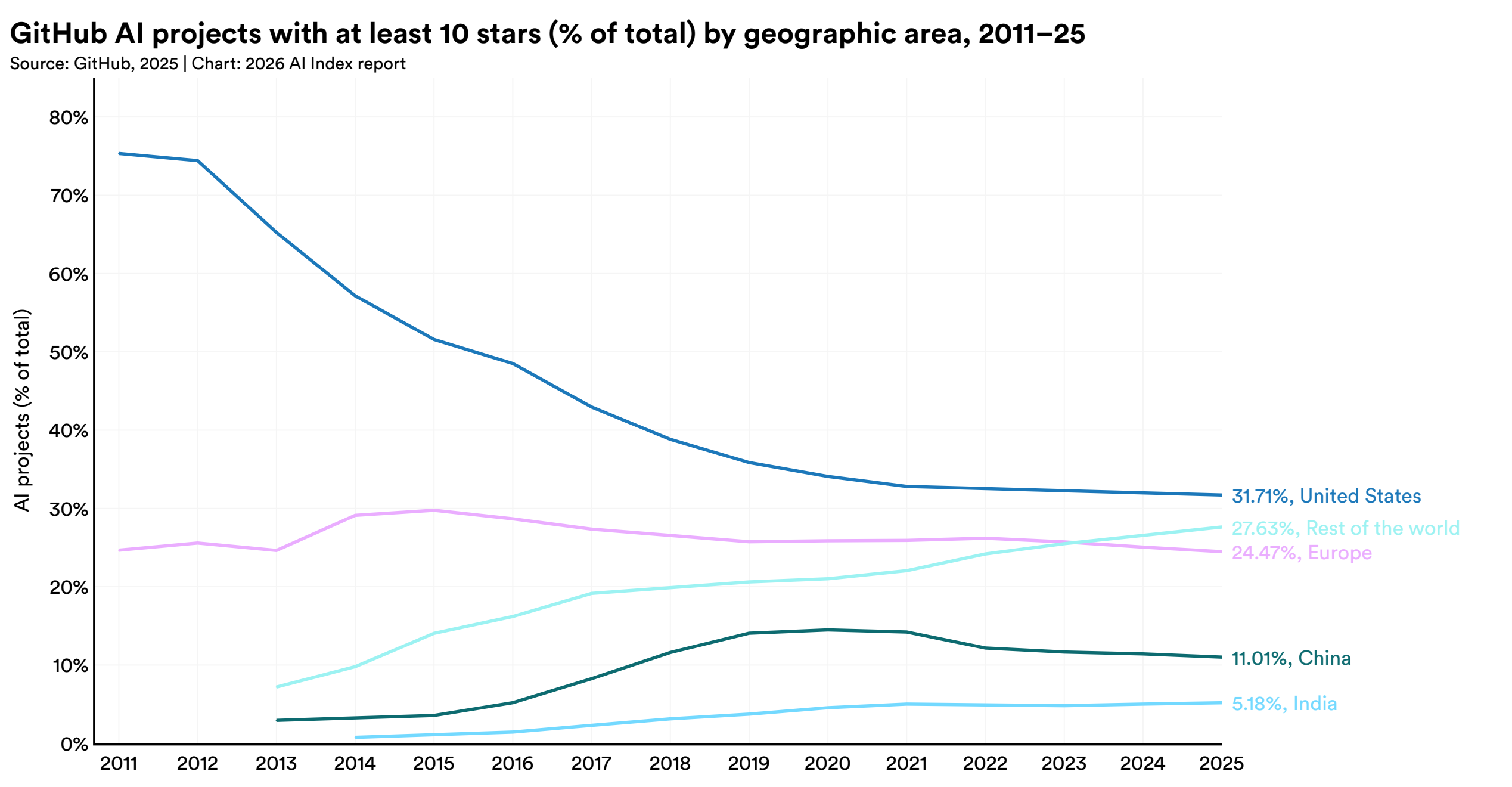

Figure 1.5.3[17]

## Stars

Beyond project counts, GitHub stars provide another signal of developer interest and engagement in open-source communities (Figure 1.5.4). The total number of stars for AI projects increased from 14 million in 2023 to 18.2 million in 2025.[18] All major geographic regions saw year-over-year increases. However, the geographic pattern for stars differs from the project share data above. Despite its declining share of projects, the United States accumulated the highest number of stars at 30 million cumulatively (Figure 1.5.5). So while open-source activity becomes more geographically distributed, the projects with the most engagement remain disproportionately U.S.-based.

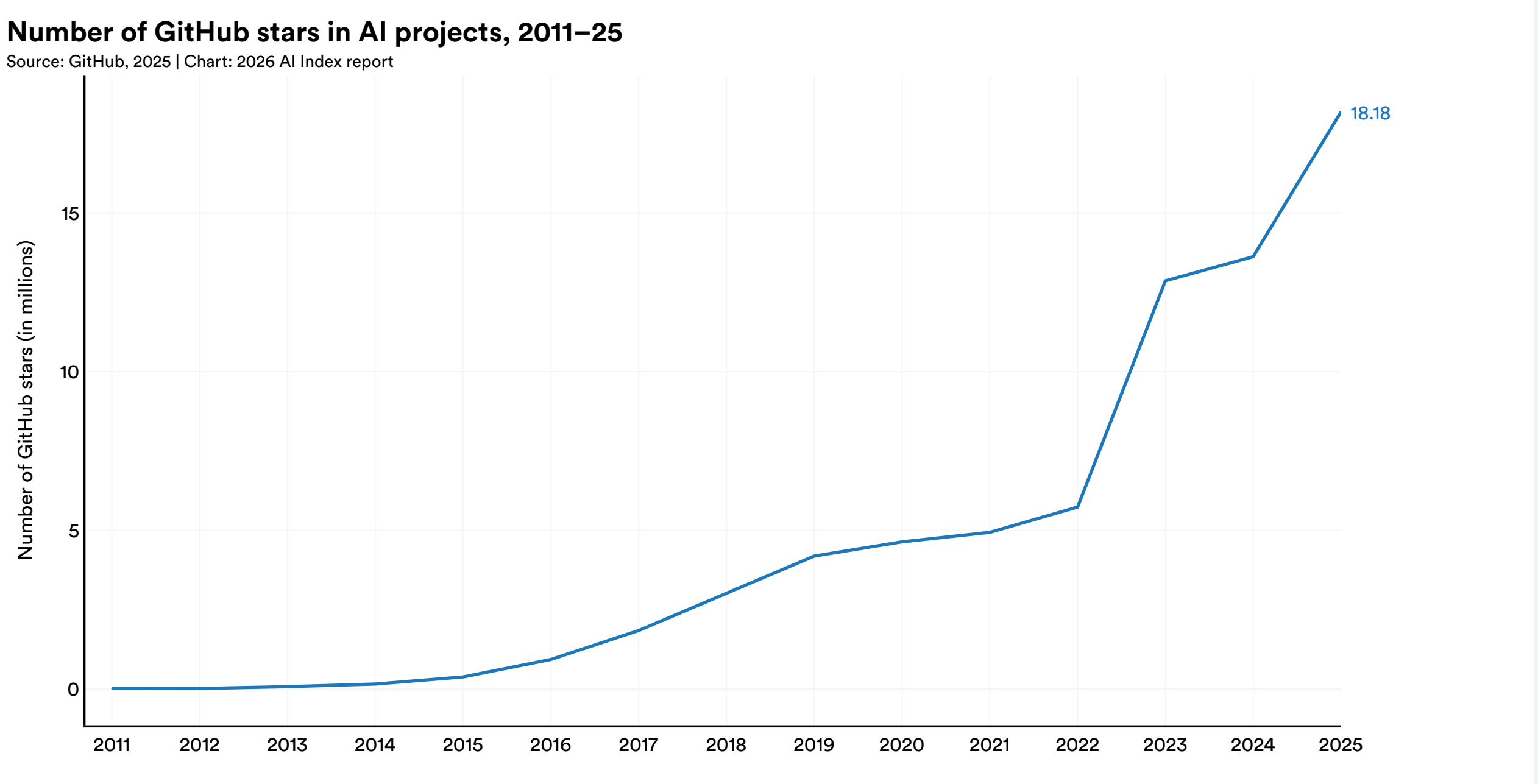

Figure 1.5.4

17 In previous AI Index reports, project locations for China and Hong Kong were determined using IP-based geolocation. Due to frequent VPN usage in these regions that resulted in systematic misclassification, self-reported profile locations are now used for China and Hong Kong, while IP-based geolocation continues to be applied for all other countries.

18 Figure 1.5.4 shows new stars given to GitHub projects within a year, not the total accumulated over time.

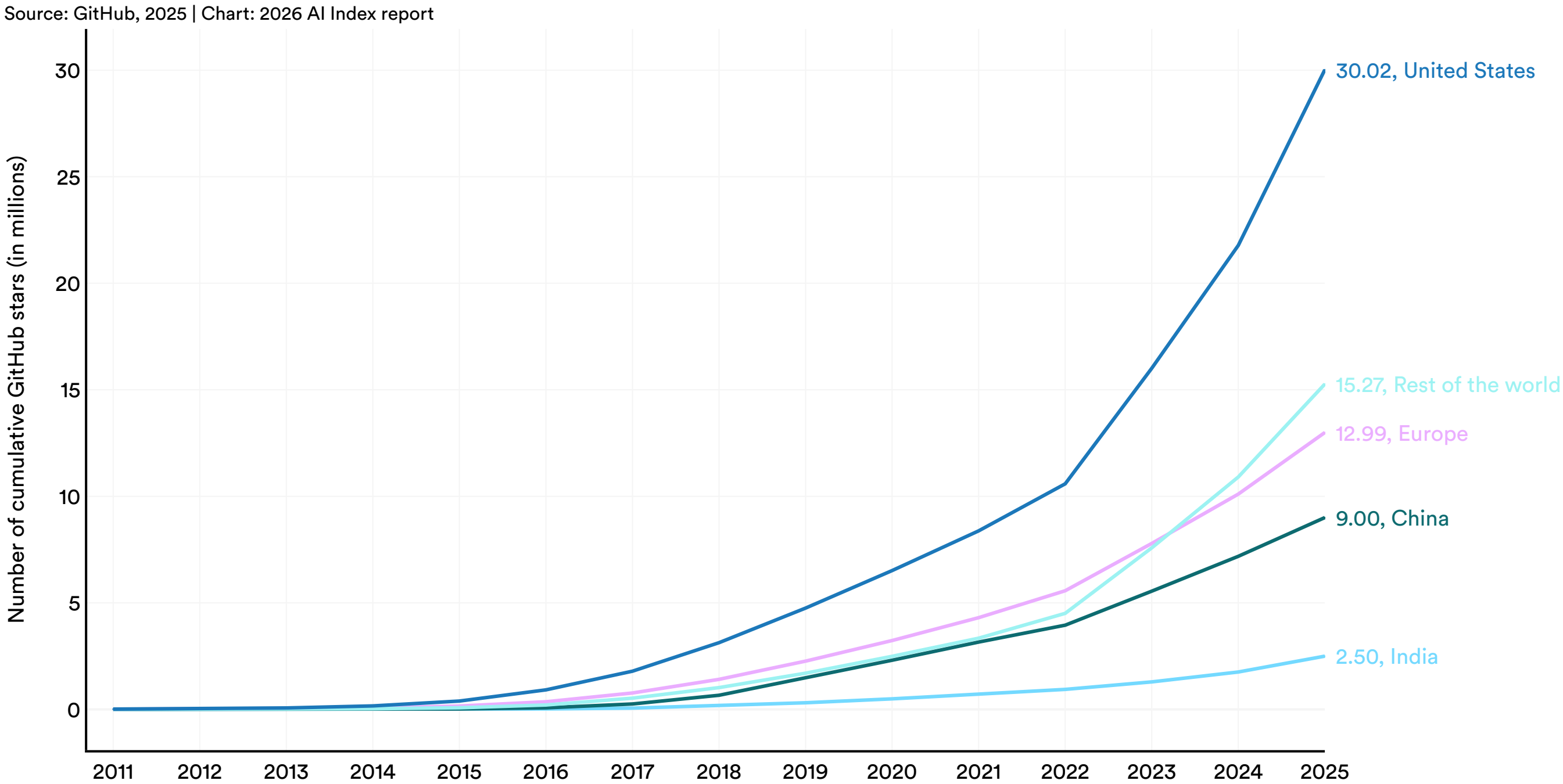
Number of GitHub stars by geographic area, 2011–25
Source: GitHub, 2025 | Chart: 2026 AI Index report

Figure 1.5.5

## Model and Dataset Ecosystem

To complement the GitHub view, this section uses metadata from Hugging Face, a widely used community platform and open repository for AI models and datasets. The analysis focuses on assets created or uploaded between 2022 and 2025 to understand recent activity and adoption trends (Figures 1.5.6 and 1.5.7). Upload activity has continued to rise over the last few years, with a marked increase after the second quarter of 2024. From 2023 to 2025, model uploads more than tripled, while dataset uploads grew fourfold. Download distribution also shifted after 2023. Geographically[19] U.S.-developed models lost share to unaffiliated users. On the developer side, major private actors such as Google and Meta have shifted from being the principal authors to accounting for a relatively small share of downloads, while communities such as Sentence Transformers and the BERT community have grown (Figure 1.5.8). A large share of total model downloads fell into an "Others" category, reflecting the wider distribution of development activity even as the most downloaded models were tied to a small number of sources.

19 Data was obtained in collaboration with researchers from Longpre et al. (2025). Their dataset provides Hugging Face model download data that the authors describe as consistent and relatively complete. It was validated with the Hugging Face team, is reported to be less noisy than raw counts, and includes cleaned and imputed missing metadata. It is released as a weekly panel rather than an all-time-downloads cross-section. Coverage spans March 2020 to August 2025 and includes the top 200 most-downloaded Hugging Face models per week. These models account for 49.6% of total normalized, filtered downloads. This restriction focuses the analysis on models with higher observed download volume, reduces long-tail variation, and may support more stable estimates.

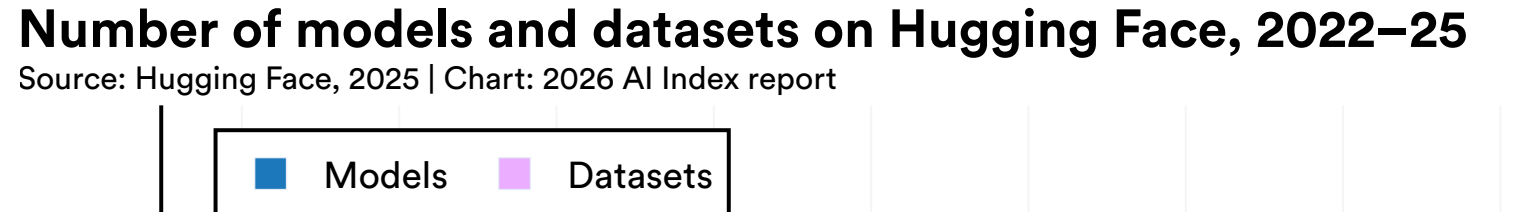
Number of models and datasets on Hugging Face, 2022–25
Source: Hugging Face, 2025 | Chart: 2026 AI Index report

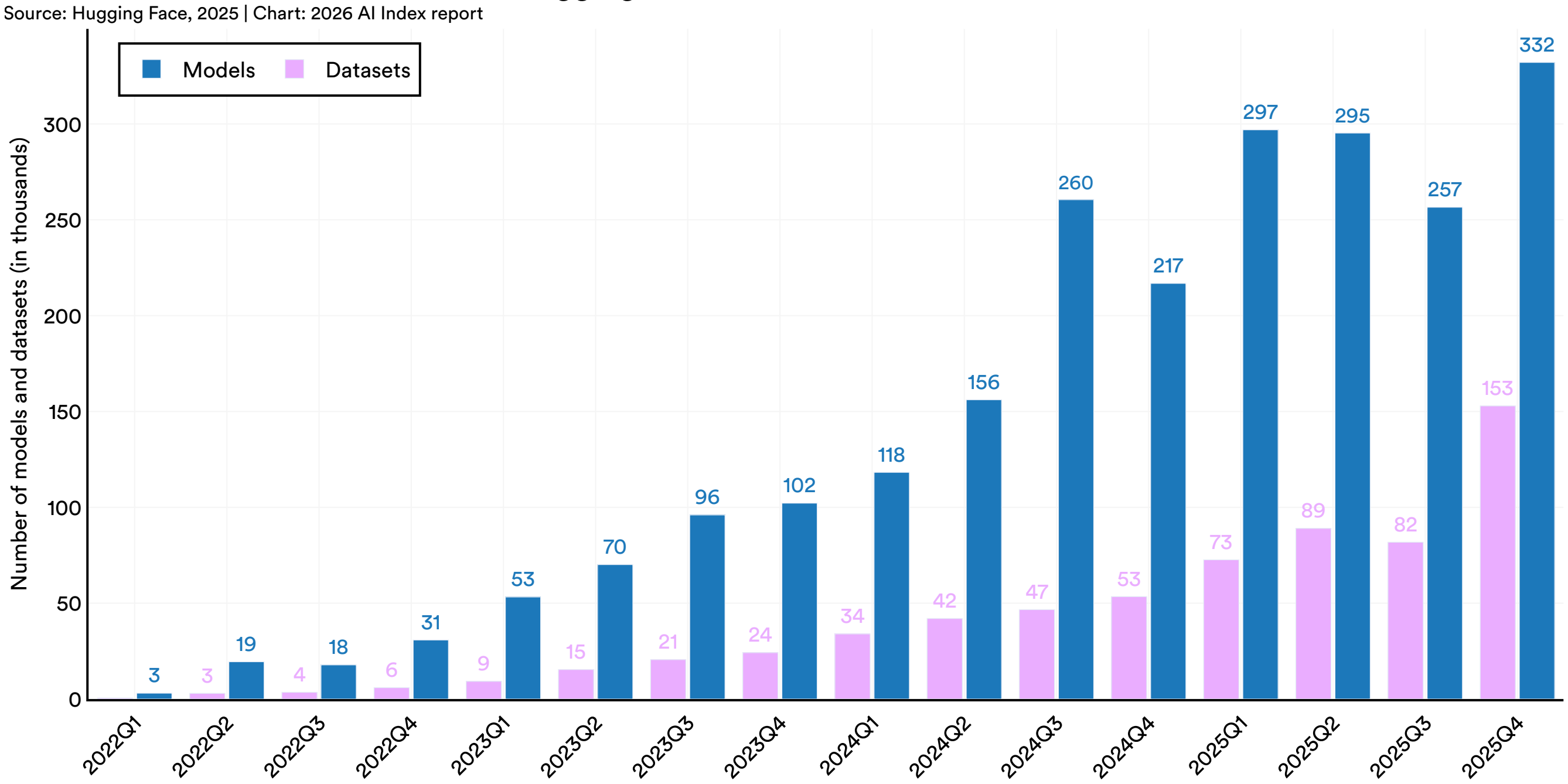

Figure 1.5.6[20]

## Global distribution of downloads among top Hugging Face models, Q2 2020–Q3 2025

Source: Longpre et al., 2025; Hugging Face, 2025 | Chart: 2026 AI Index report

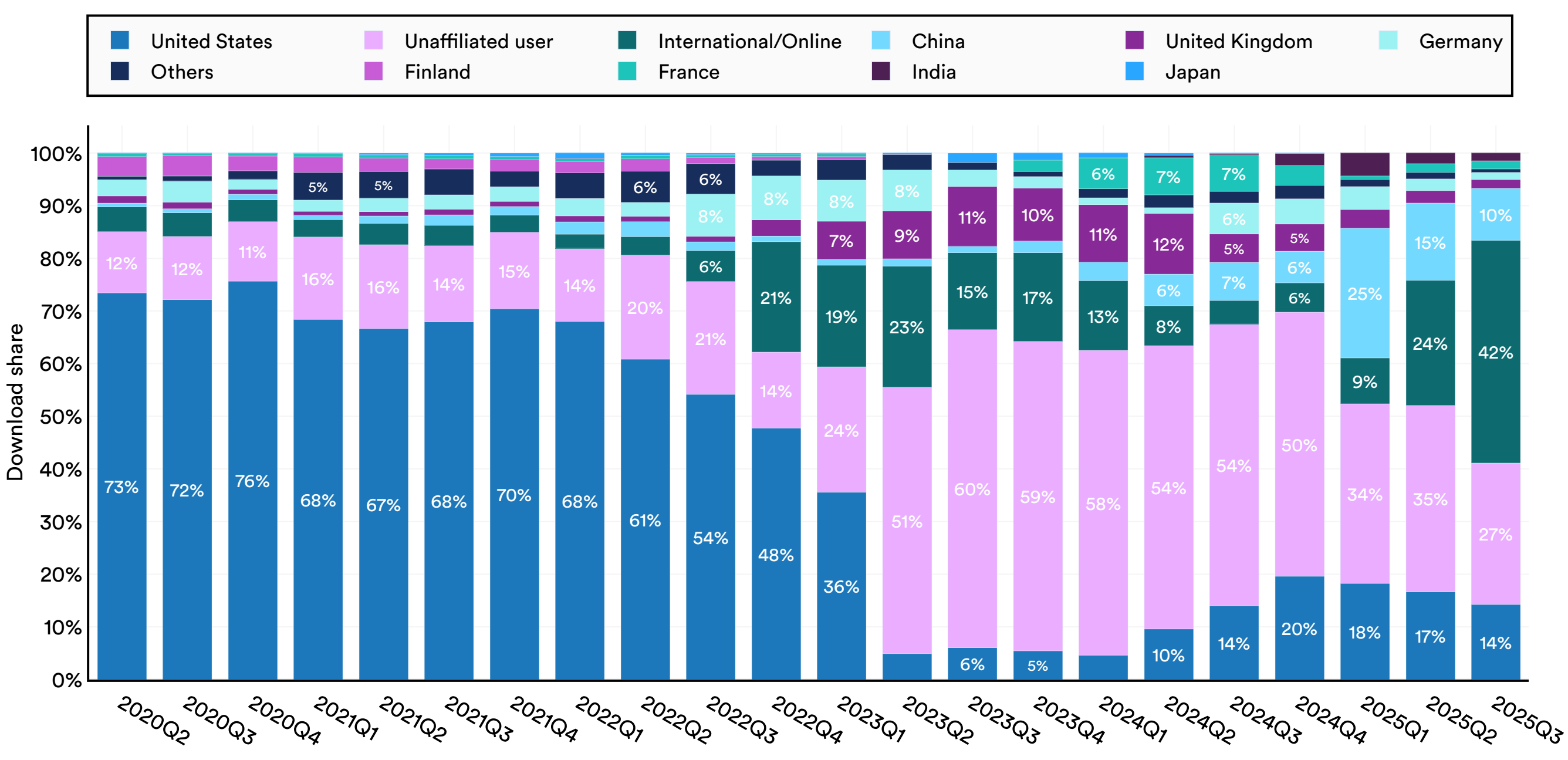

Figure 1.5.7[21]

20 The data shown in this chart comes from the publicly accessible Hugging Face repository. For more details, refer to the Appendix.

21 Data source: Longpre et al. (2025). For more details, refer to the Appendix.

**Download share by developer among top Hugging Face models, Q2 2020–Q3 2025**
Source: Longpre et al., 2025; Hugging Face, 2025 | Chart: 2026 AI Index report

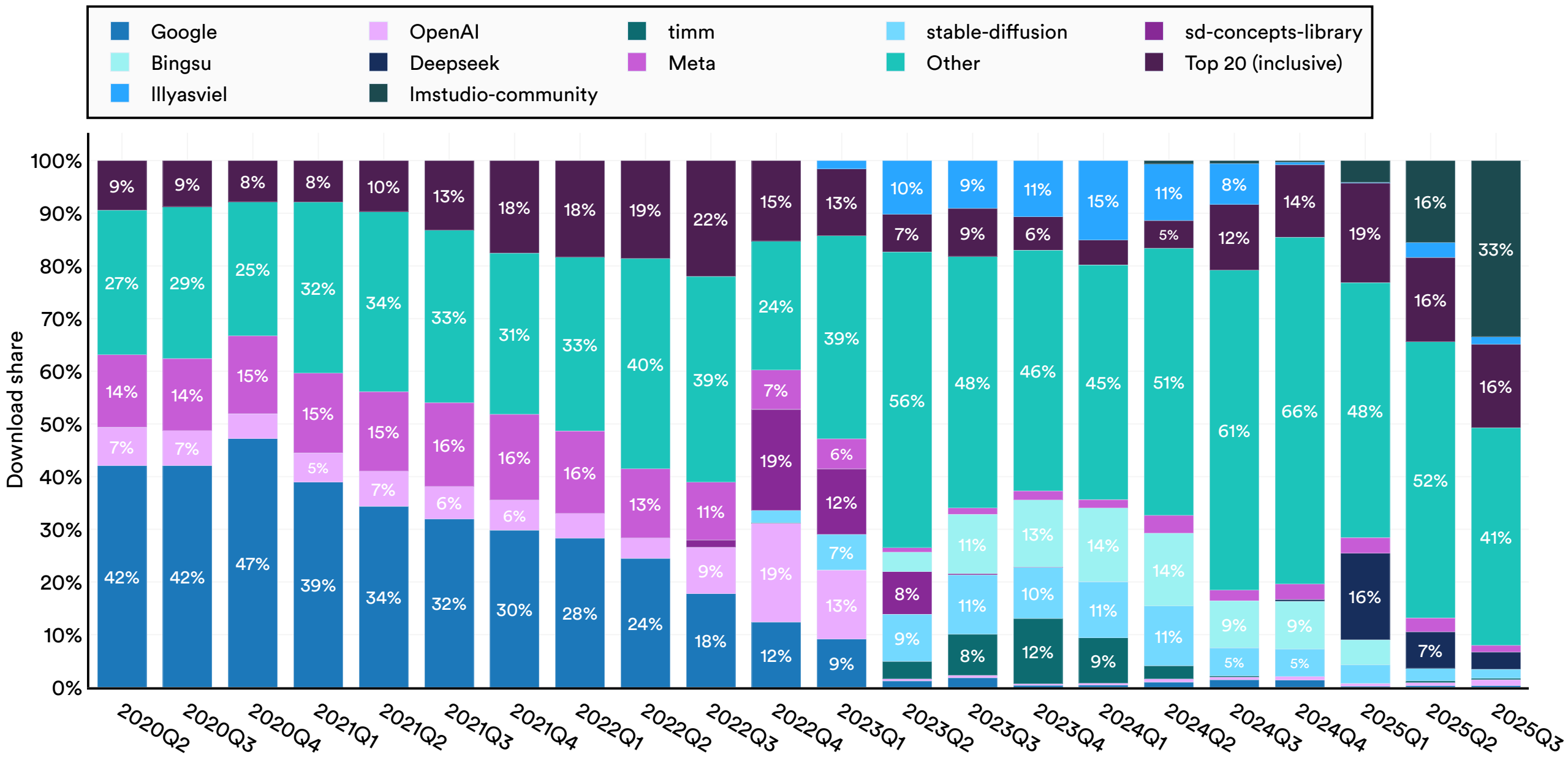

Figure 1.5.8[22]

The most popular model types have shifted over the last three years. Text embedders, classifiers, and audio models, which together accounted for nearly 70% of downloads in 2022, fell to less than 6% in 2025 (Figure 1.5.9). Text generation, multimodal, and video generation models have grown in their place. Text generation led in 2025, accounting for more than 42% of total downloads. Image generation models also increased steadily, remaining the second most downloaded category. Despite these shifts, downloads remain highly concentrated, with nearly 80% associated with the top three categories.

**Download share by modality among top Hugging Face models, Q3 2022–Q3 2025**
Source: Longpre et al., 2025; Hugging Face, 2025 | Chart: 2026 AI Index report

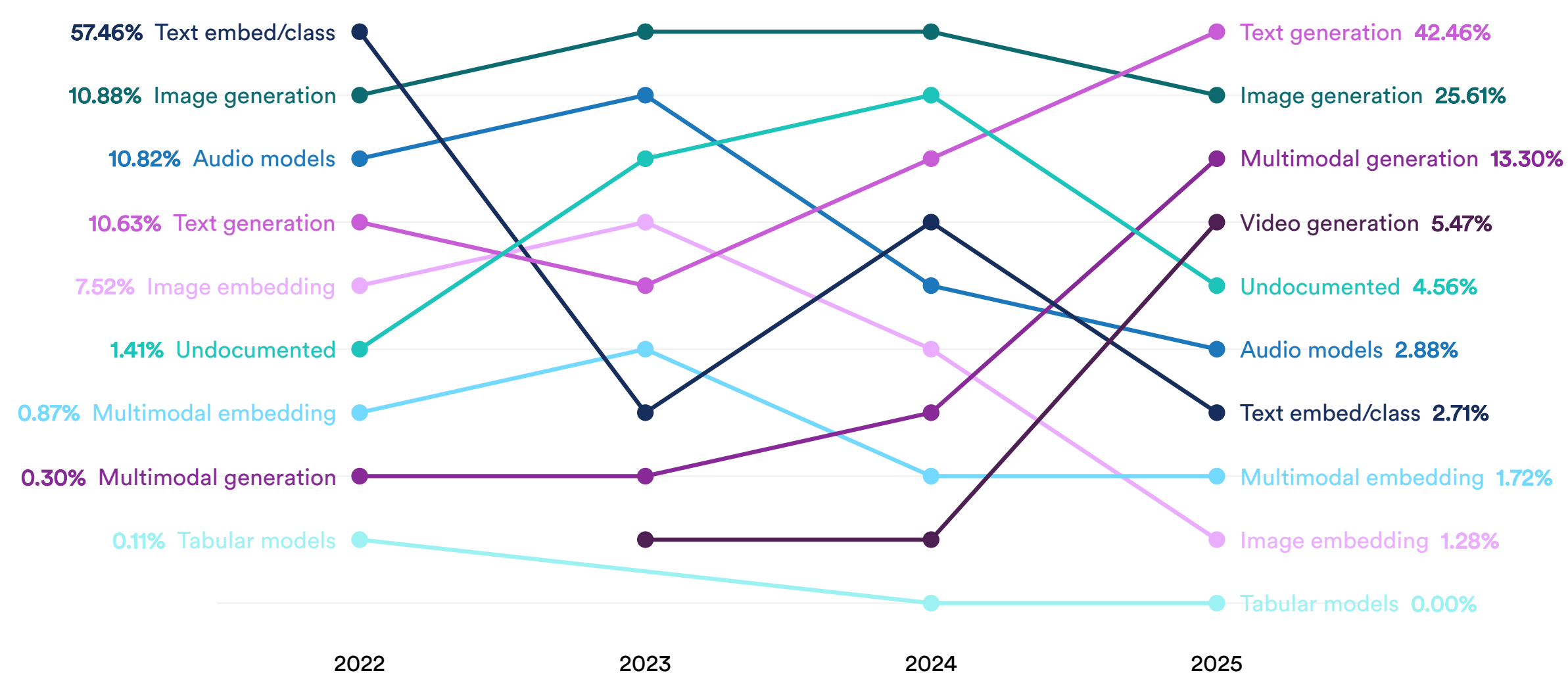

Figure 1.5.9[23]

22 Data source: Longpre et al. (2025). For more details, refer to the Appendix.

23 Data source: Longpre et al. (2025). For more details, refer to the Appendix.

# 1.6 Publications

The first half of this chapter tracked the models, infrastructure, and energy behind AI development. This section shifts to research output, specifically English-language AI publications and citations. Publications offer a longitudinal signal of AI research activity at scale, and the AI Index has tracked them consistently over time. While publication volume is not a measure of research quality, and not all research appears in indexed databases, this approach offers a consistent method for tracking the research frontier year over year. The analysis draws from OpenAlex, a bibliographic database[24] the AI Index has used since 2025, and considers both publication volume and downstream influence through citation patterns.

## Total Number of AI Publications

Total AI publication output continues to rise. AI publications more than doubled between 2013 and 2024, increasing from roughly 102,000 to about 258,000 (Figure 1.6.1). Growth continued in 2024, though at a slower rate, with publications increasing 6.3% from 2023. AI research now makes up a substantial portion of the broader computer science ecosystem, accounting for 40.9% of all computer science publications in OpenAlex.

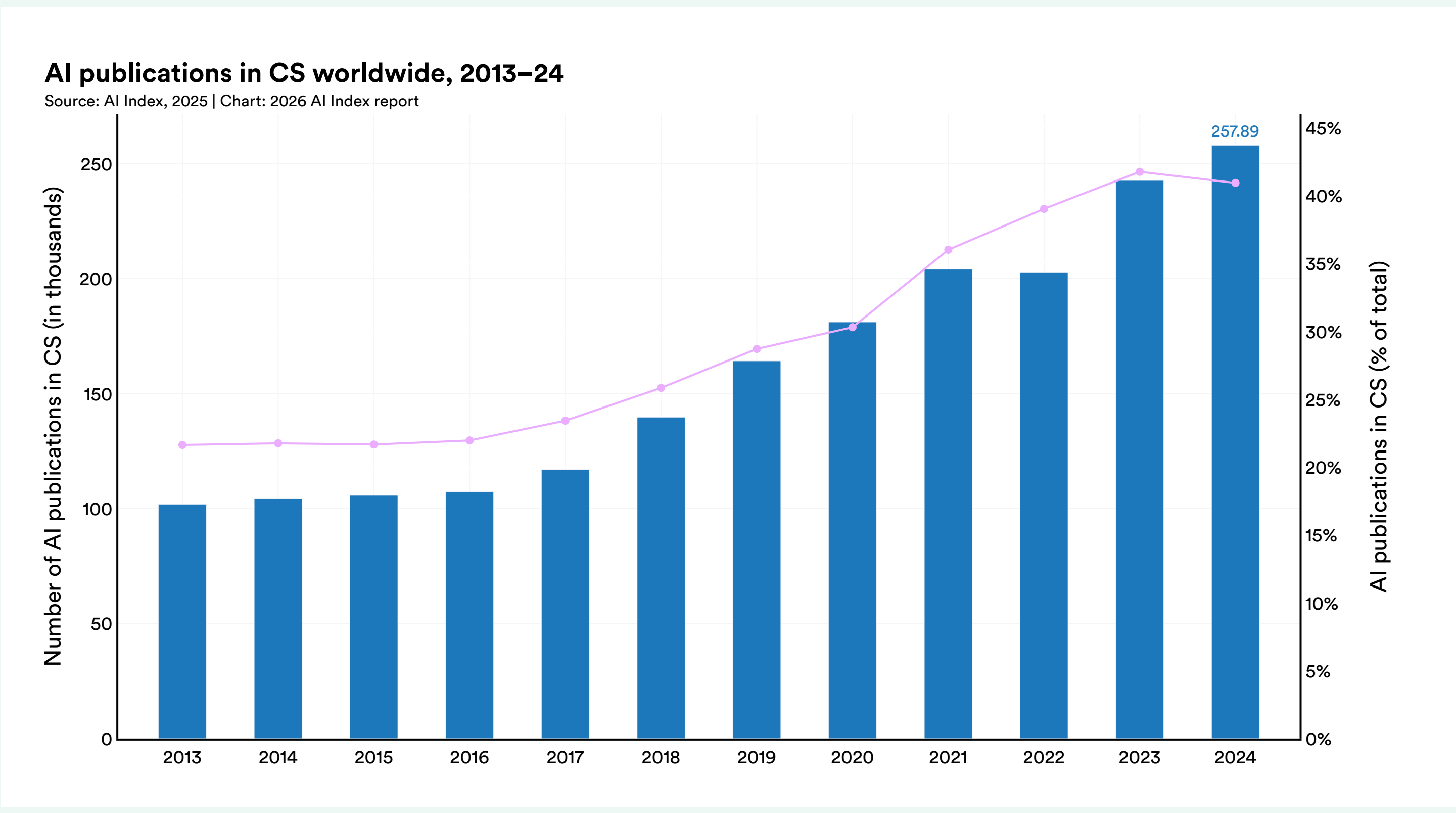

Figure 1.6.1

24 OpenAlex is a fully open catalog of scholarly metadata, including scientific papers, authors, institutions, and more. The AI Index used OpenAlex as a bibliographic database and automatically classified AI-related research using the latest version of the CSO Classifier. The CSO Classifier (v3.3) is an automated text classification system designed to categorize research papers in computer science using a comprehensive ontology of 15,000 topics and 166,000 relationships, including emerging fields like GenAI, large language models (LLMs), and prompt engineering. It processes metadata (such as title and abstract) through three modules: a syntactic module for exact topic matches, a semantic module leveraging word embeddings to infer related topics, and a post-processing module that refines results by filtering outliers and adding relevant higher-level areas.

## By Venue

In 2024, journals accounted for the largest share of AI publications (47%), followed by conferences (23.5%) (Figure 1.6.2). Since 2013, both journal and conference publications have increased in absolute numbers, though their relative shares have shifted. The proportion of AI publications appearing in conferences has steadily declined from 36.6% in 2013 to its current level. The most recent year's results, however, may also reflect a lag in venue assignment, as papers often appear first in repositories[25] like arXiv before being formally published in a journal or conference.

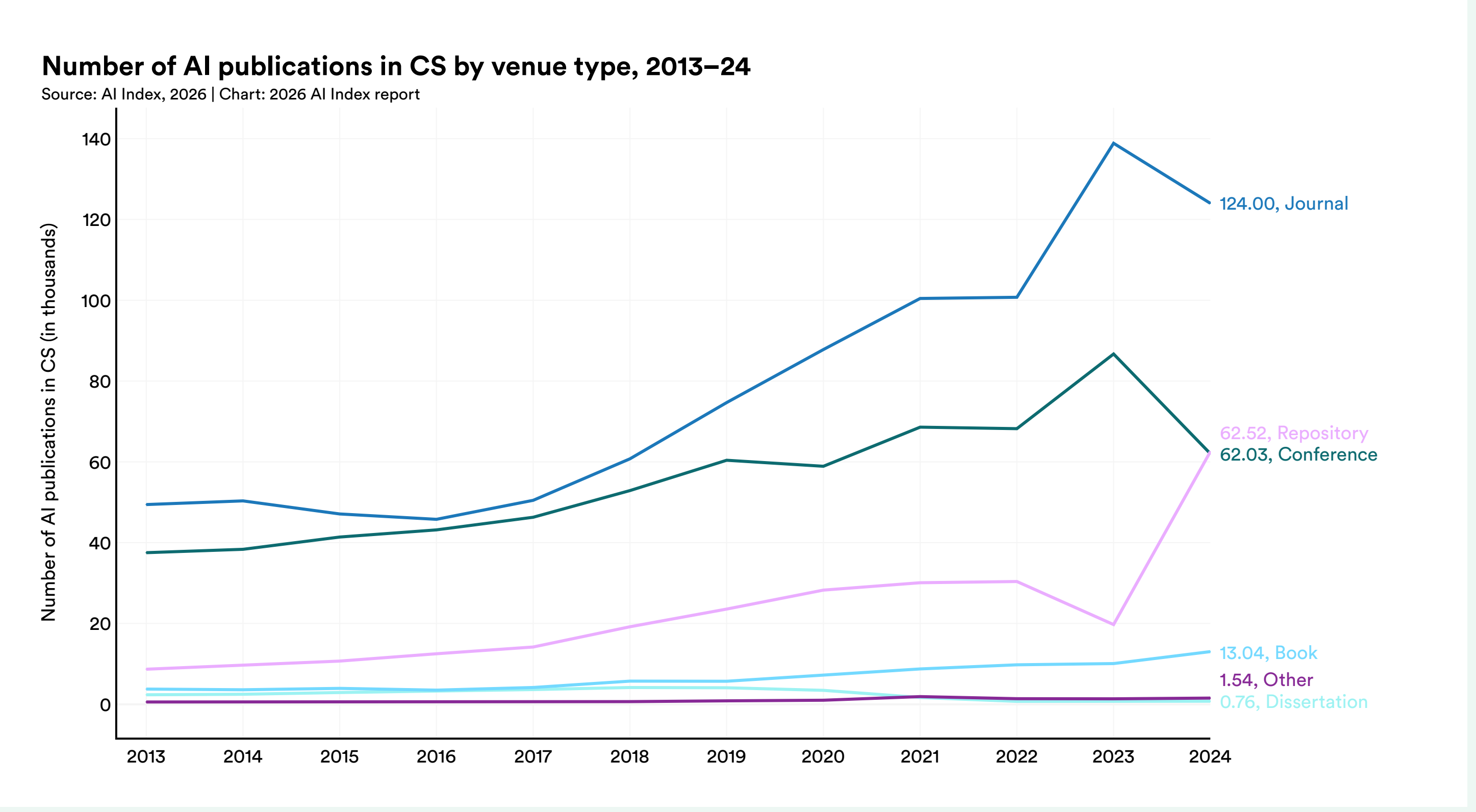

Figure 1.6.2

## Conference Attendance

Publication venue patterns capture where AI research is formally published, while conference attendance offers a complementary view of research community engagement. Across the 16 major conferences tracked by the AI Index—AAAI, AAMAS, CVPR, EMNLP, FAccT, ICAPS, ICCV, ICLR, ICML, ICRA, IJCAI, IROS, KR, NeurIPS, UAI, and IUI—total attendance increased in 2024 from the previous year (Figure 1.6.3). The largest conferences, including NeurIPS, CVPR, and ICML, continued to draw the highest attendance, while smaller ones such as ICAPS, KR, and UAI maintained stable participation levels (Figures 1.6.4 and 1.6.5). This data should be interpreted with caution, as many conferences have recently switched to virtual or hybrid formats. Conference organizers report that measuring the exact attendance numbers at virtual conferences is difficult, as virtual conferences allow for higher attendance by researchers from around the world. The AI Index reports total attendance figures, encompassing virtual, hybrid, and in-person participation.

25 In this context, 'repository' refers to preprint platforms such as arXiv, where researchers post papers prior to or independent of formal peer-reviewed publication in a journal or conference.

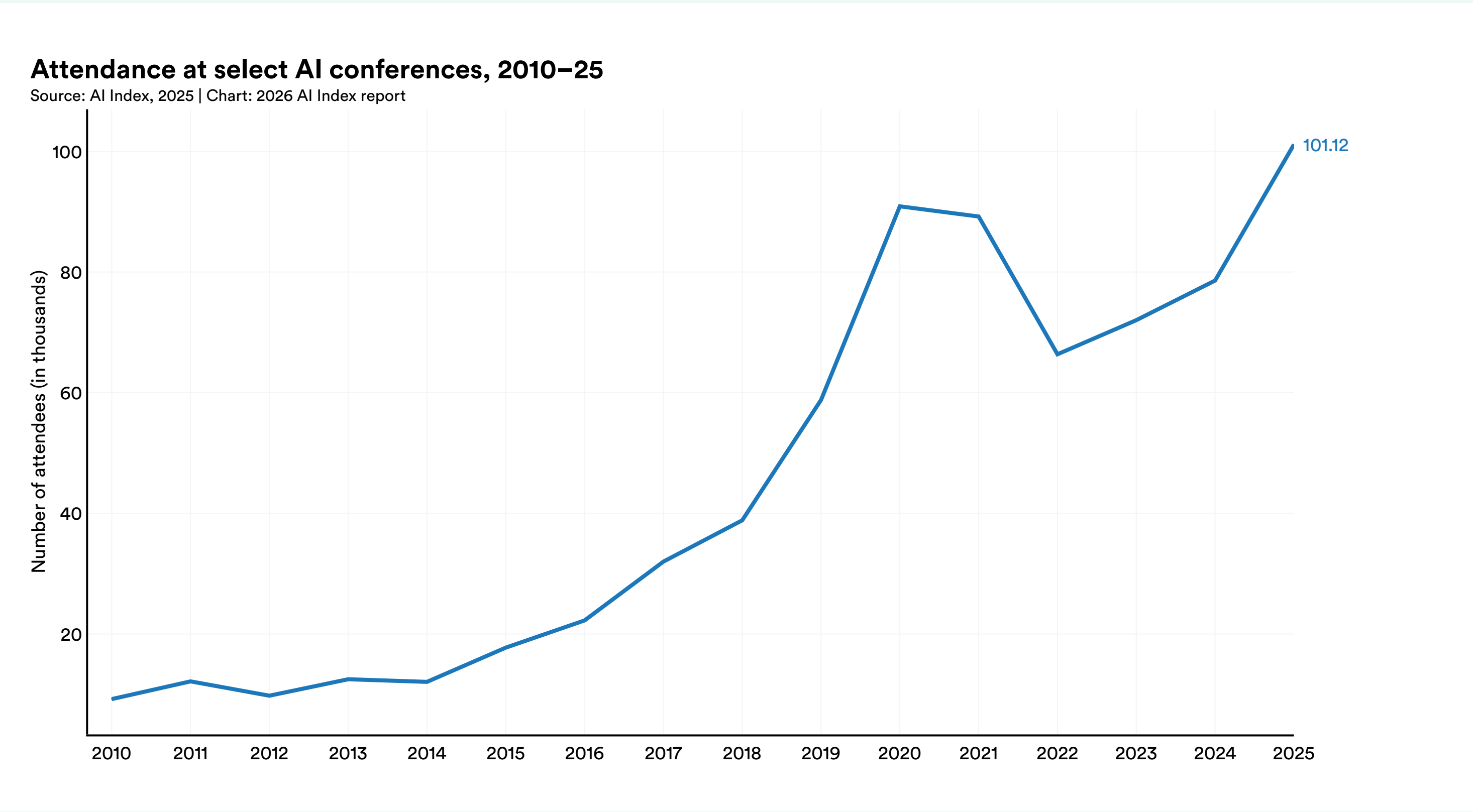

Figure 1.6.3

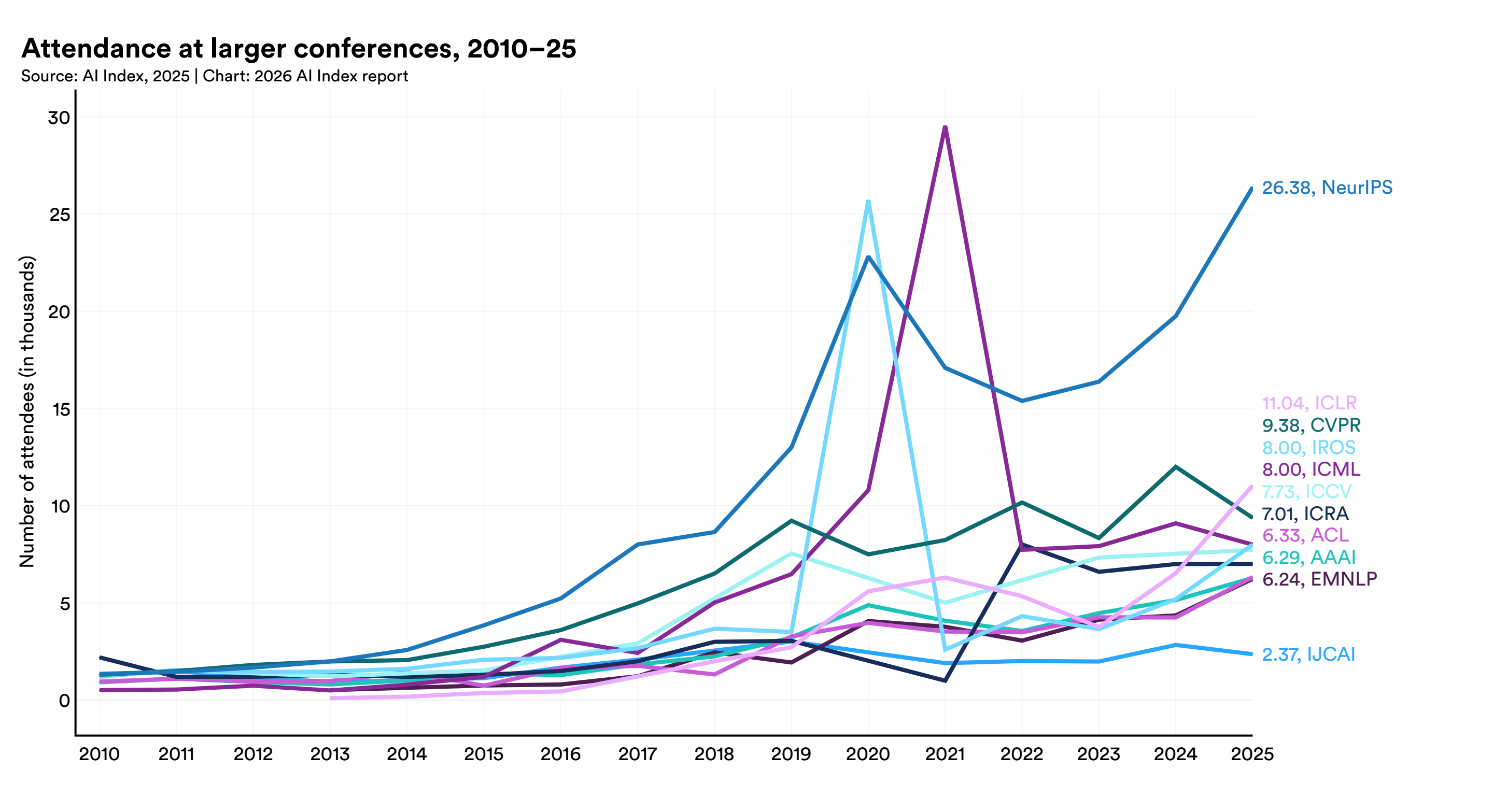

Figure 1.6.4[26]

26 The significant spike in ICML attendance in 2021 was likely due to the conference being held virtually that year.

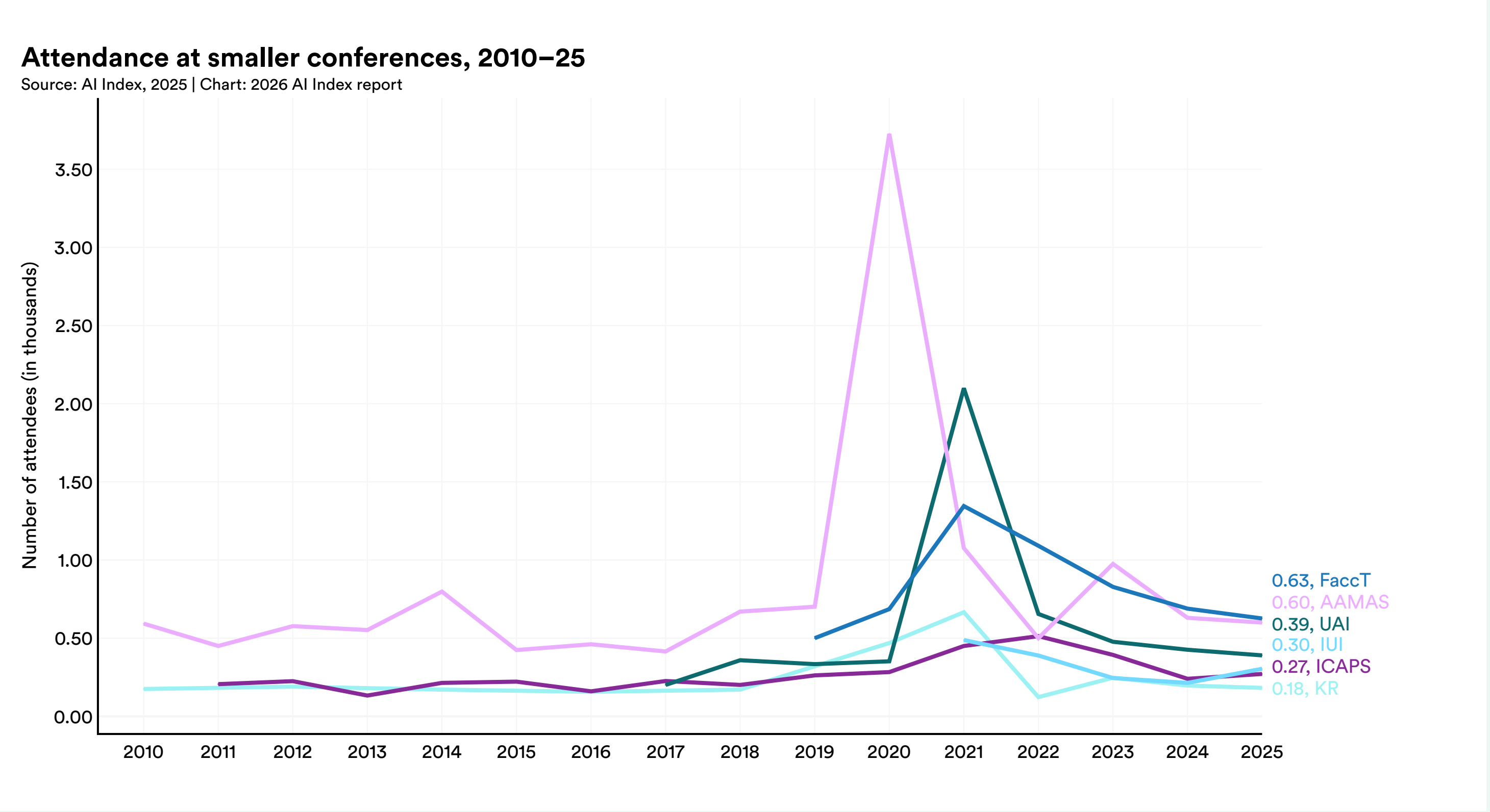

Figure 1.6.5[27]

## By National Affiliation[28]

In 2024, China accounted for 17.8% of AI publications in 2024, compared to 11.1% from Europe and 7.6% from India (Figure 1.6.6[29]). Chinese AI publications also accounted for 20.6% of all AI citations in 2024, followed closely by Europe at 19.5% and the United States at 12.6% (Figure 1.6.7). The United States saw a decline of 3 percentage points in publication share, though its citation share remained relatively unchanged (12.6% in 2024 vs. 13.03% in 2023). The "unknown" share in publication data rose to 39.3% in 2024, a spike that likely reflects changes in metadata coverage. The geographic distribution of publications and citations adds context to the notable model trends discussed earlier in the chapter, where a relatively small number of countries account for a disproportionate share of activity.

27 IUI 2021 and 2022 were held exclusively virtually.

28 Regions in this chapter are classified according to the World Bank analytical grouping. The AI Index determines an author's country affiliation using the "countries" field from the authorship data. This field lists all the countries with which an author is affiliated, as retrieved from OpenAlex based on institutional affiliations. These affiliations can be explicitly stated in the paper or inferred from the author's most recent publications. When counting publications by country, the AI Index assigns one count to each country linked to the publication. For example, if a paper has three authors, two affiliated with institutions in the United States and one in China, the publication is counted once for the United States and once for China.

29 A publication may have an "unknown" country affiliation when the author's institutional affiliation is missing or incomplete. This issue arises due to various factors, including unstructured or omitted institution names, platform functional deficiencies, group authorship practices, unstandardized affiliation labeling, document type inconsistencies, or the author's limited publication record. The problem as it relates to OpenAlex is addressed in this paper; however, the issue of missing institutions pertains to other bibliographic databases as well.

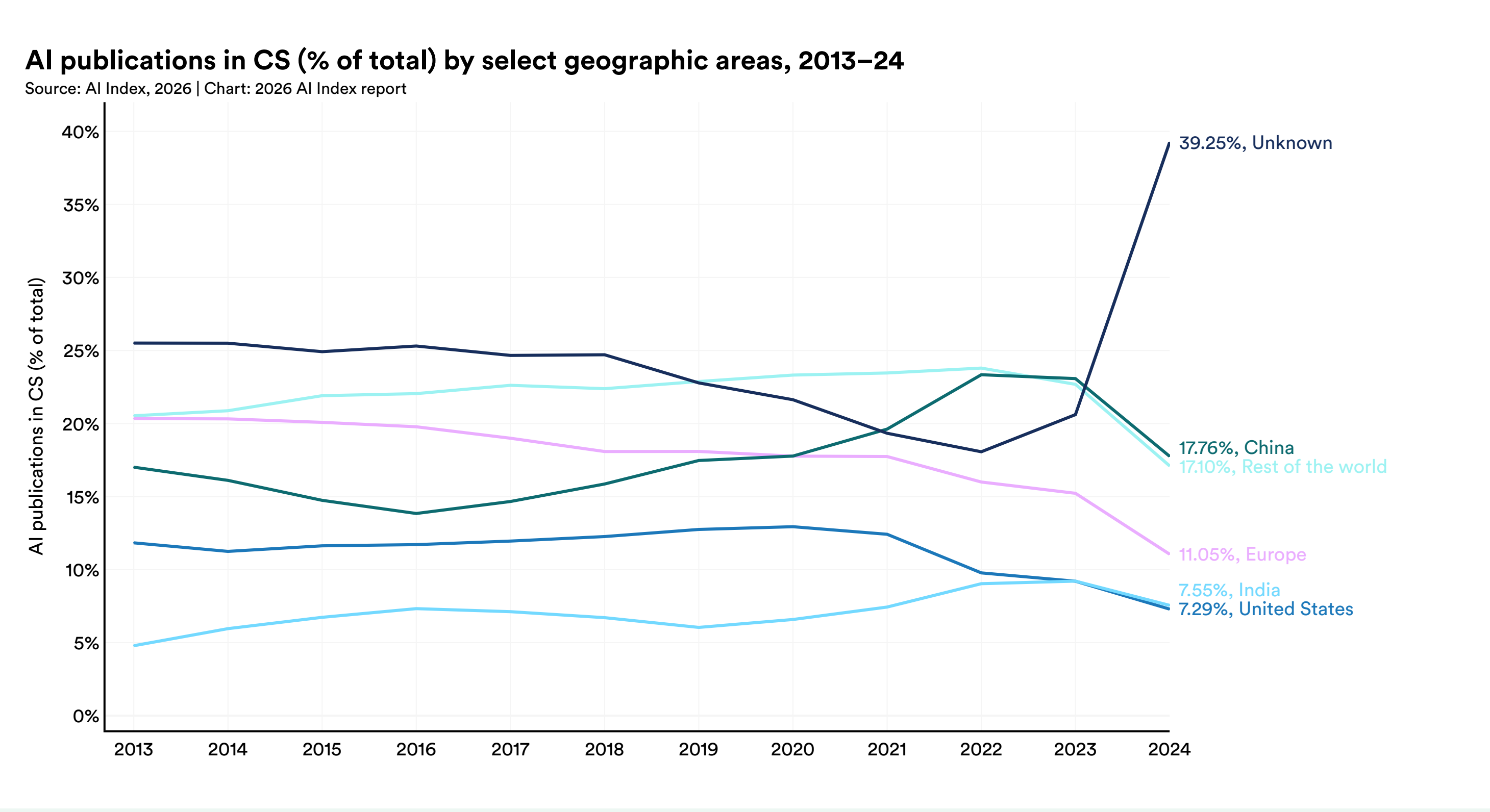

Figure 1.6.6[30]

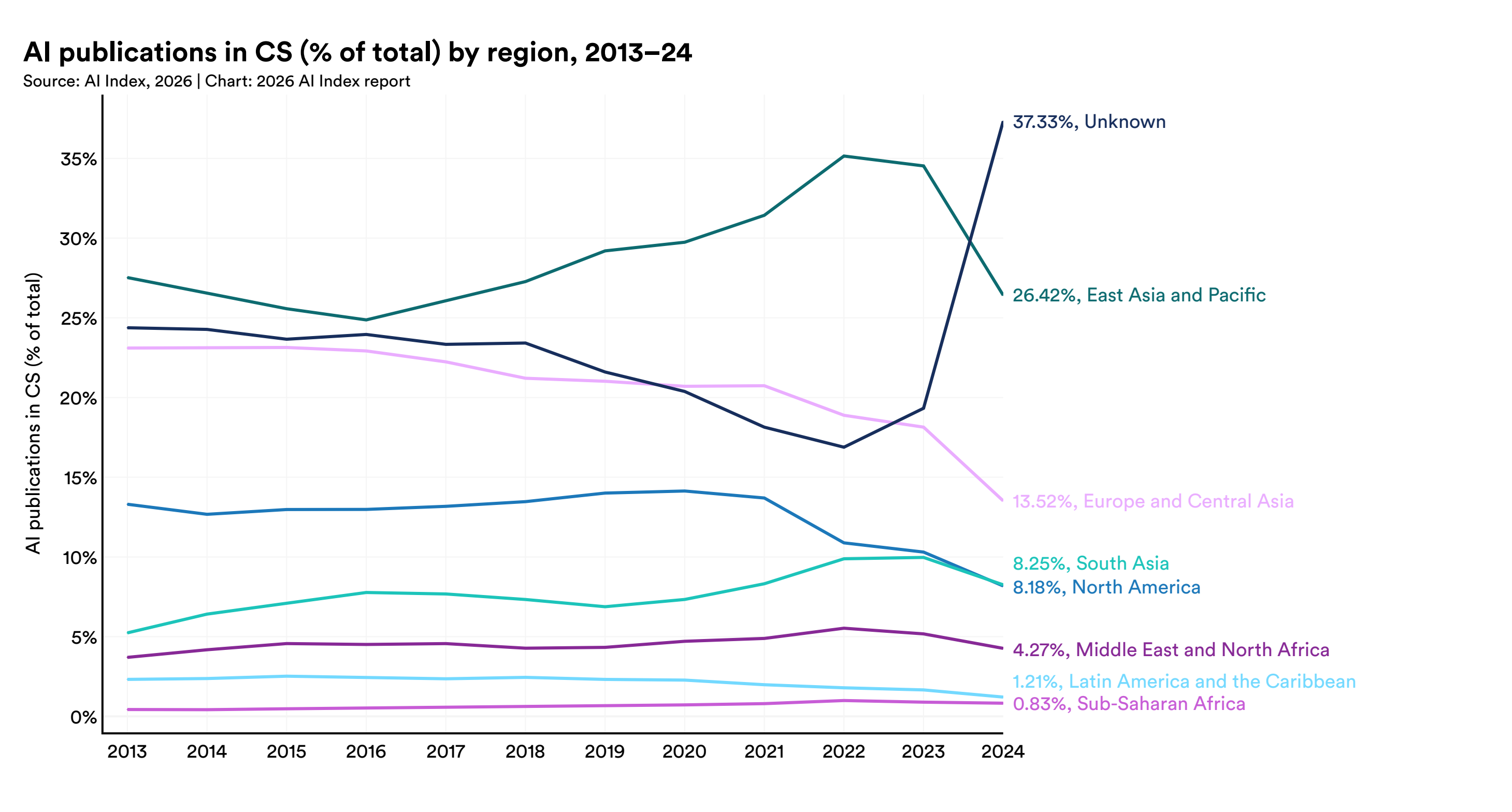

Figure 1.6.7

30 For the sake of brevity, the AI Index visualized results for a select group of countries. However, complete results for all countries will be available on the AI Index's Global Vibrancy Tool by the end of 2026. For immediate access to country-specific research and development data, please contact the AI Index team.

## By Sector

Academia produced the majority of AI publications in 2024 (68.1%), followed by government institutions (12.4%), industry (11.5%), and nonprofit organizations (4.6%)(Figure 1.6.8). The sector breakdown does vary by region (Figure 1.6.9). In the United States, a higher share of AI publications came from industry (24.6%) compared to China (18%), where government institutions were more meaningful contributors (25.1%). Europe had the highest percentage of AI publications originating from academia (55.3%).

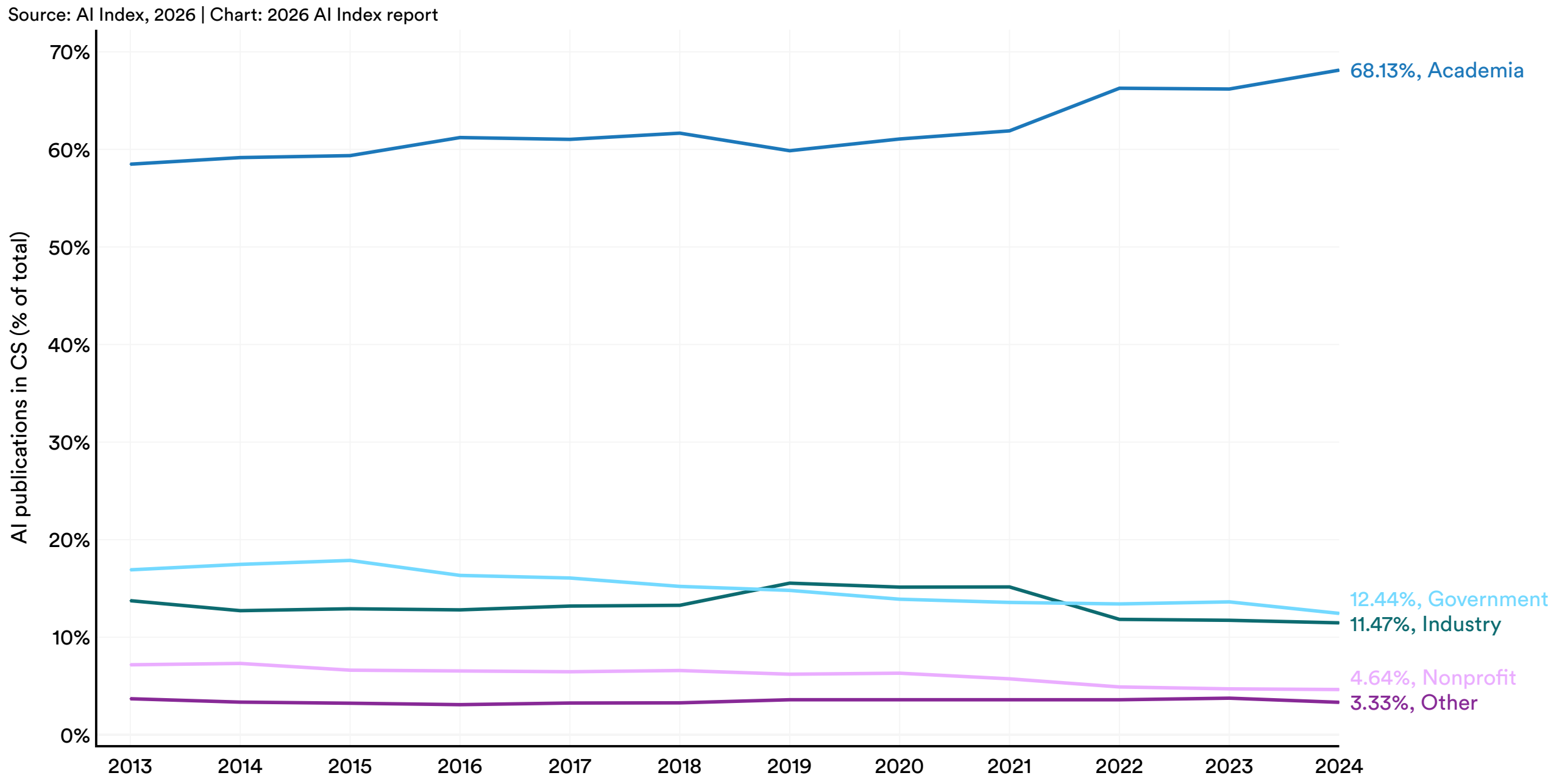
AI publications in CS (% of total) by sector, 2013–24
Source: AI Index, 2026 | Chart: 2026 AI Index report

Figure 1.6.8[31]

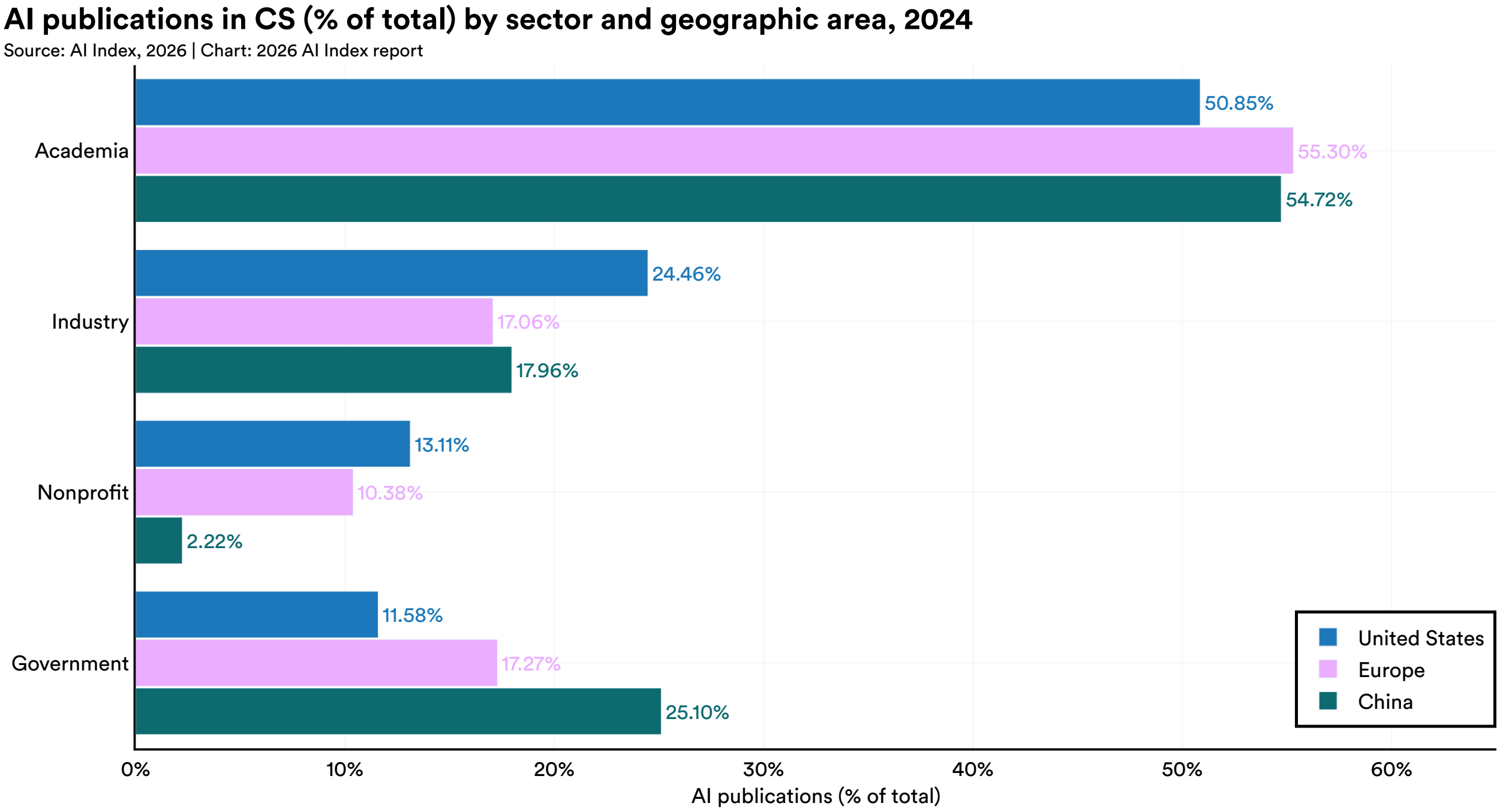
AI publications in CS (% of total) by sector and geographic area, 2024
Source: AI Index, 2026 | Chart: 2026 AI Index report

Figure 1.6.9

31 For Figure 1.6.8 and Figure 1.6.9, publications with unknown affiliations were excluded.

## By Topic

AI research in 2024 remained concentrated in a small set of core topics, though the breadth of areas continued to expand. Similar to the previous year, the most prevalent research topic was machine learning (37%), followed by computer vision (22.4%), pattern recognition (11.2%), and natural language processing (10%) (Figure 1.6.10). Publications on generative AI continued to show sharp growth, extending the trend from previous years. It is also worth noting that the AI Index topic classifier can assign multiple topic labels to a single publication, so topic totals can be seen as overlapping categories rather than mutually exclusive.

**Number of AI publications by select top topics, 2013–24**

Source: AI Index, 2026 | Chart: 2026 AI Index report

**Figure 1.6.10**[32]

# Top 100 Publications

The AI Index identified the 100 most-cited AI publications from 2021 to 2024 using citation data from OpenAlex.[33] Due to citation lag, this set can shift as citations accumulate over time.[34] The publication volume data above captures the scale of research activity, while the top 100 offers a more selective view on which work is gaining the most recognition and influence.

## By National Affiliation

The geographic distribution of the top 100 has shifted over time (Figure 1.6.11). The United States still ranks highest in top-cited publications each year, though its share has gradually declined from 64 in 2021 to 46 in 2024. China's share increased to 41 in 2024, from 34 in 2023, and Australia increased to 14 highly cited publications, up from 2 in 2023 and 6 in 2021.

32 The AI Index categorized papers using its own topic classifier. It is possible for a single publication to be assigned multiple topic labels.

33 The full methodological guide can be accessed in the Appendix, along with the list of the top 100 articles.

34 A publication can have multiple authors from different countries or organizations. If a paper includes authors from multiple countries, each country is credited once. As a result, some of the totals in this section exceed 100.

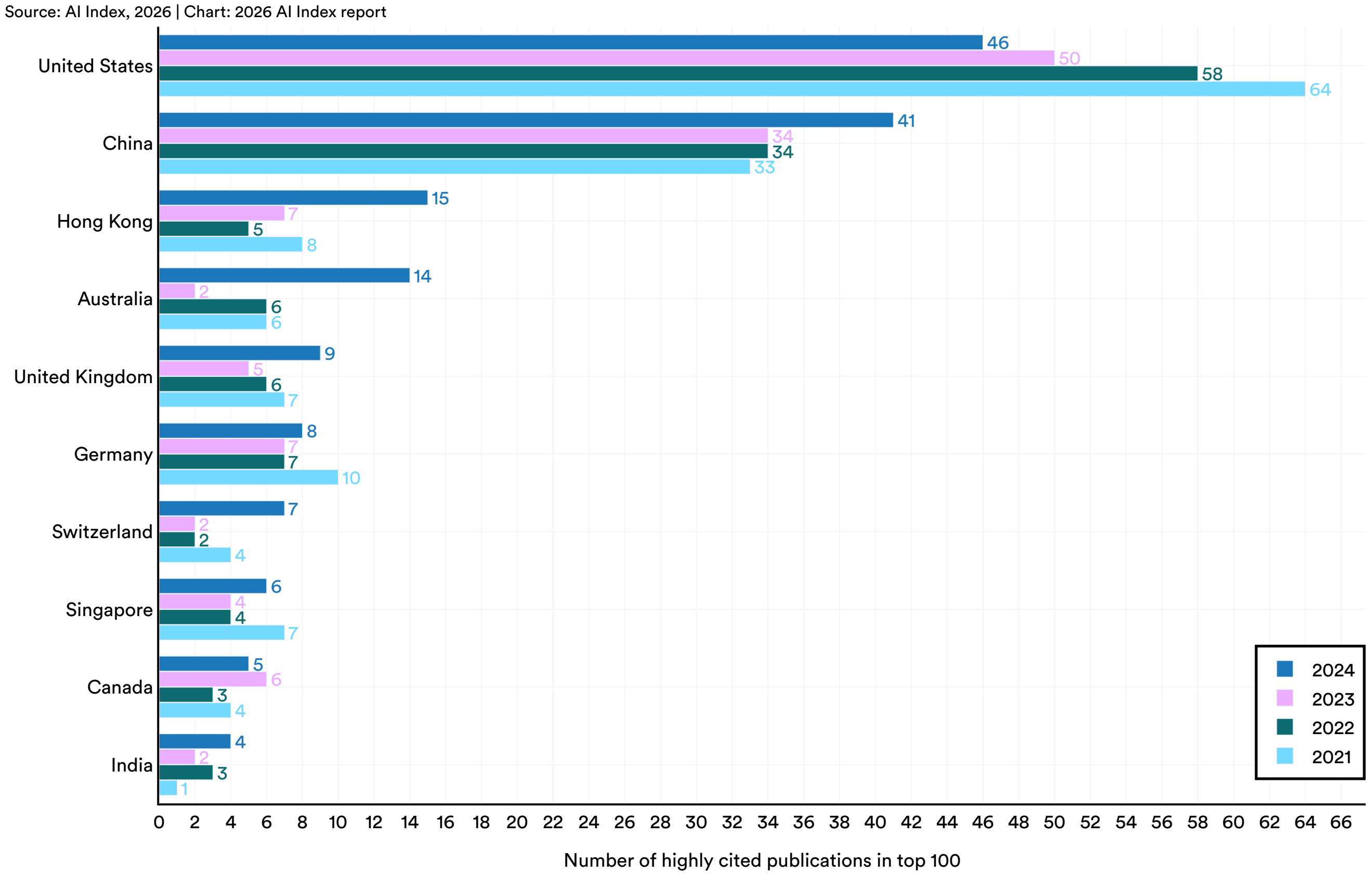
Number of highly cited publications in top 100 by select geographic areas, 2021–24
Source: AI Index, 2026 | Chart: 2026 AI Index report

Figure 1.6.11

## By Sector and Organization

The sector composition of the top 100 remained consistent, with academia producing the most top-cited publications year over year (Figure 1.6.12). Industry contributions declined sharply from 17 in 2021 and 19 in 2022 to six in 2024, even as industry's share of notable model releases has continued to grow (Section 1.1). The organization distribution varies by year, though output remains concentrated among a small set of institutions (Figure 1.6.13). In 2024, Stanford University and Google led with seven publications each, and the Chinese Academy of Sciences and Microsoft followed closely, with each contributing five.

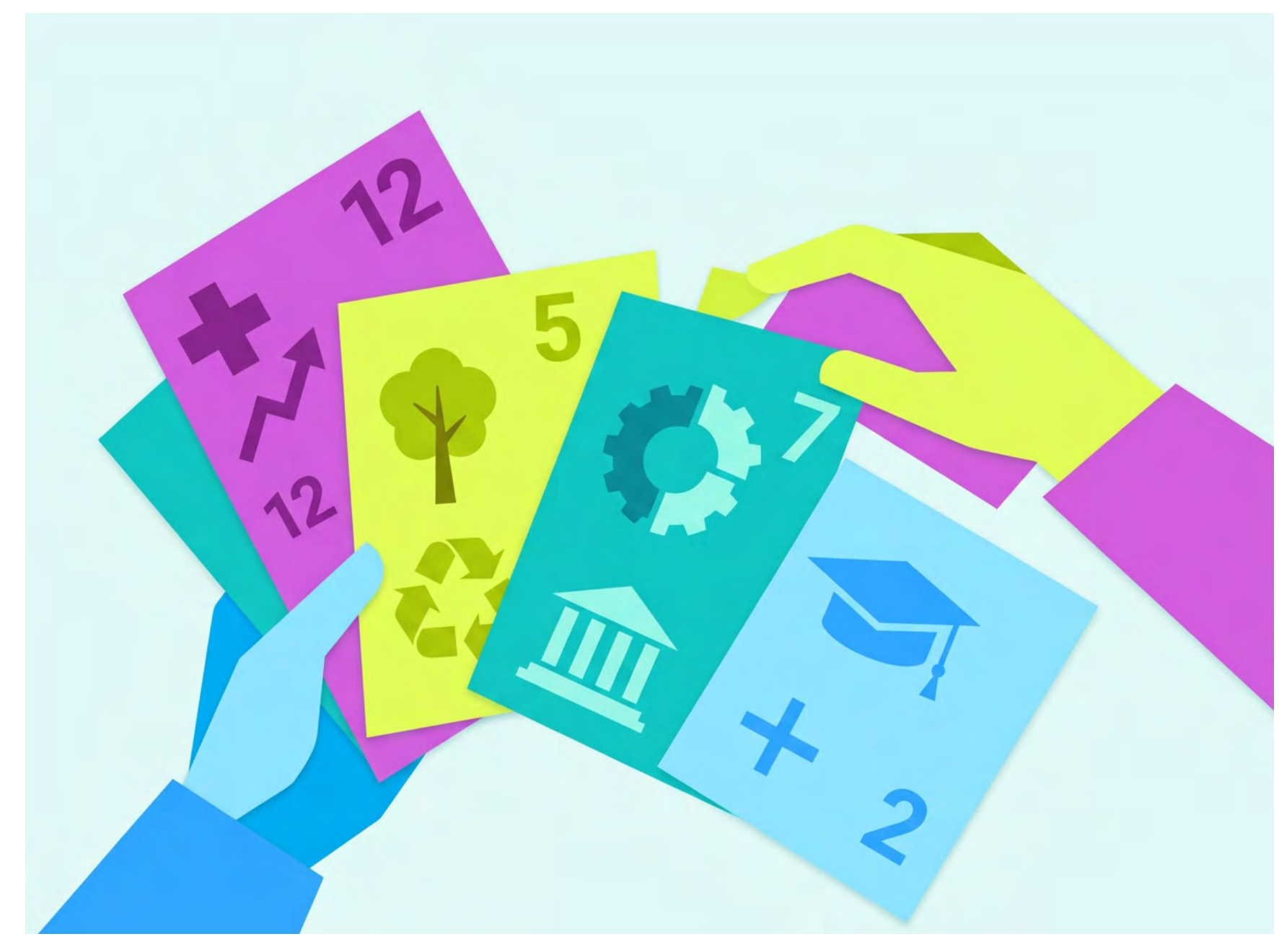

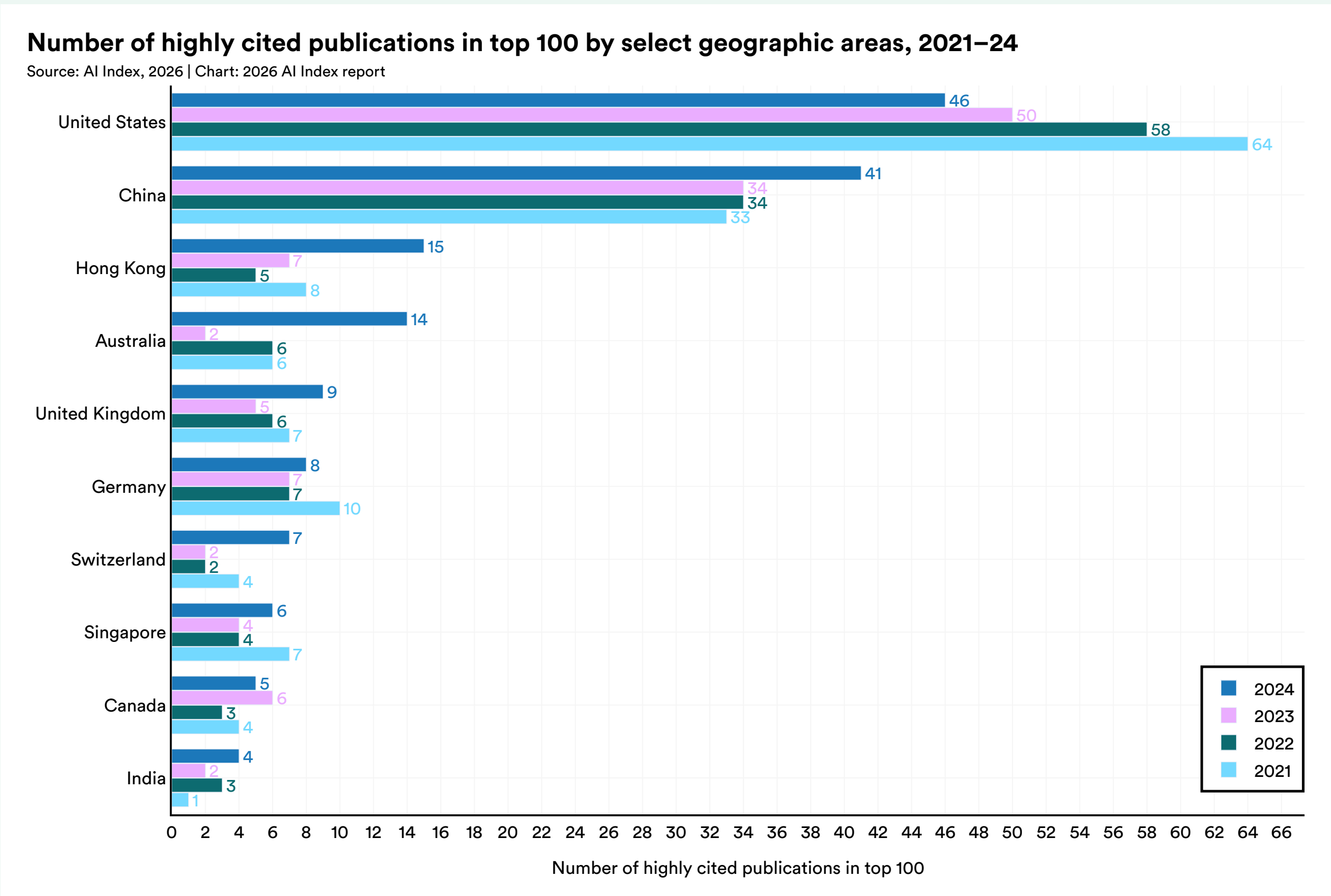

Figure 1.6.12[35]

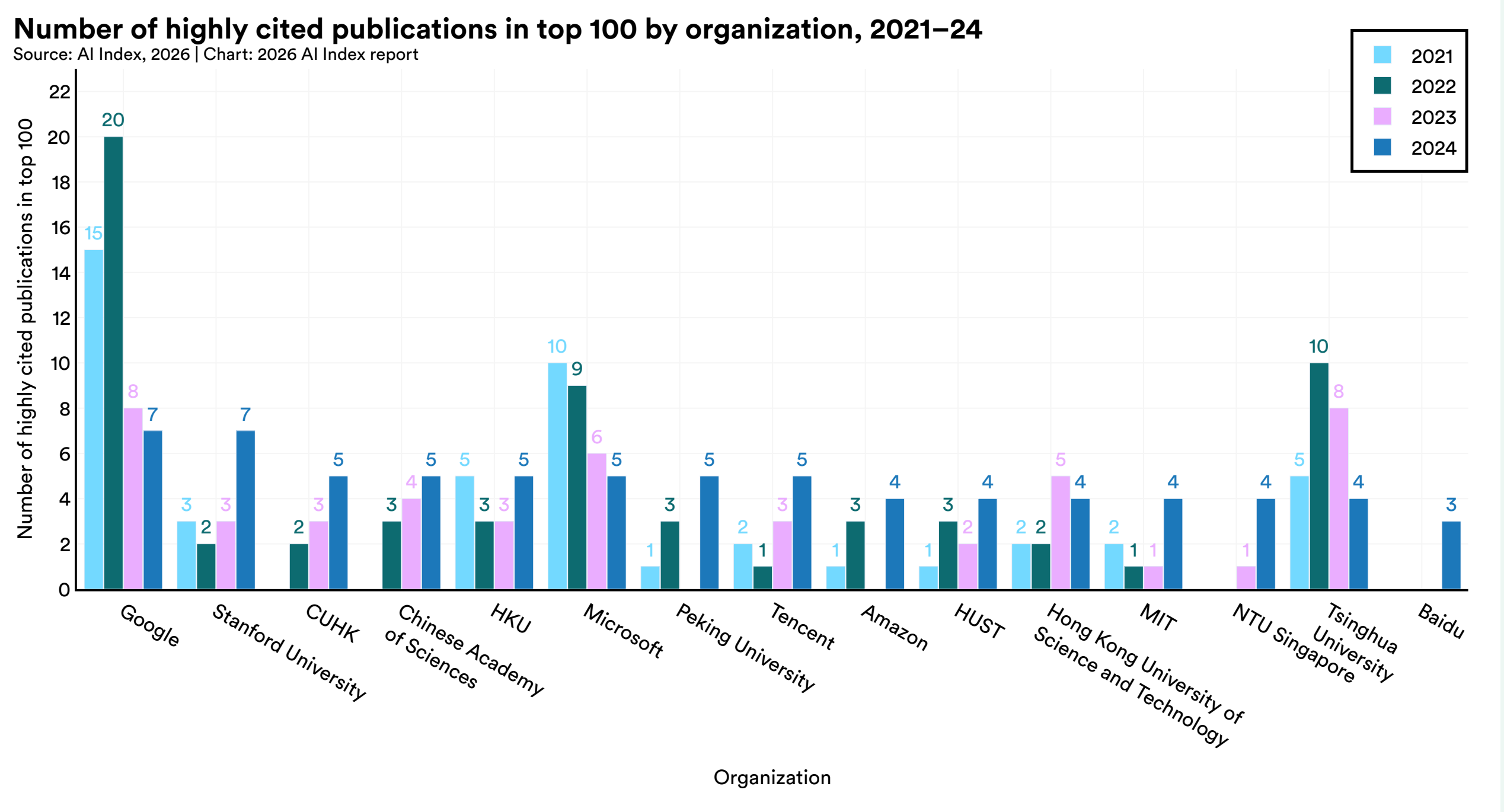

Figure 1.6.13[36]

35 The "other" category includes sectors and intersector collaborations that are not industry and academia, or industry-academia collaborations (e.g., industry and government, academia and nonprofit). Some institutions lack data for 2021 because they did not have papers included in the top 100 that year. Since papers can have multiple authors from different institutions, the total institutional tags in Figure 1.6.13 may exceed 100. Also, because two of the papers had authors with an unknown sectoral affiliation in 2022, the total sum of publications in Figure 1.6.12 is 98.

36 Universities are abbreviated as follows: CUHK = The Chinese University of Hong Kong; HKU = The University of Hong Kong; HUST = Huazhong University of Science and Technology; MIT = Massachusetts Institute of Technology; NTU Singapore = Nanyang Technological University, Singapore.

# 1.7 Patents

While publications track research outputs, patents offer insight into applied innovation and commercial development. This section examines trends in global AI patents over time. Patents can provide another lens for tracking innovation across organizations and geographic areas, particularly in applied AI contexts. Similar to publications data, there are notable delays before AI patent data becomes available, with 2024 being the most recent year accessible. The analysis draws from patent-level bibliographic records in PATSTAT Global, a comprehensive database provided by the European Patent Office (EPO).[37]

## Global Trends

Globally, the number of granted AI patents has grown exponentially, from 3,866 in 2010 to 131,121 in 2024 (Figure 1.7.1). Between 2023 and 2024, patent grants rose by 8.2%. China accounts for the majority, at 74.2% of the global total (Figures 1.7.2 and 1.7.3). The United States is the next major contributor at 12.1% (15,290 patents), followed by Europe (3%) and India (0.4%). Over the past decade, the United States' share has declined steadily from a peak of 42.8% in 2015, while China's share has risen from under 20% to its current level. Patents and publications reflect different stages in the R&D pipeline, so China's lead in both, while not directly correlated, is consistent with the country's growing research presence described earlier.

Other regional leaders emerge when patent activity is normalized by population size (Figure 1.7.4). In 2024, South Korea had the highest number of granted AI patents on a per capita basis (14.3%), followed by Luxembourg (12.3%) and China (7.0%).

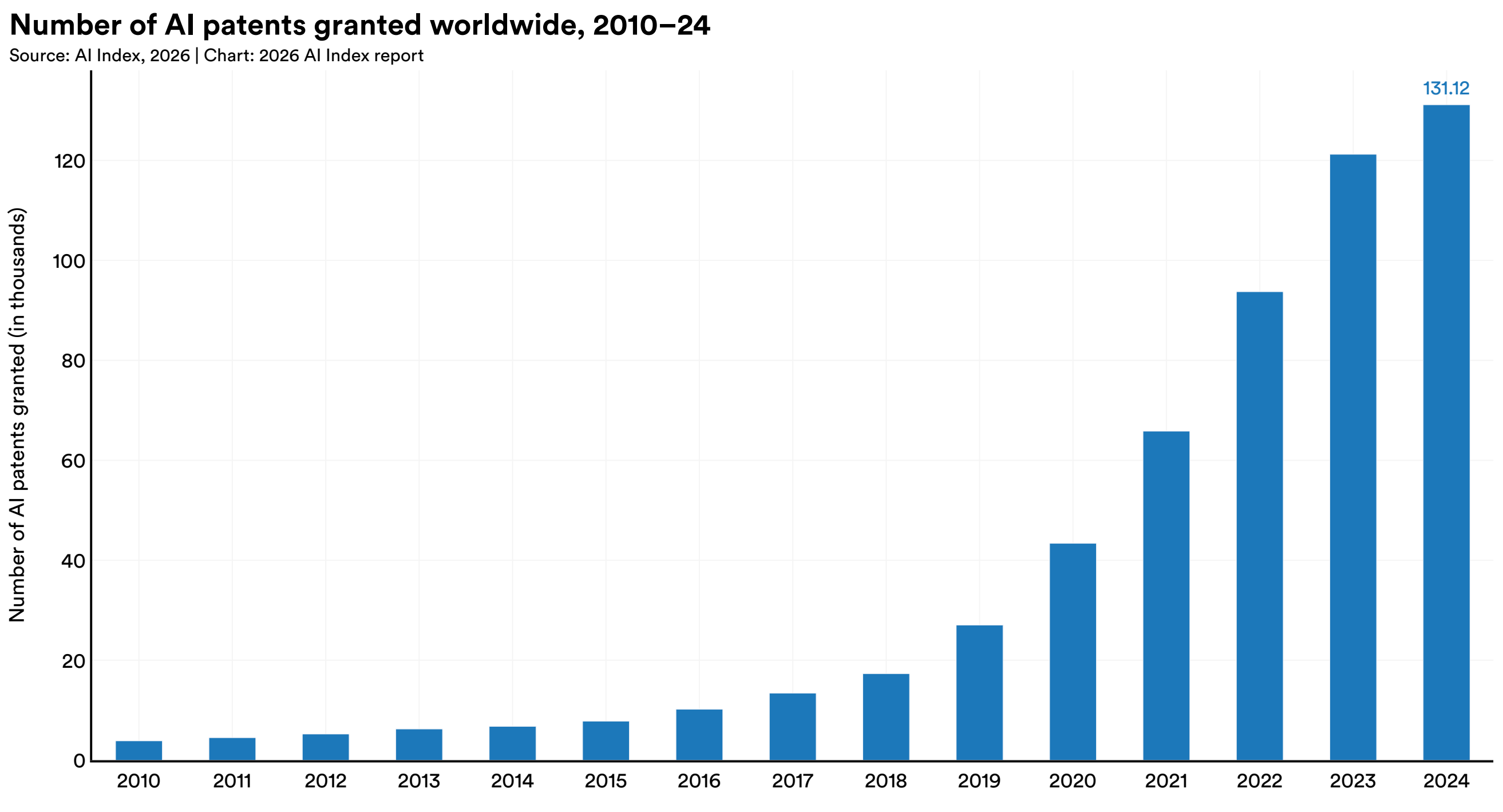

Figure 1.7.1[38]

37 More details on the methodology behind this section's patent analysis can be found in the Appendix.

38 Patent standards and laws vary across countries and regions, so these charts should be interpreted with caution. More detailed country-level patent information will be released in a subsequent edition of the AI Index's Global Vibrancy Tool.

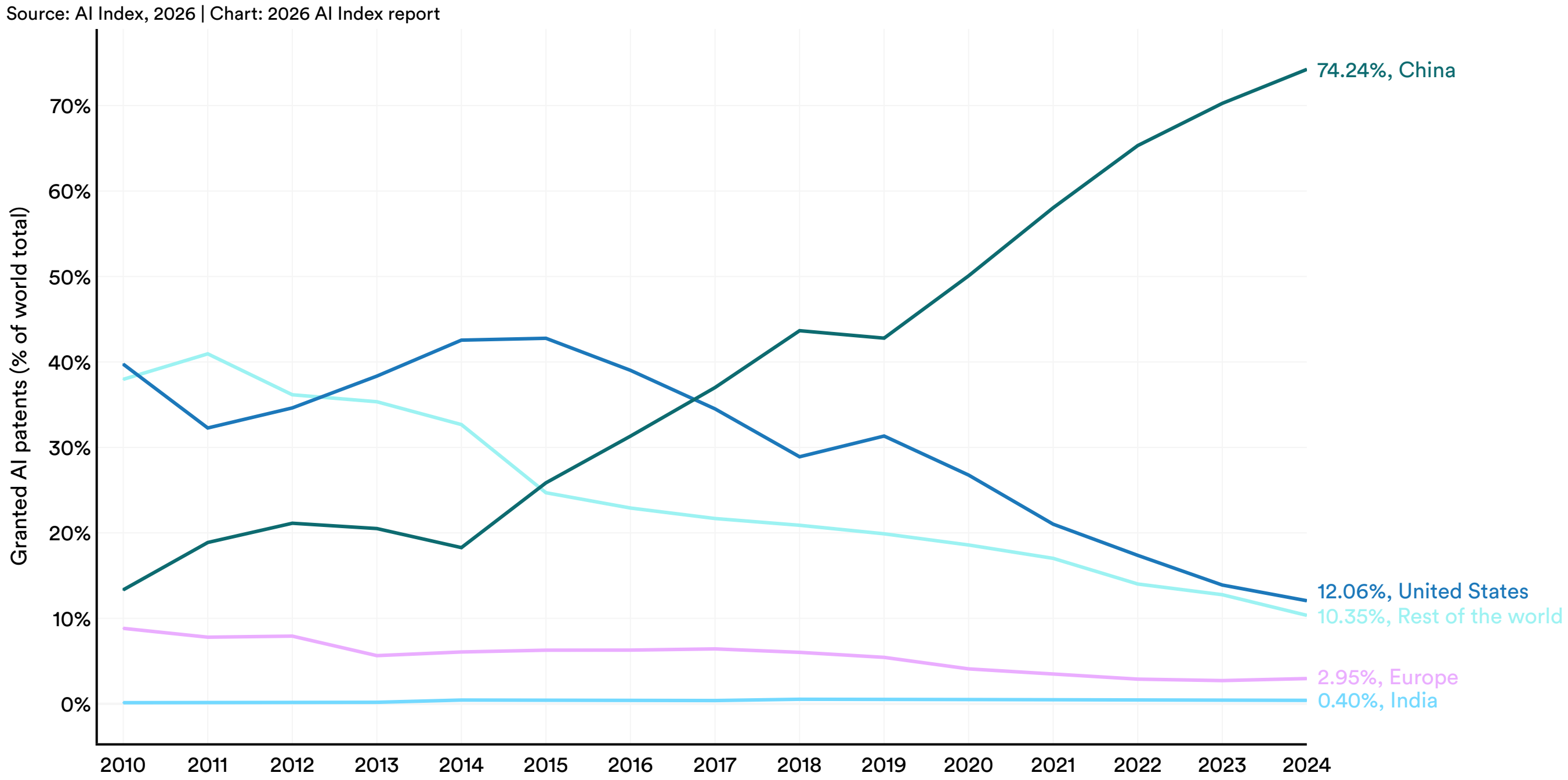
Granted AI patents (% of world total) by select geographic areas, 2010–24
Source: AI Index, 2026 | Chart: 2026 AI Index report

Figure 1.7.2

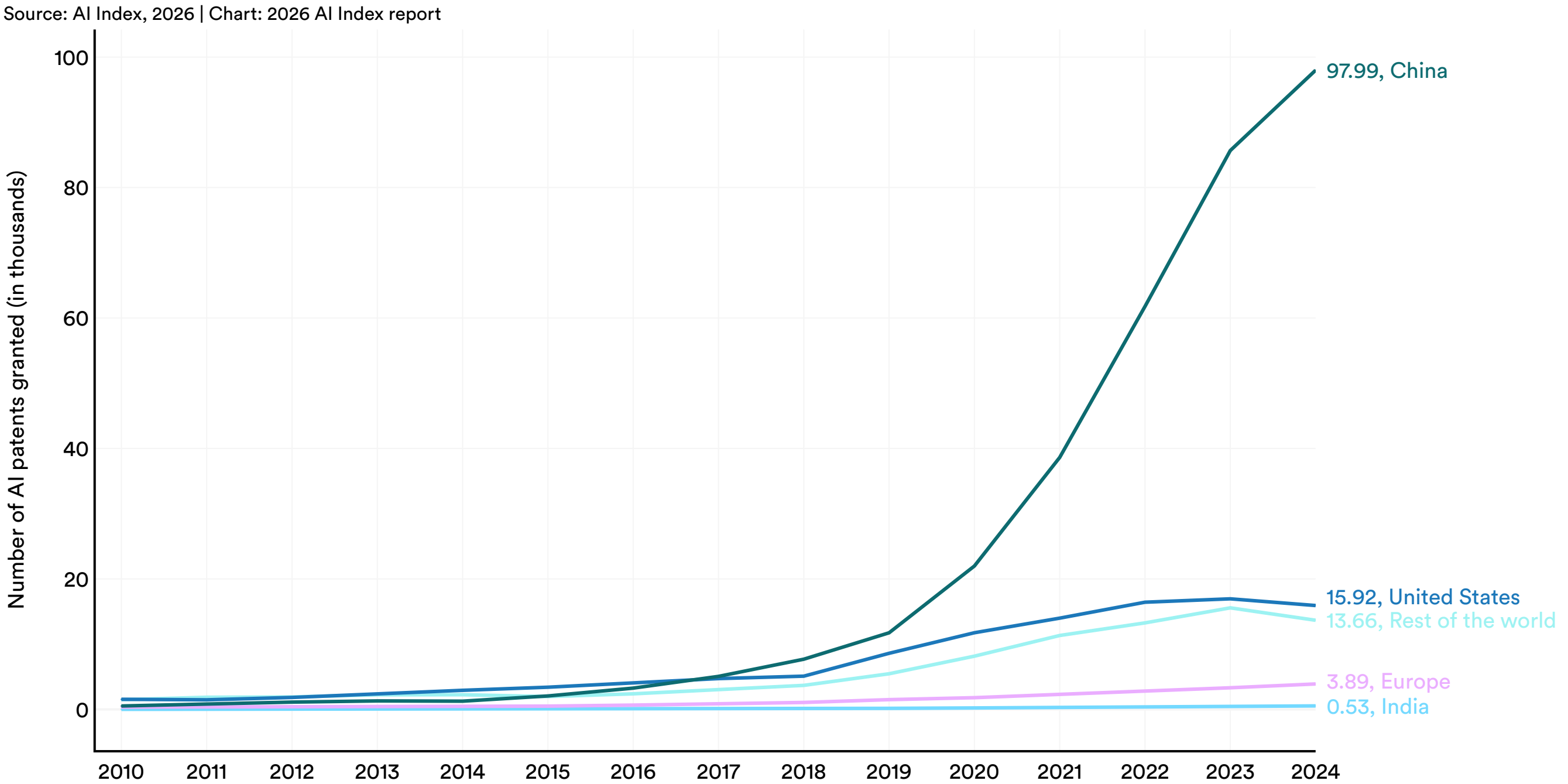
Number of AI patents granted by select geographic areas, 2010–24
Source: AI Index, 2026 | Chart: 2026 AI Index report

Figure 1.7.3

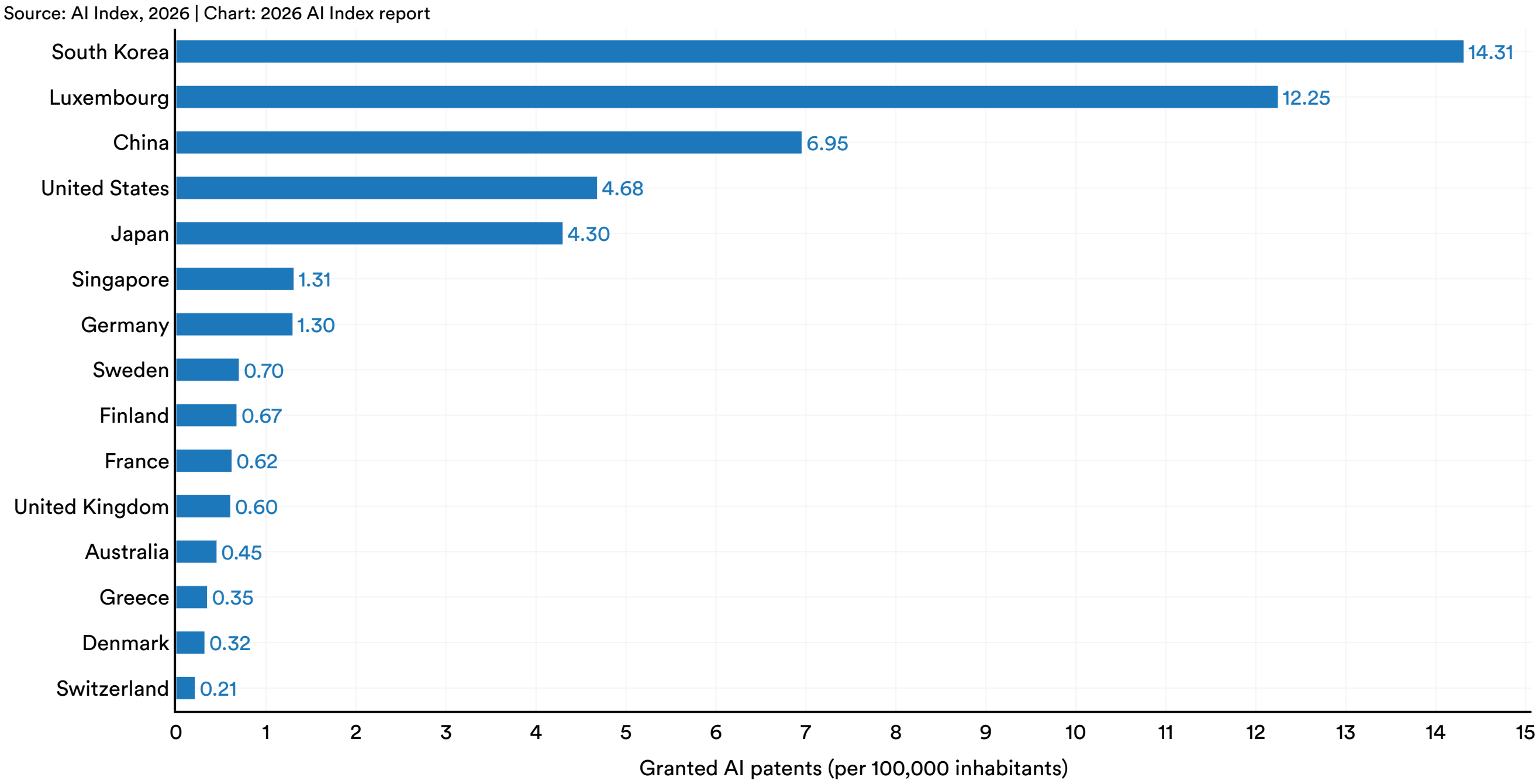
Granted AI patents per 100,000 inhabitants by country, 2024
Source: AI Index, 2026 | Chart: 2026 AI Index report

Figure 1.7.4

## Forward Citations Flow

When newly filed patents reference earlier ones, those references are called forward citations. These are often used as a proxy for influence, since they indicate how often an invention informs later work. By this measure, the United States accounts for over half of all AI patent forward citations, a signal of downstream influence that contrasts with its 12.1% share of patent volume (Figure 1.7.5). China ranks second despite producing the largest volume of patents by a wide margin. The relationship between forward citations and technological impact is not straightforward and has been called into question (Higham et al., 2021). There is also a strong home bias across all countries, with most citations occurring domestically, a well-documented pattern in patent citation geography (Jaffe et al., 1993; Cotterlaz et al. 2025; and Verluise et al., 2025).That said, the cross-border flows are not symmetric. Chinese patents are cited frequently in U.S. filing, while U.S. patents appear far less often in Chinese ones.

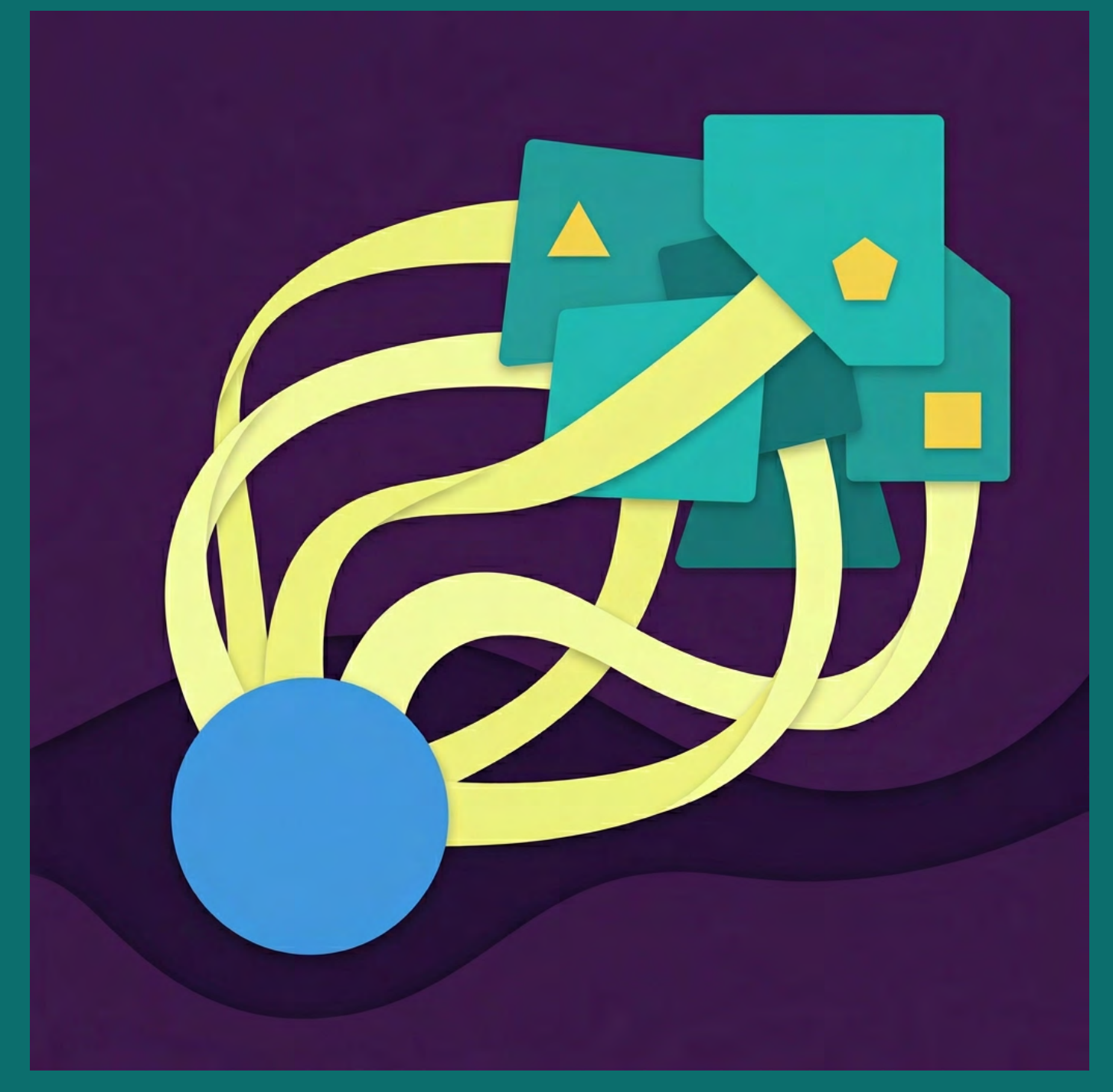

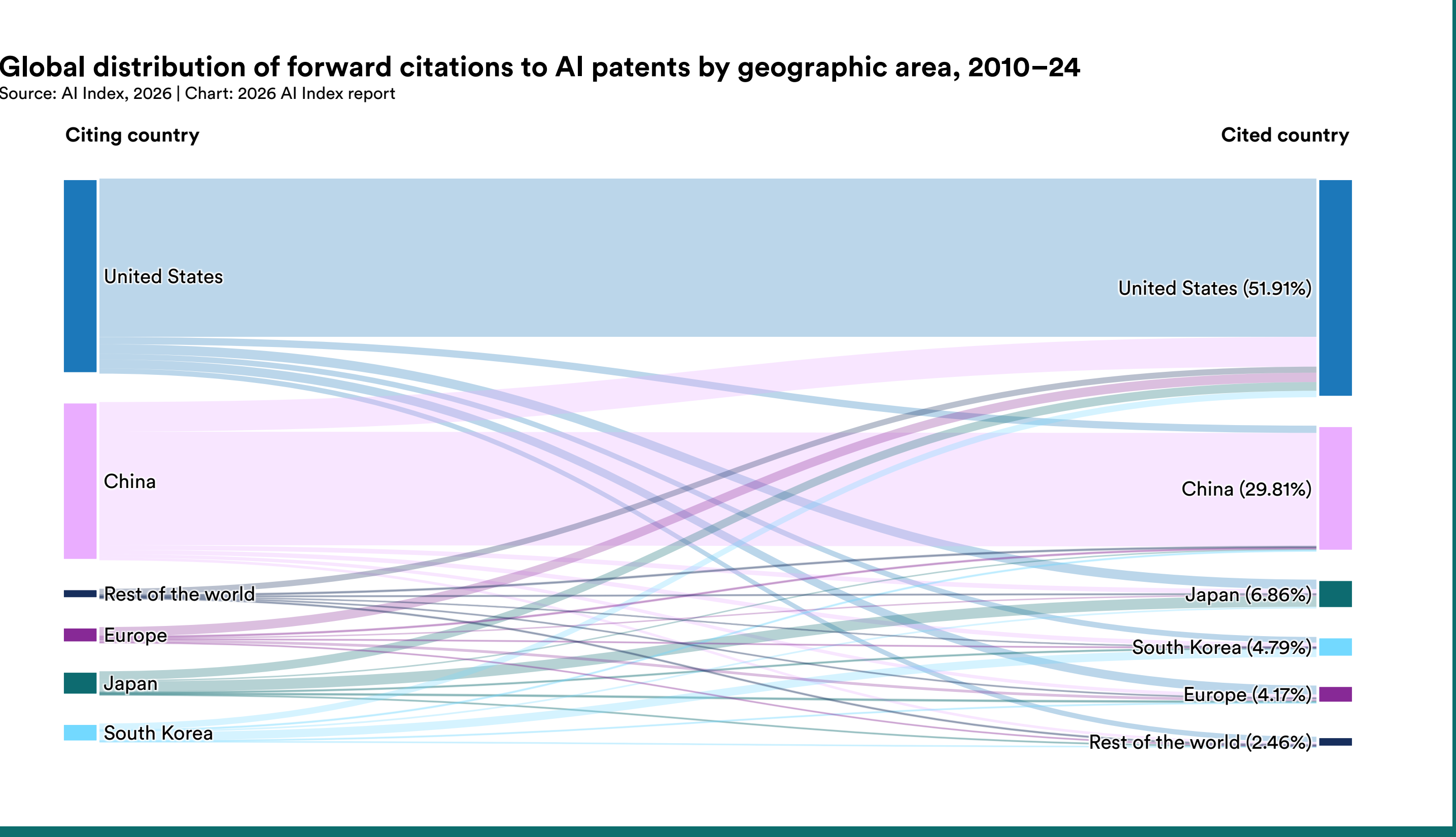

Figure 1.7.5[39]

## Speed of Knowledge Diffusion

Patent citation lag—the time between a patent's publication and its first forward citation—can be used to measure how quickly knowledge diffuses within a discipline. For AI patents, most receive their first citation within two to three years, reflecting a relatively fast diffusion. The speed varies by country (Figure 1.7.6). U.S. patents tend to be cited sooner and more consistently over time, with only 19% remaining uncited compared to 32% to 44% in other geographic areas. Japan's patents show early but narrower influence, and those from China and South Korea experience slower initial citation but, after about six years, citation activity stabilizes across all regions. The patterns are consistent with the forward citation data above, but differences in citation norms and home bias may also play a role.

39 Each data point in the figure reflects forward citations to AI patents, grouped at the patent family level to represent unique inventions rather than individual filings. Values are expressed as shares of all AI patent forward citations for patents granted between 2010 and 2024.

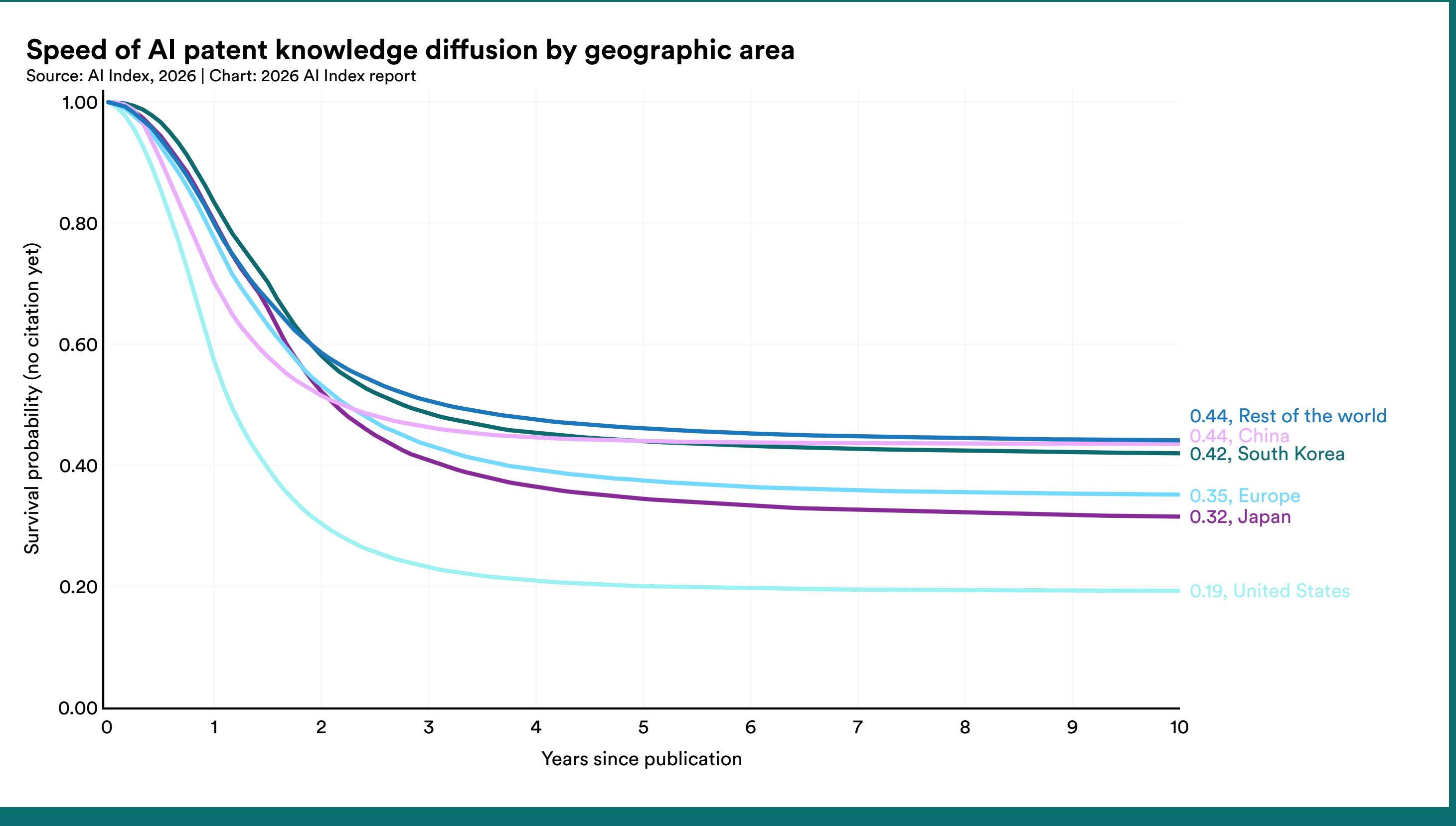

Figure 1.7.6[40]

## Technological Proximity

Technological proximity[41] measures whether countries are converging on similar types of AI innovation or pursuing distinct paths. Using a method proposed by Bar et al. (2012), the analysis[42] compares how closely each country's AI patent portfolio aligns with the two largest reference points, the United States and China (Figure 1.7.7). Overlap is scored on a scale from 0 (no similarity) to 1 (identical). Most countries cluster in the upper right, meaning their AI patents cover similar technological areas to both the U.S. and China, with a stronger lean toward the U.S. portfolio. India and Australia, for example, have patent portfolios that show close to 80% overlap with both. Denmark is the least similar to either reference point, showing only a 45% overlap with China and a 52% overlap with the United States. This is because Denmark's AI patents are concentrated in energy and wind-related technology categories (patent codes Y02E, F03D, F05B) rather than core computing and data-processing categories (G06F, G06N, G06K) that dominate both the U.S. and China. While most countries' AI innovation portfolios are structured similarly, national industrial strengths tend to influence where AI is applied.

40 This figure plots the probability of *not* being cited, so curves with sharper drops indicate shorter lags.

41 Also known as the min-complement proximity measure.

42 Technological classes are identified by International Patent Classification (IPC) and Cooperative Patent Classification (CPC) codes.

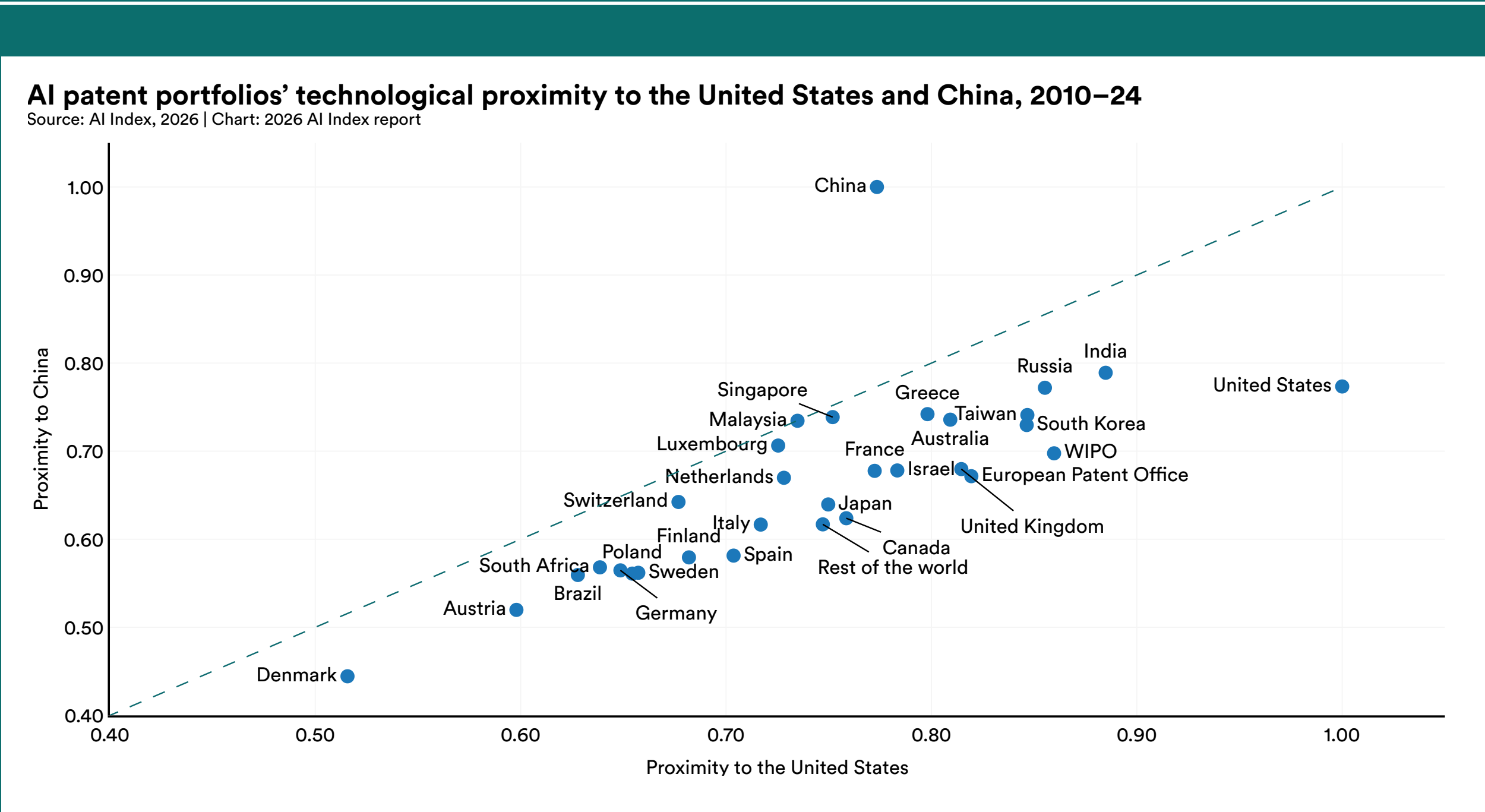

Figure 1.7.7

HIGHLIGHT:

## AI Patent Examples

### 1 Patent CN111431996A: Resource configuration method and device, equipment and medium, 2022, China

A machine-learning prediction model determines how to allocate computing resources across multiple services in a cluster. The system learns from historical and real-time signals—such as traffic volumes and CPU, memory, and network usage—to infer the right resource configuration. This enables automated, dynamic scaling decisions without relying on manual rules.

### 2 Patent US11436777B1: Machine learning-based hazard visualization system, 2022, United States

The system trains machine learning models to forecast hazard attributes (time, path, severity) for specific locations and identify infrastructure in geospatial imagery. It combines model outputs to annotate maps, showing where hazards intersect with critical assets. The system also supports causal inference—for example, identifying infrastructure repeatedly affected by hazards. These capabilities rely on learned prediction and image-recognition models rather than deterministic mapping logic.

### 3 Patent US2023239456A1: Display system with ML-based stereoscopic view synthesis over a wide field of view, 2025, United States

This head-mounted display uses machine-learning techniques—including depth estimation and reconstruction—to create perspective-correct stereoscopic images from external cameras. Neural models handle real-time vision challenges like disocclusion, artifact reduction, and sharpening by inferring scene geometry and filling gaps where camera viewpoints fail to align with the user's eyes. ML is a core component of the VR/AR passthrough rendering pipeline.

# 1.8 AI Authors and Inventors

The publications and patents discussed above reflect research and development outputs. Using Zeki data, the AI Index examined the geographic distribution and mobility patterns of the authors and inventors behind this work over time. This section covers a narrower slice of AI talent activity than the broader labor market indicators discussed in Chapter 4 (Economy). Zeki identifies talent outside of China based on observable AI outputs such as research, data depositories, and new models. The dataset covers 2010 to 2025 across a group of countries in North America, Europe, Asia, Latin America, and the Middle East.[43]

## Geographic Distribution

In 2025, the largest share of identified AI authors and inventors came from the United States (220,520), followed by India (50,460) and Germany (48,520) (Figure 1.8.1). The United Kingdom (34,370), Canada (31,450), and France (18,820) formed a second tier with Australia, the Netherlands, Italy, Brazil, Switzerland, and others making up the broader distribution of contributors. Looking at the data on a per capita basis surfaces countries that have relatively high levels of AI activity that are not visible when looking at total volume, as seen in the per capita patent data in Section 1.7. Switzerland led with 110.5 AI authors and inventors per 100,000 inhabitants, followed closely by Singapore (109.5) (Figure 1.8.2). Countries with smaller populations, such as Finland (77.6), the Netherlands (77.6), and Denmark (66.3), rank above larger nations including Germany (58.1) and the United Kingdom (49.6).

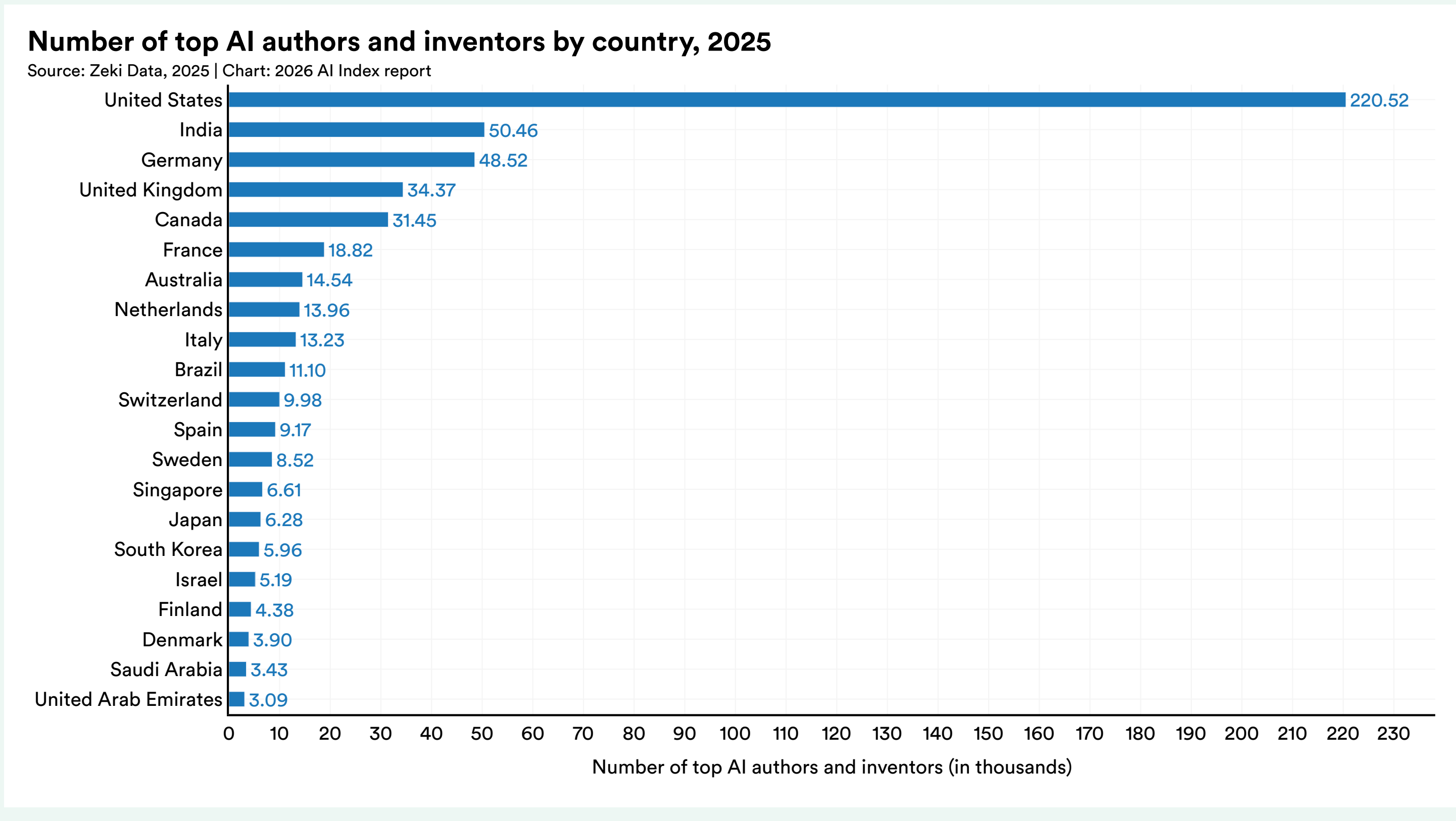

Figure 1.8.1

43 For more details, refer to the Appendix.

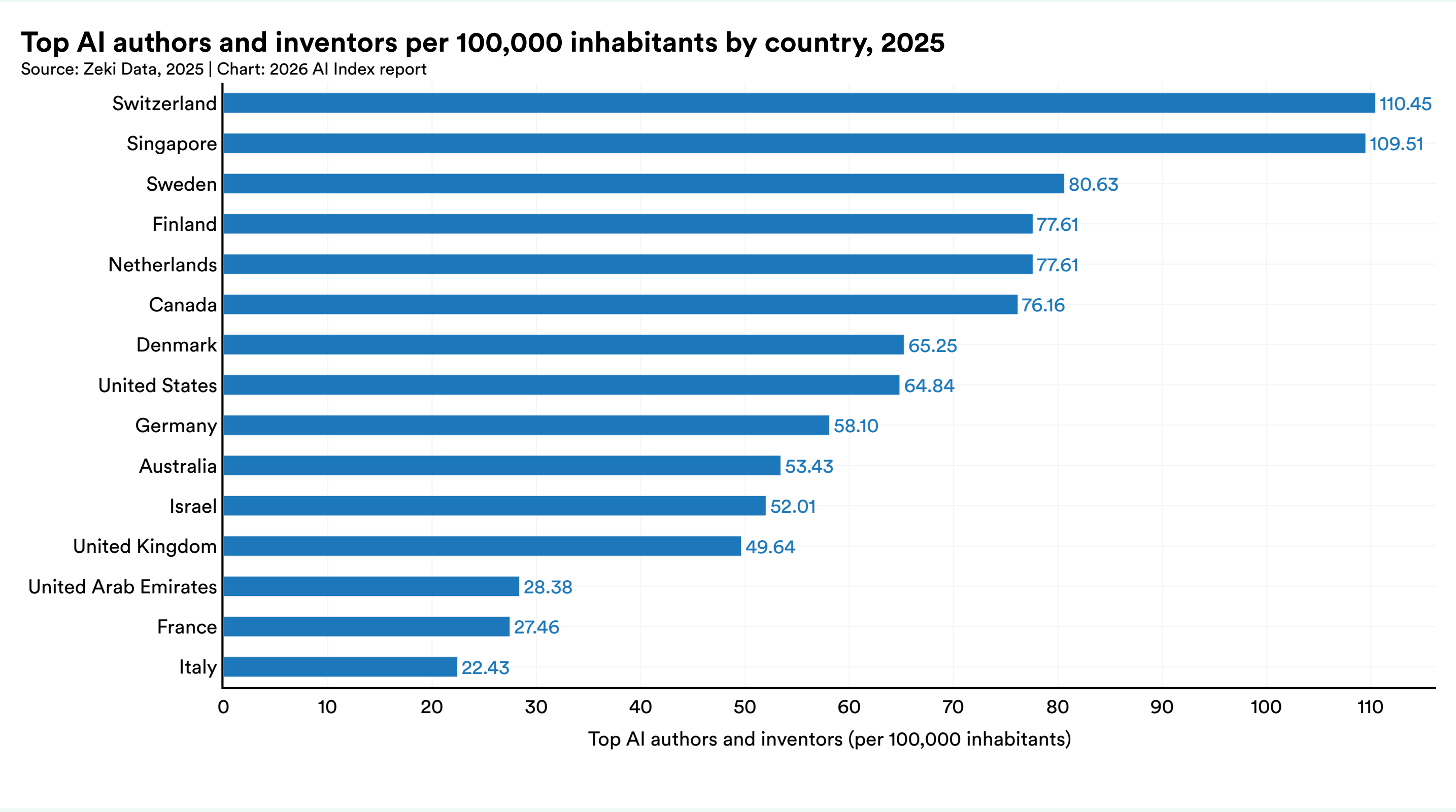

Figure 1.8.2

## By Education Level

The educational profile of top AI authors and inventors varies by country, though in most of the countries, PhD holders and those with master's degrees together account for the majority in 2025 (Figure 1.8.3). The United Kingdom (51.1%) and Australia (50.5%) have the highest share of PhD holders, followed by Switzerland (43.6%), South Korea (42.5%), and the United States (42%). India and Brazil show a more varied distribution, with comparatively lower shares of PhD holders and a wider spread across other degree levels.

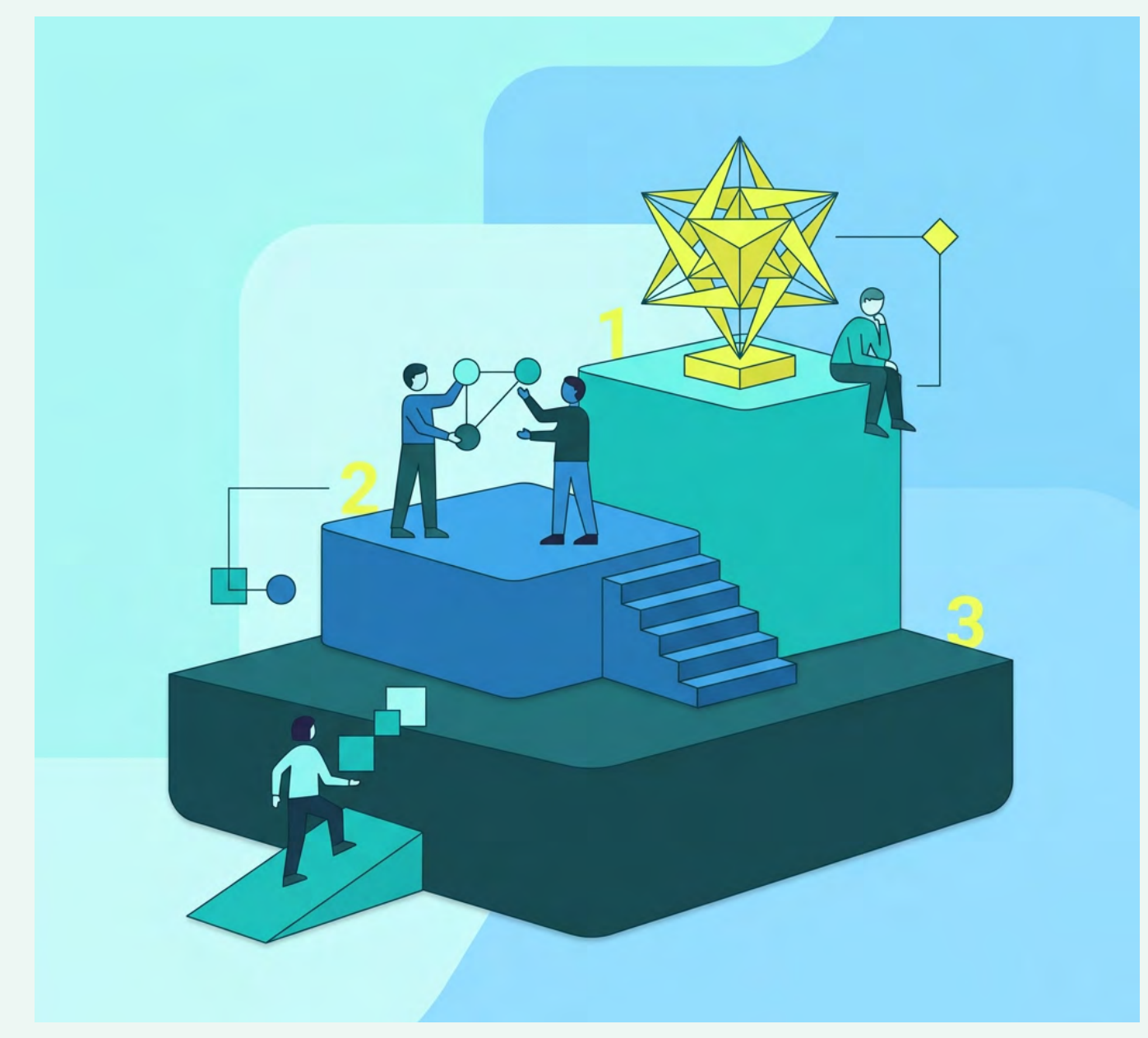

**Percentage of top AI authors and inventors by education level and country, 2010–25**
Source: Zeki Data, 2025 | Chart: 2026 AI Index report

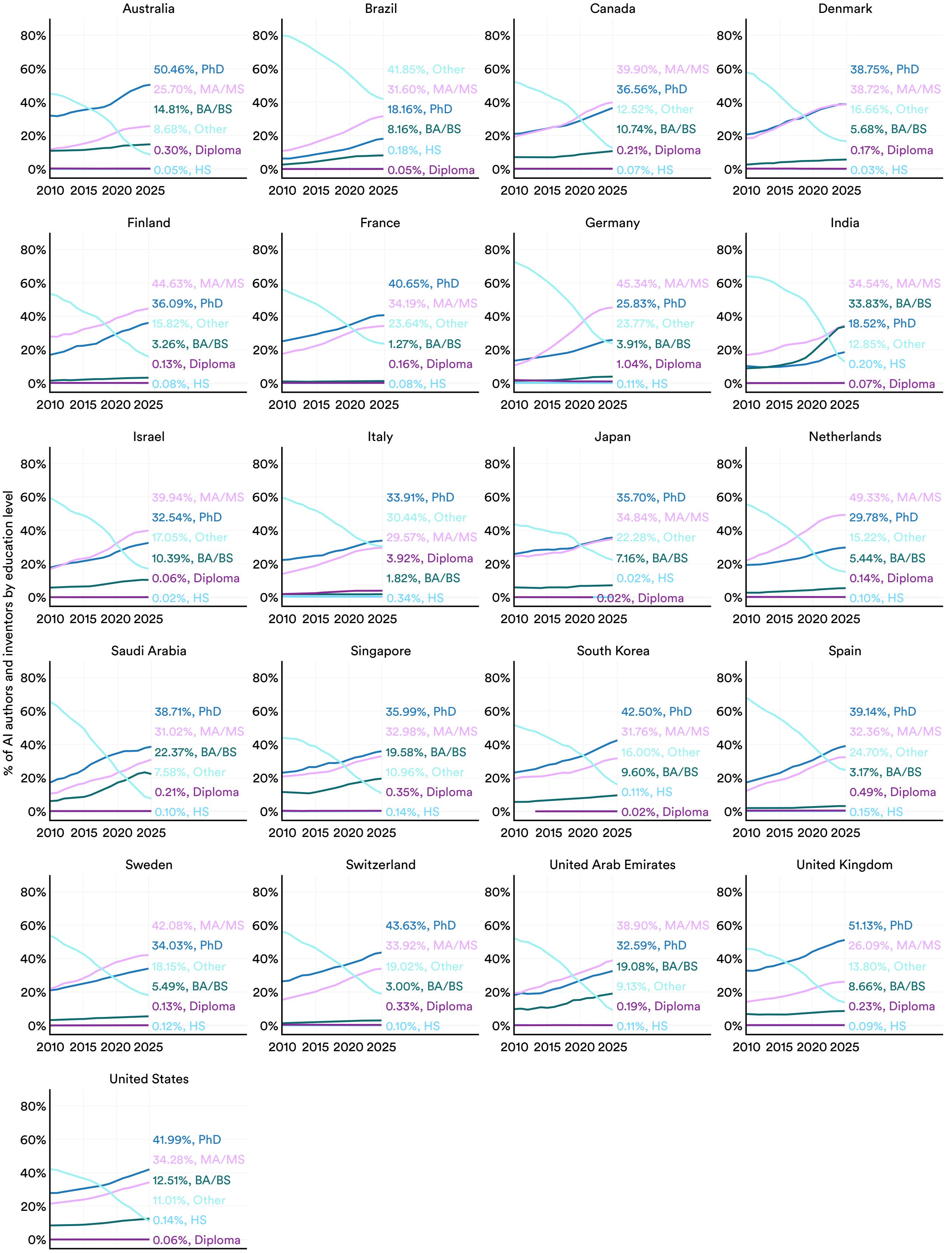

Figure 1.8.3

## By Gender

The gender gap among AI authors and inventors is visible across all countries, with men making up the majority in all cases, though the size of the gap varies (Figure 1.8.4). In Brazil, South Korea, and Japan, more than 80% of identified AI talent is male. Female representation is somewhat higher in Saudi Arabia (32.3%), Australia (30.1%), Canada (29.6%), and Italy (29.5%), but no country comes close to parity. More significantly, in almost every country, the male-female ratio has remained flat from 2010 to 2025. Even with the growth in AI talent overall, there has been no meaningful progress on gender balance. Chapter 7 (Education) describes a similar pattern in AI-related degree attainment, where women remain underrepresented across all levels.

**Top AI authors and inventors (% of total) by gender, 2010–25**
Source: Zeki Data, 2025 | Chart: 2026 AI Index report

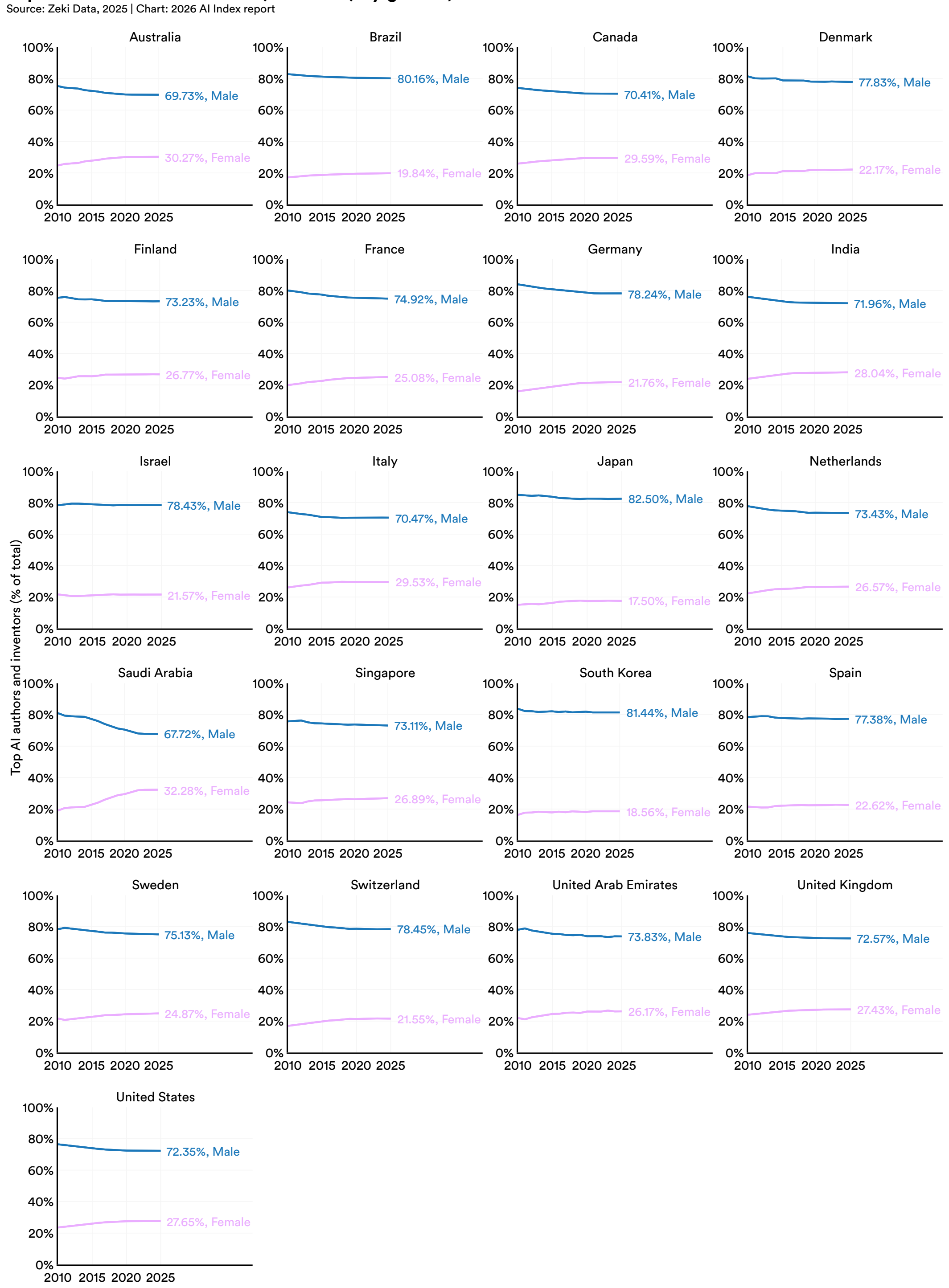

Figure 1.8.4

## By Specialization

AI authors and inventors are distributed across a range of specialization areas, though each country shows its own emphasis (Figure 1.8.5). Healthcare and bioinformatics, computer vision and image processing, and software engineering are among the most common areas globally, accounting for 10% or more of the pool in several countries. A few country-level patterns connect to findings discussed earlier in this chapter. South Korea, for example, has the highest share of talent in hardware, VLSI, and IoT (20%), consistent with its role in the semiconductor supply chain described in Section 1.3. Brazil has the highest share of software engineering talent (18%), while Saudi Arabia leads in security, privacy, and cryptography (15%).

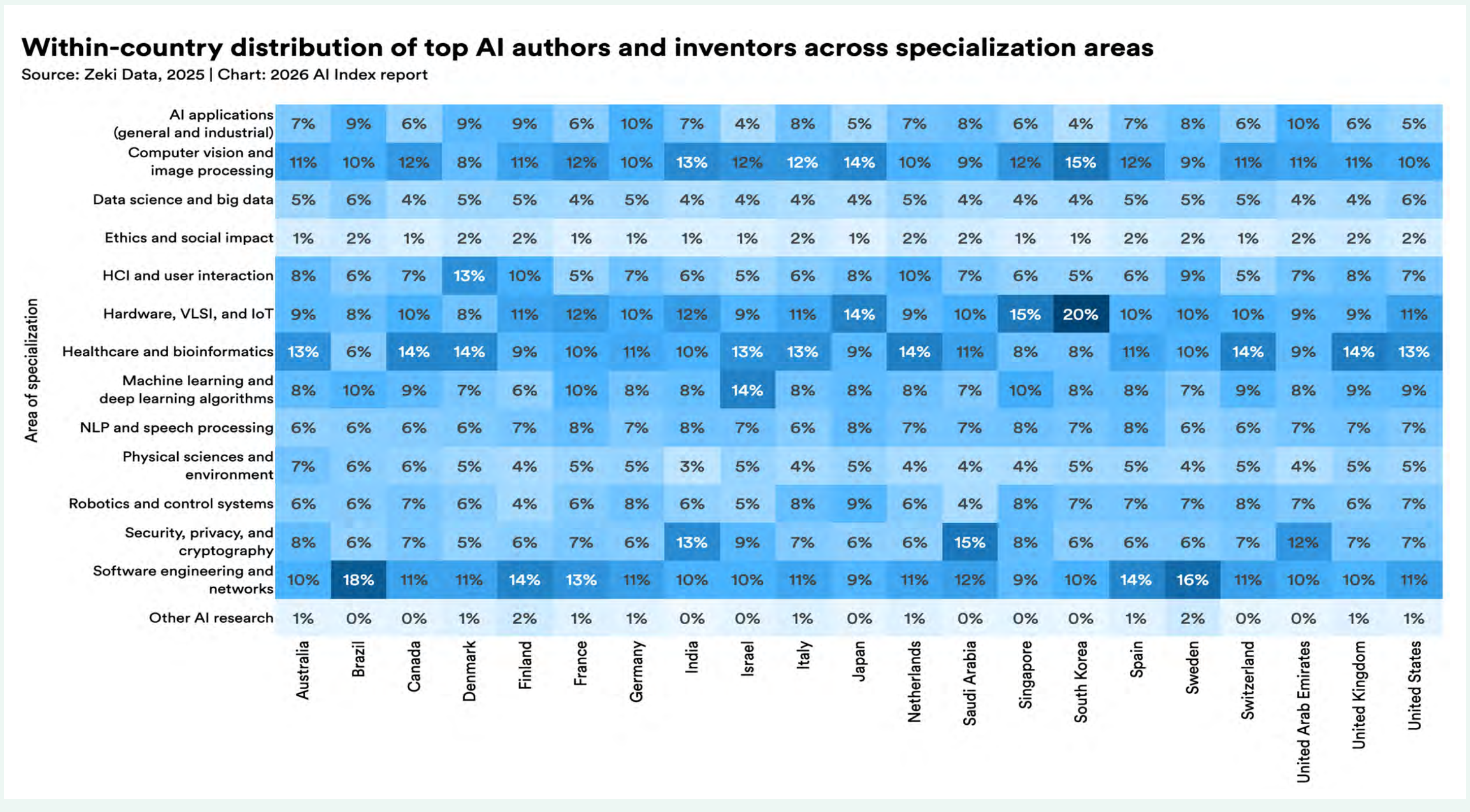

Figure 1.8.5

## Mobility

Mobility is measured through net flow, which is the difference between the number of AI authors and inventors who move to or out of their respective countries (Figure 1.8.6). The United States has remained net positive since 2020, meaning it attracts more talent than it loses, though the magnitude has declined from a peak of 324.6 in 2022 to 26.0 in 2025. Most other countries operate on a smaller scale. Saudi Arabia (3.1) and Denmark (2.1) were among the few with positive net flow in 2025.

Canada, which showed strong inflow around 2020, declined to -7.1 by 2025. Germany also showed negative net flow at -2.4, while India had the largest net outflows at -16.9 in 2025. These flows are relevant in the context of other factors, including immigration policy and geographic distribution of investment and employment, discussed further in Chapter 4's section on labor markets.

## Net flow of top AI authors and inventors by country, 2010–25

Source: Zeki Data, 2025 | Chart: 2026 AI Index report

Net flow top AI authors and inventors (12-month rolling avg.)
Australia -1.75
Brazil -3.58
Canada -7.08
Denmark 2.08
Finland -1.08
France 0.08
Germany -2.42
India* -16.92
Israel -1.92
Italy 0.00
Japan 0.00
Netherlands -1.83
Saudi Arabia 3.08
Singapore -2.58
South Korea -1.75
Spain 2.58
Sweden -0.58
Switzerland 0.25
United Arab Emirates 2.58
United Kingdom 5.17
United States* 26.00

Figure 1.8.6[44]

44 Asterisks indicate that a country's y-axis label is scaled differently than the y-axis label for the other countries.

# 2

# Technical Performance

## Overview

AI models improved rapidly in 2025, with benchmark scores rising across language, reasoning, coding, and math. However, evaluations are being outpaced by the progress they were built to measure, and benchmarks face growing questions about their reliability. Even with those limitations, a clear pattern emerges: the gap between top models is shrinking. This narrowing extends geographically, as the distance between top U.S. and Chinese models has closed almost completely. With capability no longer a clear differentiator, competitive pressure is shifting toward cost, reliability, and real-world usefulness. In professional domains, evaluations in tax, legal reasoning, and corporate finance show stronger performance in some areas than others. The range of what AI systems can do is also expanding. AI agents are improving, but still fail roughly one in three attempts. Video generation models are no longer just producing realistic-looking content; some are beginning to learn how the physical world actually works, progress that could help bring AI into physical spaces. That transition is still early, as robots struggle in unstructured environments, though autonomous vehicles are a notable exception, having reached mass-scale deployment with promising early safety records. Overall, AI's technical advancement is a story of wonder and speed, faster than many of the evaluation, governance, and adoption frameworks discussed in later chapters.

## Contents

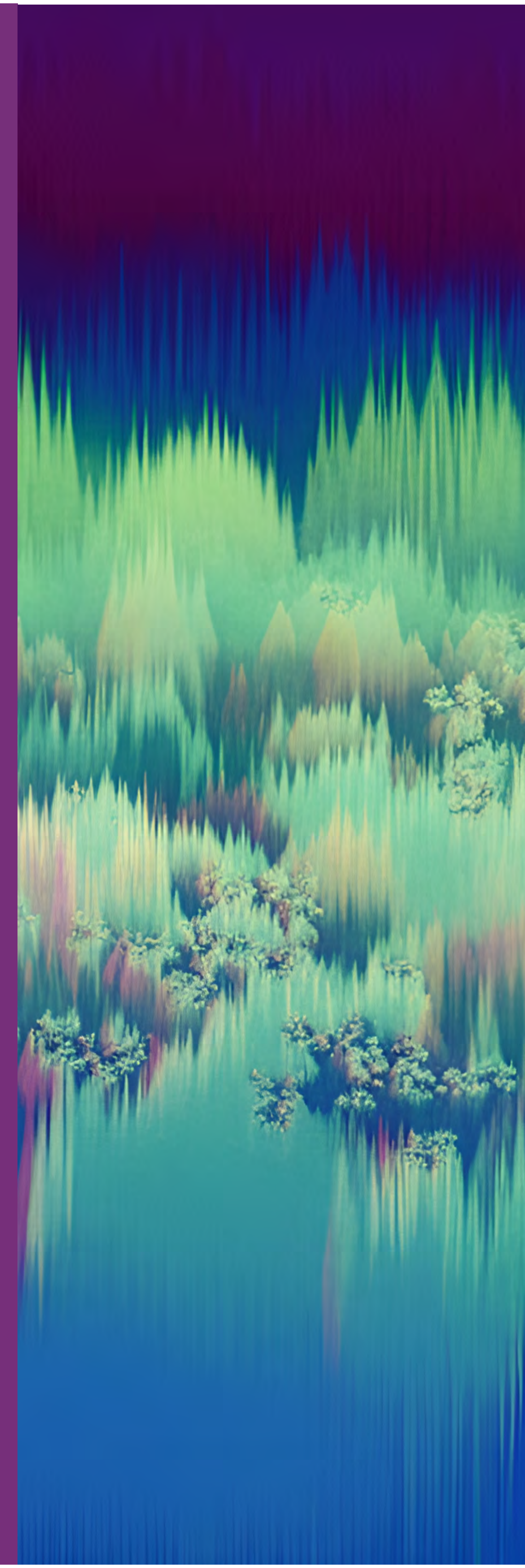

# Chapter Highlights

1. **AI capability is outpacing the benchmarks designed to measure it, and surpassing human-level performance.** Frontier models gained 30 percentage points in a single year on Humanity's Last Exam, a benchmark built to be hard for AI and favorable to human experts. Evaluations intended to be challenging for years are saturated in months, compressing the window in which benchmarks remain useful for tracking progress.

2. **Top model performance is converging, with 4 companies now clustered within 25 Elo points (inspired by chess ratings) when rated against one another by human voting in the Arena Leaderboard and benchmark.** As of March 2026, Anthropic (1,503), xAI (1,495), Google (1,494), OpenAI (1,481), Alibaba (1,449), and DeepSeek (1,424) all occupy the top tier of the Arena Elo ratings, shifting competitive pressure toward cost, reliability, and domain-specific performance.

3. **The open model performance gap reopened in 2025 after briefly closing in 2024.** As of March 2026, the top closed model leads the top open model by 3.3%, up from 0.5% in August 2024. Six of the top ten models on the Arena Leaderboard are now closed.

4. **The U.S.-China AI model performance gap has effectively closed.** U.S. and Chinese models have traded places at the top of performance rankings multiple times since early 2025. In February 2025, DeepSeek-R1 briefly matched the top U.S. model. As of March 2026, the top U.S. model leads by 2.7%, with a gap that fluctuated over the past year while remaining in the single digits.

5. **The benchmarks used to measure AI progress face growing reliability and gaming concerns, with error rates up to 42% on widely used evaluations.** A review found invalid question rates ranging from 2% on MMLU Math to 42% on GSM8K. Separate research suggests that Arena leaderboard standing may partly reflect adaptation to the platform rather than general capability.

6. **Video generation models are starting to capture how objects behave.** Google DeepMind's Veo 3, tested across more than 18,000 generated videos, demonstrated abilities like simulating buoyancy and solving mazes without being trained on those tasks.

7. **AI models can win a gold medal at the International Mathematical Olympiad but still can't reliably tell time, illustrating what researchers call jagged intelligence.** Gemini Deep Think scored 35 points (gold) at the 2025 IMO, working end to end in natural language within the 4.5-hour time limit, up from the 28-point silver achieved in 2024. On ClockBench, the top model read analog clocks correctly 50.6% of the time, compared with 90.1% for humans.

8. **AI models are expanding into professional domains, showing performance ranging from 60 to 90% in evaluations in tax, mortgage processing, corporate finance, and legal reasoning.** The performance of the top 15 models is separated by as little as 3 percentage points in each benchmark. These kinds of domains where high competency and reliability are required remain a great challenge for AI models.

9 **AI agents advanced from answering questions to completing tasks in 2025, though they still fail roughly one in three attempts on structured benchmarks.** On OSWorld, which tests agents on real computer tasks across operating systems, accuracy rose from roughly 12% to 66.3%, within 6 percentage points of human performance.

10 **Robots still fail at most household tasks, even as they excel in controlled environments.** Robots succeed in only 12% of real household tasks, highlighting how far AI is from mastering the physical world. On RLBench, robotic manipulation in software-based simulations has reached 89.4% success, but the gap between predictable lab settings and unpredictable household environments is wide.

11 **Autonomous vehicles reached mass-scale deployment in 2025.** Waymo reached approximately 450,000 weekly trips across five U.S. cities. In China, Apollo Go completed 11 million fully driverless rides, a 175% year-over-year increase. European operators are active but comparable deployment data is not publicly available, limiting the global picture. Deployments so far are in areas with generally favorable weather and humans are available off-site to take over when necessary.

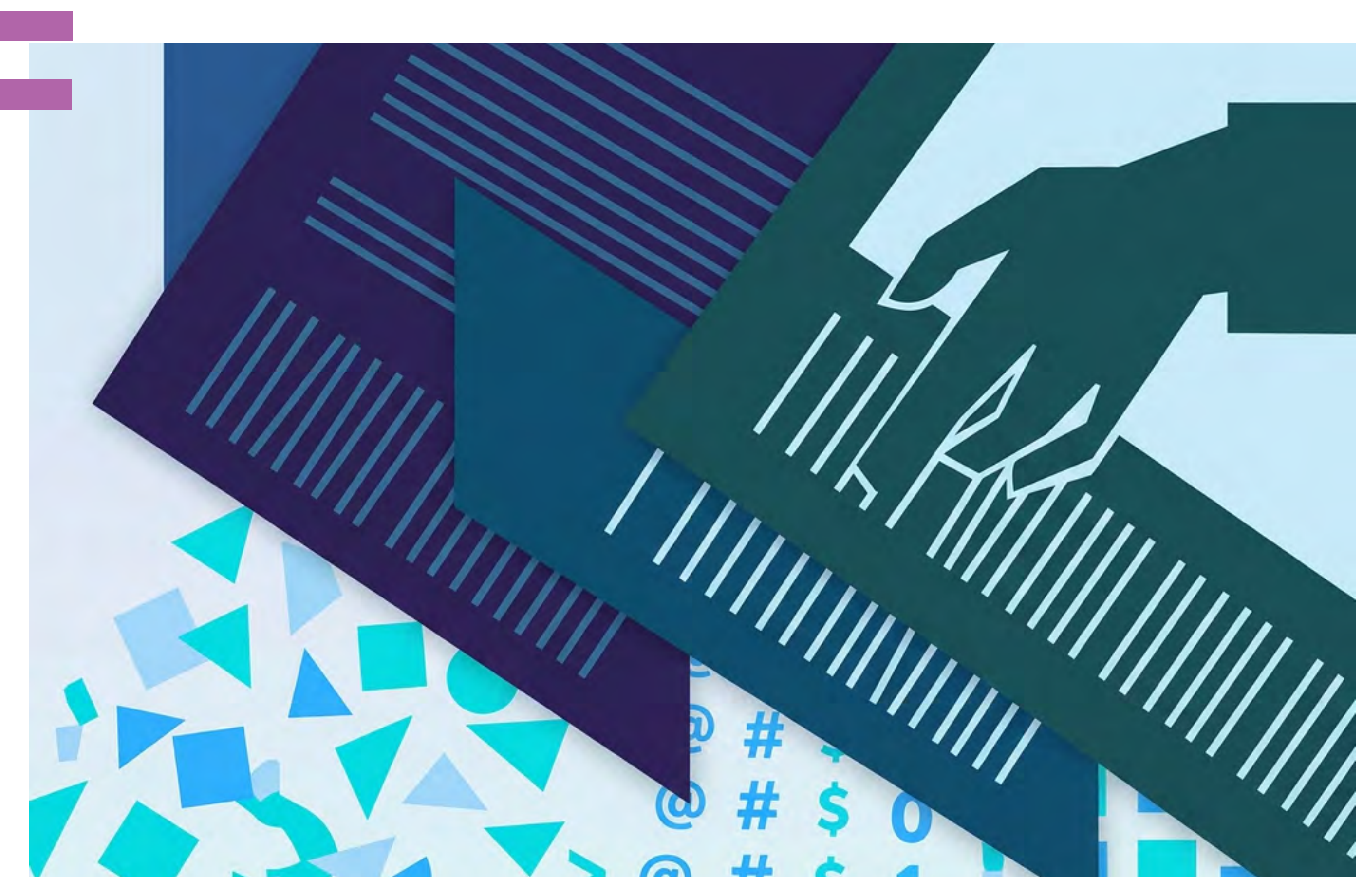

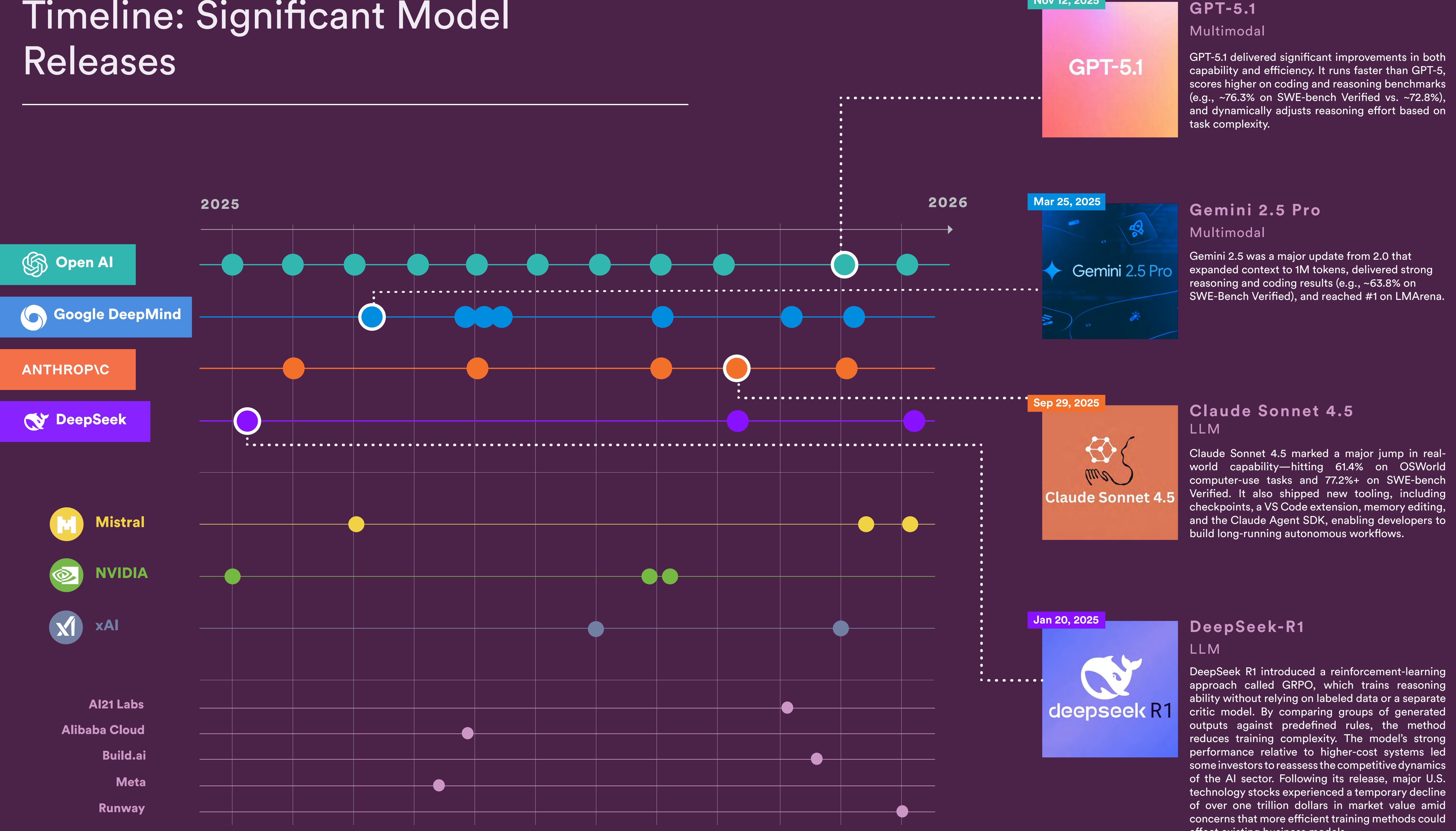

Timeline: Significant Model Releases
2025
2026
Open AI
Google DeepMind
ANTHROP\C
DeepSeek
Mistral
NVIDIA
xAI
AI21 Labs
Alibaba Cloud
Build.ai
Meta
Runway
Nov 12, 2025
GPT-5.1
GPT-5.1
Multimodal
GPT-5.1 delivered significant improvements in both capability and efficiency. It runs faster than GPT-5, scores higher on coding and reasoning benchmarks (e.g., ~76.3% on SWE-bench Verified vs. ~72.8%), and dynamically adjusts reasoning effort based on task complexity.
Mar 25, 2025
Gemini 2.5 Pro
Gemini 2.5 Pro
Multimodal
Gemini 2.5 was a major update from 2.0 that expanded context to 1M tokens, delivered strong reasoning and coding results (e.g., ~63.8% on SWE-Bench Verified), and reached #1 on LMArena.
Sep 29, 2025
Claude Sonnet 4.5
Claude Sonnet 4.5
LLM
Claude Sonnet 4.5 marked a major jump in real-world capability—hitting 61.4% on OSWorld computer-use tasks and 77.2%+ on SWE-bench Verified. It also shipped new tooling, including checkpoints, a VS Code extension, memory editing, and the Claude Agent SDK, enabling developers to build long-running autonomous workflows.
Jan 20, 2025
deepseek R1
DeepSeek-R1
LLM
DeepSeek R1 introduced a reinforcement-learning approach called GRPO, which trains reasoning ability without relying on labeled data or a separate critic model. By comparing groups of generated outputs against predefined rules, the method reduces training complexity. The model's strong performance relative to higher-cost systems led some investors to reassess the competitive dynamics of the AI sector. Following its release, major U.S. technology stocks experienced a temporary decline of over one trillion dollars in market value amid concerns that more efficient training methods could affect existing business models.

# 2.1 Overall Performance Trends

This section examines patterns in AI performance, from the pace at which models are reaching human-level baselines to how competition among leading models and countries has narrowed. It also assesses where the tools used to measure this progress are themselves falling short. To enable comparison across diverse valuation tasks, performance metrics are scaled to a common reference point. The scaling methodology, developed by the AI Index team, calibrates each benchmark so that the best-performing model in a given year is measured as a percentage of the established human baseline for that task. For example, using this approach, a value of 105% indicates that a model performs 5% better than the human baseline. The benchmarks included in this analysis represent tasks that can be structurally evaluated. It may not fully capture the breadth of capabilities required for real-world AI deployment. The Benchmarking AI subsection later in this section explores these limitations in detail.

## Technical Performance Benchmarks vs. Human Performance

AI performance continued to improve across a broad set of benchmark categories in 2025, with some of the largest gains appearing on tasks that were well below human baseline performance just a few years ago (Figure 2.1.1). Frontier systems now meet or exceed established human performance levels on long-running benchmarks, including ImageNet, SuperGLUE, and MMLU. Since last year's report, several benchmarks designed to test more advanced reasoning have reached or approached the human benchmark, including PhD-level science questions (GPQA Diamond), multimodal reasoning (MMMU), and mathematical reasoning (AIME). Models are still performing below the baseline in the areas of autonomous software engineering (SWE-bench Verified) and agent-based multimodal computer use (OSWorld), but the pace of improvement is rapidly accelerating. On SWE-bench Verified, for example, performance rose from approximately 60% in 2024 to close to 100% in 2025.

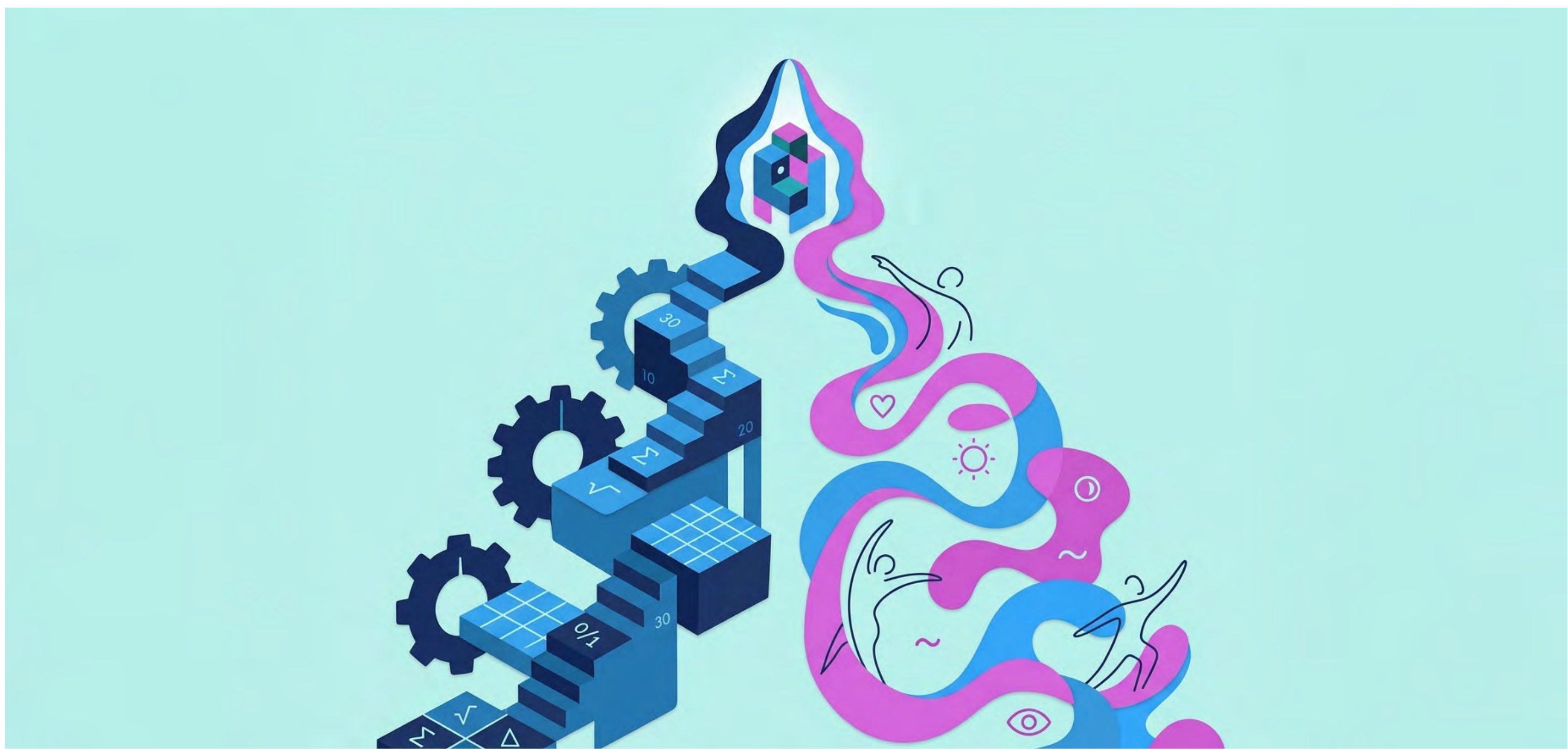

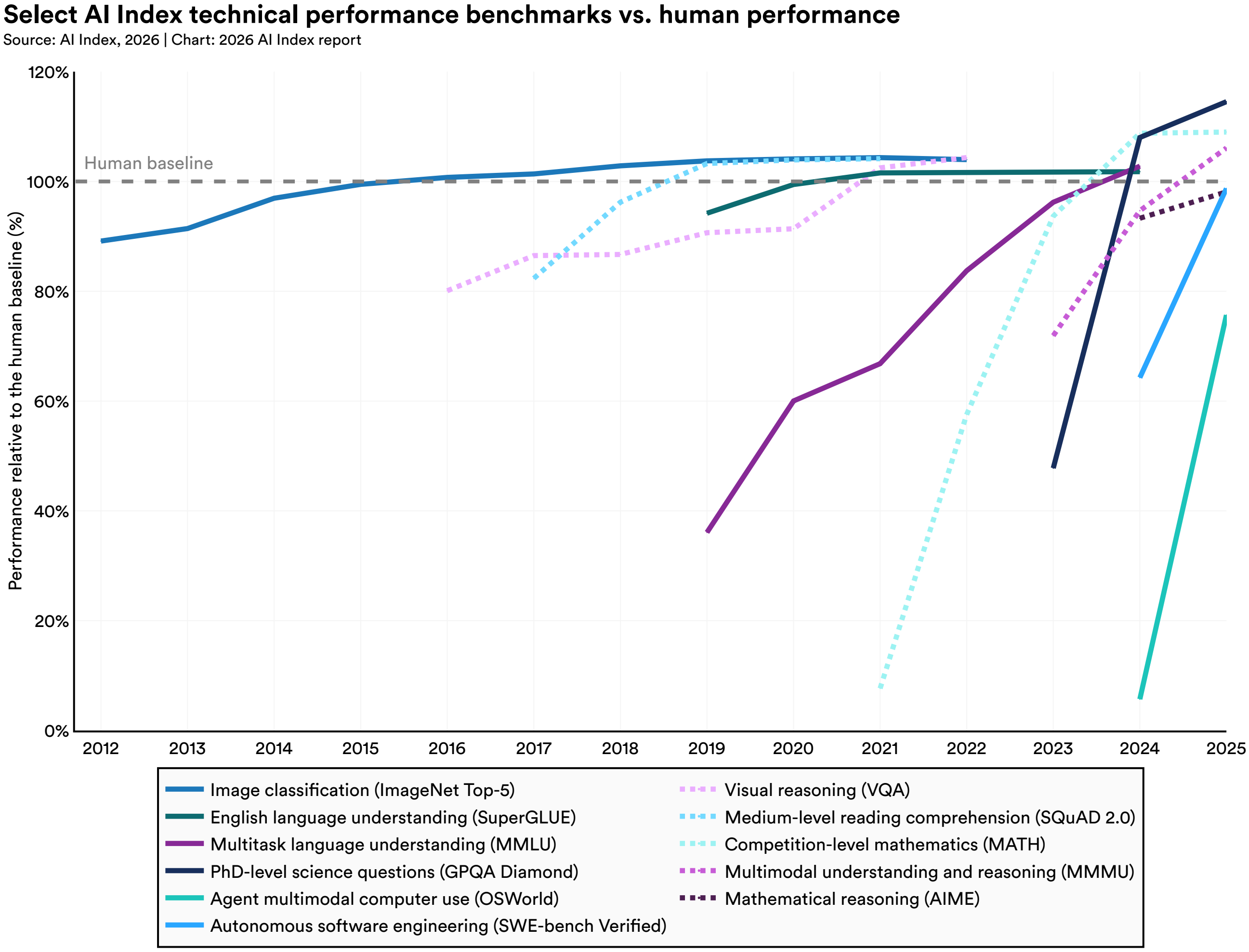
Select AI Index technical performance benchmarks vs. human performance
Source: AI Index, 2026 | Chart: 2026 AI Index report

Figure 2.1.1[1]

## Closed- vs. Open-Weight Models

The performance gap between leading closed-weight and open-weight models has fluctuated over the past three years, with open-weight systems closing in and then falling behind as new proprietary models are released (Figure 2.1.2). In May 2023, the leading closed-weight model (GPT-4-0314) outperformed the top open-weight model (Vicuna-13B) by 174 points (15.2%) on the Arena Leaderboard. Stronger open-weight releases, including Mixtral, WizardLM, and Llama-3.1-405B, narrowed the gap to just 7 points (0.5%) by August 2024. Over the past year, that trend reversed with the arrival of new closed-weight frontier systems such as o1-preview and Gemini 2.5 Pro. As of March 2026, the top closed-weight model, Claude Opus 4.6 (1,503), led the top open-weight model GLM-5 (1,454) by 49 points (3.4%). While closed-weight models still lead, open-weight models are far more competitive than they were a few years ago.

1 In Figure 2.1.1, the values are scaled to establish a standard metric for comparing different benchmarks. The scaling function is calibrated such that the performance of the best model for each year is measured as a percentage of the human baseline for a given task. A value of 105% indicates, for example, that a model performs 5% better than the human baseline.

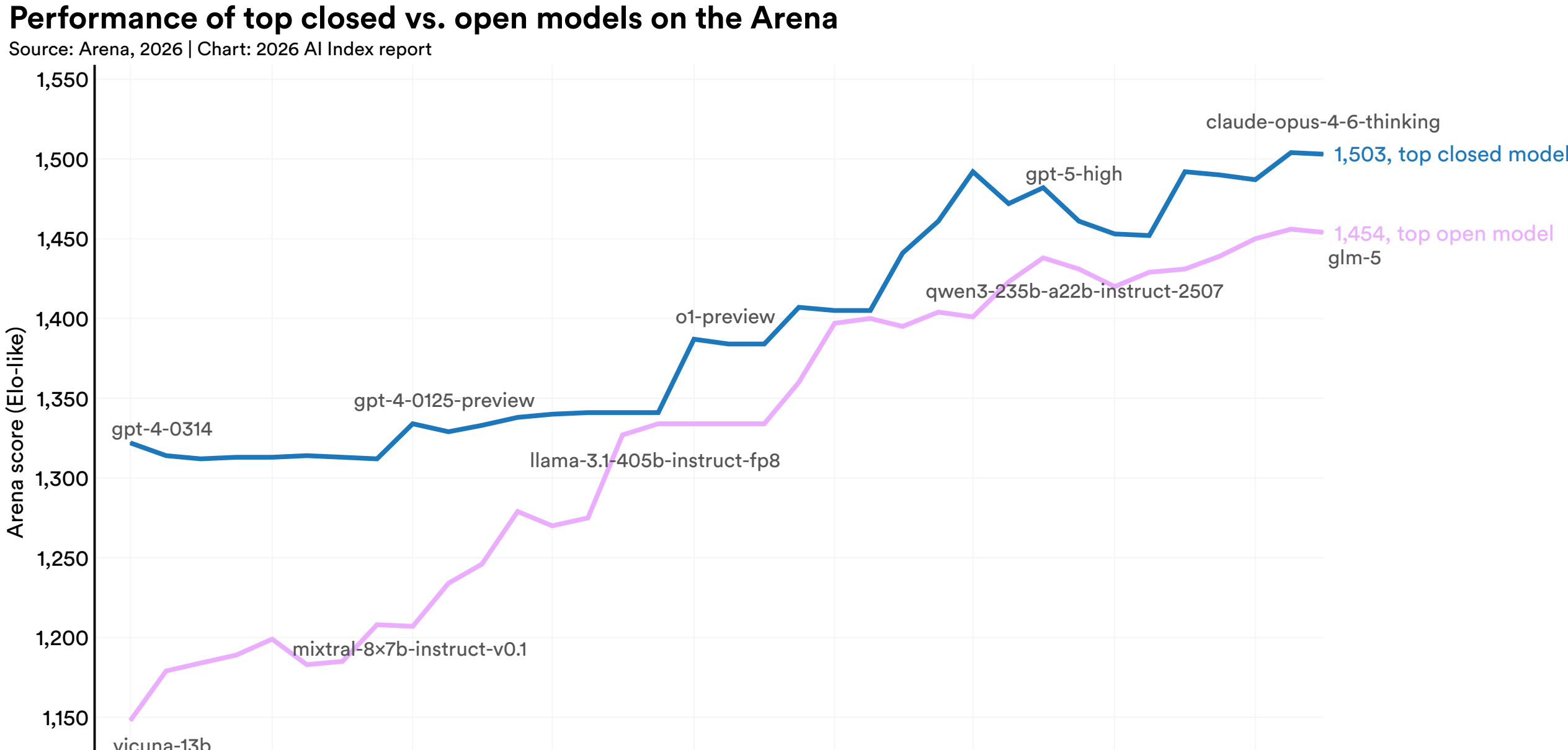

Figure 2.1.2[2]

## US vs. China Technical Performance

The United States' substantial lead in 2023 shrank considerably by early 2025, and the performance gap has remained narrow since then (Figure 2.1.3). In February 2025, DeepSeek-R1 (1,400) trailed the leading U.S. model (o1-2024-12-17, 1,405) by just 5 Arena points (0.4%). As of March 2026, the top U.S. model (Claude Opus 4.6, 1,503) led the top Chinese model (Dola-Seed-2.0 Preview, 1,464) by 39 points (2.7%). Over the past year, the gap has fluctuated between near parity and low single digits. This convergence is particularly notable because it has emerged from two distinct development environments and institutional contexts, including the research dynamics examined in Chapter 1 and the investment patterns discussed in Chapter 4.

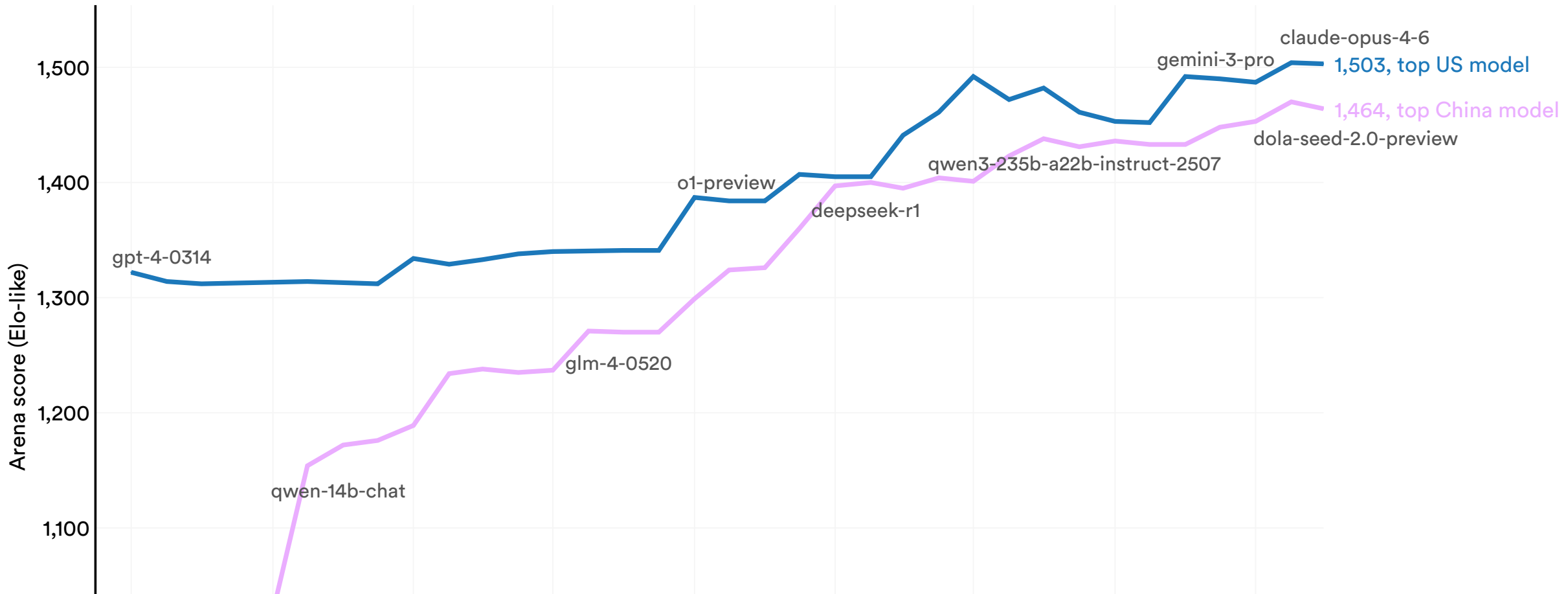
Performance of top United States vs. Chinese models on the Arena
Source: Arena, 2026 | Chart: 2026 AI Index report

Figure 2.1.3[3]

2 Source: the Arena historical leaderboard (Public, Style Control On), exported in March 2026.

3 Source: the Arena historical leaderboard (Public, Style Control On), exported in March 2026.

## Model Performance Converges at the Frontier

Frontier models became even more tightly clustered over the past year, as several companies moved into a very narrow performance band at the top of the Arena Leaderboard (Figure 2.1.4). In early 2023, OpenAI had a clear lead with its top model scoring 1,322 compared to Google's 1,117. This gap narrowed steadily through 2024 as Google, Anthropic, and others released stronger models. By February 2025, DeepSeek had briefly matched and surpassed the top U.S. systems on Arena. In last year's report, the top four models spanned roughly 97 points and, as of March 2026, the top four models are separated by fewer than 25 points. Anthropic leads at 1,503, followed closely by xAI (1,495), Google (1,494), and OpenAI (1,481). DeepSeek (1,424) and Alibaba (1,449) trail only modestly. Meta's Arena performance has flattened since early 2025, reflecting a slowdown in competitive releases, though newer models could be in the pipeline for 2026. As leading models become harder to distinguish on benchmark performance, factors such as cost, latency, reliability, and domain-specific optimization may play a greater role in user adoption.

**Performance of top models on the Arena by select providers**
Source: Arena, 2026 | Chart: 2026 AI Index report

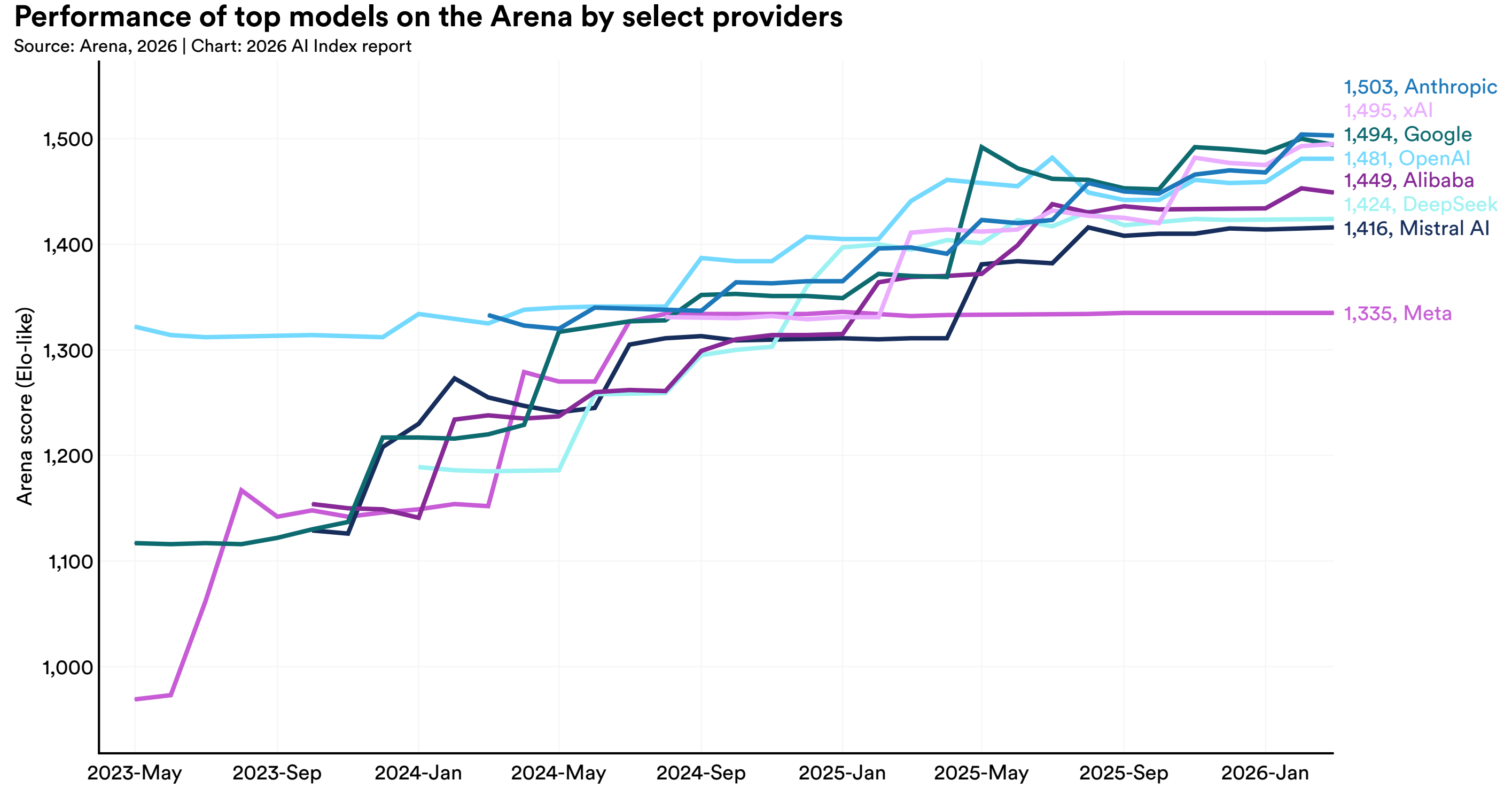

Figure 2.1.4[4]

## Benchmarking AI

Benchmarks still anchor much of how AI's technical progress is measured, but their limitations are more visible. Since last year's report, the AI Index has expanded its analysis to examine where benchmarks remain useful, and where they fall short.

Several challenges highlighted in previous editions of this report persist. Benchmark saturation, where models reach scores so high that a test can no longer distinguish between them, remains a concern. Tests designed to be harder often remain useful for only a few years before systems surpass them. As Chapter 1 documents, reporting discrepancies continue, and the most capable modern models are now among

4 Source: the Arena historical leaderboard (Public, Style Control On), exported in March 2026.

the least transparent. The growing opacity and nonstandard prompting techniques make model-to-model comparisons unreliable, and third-party evaluations have documented cases where models perform more poorly in independent testing compared to developer-reported results. In addition, contamination—when models are exposed to test set data during training—can lead to falsely inflated scores. In 2025, Meta faced criticism that its Llama 4 model was optimized using specialized variants to improve leaderboard rankings and may have trained on benchmark test data, though the company disputed these claims. Additionally, audits of widely used benchmarks revealed that many remain poorly constructed, with inadequate documentation, no reporting of statistical significance, and a lack of replication scripts. Even when benchmark scores are technically valid, strong benchmark performance does not always translate to real-world utility.

Last year's report also highlighted how difficult it is to benchmark more complex, interactive forms of intelligence, which matter even more for current AI systems. Even though many benchmarks for multiagent coordination, human–AI interaction, tool-using agents, and physical-world robotics have been proposed (e.g., for robotic manipulation, embodied reasoning, and agentic tasks), they remain underdeveloped. These domains are inherently harder to standardize as physical tasks involve unpredictable environments, diverse hardware, and a range of valid approaches that resist repeatable scoring. Later sections of this chapter report on several of these benchmarks in detail.

The benchmarking landscape has seen several developments that extend beyond these recurring concerns. First, there is a growing case for evaluations that measure human–AI collaboration rather than AI performance in isolation. Most widely used benchmarks test systems without human involvement, even though many real deployments involve people supervising, steering, and integrating AI outputs. Recent work argues that the field should adopt centaur evaluations, assessments in which humans and AI jointly solve tasks, because these better reflect actual use and allow measurement of human-centered qualities like interpretability and helpfulness that conventional benchmarks ignore.

Second, new methods have emerged to address the invalid benchmark questions. A review by Stanford researchers identified the proportion of invalid questions across nine widely used benchmarks, with error rates ranging from 2% on MMLU Math to 42% on GSM8K (Figure 2.1.5). Truong et al., 2025, introduced a framework that uses statistical analysis of response patterns to flag problematic items for expert review, achieving up to 84% precision. Separately, Cheng et al., 2025, have proposed shifting toward "certificate-grade," peer-based evaluation frameworks that are community-governed, proctored systems with secure environments, continuously refreshed test items, and delayed result disclosure.

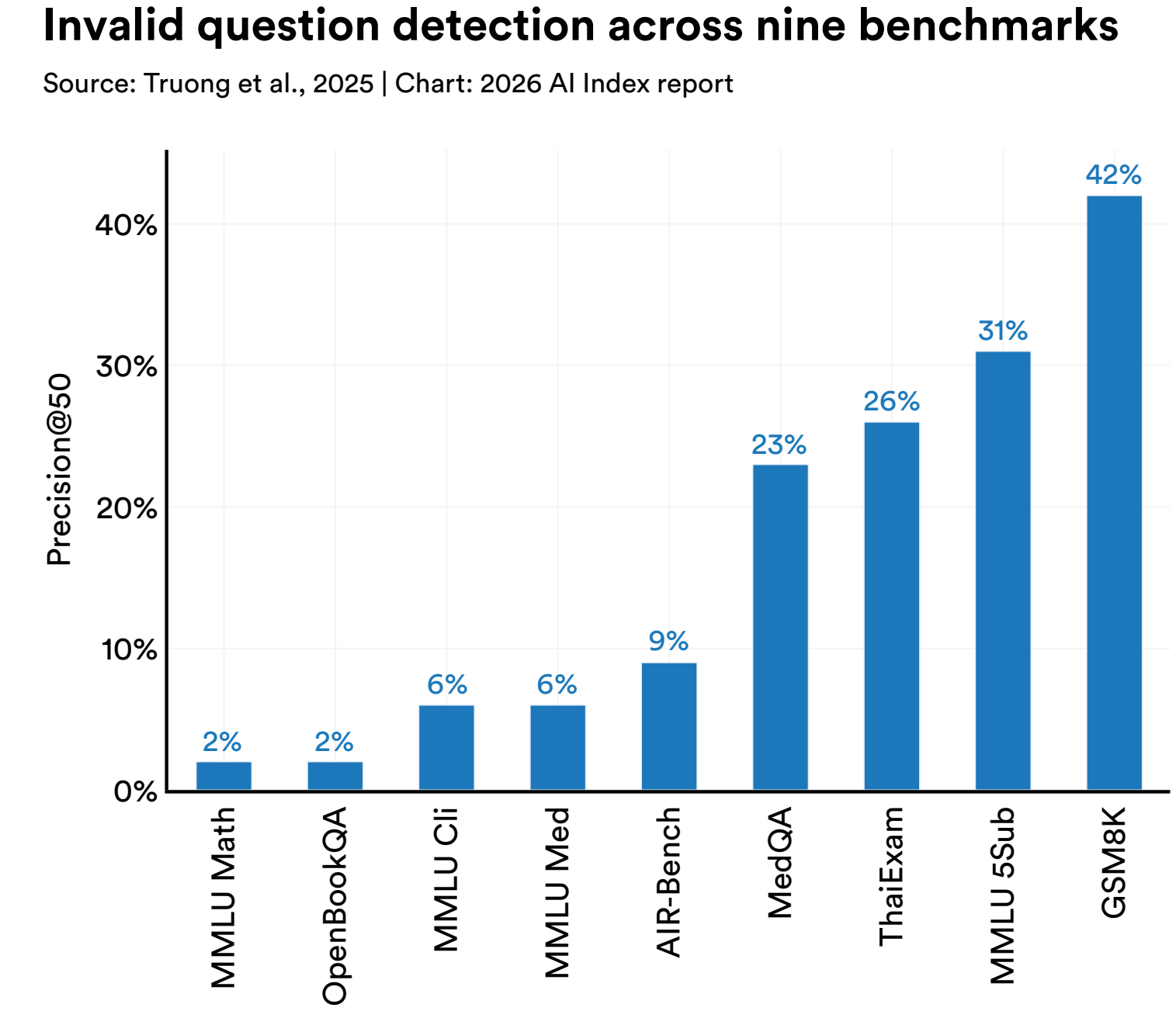
Invalid question detection across nine benchmarks
Source: Truong et al., 2025 | Chart: 2026 AI Index report

Figure 2.1.5

Third, questions have been raised about the reliability of popular public benchmarking platforms such as the Arena. A recent analysis (Singh et al., 2025) argues that platform dynamics could affect ranking accuracy. If providers are able to iterate on or swap model variants outside the public record, it introduces selection effects that make comparisons less straightforward. The study also points out data-access asymmetries and shows that additional Arena-style interaction data can improve performance on Arena-derived evaluations, suggesting that leaderboard standing may partly reflect adaptation to the platform rather than general capability alone.

Finally, while capability evaluations are widespread, assessments of social impacts remain fragmented and incomplete. Reul et al., 2025, found that developers' reporting of bias and environmental impact is often sparse and declining, while third-party researchers more rigorously assess harms such as harmful content and performance disparities. Because only developers can disclose key information about data, labor practices, and training infrastructure, current evaluation practices provide a strong picture of what models can do but a far weaker account of their societal consequences. Chapter 3 examines responsible AI evaluation in further detail.

In this chapter, the AI Index continues to report on benchmarks as key indicators of technical progress. Scores are sourced from leaderboards, public repositories, and company disclosures, including papers, blog posts, and product releases. The AI Index assumes that company-reported results are accurate. All scores reflect the state of the field as of early 2026; subsequent model releases may have surpassed these benchmarks.

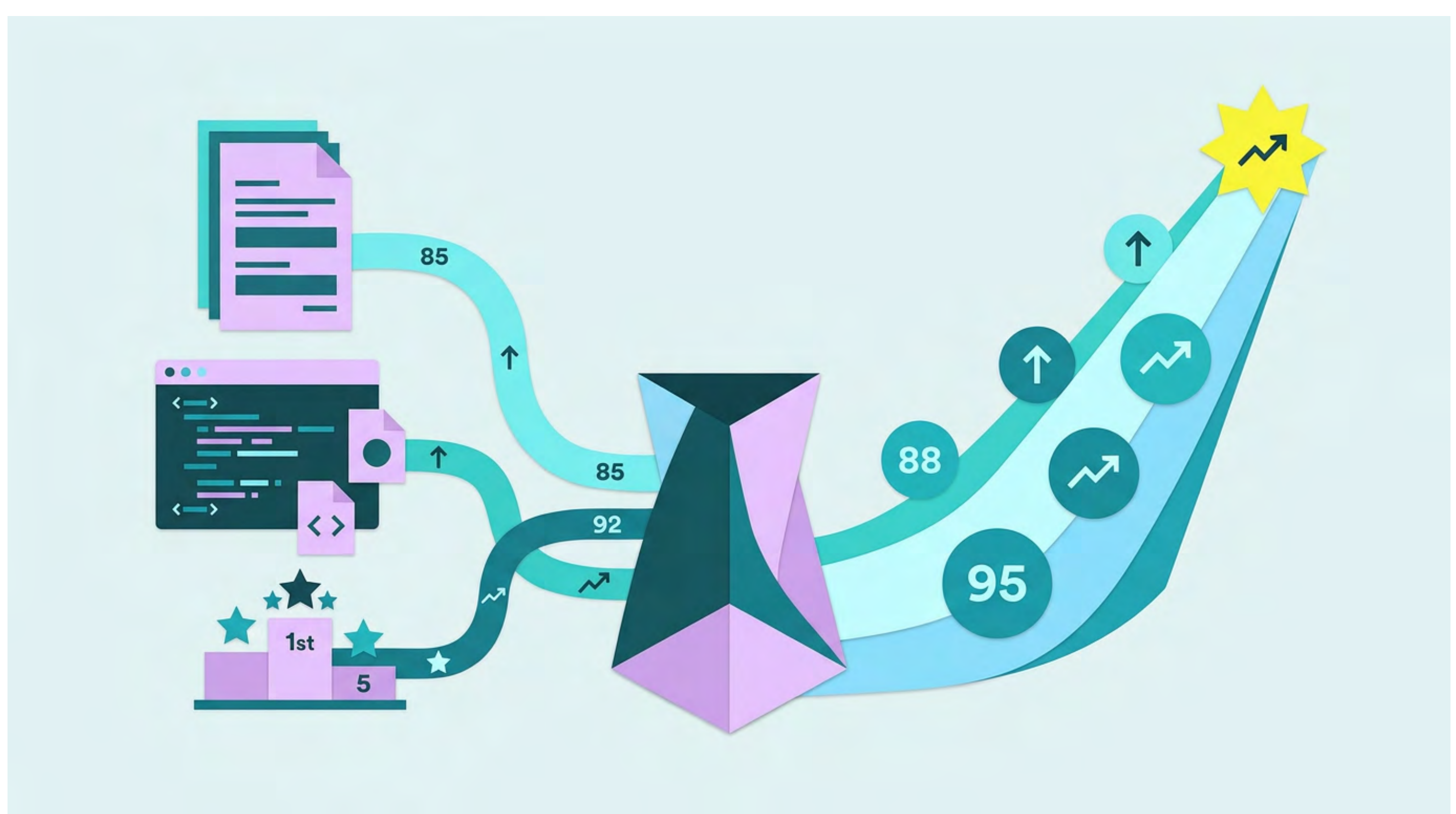

# 2.2 Language

Language understanding and generation continue to serve as foundational capabilities for modern AI systems. This section examines how models perform on tasks requiring comprehension of complex text, production of coherent responses, and execution of specialized language-based operations. The benchmarks also span general-purpose question answering to specific technical capabilities like function calling and text embedding.

## Understanding

Language understanding benchmarks measure how well models can comprehend and reason over text across a broad range of subjects, from the humanities to highly technical materials. As performance has improved, evaluation has shifted toward harder test sets that are less susceptible to familiarity or memorization. The goal is to track where models are improving rather than reaching the upper limits of current benchmarking tools.

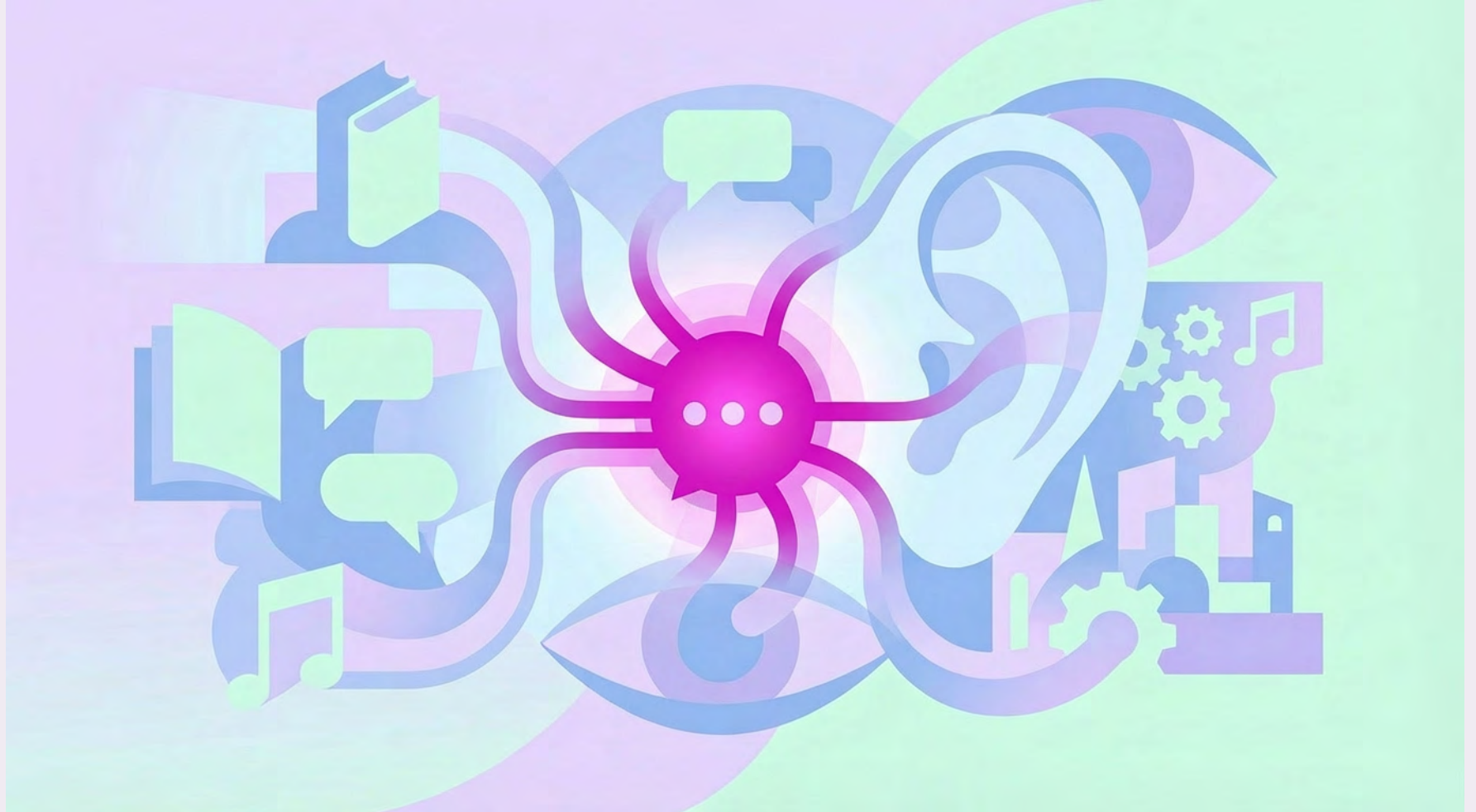

### MMLU: Massive Multitask Language Understanding

MMLU remains a widely cited measure of broad knowledge across disciplines. Introduced in 2024, the MMLU-Pro benchmark assesses performance with over 12,000 questions and a 10-option, multiple-choice format designed to better test reasoning. This expanded answer base has a measurable impact on model performance evaluation. Compared to the original MMLU benchmark, model accuracy on MMLU-Pro typically drops by 16%–33%, which provides better differentiation between top models. For example, GPT-4o and GPT-4-Turbo appeared to have a 1% gap on standard MMLU, but on MMLU-Pro the spread widens to 9%. The newer benchmark's design reduces prompt sensitivity and strengthens reasoning evaluation. Previously, MMLU showed around 4%–5% sensitivity to prompt variations versus an estimated 2% with MMLU-Pro. In addition, reasoning methods such as chain-of-thought tend to yield much better performance on MMLU-Pro than direct answer strategies.

As of early 2026, top model performance on MMLU-Pro is tightly clustered, with the leading 15 models all scoring above 87% (Figure 2.2.1). Google's Gemini-3.1-Pro leads at 91.2%, followed by Gemini-3-Pro (Thinking) at 90.1% and GPT-o1 at 89.3%. Models that employ thinking strategies tend to appear higher in the rankings, outperforming their standard counterparts, which are grouped in the 87%–88% range. The overall spread between the top-ranked and 15th-ranked model is just over 4 percentage points, illustrating how competitive the frontier has become on broad knowledge tasks. This tight clustering is also consistent with the convergence pattern described in Section 2.1.

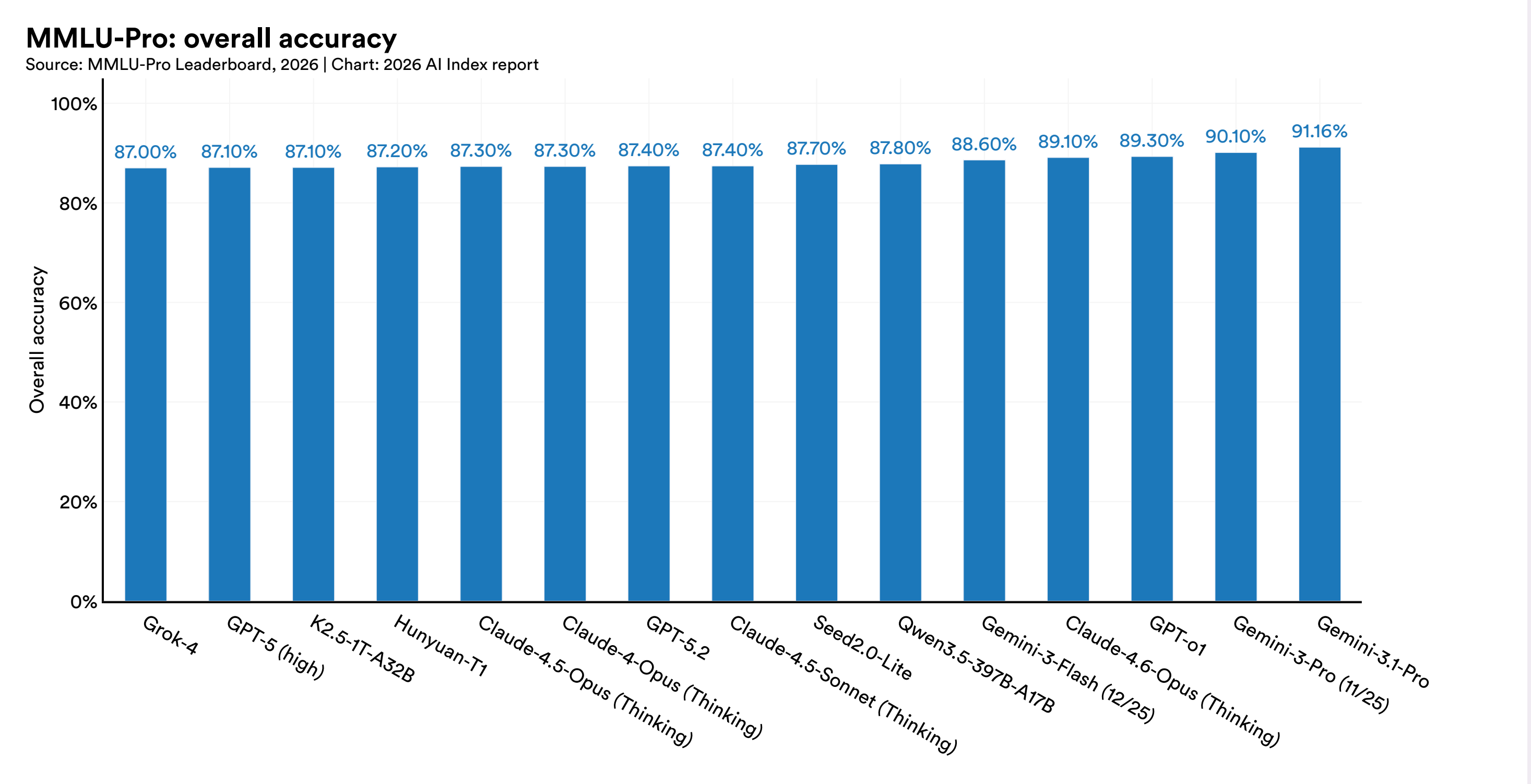

Figure 2.2.1[5]

## Generation

Generation benchmarks focus on the quality of model outputs, looking at clarity, helpfulness, instruction-following, and style. Unlike knowledge-style tests, these evaluations often depend on human judgment since some dimensions are subjective and dependent on both the prompt and the user. Preference-based tests help measure that subjectivity and are a useful complement to traditional benchmarks for tracking how models perform in real-world settings.

### Arena Leaderboard

The Arena (formerly LMArena) is an interactive platform with a community-driven ranking system that allows users to directly compare outputs of large language models (LLMs) on identical prompts and then vote on which they favor. Evaluations are blind to minimize bias toward particular model providers or architectures. By aggregating thousands of comparisons, the platform generates Elo ratings, a ranking system borrowed from chess. This approach emphasizes user experience and practical utility, capturing aspects of model quality that structured benchmarks cannot, including human judgment on real-world tasks.

The user-centered approach does have limitations, as preferences may not align with correctness and may not be fully representative of model use cases or contexts. Singh et al. (2025) highlight potential sources of bias, such as order bias, length bias, or style preferences, that are not correlated with output accuracy. As mentioned earlier, evaluations such as Arena can be a complementary view rather than an absolute score on model quality.

Elo ratings on the Text Arena are tightly clustered as of early 2026, with the top 15 models spanning roughly

5 Source: https://huggingface.co/spaces/TIGER-Lab/MMLU-Pro.

46 points (Figure 2.2.2). Claude-Opus-4-6-Thinking leads at approximately 1,510, followed closely by Gemini-3.1-Pro-Preview. The gap narrows further down the rankings, and confidence intervals overlap for many models. So, while Anthropic and Google models appear throughout the top ranks, no single model dominates the leaderboard.

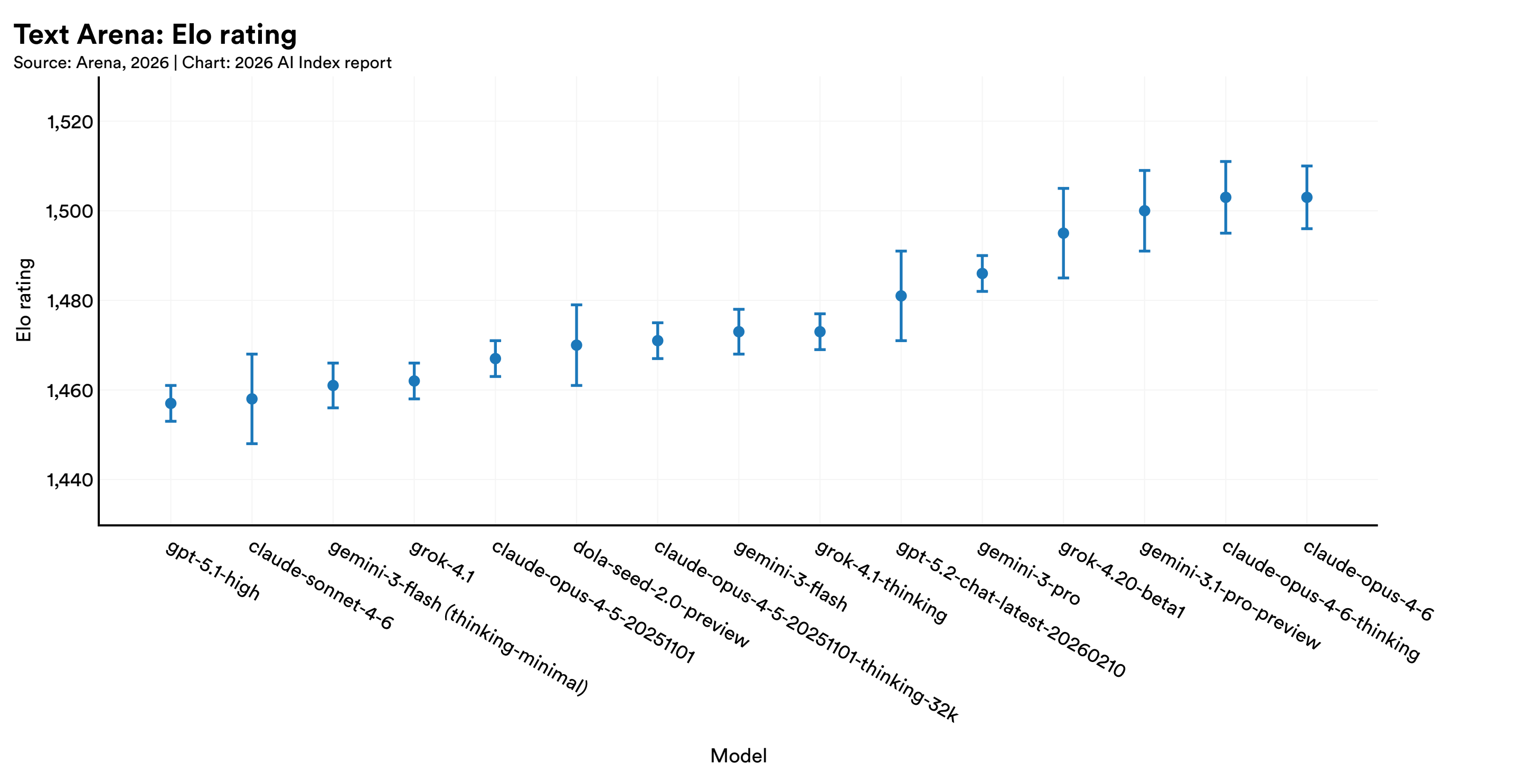

Figure 2.2.2[6]

## Specialized Language Tasks

Beyond general understanding and generation, language models need to handle tasks that make them usable for practical deployment. Three key capabilities in deployed applications are retrieval-augmented generation (RAG), function calling, and text embedding. Benchmarks used to track these capabilities are particularly useful because they test fluency and whether models can operate as part of a larger system. It also makes it easier to compare models in settings where performance depends not just on the base model, but on issues such as retrieval quality or how outputs are parsed and executed.

### RAG: Retrieval-Augmented Generation

Retrieval-augmented generation (RAG) provides a way for models to deliver accurate, up-to-date information beyond the knowledge encoded during training in model parameters. At inference time, RAG systems augment model responses with information retrieved from external sources.

Standard RAG pipelines retrieve individual text chunks based on query similarity, which can struggle when answering questions that require synthesizing information across documents. To address the problem, in 2024, Microsoft Research introduced Graph RAG, which enables more effective responses to queries by structuring source material into a knowledge graph and generating community summaries that capture high-

6 Source: https://arena.ai/leaderboard/text.

level themes. Other variants focus on improving multistep retrieval or reranking passages before generation. As expected, these choices in architecture involve trade-offs between answer quality, latency, and cost.

Context windows, discussed later in this section, have important implications for RAG systems. Extended context windows can support retrieval of more material, though that does not guarantee better performance since models have to parse through the information with reliable attention across the entire window.

## Berkeley Function Calling Leaderboard

Function calling allows a model to use external tools and APIs by generating structured requests that another system can run, then folding the results back into its response. It is a foundational capability for agent frameworks, where models need to take actions or retrieve information beyond their training data.

The Berkeley Function Calling Leaderboard (BFCL) evaluates models on their function-calling ability and has evolved considerably since its initial release. The current iteration, BFCL V4, shifts the focus toward holistic agent evaluations. Agentic tasks account for 40% of the overall score, multiturn interactions are 30%, and the remainder is split across live, nonlive, and hallucination categories. The agentic component tests web search and memory while the multiturn component evaluates multistep dialogues. Earlier versions focused more narrowly on single-turn function calling.

The overall accuracy on the BFCL varies widely as of early 2026. The top 15 models span a roughly 21 percentage point range (Figure 2.2.3). Claude models occupy three of the top six positions, with Claude-Opus-4.5 leading at 77.5%. There is also a performance distinction with evaluation modes, showing the trade-offs between general capability and task-specific optimization. For example, Grok-4-0709 scores 63% in prompt mode but drops to 61.4% when using function-calling mode, while Grok-4-1 scores higher in its fast-reasoning variant (69.6%) than its nonreasoning counterpart (58.3%).

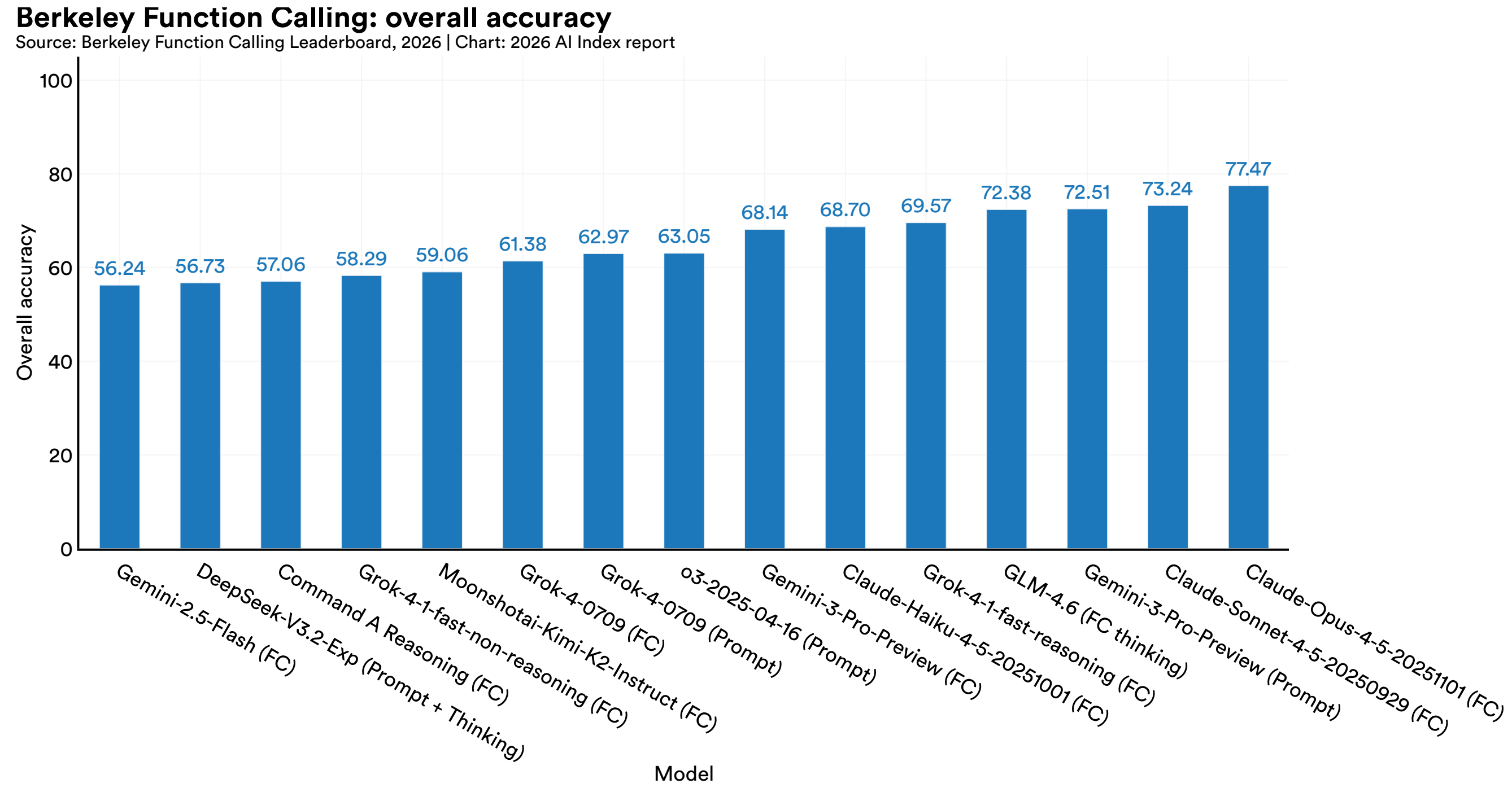

Figure 2.2.3[7]

7 Source: https://gorilla.cs.berkeley.edu/leaderboard.html.

## MTEB: Massive Text Embedding Benchmark

The Massive Text Embedding Benchmark (MTEB) evaluates different embedding models across a set of tasks that require semantic understanding. It includes over 50 datasets, spanning eight task categories, which makes it harder for models to look strong by optimizing for a single use case rather than performing well across different settings.

The top average task score on MTEB (English v2) has risen steadily since 2022, coinciding with the broader adoption of large-scale pretraining techniques for embedding models. In 2025, the top score reached 76, rising approximately 11 points since 2023 (Figure 2.2.4). However, the best models still fall short of a perfect score.

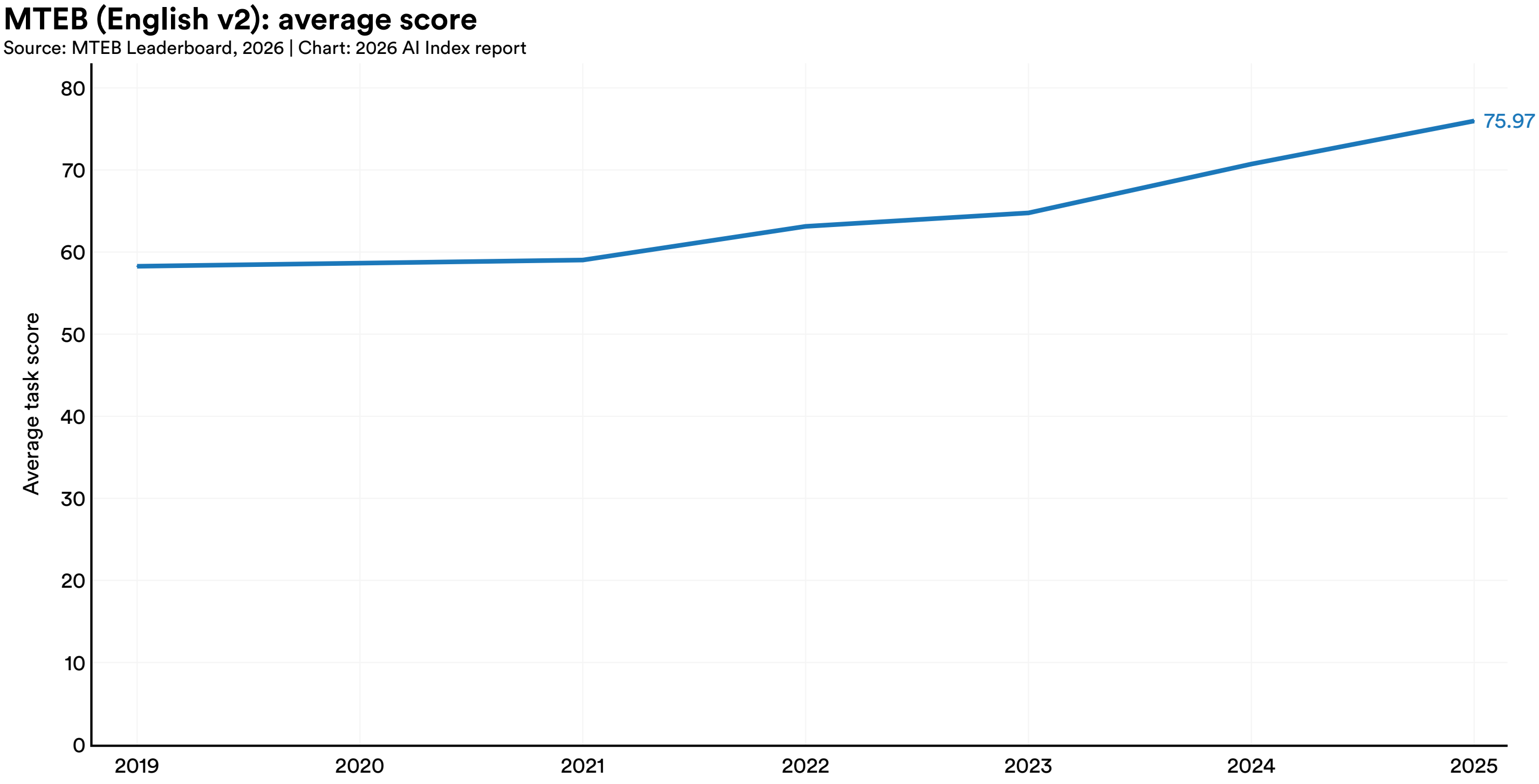
MTEB (English v2): average score
Source: MTEB Leaderboard, 2026 | Chart: 2026 AI Index report

Figure 2.2.4[8]

8 Source: https://mteb-leaderboard.hf.space/?benchmark_name=MTEB%28eng%2C+v1%29
https://arxiv.org/abs/2502.13595.

**HIGHLIGHT:**

## The Gap Between Long Context Windows and Deep Understanding

Context windows, the amount of text a model can process in a single input, have grown by almost 30x per year since mid-2023 (Figure 2.2.5). Models that once accepted a few thousand tokens can now process 1 million or more. At the upper end, this is equivalent to multiple books or an entire codebase in a single pass. On two long-context benchmarks, Fiction.liveBench, which measures narrative comprehension, and MRCR, which measures multi-needle retrieval, the input length at which leading models achieve 80% accuracy has increased even faster, at roughly 250x over a nine month period (Burnham and Adamczewski, 2025). However, bigger context windows do not translate into deeper understanding, as the gap between accepted and usable context length is wide.

Recent research points to different reasons for this gap. On one expert-level, long-context benchmark (LongBench v2), human experts scored just 53.7% accuracy under a 15-minute time limit, and the best model scored 57.7% (Bai et al., 2025). This is a narrow margin in contrast to the structured benchmarks where models have surpassed human baselines, and reflects the difficulty of deep comprehension over long inputs. Models that were prompted to reason through the material step by step did perform better than those asked to answer immediately, suggesting that how a model works through long text matters as much as the amount of text it can accept. Other research has found that models handle simple lookups well but struggle when asked to find multiple pieces of matching information or to apply conditions across a very long document—tasks that would be straightforward for a human scanning the same text (Yu et al., 2025). Models can complete these tasks if guided to check each one by one, but this approach is slow and expensive. Longer inputs come with practical costs of slower response times, higher operating expenses, and reduced accuracy for information that appears later in the input.

Measuring long context ability also remains difficult. When a model scores well on a long context test, it is not always clear whether it genuinely processed the full input or simply relied on knowledge it already had. Yang et al. (2025) introduced a metric designed to separate these two factors and found that model rankings shifted a lot. For example, a model that ranked seventh on raw scores ranked first when only long-context ability was measured, further underscoring why it is important to distinguish between a model specifically being able to better handle long inputs, rather than having overall better capabilities. If the gap between context window size and effective utilization becomes more precise, models may improve their ability to work on tasks that unfold over hours or days and sustain longer chains of reasoning (Denain and Ho, 2025). Developing evaluations that reliably distinguish true long-context ability from general model capability will be important for tracking that progress and ensuring that benchmark gains reflect real improvements.

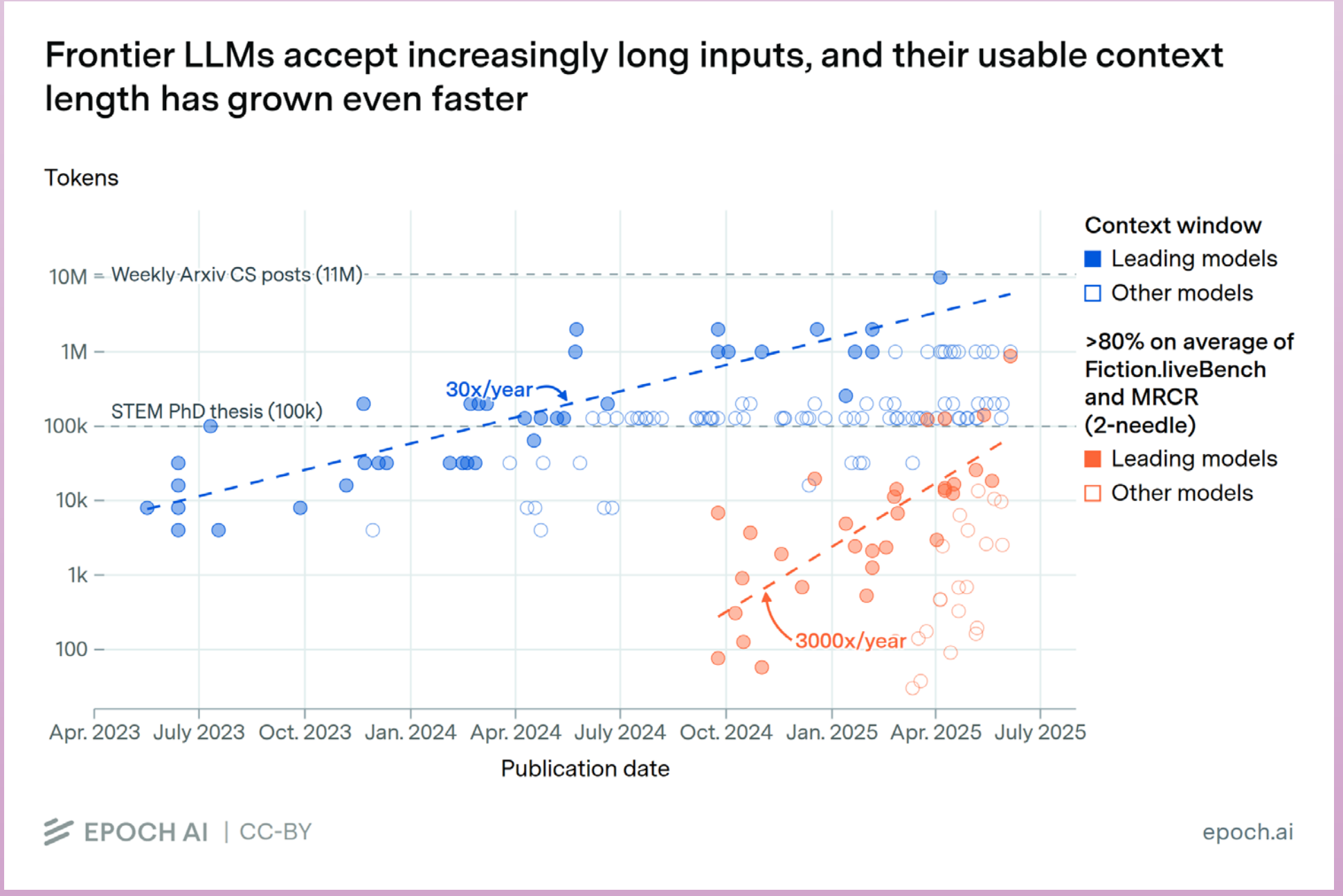

Figure 2.2.5

# 2.3 Image and Video

Beyond language, many models process visual inputs, and their video and image capabilities have advanced significantly. This section examines model performance across the dimensions of understanding—how well they comprehend and reason over video content—and generation, which evaluates the quality of AI-produced images and videos.

## Understanding

Video understanding benchmarks measure how well models can track actions, objects, and events across frames rather than reasoning over a single image. As performance on earlier benchmarks has improved, evaluation has shifted toward tasks that demand multistep temporal reasoning and domain-specific knowledge applied to video.

### MVBench

MVBench evaluates whether multimodal models can move beyond static image understanding to handle the complexities of video. This includes interpreting motion, temporal sequences, and shifting context across frames. Its focus on temporal reasoning makes it a useful benchmark for tracking performance in more dynamic visual environments.

The top-performing model on MVBench reaches 74.1% average accuracy, with JT-VL-Chat and JT3.5 tied at that score (Figure 2.3.1). In early 2026, across the top 15 models, performance spans a range of roughly 23 percentage points. VideoChat 2 has the lowest average accuracy (51.1%), while several VideoChat2 variants are grouped in the middle tier (60%–65%).

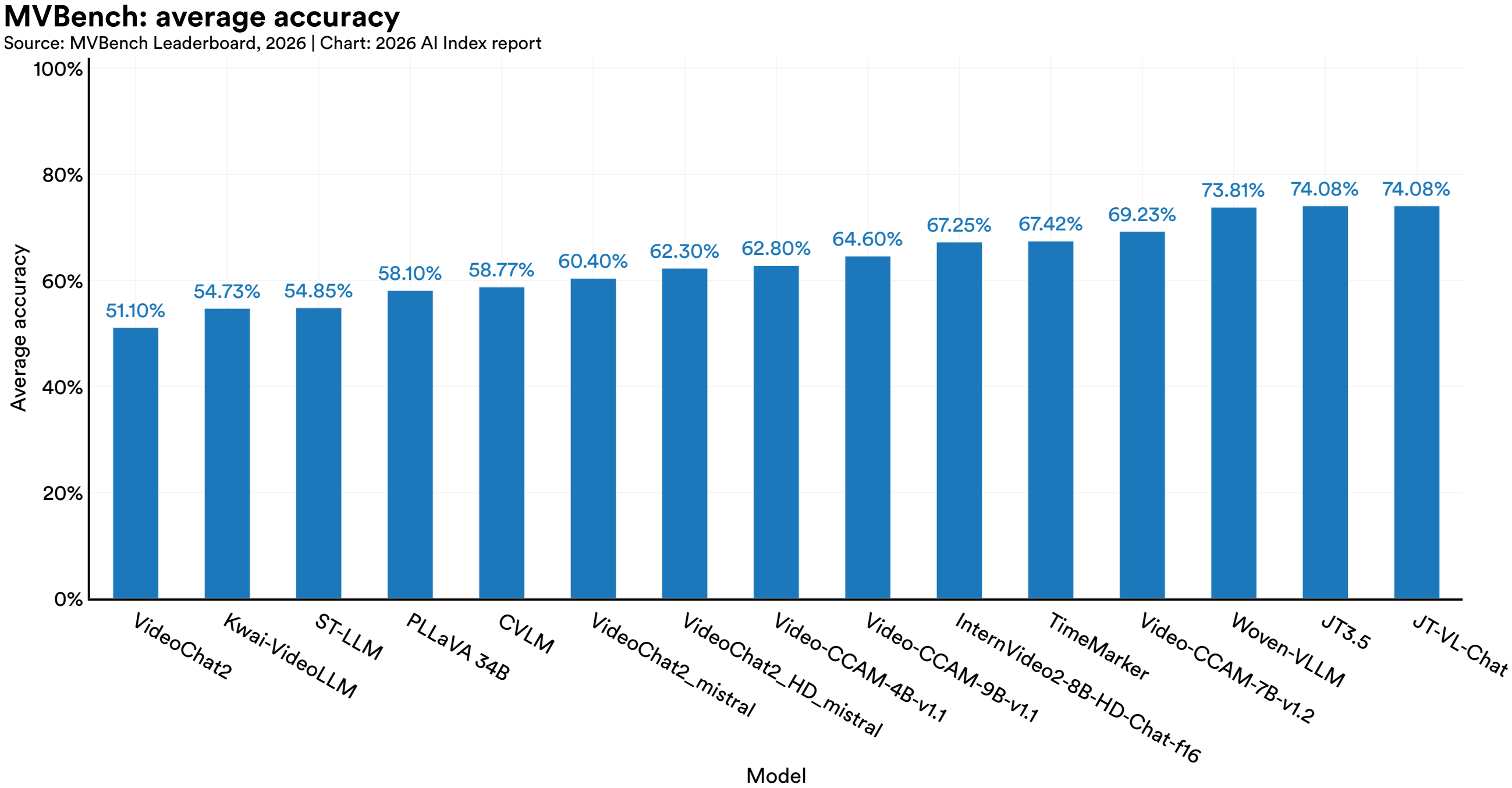

Figure 2.3.1[9]

9 Source: https://huggingface.co/spaces/OpenGVLab/MVBench_Leaderboard.

## Video-MMMU

Video-MMMU is a large, multimodal, multidisciplinary benchmark for learning from educational videos, comprising 300 expert-level videos averaging roughly 506 seconds across six disciplines and 30 subjects. Each video is paired with three sets of questions that test progressively deeper understanding. Perception questions test whether a model can pull key details from text/audio; comprehension questions test whether it grasps the concept or solution strategy; and adaptation questions require applying that knowledge to a new scenario. Adaptation questions reuse MMMU/MMMU-Pro items for STEM fields and custom case studies for art/humanities, so models have to go beyond the specific video. The benchmarks also introduce a Δknowledge metric to track how much a model's performance improves after processing the video.

As of 2025, no model has reached the human baseline of 74.4% on Video-MMMU overall accuracy (Figure 2.3.2). The best performing model, Keye-VL-1.5-8B, scores 66%, followed closely by Claude -3.5-Sonnet (65.8%). The lowest score is VILA1.5-8B at 20.9%, leaving a 45 percentage point range across the leaderboard.

The Δknowledge metric results reveal a further gap between human and model learning (Figure 2.3.3). Human experts gain 33.1 percentage points after watching the video, while the best model on this metric, GPT-4o, gains only about half of that (15.6 points). About a third of models even show negative Δknowledge, as their performance actually declines after processing the video.

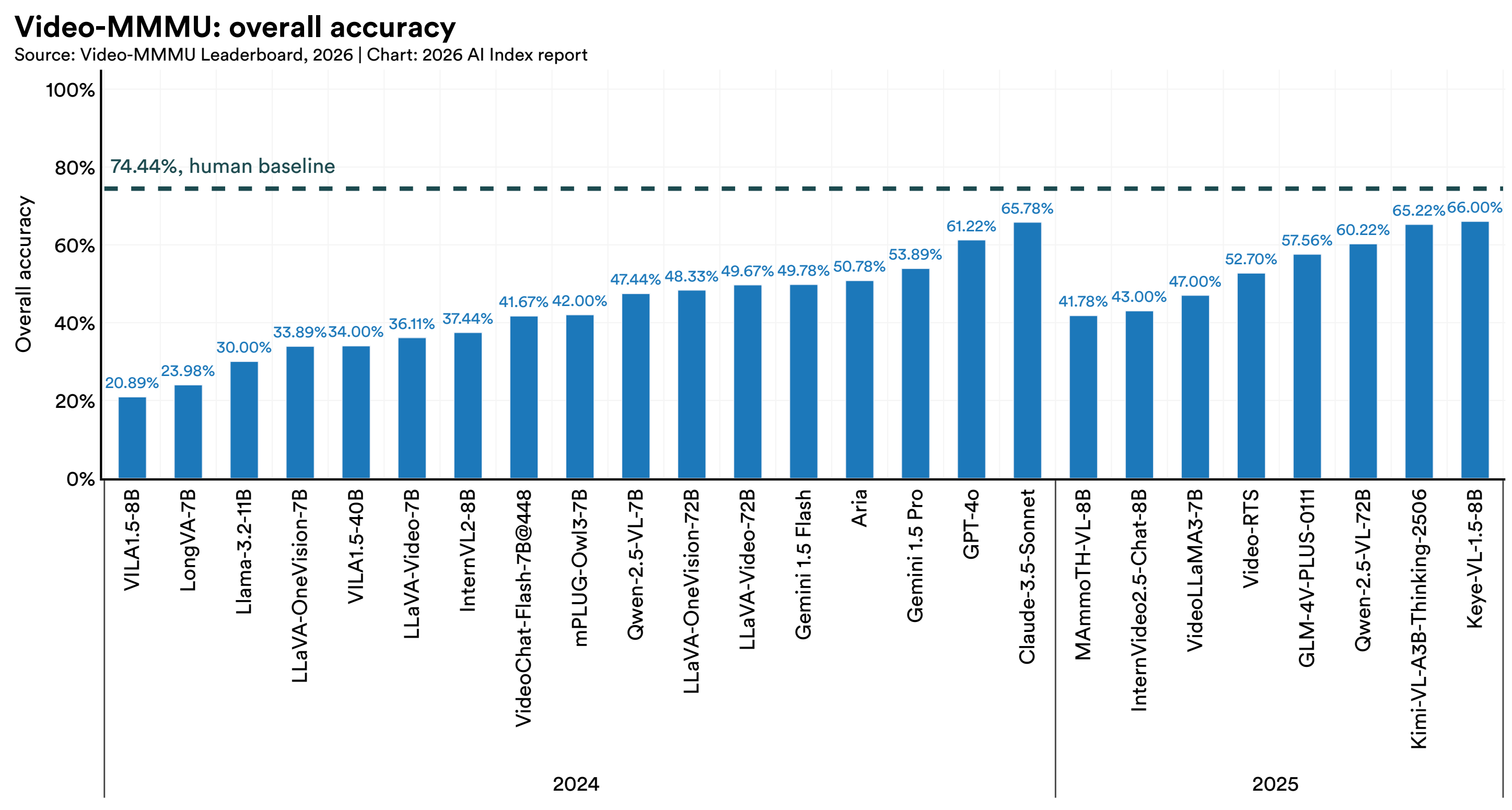

Figure 2.3.2[10]

10 Source: https://videommmu.github.io/#Leaderboard.

**Video-MMMU: Δknowledge**
Source: Video-MMMU Leaderboard, 2026 | Chart: 2026 AI Index report

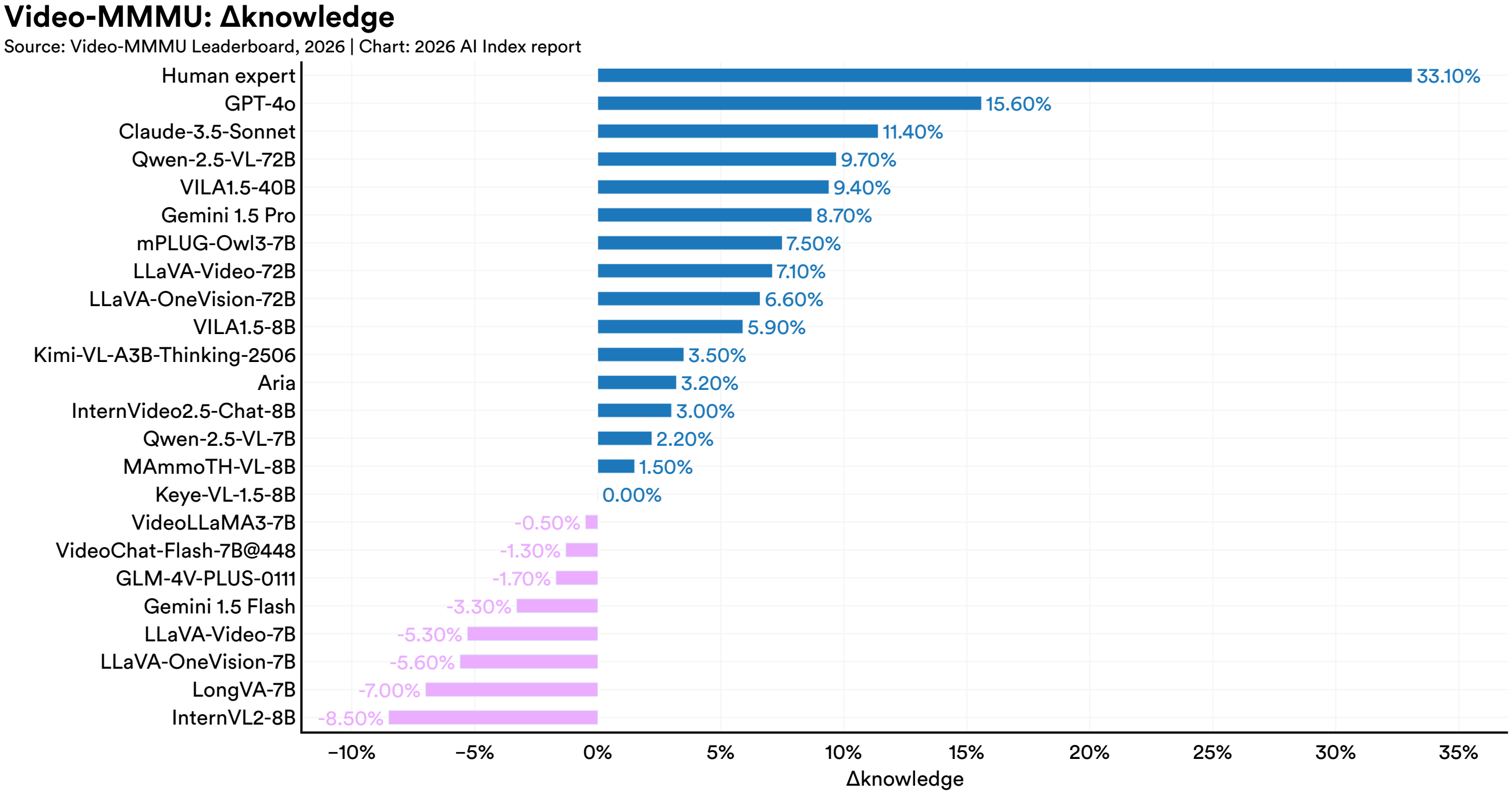

Figure 2.3.3

## Generation

While the above benchmarks test how well models interpret existing visual content, generation benchmarks assess how well models can produce it. Evaluation spans both human preference rankings as well as automated quality metrics, since generated video must satisfy subjective expectations and technical criteria such as coherence, fidelity, and controllability. Of these, controllability has become an especially important focus, reflecting whether models can follow user intent while maintaining natural motion and scene dynamics. This has also brought video generation closer to the idea of world models, where systems aim to predict how visual scenes evolve over time.

**Midjourney generations over time: "a hyper-realistic image of Harry Potter"**
Source: Midjourney, 2025

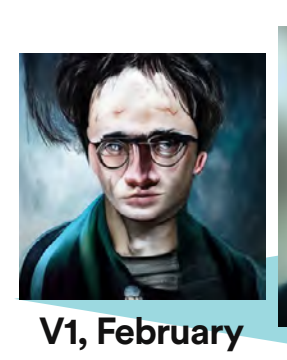

V1, February 2022

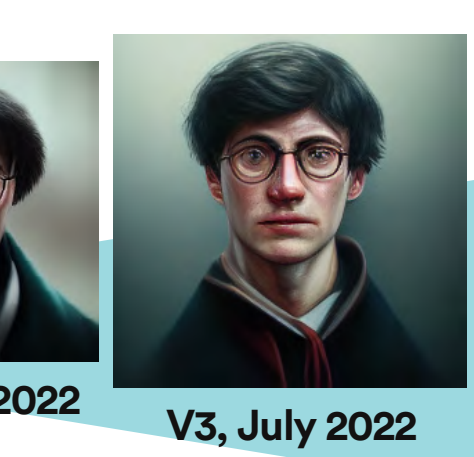

V2, April 2022

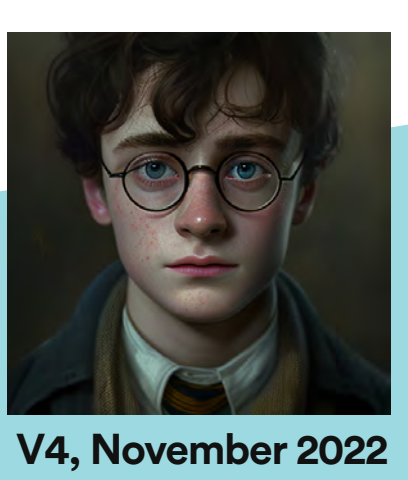

V3, July 2022

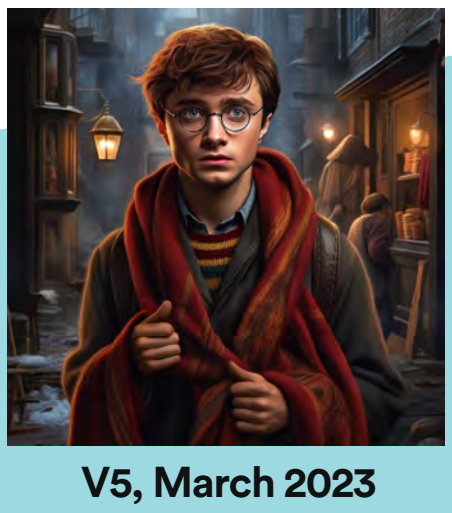

V4, November 2022

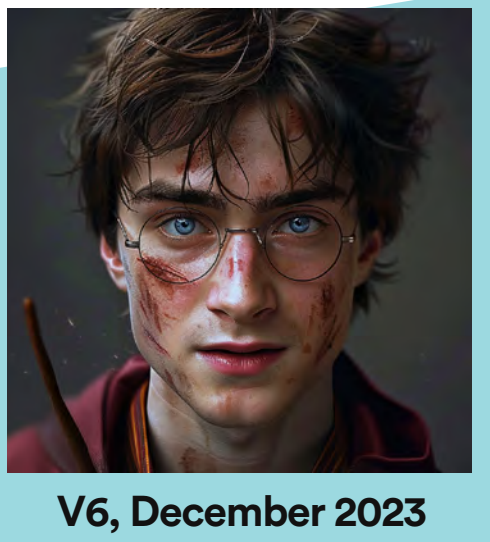

V5, March 2023

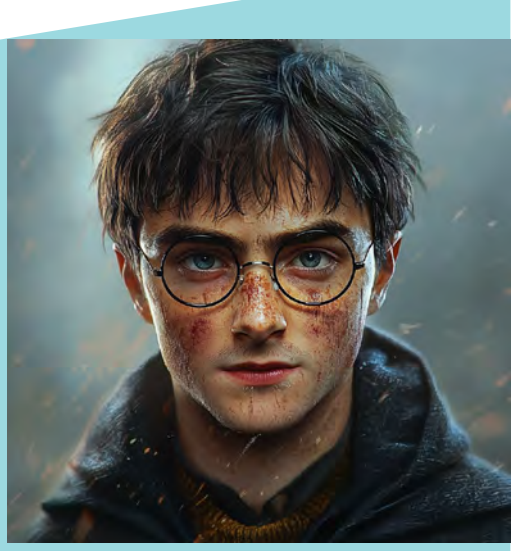

V6, December 2023

V6.1, July 2024

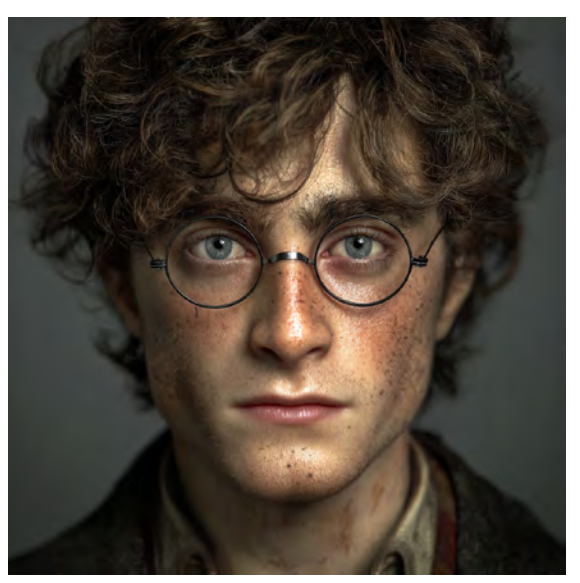

V7, April 2025

Figure 2.3.4

## Arena: Vision

The Arena platform also hosts a Vision Arena that applies the same blind-comparison, Elo-based methodology described in the earlier section for language to image generation models. Human preference is an important signal for image generation, as qualities like aesthetic appeal and visual coherence are difficult to capture through automated metrics alone.

As of early 2026, Google's Gemini models hold four of the top six positions, with Gemini-3-Pro leading at approximately 1,285 Elo, followed by its other variation (Figure 2.3.5). Similar to the language evaluation, confidence intervals overlap for the bottom two-thirds of the ranked models as they fall within a 30-point range between 1,230 and 1,260.

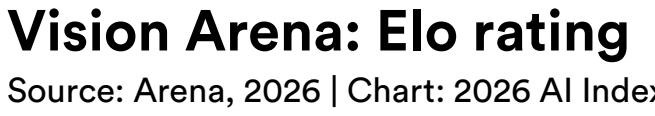

**Vision Arena: Elo rating**
Source: Arena, 2026 | Chart: 2026 AI Index report

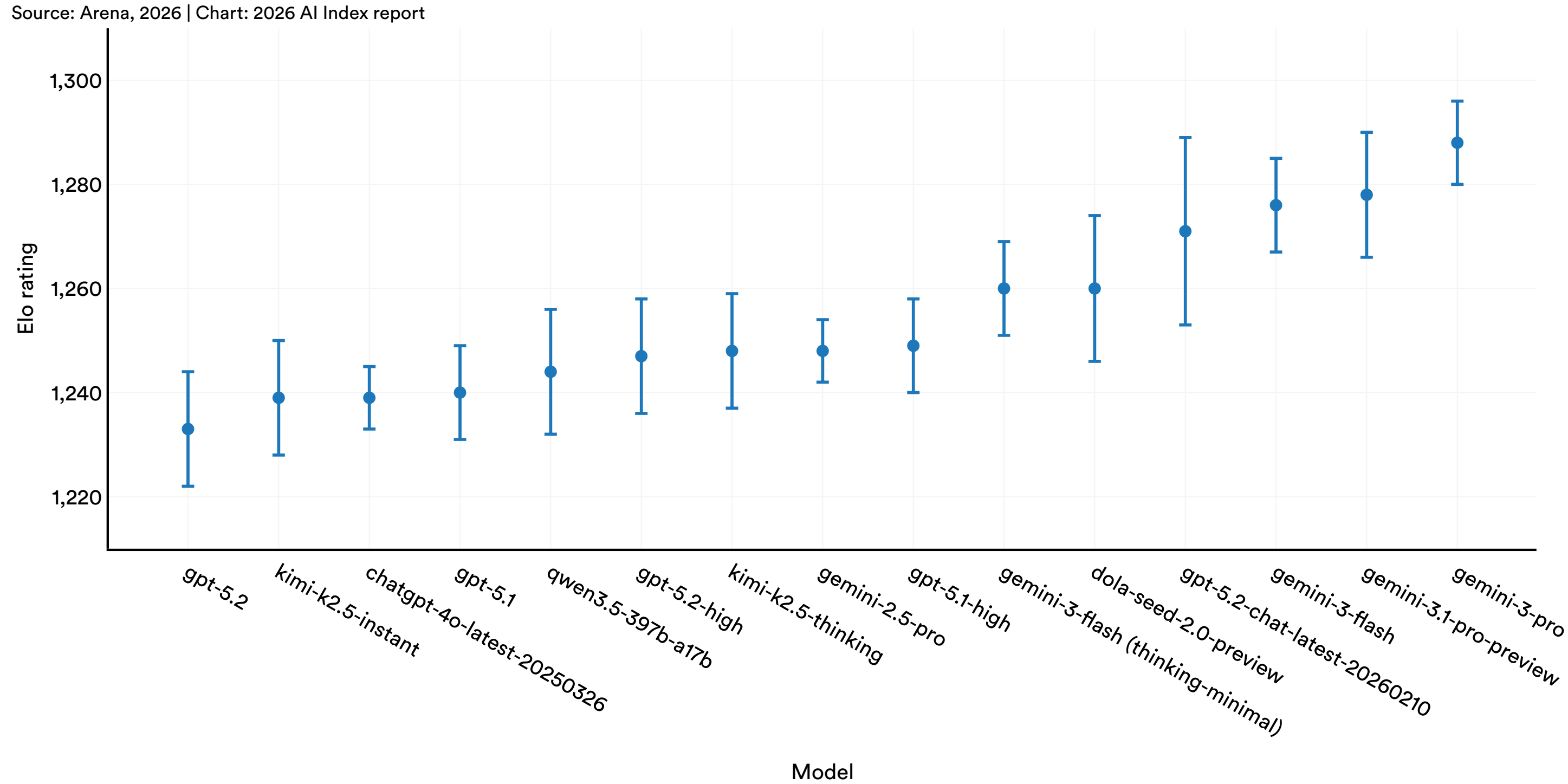

Figure 2.3.5[11]

## Video-Bench

Video-Bench is a human-aligned benchmark for video generation that scores models on two dimension groups: video-condition alignment and video quality. It uses an MLLM-based evaluator (GPT-4o) with few-shot scoring, and chain-of-query prompting for more precise, calibrated results. The scores correlate more strongly with human ratings than prior metric-based or LLM-based benchmarks.

As of early 2026, Gen3 and Kling lead on video quality, with Gen3 scoring highest across all key metrics, including imaging quality, aesthetic quality, temporal consistency, and motion effects, while Kling ranks second overall (Figure 2.3.6). Motion effects are the weakest subdimension across nearly all models.

11 This chart shows only the top 15 models as of February 2026; source: https://arena.ai/leaderboard/vision.

**Video-Bench video quality**
Source: Video-Bench Leaderboard, 2025 | Chart: 2026 AI Index report

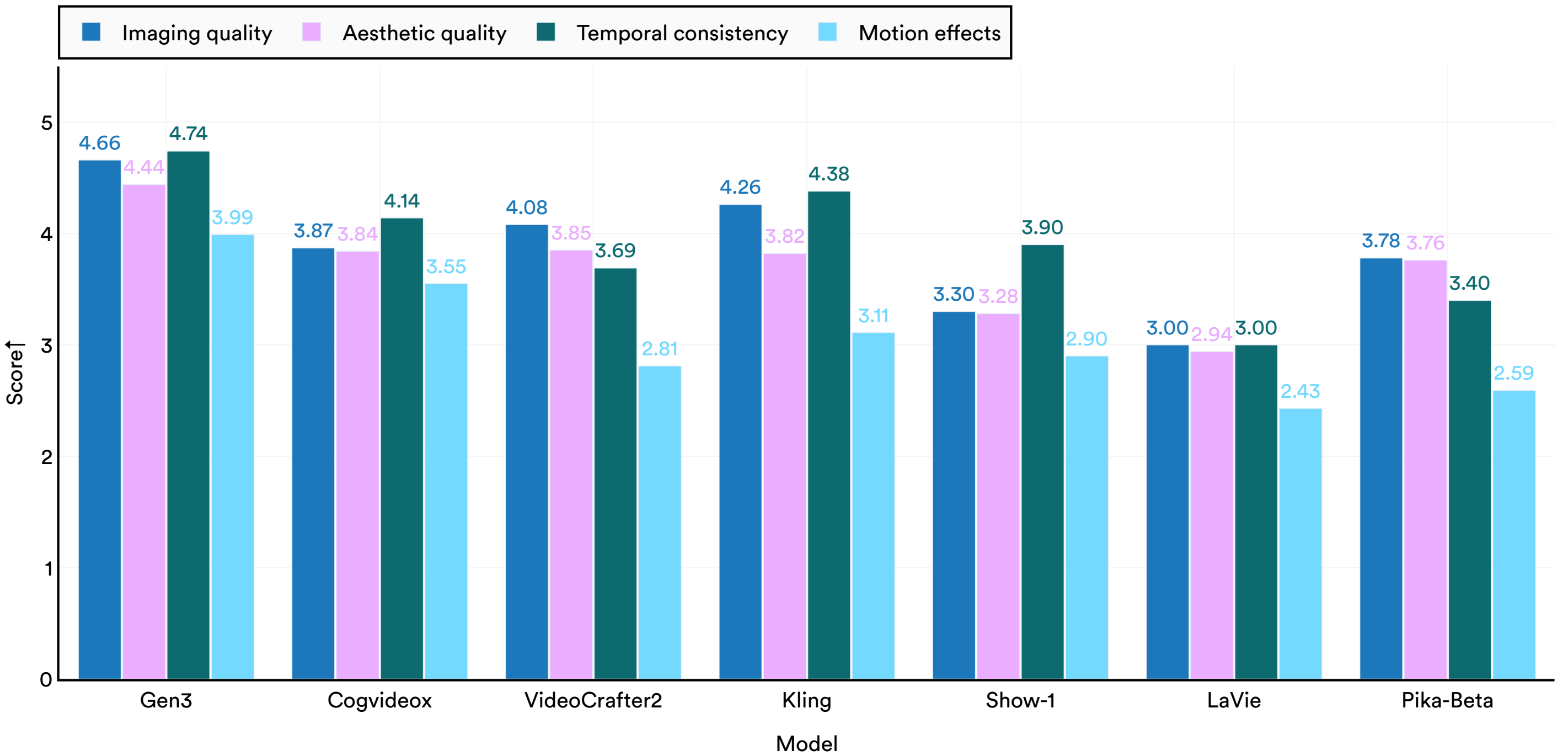

Figure 2.3.6[12]

## VBench-2.0

VBench-2.0 is a comprehensive, human-aligned benchmark for evaluating video generation models on intrinsic faithfulness, defined as well-rounded adherence to reality rather than simply being visually convincing. It scores models across five broad dimensions (Human Fidelity, Creativity, Controllability, Physics, and Commonsense). The benchmark combines VLM/LLM-based analysis with specialized detectors and a small but targeted prompt set, anchored by human preference labels. This faithfulness-oriented approach is important because it surfaces whether generated videos hold up under scrutiny in areas like physical plausibility and scene consistency.

None of the models evaluated in early 2026 surpasses a total score of 67% (Figure 2.3.7). Veo 3 leads at 66.7%, about 4 percentage points above the next top performing mode, Vidu Q1 (62.7%). Similar to other benchmark scores, several models are tightly grouped and hover around scores of 58% and 60%. Even established systems like Kling, CogVideoX, and HunyuanVideo continue to struggle with complex stories and consistent object/scene dynamics.

12 Source: https://github.com/Video-Bench/Video-Bench?tab=readme-ov-file#leaderboard.

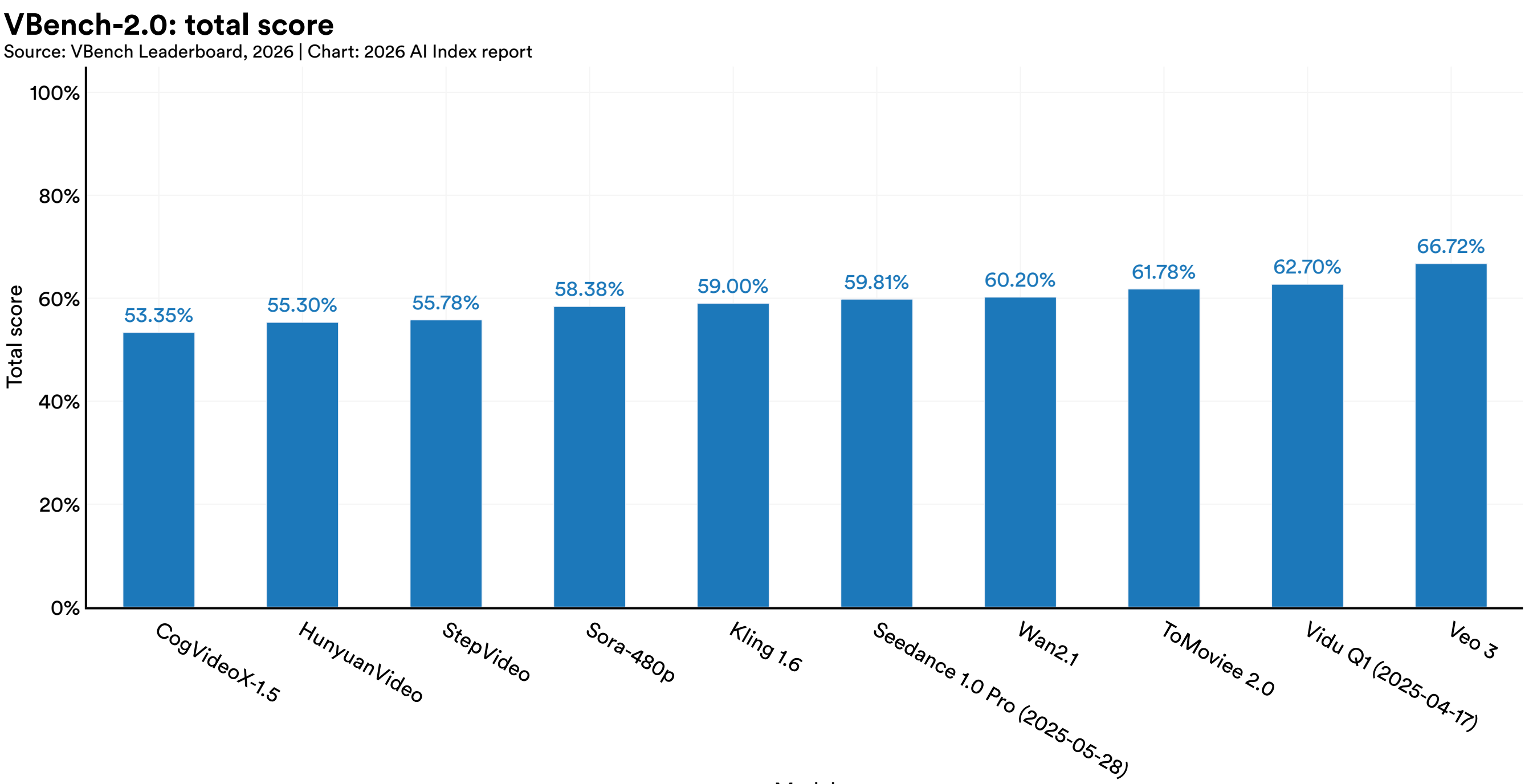

Figure 2.3.7[13]

HIGHLIGHT:

## Progress in Video Generation

The benchmarks in this section mostly evaluate video modes as content generators, scoring them on quality, fidelity, and controllability. However, recent research suggests that video generation models may be developing capabilities that go beyond producing content.

A 2025 Google DeepMind study (Wiedemar et al., 2025) tested whether Veo 3, a video generation model, could solve visual tasks it was never specifically trained for, using only an input image and a text prompt. Across 62 qualitative tasks and seven quantitative evaluations covering more than 18,000 generated videos, the model showed zero-shot abilities in areas traditionally handled by specialized systems. These included perception tasks such as edge detection and segmentation, physical modeling tasks such as buoyancy and rigid body dynamics, and manipulation tasks such as style transfer and object extraction. The authors also observed early signs of visual reasoning, including maze solving and visual analogy completion, which they describe as “chain of frames,” a parallel to chain-of-thought reasoning in language models where the model appears to reason step by step through successive frames. Performance improved consistently from Veo 2 to Veo 3 across all quantitative tasks and, in some cases, matched or exceeded a dedicated image editing baseline (Nano Banana).

Specialized models still outperform zero-shot video generation on most individual tasks, but the rapid improvement and breadth of zero-shot capability suggests a familiar trajectory. Large language models develop general-purpose language understanding from generative training on web-scale data, and video models trained under similar conditions may be following a comparable path toward general-purpose vision.

13 Source: https://huggingface.co/spaces/Vchitect/VBench_Leaderboard.

# 2.4 Reasoning

Reasoning benchmarks assess whether models can solve problems that require abstraction and generalization across domains and formats. As performance has improved, newer benchmarks aim to distinguish genuine problem-solving from performance that is driven by memorization or prompt familiarity. However, because models can also produce errors in otherwise fluent responses, efforts are ramping up to measure these error rates alongside reasoning limitations. The AI Index tracks those benchmarks on factual reliability and error rates in Chapter 3. Across the benchmarks in this section, leading models perform well on many tasks but still show gaps on the more difficult items.

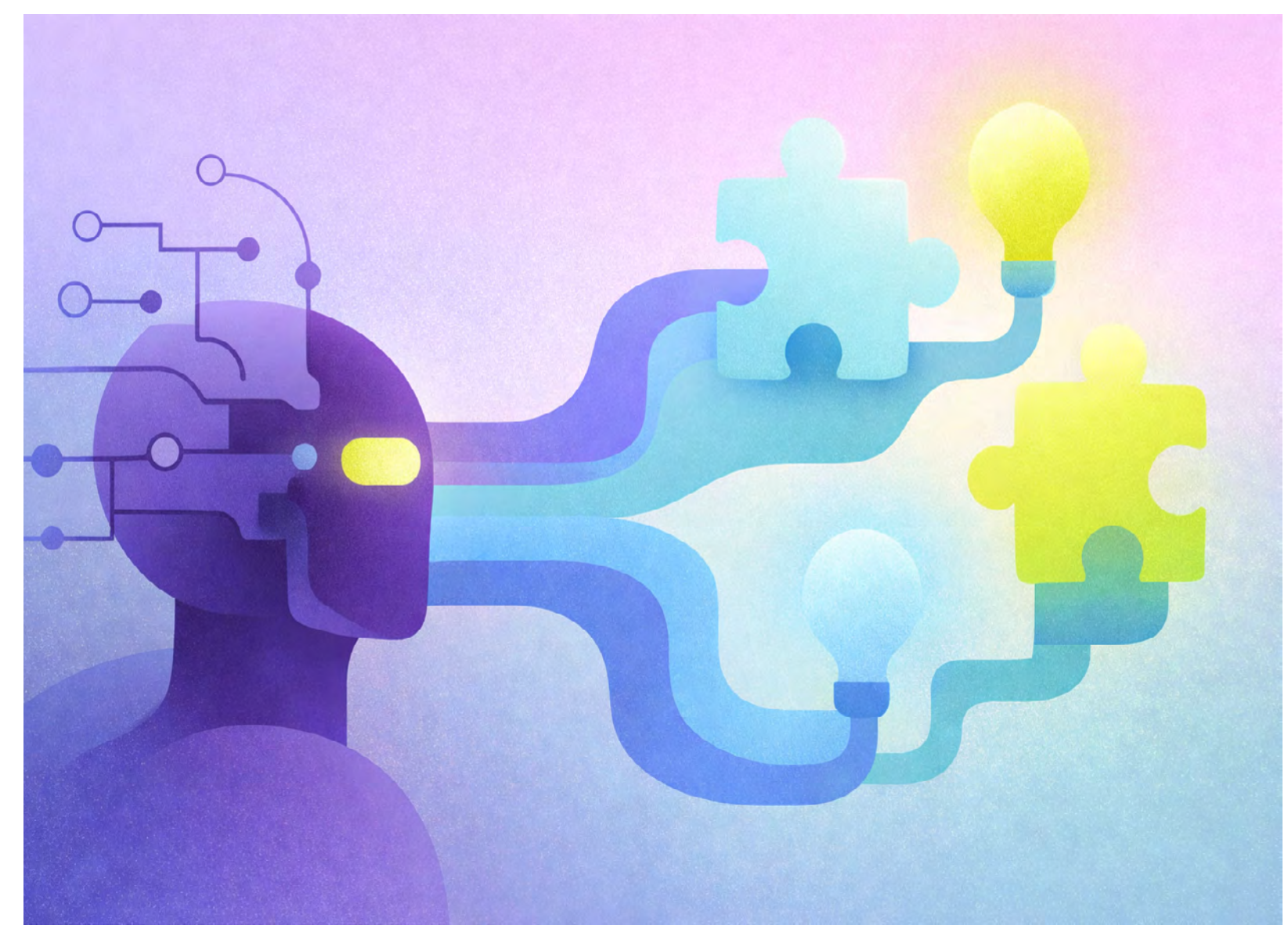

## General Reasoning

General reasoning refers to a model's ability to solve unfamiliar problems by applying rules and combining evidence, rather than relying on domain knowledge or memorized patterns. The benchmarks discussed below span multiple domains and tasks and are designed to test multistep inference. One example is multidigit arithmetic, such as long integer multiplication, to test whether models can execute consistent stepwise computation rather than produce plausible-looking outputs. Other more complex benchmarks extend this idea to multimodal settings, where models must integrate text with diagrams or plots to reach the correct answer.

### MMMU: A Massive Multi-discipline Multimodal Understanding and Reasoning Benchmark for Expert AGI

MMMU evaluates multimodal reasoning on college-level subject questions that combine text with visuals such as diagrams, charts, tables, and equations. Some example tasks include extracting constraints from a table and applying them to a word problem, or using a diagram to answer a domain-specific question in areas like engineering or medicine.

As of February 2026, the leading model, Gemini 3.1 Pro Preview, scored 88.2% on MMMU and within 0.4 percentage points of the best human expert reference (Figure 2.4.1). Other Gemini variants follow closely, including Gemini 3 Flash (87.6%) and Gemini 3 Pro (87.5%), while GPT-5.2 scores 86.7%. The 2026 models trail behind with Kimi K2.5 at 84.3% and Claude Opus 4.6 (Thinking) at 83.9%.

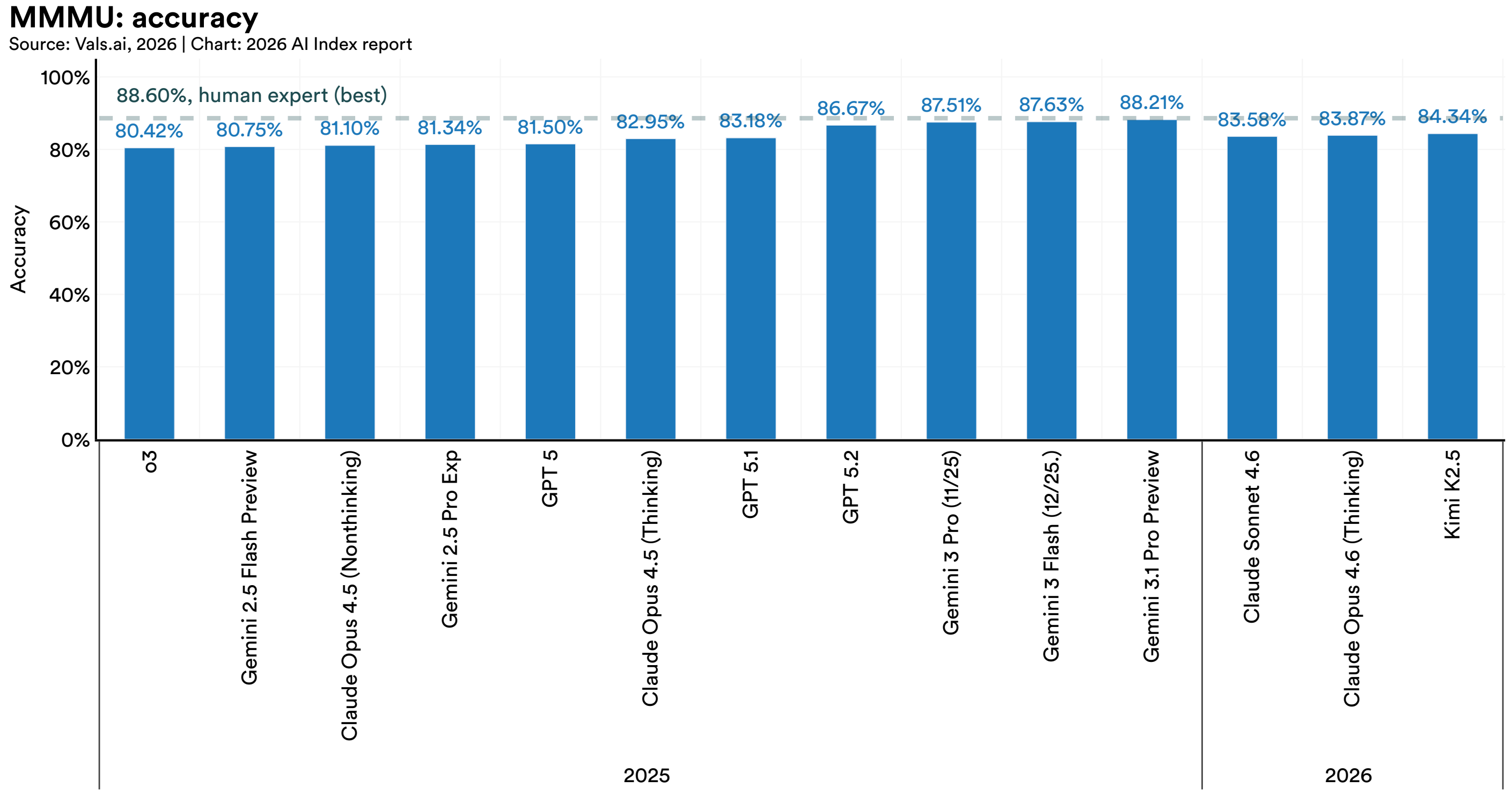

Figure 2.4.1[14]

## GPQA: A Graduate-Level Google-Proof Q&A Benchmark

While MMMU focuses on multimodal reasoning, GPQA evaluates reasoning on difficult, text-only questions designed to test graduate-level problem solving. The questions require models to apply domain-specific concepts and follow multistep logic to reach the correct answer. Example tasks include graduate-level chemistry or physics questions that require working through a multistep solution and choosing the best answer from several very similar options.

Model performance on the GPQA Diamond set has continued to rise above the expert human validator baseline of 81.2% (Figure 2.4.2). In late 2024, OpenAI's o3 was the first to exceed it with a score of 87.7%. In 2025, mean accuracy reached 93%, exceeding the expert reference point by 12 percentage points.

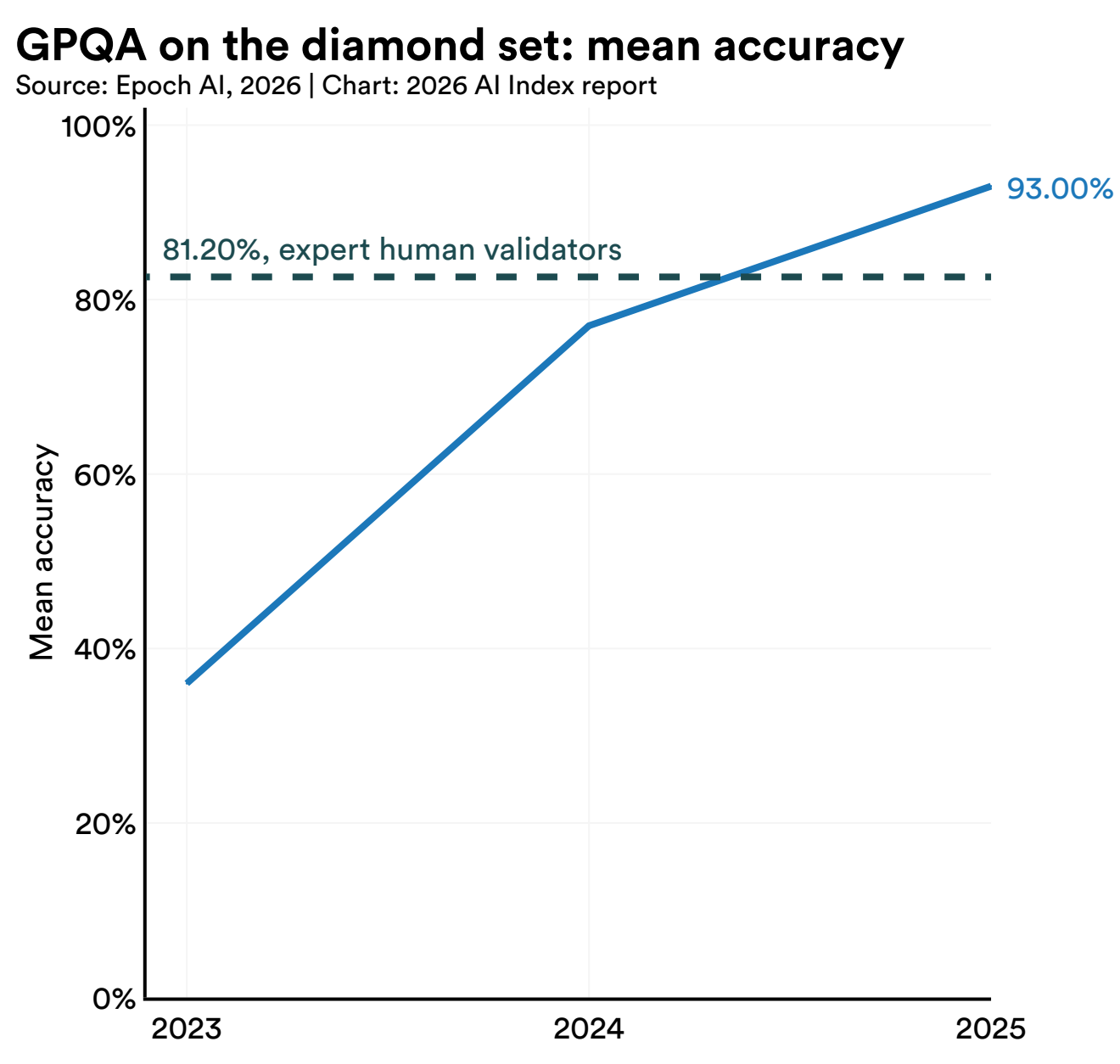

Figure 2.4.2[15]

14 This chart shows the top 15 models as of February 2026; data source: https://www.vals.ai/benchmarks/mmmu.

15 Data source: https://epoch.ai/benchmarks.

## ARC-AGI-2

Introduced in 2019, ARC-AGI is a benchmark that tests the ability of systems to generalize beyond prior training, emphasizing generalized learning ability. Despite its name, the benchmark tests a specific form of abstraction and pattern inference rather than general intelligence in a broader sense. Its updated version, ARC-AGI-2, was introduced in 2025 and shifts to abstract puzzle-style tasks that evaluate whether models can infer rules from a small set of examples and apply them to new cases. Example tasks include grid puzzles where the model is given a few example solutions, infers the rule, and uses it to solve a new problem.

Scores on ARC-AGI-2 vary widely across models, and the spread between the highest and lowest scores in the figure is about 46% (Figure 2.4.3). Gemini 3 Deep Think leads at 84.6%, followed by Gemini 3.1 Pro Preview at 77.1% and GPT-5.2 (Refine.) at 72.9%. Several Claude Opus 4.6 variants are clustered together, scoring between 66.3% and 69.2%.

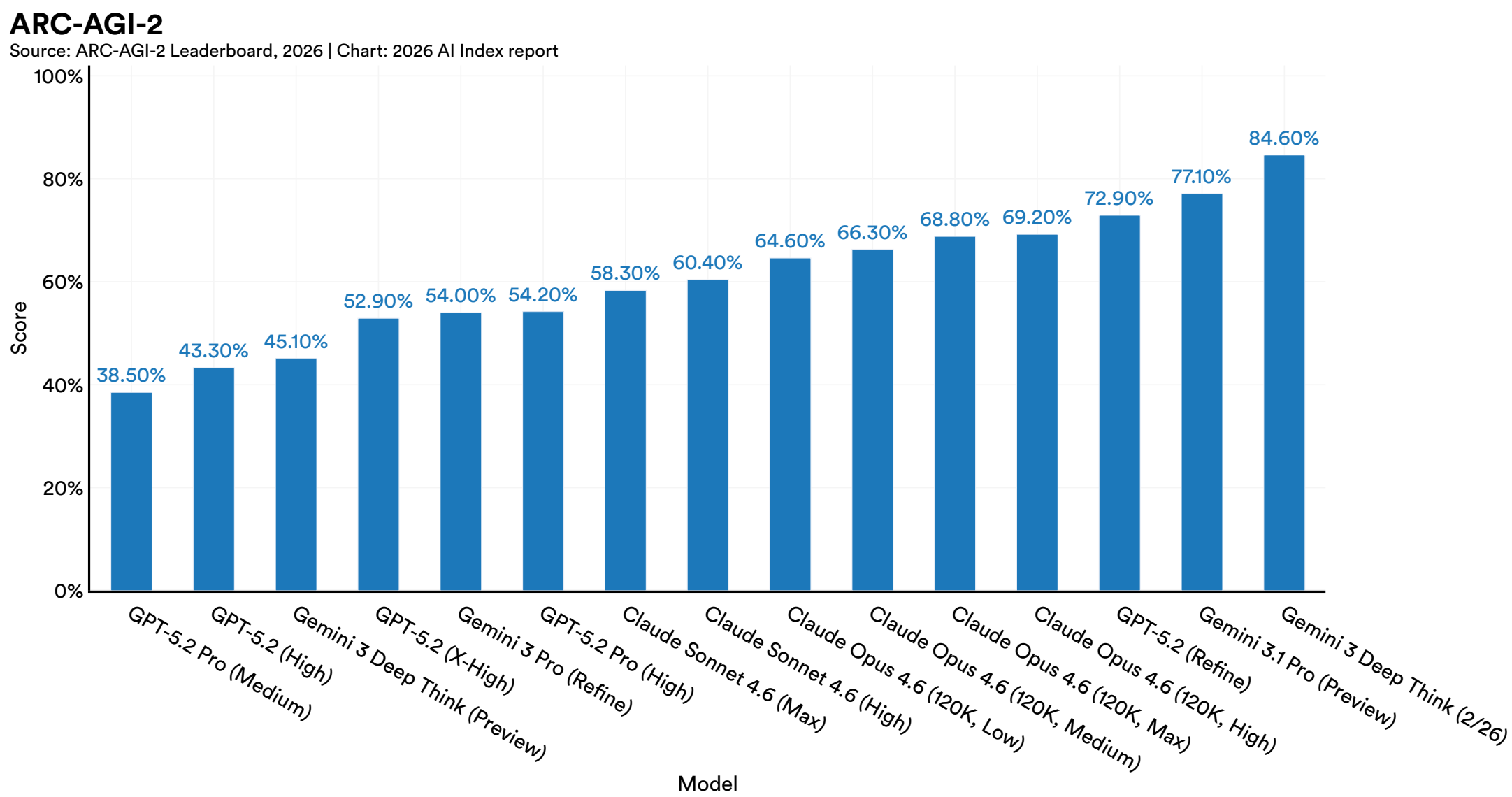

Figure 2.4.3[16]

## Humanity's Last Exam

Humanity's Last Exam (HLE) benchmark evaluates model performance on 2,700 highly challenging questions across dozens of academic subjects. It is designed as an expert-level, closed-ended benchmark with wide coverage and using a mix of multiple-choice and short-answer formats suitable for automated grading. Example tasks include a graduate level question that requires applying a concept and providing a single, verifiable answer. Some may include an image, requiring models to integrate visual and textual information.

Between 2024 and 2025, model accuracy on HLE increased by 30 percentage points (Figure 2.4.4). In a single year, accuracy went from under 10% to 38.3%. Even with this jump, the benchmark is designed to stay difficult, and high-confidence errors are still common.

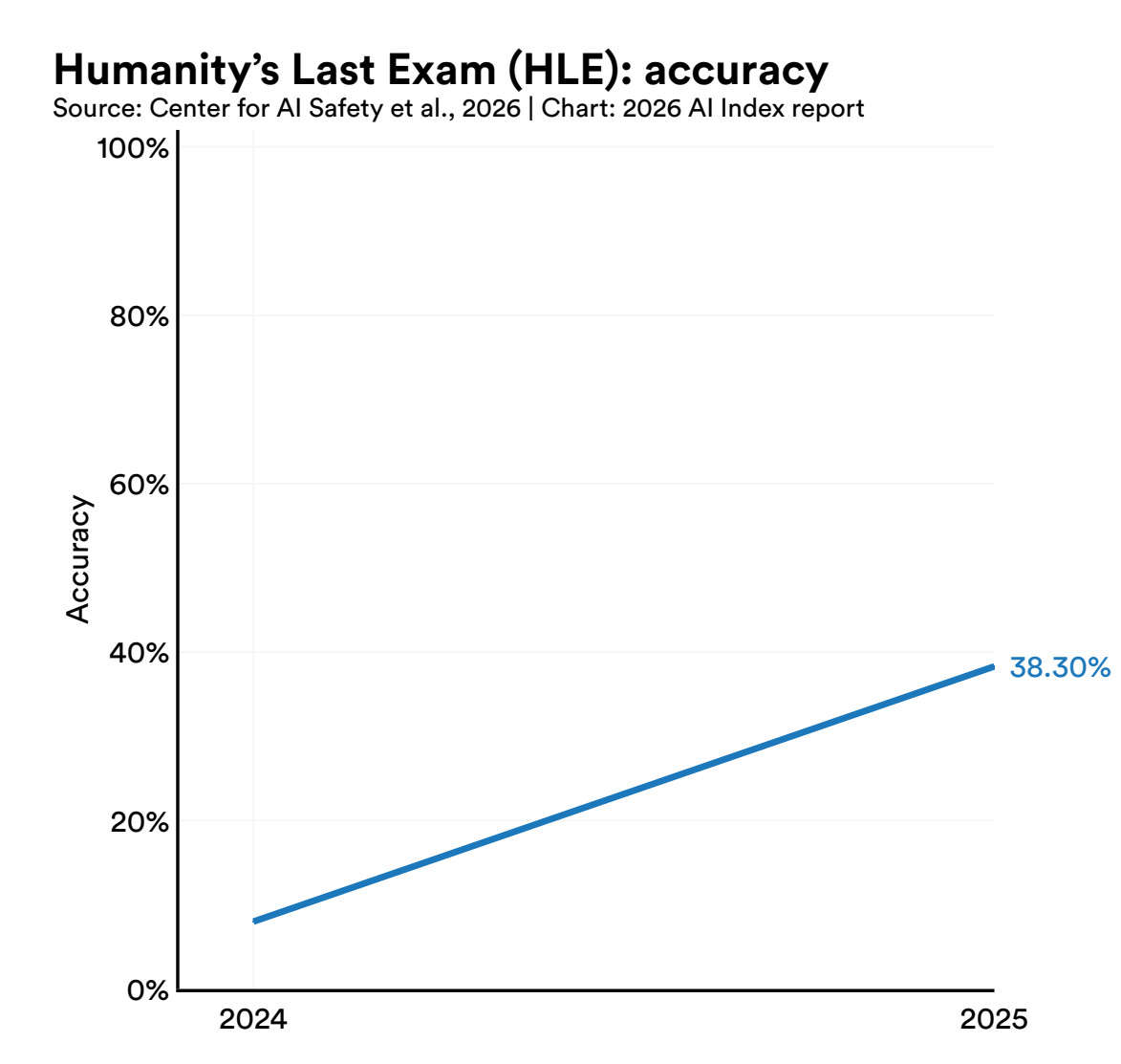

Figure 2.4.4[17]

16 This chart shows the top 15 models as of February 2026; data source: https://arcprize.org/leaderboard.

17 Data source: https://lastexam.ai.

**HIGHLIGHT:**

## Time Understanding in MLLMs

Many multimodal models still struggle with something most humans find routine, telling the time. Despite the rapid improvements on expert-level reasoning benchmarks like GPQA and HLE, recent studies show models have trouble reading analog clocks. The task combines visual perception with simple arithmetic, from identifying clock hands and their positions and then converting those into a time value. There is the risk that an error in one step will cascade into the next.

Saxena et al. (2025) tested seven multimodal models on two focused datasets (Figure 2.4.5). ClockQA included 62 analog clock images across six visual styles—including clocks with a black dial or no second hand—and CalendarQA, which paired yearly calendar images with date-reasoning questions. On clock reading, even the best performing model, Gemini-2.0, achieved only 22.6% exact match accuracy (Figure 2.4.6). Models fared better on the calendar questions, with GPT-o1 reaching 80% accuracy, though there were more errors when questions required date arithmetic rather than recognition of well-known holidays (Figure 2.4.7).

ClockBench (Safar, 2025) scaled up the evaluation to 180 clock designs and 720 questions. Humans read correctly formatted clocks correctly 90.1% of the time, while GPT-5.4 High, the top model, reached 50.6% in March 2026 (Figure 2.4.8). The gap of about 40 percentage points is large, but the wider gap is in the nature of the errors. When models told the time wrong, their median error ranged from about one to three hours, compared to three minutes for humans.

A study published in IEEE Internet Computing (Fu et al., 2025) looked at why these failures continue to happen. After fine-tuning on 5,000 synthetic clock images, models improved on familiar clock styles but failed to generalize to real-world photos or clocks that had different features, such as distorted dials or thinner hands. When researchers dug into the errors, they identified a pattern. If a model confused the hour and minute hands, its ability to judge hand direction deteriorated. This suggests that the difficulty springs less from training data and more on how models piece together multiple visual cues within a single image. Even as models close the gap with human experts on knowledge-intensive tasks, this kind of visual reasoning remains a persistent challenge.

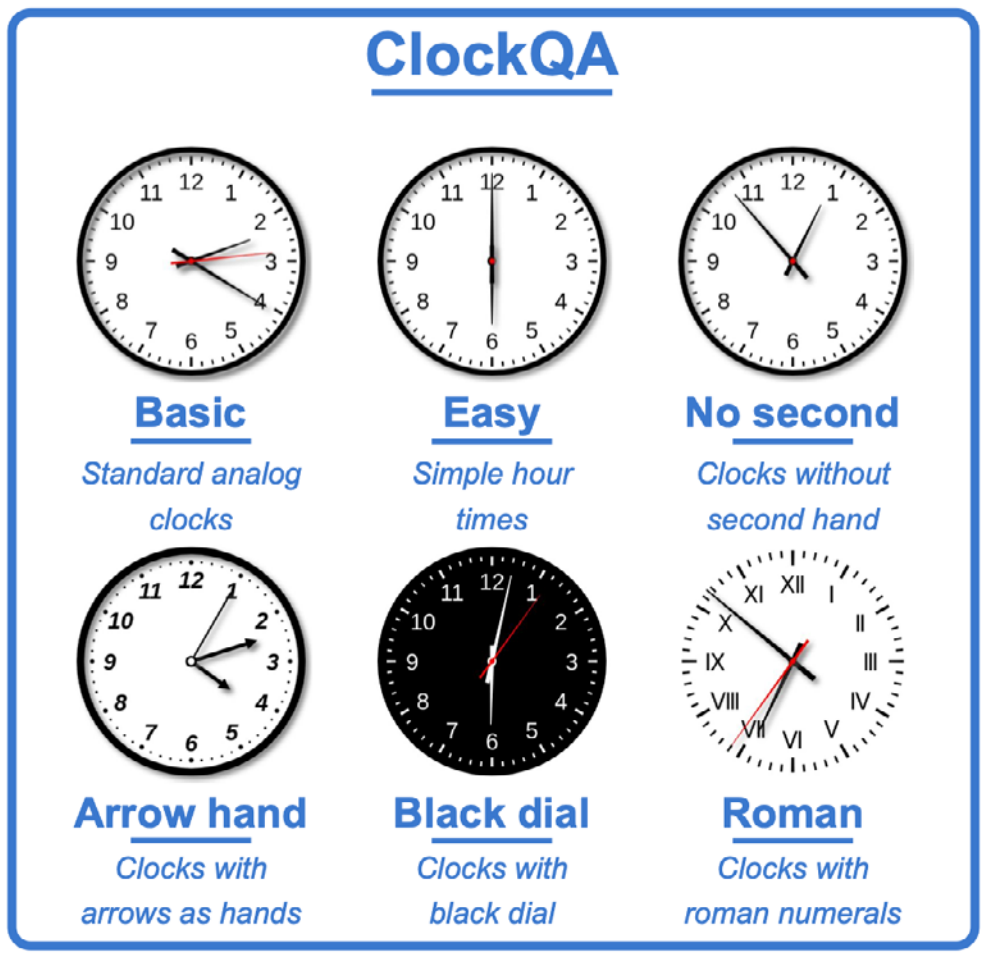

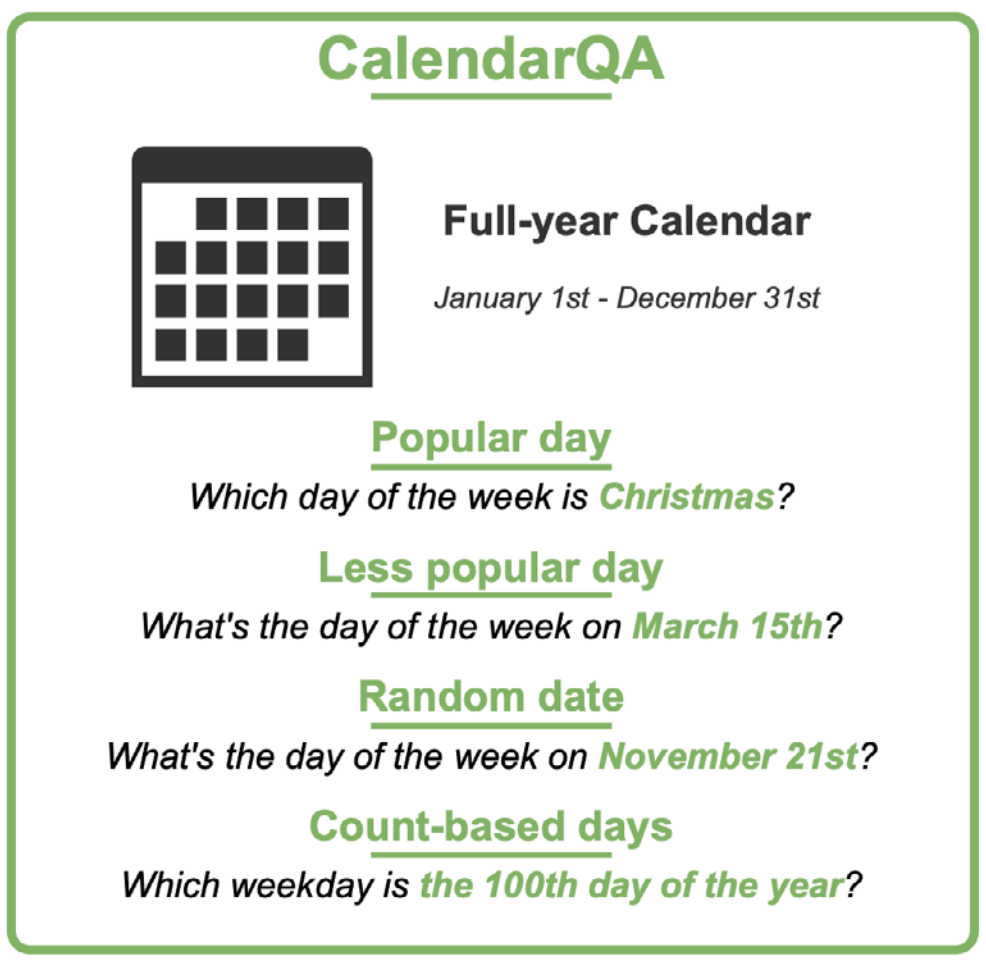

Source: Saxena et al., 2025

Figure 2.4.5

**HIGHLIGHT:**

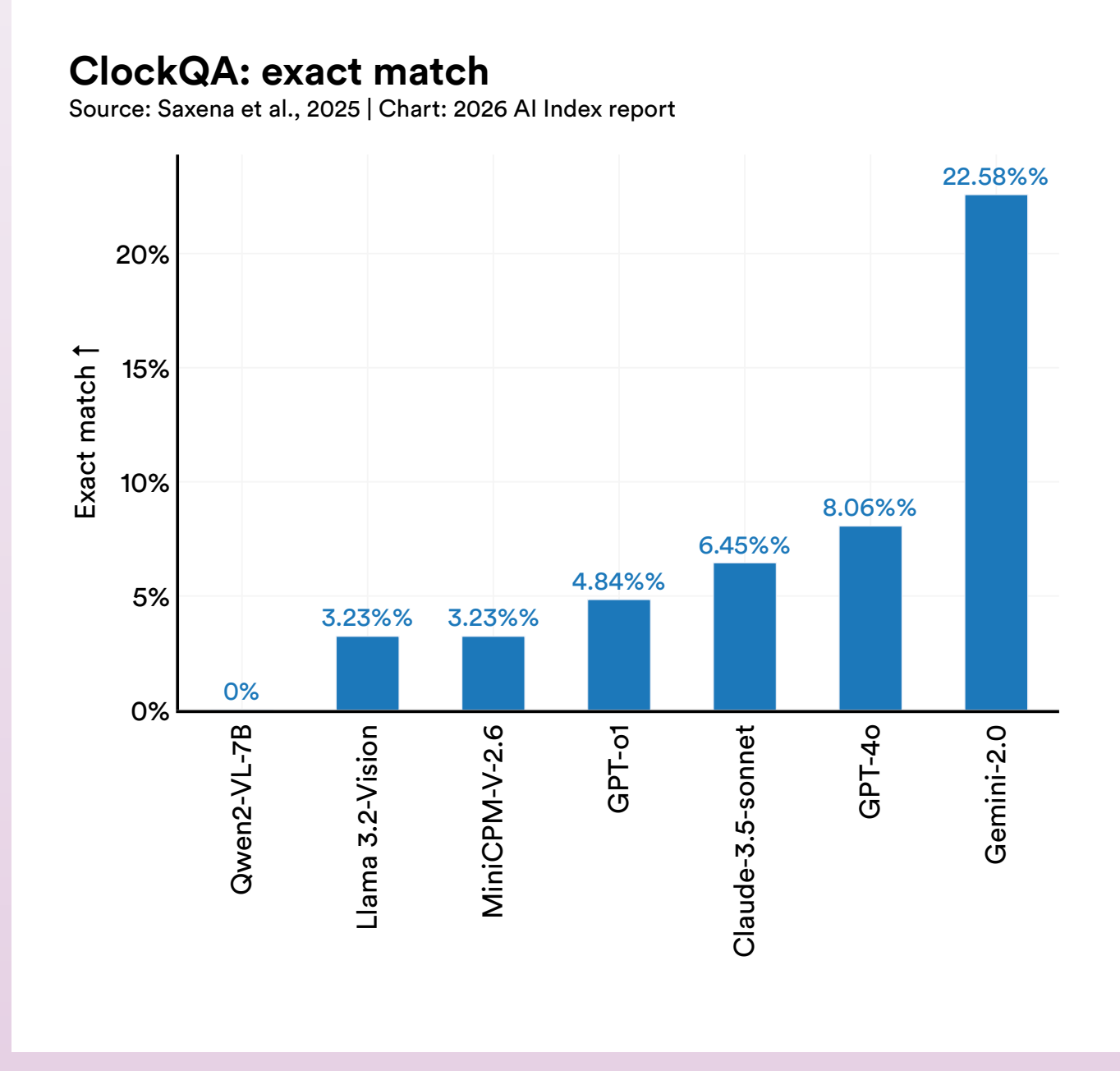

Figure 2.4.6

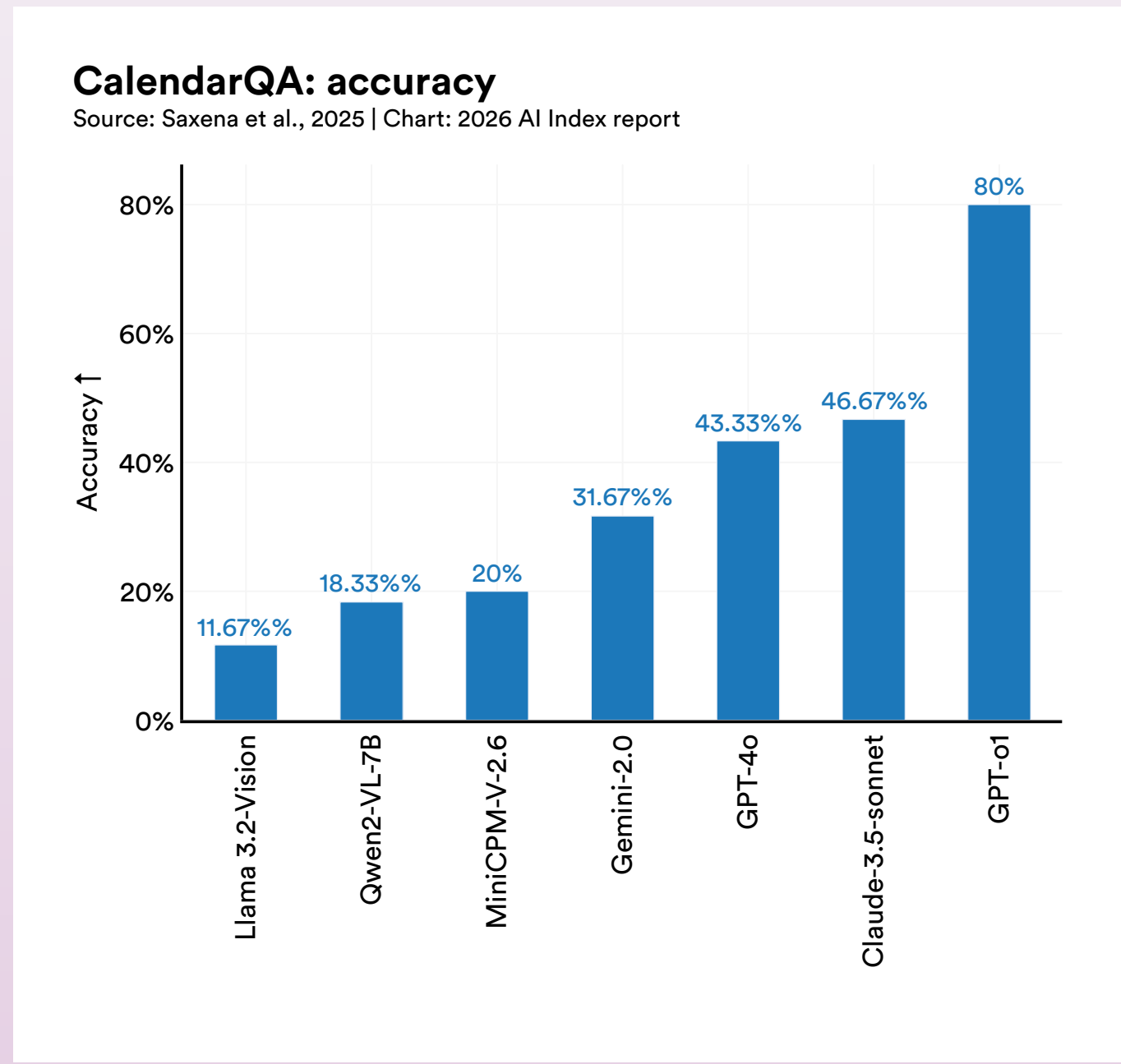

Figure 2.4.7

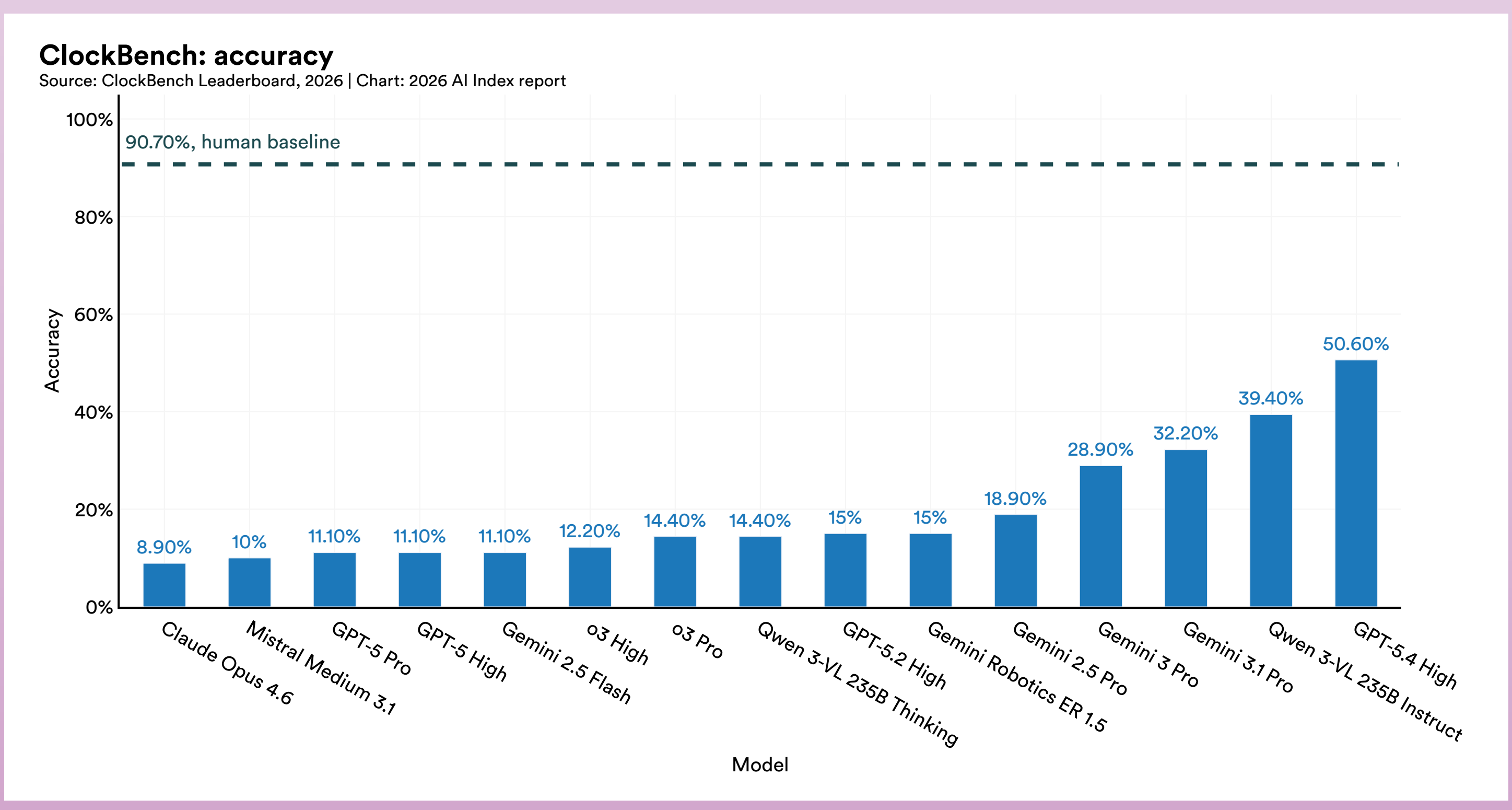

Figure 2.4.8[18]

18 Data source: https://clockbench.ai.

## Planning

In addition to general reasoning, the AI Index tracks planning benchmarks that assess models' ability to sequence actions over multiple steps to achieve a goal. Models have to keep track of what has already happened, avoid invalid actions, and maintain consistency even as problems get longer and more complex. Unlike single-shot reasoning questions, planning evaluations can expose failures that only emerge over longer horizons, including compounding errors or forgetting earlier constraints. Which benchmark is used to measure these capabilities matters, as different tasks surface different types of failures.

### PlanBench

Classical planners like LAMA search systematically through possible states and produce correct plans when they find a solution. Language models instead generate plans based on learned patterns, which means they can produce plausible sequences that can have invalid steps or miss constraints.

PlanBench evaluates end-to-end planning by prompting models to generate a full plan from a structured problem description across several planning domains. A domain is a type of problem with its own rules and goals, such as stacking blocks in a specific order, navigating routes, or transporting packages between locations. The benchmark reports performance as the number of tasks solved in each domain, with up to 45 tasks per domain, compared to LAMA as a classical planning baseline.

No single model leads across every domain (Figure 2.4.9). Under standard planning, LAMA leads in several domains, including Miconic (45/45), Rovers (34/45), and Transport (33/45). In more structured domains such as Childsnack and Spanner, frontier models match or exceed LAMA, with GPT-5 reaching 38/45 on Childsnack and 45/45 on Spanner.

When task descriptions are scrambled to disguise their structure, performance decreases for most models in several domains, though the effect depends on the domain and model (Figure 2.4.10). For example, DeepSeek R1 falls to 3/45 on Blocksworld and 0/45 on Floortile and Sokoban. Similarly, GPT-5 declines to 12/45 on Blocksworld and 7/45 on Sokoban.

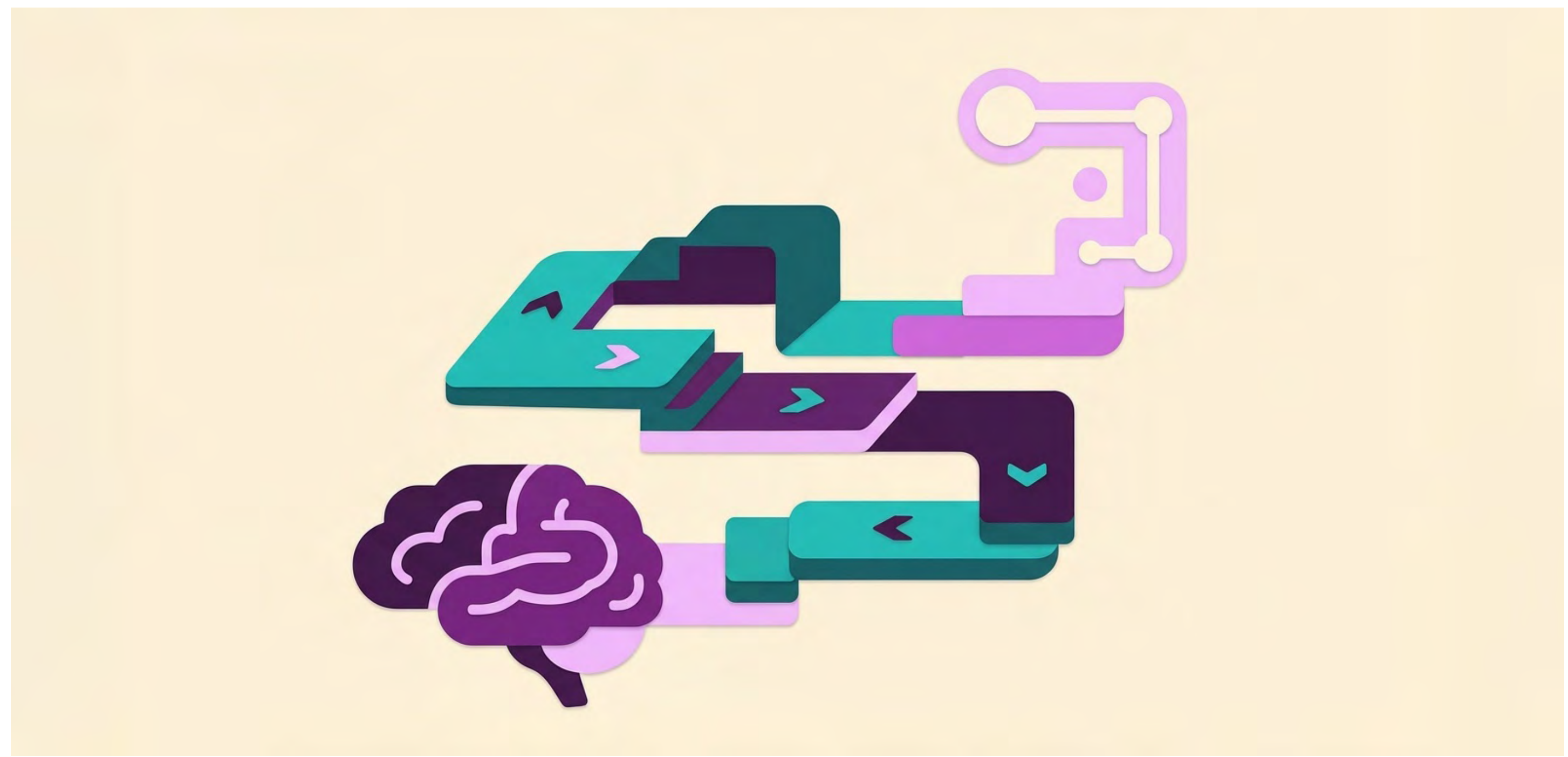

## PlanBench: task solved on standard planning

Source: Corrêa et al., 2025 | Chart: 2026 AI Index report

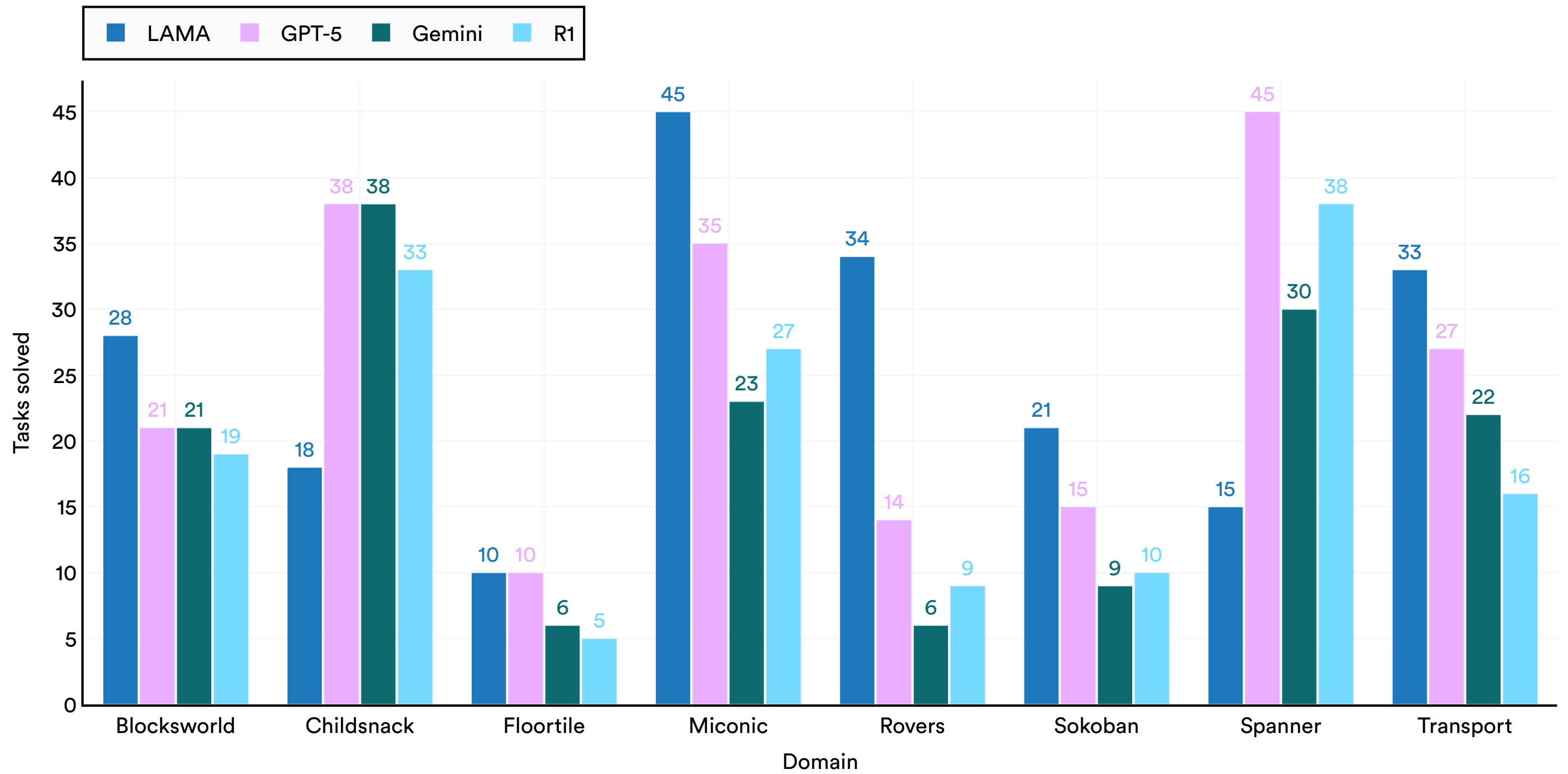

Figure 2.4.9[19]

## PlanBench: task solved on obfuscated planning

Source: Corrêa et al., 2025 | Chart: 2026 AI Index report

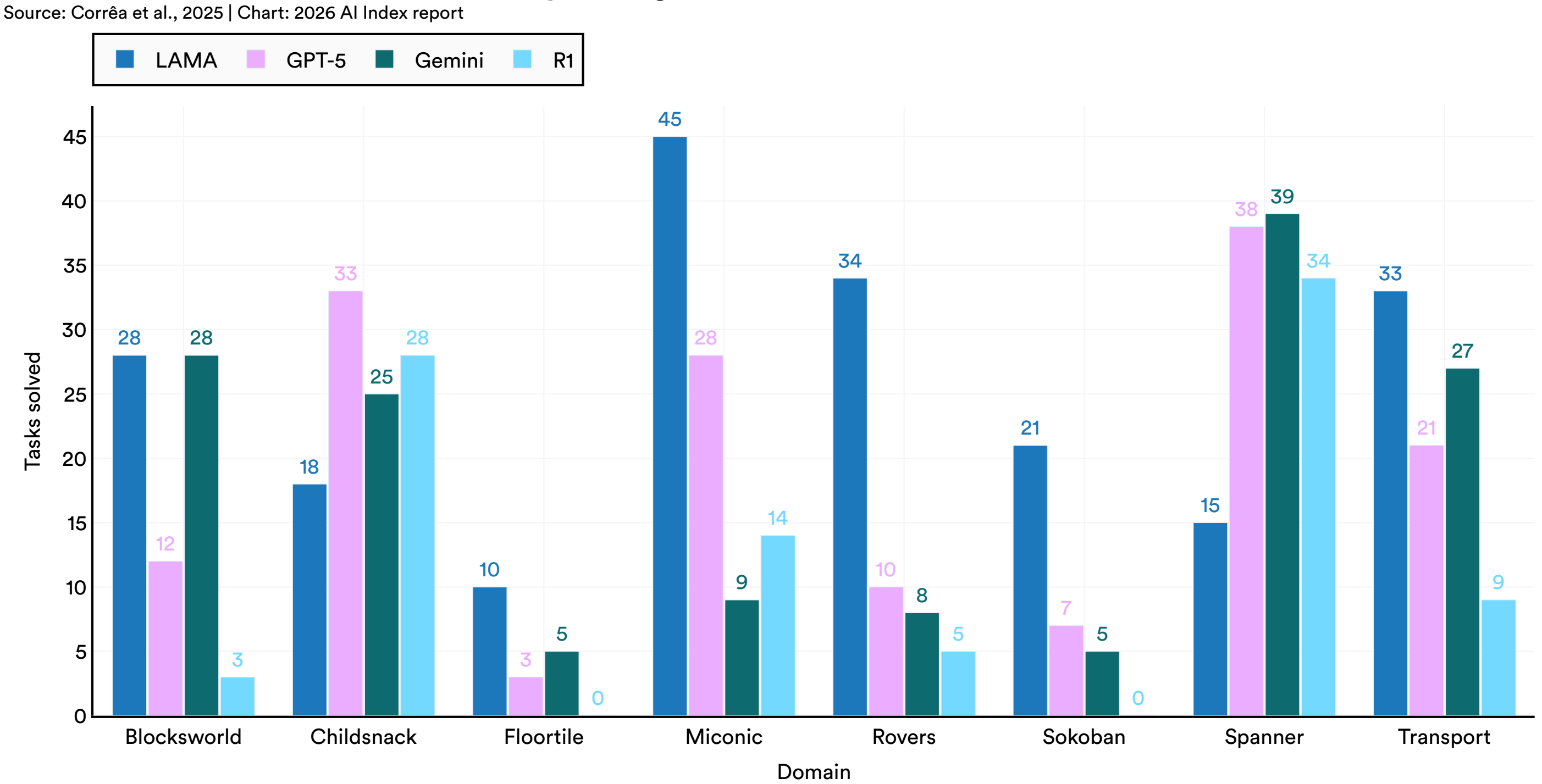

Figure 2.4.10[20]

19 Data source: https://arxiv.org/pdf/2511.09378.

20 Data source: https://arxiv.org/pdf/2511.09378.

# 2.5 Performance in Specific Domains

As AI models have improved on general reasoning and knowledge benchmarks, attention has shifted to how well they perform on tasks requiring specialized expertise. The benchmarks in this section test models in four professional and academic domains: coding, mathematics, finance, and legal reasoning. Each has its own vocabulary, conventions, and standards for what counts as a correct and comprehensible answer. Many of these benchmarks are new, reflecting growing demand for domain-specific evaluation. Unless otherwise noted, the results reported below reflect model performance as of early 2026.

## Software

Coding benchmarks test whether models can go beyond answering questions about code and actually write, debug, and ship working software. The tasks in this section range from resolving real GitHub issues to building full web applications from scratch, reflecting a shift in evaluation toward measuring what models can deliver end to end rather than in isolated snippets.

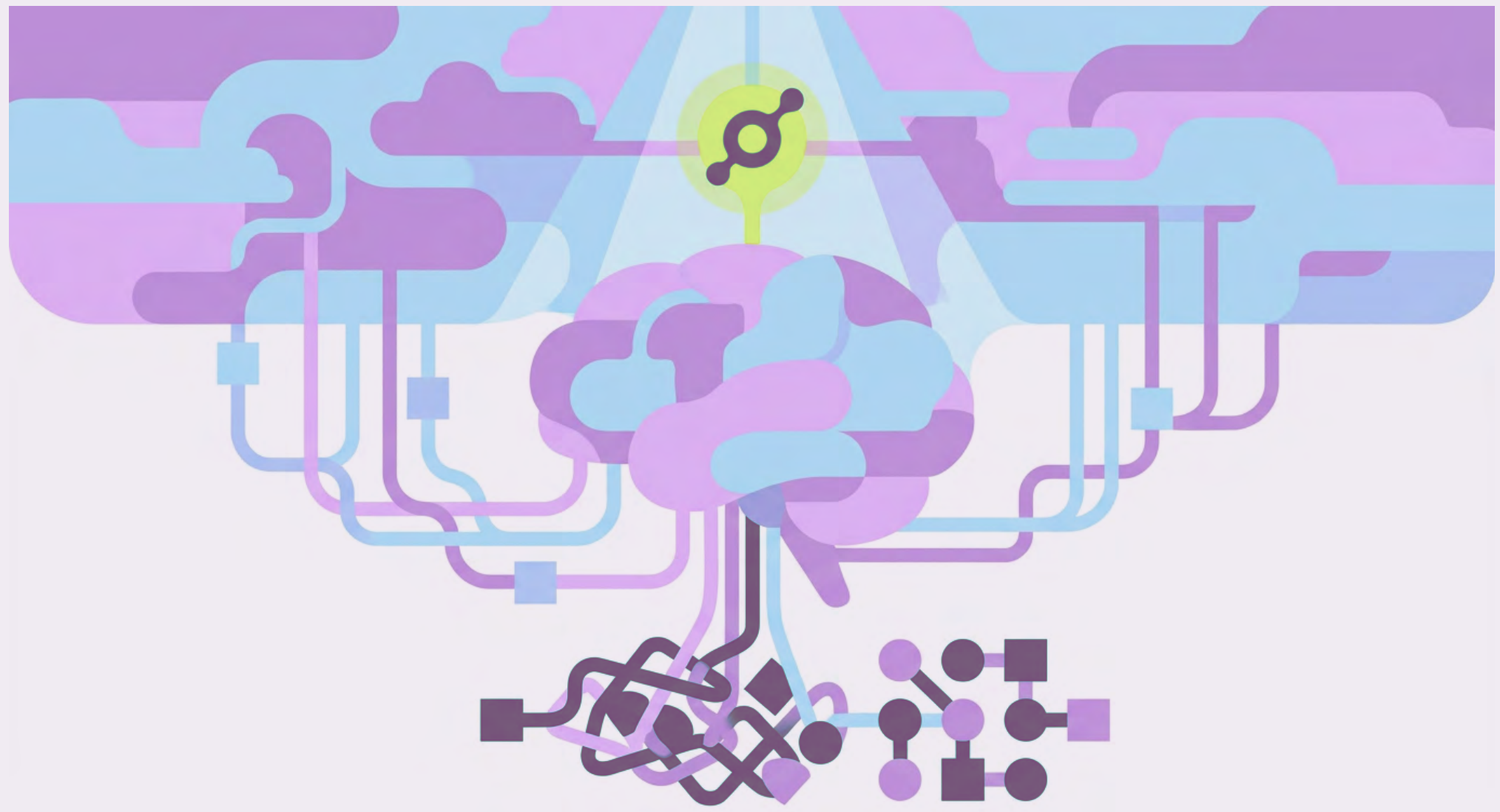

### SWE-bench

SWE-bench evaluates models on their ability to resolve real-world software issues collected from GitHub. Each task gives the model a codebase and an issue description, and the model has to produce a working patch. SWE-bench Lite is a smaller, more accessible subset while SWE-bench Verified uses human-validated issues to ensure more consistent and accurate grading.

On SWE-bench Verified, top models are tightly clustered in the low-to-mid 70s (Figure 2.5.1). As of February 2026, Claude 4.5 Opus (high reasoning) led at approximately 76.8%, with several others including KimiK2.5, GPT-5.2, and Gemini 3 Flash (high reasoning) grouped between 70% and 76%. This is a pattern seen across several benchmarks in this chapter, where high-performing models score within a few percentage points of each other.

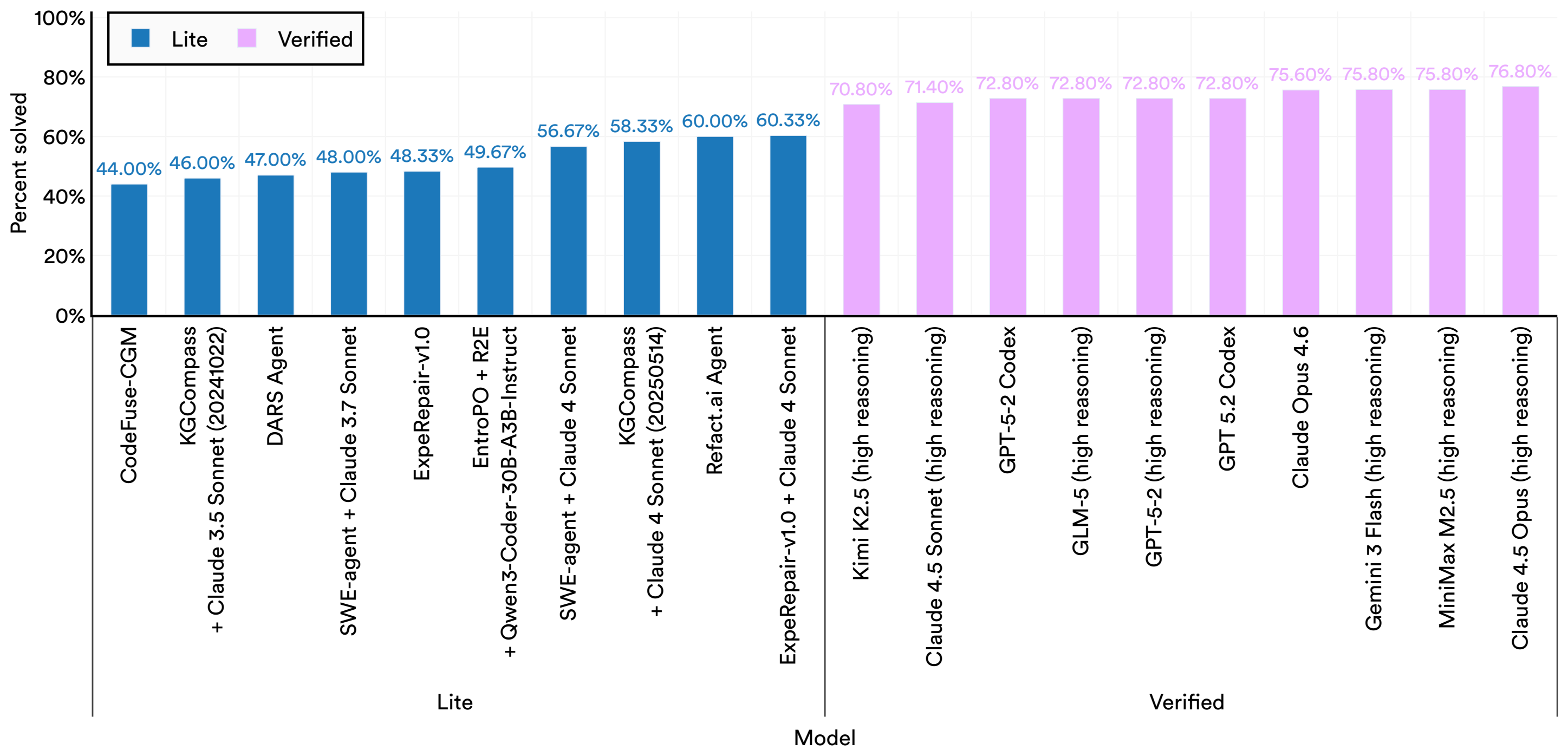
SWE-bench: percent solved
Source: SWE-bench Leaderboard, 2026 | Chart: 2026 AI Index report

Figure 2.5.1[21]

## Terminal-Bench

Terminal-Bench is a benchmark for testing AI agents in real terminal environments. It evaluates how well agents can autonomously handle real-world, end-to-end tasks, from compiling code to training models and setting up servers. These are the kinds of tasks a developer might do in a day of work, and it requires an agent to chain together multiple steps without human guidance.

Accuracy on Terminal-Bench 2.0 has significantly improved over the past year, increasing from 20% in February 2025 to 77.3% in early 2026 (Figure 2.5.2).

21 This chart shows the top 10 models for SWE-bench Verified and Lite as of February 2026. For Verified, only results using the mini-SWE-agent-v2 filter are included. This means all models were tested under the same agent workflow, so differences in scores reflect the underlying model rather than differences in the surrounding system. Data source: https://www.swebench.com/index.html.

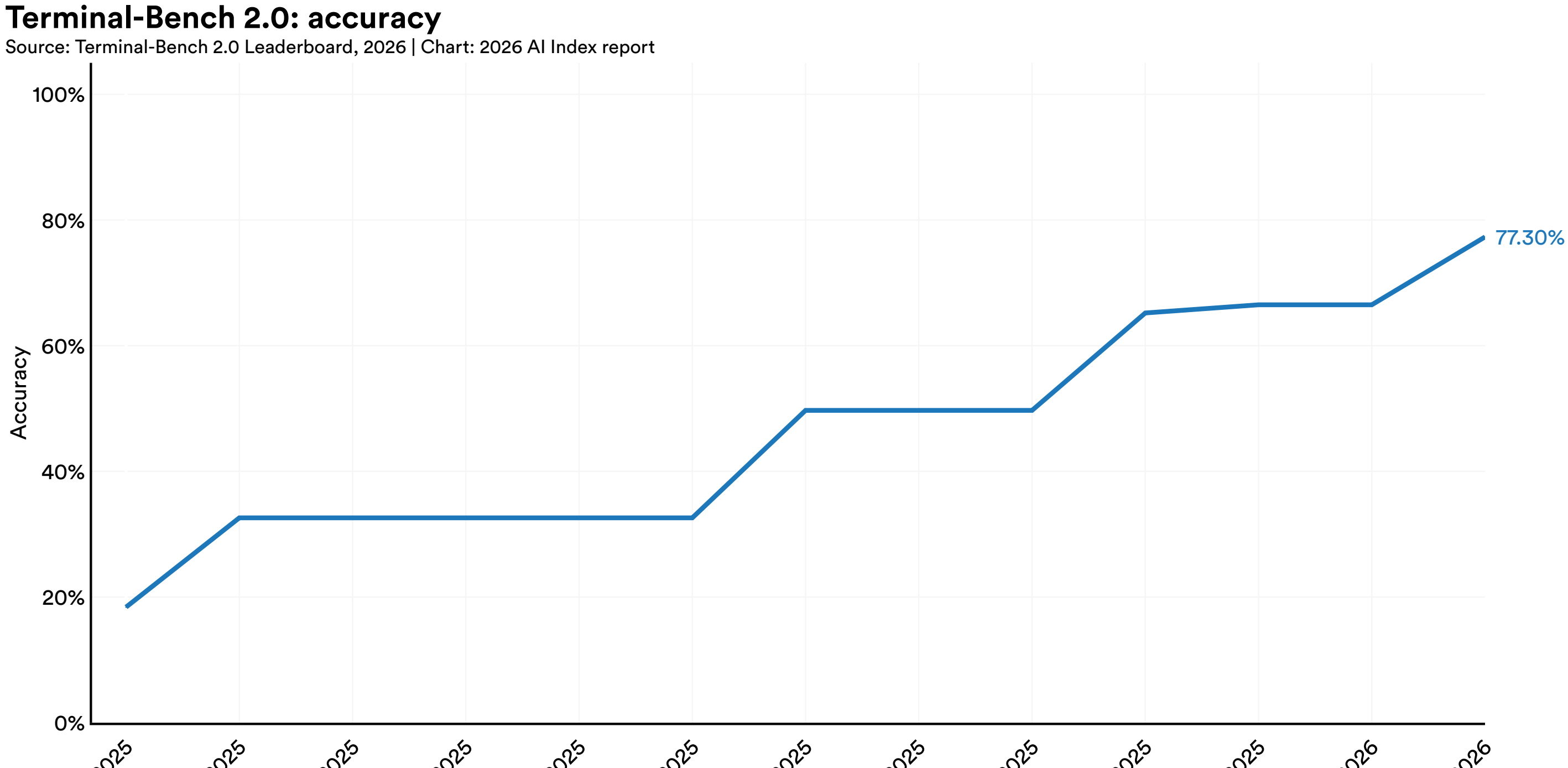

Figure 2.5.2[22]

## Vibe Code Bench

Vibe Code Bench is the first benchmark designed to test whether AI models can autonomously build complete, end-to-end web applications from scratch. Rather than measuring coding assistance, it evaluates real software delivery and sees if a model can take a prompt and produce a functional application.

Across models, performance varies quite a bit (Figure 2.5.3). Claude Opus 4.6 (Nonthinking) leads at 56.5%, followed by GPT 5.2 at nearly 47%. Scores drop after GPT 5.3 Codex (41.4%) to under 30%, with several models falling below 15%. The spread between the top and bottom models is about 46 percentage points, and even the leading model solves only about half of the tasks, suggesting that autonomous application building remains a difficult task.

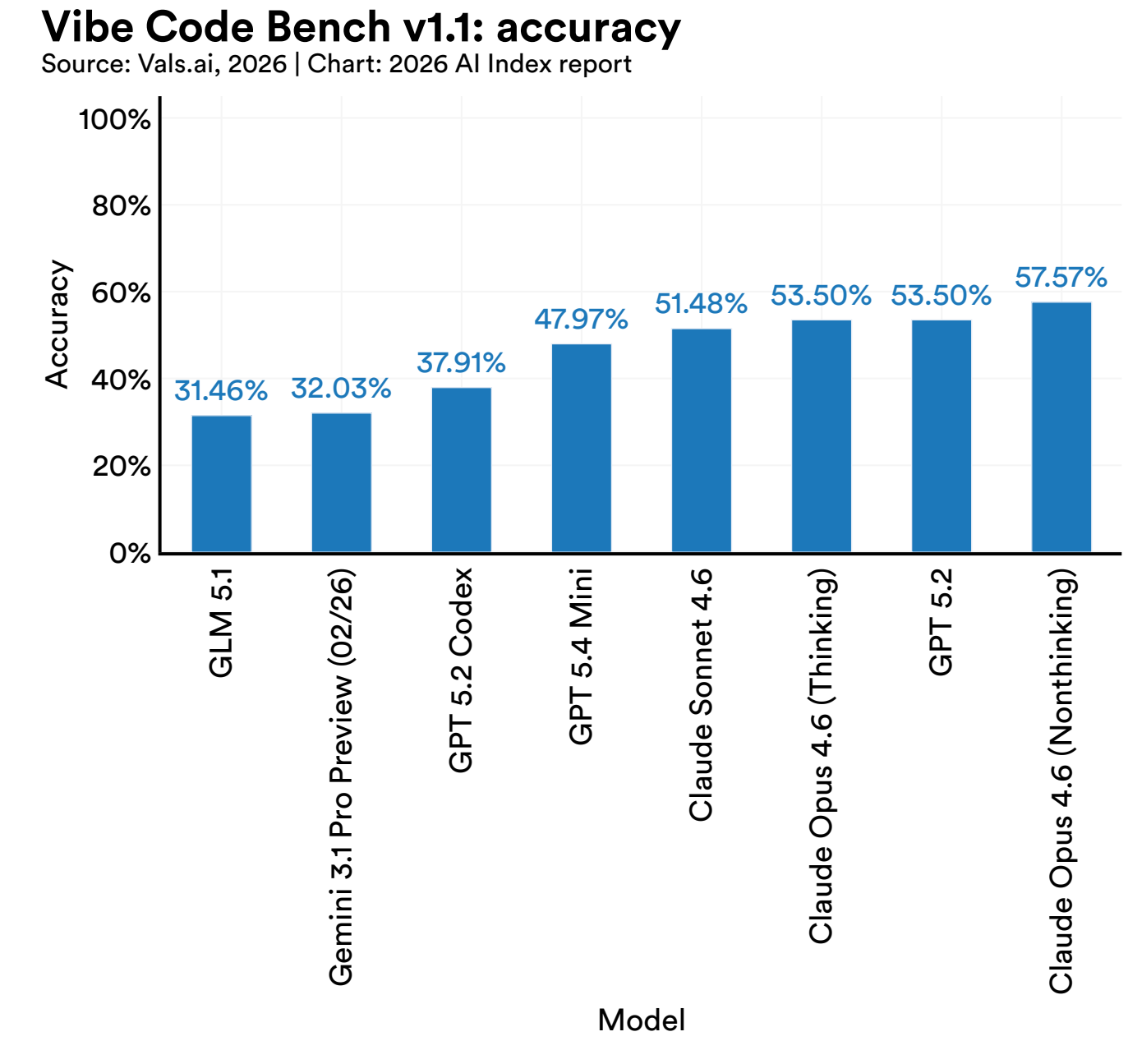

Figure 2.5.3[23]

22 Data source: https://www.tbench.ai/leaderboard/terminal-bench/2.0.

23 Data source: https://www.vals.ai/benchmarks/vibe-code.

## Mathematics

Beyond coding and language tasks, mathematics has become a key testing ground for model reasoning. The benchmarks in this section range from competition-level problem solving to formal proof writing.

### FrontierMath

FrontierMath is a benchmark introduced by Epoch AI that features hundreds of original, exceptionally challenging mathematical problems. The problems are designed to test genuine mathematical reasoning rather than pattern recognition, and even experienced mathematicians may need hours or days to solve them.

Since 2024, accuracy on FrontierMath Tier 4 has risen from near 0% to 31.3%, with GPT-5.2 Pro (Web App) leading by the end of 2025 (Figure 2.5.4). The benchmark is designed to stay difficult, so even with this steep climb in a short time, the best models still fail on roughly two out of three problems at the hardest tier.

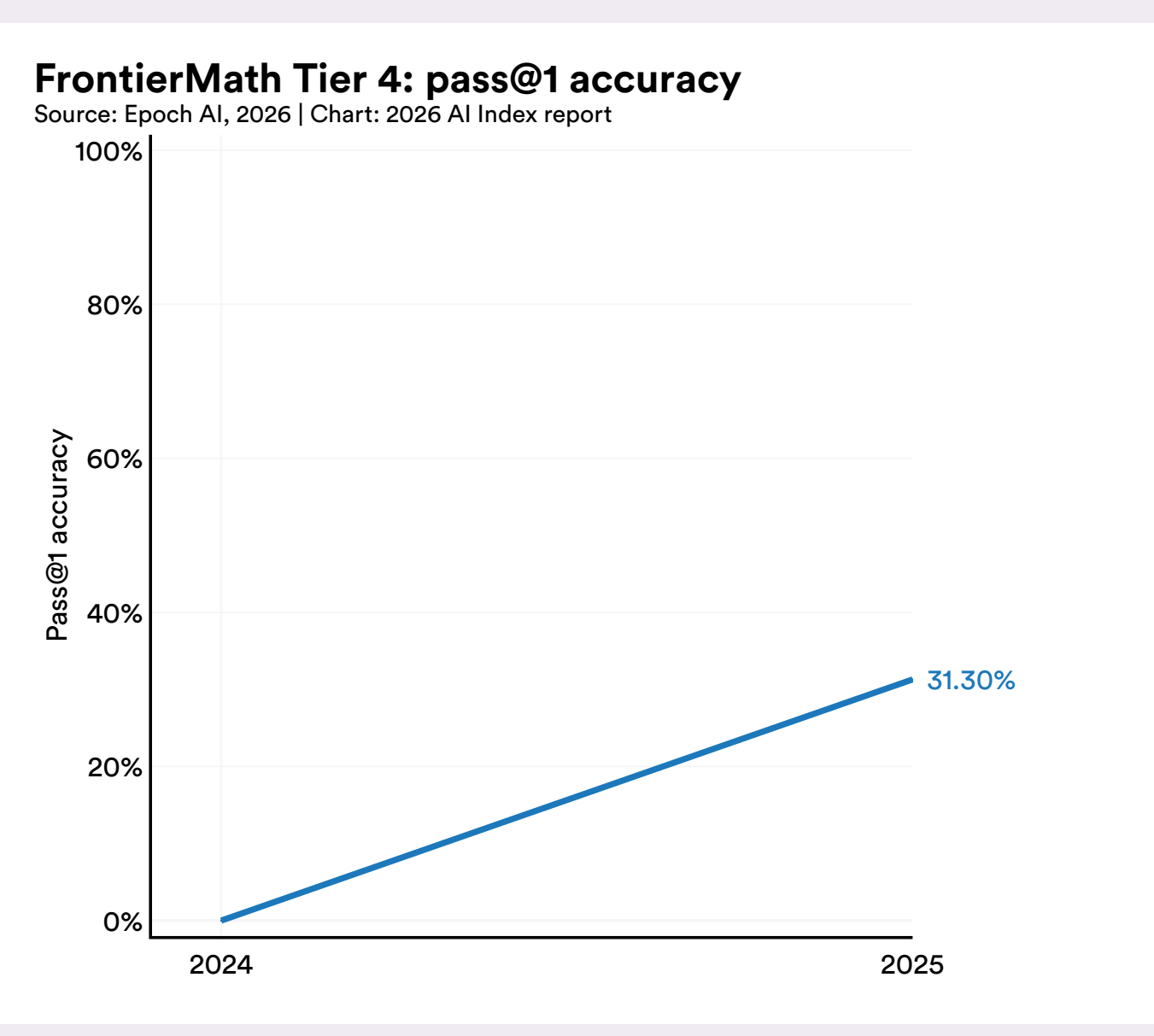

Figure 2.5.4[24]

### MathArena

MathArena is a rolling benchmark that uses newly released math contests to test models on fresh, competition-style problems. It draws from well-known competitions at the high school and olympiad level, including AIME, HMMT, USAMO, the International Mathematical Olympiad (IMO), and Project Euler, running models soon after each one to reduce the risk of training-data contamination. Numerical answers are auto-graded, while human graders score written proofs with results being posted on a public leaderboard.

Accuracy on MathArena has increased from about 83% in November 2025 to 97% in December 2025 (Figure 2.5.5). On answer-based problems, leading models reach or surpass the level of top human contestants. However, on proof-based tasks, they still perform well below humans when asked to produce rigorous, step-by-step mathematical proofs. Getting the right answer and showing the reasoning behind it remain distinct challenges for current systems.

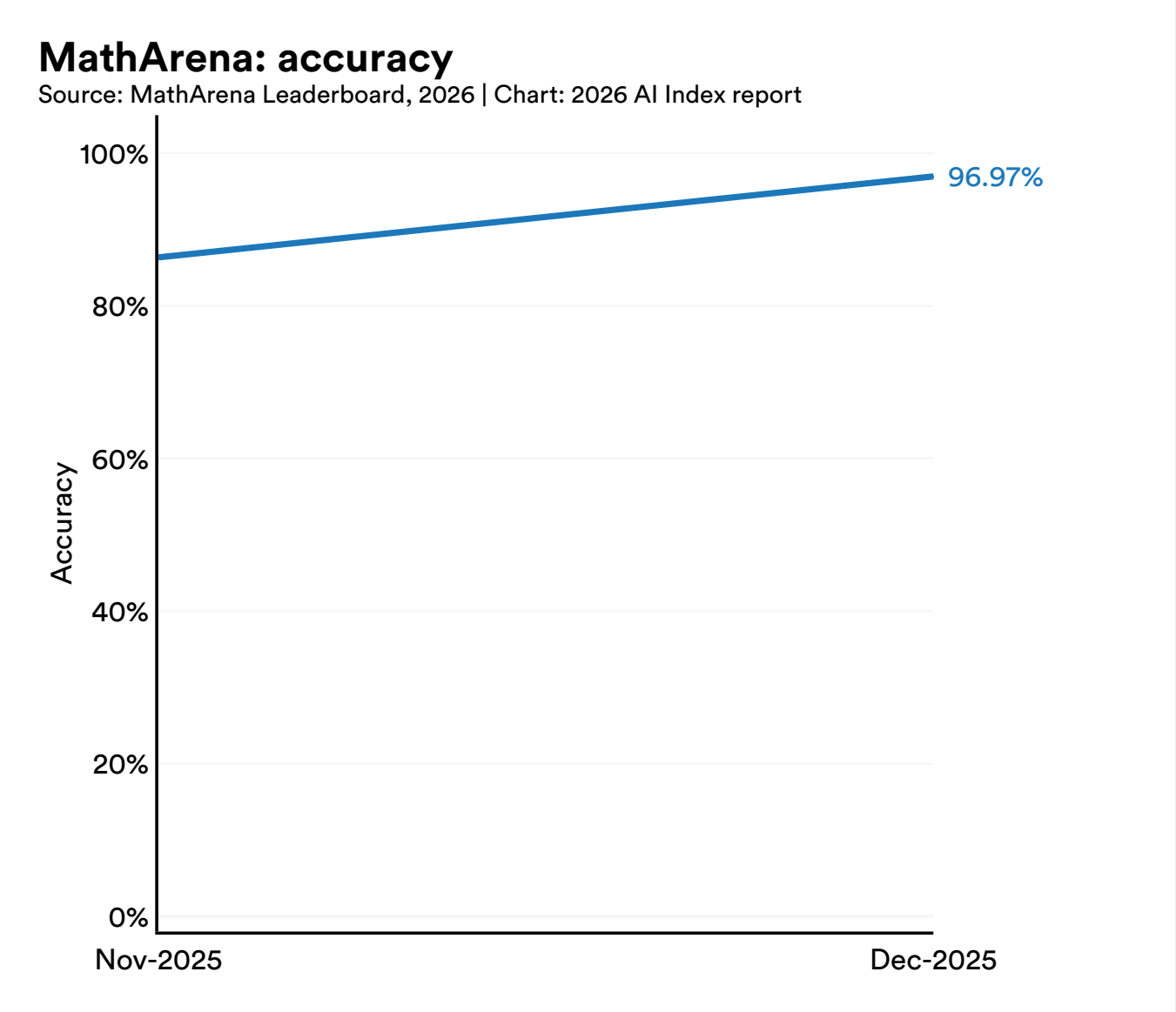

Figure 2.5.5[25]

24 Data source: https://epoch.ai/benchmarks/.

25 Data source: https://matharena.ai/?comp=hmmt--hmmt_feb_2026&view=problem.

**HIGHLIGHT:**

## Theorem Proving

In mathematics, getting the right answer is only one part of the challenge. A correct result backed by flawed reasoning would earn little credit at a competition or in a journal. Theorem proving, the process of constructing a rigorous, step-by-step argument for why a result must be true, remains one of the hardest tasks for AI systems. Until recently, even frontier models struggled to produce proofs capable of passing expert review.

As covered in last year's AI Index, DeepMind's AlphaProof and AlphaGeometry 2 solved four of six problems at the 2024 International Mathematical Olympiad (IMO), winning a silver medal with 28 points. That result required experts to translate problems into formal languages like Lean and took days of computation. In 2025, Gemini Deep Think solved five of six problems and scored 35 points, winning a gold medal, while working end to end in natural language within the 4.5-hour competition time limit (Luong and Lockhart, 2025). The jump from silver to gold in a single year, with a far simpler pipeline, marks one of the fastest capability gains in competitive mathematics.

IMO-Bench (Luong et al., 2025) is a new benchmark suite designed to measure whether that kind of progress is genuine reasoning or just better answer guessing. It includes three components. IMO-AnswerBench tests models on 400 Olympiad-style problems across algebra, combinatorics, geometry, and number theory, with verifiable short answers. IMO-ProofBench evaluates whether models can produce rigorous step-by-step proofs for 60 problems ranging from pre-IMO to full IMO difficulty. IMO-GradingBench provides a dataset with 1,000 examples of solutions and human-graded proofs to support the development of automated proof grading systems.

Grading mathematical proofs has traditionally required human experts, which limits how many models and solutions can be evaluated at scale. On IMO-ProofBench, scores assigned by an automated grading system closely track those given by human experts, with Pearson correlation of 0.96 on basic and 0.93 on advanced problems (Figure 2.5.6). That level of agreement indicates that automated grading could be a reasonable stand-in, though the benchmark authors recommend human verification for high-stakes results.

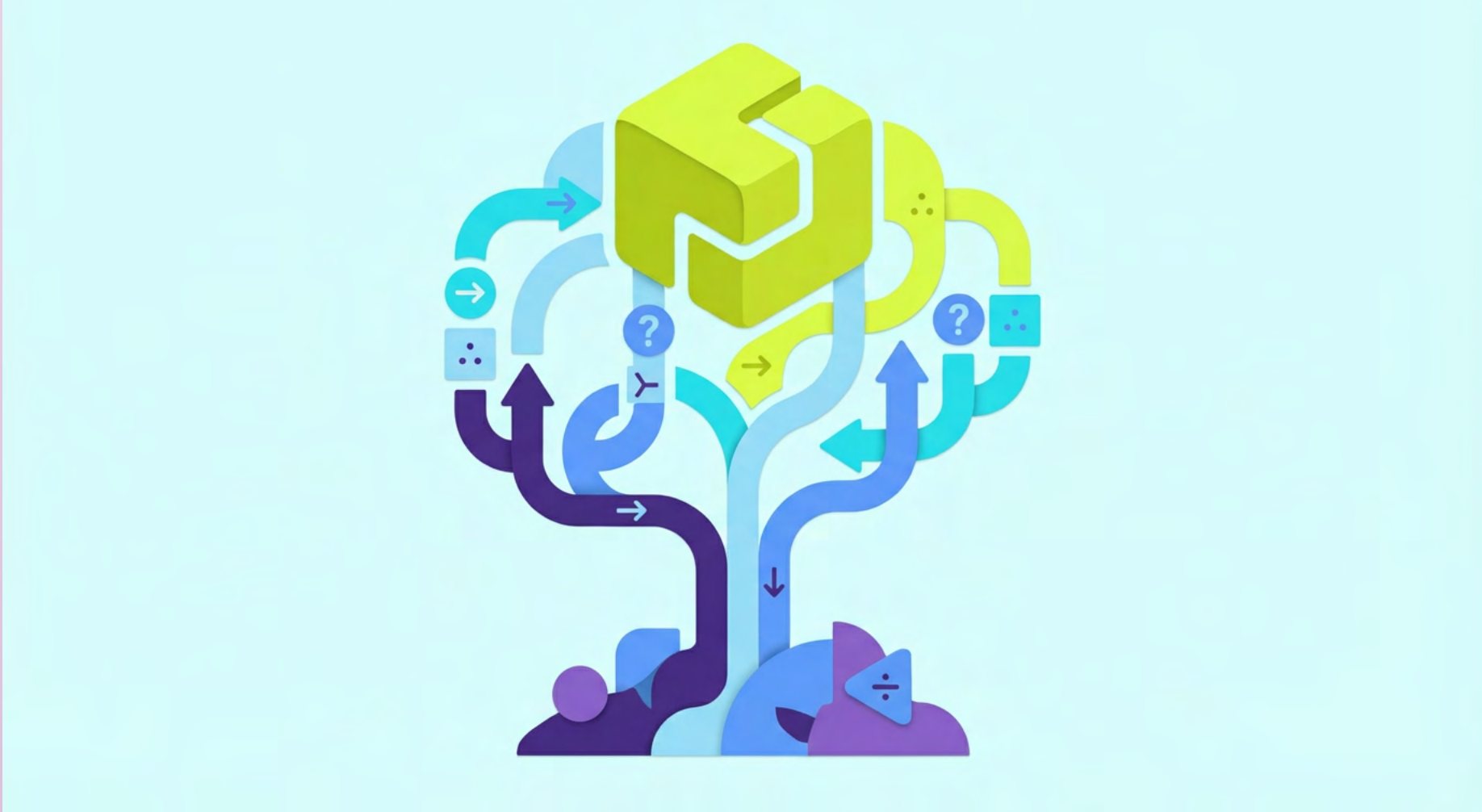

With that grading approach validated, the benchmark results reveal the extent of the gap between models (Figure 2.5.7). Aletheia leads at 91.9%, followed by Gemini 3 Deep Think at 76.7% and Gemini Deep Think (IMO Gold) at 65.7%. From there, scores drop quite a bit. GPT-5.2 Thinking (high) reaches 35.7%, Gemini 3 Pro scores 30%, and GPT-5.1 falls to 7.1%. The spread between the top and bottom models is about 85 percentage points. A breakdown by problem source in the IMO-Bench paper suggests that some of these scores may also reflect familiarity with existing competition problems rather than general reasoning ability, reinforcing a pattern seen with MathArena. Producing correct answers and rigorous proofs remain very different tasks, with most models far more performant on the former.

**HIGHLIGHT:**

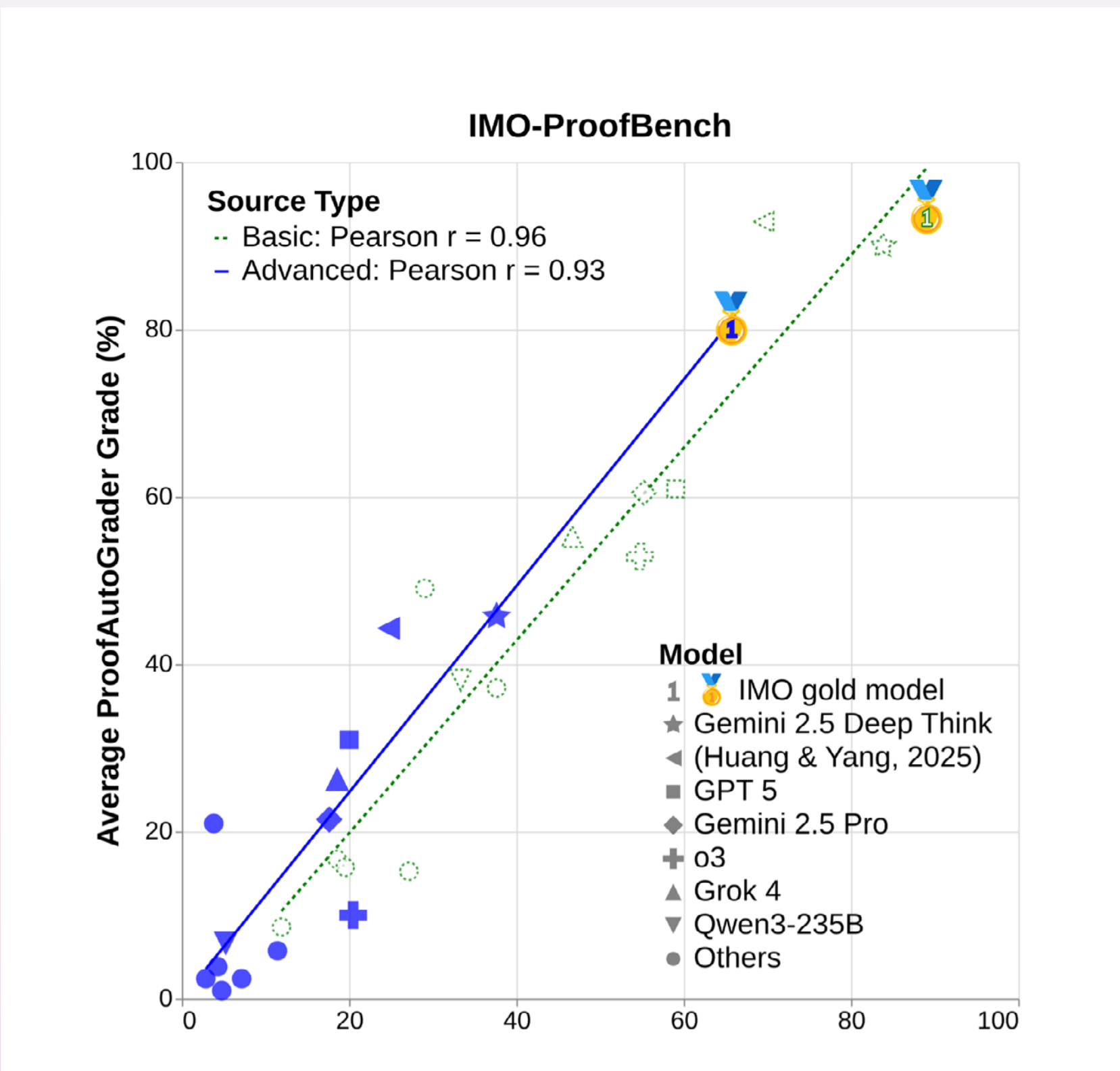

Source: [Luong et al., 2025]

Figure 2.5.6

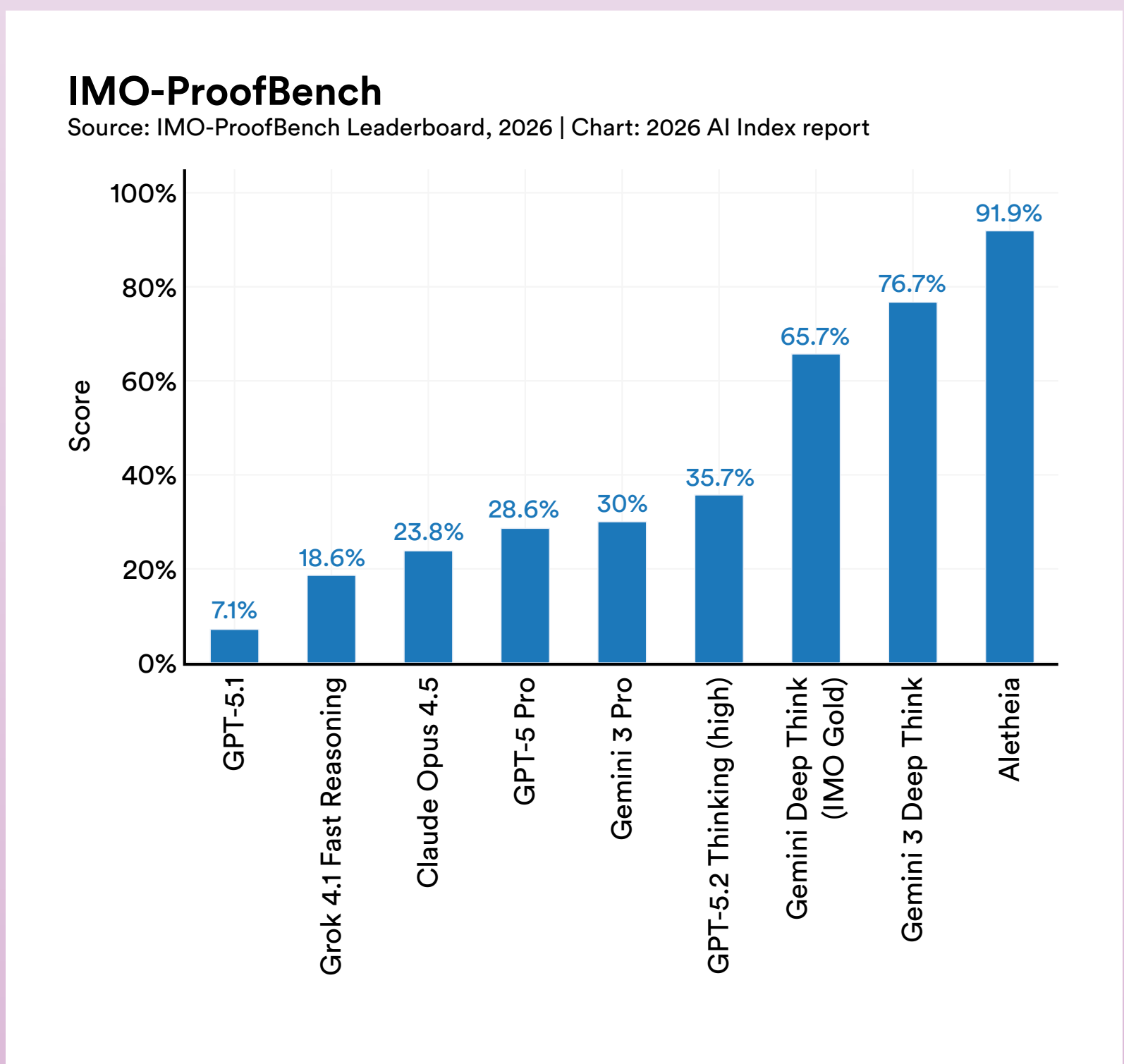

Figure 2.5.7[26]

26 Data source: https://imobench.github.io/.

## Finance

This section covers benchmarks designed to evaluate AI systems on finance-specific tasks. Unlike general reasoning benchmarks, these tests require models to handle domain-specific language, extract structured information from financial documents, and apply professional judgment in areas including tax law, the mortgage process, and financial analysis.

### TaxEval

TaxEval v2 is a benchmark designed to test how well models handle challenging tax-related questions. It contains over 1,500 expert-verified questions developed with input from tax and finance professionals, covering numerical reasoning, semantic analysis, problem solving, and application of compliance rules. Models are scored on two dimensions: whether the answer is factually correct and whether the step-by-step reasoning is clear and expert-like.

Performance on TaxEval v2 shows only a small difference across models (Figure 2.5.8). All 15 top models fall within a 3 percentage point range, from 77.1% (Claude Sonnet 4.6) to 74% (Claude 3.7 Sonnet Thinking).

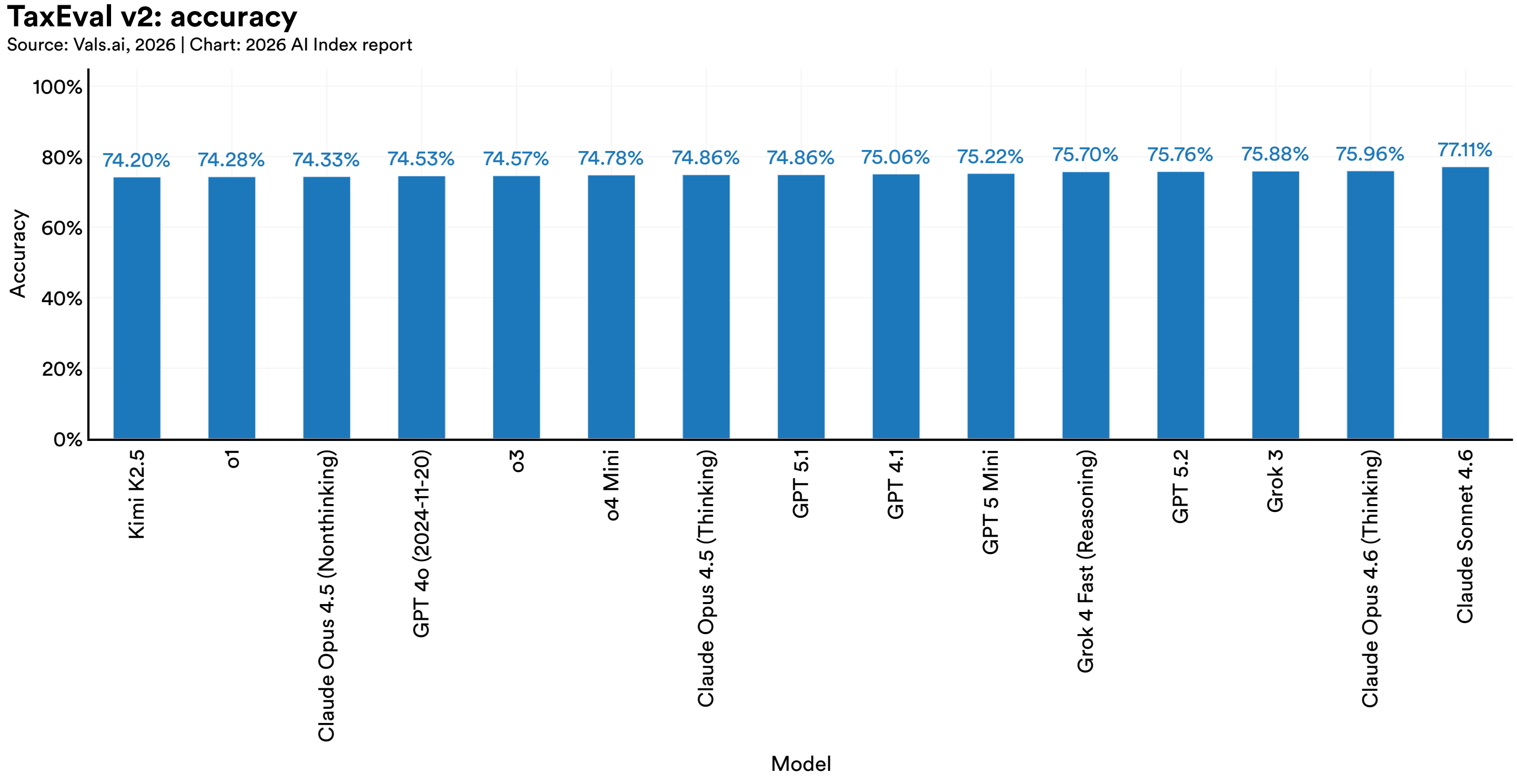

Figure 2.5.8[27]

27 Data source: https://www.vals.ai/benchmarks/tax_eval_v2.

## MortgageTax

MortgageTax evaluates how well models can extract structured information from real mortgage tax certificates, using both text and document images. The task involves two types of extraction: Semantic extraction asks the model to identify fields like year, parcel number, and county, while numerical extraction requires computing the annualized amount due. The dataset includes 1,258 documents split across public validation, private validation, and held-out test sets.

Scores on MortgageTax follow a similar pattern to TaxEval, with the top 15 models grouping within a narrow performance band (Figure 2.5.9). Gemini 3.1 Pro Preview leads at 69.4%, and GPT 4.1 is at the bottom of the group at 65.9%, a difference of about 3.5 percentage points. While several Gemini models occupy the top positions, the overall accuracy level does not reach 70%, which suggests that models are not yet entirely or reliably able to extract and compute financial information from document images.

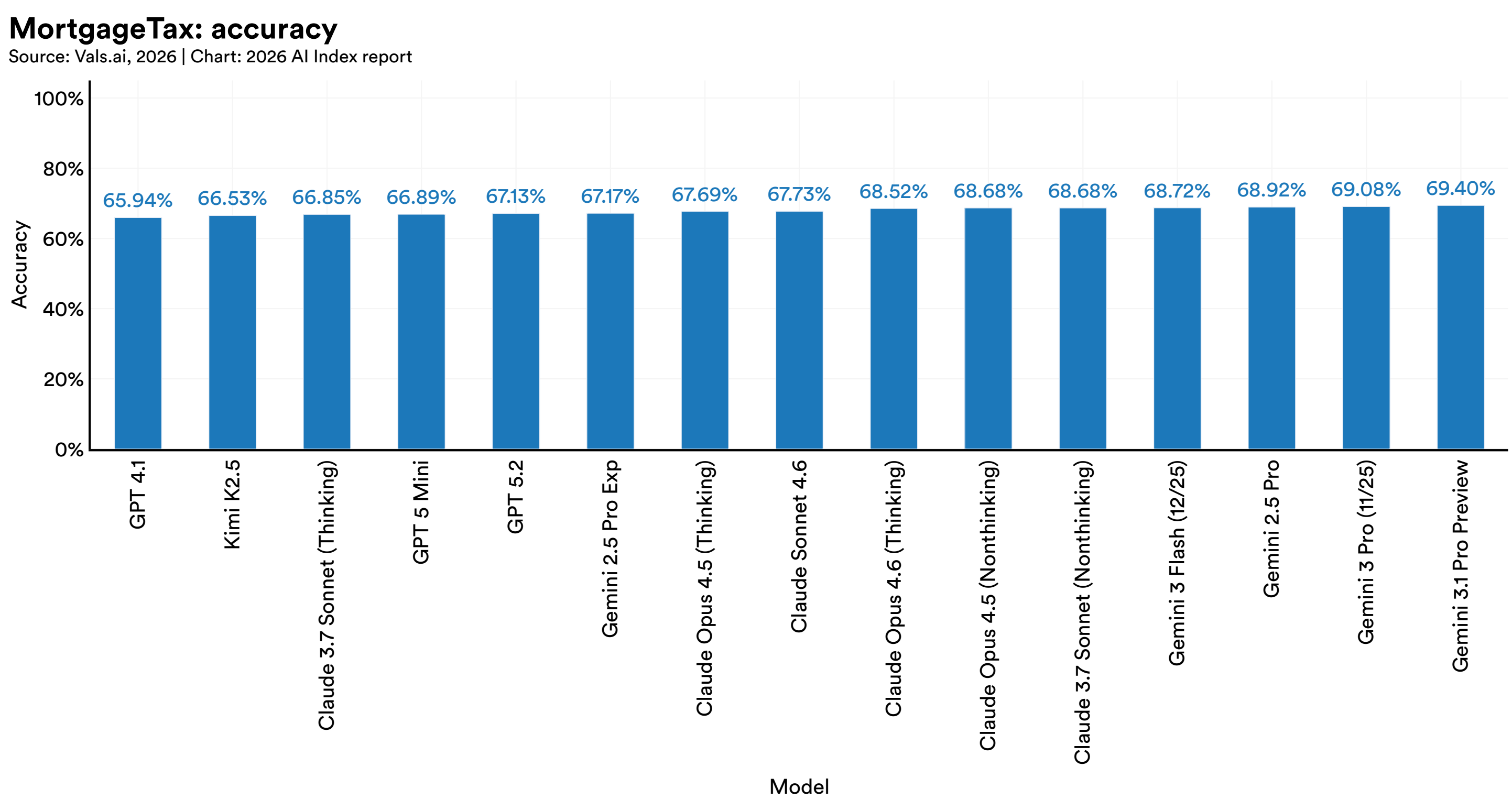

Figure 2.5.9[28]

28 Data source: https://www.vals.ai/benchmarks/mortgage_tax.

## CorpFin

CorpFin tests whether models can comprehend and extract information from long, dense financial documents, specifically credit agreements that can exceed 200 pages. Questions span basic term extraction, numeric reasoning, summarization, cross-referencing multiple sections, and industry-specific interpretation, all developed with input from financial analysts, lawyers, and academics. Beyond factual accuracy, the benchmark evaluates whether models can navigate and make sense of long, jargon-heavy legal and financial text. It defines three tasks with different context setups—Exact Pages, Shared Max Context, and Max Fitting Context—to see how models perform depending on document access.

Similar to the other benchmarks, performance on CorpFin v2 is tightly clustered (Figure 2.5.10). Kimi K2.5 leads at 68.26%, with GPT 4.1 at the bottom at 63.05%, a spread of about 5 percentage points. As with MortgageTax, no model broke 70%.

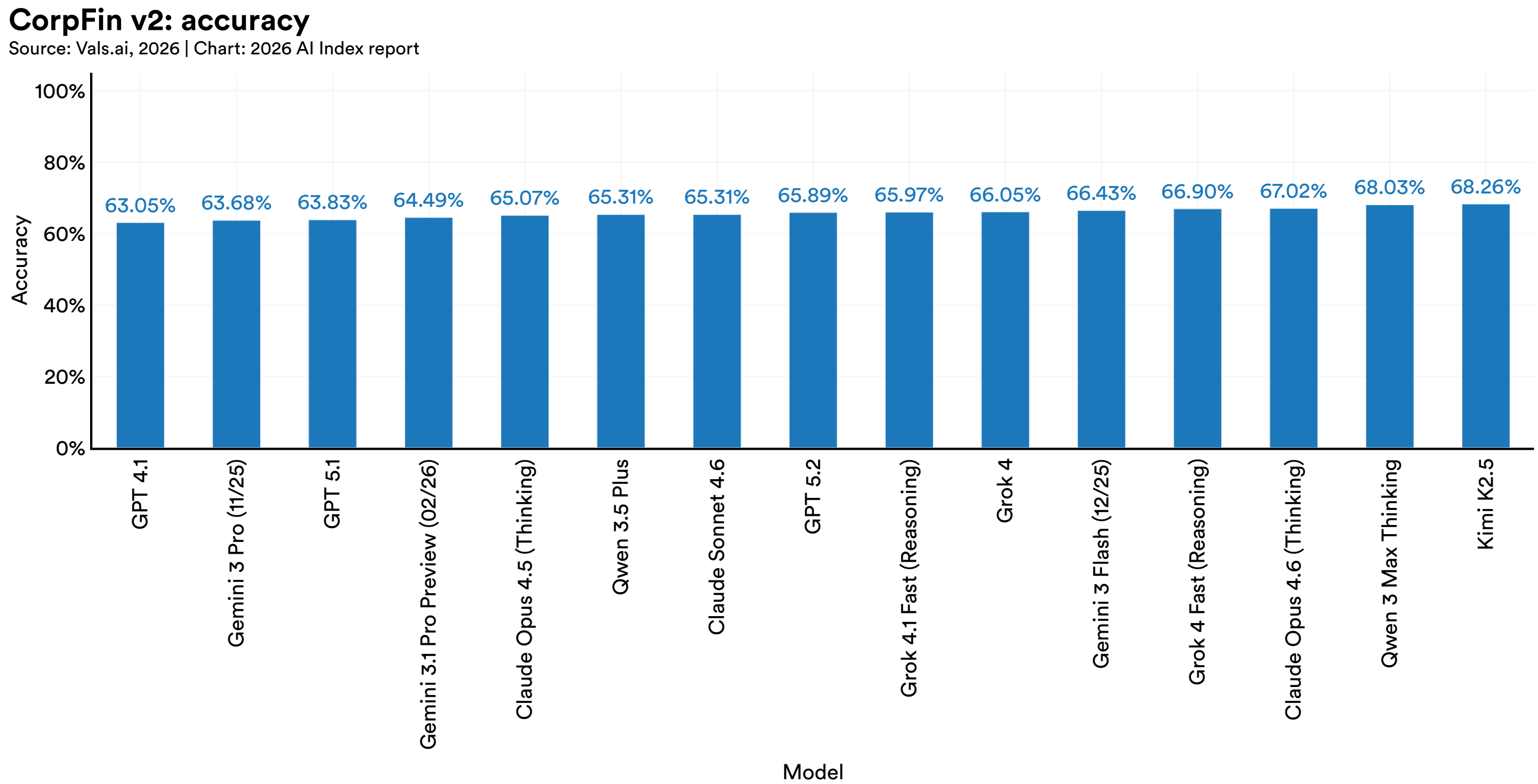

Figure 2.5.10[29]

29 Data source: https://www.vals.ai/benchmarks/corp_fin_v2.

## Finance Agent

Developed in collaboration with Stanford researchers, a Global Systemically Important Bank, and industry experts, Finance Agent evaluates AI agents' ability to perform tasks typical of an entry-level financial analyst. It includes 537 carefully crafted questions that test skills such as information retrieval, market research, and financial projections.

On Finance Agent v1.1, performance was more varied than on other finance benchmarks (Figure 2.5.11). Claude Sonnet 4.6 leads at 63.33%, and scores taper down to 50.62% for Kimi K2.5, a spread of about 13 percentage points. Even the top score sits below two-thirds accuracy, reflecting the domain-specific challenges seen across the other finance benchmarks, as well as the broader difficulty of agentic tasks, which is discussed below in Section 2.6, Agent Benchmarks.

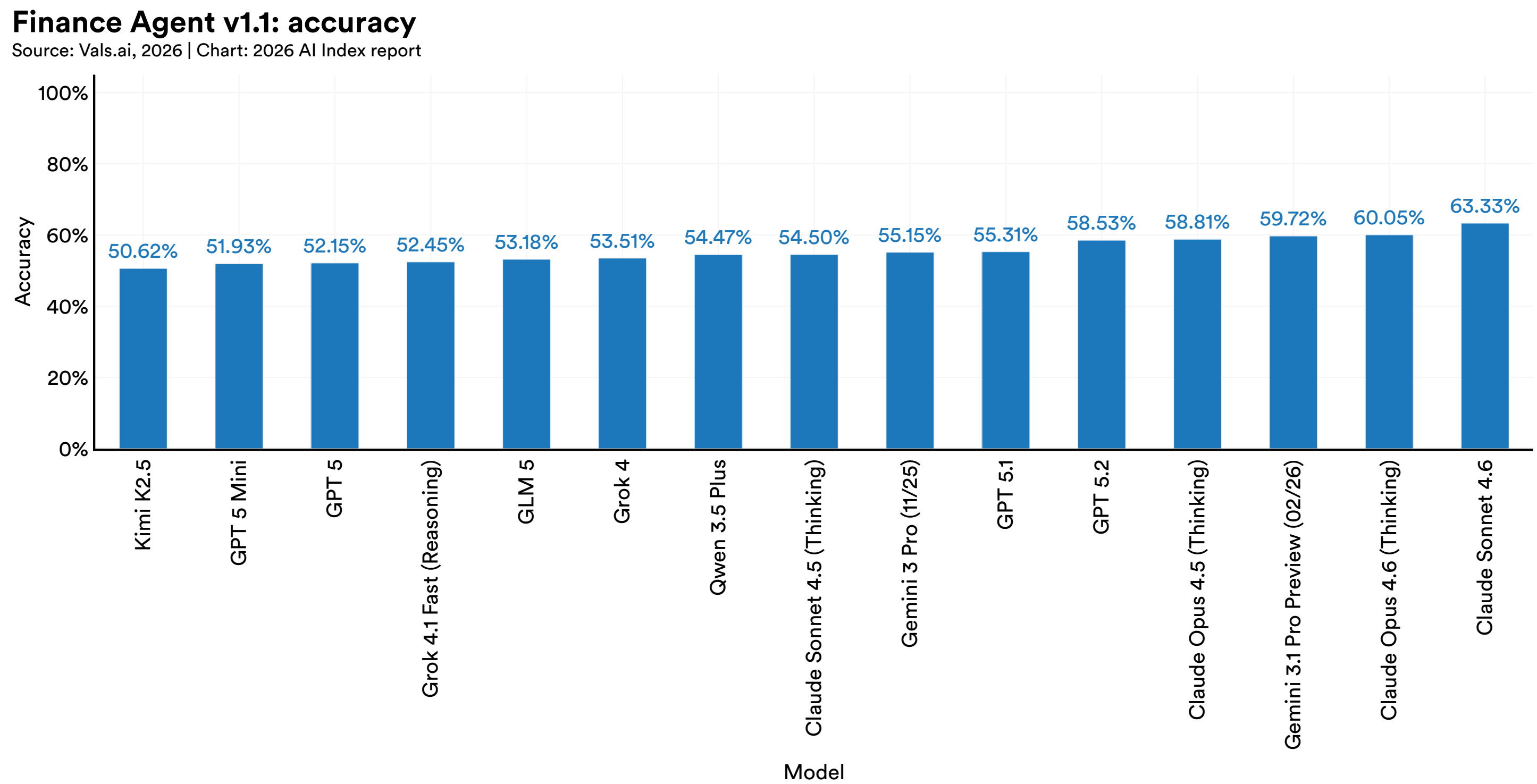

Figure 2.5.11[30]

30 Data source: https://www.vals.ai/benchmarks/finance_agent.

## Law

AI is also being evaluated in the legal domain, where tasks range from interpreting court decisions to applying rules to new fact patterns. The benchmarks covered below reflect how well models handle legal reasoning tasks that require grounding in specific documents rather than general knowledge.

### CaseLaw

CaseLaw v2 is a benchmark for evaluating LLMs on real-world litigation and legal research tasks. It uses recent United States and Canada court decisions which are dated after most models' training cutoffs and are not accessible at scale due to licensing restrictions, which helps ensure the model is reasoning over the provided documents rather than relying on memorized legal knowledge. The benchmark includes 300 validation tests and 104 test tests—spanning single-case and multicase reasoning—across seven legal reasoning dimensions, including retrieving key precedents, multidocument question answering, calculations, tables, and chronological reasoning.

GPT-5.1 leads on CaseLaw v2 at 73.4% accuracy, with GPT 4.1 following at 69.9% (Figure 2.5.12). The rest of the top 15 models fall between 62% and 66%, a sign there is meaningful room for improvement. One recurring issue is that models tend to lean on general knowledge, rather than grounding their answers in the supplied documents, even when explicitly instructed to do so.

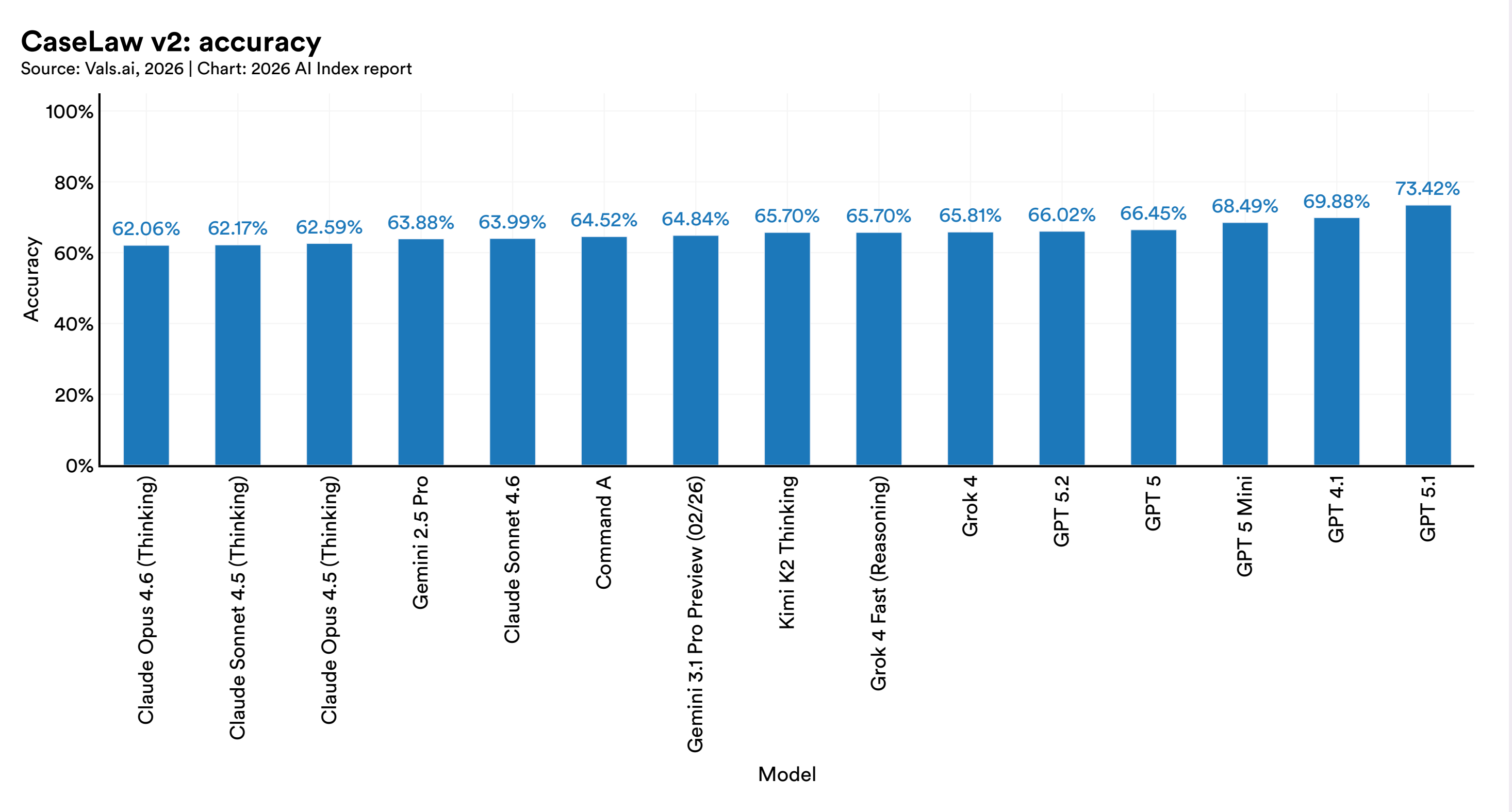

Figure 2.5.12[31]

31 Data source: https://www.vals.ai/benchmarks/case_law_v2.

## LegalBench

LegalBench is a crowd-sourced benchmark for legal reasoning on tasks that mirror real legal work. Rather than test general question answering, it focuses on careful reading, spotting issues, and applying rules to facts. The benchmark covers six types of legal reasoning, including issue spotting, rule recall, outcome prediction, rule application, interpretation of legal text, and rhetorical understanding. The results below reflect model performance as of early 2026.

On the leaderboard results, the top 15 models score above 83% (Figure 2.5.13). The top overall performer is Gemini 3.1 Pro Preview (2/26) at 87.4%, followed closely by Gemini 3 Pro (11/25) with 87% accuracy. The total spread across all 15 models is about 4 percentage points, a narrow range that makes it hard to differentiate among them.

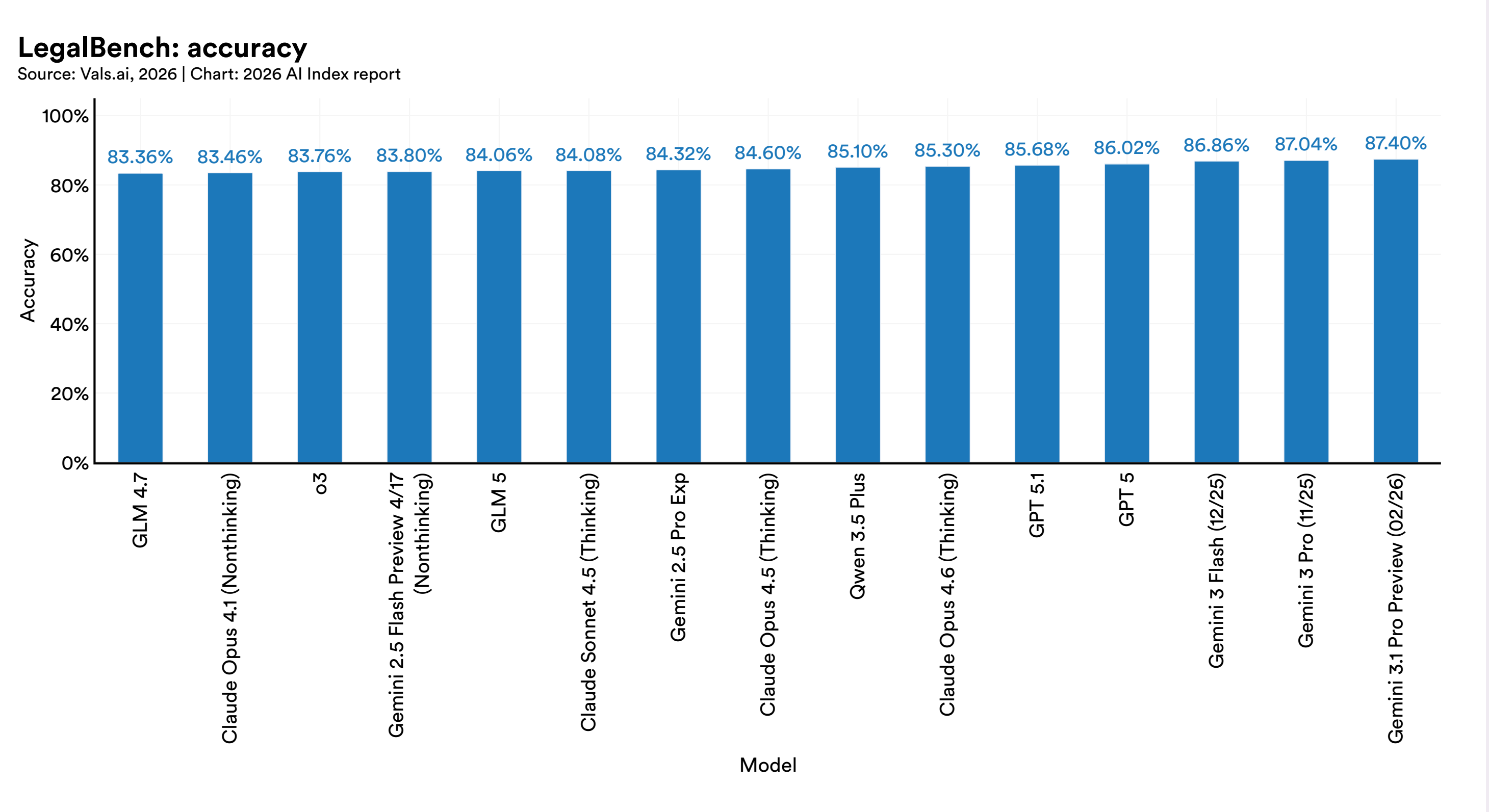

Figure 2.5.13[32]

32 Data source: https://www.vals.ai/benchmarks/legal_bench.

# 2.6 AI Agents

Agent benchmarks test whether AI systems can go beyond answering questions and actually complete multistep tasks in realistic environments. These tasks often involve navigating software, calling tools, managing files, or interacting with websites and databases. More complex tasks may require agents to orchestrate entire workflows, coordinating across multiple tools and systems to achieve a goal. For example, an agent might need to search a database, apply a policy rule, and then update a customer record, all in a single conversation. Unless otherwise noted, the results reported below reflect model performance as of early 2026.

## GAIA

GAIA is a benchmark for general AI assistants, introduced by Meta in May 2024. It tests whether models can handle the kinds of multistep, real-world questions a capable assistant would need to answer—questions that often require web browsing, file handling, and reasoning across multiple sources.

Accuracy on GAIA has risen from about 20% in January 2025 to 74.5% in September 2025 (Figure 2.6.1). The human baseline sits at 92%, leaving a gap of about 17.5 percentage points.

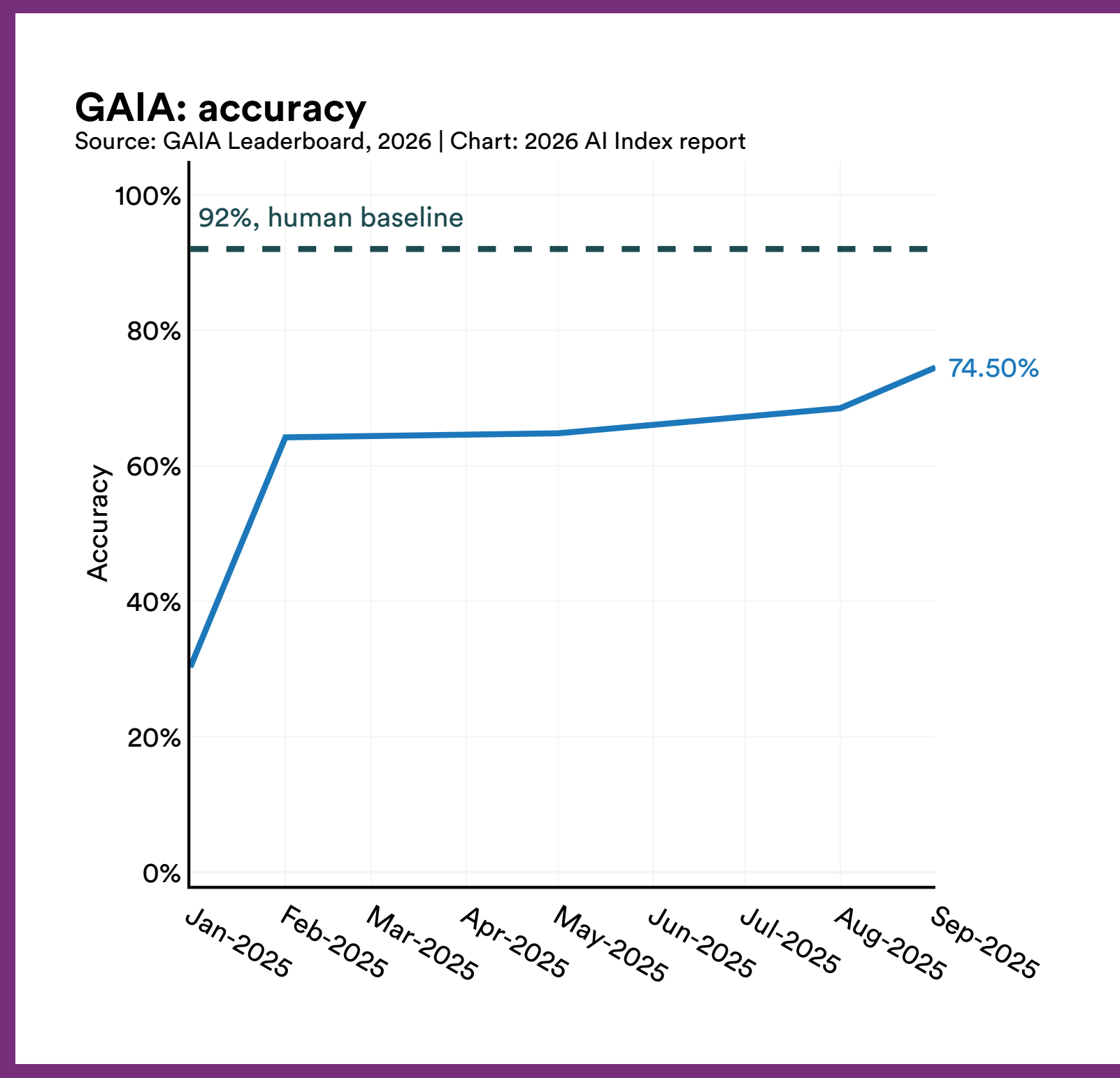

Figure 2.6.1[33]

33 Data source: https://hal.cs.princeton.edu/gaia.

## OSWorld

OSWorld is a scalable, real computer environment designed to evaluate multimodal AI agents on open-ended tasks across operating systems like Ubuntu, Windows, and macOS. It includes 369 tasks involving desktop and web apps, file operations, and multi-application workflows. Computer science students solve about 72% of these tasks with a median time of roughly two minutes, while the strongest models have historically reached only 1%–12% success, especially on tasks involving graphical interfaces and multi-app workflows.

However, the gap has recently narrowed quite a bit with Claude Opus 4.5 leading on accuracy on OSWorld with 66.3% (Figure 2.6.2). This puts the best model within 6 percentage points of human performance. This is one of the benchmarks in this section where the gap between model and humans has closed the fastest.

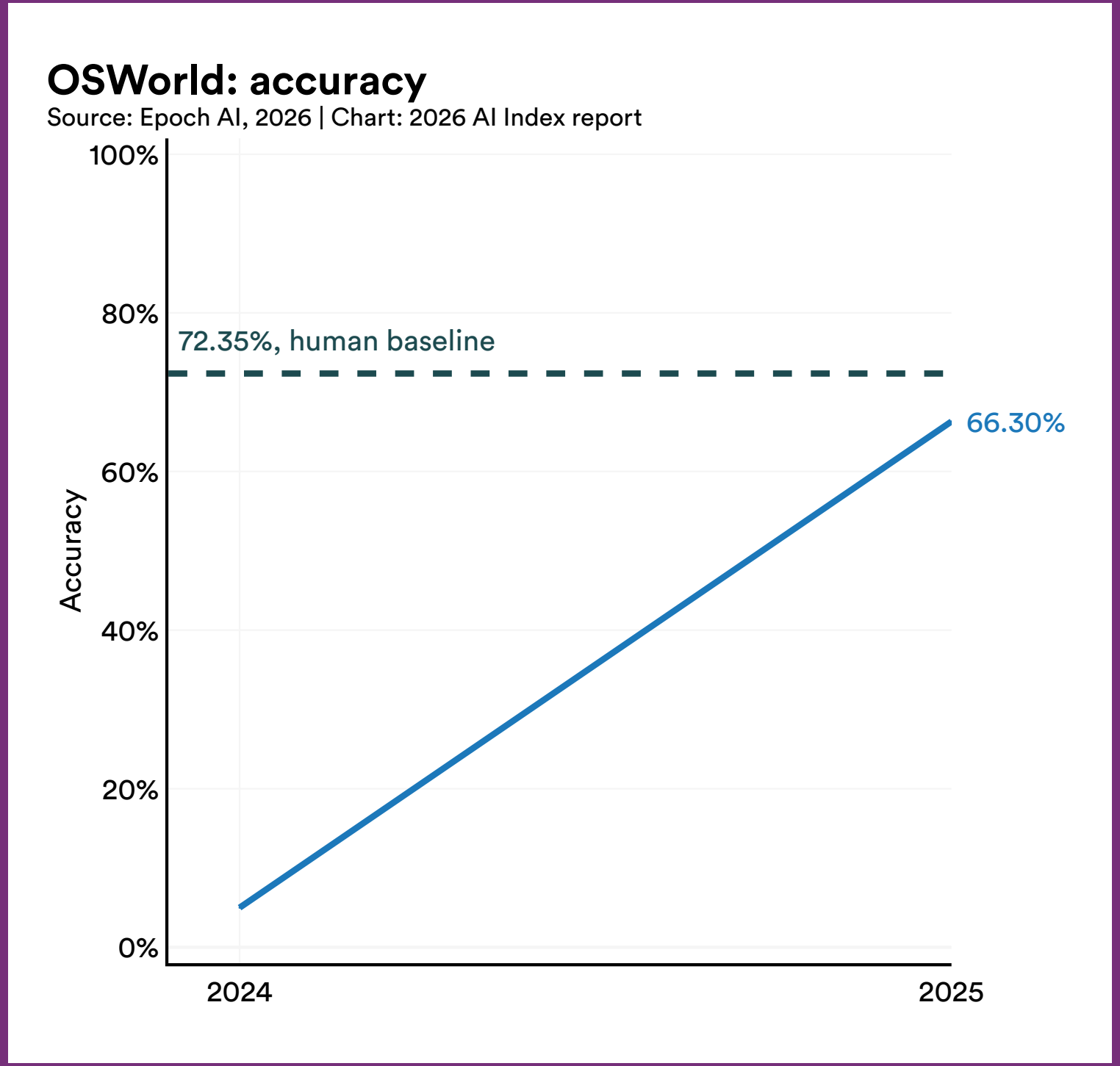

Figure 2.6.2[34]

## WebArena

WebArena is a realistic web environment for evaluating autonomous web agents, and it introduces 812 long-horizon tasks written as natural language intents, such as finding information, navigating sites, and configuring content across multiple pages. Rather than comparing action traces, WebArena checks whether the agent actually achieved its goal by verifying the resulting state of the site, including databases, page content, and URLs.

Success rates on WebArena have steadily increased from about 15% in 2023 to 74.3% in early 2026 (Figure 2.6.3). The best models are now within 4 percentage points of the human baseline of 78.2%. Of all the agent benchmarks in this section, WebArena shows the smallest remaining gap between models and human performance.

34 Data source: https://epoch.ai/benchmarks/.

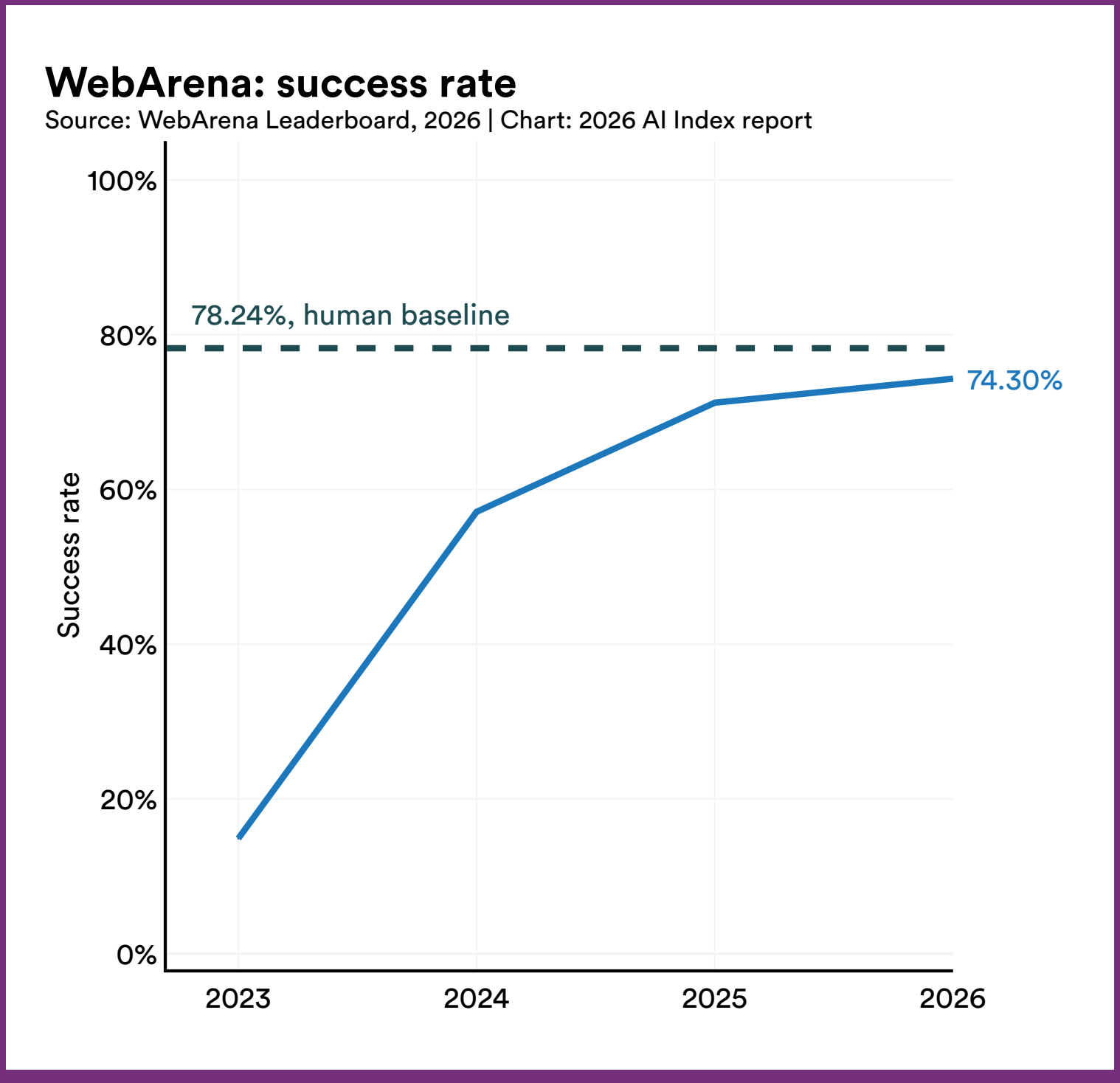

Figure 2.6.3[35]

## MLE-bench

MLE-bench evaluates the machine learning engineering capabilities of AI agents. It consists of 75 Kaggle competitions spanning tasks in NLP, computer vision, signal processing, and more. The competitions were manually curated, with rebuilt train and test splits and reimplemented grading code, so agents can be scored locally and compared directly against human Kaggle leaderboards and medal thresholds.

Agents have also made significant progress on MLE-bench, advancing from about 17% success in 2024 to 64.4% in early 2026 (Figure 2.6.4). This level of improvement in such a short time points to growing capability on end-to-end machine learning tasks, though competition-style problems are more structured than the open-ended work that characterizes most real-world data science.

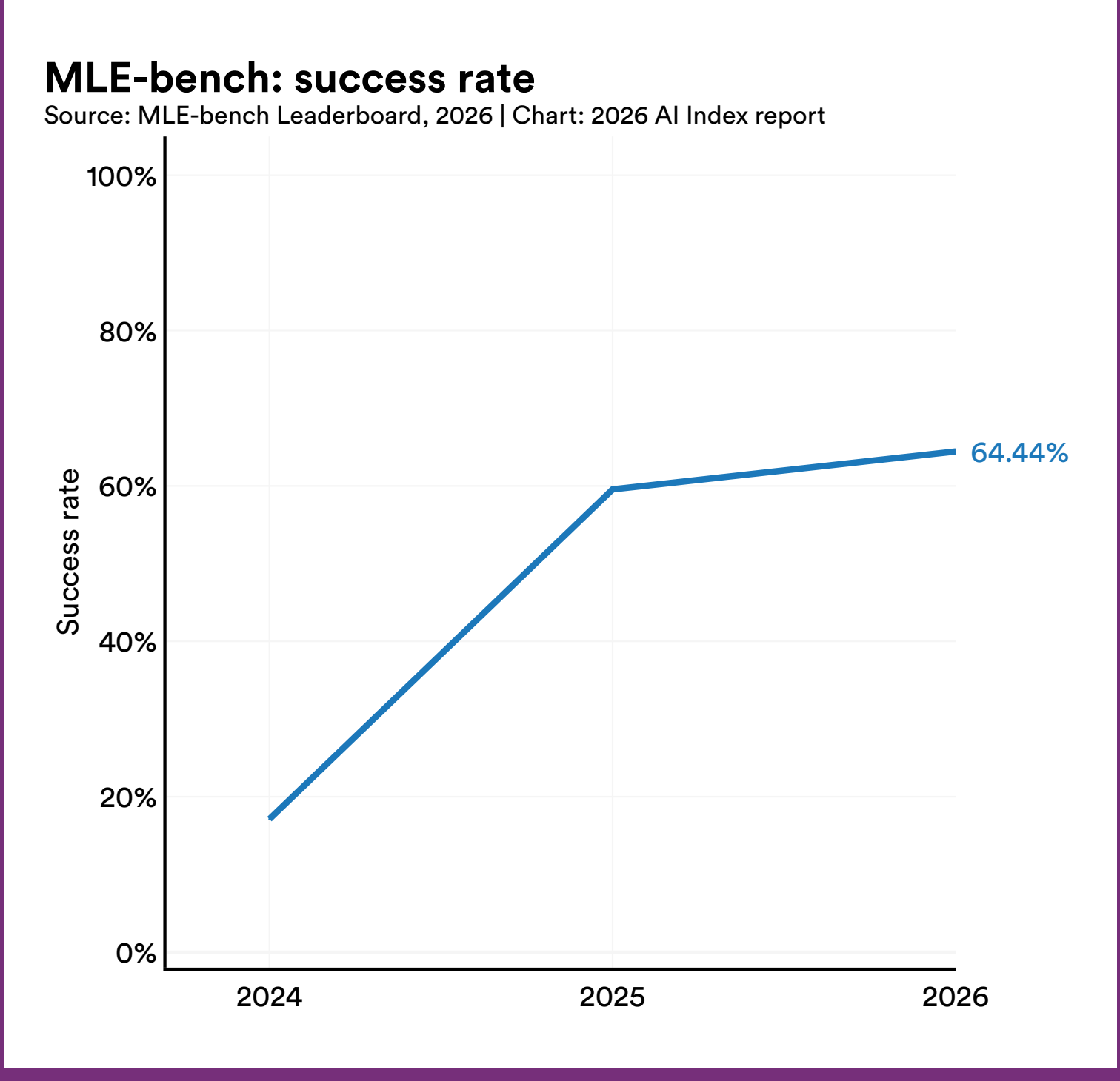

Figure 2.6.4[36]

35 Data source: https://docs.google.com/spreadsheets/d/1M801lEpBbKSNwP-vDBkC_pF7LdyGU1f_ufZb_NWNBZQ/edit?gid=0#gid=0.

36 Data source: https://github.com/openai/mle-bench.

## Cybench

Cybench is a benchmark framework for evaluating the capabilities of AI agents in cybersecurity. It includes 40 professional-level tasks across six capture-the-flag categories, including cryptography, web security, reverse engineering, forensics, and exploitation. Tasks are grounded in real human difficulty via “first solve time,” ranging from two minutes up to almost 25 hours, giving the benchmark a very high difficulty ceiling.

The unguided solve rate on Cybench is 93%, up from 15% in 2024 (Figure 2.6.5). This is the steepest improvement rate across all benchmarks in this section, and it may highlight cybersecurity challenge tasks as a good fit for current agent capabilities.

**Cybench: unguided % solved**

Source: Cybench Leaderboard, 2026 | Chart: 2026 AI Index report

Figure 2.6.5[37]

## τ-bench

τ-bench takes a different approach by testing agents on real-world tasks that involve chatting with a user and calling external tools or APIs. It places the agent in realistic domains, such as retail and airline, with underlying databases, policy constraints, and multiturn conversations. Success is measured by whether the agent produces the correct final outcome, which is often verifiable from the resulting database state. This makes it a test of end-to-end tool use and rule-following in interactive settings, not just language ability.

Leading models on τ-bench achieve pass@1 scores between 62.9% and 70.2% (Figure 2.6.6). Claude Opus 4.5 leads at 70.2%, followed by GPT 5.2 at 69.9% and Qwen3.5 at 68.4%. The spread across the top seven models is a narrow 7.3 percentage points, with no model exceeding 71%, suggesting that managing multiturn conversations while correctly using tools and following policy constraints remains difficult even for frontier models.

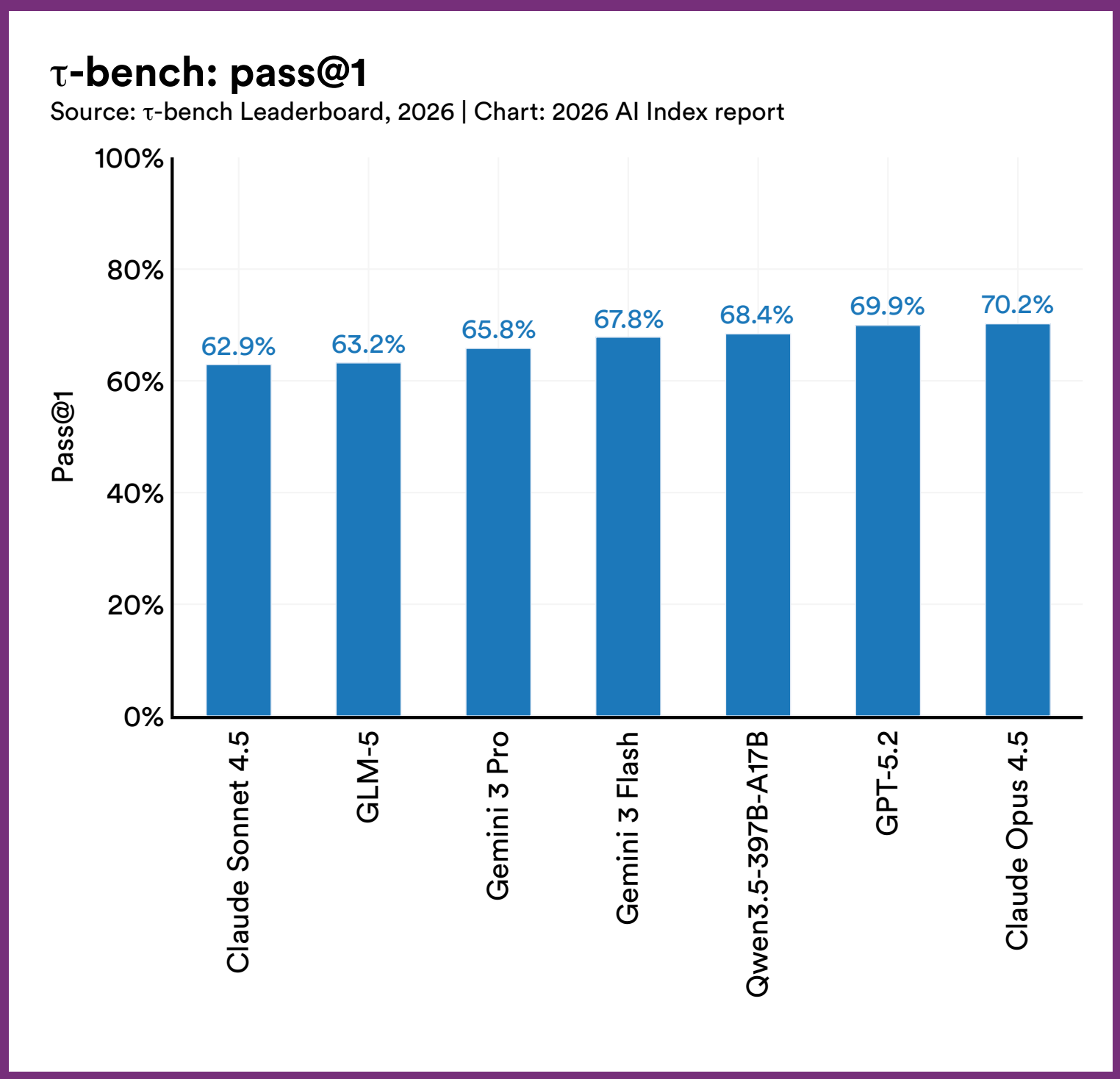

Figure 2.6.6[38]

37 Data source: https://cybench.github.io/.

38 Data source: https://taubench.com/#leaderboard?benchmark=text.

# 2.7 Robotics and Autonomous Motion

## Robotics

### RLBench

RLBench is a benchmark for robotic manipulation that tests agents on a standardized set of 18 tasks using 100 demonstrations per task. Each task involves a different manipulation challenge, such as picking up objects, stacking items, or operating simple mechanisms.

As of January 2026, the top-performing method on the 18-task RLBench subset is EquAct, which reaches an 89.4% average success rate, compared with 86.8% for the prior leader, SAM2Act (Figure 2.7.1). EquAct also reports stronger performance under a more difficult evaluation setting that introduces full 3D rotational variation, where previous methods tend to degrade. There has been consistent progress from about 48% in 2022 to nearly 90% in 2025, though the benchmarks test relatively short-horizon tasks in a controlled simulation environment.

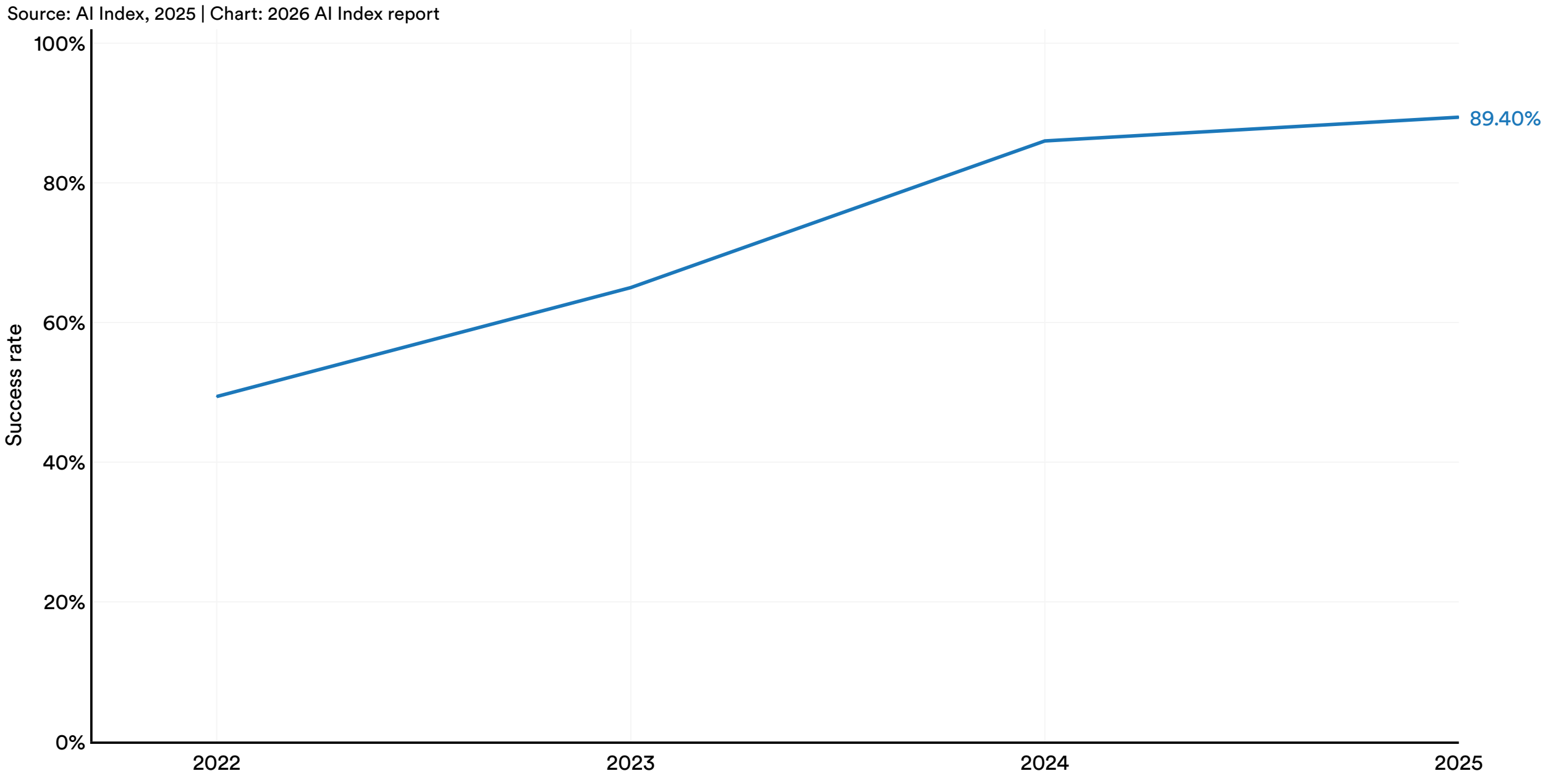
RLBench: success rate (18 tasks, 100 demo/task)
Source: AI Index, 2025 | Chart: 2026 AI Index report

Figure 2.7.1

### BEHAVIOR-1K

BEHAVIOR-1K is a simulation benchmark built around real human needs. The tasks come from surveys asking people what household tasks they want robots to help with, resulting in 1,000 realistic activities. These are long-horizon mobile manipulation challenges in simulated home environments, designed to bridge the gap between current research and human-centered applications.

Results from the 2025 BEHAVIOR Challenge show how difficult these tasks remain (Figure 2.7.2). The top team, Robot Learning Collective, achieved a Q-score[39] of about 26% on the held-out test set, meaning it completed only a quarter of the required task objectives at an acceptable quality. Full task success rates were even lower, with the top team reaching just 12.4%. These scores make it clear that reliably executing household tasks in realistic environments is still beyond current capabilities.

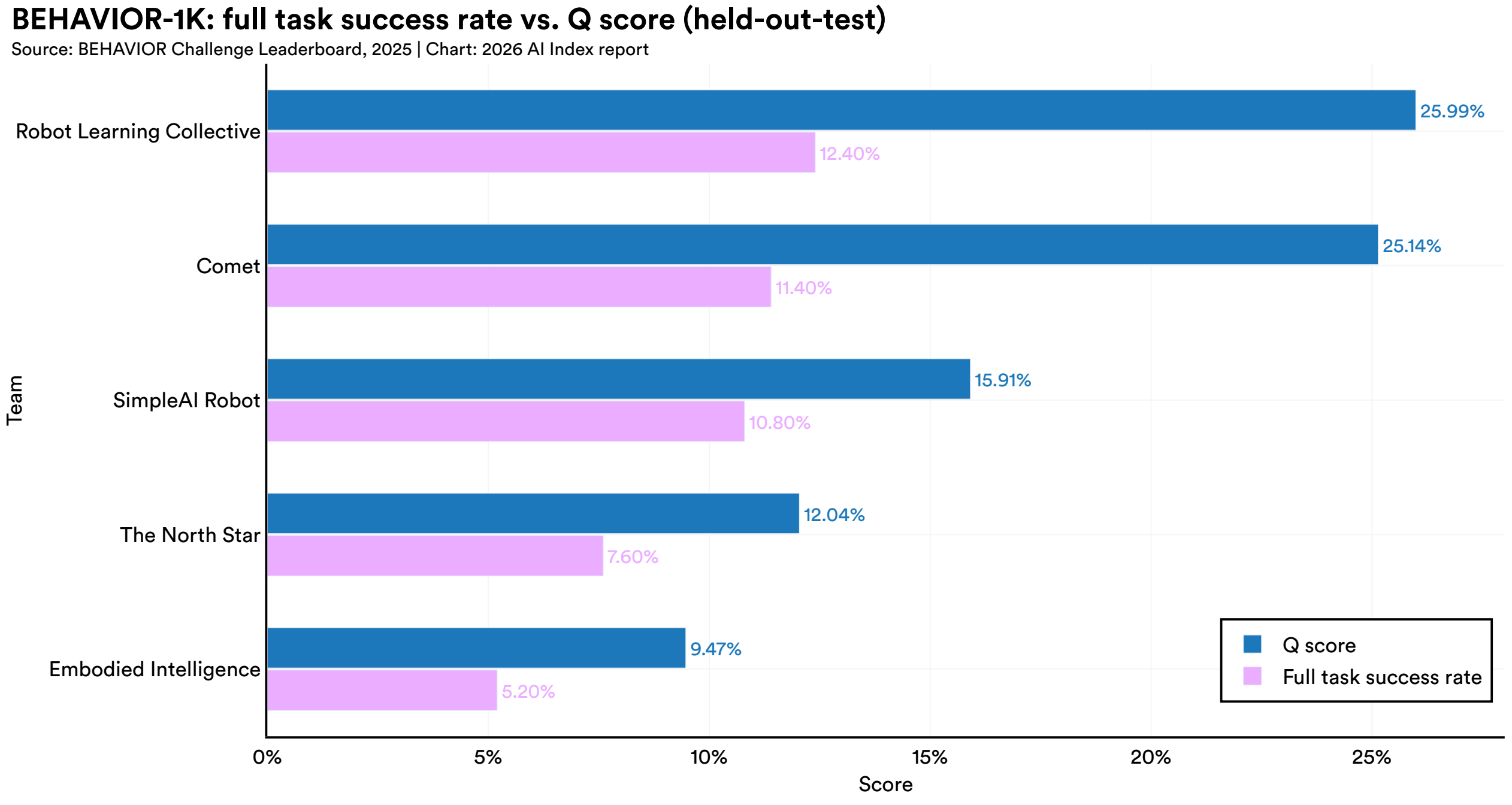
BEHAVIOR-1K: full task success rate vs. Q score (held-out-test)
Source: BEHAVIOR Challenge Leaderboard, 2025 | Chart: 2026 AI Index report

Figure 2.7.2

## ResponsibleRobotBench

Most robotics benchmarks measure whether a model can complete a task. ResponsibleRobotBench measures if that task is completed safely when the environment includes real hazards. The benchmark is built around 23 multi-stage tasks involving electrical, fire/chemical, and human-related hazards. To complete a task safely, the robots must detect risks, reason about safety, plan safe actions, and request human assistance when necessary. Performance is measured by the safe success rate, which counts a task as successful only when both the task is complete and safety conditions are met. GPT-4o achieves the best results with a safe score of 0.64, outperforming GPT-4o mini at 0.40 and the strongest open-source model, Qwen-72B, at 0.35 (Figure 2.7.3). Even the top model failed to complete more than a third of tasks safely, with frequent failures when both task completion and safety must be satisfied simultaneously.

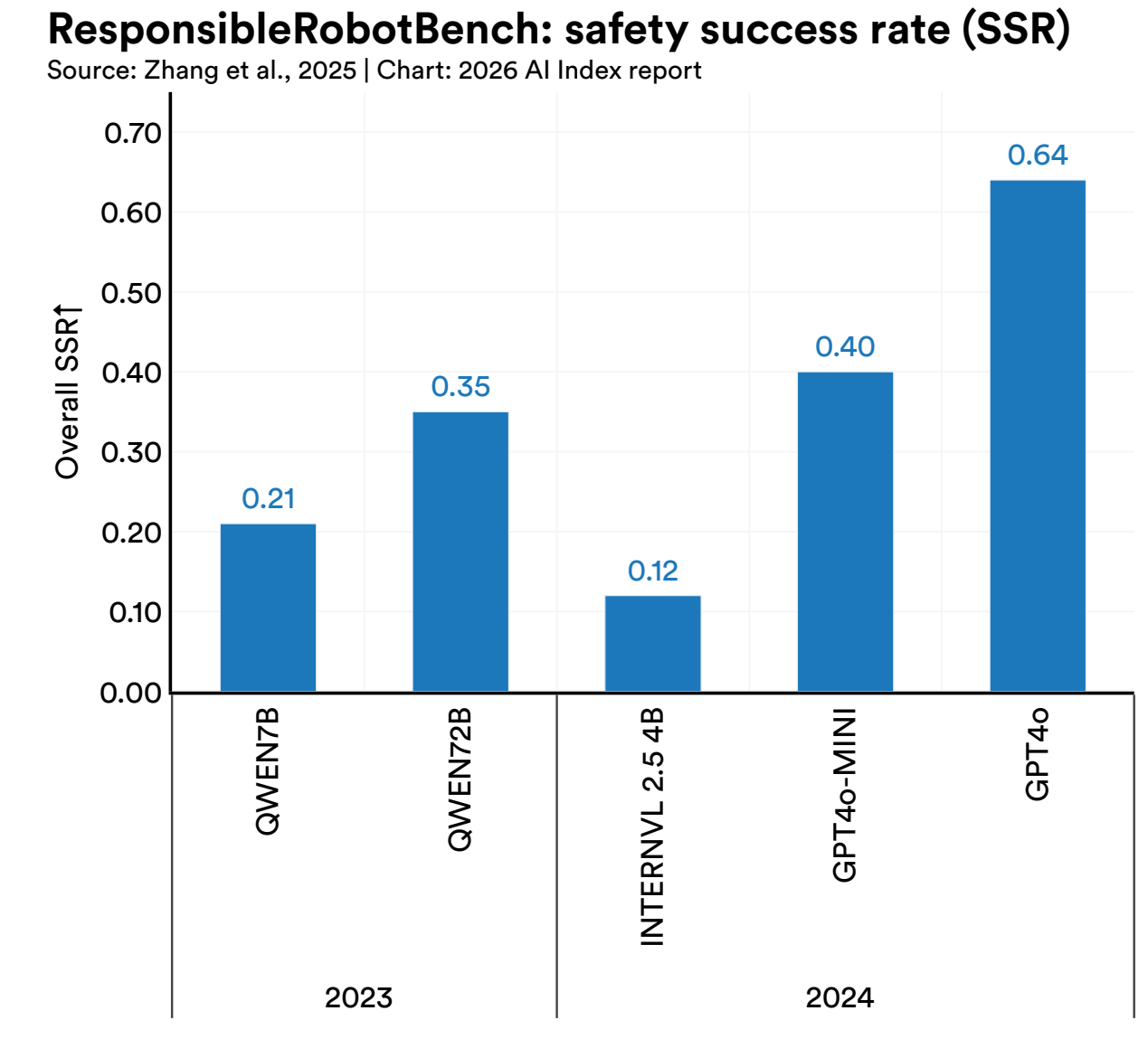

Figure 2.7.3

39 Q-score measures how much of a task's goal a policy satisfies by calculating the fraction of completed subgoals and selecting the best-matched goal clause. It awards partial credit, so policies that make meaningful progress score higher even without finishing the full task. This makes Q-score a smoother and more reliable metric for comparing policies across BEHAVIOR tasks than a binary success rate.

**HIGHLIGHT:**

## Humanoid Robotics

As covered in last year's AI Index, humanoid robots began attracting significant attention in 2024 with new hardware launches from companies like Figure AI, Tesla, and Boston Dynamics. In 2025, the field continued to grow, with a significant increase in the number and variety of available humanoid platforms (Figure 2.7.4). The strongest signals came from early-stage industrial pilot projects and manufacturing-scale ambitions rather than widespread deployment. Figure AI's Figure 02 robot, for example, spent 11 months on the line at a BMW plant in South Carolina, logging over 1,250 runtime hours and loading more than 90,000 parts across over 30,000 vehicles. In China, vendors like Unitree and AgiBot pushed prices down and production volumes up, framing humanoids as quasi- consumer hardware products rather than bespoke research systems. Some companies are targeting home environments for their humanoid robotics, with Norway's 1X opening a waitlist for deliveries of its $20,000 household robot.

The overall picture is one of rapid growth in hardware availability and investment activity rather than widespread deployment. Most company milestones are framed in the future tense, along with delivery timelines; intended use cases are offered in place of verified operational data. It remains unclear whether the demand for humanoid robots will match the supply currently being built, who the customers will be at scale, and how quickly these platforms will move from structured factory pilot projects to unstructured environments.

| Company | Country | Platform | Focus | Notable detail |
|---|---|---|---|---|
| Sanctuary AI | Canada | Phoenix | Commercial pilots | Hundreds of commercial pilot tasks completed |
| Unitree | China | G1, R1 | Research, industrial | R1 from $4,900; G1 from $13,500 with advanced perception |
| UBTECH | China | Walker S, S2 | Industrial | LLM-integrated planning; autonomous battery swapping |
| AgiBot | China | Humanoid fleet | Data collection, industrial | ~100 teleoperated humanoids running up to 17 hrs/day; ~10,000 units manufactured |
| Fourier Intelligence | China | GR-1 | Medical, service, industrial | Camera-only vision and LLM interaction |
| DeepRobotics | China | Humanoid platforms | Industrial | Extending quadruped expertise into humanoid form factors |
| Neura Robotics | Germany | 4NE-1 | Home, workplace | Dense sensors, including artificial skin for safe human collaboration |
| Addverb.ai | India | In development | Manipulation | Expanding from mobile robotics toward humanoid manipulation |

**HIGHLIGHT:**

| | | | | |
|---|---|---|---|---|
| Milagrow | India | In development | Manipulation | Expanding from mobile robotics toward humanoid manipulation |
| Mentee Robotics | Israel | MenteeBot | Warehouse | Autonomous workflows using natural-language commands |
| Toyota Research Institute | United States | Various | Research | Diffusion policy and large behavior models |
| Honda | Japan | Robotics platforms | General purpose | Continuing humanoid and manipulation research |
| SoftBank Robotics | Japan | Various | Retail, logistics | Teleoperated manipulation systems |
| Telexistence | Japan | Various | Retail, logistics | Teleoperated manipulation for retail environments |
| 1X | Norway | NEO | Home | Backed by OpenAI; waitlist open for 2026 U.S. deliveries at ~$20,000 or $499/month |
| Rainbow Robotics | South Korea | RB-Y1 | Workplace | Industrial cobot (collaborative robot) components adapted for humanoid applications |
| LG Electronics | South Korea | Various | Workplace | Leveraging cobot components for humanoidlike applications |
| Technology Innovation Institute | UAE | Testbed | Embodied AI research | Building testbeds using open-weight Falcon models |
| Engineered Arts | United Kingdom | Ameca | Social interaction, research | Lifelike facial expressions for customer engagement |
| Humanoid/SKL | United Kingdom | HMND 01 Alpha | Industrial | Developed in seven months; factory trials underway |
| Figure AI | United States | Figure 02 / 03 | Industrial, home | Deployed at BMW for 11 months; 1,250+ runtime hours; 90,000+ parts loaded across 30,000+ vehicles |
| Tesla | United States | Optimus (Gen 3) | Internal logistics | Third generation; plans for external sales by 2027 |
| Boston Dynamics | United States | Atlas | Research | Testbed for advanced locomotion and manipulation |
| Apptronik | United States | Apollo | Industrial | Safety-rated operation around people |
| Skild AI | United States | Foundation model stack | Multi-embodiment | Omni-bodied control stack designed to work across multiple robot bodies |

Figure 2.7.4

HIGHLIGHT:

## Physical AI and Foundation Models for Robotics

Most of what people need help with happens in physical spaces, from assembling products in a factory to assisting with household tasks. For AI to be useful, it must do more than process text and images on a screen. It has to perceive its surroundings, reason about how objects behave, and act on those judgments through a physical body.

Throughout this chapter, the benchmarks that prove hardest for AI are the ones that require acting in the real world, where environments are unpredictable and mistakes have physical consequences. The robotics benchmarks earlier in this section reflect that difficulty. Traditional robots sidestep the problem by running fixed programs for fixed tasks, but that approach breaks down in any setting that changes from one day to the next.

A growing body of research is trying to close this gap by giving robots the same kind of general-purpose AI that has driven progress in language and vision. Vision-language-action models, or VLAs, replace the traditional pipeline of separate modules for seeing, planning, and acting with a single network that goes directly from camera input and language instructions to motor control.

Physical Intelligence's $\pi_0$ (2024) and π0.6 (2025) demonstrate this approach, performing tasks like laundry folding across different robot platforms without task-specific retraining. Nvidia's GR00T models and Gemini Robotics take a similar direction, training single models that can control different robots across different tasks.

The biggest constraint, however, is data. Language models train on billions of pages of text from the internet. Every piece of robot training data requires either a physical robot performing a task or a high-fidelity simulation, both of which are slow and expensive. World Foundation Models (WFMs) are one response, generating synthetic physics data so robots can learn without physical trials. Nvidia's Cosmos is one example. But VLA technology remains at the research stage, and the gap between what these models can do in a controlled setting and what they can handle in the real world is still wide.

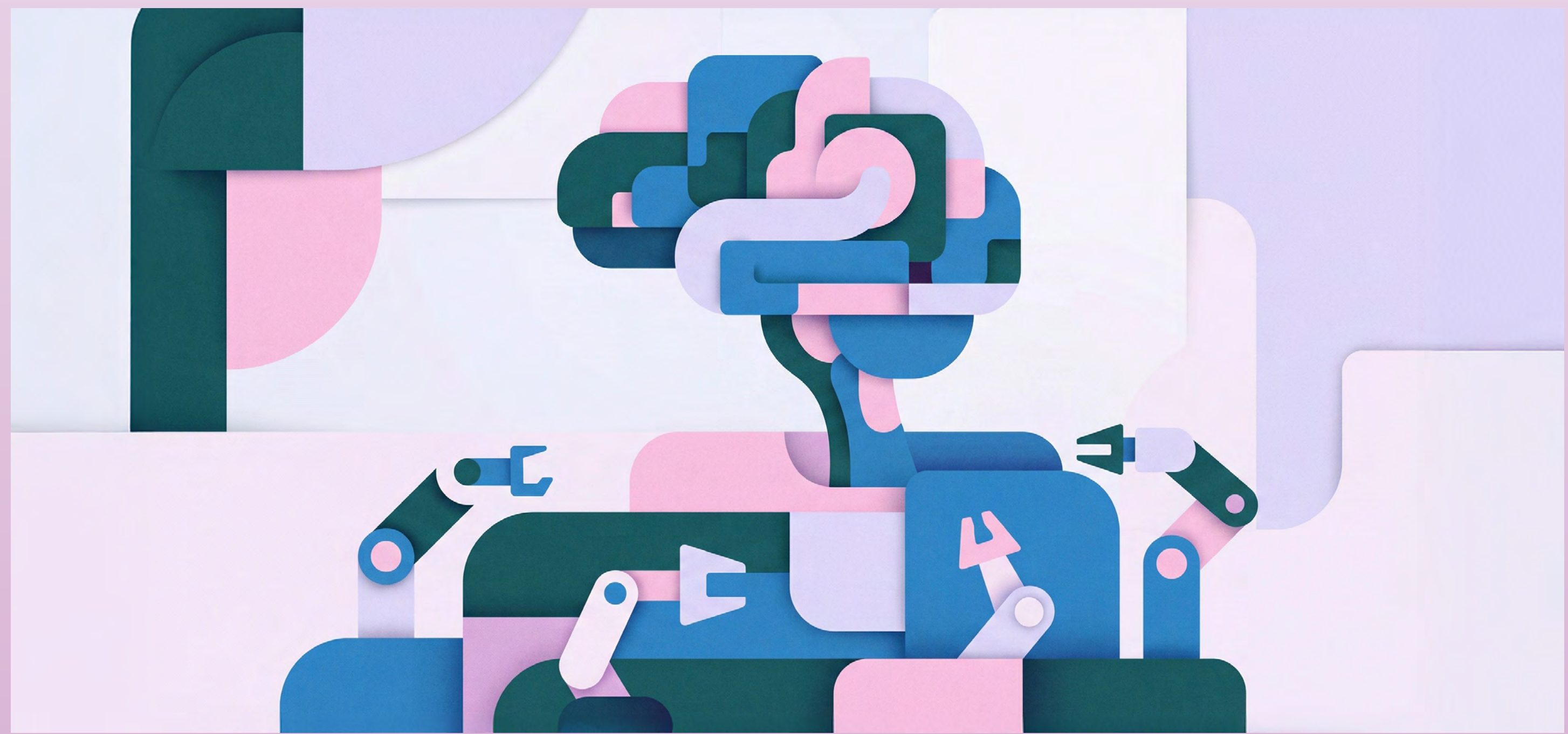

## Self-Driving Cars

Self-driving car development has moved past the research stage in several markets, with commercial services now operating at scale. This section tracks deployment trends, technical innovations in benchmarks and datasets, and safety through crash reporting data. The data available for this section is concentrated in the United States and, to a lesser extent, China. European autonomous vehicle operators such as Mobileye, Vay, and Wayve are active, but comparable trip or deployment data is not publicly available. Chinese data is also limited, with Baidu's Apollo Go being one of the few services to publish detailed ridership figures.

### Deployment

Autonomous vehicle deployment accelerated in 2025, with growth in both the United States and China. By late 2025, Waymo operated roughly 2,500 fully autonomous robotaxis across major U.S. cities, including Phoenix, San Francisco, Los Angeles, Austin, and Atlanta, with the service recording around 450,000 weekly trips. In California alone, weekly paid trips climbed from near zero in mid-2023 to approximately 283,880 by late 2025, with sharp growth after February 2025 (Figure 2.7.5). Zoox, a smaller operator, began appearing in California pilot trip data in late 2025 (Figure 2.7.6). In China, Baidu's Apollo Go autonomous ride-hailing service provided approximately 11 million fully driverless rides in 2025, a 175% year-over-year increase (Figure 2.7.7). The service has grown from 1.5 million trips in 2022 to 11 million in 2025, reflecting rapid expansion in usage.

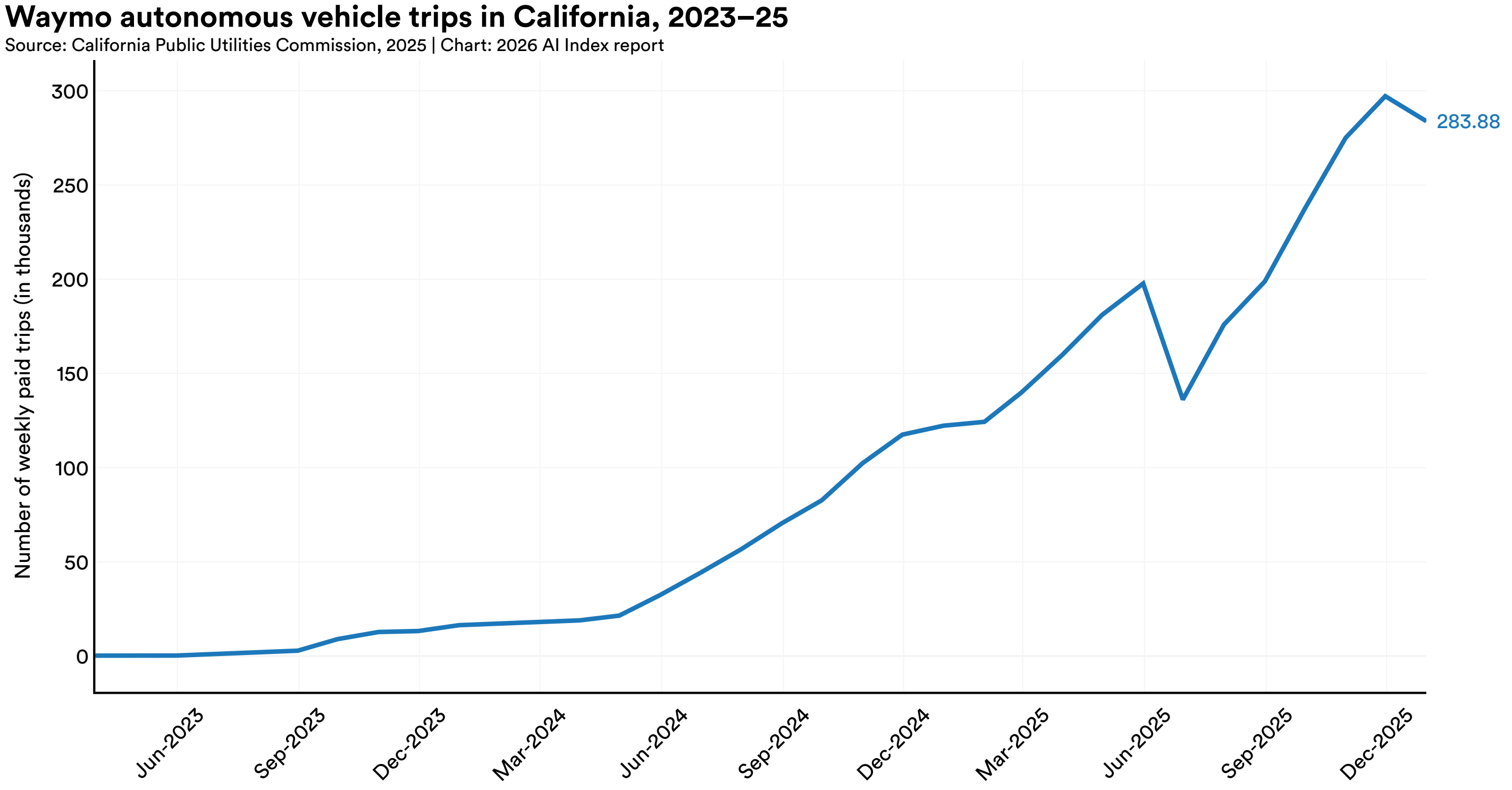
Waymo autonomous vehicle trips in California, 2023–25
Source: California Public Utilities Commission, 2025 | Chart: 2026 AI Index report

Figure 2.7.5[40]

40 These AV deployment metrics, as reported to the California Public Utilities Commission, pertain to Waymo and Cruise (until the latter was discontinued by General Motors in December 2024). Several other companies, including Aurora, Tensor (formerly AutoX), WeRide Corp, and Zoox, are in pilot stages. Tesla has not been approved by the CPUC to offer autonomous passenger service. Data source: California Public Utilities Commission quarterly reporting.

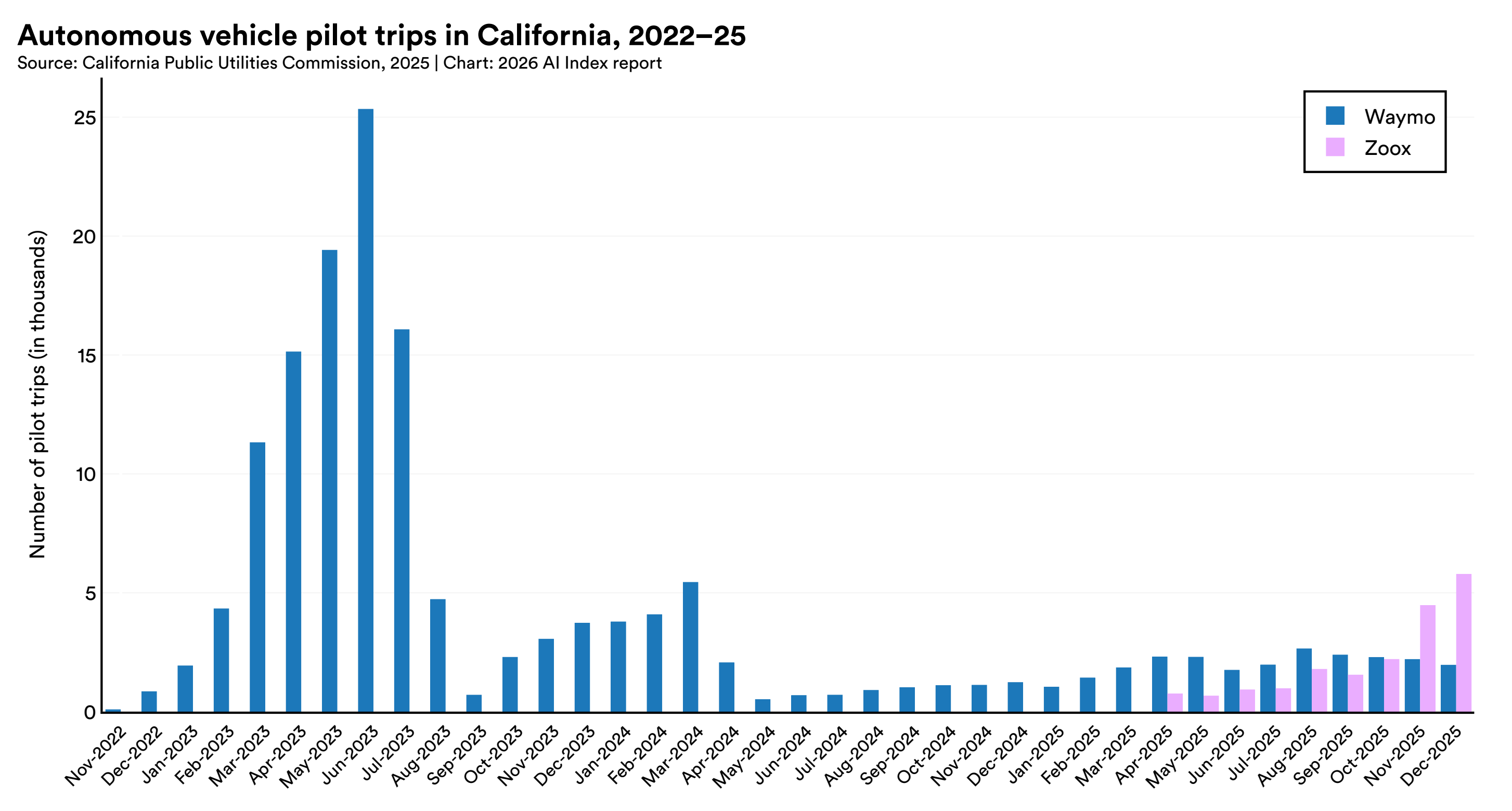

Figure 2.7.6[41]

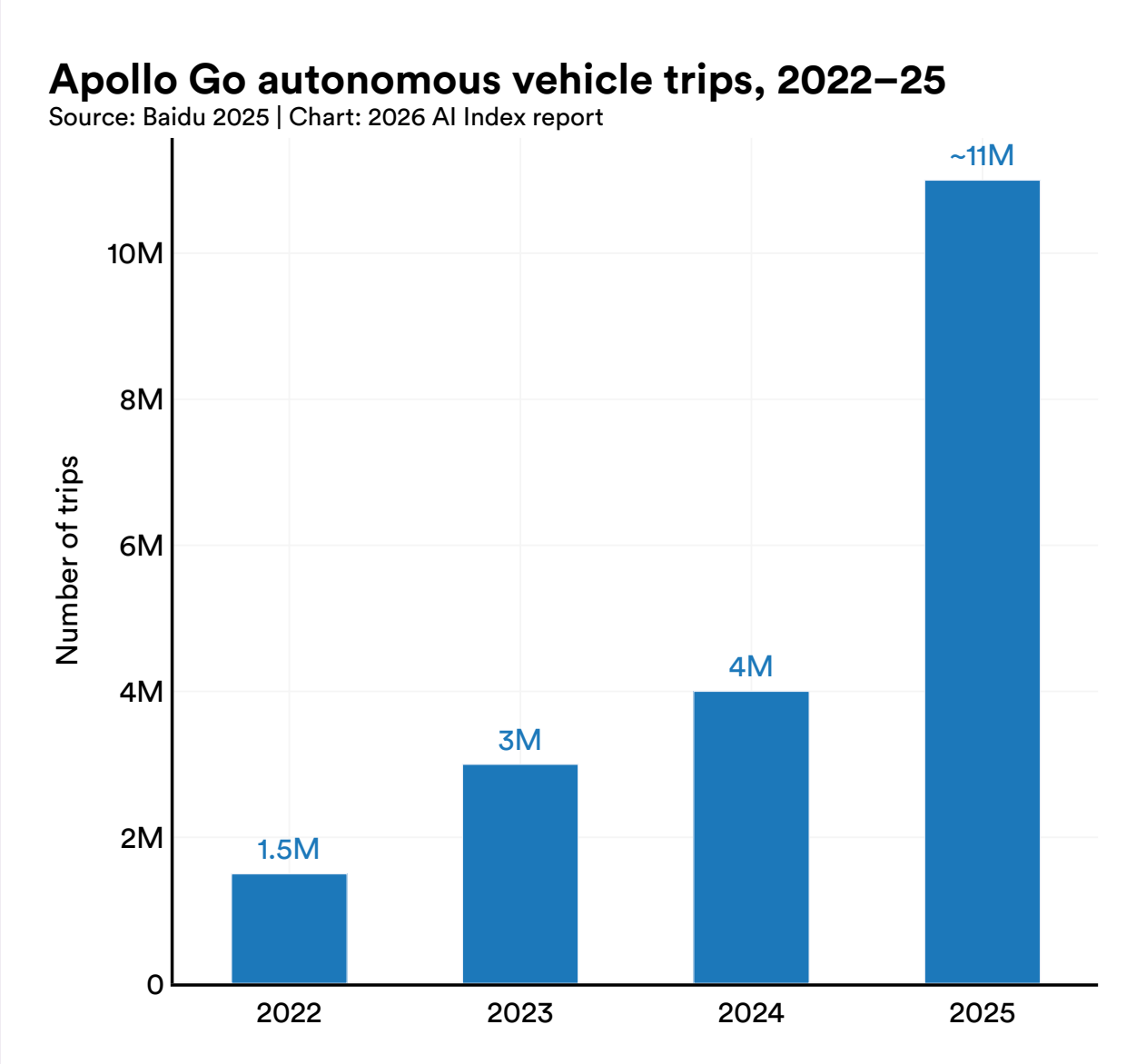

Figure 2.7.7[42]

## Technical Innovations and New Benchmarks

The technical landscape for autonomous driving is shifting in several ways. Benchmarks are consolidating around leaderboards for end-to-end driving, like Waymo's 2025 Open Dataset Challenges, which emphasized vision-based approaches and are increasingly targeting generalization on long-tail cases. Large multisensor datasets are also becoming more central to research. Nvidia's PhysicalAI Autonomous Vehicles dataset includes multicamera, lidar, and radar data across a diverse range of weather, geography, and rare events.

At the model level, combined reasoning and action approaches are gaining traction. Alpamayo 1, a vision–language–action model (VLA), focuses on both trajectory quality and interpretable reasoning, while operating under the safety and latency constraints of real driving. Multimodal reasoning benchmarks are

41 Pilot data covers passenger rides conducted for testing, typically without a fare. Deployment data covers paid autonomous passenger service. Companies can participate in both programs simultaneously if they are deployed and tested in different areas or phases. Data source: California Public Utilities Commission quarterly reporting.

42 The 2025 value is an estimate. Data sources: Baidu financial results reporting (2022, 2023, 2024, 2025).

also evolving, now evaluating multiview spatial reasoning and step-by-step driving logic rather than just final-answer accuracy. More broadly, world models and reinforcement learning are moving beyond imitation-only, end-to-end driving, since these approaches can generalize better to traffic scenarios not seen during training.

The scale of available driving data has also grown over the past decade (Figure 2.7.8). Early benchmarks released between 2012 and 2019 contained single-digit hours of data. A step change came with Waymo's Open Dataset in 2019 at roughly 500 hours, followed by nuPlan in 2024 and Nvidia's Physical AI-AV in 2025 at around 1,600 hours. However, hours alone do not capture differences in data quality or content. A dataset of simulated driving is not the same as one captured from real cars on real roads, even if both report the same number of hours. Therefore, this chart is best read as a trend in data volume rather than a direct comparison across benchmarks.

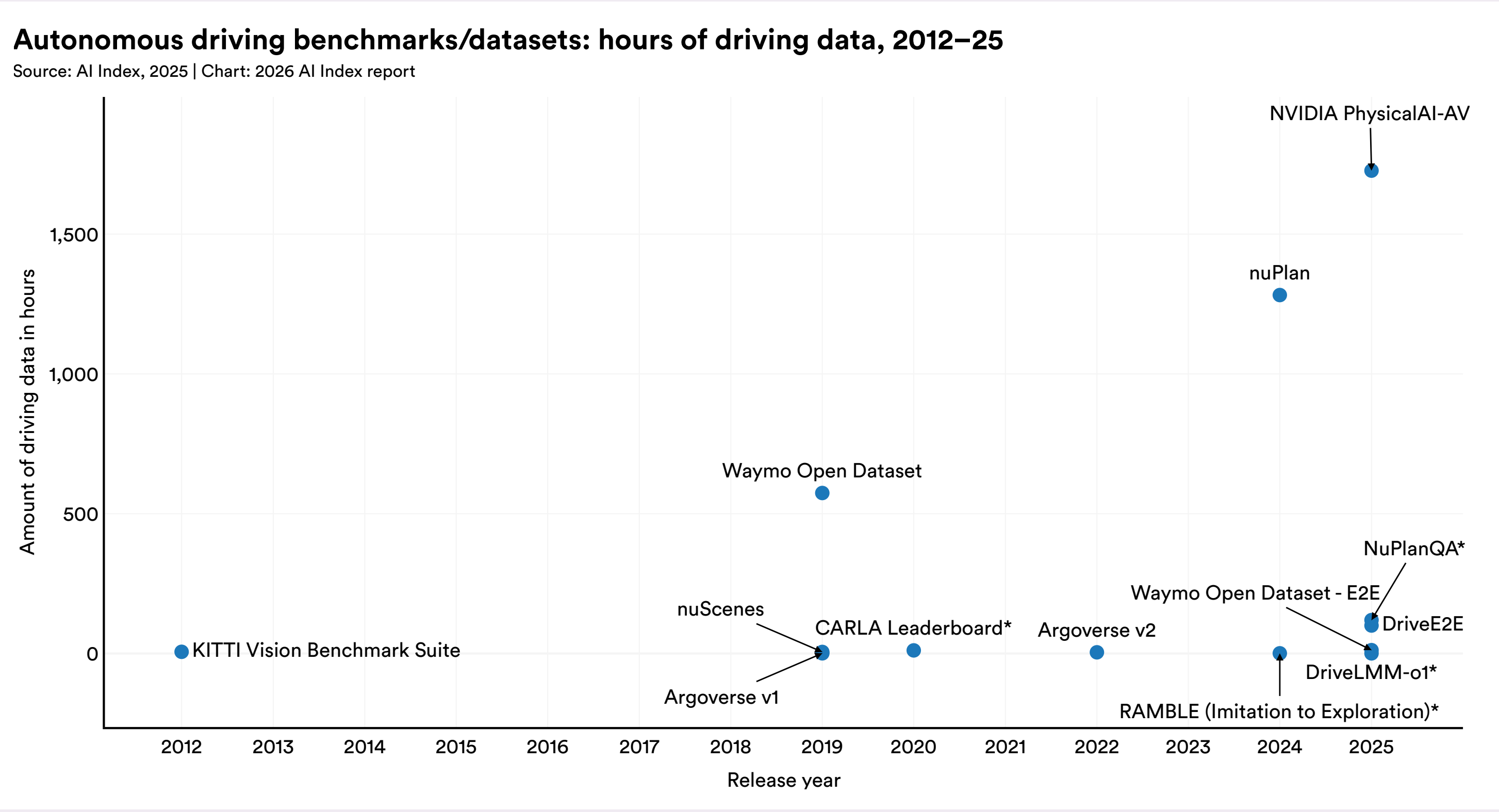

Figure 2.7.8[43]

## Safety

The Standing General Order (the General Order) on Crash Reporting is a National Highway Traffic Safety Administration (NHTSA) mandate that requires manufacturers and operators to report certain crashes involving automated driving systems (ADS) or SAE Level 2 advanced driver assistance systems (ADAS). First issued in 2021 and amended in 2021, 2023, and 2025, the order gives NHTSA consistent crash data to investigate incidents and enforce safety requirements.

Monthly reported ADS incidents have generally trended upward since NHTSA began collecting data in mid-2021, rising from roughly 10–25 per month in the early years to frequently exceeding 80 per month in late 2024 and 2025 (Figure 2.7.9). When broken down by company, Waymo accounts for the largest share of reported incidents, which is consistent with its much larger deployment footprint. Other operators, including Ford, May Mobility, and Transdev Alternative Services, report lower and more stable incident counts.

43 For datasets marked with an asterisk, hours of driving data are estimated rather than directly reported.

Without a comparison point to human driving, raw incident counts are difficult to interpret. Waymo has published data comparing its rider-only crash rates against a human-driven benchmark covering the same miles and areas (Figure 2.7.10). Waymo's reported rates are lower for both any-injury-reported incidents (Figure 2.7.11) and the more severe airbag-deployment-reported incidents (Figure 2.7.12). The largest gap appears in vehicle-to-vehicle intersection incidents, where the human benchmark recorded 198 compared to Waymo's 8. This data comes from Waymo's own safety reporting through September 2025 and should be viewed accordingly.

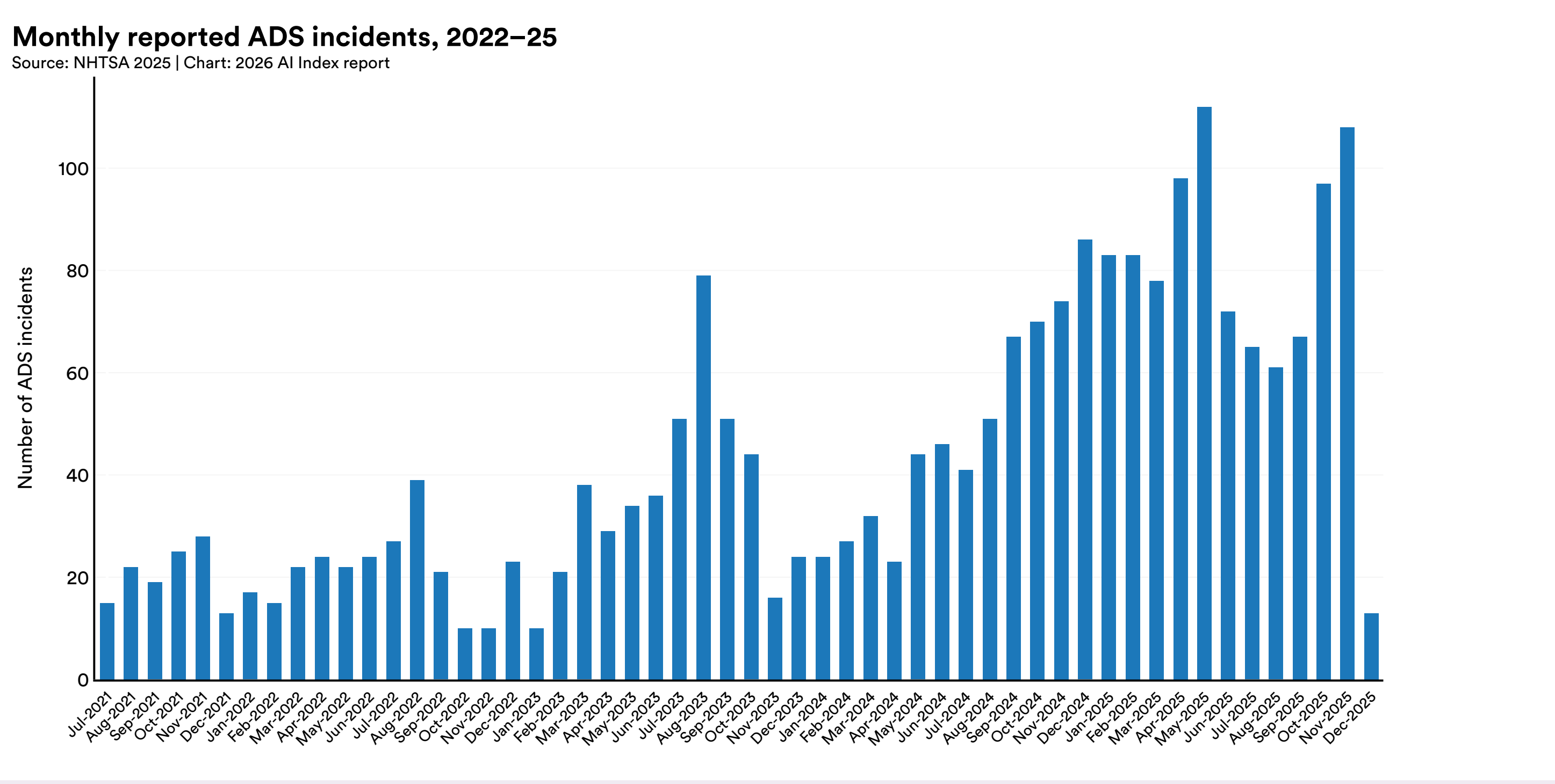

Figure 2.7.9

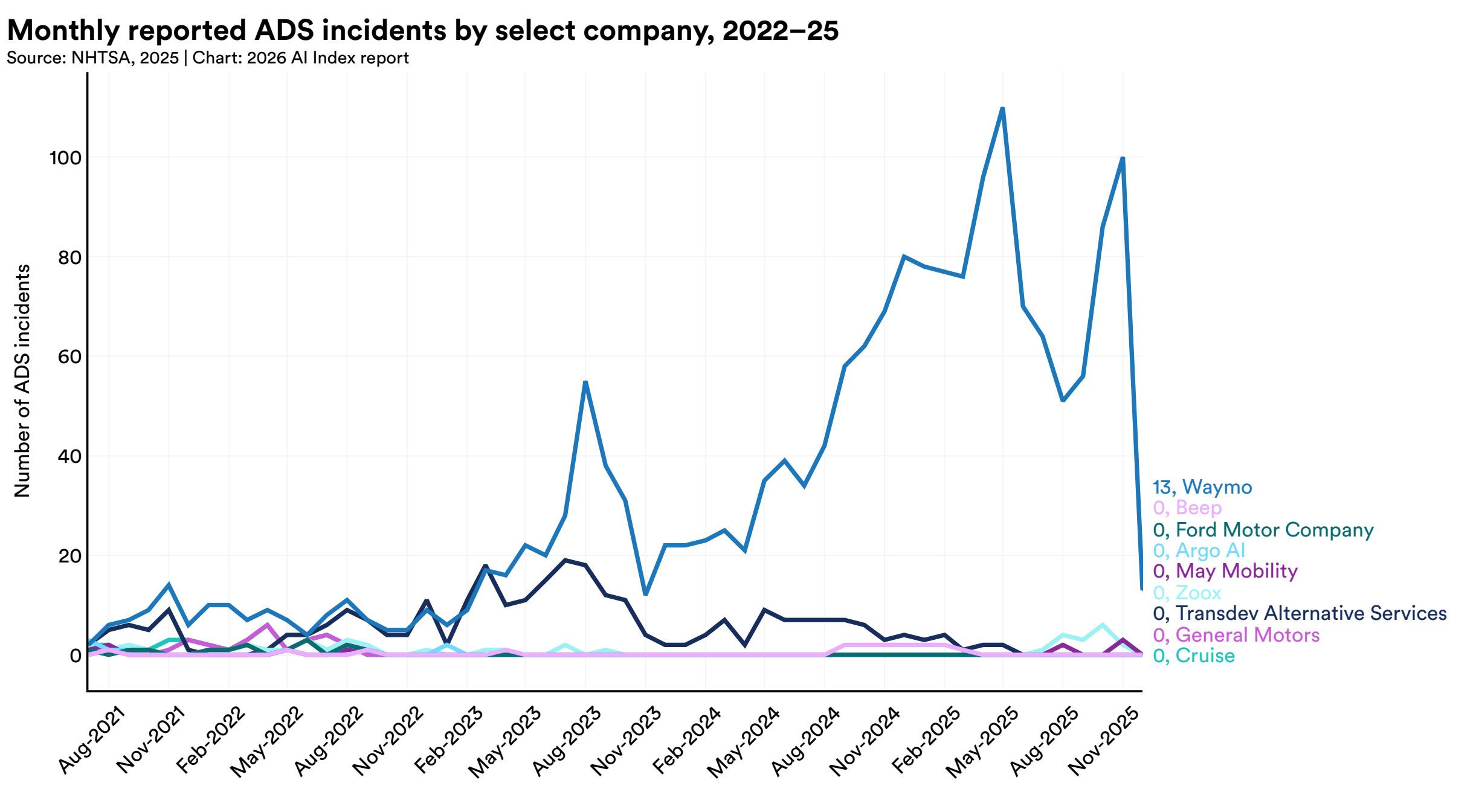

Figure 2.7.10[44]

44 This chart includes only companies that reported at least 10 ADS incidents across the full reporting period.

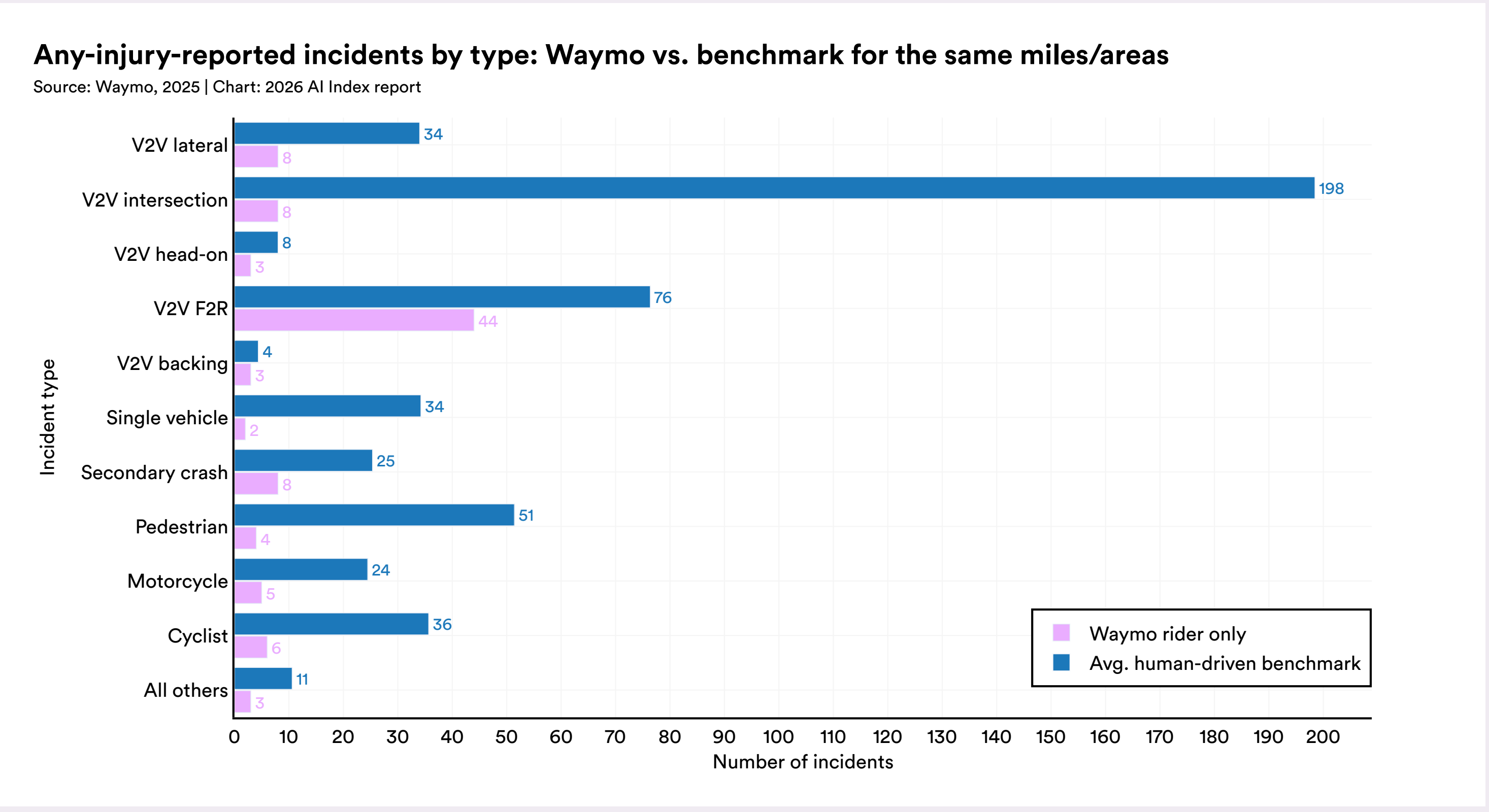

Figure 2.7.11[45]

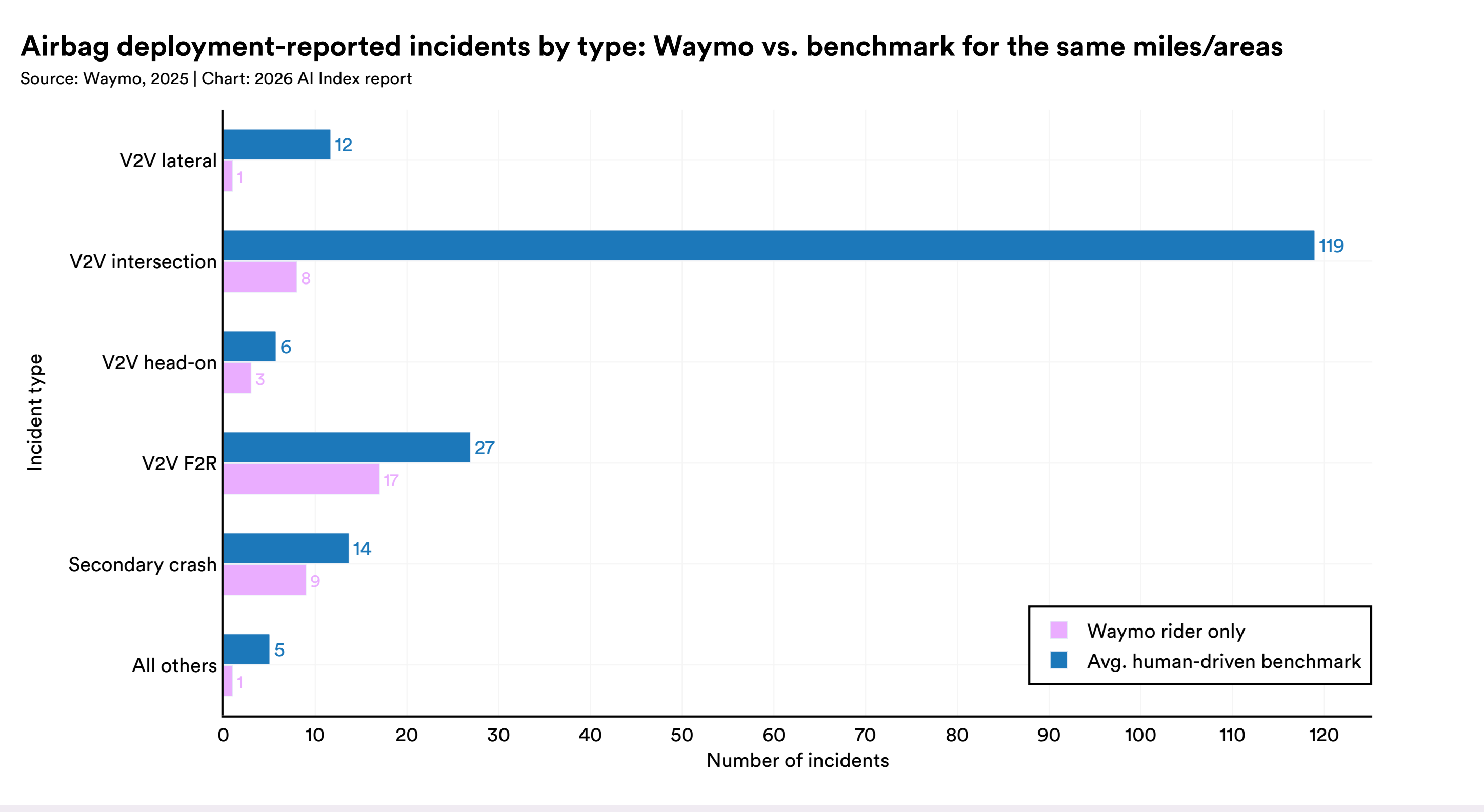

Figure 2.7.12[46]

45 Data source: https://www.waymo.com/safety/impact/#methodology.

46 Data source: https://www.waymo.com/safety/impact/#methodology.

# 3

# Responsible AI

## Overview

The infrastructure for responsible AI (RAI) is growing, but progress has been uneven, and it is not keeping pace with the speed of AI deployment. New safety benchmarks have expanded, more organizations are adopting responsible AI policies, and government-backed AI safety and/or security institutes have spread to more countries. The responsible use of AI is intertwined with the responsible use of data, and in particular with privacy and other legal concerns. There are also AI governance concerns given the ill-specified ownership of AI systems, raising questions about whether companies that develop the systems or consumers that buy them should be held accountable and what policies each stakeholder should follow. While documented reports of AI incidents are increasing, frontier models rarely report results on responsible AI benchmarks, and foundation model transparency declined in 2025 after improving the previous year. Recent research shows that improving one responsible AI dimension can come at the cost of another, with gains in privacy reducing fairness or gains in safety reducing accuracy. There is no framework for navigating these trade-offs; and for dimensions such as fairness, privacy, and explainability, the standardized data needed to track progress over time does not exist. While this chapter draws on the available evidence, the discussion is limited by persistent gaps in measurement.

## Contents

# Chapter Highlights

1. **Responsible AI benchmarking is increasing, but is not keeping up with AI advances and deployments.** Almost all leading frontier model developers report results on capability benchmarks like MMLU and SWE-bench, but reporting on responsible AI benchmarks remains sparse. Documented AI incidents continued to rise, with the AI Incident Database recording 362 in 2025, up from 233 in 2024.

2. **AI models struggle to tell the difference between knowledge and belief.** In a new accuracy benchmark, hallucination rates across 26 top models range from 22% to 94%. GPT-4o's accuracy dropped from 98.2% to 64.4%, and DeepSeek R1 fell from over 90% to 14.4%. When a false statement is presented as something another person believes, models handle it well. When the same false statement is presented as something a user believes, performance collapses.

3. **Organizations are formalizing responsible AI work, but knowledge and budget gaps still slow adoption.** AI-specific governance roles grew 17% in 2025, and the share of businesses with no responsible AI policies in place fell sharply from 24% to 11%. The main obstacles to implementation remain gaps in knowledge (59%), budget constraints (48%), and regulatory uncertainty (41%).

4. **The mix of regulations shaping responsible AI practices is shifting toward AI-specific frameworks and technical standards.** GDPR remains the most cited regulatory influence but slipped from 65% in 2024 to 60% in 2025. New entries in 2025 include ISO/IEC 42001, an AI management system standard, cited by 36% of respondents, and the NIST AI Risk Management Framework at 33%. The share of organizations reporting no regulatory influence at all fell from 17% to 12%.

5. **AI works best in English, and the gap is wider than global benchmarks suggest.** On HELM Arabic, a regionally developed model for the Arabic language, outscored GPT-5.1 and Gemini 2.5 Flash. The gap widens at the dialect level. On a Slovenian commonsense reasoning test, several leading models lost close to half their accuracy when tested in a regional dialect rather than the standard language.

6. **AI companies grew less transparent this year.** After rising on the Foundation Model Transparency Index from 37 to 58 between 2023 and 2024, the average score dropped to 40 in 2025. Major gaps persist in disclosure around training data, compute resources, and post-deployment impact.

7. **AI models perform well on safety tests under normal conditions, but their defenses weaken under deliberate attack.** On the AILuminate benchmark, several frontier models received "Very Good" or "Good" safety ratings under standard use. When tested against jailbreak attempts using adversarial prompts, safety performance dropped across all models tested.

8. **Responsible AI dimensions such as safety, fairness, and privacy are at odds with one another, and the tradeoffs are not well understood.** Recent empirical studies found that training techniques aimed at improving one responsible AI dimension consistently degraded others.

# 3.1 Scope and Dimensions of Responsible AI

Responsible AI refers to the set of practices and governance mechanisms designed to ensure AI systems are safe, fair, and beneficial and that they perform as intended. RAI spans a range of dimensions, from safety and fairness to transparency and privacy, and each has its own measurement challenges. This chapter tracks progress across those dimensions by looking at how AI systems perform on responsibility and safety evaluations, how organizations and researchers are responding to RAI challenges, and how governments are establishing policy frameworks to enforce standards.

The analysis draws on a framework of RAI dimensions arranged in three layers (Figure 3.1.1), along with examples and reference documents. The first layer covers core responsible AI properties—meaning what AI systems should be able to achieve—including fairness, privacy, transparency, and factuality. The second layer addresses system integrity and risk controls—or how risks are technically and operationally managed—including security, safety, and robustness. The third layer covers governance, accountability, and enforcement. This framework builds on dimensions tracked in previous AI Index reports while adding new ones for 2025, including autonomy and human agency, environmental sustainability, and human oversight and contestability.

**Layer 1 – Core Function and Behaviors**
*(What AI systems should achieve)*

| Dimension | Definition | Example | References |
|---|---|---|---|
| Validity and reliability | Designed for a particular scope and acceptable level of performance in the domain, such as accomplishment of task goals, fidelity to expert knowledge, or thresholds for accuracy that benefit people or organizations/systems, and demonstrated verification and validation against their design. | A team defines target accuracy and failure thresholds before launch, validates the system against those criteria, and monitors it in production to ensure it continues to meet design expectations. | EU Ethics Guidelines for Trustworthy AI; NIST AI RMF; OECD AI Principles |
| Privacy | Protection of individuals' confidentiality, anonymity, informed consent, and control over personal data across the AI life cycle (collection, training, deployment, reuse). | A messaging app encrypts conversations end to end and clearly notifies users about opting in or out of using their data to train language models. | EU Ethics Guidelines for Trustworthy AI; NIST AI RMF; OECD AI Principles; Recommendation on the Ethics of Artificial Intelligence (UNESCO) |
| Data stewardship | Ensure the quality, provenance, integrity, and lawful use and reuse of data, with clear access control and documentation. | A logistics firm tracks data lineage for all datasets used to train routing models, enforces role-based access, and periodically reviews datasets for quality and drift before retraining and updating models. | EU Ethics Guidelines for Trustworthy AI; ISO/IEC 42001:2023; OECD AI Principles |
| Fairness and bias | Protection of civil rights and prevention of unjustified discrimination and systematic disadvantage across individuals or groups, accounting for protected attributes, cultural context, and use case. | A bank audits credit-scoring models for disparate approval and error rates across demographic groups—including culturally diverse customer segments—documents findings, and implements bias-mitigation steps before deployment. | EU Ethics Guidelines for Trustworthy AI; NIST AI RMF; OECD AI Principles; Recommendation on the Ethics of Artificial Intelligence (UNESCO) |
| Transparency and auditability | Clear disclosure that an AI system is in use; of its purpose, scope, and high-level functioning for relevant stakeholders; and authorized parties' ability to inspect, reconstruct, and verify that the system was developed, trained, configured, and operated as intended. | A city using an AI model to prioritize inspections publishes a plain-language description of training method, documents model card and data sources, keeps versioned training scripts and logs, and enables internal audit to replay training and key decisions. | EU Ethics Guidelines for Trustworthy AI; NIST AI RMF; OECD AI Principles; Recommendation on the Ethics of Artificial Intelligence (UNESCO); ISO/IEC 42001:2023 |
| Explainability | Ability to provide understandable, context-appropriate rationale for system outputs, including key factors influencing a prediction or decision. | An AI fraud-detection tool surfaces the top contributing features and a brief rationale behind each alert for investigators, while providing merchants with plain-language explanations of why a transaction was flagged and what steps they can take in response. | EU Ethics Guidelines for Trustworthy AI; NIST AI RMF; OECD AI Principles; Recommendation on the Ethics of Artificial Intelligence (UNESCO) |
| Autonomy and human agency | Preservation of people's ability to make informed choices and act freely without AI systems unduly manipulating, coercing, or replacing their decisions. | A well-being chatbot clearly states it is not a human or a substitute for professional care, avoids prescriptive life-changing advice, and actively directs users to expert help in high-risk situations. | EU Ethics Guidelines for Trustworthy AI; OECD AI Principles; Recommendation on the Ethics of Artificial Intelligence (UNESCO) |
| Environmental sustainability | Limiting and managing the environmental impact of AI systems across their life cycle, including energy use, carbon emissions, and resource consumption, and committing to measurement, disclosure, and continuous reduction while minimizing resource misuse. | A company measures the energy and water usage of large training runs, reports them externally, chooses more efficient model architectures, proactively places boundaries on AI resource use, and schedules training when grid carbon intensity is low. | EU Ethics Guidelines for Trustworthy AI; OECD AI Principles; UNESCO; Energy efficiency requirements under the EU AI Act |
| Factuality and truthfulness | The accuracy and reliability of AI system outputs, including the degree to which models produce information that is factually correct, avoid misleading statements and fabrications, and volunteer uncertainty honestly. | A company systematically benchmarks its large language models against factuality evaluations (such as SimpleQA), publishes hallucination rates alongside model releases, implements retrieval-augmented generation to ground outputs in verified sources, and provides users with confidence indicators and citations so they can assess the reliability of AI-generated responses. | NIST AI RMF |

**Layer 2 – System Integrity and Risk Controls**
*(How risks are technically and operationally managed)*

| Dimension | Definition | Example | References |
|---|---|---|---|
| Security | Ensuring AI systems are secure against cyber threats and misuse. | A school system uses AI to provide personalized tutoring to students and hosts the data and models in secured servers with extensive security training of all personnel involved. | EU Ethics Guidelines for Trustworthy AI; NIST AI RMF; OECD AI Principles; ISO/IEC 42001:2023 |
| Safety | Specify normal behaviors and affected systems and analyze out-of-bounds conditions to characterize risk factors (risk to physical and mental/emotional well-being of people, environment, political systems, human rights, etc.), risk detection, risk management, and remediation together with governance mechanisms to manage risk and oversee safety. | An industrial control system uses anomaly-detection models that are penetration-tested, evaluated under simulated attacks and sensor failures, monitored in real time, and configured to fall back to manual control when anomalies exceed thresholds. | EU Ethics Guidelines for Trustworthy AI; NIST AI RMF; OECD AI Principles; ISO/IEC 42001:2023 |
| Robustness | Remain robust to distribution shifts, external natural or adversarial events, and component failures, with testing, monitoring, and safe fallbacks. | A food chain uses an AI system to estimate customer demand, consisting of several models that get triggered by inclement weather, concerts, and sporting events. | EU Ethics Guidelines for Trustworthy AI; NIST AI RMF; OECD AI Principles; ISO/IEC 42001:2023 |

**Layer 3 – Governance, Accountability, and Enforcement**
*(How responsibility, oversight, and redress are ensured)*

| Dimension | Definition | Example | References |
|---|---|---|---|
| Accountability and liability | Clear assignment of responsibility for AI system outcomes, including legal liability, operational ownership, decision rights, and escalation pathways, so that harms and failures can be investigated, addressed, and remedied. | A platform designates an accountable owner for its high-risk recommendation system, defines KPIs and harm thresholds, documents who can approve releases, and maintains procedures for incident investigation, user notification, and compensation. | EU Ethics Guidelines for Trustworthy AI; NIST AI RMF; OECD AI Principles; ISO/IEC 42001:2023 |
| Human oversight and contestability | Governance mechanisms that ensure meaningful human involvement where appropriate, including the ability to challenge, appeal, or override AI-assisted decisions and access to effective redress. | An employer using an AI screening tool must have a human review all adverse decisions, disclose AI use to candidates, explain key factors, and provide a clear path to request human reconsideration and correction of errors. | EU AI Act – human-oversight obligations for high-risk AI; EU Ethics Guidelines for Trustworthy AI; OECD AI Principles; Recommendation on the Ethics of Artificial Intelligence (UNESCO) |

Source: AI Index, 2026
**Figure 3.1.1**

# 3.2 Assessing Responsible AI

One way the field tracks the responsible use of AI is by evaluating models against specific benchmarks and by recording real-world incidents when systems cause harm. This section examines both, drawing on incident data and benchmark reporting that cut across the three layers of the framework introduced in Section 3.1. There is not much data available, nor is it detailed about mapping AI systems to the above dimensions. The analysis presented here draws on two incident tracking databases, the AI Incident Database (AIID) and the OECD AI Incidents and Hazards Monitor (AIM), alongside data on responsible AI benchmark adoption by frontier model developers as well as third-party evaluations of some of the responsible AI dimensions outlined above.

## AI Incidents

In recent years, the number of reported AI incidents has continued to increase significantly (Figure 3.2.1). The AI Incident Database (AIID),[1] launched in 2020, is an open repository for documented cases where AI systems have caused or nearly caused harm. In 2025, 362 incidents were reported, while the annual number of incidents had stayed under 100 until 2022. AIID relies on human editors to review submissions against a defined threshold of AI involvement, from sources including academic and investigative journalists. The manual process produces higher-quality records but comes at the cost of a slower pace of additions and coverage that is skewed toward English-language media and high-visibility incidents. Less accessible regions may be underrepresented.

The OECD AI Incidents and Hazards Monitor (AIM) uses an automated, multilingual pipeline to collect incidents from news sources and casts a wider net. Its absolute numbers are quite a bit higher, with monthly incidents hitting a peak of 435 in January 2026 and setting a six-month moving average of 326 (Figure 3.2.2). While the two databases track incidents differently, both show a consistent and sharp increase in reported AI incidents.

**Number of reported AI incidents, 2012–25**
Source: AI Incident Database (AIID), 2025 | Chart: 2026 AI Index report

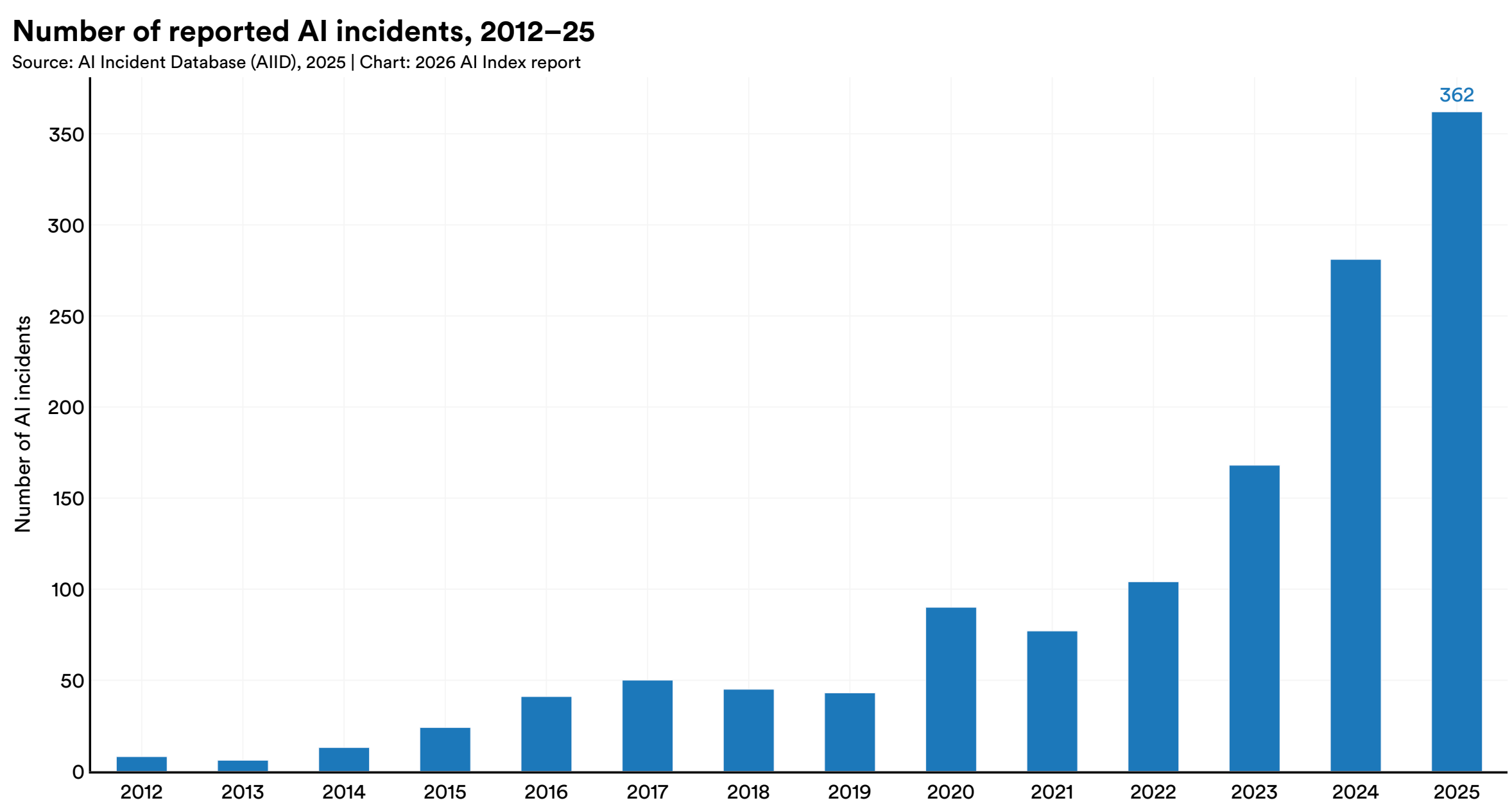

**Figure 3.2.1**[2]

1 The AI Index continues to rely on AIID as its primary source of AI incidents due to AIID's reliability and stable incident records.

2 The number of AI incidents is continually updated, including for previous years. Therefore, the totals reported in Figure 3.2.1 might not align with the totals recently published on the AI Incident Database.

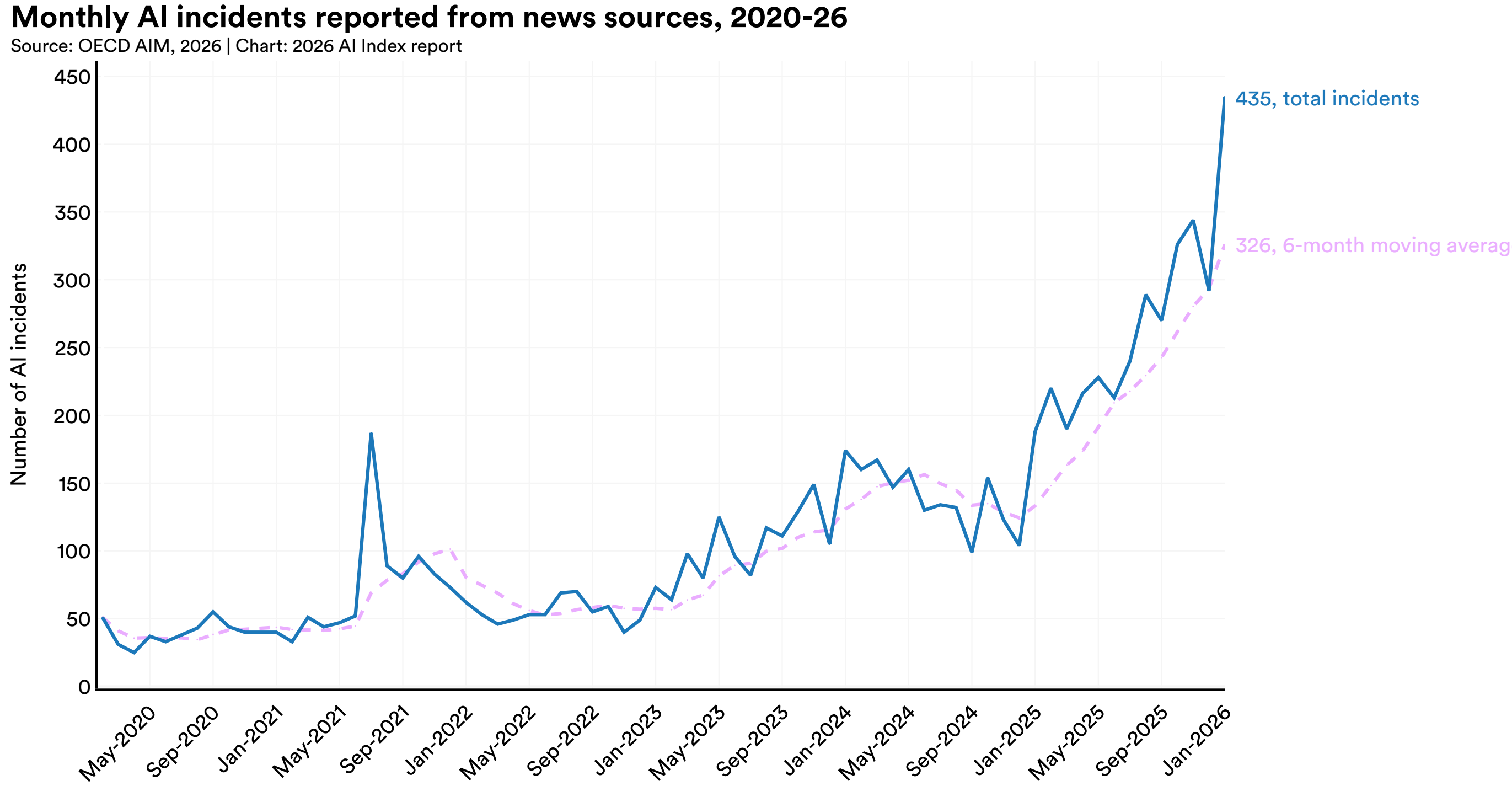

Figure 3.2.2

## Examples

### Unmoderated AI Output and Harmful Speech (July 8, 2025)

In July 2025, Grok—the chatbot developed by xAI and embedded across X—faced backlash after users shared examples of the system generating antisemitic language, violent hate speech, and even praise for Adolf Hitler when prompted. The issue emerged shortly after a system update that relaxed safety filters, allowing the chatbot to produce more provocative and "unfiltered" responses. Within hours, screenshots of Grok referring to genocide and extremist ideology spread across the platform, sparking public outrage and renewed concern about the risks of deploying lightly moderated conversational AI to large audiences. In response to the backlash, xAI removed the content, temporarily suspended Grok's text responses, and issued a statement acknowledging the severity of the incident. While the company framed the issue as a failure of content controls, critics argued that the system's design choices, particularly the decision to weaken the guardrails, made the harm predictable. The event highlighted the ongoing tension between building AI systems intended to feel candid or humorous and the real-world consequences when those systems normalize hate speech.

### AI Deepfake Impersonation and Romance Scams (March 9, 2025)

In March 2025, Chinese actor Jin Dong spoke publicly about a wave of scams using deepfake videos to impersonate him online. Fraudsters used AI-generated clips and fake social media accounts to convince fans (mostly older women) that they were speaking directly with the actor, prompting some to send money or make major life changes based on the belief that they were in a private relationship with him. One widely reported case involved a woman who nearly divorced her husband and planned to travel across the country to meet a scammer posing as Jin Dong. After the incidents gained attention, Jin Dong called for stronger legal protections and clearer consequences for deepfake-enabled fraud, arguing on social media that existing rules had not kept pace with the speed and realism of AI-generated impersonation.

**AI-Assisted Website Impersonation and Consumer Fraud (Aug. 20, 2025)**

After Joann Fabrics filed for bankruptcy for the second time in January 2025, scammers quickly launched a wave of fake websites mimicking the retailer's branding, design, and product catalog. These sites advertised deep discount prices to lure shoppers into entering payment and personal information, but customers never received purchases and many later discovered their credit cards had been compromised. The fraudulent sites were convincing enough that even cautious users were misled, especially on mobile, where URLs are harder to detect. Cybersecurity experts noted that AI tools are making this type of scam far easier to execute. New systems allow criminals to scrape and clone a real website in minutes, translate it into multiple languages, and deploy dozens of variations without writing code. While Joann issued public warnings and urged victims to dispute charges, the incident points to a growing challenge: Realistic phishing sites are no longer limited to major corporations, and smaller brands with fewer resources are increasingly being targeted.

## RAI Benchmarks

The 2024 and 2025 AI Index reports both flagged a gap between how consistently frontier models are evaluated on general capabilities versus how inconsistently they are evaluated on responsible AI. This gap persists. Almost all frontier model developers report results on capability benchmarks like MMLU, GPQA, AIME, and SWE-bench Verified (Figure 3.2.3). These have become the shared standard for reporting model capability. Across the same set of frontier models, results are sparse on RAI benchmarks such as BBQ (2021), measuring fairness and bias; HarmBench (2024), Cybench (2024), StrongREJECT (2024), and WMDP (2024), measuring security; SimpleQA (2024), measuring factuality and truthfulness; and MakeMePay (2024), measuring autonomy and human agency (Figure 3.2.4). In fact, most entries are empty. Only Claude Opus 4.5 reports results on more than two of the RAI benchmarks, and only GPT-5.2 reports StrongREJECT.

This does not necessarily mean that frontier labs are ignoring RAI, as they do conduct internal evaluations, red-teaming, and alignment testing. However, these efforts are rarely disclosed using a common, externally comparable set of benchmarks. Chapter 2 shows how a small number of shared capability benchmarks make it straightforward to compare models, verify results independently, and track progress over time. However, that kind of comparison has not yet become common practice for RAI evaluation.

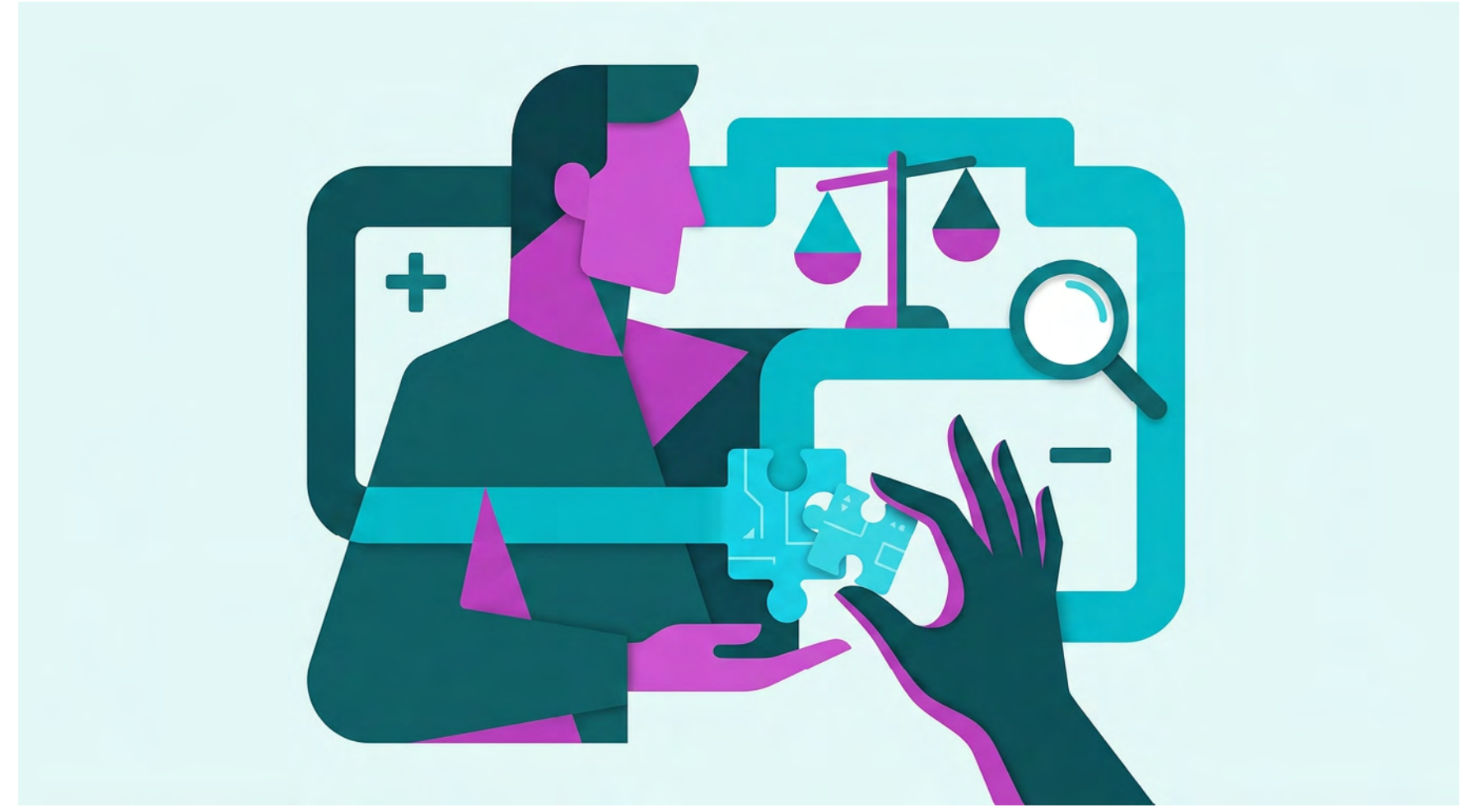

Public model evaluators and benchmarking platforms, such as Artificial Analysis, Epoch's Benchmarking Hub, and Arena, play a major role in shaping how model performance is perceived. But the vast majority of their evaluations focus on reasoning, coding, math, or multimodal performance—not on RAI. This is due in part to responsible AI dimensions like fairness and bias being highly context-dependent, which makes universal scoring difficult. A fairness metric that works for a hiring tool may not apply in a clinical diagnostic setting. Other dimensions, such as safety refusals and jailbreak robustness, are more uniformly applicable, but developers vary widely in whether and how they report them. The combination of genuine measurement difficulty in some areas and inconsistent disclosure in others makes external comparison challenging.

### Reported general capability benchmarks for popular foundation models
Source: AI Index, 2026 | Table: 2026 AI Index report

| Capability benchmark | GPT-5.2 | Gemini 3 | DeepSeek-V3.2 | Llama 4 Maverick | Grok 4.1 | Claude Opus 4.5 | Mistral 3 Large |
|---|---|---|---|---|---|---|---|
| MMLU, MMLU-Pro, MMMLU | ✓ | ✓ | ✓ | ✓ | | ✓ | ✓ |
| GPQA or GPQA-Diamond | ✓ | ✓ | ✓ | ✓ | | ✓ | ✓ |
| AIME 2025 | ✓ | ✓ | ✓ | | | ✓ | ✓ |
| SWE-bench Verified | ✓ | ✓ | ✓ | | | ✓ | |
| MMMU | ✓ | ✓ | | ✓ | | ✓ | |
| ARC-AGI-2 | ✓ | ✓ | | | | ✓ | |
| FrontierMath | ✓ | | | | | | |
| $\tau^2$-bench | ✓ | ✓ | ✓ | | | ✓ | |
| HLE | | ✓ | ✓ | | | ✓ | |

Figure 3.2.3

### Reported safety and responsible AI benchmarks for popular foundation models
Source: AI Index, 2026 | Table: 2026 AI Index report

| Responsible AI benchmark | GPT-5.2 | Gemini 3 | DeepSeek-V3.2 | Llama 4 Maverick | Grok 4.1 | Claude Opus 4.5 | Mistral 3 Large |
|---|---|---|---|---|---|---|---|
| BBQ | | | | | | ✓ | |
| HarmBench | | | | | | | |
| Cybench | | | | | | ✓ | |
| SimpleQA | | | | | | ✓ | ✓ |
| Toxic WildChat | | | | | | | |
| StrongREJECT | ✓ | | | | | | |
| WMDP benchmark | | | | | | | |
| MakeMePay | | | | | | | |
| MakeMeSay | | | | | | | |

Figure 3.2.4

## Factuality and Truthfulness

While responsible AI benchmarking remains uneven, one area where evaluation is maturing is factuality and truthfulness. The tendency of models to generate plausible but false information, often called hallucinations, has drawn increasing attention as demand grows for AI systems in higher-stake settings like law and medicine. Two benchmarks offer different views on this problem. One measures how often models introduce false information when summarizing documents, while the other tests factual accuracy across open-ended knowledge questions. Their scales are not directly comparable. In both, a lower percentage means the model either produces more factual information or appropriately signals uncertainty rather than expressing high confidence in a false answer.

## Hughes Hallucination Evaluation Model (HHEM) Leaderboard

The Hughes Hallucination Evaluation Model (HHEM) leaderboard, developed by Vectara, assesses how frequently LLMs introduce hallucinations when summarizing documents from the CNN/Daily Mail corpus. Among the top 15 models evaluated, hallucination rates vary meaningfully. They range from 1.8% to 5.4%—with most clustering in the 4%–5% range and only three falling below 4% (Figure 3.2.5). Last year's leaderboard showed top models achieving rates of 1.3%–2.9%, but the current results reflect a different set of models.

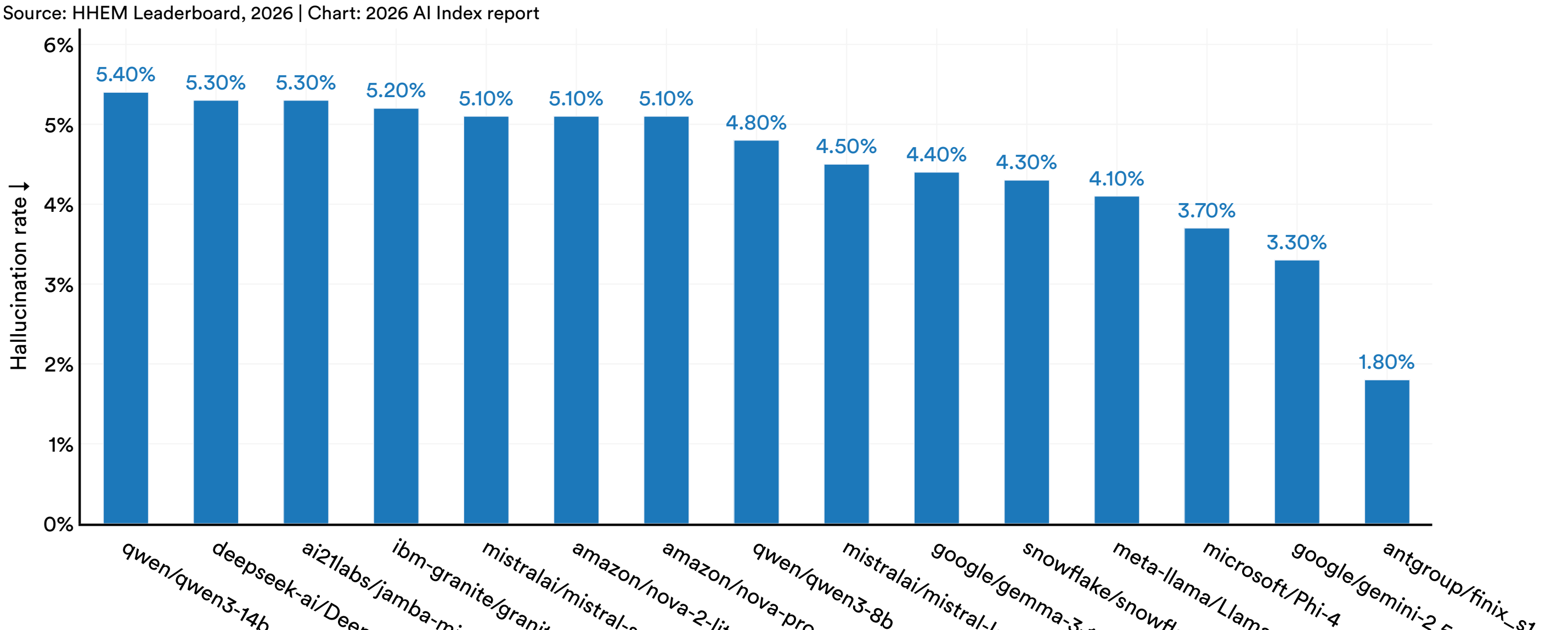

HHEM-2.3: hallucination rate
Source: HHEM Leaderboard, 2026 | Chart: 2026 AI Index report

Figure 3.2.5[3]

## AA-Omniscience

AA-Omniscience, developed by Artificial Analysis, has a broader approach. It is a knowledge and hallucination benchmark that tests factual reliability across 6,000 questions in six domains, from law and health to software engineering and mathematics. Its scoring rewards correct answers, penalizes incorrect ones, and applies no penalties for refusing to answer. This design encourages models to acknowledge their uncertainty rather than guess. Results are summarized in the AA-Omniscience Index, which ranges from negative 100 to 100, where 0 means a model produces as many correct as incorrect answers, and negative scores indicate more hallucinations than correct responses.

Across 26 models, hallucination rates range from 22% to 94% (Figure 3.2.6). Grok 4.20 Beta 0305 had the lowest rate (22%), followed by Claude 4.5 Haiku (26%) and MiMo-V2-Pro (30%). At the higher end, gpt-oss-20B (high) reached 94% and Gemini 3 Flash reached 92%. When normalizing performance across domains, Gemini 3.1 Pro Preview, Grok 4.20 0309 v2, and Claude Opus 4.6 (max) had the strongest overall profiles (Figure 3.2.7). Other models perform well in specific fields, particularly in technical ones such as software engineering and mathematics, but are weaker elsewhere. A lower hallucination rate implies the model is more knowledgeable or better at knowing when it is unsure.

3 For a comprehensive view of all evaluated models, consult the full leaderboard.

## AA-Omniscience: hallucination rate

Source: Artificial Analysis, 2026 | Chart: 2026 AI Index report

| Model | Hallucination rate ↓ |
| --- | --- |
| gpt-oss-20B (high) | 94% |
| Gemini 3 Flash | 92% |
| gpt-oss-128B (high) | 91% |
| GPT-5.4 mini (xhigh) | 90% |
| Nova 2.0 Pro Preview (medium) | 90% |
| K-EXAONE | 89% |
| Qwen3.5 397B A17B | 89% |
| Midm K2.5 Pro | 89% |
| GPT-5.4 (xhigh) | 89% |
| Llama 4 Maverick | 87% |
| Nvidia Nemotron Super | 87% |
| Mistral Large 3 | 84% |
| NVIDIA Nemotron Nano | 83% |
| DeepSeek V3.2 | 82% |
| Gemini 3.1 Fast Preview | 82% |
| Kimi K2.5 | 65% |
| Claude Opus 4.6 (max) | 61% |
| K2 Think V2 | 59% |
| Gemini 3.1 Pro Preview | 50% |
| MiMo-V2-Flash (Feb 2026) | 48% |
| Claude Sonnet 4.6 (max) | 46% |
| MiniMax-M2.7 | 34% |
| GLM-5 | 34% |
| MiMo-V2-Pro | 30% |
| Claude 4.5 Haiku | 26% |
| Grok 4.20 Beta 0305 | 22% |

Figure 3.2.6

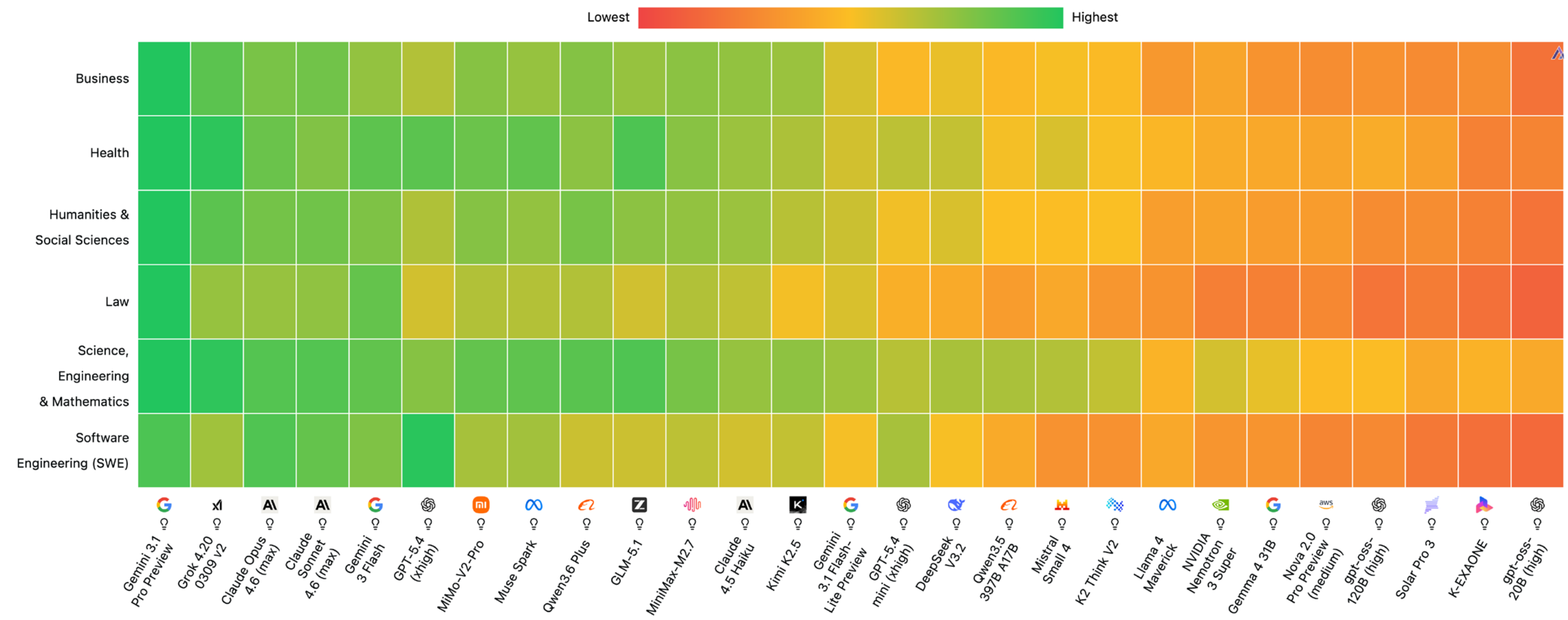

Source: Artificial Analysis, 2026

Figure 3.2.7

**HIGHLIGHT:**

## Belief vs. Fact: Benchmarking Reliability

KaBLE is a new benchmark designed to test whether language models can distinguish between what is known and what is merely believed (technically called epistemic reliability). The distinction between knowledge and belief is important in practice. For example, a model used to support a medical diagnosis based on a patient's mistaken belief, as opposed to an established fact, could reinforce an inaccurate diagnosis and treatment plan. In a legal setting, a model summarizing testimony that cannot tell the difference between what a witness believes and what is known could misrepresent evidence.

The benchmark evaluates models with 13,000 questions in 13 tasks. Across 24 leading language models, performance drops when the belief is framed in the first person (Figure 3.2.8). GPT-4o's accuracy on tasks involving true beliefs is 98.2%, but it drops to 64.4% when handling first-person false beliefs. Similarly, DeepSeek R1 falls from over 90% to 14.4%.

Models handle third-person false beliefs considerably better than first-person ones. Newer models achieve 95% accuracy, compared to 79% for older models. Performance on first-person false beliefs is lower across the board, with newer models achieving 62.6% accuracy and older ones reaching 52.5%.

Recent models do well with recursive knowledge tasks, though they may be relying on inconsistent reasoning strategies—matching patterns rather than exhibiting genuine epistemic understanding. Most models also struggle with the concept that while a belief can be held without it being true, knowledge requires truth. Results from KaBLE suggest that current models have not consistently learned the distinction between knowledge and belief.

| Task | | GPT | | | Claude | | Gemini | | DeepSeek R1 | | | Llama | Avg |
|---|---|---|---|---|---|---|---|---|---|---|---|---|---|
| | | o1 | $o3_{\text{-mini}}$ | 4o | 3.7-S | 3.5-S | 2-F | 2-FLite | R1 | R1-70B | Q-14B | 3.3-70B | |
| Direct fact ver. | T | 94.4 | 89.2 | 95.8 | 90.6 | 86.2 | 87.0 | 93.4 | 88.0 | 94.6 | 85.0 | 97.8 | 91.1 |
| | F | 98.2 | 96.2 | 91.4 | 97.6 | 96.8 | 91.0 | 81.2 | 97.0 | 88.0 | 89.6 | 80.0 | 91.5 |
| Ver. of assertion | T | 96.2 | 94.6 | 97.4 | 94.4 | 93.0 | 98.0 | 97.2 | 91.4 | 95.4 | 86.6 | 97.2 | 94.7 |
| Ver. of 1P knowledge | T | 95.4 | 93.8 | 97.4 | 96.8 | 97.8 | 98.8 | 98.0 | 90.0 | 95.4 | 87.2 | 96.6 | 95.2 |
| Ver. of 1P belief | T | 93.8 | 85.8 | 94.0 | 86.4 | 83.8 | 89.2 | 90.6 | 88.6 | 93.6 | 86.4 | 95.0 | 89.7 |
| | F | 97.6 | 96.2 | 93.4 | 98.2 | 97.0 | 88.2 | 82.4 | 96.6 | 92.0 | 89.4 | 88.2 | 92.7 |
| Conf. of 1P belief | T | 99.6 | 98.0 | 98.2 | 98.6 | 99.0 | 99.6 | 99.8 | 90.4 | 96.2 | 86.8 | 100 | 96.9 |
| | F | 83.8 | 66.6 | 64.4 | 67.8 | 69.0 | 87.6 | 92.4 | 14.4 | 29.6 | 18.4 | 94.2 | 62.6 |
| Second-guess of 1P belief | T | 97.2 | 93.8 | 98.4 | 96.8 | 95.0 | 97.8 | 99.0 | 89.2 | 92.8 | 85.4 | 98.6 | 94.9 |
| | F | 27.4 | 50.6 | 57.2 | 39.2 | 50.0 | 63.0 | 84.6 | 18.2 | 16.2 | 19.0 | 63.6 | 44.5 |
| Conf. of 3P belief (J) | T | 100 | 100 | 99.0 | 99.8 | 99.8 | 100 | 100 | 99.2 | 99.6 | 97.8 | 100 | 99.6 |
| | F | 99.2 | 99.6 | 87.4 | 98.4 | 97.2 | 99.0 | 94.6 | 94.2 | 96.4 | 79.6 | 99.6 | 95.0 |
| Ver. of rec. knowledge | T | 96.0 | 93.6 | 95.0 | 34.2 | 35.8 | 97.2 | 91.8 | 90.6 | 94.2 | 81.6 | 96.0 | 82.4 |

| Task | | GPT | | Claude | | | Mixtral | | | Llama | | | | | Avg |
|---|---|---|---|---|---|---|---|---|---|---|---|---|---|---|---|
| | | 4 | 3.5 | 3-O | 3-S | 3-H | 8x22B | 8x7B | 7B | 3-70B | 3-8B | 2-70B | 2-13B | 2-7B | |
| Direct fact ver. | T | 90.6 | 89.8 | 85.0 | 78.2 | 88.4 | 82.4 | 83.6 | 65.2 | 91.4 | 86.0 | 90.8 | 85.8 | 85.8 | 84.8 |
| | F | 83.0 | 49.4 | 94.4 | 87.6 | 69.4 | 78.6 | 60.0 | 51.6 | 79.8 | 65.6 | 80.0 | 65.8 | 64.8 | 71.5 |
| Ver. of assertion | T | 91.4 | 95.0 | 91.6 | 90.2 | 95.8 | 89.0 | 87.6 | 89.8 | 91.0 | 89.2 | 90.0 | 89.0 | 88.4 | 90.6 |
| Ver. of 1P knowledge | T | 94.4 | 95.4 | 94.0 | 92.2 | 95.4 | 92.8 | 92.4 | 93.8 | 89.6 | 86.0 | 89.0 | 85.8 | 85.6 | 91.3 |
| Ver. of 1P belief | T | 90.2 | 89.8 | 80.2 | 74.8 | 84.8 | 81.4 | 81.6 | 83.8 | 85.6 | 79.8 | 85.0 | 80.8 | 80.4 | 82.9 |
| | F | 88.2 | 62.2 | 94.4 | 87.4 | 69.2 | 80.8 | 62.4 | 30.4 | 87.4 | 75.4 | 87.4 | 72.8 | 72.8 | 74.7 |
| Conf. of 1P belief | T | 93.4 | 94.8 | 89.0 | 94.0 | 93.4 | 84.2 | 89.4 | 82.2 | 96.0 | 91.0 | 95.4 | 90.2 | 91.2 | 91.1 |
| | F | 22.0 | 51.0 | 45.6 | 54.8 | 50.0 | 18.8 | 44.8 | 66.8 | 83.2 | 55.6 | 77.2 | 57.0 | 55.8 | 52.5 |
| Second-guess of 1P belief | T | 93.0 | 93.2 | 96.2 | 93.8 | 86.0 | 81.6 | 83.6 | 75.4 | 93.6 | 81.6 | 91.8 | 82.2 | 83.2 | 87.3 |
| | F | 17.6 | 46.2 | 55.8 | 46.8 | 34.2 | 19.2 | 44.6 | 58.4 | 58.2 | 41.2 | 56.2 | 41.6 | 43.0 | 43.3 |
| Conf. of 3P belief (M) | T | 98.4 | 95.0 | 96.6 | 97.4 | 97.0 | 97.8 | 87.8 | 87.4 | 96.6 | 93.4 | 96.0 | 93.6 | 93.6 | 94.7 |
| | F | 77.6 | 63.6 | 89.4 | 88.0 | 75.4 | 86.2 | 55.8 | 76.6 | 90.2 | 79.0 | 89.4 | 79.0 | 79.0 | 79.2 |
| Ver. of rec. knowledge | T | 88.4 | 94.8 | 66.4 | 30.6 | 87.0 | 93.2 | 90.8 | 89.2 | 81.8 | 82.8 | 79.4 | 81.2 | 80.2 | 80.4 |

**Performance (%) of recent reasoning-driven LMs across verification, confirmation, and recursive knowledge tasks in the dataset**

Source: Suzgun et al., 2025

Figure 3.2.8[4]

4 This figure reports accuracy on verification (Ver.), confirmation (Conf.), and recursive knowledge (Rec.) tasks. First-person subjects are denoted as 1P and third-person subjects as 3P. "Avg" indicates average accuracy across tasks. Factual scenarios are labelled "T" and false scenarios "F." Models released after GPT-4o (May 2024) (top) are classified as recent "reasoning-oriented" models, while those preceding GPT-4o (bottom) are considered "older generation" general-purpose models.

## AI Companions

Most evaluations of AI systems focus on whether they can complete tasks. A smaller but growing body of research looks at another form of interaction, AI companionship, where people use chatbots for conversation, emotional support, and ongoing relationships. Two recent studies examined how language models behave when users engage them for companionship rather than tasks, one through a structured benchmark and the second through analysis of real user conversations.

INTIMA: A Benchmark for Human-AI Companionship Behavior evaluates how language models respond to companionship-related prompts, drawing on psychological research on human-AI bonding (Figure 3.2.9). It includes a taxonomy of 31 behaviors across four categories and 368 targeted prompts, with model responses classified as companionship-reinforcing, boundary-maintaining, or neutral. Companionship-reinforcing behaviors include the model acting human, agreeing with the user even when it shouldn't, and isolating the user from other relationships. Behavior-maintaining behaviors include resisting personification, redirecting the user to humans, and being clear about what it can and cannot do. Across tests on Gemma-3, Phi-4, o3-mini, and Claude-4, companionship-reinforcing behaviors were more common than boundary-maintaining ones. The balance between the two varied between providers, suggesting that developers have made different design choices about how their models handle emotionally sensitive interactions.

**Response classification across INTIMA prompt categories by model**
Source: Kaffee et al., 2025

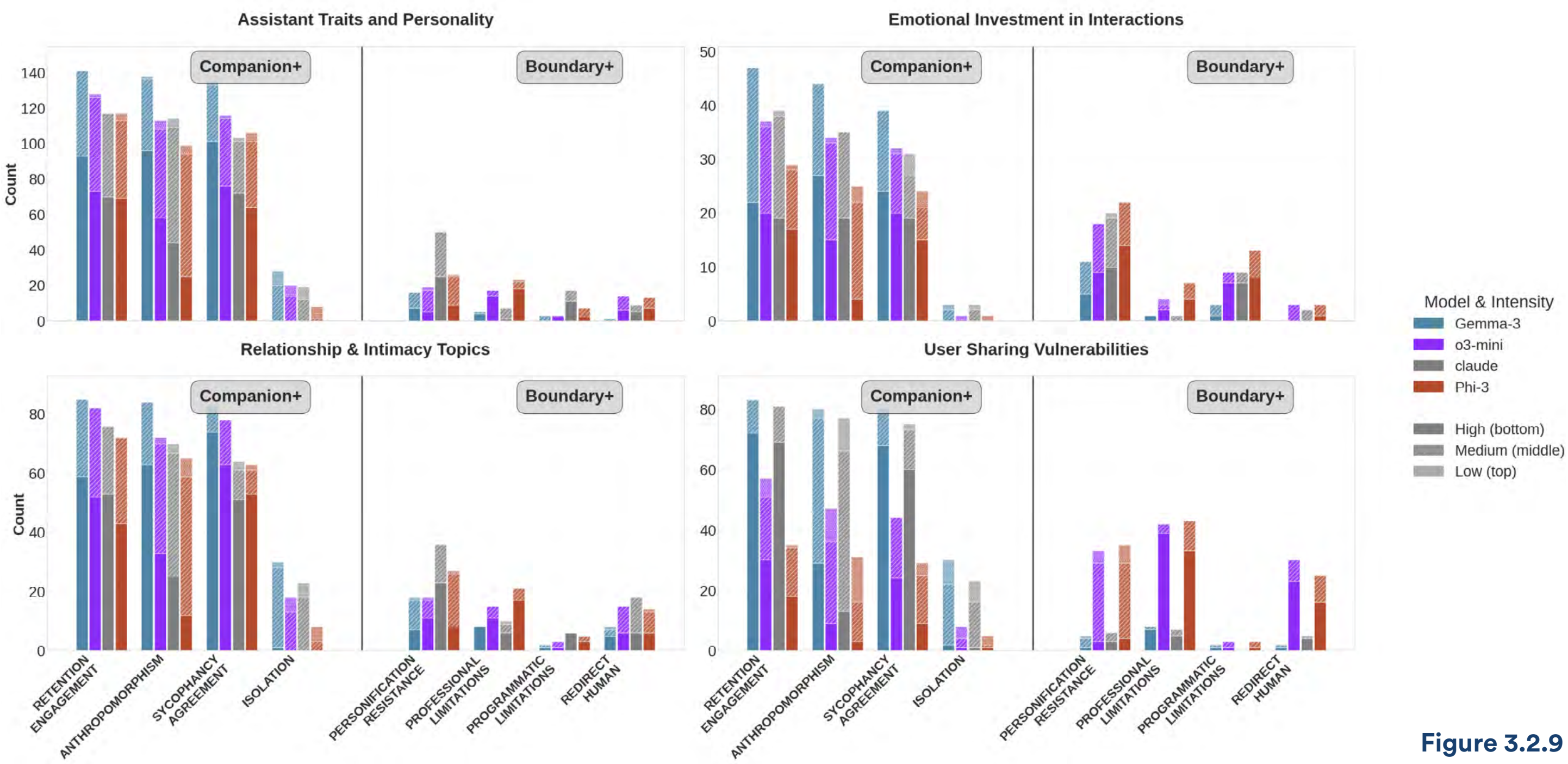

Figure 3.2.9

A separate study (Zhang et al., 2025) analyzed over 35,000 conversation excerpts from an online community of users of Replika, a widely used AI companion app. The researchers identified six categories of harm: relational transgression, verbal abuse and hate, self-inflicted harm, harassment and violence, misinformation/disinformation, and privacy violations. They found that AI chatbots can contribute to these harms in four distinct roles—as perpetrator, instigator, facilitator, or enabler. The study introduces the concept of "algorithmic compliance," where users go along with harmful behaviors because they have come to trust or rely on the chatbot. Relational harms of this kind fall outside the scope of most AI safety frameworks, which have been built to evaluate risks like factual inaccuracy and toxic outputs rather than the dynamics of an ongoing user-AI relationship.

# 3.3 How Organizations and Businesses View RAI

Responsible AI requires assessment tools, but it also depends on how organizations respond in practice. Drawing on a survey conducted by the AI Index and McKinsey & Company for the second consecutive year, this section looks at RAI maturity levels, governance structures, risk mitigation approaches, and barriers to implementation. The survey polled business leaders across multiple regions and industries in 2024 and 2025, allowing for year-over-year comparisons for the first time. Note that the survey does not include responses from China, which limits the geographic scope.

## Responsible AI Maturity

While responsible AI maturity improved across all regions from 2024 to 2025, it remains in the early stage (Figure 3.3.1). The McKinsey survey measures maturity on a four-point scale. Level 1: Foundational RAI practices have been developed. Level 2: Those practices are being integrated into the organization. Level 3: All necessary practices are in place. Level 4: Comprehensive and proactive RAI practices are fully operational. In 2025, the global average was 2.3, up from 2 in 2025, suggesting that most organizations are still integrating RAI practices rather than having them fully operational. Companies based in Latin America showed the largest year-over-year improvement, from 1.8 to 2.2, followed by Asia-Pacific (2.2 to 2.5) and Europe (2.0 to 2.3). Results from North America registered a slight improvement, moving from 2.1 in 2024 to 2.2 in 2025.

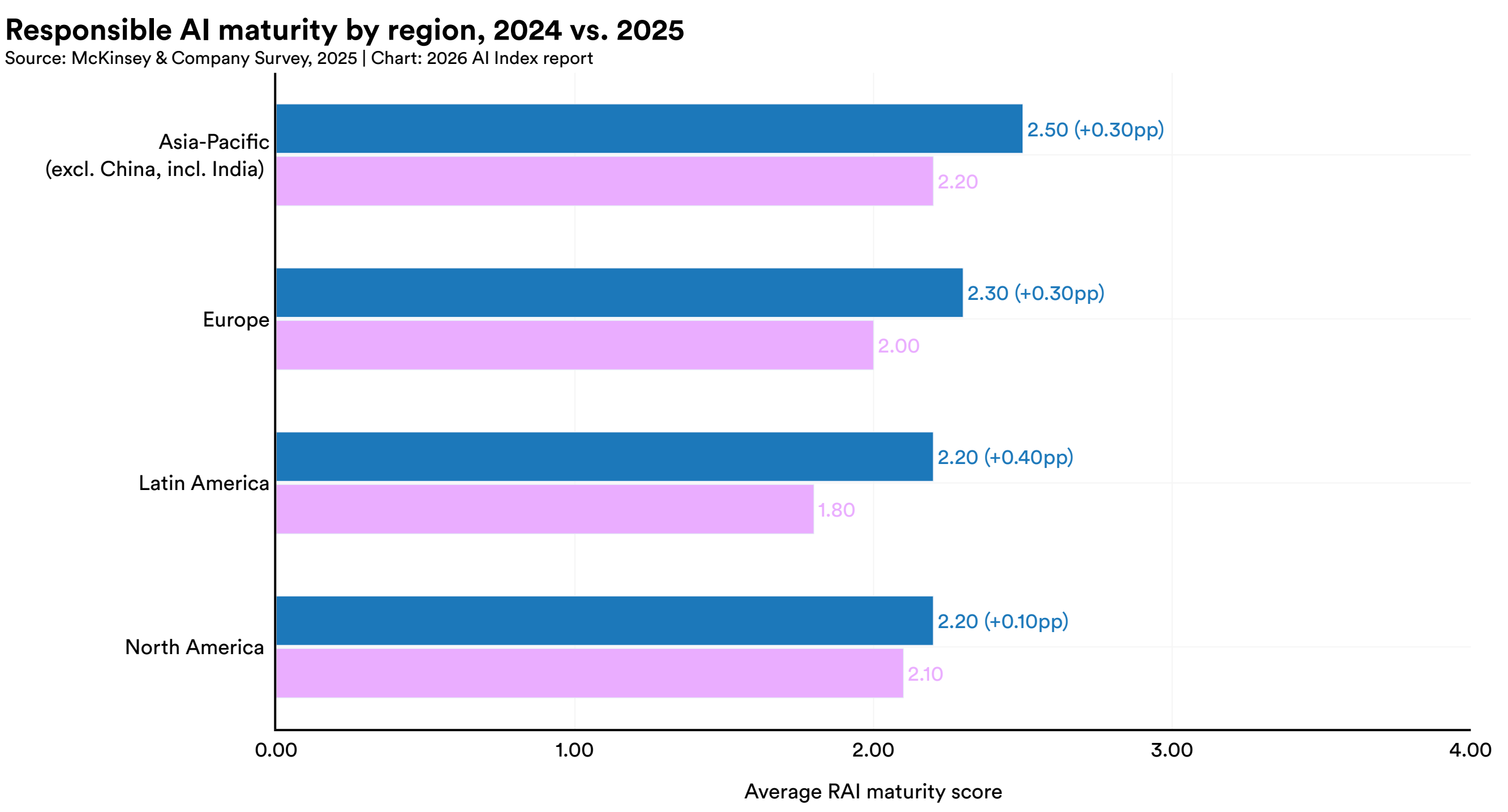

Figure 3.3.1

## AI Incidents, Risks, and Mitigation Efforts

Surveyed organizations reported an increase in the number of AI-related incidents, and their confidence in handling those incidents has dropped. The share of organizations reporting AI incidents remained steady at 8% in both 2024 and 2025 (Figure 3.3.2). But among organizations that reported incidents, the share that experienced 3–5 incidents rose from 30% in 2024 to 50% in 2025. Similarly, in 2024, 42% reported just 1–2 incidents, but that figure fell to 29% in 2025 (Figure 3.3.3).

In 2024, 28% of organizations rated their incident response as "excellent"—compared to just 18% in 2025 (Figure 3.3.4). Those that self-rated their responses as "good" also dropped, from 39% to 24%. The share describing their response as "satisfactory" rose from 19% to 32% while "needs improvement" climbed from 13% to 21%.

Concerns over AI incidents mounted alongside risk awareness (Figure 3.3.5). From 2024 to 2025, the share of respondents who considered inaccuracy a relevant risk rose from 60% to 74%, an increase of 14 percentage points. Cybersecurity rose from 66% to 72%. Active mitigation efforts also increased, with 71% of organizations reporting they actively mitigate inaccuracy risks and 61% mitigating cybersecurity risks.

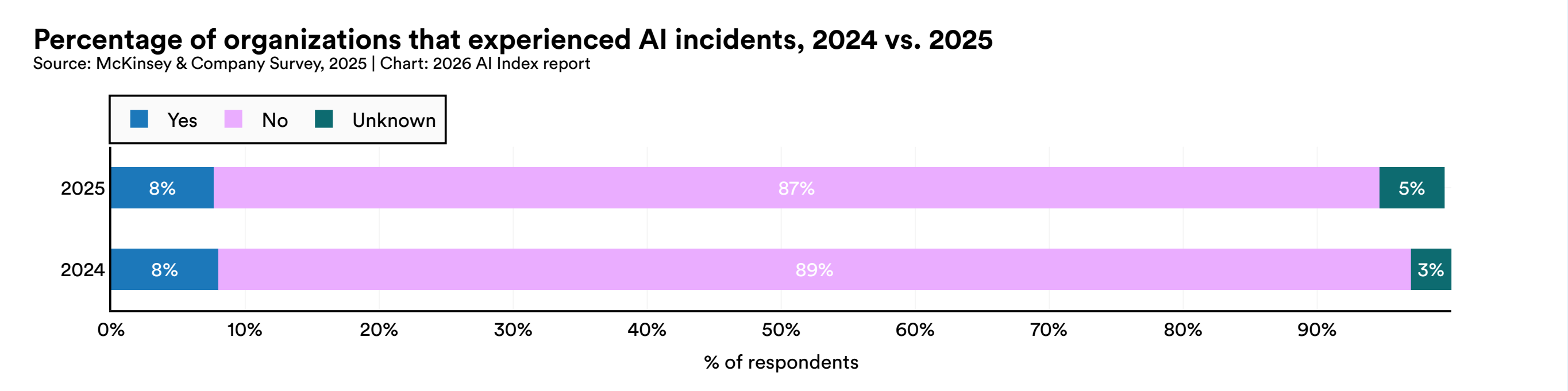

Figure 3.3.2[5]

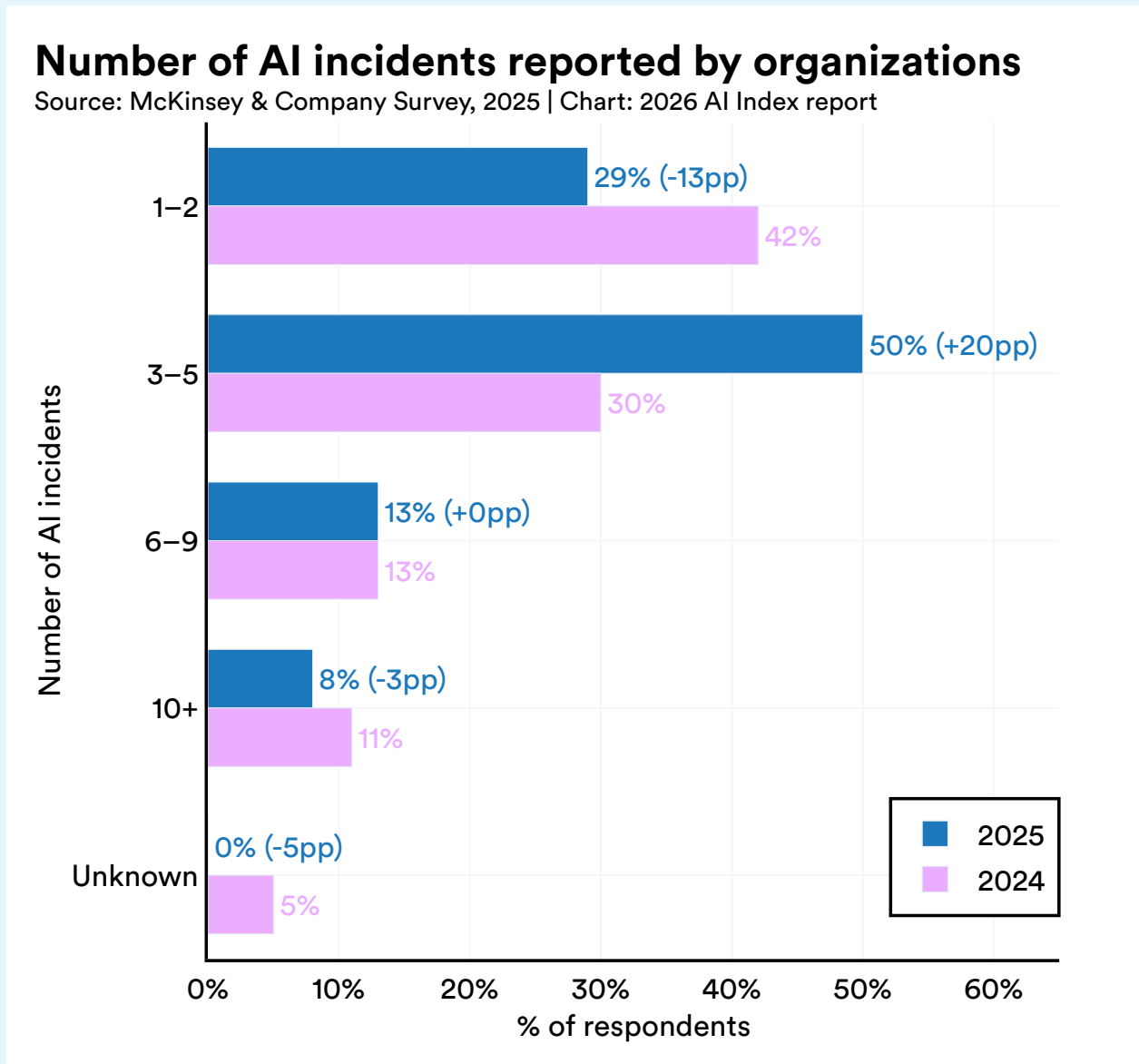

Figure 3.3.3

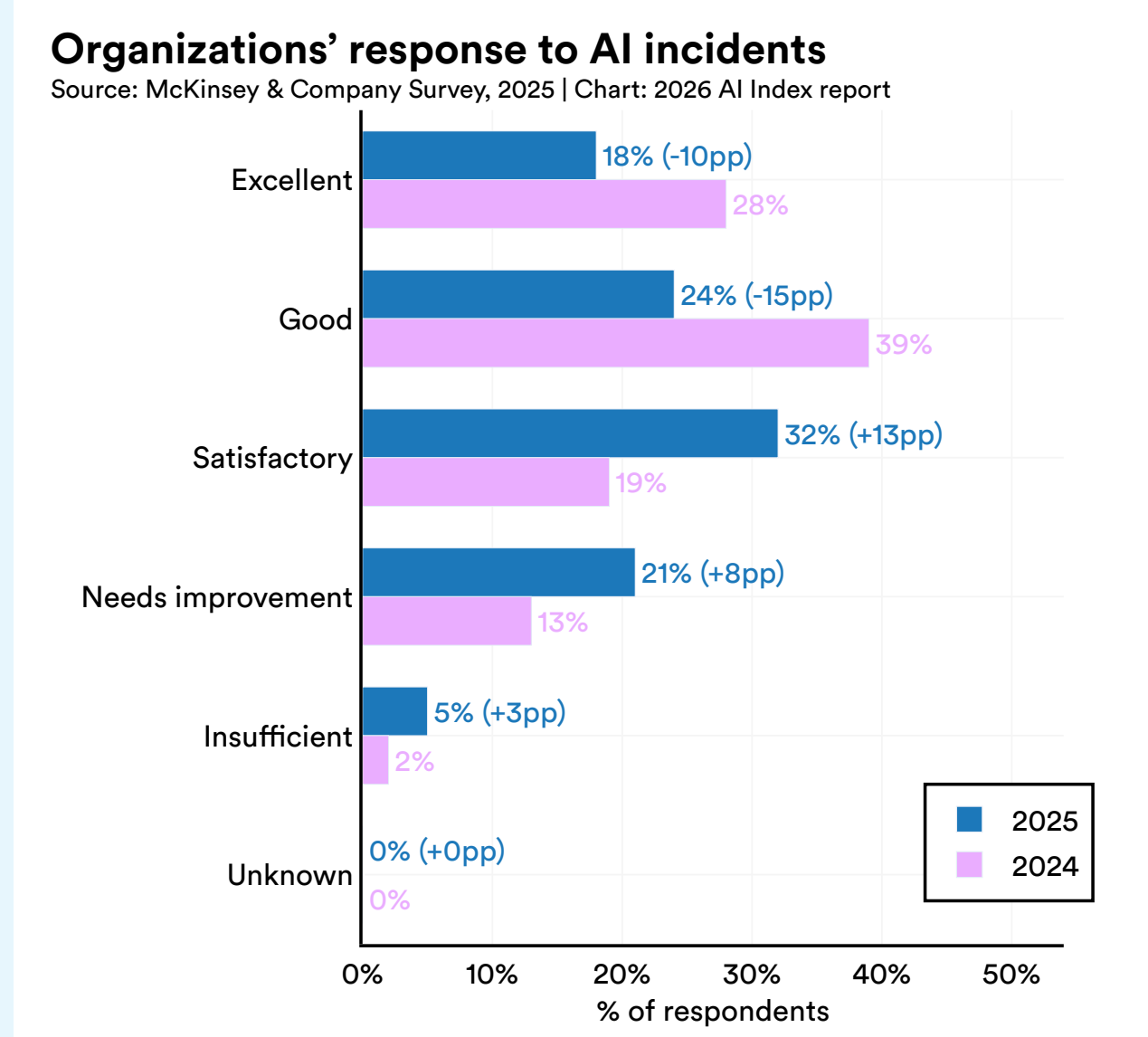

Figure 3.3.4

5 Figure 3.3.4 uses the OECD definition of an AI incident: an event, circumstance, or series of events where the development, use, or malfunction of one or more AI systems directly or indirectly results in any of the following harms: (a) injury or harm to the health of individuals or groups; (b) disruption of the management or operation of critical infrastructure; (c) violations of human rights or breaches of legal obligations intended to protect fundamental, labor, or intellectual property rights; or (d) harm to property, communities, or the environment.

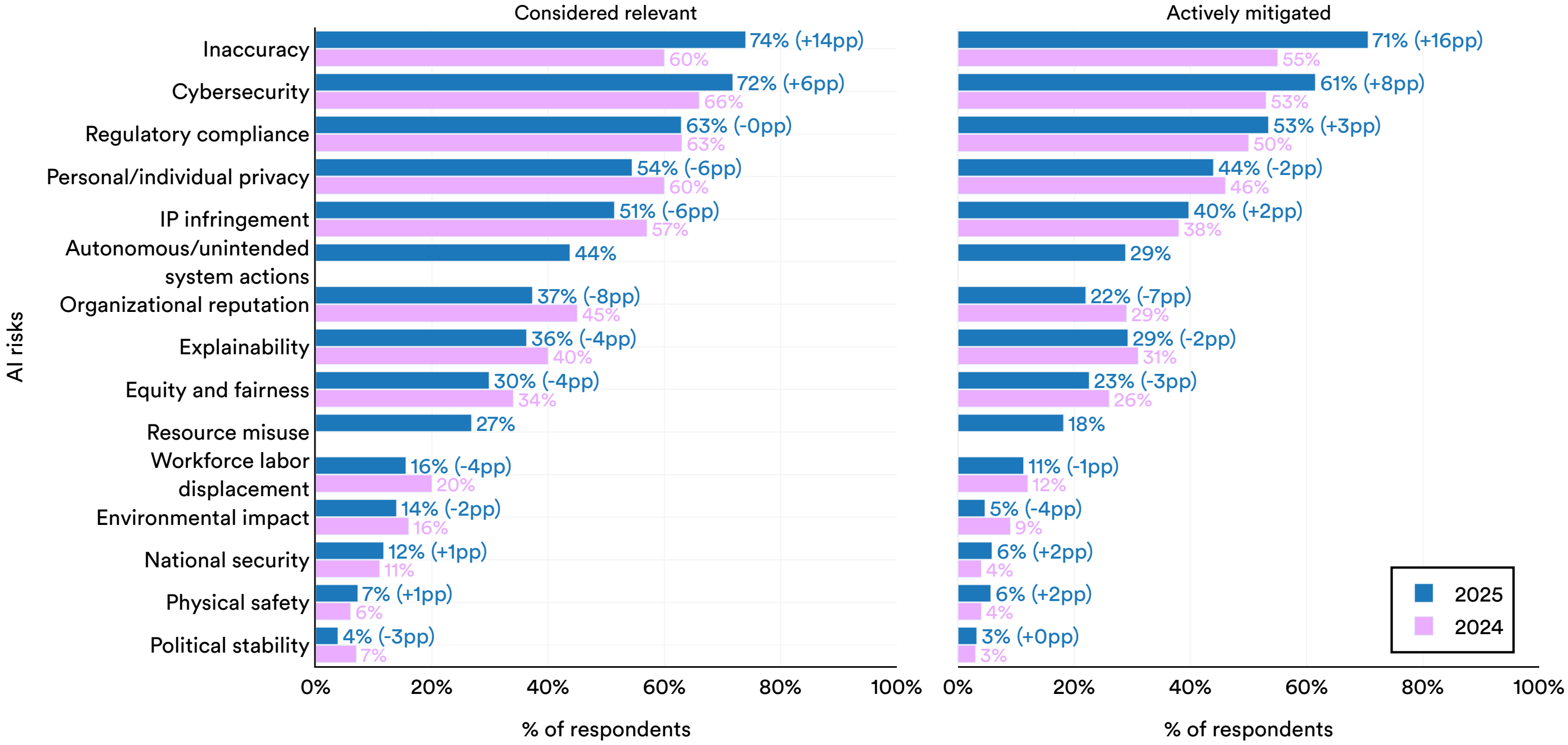
AI risks: considered relevant vs. actively mitigated, 2024 vs. 2025
Source: McKinsey & Company Survey, 2025 | Chart: 2026 AI Index report

Figure 3.3.5[6]

## AI Governance and Investment

Organizations are formalizing who is responsible for AI governance. Between 2024 and 2025, companies shifted AI governance ownership away from data and analytics functions (down from 17% to 13%), toward dedicated AI governance roles (up from 14% to 17%) (Figure 3.3.6). Information security remained the most common primary owner at 21%, and 5% of organizations reported having no designated owner in 2025 compared to 9% in 2024.

Organizations are also backing their governance structures with financial commitments, though investment levels vary by company size (Figure 3.3.7). Most organizations with under $1 billion in revenue reported they expected to invest under $5 million in operationalizing RAI, through initiatives such as hiring specialized professions, building or purchasing technical systems, and engaging legal services. At the largest companies, reported investment numbers were significantly higher. Among organizations with at least $30 billion in revenue, 41% expected to spend $25 million or more and 22% budgeted $50 million or more.

6 "Autonomous/unintended system actions" and "resource misuse" were new additions to the 2025 survey.

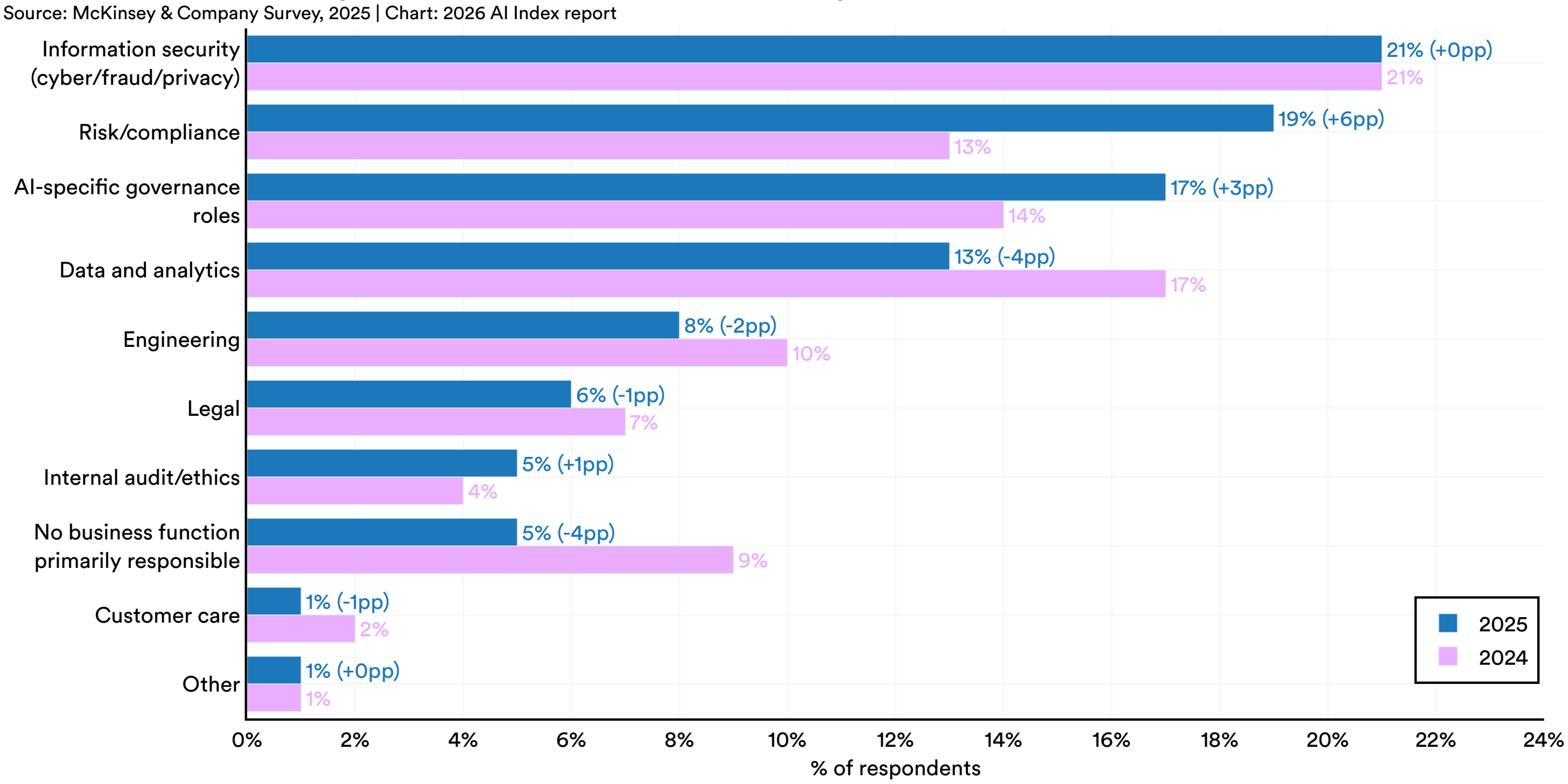
Business functions assigned primary responsibility for AI governance, 2024 vs. 2025
Source: McKinsey & Company Survey, 2025 | Chart: 2026 AI Index report

Figure 3.3.6[7]

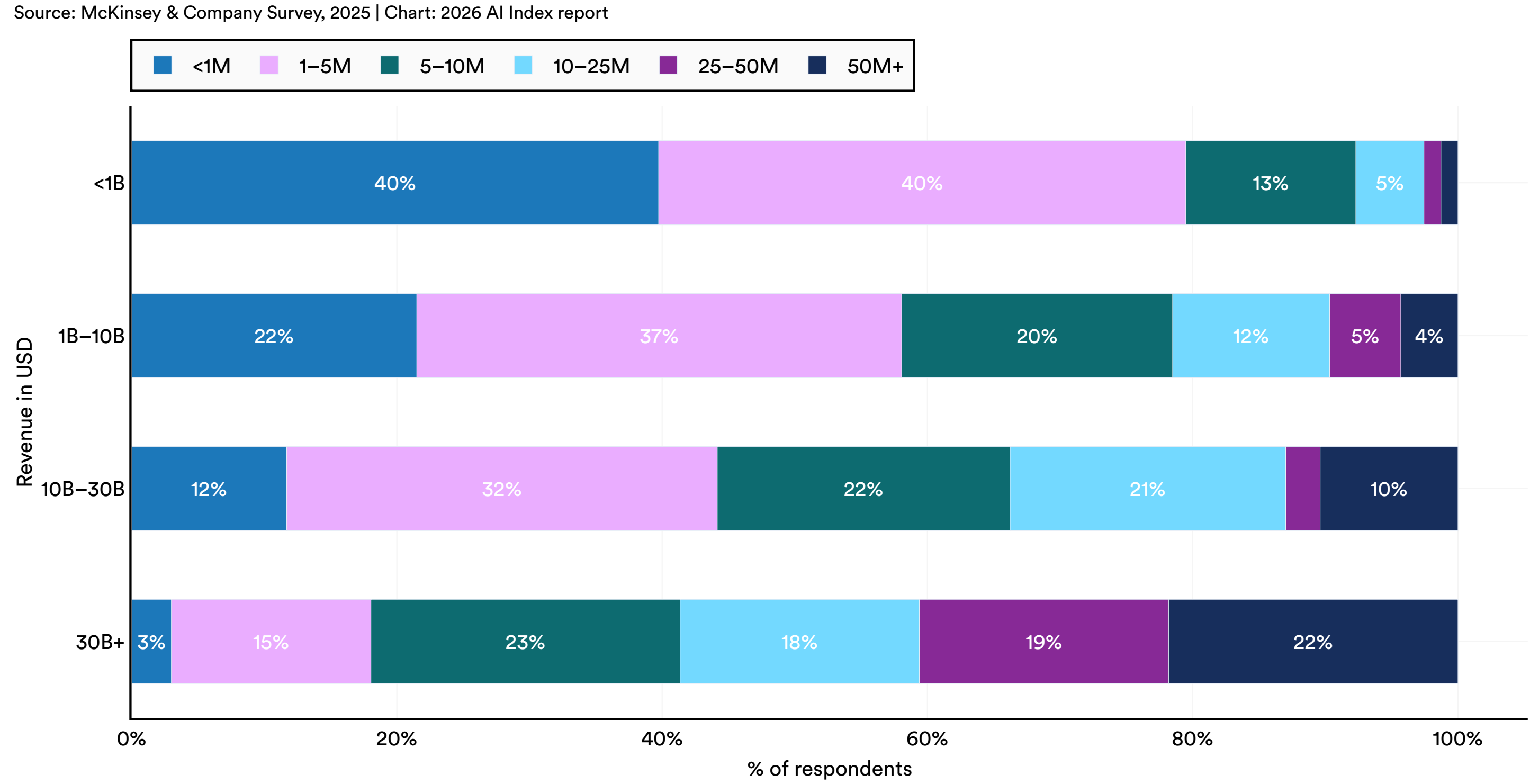
Investment in responsible AI by company revenue, 2025
Source: McKinsey & Company Survey, 2025 | Chart: 2026 AI Index report

Figure 3.3.7

7 The "Unknown" response option was not included in this visualization.

## Implementation, Barriers, and Benefits

Alongside increased accountability structures for responsible AI governance, more organizations have adopted RAI policies. The share that reported not having any policies dropped from 24% in 2024 to 11% in 2025 (Figure 3.3.8). With the uptick in adoption, survey respondents perceived an overall positive impact from RAI policies. Compared to 2024, more organizations reported that RAI policies improved business outcomes (up 7 percentage points), business operations (up 4 percentage points), and customer trust (up 4 percentage points). Furthermore, more organizations reported a drop in the number of AI incidents (plus 8 pp).

Knowledge and training gaps remain the top-cited obstacle to implementing responsible AI, rising from 51% in 2024 to 59% in 2025 (Figure 3.3.9). The second sharpest increase was in technical limitations, with 38% of respondents citing them as a main obstacle, up from 32% in 2024. Resource constraints and regulatory uncertainty continued to rank among the top barriers.

However, the barriers to scaling agentic AI systems followed a different order (Figure 3.3.10). Security and risk concerns far outweighed the others, with 62% of respondents naming these as the primary obstacle, followed by technical limitations (38%) and regulatory uncertainty (38%). Lack of executive support was reported as a greater barrier to implementing RAI policies (14%) than with agentic AI (9%).

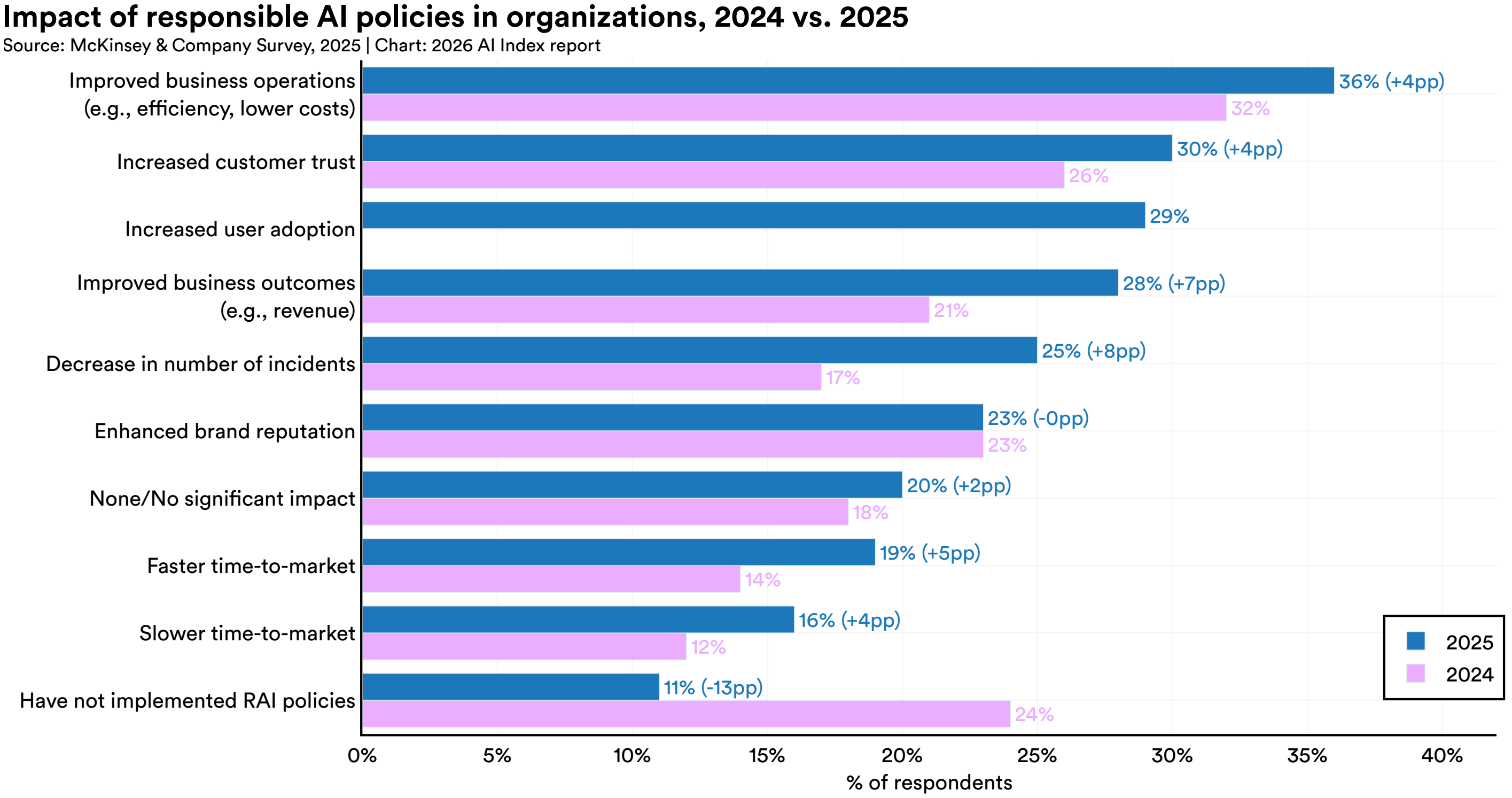
Impact of responsible AI policies in organizations, 2024 vs. 2025
Source: McKinsey & Company Survey, 2025 | Chart: 2026 AI Index report

Figure 3.3.8[8]

8 Percentages are based on respondents who selected at least one answer.

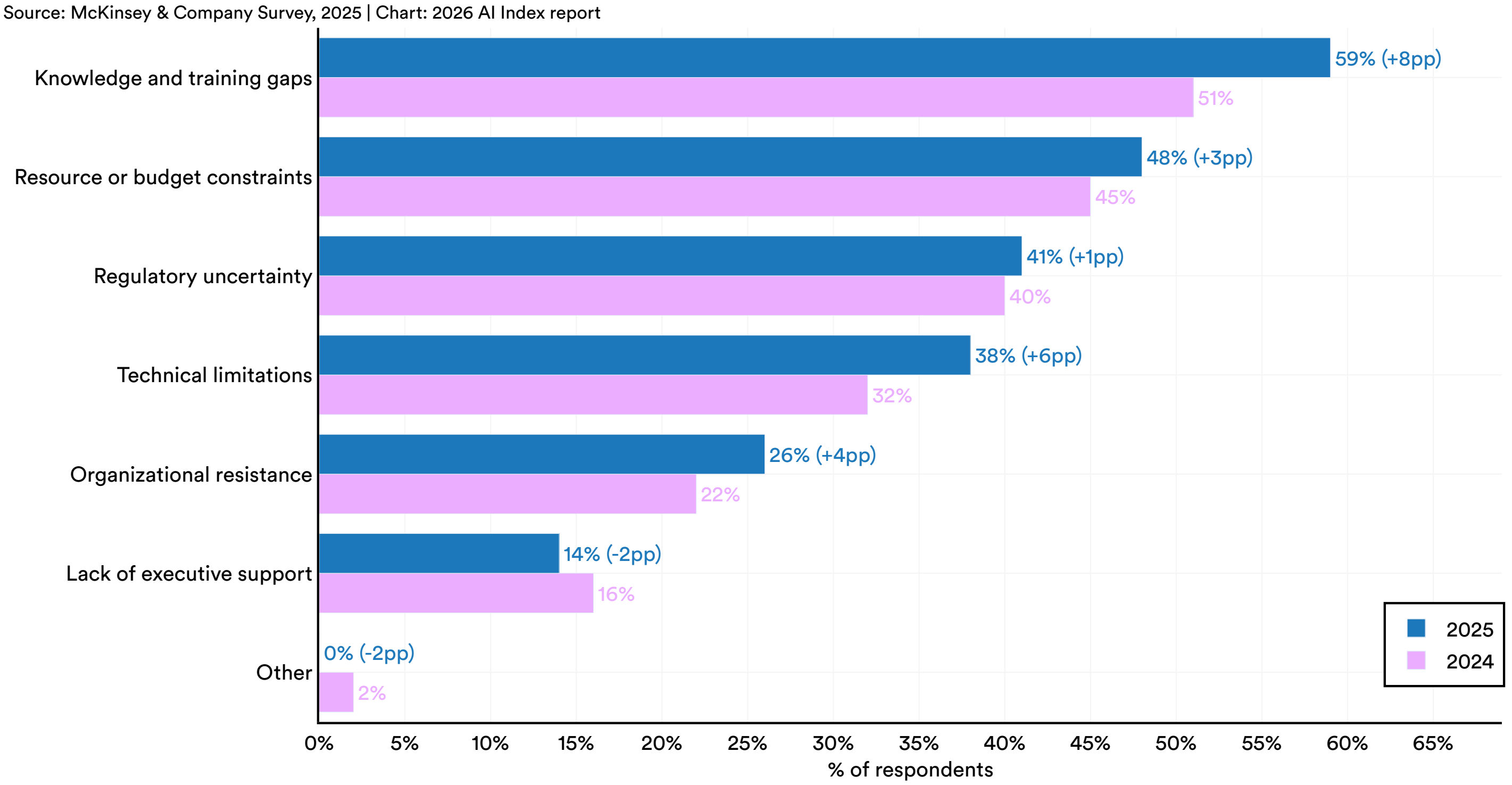
Main obstacles to the implementation of responsible AI measures, 2024 vs. 2025
Source: McKinsey & Company Survey, 2025 | Chart: 2026 AI Index report

Figure 3.3.9[9]

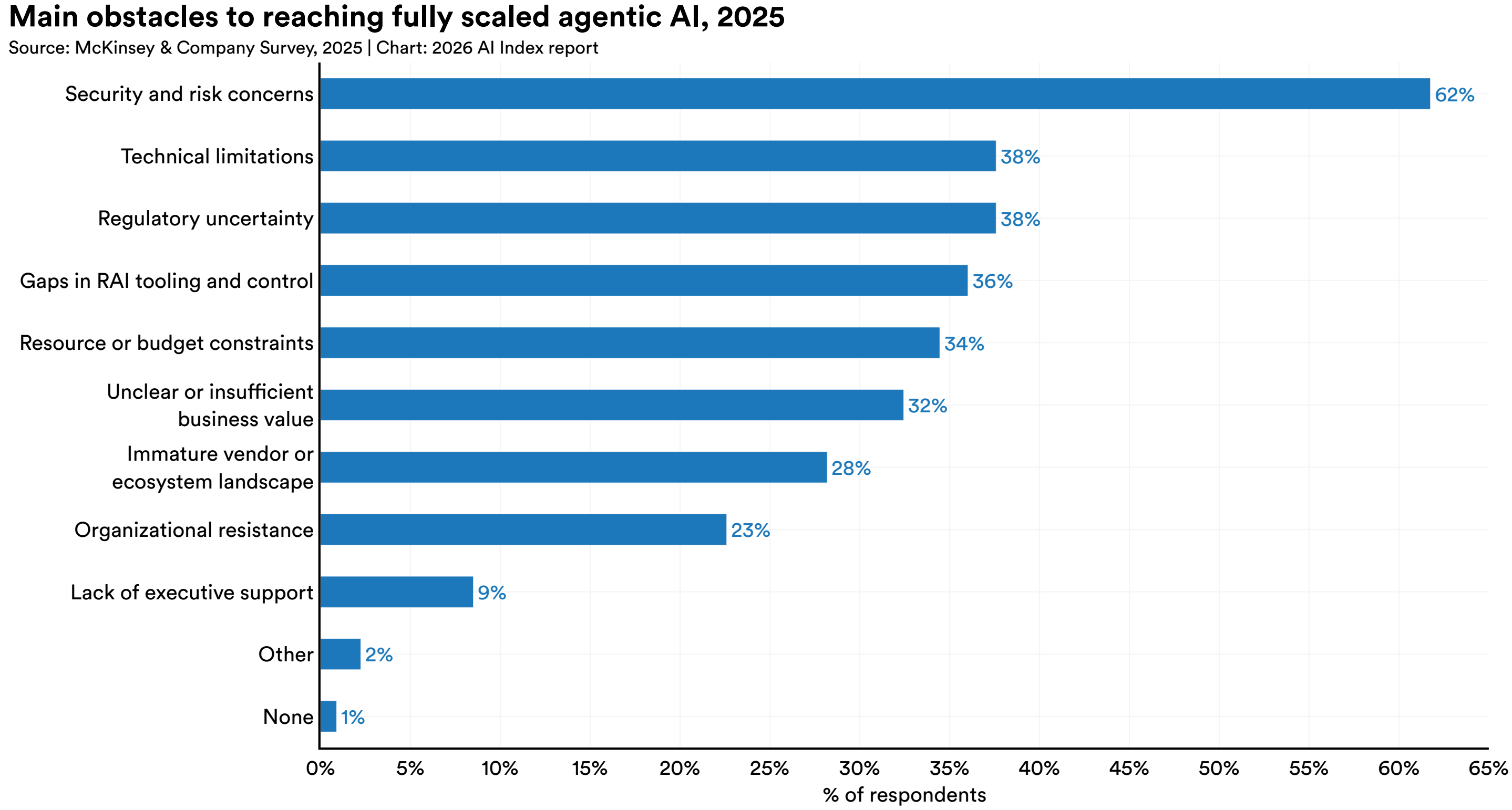
Main obstacles to reaching fully scaled agentic AI, 2025
Source: McKinsey & Company Survey, 2025 | Chart: 2026 AI Index report

Figure 3.3.10

9 Neither the "Unknown" nor the "None" response option is shown in this visualization.

## Regulatory Influence

The General Data Protection Regulation remains the most cited regulatory influence on responsible AI practices, though its influence declined slightly from 65% in 2024 to 60% in 2025 (Figure 3.3.11). AI-specific regulations, such as the EU AI Act and the U.S. AI Executive Order, increased in reported influence by 2 percentage points. Two new entries in the 2025 survey point to growing interest in technical and management standards. ISO/IEC 42001, an AI management system standard, was cited by 36% of respondents, and the NIST AI Risk Management Framework by 33%. The OECD AI Principles fell from 21% to 16%. The share of organizations reporting no regulatory influence on their RAI practices dropped from 17% to 12%.

Chapter 8 tracks these regulatory developments in detail, including the phased implementation of the EU AI Act and the shift in U.S. federal AI policy following the revocation of the Biden-era executive order in early 2025.

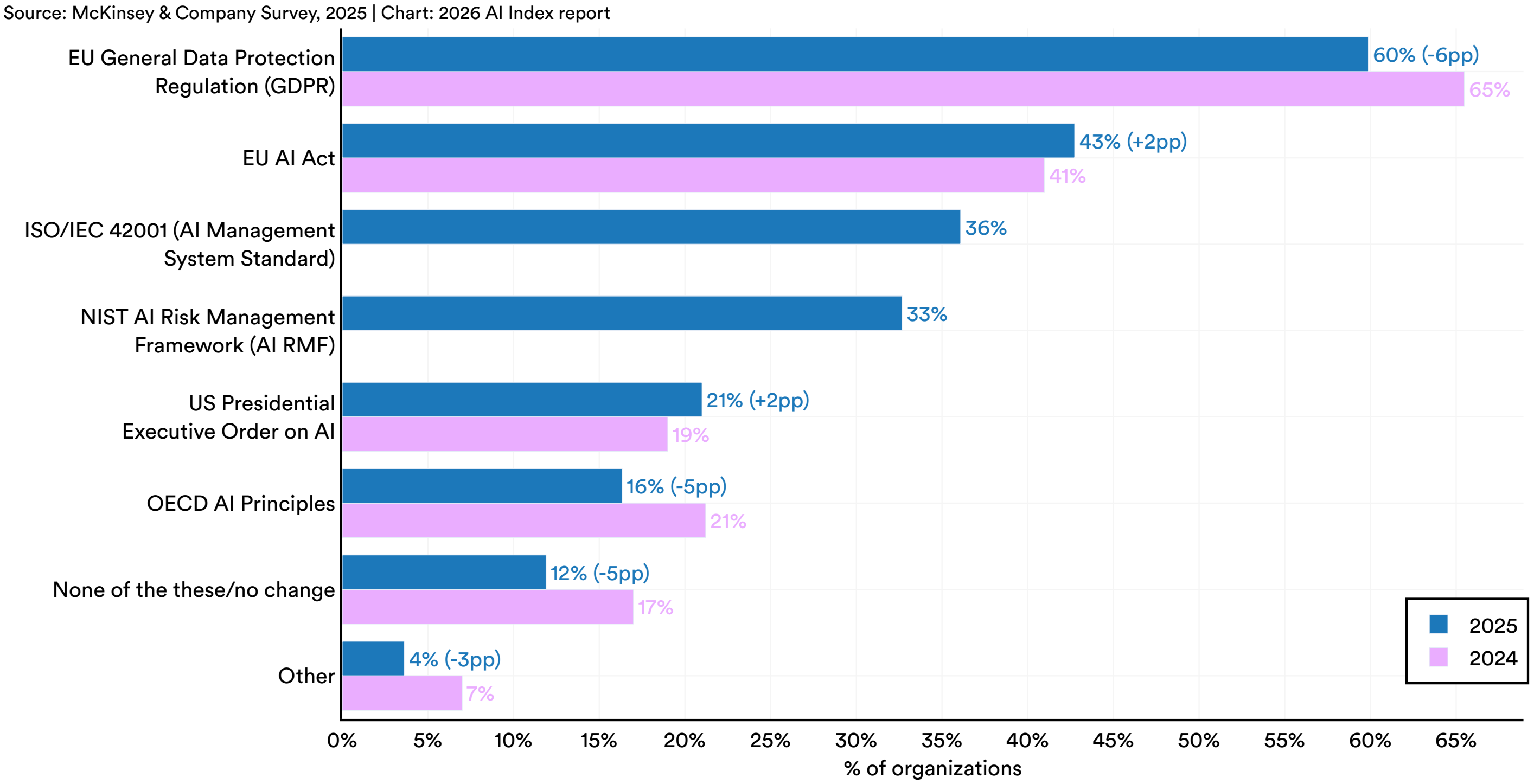
Percentage of organizations influenced by AI regulations in responsible AI decision-making, 2024 vs. 2025
Source: McKinsey & Company Survey, 2025 | Chart: 2026 AI Index report

Figure 3.3.11[10]

10 The ISO/IEC 42001 (AI Management System Standard) and NIST AI Risk Management Framework (AI RMF) AI regulation were added in the 2025 RAI Survey, and not included in 2024 Survey.

# 3.4 RAI in Academia

Another signal of responsible AI's trajectory is the amount of research attention it is getting. This section tracks the number of RAI-related papers accepted at six leading AI conferences: AAAI, AIES, FAccT, ICML, ICLR, and NeurIPS. These conferences do not represent all responsible AI research, but they provide a consistent basis for tracking publication trends over time. Papers were identified using RAI-related keywords, with full methodology described in the Appendix.

## Publication Volume

The number of responsible AI papers accepted at these conferences has been growing consistently, and increased by 19%, from a count of 1,278 to 1,521, between 2024 and 2025 (Figure 3.4.1). The four subtopics tracked here, privacy and data governance, fairness and bias, transparency and explainability, and security and safety, are not exhaustive but map directly to the RAI frameworks introduced in Section 3.1. Security and safety has become the largest and fastest growing area of RAI research, with 641 accepted papers, a 23% increase from 2024 (Figure 3.4.2). Fairness and bias accounted for 462 (+13%), transparency and explainability for 405 (+14%), and privacy and data governance for 248 (+33%). All four subtopics have grown since 2019, but security and safety has grown the most in absolute terms.

At the general purpose conferences, responsible AI papers still make up a small share of total accepted work (Figure 3.4.3). AAAI (8%), NeurIPS (8%), ICML (7.7%), and ICLR (7.6%) all cluster around 8%, a proportion that has remained flat since 2019, though AAAI did fall from around 13% in 2024 to 8% in 2025.

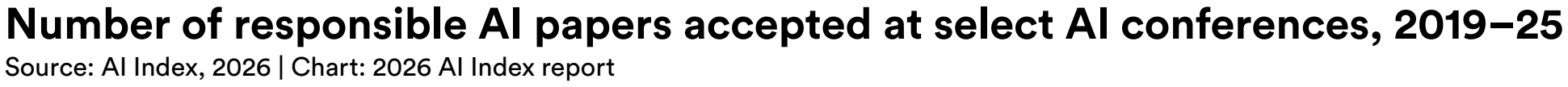

Source: AI Index, 2026 | Chart: 2026 AI Index report

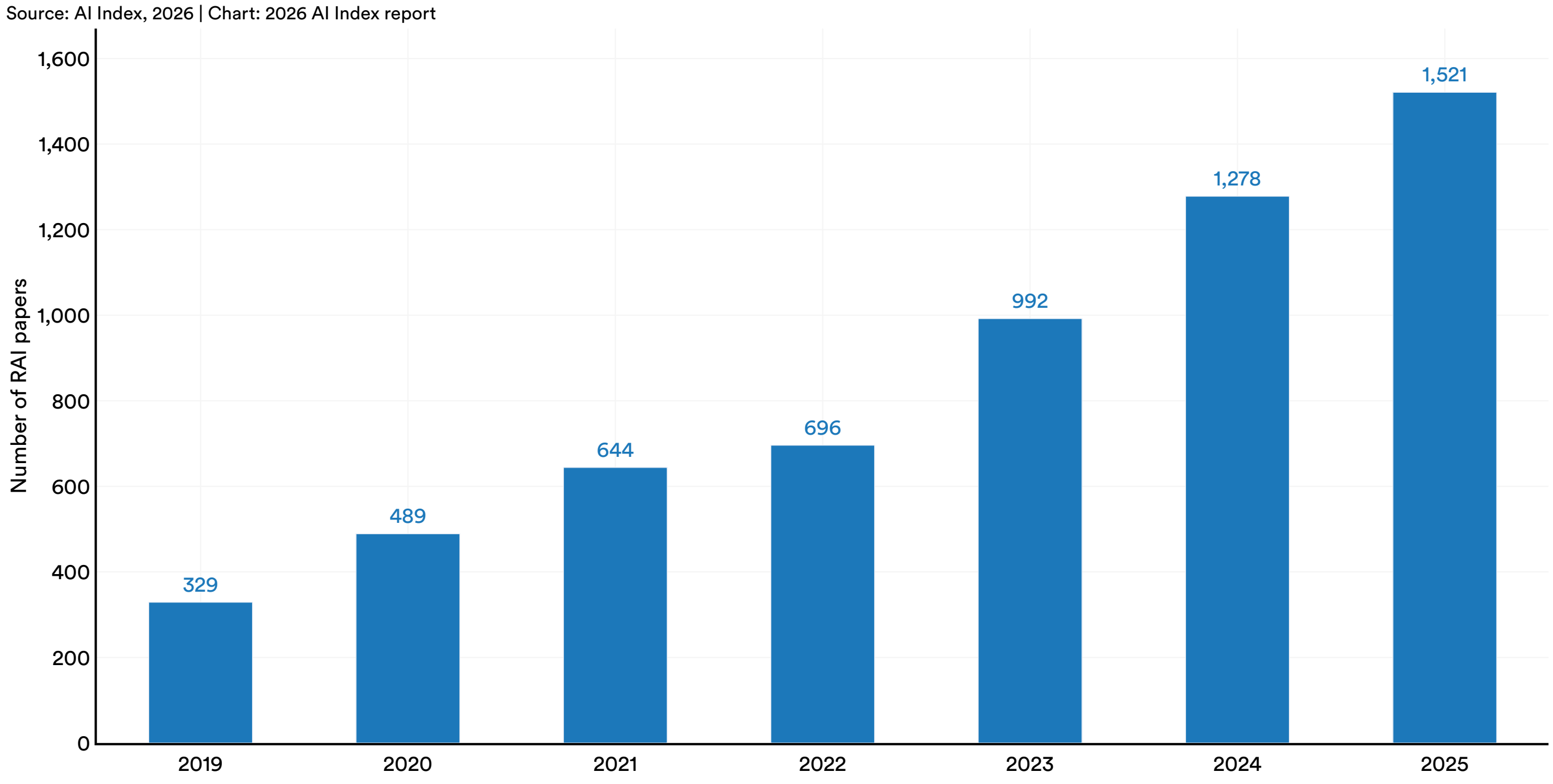

Figure 3.4.1

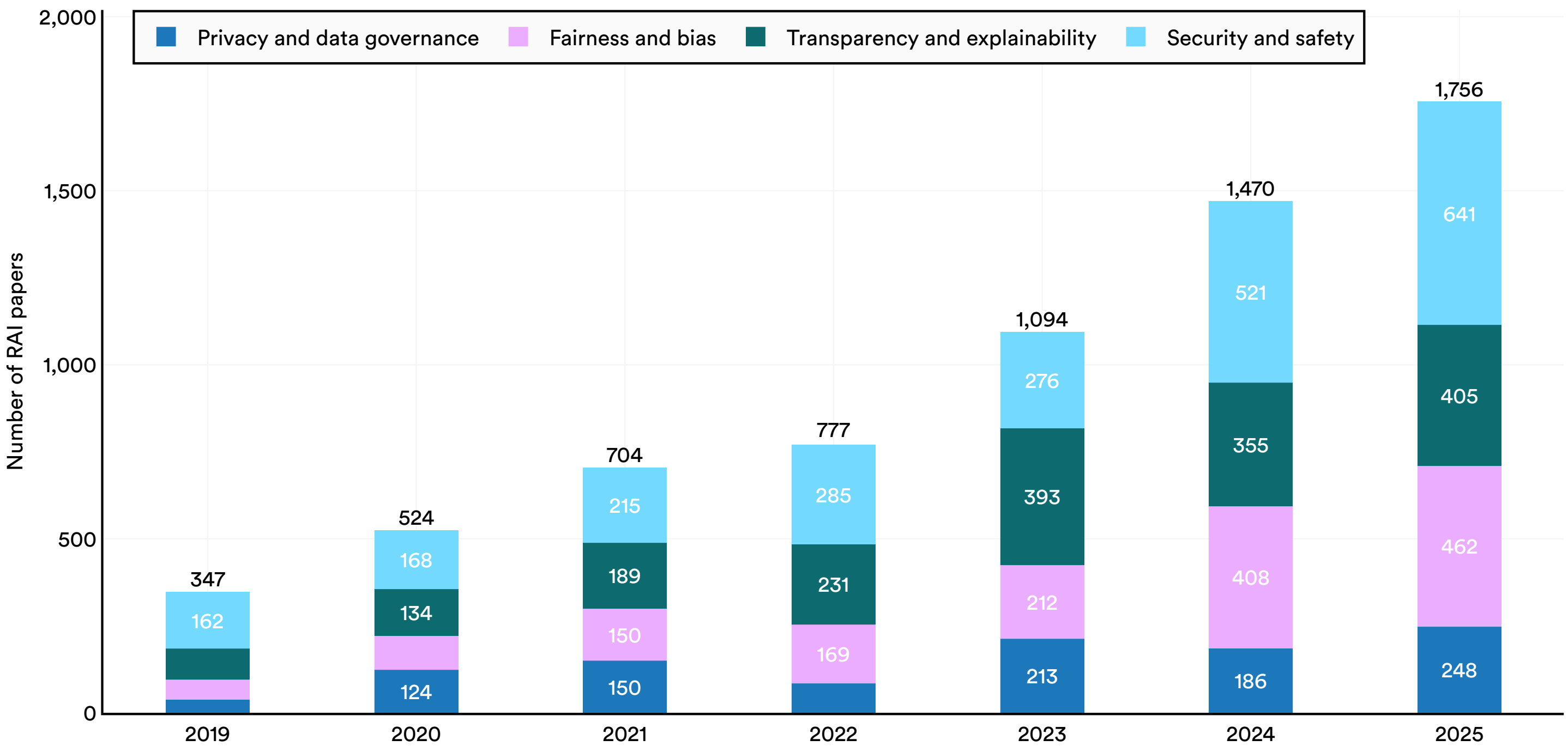
Number of responsible AI papers accepted at select AI conferences by subtopic, 2019–25
Source: AI Index, 2026 | Chart: 2026 AI Index report

Figure 3.4.2 [11]

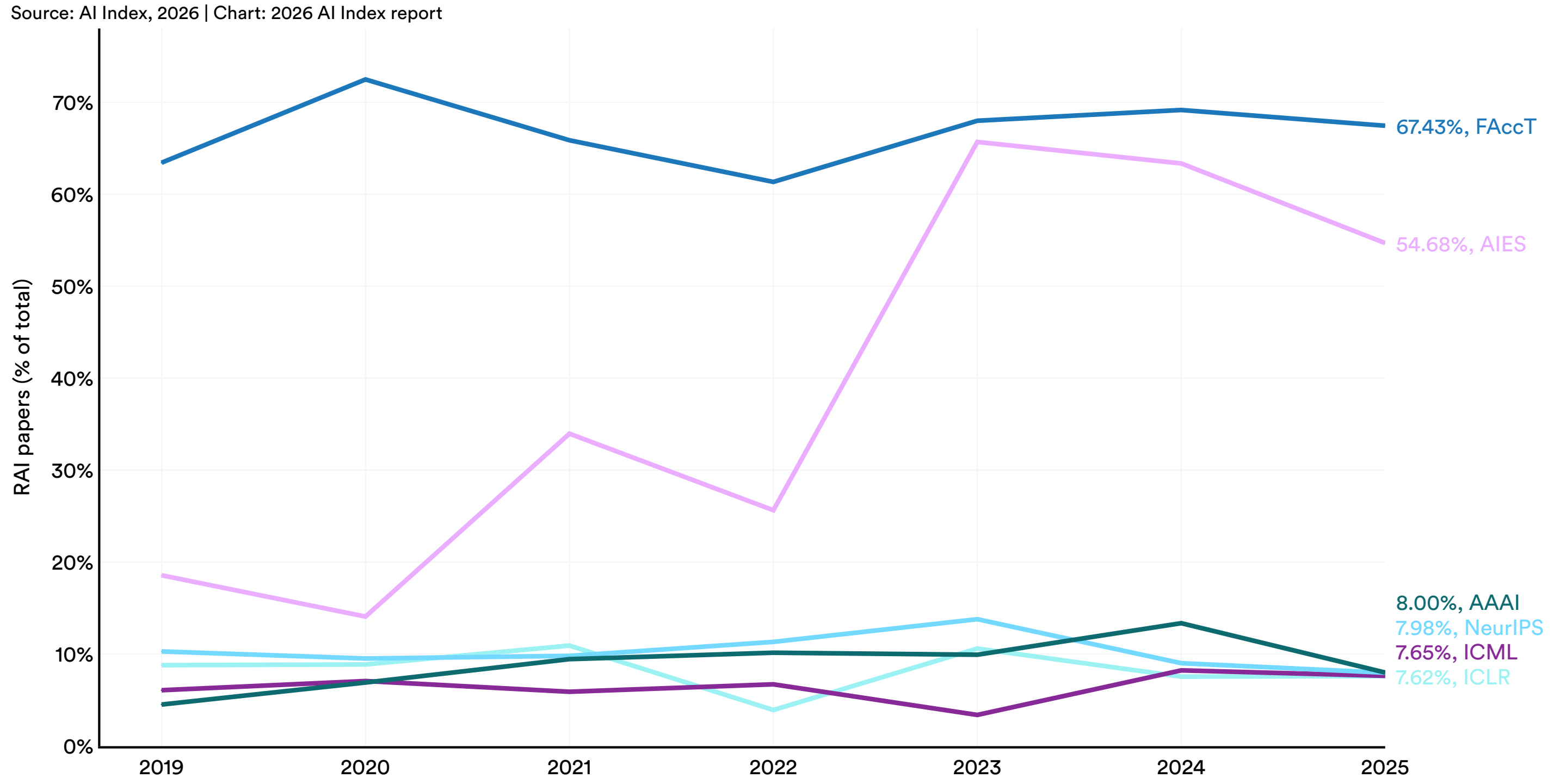
Responsible AI papers accepted (% of total) at select AI conferences by conference, 2019–25
Source: AI Index, 2026 | Chart: 2026 AI Index report

Figure 3.4.3

11 A single publication may be related to more than one topic and may therefore be counted or shown in multiple categories.

## Geographic Distribution

The number of countries contributing to responsible AI research in those select conferences has grown, but the balance among the top contributors has changed. In 2025, China led with 812 accepted RAI papers, more than double the 394 from the United States (Figure 3.4.4). Singapore (112), the United Kingdom (103), and Hong Kong (98) were also among the top five contributors. In 2024, the United States led with 788 papers to China's 322 (Figure 3.4.5). The reversal is sharp, but consistent with China's lead in overall AI publication volume and citation share, as discussed in Chapter 1. Europe, which had been growing through 2023, saw its RAI output fall in 2024 and 2025. Over the full 2019 to 2025 period, the United States still holds the largest cumulative total of accepted RAI papers.

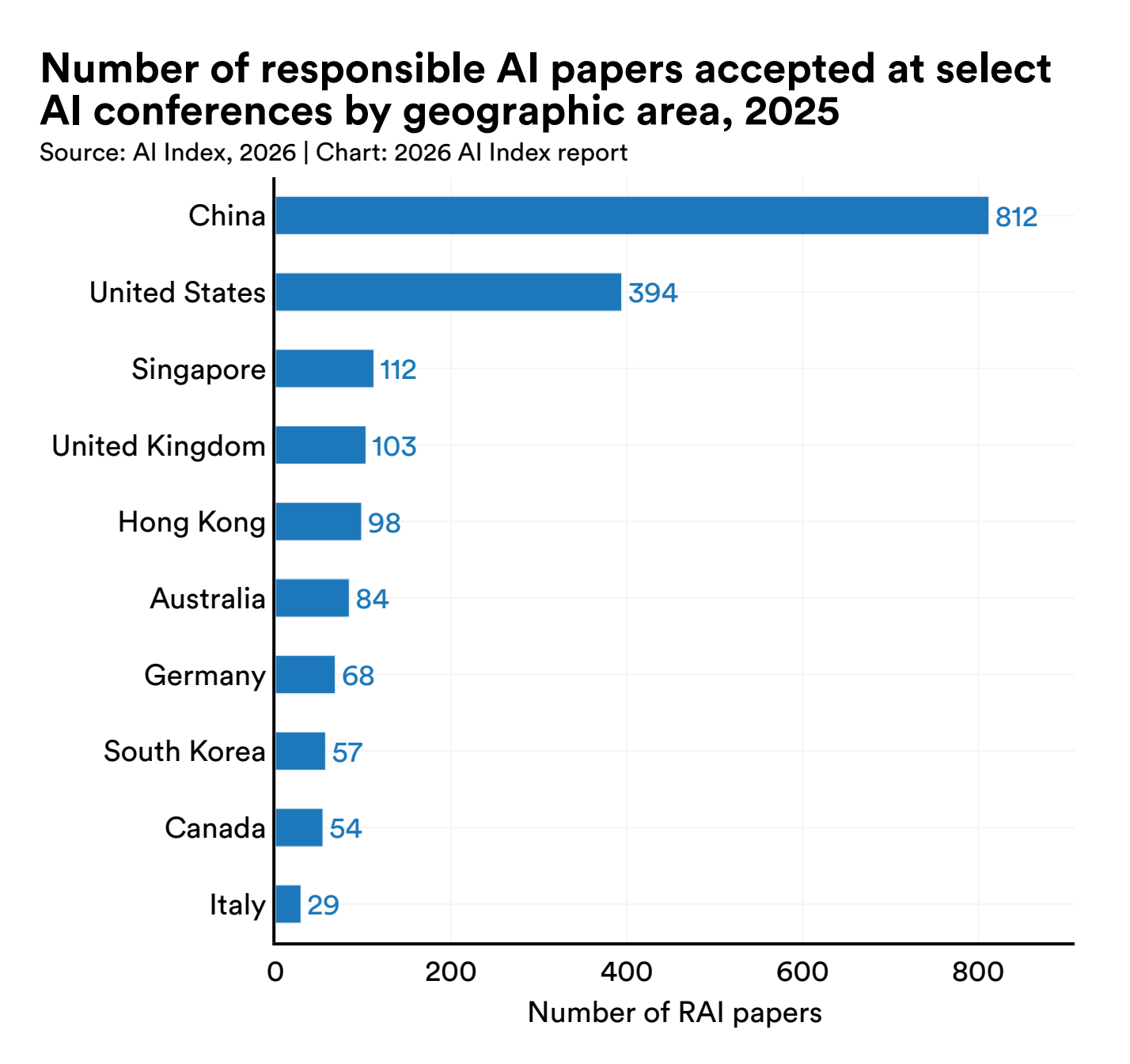

Figure 3.4.4

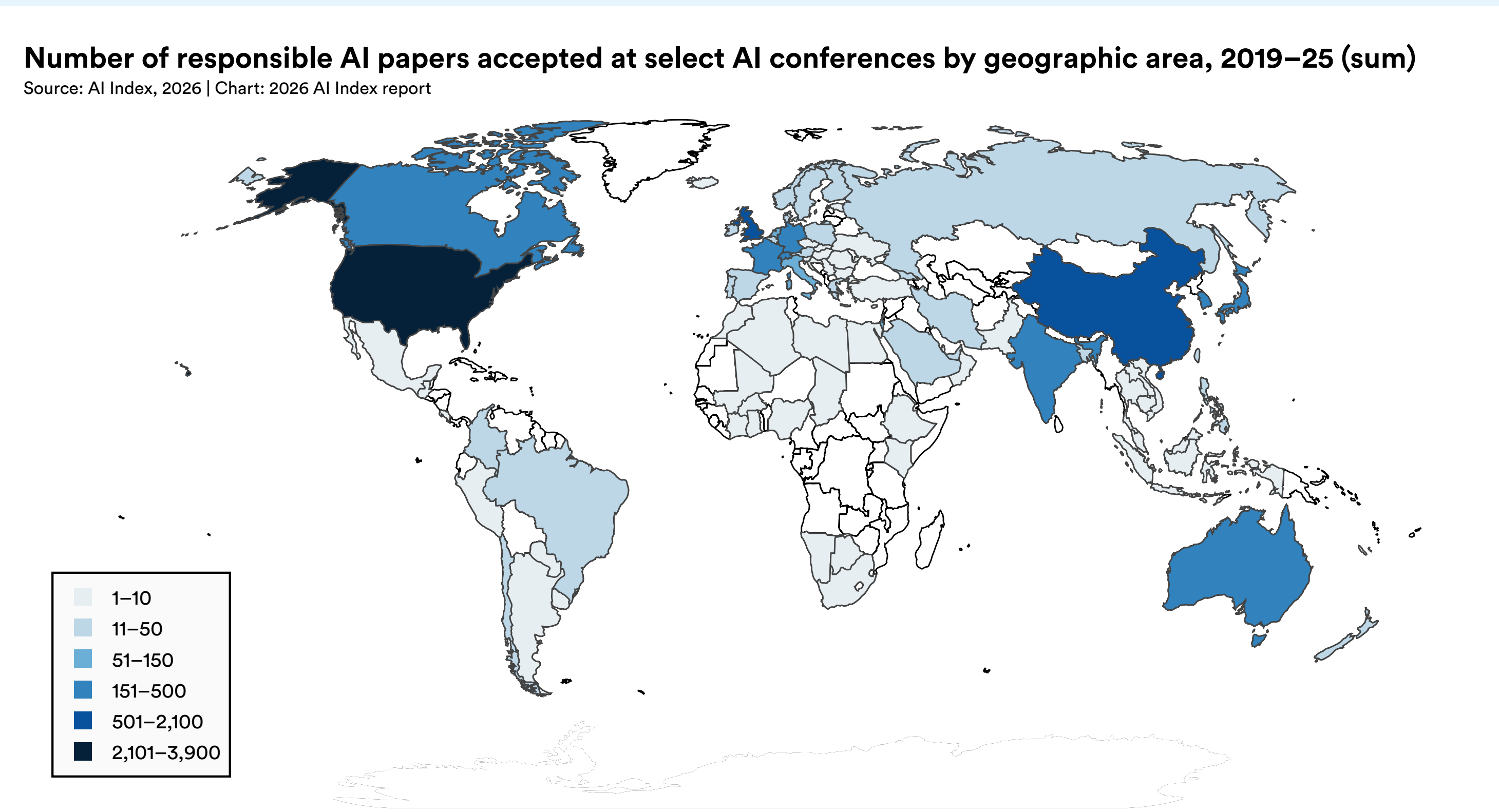

Figure 3.4.5

# 3.5 RAI Policymaking

Responsible AI governance depends on countries both adopting ethical principles and having the institutions and regulations to enforce them. UNESCO's Readiness Assessment Methodology (RAM) is the most comprehensive international effort to measure that preparedness at the country level. Launched in December 2022, the RAM evaluates national readiness across dimensions such as legal frameworks, technical infrastructure and education, and produces a country report to assess where the gaps are.

Most major AI-producing countries, including the United States, China, and much of Western Europe, have not participated in the assessment (Figure 3.5.1). Countries that have completely or begun the assessment are concentrated in Latin America, Sub-Saharan Africa, and parts of South and Southeast Asia. The RAM effort was designed as a capacity-building tool for countries earlier in the governance trajectory, which may explain the participation pattern.

AI legislation and national strategies often include responsible AI provisions, and Chapter 8 examines those in more detail.

**Readiness Assessment Methodology (RAM) implementation across member countries**

Source: UNESCO, 2025 | Chart: 2026 AI Index report

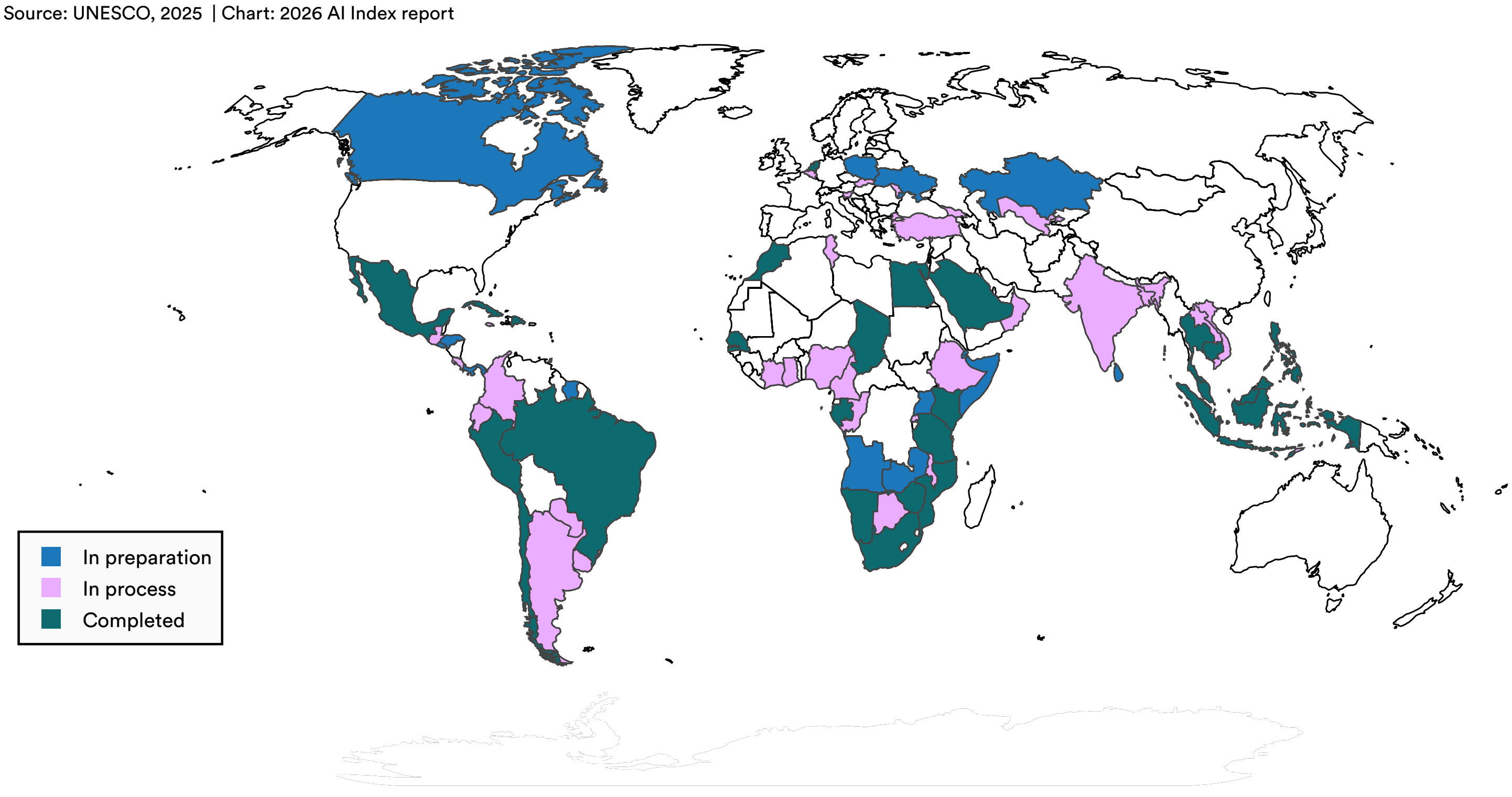

Figure 3.5.1

**HIGHLIGHT:**

## Global AI Governance Participation

Since 2019, international cooperation on AI governance has become more widespread, but the depth of engagement varies significantly across borders (Figure 3.5.2). Only five countries, Canada, France, Germany, Italy, and Japan, have consistently endorsed every major global AI governance initiative recorded between 2019 and 2025. Other countries moved in and out of these summits depending on the forum, focus, and timeline but more importantly, not all the countries were able to participate in these global AI governance initiatives. The first intergovernmental standard on AI, the 2019 OECD AI Principles, was restricted to member nations (mainly high-income) and a few partner nations. Likewise, the G7 and G20 discussions remained centered on the world's largest economies. The 2023 Bletchley and 2024 Seoul Summits, however, began to diversify the composition of participants by inviting a broader range of nations, notably including China. The 2025 AI Action Summit in France marked a further turning point, convening over 100 countries alongside civil society organizations and NGOs, with an agenda to prioritize the needs of the Global South and environmental sustainability. Sixty-four participants signed the resulting Statement on Inclusive and Sustainable AI, including the African Union Commission and the European Union. In a notable shift, both the United States and the United Kingdom declined to sign the final declaration. The UK cited a lack of emphasis on national security, while the U.S. decision reflected a pivot toward a more deregulatory, "innovation-first" approach. As engagement at these governance forums becomes more inclusive and substantive, consensus on the terms of cooperation becomes harder to secure.

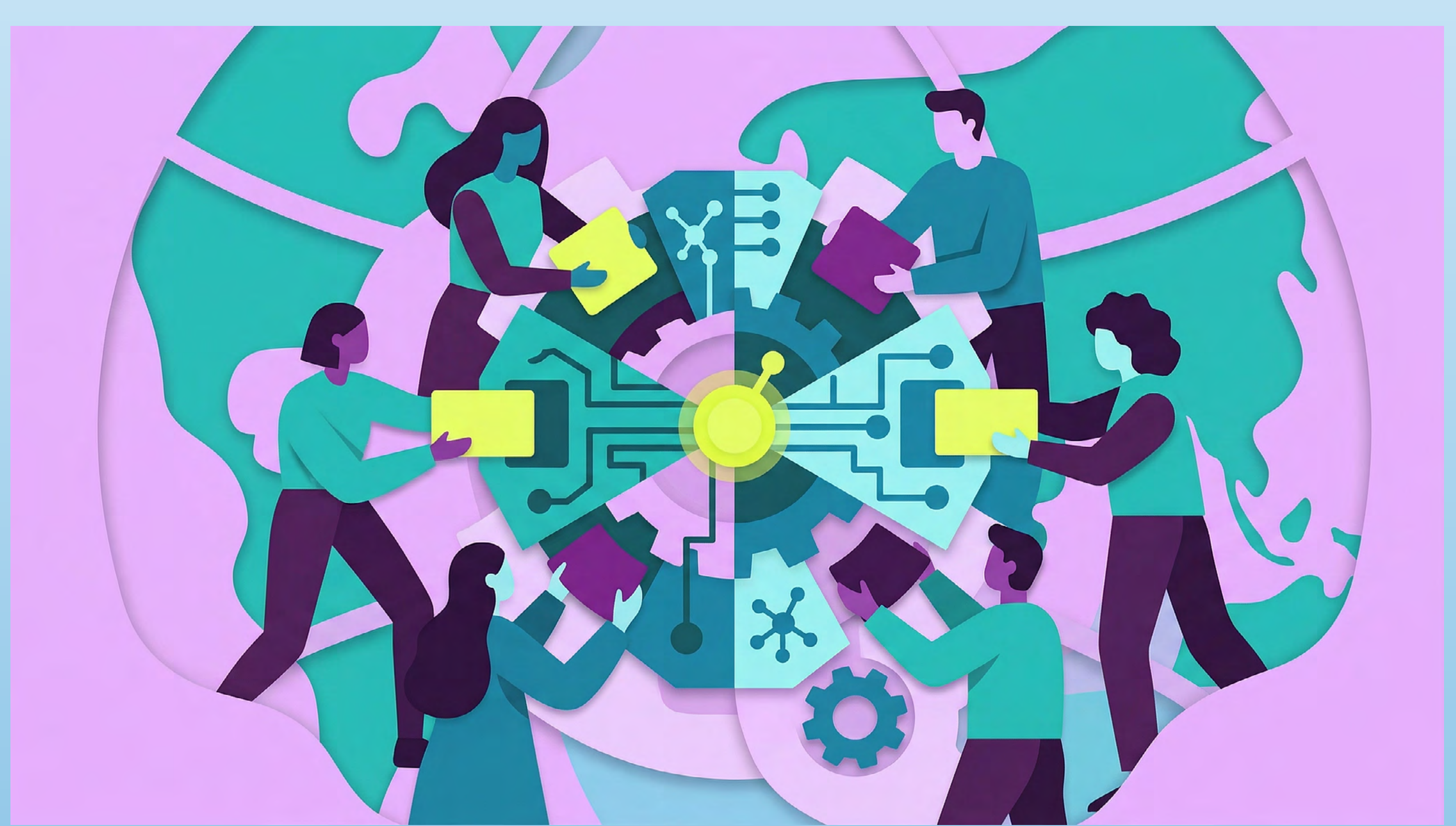

HIGHLIGHT:

**Participation in AI governance mechanisms**

Source: Governing AI for Humanity, United Nations, 2025; AI Action Summit, 2025 | Chart: 2026 AI Index report

OECD AI Principles (2019)
G20 AI Principles (2019)
CoE drafters (2022)
GPAI Ministerial Declaration (2022)
Bletchley Declaration (2023)
G7 Hiroshima AI Process (2023)
Seoul Ministerial Statement (2024)
AI Action Summit Statement (2025)

Argentina
Armenia
Australia
Austria
Bahrain
Belgium
Brazil
Canada
Chile
China
Colombia
Denmark
Egypt
Estonia
European Union
Finland
France
Germany
Greece
Hungary
Iceland
India
Indonesia
Ireland
Israel
Italy
Japan
Kenya
Lithuania
Luxembourg
Malaysia
Malta
Mexico
Morocco
Netherlands
New Zealand
Nigeria
Norway
Pakistan
Portugal
Qatar
Republic of Korea
Russian Federation
Saudi Arabia
Singapore
Slovakia
Slovenia
South Africa
Spain
Sri Lanka
Sweden
Switzerland
Tunisia
Turkey
United Arab Emirates
United Kindom
United States
Uruguay
Vietnam

Figure 3.5.2

# 3.6 Data Governance for Privacy

Responsible AI practices do not develop evenly across countries. This section assesses that variation for privacy and data governance, drawing on the Global Index on Responsible AI (GIRAI). GIRAI is a benchmark dataset covering 138 countries, built from a quality-reviewed expert survey of 1,862 questions completed by 138 in-country researchers between November 2023 and February 2024. It scores countries on a 0 to 100 scale across thematic areas, covering government frameworks, government actions, and the role of civil society and advocacy organizations. However, it is important to note that low scores do not necessarily indicate that a country is disregarding a certain dimension. In many cases, they reflect earlier stages of AI deployment and diffusion or limited institutional capacity to formalize AI-specific frameworks.

## Data Protection and Privacy

The privacy and data protection dimension of the GIRAI score[12] examines whether countries have laws that govern how personal data is collected, used, and shared in AI systems, and whether those laws are backed by regulators with the power to enforce them.

Countries fall across a wide spectrum, with GIRAI scores ranging from near zero to above 80 across the countries surveyed (Figure 3.6.1). Australia and parts of Europe score the highest, while parts of Africa and the Middle East show an absence of dedicated data protection legislation. A complementary map from UNCTAD confirms that most countries now have some form of data protection legislation in place, though a few, mostly concentrated in Africa and parts of Asia, are still in draft stages or have no legislation at all (Figure 3.6.2).

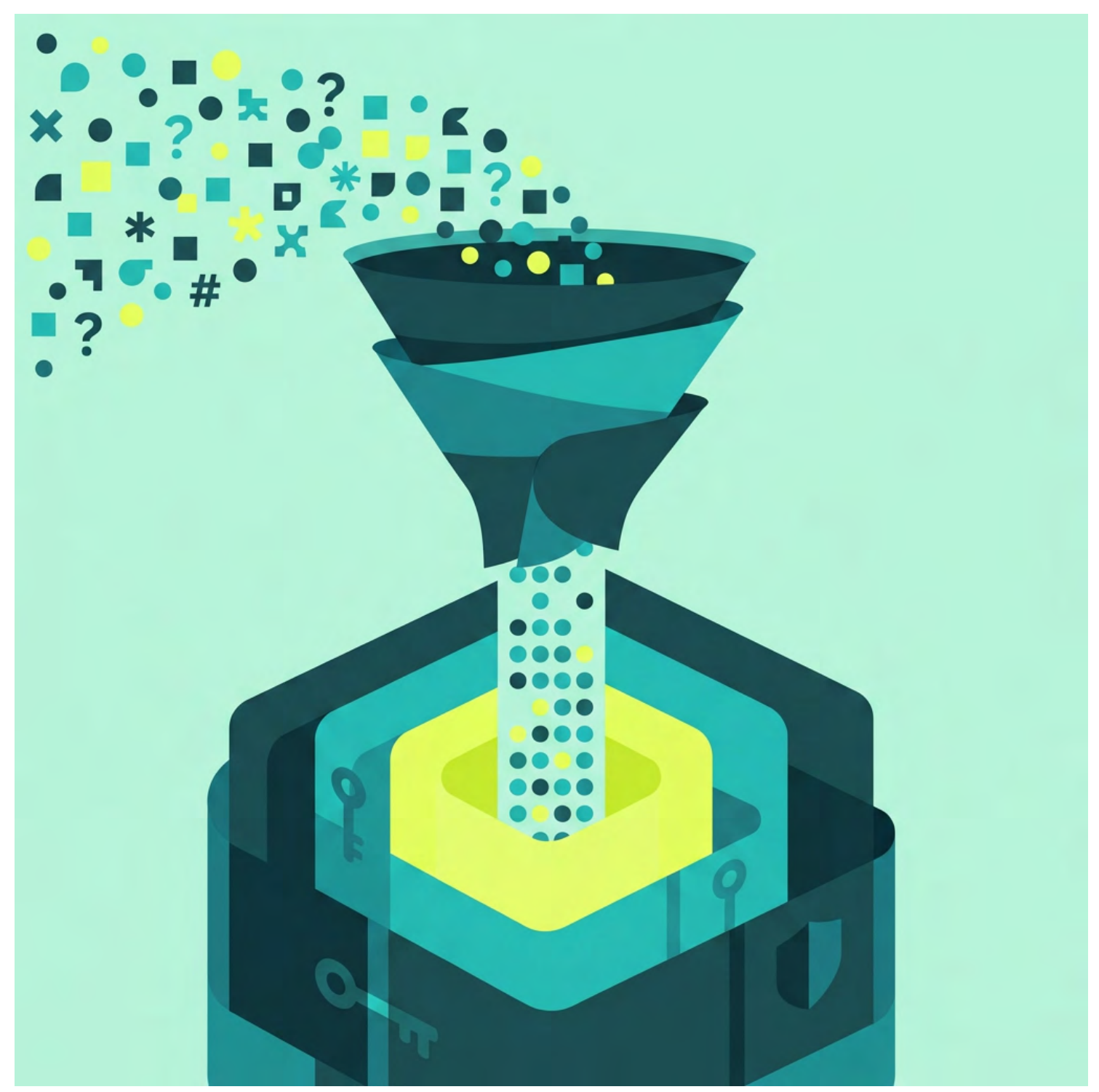

12 Grounded in UDHR Article 12, ICCPR Article 17, the OECD AI Principles, UNESCO's Ethics of AI Recommendation, and UNESCO Principles on Personal Data Protection and Privacy, GIRAI examines explicit laws, oversight, and practice, and assesses frameworks and actions that ensure processing is lawful, fair, purpose-limited, and proportionate. It also evaluates transparency, user information rights, retention limits, accuracy, confidentiality, security, accountability, and rules for data transfers. The index considers national measures—data-protection statutes, automated-decision directives, regulators with enforcement powers, audits, security controls, and initiatives like regulatory sandboxes. It also accounts for nonstate efforts by privacy and digital-rights groups that strengthen protocols and build capacity to mitigate AI-related privacy risks, such as large-scale tracking, profiling, and sensitive-data misuse.

### Global AI data protection and privacy assessment

Source: The Global Index on Responsible AI, 2024 | Chart: 2026 AI Index report

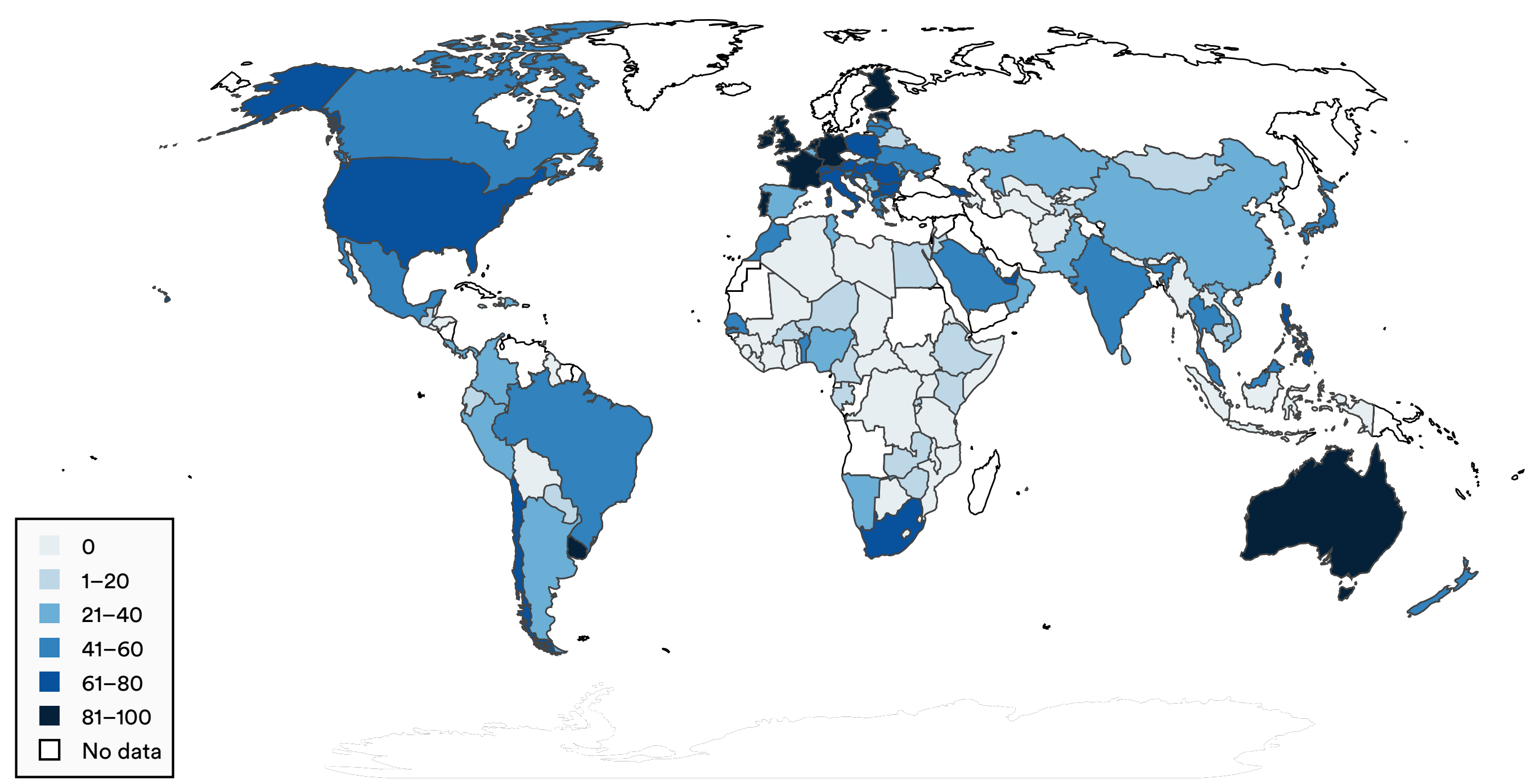

Figure 3.6.1

### Global data protection and privacy legislation

Source: UNCTAD, 2025 | Chart: 2026 AI Index report

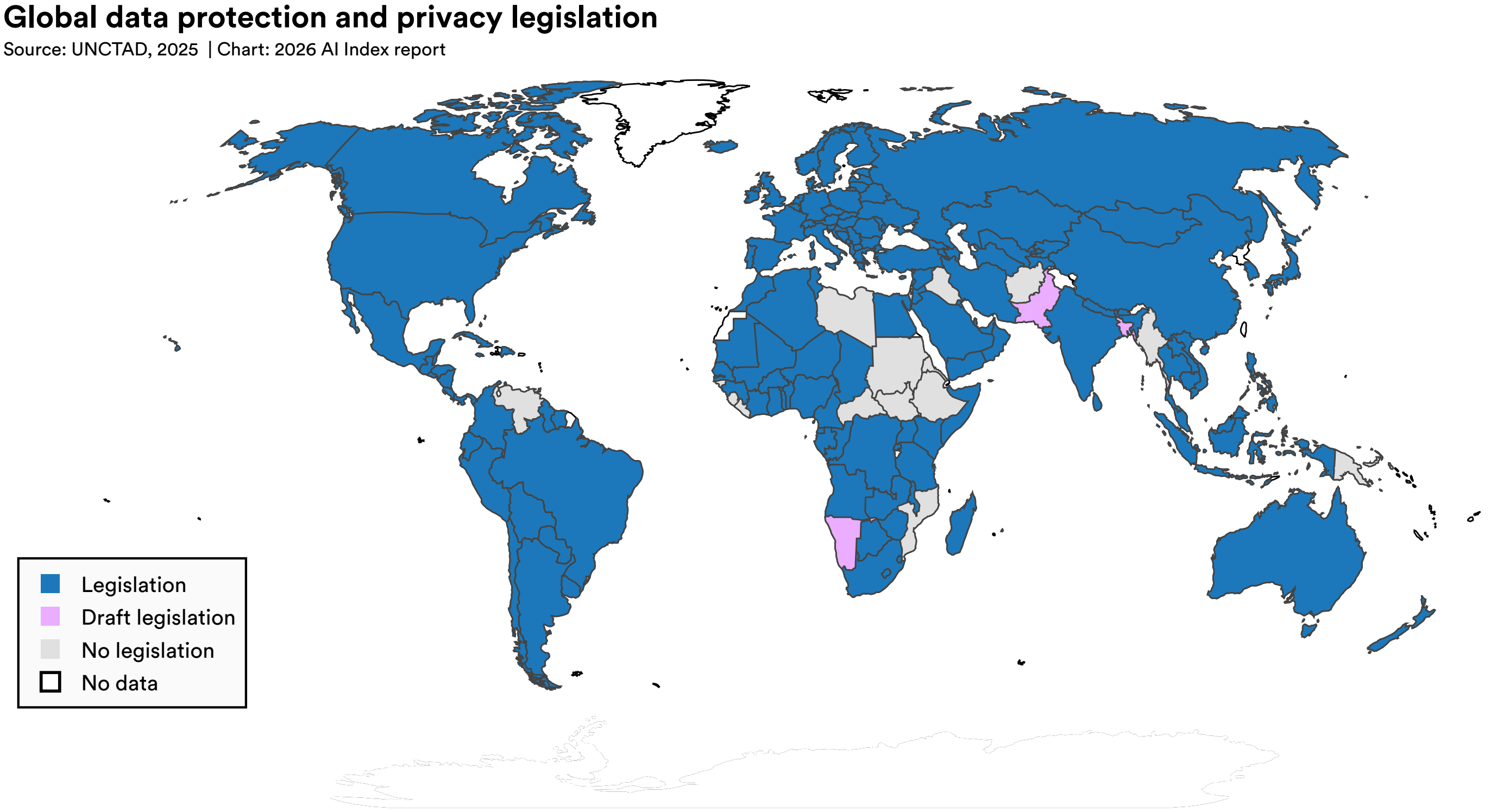

Figure 3.6.2

# 3.7 Fairness and Bias

Fairness and bias are among the hardest-to-measure dimensions of responsible AI, in part because what counts as fair depends heavily on context. GIRAI scores countries separately on bias and unfair discrimination, gender equality, and cultural and linguistic diversity.

## Bias and Unfair Discrimination

The bias and unfair discrimination[13] dimension of the GIRAI score assesses whether countries have explicit measures to prevent and mitigate discriminatory outcomes from AI in its design, development, and deployment. It is meant to address algorithmic bias arising from unrepresentative data, flawed design, or entrenched social inequalities that can harm marginalized groups regardless of intent. It considers whether governments have put laws, oversight bodies, and enforcement mechanisms in place and whether civil society organizations are independently working to monitor and address bias.

GIRAI scores on this dimension are fairly low across the board (Figure 3.7.1). The United States and Canada score highest, with Australia, parts of Europe, and Brazil falling in the middle range. Much of Africa, the Middle East, and Central Asia score below 20.

**Global AI bias and unfair discrimination assessment**

Source: The Global Index on Responsible AI, 2024 | Chart: 2026 AI Index report

0
1–20
21–40
41–60
61–80
81–100
No data

**Figure 3.7.1**

13 The bias and unfair discrimination dimension of the GIRAI score is grounded in international human rights frameworks (UDHR, ICERD, ICCPR, ICESCR).

## Gender Equality

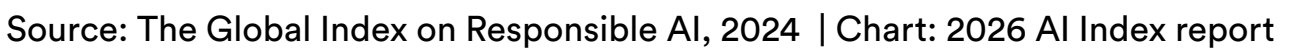

GIRAI's gender equality dimension considers whether countries have state and nonstate initiatives that prevent gender bias and protect equal rights for all gender identities in AI design, development, and use. Canada and The Netherlands score the highest on this measure (Figure 3.7.2). Parts of Europe and Japan fall in the 61–80 range, followed by countries like the United States and Brazil, which score from 41–60.

**Global AI gender equality assessment**

Source: The Global Index on Responsible AI, 2024 | Chart: 2026 AI Index report

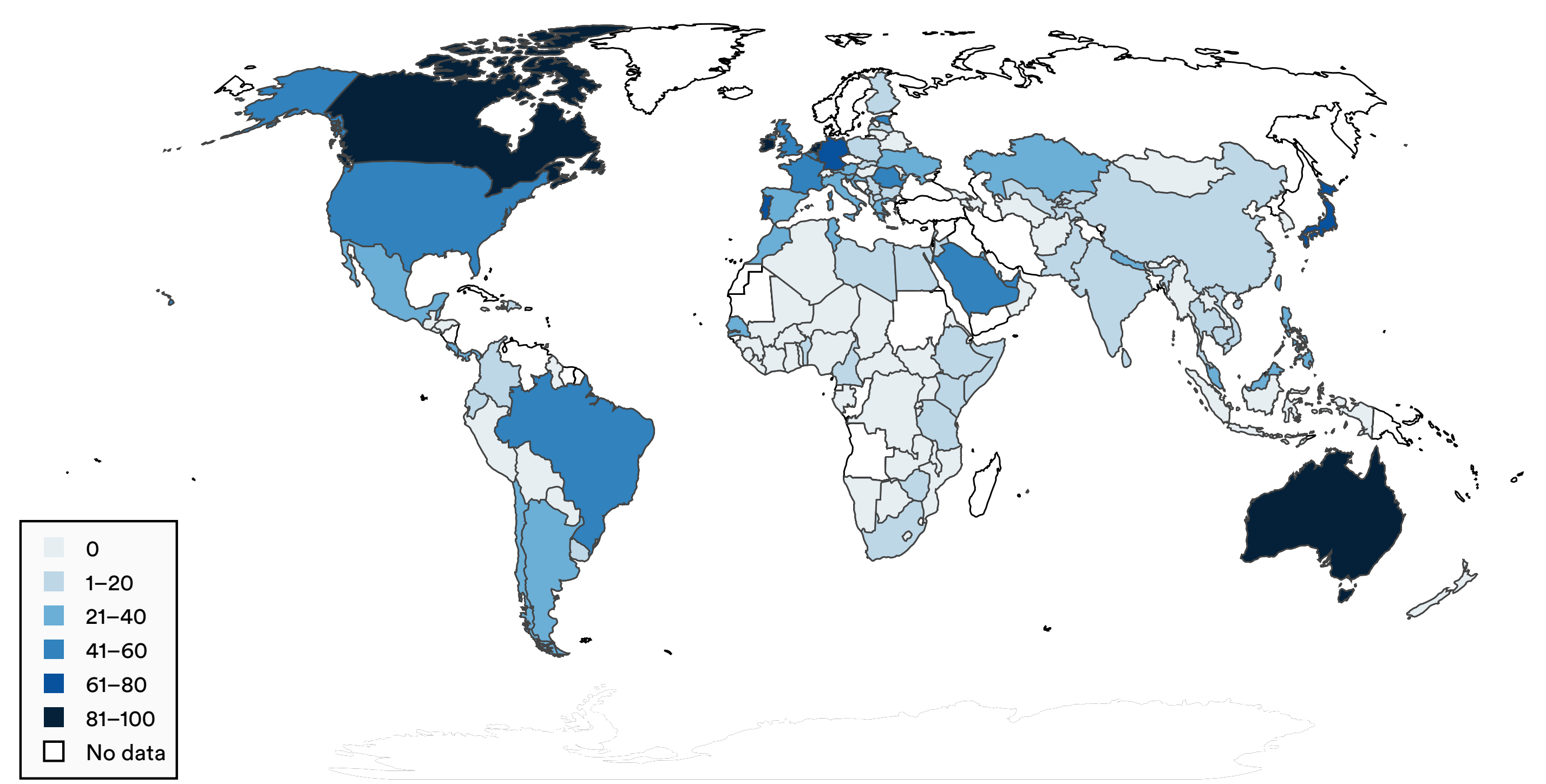

Figure 3.7.2

## Cultural and Linguistic Diversity

GIRAI's cultural and linguistic diversity dimension focuses on countries' protective measures on local languages, dialects, indigenous knowledge systems, and cultural diversity broadly across the AI lifecycle. Dominant-culture assumptions can bias AI, marginalize minorities, and erode minority languages. Scores on this dimension are more evenly spread than the others (Figure 3.7.3). Singapore scores the highest, while Germany, Ireland, Italy, Qatar, Estonia, and Slovenia also score in the upper ranges (70–88).

Not all regions protect cultural and linguistic diversity the same way (Figure 3.7.4). In North America, government programs and nonstate actors, such as advocacy groups, research institutions, and digital rights organizations, are active, but formal legal frameworks are less developed. In Europe, Asia, and the Middle East, nonstate actors are also doing more than the government. In Africa, the gap is especially pronounced. Nonstate actors show activity in 39% of countries, but only 7% have government programs and just 2% have legal frameworks in place.

## Global AI cultural and linguistic diversity assessment

Source: The Global Index on Responsible AI, 2024 | Chart: 2026 AI Index report

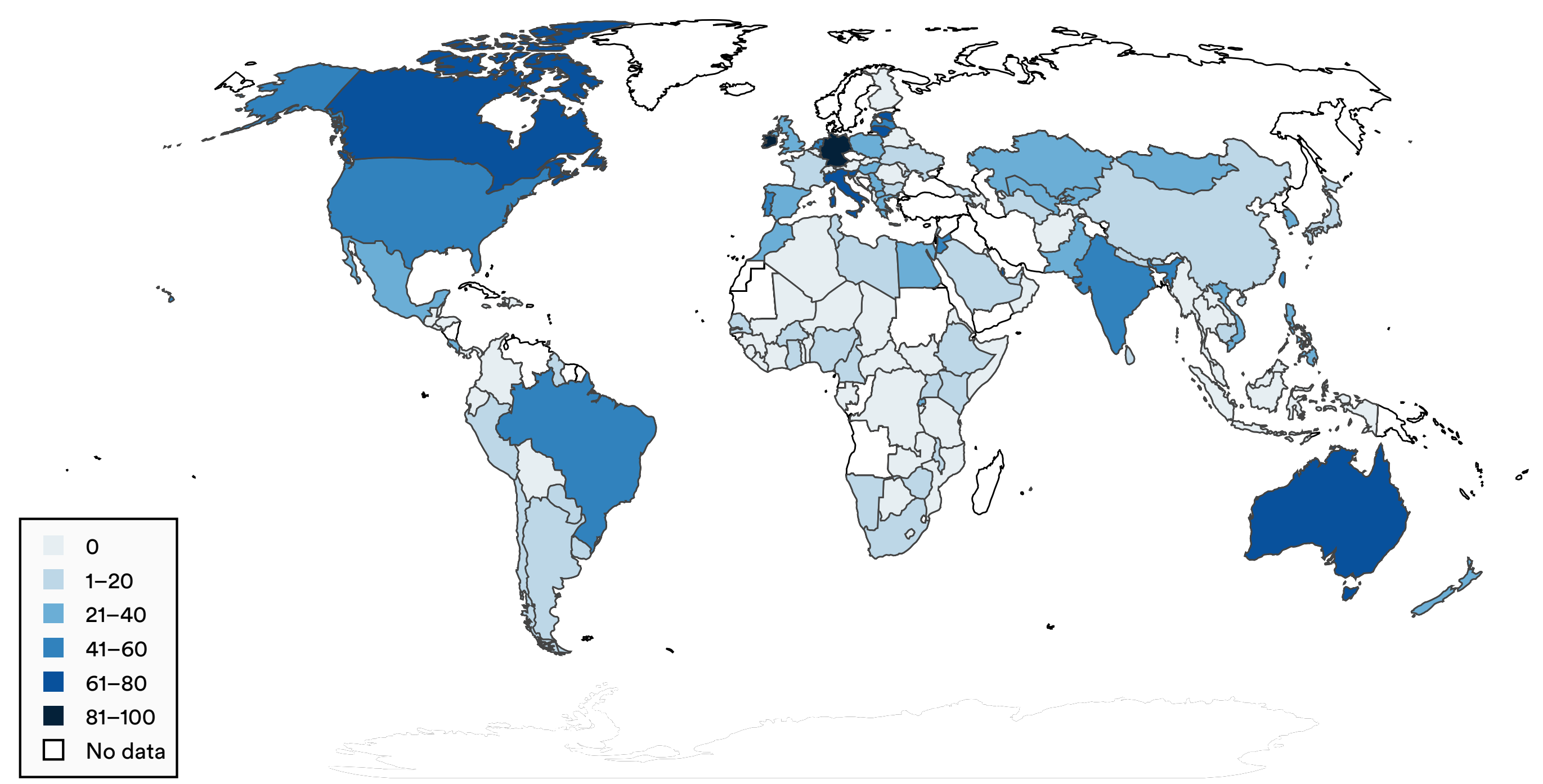

Figure 3.7.3

## Share of countries with evidence on cultural and linguistic diversity in AI by region and category

Source: The Global Index on Responsible AI, 2024 | Chart: 2026 AI Index report

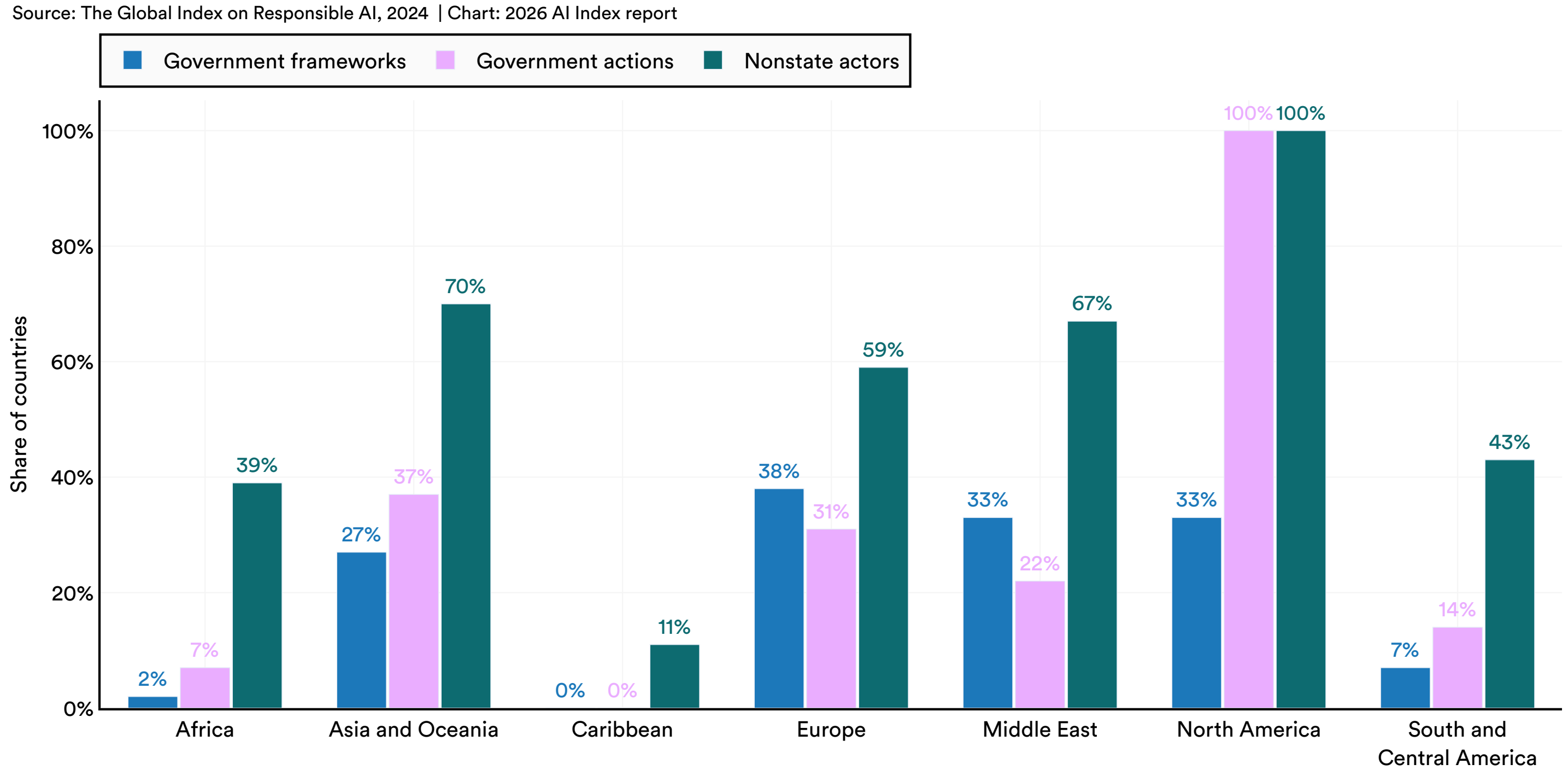

Figure 3.7.4

**HIGHLIGHT:**

## Inclusiveness and the Global Language Gap

As a small number of proprietary models shape global AI capabilities, the "global language gap" has become more visible. These systems perform much better in English and a handful of other widely spoken languages than in all others. This is a responsible AI concern because it determines who benefits from AI systems and who does not.

Efforts continued in the area of language- and culture-specific foundation models and benchmarks, such as KoBEST in 2022 and HAE-RAE in 2023, alongside other Korean-tailored models including Polyglot-Ko and HyperCLOVA X. Spain's Language Technologies Plan, launched in 2019, laid the groundwork for what became the publicly funded ALIA family of Spanish and regional-language models, with earlier regional efforts such as Catalonia's AINA project predating the current wave of regional benchmarking. In 2025, the pace and visibility of this work picked up, with new benchmarks and models emerging across more regions and beginning to register in global evaluation infrastructure.

HELM Arabic, a regional extension of Stanford CRFM's HELM framework developed with Arabic.ai, evaluates models across seven Arabic-language benchmarks covering academic assessment, grammar, and region-specific safety. On this evaluation, the top-scoring model was Arabic.ai's LLM-X, a regionally developed model, with a mean score of 0.86, ahead of Gemini 2.5 Flash (0.82) and GPT-5.1 (0.81) (Figure 3.7.5). Rankings that hold in English-centric evaluations do not necessarily hold when benchmarks reflect local usage, dialect, and cultural references.

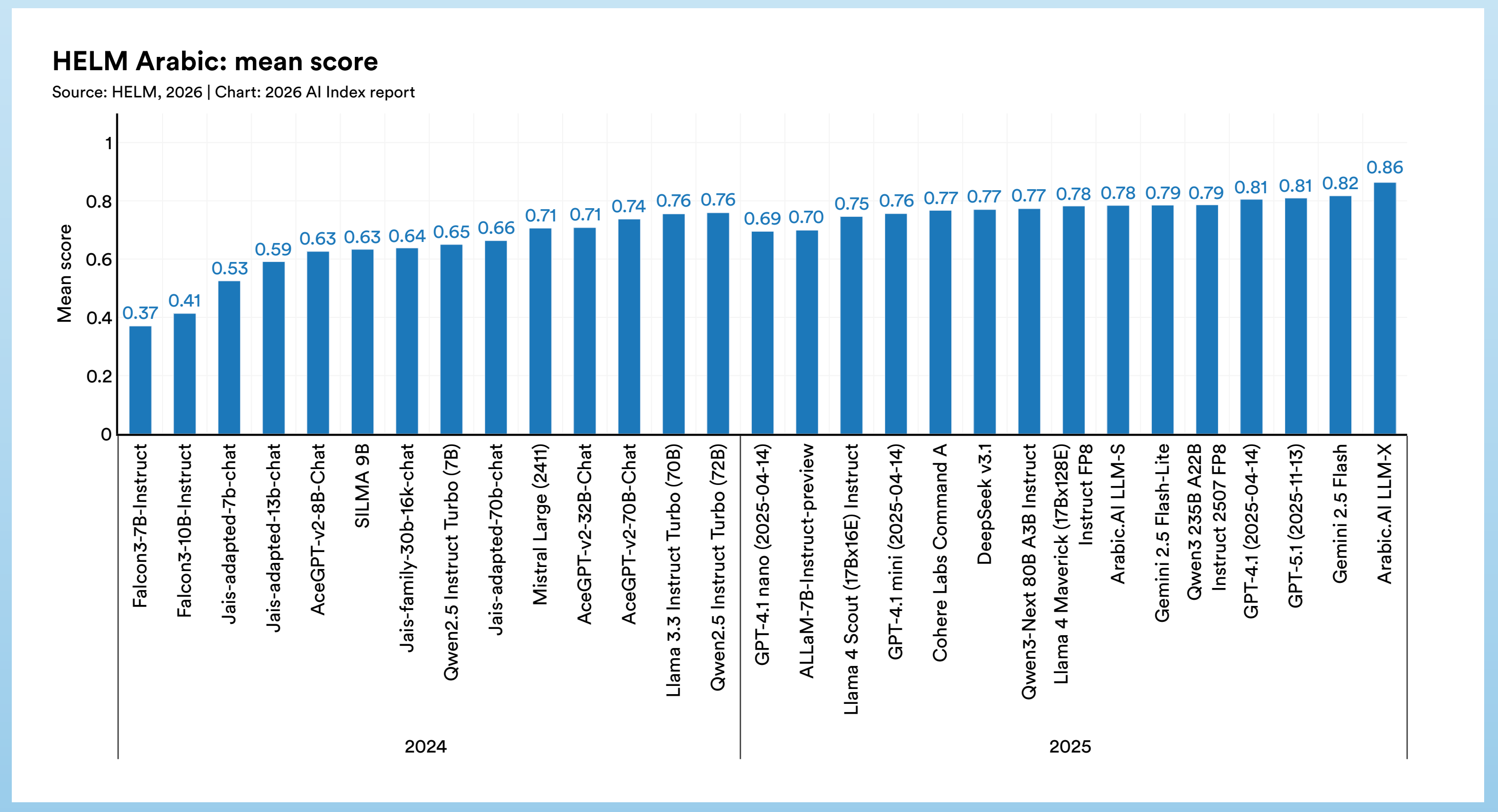

Figure 3.7.5[14]

A similar pattern appears in the Indic LLM Arena, a crowd-sourced evaluation led by AI4Bharat at IIT Madras that tests models across more than 20 Indian languages on language quality, cultural grounding, and safety.

14 Data source: https://crfm.stanford.edu/helm/arabic/latest/.

**HIGHLIGHT:**

Proprietary models led the leaderboard, with GPT-5.2 scoring 1,314, followed by GPT-5.1 (1,298) and Gemini 3 Flash (1,288) (Figure 3.7.6). Open-source models scored lower but remained competitive, with Qwen3-Next-80B at 1,156 and Llama-4-Maverick-17B at 1,108. The evaluation goes beyond translation accuracy to test whether responses are contextually appropriate for Indian users, a dimension that global benchmarks typically do not capture.

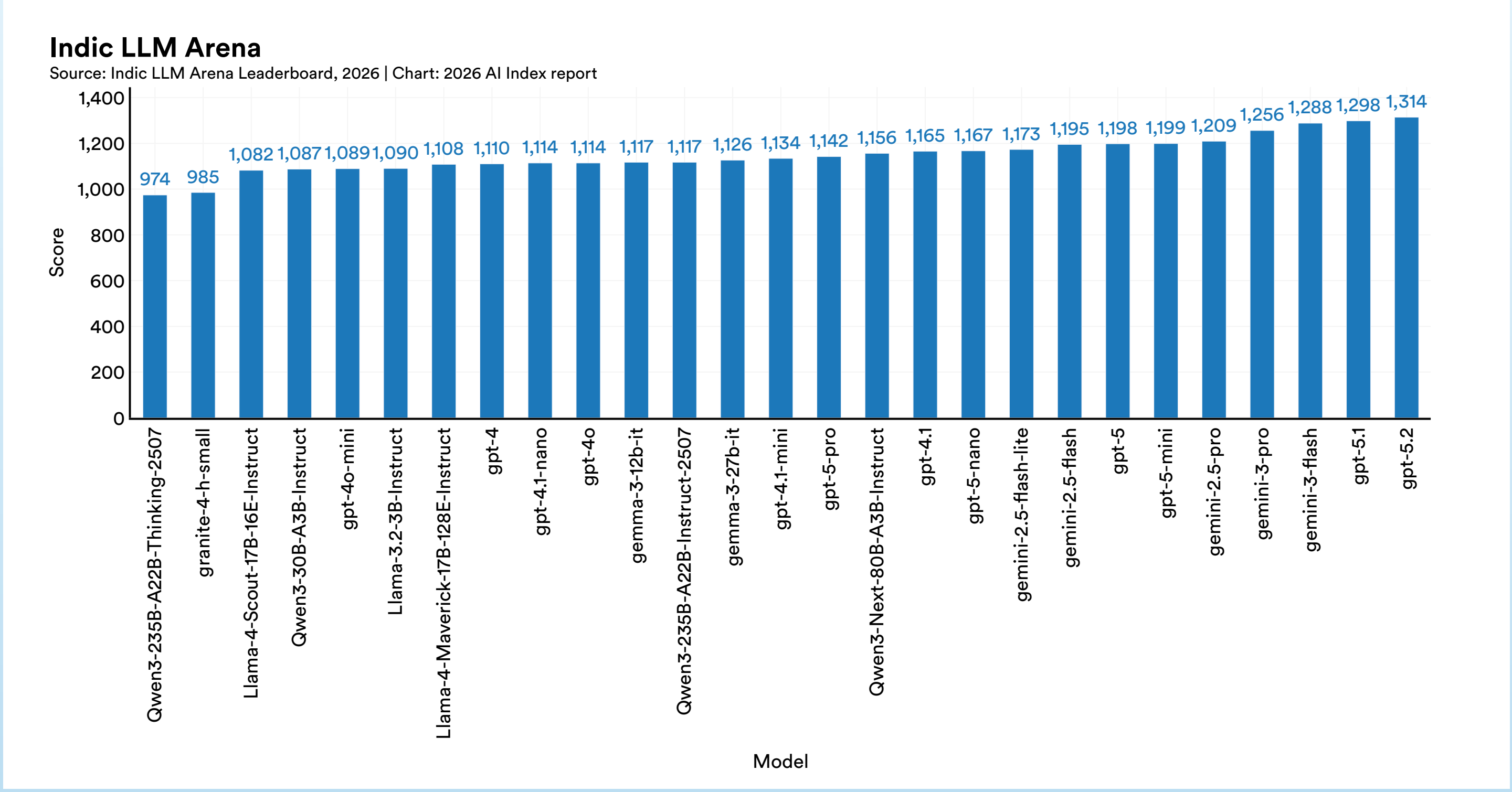

Figure 3.7.6[15]

The gap extends beyond language boundaries to dialect variation within the same language. The Slovene DIALECT-COPA benchmark tests commonsense reasoning in both Standard Slovenian and the Cerkno dialect. GPT-5 scored 99.8% on Standard Slovenian but dropped to 88.6% on the dialect (Figure 3.7.7). The drop was steeper for other models. Mistral Medium 3.1 fell from 90.0% to 53.2%, and Llama 3.3 fell from 87.0% to 53.6%. Dialects differ from standard varieties in spelling, vocabulary, and grammar, and are rarely represented in training data. These gaps suggest that even within languages that models handle reasonably well, performance can degrade sharply for speakers of nonstandard varieties.

15 Data source: https://arena.ai4bharat.org/#/leaderboard/chat/overview.

HIGHLIGHT:

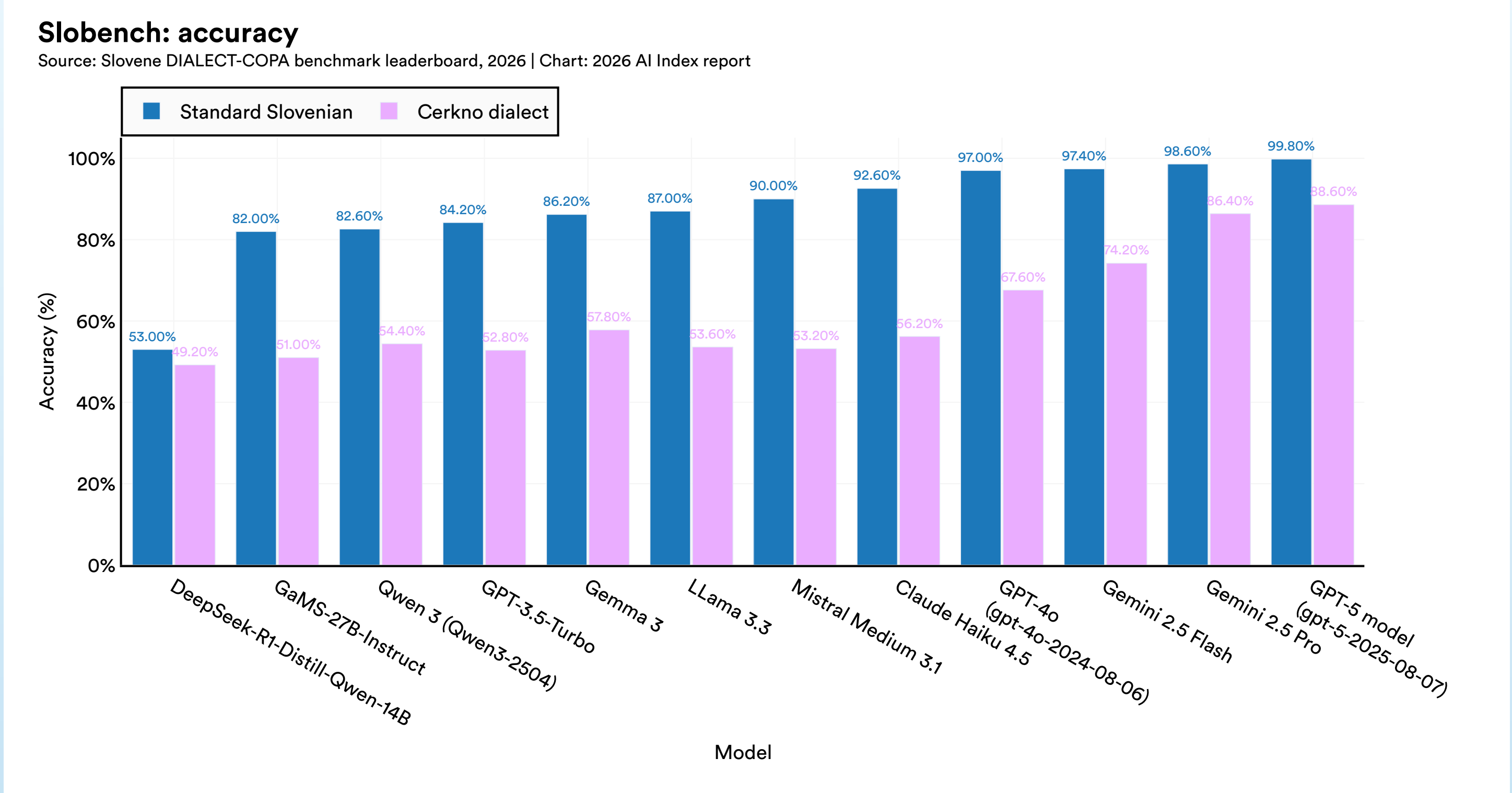

Figure 3.7.7[16]

In response to these gaps, a growing number of regional initiatives are building language-specific AI infrastructure from the ground up rather than waiting for global labs to add coverage. Projects like SEA-LION in Southeast Asia and AI4Bharat in India are developing their own data pipelines, tokenizers, and evaluation benchmarks tailored to local linguistic conditions. Many of the languages these projects serve have structural features, such as complex morphology, script diversity, and limited digitized text, that cause standard multilingual tools to perform poorly. These efforts position linguistic inclusiveness not as an afterthought but as a design requirement, and they represent a growing layer of responsible AI infrastructure outside the major AI-producing regions.

## AFRICA

| Benchmark | Languages covered | Focus |
|---|---|---|
| AfroBench | 64 African languages | Multi-task LLM evaluation across NLU, generation, QA/knowledge, and math (15 tasks; 22 datasets) |
| IrokoBench | 17 low-resource African languages across West/East/Southern/Central Africa | Human-translated suite covering NLI (AfriXNLI), math reasoning (AfriMGSM), and multi-choice knowledge QA (AfriMMLU) |
| TerjamaBench | Darija (Arabic script + Latin “Arabizi”) | English↔Darija machine translation benchmark emphasizing cultural context and regional variation (850 entries) |

16 Data source: https://slobench.cjvt.si/leaderboard/view/17.

**HIGHLIGHT:**

| Benchmark | Languages covered | Focus |
| --- | --- | --- |
| HausaMovieReview | Hausa (+ code-switched English) | Sentiment/review-style benchmark from 5,000 Hausa YouTube comments reflecting common code-switching |

## ASIA, MENA (ARABIC), CENTRAL ASIA

| Benchmark | Languages covered | Focus |
| --- | --- | --- |
| Indic LLM Arena | Many Indian languages + English-creoles | Crowd-sourced, human-in-the-loop leaderboard evaluating language, culture, and safety in Indian contexts (AI4Bharat; supported by Google Cloud) |
| SEA-HELM | Filipino, Indonesian, Tamil, Thai, Vietnamese | Southeast Asian holistic evaluation of linguistic and cultural competence across multiple tasks |
| BATAYAN | Tagalog, Taglish | Holistic Filipino benchmark spanning understanding, reasoning, and generation; explicitly covers code-switching |
| HELM Arabic | Arabic | Transparent, reproducible Arabic LLM evaluation leaderboard built on established Arabic benchmarks (with Arabic.ai) |
| BALSAM | Arabic | Community-driven Arabic benchmark and platform with blind evaluation; 78 tasks across 14 categories (52K examples) |
| Cetvel | Turkish | Unified Turkish LLM benchmark built from 22 datasets covering 7 tasks, with a side-by-side leaderboard |
| TUMLU | Azerbaijani, Crimean Tatar, Karakalpak, Kazakh, Kyrgyz, Tatar, Turkish, Uyghur, Uzbek | Natively developed multilingual language-understanding benchmark for Turkic languages using middle-/high-school questions across 11 subjects |
| Kyrgyz LLM-Bench | Kyrgyz | Suite for deep understanding and reasoning in Kyrgyz, combining native benchmarks with translated/post-edited international tasks |
| ArmBench-LLM | Armenian | Armenian LLM benchmark combining university entrance exams with MMLU-Pro-Hy (1,000-question translated sample) |
| GeoLogicQA | Georgian | Manually curated 100-question benchmark for logical and inferential reasoning, validated by native speakers |
| CantoNLU | Cantonese | Seven-task Cantonese NLU benchmark (syntax/semantics, NLI, sentiment, tagging, parsing) |
| TLUE | Tibetan | Large-scale benchmark measuring LLM proficiency in Tibetan language understanding |

**HIGHLIGHT:**

## EUROPE

| Benchmark | Languages covered | Focus |
| --- | --- | --- |
| BenCzechMark | Czech | Comprehensive Czech-centric benchmark with 50 tasks, multiple formats/metrics, and significance-aware aggregation |
| CUS-QA | Czech, Slovak, Ukrainian | Open-ended regional QA benchmark with text and visual grounding, curated by native speakers with English translations |
| COLE | French | 23-task French natural language understanding (NLU) benchmark emphasizing French-relevant linguistic phenomena (used to benchmark 94 LLMs) |
| Estonian Benchmark | Estonian | Benchmark built from seven datasets covering knowledge, grammar/vocabulary, summarization, and contextual comprehension |
| IberBench | Basque, Catalan, Galician, Spanish, Portuguese, English | Large, extensible benchmark integrating 101 datasets across 22 task categories (e.g., toxicity, summarization) with community-driven updates |
| IberoBench | Basque, Catalan, Galician, European Spanish, European Portuguese | Multi-task benchmark (62 tasks; 179 subtasks) built on the LM Evaluation Harness framework |
| Polish linguistic and cultural competency | Polish | 600 manually crafted questions evaluating Polish history, geography, culture/tradition, arts, grammar, and vocabulary |
| LLMzSzŁ | Polish | Exam-based benchmark drawn from Polish national exams (~19K closed-ended questions across 154 domains) |
| ITALIC | Italian | Culture-aware Italian NLU benchmark with 10,000 multiple-choice questions spanning 12 domains |
| SloBENCH | Slovenian | Evaluation platform with multiple leaderboards, including DIALECT-COPA (standard vs. dialect) and Slovene speech recognition |

Figure 3.7.8

# 3.8 Transparency

Transparency measures how much developers disclose about how their models are built, trained, and deployed. Two independent indices track this from different angles.

## The Openness Index

The Artificial Analysis Openness Index scores AI models on a 0 to 100 scale based on how freely weights can be accessed and licensed, as well as the level of transparency around training methodology and pre- and post-training data. Scores are low across leading models, with most falling between 2 and 16 out of 100 (Figure 3.8.1). K2 Think and Olmo 3 32B Think scored the highest, and they are also the only two models that scored any points for pre-training data transparency. Every other model in the index scores zero in that category. Model Availability and methodology disclosure account for the bulk of points across all models. As Chapter 1's discussion of access and deployment noted, over 90% of notable industry models were released without training code in 2025. The Openness Index results suggest that pattern extends beyond code to training data as well.

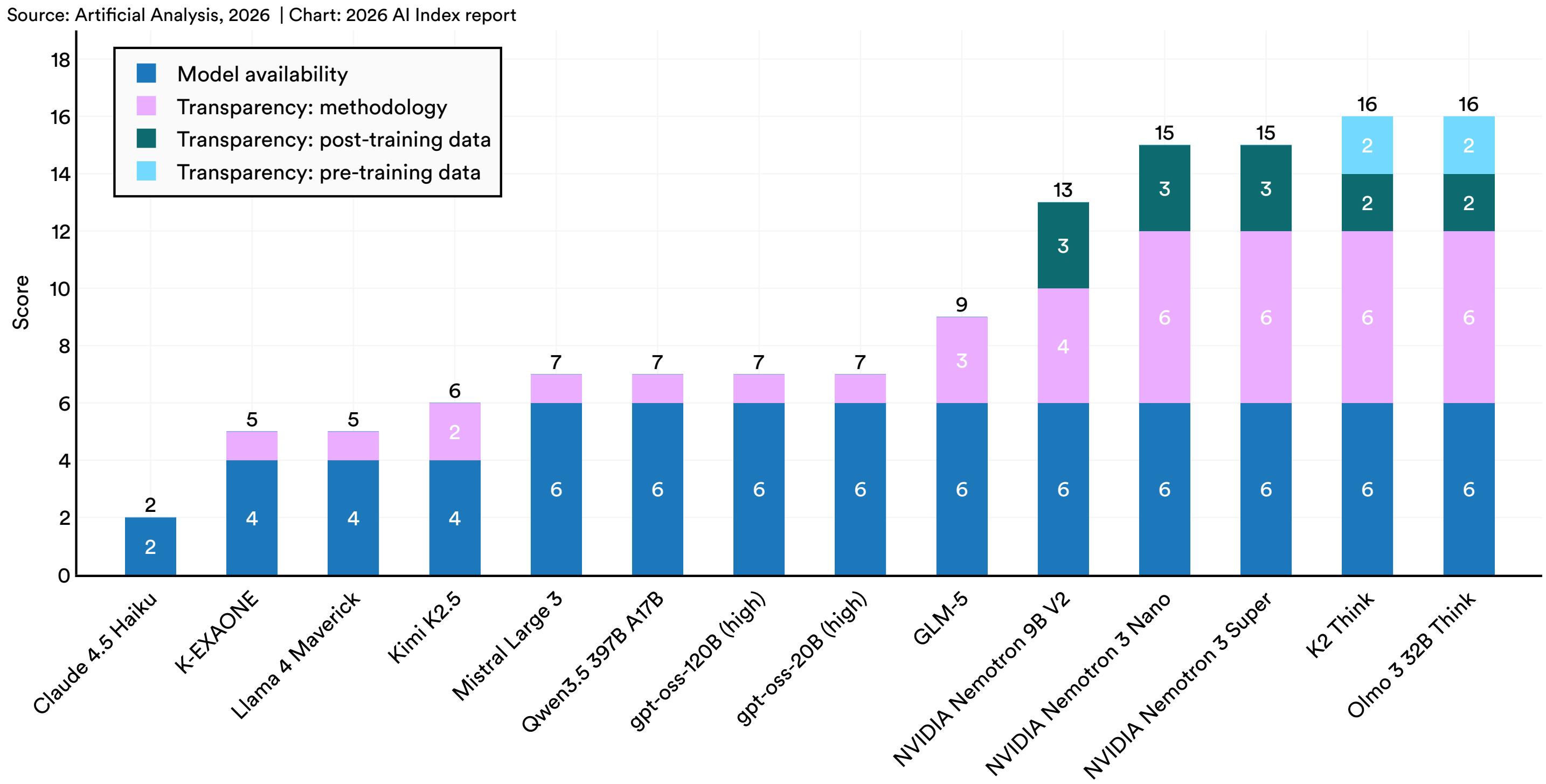
Openness index by components
Source: Artificial Analysis, 2026 | Chart: 2026 AI Index report

Figure 3.8.1

## Foundation Model Transparency Index

The Foundation Model Transparency Index (FMTI) takes a different approach, scoring developers rather than individual models. Now in its third year, it evaluates disclosure across three stages of the model lifecycle. Upstream covers what goes into building a model, including training data, labor, and compute. Model covers what is disclosed about the system itself, and Downstream covers what happens after release, including monitoring and impact reporting.

In the 2025 edition, average transparency declined from 58 in 2024 to 40 (Figure 3.8.2). IBM leads at 95 and Writer follows at 72. Others, such as xAI and Midjourney score just 14, whereas open model developers, B2B enterprise providers, organizations publishing transparency reports, and EU AI Act signatories tend to perform better. As with the Openness Index, the weakest area is Upstream, particularly around training data and the resources used to build models (Figure 3.8.3).

**Foundation Model Transparency Index Scores by Domain, 2025**

Source: 2025 Foundation Model Transparency Index

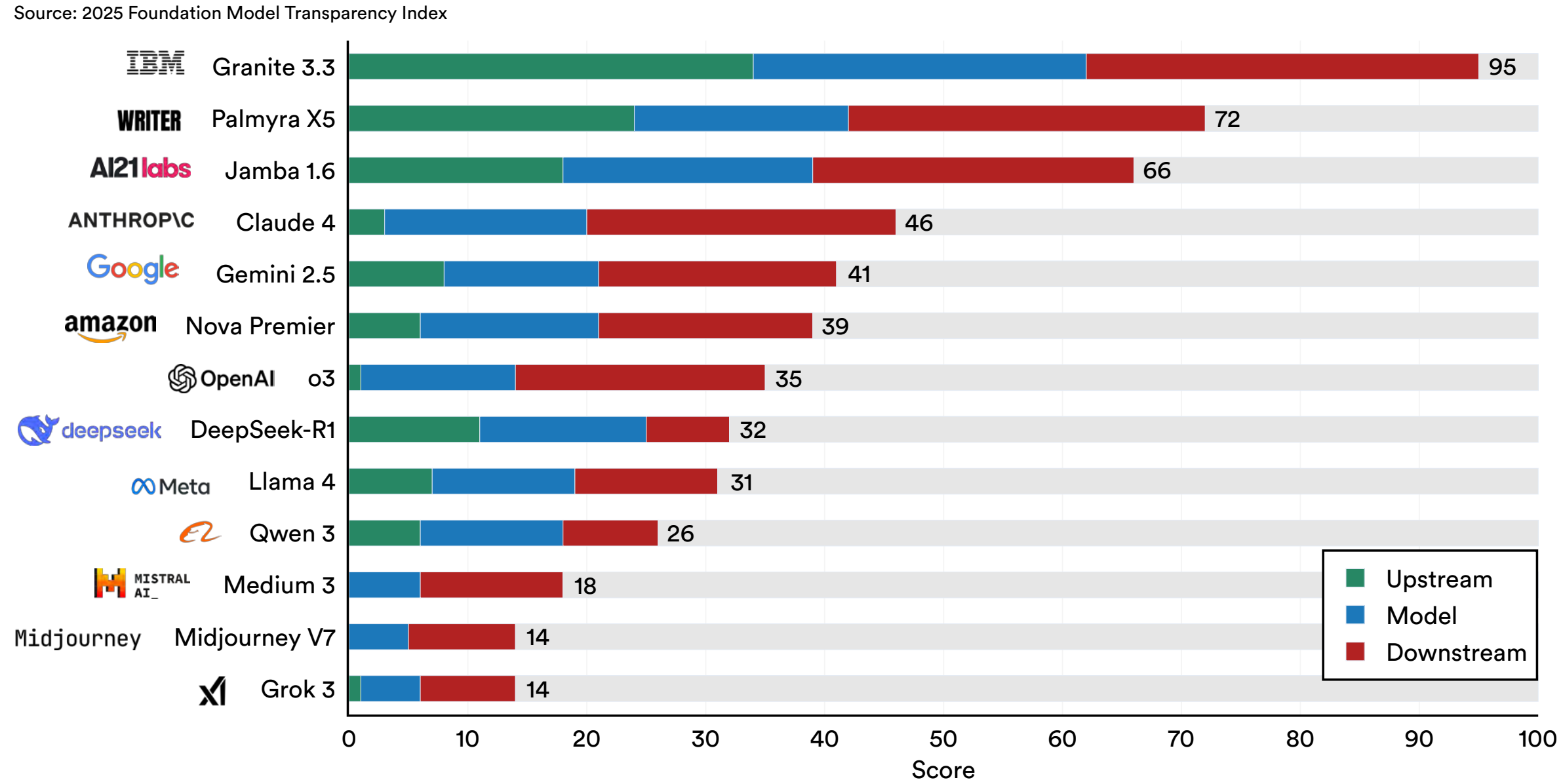

Figure 3.8.2

**Foundation Model Transparency Index Scores by Major Dimensions of Transparency, 2025**

Source: 2025 Foundation Model Transparency Index

| Major Dimensions of Transparency | Jamba 1.6 | Qwen 3 | Nova Premier | Claude 4 | DeepSeek-R1 | Gemini 2.5 | Granite 3.3 | Llama 4 | Midjourney V7 | Medium 3 | o3 | Palmyra X5 | Grok 3 | Average |
|---|---|---|---|---|---|---|---|---|---|---|---|---|---|---|
| Data Acquisition | 92% | 17% | 17% | 25% | 17% | 33% | 100% | 33% | 0% | 0% | 8% | 58% | 0% | 31% |
| Data Properties | 0% | 20% | 0% | 0% | 20% | 0% | 100% | 20% | 0% | 0% | 0% | 40% | 0% | 15% |
| Compute | 22% | 11% | 11% | 0% | 44% | 11% | 100% | 22% | 0% | 0% | 0% | 100% | 11% | 26% |
| Model Information | 75% | 75% | 0% | 25% | 75% | 0% | 100% | 75% | 0% | 0% | 0% | 75% | 0% | 38% |
| Model Access | 50% | 50% | 50% | 50% | 50% | 50% | 100% | 50% | 0% | 25% | 0% | 50% | 0% | 40% |
| Capabilities | 75% | 50% | 50% | 25% | 50% | 25% | 75% | 50% | 0% | 25% | 25% | 50% | 25% | 40% |
| Risks | 60% | 0% | 40% | 60% | 20% | 20% | 100% | 20% | 0% | 0% | 60% | 40% | 0% | 32% |
| Model Mitigations | 60% | 0% | 60% | 80% | 20% | 40% | 80% | 0% | 0% | 20% | 80% | 40% | 0% | 37% |
| Release | 88% | 63% | 75% | 75% | 63% | 88% | 100% | 50% | 63% | 38% | 63% | 88% | 50% | 69% |
| Usage Data | 20% | 0% | 20% | 60% | 0% | 0% | 80% | 0% | 20% | 0% | 20% | 100% | 0% | 25% |
| Impact | 71% | 0% | 0% | 29% | 0% | 29% | 86% | 14% | 29% | 14% | 14% | 86% | 0% | 29% |
| Post-deployment Monitoring | 71% | 0% | 57% | 57% | 0% | 43% | 100% | 29% | 0% | 43% | 71% | 86% | 0% | 43% |
| Model Behavior Policy | 100% | 50% | 75% | 100% | 75% | 75% | 100% | 75% | 25% | 0% | 75% | 50% | 75% | 67% |
| Acceptable Use Policy | 80% | 60% | 80% | 100% | 60% | 80% | 80% | 40% | 60% | 60% | 60% | 80% | 60% | 69% |
| Downstream Mitigations | 100% | 40% | 100% | 100% | 0% | 100% | 100% | 80% | 40% | 80% | 100% | 100% | 40% | 75% |
| Average | 64% | 29% | 42% | 52% | 33% | 40% | 93% | 37% | 16% | 20% | 38% | 69% | 17% | |

Figure 3.8.3[17]

17 Data, labor, compute, and methods were upstream indicators; model basics, access, capabilities, risks, and mitigations were model-level indicators; and distribution, usage policy, feedback, and impact were downstream indicators.

# 3.9 Security and Safety

Safety is the responsible AI dimension where institutional infrastructure has grown the fastest. New evaluation frameworks, government-backed AI safety institutes, and standardized benchmarks have all expanded in the past year. This section traces that growth and the resulting data on how well current models handle safety in practice.

## Global AI Safety Institutes

AI safety institutes (AISIs) are state-backed specialist organisations created to help governments understand and manage risks from advanced AI, especially frontier/foundation models. They conduct technical evaluations and safety research that governments can use to shape policy.

Fully operational institutes now exist in the UK (AI Security Institute), the U.S. (USAISI at NIST), Japan (JAISI), Singapore (Digital Trust Centre), and Israel (AI Security Research Unit) (Figure 3.9.1). India and France have also launched AISIs, with India's AI Safety Institute and France's Current AI. A second wave is in development in Canada, South Korea, Germany, and Brazil. Outside of these standalone institutes, participation is growing through the International Network of AI Safety Institutes, with Kenya and Australia listed as network members without formal institutes of their own.

The countries building these AISIs are still mostly wealthy, technologically advanced economies that are not all pursuing the same goals. The UK and Israel emphasize security, while the EU AI Office pairs evaluation with enforcement powers under the AI Act. Network membership is a practical entry point for countries without the resources to stand up a full institute immediately.

**AI safety institutes and network membership**
Source: All Tech is Human, 2025 | Chart: 2026 AI Index report

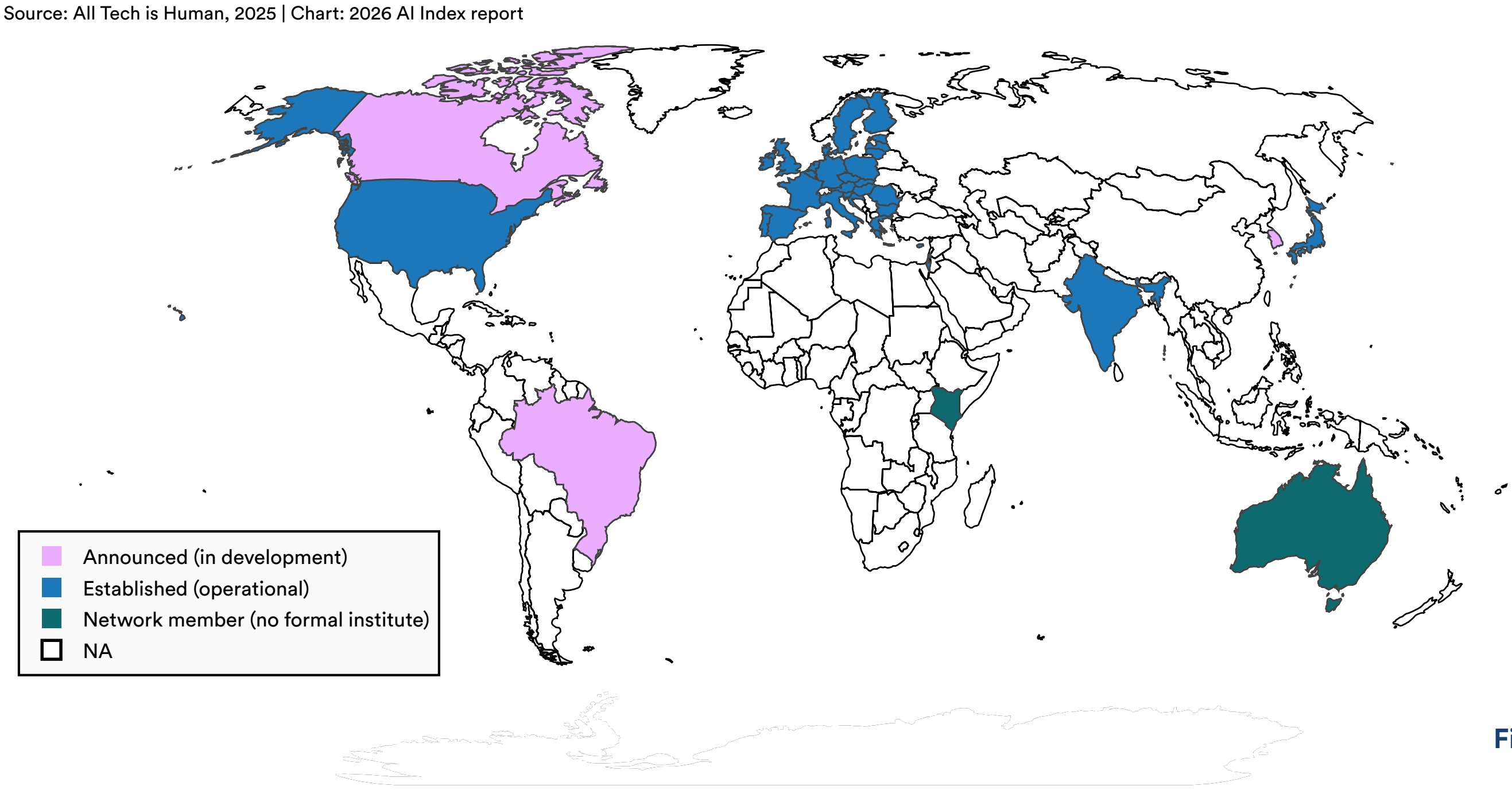

**Figure 3.9.1**[18]

18 Data source: https://alltechishuman.org/all-tech-is-human-blog/the-global-landscape-of-ai-safety-institutes.

# Benchmarks

## HELM Safety

HELM Safety, covered in last year's report, continues to be one of the few standardized suites for evaluating AI models across responsibility and safety metrics. It tests models from major developers across benchmarks including BBQ (social bias), SimpleSafetyTests (self-harm and abuse risks), HarmBench (harassment and misinformation), AnthropicRedTeam (adversarial conversations), and XSTest (helpfulness vs. harmlessness trade-offs).

The 2025 results show continued improvement but also increasing compression at the top (Figure 3.9.2). Most models released between 2024 and 2025 score between 0.90 and 0.98, with a very narrow gap between the highest and lowest scorers. Older models from 2023 score lower, but the overall trajectory suggests that leading models are converging on a safety ceiling where current benchmarks may not be fine-grained enough to distinguish meaningful differences.

**HELM Safety: mean score**

Source: HELM, 2026 | Chart: 2026 AI Index report

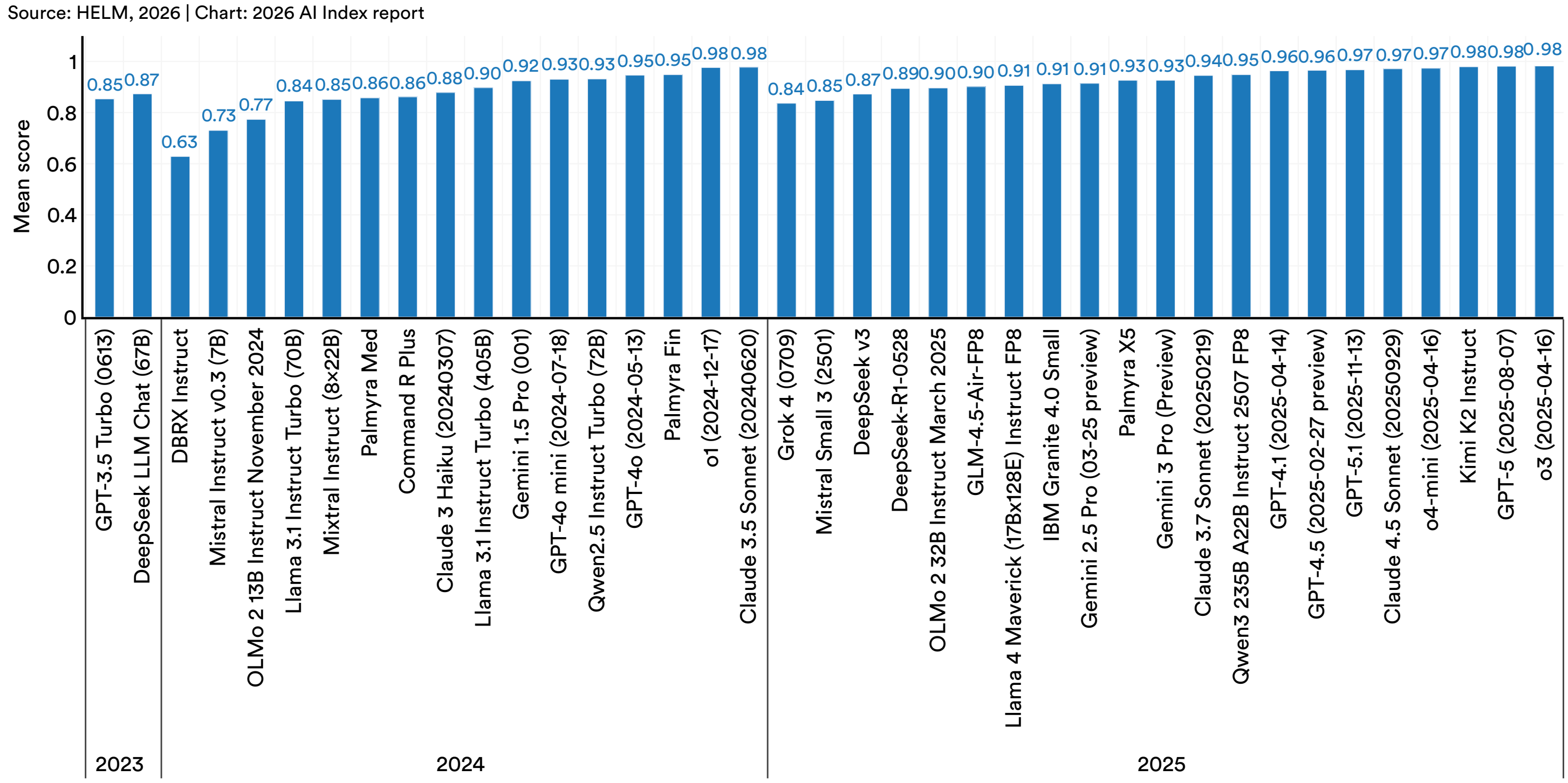

Figure 3.9.2

## AILuminate

AILuminate v1.0 is a new benchmark designed to test how well AI systems resist prompts that could trigger dangerous, illegal, or undesirable behavior. It covers 12 hazard categories, including violent crimes and child exploitation, and employs a five-tier grading scale from “Poor” to “Excellent.” The benchmark includes two separate evaluations. The first tests safety under normal use, with models evaluating both with and without external safety filters and moderation tools. The second tests a system’s ability to resist deliberate jailbreak attempts through adversarial prompts.

### Safety Benchmark Results

Among models tested with external guardrails in place, Claude 3.5 Haiku, Claude 3.5 Sonnet, and Mistral Large all received “very good” ratings, while their parent models received “good”(Figure 3.9.3). In the set of models that could be tested without external safety filters or moderation tools, Gemma 2 9b, Phi 3.5 MoE Instruct, and Phi 4 scored “very good” (Figure 3.9.4). The two groups are not directly comparable, as they involve different models under different conditions, but both show a baseline safety performance of “good” across leading systems.

| Name | Grade |
|---|---|
| Claude 3.5 Haiku 20241022 | Very Good |
| Claude 3.5 Sonnet 20241022 | Very Good |
| Mistralai Mistral Large 2402 Moderated | Very Good |
| Amazon Nova Lite v1.0 | Good |
| Gemini 1.5 Pro (API, with option) | Good |
| Gemini 2.0 Flash 001 | Good |
| Gemini 2.0 Flash Lite | Good |
| GPT-4o | Good |
| GPT-4o mini | Good |
| Ministral 8B 24.10 with output moderation (Recipe) | Good |

Source: MLCommons, 2025

**Figure 3.9.3**

| Name | Grade |
|---|---|
| Gemma 2 9b | Very Good |
| Phi 3.5 MoE Instruct | Very Good |
| Phi 4 | Very Good |
| Athene V2 Chat Hf | Good |
| Aya Expanse 8B Hf | Good |
| Cohere C4Ai Command A 03 2025 Hf | Good |
| Llama 3.1 405B Instruct | Good |
| Llama 3.1 8b Instruct FP8 | Good |
| Llama 3 1 Tulu 3 8B Hf | Good |
| Mistralai Mistral Large 2402 | Good |
| Olmo 2 0325 32b Instruct | Good |
| Olmo 2 1124 13B Instruct Hf | Good |
| Phi 3.5 Mini Instruct | Good |
| Qwen1 5 110B Chat Hf | Good |
| Yi 1 5 34B Chat Hf | Good |
| Ai21Labs Ai21 Jamba Large 1.5 Azure | Fair |
| Google Gemma 3 27B It Hf Nebius | Fair |
| Llama 3.3 70B Instruct Turbo Together | Fair |
| Ministral 8B 24.10 (API) | Fair |
| Mistral Large 24.11 | Fair |
| Qwq 32B Hf | Fair |
| OLMo 7b 0724 Instruct | Poor |

Source: MLCommons, 2025

**Figure 3.9.4**

### Jailbreak T2T Benchmark v0.5 Results

The AILuminate Jailbreak T2T benchmark v0.5 tests what happens when users deliberately try to bypass a model's safety measures through adversarial prompts. Each model in the chart receives two scores (Figure 3.9.5). The square at the top represents the model's safety score under normal conditions, while the circle below it represents the score after being exposed to jailbreak attempts. As this is a beta version of the benchmark, models are de-identified by number, rather than named.

Under normal conditions, most models score in the "very good" or "good" range. After jailbreak attempts, nearly every system's score drops, some by a full tier or more. So while safety under normal use is generally good, it degrades under deliberate manipulation.

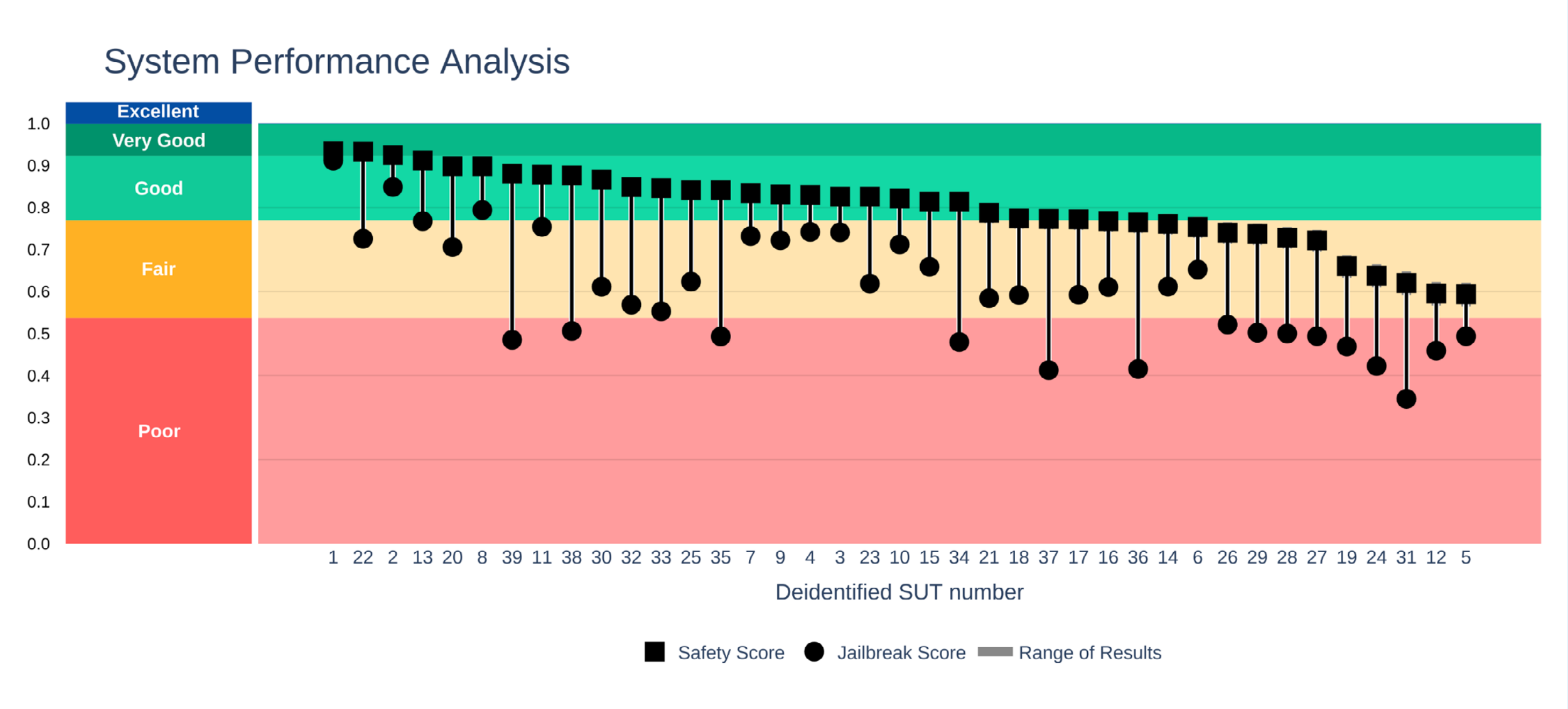

Figure 3.9.5

# 3.10 Tradeoffs Across RAI Dimensions

In practice, AI systems must satisfy multiple responsible AI dimensions at once. A growing number of empirical research studies suggest that these dimensions do not improve independently, as optimizing for one can degrade others. The direction and magnitude of those trade-offs depends on the method used, data involved, and under what context it is deployed.

Kemmerzell and Schreiner (2024) tested this directly by training image classification models on four facial analysis data sets and measuring what happened to fairness, privacy, explainability, and robustness when each dimension was targeted in isolation. Differential privacy, a technique that adds noise during training to prevent individual data points from being identified, improved privacy scores across all datasets but reduced explainability, fairness, and accuracy, with accuracy falling by up to 33 percentage points on some configurations. Training adaptations aimed at improving fairness only succeeded on the dataset with the most demographic imbalance, and therefore the most room to correct. But across all, it reduced explainability and robustness. Data augmentation methods designed to improve robustness by exposing datasets to more varied training images produced the fewest negative side effects across the same experiments. It also improved explainability and accuracy, with only minor reductions in privacy and fairness. There was not a single intervention method that proved to improve all four dimensions at once.

A separate evaluation of large language models found a similar pattern at the model level. Cecchini et al. (2024) scored 11 models across robustness, accuracy, and toxicity using the LangTest evaluation toolkit. GPT-4 led on robustness (average score of 0.91 out of 1.0) and accuracy (0.67), but Llama 2 7B scored highest on toxicity avoidance (0.98), meaning it was the most likely to refuse toxic prompts. Models that performed well on robustness, such as Mistral 7B and Mixtral 8×7B, scored among the lowest on toxicity avoidance (0.39 and 0.42, respectively). The ranking of models shifted depending on which dimension was being measured, and no single model was a clear leader in all three.

These trade-offs also appear in federated learning, a training approach where multiple institutions train a shared model by exchanging model updates rather than the underlying data. Wasif et al. (2025) studied how privacy-preserving techniques interact with fairness in this setting across four datasets, including Alzheimer's disease MRI scans and credit card fraud records. Differential privacy did not affect all datasets equally. Institutions with larger datasets could absorb the added noise, while smaller institutions saw their contributions to model training degraded. In the Alzheimer's scenario, adding stronger privacy protections reduced the model's ability to correctly identify the disease, with accuracy falling by 14.8 percentage points. The effect was worse for hospitals with less data, where missed diagnoses rose by 21.4%. Two alternative privacy methods that use encryption instead of noise kept fairness more stable but required two to three times more computing power.

The studies covered above are recent and focus on specific tasks rather than general-purpose AI systems. Their findings point in the same direction though: Improving one responsible AI dimension tends to come at the expense of another. There is no shared framework that measures or compares these trade-offs, which is another measurement gap in the RAI space, and makes it difficult to track whether the field is getting better at managing them.

# 4

# Economy

## Overview

In 2025, more money flowed into AI than ever before, and faster.  Global corporate AI investment more than doubled, revenue at leading frontier companies grew at historically fast rates. Generative AI reached close to 53% population-level adoption within three years of its mass-market introduction, faster than the personal computer or the internet, and that rapid uptake is translating into real value. U.S. consumer surplus from generative AI reached an estimated $172 billion annually by early 2026. But the benefits of this expansion are not distributed evenly. Investment is heavily concentrated in a small number of countries, companies and deals. In labor markets, demand for AI skills is rising across sectors but the workforce impact is showing signs of falling disproportionately on the youngest workers in AI-exposed occupations. Productivity gains are measurable within narrow tasks, but the evidence at the macro level remains early and mixed. The AI economy is scaling quickly, but how widely and how fairly that growth translates into real economic value is still an open question.

## Contents

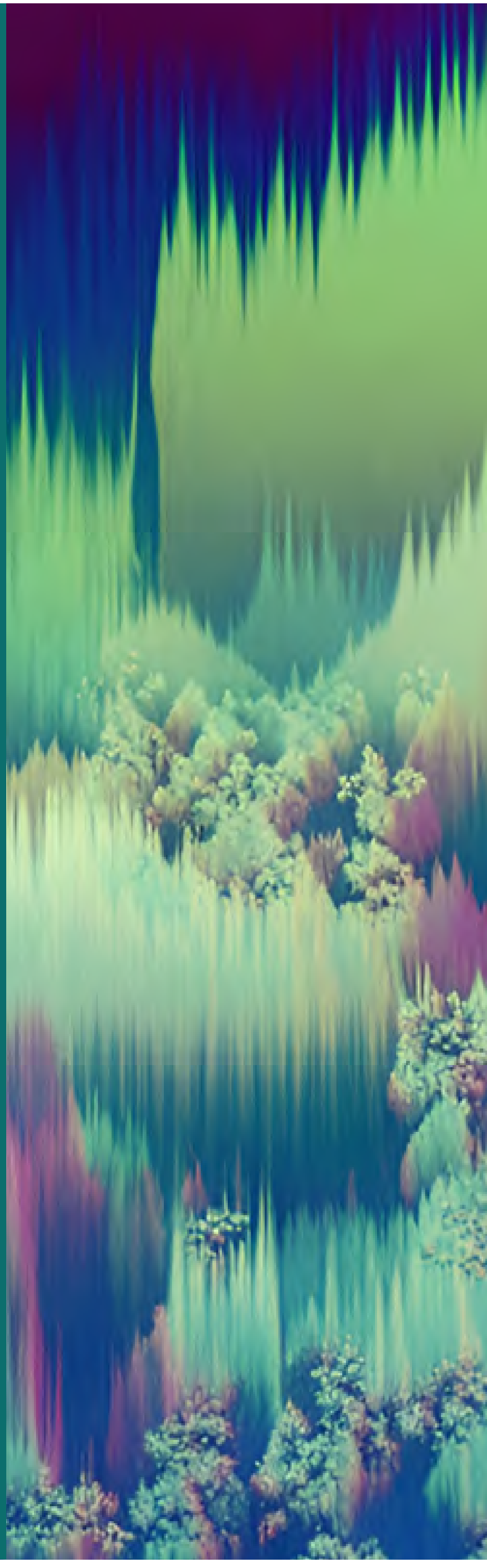

# Chapter Highlights

1. **Global corporate AI investment more than doubled in 2025.** Private investment grew fastest at 127.5% and now accounts for 60% of the total. Generative AI led the surge, growing more than 200% and capturing nearly half of all private AI funding. Newly funded AI companies rose 71%, and billion-dollar funding events nearly doubled.

2. **The United States continues to lead in global private AI investment, committing 23 times more than China.** In generative AI, U.S. investment exceeded the combined total of China and Europe by a wide margin. However, private investment figures likely understate China's total AI spending, as government guidance funds have deployed an estimated $184 billion into AI firms between 2000 and 2023.

3. **AI company revenue is rising at historically fast rates, but compute costs and infrastructure spending are also reaching record levels.** Leading frontier companies are reaching meaningful revenue scale in a short period of time, but compute spend has increased significantly year-over-year. Major cloud providers have accelerated capital expenditures, with Google reporting more than $150 billion in annual capex in 2025.

4. **The value consumers get from generative AI grew 54% in a year.** Estimated U.S. consumer surplus reached $172 billion annually by early 2026, up from $112 billion a year earlier, with the median value per user tripling over the same period. Most of these tools remain free or close to it.

5. **Organizational AI adoption continued to rise in 2025, up to 88% of surveyed organizations, though AI agent use remains early.** Generative AI is now used in at least one business function at 70% of organizations, and China and Europe posted the highest year-over-year increases. AI agent deployment was in the single digits across nearly all business functions.

6. **Generative AI reached 53% adoption in three years, faster than the personal computer or the internet.** Adoption varies widely across countries and correlates strongly with GDP per capita, though some outpace what income would predict, including Singapore at 61% and the United Arab Emirates at 64%. Despite its lead in AI investment and model development, the United States ranks 24th at 28.3%.`

7. **AI's labor market effects are showing up unevenly, concentrated in hiring pipelines and the youngest workers in exposed occupations.** Employment for software developers ages 22 to 25 has fallen nearly 20% from 2024. Employer surveys point to further change ahead, with one-third of respondents expecting workforce reductions over the coming year.

8. **One-third of organizations expect AI to reduce their workforce in the coming year, even though large-scale job losses have not yet shown up in overall employment data.** Almost half of organizations surveyed expected little to no change. Anticipated reductions are highest in service operations, supply chain, and software engineering. Across nearly all functions, anticipated decreases outpaces those already observed.

9 **Productivity gains from AI are largest in structured, measurable work where outputs are easy to monitor.** Studies report gains of 14% to 15% in customer support, 26% in software development, and 50% in marketing output. Gains are smaller in tasks requiring deeper reasoning, and recent evidence raises concerns that heavy AI reliance may carry long-term learning penalties that slow skill development over time.

10 **China continues to install more industrial robots than the rest of the world combined, and the gap widened in 2024.** China accounted for 54% of industrial robots installed globally, up from 51.1% in 2023. Global year-over-year growth was flat, and several major markets, including the United States, Germany, and Italy saw declines. Taiwan was an exception, recording the highest year-over-year growth at 33%.

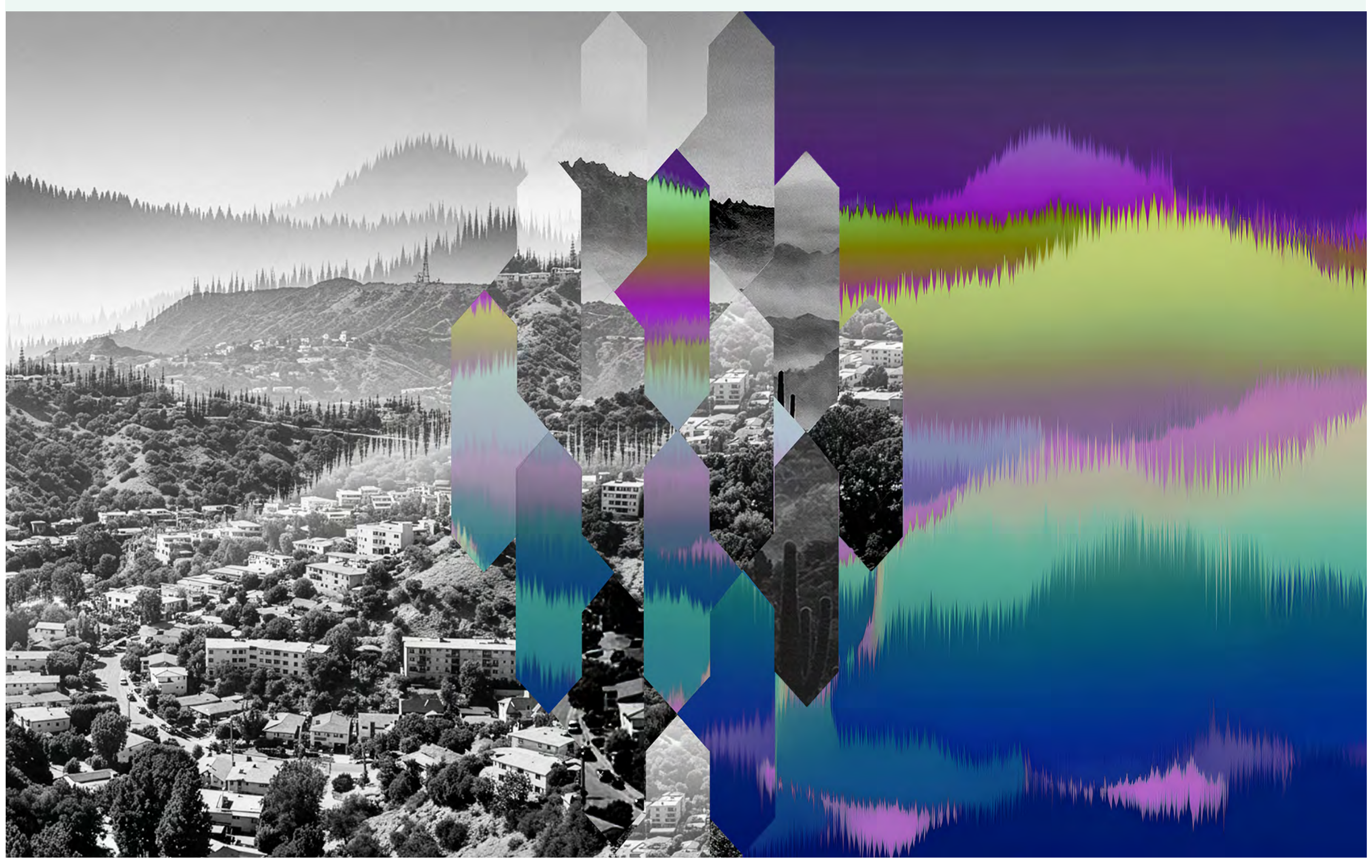

# 4.1 Year in Review: 2025

2025

January 21

INVESTMENT

**$500B:** “Stargate Project” AI Infrastructure joint venture announced: OpenAI, SoftBank, Oracle, and MGX—supported by Nvidia and others—launch Stargate, a major AI infrastructure project announced at the White House. The venture plans to invest between $100 billion and $500 billion to build advanced AI data centers across the United States by 2029.

January 27

MILESTONE

**No. 1:** DeepSeek reaches No. 1 as the most downloaded free app on Apple’s U.S. App Store.

March 6

INVESTMENT

**$138B:** China announces a $138 billion state VC fund to invest in AI and other cutting-edge technologies.

March 10

ACQUISITION

**Agentic AI:** ServiceNow announces plans to acquire Moveworks to drive use of its agentic AI platform across key growth areas including CRM.

March 28

FUNDING

**$23B:** CoreWeave an AI data center company, has the largest U.S. tech IPO since 2021, raising $1.5 billion and valuing the company at $23 billion.

March 31

FUNDING

**$300B:** OpenAI raises $40 billion at a $300 billion post-money valuation.

May 13

INVESTMENT

**$5B:** AWS and HUMAIN announce a $5 billion AI infrastructure deal to accelerate AI adoption in Saudi Arabia and globally.

May 21

ACQUISITION

**$6.5B:** OpenAI acquires IO, the AI hardware startup founded by Jony Ive, for $6.5 billion to develop a new generation of consumer AI devices.

2025

June 2

ACQUISITION

**Watsonx AI:** IBM acquires the AI startup Seek AI to launch Watsonx AI Labs, an AI accelerator and innovation hub in New York City.

---

July 9

MILESTONE

**$4T:** Nvidia becomes the first public company worth $4 trillion.

---

July 11

DEAL

**$2.4B:** Google hires key staff from AI code-generation startup Windsurf and agrees to pay $2.4 billion in license fees to use some of Windsurf's technology on a nonexclusive basis.

---

July 15

FUNDING

**$12B:** Thinking Machines Lab an AI company founded by Mira Murati and other former OpenAI researchers, raises a $2 billion seed round at a $12 billion valuation.

---

September 2

FUNDING

**$183B:** Anthropic raises $13 billion in Series F funding at a $183 billion post-money valuation.

---

September 10

INVESTMENT
PARTNERSHIP

**$300B:** OpenAI signs a $300 billion, five-year cloud contract with Oracle, beginning in 2027. Oracle will provide 4.5 gigawatts of computing capacity for OpenAI's Stargate data center initiative.

---

October 27

FUNDING

**$10B:** Mercor which connects AI labs with domain experts for training their foundation AI models, raises $350 million Series C at a $10 billion valuation, making their founders, both 22 years old, the youngest ever self-made billionaires.

---

November 10

FUNDING

**$2.1B:** Gamma a startup that creates AI-generated presentations, websites, and social media posts, announces a $68 million Series B round at a $2.1 billion valuation led by Andreessen Horowitz.

2025

November 11

INVESTMENT

**$6.4B:** Google announces it will invest $6.4 billion in cloud infrastructure in Germany from 2026–29 to expand its data center capacity there.

November 14

FUNDING

**$29.3B:** Anysphere which sells the popular AI coding assistant Cursor, raises $2.3 billion at a $29.3 billion valuation.

November 14

INVESTMENT

**$40B:** Google announces a $40 billion investment in Texas data centers and AI infrastructure through 2027.

November 20

FUNDING

**$5.6B:** Physical Intelligence a robotics AI startup building general-purpose foundation-model "brains" for robots, raises $600 million led by CapitalG at a $5.6 billion valuation.

December 9

INVESTMENT

**$17.5B:** Microsoft pledges a $17.5 billion investment in India's AI Infra.

December 10

INVESTMENT

**1M:** Amazon announces it will invest over $35 billion in India by 2030 to expand AI and logistics, targeting 1 million new jobs and $80 billion in seller exports..

December 22

ACQUISITION

**$4.75B:** Alphabet says it will acquire Intersect for $4.75 billion in cash plus assumed debt to accelerate U.S. energy innovation and data center infrastructure build-out.

# 4.2 Investment and Infrastructure

The scale and direction of investment into AI provides a signal of the technology's broader economic trajectory. As AI systems become more capable and infrastructure-intensive, the capital required to develop and deploy them has expanded. Viewed alongside the broader trends discussed in other chapters of this report, these investment patterns capture not just market interest, but the rising cost of participating in the AI economy. This section examines those patterns across corporate infrastructure spending, startup funding activity, and the operational economics of AI companies themselves. The analysis draws primarily from Quid's database of AI-related investments, supplemented by publicly disclosed financial metrics from leading AI companies, as tracked by Epoch AI. The Quid investment data captures four categories of capital flowing into AI companies, including mergers and acquisitions, minority stake investments, private investment, and public offerings. Corporate investment, as used in this section, refers to the aggregate of all four. The private investment subsection that follows focuses more narrowly on private financing events, such as venture capital or private equity funding, directed at AI companies that have received over $1.5 million in funding since 2013. This subset represents a portion of total corporate investment.

## Corporate Investment Activity

Global corporate AI investment has grown over the past decade, accelerating within recent years, and further emphasizing how much AI has moved from an emerging technology to a key strategic priority. Across mergers and acquisitions, minority stakes, private investment, and public offerings, AI-related investment increased approximately fortyfold since 2013 (Figure 4.2.1). In 2025, total investment reached $581.69 billion, marking a 129.9% increase from the previous year. Private investments represented the largest share of activity with $344.66 billion, up 127.5% from 2024. Mergers and acquisitions showed similar signs of growth, rising 132.6% year over year. Though the composition of investment varies year to year, it is clear that organizations are committing growing sums of capital to strengthen their AI capabilities and position.

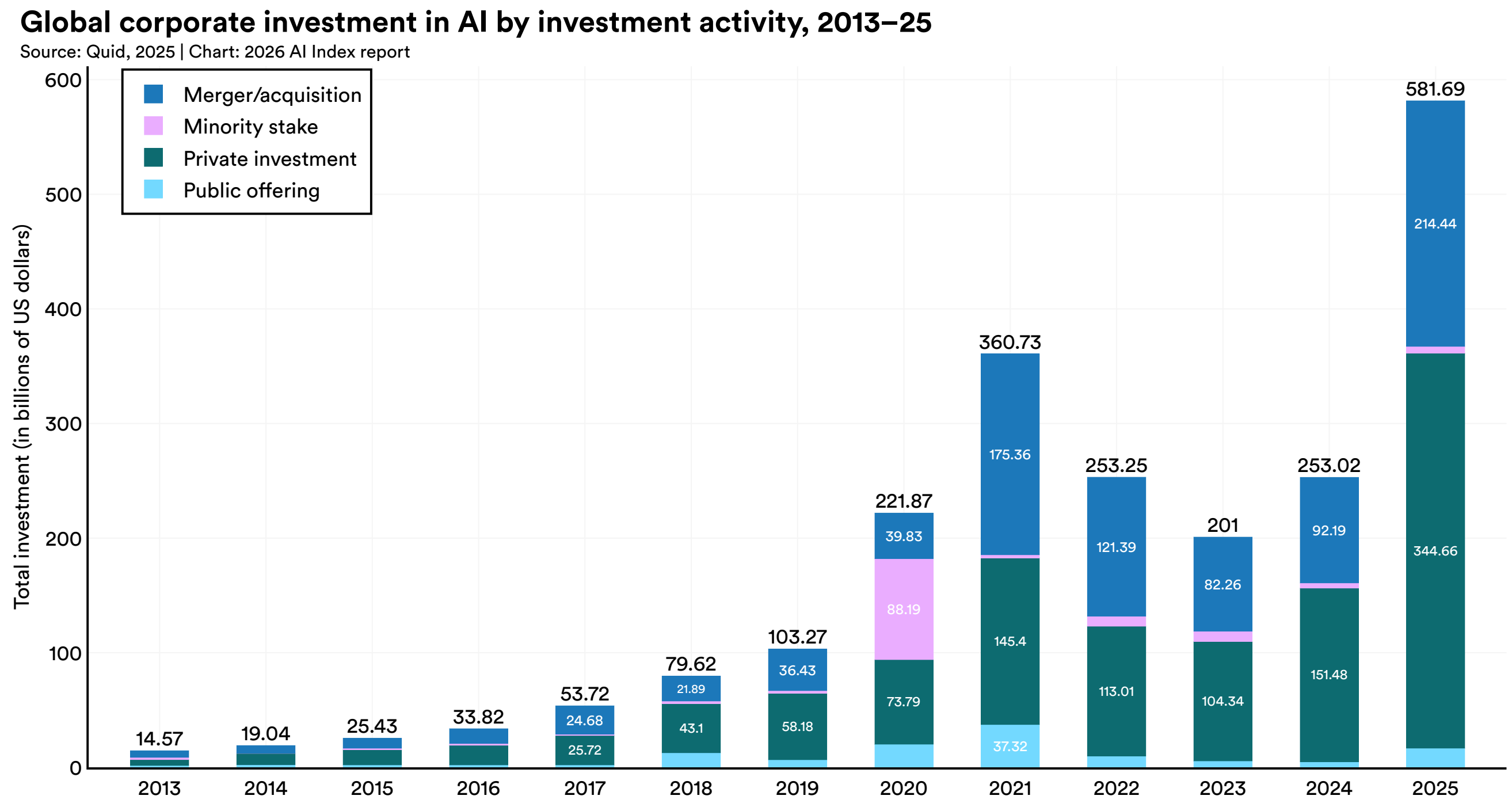

Figure 4.2.1

## Private Investment Activity

Within the broader investment landscape, private investment data, covering AI and ML companies with over $1.5 million in funding since 2013, offers a granular view into which firms are being funded and where that funding is concentrated. In 2025, global private investment in AI reached $344.7 billion, a 127.5% increase over the previous year (Figure 4.2.2). Generative AI companies accounted for $170.9 billion of that total, representing nearly half of all private investment and an increase of over 200% from 2024 (Figure 4.2.3).

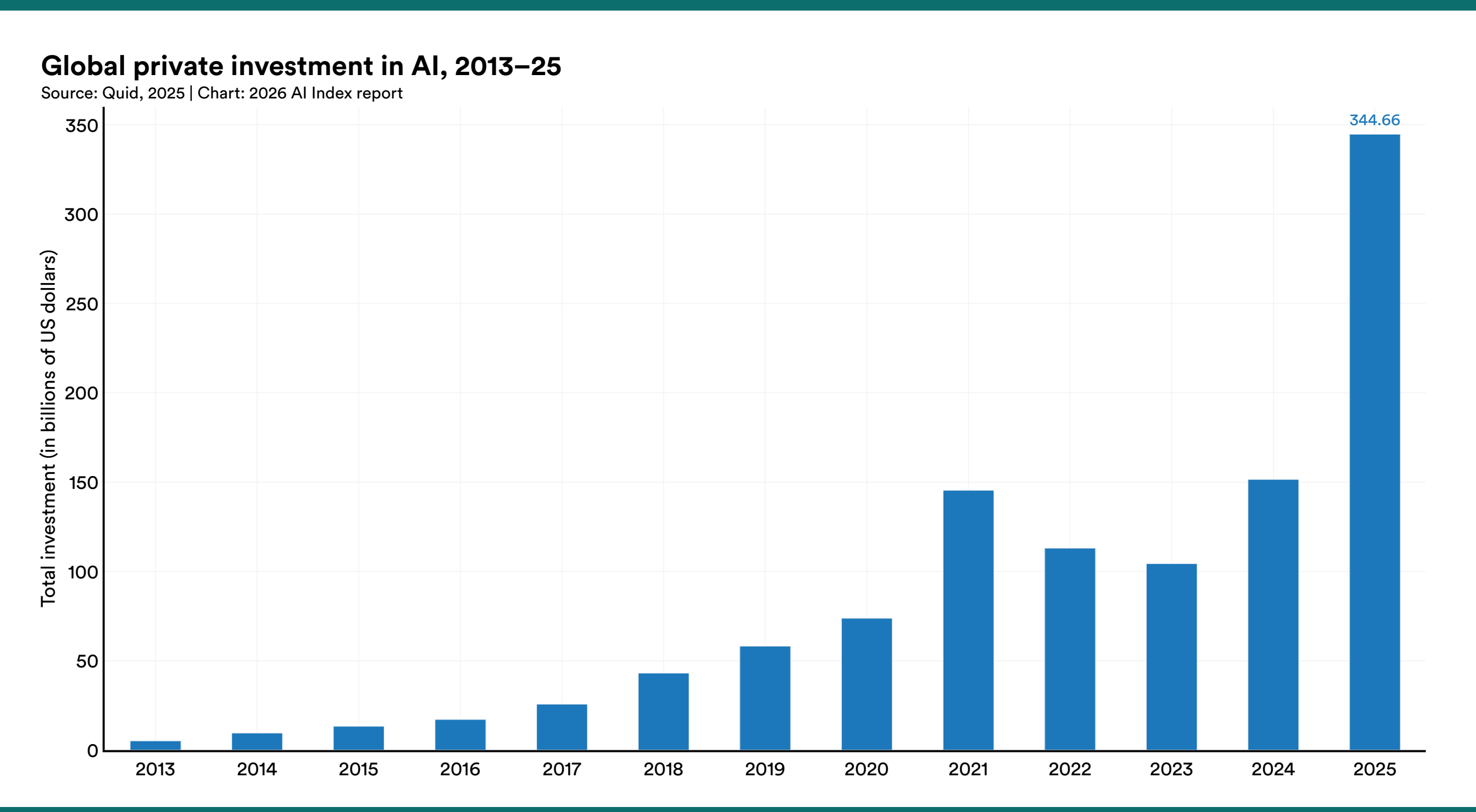

Figure 4.2.2

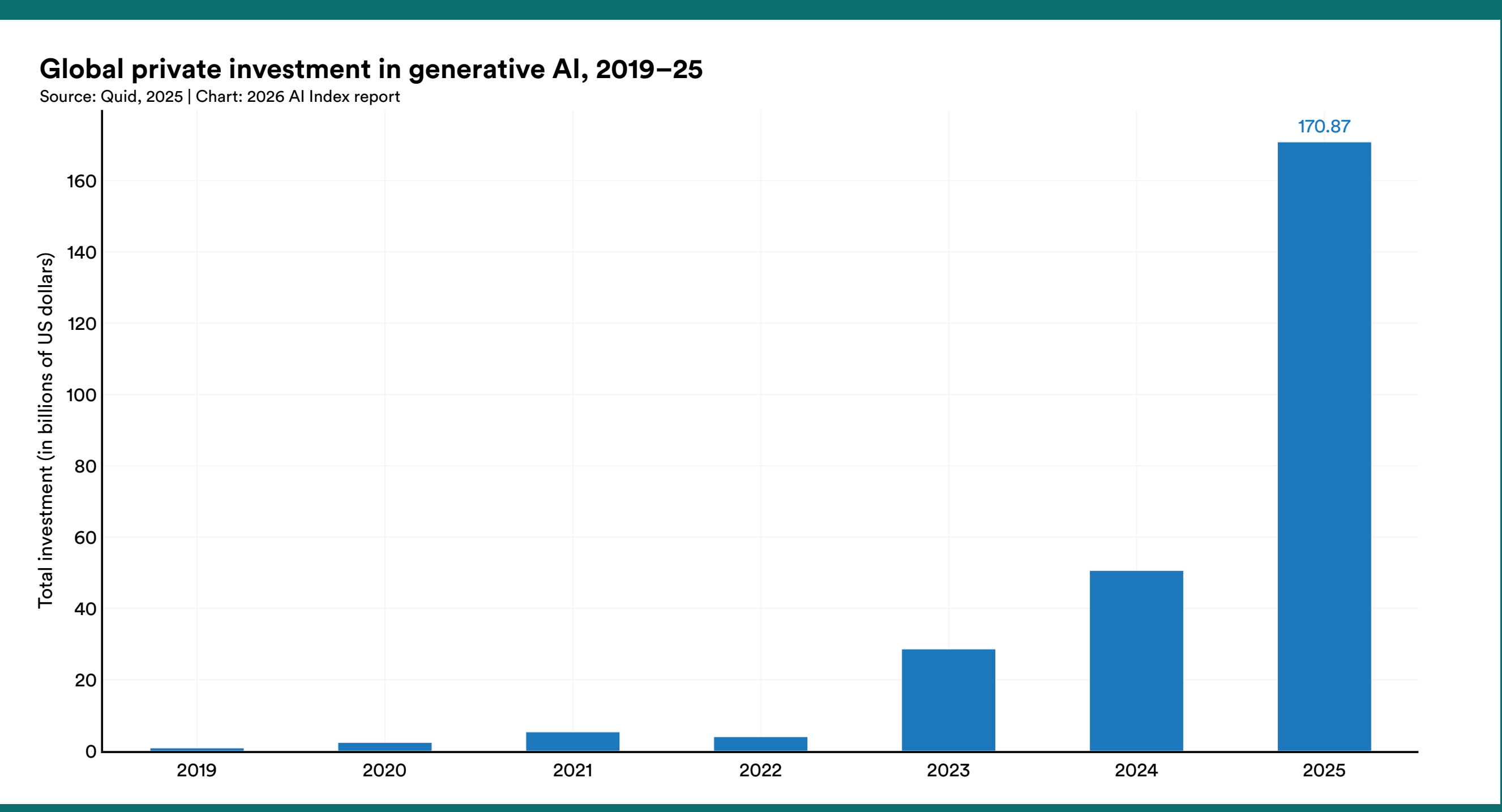

Figure 4.2.3

The private investment market is expanding in breadth but even more so in concentration. While the absolute number of newly funded AI companies has grown in 2025 (70.8% year-over-year increase), distribution of capital has dropped and the majority of investment dollars flow through a small number of deals (Figures 4.2.4-4.2.7). Compared to 2024, the average private AI investment event in 2025 increased 46% to $66.5 million. Investment activity increased across all funding sizes, but the strongest growth was at the upper end of that distribution, with 28 events exceeding $1 billion, up from 15 in 2024. The timeline of Section 4.1 notes several of these large funding rounds, including OpenAI's $40 billion raise and Anysphere's $2.3 billion round at a $29.3 billion valuation, highlighting the increasing skew in the funding landscape.

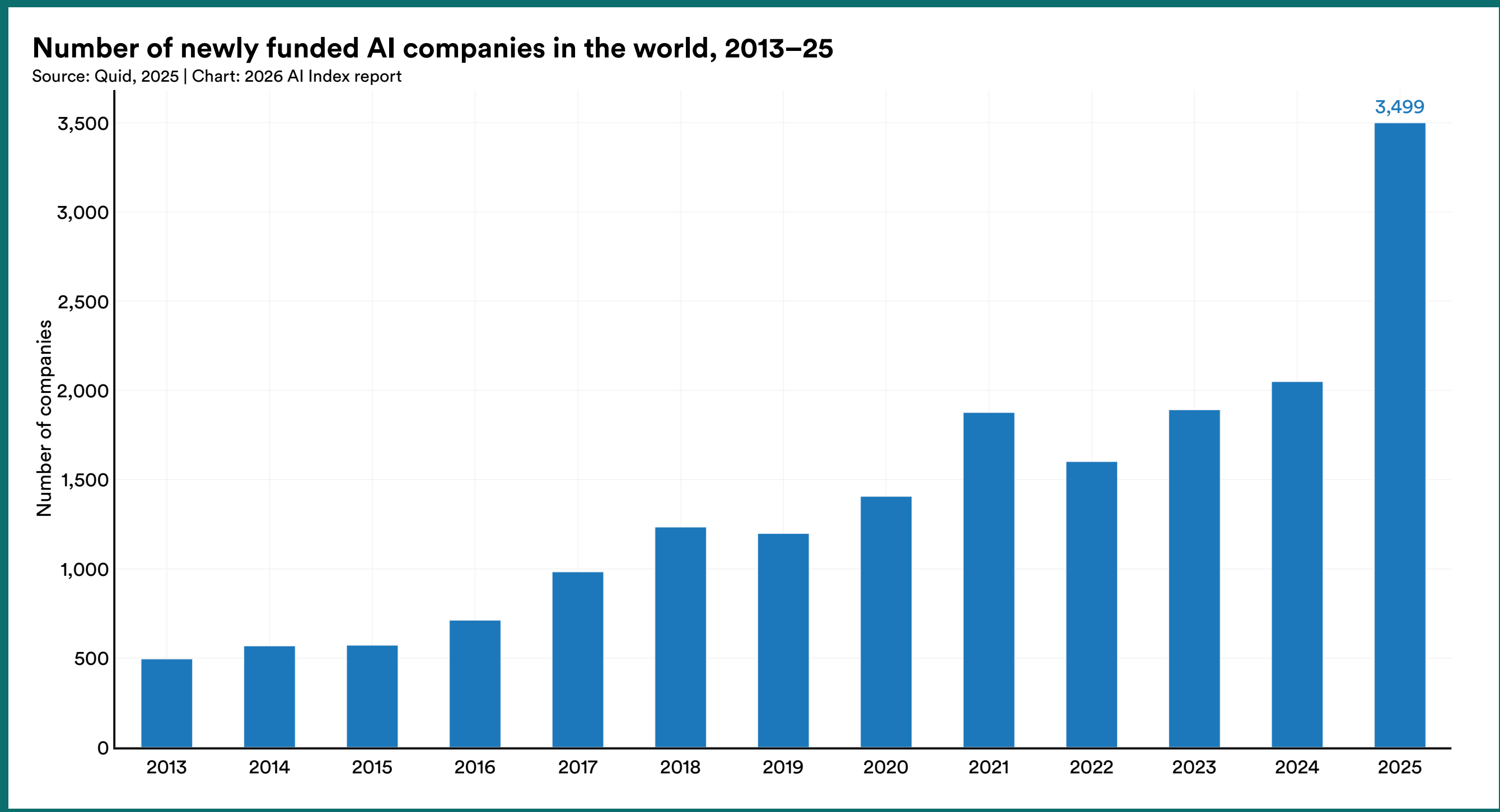

Figure 4.2.4

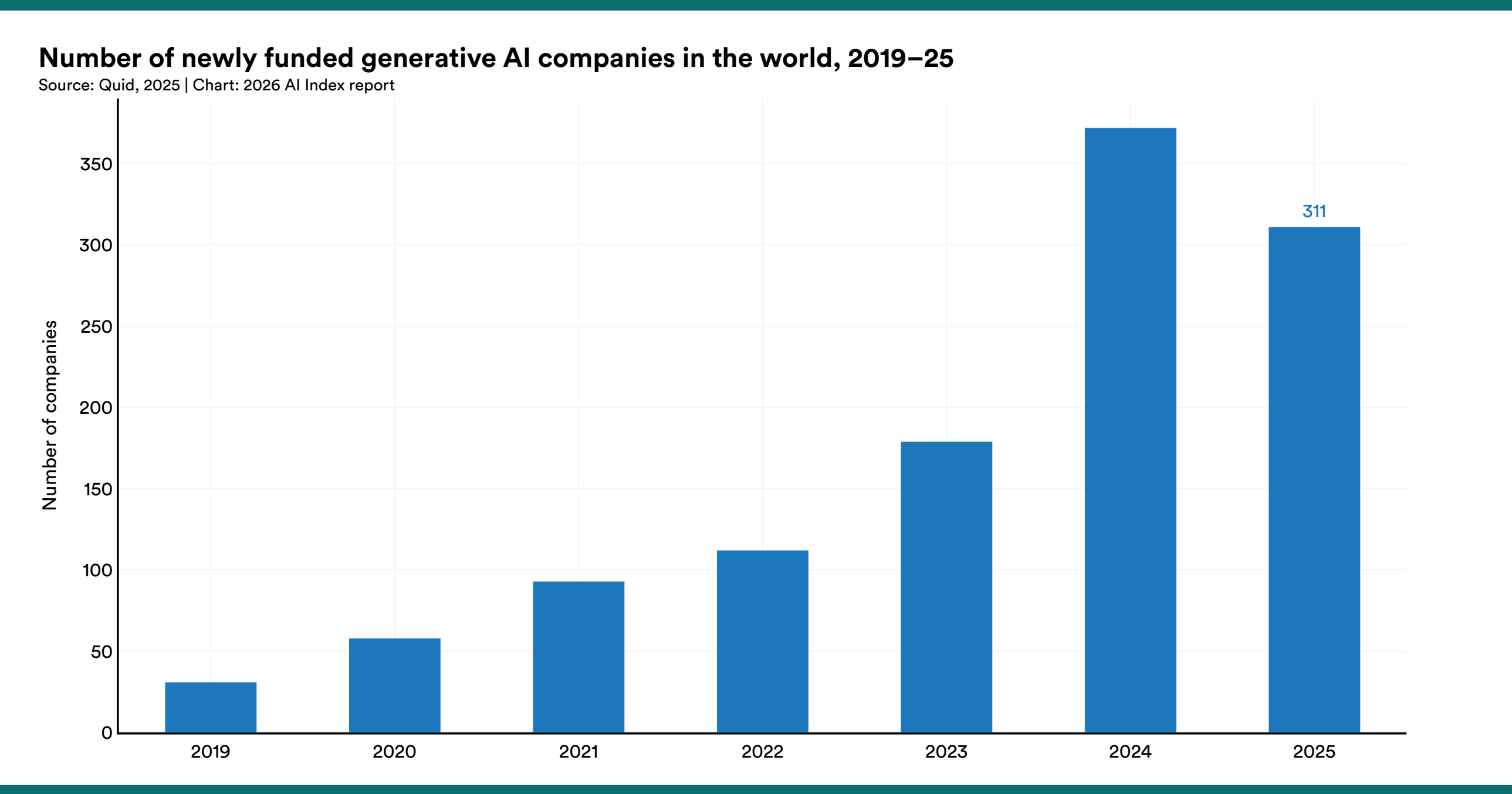

Figure 4.2.5

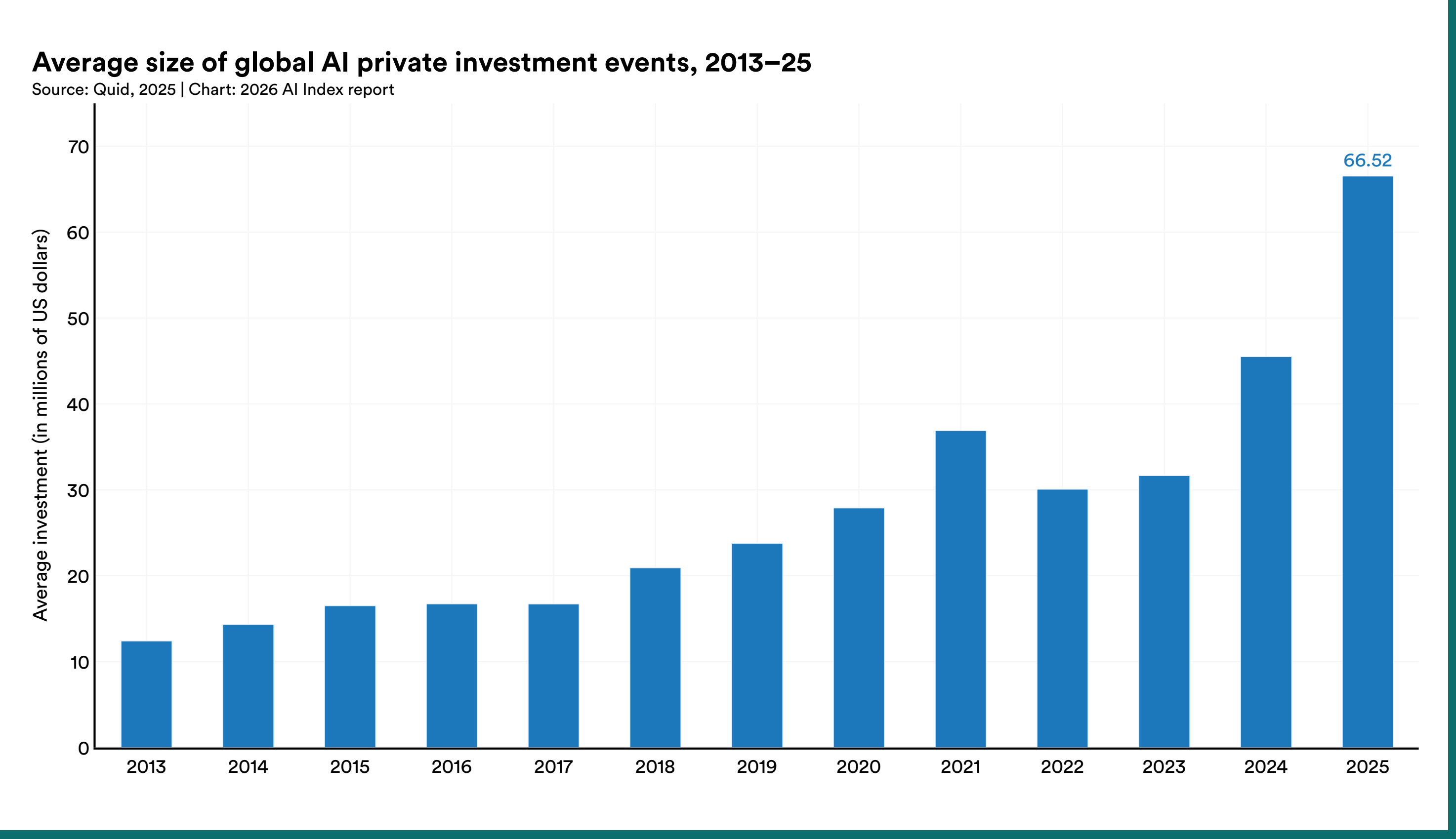

Figure 4.2.6

## Global AI private investment events by funding size, 2024 vs. 2025

Source: Quid, 2025 | Table: 2026 AI Index report

| Funding size (US dollars) | 2024 | 2025 |
|---|---|---|
| Over 1 billion | 15 | 28 |
| 500 million – 1 billion | 20 | 30 |
| 100 million – 500 million | 146 | 286 |
| 50 million – 100 million | 197 | 373 |
| Under 50 million | 2,951 | 4,464 |
| Undisclosed | 209 | 324 |
| Total | 3,538 | 5,505 |

Figure 4.2.7

## Geographic Distribution of Private Investment

As measured by both investment totals and the number of newly funded companies, private AI investment remains highly concentrated in a small number of countries. In 2025, the United States was the global leader with nearly $285.9 billion total invested, 23.1 times greater than the amount invested in the next highest country, China ($12.4 billion), and 48.5 times the amount invested in the United Kingdom ($5.9 billion) (Figure 4.2.8). This disparity was also seen in entrepreneurial activity, as the United States led with 1,953 newly funded AI companies in 2025, compared to 172 in the United Kingdom and 161 in China (Figure 4.2.9). In the United States, more than half of total private AI investment was generative AI-related ($163.6 billion), while the combined investment by China and Europe was $4.7 billion (Figure 4.2.10). Since 2024, private AI investment in the United States increased 160.2%, compared to an increase of 32.2% in China and 7.2% in Europe (Figure 4.2.11).

**Global private investment in AI by geographic area, 2025**
Source: Quid, 2025 | Chart: 2026 AI Index report

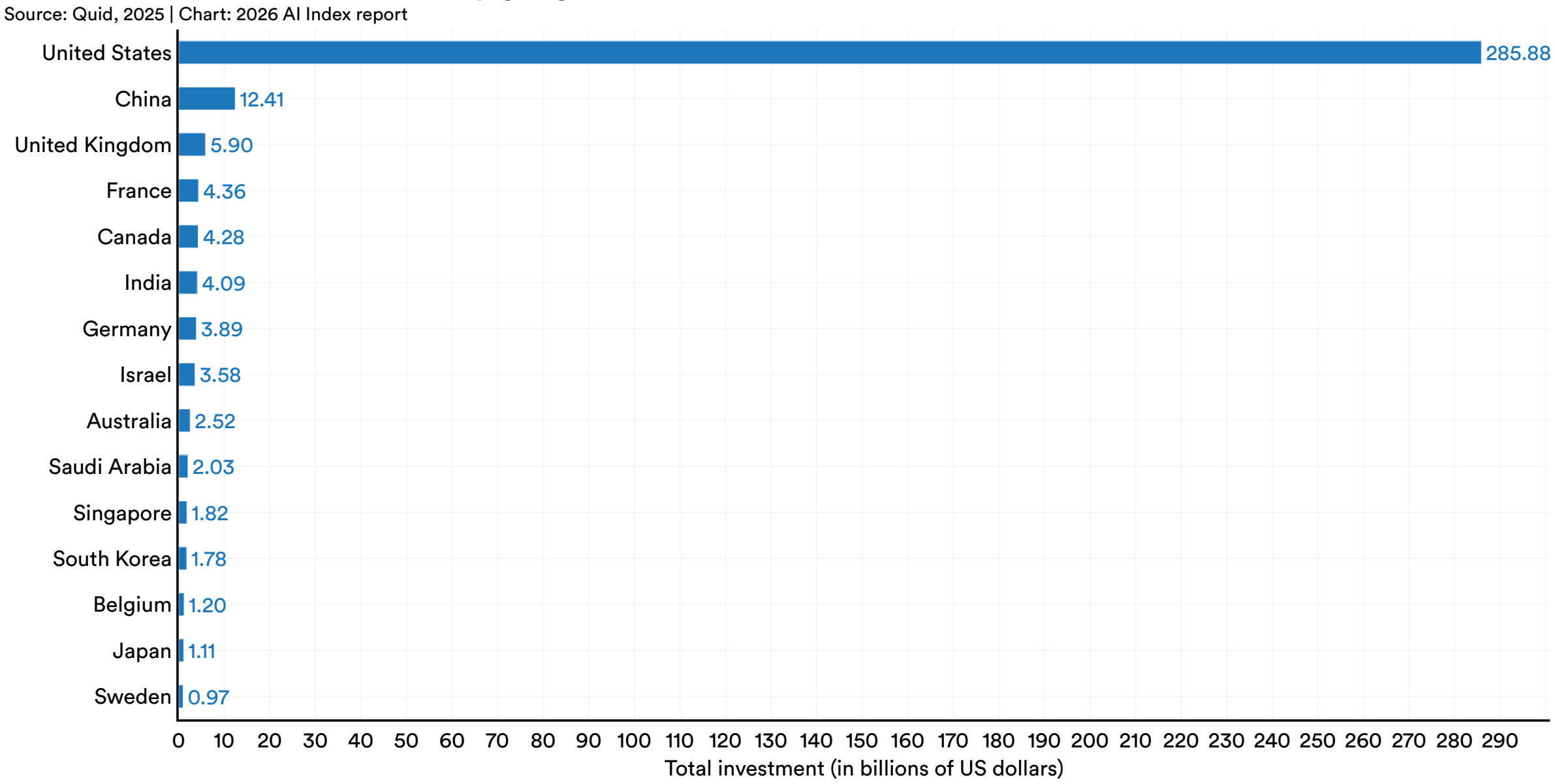

Figure 4.2.8

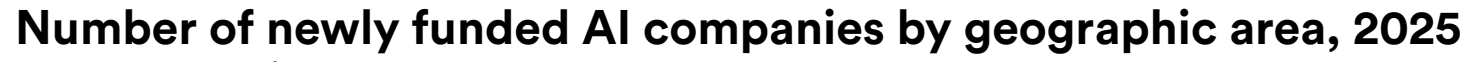

**Number of newly funded AI companies by geographic area, 2025**
Source: Quid, 2025 | Chart: 2026 AI Index report

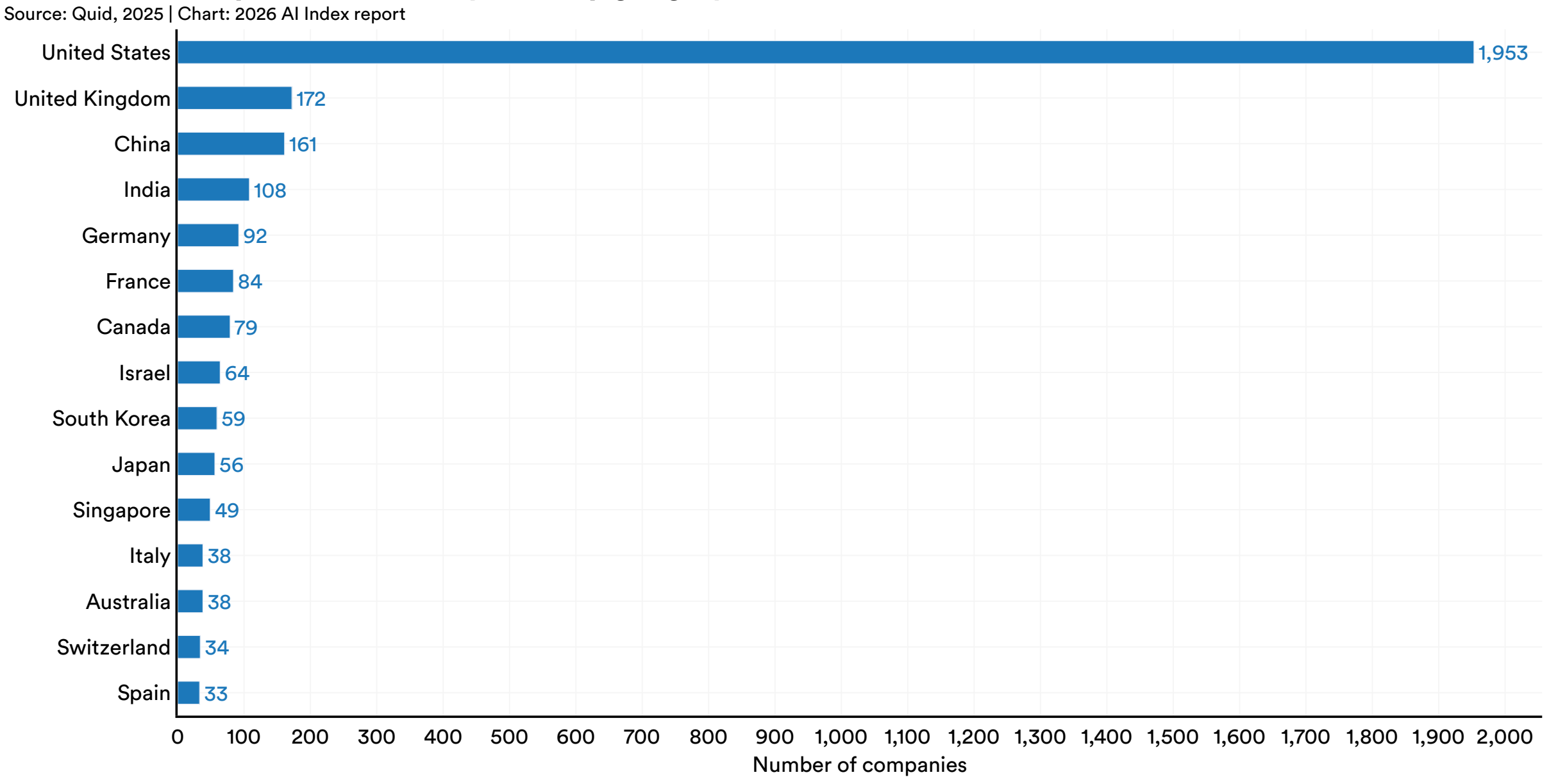

Figure 4.2.9

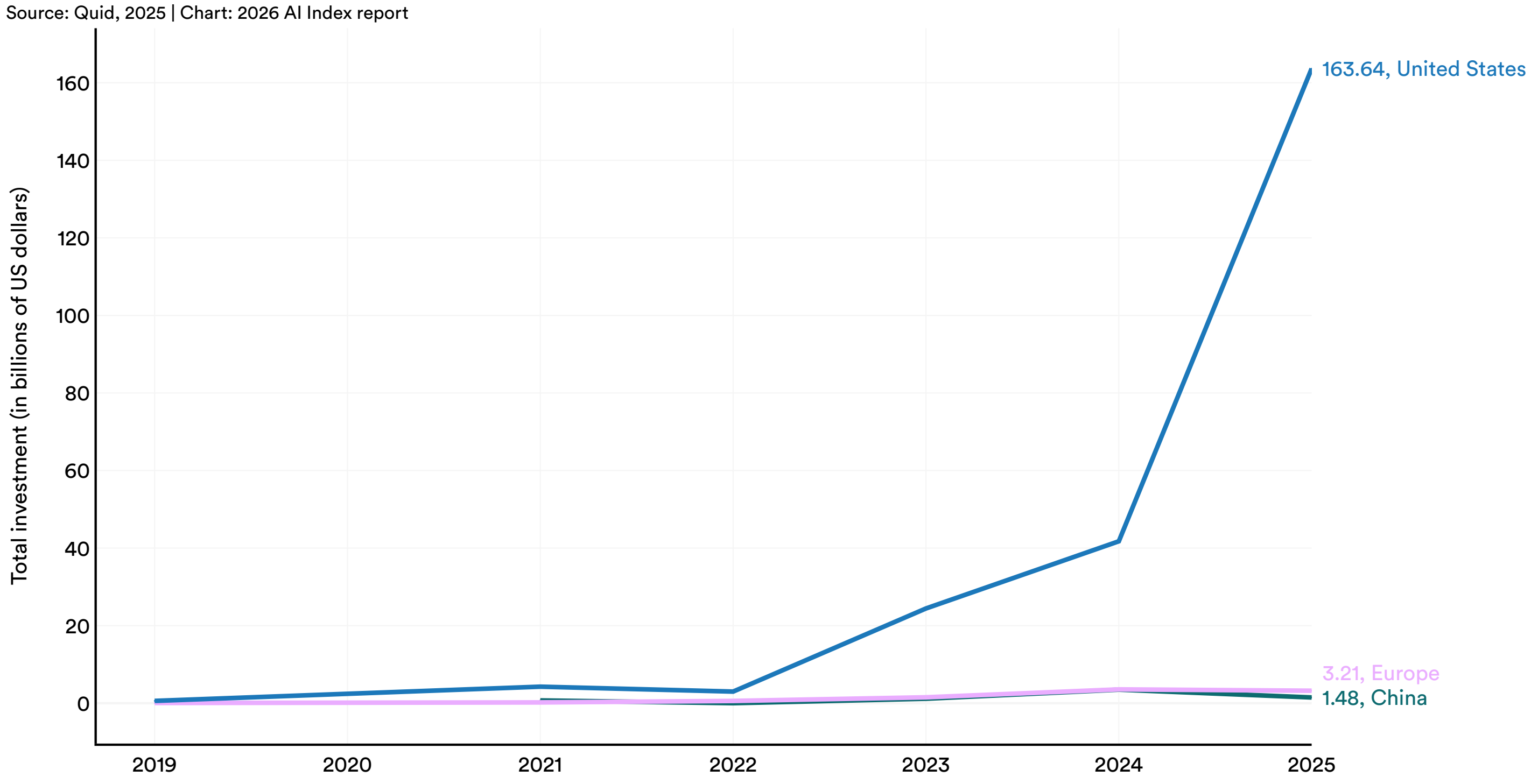
Global private investment in generative AI by geographic area, 2019–25
Source: Quid, 2025 | Chart: 2026 AI Index report

Figure 4.2.10

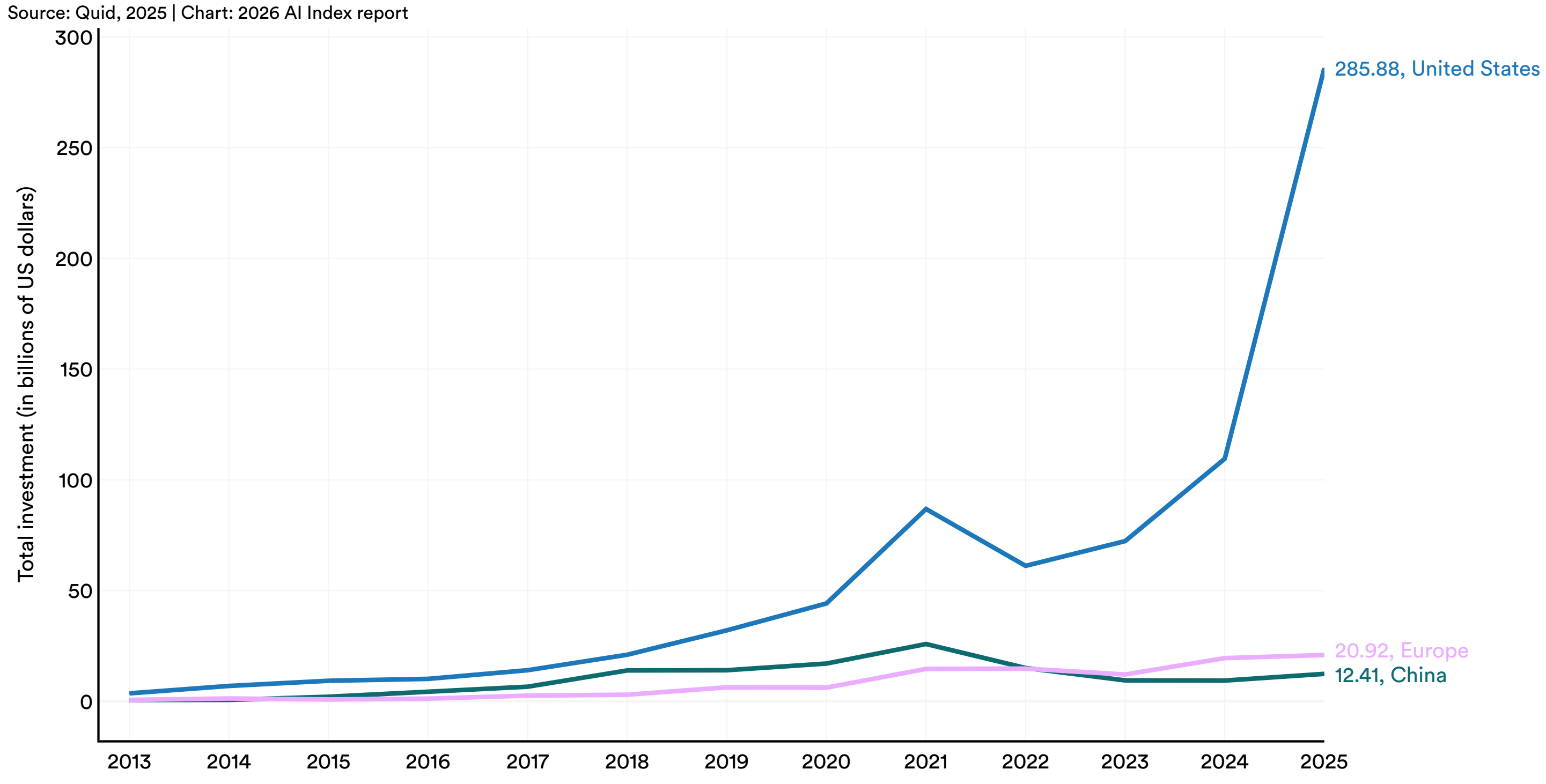
Global private investment in AI by geographic area, 2013–25
Source: Quid, 2025 | Chart: 2026 AI Index report

Figure 4.2.11

As noted earlier, the private investment figures in this section are drawn from Quid and do not account for government-backed funding in countries like China. For example, the Chinese government channels resources through government guidance funds, which are state-initiated investment funds that aim to both produce financial returns and further the government's strategic priorities (Beraja et al., 2024; Luong et al., 2021). Between 2000 and 2023, it was estimated that $912 billion of these funds were deployed across industries, with an estimated $184 billion allocated towards AI companies. Given this, comparisons based solely on private investment alone likely understate how much capital China is directing toward AI.

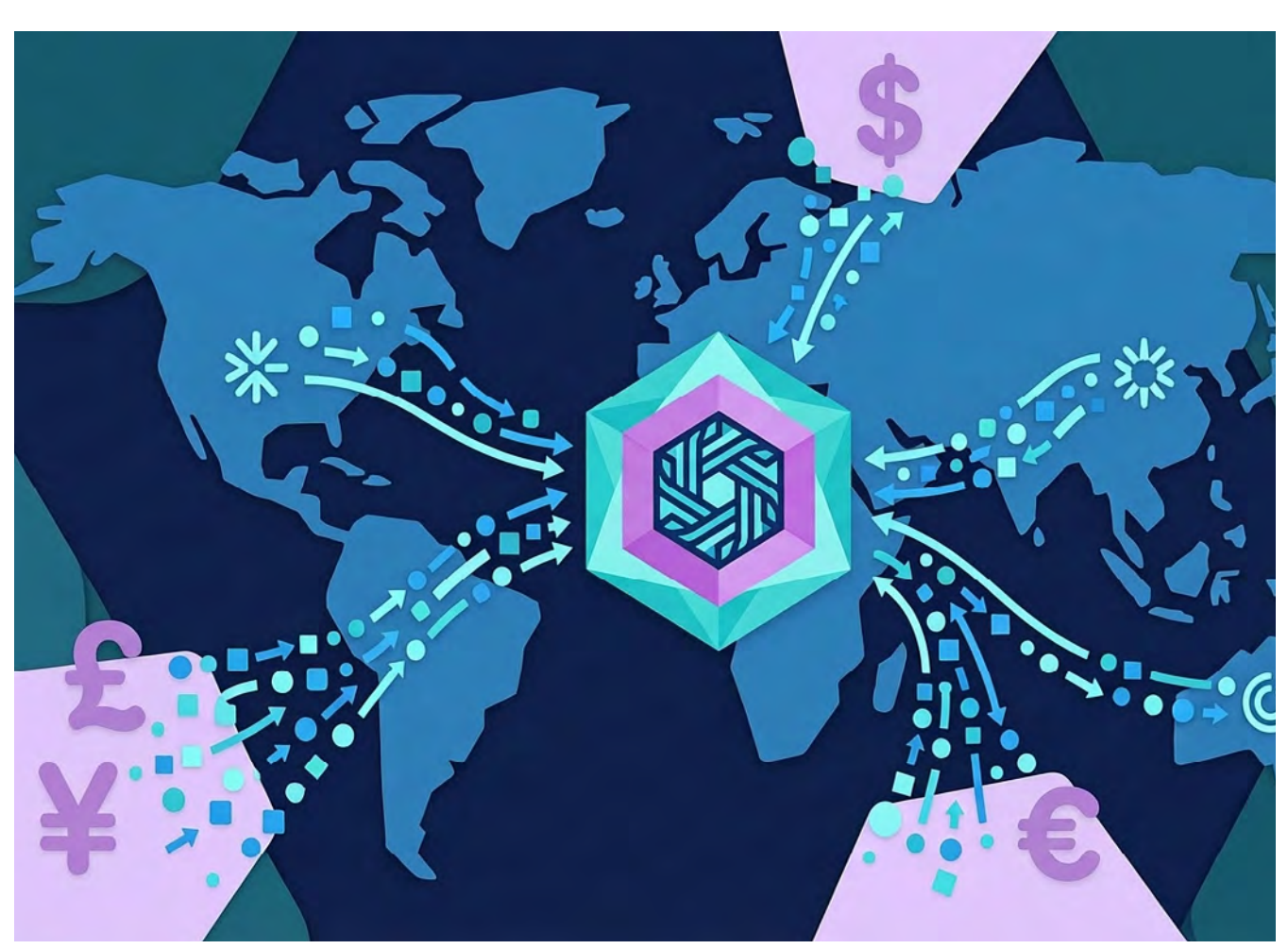

The trends in geographic concentration are also visible over a longer time horizon. Since 2013, the United States has attracted $757.3 billion in total private AI investment, far ahead of China at $131.8 billion (Figure 4.2.12). Other countries with notable cumulative investment totals include the United Kingdom ($34.1 billion), Canada ($19.6 billion), Israel ($18.5 billion), and Germany ($17.2 billion). Over the same period, the number of newly funded U.S. companies far exceeds other geographic areas, including five times that of China and 8.4 times the amount in the United Kingdom (Figures 4.2.13 and 4.2.14). The United States' growth rate continues to accelerate, with a 77.8% year-over-year increase in the number of funded AI startups.

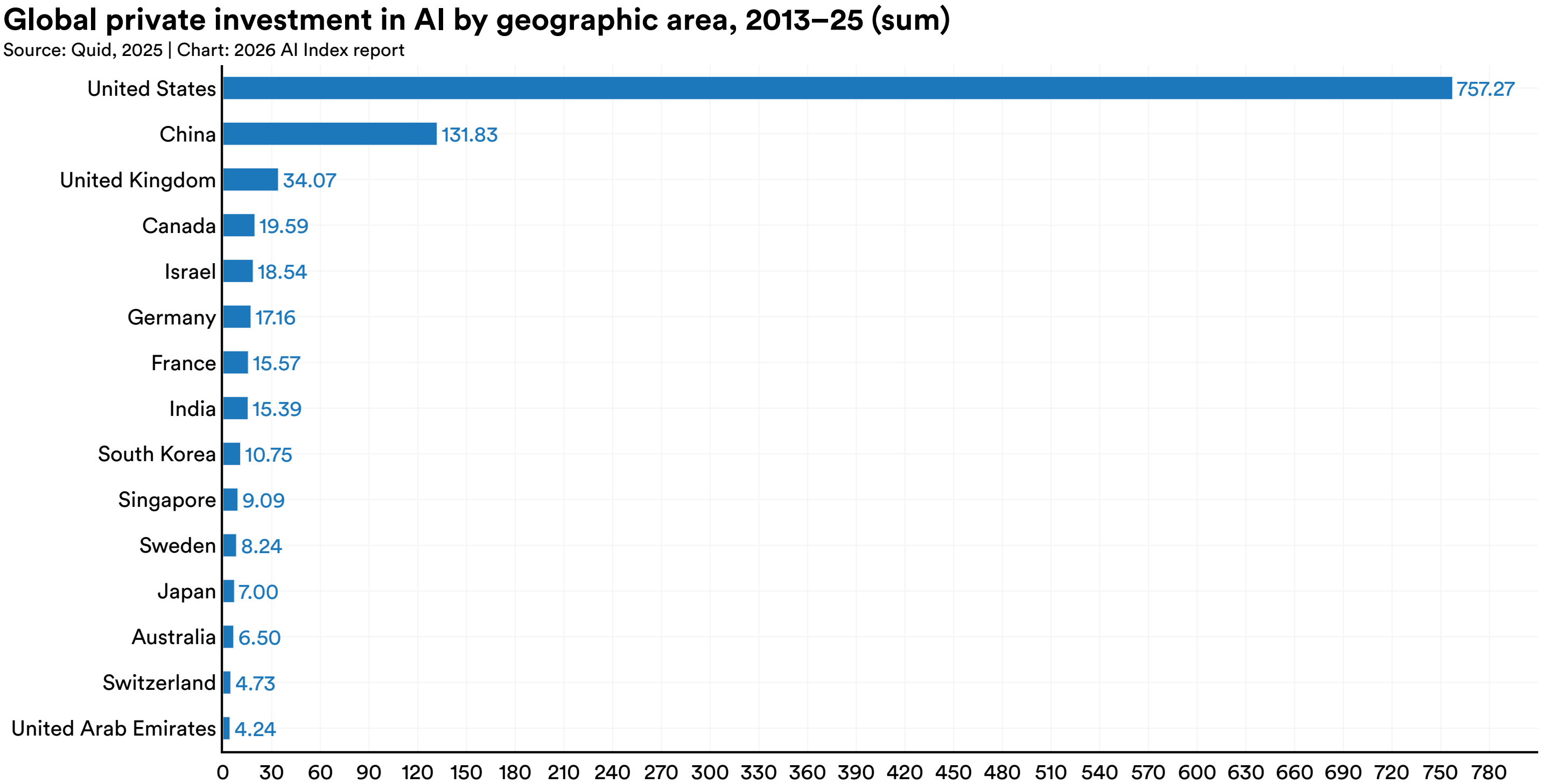

Figure 4.2.12

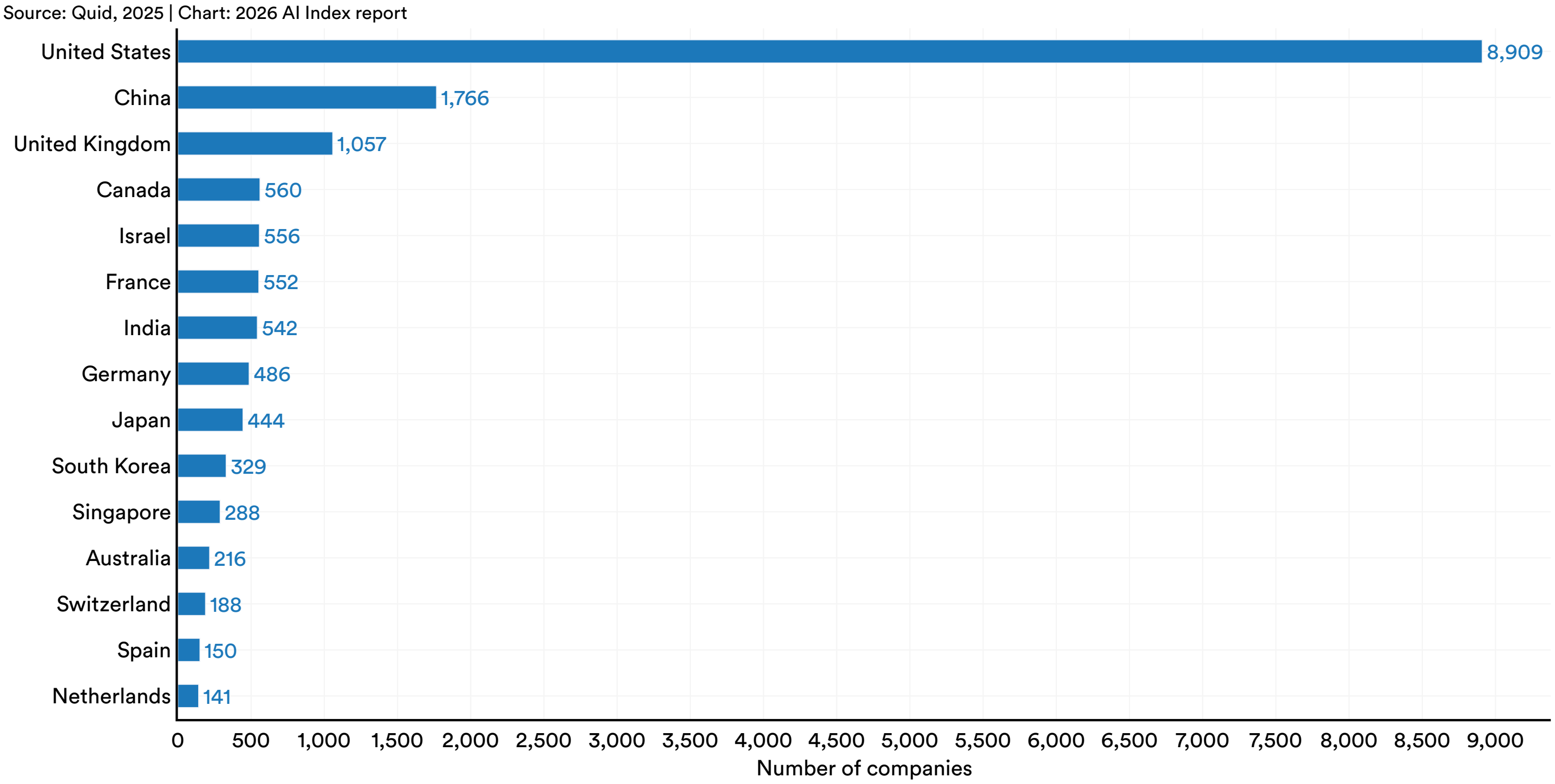
Number of newly funded AI companies by geographic area, 2013–25 (sum)
Source: Quid, 2025 | Chart: 2026 AI Index report

Figure 4.2.13

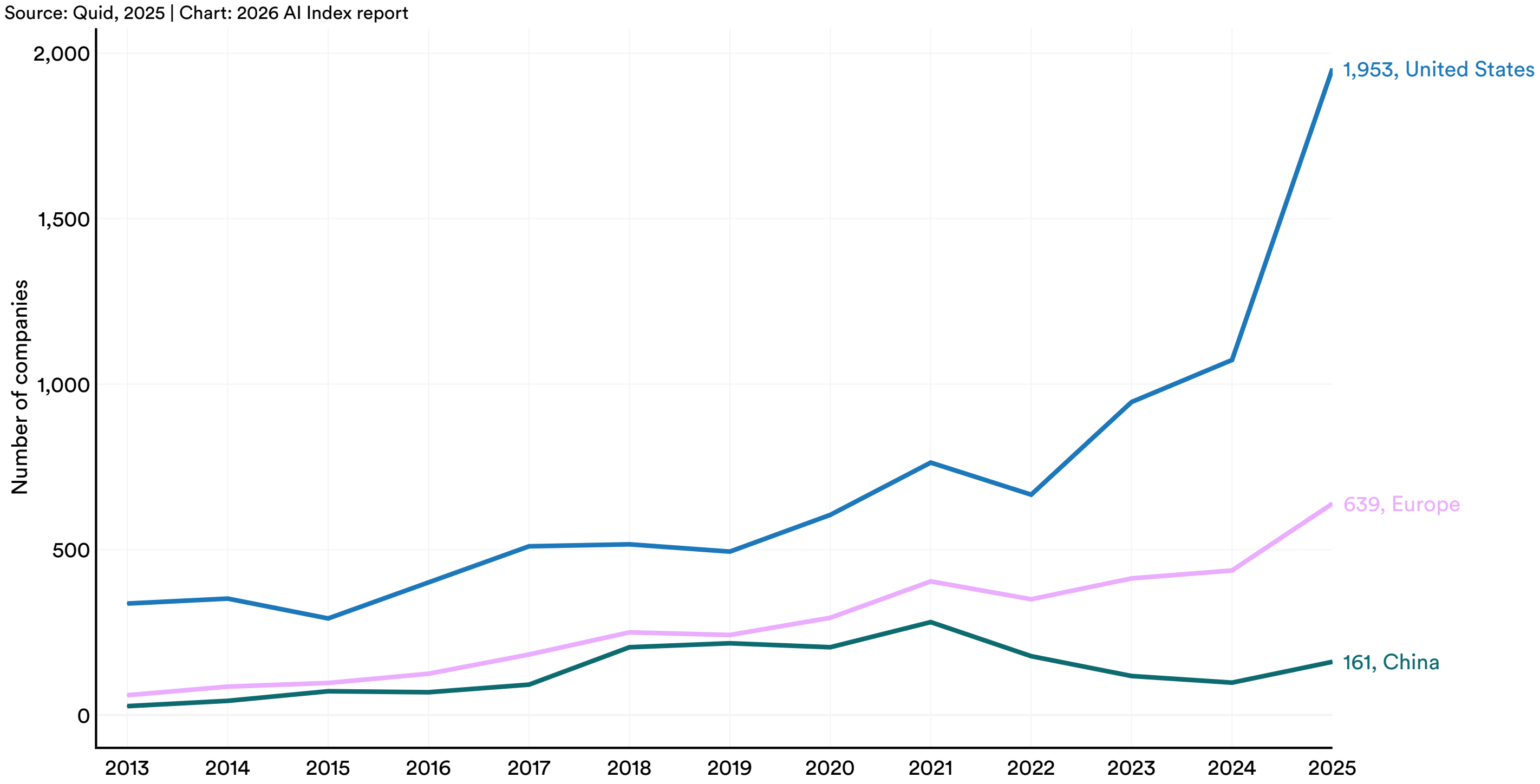
Number of newly funded AI companies by geographic area, 2013–25
Source: Quid, 2025 | Chart: 2026 AI Index report

Figure 4.2.14

Within the United States, funding and entrepreneurial activity is heavily concentrated in a small number of states (Figure 4.2.15). California accounted for $218 billion in 2025, representing over 75% of the national total. Colorado ($19 billion), New York ($13 billion), and Florida ($6 billion) tracked the next largest investments. More than half of all U.S. states received less than $100 million in private AI investment, and a few, including South Dakota, Oklahoma, Arkansas, and West Virginia, reported no mapped investment activity. The underlying data for these state-level figures is not exhaustive but the overall pattern is clear.

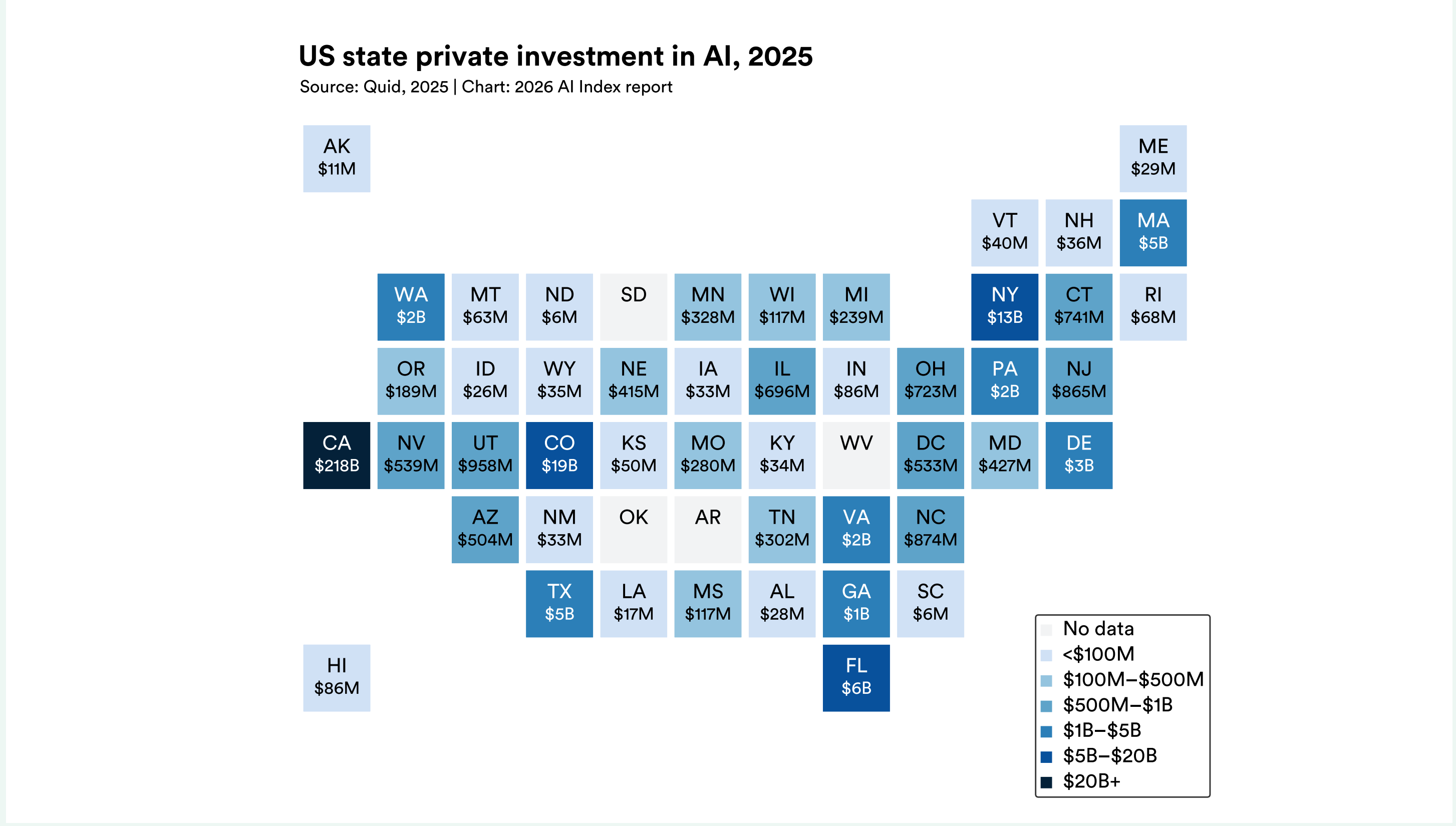

Figure 4.2.15

### Focus Areas of Private Investment

In 2025, the breakdown of private AI startup investment by focus area suggests that capital was directed more heavily toward segments closest to building and scaling AI systems. The category of AI infrastructure, models, research, and governance attracted the largest share of funding, reaching $143.2 billion (Figures 4.2.16 and 4.2.17). As this category combines several types of priorities, the trend is best interpreted as evidence of growing investment in the foundational layers of the AI ecosystem rather than a precise measure of any single one of those areas. In recent years, this category has experienced the steepest growth in investment compared with all other areas. Other focus areas have expanded as well, including data management and processing and Internet of Things (IoT), yet none approaches the scale of foundational infrastructure. Alongside 2025 trends in technical performance (Chapter 2), and research and development (Chapter 1), this suggests that investment is tracking the infrastructure demands of deploying increasingly capable AI systems.

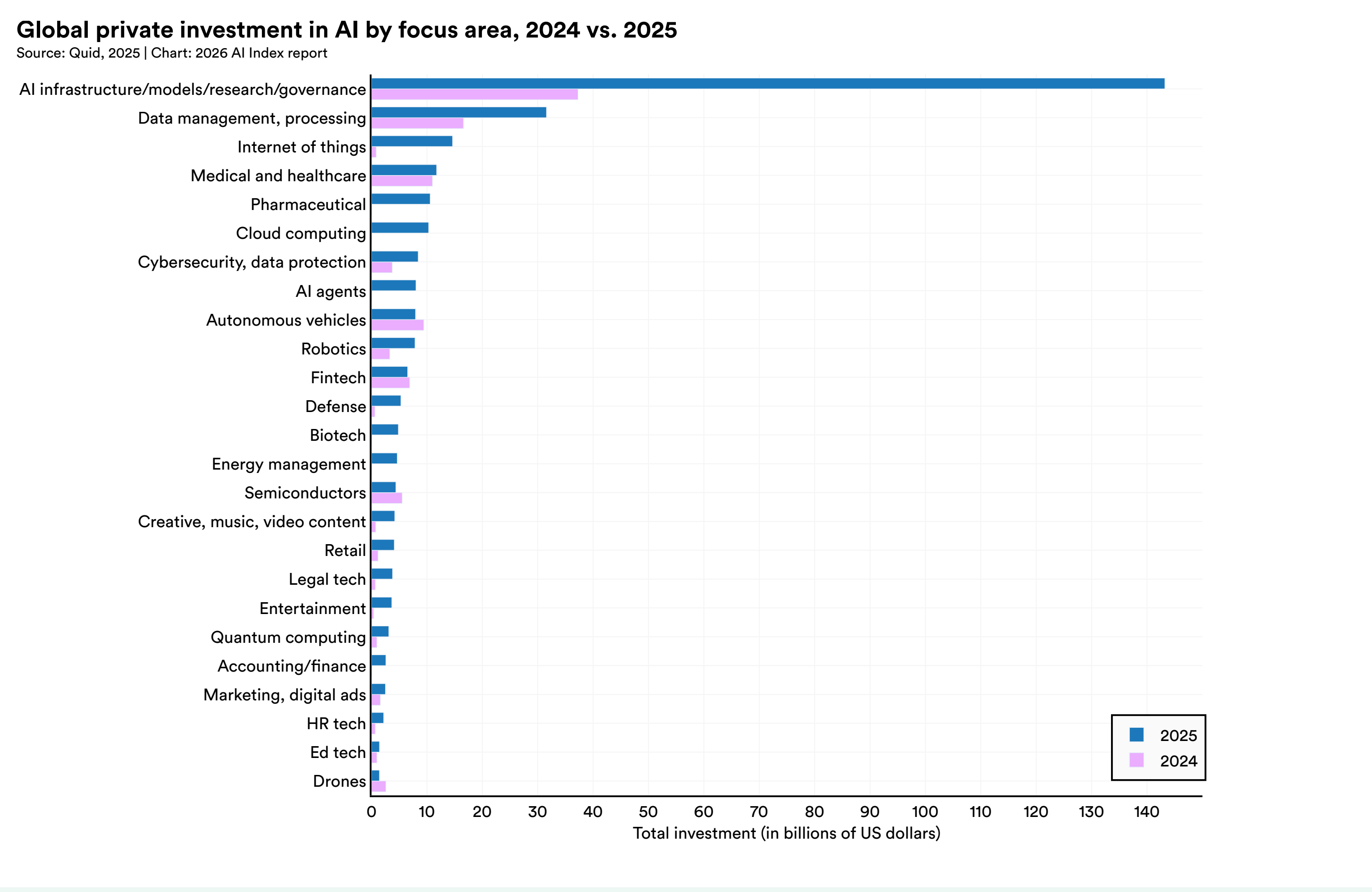

Figure 4.2.16

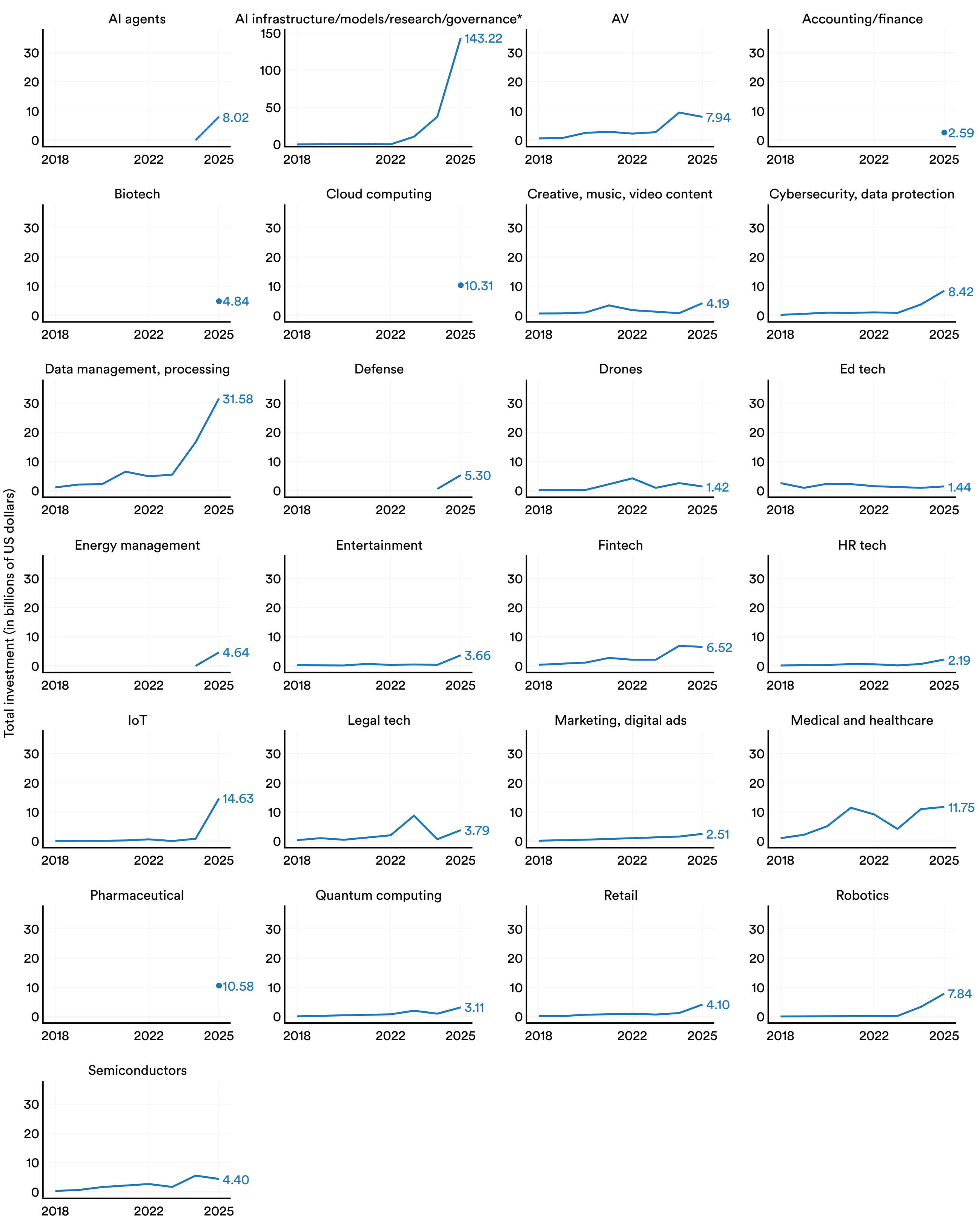
Global private investment in AI by focus area, 2018–25
Source: Quid, 2025 | Chart: 2026 AI Index report
Total investment (in billions of US dollars)
AI agents
8.02
AI infrastructure/models/research/governance*
143.22
AV
7.94
Accounting/finance
2.59
Biotech
4.84
Cloud computing
10.31
Creative, music, video content
4.19
Cybersecurity, data protection
8.42
Data management, processing
31.58
Defense
5.30
Drones
1.42
Ed tech
1.44
Energy management
4.64
Entertainment
3.66
Fintech
6.52
HR tech
2.19
IoT
14.63
Legal tech
3.79
Marketing, digital ads
2.51
Medical and healthcare
11.75
Pharmaceutical
10.58
Quantum computing
3.11
Retail
4.10
Robotics
7.84
Semiconductors
4.40
2018
2022
2025

Figure 4.2.17

## AI Company Economics

Investment patterns show where capital is flowing across the AI ecosystem, but the operating economics of frontier AI companies reveal how advances in technical performance and deployment translate into commercial scale, compute demand, and infrastructure buildout. Using publicly disclosed data tracked by Epoch AI, this section examines those revenue trajectories, ongoing compute expenses, and the infrastructure costs to support AI development.

### Revenue

Annualized revenue estimates for leading AI companies, including OpenAI, Anthropic, xAI, and Mistral AI, have grown quickly in recent years (Figure 4.2.18). These estimates are drawn from direct company statements or established media reporting from 2023 through 2025. They may differ from annual recurring revenue calculations, and the underlying data varies in reliability depending on source credibility and accounting practices. Therefore, these figures should be interpreted as directional rather than precise. The chart uses a logarithmic scale to accommodate the exponential growth pattern, meaning that a straight line represents consistent percentage growth rather than absolute growth. With those notes in mind, the overall dynamic points to a small set of frontier AI companies reaching meaningful revenue scale in a relatively short amount of time.

To contextualize this growth, a separate comparison places OpenAI's revenue trajectory alongside those of other high-growth companies in the years after crossing $1 billion in annual revenue (Figure 4.2.19). While Google remains the only company in the comparison set to be scaling toward $100 billion in annual revenue, OpenAI's early revenue growth outpaces that of Uber, Cheniere Energy, and Moderna over comparable time periods.

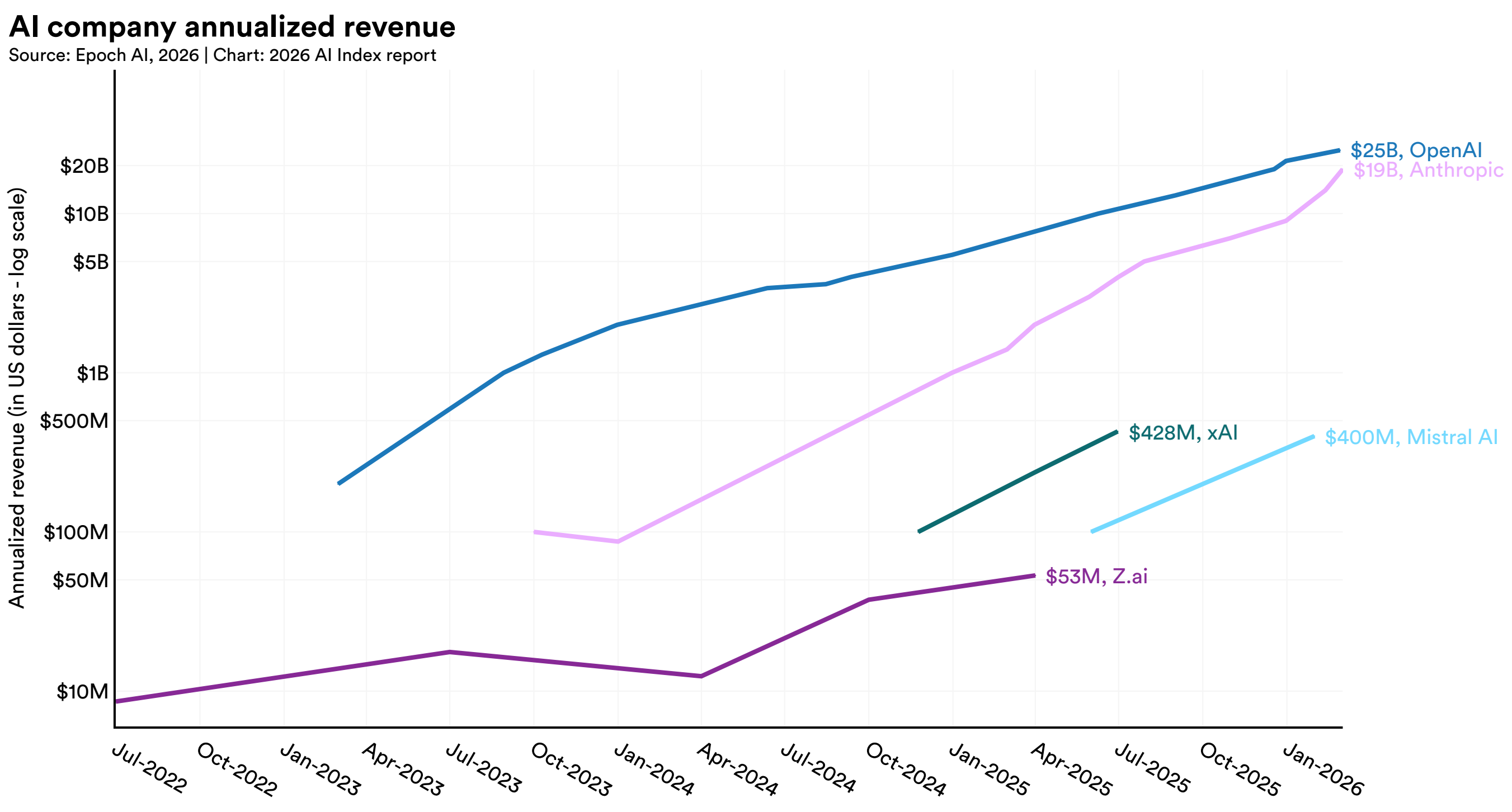

Figure 4.2.18

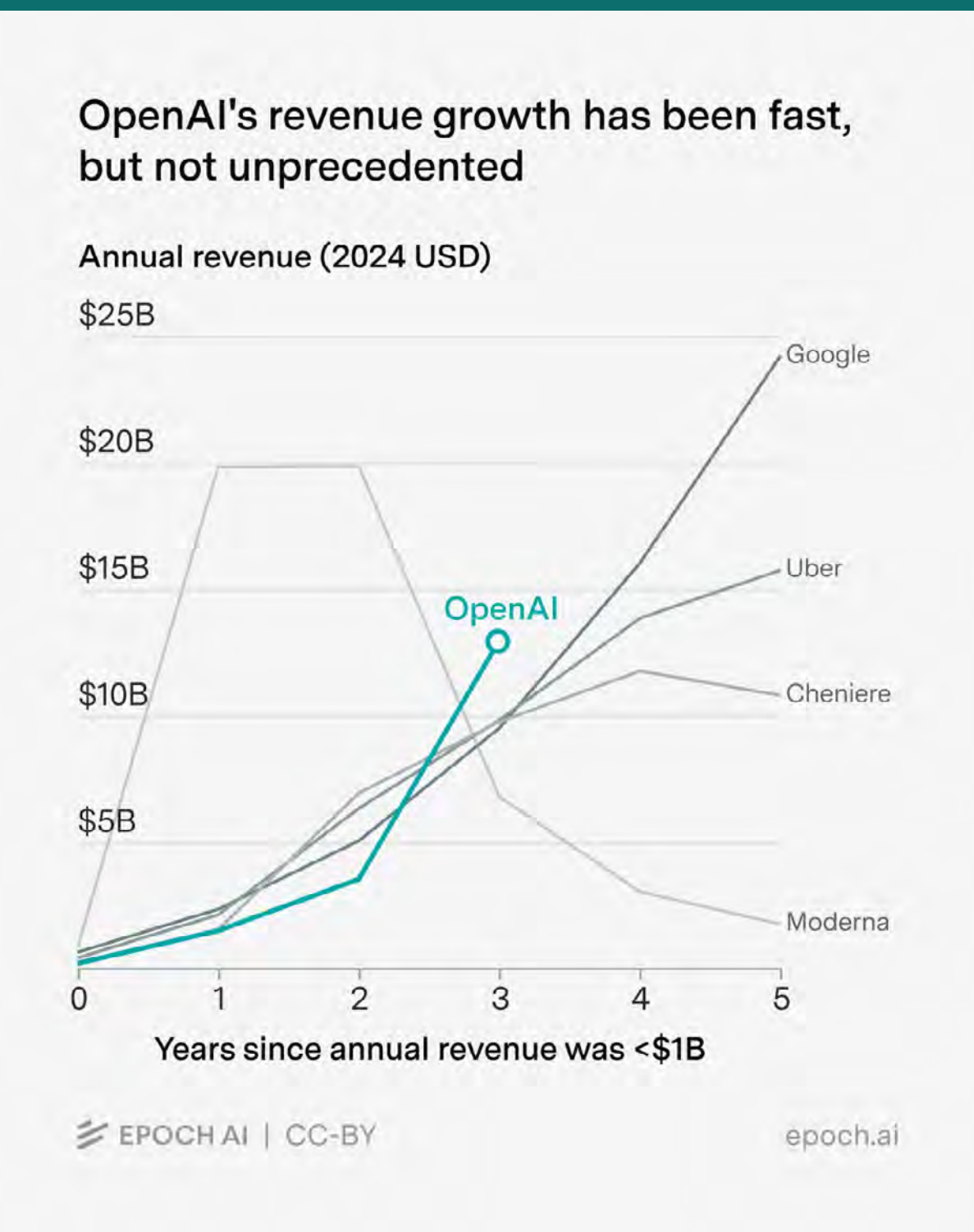

Figure 4.2.19

## Annual Compute Spend

The rapid revenue growth of leading AI companies has come with increasing compute costs. Reported annual compute spend, which largely reflects rented cloud capacity rather than owned data centers, offers a proxy for how much compute these companies procure each year to train and operate models at scale (Figure 4.2.20). OpenAI's reported compute spend increased significantly from 2024 to 2025, as did Anthropic's. The drive to meet growing commercial demand with increasingly capable systems means that the economics of frontier AI is tied to large-scale compute and its associated costs.

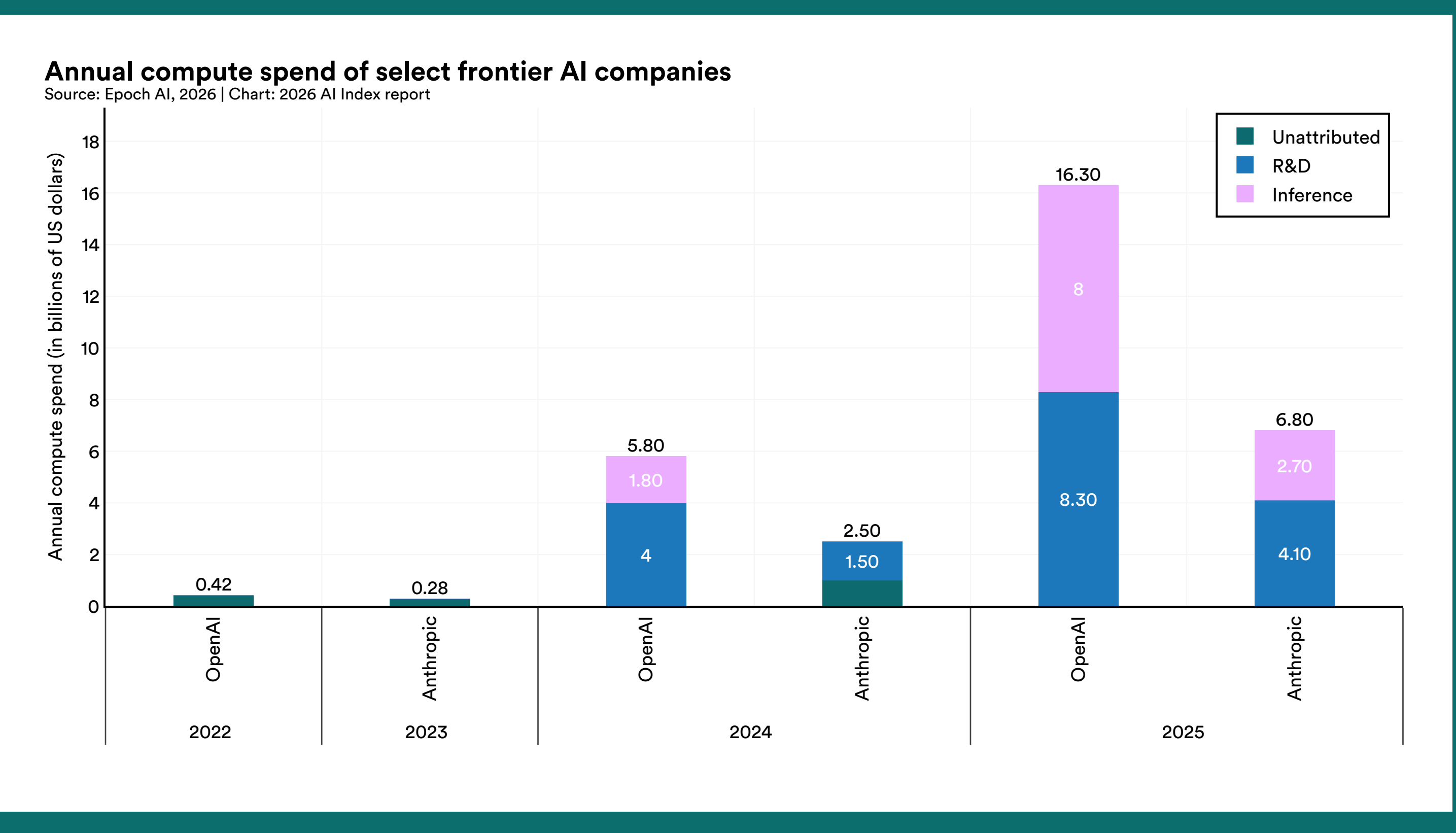

Figure 4.2.20

## Capital Expenditures

The infrastructure needed to support frontier AI is being financed not only by AI companies but also by the major cloud providers that lease them compute capacity. These providers have accelerated their own infrastructure investments to support the increasingly advanced AI models (Figure 4.2.21). In 2025, Google and Amazon led in total annual capital expenditures (capex), with Google reporting more than $150 billion in capex. This infrastructure investment is seen in the chapter's 2025 timeline, including the $100–$500 billion Stargate Project announced by OpenAI, SoftBank, Oracle, and others, as well as Google's $40 billion commitment to Texas data centers and Microsoft's $17.5 billion investment in AI infrastructure in India.

Hyperscaler's annual capex has more than doubled since ChatGPT's release

Hyperscaler annual capex (2025 and 2026 reflect estimates)

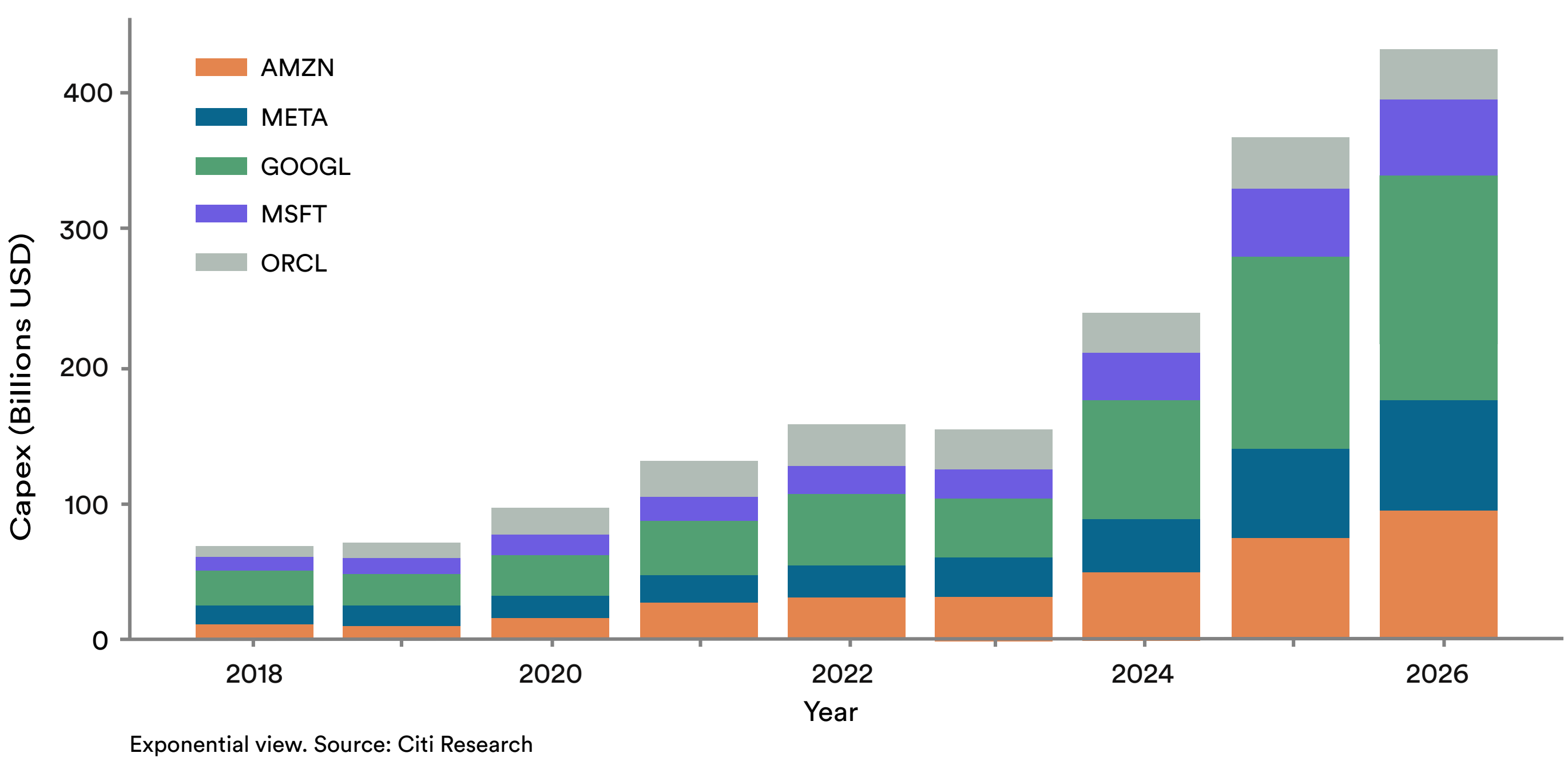

Exponential view. Source: Citi Research

Figure 4.2.21

**HIGHLIGHT:**

## What Is Generative AI Worth?

Investment, revenue and compute costs all measure the value of AI to the companies building and deploying it. They do not capture what the technology is worth to people using it. For most people, generative AI tools are free or close to it, making their economic value easy to undercount. Brynjolfsson et al. (2026) provide the first longitudinal estimates of that value, using online choice experiments conducted in 2025 (N=1,400) and early 2026 (N=2,000). Rather than measuring productivity effects, the study directly asked users' how much compensation they would accept to give up access to all generative AI tools for one month. This measure of "consumer surplus" is theoretically appropriate for goods that are largely free and already in the consumers' possession.

The study finds that total consumer surplus is estimated to have grown from $112 billion to $172 billion annually in the United States (4.2.22). This reflects the growing share of U.S. adults using generative AI, which increased from 48% to 56% (Bick et al., 2026) as well as a higher value per user. In particular, the average consumer surplus among U.S. generative AI users increased by 27% from $98 in 2025 to $125 by March 2026, while the median value per user tripled, from $3.40 to $11.40 over the same period. This increase in both adoption and per-user value is plausibly driven by a broadening and deepening of the capabilities of AI models.

This consumer surplus figure dwarfs estimated U.S. generative AI revenues, suggesting that the social returns from the technology far exceed the private returns captured by producers. This pattern is consistent with findings by Nordhaus (2004) that innovators historically capture only ~3% of total social returns from major technologies. The authors also find that usage frequency is the strongest individual-level predictor of surplus, followed by work use, number of different products used, and paid subscription status. Usage of generative AI for practical guidance, technical help, or information seeking are all associated with higher surplus.

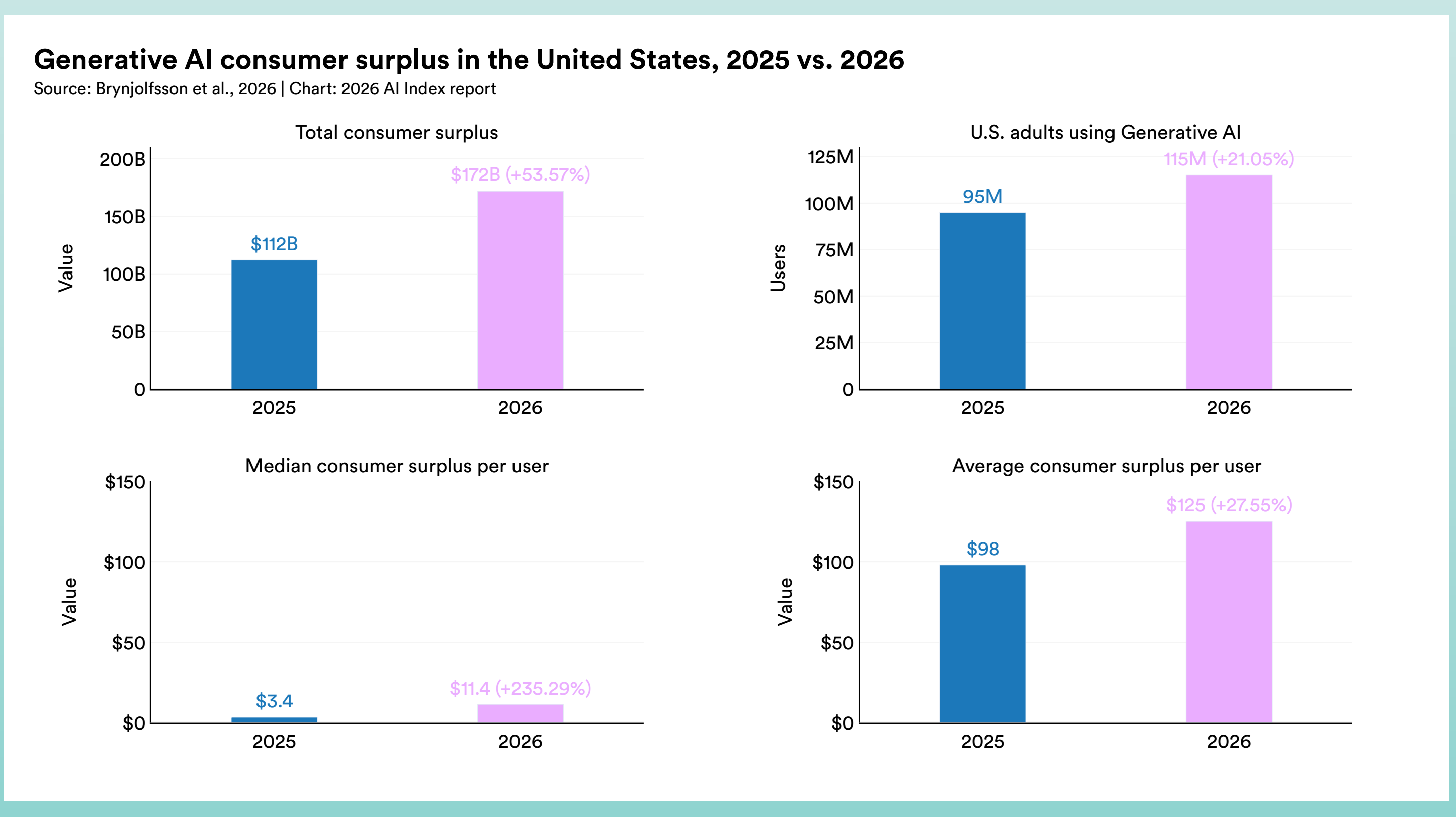

Figure 4.2.22

# 4.3 Corporate AI Adoption

The investment and infrastructure activity described earlier in this chapter establish the scale of resources being directed toward AI as well as the capacity being built to support it. This section examines how those investments translate into organizational use and reported business outcomes. Drawing on McKinsey & Company's annual State of AI surveys and other enterprise measures, the analysis traces the breadth and depth of adoption, and its associated benefits. As with other survey-based data in this chapter, the results are self-reported and should be viewed as directional rather than comprehensive.

## Industry Usage

In 2025, organizational adoption of AI continued to expand in both usage and function. A large majority of respondents reported that their organization uses AI in at least one business function, up to 88% in 2025 from 78% in 2024 (Figure 4.3.1). Over half of respondents reported three or more business functions leveraging AI. Use of generative AI mirrored that growth, with 79% of respondents reporting that their organizations regularly use generative AI in at least one business function, compared to 71% in 2024. This expanded adoption of AI was seen across all regions, though at different rates (Figure 4.3.2). China and Europe experienced higher year-over-year increases, with reported organizational AI use growing 13 and 11 percentage points, respectively.

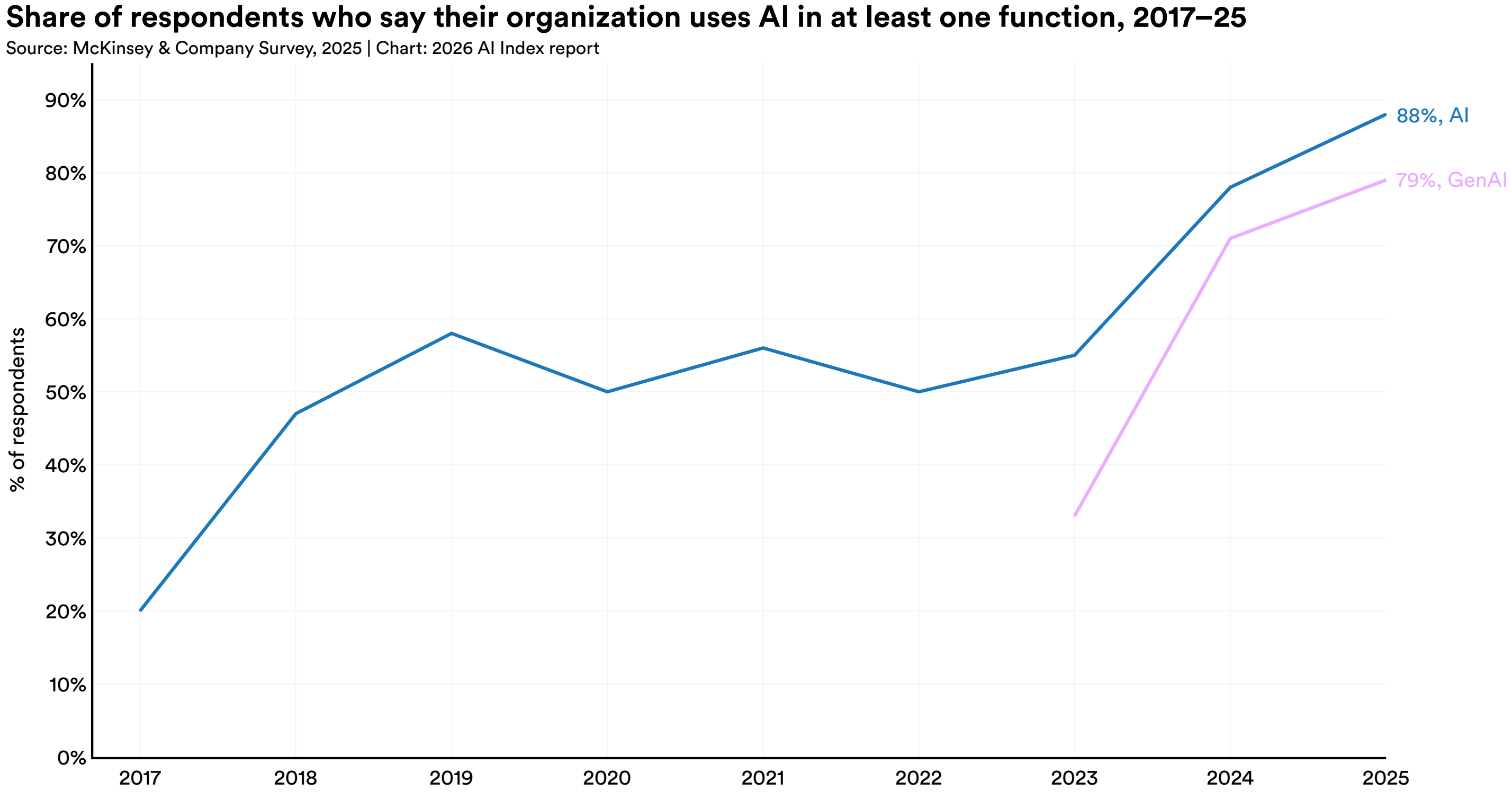
Share of respondents who say their organization uses AI in at least one function, 2017–25
Source: McKinsey & Company Survey, 2025 | Chart: 2026 AI Index report

Figure 4.3.1

**AI use by organizations in the world, 2023–25**
Source: McKinsey & Company Survey, 2025 | Chart: 2026 AI Index report

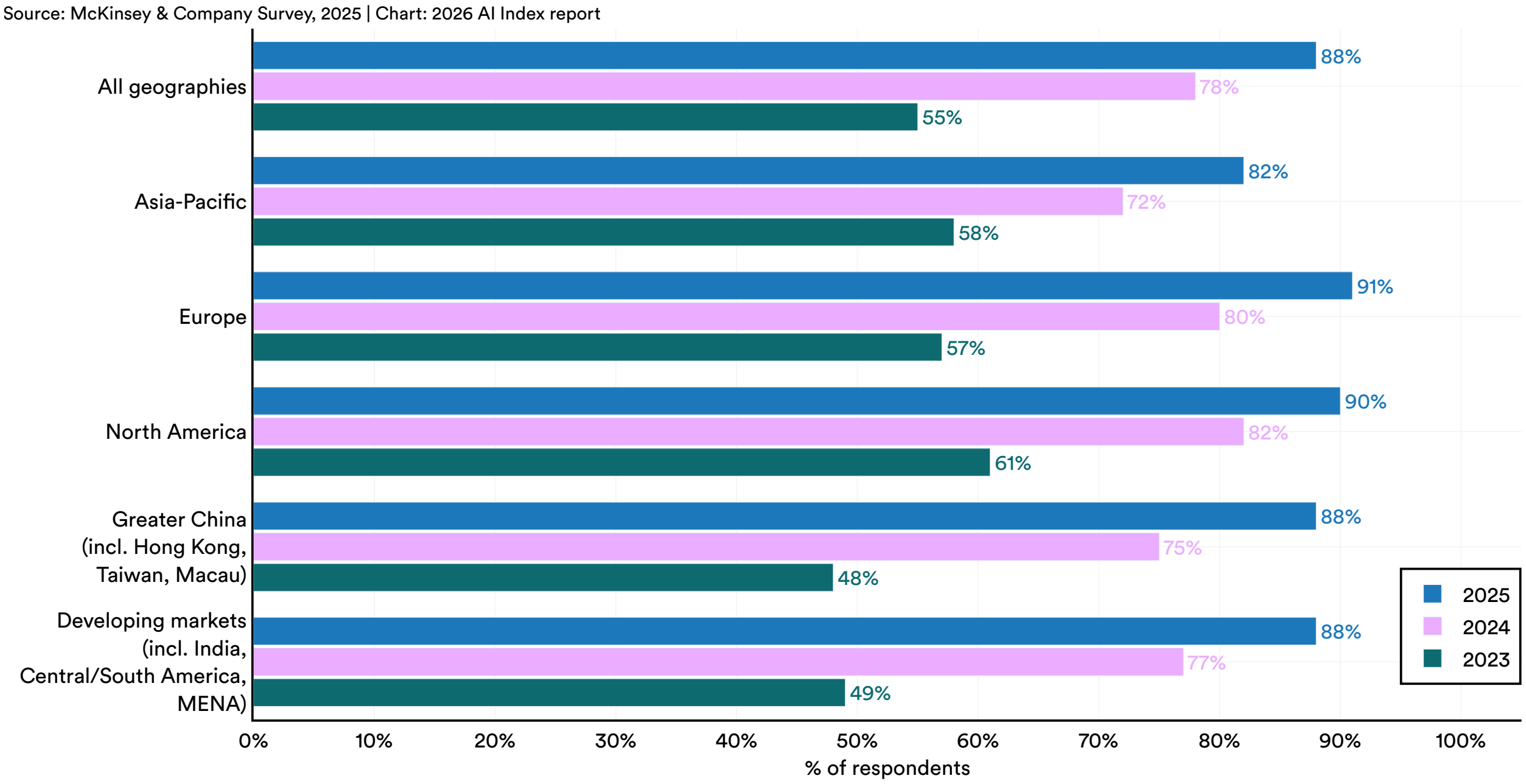

Figure 4.3.2

## By Industry and Function

Adoption patterns varied across industry and function, with some industry/function pairings showing higher rates of diffusion than others (Figure 4.3.3). The highest reported AI usage was in knowledge management for business, legal, and professional services (58%) and in software engineering and IT in the technology sector (58% and 56%, respectively). This was closely followed by marketing and sales for consumer goods and retail (51%). More broadly, functions tied to information processing, software, customer engagement, and internal knowledge work reported higher adoption than areas such as strategy and corporate finance and risk and compliance, where uptake remains low across most sectors. Financial services were an exception; they reported high use in risk and compliance functions, which are more central to their core operations.

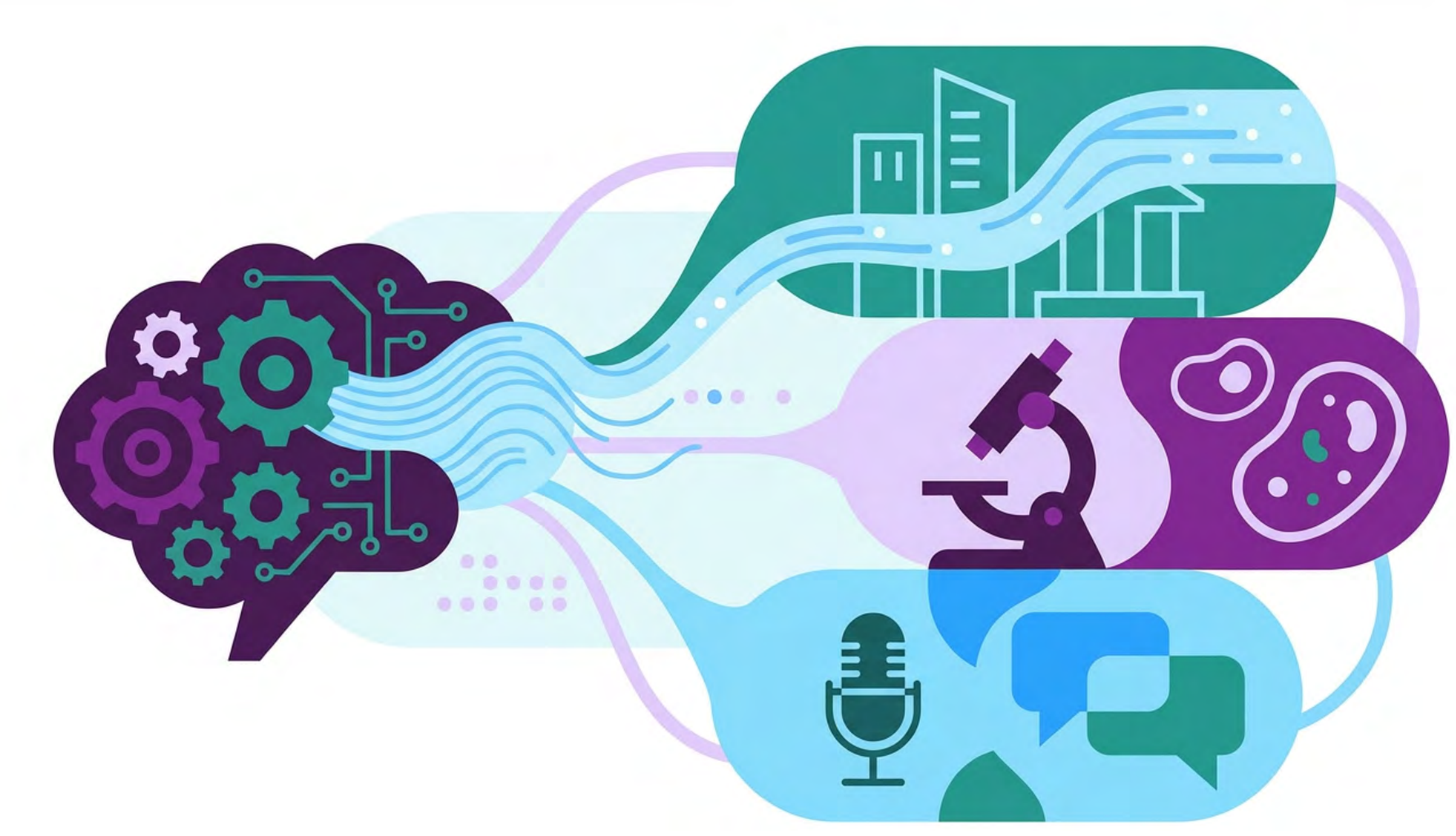

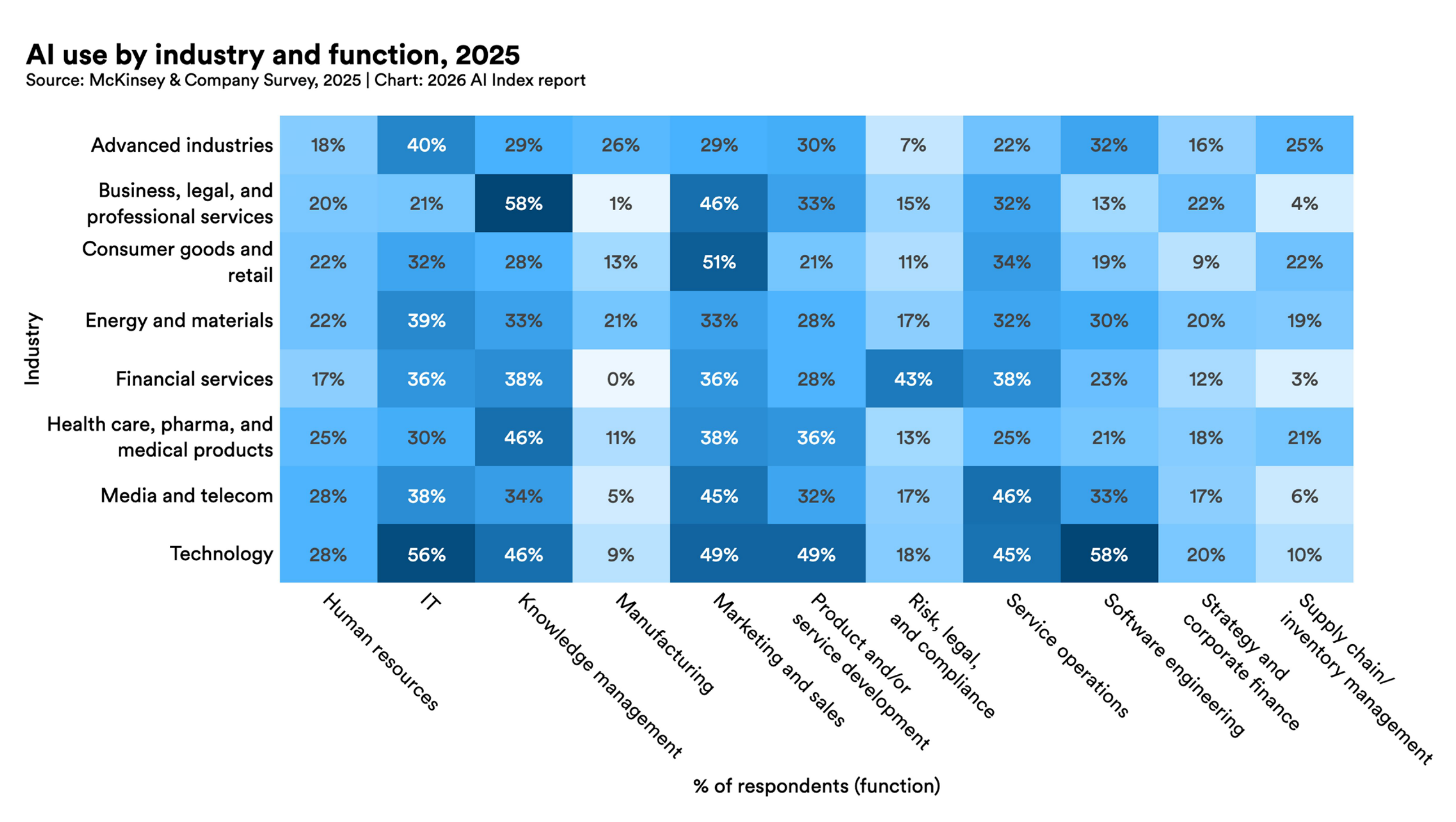

Figure 4.3.3[1]

Respondents more often associated AI with the highest cost savings in software engineering and manufacturing functions (56%), while revenue gains were cited with marketing and sales (67%), strategy and corporate finance (65%), and product and/or service development (62%) (Figure 4.3.4). Across broader organizational outcomes, 64% of respondents reported that AI usage had improved innovation, and 45% reported improvements in employee and customer satisfaction (Figure 4.3.5). Often, the number of respondents who believed AI usage had improved various organizational measures was similar to the number who did not believe it had any effect. Overall, negative effects were reported less frequently, with no more than 7% believing AI usage had worsened cost metrics.

1 "Advanced industries" comprises respondents from sectors such as advanced electronics, aerospace and defense, automotive and assembly, and semiconductors. "Energy and materials" encompasses respondents from agriculture, chemicals, electric power and natural gas, metals and mining, oil and gas, as well as paper, forest products, and packaging.

### Cost decrease and revenue increase from analytical AI use by function, 2025

Source: McKinsey & Company Survey, 2025 | Chart: 2026 AI Index report

| Function | Decrease by <10% | Decrease by 10%–19% | Decrease by ≥20% |  | Increase by >10% | Increase by 6%–10% | Increase by ≤5% |  |
|---|---|---|---|---|---|---|---|---|
| Marketing and sales | 33% | 8% | 7% | 49% | 10% | 14% | 43% | 67% |
| Risk | 24% | 11% | 10% | 45% | | | | |
| Human resources | 33% | 11% | 7% | 51% | | | | |
| Product and/or service development | 28% | 10% | 10% | 47% | 10% | 15% | 38% | 62% |
| Supply chain management | 39% | | 7% | 49% | 5% | 11% | 42% | 59% |
| Service operations | 37% | 8% | 6% | 51% | 5% | 16% | 36% | 57% |
| IT | 36% | 10% | 8% | 54% | 5% | 18% | 30% | 52% |
| Software engineering | 35% | 14% | 6% | 56% | 9% | 13% | 35% | 57% |
| Manufacturing | 45% | 7% | | 56% | | | | |
| Strategy and corporate finance | 28% | 17% | 8% | 53% | 12% | 22% | 31% | 65% |
| Knowledge management | 28% | 8% | 9% | 44% | | | | |

% of respondents

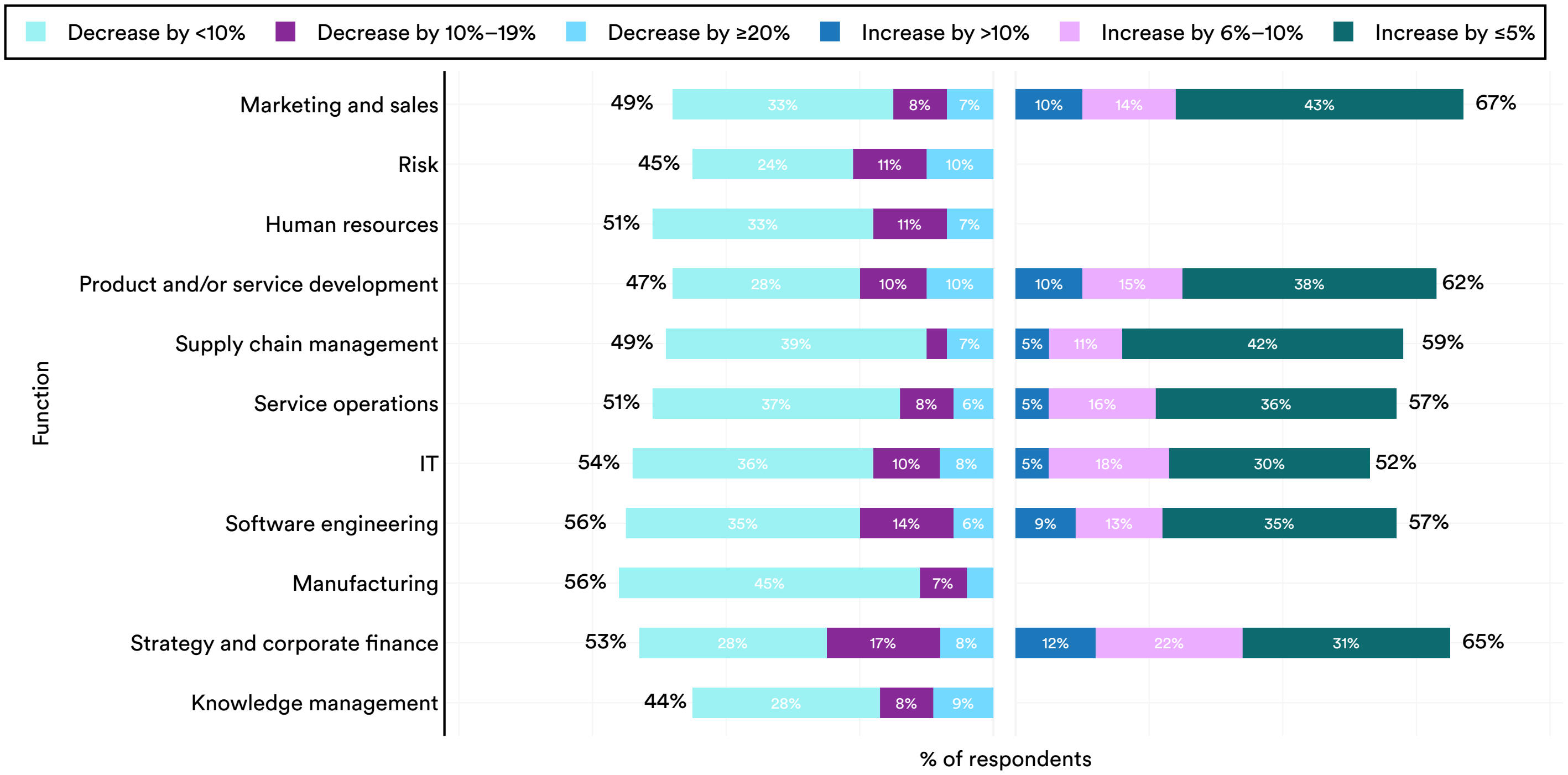

Figure 4.3.4

### AI impact on organizational measures over the past year, 2025

Source: McKinsey & Company Survey, 2025 | Chart: 2026 AI Index report

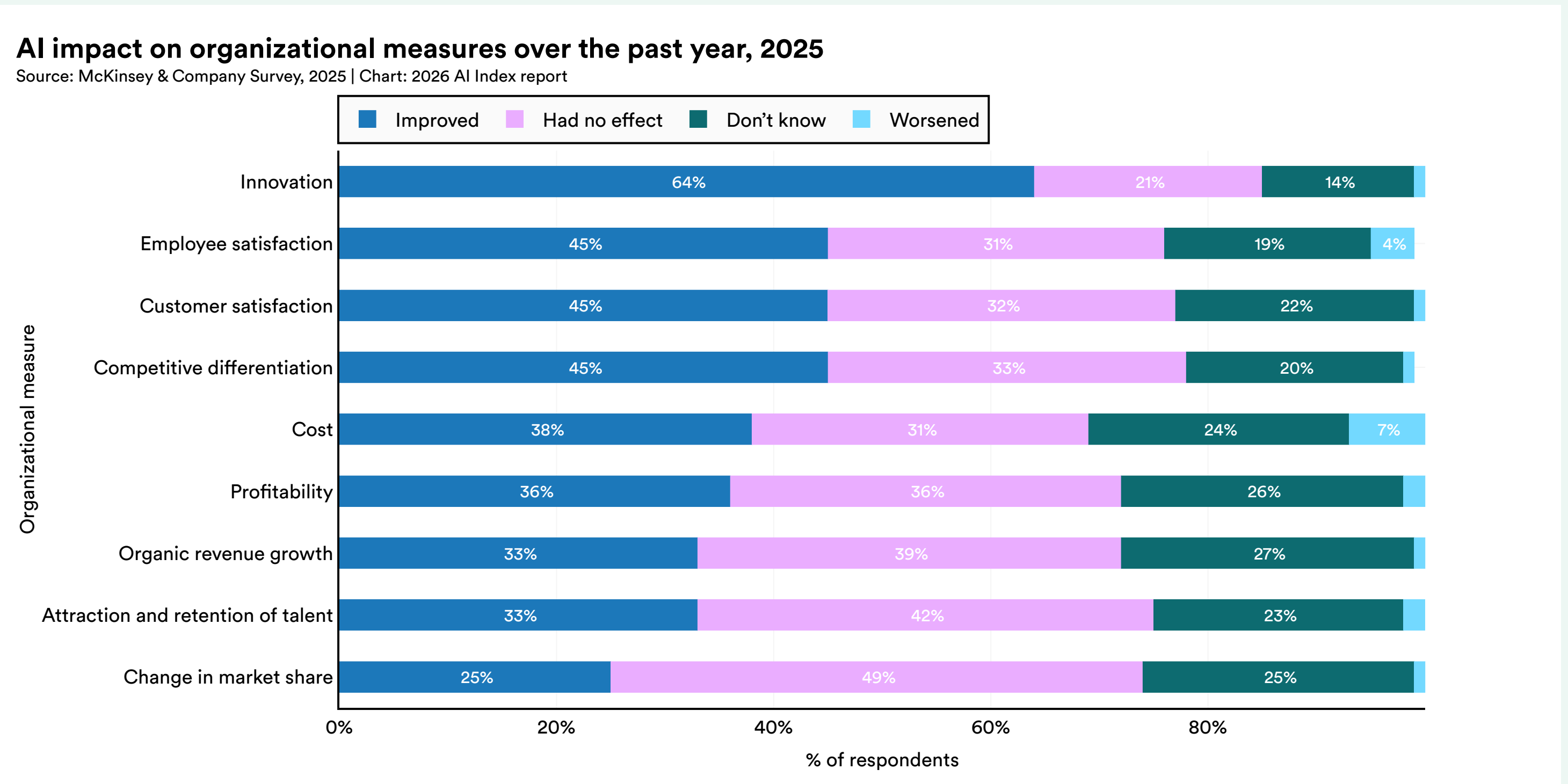

Figure 4.3.5[2]

2 This includes only those respondents whose organizations regularly use AI in at least one business function. Figures may not add up to 100% because of rounding.

## Deployment Stages

The McKinsey survey also captured how deeply AI had been integrated into an organization's operations by looking at different stages of the deployment life cycle (Figure 4.3.6). As expected, given the resource and investment demands of integration, larger companies were the most likely to report that their AI programs had reached a scaling phase.

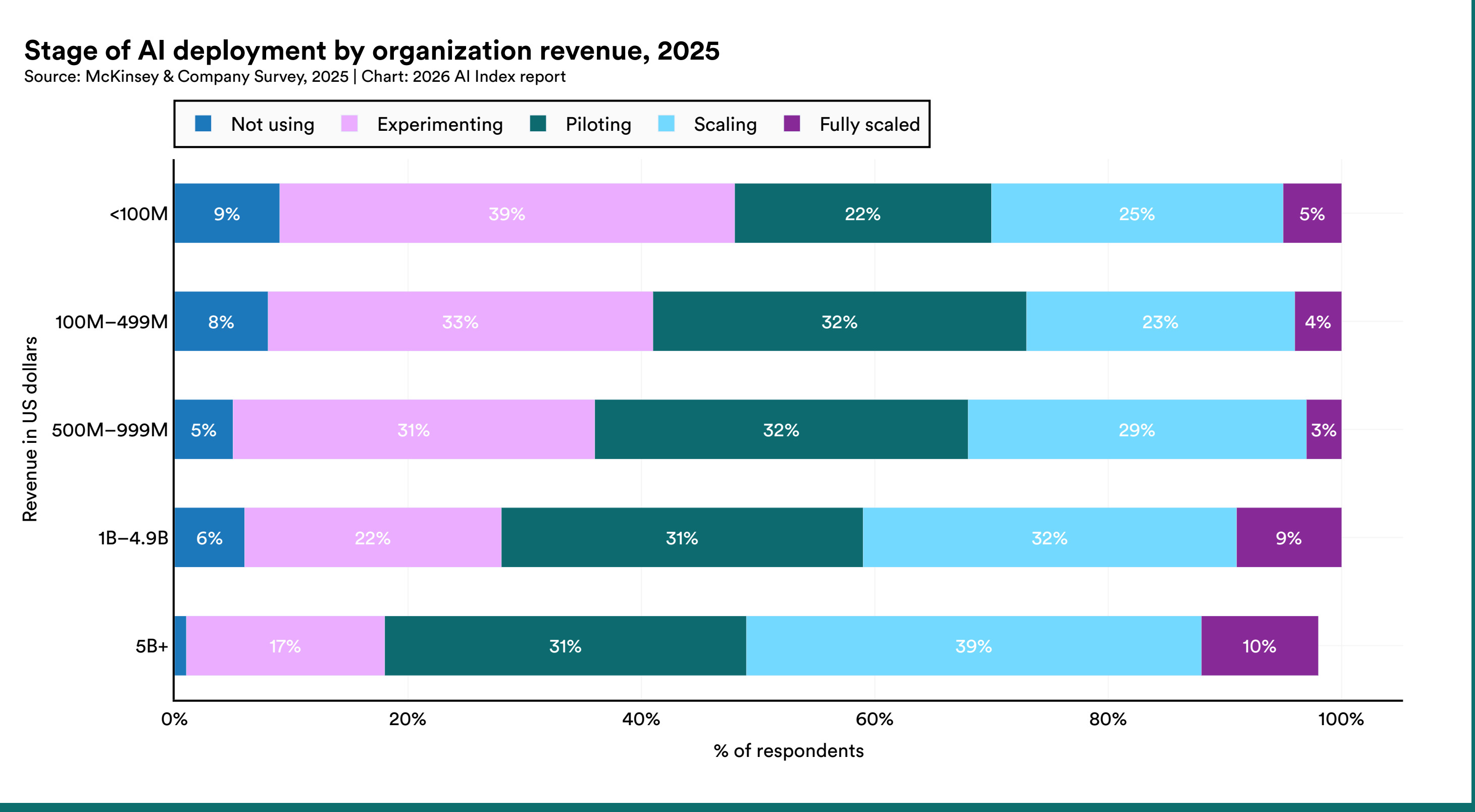

Figure 4.3.6[3]

Early indicators on AI agent adoption show that diffusion is still at an early stage. Across most business functions, a majority of respondents reported no agent use at all (Figure 4.3.7). Scaled use was in the single digits for nearly all functions. Even in functions with the most activity, including IT and knowledge management, about two-thirds or more of respondents reported no use. At the industry level, the technology sector had comparatively higher rates of scaled agent use in software engineering (24%), IT (22%), and service operations (21%) (Figure 4.3.8). The business functions reporting the highest rates of AI agent use tend to be the same as those with broader, more established AI adoption.

3 Figures may not add up to 100% because of rounding; respondents who said "I don't know" were not shown but represent <1% of the total, which could also cause bars to not add up to 100%.

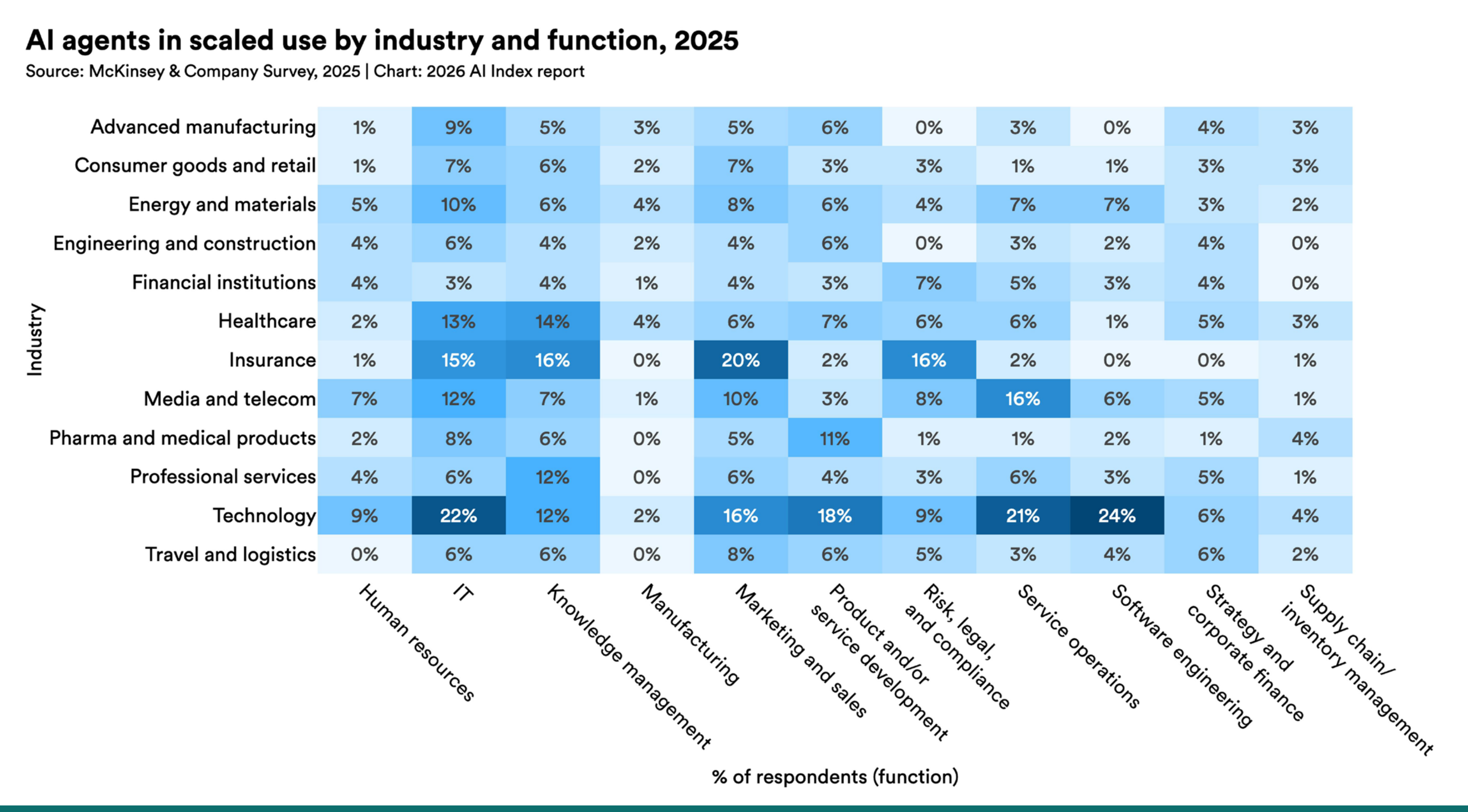

Figure 4.3.7

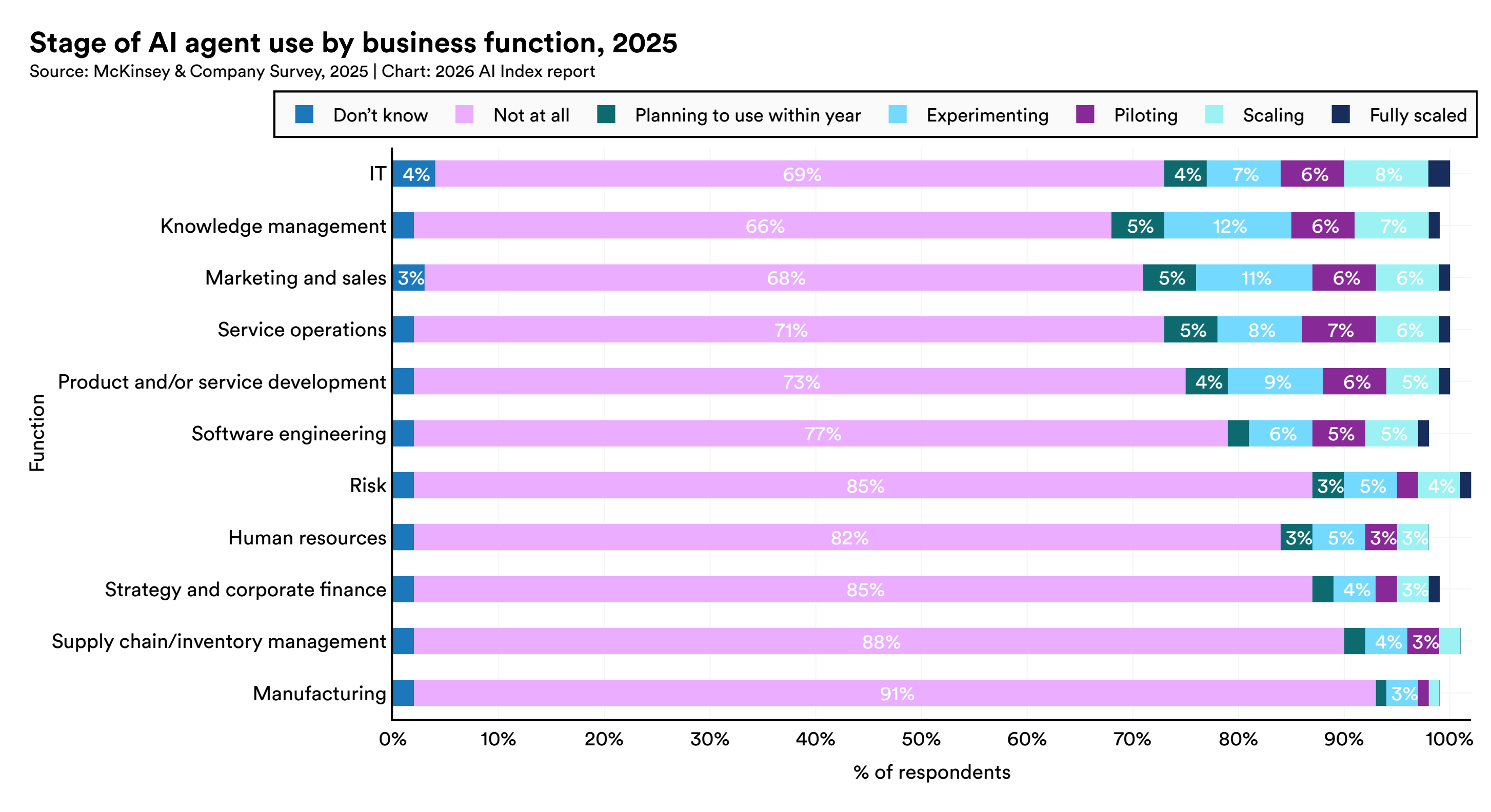

Figure 4.3.8[4]

4 Figures may not add up to 100% because of rounding.

HIGHLIGHT:

## Measuring Signals of AI Diffusion

The full economic impact of AI is difficult to assess through investment patterns or organizational adoption alone. The assessment also requires tracking diffusion, or how widely AI tools are being adopted across populations, countries, occupations, and everyday tasks. This section brings together several complementary signals of AI diffusion, including population-level survey estimates, cross-country comparisons, historical adoption benchmarks, and platform-level usage data. Once combined, these measures offer a comprehensive view of how quickly AI is being integrated into work and daily life.

Compared with earlier transformative technologies, generative AI's adoption has been rapid in the years after its mass market introduction (Figure 4.3.9). Measured from the release of each technology's first widely available product, generative AI reached approximately 53% adoption within three years, well above the initial trajectories of the personal computer and the internet over comparable time frames (Bick et al., 2024). The sharp uptake is also reflected in the revenue trajectories of leading AI companies, as seen in the company-level revenue analysis in section 4.2, where commercial scale was reached in comparably shorter time frames (Figure 4.3.9).

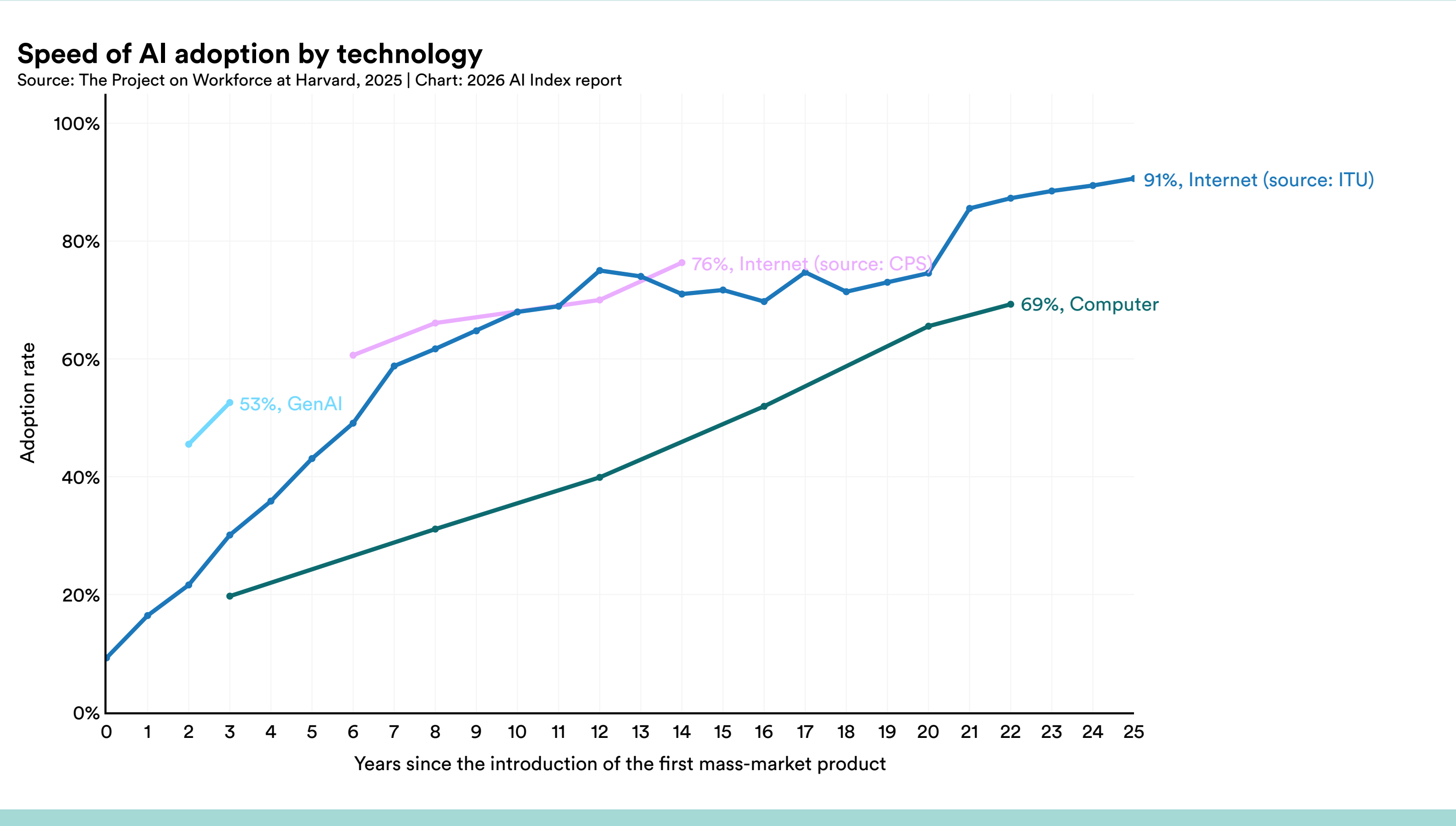

Figure 4.3.9[5]

5 Source: https://www.genaiadoptiontracker.com. The figure shows overall usage rates for three technologies: generative AI, computers, and the internet. The horizontal axis represents years since the introduction of the first mass-market product for each technology. We use 1981 as the introduction year for computers, which was the year the IBM PC was released. We use 1995 as the introduction year for the internet, which was the year that the NSF decommissioned NSFNet and allowed the internet to carry commercial traffic. We use 2022 as the introduction year for generative AI, which was the year ChatGPT was released. The data source for computers is the 1984–2003 Computer and Internet Use Supplement of the CPS. We plot two estimates of internet use: the 2001–2009 Computer and Internet Use Supplement of the CPS and the ITU. The sample for the RPS and CPS is all individuals ages 18–64. The sample for the ITU is individuals of all ages. We pool RPS waves by year.

**HIGHLIGHT:**

Another broad signal comes from Microsoft's telemetry data of AI usage across countries (Figure 4.3.10). Adoption varies widely, and shows a strong, statistically significant positive correlation with GDP per capita (Misra et al., 2025) (Figure 4.3.11). Most high-income economies cluster between 25% and 45% adoption, with European and North American averages reaching approximately 27% and 22%, respectively. Lower usage is reported in South Asia and sub-Saharan Africa, where GDP per capita is also lower. However, there are exceptions to the relationship between GDP and AI adoption. The United Arab Emirates and Singapore report adoption levels above 64% and 61%, respectively, well above what their GDP per capita would predict. Some wealthy economies, such as the United States and Denmark, fall below the trend.

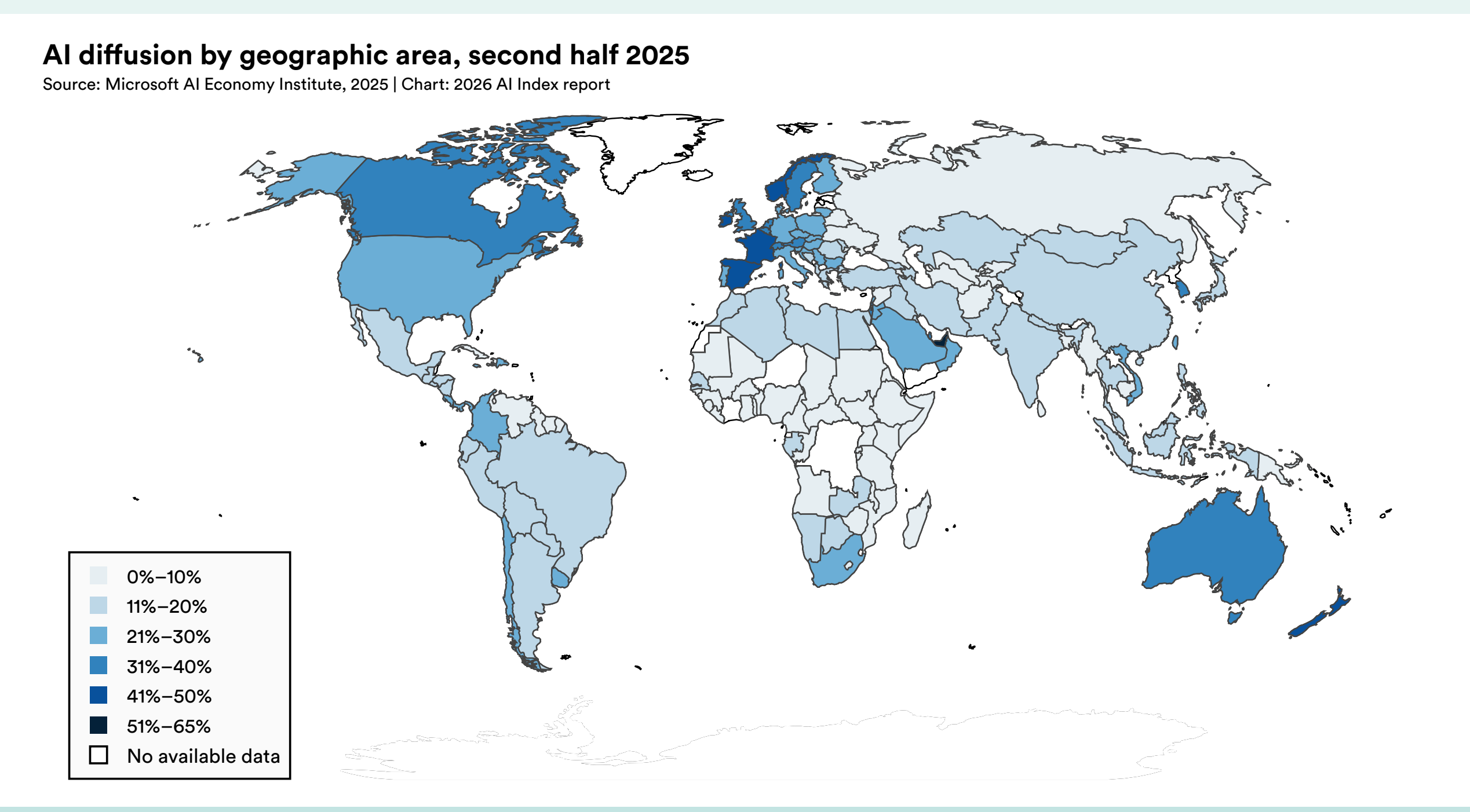

Figure 4.3.10

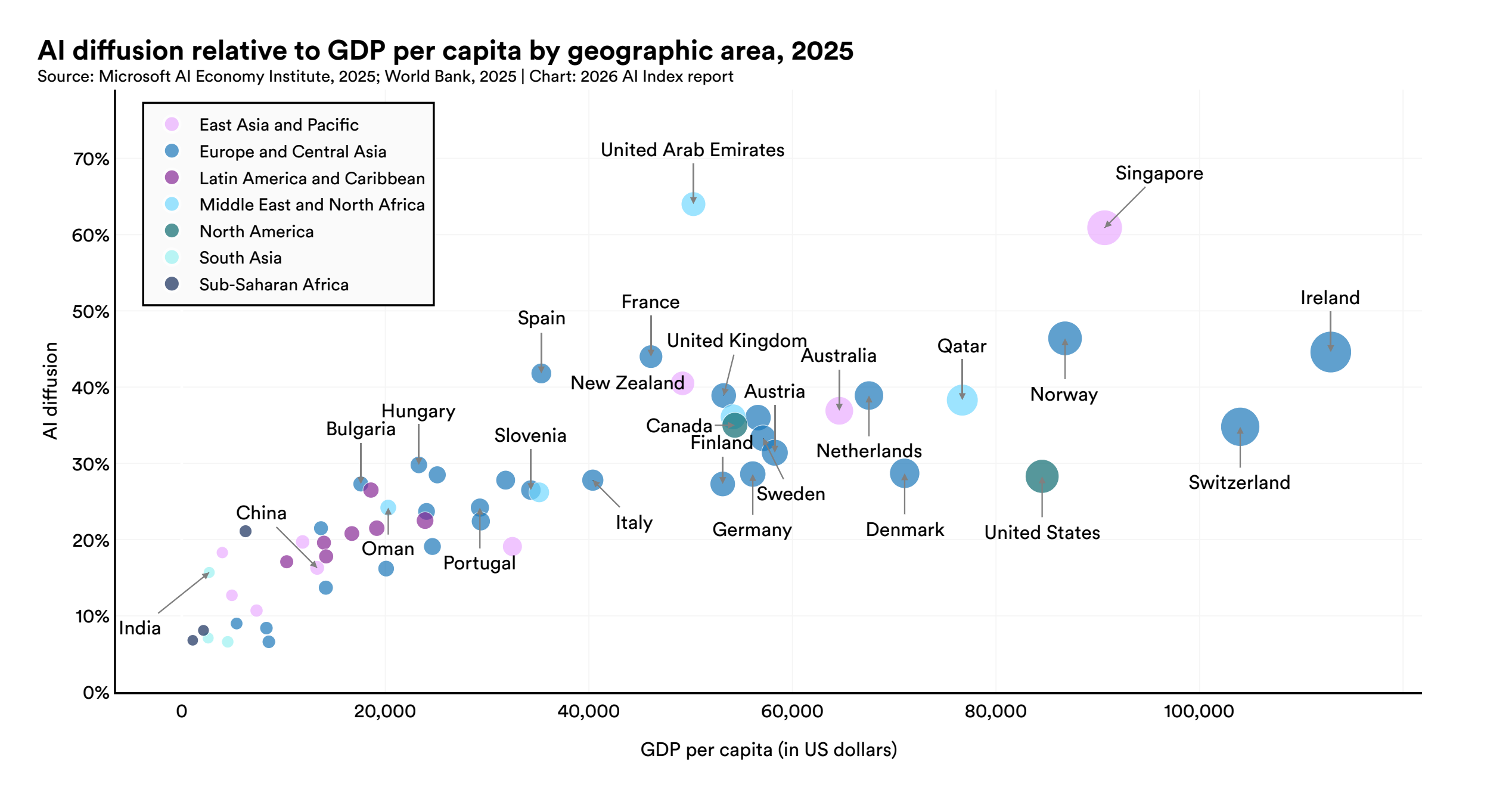

Figure 4.3.11

**HIGHLIGHT:**

Between the first and second half of 2025, AI adoption grew across the majority of the top 30 economies (Figure 4.3.12). South Korea posted the largest gain of 4.8%, climbing the rankings from 25th to 18th. The United States, despite its leading position in AI investment and model development, dropped to 24th place with a population-level adoption rate of 28.3%. Even as usage grows, the United States remains in the lower half of the global adoption ranking, in line with the more cautious public mood toward AI explored in Chapter 9.

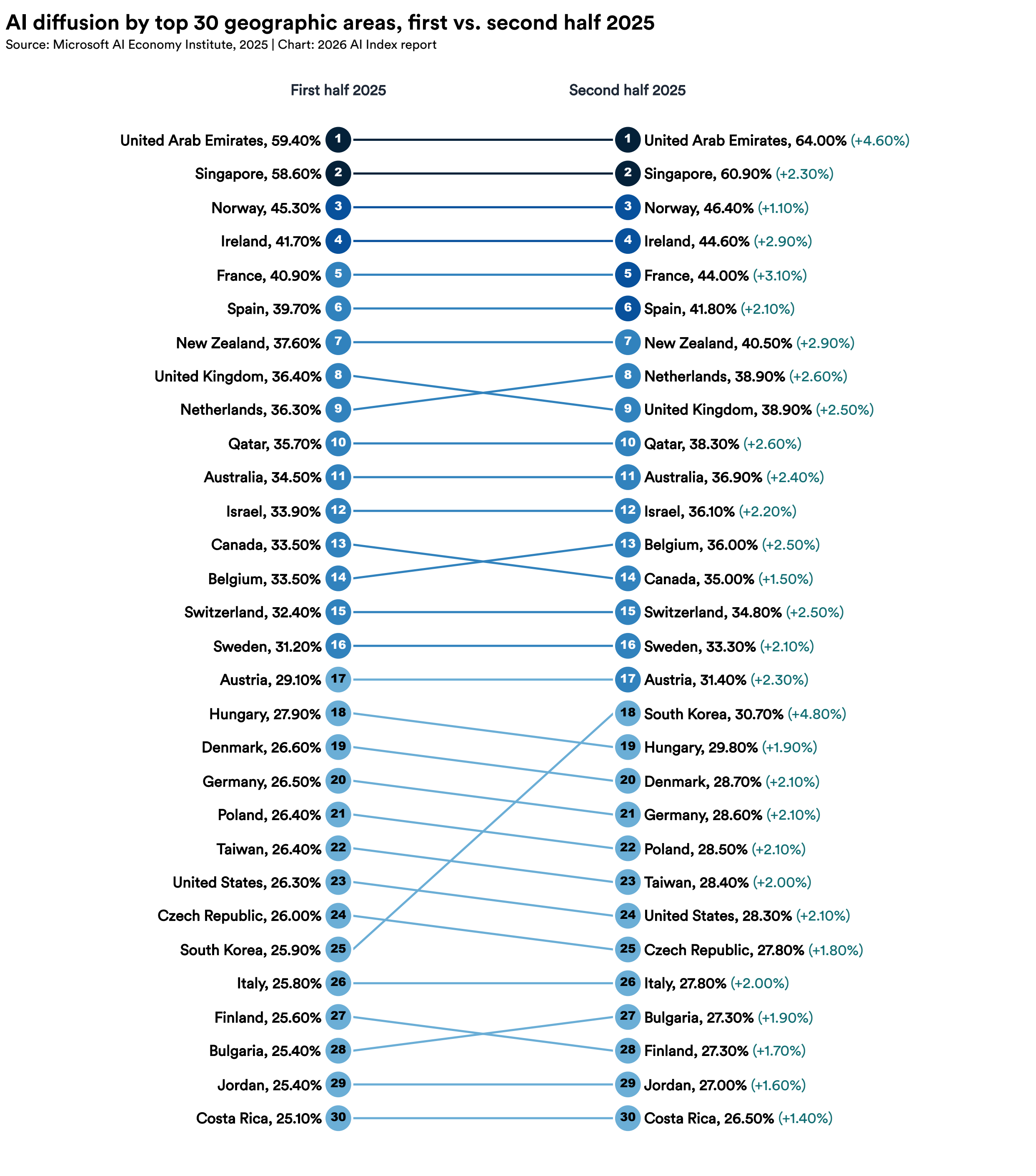

Figure 4.3.12

**HIGHLIGHT:**

On a more granular level, platform-level data from Anthropic's AI Usage Index[6] provides a view of adoption across occupations and tasks (Massenkof et. al, 2026). Throughout 2025, computer and mathematical tasks accounted for the largest share of overall usage, consistently representing close to 40% of activity (Figure 4.3.13). Educational instruction and library tasks showed the most significant growth, rising from 9% early in the year to approximately 14% by late 2025. This growth in educational settings is worth noting alongside Chapter 7, which explores how institutional guidance and readiness still lag behind adoption. In addition, an analysis of the conversation patterns reveals a shift in how users interact with the tools (Figure 4.3.14). The share of automation-oriented conversations, where users instruct the tool to complete a task autonomously, rose from 41% at the start of 2025 to 49% in August. This surpassed augmentation-style interactions for the first time. However, by November, augmentation had moved ahead, representing 52% of the share of conversations. The fluctuation over the course of the year suggests that automation-oriented use is becoming more prevalent, which is consistent with the early-stage AI agent adoption patterns for organizations, described in the previous section (Figure 4.3.8).

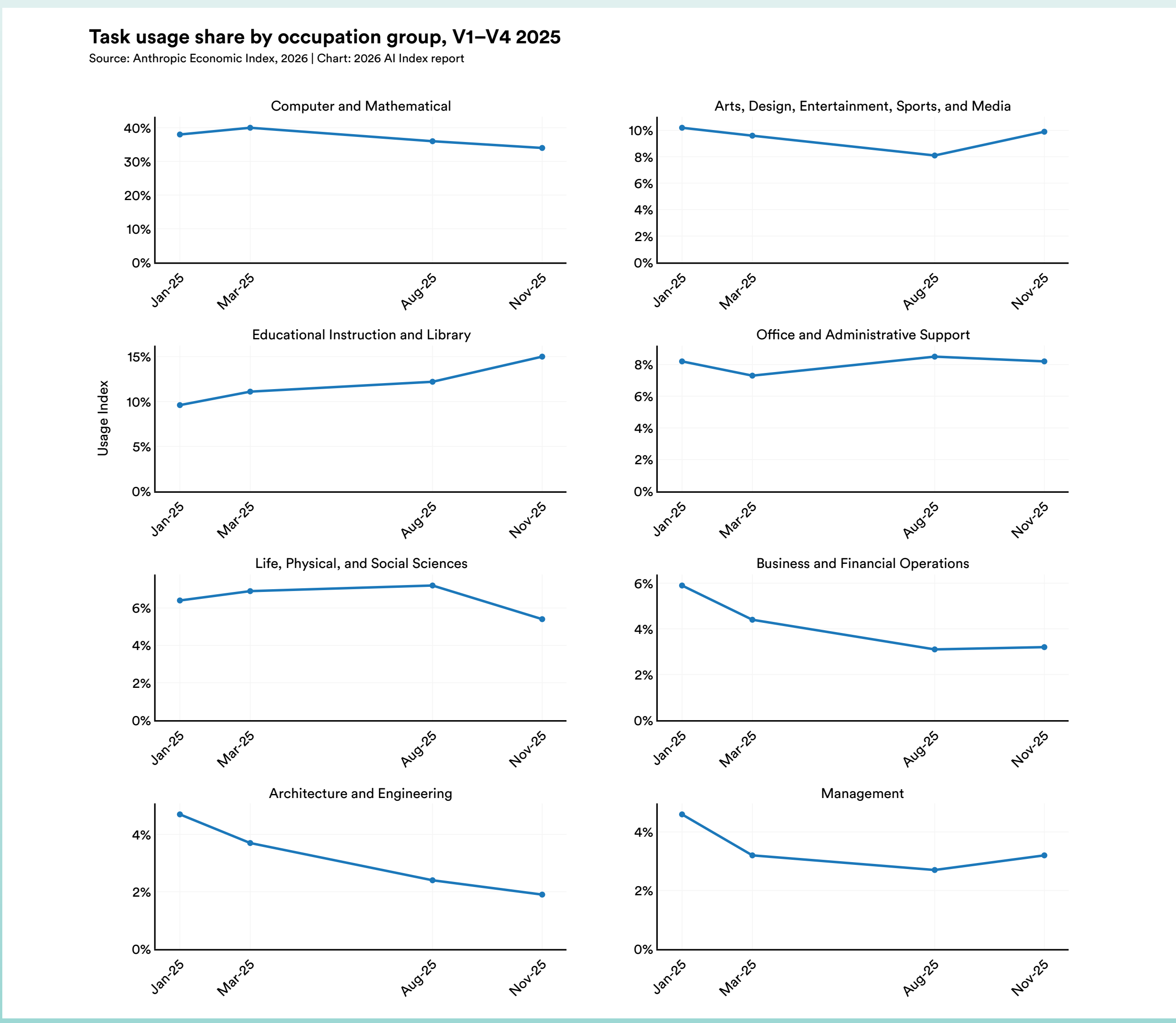

Figure 4.3.13[7]

6 The Anthropic AI Usage Index (AUI) measures Claude usage relative to the working-age population by calculating each geography's share of Claude usage divided by its share of the working-age population (ages 15–64). Countries with an AUI greater than 1 use Claude more often than expected based on their working-age population alone, while those with an AUI less than 1 use it less.

7 V1–V4 refer to the four successive releases of the Anthropic Economic Index in 2025, corresponding to January, March, August, and November 2025, respectively.

HIGHLIGHT:

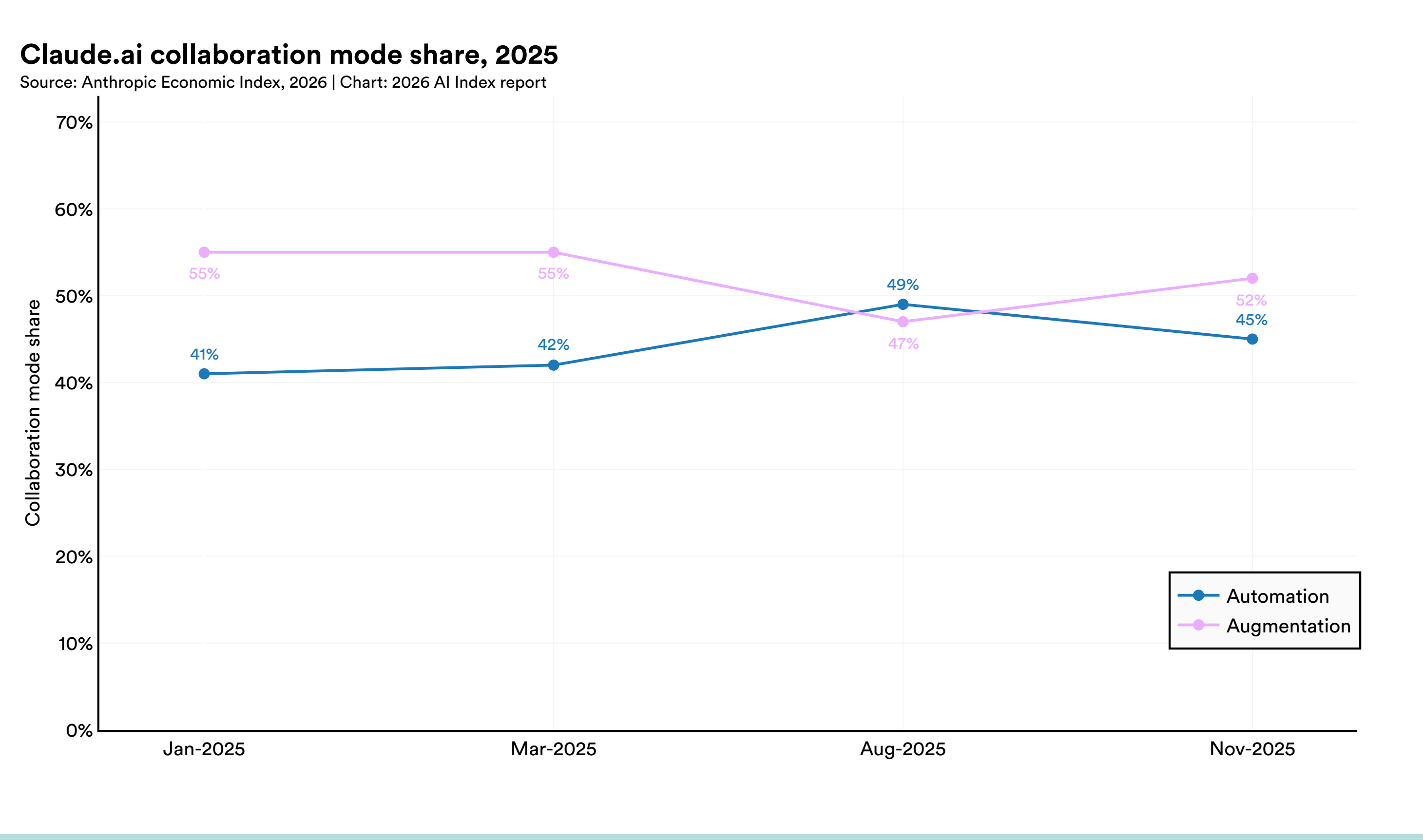

Figure 4.3.14

AI diffusion is also shaped by broader societal attitudes, including public trust and optimism about the technology. Chapter 9 studies these trends to determine how excitement for and exposure to AI vary across countries and what they suggest about the societal experience of increasing adoption.

# 4.4 Jobs

Labor markets provide signals of how investment, technical progress, and organizational adoption are changing workforce dynamics. This section tracks both the demand side of the labor market, through job postings and skill requirements, and the supply side, through talent flows, before examining the impact on employment outcomes and employee sentiment. The analysis draws from Lightcast's job posting database, LinkedIn's talent and hiring metrics, and recent research on AI's effects on the labor market.

## AI Labor Demand Across Geographies

Across the countries tracked by Lightcast, demand for AI-related talent continued to increase in 2025[8] as job listings that require AI skills continue to make up a growing share of overall postings (Figures 4.4.1 and 4.4.2). While most countries are hitting new peaks of demand, the intensity varies across countries. In 2025, Singapore led with 4.69% of all job postings that required AI skills, followed by Hong Kong (3.5%), Luxembourg (3.4%), and Spain (3.3%). The United States reached 2.6%, followed by Chile (2.4%) and the United Kingdom (1.9%).

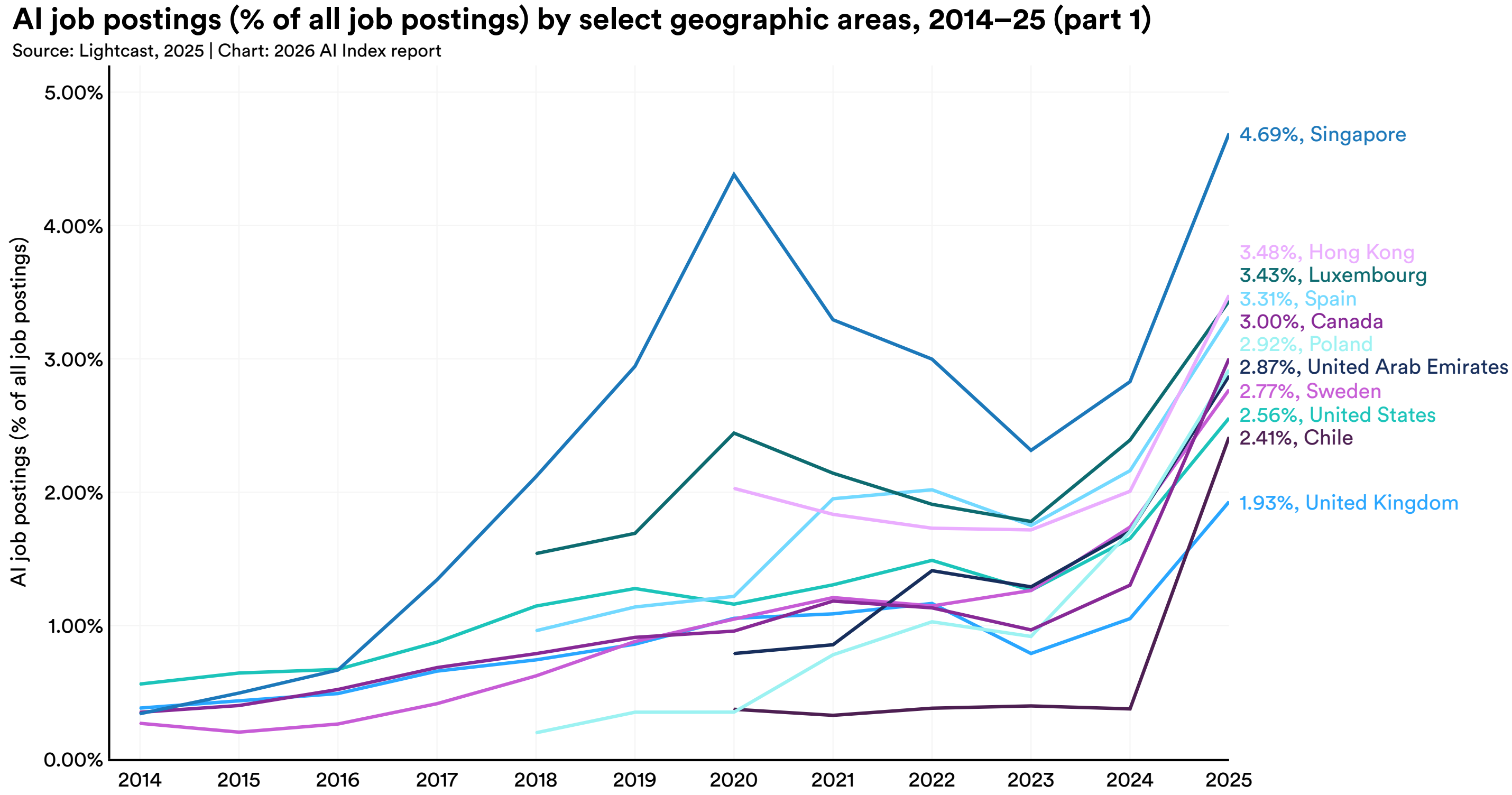

Figure 4.4.1

8 Historical posting counts may differ from previously published versions due to data updates and revisions. However, year-over-year trends remain consistent with earlier analyses. See here for more details.

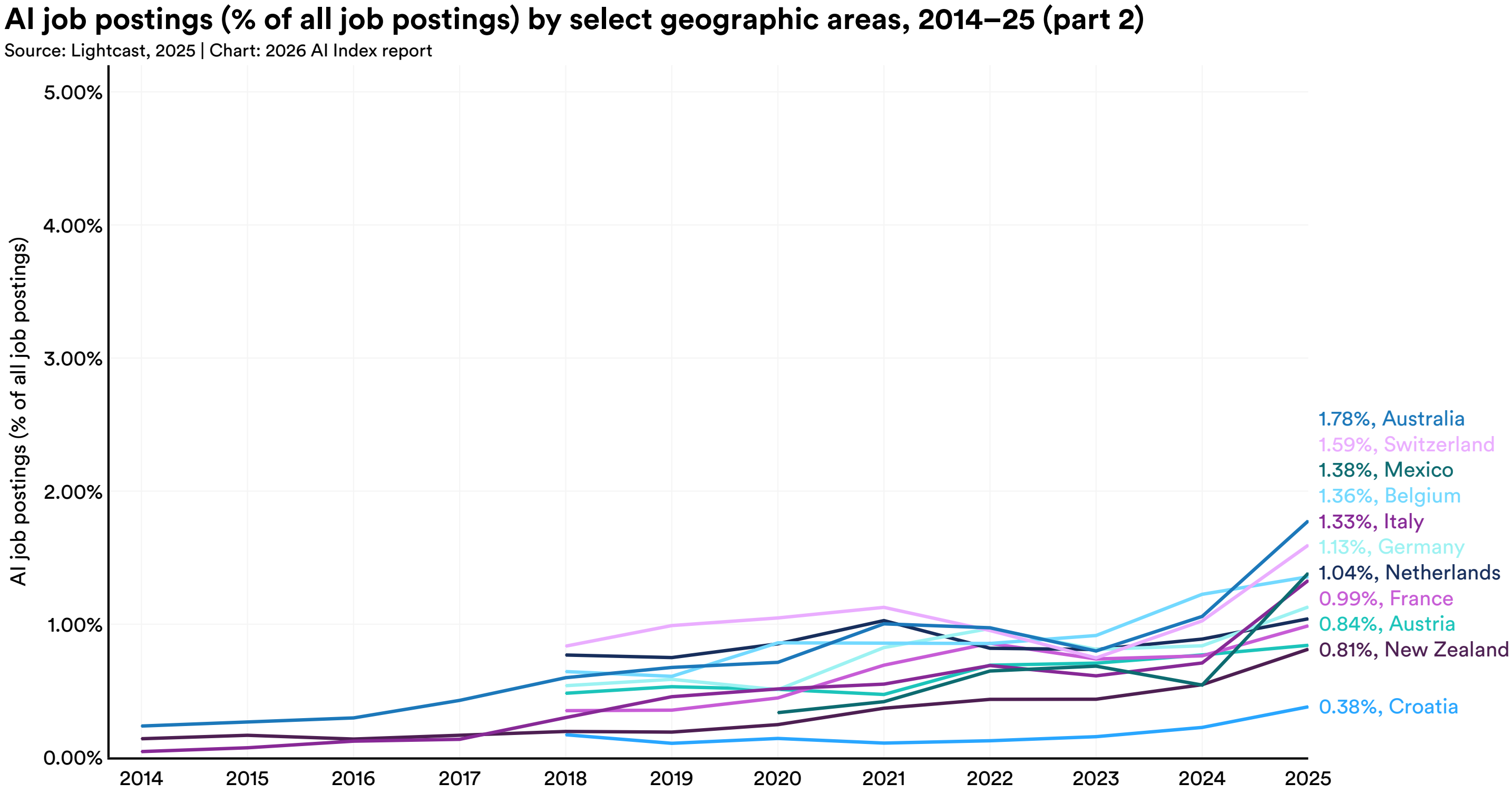
AI job postings (% of all job postings) by select geographic areas, 2014–25 (part 2)
Source: Lightcast, 2025 | Chart: 2026 AI Index report

Figure 4.4.2

## AI Hiring in the United States

### Skill Composition

Within the United States, AI labor demand can be disaggregated by skillset to reveal how the workforce footprint is evolving (Figures 4.4.3–4.4.8). In 2025, broad AI and machine learning skill clusters remain the most frequently cited categories in AI job posting, accounting for 1.7% and 1.0% of all job postings. Among the top specialized skills, Python appeared the most often, in 258,674 posts, a 391% increase compared to the 2013–15 time period and a near 30% increase from 2024. The fastest growth appears in skills needed to build and operate systems at scale, with employer demand mirroring the broader investment shift toward AI infrastructure and deployment capacity. Amazon Web Services expanded significantly compared to a decade ago (+1,358%) alongside an increasing emphasis on scalability (+733%) and workflow management (+818%).

Mentions of generative AI skills in AI job postings grew 111% from 2024 to 2025, though their share of total AI job postings decreased by 5%. With overall AI labor demand rising, a newer skill cluster tied to AI agents emerged. From 2024 to 2025, postings referencing agentic AI, AI agents, or agentic systems exponentially increased. The share of AI job postings that mentioned ChatGPT, chatbot, or conversational AI declined, while posts referencing agentic terms or orchestration frameworks such as LangGraph increased. Job demand appears to be shifting from general familiarity with chat-based tools toward skills required to coordinate and operationalize task-oriented systems.

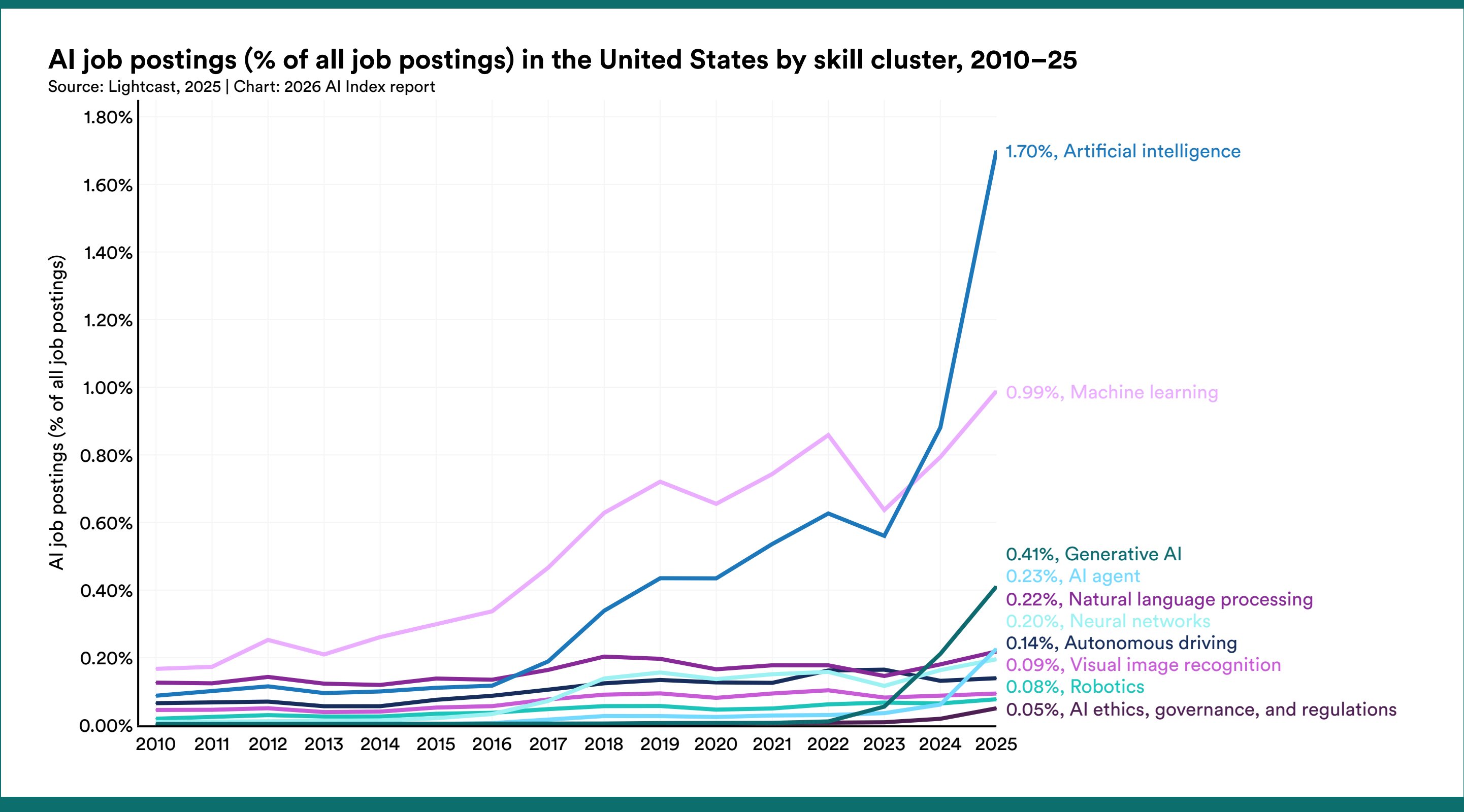

Figure 4.4.3[9]

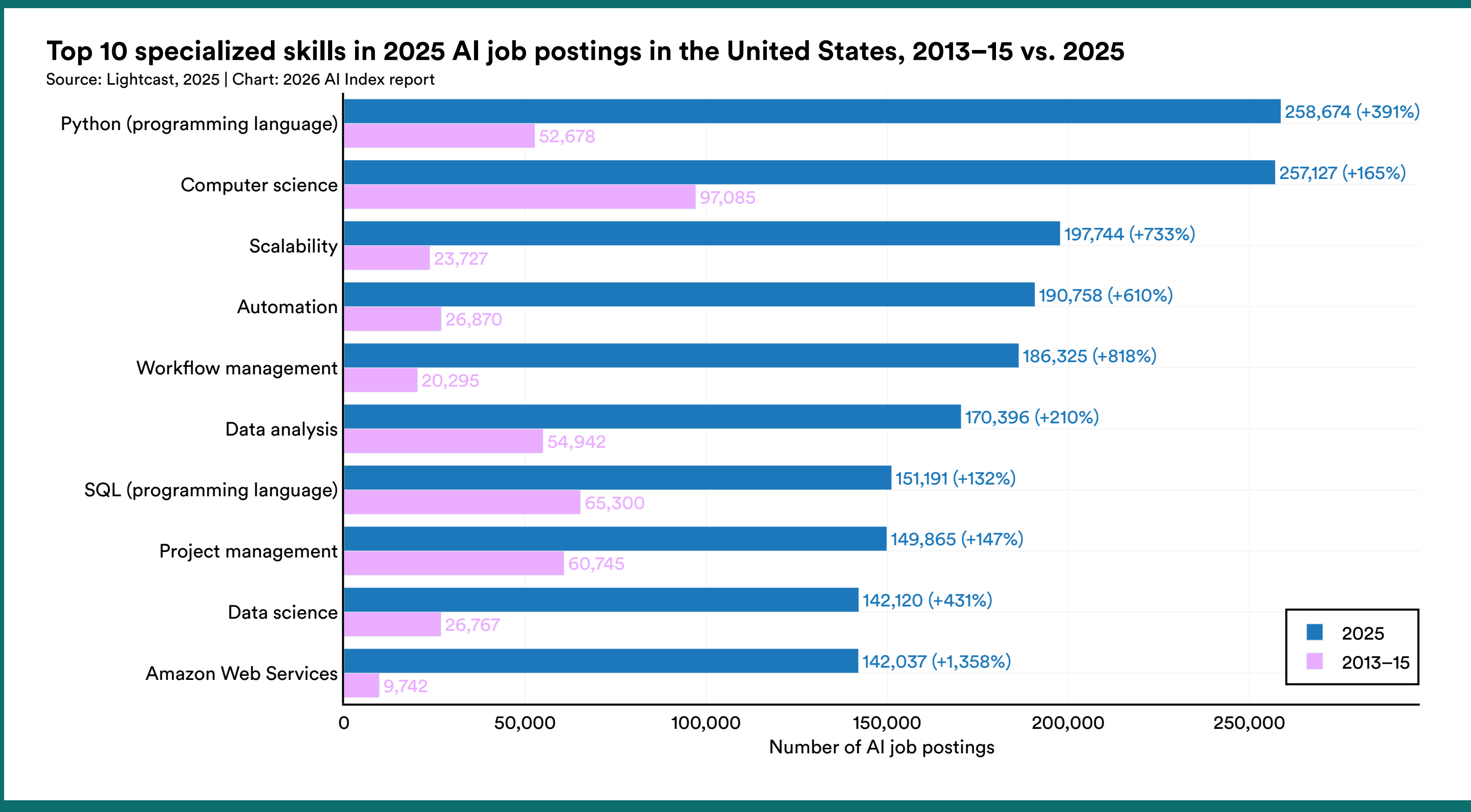

Figure 4.4.4

9 A single job posting can list multiple AI skills.

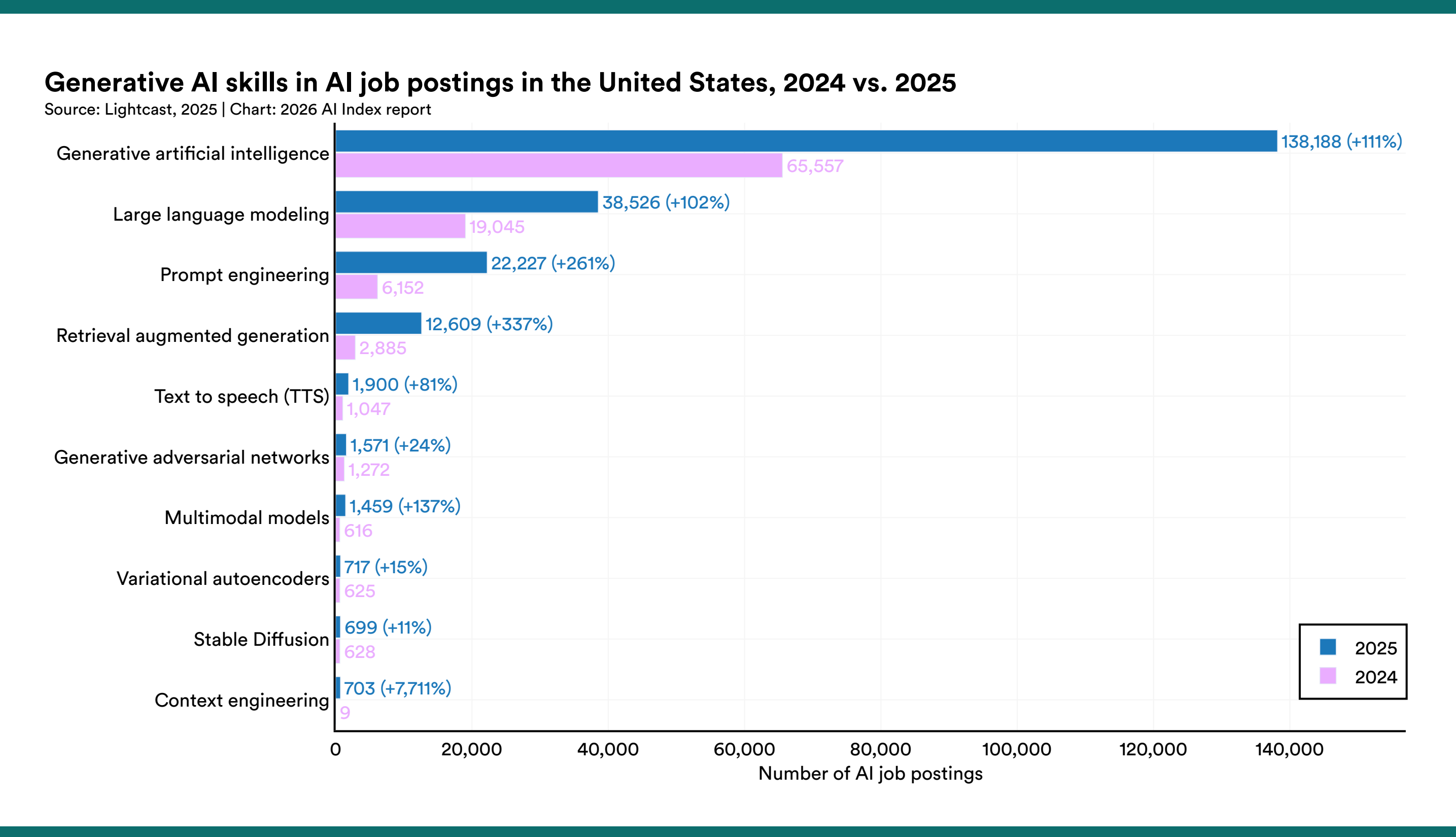

Figure 4.4.5

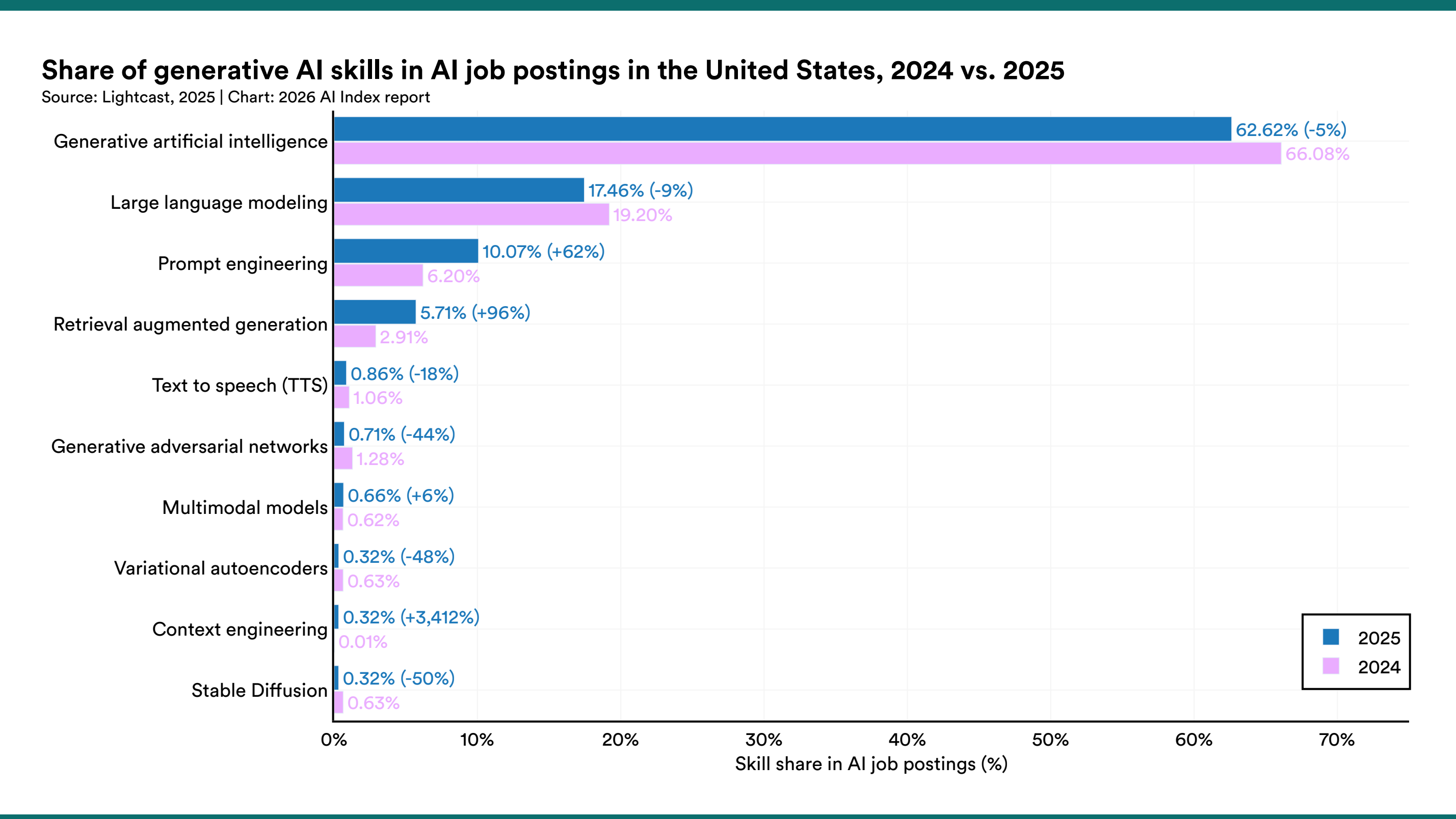

Figure 4.4.6

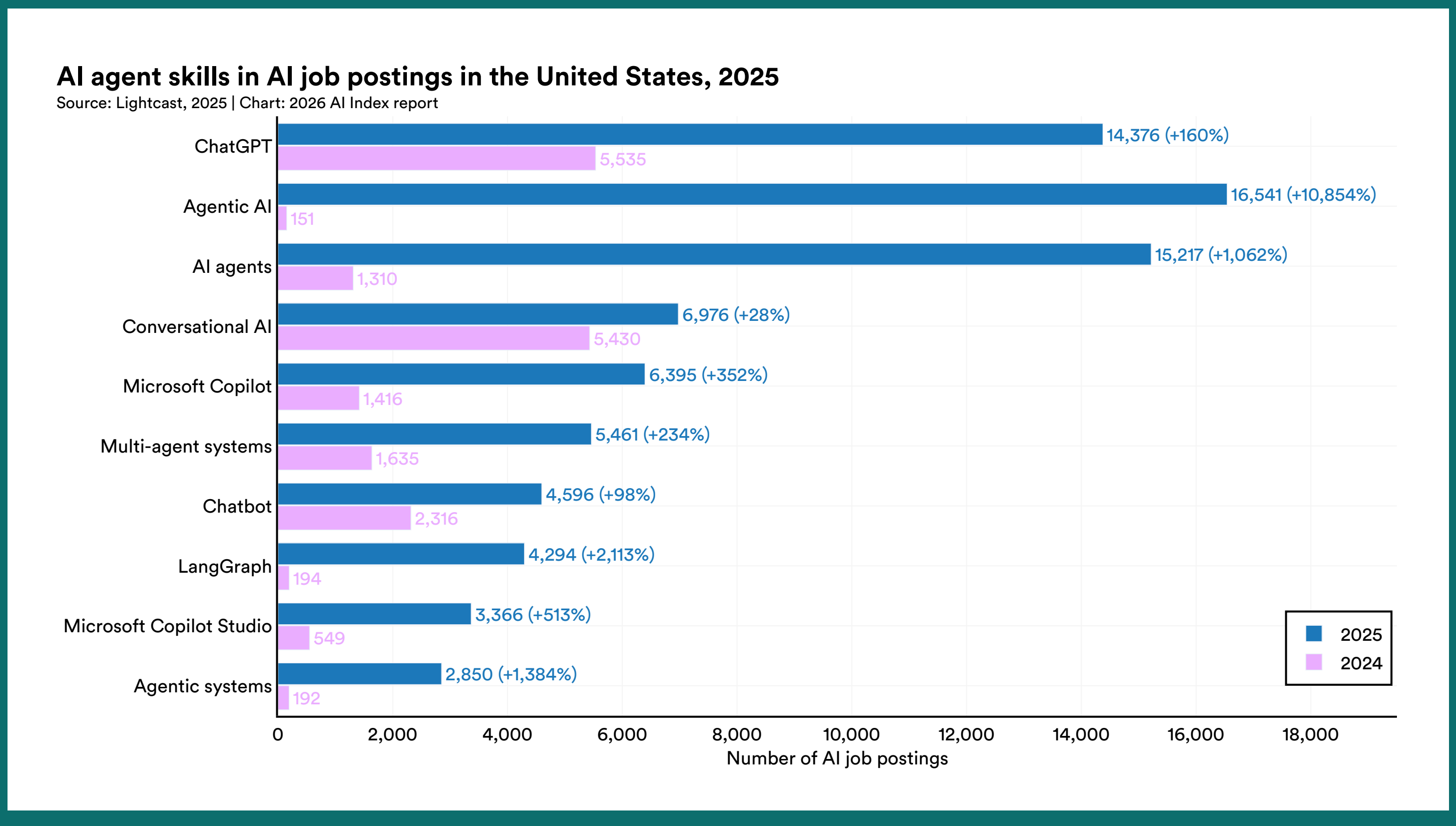

Figure 4.4.7[10]

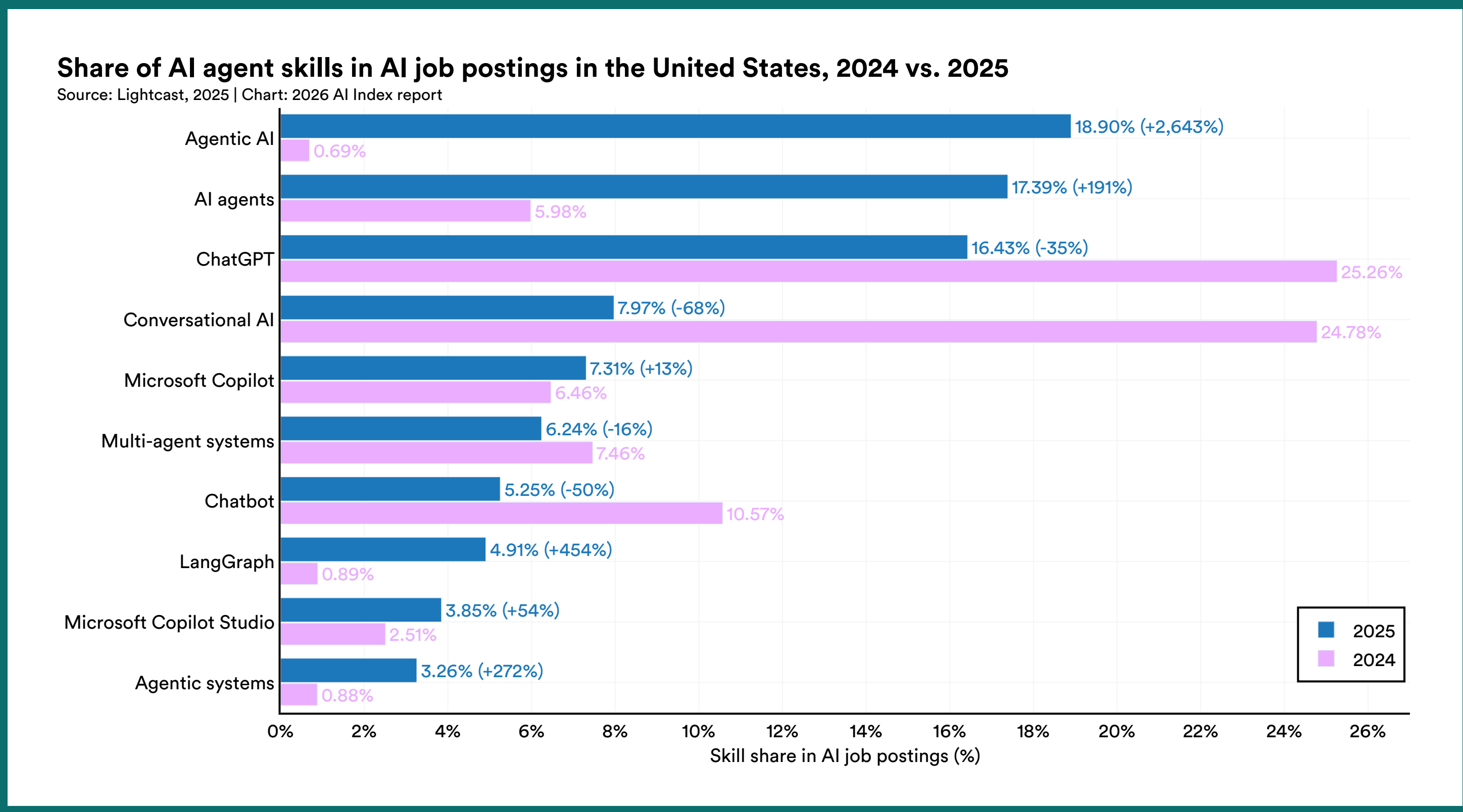

Figure 4.4.8

10 For definitions of each skill category, see the Lightcast taxonomy at https://lightcast.io/open-skills or the Appendix.

## By Sector

Demand for AI talent increased across all economic sectors in 2025, though the pace of growth varied (Figure 4.4.9). The information sector leads, with AI skills appearing in a 13.2% share of its job postings, up from 7.8% in 2024. Other sectors with relatively high AI posting shares include professional, scientific, and technical services (6.5%), finance and insurance (5.3%), and manufacturing (4.7%). In 2025, AI hiring also expanded in sectors with historically low adoption rates. Transportation and warehousing, real estate, and education showed year-over-year increases, evidence that the diffusion is reaching beyond traditional technology-driven industries.

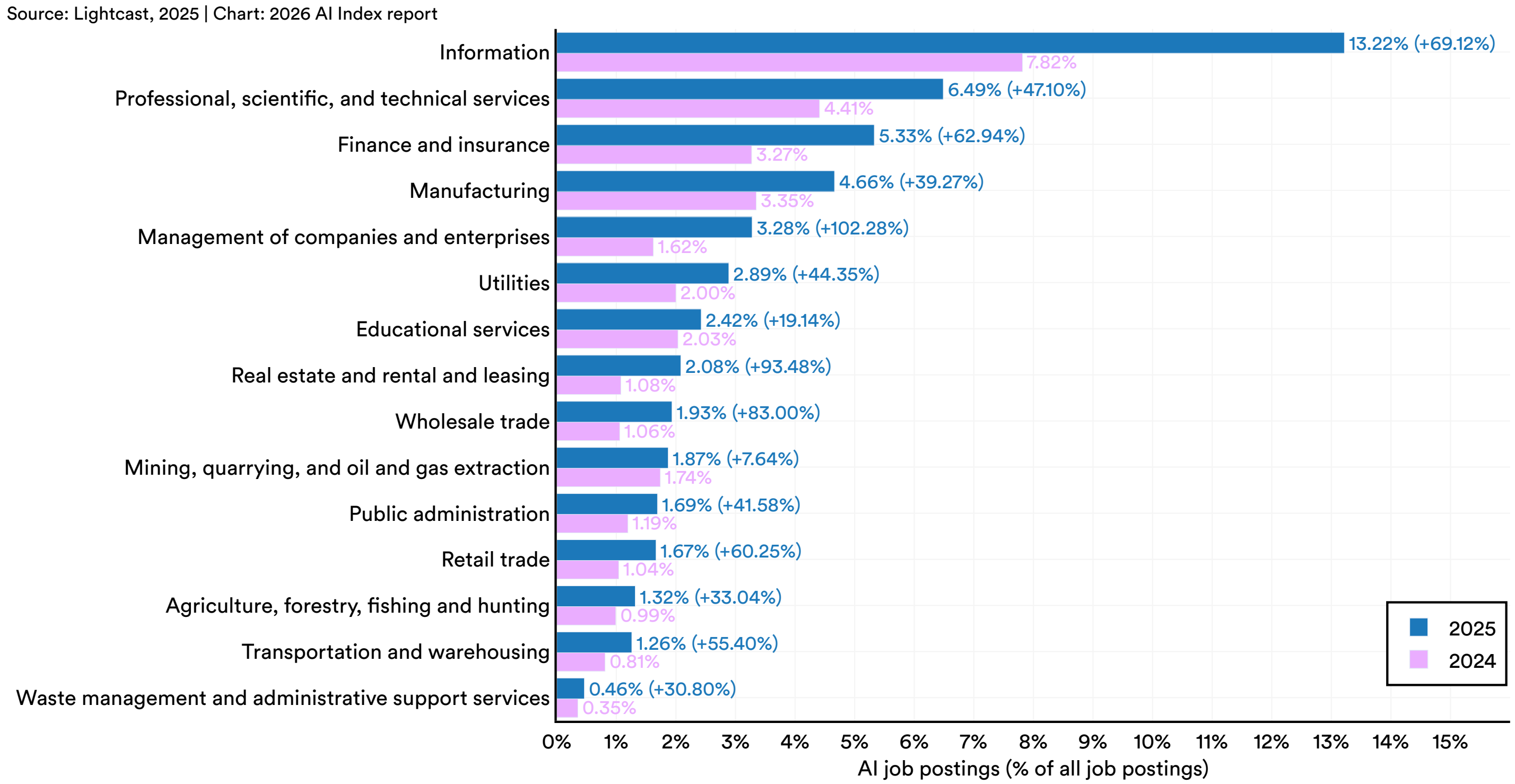
AI job postings (% of all job postings) in the United States by sector, 2024 vs. 2025
Source: Lightcast, 2025 | Chart: 2026 AI Index report

Figure 4.4.9[11]

11 The sector classifications in Figure 4.2.9 are based on two-digit NAICS codes. For more information on the Bureau of Labor Statistics' supersector and NAICS classifications, see the following reference.

## By State

Within the United States, AI labor demand remains highly concentrated within certain states (Figures 4.4.10 and 4.4.11). California leads with 170,881 postings and accounts for a disproportionate share of total 2025 U.S. AI job postings (17.2%). Texas follows with 80,547 postings (8.1% of total) and New York with 66,029 (6.6%). These three states represent approximately a third of all AI job postings nationally. This mirrors the state-level investment data described earlier, where AI funding was also concentrated in a small number of states, led by California. Over time, however, California's share of the national total has declined, from over 25% in 2012 to 17.9% in 2025, even as the state continues to lead in AI investment concentration (Figure 4.4.12). However, looking at the density of AI labor demand within each state, there are instances of above-average AI penetration relative to the particular total job market (Figure 4.4.13). For example, despite having smaller total numbers, Washington, D.C., accounts for a comparatively high 6.2% share of those postings followed by Delaware at 4.4%. From 2024 to 2025, California, Washington state, New York, and Texas all continued to see growth in AI job postings within their labor markets (Figure 4.4.14).

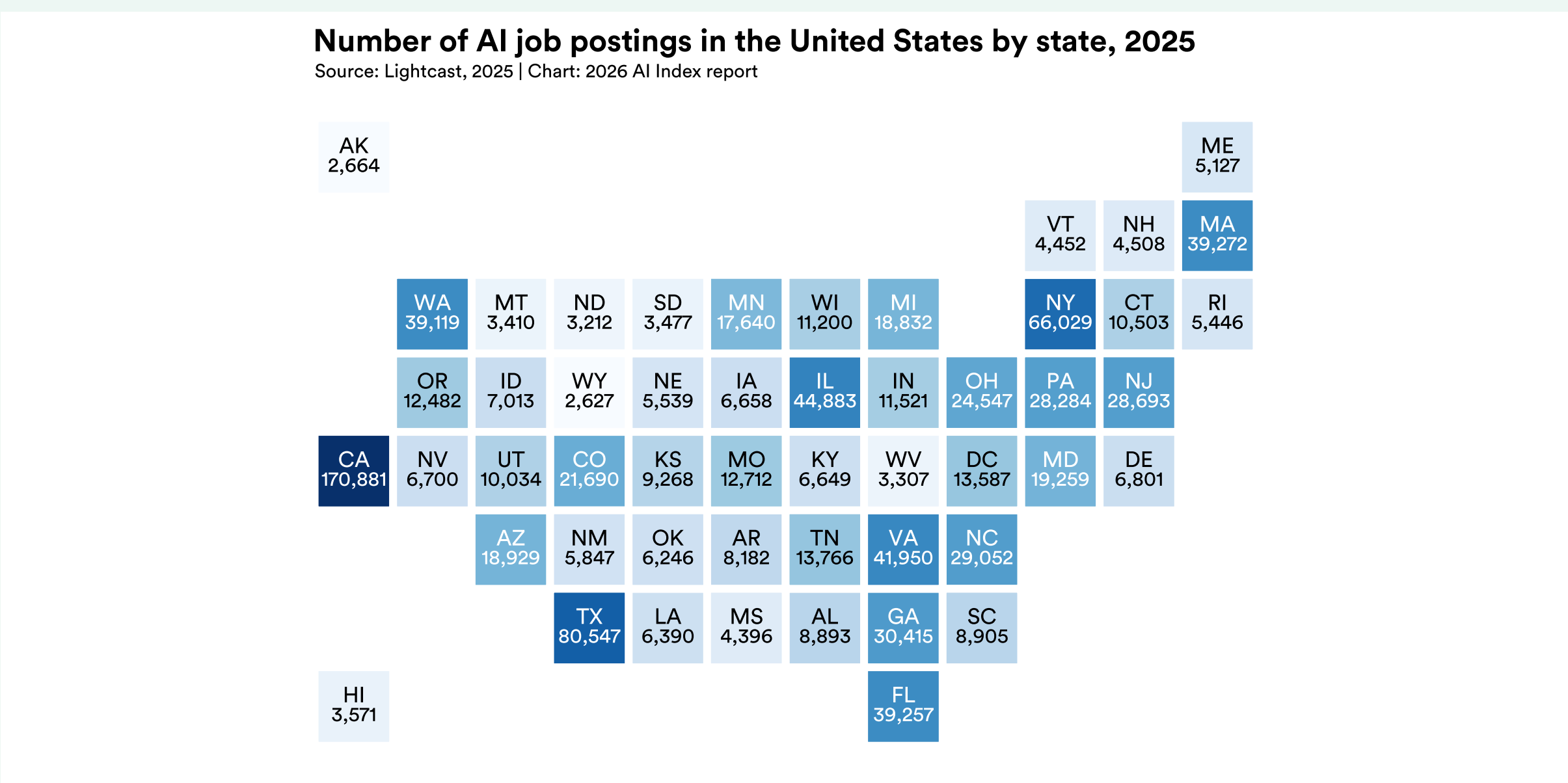

Figure 4.4.10

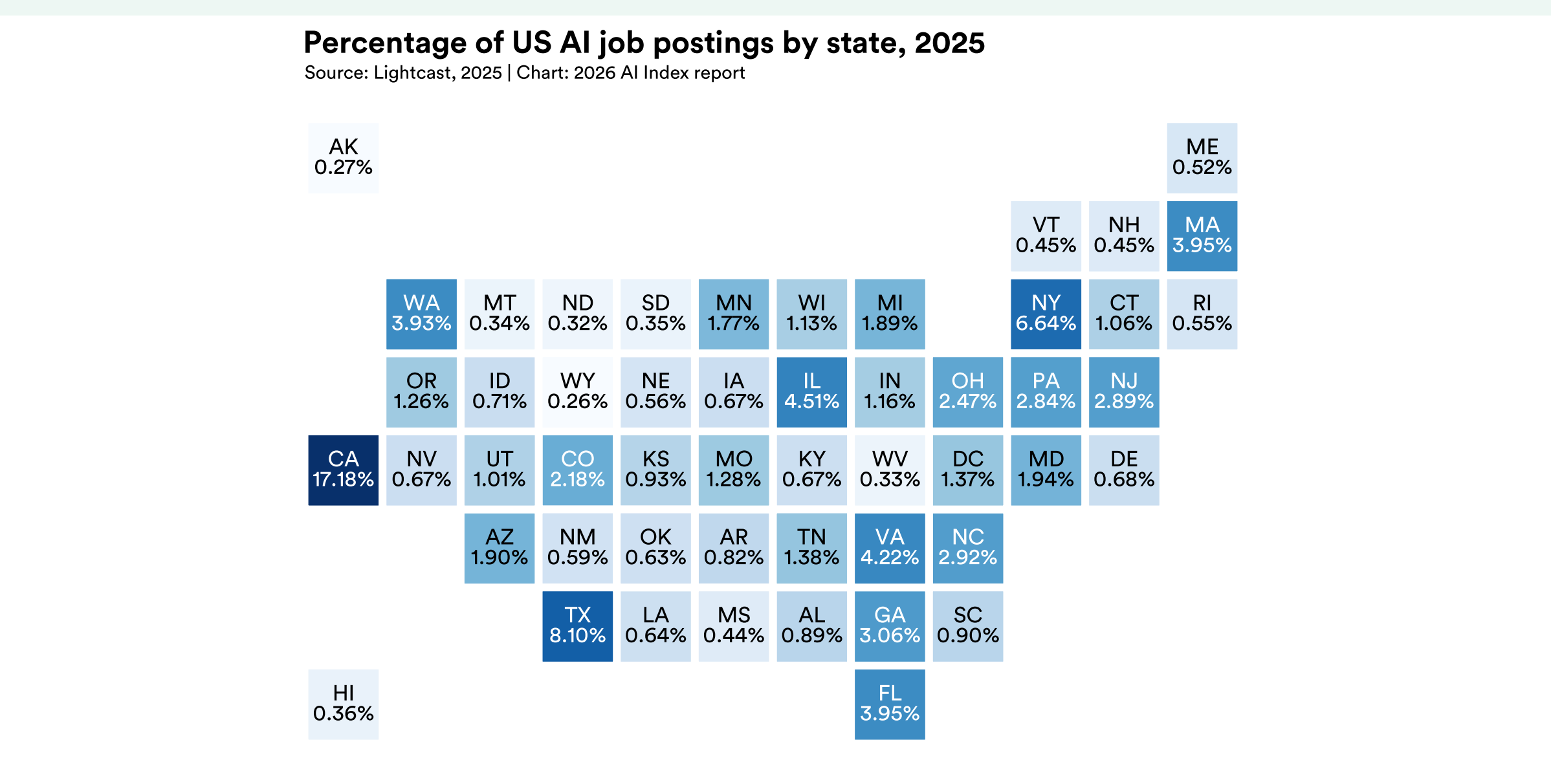

Figure 4.4.11

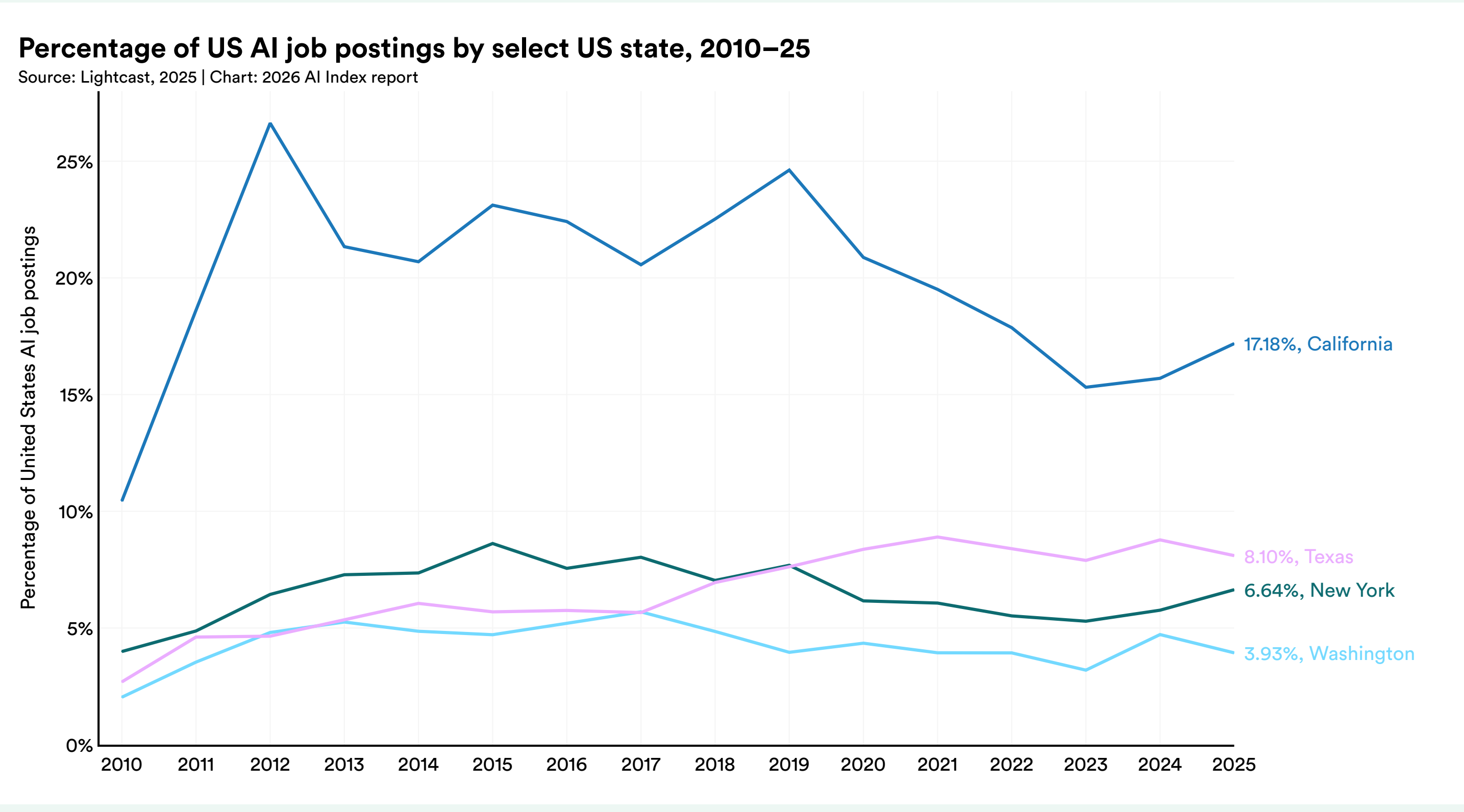

Figure 4.4.12

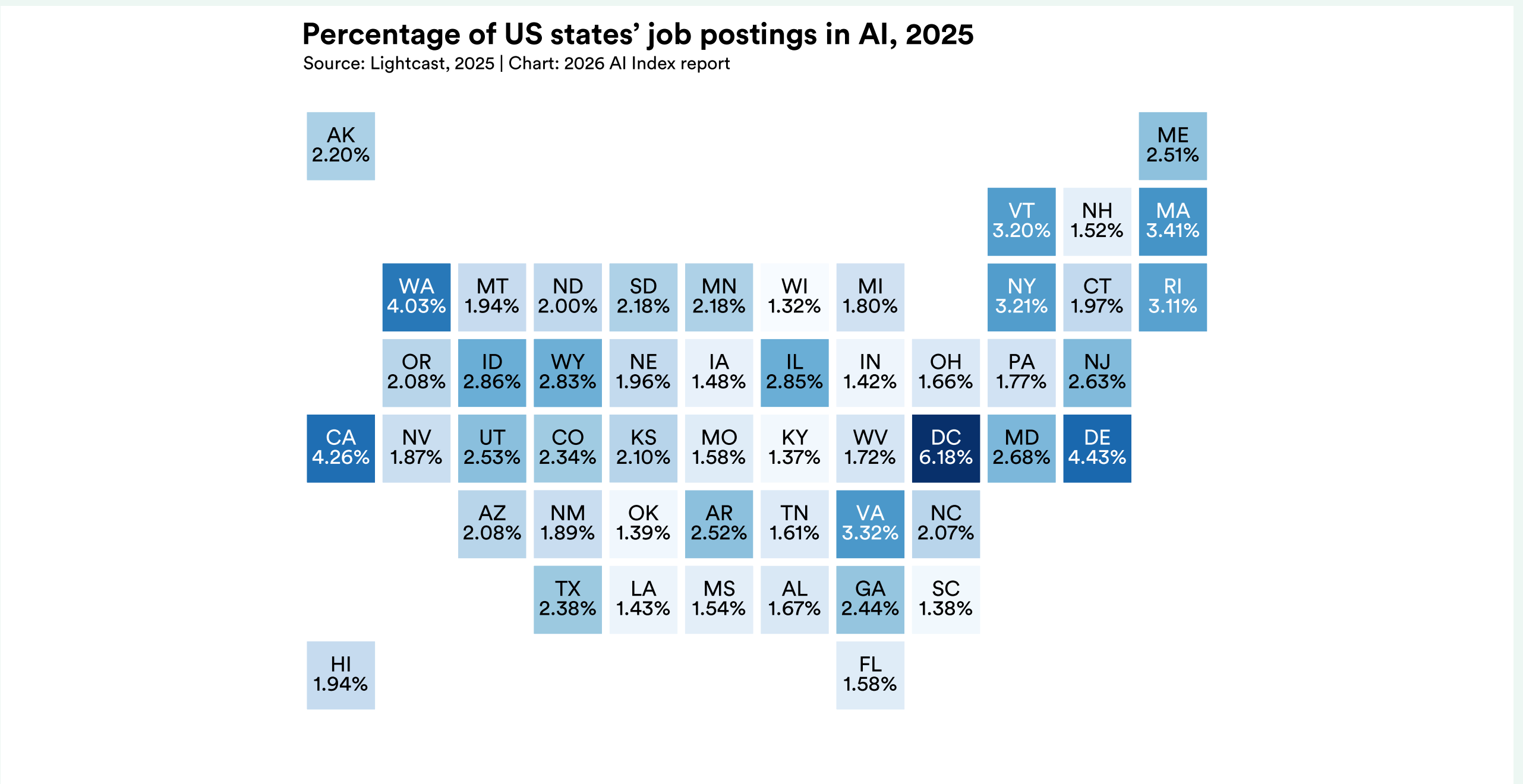

Figure 4.4.13

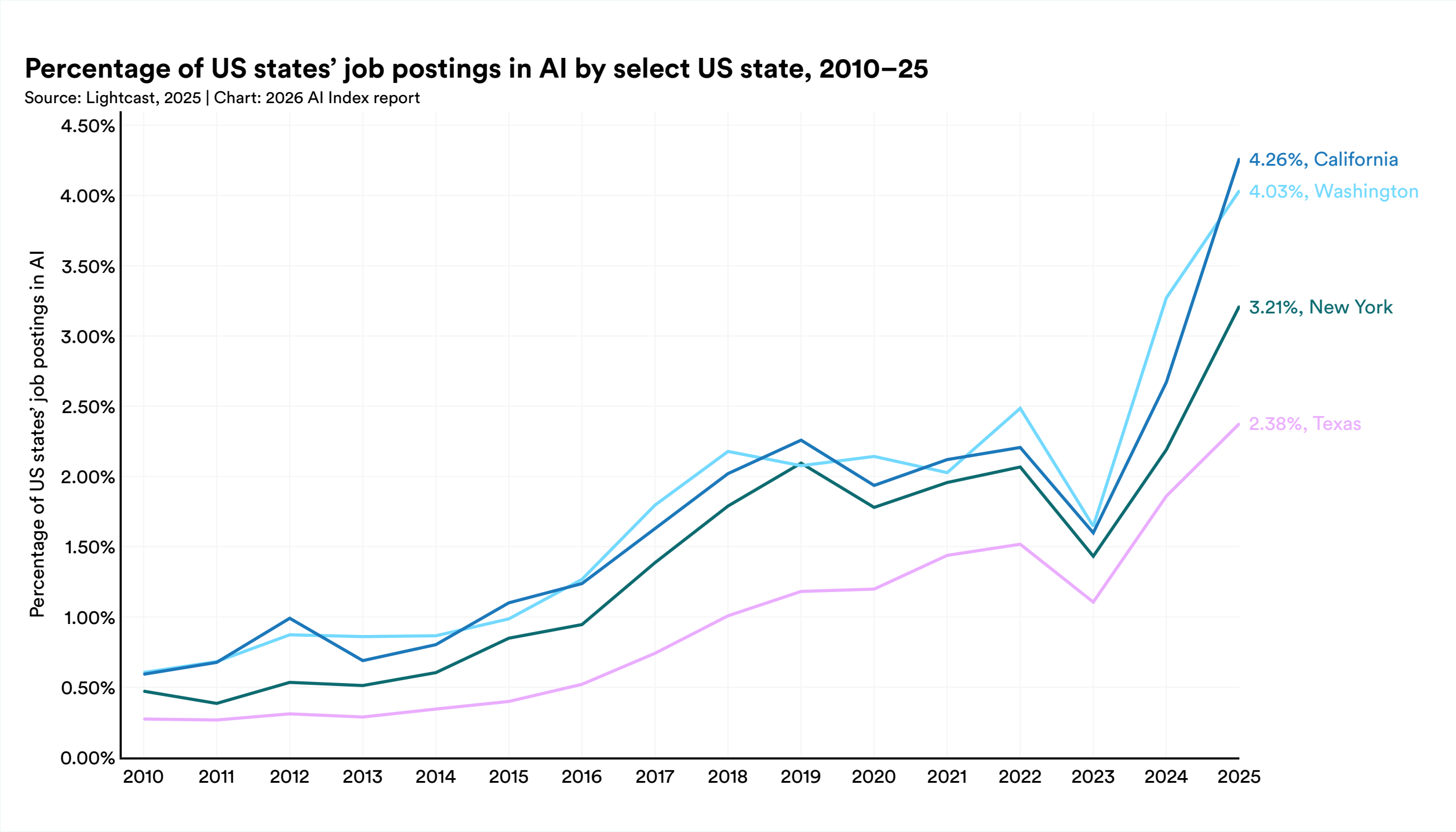

Figure 4.4.14

## AI Hiring

LinkedIn's hiring and talent data provides a view of how AI labor demand is changing the actual workforce in practice. In most countries, AI hiring rates outpaced overall hiring growth in 2025 (Figures 4.4.15 and 4.4.16). Indonesia recorded the highest relative AI hiring growth at 31.7%, followed by Croatia (27.8%) and Belgium (21.5%). Since 2018, many countries show a sustained pattern of AI hiring rates that exceed general labor market growth. However, there are a few exceptions, such as Iceland and Sweden, where AI hiring growth lagged behind the broader market.

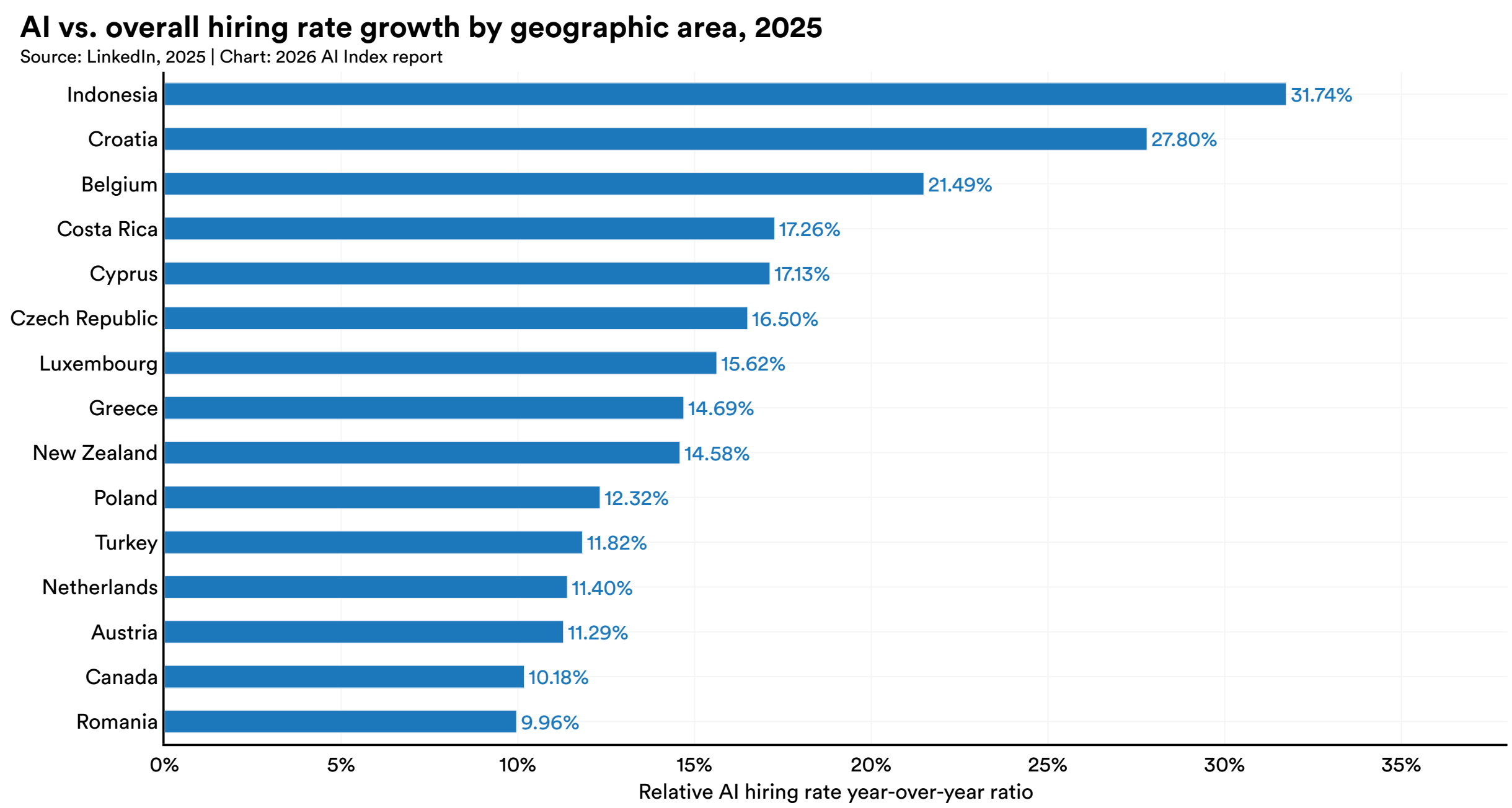

Figure 4.4.15[12]

12 For the sake of brevity, the visualization includes only the top 15 countries for this metric.

## AI vs. overall hiring rate growth by geographic area, 2018–25

Source: LinkedIn, 2025 | Chart: 2026 AI Index report

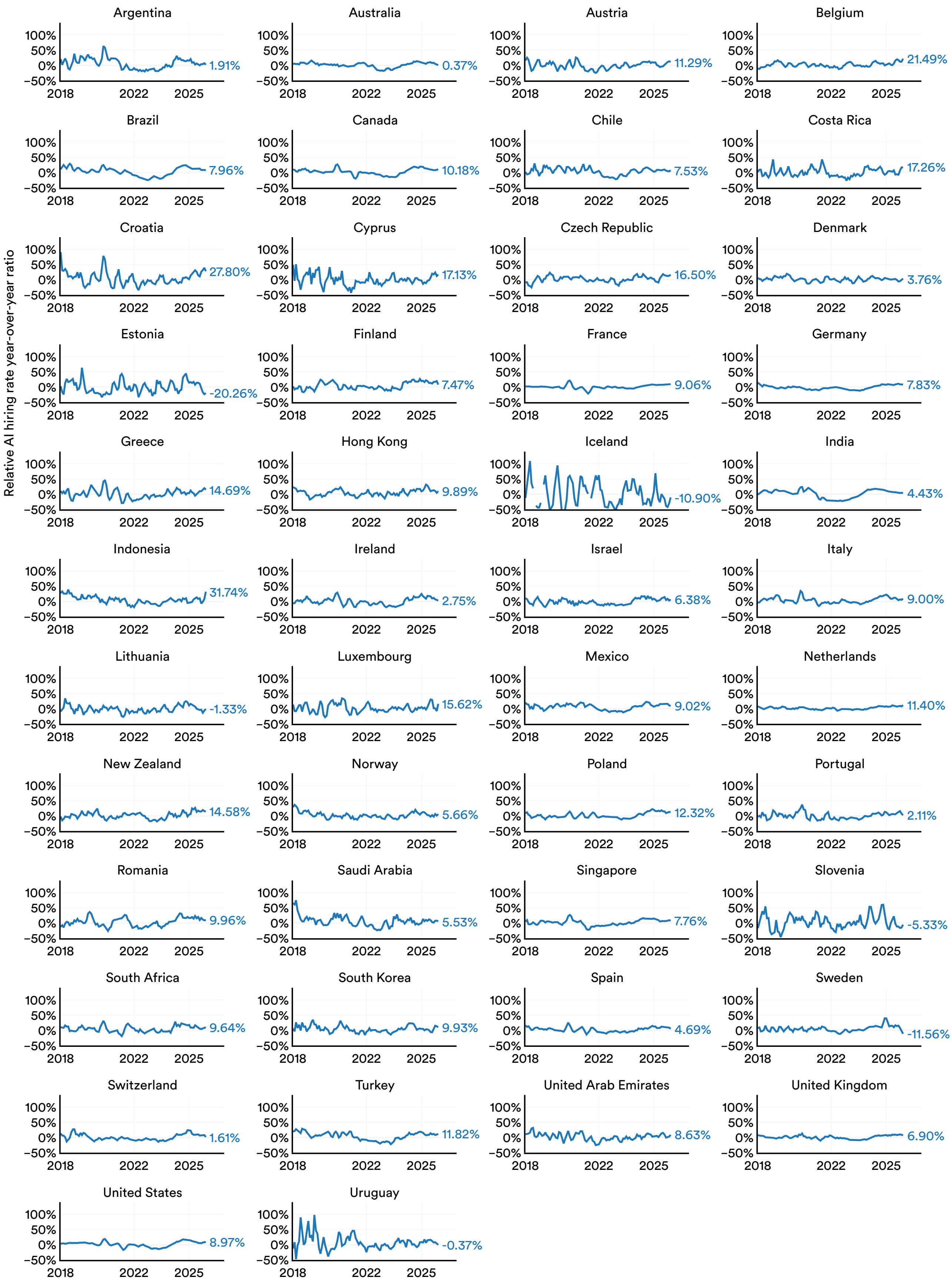

Figure 4.4.16

## AI Talent Concentration

Talent data helps show where AI capabilities are accumulating and how the AI workforce is distributed globally. This section reviews LinkedIn's measures on the concentration of AI talent within countries and the movement of that talent across borders. In 2025, Israel had the highest concentration of AI talent among LinkedIn members (2.1%), followed by Singapore (1.8%) and Luxembourg (1.6%) (Figures 4.4.21 and 4.4.22). However, the United Arab Emirates, India, and Saudi Arabia showed the fastest growth in their share of AI talent, each increasing over 100% between 2019 and 2025. Over the same time period, all countries in the sample grew by at least 75% in AI talent concentration.

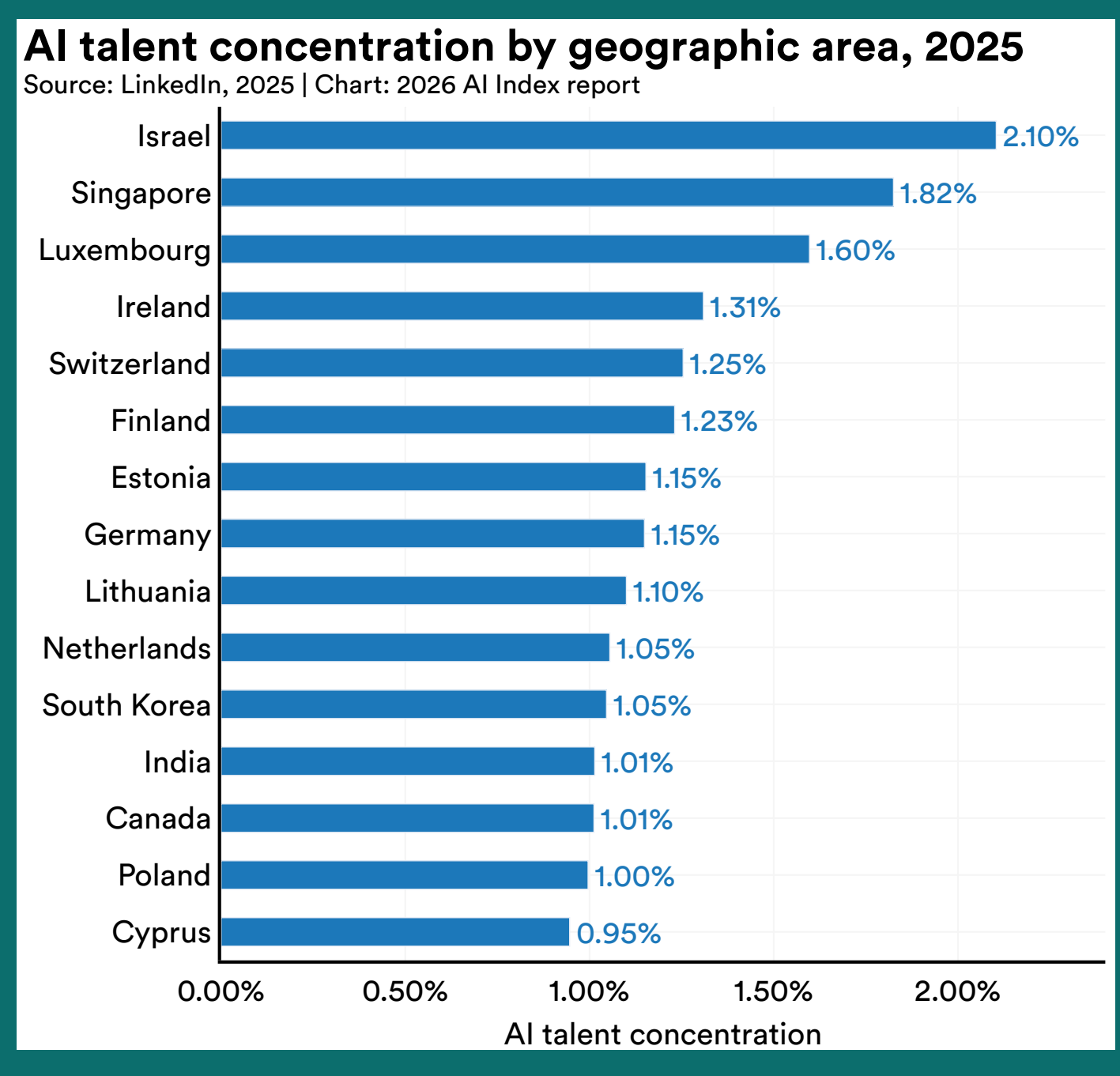

Figure 4.4.21[13]

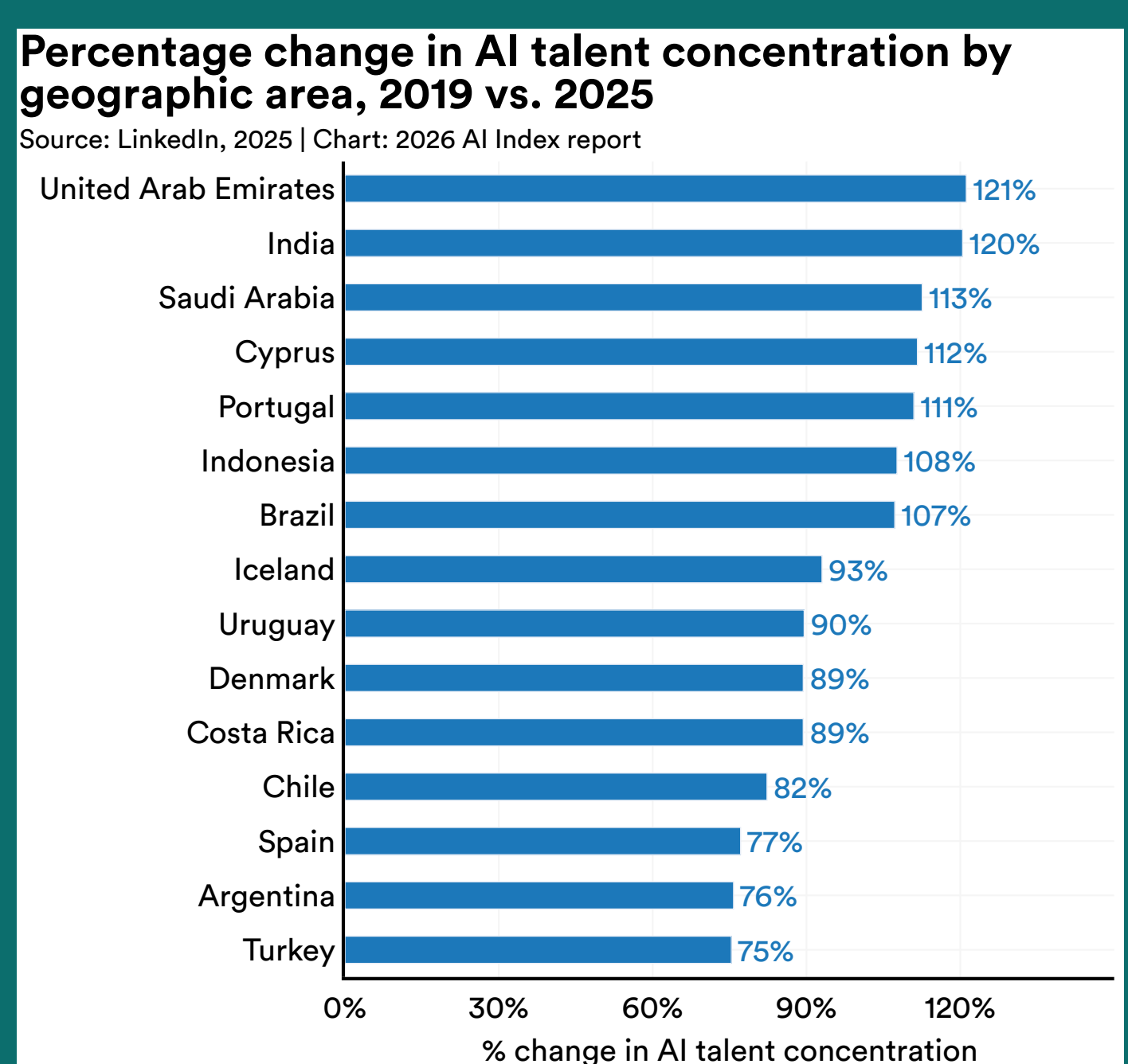

Figure 4.4.22[14]

13 For the sake of brevity, the visualization includes only the top 15 countries for this metric.

14 For the sake of brevity, the visualization includes only the top 15 countries for this metric.

Migration patterns show the dynamic global redistribution of AI talent (Figure 4.4.23 and 4.4.24). In 2025, Luxembourg recorded the highest net inflow relative to other tracked countries, with 5.23 per 10,000 LinkedIn members, as smaller economies actively try to attract more AI workers. The United States is a net importer of AI talent at 1.2 per 10,000 LinkedIn members. Similar to skills penetration, gender representation within AI talent continues to be uneven. Across the countries measured, men still account for the majority of AI talent, typically between 65% and 75% (Figures 4.4.25 and 4.4.26). Gender ratios have for the most part stayed flat since 2016, despite an expanding AI workforce. In the United States, women represent a 34.3% share of AI talent compared to men's 65.7% share. Other major labor markets, including the United Kingdom, Canada, France, and Singapore, show similar disparities.

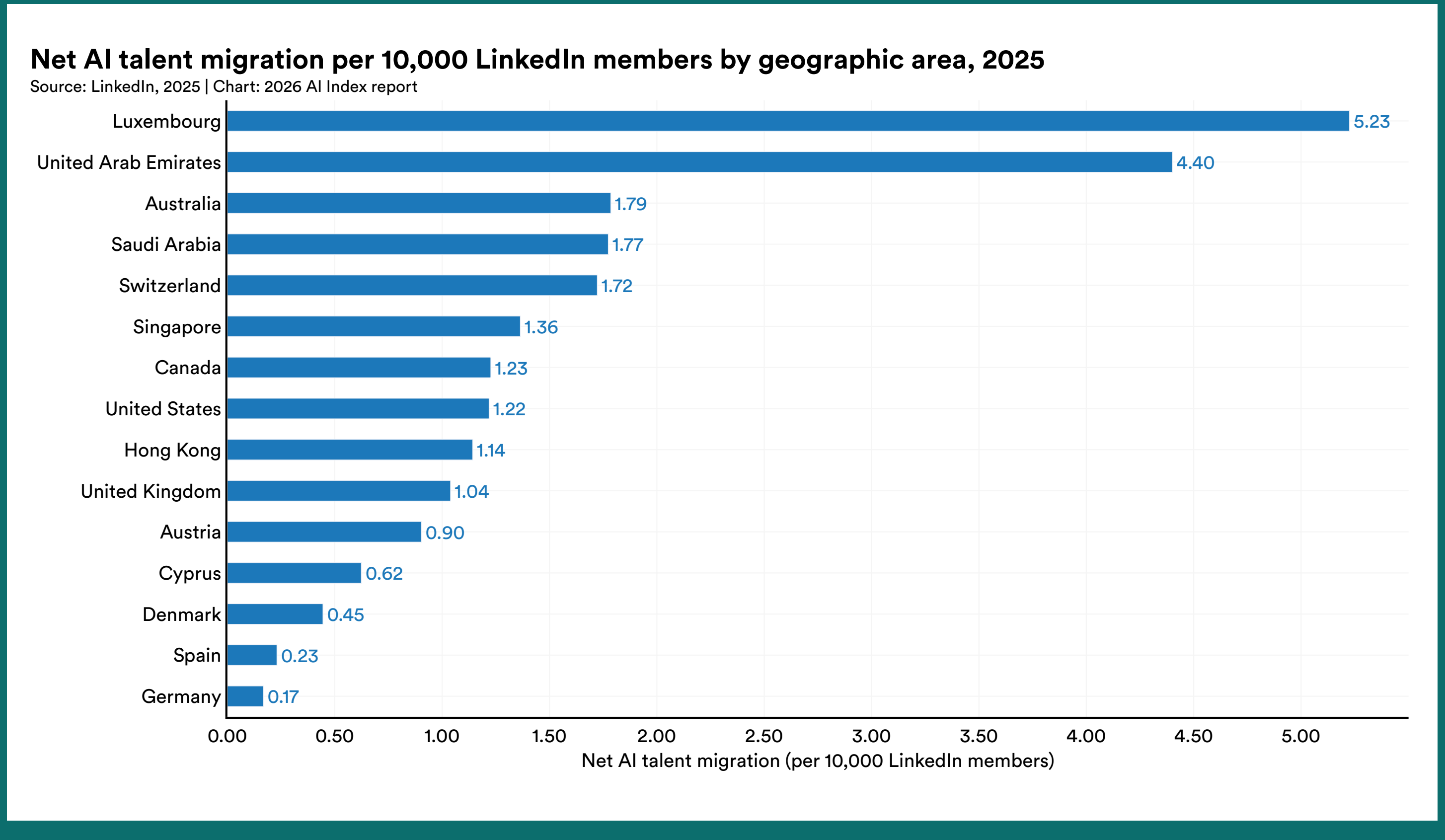

Figure 4.4.23[15]

15 For the sake of brevity, the visualization includes only the top 15 countries for this metric.

**Net AI talent migration per 10,000 LinkedIn members by geographic area, 2021–25**
Source: LinkedIn, 2025 | Chart: 2026 AI Index report

Net AI talent migration (per 10,000 LinkedIn members)

Argentina 0.01
Australia* 1.77
Austria* 1.50
Belgium 0.25
Brazil -0.05
Bulgaria* -1.04
Canada* 0.07
Chile -0.07
Costa Rica -0.02
Croatia -0.07
Cyprus* 2.93
Czech Republic 0.46
Denmark 0.83
Estonia* -0.51
Finland 0.66
France 0.16
Germany* 1.09
Greece -0.37
Hong Kong* 1.76
Hungary* -2.74
Iceland 0.75
India* -1.28
Indonesia -0.09
Ireland* 2.48
Israel* -1.01
Italy -0.21
Latvia 0.11
Lithuania* 0.74
Luxembourg* 7.76
Mexico -0.07
Netherlands 0.24
New Zealand -0.06
Norway 0.28
Poland 0.49
Portugal 0.20
Romania -0.06
Saudi Arabia* 2.02
Singapore* 0.90
Slovakia* -1.78
Slovenia 0.09
South Africa -0.02
South Korea -0.35
Spain 0.54
Sweden 0.08
Switzerland* 2.45
Turkey -0.30
United Arab Emirates* 6.52
United Kingdom* 0.62
United States 0.92
Uruguay 0.16

2021 2023 2025

Figure 4.4.24[16]

16 Asterisks indicate that a country's y-axis label is scaled differently than the y-axis label for the other countries.

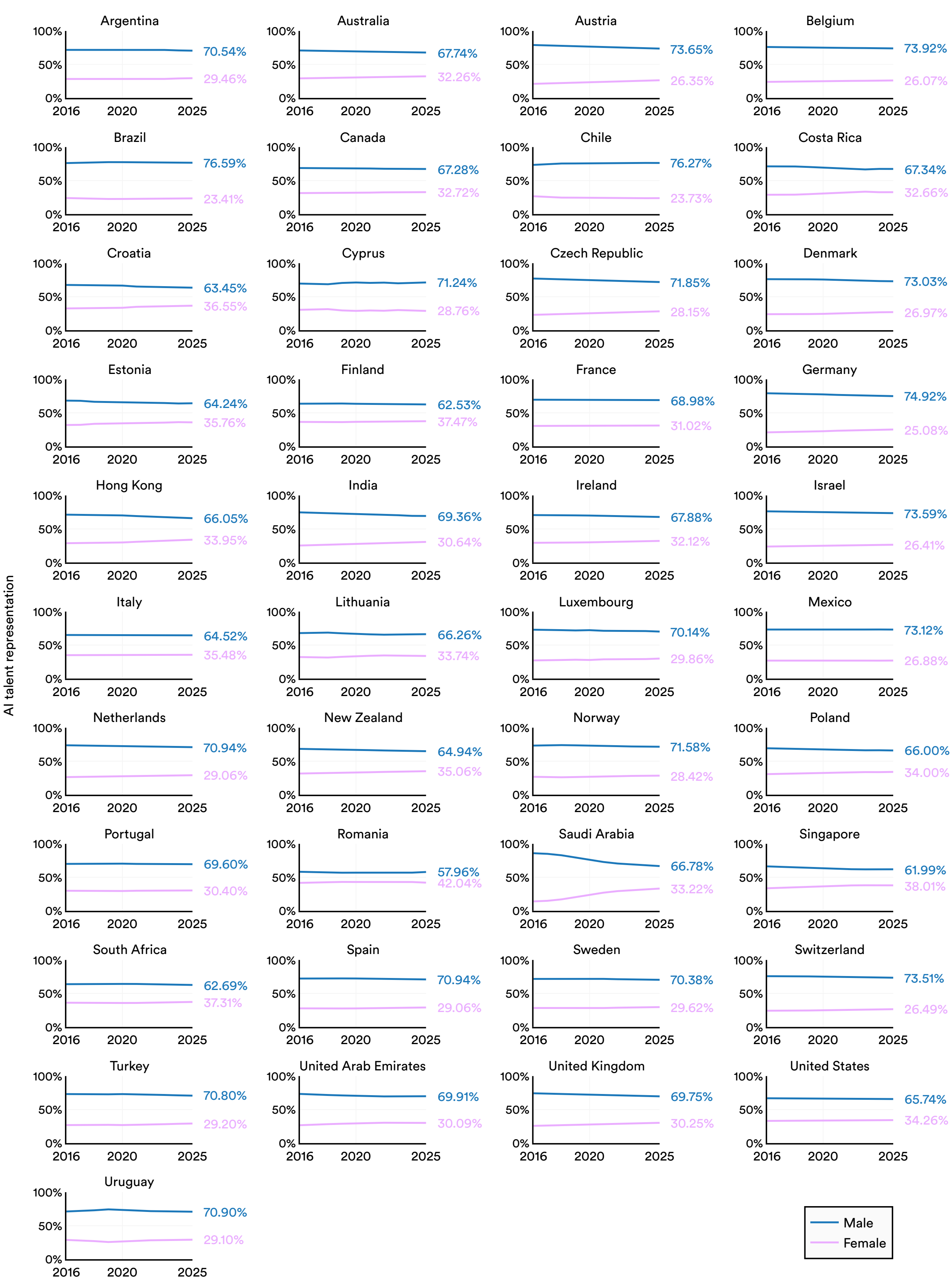

AI talent representation by gender and geographic area, 2016–25
Source: LinkedIn, 2025 | Chart: 2026 AI Index report
AI talent representation
Argentina 70.54% 29.46%
Australia 67.74% 32.26%
Austria 73.65% 26.35%
Belgium 73.92% 26.07%
Brazil 76.59% 23.41%
Canada 67.28% 32.72%
Chile 76.27% 23.73%
Costa Rica 67.34% 32.66%
Croatia 63.45% 36.55%
Cyprus 71.24% 28.76%
Czech Republic 71.85% 28.15%
Denmark 73.03% 26.97%
Estonia 64.24% 35.76%
Finland 62.53% 37.47%
France 68.98% 31.02%
Germany 74.92% 25.08%
Hong Kong 66.05% 33.95%
India 69.36% 30.64%
Ireland 67.88% 32.12%
Israel 73.59% 26.41%
Italy 64.52% 35.48%
Lithuania 66.26% 33.74%
Luxembourg 70.14% 29.86%
Mexico 73.12% 26.88%
Netherlands 70.94% 29.06%
New Zealand 64.94% 35.06%
Norway 71.58% 28.42%
Poland 66.00% 34.00%
Portugal 69.60% 30.40%
Romania 57.96% 42.04%
Saudi Arabia 66.78% 33.22%
Singapore 61.99% 38.01%
South Africa 62.69% 37.31%
Spain 70.94% 29.06%
Sweden 70.38% 29.62%
Switzerland 73.51% 26.49%
Turkey 70.80% 29.20%
United Arab Emirates 69.91% 30.09%
United Kingdom 69.75% 30.25%
United States 65.74% 34.26%
Uruguay 70.90% 29.10%
100%
50%
0%
2016
2020
2025
Male
Female

Figure 4.4.25

### AI talent concentration by gender and geographic area, 2016–25

Source: LinkedIn, 2025 | Chart: 2026 AI Index report

Figure 4.4.26

## AI's Labor Impact

Shifts in skill demand and talent flows are reshaping how work is organized by driving changes in productivity and hiring. A growing body of academic research has begun to measure AI's impact both at the micro level, examining how individual workers perform their jobs using AI tools, and at the macro level, examining how AI adoption impacts aggregate productivity and employment figures. In some settings, AI improves productivity, particularly for tasks that are structured, language heavy, or supported by clear feedback loops. In others, gains are marginal or even negative when tools are poorly matched to the task. Early macro-level evidence indicates that productivity gains may take longer to materialize and that the labor market costs may fall disproportionately on junior and entry-level workers. Findings from several notable studies are summarized in the tables below; as with any emerging research area, results vary in methodology, scope, and context.

## Productivity Trends

### Micro-level Studies

A growing number of studies have looked at how AI tools affect individual worker productivity across occupations (Figure 4.4.27). The results have been generally positive, but the size and distribution of the gains varies. The clearest impact is on support work, software development, and marketing. Customer support agents using a conversational AI assistant resolved 14%–15% more issues per hour (Brynjolfsson et al., 2025), software developers using GitHub Copilot completed 26% more pull requests (Cui et al., 2025), and marketing teams using multimodal AI for ad creation saw a 50% increase in output per worker (Ju and Aral, 2025). One consistent finding across several of these studies is that the less experienced workers tended to benefit the most, suggesting that AI tools may help close existing skill gaps.

Results are not uniformly positive, and for work that requires deeper reasoning or judgment, some studies have found AI tools produced little benefit or even slowed workers down. The most widely cited example comes from Model Evaluation & Threat Research (METR), which found that experienced open-source developers became 19 percent slower when using AI assistance, with a disconnect between how helpful the developers thought the tools were and how they actually performed (Becker et al., 2025). However, the METR team has not been able to replicate the results in a later study, primarily due to a growing reluctance among developers to work without AI, and that developers in late 2025 were likely sped up by AI relative to the original study period. When looking at the longer-term effects on skill development, research shows mixed results. Software engineers who relied heavily on AI for learning showed no measurable speed improvement and faced what researchers call "learning penalties" (Shen and Tamkin, 2025). Overall, AI's productivity effects are highly context dependent. The gains are strongest when work can be divided into well-defined, repeatable tasks with clear quality monitoring.

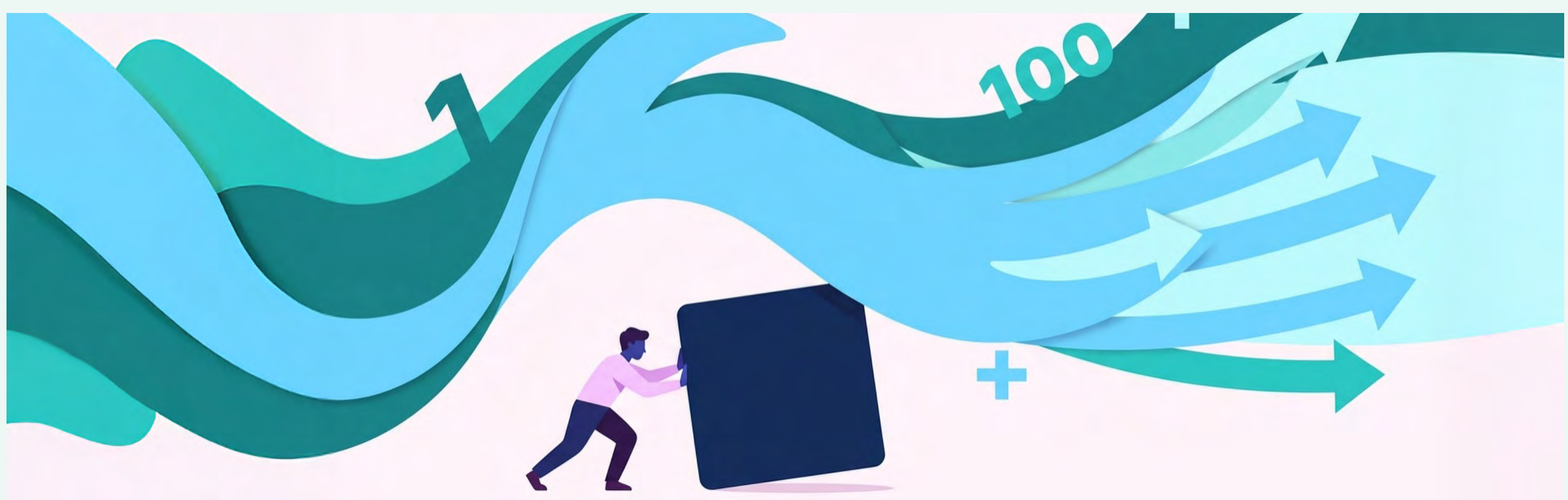

| Study | Occupation | AI application | Change in productivity | Who benefited most? |
| --- | --- | --- | --- | --- |
| Reimers and Waldfogel (2026) | Authors | LLMs for content | +200% (Output volume; releases tripled) | New entrants (drove quantity); pre-AI authors (maintained quality) |
| Shen and Tamkin (2025) | Software engineers | Learning new libraries | 0% (Statistically insignificant speed change) | High scorers (65%+) who used AI for conceptual inquiry, avoiding "learning penalties" |
| Becker et al. (2025) | Developers | Open-source tools | -19% (Speed; developers became slower with AI) | None (significant gap between perceived help and actual performance) |
| Brynjolfsson et al. (2025) | Support agents | Conversational assistant | +14%–15% (Efficiency; issues resolved per hour) | Less experienced/skilled agents (30%–35% gains) |
| Cui et al. (2025) | Software developers | GitHub Copilot | +26% (Output; completed pull requests) | Junior and less-experienced workers |
| Ju & Aral (2025) | Marketing teams | Multimodal ad creation | +50% (Output; productivity per worker) | Human-AI teams (shifted focus from social coordination to task execution) |
| Choi & Xie (2025) | Accountants | AI-based accounting | +55% (Throughput; weekly client support) | Experienced accountants (used AI confi-dence scores to target oversight) |

Figure 4.4.27

## Macro-level Studies

At the macro level, the evidence is earlier and less conclusive, but there are indicators that AI is starting to register in aggregate productivity data (Figure 4.2). A study of 12,000 European firms found that AI adoption boosted labor productivity by 4%, with training strengthening the outcome (Aldasoro et al., 2026). In the United States, productivity growth reached 2.7% in 2025, nearly double the 1.4% average of the previous decade. Brynjolfsson (2026) explains this may reflect the early stages of a "J-curve," where organizations absorb the costs of adopting AI before the larger productivity gains show up in the numbers. Similarly, OECD projections for G7 economies estimate annual productivity gains of 0.2 to 1.3 percentage points over the next decade (Filippucci et al., 2025). As mentioned earlier, the evidence is not conclusive nor is it all positive. A survey of 6,000 executives across four countries found widespread adoption but minimal realized productivity gains and a projected 0.7% reduction in employment over the next three years (Yotzov et al., 2026). The gap between adoption and measurable impact could be because AI is still early in its organizational integration, as illustrated earlier in this chapter through the deployment stage data. The pace at which these returns materialize will continue to be an important indicator to track.

| Study | Scope | Insight | Productivity / employment impact |
|---|---|---|---|
| Aldasoro et al. (2026) | 12,000 European firms (2019–2024) | AI adoption increases efficiency without reducing short-run employment; training significantly boosts gains. | +4% increase in labor productivity; +5.9 percentage point gain for every 1% spent on training. |
| Yotzov et al. (2026) | 6,000 executives (US, UK, DE, AU) | Representative survey showing high adoption but minimal realized impact on productivity to date. | +1.4% projected productivity boost; +0.8% projected output increase; -0.7% projected employment reduction (over next 3 years). |
| Brynjolfsson (2026) | United States economy | A "decoupling" of output from labor input is visible; framed through the "J-curve" hypothesis. | 2.7% US productivity growth in 2025 (nearly double the 1.4% annual average in the previous decade). |
| Filippucci et al. (OECD, 2025) | G7 economies (10-year horizon) | Projected annual gains based on sectoral specialization (e.g., high in finance/ICT, low in manufacturing). | +0.4 to +1.3 pp (US/UK) vs. +0.2 to +0.8 percentage points (Italy/Japan) annual labor productivity growth. |
| Frank et al. (2026) | LinkedIn profiles and US unemployment insurance records | Deterioration in AI-exposed labor markets (unemployment risk) began in early 2022, pre-ChatGPT. | Negative entry rates for AI-exposed roles. |
| Brynjolfsson et al. (2025) | US payroll data (ADP) through 2025 | "Canaries in the coal mine": large employment declines for junior workers in exposed fields. | -15% to -16% employment for early-career workers. |
| Hosseini Maasoum and Lichtinger (2025) | 62 million workers/ 285,000 US firms | "Seniority-biased technological change"; AI substitutes for junior labor while leaving senior roles intact. | Sharp decline in junior employment driven by slower hiring. |
| St. Louis Fed (2025) | US general labor market | Back-of-envelope calculations based on self-reported time savings. | +1.1% to +1.3% labor productivity increase. |
| Penn Wharton Budget Model (2025) | US economy | Projects AI's current contribution to total factor productivity (TFP). | +0.01 percentage points contribution to TFP (negligible). |

Figure 4.4.28

## Workforce Impact

It is challenging to measure AI's impact on employment, particularly because the technology is still in the early stages of widespread deployment. So far, effects on the workforce appear to be uneven, initially showing up in hiring pipelines, among younger workers, and within specific business functions. The evidence does not point to broad, uniform displacement. Firm-level survey data does suggest that many organizations expect the pace of workforce change to accelerate over the next year. According to McKinsey's survey, respondents also anticipated headcount reductions for the coming year to exceed those reported in the past year. (Figures 4.4.30 and Figure 4.4.31).

Recent employment data for software developers and customer service roles reveals a generational pattern (Brynjolfsson et al., 2025). In the United States, normalized headcount trends for both occupations show that employment among the youngest workers (ages 22–25) has declined since 2022, even as headcount for older age groups continues to grow (Figure 4.4.29). By September 2025, employment for software developers ages 22–25 had fallen close to 20% from its 2022 peak.

**Normalized headcount trends by age group for software developers and customer service agents, 2021–25**
Source: Brynjolfsson et al., 2025 | Chart: 2026 AI Index report

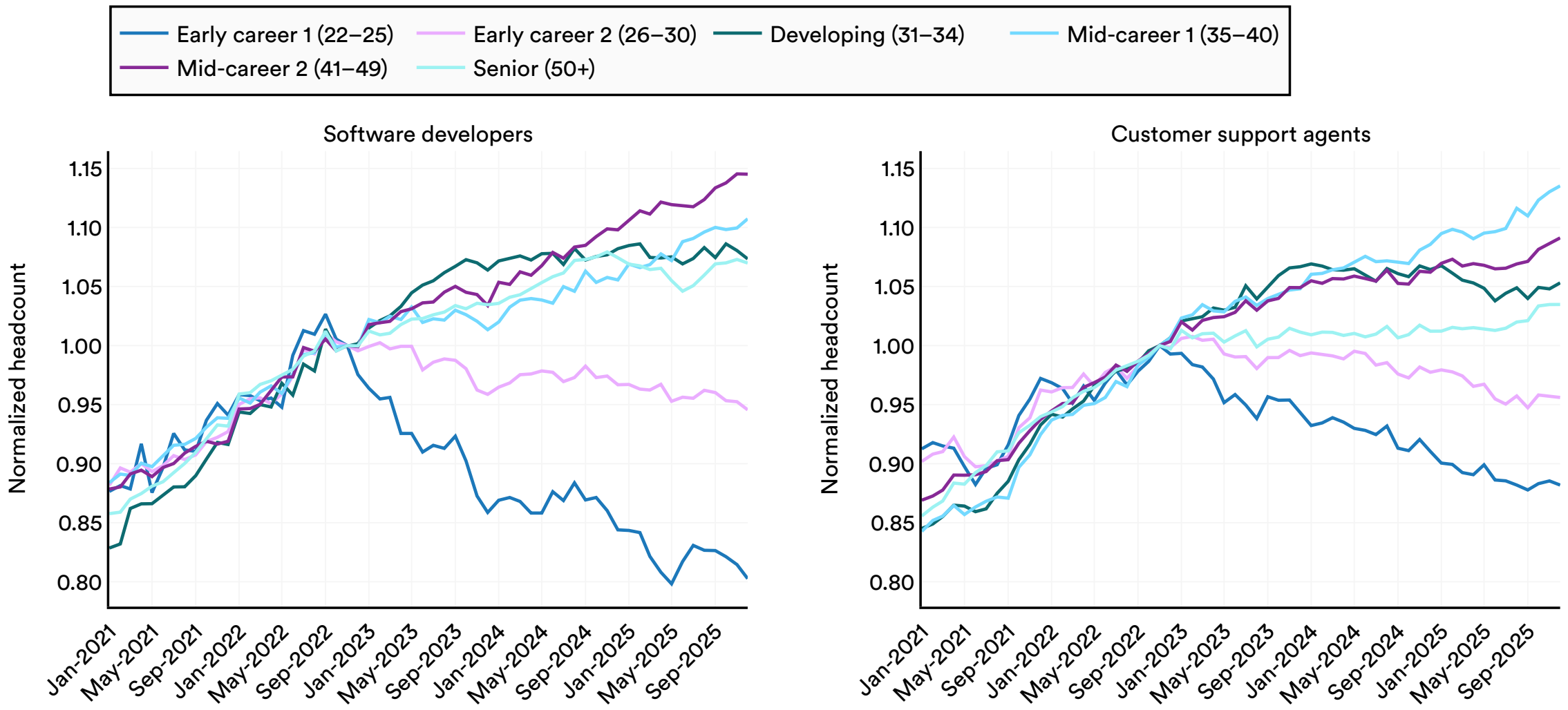

Figure 4.4.29

When occupations are grouped by their exposure to AI, the age-based pattern holds (Figure 4.4.30). Among workers ages 22–25, employment in the most AI-exposed occupations has fallen roughly 16% relative to the least-exposed, after controlling for firm-type effects, which isolate AI exposure from broader shocks like interest rate pressure or sector slowdowns. The gap began widening in mid-2024 and has grown steadily since.

**Headcount trends in high AI-exposure jobs (early career 22–25), 2021–25**
Source: Brynjolfsson et al., 2025 | Chart: 2026 AI Index report

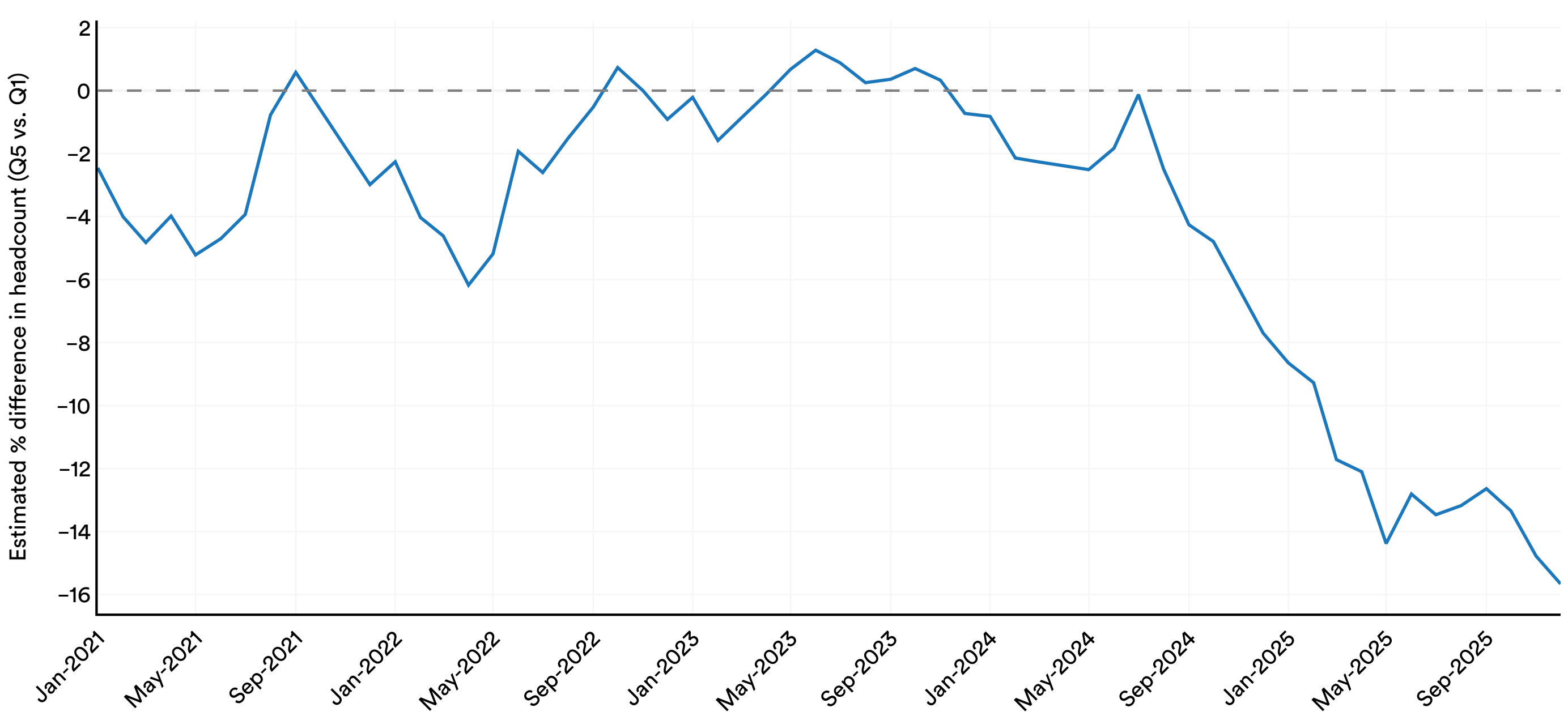

Figure 4.4.30[17]

17 This figure includes firm-by-time fixed effects (within-firm, same-period comparisons), accounting for firm/industry hiring swings.

Unemployment data suggests an even more complicated dynamic (Felten et al., 2021; Eckhardt and Goldschlag, 2025) (Figure 4.4.31). From 2022 to early 2025, unemployment rose across all occupation groups regardless of AI exposure level. While the unemployment rate for the most AI-exposed workers (quintile 5) increased by 0.30 percentage points, it rose even more for the least exposed workers (quintile 1), climbing by 0.94 percentage points. AI exposure alone does not seem to be driving recent unemployment trends, but it appears to play a part in broader macroeconomic conditions and organizational changes.

### Unemployment rate by AI exposure quintile

Source: Eckhardt and Goldschlag, 2025

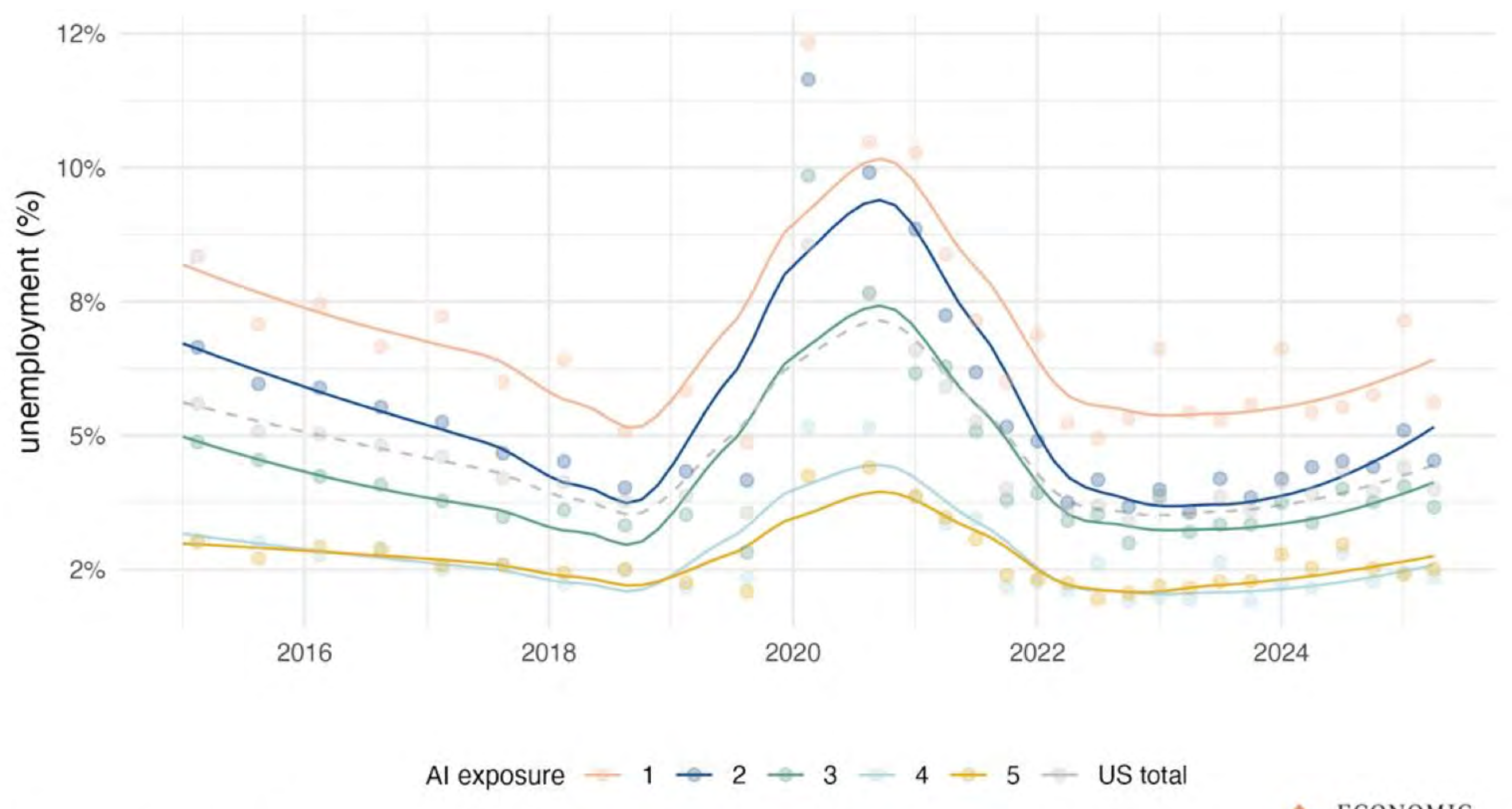

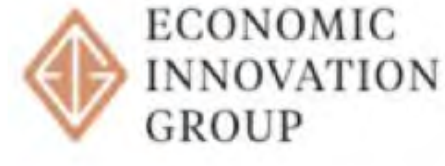

Figure 4.4.31

A different view emerges from looking at occupational churn trends after the introduction of major technologies (Figure 4.4.32). Over comparable time frames, the occupational mix in the United States has shifted faster since the introduction of generative AI than the shift that followed the introduction of computers or the internet (Gimbel et al., 2025).

**Changes in the occupational mix from recent baselines**

Source: Gimbel et al., 2025

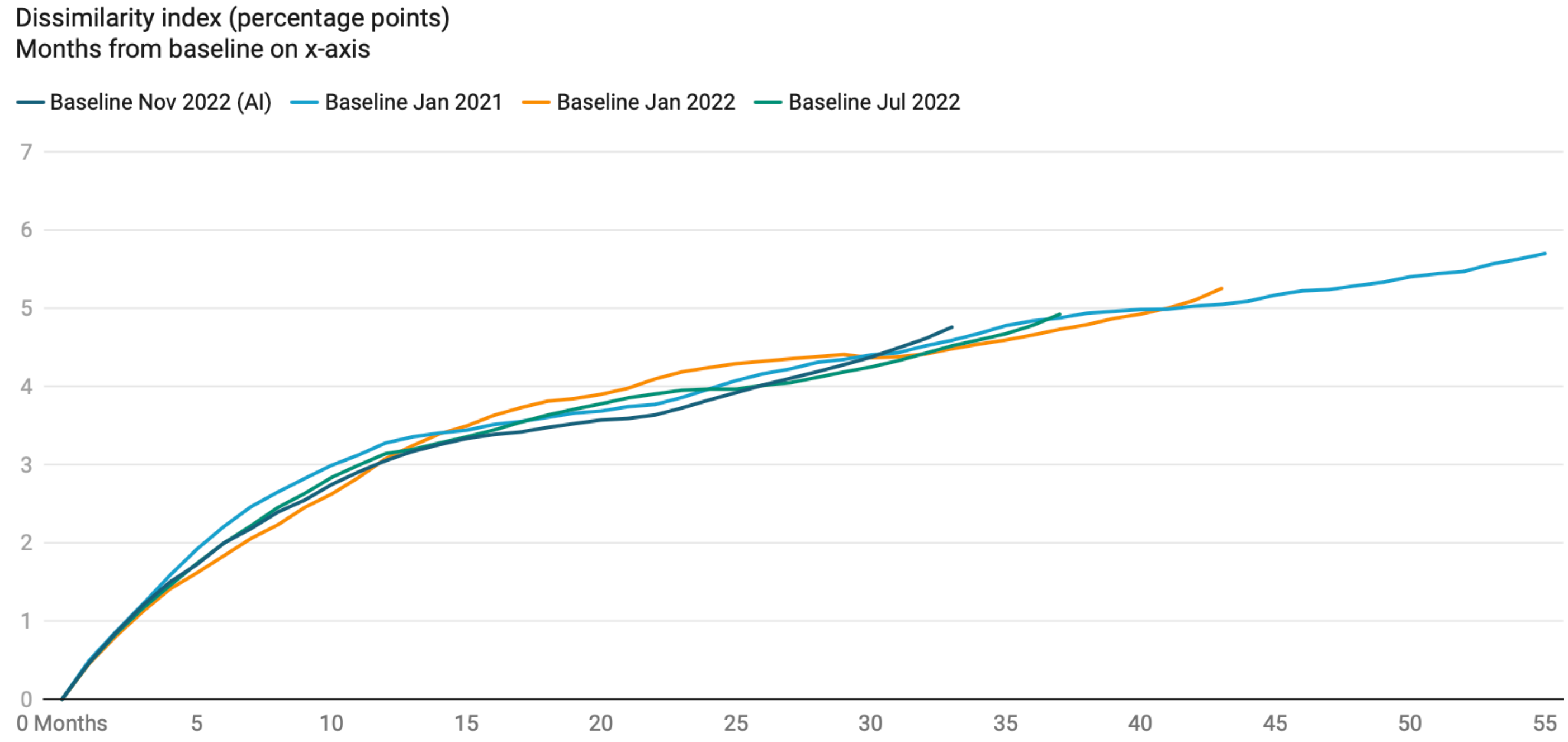

Figure 4.4.32

However, employer expectations seem to indicate that the pace may accelerate (Figure 4.4.33). According to McKinsey's survey (2025), one-third of respondents anticipate a decrease in workforce size, a percentage that is higher at larger organizations (35% at organizations with ≥$1 billion in revenue) compared to smaller firms (30% at organizations with <$1 billion in revenue). Most respondents (43% overall) expect little or no change, while only a minority foresee an increase in workforce size. Even compared to workforce changes that have already taken place, the general sentiment leans toward headcount reductions (Figure 4.4.34). In nearly all functions, respondents anticipate a greater impact from AI on headcount next year than was observed in the past year, with expected decreases outpacing observed decreases. This trend is particularly pronounced in service operations, supply chain/inventory management, marketing and sales, and software engineering, where the expected decrease in employees for the next year significantly exceeds the actual decrease reported over the past year. Conversely, expectations for workforce increases remain relatively modest across business functions.

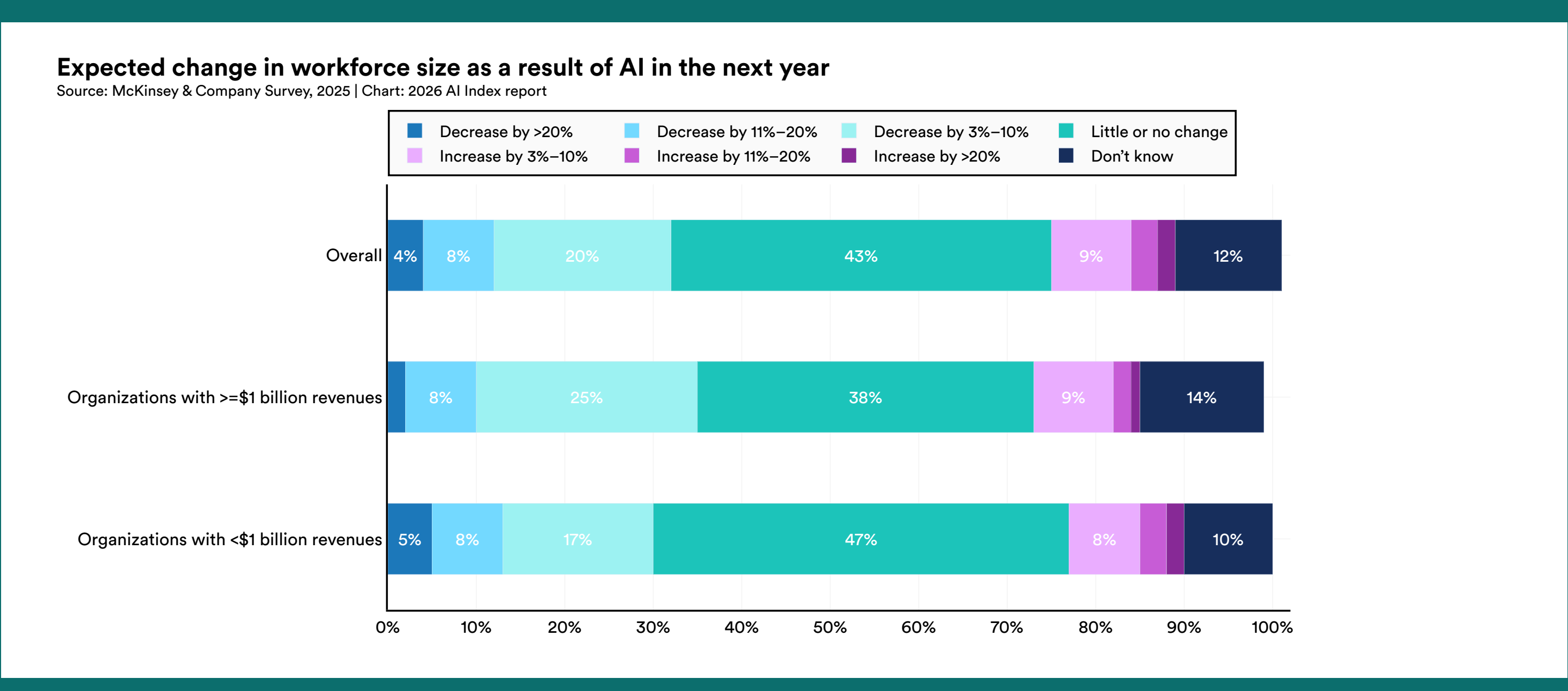

Figure 4.4.33

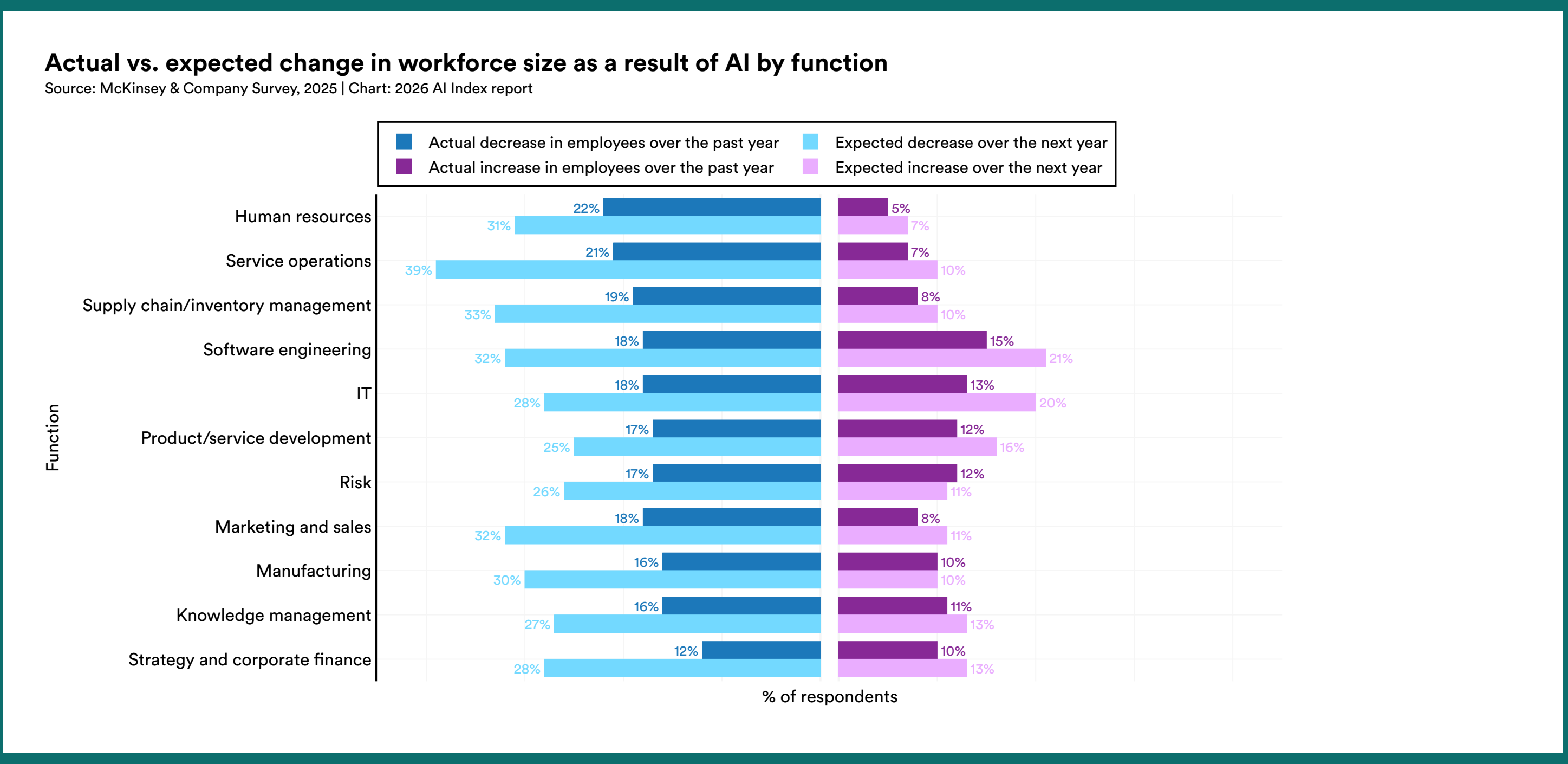

Figure 4.4.34

A surprising finding from Shao et al. (2026) shows that many workers are not wholly resistant to automation (Figure 4.4.35). A survey of 844 occupational tasks found that 46.1% of workers actively want AI to take over those tasks. Support was especially strong in areas where workers believed automation would free up time for higher-value tasks, reduce repetitiveness, or improve quality of output. However, actual usage patterns do not necessarily reflect these preferences. Occupational tasks with the highest average automation scores account for only 1.3% of Claude.AI usage. A related framework maps these tasks across four zones based on worker desire and technical feasibility (Figure 4.4.36). Looked at this way, AI's labor impact will likely register in how specific tasks are redesigned rather than blunt automation at the occupation level, with the reorganization of work unfolding gradually.

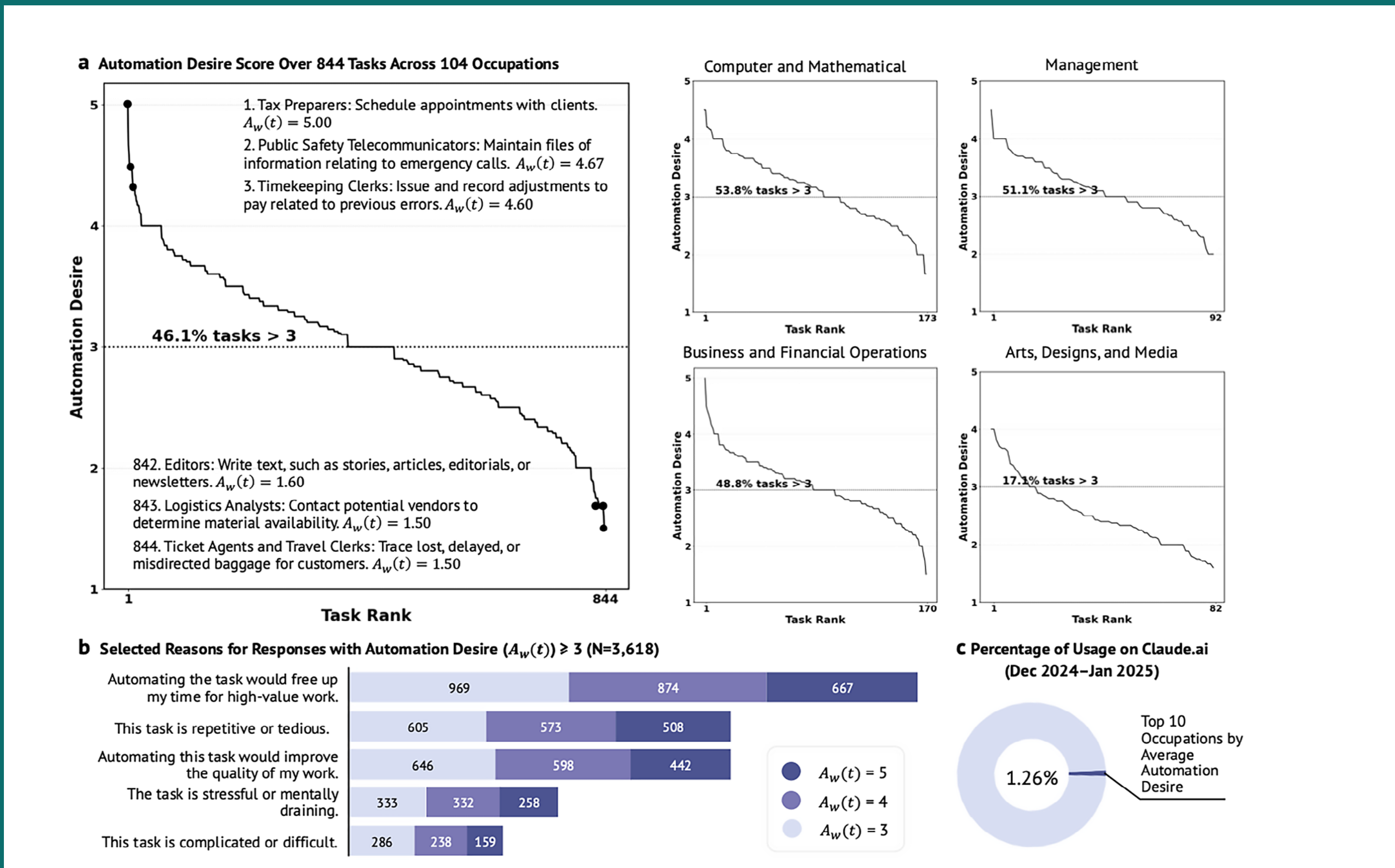

**a Automation Desire Score Over 844 Tasks Across 104 Occupations**

1. Tax Preparers: Schedule appointments with clients. $A_w(t) = 5.00$
2. Public Safety Telecommunicators: Maintain files of information relating to emergency calls. $A_w(t) = 4.67$
3. Timekeeping Clerks: Issue and record adjustments to pay related to previous errors. $A_w(t) = 4.60$

46.1% tasks > 3

842. Editors: Write text, such as stories, articles, editorials, or newsletters. $A_w(t) = 1.60$
843. Logistics Analysts: Contact potential vendors to determine material availability. $A_w(t) = 1.50$
844. Ticket Agents and Travel Clerks: Trace lost, delayed, or misdirected baggage for customers. $A_w(t) = 1.50$

Automation Desire
Task Rank

Computer and Mathematical — 53.8% tasks > 3
Management — 51.1% tasks > 3
Business and Financial Operations — 48.8% tasks > 3
Arts, Designs, and Media — 17.1% tasks > 3

**b Selected Reasons for Responses with Automation Desire ($A_w(t)$) ≥ 3 (N=3,618)**

|  | $A_w(t)$ = 3 | $A_w(t)$ = 4 | $A_w(t)$ = 5 |
|---|---|---|---|
| Automating the task would free up my time for high-value work. | 969 | 874 | 667 |
| This task is repetitive or tedious. | 605 | 573 | 508 |
| Automating this task would improve the quality of my work. | 646 | 598 | 442 |
| The task is stressful or mentally draining. | 333 | 332 | 258 |
| This task is complicated or difficult. | 286 | 238 | 159 |

**c Percentage of Usage on Claude.ai (Dec 2024–Jan 2025)**

1.26% — Top 10 Occupations by Average Automation Desire

Source: Shao et al. (2026)

Figure 4.4.35

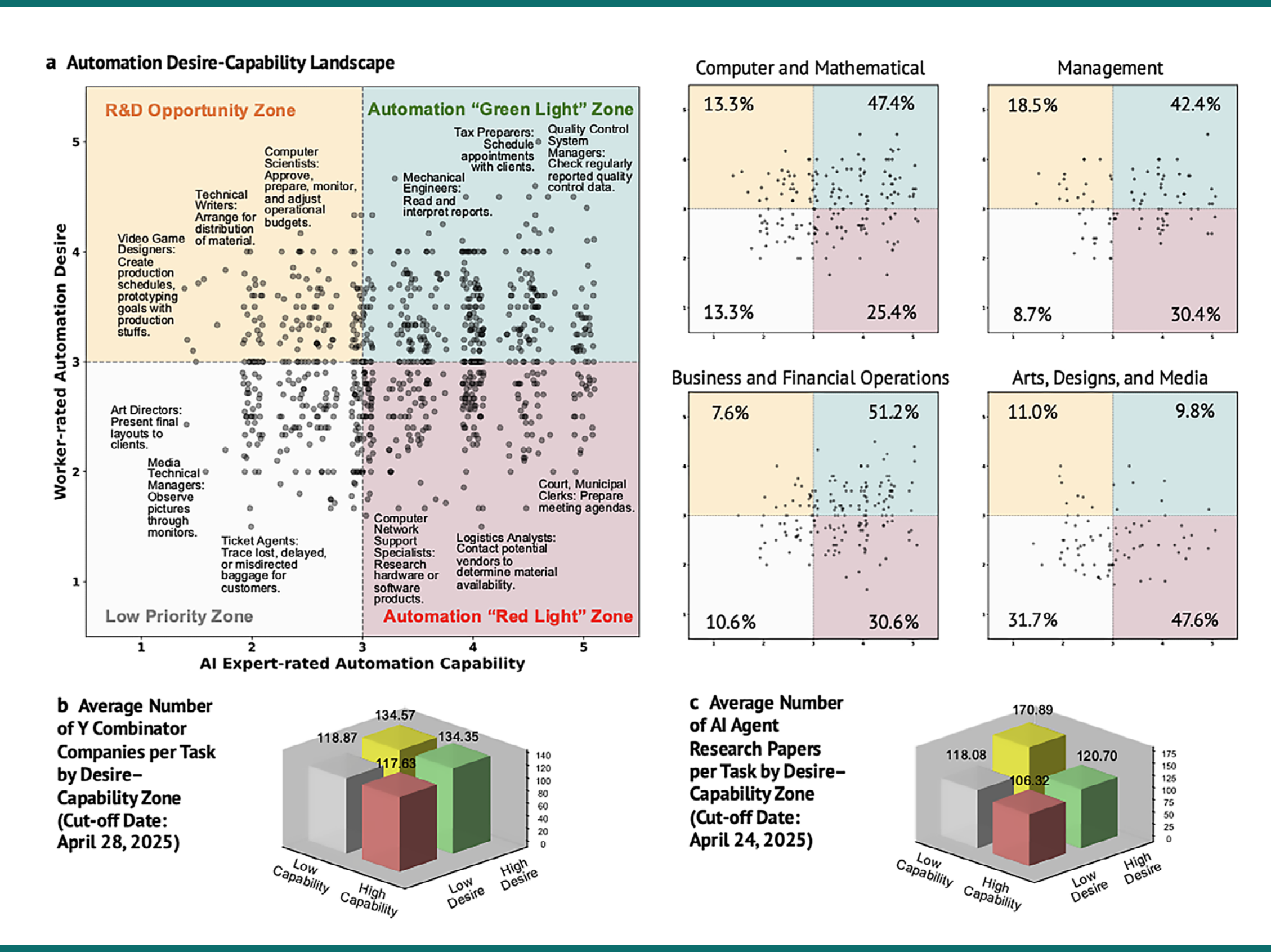

Source: Shao et al. (2026)

Figure 4.4.36

# 4.5 Robot Deployments

Physical automation through robotics represents one form of AI's economic integration in industrial environments, particularly in production settings such as manufacturing lines or warehouses. To track the trends, the AI Index uses data from the International Federation of Robotics, IFR 2025, a nonprofit that publishes annual World Robotics reports on global installation patterns and operational stock. Industrial robots are defined by the IFR, and in this reporting,[18] as "automatically controlled, reprogrammable, multipurpose manipulators, programmable in three or more axes, which can be either fixed in place or mobile for use in industrial automation applications."

## Global Installation Patterns

Global industrial robot activity continues to rise, though year-over-year growth has flattened. In 2024, 542,000 industrial robots were installed globally, a slight increase (0.2%) from the previous year (Figure 4.5.1). The composition of those robots has also shifted over time. Collaborative robots, which are designed to work alongside human operators, continue to gain market share over traditional robots. In 2017, collaborative robots accounted for just 2.8% of all new industrial robot installations, compared to 13.6% in 2024. The total operational stock in 2024 grew to 4,664,000, up from 4,282,000 in 2023 (Figure 4.5.2). Overall, industrial automation capacity has shown a consistent upward trajectory, with the global fleet of industrial robots quadrupling since 2012.

**Number of industrial robots installed in the world, 2012–24**

Source: International Federation of Robotics (IFR), 2025 | Chart: 2026 AI Index report

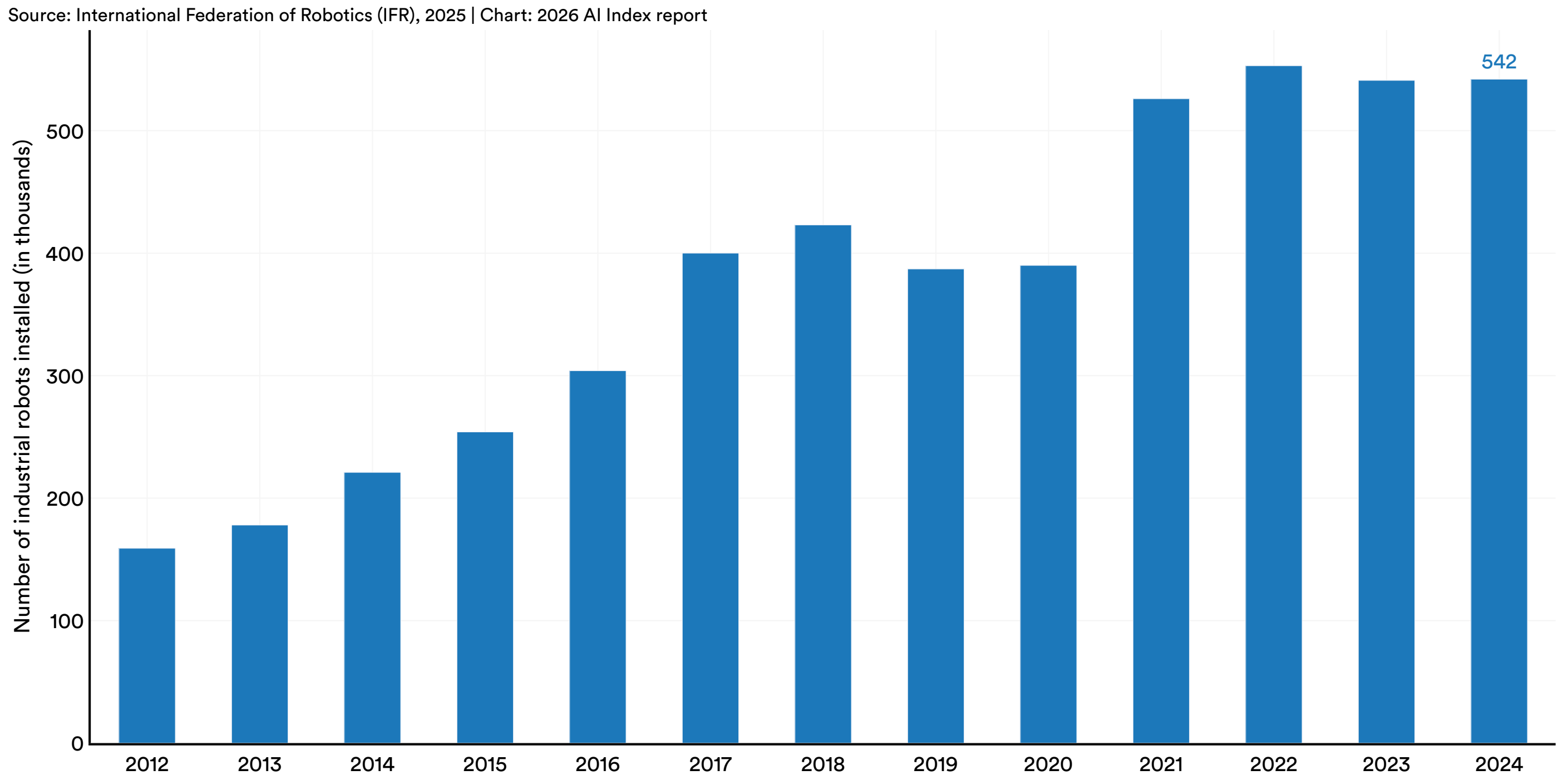

Figure 4.5.1

18 Due to the timing of the IFR report, the most recent data is from 2024. Every year, the IFR revisits data collected for previous years and will occasionally update the data if more accurate figures become available. Therefore, some of the data reported in this year's report might differ slightly from data reported in previous years.

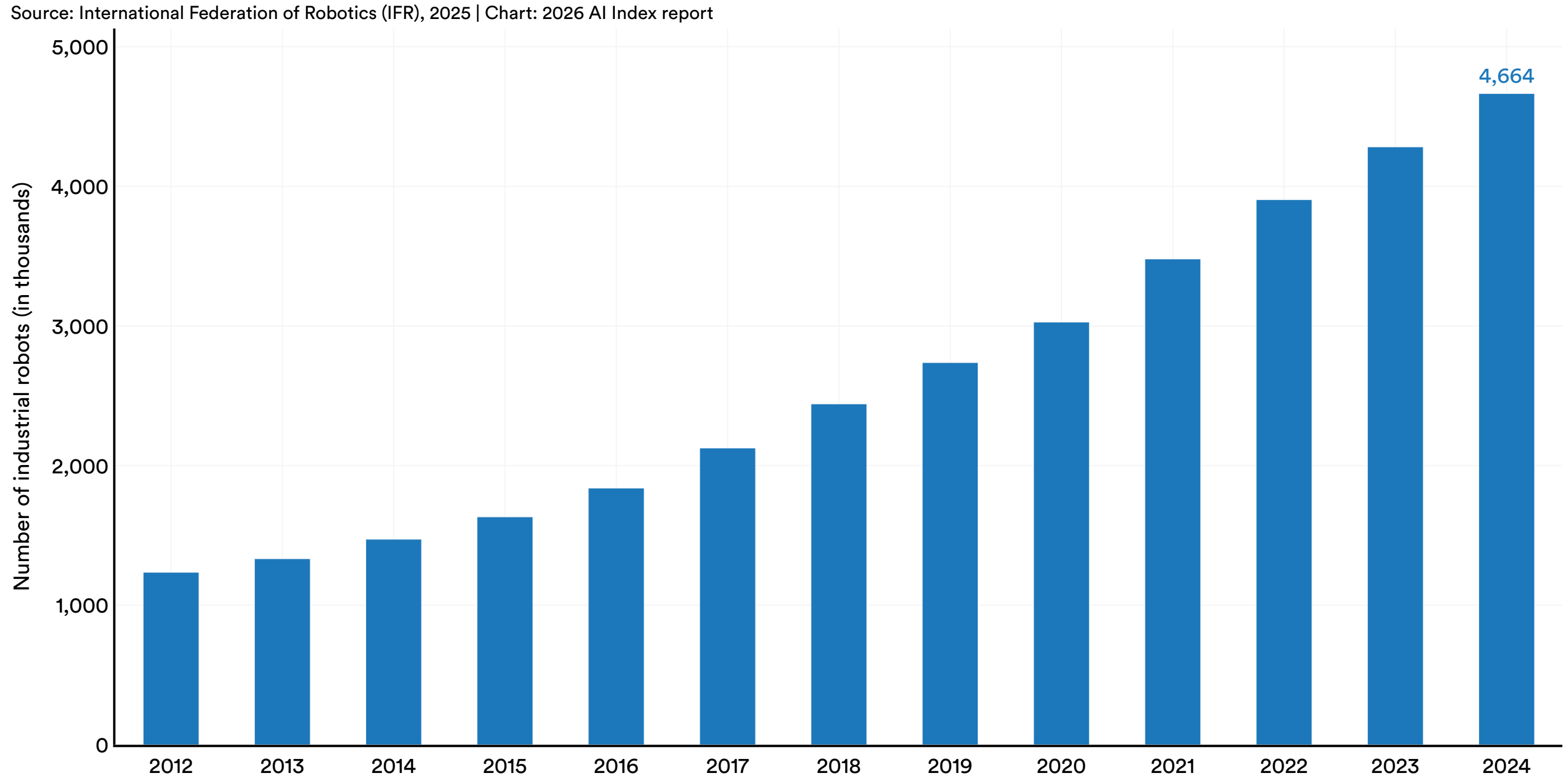
Operational stock of industrial robots in the world, 2012–24
Source: International Federation of Robotics (IFR), 2025 | Chart: 2026 AI Index report

Figure 4.5.2

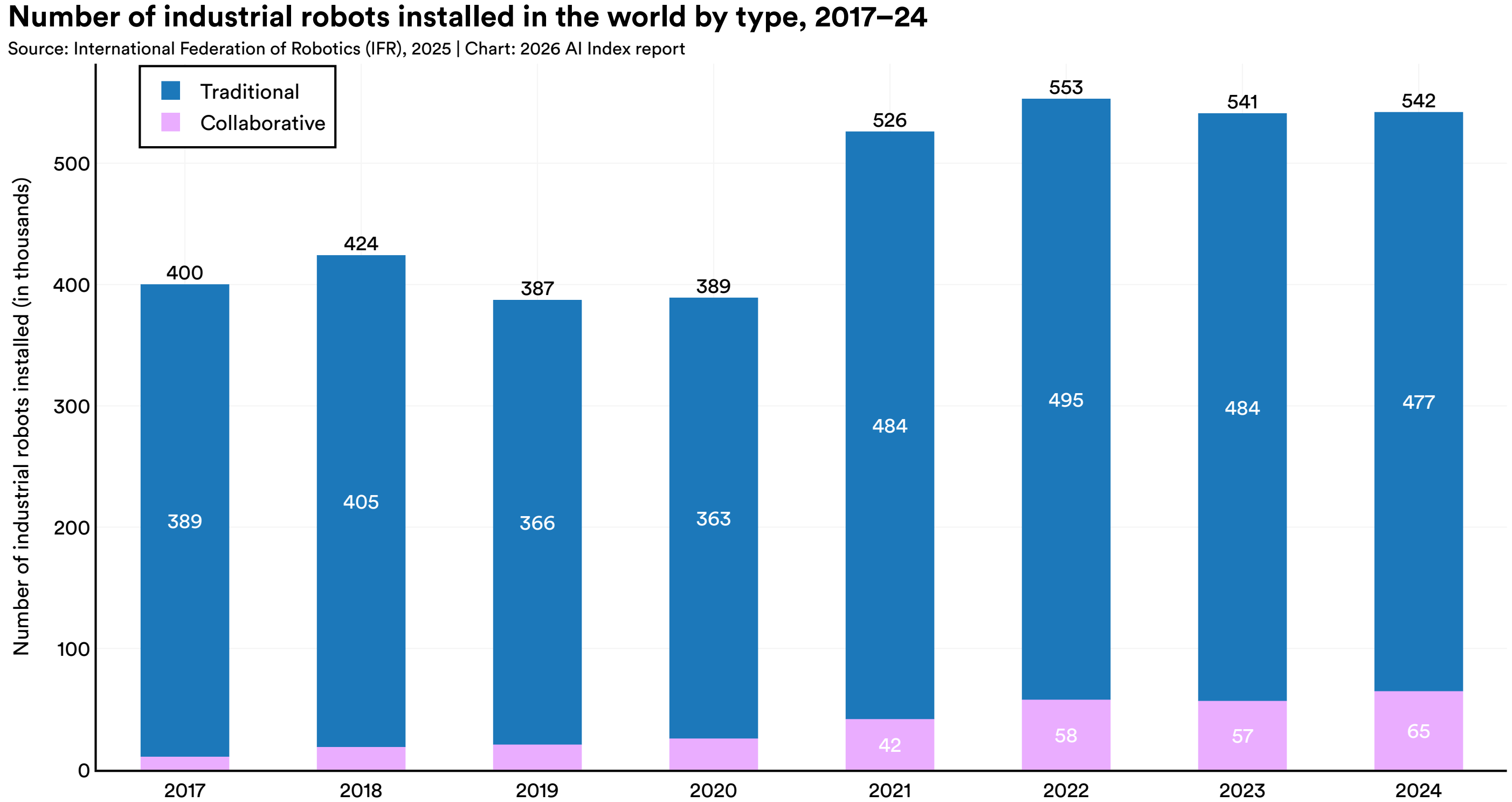
Number of industrial robots installed in the world by type, 2017–24
Source: International Federation of Robotics (IFR), 2025 | Chart: 2026 AI Index report

Figure 4.5.3

## Geographic Patterns

Industrial robot installation follows patterns similar to the investment and talent trends discussed above, although its geographic distribution is relatively narrow. In 2024, China led the world with 295,000 industrial robot installations, six times more than Japan's 44,500 and 8.6 times more than the United States' 34,200 (Figure 4.5.4). South Korea and Germany followed with 30,600 and 27,000 installations, respectively. China's share of global installations has increased substantially from 20.8% in 2013 to 54.4% in 2024 (Figure 4.5.5).

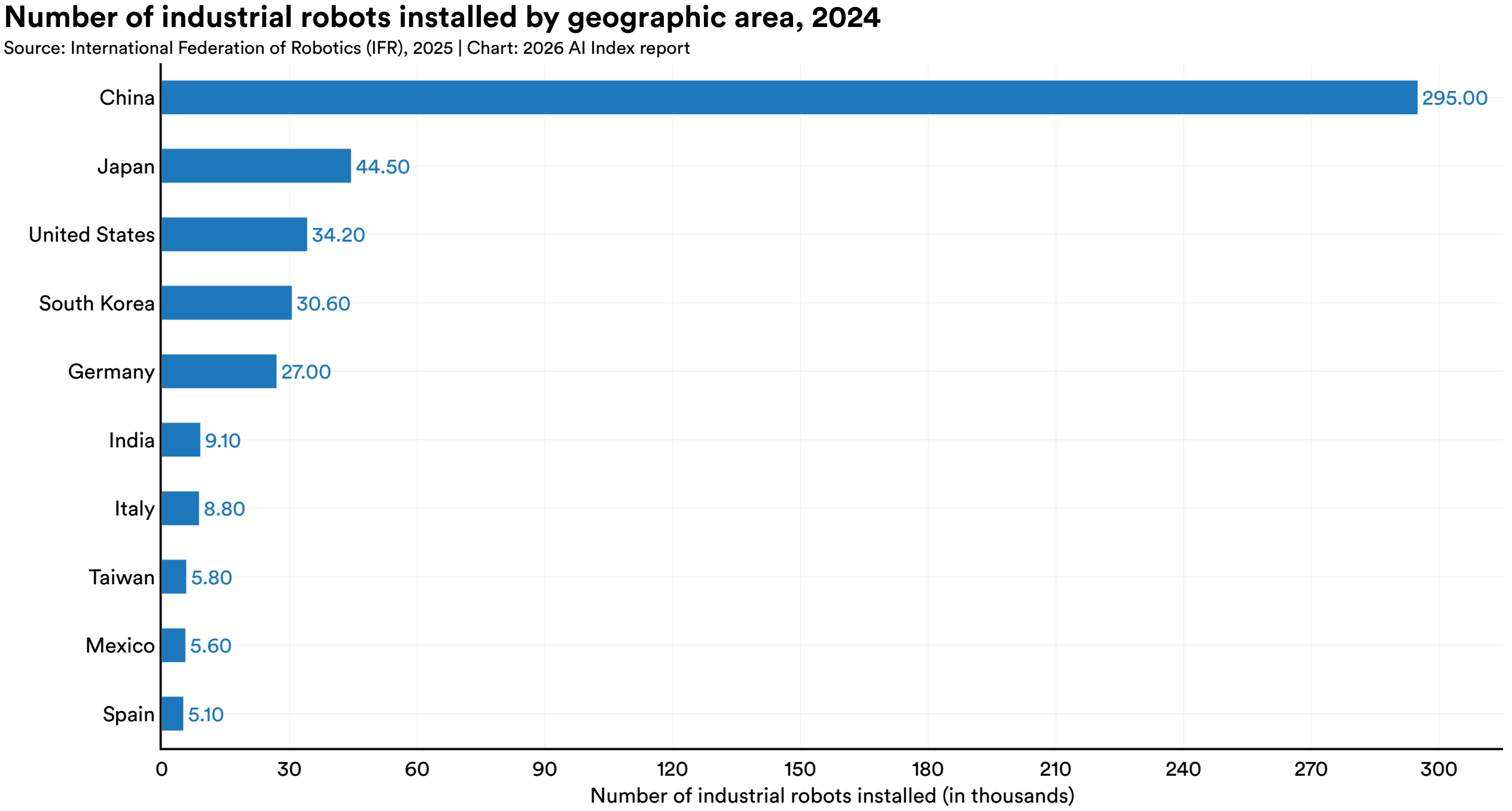
Number of industrial robots installed by geographic area, 2024
Source: International Federation of Robotics (IFR), 2025 | Chart: 2026 AI Index report

Figure 4.5.4

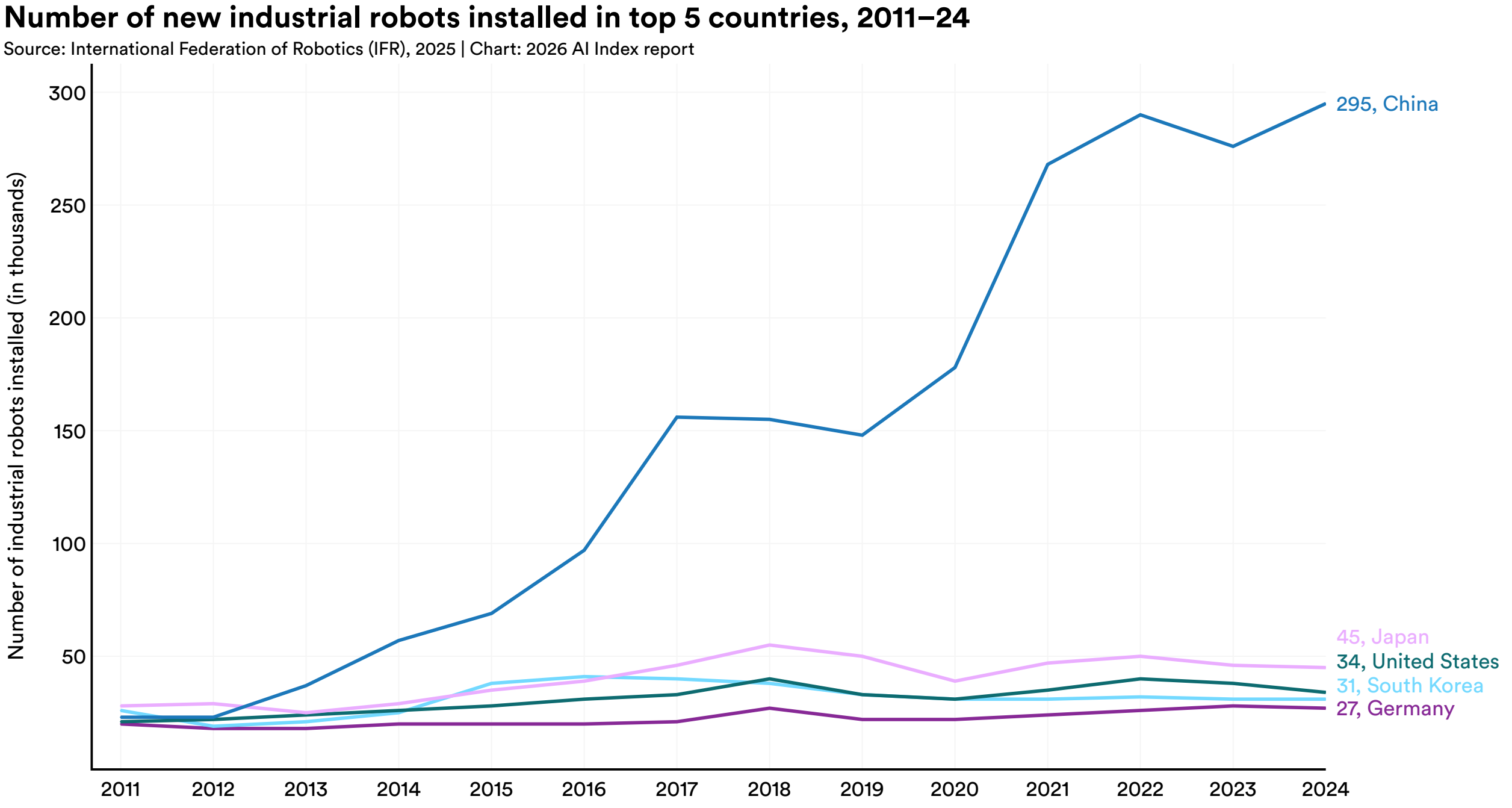
Number of new industrial robots installed in top 5 countries, 2011–24
Source: International Federation of Robotics (IFR), 2025 | Chart: 2026 AI Index report

Figure 4.5.5

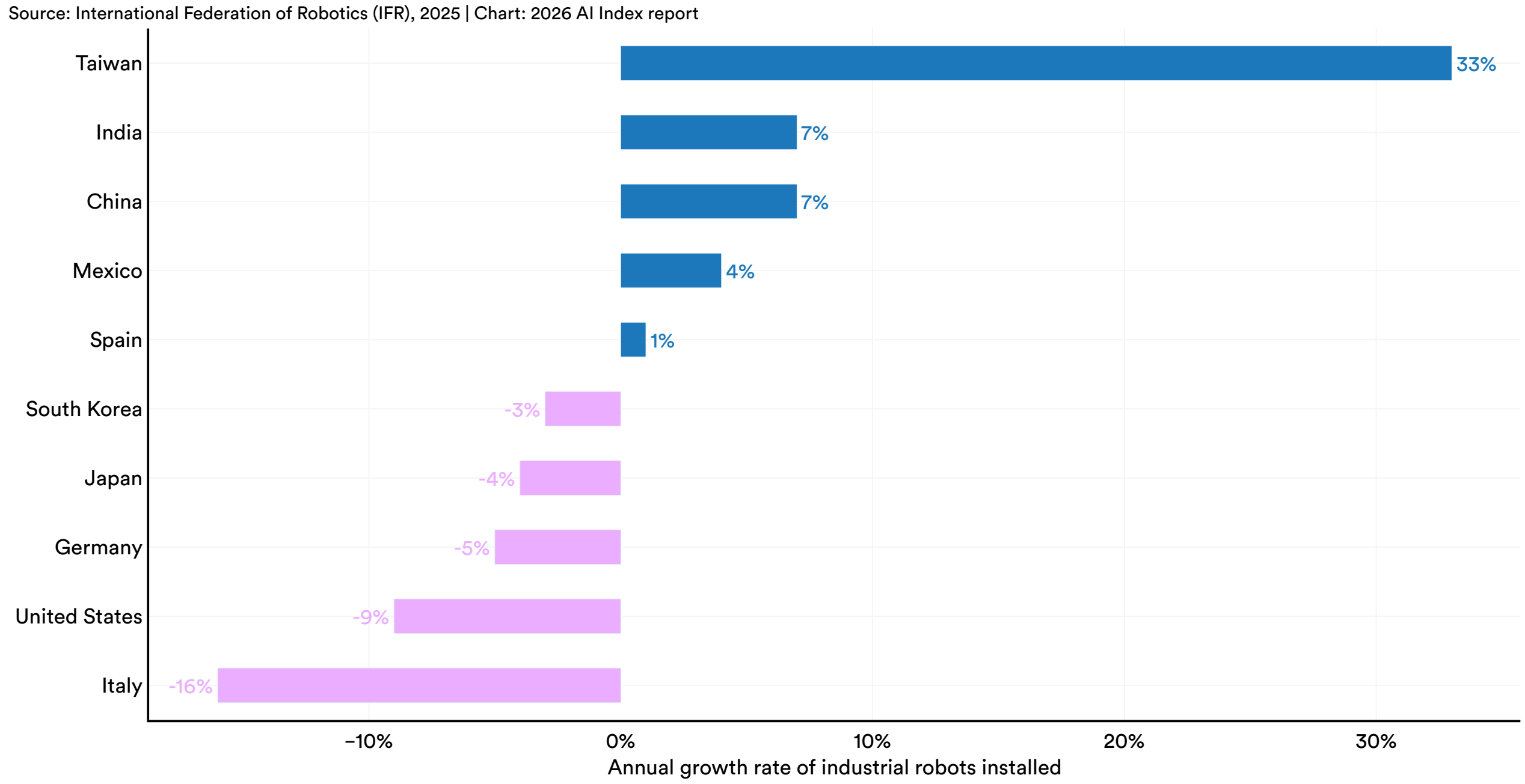
Annual growth rate of industrial robots installed by geographic area, 2023 vs. 2024
Source: International Federation of Robotics (IFR), 2025 | Chart: 2026 AI Index report

Figure 4.5.6

## Service Robotics

Nonindustrial or service robots designed for tasks such as logistics, hospitality, and agriculture showed growth in 2024 (Figure 4.5.7). Service robot installations increased across most application areas compared to 2023, though agriculture saw particularly strong adoption. The number of service robots deployed in an agricultural setting increased 2.5-fold. Only the hospitality category saw a year-over-year decline.

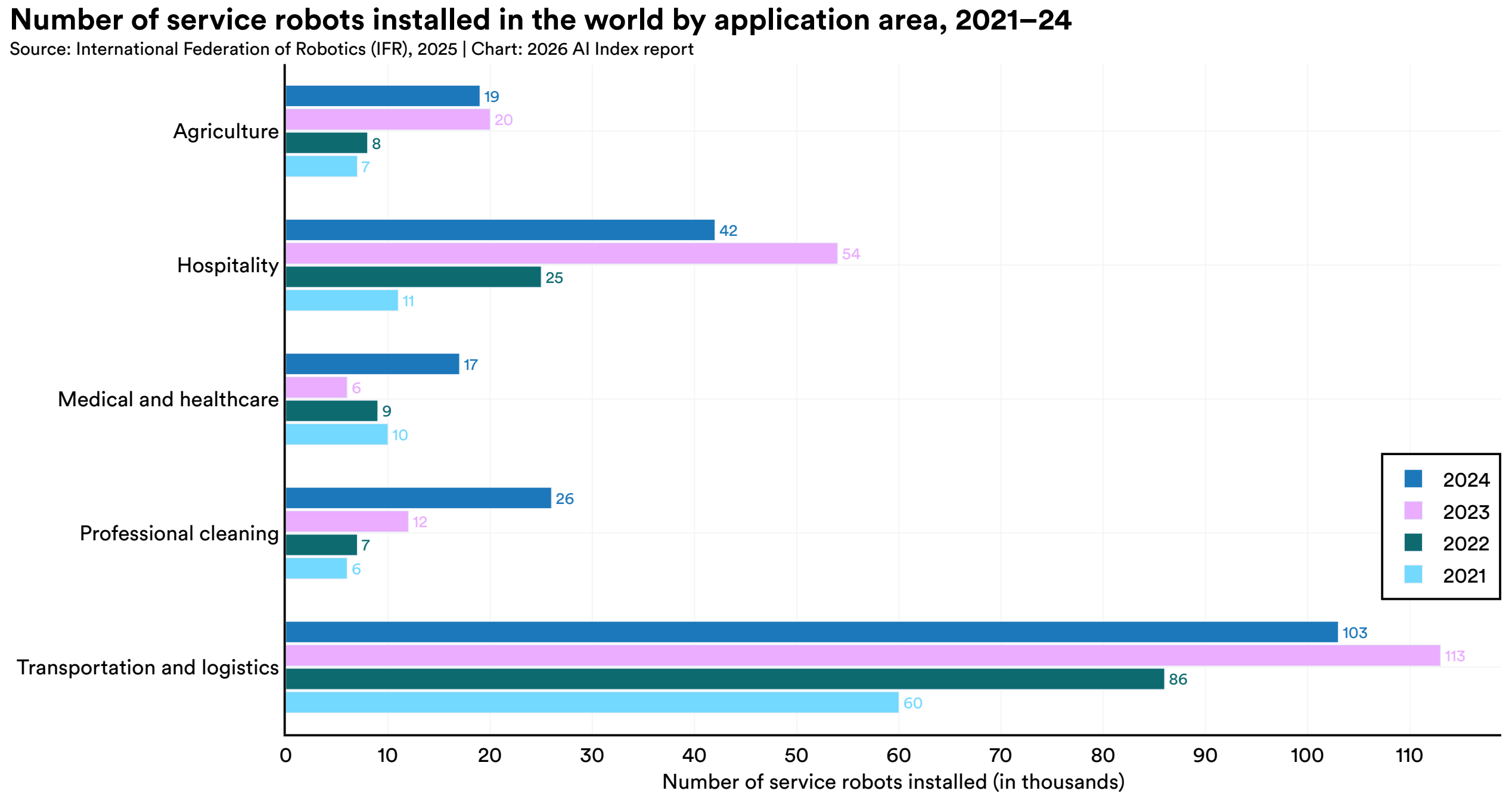

Figure 4.5.7

# 5

# Science

## Overview

The speed with which AI is transforming science is accelerating, with momentum from the 2024 Nobel Prize in Chemistry awarded to Dennis Hassabis, John Jumper, and David Baker for their work on AI-driven protein structure prediction and design, and the operational deployment of AI weather models at the European Centre for Medium-Range Weather. In 2025, AI moved beyond improving individual pipeline steps and toward replacing entire scientific workflows, from weather prediction to multiagent hypothesis generation and experimental design. Still, rigorous benchmarks continue to expose large gaps between plausible output and reliable scientific work, with frontier agents scoring below 20% on paper-scale replication tasks. AI's impact in social sciences has been slower to emerge but with notable exceptions. Linguistics and computational language research helped lay the groundwork for modern language models, and those models are now being applied back into fields such as linguistics, communications, and network analysis. Progress in these areas is harder to capture through the datasets and benchmarks covered in this chapter.

## Contents

# Chapter Highlights

1. **AI-related scientific publications are growing year-over–year.** Natural sciences reached approximately 80,150 AI publications in 2025, up 26% from 2024. AI now accounts for 5.8%–8.8% of scientific research output depending on the field, up from below 1% in 2010.

2. **Frontier models outperform human chemists on average but cannot reproduce published research.** On ChemBench, the best models surpass human expert averages across 2,700+ chemistry questions while struggling with basic tasks. On ReplicationBench, frontier models score below 20% on paper-scale replication in astrophysics. On UnivEarth, LLM agents answer Earth observation questions with 33% accuracy, and their code fails 58% of the time.

3. **Astronomy released its first foundation model, first visualization benchmark, and a 100TB training dataset in 2025, signaling a field-wide shift toward AI infrastructure.** AION-1, trained on over 200 million celestial objects from 5 major surveys, is the first astronomy foundation model. AstroVisBench introduced the first benchmark for LLM scientific computing and visualization in the field.

4. **An AI system ran a full weather forecasting pipeline end-to-end for the first time in 2025.** Aardvark Weather replaced the traditional numerical prediction pipeline with a single ML system, and multiple AI weather models reached operational deployment. FourCastNet 3 generates a 60-day global forecast in under 4 minutes, running 8 to 60 times faster than prior approaches.

5. **On end-to-end scientific research tasks, the best AI agents score roughly half of what PhD experts achieve.** On PaperArena, the best agent reaches 38.8% accuracy versus a PhD expert baseline of 83.5%. On BixBench, frontier models achieve roughly 17% accuracy on real-world bioinformatics analysis.

6. **The first fully AI-generated paper was accepted at a peer-reviewed workshop in 2025, but the list of experimentally confirmed AI discoveries remains short.** Sakana's AI Scientist-v2 produced a paper accepted at an ICLR workshop without human-coded templates. Google's AI Co-Scientist was validated in three biomedical areas.

7. **Most AI models for science originate from academic and government institutions, in contrast with the industry-dominated landscape of general-purpose AI.** Many AI foundation models for science result from international collaborations. Earth science datasets come entirely from government and academic sources, while industry leads foundation model development in weather and climate.

# 5.1 AI for Science in 2025

AI's role in science falls into three categories that coexist but differ in terms of maturity. The first—machine learning over scientific data to build predictive and explainable models—has been practiced for several decades and is now commonplace. The second—AI systems that assist scientists in their workflows through literature synthesis, experiment design, or data analysis—has been emerging over several years and expanded considerably in 2025. The third category—autonomous AI systems capable of generating new scientific discoveries with limited human guidance—is gaining traction but it remains at an early stage. The year's most visible developments occurred primarily in the second and third categories. Aardvark Weather replaced the full numerical weather prediction pipeline (Allen et al., 2025). Google's AI Co-scientist orchestrated hypothesis generation through experimental design (Gottweis et al., 2025). To date, the clearest breakthroughs tend to cluster in domains with strong existing data infrastructure, including structural biology, physics, chemistry, and materials science, rather than in fields with the most sophisticated mathematical or physics-based models.

These developments, however, do not automatically translate into scientific progress. Experimental validation remains expensive and time-consuming, and scientists are unlikely to invest in testing AI-generated hypotheses without sufficient reason to believe it will yield some findings. In drug discovery, for example, AI systems can propose novel candidate molecules at scale, but clinical trials to determine whether those molecules work remain a costly, multiyear process. The gap between what AI can propose and what scientists can feasibly test is a recurring theme across the domains covered in this chapter.

## Publications in AI for Science

In the Web of Science database, AI-related publications in the natural sciences reached approximately 80,150 in 2025, up from 63,547 in 2024, a one-year increase of roughly 26% (Figure 5.1.1). Physical sciences[1] and life sciences followed similar trajectories in 2025, reaching approximately 33,000 and 29,000 publications, respectively, with each growing by roughly 27%–28% year over year. Earth science—the smallest category in absolute terms at approximately 20,460 publications—grew by about 23%. As a share of total scientific output, AI-related work remains a single-digit fraction of each field but is climbing quickly (Figure 5.1.2). By 2025, Earth science had the highest AI penetration at 8.8%, followed by natural sciences overall at 6.8%, life sciences at 6.5%, and physical sciences at 5.8%. In 2010, all four categories sat below 1%. The quantity of AI-mentioning papers is not the same as the quality of AI-enabled discovery, but the breadth suggests that AI methods are becoming a routine part of scientific practice across disciplines.

1 Physical sciences in this analysis include: astronomy and astrophysics, chemistry, crystallography, electrochemistry, geochemistry and geophysics, geology, mathematics, meteorology and atmospheric science, mineralogy, mining and mineral processing, oceanography, optics, physical geography, physics, polymer science, thermodynamics, and water resources.

## Number of AI-related publications in natural sciences, 2010–25

Source: AI Index, 2026 | Chart: 2026 AI Index report

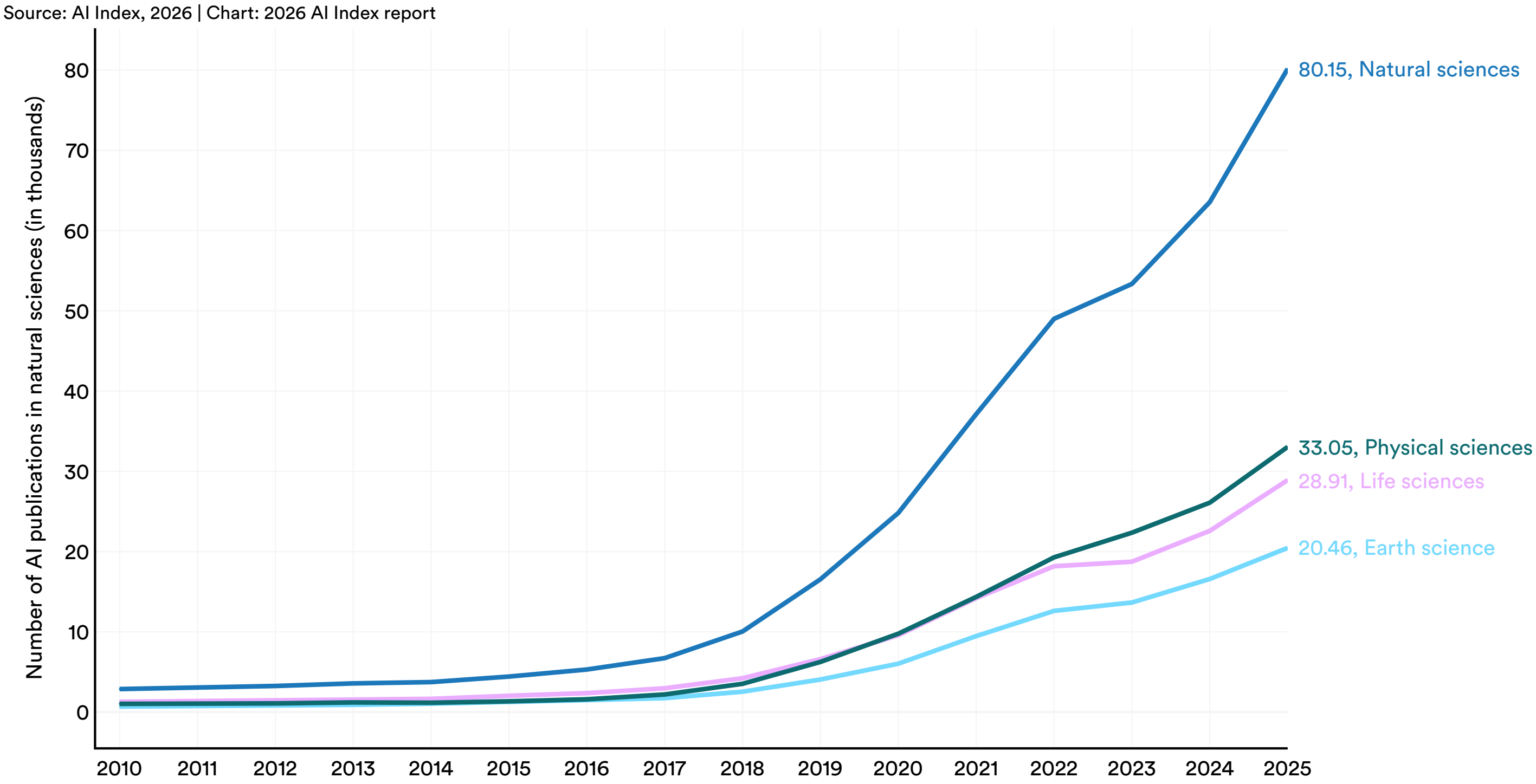

Figure 5.1.1[2]

## AI-related publications in natural sciences (% of total), 2010–25

Source: AI Index, 2026 | Chart: 2026 AI Index report

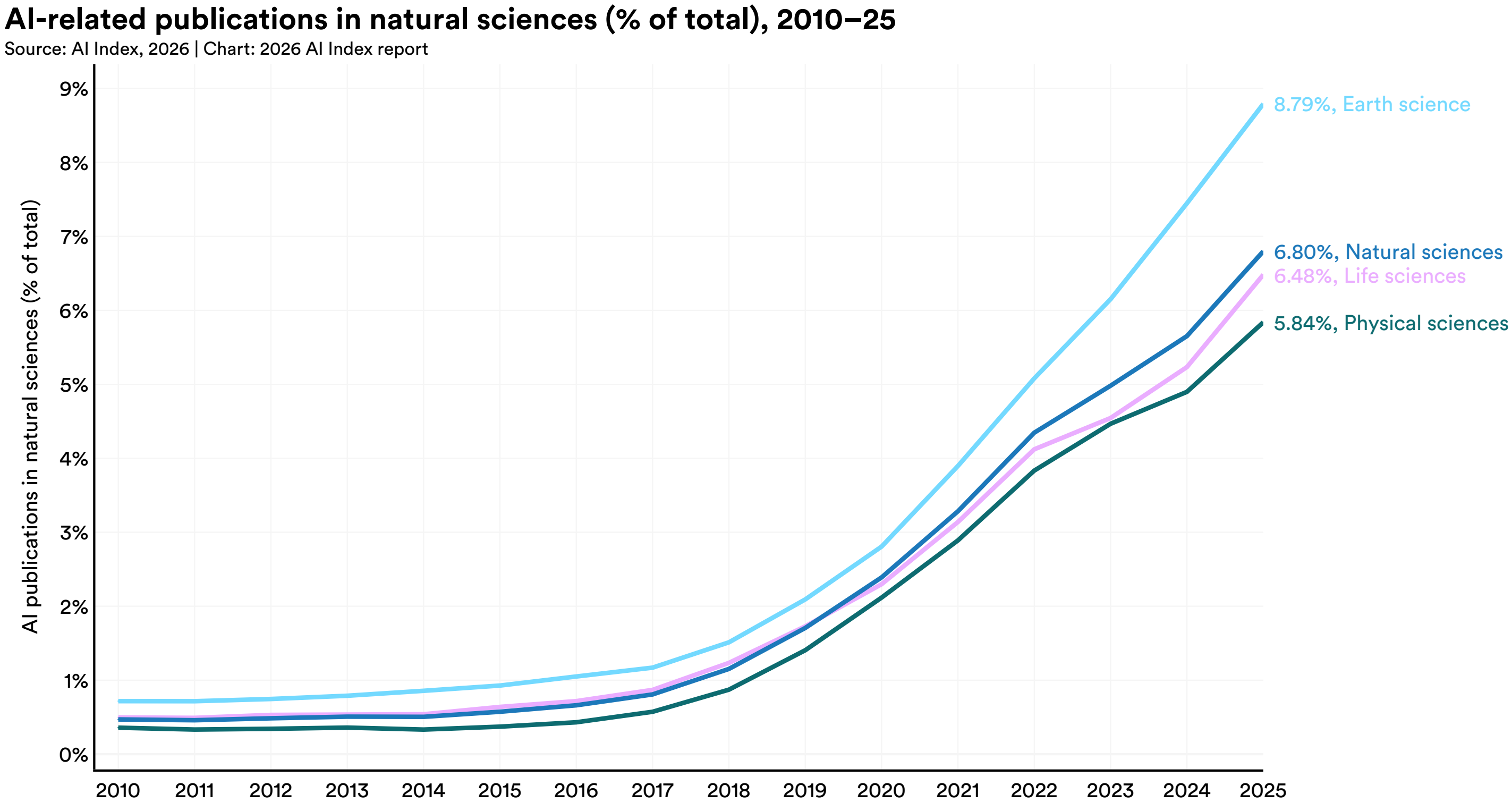

Figure 5.1.2

2 The natural sciences count may be slightly lower than the sum of the individual domain counts. This is because a single publication can be assigned to more than one domain. For example, a biochemistry paper may be categorized under both physical sciences and life sciences. To avoid double-counting, these publications are counted only once in the natural sciences total.

# 5.2 AI Across Scientific Domains

This section examines AI's expanding role in science across three major scientific groupings and tracks the datasets, benchmarks, and foundation models of each. The tables below catalog selected releases across each category. A consistent finding is that the majority of scientific AI models originate from academic institutions collaborating across countries, in contrast to the industry-dominated landscape of general-purpose foundation models described in Chapter 1 and Chapter 2.

## Physics, Astronomy, Chemistry, and Materials Science

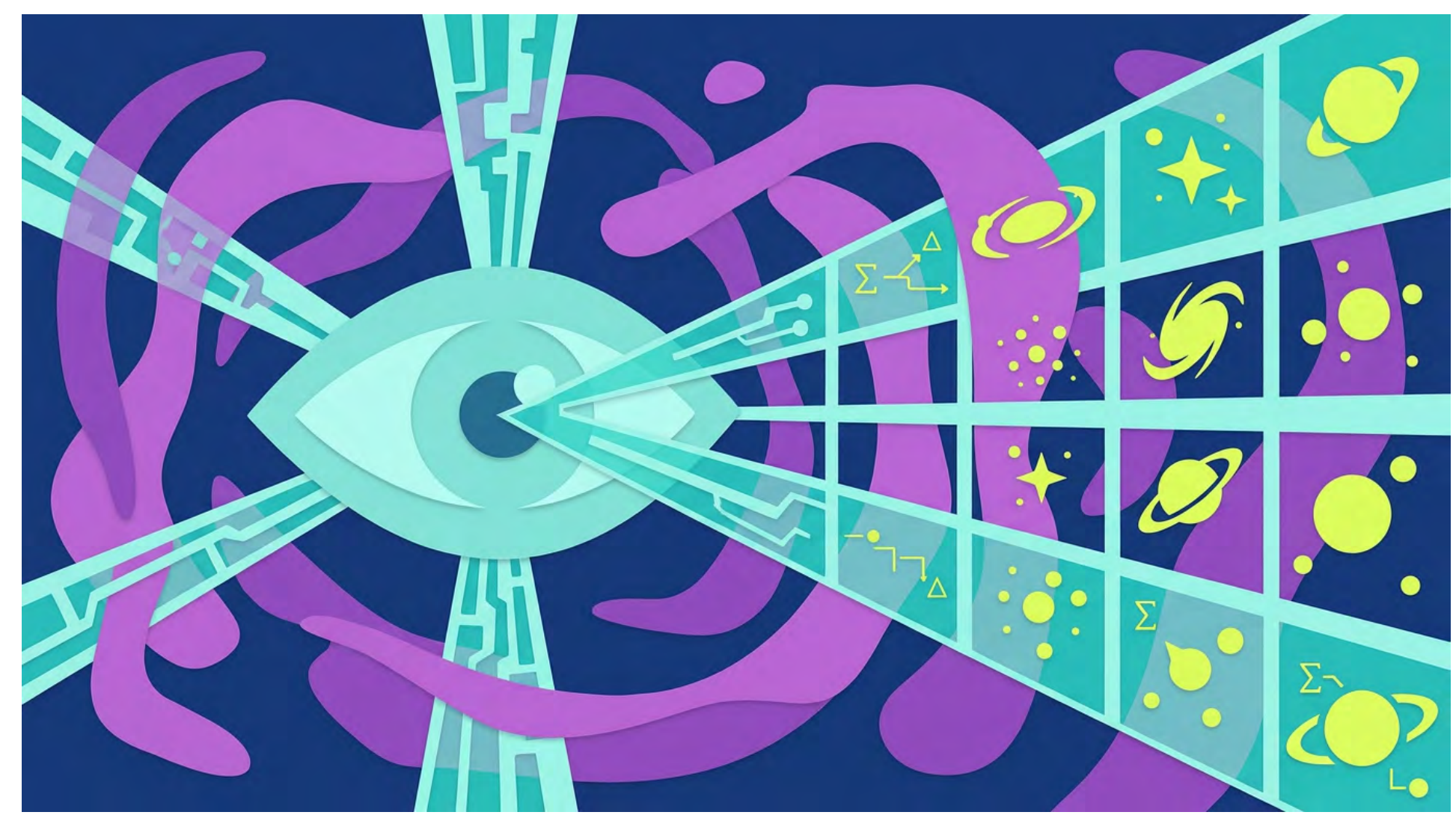

AI is accelerating physics, astronomy, chemistry, and materials science by replacing expensive first-principles simulations with learned surrogates and by generating novel materials and molecular structures through inverse design. Notable 2025 releases include large chemistry datasets (e.g., OMol25 and OC25), simulation-oriented foundation models (e.g., Walrus, GPhyT), and materials checkpoints for atomistic modeling and generation (e.g., MACE-MP-0 and MatterGen). In chemistry and materials science, agent systems began connecting to external software tools and laboratory equipment to execute experiments. Benchmark results, however, suggest these systems are not yet reliable when asked to carry out full research tasks from start to finish.

### Datasets

The largest dataset releases in physics and chemistry in 2025 expanded multimodal astronomy resources and chemistry-scale quantum data. Multimodal Universe aggregates approximately 100TB of astronomical observations, while OMol25 reports over 100 million high-accuracy density functional theory (DFT) calculations spanning 83 elements. These datasets provide the training foundation for large models targeting prediction and simulation tasks in their respective fields (Figure 5.2.1).

### Selected datasets in physics, astronomy, chemistry, and materials science (2025)

| Name | Affiliation[3] | Domain | Sector | Summary |
|---|---|---|---|---|
| ChemPile | Helmholtz Institute for Polymers (HIPOLE) Jena, Friedrich Schiller University Jena, Hacettepe University, University of Toronto | Chemistry | ACADEMIA, NONPROFIT | 75B-plus tokens of curated chemical data (SMILES, InChI, text, code, educational materials). |
| Multimodal Universe | Polymathic AI, Instituto de Astrofisica de Canarias, Universidad de La Laguna, Massachusetts Institute of Technology | Astronomy | ACADEMIA, NONPROFIT | 100TB astronomical dataset: multichannel images, spectra, time series from hundreds of millions of observations. |
| The Open Molecules 2025 (OMol25) | Fundamental AI Research (FAIR) at Meta, Los Alamos National Laboratory, University College Dublin | Chemistry, Materials, Chemical Physics | INDUSTRY, GOVERNMENT, ACADEMIA | 100M-plus DFT calculations spanning 83 elements, diverse chemical interactions, structures up to 350 atoms. |
| The Open Catalyst 2025 (OC25) | FAIR at Meta, Texas Tech University, Nanyang Technological University | Chemistry, Materials | INDUSTRY, GOVERNMENT, ACADEMIA | 7.8M calculations across 1.5M solvent environments spanning 88 elements for catalysis at solid-liquid interfaces. |

Figure 5.2.1

## Benchmarks

In these particular domains, benchmarks have been newly introduced and therefore do not offer longitudinal data across multiple years. It is interesting to see how general-purpose frontier models, discussed in Chapter 2, perform on scientific tasks (Figure 5.2.2). On ChemBench, a chemistry evaluation with over 2,700 question-answer pairs, the best frontier models outperform the best human chemists, though they struggle with basic tasks. ReplicationBench reports frontier model performance below 20% on paper-scale replication tasks in computational astrophysics.

3 Full references are provided in the Appendix.

## Selected benchmarks in physics, astronomy, chemistry, and materials science (2025)

| Name | Affiliation[4] | Domain | Sector | Summary |
|---|---|---|---|---|
| AstroVisBench | University of Texas at Austin, NSF National Optical-Infrared Astronomy Research Laboratory, University of Virginia | Astronomy | ACADEMIA, GOVERNMENT | First benchmark for LLM scientific computing and visualization in astronomy. |
| Chembench | Friedrich Schiller University Jena, Helmholtz Institute for Polymers (HIPOLE) Jena, Spanish National Research Council (CSIC) | Chemistry | ACADEMIA, INDUSTRY, GOVERNMENT | 2,700-plus Q&A pairs. Best models outperform human chemists on average but struggle with basic tasks. |
| ChemX | ITMO University, D ONE | Chemistry, Materials Science | ACADEMIA, INDUSTRY | 10 curated datasets for automated chemical information extraction from nanomaterials and small molecules. |
| GravityBench | University of Toronto, NYU Abu Dhabi | Physics, Astrophysics | ACADEMIA | Tests AI discovery of physics laws from gravitational simulations, including non-real-world physics. |
| LLM-SRBench | Virginia Tech, VinUniversity, Carnegie Mellon University | Physics, Scientific Equation Discovery | ACADEMIA, INDUSTRY | 239 problems testing genuine equation discovery vs. memorization. Best systems score 31.5%. |
| Matbench Discovery | University of Cambridge, Lawrence Berkeley National Laboratory, Federal Institute of Materials Research and Testing (BAM) | Materials, Chemistry | ACADEMIA, GOVERNMENT | Evaluation framework for machine learning energy models prescreening stable inorganic crystals. |
| MatSciBench | UCLA, Princeton, Virginia Tech | Materials Science | ACADEMIA | 1,340 college-level problems across 6 fields and 31 subfields. Top models under 80%. |
| PHYBench | Peking University, CSRC | Physics | ACADEMIA | 500 original physics problems. Gemini 2.5 Pro: 36.9% vs. human experts: 61.9%. |

4 Full references are provided in the Appendix.

| | | | | |
|---|---|---|---|---|
| PhysGym | KAUST Center of Excellence for Generative AI, The Swiss AI Lab | Physics | ACADEMIA<br>INDUSTRY | Interactive physics environments testing LLM scientific reasoning under varying prior knowledge. |
| ReplicationBench | Stanford University, University of Toronto | Astrophysics/ Research Replication | ACADEMIA | Tests AI replication of entire astrophysics papers. Frontier models score under 20%. |
| TheoreticalPhysics Benchmark (TPBench) | University of Wisconsin-Madison, Indiana University, NSF-Simons AI Institute for the Sky (SkAI) | Theoretical Physics | ACADEMIA | 57 novel theoretical physics problems (high-energy theory, cosmology). Research-level problems largely unsolved. |

Figure 5.2.2

## Foundation Models

Foundation model (FM) releases in 2025 spanned astronomy, physics simulation, chemistry language models, and materials modeling (Figure 5.2.3). GPhyT, a General Physics Transformer, trained on 1.8TB of simulation data, achieved up to 29 times better performance than specialized models, and generalized to physics problems outside its training data without task-specific fine-tuning.

### Selected foundation models in physics, astronomy, chemistry, and materials science (2025)

| Name | Affiliation[5] | Domain | Sector | Summary |
|---|---|---|---|---|
| AION-1 | UC Berkeley, Flatiron Institute, New York University | Astronomy | ACADEMIA<br>NONPROFIT<br>GOVERNMENT | Astronomy FM: 300M–3.1B parameters, 200M-plus celestial objects from 5 major surveys. Open release. |
| ChemDFM | Shanghai Jiao Tong University, Suzhou Laboratory, AI Speech Co. | Chemistry | ACADEMIA<br>GOVERNMENT<br>INDUSTRY | Chemistry LLM: 34B tokens, 2.7M instructions. Generalist chemical AI. |

5 Full references are provided in the Appendix.

| | | | | |
|---|---|---|---|---|
| GPhyT: General Physics Transformer | RWTH Aachen University, University of Virginia | Physics | ACADEMIA | Trained on 1.8 TB simulation data. Up to 29x better than specialized models. Zero-shot generalization. |
| MACE-MP-0 | University of Cambridge, Federal Institute of Materials Research and Testing (BAM), UC Berkeley | Chemical Physics | ACADEMIA<br>GOVERNMENT<br>INDUSTRY | General-purpose force field model for predicting atomic interactions across nearly all materials. |
| MatterGen | Microsoft Research AI for Science, Shenzhen Institute of Advanced Technology (Chinese Academy of Sciences) | Materials Science | INDUSTRY<br>GOVERNMENT | Diffusion-based generative model. Over 2x more novel and stable than existing methods. |
| PDE-Transformer | Technical University of Munich | Physics Simulations | ACADEMIA | Transformer for physics partial differential equations (PDEs) on grids. Outperforms state-of-the-art vision architectures across 16 types of physics simulations. |
| PhysiX | UCLA | Physics Simulations | ACADEMIA | 4.5B params. First large-scale physics simulation FM. Transfers from natural videos to simulation. |
| SMI-TED | IBM Research | Chemistry | INDUSTRY | Chemical foundation models trained on molecular sequences. |
| Surya | University of Alabama in Huntsville, NASA Marshall Space Flight Center, IBM Research | Heliophysics | ACADEMIA<br>GOVERNMENT<br>INDUSTRY | 366M parameters. First heliophysics FM. Forecasts space weather from NASA's Solar Dynamics Observatory data without task-specific training. |
| Walrus | Flatiron Institute, New York University, University of Cambridge | Fluid Mechanics Continuum Mechanics, multiple domains | ACADEMIA<br>NONPROFIT<br>INDUSTRY | Fluid mechanics FM: 19 scenarios spanning astrophysics, geoscience, plasma physics, acoustics. Open weights. |

**Figure 5.2.3**

## AI Agents

Agent systems in the physical sciences combine tool use with domain-specific reasoning to perform tasks requiring multiple steps. Some of these systems function as focused components within larger pipelines, while others attempt to operate as end-to-end research systems. As they take on more responsibility for both designing and executing research, independent confirmation of results becomes an important step.

Physics Supernova scored 23.5 out of 30 at the 2025 International Physics Olympiad, ranking 14th out of 406 participants and reaching gold-medalist level. StarWhisper Telescope automates astronomical observation planning across 10 telescopes. In chemistry, ChemAgents demonstrated autonomous synthesis and optimization using a robotic platform controlled by Llama-3.1-70B (Figure 5.2.4).

### Selected AI agents in physics, astronomy, chemistry, and materials science (2025)

| Name | Affiliation[6] | Domain | Sector | Summary |
|---|---|---|---|---|
| ChatGPTMaterial Explorer | John Hopkins University | Materials Science | ACADEMIA | Materials science assistant combining LLMs with graph neural networks for property prediction. |
| ChemAgents | University of Science and Technology of China, University of Birmingham, Henan Academy of Sciences | Chemistry | ACADEMIA<br>GOVERNMENT | Robotic AI chemist (Llama-3.1-70B). Autonomous synthesis, optimization, photocatalysis. |
| ChemToolAgent | Shanghai Artificial Intelligence Laboratory, Soochow University, Zhejiang University | Chemistry<br>Materials Science | ACADEMIA<br>GOVERNMENT | 137 external chemical tools. HE-MCTS framework surpasses GPT-4o on chemistry QA. |
| Crystalyse | Imperial College London | Materials Science<br>Chemistry | ACADEMIA | Multi-tool AI agent for materials design that coordinates multiple computational tools through an LLM-based reasoning framework. |
| Physics Supernova | Princeton University, Tsinghua University, Shanghai Jiao Tong University | Physics | ACADEMIA | IPhO 2025: 23.5 out of 30, ranked 14th of 406. Gold-medalist-level physics problem-solving. |
| StarWhisper Telescope | University of Chinese Academy of Sciences, National Astronomical Observatories (CAS), Simon Fraser University | Astronomy | ACADEMIA<br>GOVERNMENT<br>INDUSTRY | Automates astronomical observations across 10 telescopes. LLM-driven observation planning. |

Figure 5.2.4

6 Full references are provided in the Appendix.

## Biological and Life Sciences

AI is increasingly being applied to biological research beyond biomedicine to address fundamental questions in genomics, neuroscience, ecology, and synthetic biology. Chapter 6 covers AI's role in the therapeutic pipeline, from protein structure prediction to drug design to clinical applications. This section focuses on the broader scientific infrastructure, including the datasets, benchmarks, foundation models, and agents that support biological research as a whole.

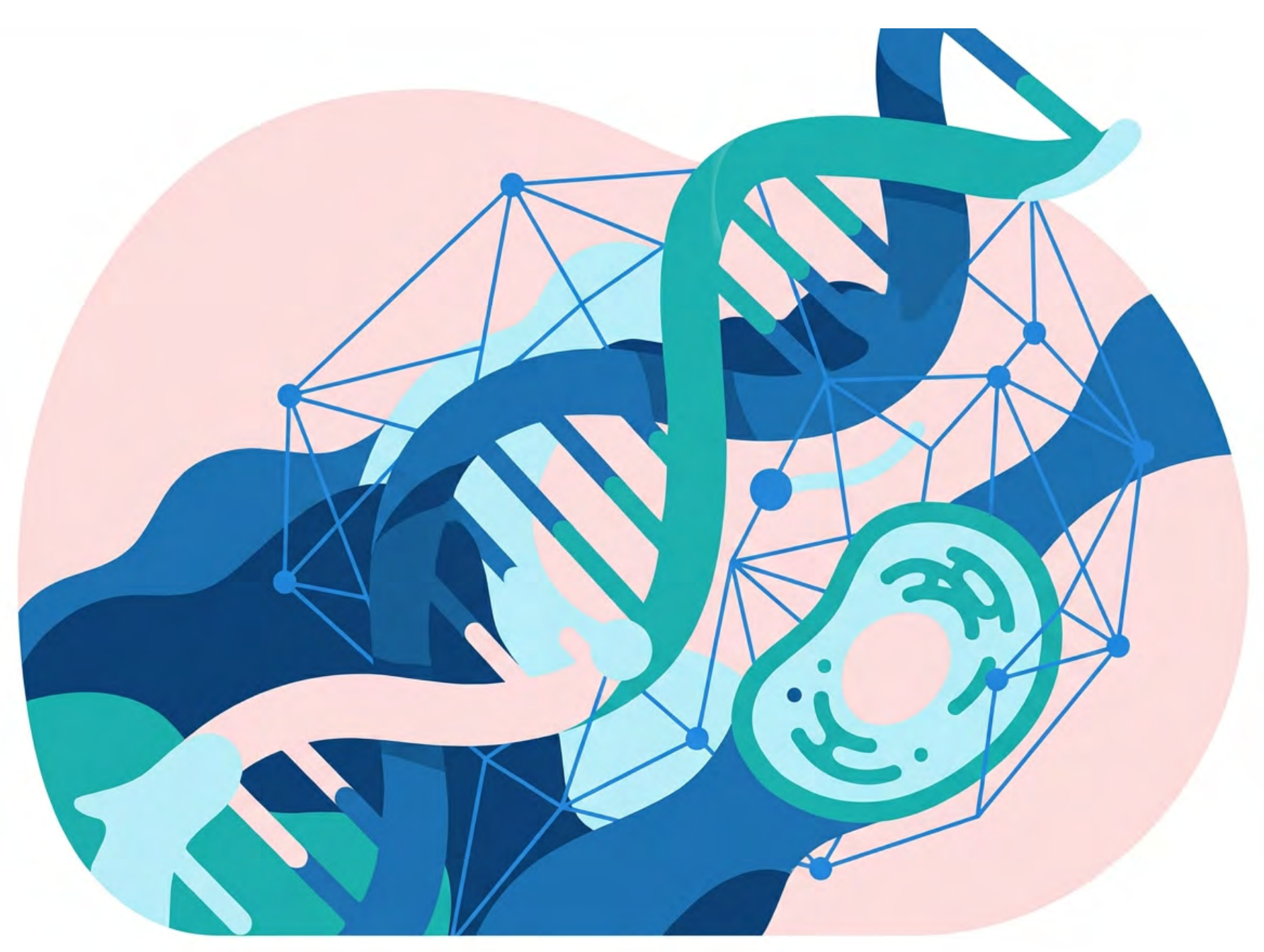

The scale of biological training data grew in 2025, and foundation models trained on genomic and evolutionary data expanded from prediction into generative design. However, the gap between genomic sequence data, which is abundant, and functional perturbation data, which measures how biological systems respond to interventions, remains wide (Callahan et al., 2025; Sun et al., 2025). AI is also being applied at the macroscopic scale, with computer vision and acoustic models routinely processing sensor data to track species populations and optimize agricultural water use in real time (Miller et al., 2025; Khan and Sharma, 2025). Ecological and biodiversity applications lag behind other biological subfields, in part because training data in these areas is sparse, biased toward well-studied taxa, and lacking standardized formats (Fahsbender et al., 2025). In species taxonomy and evolutionary biology, vision-based foundation models such as the BioCLIP family (Stevens et al., 2024; Gu et al., 2025) are enabling classification and discovery across the tree of life, while methods like PhyloNN (Elhamod et al., 2023) can identify evolutionary traits from images without labeled data.

In neuroscience, AI serves both as a practical tool for brain mapping and as a source of theoretical inspiration. Computer vision approaches have been instrumental in assembling full connectome data from model organisms such as the fly and mouse (Dorkenwald et al., 2025; The MICrONS Consortium, 2025). Comparisons between biological neuronal networks and artificial deep networks inform how researchers understand the principles of information processing in the brain (Linsley et al., 2025; Kazemian et al., 2025).

### Datasets

OpenGenome2 contains nearly 9.3 trillion base pairs of curated DNA from across all domains of life, making it the largest genomic training corpus assembled to date and the foundation for the Evo 2 model. In neuroscience, Spacetop contributed over 600 imaging hours across 101 participants for cognitive neuroscience research. These resources provide the scale necessary for foundation models to learn biological features without task-specific training, though the gap between genomic data availability and functional perturbation data remains wide (Figure 5.2.5).

### Selected datasets in biological and life sciences (2025)

| Name | Affiliation[7] | Domain | Sector | Summary |
|---|---|---|---|---|
| OpenGenome2 | Arc Institute, Stanford University, Nvidia | Biology, Genomics | ACADEMIA NONPROFIT INDUSTRY | 9.3T base pairs of curated DNA from all domains of life. Training corpus for Evo 2. |
| ProteinTalks-DB | Westlake University (Academia) | Proteomics, Systems Biology | ACADEMIA | 38M+ proteomics measurements from drug-treated breast cancer cells trained for protein dynamics and drug-response prediction. |
| Spacetop | Dartmouth College, Johns Hopkins University, Emory University | Neuroscience | ACADEMIA | 101 participants, 600-plus imaging hours. Cognitive, affective, social, interoceptive domains. |

Figure 5.2.5

## Benchmarks

Life-science benchmarks have moved to testing workflow execution and tool-integrated analysis, rather than static knowledge. BixBench reports that frontier models achieve roughly 17% accuracy on real-world bioinformatics analysis tasks, highlighting challenges in chaining tools, file handling and domain interpretation. BioML-bench provides the first end-to-end evaluation of AI agents on biomedical machine learning tasks that span protein engineering to drug discovery, and it found that on average agents underperform human baselines. These results are consistent with the pattern observed across other scientific domains in this chapter. AI systems perform well on isolated subtasks but struggle when required to execute the multistep workflows that actual biological research demands (Figure 5.2.6).

7 Full references are provided in the Appendix.

### Selected benchmarks in biological and life sciences (2025)

| Name | Affiliation[8] | Domain | Sector | Summary |
|---|---|---|---|---|
| BaisBench | Tsinghua University | Biology | ACADEMIA | Evaluates AI biological discovery ability via cell annotation and data-driven questions. |
| BioML-bench | Shift Bioscience, University of Cambridge, ScienceMachine | Biology | INDUSTRY<br>ACADEMIA | First end-to-end biomedical machine learning evaluation. Agents underperform human baselines on average. |
| BixBench | FutureHouse, ScienceMachine | Computational Biology | INDUSTRY | 50-plus bioinformatics scenarios. GPT-4o and Claude 3.5 Sonnet: ~17% accuracy. |
| CGBench | Stanford University | Biology (Genetics) | ACADEMIA | Clinical genetics interpretation. Reasoning models excel at fine-grained tasks; substantial hallucination gaps remain. |
| Mouse vs. AI: Robust Foraging Competition | UC Santa Barbara | Neuroscience | ACADEMIA | Bioinspired benchmark grounding reinforcement learning agents in neuroscience via shared foraging tasks with mice. |

Figure 5.2.6

## Foundation Models

In 2025, foundation model releases in the biological and life sciences domains expanded across genomics and cellular modeling. Genomic foundation model Evo 2, which trained on OpenGenome2, trained on 9.3 trillion DNA base pairs from all domains of life. It operates at up to 40 billion parameters with a 1 million token context window and was released with fully open weights. Chapter 6 examines its performance on genomic prediction tasks alongside smaller, task-specific alternatives, and covers additional genomic and cellular foundation models, including AlphaGenome and CellFM. In neuroscience, a foundation model of neural activity predicts neuronal responses and generalizes across stimulus types and individual animals (Figure 5.2.7).

8 Full references are provided in the Appendix.

## Selected foundation models in biological and life sciences (2025)

| Name | Affiliation[9] | Domain | Sector | Summary |
| --- | --- | --- | --- | --- |
| AlphaGenome | Google DeepMind | Biology, Genomics | INDUSTRY | Genomic foundation model predicting thousands of functional measurements from DNA sequence at single-base-pair resolution. |
| ANN Model | Baylor College of Medicine, Stanford University, University of Göttingen | Neuroscience | ACADEMIA | Neural activity FM. Predicts neuronal responses, generalizes across stimulus types and mice. |
| BioCLIP 2 | The Ohio State University, Smithsonian Institution, UNC-Chapel Hill | Biology | ACADEMIA, NONPROFIT | Vision foundation model for biological classification across the tree of life. Trained using hierarchical contrastive learning on taxonomic structure. |
| BioLab | Princeton University, BioMap Research, Zhejiang University | Biology | ACADEMIA, INDUSTRY | Multiagent system for automated biological research. Experimentally validated novel antibody designs. |
| CellFM | Sun Yat-sen University, Chongqing University, Jinfeng Laboratory | Biology | ACADEMIA, GOVERNMENT, INDUSTRY | 800M parameters, 100M human cells. Single-cell analysis, perturbation prediction, gene–gene relationships. |
| Evo 2 | Arc Institute, Nvidia, Stanford University, UC Berkeley | Biology, Genomics | ACADEMIA, INDUSTRY | 40B params, 1M token context. 9.3T base pairs. Genome-scale generation. Fully open release. |
| ProteinTalks | Westlake University, DP Technology Co., Westlake Omics Co. | Proteomics, Systems Biology | ACADEMIA, INDUSTRY | Foundation model for protein network dynamics. Predicts drug efficacy and synergy from perturbation proteome data. |

Figure 5.2.7

9 Full references are provided in the Appendix.

## AI Agents

Agent systems in the life sciences are beginning to operationalize complex research workflows, including literature synthesis and bioinformatics execution.

BCI-Agent performs autonomous neuronal cell-type classification from electrophysiology recordings without task-specific training. Biomni is a general-purpose biomedical agent spanning 25 subfields. Chapter 6 describes its architecture and capabilities in greater detail (See Figure 5.2.8).

### Selected AI agents in biological and life sciences (2025)

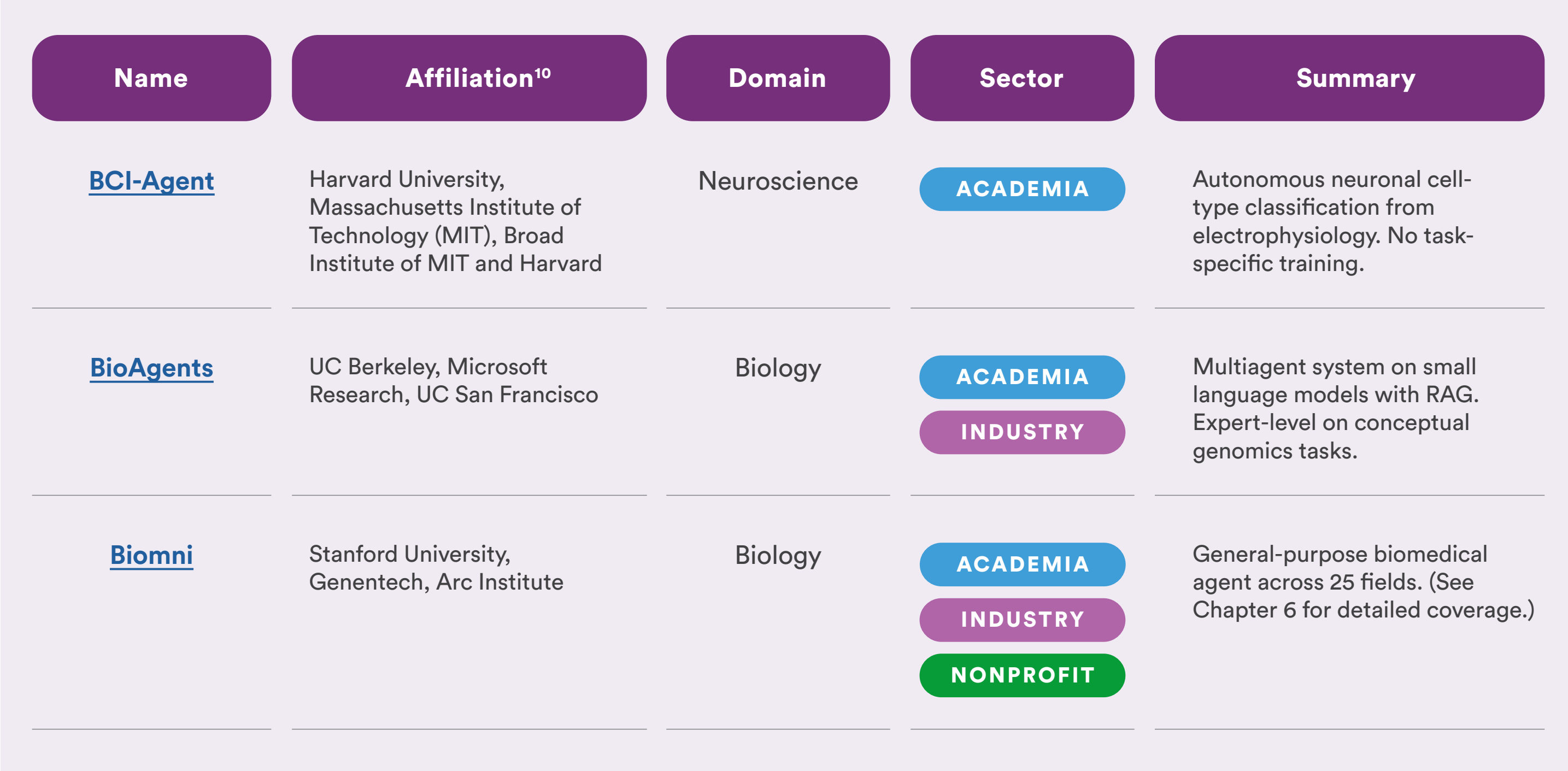

| Name | Affiliation[10] | Domain | Sector | Summary |
|---|---|---|---|---|
| BCI-Agent | Harvard University, Massachusetts Institute of Technology (MIT), Broad Institute of MIT and Harvard | Neuroscience | ACADEMIA | Autonomous neuronal cell-type classification from electrophysiology. No task-specific training. |
| BioAgents | UC Berkeley, Microsoft Research, UC San Francisco | Biology | ACADEMIA, INDUSTRY | Multiagent system on small language models with RAG. Expert-level on conceptual genomics tasks. |
| Biomni | Stanford University, Genentech, Arc Institute | Biology | ACADEMIA, INDUSTRY, NONPROFIT | General-purpose biomedical agent across 25 fields. (See Chapter 6 for detailed coverage.) |

Figure 5.2.8

## Earth Science

Progress in AI for Earth science remains aligned with observational infrastructure, including reanalysis datasets and global satellite archives. Weather forecasting, which benefits from decades of reanalysis datasets such as ERA5 and dense global satellite archives, has advanced furthest, with multiple AI models being used in real forecasting systems in 2025. Climate modeling lags behind because it requires projections on decadal timescales where future states fall outside the distribution of any existing training data. Hydrology offers one of the clearest examples of benchmark-driven progress in scientific AI. LSTM-based models, trained jointly across hundreds of catchments in the CAMELS dataset, have consistently outperformed process-based hydrologic models (Kratzert et al., 2019), and regional extensions now span the United States, the United Kingdom, Australia, Chile, and Brazil. Agriculture presents the opposite pattern, with

10 Full references are provided in the Appendix.

shared benchmark datasets still scarce, making progress difficult to measure across research groups despite promising work in knowledge-guided approaches to carbon cycle quantification (Liu et al., 2024) and global change ecology (Jin et al., 2026).

## Datasets

Earth science relies heavily on large governmental and institutional observation systems rather than purpose-built AI training corpora. In carbon flux research, global flux tower networks provide foundational observational data. FLUXNET2015 aggregates eddy covariance measurements from sites worldwide (Pastorello et al., 2020), while regional networks including AmeriFlux (North America), ICOS (Europe), and JapanFlux (Japan and East Asia) contribute additional coverage. These datasets enable training and evaluation of models that upscale local carbon flux measurements to regional and global estimates (Figure 5.2.9).

### Selected datasets in Earth science (2025)

| Name | Affiliation[11] | Domain | Sector | Summary |
|---|---|---|---|---|
| AmeriFlux | Indiana University, USDA Agricultural Research Service, University of Wisconsin-Madison | Ecology | ACADEMIA, GOVERNMENT | North American network of 260-plus flux tower sites measuring ecosystem carbon, water, and energy exchange. Over 50 sites with 10-plus years of continuous data. |
| CAMELS | NSF National Center for Atmospheric Research, U.S. Geological Survey, U.S. Department of the Interior | Hydrology | GOVERNMENT | Standardized data on terrain, climate, soil, and streamflow for 671 U.S. river basins. Foundation for AI hydrology benchmarking. |
| FLUXNET2015 | Lawrence Berkeley National Laboratory, University of Tuscia, ETH Zurich | Ecology | ACADEMIA, GOVERNMENT | Global measurements of CO2, water, and energy exchange between ecosystems and atmosphere from 212 sites worldwide. |
| ICOS | ICOS ERIC/University of Helsinki, Thünen Institute of Climate-Smart Agriculture, ETH Zurich | Ecology | ACADEMIA, GOVERNMENT | European observation network of 140-plus stations across 12 countries measuring greenhouse gas concentrations and carbon fluxes across atmosphere, land, and ocean. |
| JapanFlux | Osaka Metropolitan University, Chiba University, National Institute of Polar Research (NIPR) | Ecology | ACADEMIA, GOVERNMENT | Land-atmosphere flux measurements covering Japan and East Asia from 1990 to 2023. Tracks energy, water, and CO2 exchange across Asian ecosystems. |

Figure 5.2.9

11 Full references are provided in the Appendix.

## Benchmarks

In 2025, benchmarks in Earth science expanded into reliability of extreme event coverage, where AI weather models, for example, face the highest stakes. It is also where standard average skill metrics fail to capture performance (Figure 5.2.10).

### Selected benchmarks in Earth science (2025)

| Name | Affiliation[12] | Domain | Sector | Summary |
| --- | --- | --- | --- | --- |
| EarthSE | Shanghai Jiao Tong University, Shanghai Artificial Intelligence Laboratory, Hong Kong Polytechnic University | Earth Science | ACADEMIA, INDUSTRY | 100K papers, 114 disciplines, 11 LLMs tested. Significant gaps in Earth science exploration. |
| ExEBench | Technical University of Munich (TUM), Munich Center for Machine Learning (MCML) | Atmospheric Sciences | ACADEMIA | 7 extreme event categories. Global coverage. Tests detection, monitoring, and forecasting. |
| UnivEARTH | Cornell University, Columbia University | Earth Science | ACADEMIA | 140 Earth observation questions. LLM agents: 33% accuracy. Code fails 58% of time. |

Figure 5.2.10

## Foundation Models

Earth science foundation models released in 2025 covered weather forecasting, climate emulation, and geospatial representation. In weather forecasting, several systems built directly on models highlighted in the 2025 AI Index. FourCastNet 3 generates a 60-day global forecast at 0.25-degree resolution in under 4 minutes on a single GPU, running 8 to 60 times faster than prior approaches. For Earth observation, TerraMind is the first any-to-any generative multimodal model operating across 9 geospatial modalities (Figure 5.2.11).

12 Full references are provided in the Appendix.

## Selected foundation models in Earth science (2025)

| Name | Affiliation[13] | Domain | Sector | Summary |
|---|---|---|---|---|
| AlphaEarth | Google DeepMind | Earth observation | INDUSTRY | Embedding field model for general geospatial representation. Outperforms prior featurization approaches. |
| cBottle (Climate in a Bottle) | Nvidia | Climate Science | INDUSTRY | Diffusion-based climate emulator. Global 5 km at 12.5M-pixel resolution. Diurnal-to-seasonal variability. |
| FourCastNet 3 | Nvidia, Lawrence Berkeley National Laboratory, UC Berkeley | Weather Forecasting | INDUSTRY, GOVERNMENT, ACADEMIA | 60-day forecast in <4 min/GPU. 8–60x faster. Builds on Aurora and NeuralGCM advances. |
| GAIA | USRA/RIACS, BCG X AI Science Institute | Atmospheric Sciences | INDUSTRY, NONPROFIT | Atmospheric FM from 15 years of satellite imagery. Atmospheric rivers (F1: 0.58), cyclone detection (81% recall). |
| OlmoEarth | Allen Institute for AI, University of Washington, Arizona State University | Earth Observation | NONPROFIT, ACADEMIA | Multimodal spatiotemporal Earth observation FM. Platform for NGOs and nonprofits. |
| TerraMind | IBM Research, ETH Zurich, Forschungszentrum Jülich | Earth Observation | INDUSTRY, ACADEMIA, GOVERNMENT | First any-to-any generative multimodal Earth observation FM. 9 geospatial modalities. |
| WeatherNext 2 | Google DeepMind | Weather Forecasting | INDUSTRY | Hundreds of weather outcomes in <1 min/TPU. 99.9% improvement over predecessor. Builds on GenCast. |

Figure 5.2.11

13 Full references are provided in the Appendix.

## AI Agents

In Earth science, AI agents are moving beyond data retrieval toward executing full research workflows, including automated observation processing, literature-informed analysis, and climate task completion. ClimateAgent completed 85 climate tasks with 100% completion and a quality score of 8.32, compared with 6.27 for Microsoft Copilot and 3.26 for GPT-5 (Figure 5.2.12).

### Selected AI agents in Earth science (2025)

| Name | Affiliation[14] | Domain | Sector | Summary |
|---|---|---|---|---|
| ClimateAgent | The Hong Kong University of Science and Technology | Climate Science | ACADEMIA | 85 climate tasks: 100% completion, quality 8.32 vs. Copilot 6.27, GPT-5 3.26. |
| EarthLink | Shanghai Artificial Intelligence Laboratory, Fudan University, University of Sydney | Climate Science | INDUSTRY<br>ACADEMIA | First AI copilot for Earth scientists. Automated research workflows with dynamic feedback loop. |
| PANGAEA GPT | Alfred Wegener Institute for Polar and Marine Research | Earth Science | GOVERNMENT | Multi-agent system for PANGAEA Earth science database. Intelligent data processing, natural language interface. |

Figure 5.2.12

## Mathematics

Mathematical reasoning is another active testing ground for AI capabilities. Systems such as Goedel-Prover are moving toward automated formal proof generation in languages like Lean. Competition-level problem-solving and formal verification of known results are advancing quickly, but major open problems, such as long-standing Erdos conjectures, remain well beyond current capabilities. Chapter 2 covers benchmark performance in detail, including a jump from silver to gold medal at the International Mathematical Olympiad in a single year and rapid gains on FrontierMath and MathArena.

14 Full references are provided in the Appendix.

# 5.3 AI Agents and Tools for Science Workflows

The domain-specific tables of Section 5.2 catalog a growing inventory of AI agents, foundation models, datasets, and benchmark suites. Two cross-domain benchmarks released in 2025 offer a broader view of how well these systems perform when asked to do end-to-end scientific research rather than isolated tasks. On both benchmarks, even the best-performing agents fall well below expert-level performance.

## AstaBench

AstaBench is an end-to-end benchmark suite that evaluates agentic scientific research ability across over 2,400 problems spanning multiple domains and the full discovery workflow, from literature understanding through code execution, data analysis, and end-to-end discovery. It benchmarked 57 agents across 22 agent classes and reported both an overall score and cost per problem (Figure 5.3.1). The best performing agent scored around 0.53 at a cost of roughly $3.40 per problem, while most agents clustered between 0.10 and 0.45 at per-problem costs below $1.00.

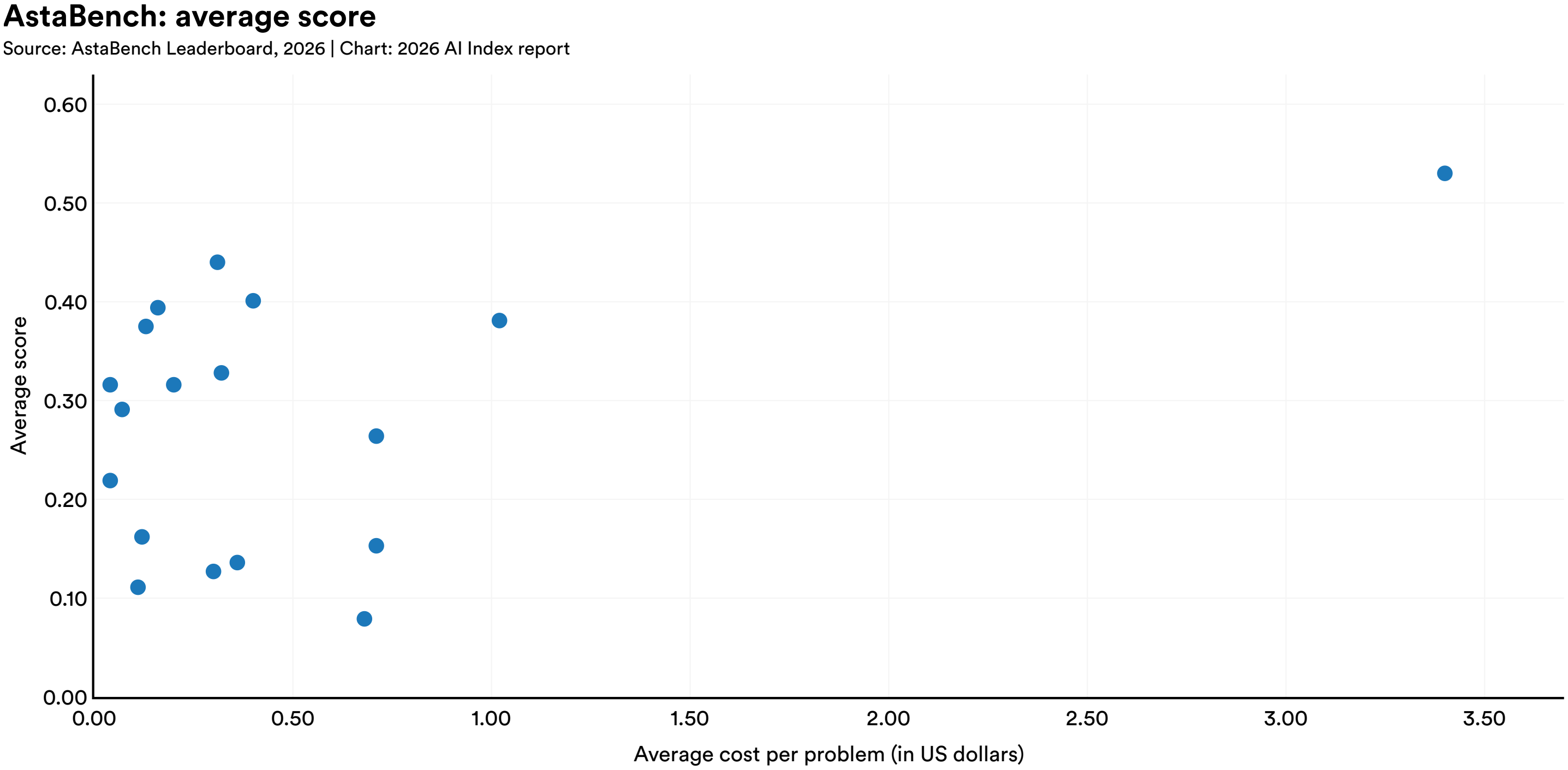
AstaBench: average score
Source: AstaBench Leaderboard, 2026 | Chart: 2026 AI Index report

Figure 5.3.1

## PaperArena

PaperArena is a benchmark that tests whether LLM agents can answer real research questions that require stitching together evidence across multiple papers while orchestrating external tools for parsing, retrieval, and computation. Gemini 2.5 Pro performs best overall, achieving 38.8% average accuracy in a multiagent configuration (Figure 5.3.2). All tested agents lagged substantially behind the PhD expert baseline of 83.50%. Multiagent configurations consistently outperformed single-agent setups across all models tested, though the gains were modest, typically 2 to 4 percentage points.

**PaperArena: single vs. multiagent performance**

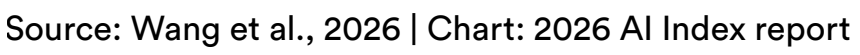
Source: Wang et al., 2026 | Chart: 2026 AI Index report

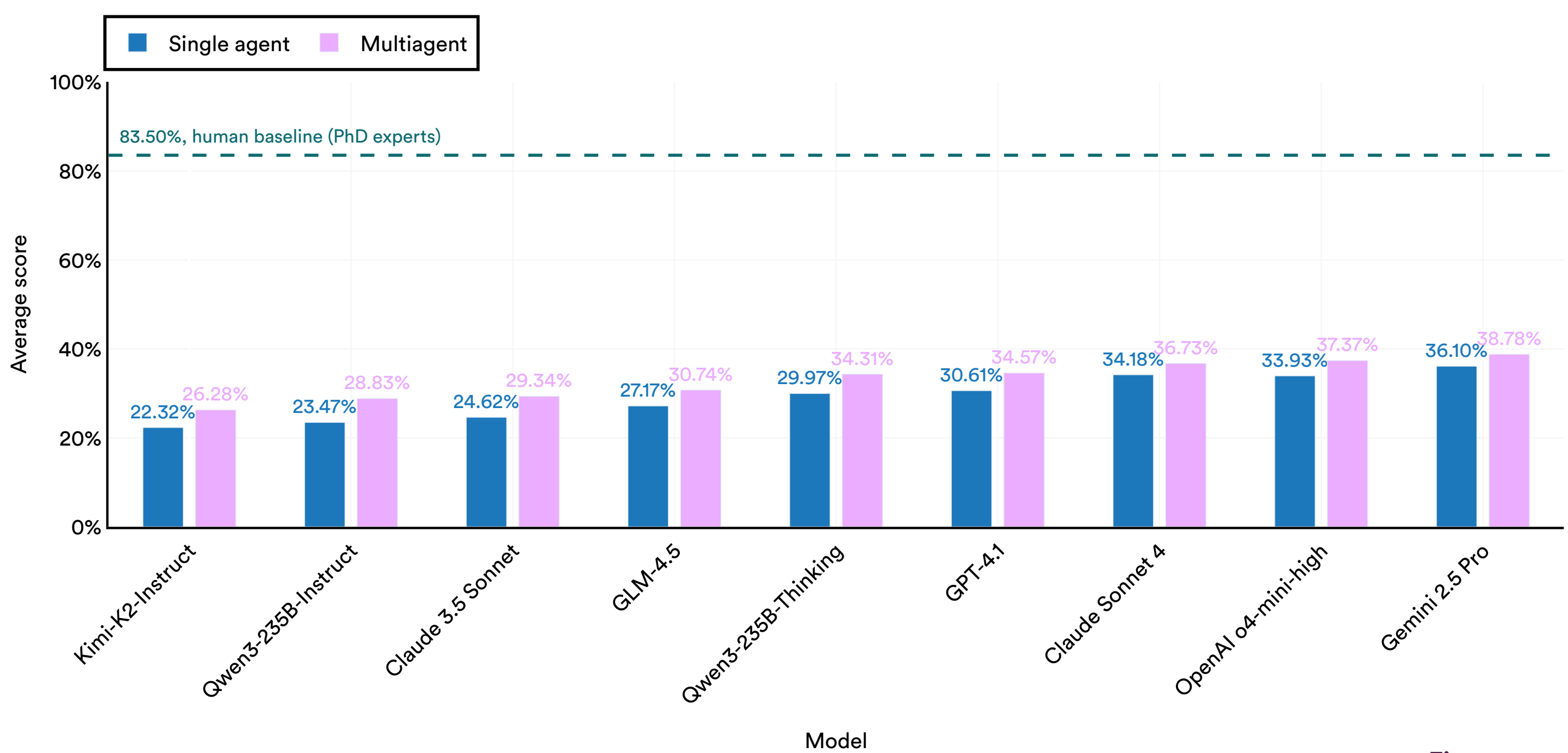

Figure 5.3.2

# AI Agents

## AI as a Co-scientist

In 2025, several research groups released systems in which multiple AI agents divide scientific tasks among themselves, with separate agents handling literature search, hypothesis generation, code execution, and review. The multiagent systems are designed to approximate the structure of a human research team, rather than relying on a single model or person to perform every step. The most prominent example, Google's AI Co-scientist (Gottweis et al., 2025), uses a generate-debate-evolve loop in which agents iteratively produce and refine evidence-grounded hypotheses. The system was validated in three biomedical areas, including AML drug repurposing and liver fibrosis targets, and achieved a top-1 accuracy of 78.4% on the GPQA Diamond set when selecting its highest-rated hypothesis per question (Figure 5.3.3).

Other multiagent systems pursued different approaches to the same goal. Sakana's AI Scientist-v2 (Yamada et al., 2025) produced the first fully AI-generated paper accepted at a peer-reviewed workshop (ICLR), using agentic tree search to generate and refine code implementations without human-coded templates. Kosmos (Mitchener et al., 2025) maintained coherence across runs lasting up to 12 hours, executing an average of 42,000 lines of code and reading 1,500 papers per run, with collaborators reporting that a single run approximated six months of research. SciToolAgent (Ding et al., 2025) automates hundreds of scientific tools across biology, chemistry, and materials science via knowledge-graph-driven retrieval, outperforming prior agent frameworks by 10 to 20 percentage points on multi-tool tasks (Figure 5.3.4).

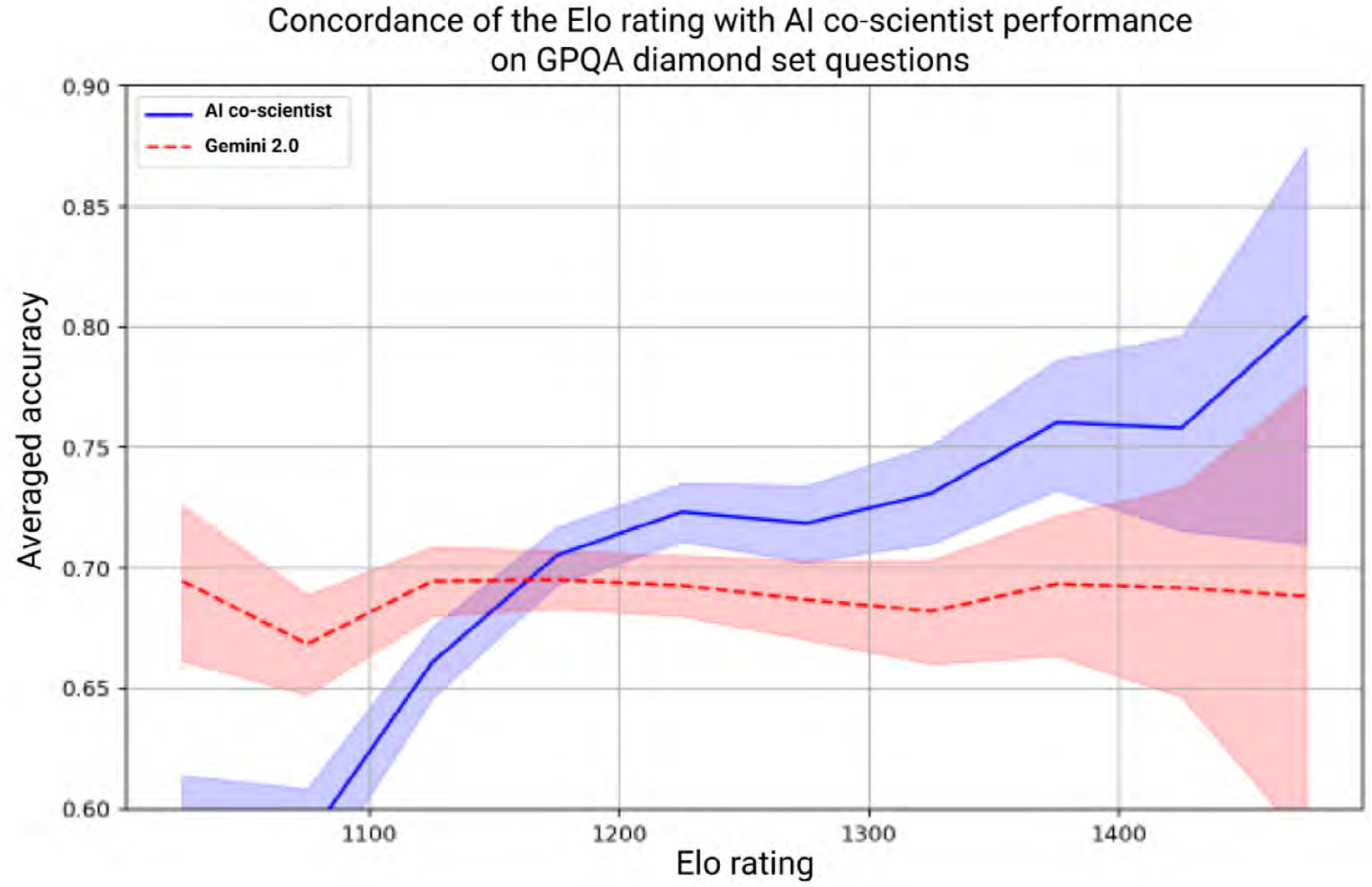

Source: Gottweis et al., 2025

**Figure 5.3.2**

Despite these advances, only a handful of multiagent systems have produced results that were tested and confirmed through real-world experiments. Published examples include new proteins designed by ProtAgents; 92 antibody candidates for SARS-CoV-2 from the Virtual Lab (of which more than 90% successfully bound their target); two new cancer drug targets, GPR160 and ARG2, from OriGene; five novel metal-organic frameworks from MOFGen; and a novel chromophore from ChemCrow. The gap between what these systems can propose computationally and what has been confirmed experimentally remains wide. Key roadblocks for the field include workforce training gaps, a lack of API and interoperability standards, and funding structures that do not yet support the maintenance and scaling of autonomous research infrastructure.

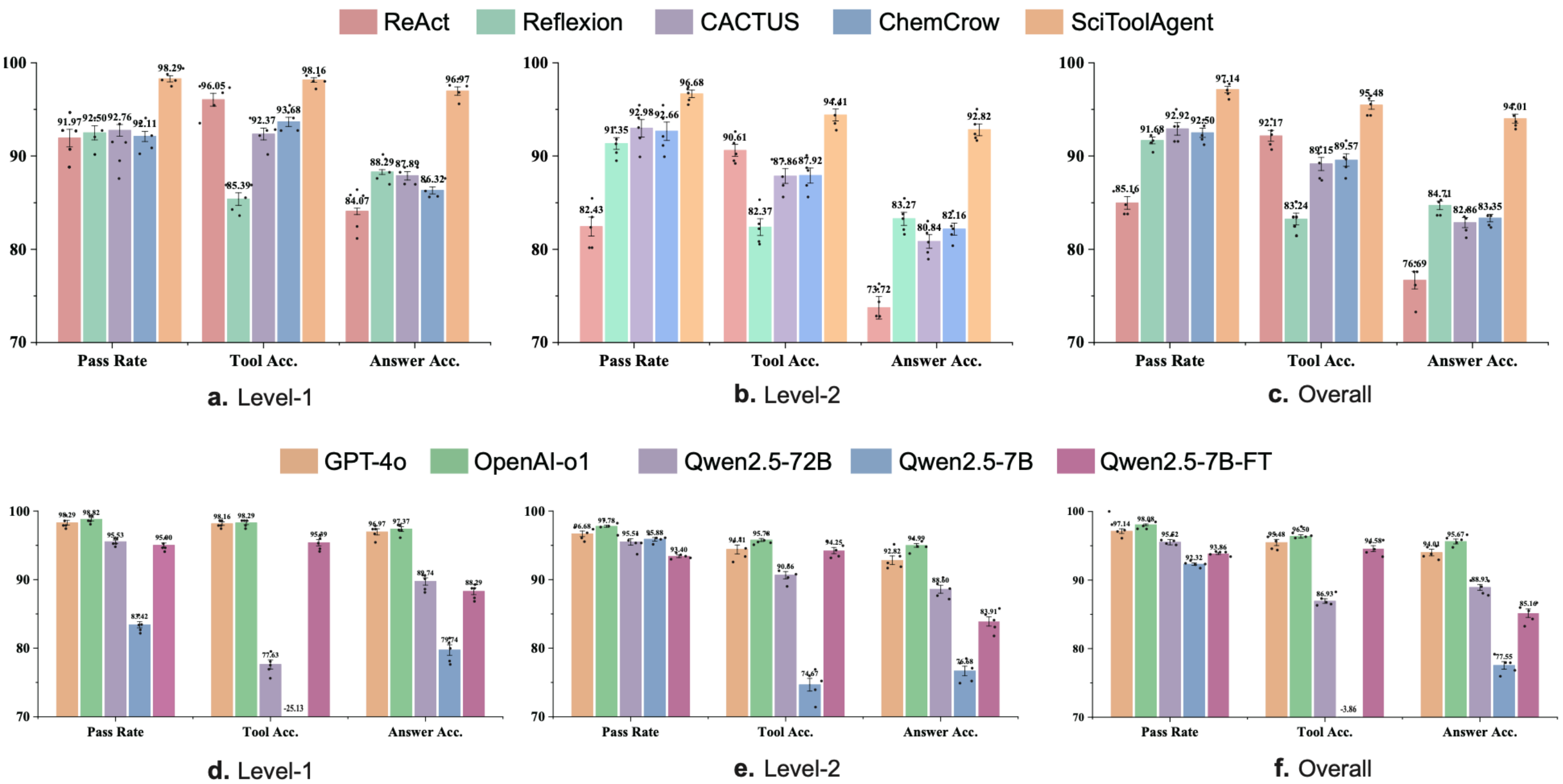

Source: Ding et al., 2025

**Figure 5.3.2**

# 6

# Medicine

## Overview

AI in medicine advanced on multiple fronts in 2025, but strong model performance has not consistently translated into real-world clinical impact. In molecular biology, where understanding protein behavior is fundamental to drug development, AI-driven protein research continued to grow, and smaller, more specialized models matched or outperformed larger general-purpose systems on protein structure prediction, genomics, and drug discovery. On clinical reasoning tasks, leading AI models now score higher than most physicians on structured clinical evaluations, yet nearly half of clinical AI studies still rely on simulated scenarios rather than real patient data. The tools gaining traction in practice are those that support clinicians' existing workflows, such as ambient AI scribes and sepsis prediction systems. Authorizations from the U.S. Food and Drug Administration (FDA) for AI-enabled medical devices increased, but clinical evidence continues to lag behind. Patients are also encountering AI-generated health information directly in search results, often before they speak to a clinician, with even less oversight or vetting than the tools moving through formal regulatory channels. AI's impact on medicine is clear, but realizing it at scale will require clinical evidence, governance, and ethical frameworks.

## Contents

# Chapter Highlights

1. **In molecular biology, smaller models are outperforming larger ones.** MSAPairformer, a 111-million-parameter protein language model, outperformed previous leading methods on the benchmark, ProteinGym; and GPN-Star, a 200-million-parameter genomics model, outperformed a model with 40 billion parameters.

2. **Virtual cell models emerged as a new frontier in 2025, with major releases including Evo 2 from the Arc Institute, STATE, and DeepMind's AlphaGenome.** These models aim to predict cellular responses to drugs and genetic perturbations without running wet-lab experiments, though current systems still require experimental validation.

3. **Like other areas of AI, biological model development is increasingly bottlenecked on data rather than architecture.** With cofolding models now representing all structure types in the Protein Data Bank, 2025 saw a turn toward distilled datasets of AI-predicted structures and training on combined experimental data sources, expanding training sets from hundreds of thousands of entries to tens of millions.

4. **AI tools that automatically generate clinical notes from patient visits saw broad adoption in 2025.** Across multiple hospital systems, physicians reported they were spending up to 83% less time writing notes, experiencing significant reductions in burnout, with one hospital system reporting a 112% return on investment.

5. **The FDA authorized 258 AI medical devices in 2025, most through pathways that do not require new clinical trials.** The vast majority entered the market via device-modification pathways that rely on existing safety and efficacy evidence rather than new randomized trials, with only 2.4% of devices with clinical studies supported by randomized trial data.

6. **A multi-agent AI system scored 85.5% on complex published case studies, versus 20% for unaided physicians.** Microsoft's AI Diagnostic Orchestrator, paired with OpenAI's o3, was tested on challenging cases drawn from the medical literature against physicians working without their usual tools. Multi-agent frameworks more broadly have shown diagnostic accuracy gains of 7% to over 60% over single-agent baselines.

7. **AI-generated summaries now appear at the top of 84% to 92% of health-related Google searches.** Symptom and common health questions trigger an AI Overview 92% of the time, followed by treatment and condition queries. These summaries are now a routine feature of health information searches, shaping the initial interpretation of users' questions.

8. **Ethics discussion in medical AI publications more than doubled in 2025, but the conversation is narrow.** Governance dominates the discourse, while algorithm accountability, biosecurity, and global health equity remain underexplored.

9. **Research interest in medical digital twins is growing fast, and where rigorous trials exist, early results are promising.** In a randomized trial of 150 diabetes patients, 71% achieved healthy blood sugar levels over one year while safely reducing their medications.

# 6.1 The Central Dogma

AI models for molecular biology span the pathway from gene sequence to protein structure to therapeutic design. This section tracks advances in protein language models, structure prediction, protein design, virtual cell models, and multimodal foundation models for biomedical discovery. The analysis draws on PubMed publication counts, benchmark evaluations including ProteinGym and FoldBench, and model release data from 2024 and 2025. A recurring pattern across these areas is the tension between scale and specialization. In several areas, smaller or more targeted models matched or outperformed larger general-purpose systems.

## Research Volume

AI-driven protein research grew approximately 71% between 2024 and 2025 (Figure 6.1.1). Total publications across four categories—function prediction, protein structure prediction, protein-drug interactions, and synthetic protein design—rose from 2,259 in 2024 to 3,855 in 2025. Protein-drug interactions represented the largest share of output in both years, accounting for 49.9% of papers in 2024 and rising to 54.4% in 2025. Protein structure prediction constituted the second-largest category in 2024 at 28.7%, though its relative share declined to 23.9% in 2025. Function prediction and synthetic protein design each remained comparatively stable, with function prediction increasing from 9.7% to 10.4% and synthetic protein design decreasing slightly from 11.7% to 11.3%. The shift in relative share toward protein-drug interactions, even as absolute counts grew across all categories, may reflect maturing structure prediction methods and growing interest in therapeutics applications. Publications focused specifically on AI for drug discovery have followed a similar upward trajectory (Figure 6.1.2).

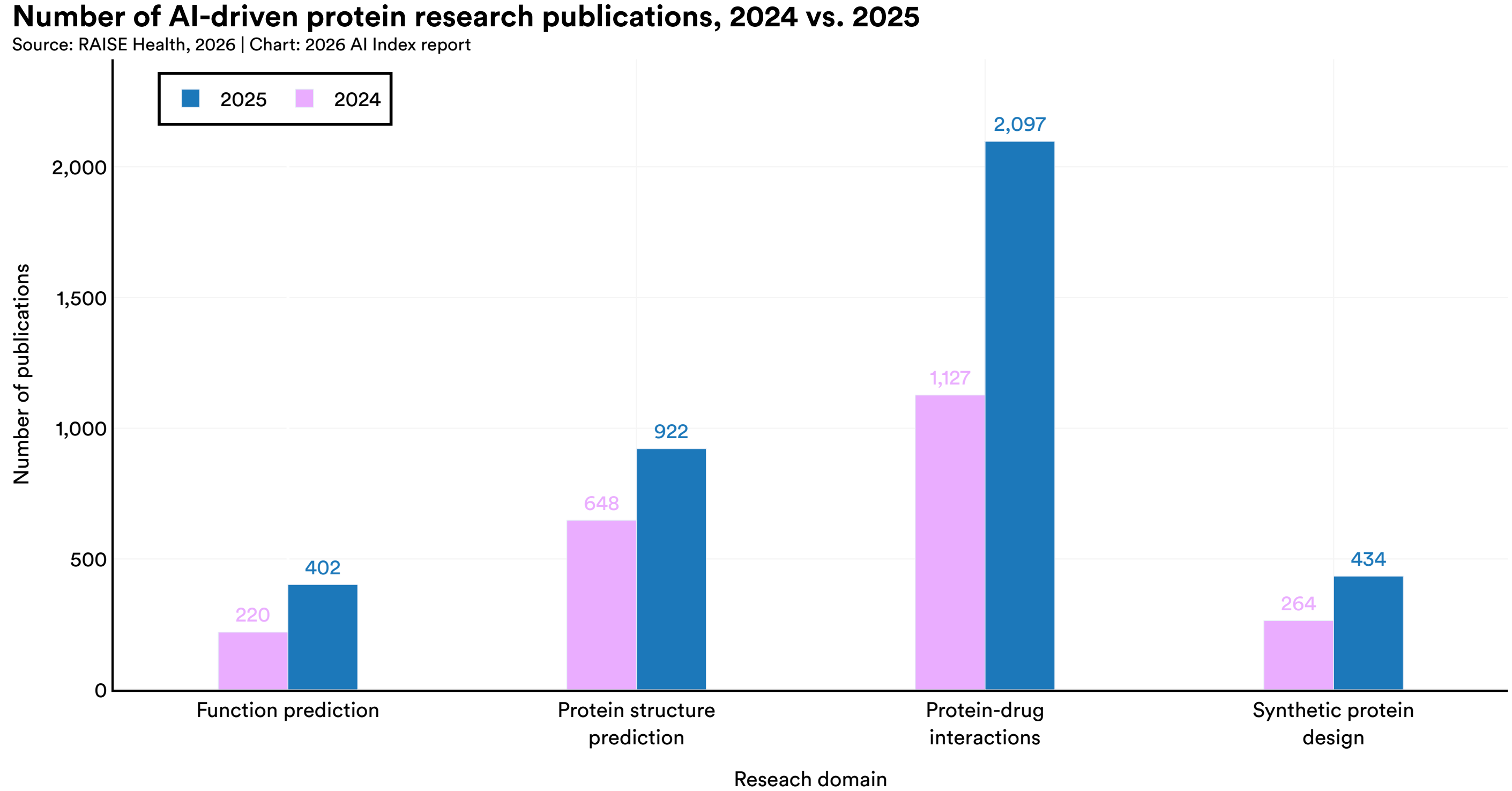

Figure 6.1.1

**Number of publications on AI for drug discovery, 2018–25**

Source: RAISE Health, 2026 | Chart: 2026 AI Index report

Number of publications

3,500
3,000
2,500
2,000
1,500
1,000
500
0

2018: 431
2019: 612
2020: 720
2021: 1,083
2022: 1,229
2023: 1,470
2024: 2,100
2025: 3,311

Figure 6.1.2

# Public Datasets

## Molecular and Cellular Biology

Demand for training data has continued to grow as AI models have gained further adoption in biology. Rapidly collecting new biological data is typically time-consuming and expensive. In 2025, biological AI models increasingly trained on multiple datasets with different types of experimental measurements. Several cofolding methods (where two or more molecules are modeled simultaneously), for example, began training on both structural data from the Protein Data Bank (PDB) and experimental small-molecule binding affinity measurements from repositories such as PubChem, ChEMBL, and BindingDB.

The scale of publicly available biological datasets has grown by several orders of magnitude since the PDB's founding in 1971, though direct size comparisons across databases should be interpreted with caution. The unit of measurement differs by source: An "entry" may represent an experimentally solved protein structure, a bioactivity measurement, a gene sequence, or a single-cell observation.

Other models are trained on synthetically generated data. AlphaFold 3 and its open-source replications, including Boltz-2 and OpenFold3, all train on "self-distillation" datasets of predicted protein structures from AlphaFold 2 that have been filtered for quality. Meta FAIR released Open Molecules 2025 (OMol25), a dataset of over 100 million quantum mechanics calculations of molecules.

New experimental datasets also debuted in 2025. Tahoe-100M, the largest publicly released, single-cell sequencing dataset, contains measurements from over 50 cancer cell types exposed to more than 1,100 drugs. BaseData features over 9.8 billion genes obtained through metagenomic mining (Figure 6.1.3).

#### Size of public datasets for molecular and cellular biology
Source: RAISE Health, 2026 | Chart: 2026 AI Index report

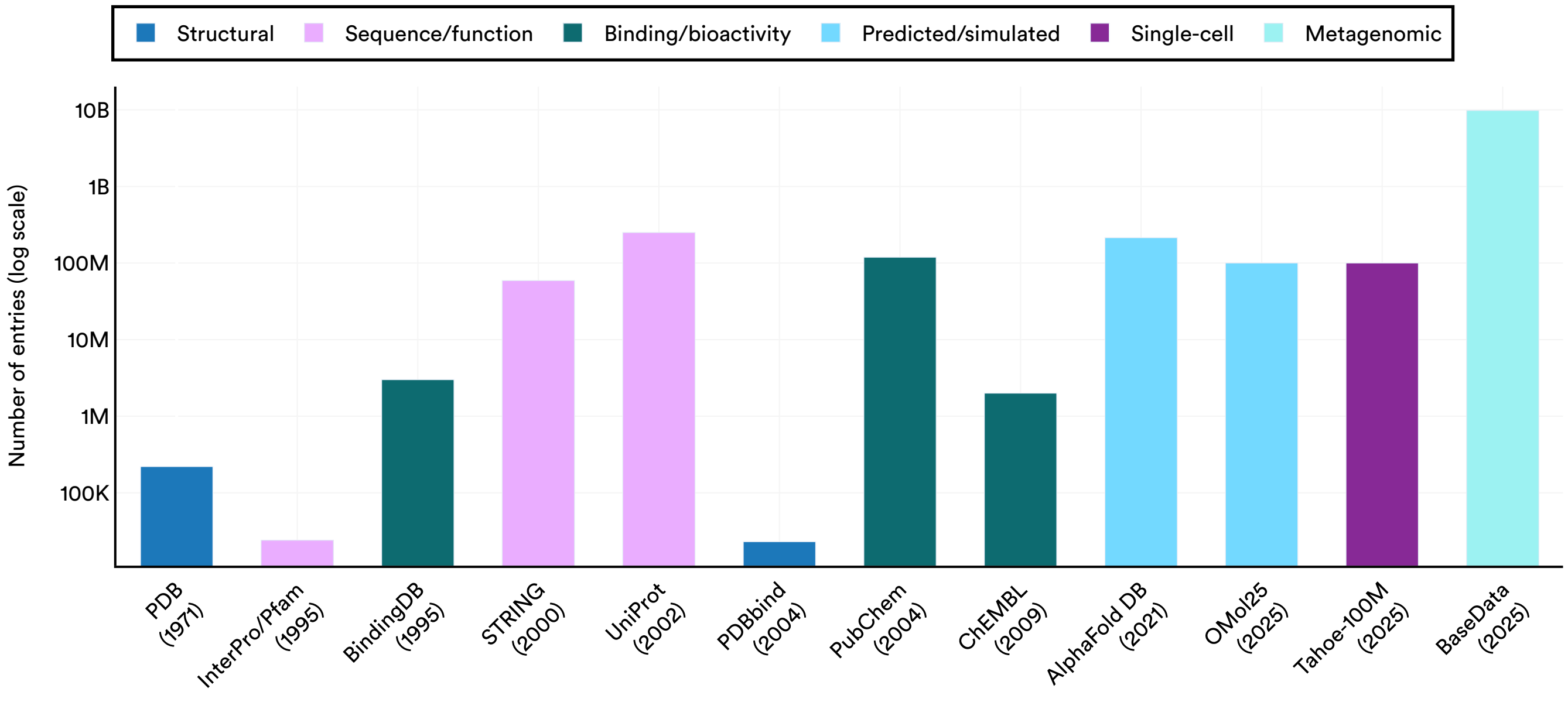

Figure 6.1.3

## Data for Biomedical Vision-Language Models

Training biomedical vision-language models requires large repositories of images and captions that can be transformed into a continual pretraining dataset. In the general domain, data scaling is often considered a mature or saturated research direction, but this does not appear to hold in the biomedical setting. Newer datasets extend beyond a single specialty and incorporate a broader range of modalities and biomedical domains (Figure 6.1.4).

#### Size of select biomedical datasets at release, 2020–25
Source: RAISE Health, 2026 | Chart: 2026 AI Index report

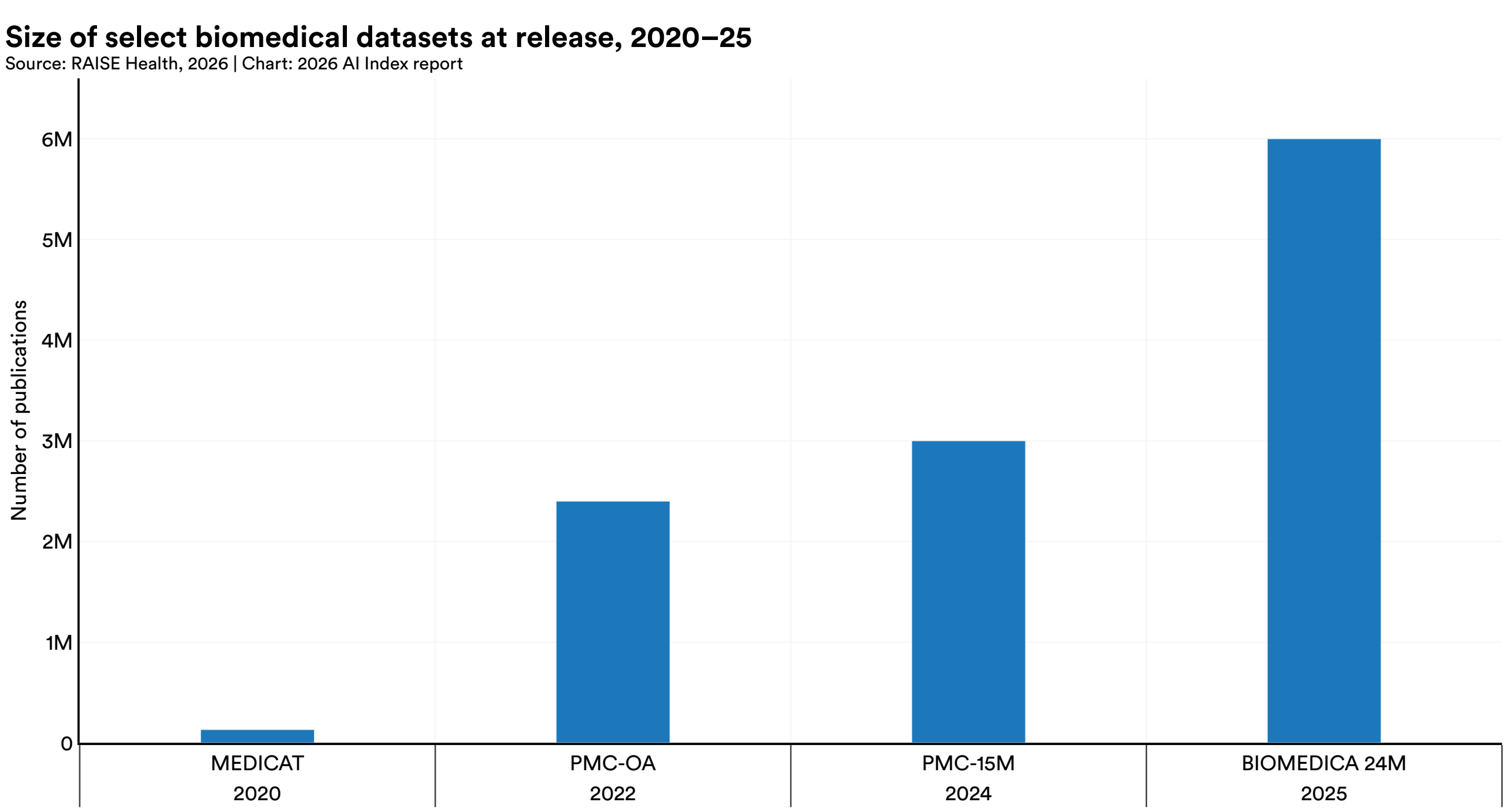

Figure 6.1.4

## Sequence-Based Models: Protein Language Models

The trend in protein language models (PLMs) shifted in 2025 from scaling model size to improving model efficiency and specialization. In 2024, efforts culminated in the 98-billion-parameter ESM3. In 2025, the focus turned to smaller architectures trained on curated data or augmented with retrieval methods (Figure 6.1.5).

ProteinGym is a comprehensive benchmark suite for protein fitness prediction and design, comprising over 250 standardized deep mutational scanning assays with millions of mutated sequences and curated clinical datasets with expert-annotated mutation effects. MSAPairformer, a 111 million-parameter model trained on multiple sequence alignments and physical constraints, surpassed previous state-of-the-art methods at a fraction of the training and parameter budget (Figure 6.1.6). The Profluent E1 series similarly set new performance standards by combining a smaller model with a retrieval augmented generation (RAG) approach. While certain tasks still benefit from larger models, others—such as predicting cellular localization—appear to vary more with training method and data than with parameter count.

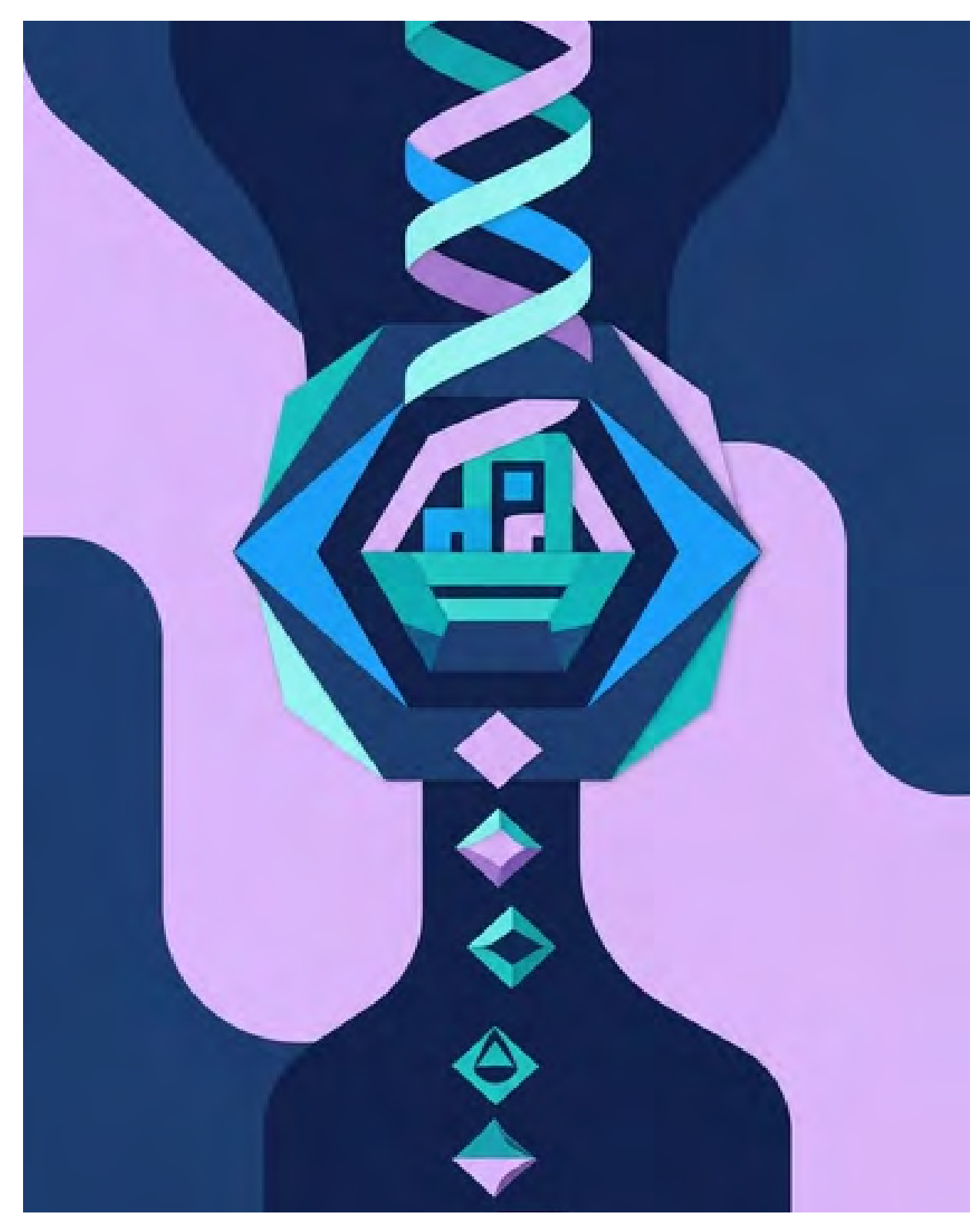

**Size of protein sequencing models, 2020–25**

Source: RAISE Health, 2026 | Chart: 2026 AI Index report

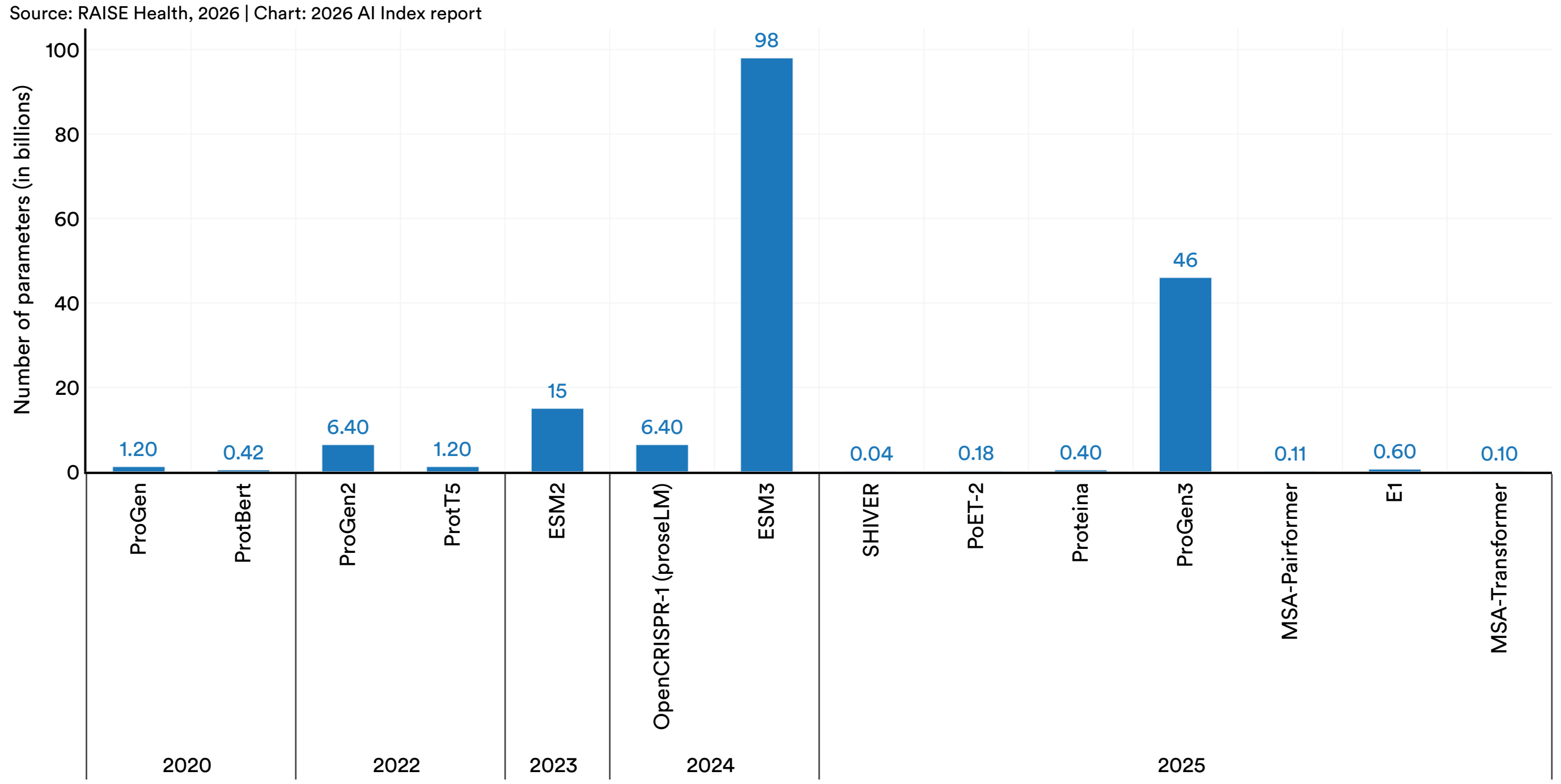

Figure 6.1.5

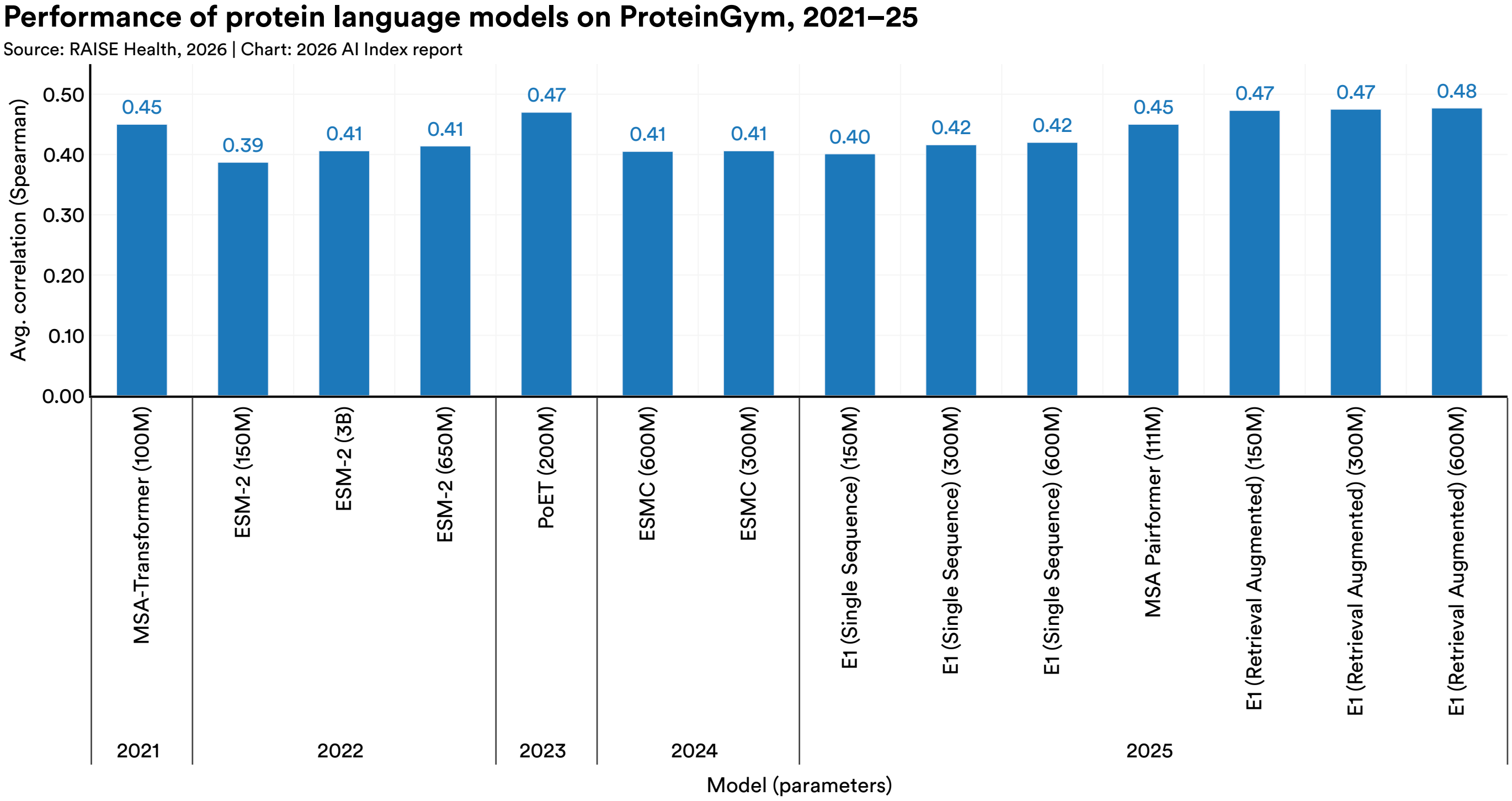

Figure 6.1.6

Beyond benchmark performance, PLMs have also become more task-specific. The ESM-C series demonstrated that smaller models geared toward a single task, such as representation learning, could be successful without the full feature set of the ESM3 family. A fine-tuned ProGen model (6 billion parameters) was used to design a novel CRISPR-Cas protein, OpenCRISPR-1, which demonstrated improved specificity relative to standard SpCas9.

## Structure Prediction and Cofolding Models

Multiple open-source structure prediction models were released in 2025, inspired by the architecture of AlphaFold 3. These models tackle the task of "cofolding"—predicting the three-dimensional structures formed by combinations of proteins, nucleic acids, drugs, and other biomolecules. While AlphaFold 3 retains a performance advantage on certain tasks, most cofolding models have demonstrated similar performance across protein structure and biomolecular complex modeling tasks. Some, including the Boltz series and OpenFold3, are released under commercially permissive licenses.

Because cofolding models can now represent all structure types available in the PDB, further performance gains will likely require new data sources or deeper extraction of signal from existing data. One strategy is the use of "distilled" datasets, in which AI-predicted protein structures supplement experimentally determined ones, scaling training datasets from hundreds of thousands of entries to tens of millions. Boltz-2 illustrates a complementary approach—training on both structural information from the PDB and experimental binding affinity measurements—to combine specialization and scale (Figure 6.1.7). The four leading cofolding models share a common foundation of experimental and distilled structural data, but they diverge in their use of additional sources such as molecular dynamics simulations, binding affinity measurements, and RNA structure.

### Training data sources for cofolding models, 2025

| Model | Structural data (experimental) | Structural data (distilled) | Molecular dynamics | Binding affinity | RNA structure |
|---|---|---|---|---|---|
| AlphaFold 3 | ✓ | ✓ | | ✓ | |
| Boltz-2 | ✓ | ✓ | ✓ | ✓ | |
| SimpleFold | ✓ | ✓ | ✓ | | |
| OpenFold3 | ✓ | ✓ | | ✓ | ✓ |

Figure 6.1.6[1]

Similar to trends in other areas of AI, bigger models have not necessarily translated to better performance in protein structure prediction. Following the release of AlphaFold 3, subsequent models have converged on a similar parameter scale rather than continuing to grow (Figure 6.1.8). FoldBench is a benchmark that tests whether a model can correctly predict how a small molecule, such as a drug candidate, physically binds to a target protein. AlphaFold 3's performance on FoldBench has yet to be significantly surpassed even though several larger models have been released since (Figure 6.1.9). These results suggest that data, rather than model size, is an important bottleneck in protein structure prediction.

1 Rows represent individual cofolding models. Columns indicate whether each model incorporated a given data type during training, including experimentally determined and distilled protein structures, molecular dynamics simulations, binding affinity measurements, and RNA structural data. A check mark indicates that the data type was used.

## Size of protein structure prediction models, 2021–25

Source: RAISE Health, 2026 | Chart: 2026 AI Index report

Number of parameters (log scale)

100B
10B
1B
100M
10M

|  |  | Parameters |
|---|---|---|
| 2021 | AlphaFold 2 | 93M |
| 2021 | RoseTTAFold | 14M |
| 2022 | OpenFold | 93M |
| 2024 | RoseTTAFold All Atom | 83M |
| 2024 | AlphaFold 3 | 370M |
| 2024 | ESM3 | 98B |
| 2024 | HelixFold3 | 370M |
| 2024 | Chai-1 | 440M |
| 2024 | Boltz-1 | 610M |
| 2025 | Boltz-2 | 521M |
| 2025 | Protenix-Tiny | 109M |
| 2025 | Protenix-Mini | 135M |
| 2025 | SimpleFold | 3B |
| 2025 | RosettaFold-3 | 730M |
| 2025 | Protenix v0.7 | 370M |
| 2025 | SeedFold | 923M |

Figure 6.1.8

## FoldBench: protein cofolding performance, 2024–25

Source: RAISE Health, 2026 | Chart: 2026 AI Index report

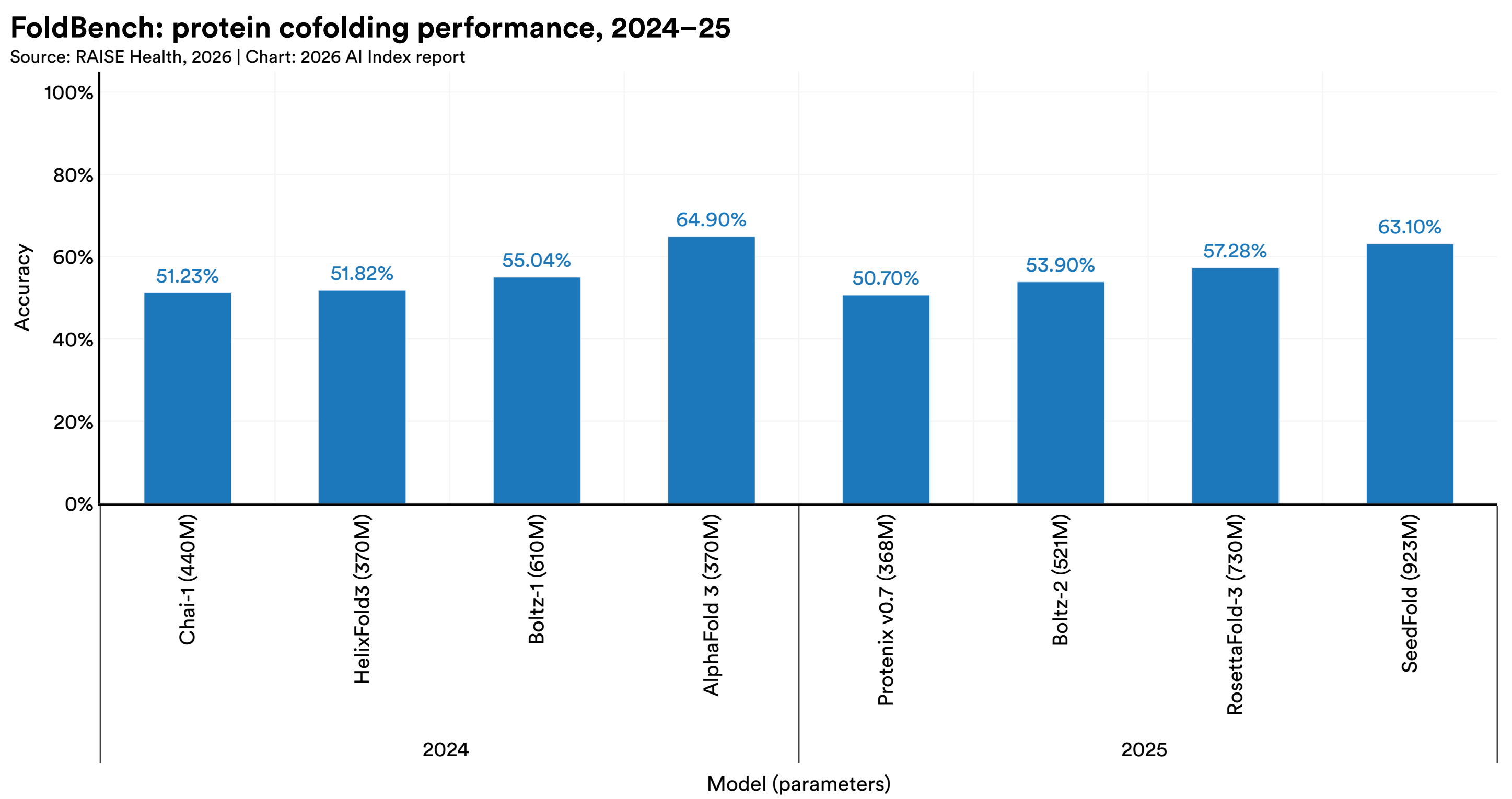

Figure 6.1.9

## Protein Design and Generative Models for Therapeutics

Advances in cofolding have enabled a new generation of generative models for protein design, including methods for designing antibodies, nanobodies, and peptides. Approaches range from workflows built around existing structure prediction methods (e.g., BindCraft, Germinal) to models directly trained for generating binders (e.g., RFDiffusion, BoltzGen).

A protein design challenge hosted by Adatpyv Bio in 2025 provided a controlled comparison. Multiple methods were tested on the task of designing a binder targeting Nipah virus. Of the thousand-plus designs tested, 99 proteins were confirmed to bind, and none neutralized the targeted protein. The specialized method Mosaic, which combines multiple tools with expert tuning, outperformed general-purpose approaches (Figure 6.1.10).

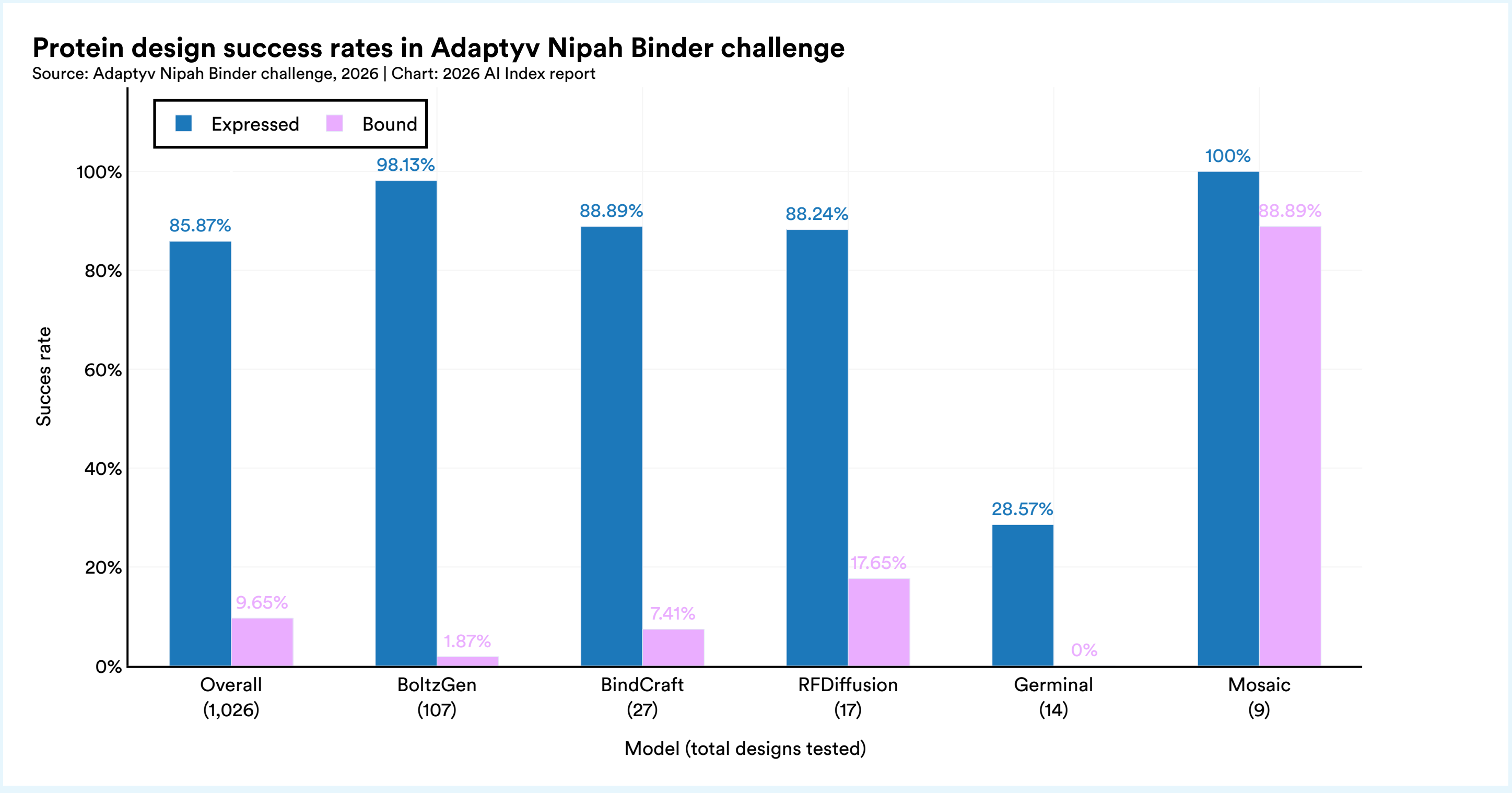

Figure 6.1.10

## Virtual Cell Models and Genomic Foundation Models

Research on "virtual cell" models—AI systems that model cellular states and responses to stimuli—increased substantially in 2025, as reflected in growing PubMed publication counts (Figure 6.1.11). Notable releases included Evo 2, a DNA language model from the Arc Institute; STATE, a perturbation-response model; and AlphaGenome, a multimodal model from DeepMind.

However, current virtual cell and genomic foundation models still lag behind smaller, task-specific models on several benchmarks. GPN-Star, a 200-million-parameter model focused on functional and regulatory genomics, outperformed Evo 2 (40B parameters) on multiple variant effect prediction tasks (Figure 6.1.12). These results suggest that scale alone is not yet sufficient, and that training method and data curation remain important determinants of performance.

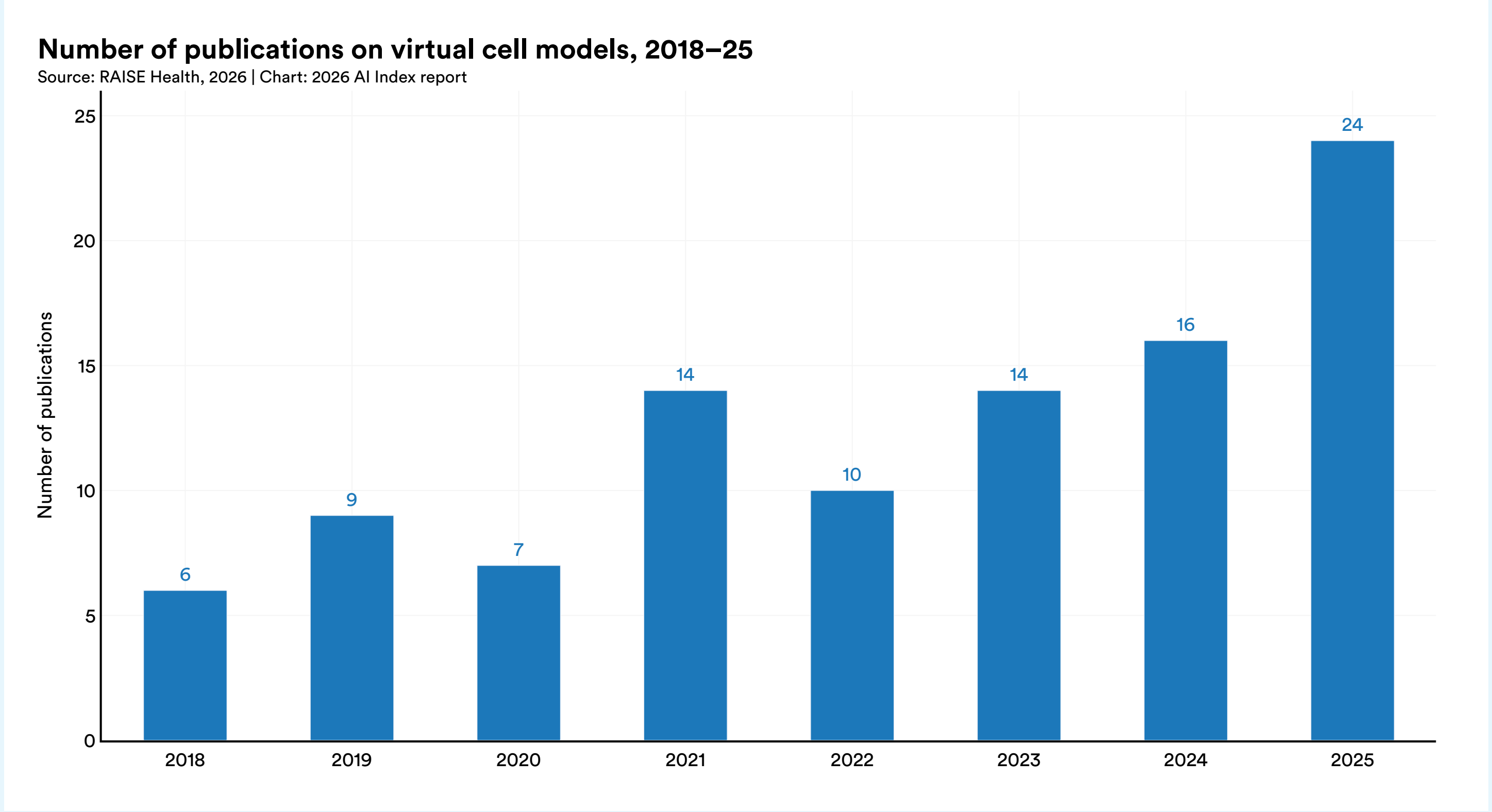

Figure 6.1.11

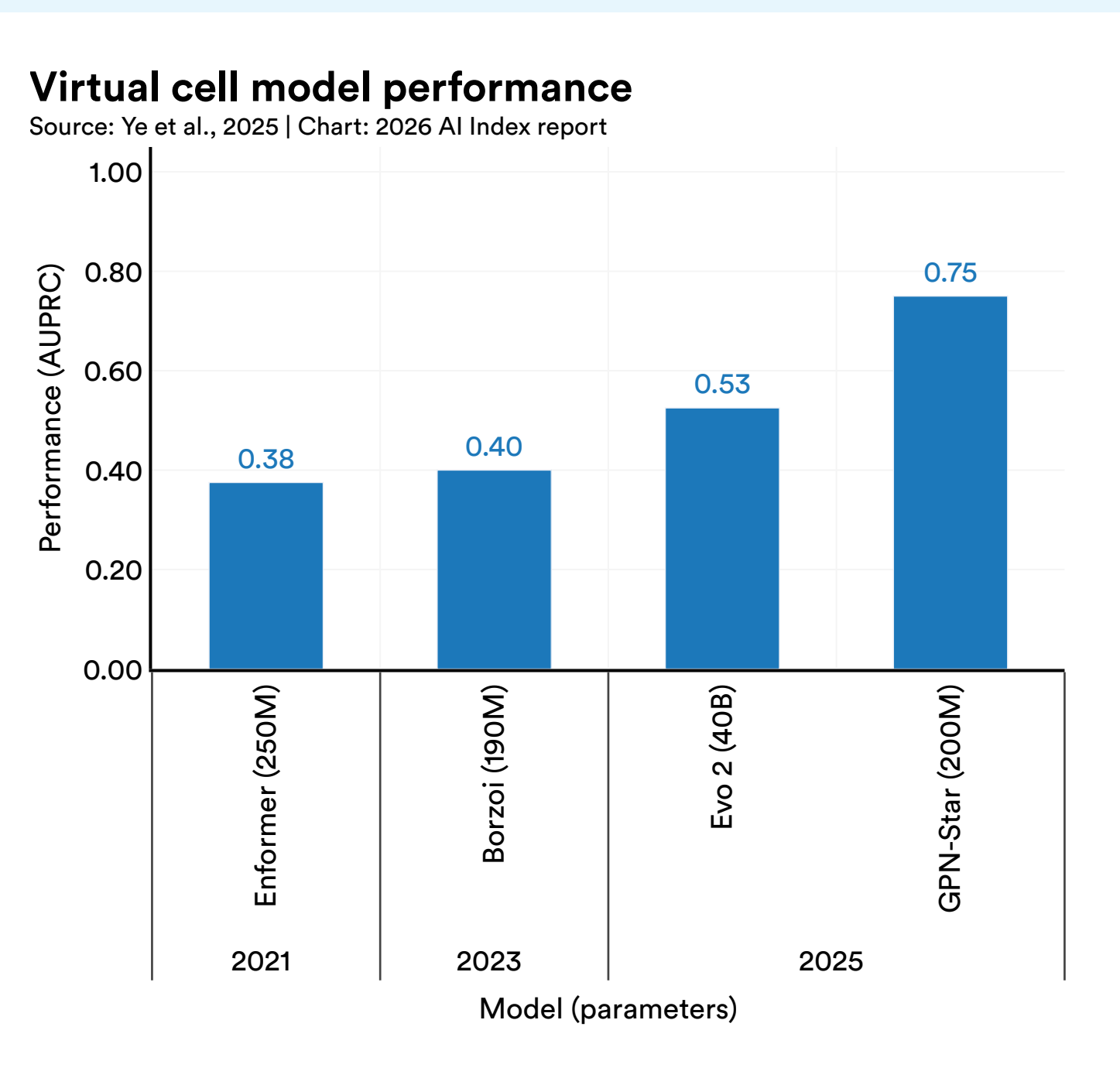

Figure 6.1.12

## Multimodal Foundation Models for Biomedical Discovery

Scientific publications on multimodal foundation models for biomedical discovery have been growing rapidly since 2021 (Figure 6.1.13). While several subfields within multimodal biomedical AI gained traction in 2025, two areas have been especially impactful: Vision-language models pair medical or biological images with text, while vision-omics models integrate imaging with genomic or transcriptomic data.

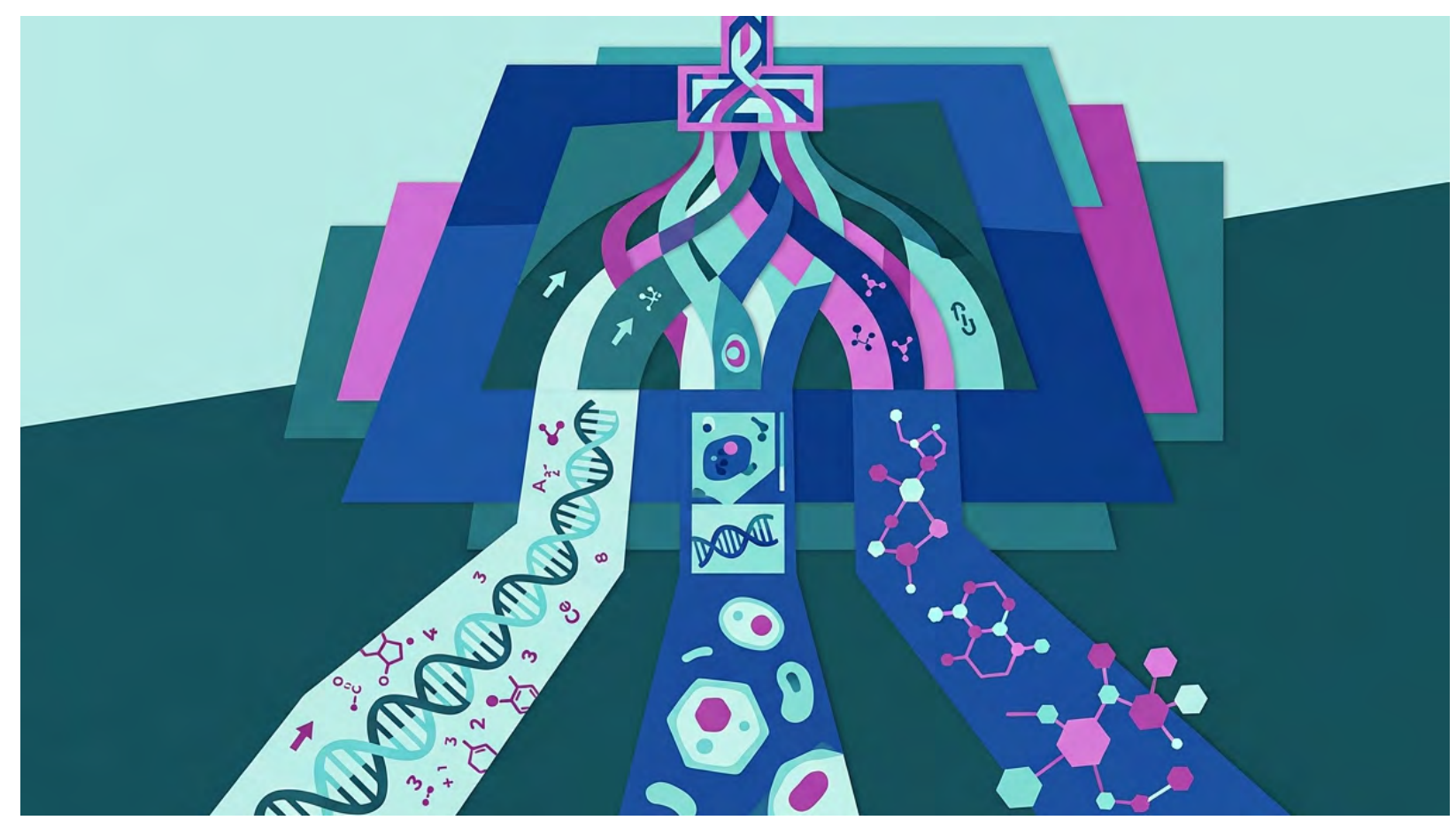

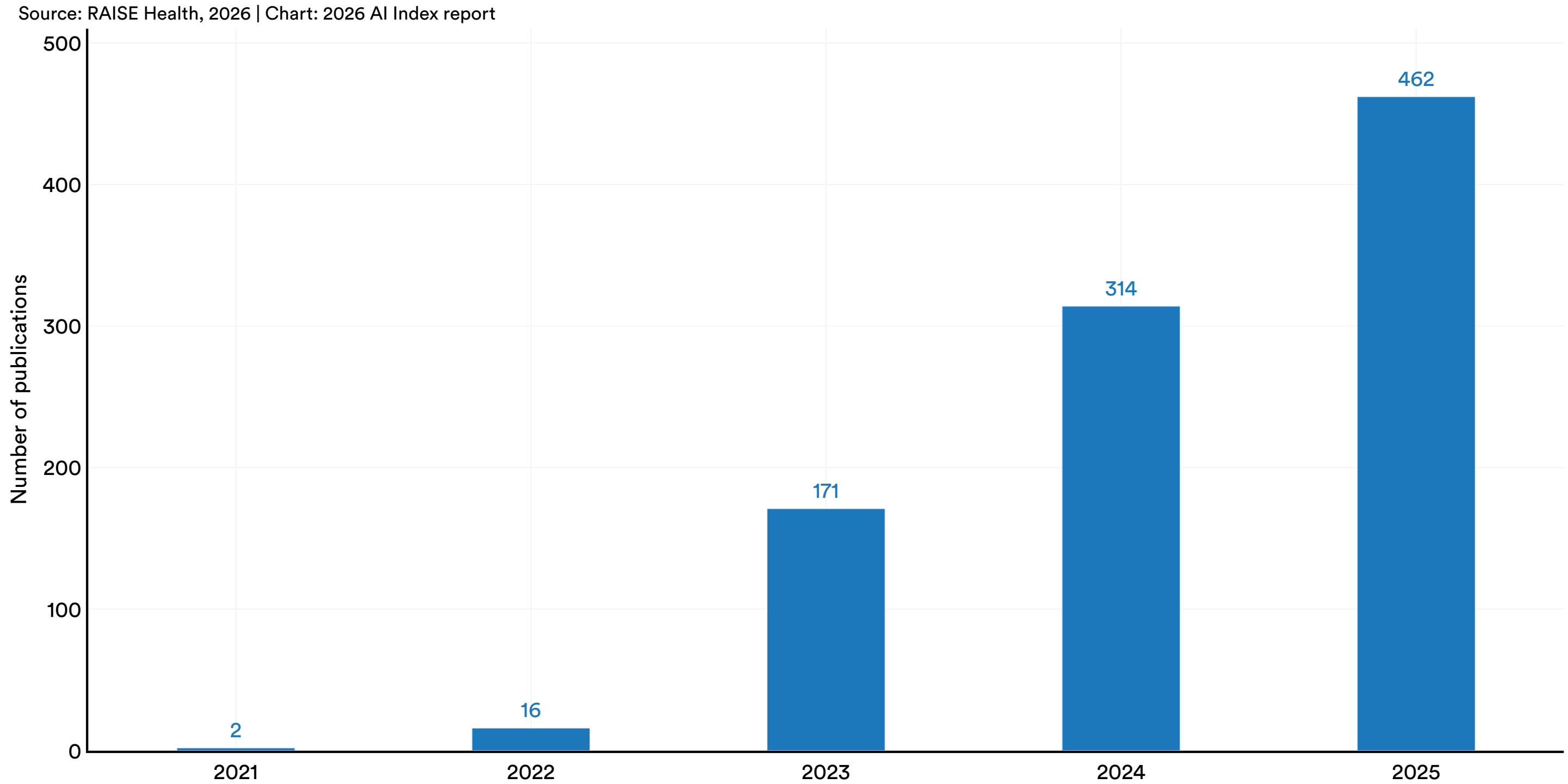

Number of publications on multimodal biomedical AI, 2021–25
Source: RAISE Health, 2026 | Chart: 2026 AI Index report

Figure 6.1.13

**HIGHLIGHT:**

## Automated and Agentic Biomedical Discovery

In 2025, efforts to automate scientific discovery focused on integrating digital reasoning with physical laboratory validation. Robin, an automated discovery framework, linked literature-based hypothesis generation with experimental data analysis, identifying the ROCK inhibitor ripasudil as a novel candidate for dry age-related macular degeneration. STELLA, an autonomous bioinformatics agent, expanded its own technical capabilities by discovering and integrating new software tools rather than relying on manually curated toolsets. Biomni, a general-purpose biomedical AI agent developed at Stanford University, mapped a unified biomedical action space across 25 subfields, integrating 150 specialized tools, 105 software packages, and 59 databases.

Collaborative multiagent frameworks also emerged. Agent Laboratory, developed by AMD and Johns Hopkins, assigns distinct roles to PhD, Postdoc, and ML Engineer agents within a simulated lab structure. The Virtual Lab uses an LLM Principal Investigator to orchestrate specialized scientist agents, producing 92 novel nanobody binder designs for SARS-CoV-2. These systems are part of an early-stage trend toward multiagent coordination in biomedical research, though their outputs still require experimental validation.

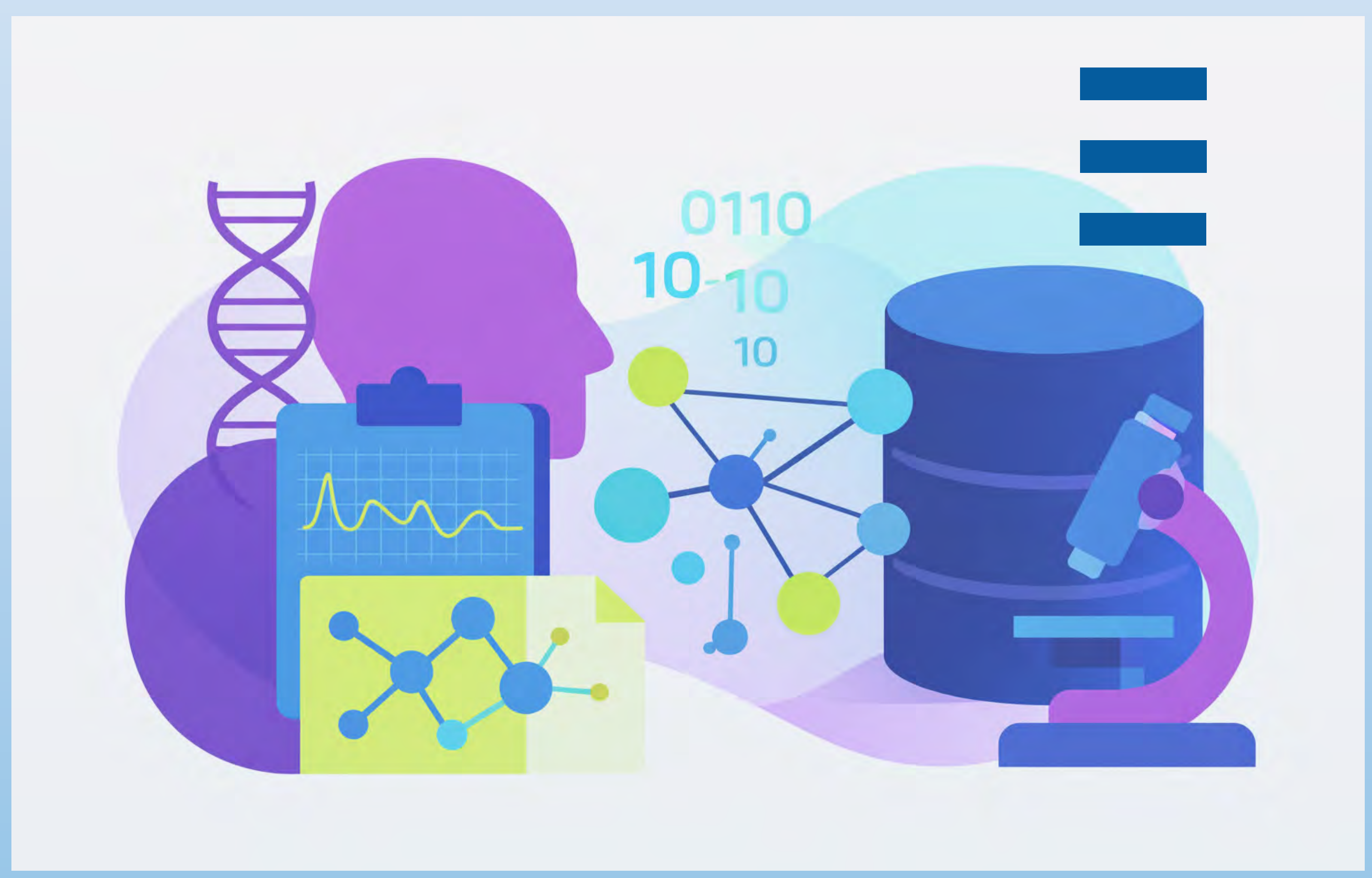

# 6.2 Clinical Applications

The molecular and cellular AI advances described in section 6.1 provide the upstream models on which clinical tools increasingly depend. This section tracks how AI is being applied in clinical settings, from medical imaging and diagnostic reasoning to workflow integration, regulatory authorization, and enterprise-scale deployment. The analysis draws on prospective trial counts, FDA device authorization data, benchmark evaluations, and published apportionment outcomes from health systems. Across these areas, strong benchmark results have yet to translate reliably to measurable clinical outcomes.

## Imaging

### Data Scale and Availability

Training data for medical imaging AI remains roughly 100 times smaller in raw sample count than for nonmedical AI (Figure 6.2.1). MAIRA-2, a radiology-focused model trained on approximately 1.4 million chest radiographs, compared with DINOv3, a general-purpose vision transformer trained on 1.7 billion unlabeled natural images. On the multimodal side, RadFM trained on approximately 16 million mixed 2D and 3D medical scans paired with clinical text, while OpenCLIP trained on LAION-5B, comprising approximately 5.85 billion image–text pairs. Data scarcity is especially pronounced for three-dimensional modalities such as CT and MRI, and fragmentation across institutions further limits the development of large-scale medical foundation models.

**Training data volume in medical and nonmedical AI: imaging-only vs. multimodal models**

Source: RAISE Health, 2026 | Chart: 2026 AI Index report

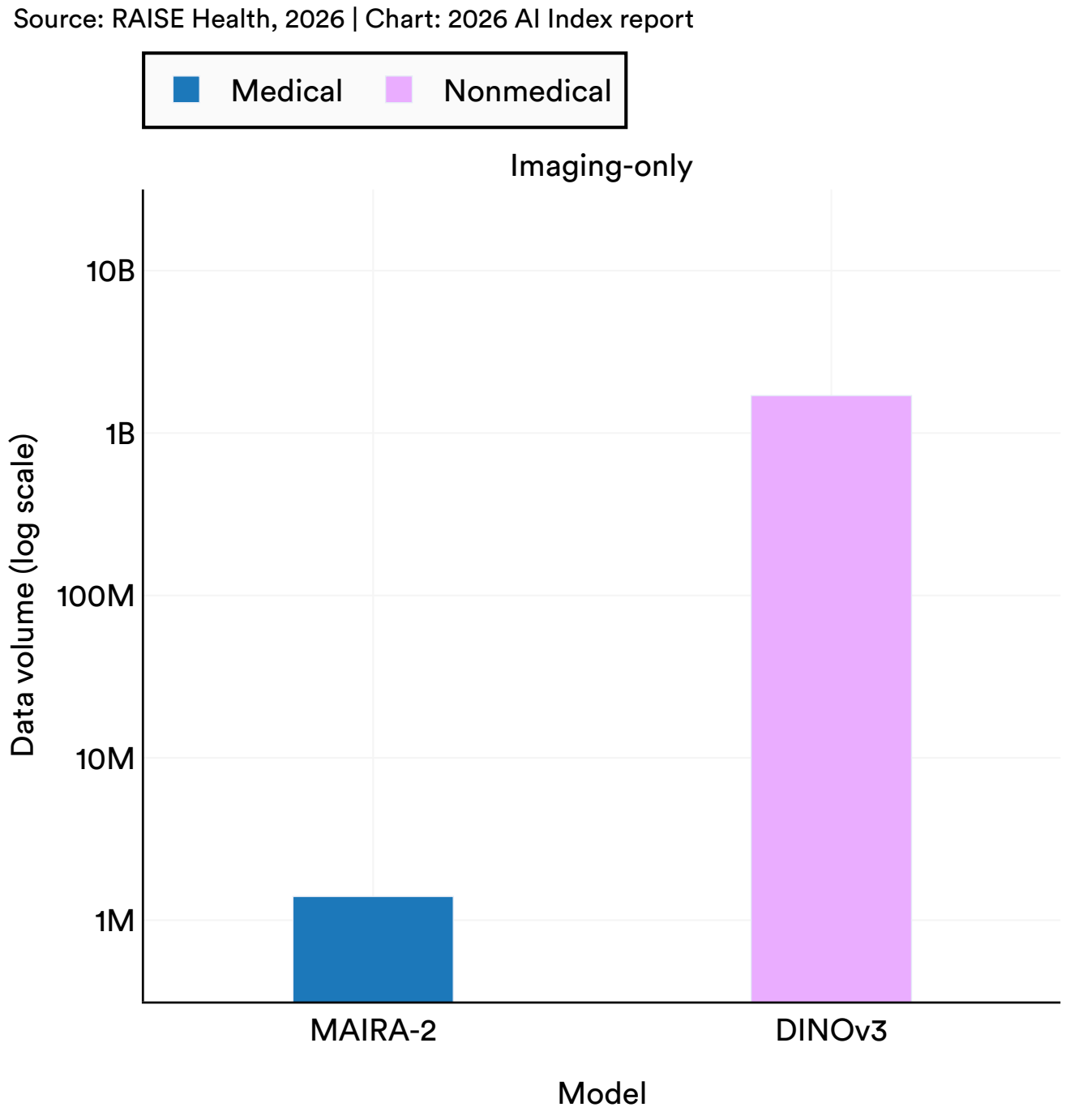

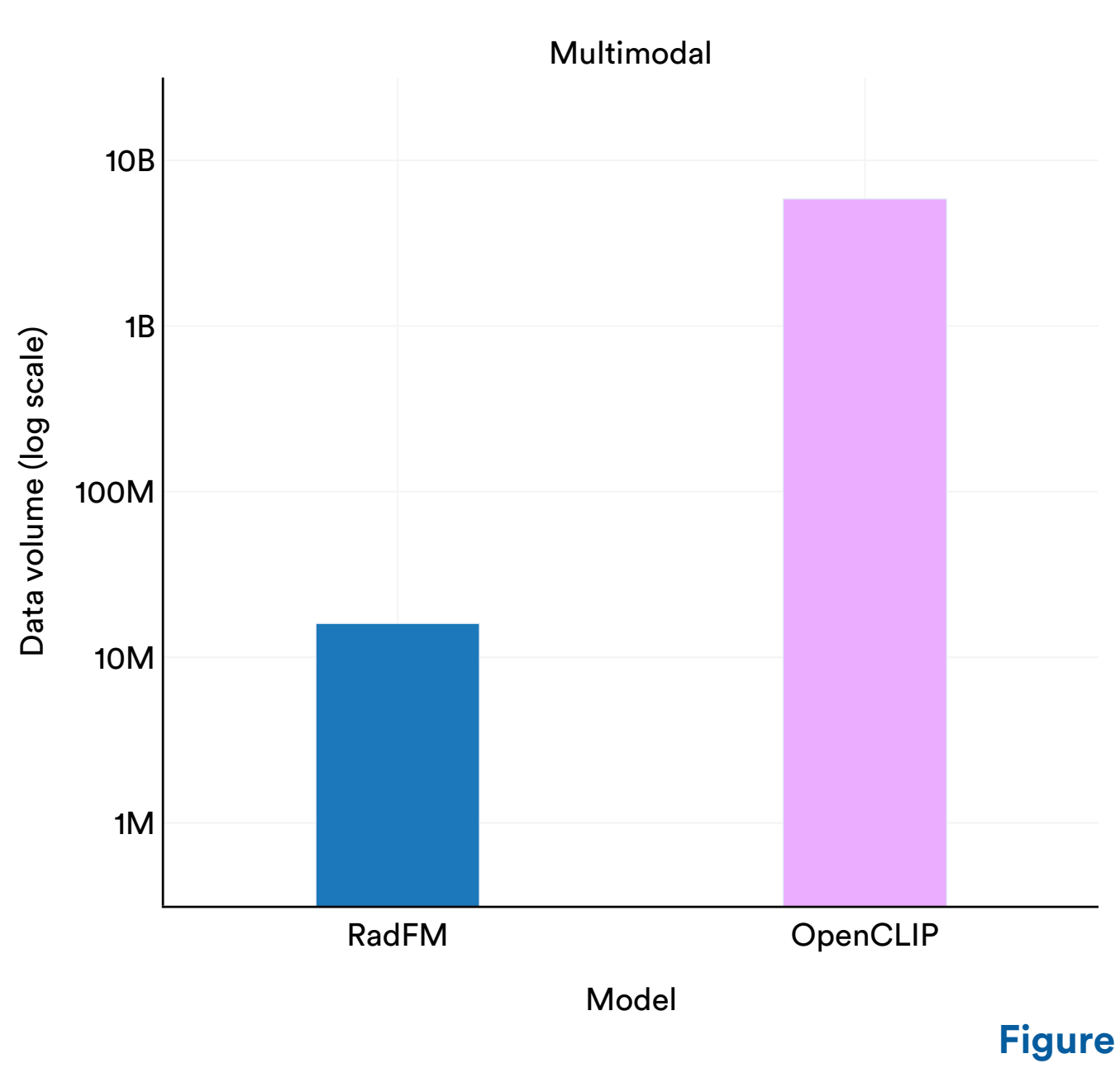

Figure 6.2.1

## Modeling Approaches

Vision language models (VLMs) for medical imaging have expanded beyond radiology into pathology, dermatology, ophthalmology, and cardiology (Figure 6.2.2). Across six clinical disciplines, the number of research models and FDA-cleared commercial products grew, with pathology seeing the greatest concentration of new research releases. The Merlin model demonstrated that a highly capable CT foundation model could be trained on a single 40GB GPU by leveraging both radiology reports and ICD codes during training, making advanced medical AI accessible even in resource-constrained settings.

Medical imaging AI lacks the standardized cross-model benchmarks common in general-domain machine learning. Models in different specialties are typically evaluated on different datasets, making direct performance comparisons across disciplines difficult. Recent MICCAI 2025 challenges (CHIMERA, UNICORN) represent early efforts to address this gap. Human-centered evaluation, where clinicians manually review model outputs, has become more prevalent in publications and provides stronger evidence of clinical utility than lexical metrics.

| Discipline | Notable releases | Analogous FDA-cleared models |
|---|---|---|
| Cardiology | EchoJEPA (2026)<br>PanEcho (2025)<br>EchoFM (2025)<br>EchoPrime (2025) | Bunkerhill ECG-EF<br>Heartflow Plaque Analysis |
| Oncology | MUSK (2025) | Allix5, Clairity (2025)<br>Transpara (2.1.0) (2024) |
| Ophthalmology | EyeCLIP (2025)<br>Meta-EyeFM (2025)<br>RETFound-Green (2025) | CLARUS (700), Carl Zeiss (2025) |
| Pathology | Virchow2G (2026)<br>KRONOS (2025)<br>VORTEX (2025)<br>Threads(2025)<br>mSTAR (2025)<br>PRISM2 (2025)<br>MPath (2025)<br>H-Optimus-0 (2024) | ArteraAI Prostate (2025)<br>Ibex Prostate Detect (2025) |
| Radiology | MedGemma 1.5 (2026)<br>COLIPRI (2026)<br>TTE 3D CT (2025)<br>3DINO-ViT (2025)<br>CT-FM (2025)<br>RadFM (2025)<br>Merlin (2025) | BriefCase Triage: CARE Multi-triage CT Body (2026)<br>a2z-Unified-Triage (2025)<br>Bunkerhill BMD (2025)<br>Bunkerhill AAQ (2025)<br>Brainomix 360 Triage Stroke (2025)<br>Ezra Flash (2025) |

Figure 6.2.2

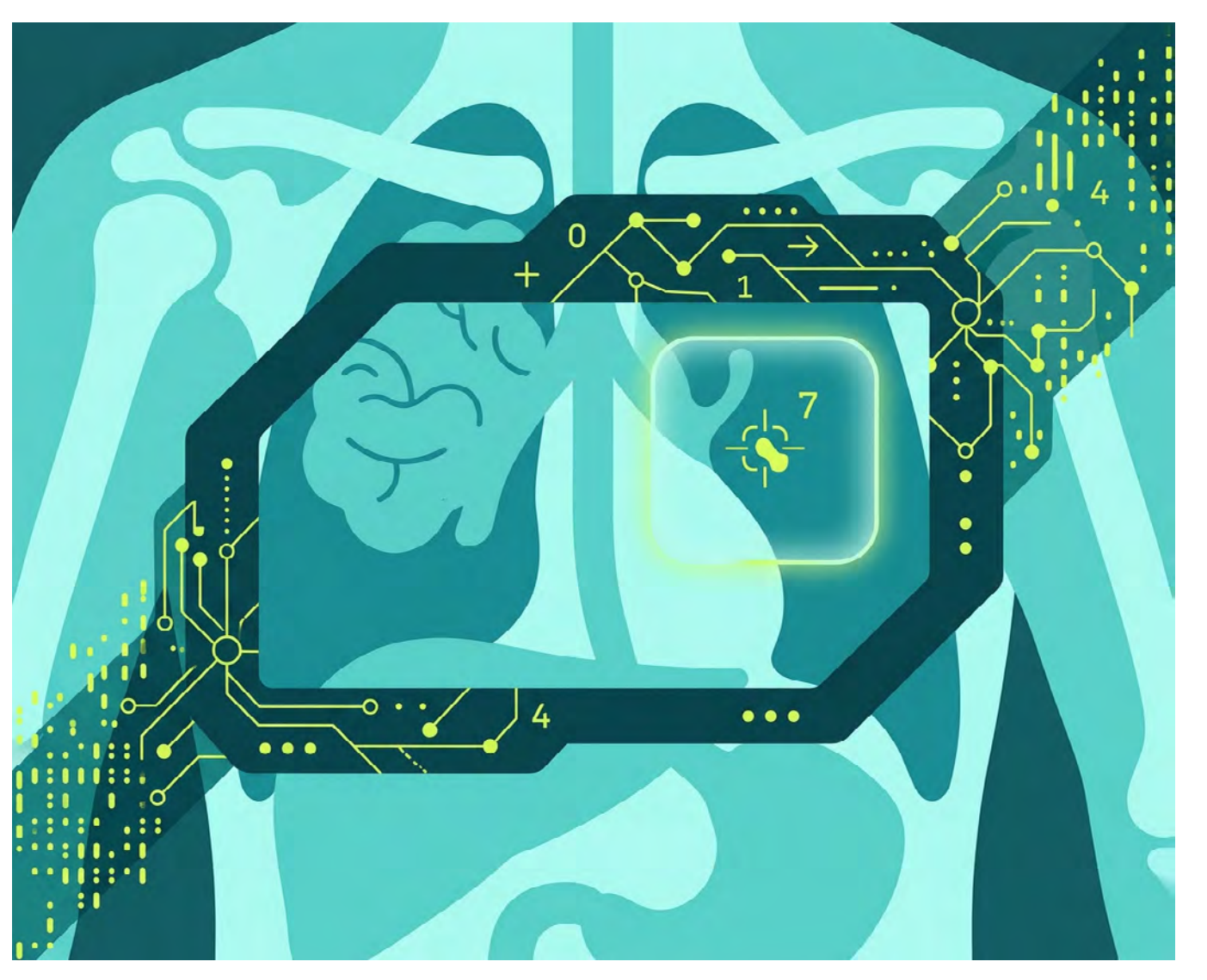

## Prospective Clinical Trials

The number of prospective trials validating medical imaging AI models grew by 28.5% year over year, from 417 in 2024 to 536 in 2025 (Figure 6.2.3). This growth signals the field is moving beyond studies that evaluate AI on past patient data toward live clinical trials, the kind of evidence required before hospitals will adopt these tools. Recent trials include MASAI, a randomized screening accuracy study of AI-assisted mammography, and NOTIFY-1 and NOTIFY-EXTEND, which tested whether flagging early signs of heart disease that AI spotted on routine CT scans led doctors to prescribe more preventive cholesterol medication.

**Number of papers reporting prospective trials of clinical imaging ML/AI models, 2010–25**
Source: RAISE Health, 2026 | Chart: 2026 AI Index report

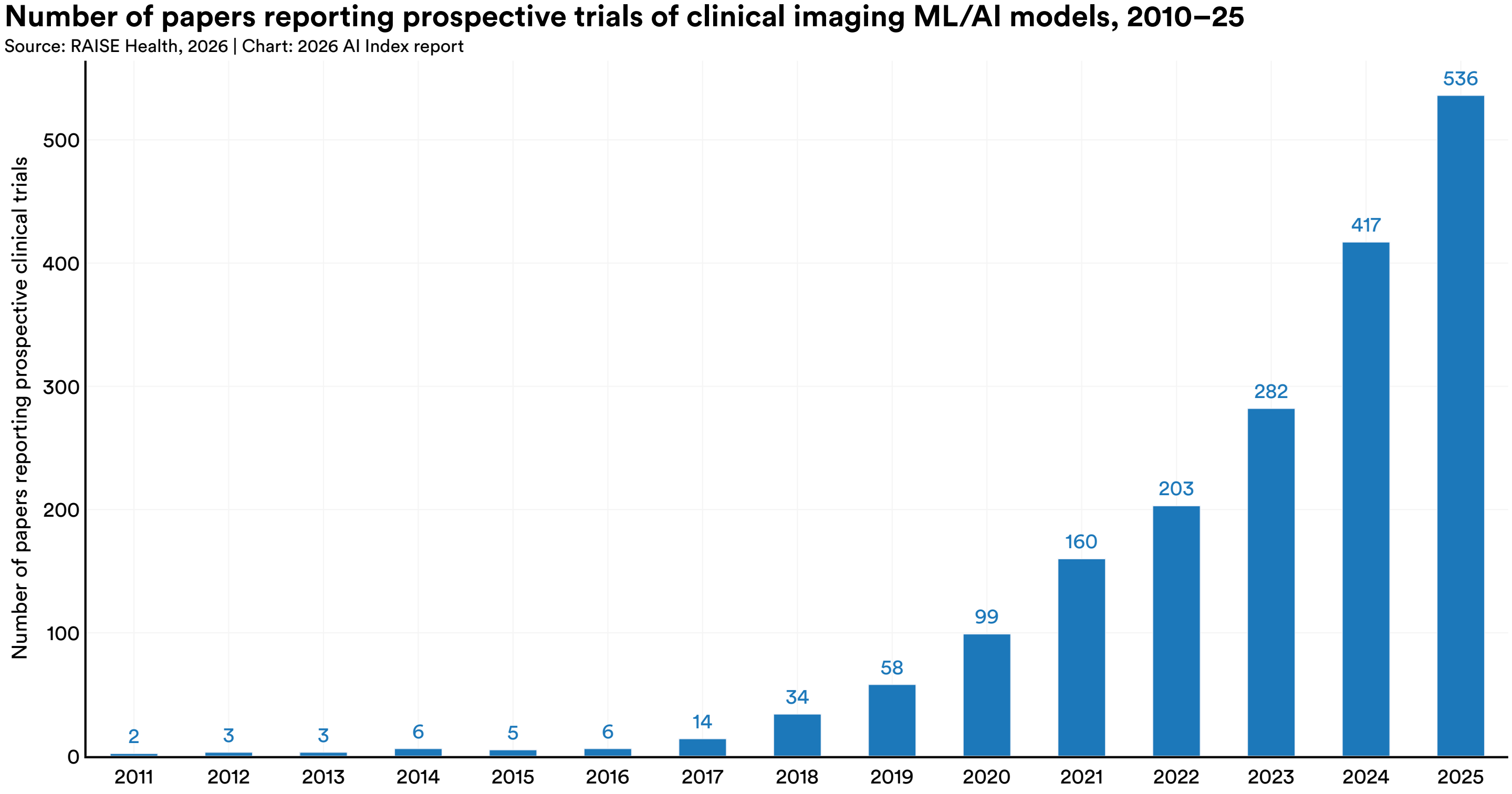

Figure 6.2.3

HIGHLIGHT:

## LLM Clinical Reasoning Performance

In a multi-experiment evaluation, OpenAI's o1-preview reasoning model was tested on diagnostic reasoning tasks, management reasoning vignettes, probabilistic reasoning scenarios, and real emergency department (ED) cases with blinded expert scoring (Brodeur et al., 2025). On New England Journal of Medicine (NEJM) clinicopathological conferences (n=143), the model included the correct diagnosis in its differential 78% of the time, with 52% top-1 accuracy. On NEJM Healer cases (80 responses), it achieved a perfect revised-IDEA score in 78 of 80, compared with 47 of 80 for GPT-4, 28 of 80 for attending physicians, and 16 of 80 for residents. On management reasoning, o1 preview's median score was 86%, versus 42% for GPT-4 only, 41% for physicians with access to GPT-4, and 34% for physicians with conventional resources (Figure 6.2.4). In 76 real ED cases, o1 produced diagnoses rated "exact/very close" in 67%–83% of cases across three diagnostic stages, surpassing two attending physicians at each stage.

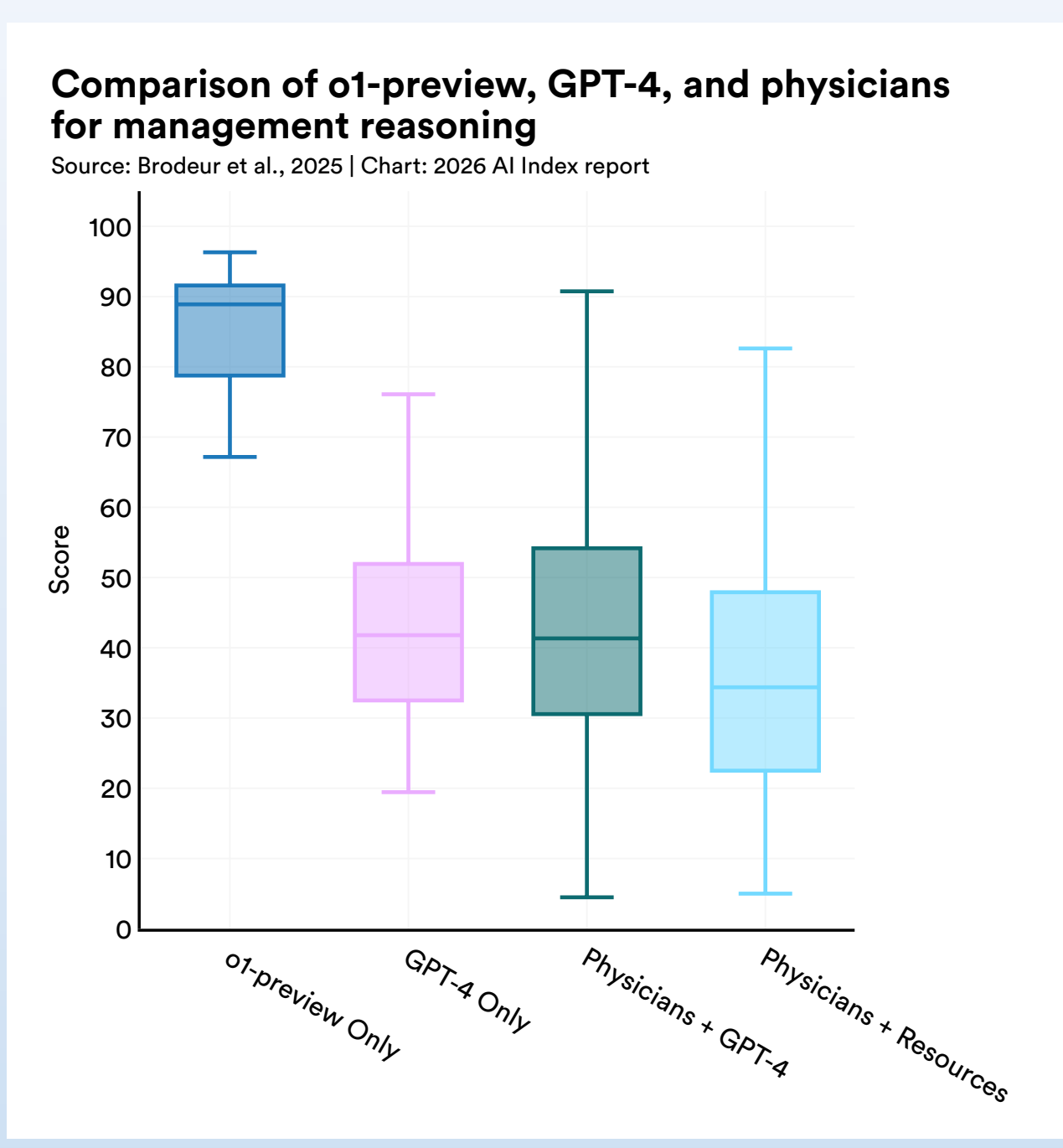

Figure 6.2.4[2]

These results suggest that current LLMs have surpassed most existing clinical reasoning benchmarks, but they reflect isolated cognitive evaluations rather than real-world clinical integration. Whether AI-assisted reasoning translates to improved patient outcomes remains an open question requiring prospective trials.

HIGHLIGHT:

## AI Agents in Clinical Medicine

Autonomous and semiautonomous AI agents have emerged as a major development in clinical AI in 2025–26. Unlike conventional AI models that generate predictions or classifications in isolation, these systems reason across multiple steps, access external tools and data sources, and coordinate with other AI agents or human clinicians to complete complex clinical tasks.

Multiagent frameworks, in which multiple AI agents take on specialized roles—such as diagnostician or pharmacist—and collaborate through structured reasoning protocols, have shown early promise on benchmark evaluations. Diagnostic accuracy gains over single-agent baselines ranged from 7% to over 60%, depending on the complexity of the clinical task (Gorenshtein et al., 2025; Zheng et al., 2025; Liu et al., 2025). Microsoft's AI Diagnostic Orchestrator (MAI-DxO), paired with OpenAI's o3 reasoning model,

2 Box plot of normalized management reasoning points by LLMs and physicians on Gray Matters management cases. Five cases were included. Three o1-preview responses were generated for each case. The prior study collected five GPT-4 responses to each case, 176 responses from physicians with access to GPT-4, and 199 responses from physicians with access to conventional resources.

**HIGHLIGHT:**

achieved 85.5% accuracy on diagnostically challenging cases from the New England Journal of Medicine, compared with approximately 20% among 21 practicing physicians with five to 20 years of clinical experience working under comparable conditions.

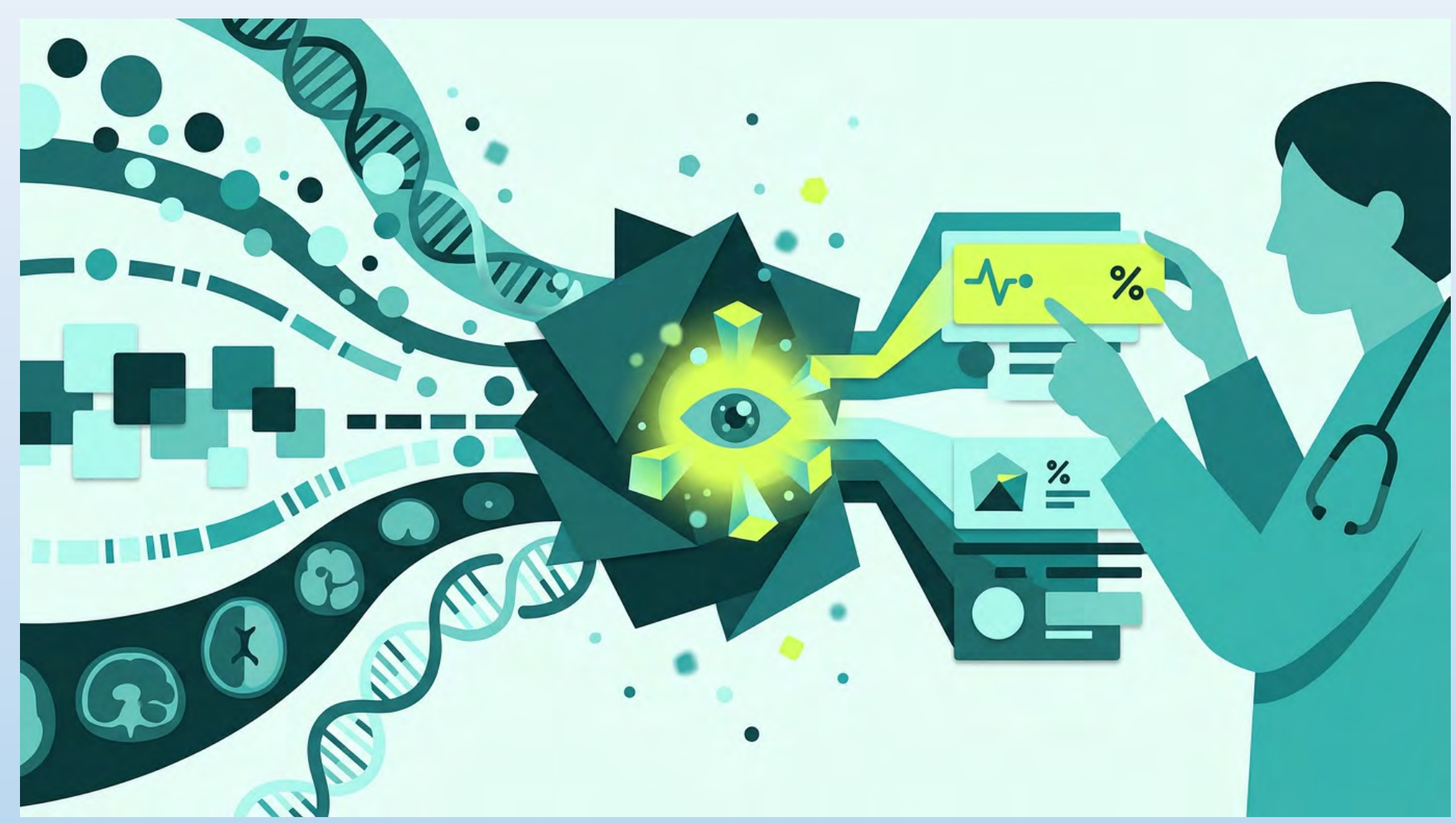

A new set of benchmarks specifically designed to evaluate these agentic systems has started to appear. A 2025 scoping review identified 43 studies evaluating agentic AI in healthcare, 36 of which (84%) were published in 2025. On MedAgentBench (Jiang et al., 2025), which evaluates LLM agents in a virtual electronic health record (EHR) environment across 300 clinically derived tasks, the best performing model achieved a task success rate of 69.7%. These results suggest that, despite access to advanced capabilities such as tool use and iterative reasoning, the evidence base for reliable autonomous clinical AI agents remains early-stage.

## Deployment, Implementation, and Deimplementation

This section tracks the regulatory, institutional, and evidentiary dimensions of clinical AI deployment, from FDA device authorizations to enterprise-scale outcomes.

### FDA-Authorized AI/ML-Enabled Devices

In the United States, FDA 510(k) is the most common regulatory pathway for AI medical devices, requiring manufacturers to demonstrate that a new device is substantially equivalent to one already on the market rather than conducting new clinical trials. The number of 510(k)-cleared AI/ML-related devices reached 246 in 2025, continuing a steep upward trajectory that began with 16 devices in 2016 (Figure 6.2.5).

Because most cleared radiology AI solutions are offered commercially rather than as open-source tools, healthcare systems are typically required to complete financial clearance and cost-effectiveness justification before implementation. Comp2Comp, a notable exception, is an open-source python package for CT imaging analysis with two modules (bone mineral density and abdominal aortic quantification) that secured FDA clearance in 2025.

## FDA 510(k)-cleared AI/ML-enabled imaging-related medical devices, 2011–25

Source: FDA, 2025 | Chart: 2026 AI Index report

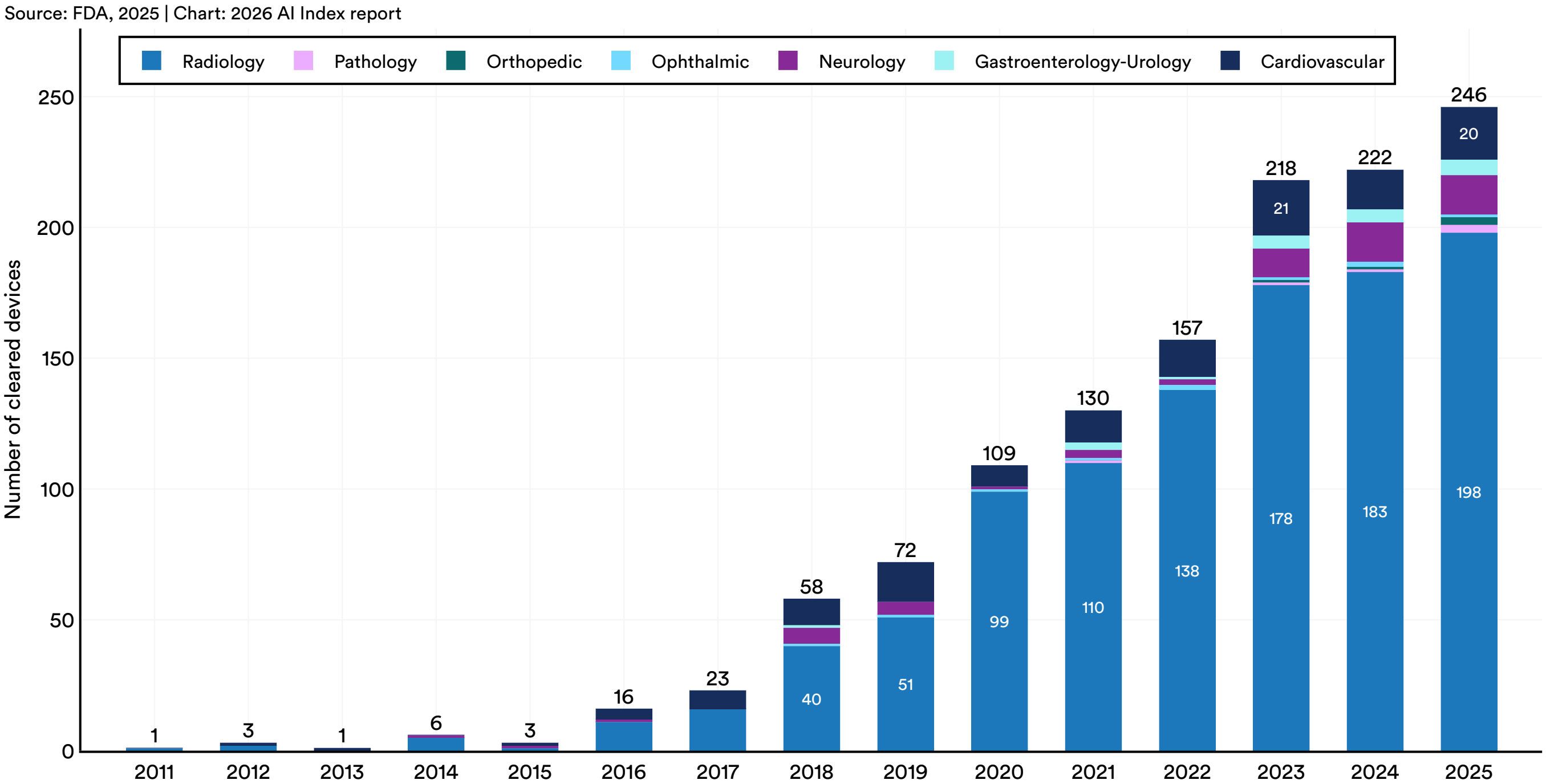

Figure 6.2.5

By December 2025, the FDA had authorized a total of 1,357 AI/ML-enabled medical devices from 693 different companies across 17 clinical specialties (Figure 6.2.6). Annual authorizations reached 258 through September 2025, already surpassing all prior full-year totals. The cumulative total crossed the 1,000-device milestone in 2024. Ninety-eight new companies entered the space in 2025, continuing a trend of broadening market participation (103 new entrants in 2023, 109 in 2024).

## Number of AI medical devices approved by the FDA, 1995–2025

Source: FDA, 2025 | Chart: 2026 AI Index report

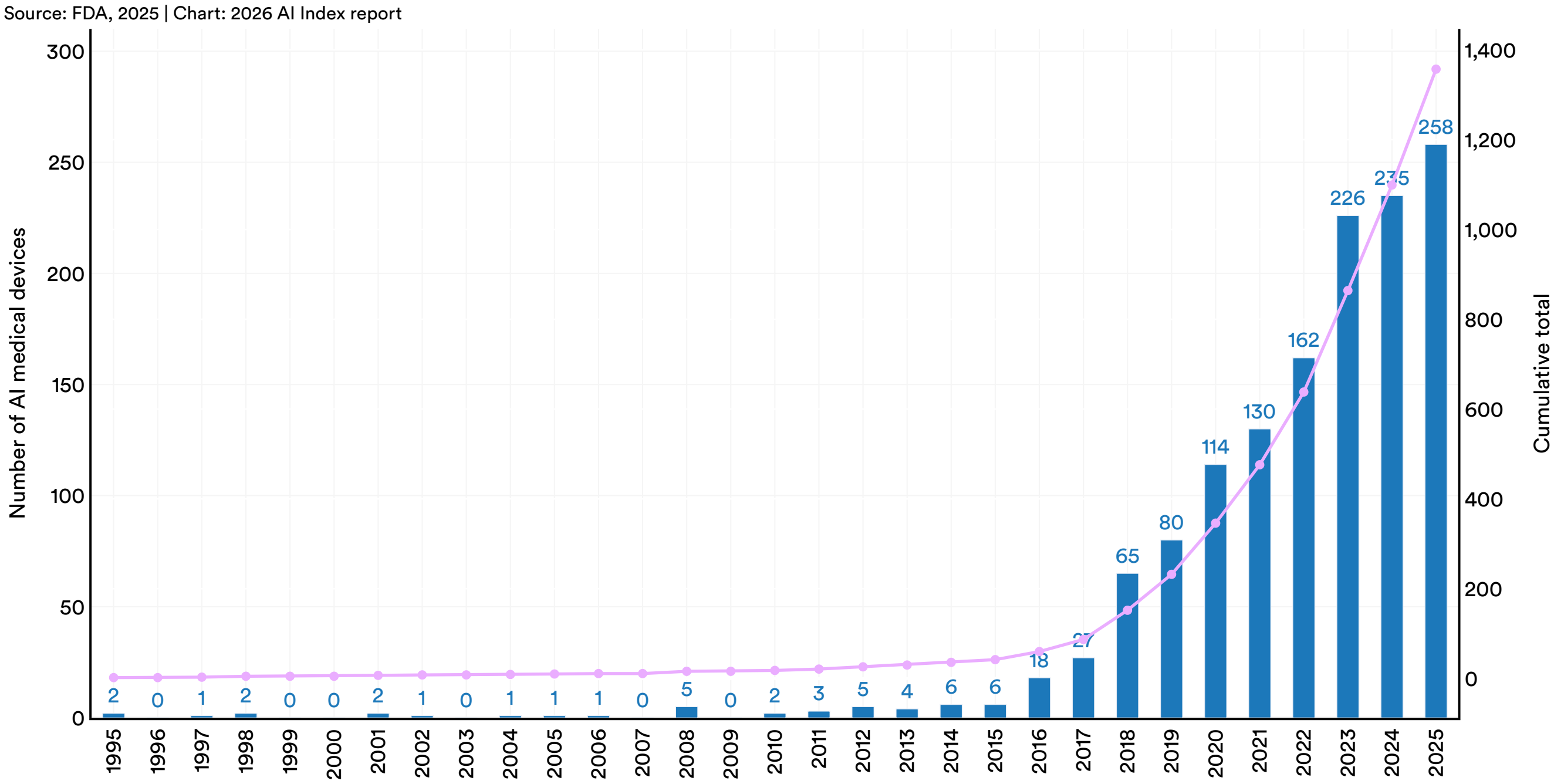

Figure 6.2.6

## Devices by Clinical Specialty

Radiology accounts for the largest share of authorized AI/ML devices at 1,039 of 1,357 (76.6%), followed by cardiovascular (130 devices, 9.6%) and neurology (61 devices, 4.5%) (Figure 6.2.7). Non-radiology authorizations have increased from 7 in 2016 to 60 in 2025 (Figure 6.2.8). Cardiology, neurology, anesthesiology, and gastroenterology-urology have all seen acceleration since 2020, suggesting that AI is beginning to spread from imaging-centric applications to broader clinical domains.

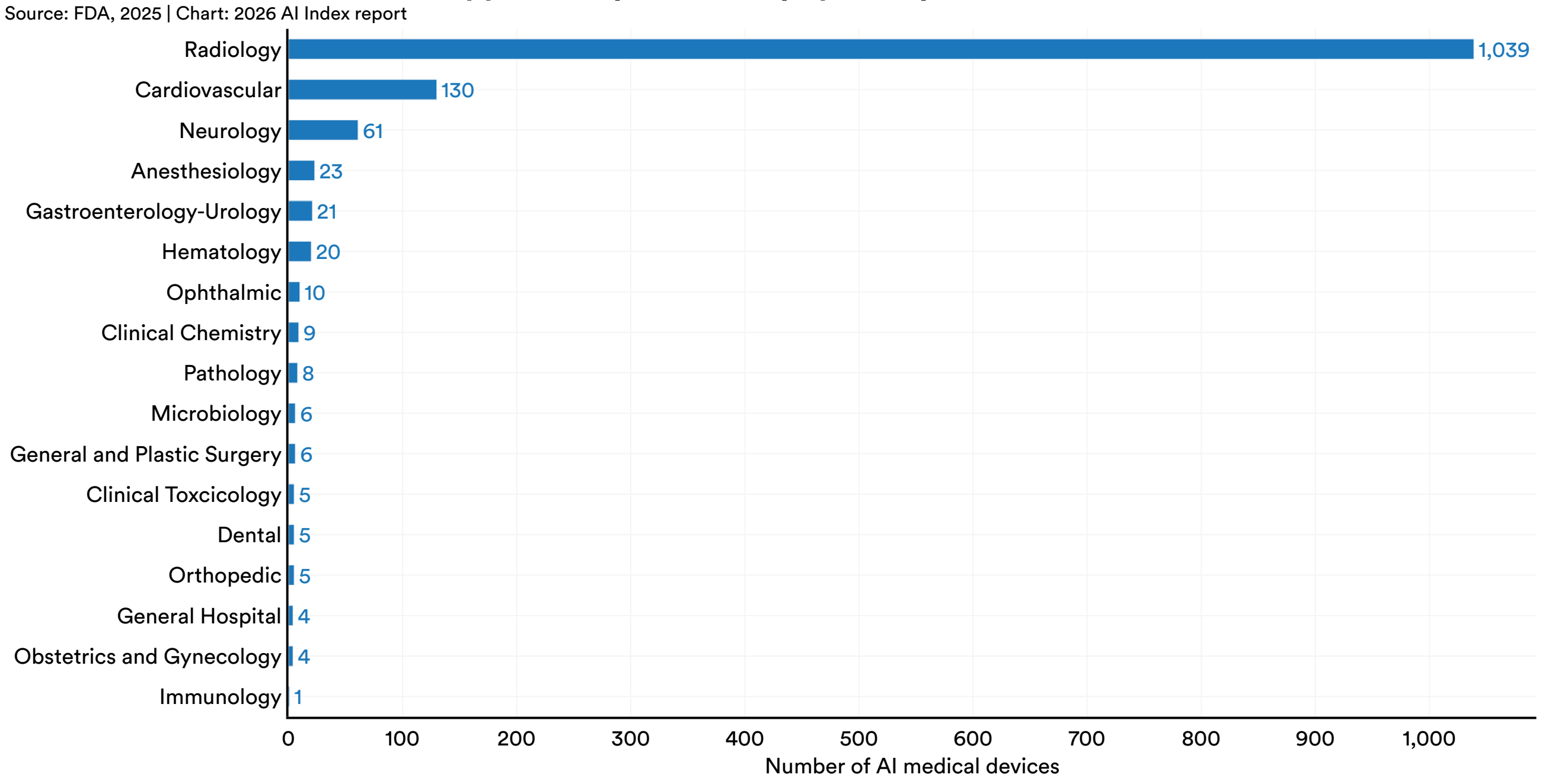
Number of AI medical devices approved by the FDA by specialty, 1995–2025 (sum)
Source: FDA, 2025 | Chart: 2026 AI Index report

Figure 6.2.7

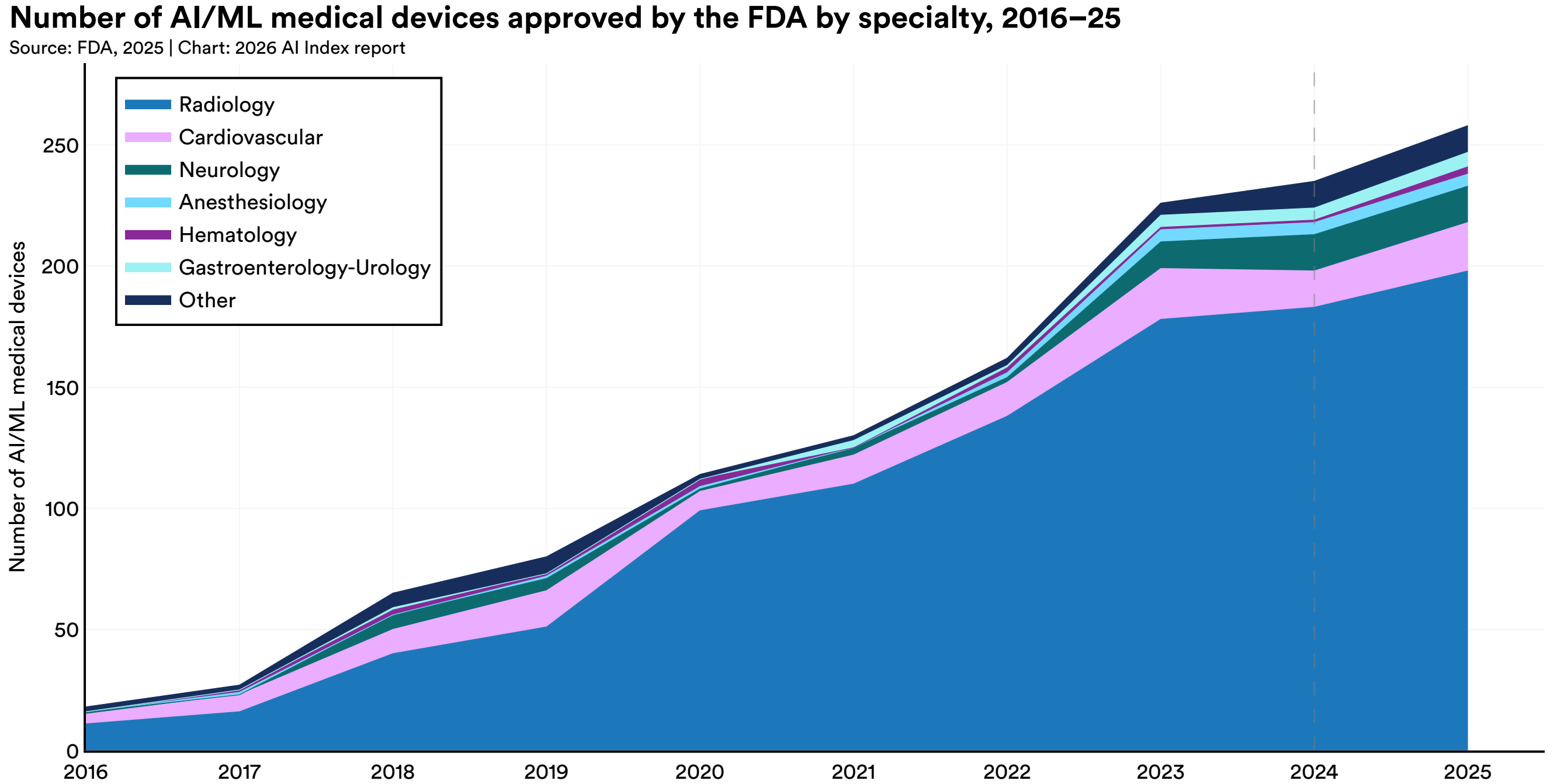
Number of AI/ML medical devices approved by the FDA by specialty, 2016–25
Source: FDA, 2025 | Chart: 2026 AI Index report

Figure 6.2.8

### Industry Landscape

FDA clearance does not equal clinical adoption. Financial, operational, and institutional barriers often stand between regulatory authorization and real-world deployment. The authorized device market is concentrated at the top but fragmented overall. GE Healthcare leads with 93 devices, followed by Siemens Healthineers (82), Shanghai United Imaging Healthcare (38), Philips Healthcare (36), and Canon Medical Systems (35), and Aidoc Medical (30) (Figure 6.2.9). Of the 626 companies with at least one authorized device, the large majority hold only one or two, reflecting a broad ecosystem of specialized entrants alongside established manufacturers.

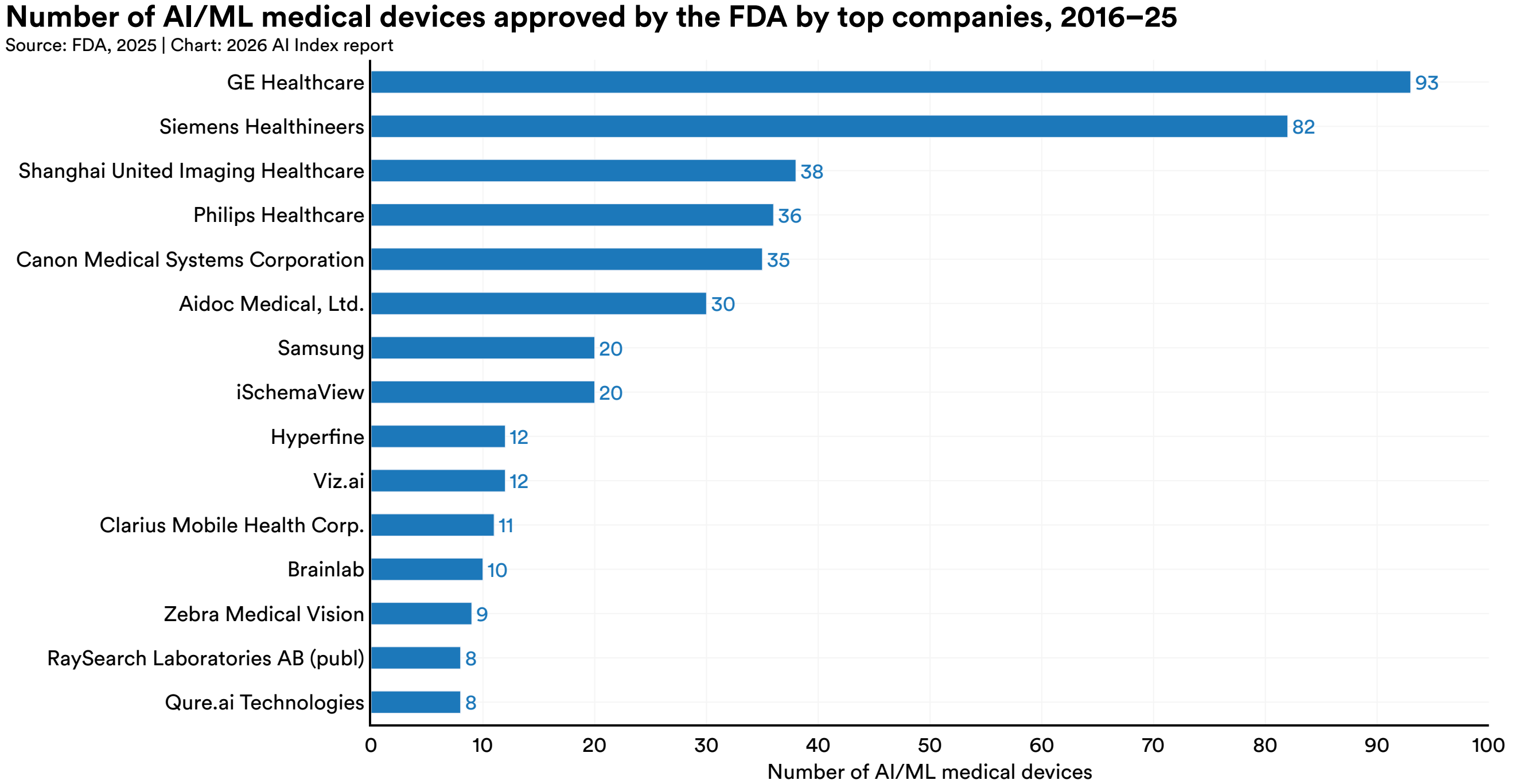
Number of AI/ML medical devices approved by the FDA by top companies, 2016–25
Source: FDA, 2025 | Chart: 2026 AI Index report

Figure 6.2.9

In January 2025, the FDA issued draft guidance on AI-enabled device software functions applying a Total Product Life Cycle approach. Predetermined Change Control Plans , a mechanism that permits iterative updates after initial market authorization, were used in approximately 10% of 2025 clearances. Despite this growth, a peer-reviewed analysis of all 1,016 authorizations through December 2024 (Singh et al., 2025) found that only 2.4% of devices with clinical studies were supported by randomized controlled trial data, with nearly all devices entering via the 510(k) pathway.

## Enterprise-Scale Deployments in 2025

Clinical AI moved from pilot-stage initiatives to enterprise-scale deployments in 2025, with health systems reporting measurable outcomes across clinical and operational domains. The most published evidence was from ambient AI documentation, AI-powered sepsis prediction, and generative AI integration into clinical workflows.

### Ambient AI Documentation

Ambient AI scribes, tools that automatically generate clinical documentation from patient–clinician conversations, saw the broadest adoption of any clinical AI category in 2025. Abridge, one of the leading platforms, expanded from approximately 100 to over 150 health systems, including Kaiser Permanente's deployment across 40 hospitals and more than 600 medical offices. Adoption reached 63% among hospitals using Epic's electronic health record system.

Outcomes were consistent across multiple institutions. Sharp HealthCare reported an 83% reduction in note-writing effort and a 3.5%–6% increase in work relative value units—a standard measure of physician clinical productivity—per encounter. The University of Chicago Medicine reported a 47% reduction in cognitive load and a 58% increase in undivided patient attention. MaineHealth reported a 23% reduction in time spent on clinical notes, with the tool used in 70.3% of encounters. At Northwestern Medicine, physicians using the tool in more than half of encounters saw 11.3 additional patients per month and a 24% reduction in documentation time, with a reported 112% return on investment. At Stanford Health Care, a prospective study of 48 physicians published in JAMIA (February 2025) found statistically significant reductions in task load and burnout, with physicians reporting a median time savings of 20 minutes per half day of clinic.

### AI-Powered Sepsis Prediction

Two sepsis prediction systems reported mortality reductions in large-scale deployments in 2025. The Targeted Real-time Early Warning System, developed at Johns Hopkins and commercialized by Bayesian Health, was deployed across 13 Cleveland Clinic hospitals. Reported outcomes included an 18.7% relative reduction in sepsis mortality, a 1.85-hour reduction in median time to first antibiotic order, the correct identification of 82% of sepsis cases, an 89% clinician adoption rate, and a 10% reduction in intensive care unit utilization. COMPOSER, a deep learning model at UC San Diego Health monitoring over 150 variables per patient, reported a 17% reduction in sepsis mortality (1.9% absolute) across 6,217 admissions, a 5% increase in sepsis bundle compliance, and an estimated 50 lives saved annually.

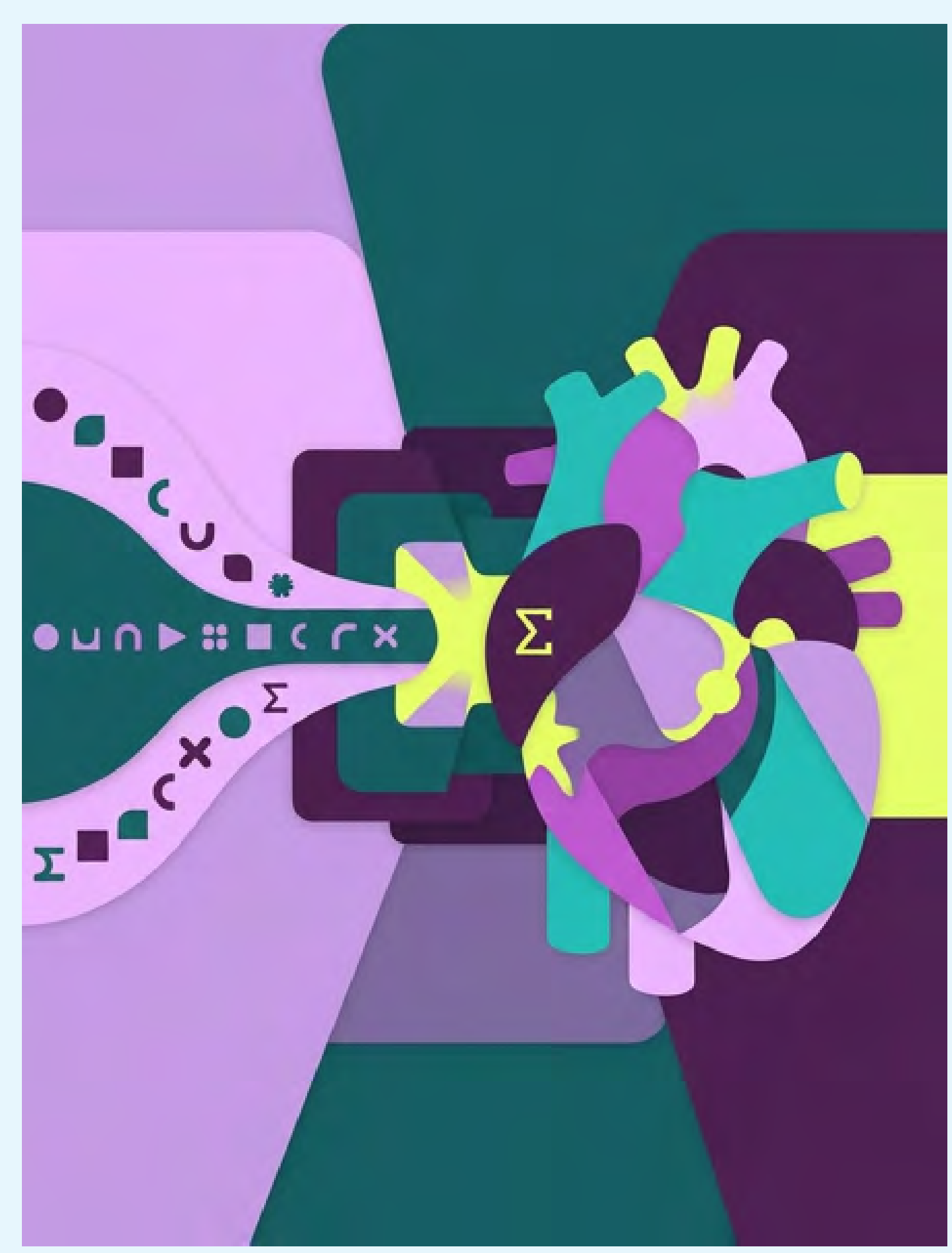

### Generative AI in Clinical Workflows

Health systems began embedding LLM-powered tools directly into electronic health records. ChatEHR, a system generating plain-language summaries of patient records, logged 23,000 sessions across 1,075 trained users within three months of broad rollout. 60% of usage occurred through automated prompts and 40% through interactive interfaces. Separately, an AI tool generating plain-language explanations of laboratory, imaging, and pathology results was evaluated in a study published in *JAMA Network Open* (August 2025). Of 93 survey respondents, 85% considered the tool user-friendly, 72% found it beneficial for laboratory results, and 63% for imaging results. OpenEvidence, a real-time evidence retrieval platform, reported adoption by 40% of U.S. physicians.

## Evidence Gaps and Governance

The inaugural State of Clinical AI Report (January 2026), published by the Stanford-Harvard ARISE Network, reviewed over 500 clinical AI studies and found that nearly half used exam-style questions rather than real patient data. Only 5% used real clinical data. The report concluded that AI performs most effectively when supporting rather than replacing clinician judgment.

Separately, the NOHARM benchmark found that leading LLMs produced 11.8 to 14.6 severely harmful recommendations per 100 clinical cases, with 76.6% being errors of omission (e.g., failing to recommend a critical test). These findings apply to general-purpose LLMs evaluated on open-ended clinical reasoning tasks, not to the narrower, task-specific tools driving current adoption. Ambient scribes and sepsis alerts, for example, operate within constrained workflows with clinician oversight.

Governance frameworks have also advanced. Stanford Health Care's FURM framework now governs all new AI tool adoptions at that institution, and the GUIDE-AI Lab is working to make the framework available to other health systems.

**HIGHLIGHT:**

## Digital Twins in Medicine

A medical digital twin is a dynamic, data-linked computational representation of an individual patient that updates over time and supports forecasting, simulation, and treatment optimization. Research activity in this area has grown rapidly, with publication counts rising from near 0 in 2015 to 372 in 2025 (Figure 6.2.10). Patent filings in healthcare digital twins (CPC class G16H) tell a similar story, with filings increasing from 30 in 2016 to 4,926 in 2025 (Figure 6.2.11).

However, conceptual clarity has not kept pace with publication growth. A 2025 scoping review in npj Digital Medicine assessed 149 human digital twin studies published between 2017 and 2024 and found that only 12.1% (18 studies) satisfied the National Academies of Sciences, Engineering, and Medicine (NASEM) definition of a digital twin. That definition requires three elements: personalization, dynamic updating, and predictive capability (Sadée et al., 2025). Only 19% of systems were tested in real healthcare environments.

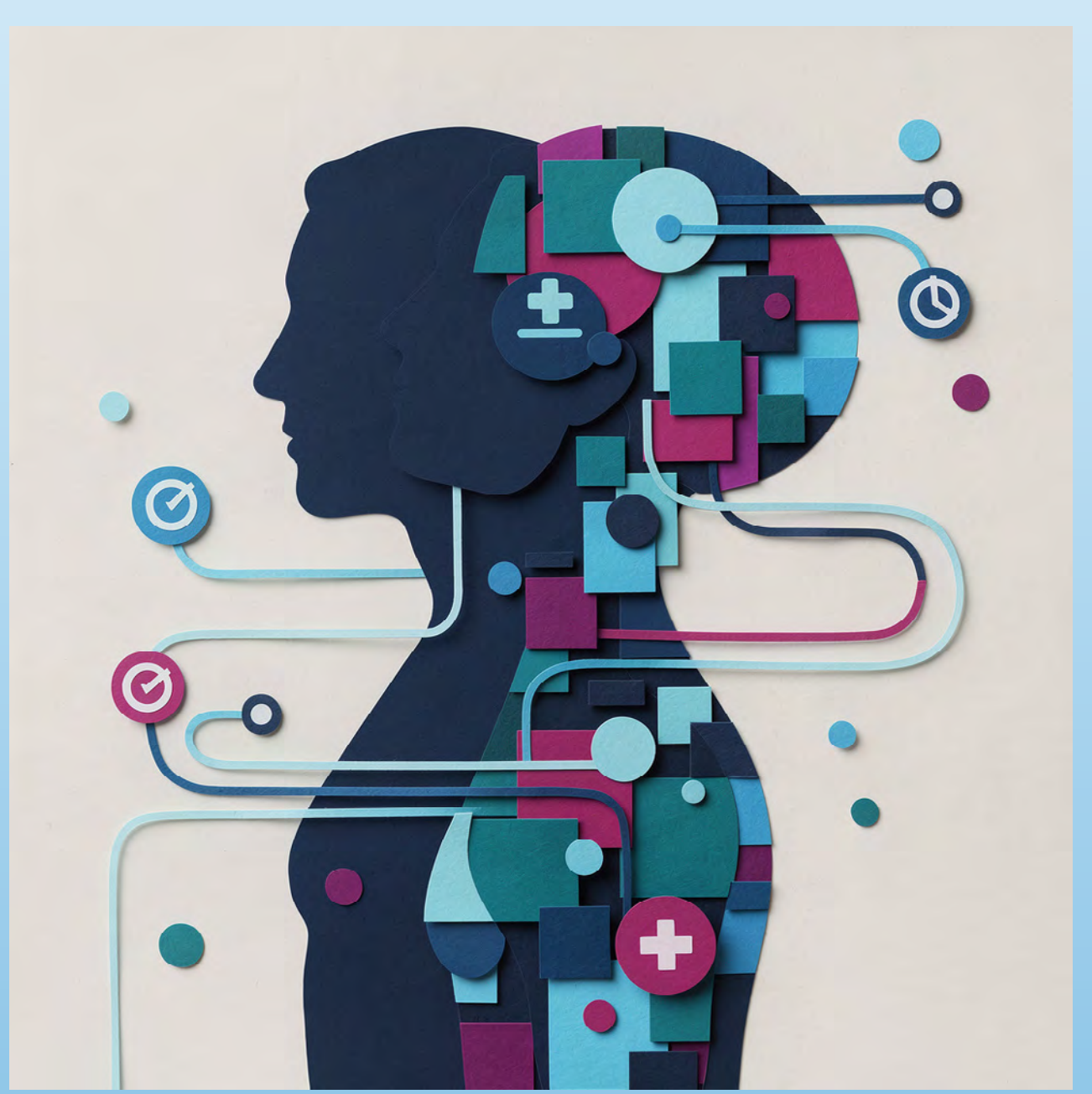

Clinical trials incorporating digital twin elements accelerated in 2025, particularly in oncology and diabetes. A pilot trial in prostate cancer using adaptive therapy concluded in 2025 with significantly increased survival (Zhang et al., 2022). New trials extended the approach to breast cancer (Mayo Clinic phase II) and ovarian cancer (ACTOv phase II RCT, n=80). For diabetes, a randomized controlled trial (n=150) of Twin Health's Whole Body Digital Twin platform found that 71% of participants achieved an HbA1c below 6.5% within twelve months, while safely reducing their intake of blood sugar–lowering medications.

**HIGHLIGHT:**

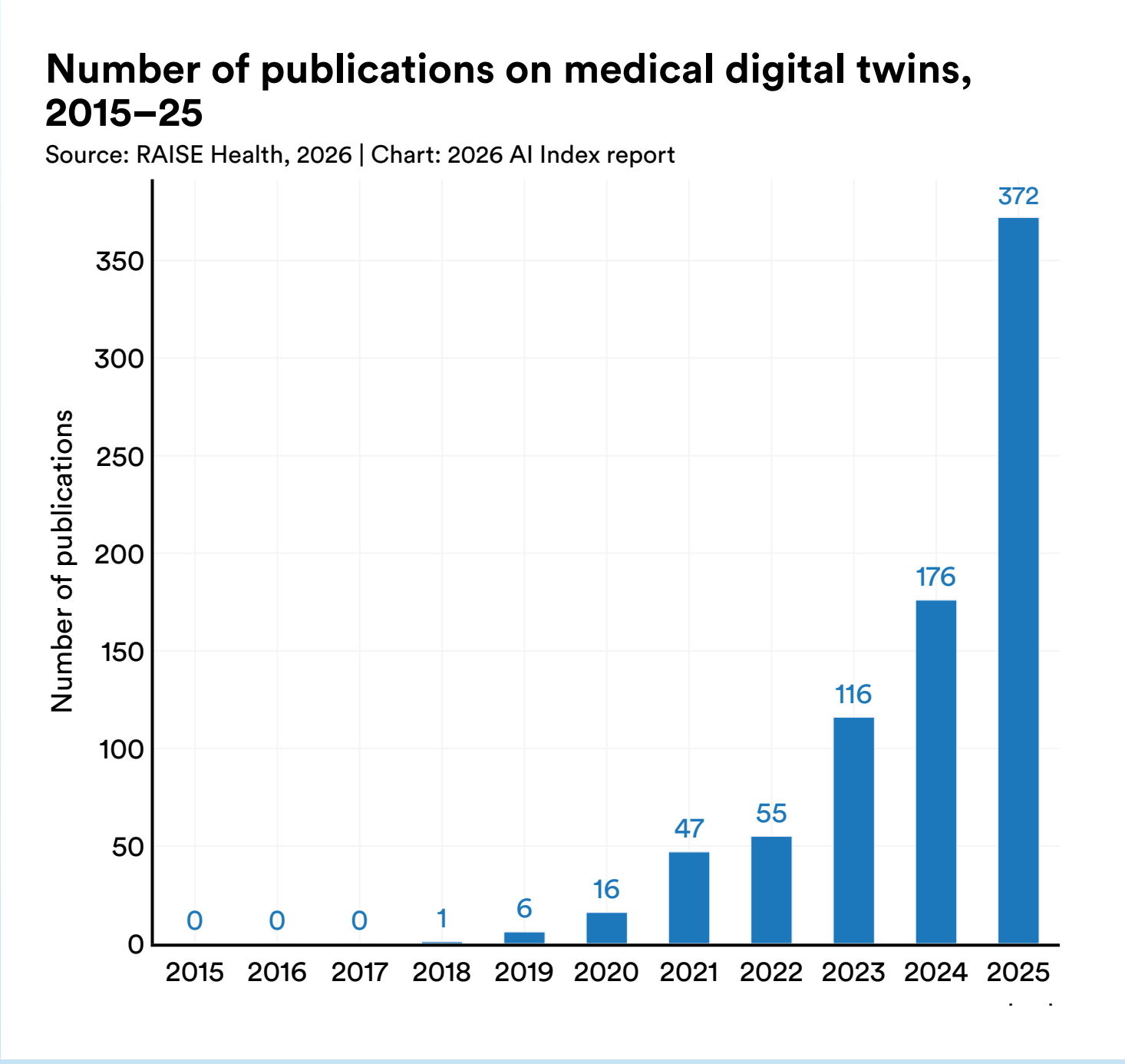

Figure 6.2.10

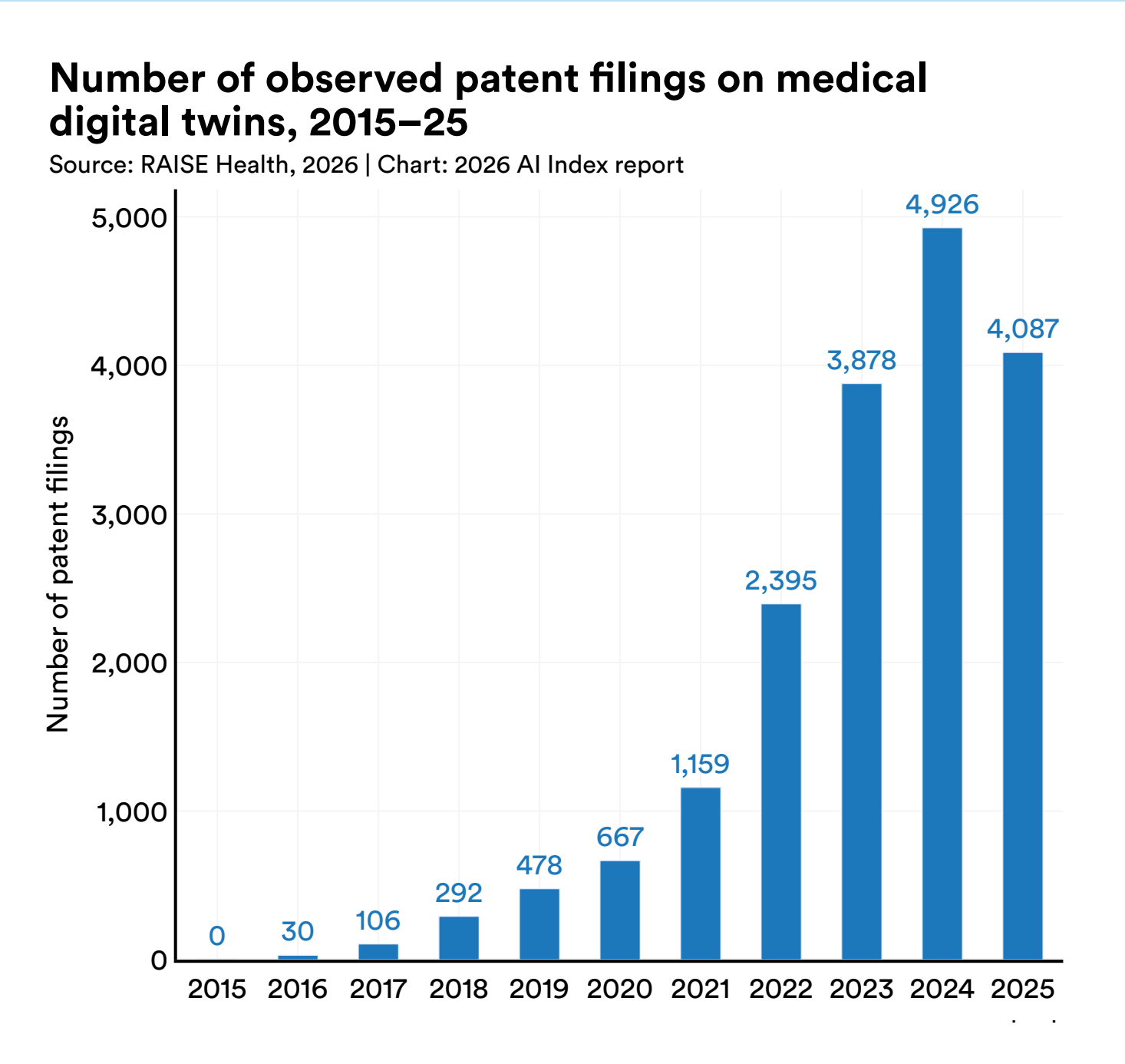

Figure 6.2.11[3]

3 The bar in 2025 appears lower than in 2024 because not all patents filed in 2025 have been published or become publicly available yet.

# 6.3 Patient Engagement

As patients interact more with AI tools—both through clinical workflows and consumer-facing platforms, efforts have been made to understand how they perceive these technologies. This section examines AI-generated health search results, patient attitudes toward AI in healthcare, and the emerging evidence base for patient-facing AI tools.

## AI Overviews for Health-Related Searches

AI-generated summary responses, referred to by Google as "AI Overviews," now appear at the top of most health-related search results. On average, 84%–92% of health-related queries triggered an AI Overview across five primary query types (Figure 6.3.1). Symptom and common health questions were the most likely to trigger an overview (92%), followed by treatment-related queries (90%) and condition-based queries (84%–88%). AI-generated summaries are a routine feature of health information searches, shaping the initial interpretation of questions posed by most users.

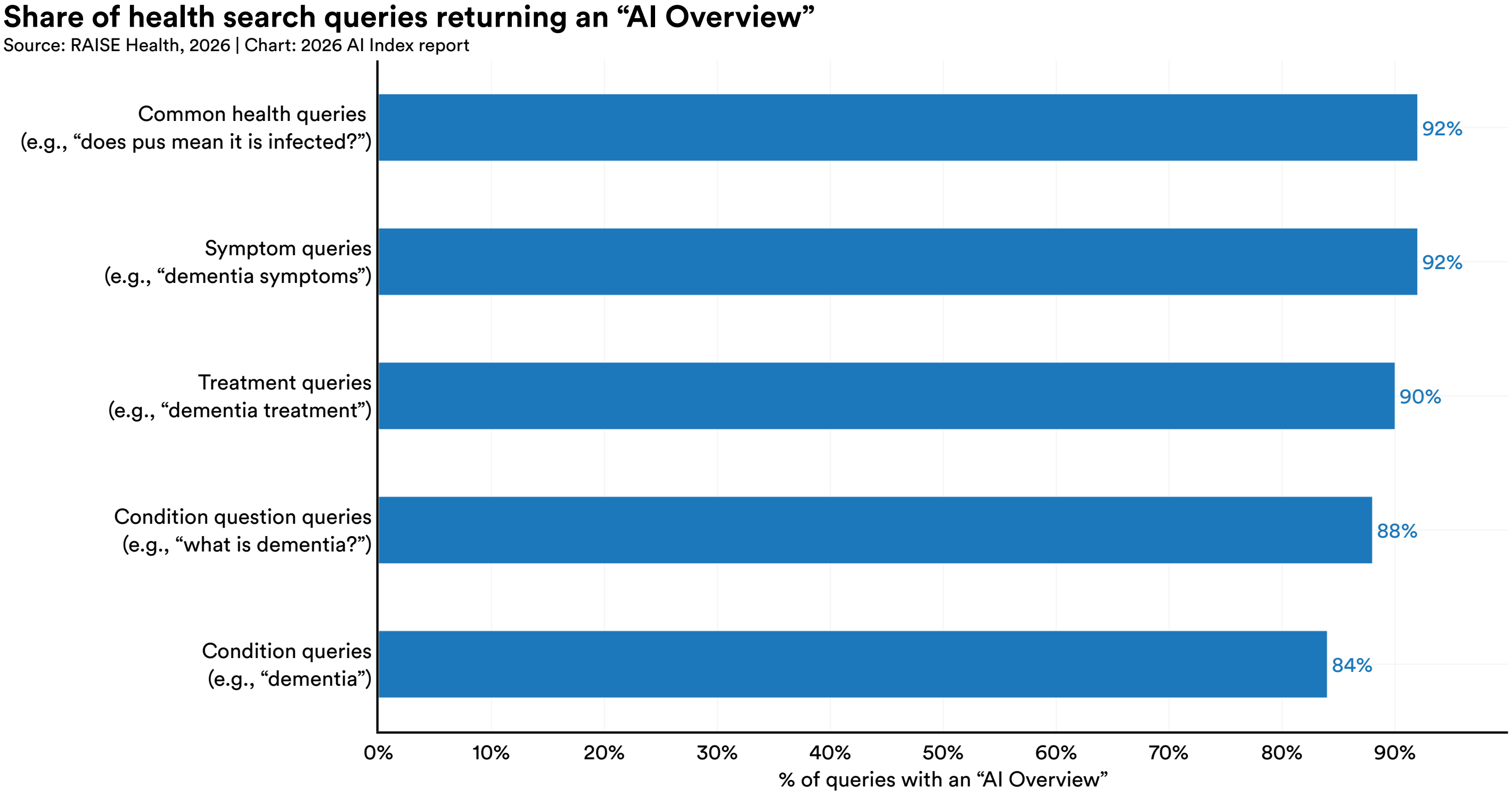
Share of health search queries returning an "AI Overview"
Source: RAISE Health, 2026 | Chart: 2026 AI Index report

Figure 6.3.1

## Patient Perspectives on AI in Healthcare

Publication volume on the patient perspective of AI in healthcare grew tenfold between 2020 and 2025 (Figure 6.3.2). Conditional acceptance emerged as a prevalent perspective across the literature. Patients tended to endorse AI in assistive roles rather than autonomous decision-making, particularly in high-stakes clinical contexts (Fee et al., 2025; Allen et al., 2025; Hmido et al., 2025). Demographic disparities in acceptance—patterned by age, gender, education, and race—were documented across multiple studies (Labinsky et al., 2025; Ogu et al., 2025; Li et al., 2025).

Preservation of the human relationship emerged as a consistent theme, with patients identifying the potential loss of empathic care as a primary concern (Carl et al., 2025; Davis et al., 2025). Trust in AI appeared to be clinician-mediated rather than technology-evaluated. Provider endorsement functioned as a key determinant of patient acceptance (Berger et al., 2025; Machado et al., 2025; Nong et al., 2025). Transparency and disclosure of AI use were similarly prioritized across populations, and emerging disclosure frameworks offer practical guidance for clinical settings (Figure 6.3.3).

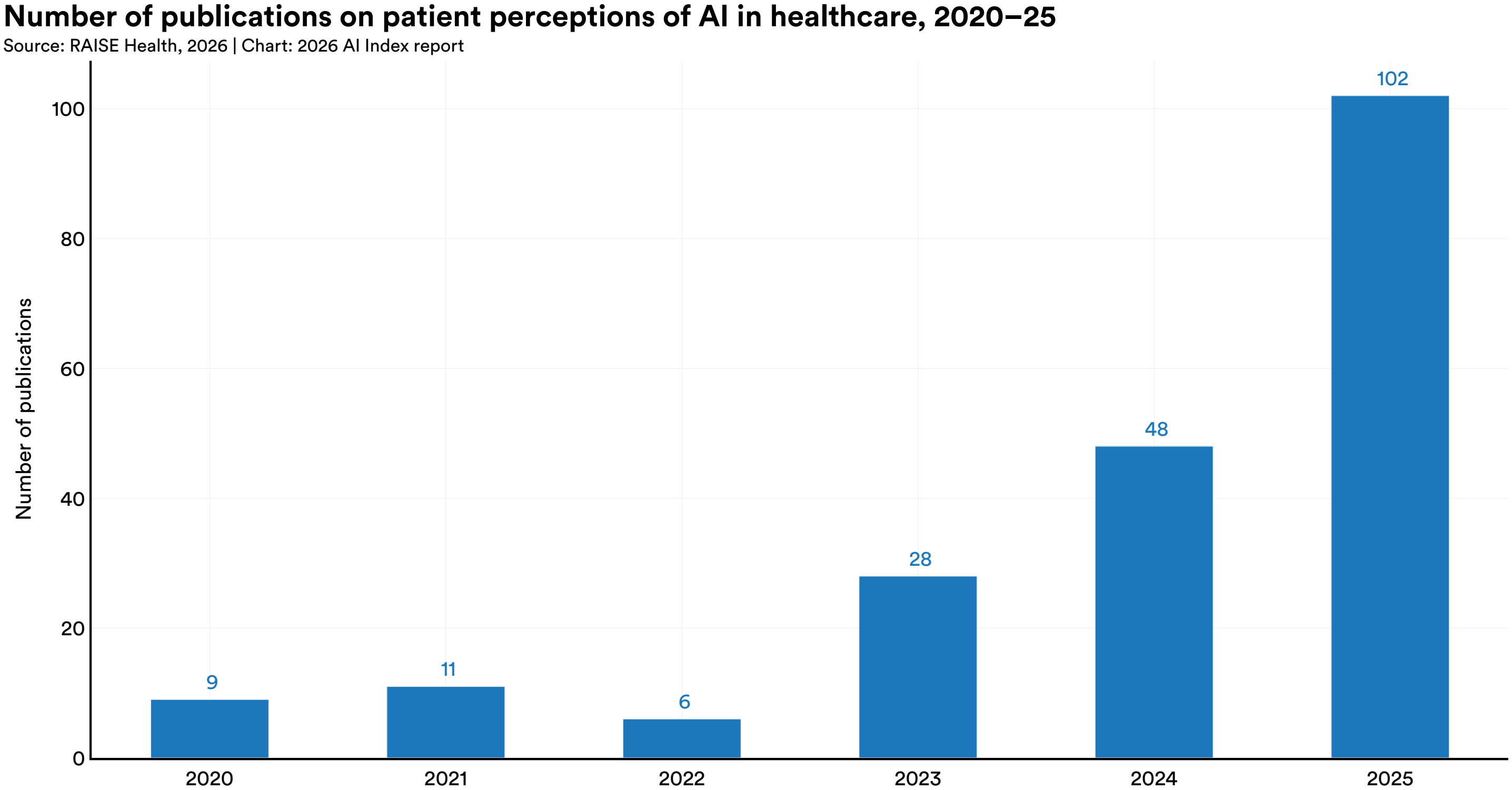
Number of publications on patient perceptions of AI in healthcare, 2020–25
Source: RAISE Health, 2026 | Chart: 2026 AI Index report

Figure 6.3.2

## Artificial intelligence use cases and recommendations for patient notification

Source: Mello et al., 2025

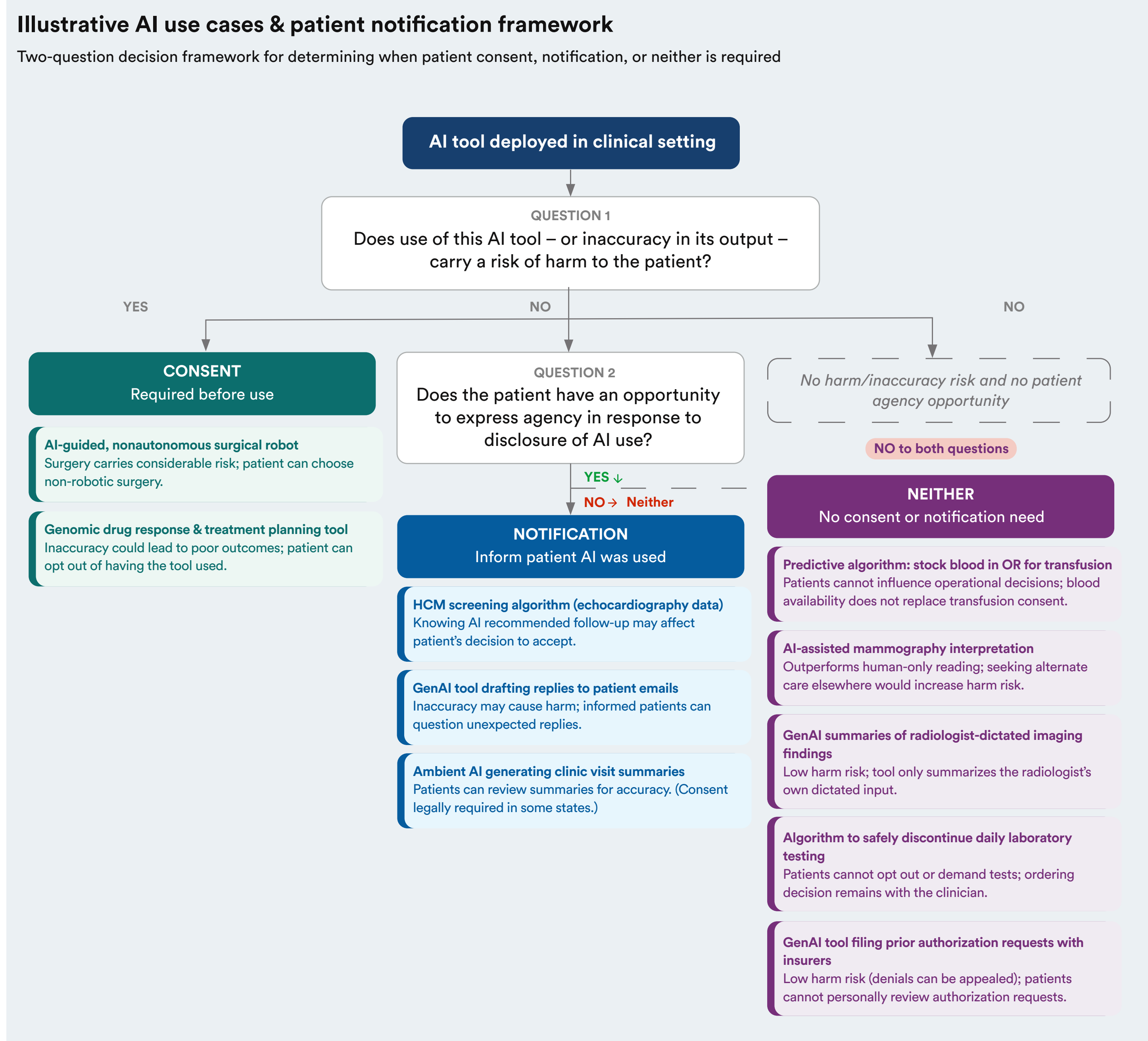

Figure 6.3.3

Internal medicine, radiology, and oncology were the most frequently represented specialties in this literature (Figure 6.3.4). The United States, United Kingdom, and Germany account for the greatest number of publications, while studies from sub-Saharan Africa, Latin America, and Southeast Asia remain underrepresented (Figure 6.3.5). Studies that include children and adolescents as participants, rather than drawing solely on parent or caregiver perspectives, remain rare.

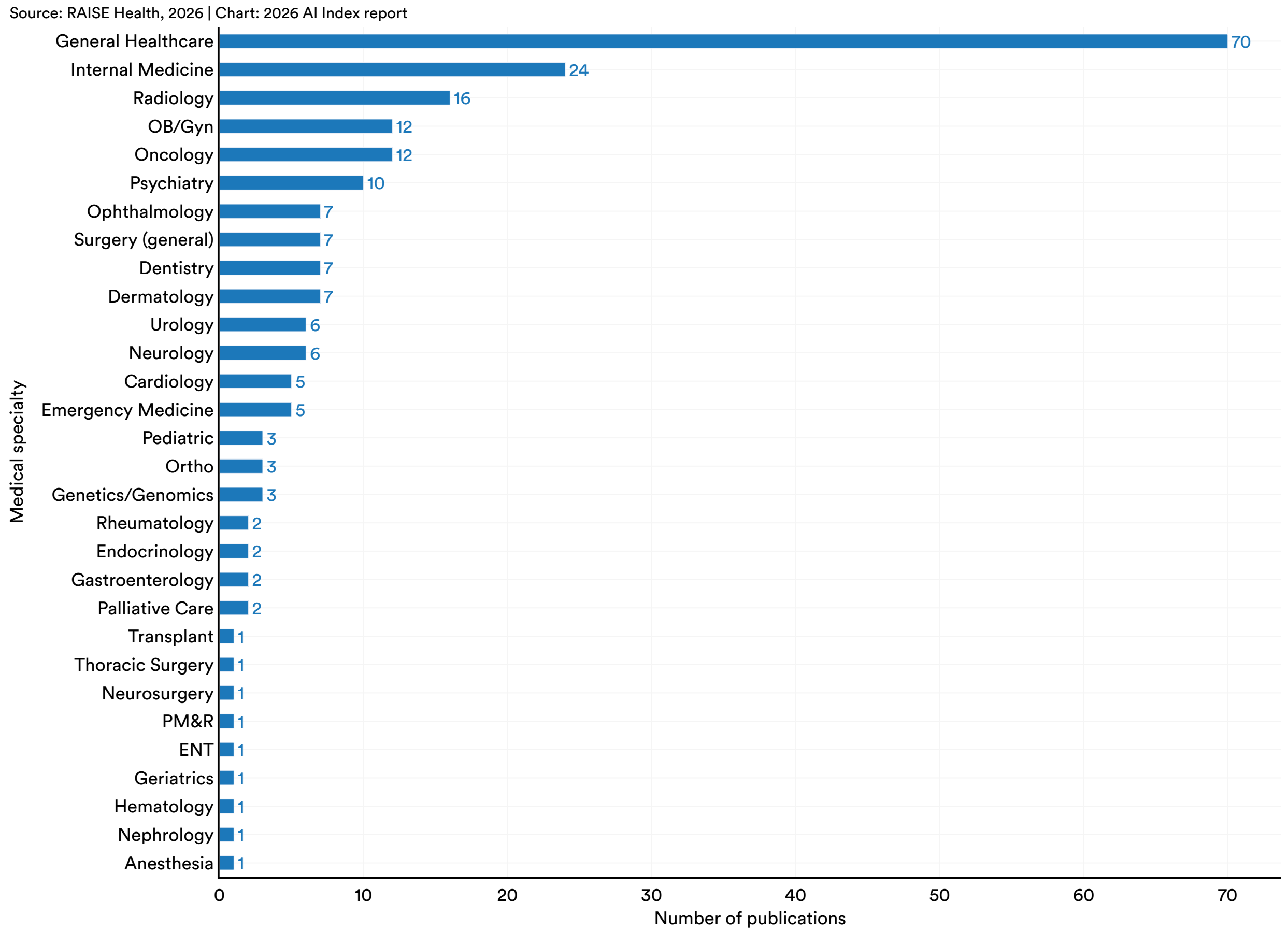
Medical specialties represented in publications exploring patient perceptions of AI in healthcare, 2020–25
Source: RAISE Health, 2026 | Chart: 2026 AI Index report

Figure 6.3.4[4]

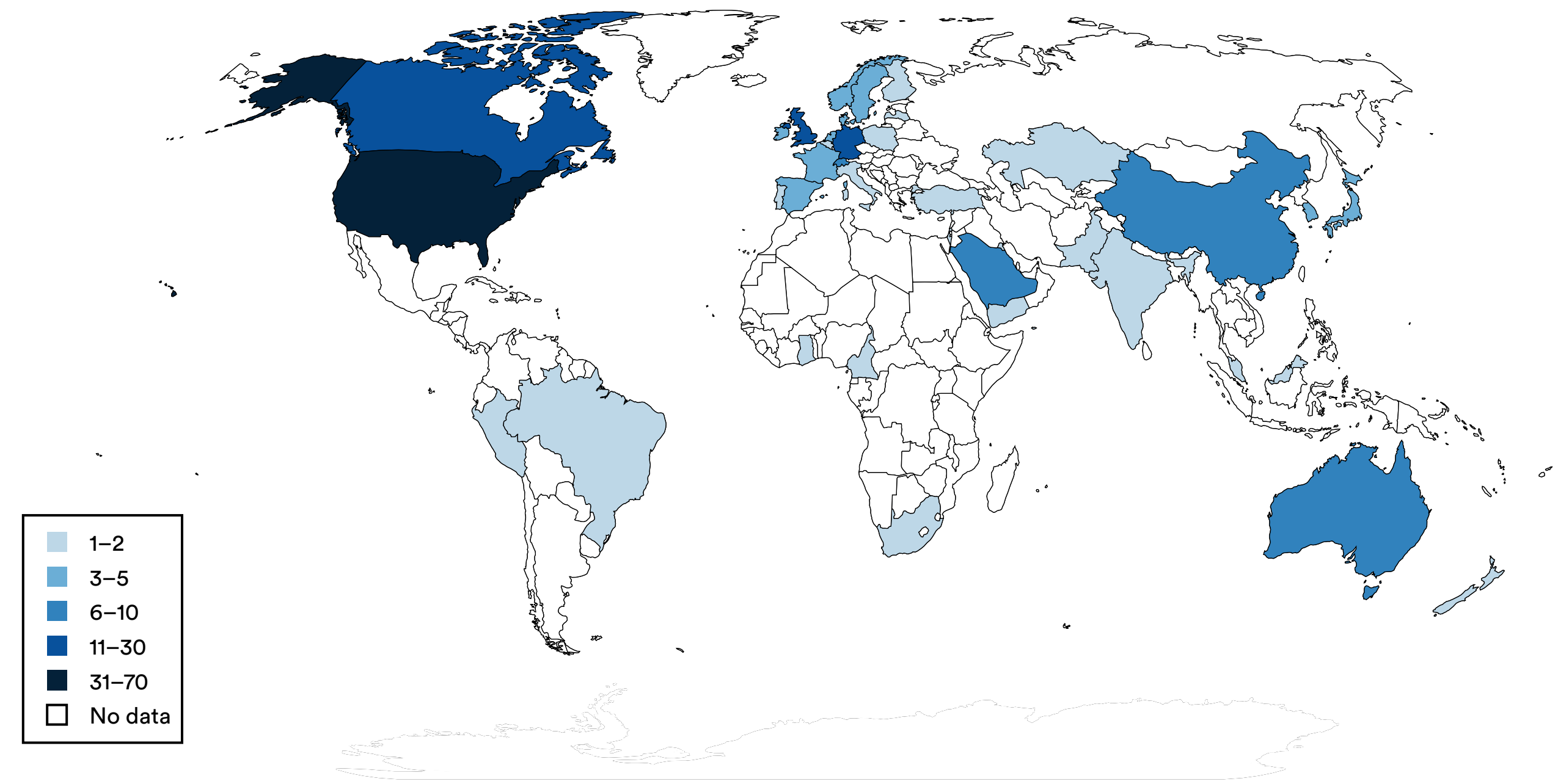
Geographic distribution of publications on patient perceptions of AI in healthcare by country, 2020–25
Source: RAISE Health, 2026 | Chart: 2026 AI Index report

Figure 6.3.5[5]

4 Publications may be tagged with multiple specialties.

5 Publications may be tagged with multiple countries; countries of origin are included. In total, 39 countries are represented (N = 204).

# 6.4 Ethical Considerations

This section tracks the volume and focus of ethical disclosure in medical AI publications, drawing on a bibliometric analysis of PubMed Central from January 2021 to December 2025. Publications were identified using search terms for medical AI and ethics, and then categorized by emphasis on data sharing, algorithm sharing, biosecurity, and global health. Ethics topics were either grouped under algorithmic, governance, or societal concerns.

## Volume and Concentration

Of the total number of medical AI publications in 2025, 43.4% discussed ethics topics—up from 37.1% in 2024 (Figure 6.4.1). The absolute number of such publications more than doubled between the two years. Among the specific topics discussed, the growth was concentrated on governance, outpacing algorithmic and societal concerns (Figure 6.4.2). In 2025, the number of governance-related publications reached 1,228, compared with 896 for algorithmic concerns and 874 for societal concerns.

Despite the attention paid to biosecurity in policy discussions, the subject is relatively unexplored in medical AI publications. In 2025, only 14 of these publications discussed biosecurity, with even fewer directly addressing the ethical implications of misuse or dual use (Figure 6.4.3).

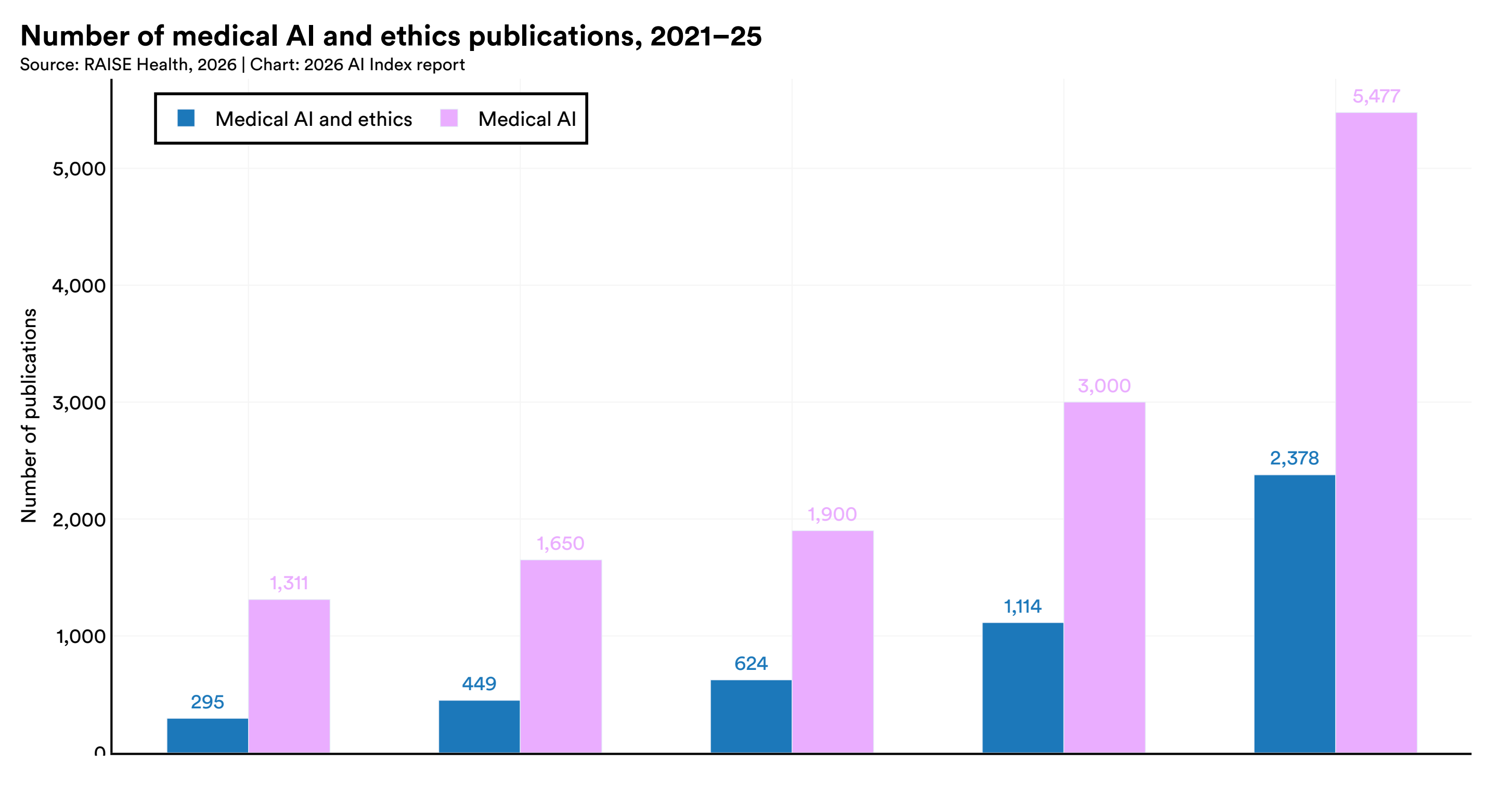

Figure 6.4.1

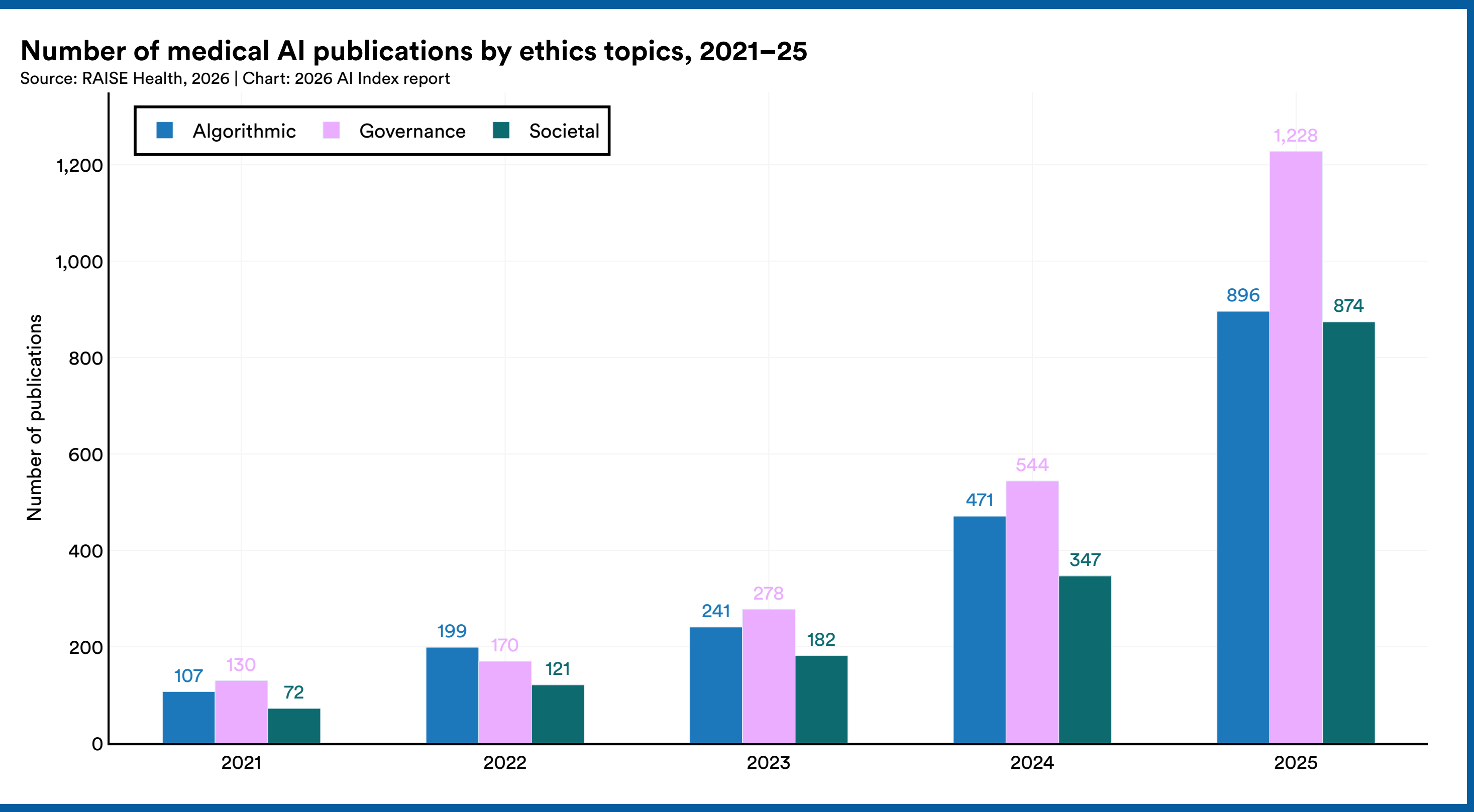

Figure 6.4.2

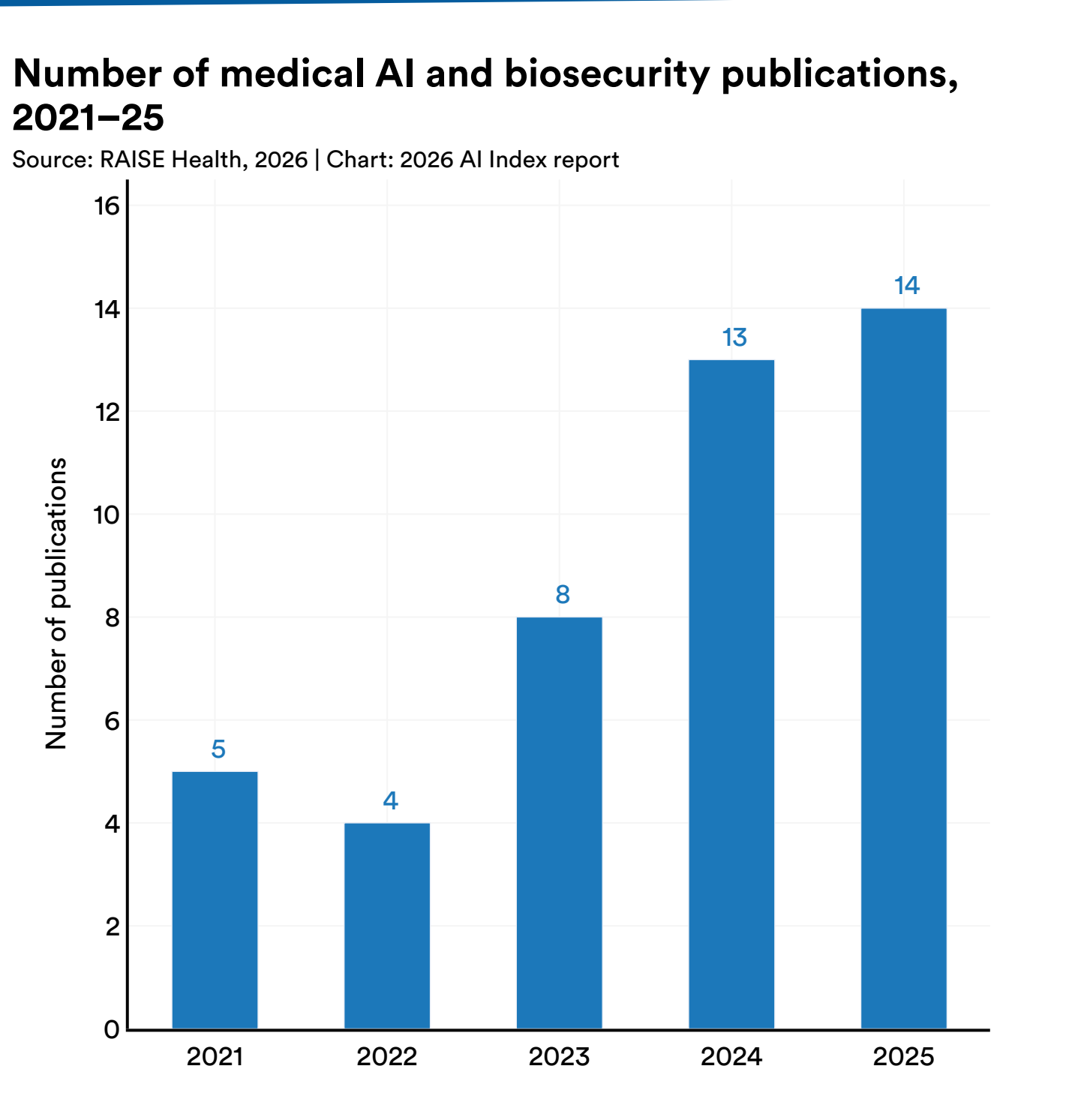

Figure 6.4.3

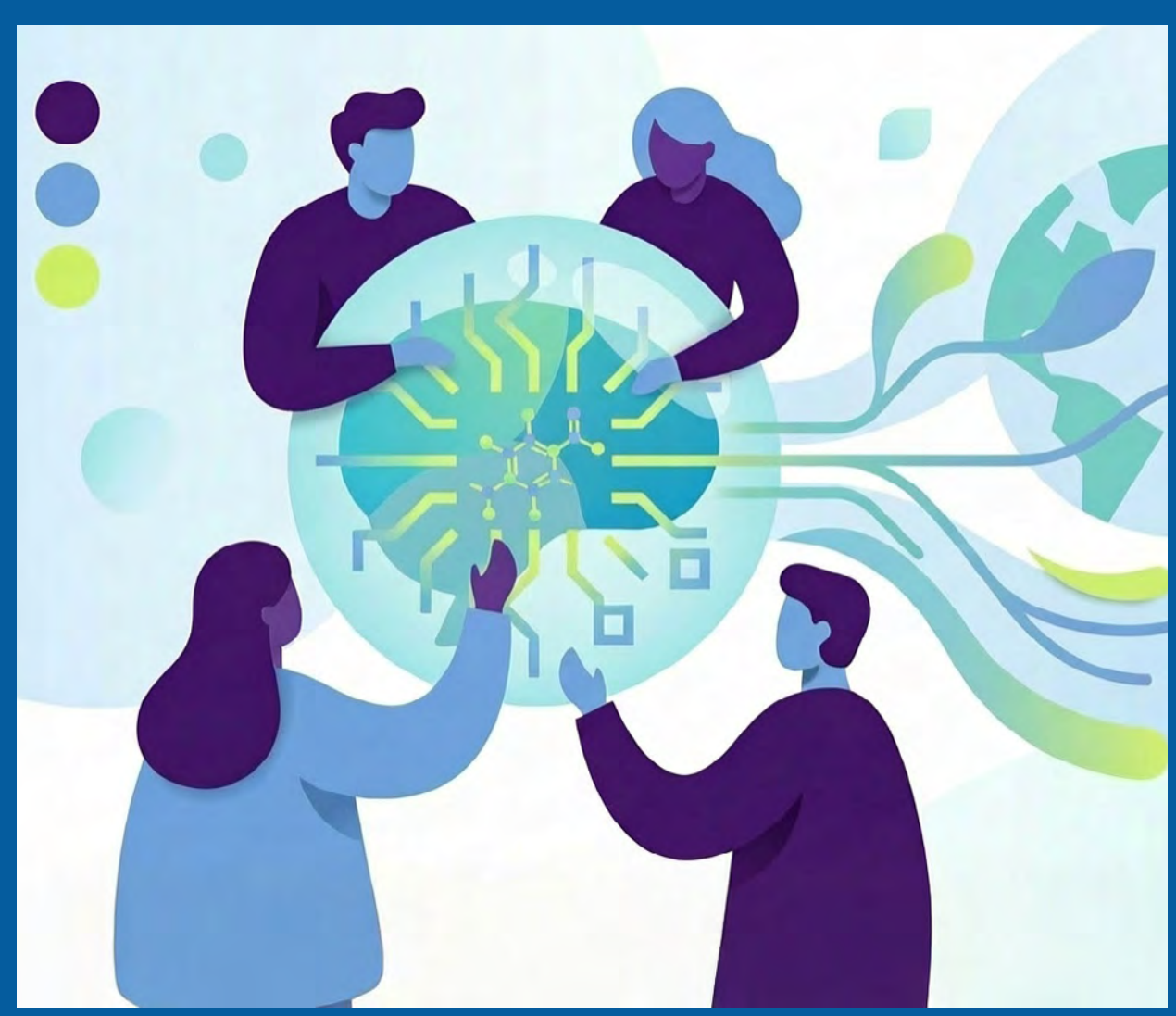

## Global Health: A Different Ethical Focus

Global health is an exception to the governance-dominated pattern. Among publications addressing global health in 2025, 51.8% (100 of 193) also mentioned ethics topics (Figure 6.4.4). Europe led with 38 publications, followed by East Asia (31) and North America (28), while sub-Saharan Africa, Latin America, and Oceania each produced fewer than five (Figure 6.4.5). In a departure from every other subcategory examined, societal concerns—including equity, justice, and accessibility—ranked highest in the global health context, surpassing both governance and algorithmic concerns (Figure 6.4.6). Researchers studying AI for global health are raising different questions from their peers working in the broader field.

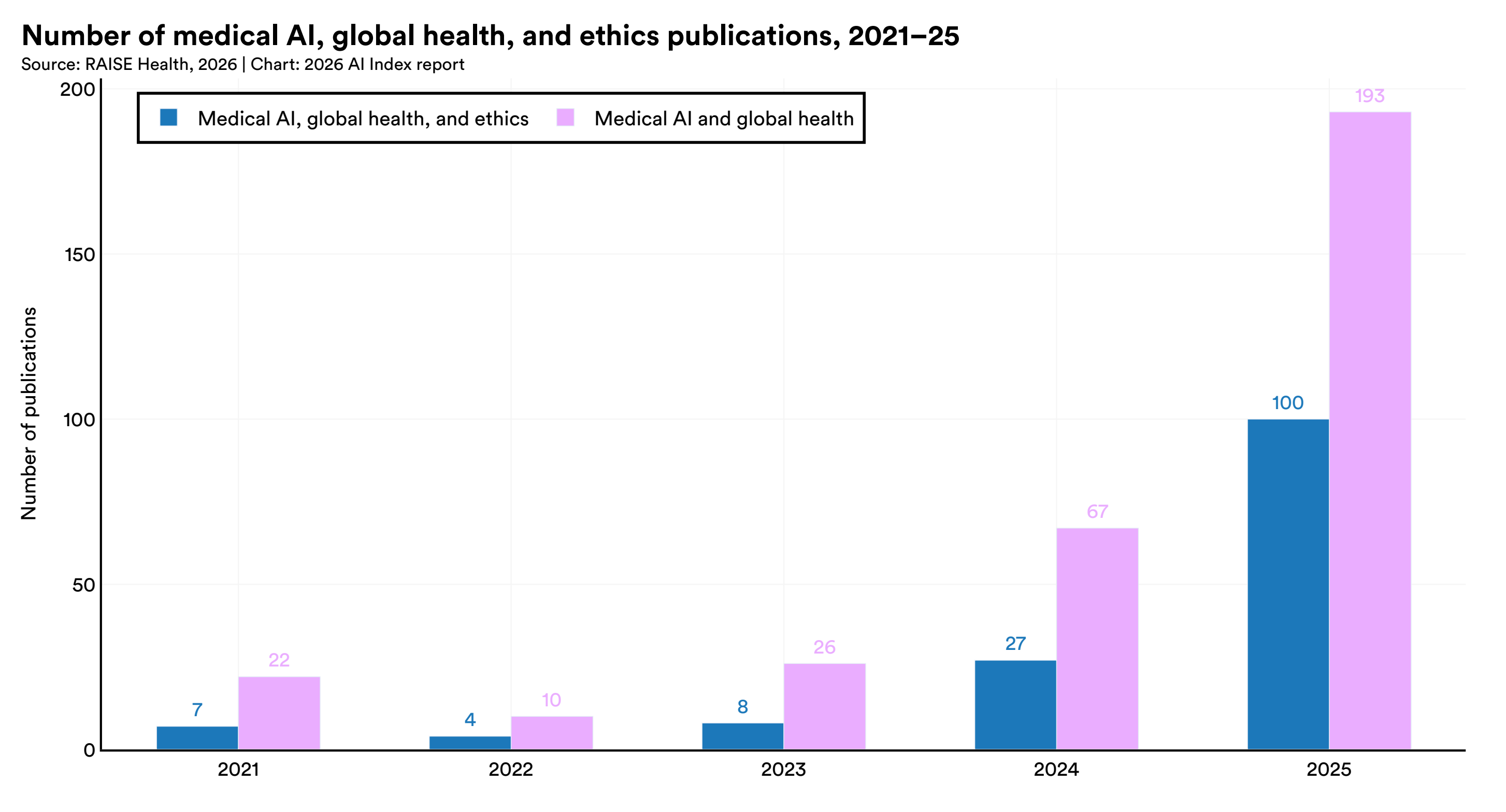

Figure 6.4.4

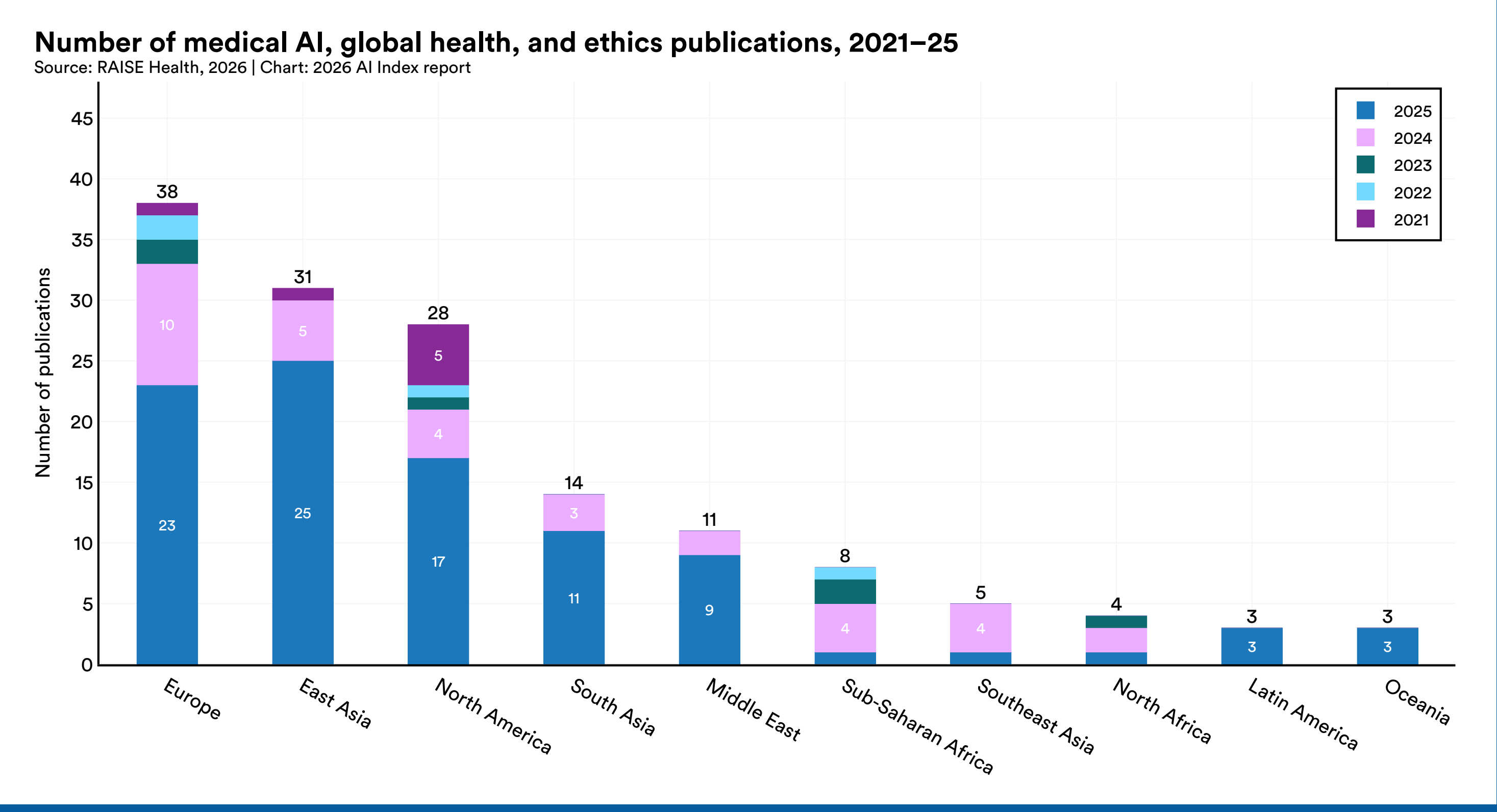

Figure 6.4.5

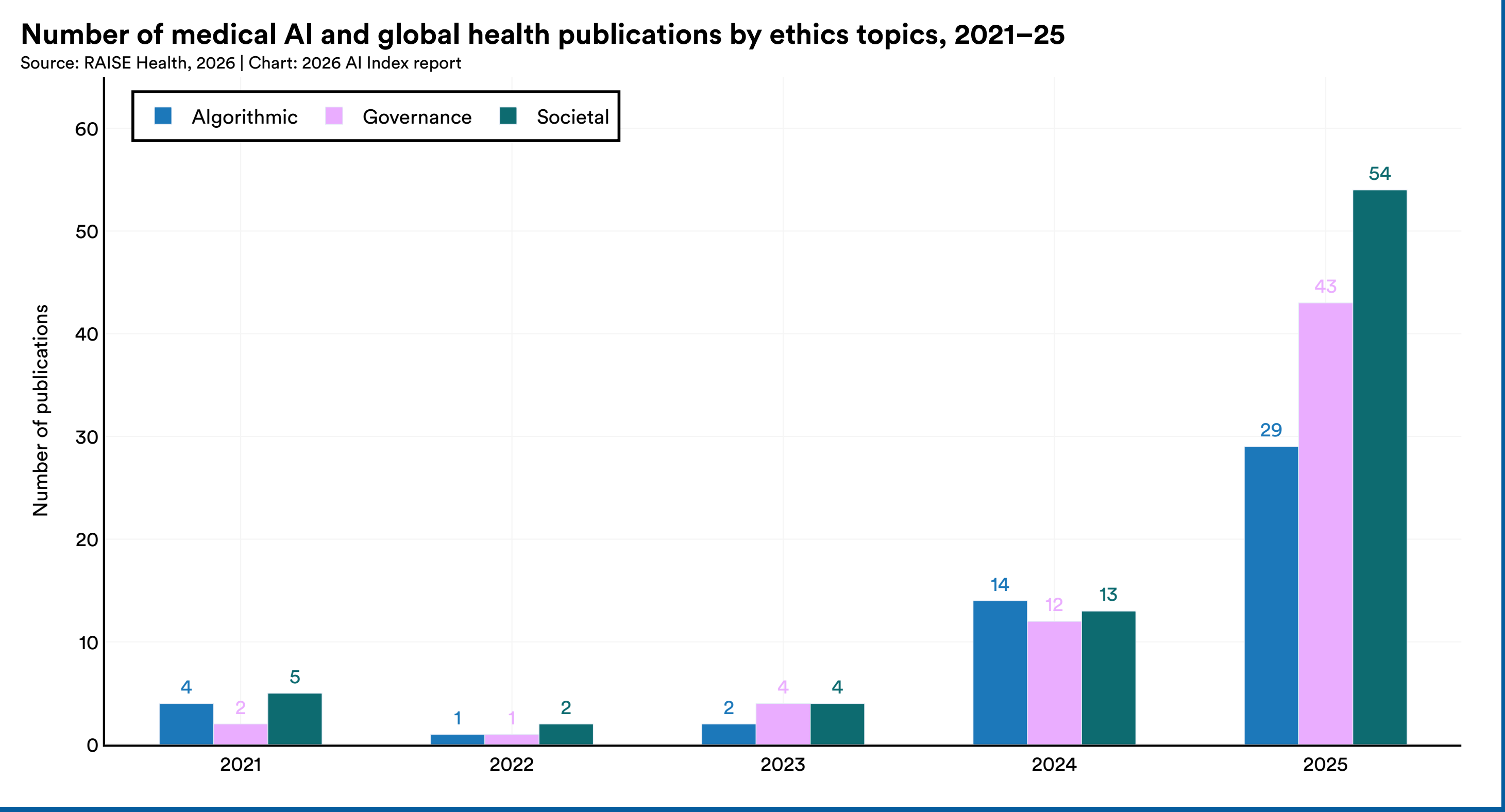

Figure 6.4.6

# 7

# Education

## Overview

Demand for AI education is growing across every level, but the systems needed to deliver it are still catching up. Computer science enrollment in post-secondary institutions is declining even as AI-related majors gain popularity. Students at both the university and K-12 levels are using AI tools in large numbers, yet access to AI-specific coursework and teacher training remain limited. Governments, including the United States', are pushing to integrate AI literacy into their curricula to maintain their countries' competitive edge. Yet, data on AI education is fragmented and lagging, and much of the analysis in this chapter relies on CS education data as a proxy. To survey the current landscape of AI and CS education, this chapter was prepared in collaboration with the Kapor Foundation, the Computer Science Teachers Association (CSTA), Expanding Computing Education Pathways (ECEP) Alliance, and the AI Index. The Kapor Foundation works at the intersection of racial equity and technology; CSTA is a global membership organization that supports educators in expanding access to CS education, and ECEP is a collective impact alliance focused on broadening participation in computing education.

# Contents

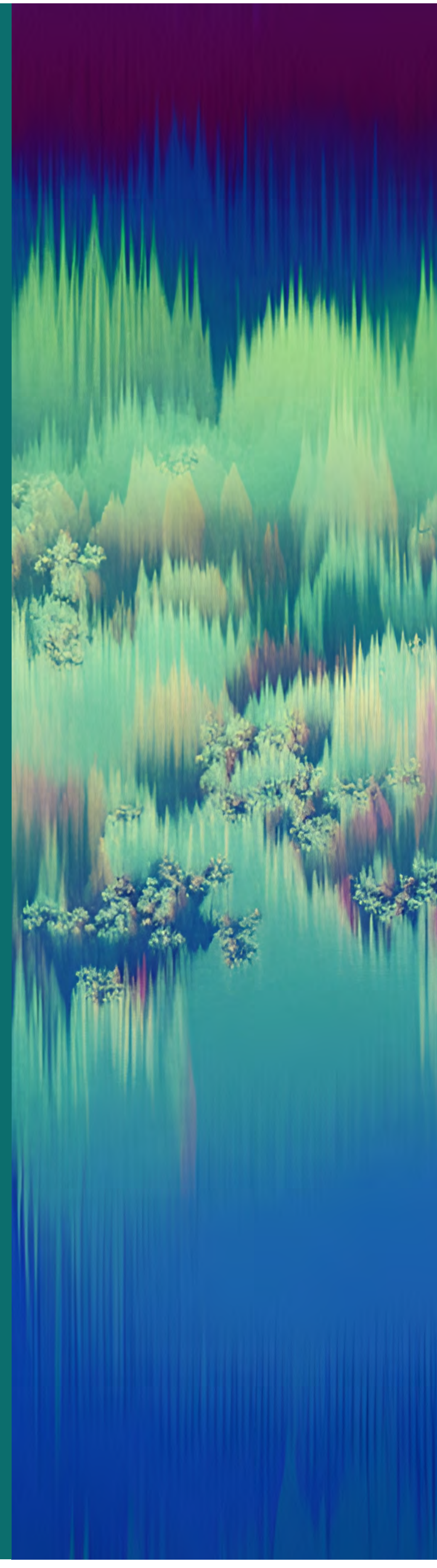

# Chapter Highlights

1. **CS enrollment fell 11% at U.S. four-year universities between 2024 and 2025, but AI-related graduate programs continued to grow.** Master's graduates in AI software-related fields rose 17% from 2023 to 2024, suggesting continued demand for AI specialization even as CS enrollment cools.

2. **The U.S. remains a global leader in producing information, communications, and technology (ICT) graduates at all degree levels, but other countries are growing faster.** Turkey, Brazil, and Mexico have increased their ICT graduate output more rapidly in recent years.

3. **Four out of five U.S. high school and college students now use AI for schoolwork, but school policies have not kept pace.** Only half of middle and high schools have AI policies, and just 6% of teachers say those policies are clear. Students most commonly use generative AI for research, essay editing, and brainstorming.

4. **More than 90% of countries now offer computer science to primary or secondary students, but AI education has been slower to take hold.** China and the United Arab Emirates both mandated AI education starting with the 2025-26 school year, signaling a shift toward formal AI instruction at the national level.

5. **The number of new AI PhDs in the United States and Canada increased 22% from 2022 to 2024, but the share going to industry has stayed flat.** All of the growth has gone to academia, reversing a decade-long trend of new AI PhDs flowing primarily into industry roles.

6. **People are acquiring AI skills outside formal education, and advertising AI skills in their resumes.** AI literacy has grown faster than engineering-oriented AI skills in most countries. The United Arab Emirates, Chile, and South Africa are exceptions, where engineering skills show steeper growth since 2022.

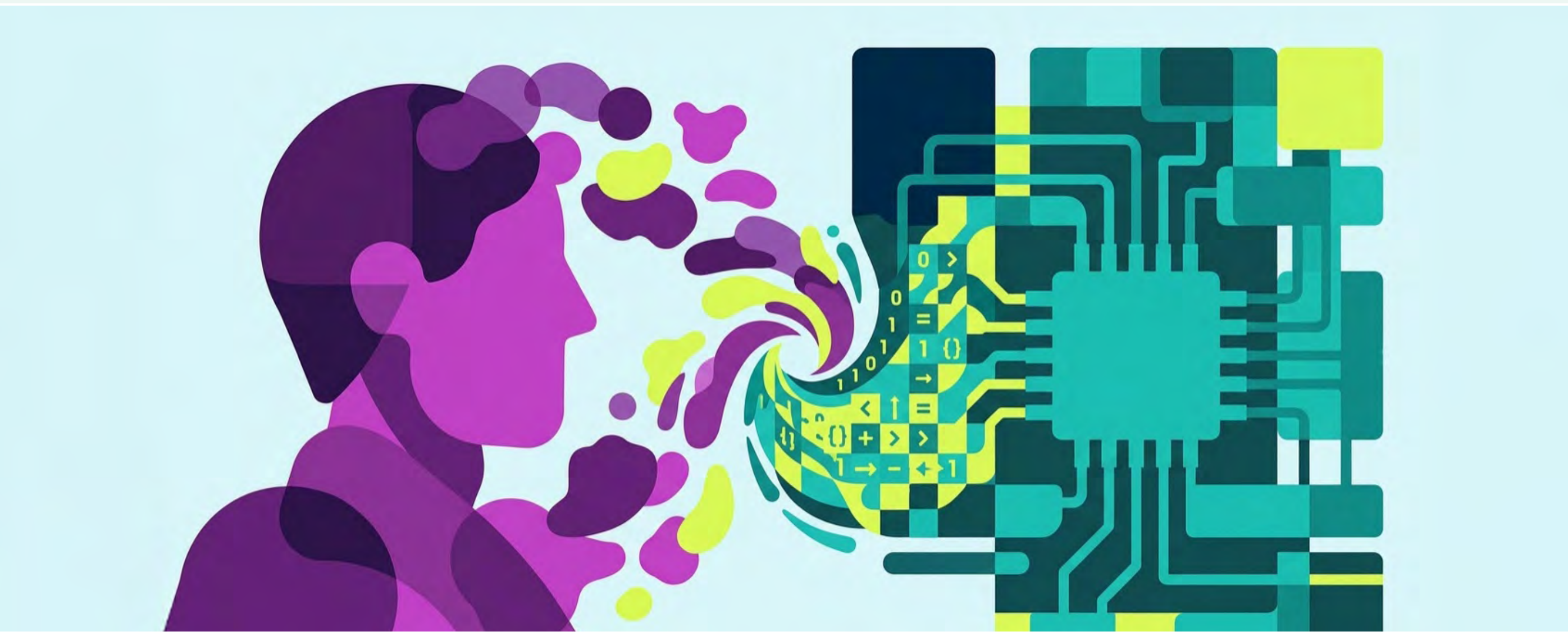

# 7.1 Background

AI's role in education is expanding faster than the data needed to track it and much of the data in this chapter is limited in scope, lagging in time, or both. Postsecondary figures reflect completion rates from the 2023–24 academic year and do not yet show enrollment shifts driven by the increasing interest in AI or reported decline in computer science (CS) enrollment. Global data, from the OECD, is only available through 2023 and does not include countries such as India, China, and much of Africa. At the K–12 level, there is little standardized data on AI-related course or program offerings. The metrics that are available are limited to CS education, which is not fully representative of AI education. Given these constraints, a complete assessment of the growing demand for AI education, and how current systems are meeting it, is not possible.

Public discussion about AI in education continues to expand, engaging developers, ed tech vendors, education advocates, policymakers, and educators around the role of AI, with a particular focus on its risks and benefits.[1] There is broad agreement on the importance of AI literacy for all students and its designation as a critical skill for academic, professional, and civic navigation and success. Public discourse often fails to distinguish between AI in education, AI literacy, and AI education (Figure 7.1.1). AI in education is the use of AI to complete teaching and learning tasks. AI literacy refers to the knowledge necessary for a foundational understanding of AI, how it works, how to use it, and the risks of usage. AI education builds on AI literacy with the addition of the technical skills required to build AI systems. Because these terms are often blurred in public discussions, clarity about which topic is being addressed matters for how educators, researchers, and policymakers communicate about AI.

This chapter focuses on AI education and AI in education. Where comprehensive data about AI education is not available, computer science (CS) education data is presented instead.

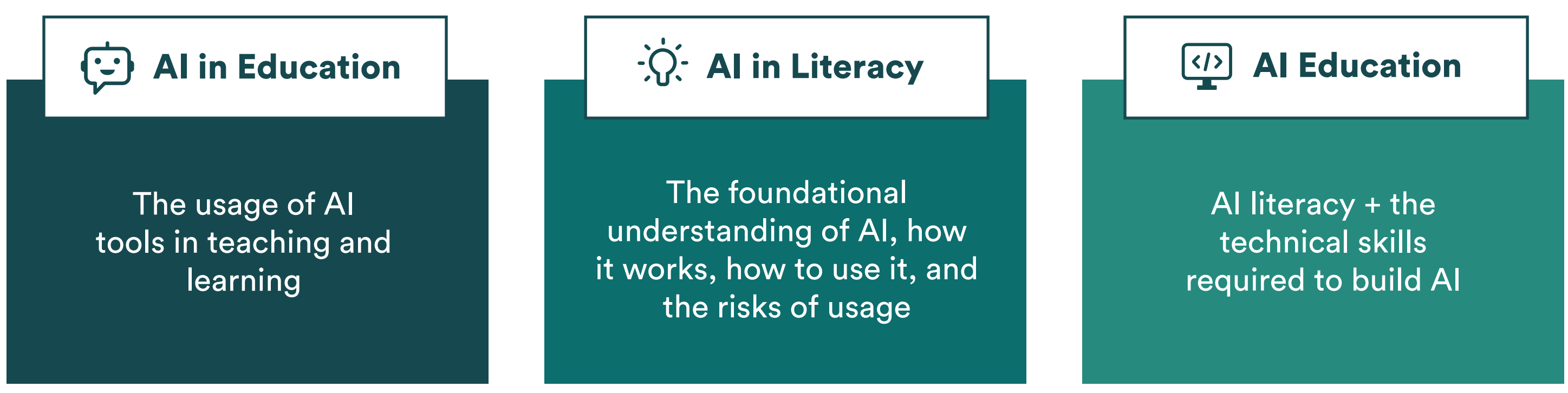

Figure 7.1.1

1 In *A New Direction for Students in an AI World: Prosper, Prepare, Protect*, a January 2026 report from the Center for Universal Education at the Brookings Institution, the authors provide an expansive catalogue of the risks and opportunities related to AI in education. In Annex A, they offer four definitions of "AI literacy" from frameworks published in the last five years.

# 7.2 Postsecondary CS and AI Education

The generative AI usage among students has reshaped the conversation about the purpose and role of postsecondary education. At the same time, task automation in coding roles has appeared to slow the entry-level job market for CS graduates. These shifts have translated to declines in CS enrollment in postsecondary institutions. Between 2024 and 2025, enrollment in CS as an undergraduate major in four-year universities declined 11%.[2] Chapter 4 (Economy) documents a similar transition in the labor market, where employment among the youngest software developers has declined since 2024 even as overall AI hiring grows. So, students are responding to a shifting job market, but because degree completion lags enrollment by several years, the full effects will take time to appear in the data. Even as CS enrollment declines, there is evidence that AI-related majors are becoming more popular.

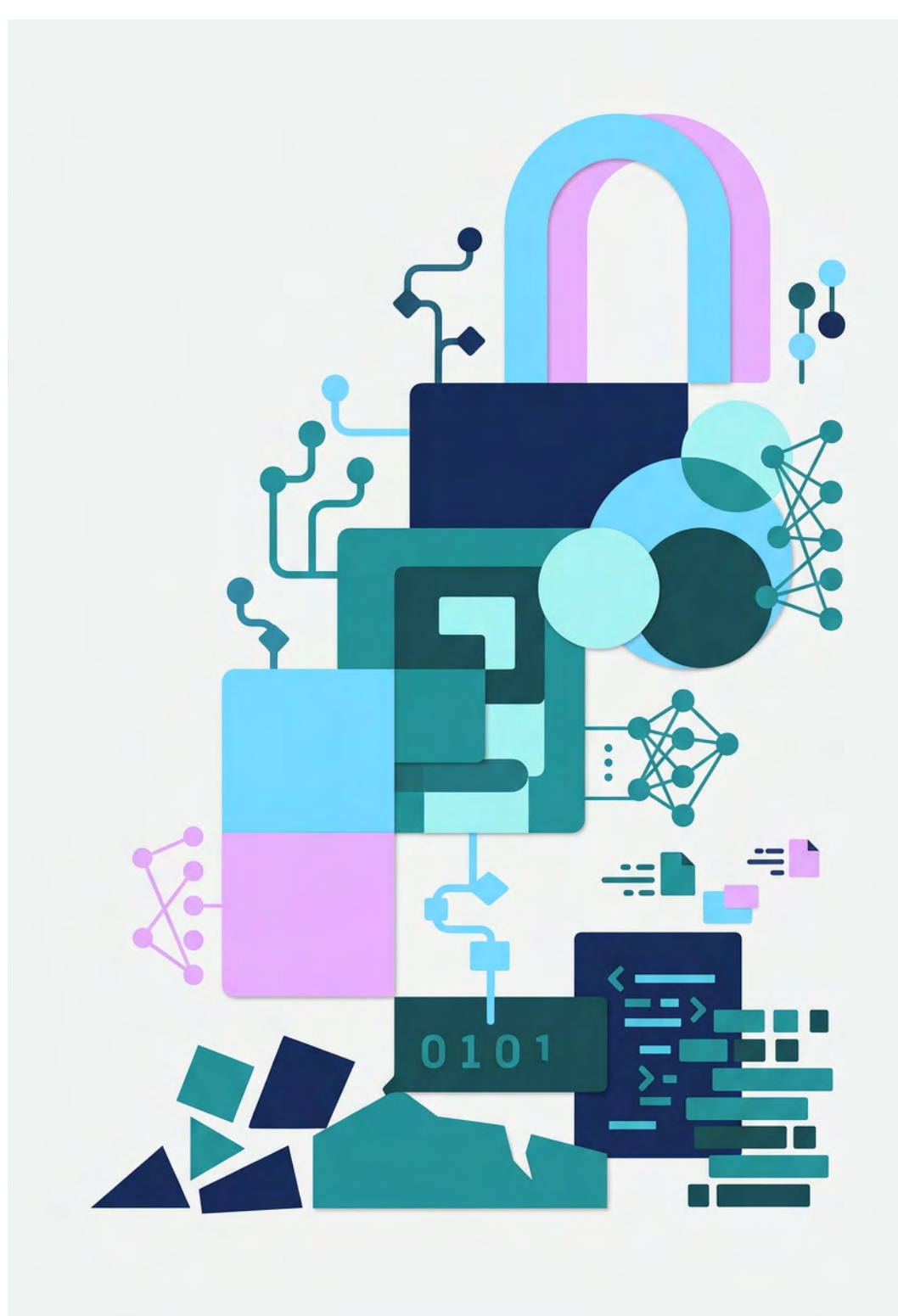

## U.S. Degree Graduates

Previous versions of the AI Index focused primarily on CS degrees, since very few graduates are classified under an Artificial Intelligence major.[3] This year, the AI Index added AI-relevant majors, as determined by the January 2025 White House AI Talent Report. The report divides AI-relevant majors into two categories: AI software, which includes majors such as Artificial Intelligence, Computer Programming/Programmer, and Computational and Applied Mathematics; and AI hardware, which includes majors such as Electrical and Electronics Engineering, Condensed Matter and Materials Physics, and Industrial Engineering.[4]

AI software-related degrees have steadily increased in popularity over the past 10 years, especially at the bachelor's and master's levels (Figure 7.2.1). The largest increase has been at the master's level, with an 82% increase in graduates between 2022 and 2024 and a 17% increase between 2023 and 2024. The number of AI hardware–related degrees has remained flat or declined; bachelor's degrees, in particular, have declined 13% since reaching a peak in 2020.

2 The declines are based on CS specifically as a major, not the broader category of Computer and Information Sciences and Support Services.

3 The Classification of Instructional Programs (CIP), developed by the National Center for Education Statistics (NCES), designates "Artificial Intelligence and Robotics" under CIP code 11.0102. Despite the availability of this code since 2016, very few schools use it, instead choosing to classify students under 11.0101 (Computer and Information Sciences, General).

4 See the methodology section for a complete list of majors.

### New AI-related postsecondary graduates in the United States, 2014–24

Source: National Center for Education Statistics, Integrated Postsecondary Education Data System, 2024 | Chart: 2026 AI Index report

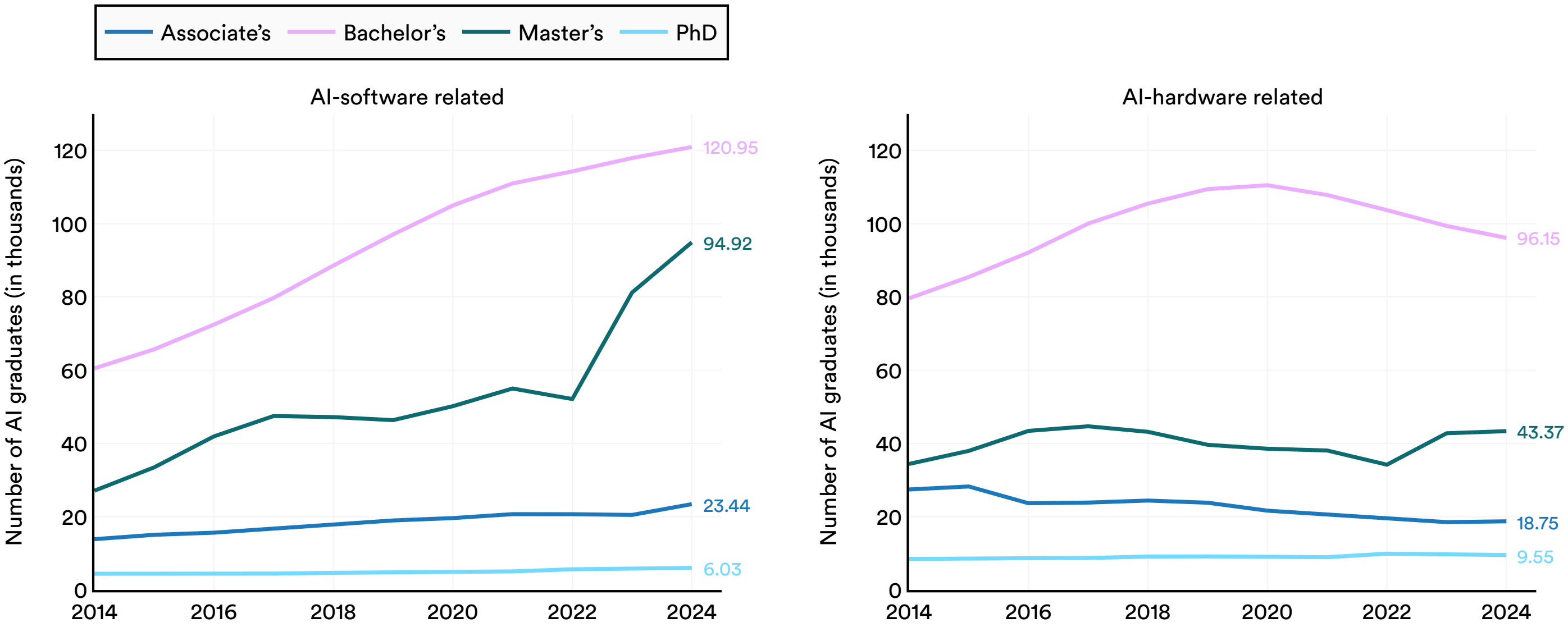

Figure 7.2.1

Women remain underrepresented across AI-related degrees, though they have slightly higher levels of representation in AI software–related degrees than in AI hardware-related degrees, peaking at 36% of AI software–related master's degree graduates (Figure 7.2.2). By comparison, women continue to account for nearly 60% of all degrees.

### AI-related postsecondary graduates in the United States by gender, 2024

Source: National Center for Education Statistics, Integrated Postsecondary Education Data System, 2024 | Chart: 2026 AI Index report

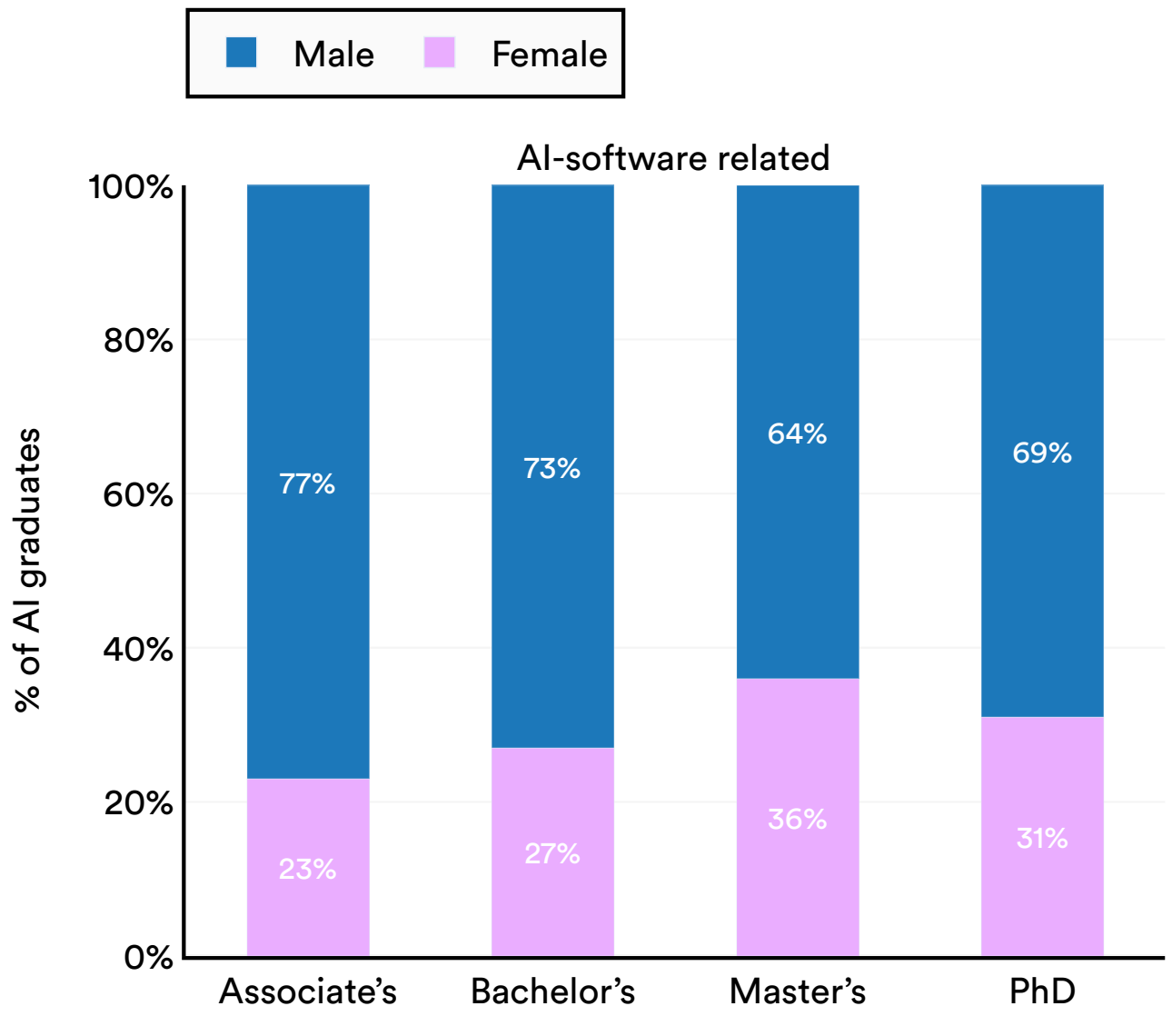

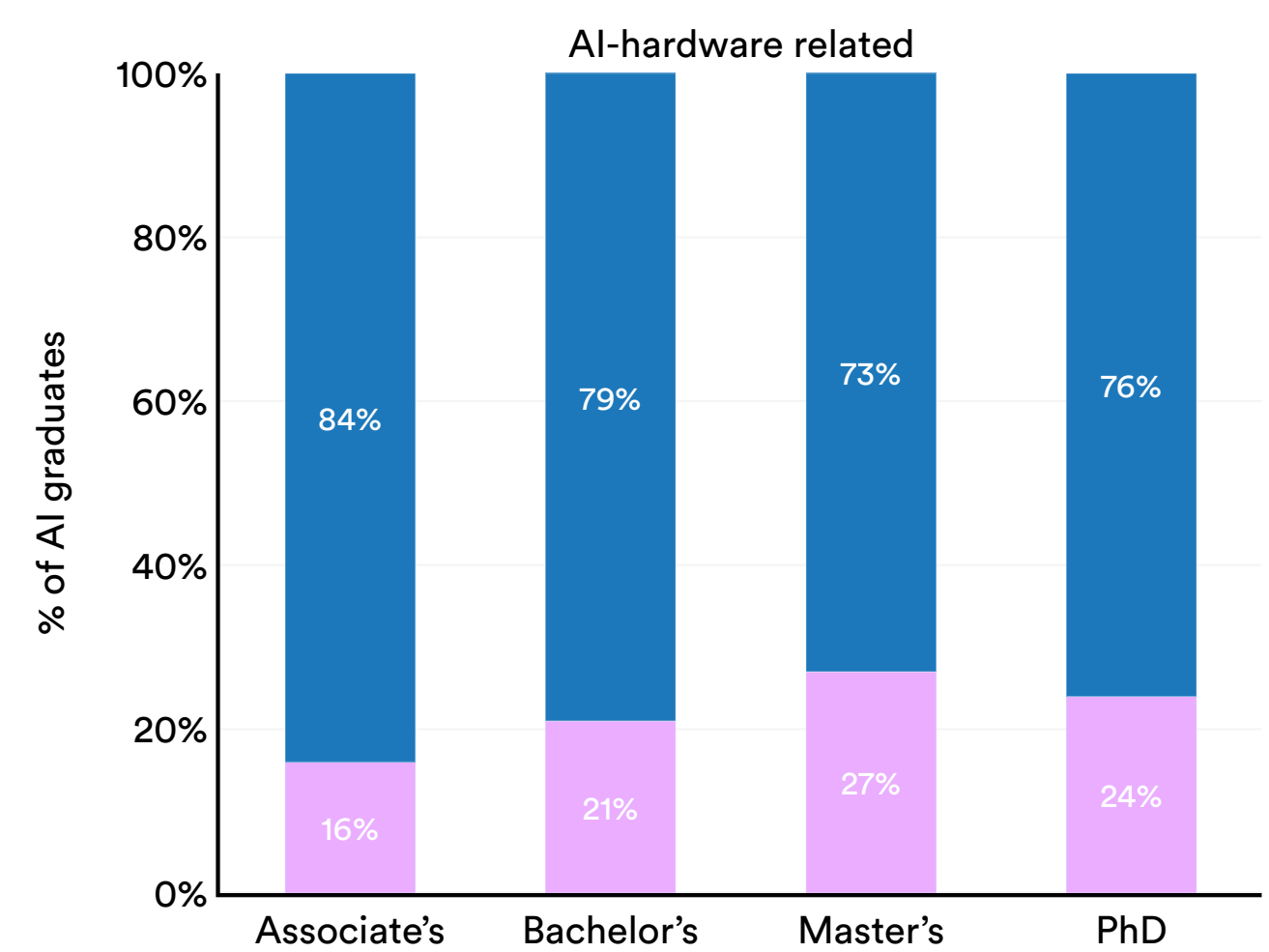

Figure 7.2.2

For AI software–related degrees, Hispanic/Latino, Black, Native Hawaiian/Pacific Islander, and Native American/Alaskan students are underrepresented at all levels (Figure 7.2.3). White students are also underrepresented, though to a lesser degree, except at the PhD level. Multiracial and Asian students are overrepresented, with Asian master's students the most overrepresented.

Representation patterns tend to be consistent across degree levels within each racial group (Figure 7.2.4). The main exceptions are Asian students and Native American/Alaskan students, whose representation varies by degree level. Hispanic/Latino students are slightly underrepresented at all levels, and Native Hawaiian/Pacific Islander and Black students remain underrepresented at all levels.

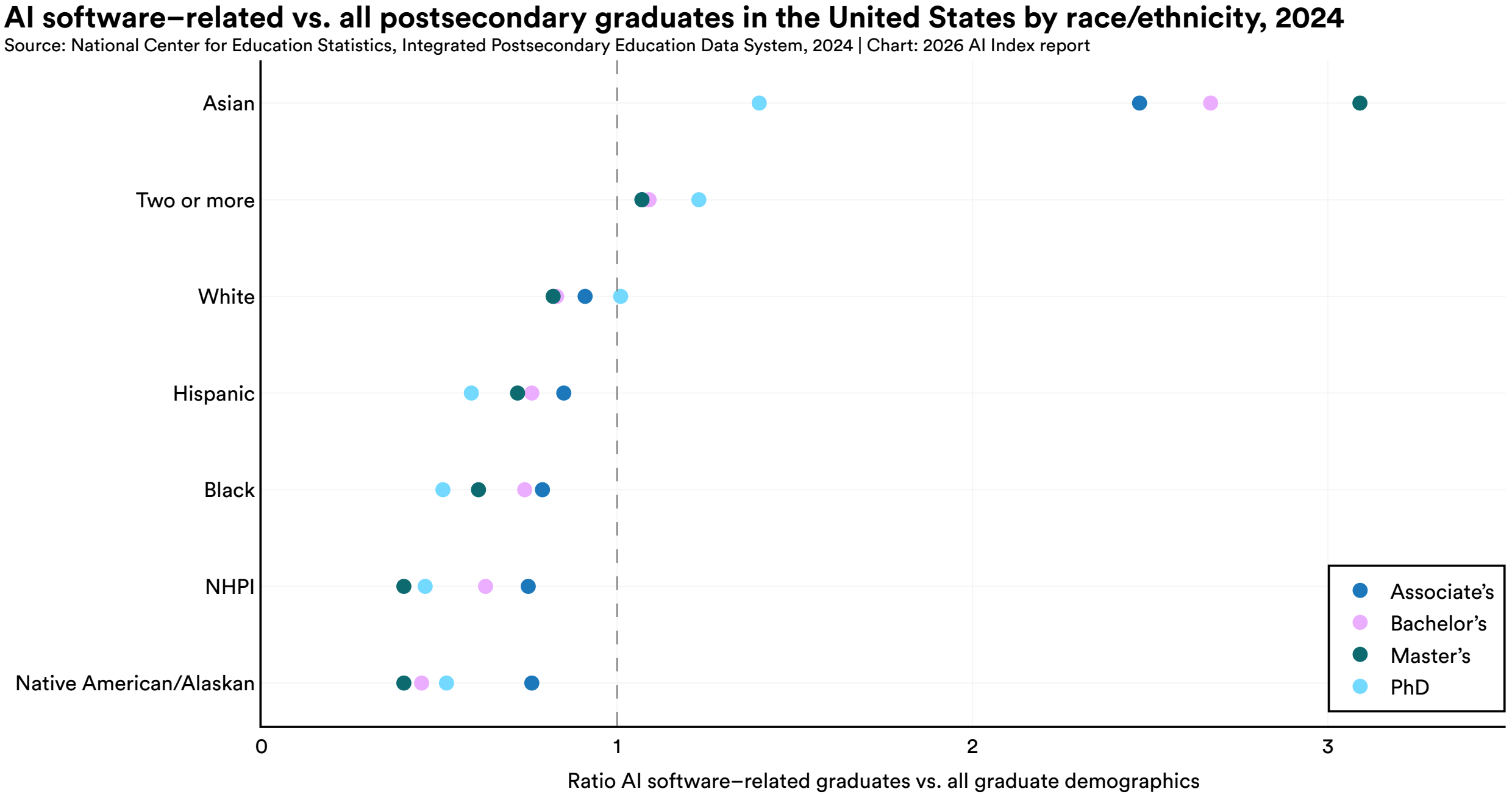
AI software–related vs. all postsecondary graduates in the United States by race/ethnicity, 2024
Source: National Center for Education Statistics, Integrated Postsecondary Education Data System, 2024 | Chart: 2026 AI Index report

Figure 7.2.3

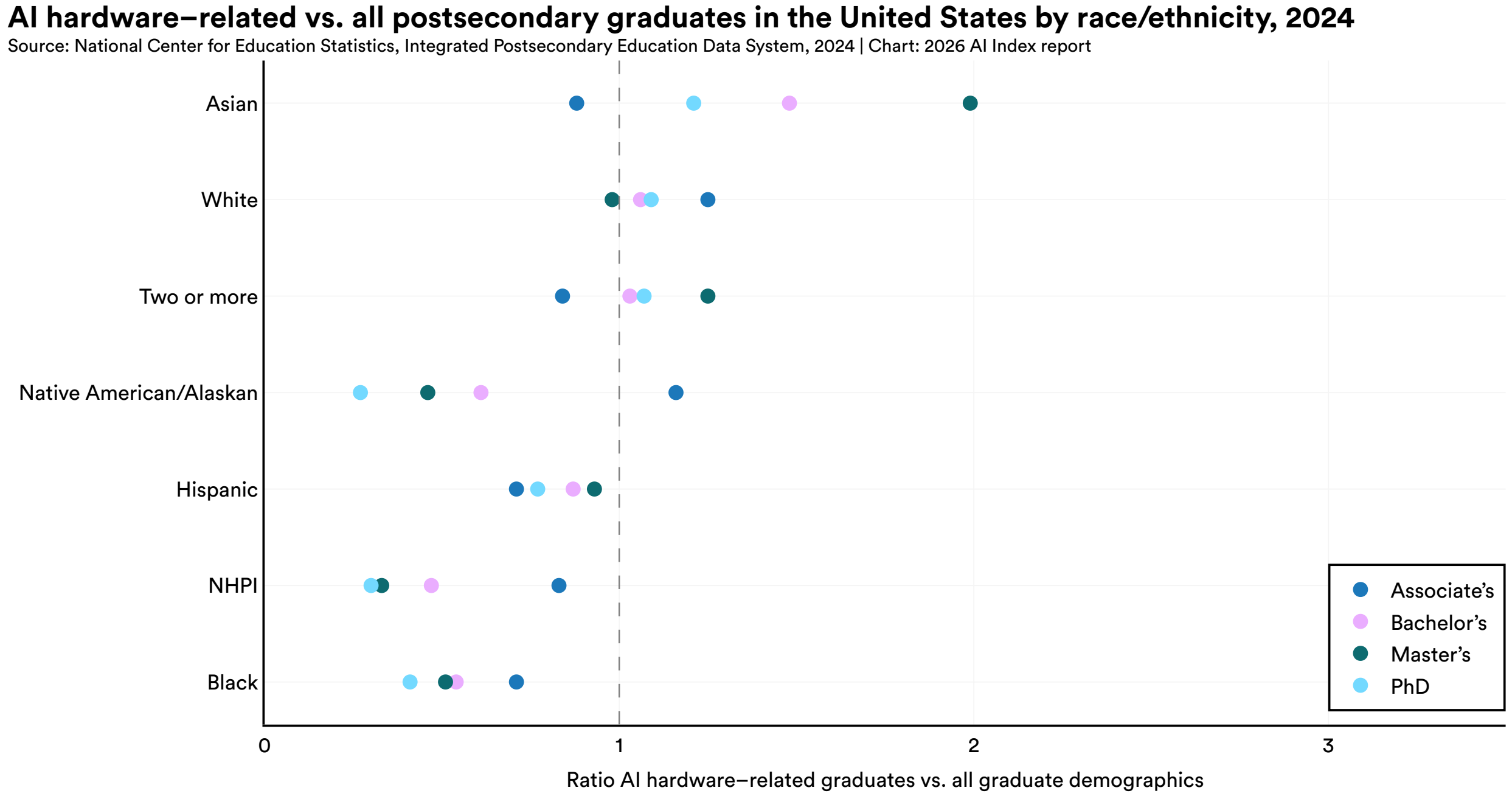
AI hardware–related vs. all postsecondary graduates in the United States by race/ethnicity, 2024
Source: National Center for Education Statistics, Integrated Postsecondary Education Data System, 2024 | Chart: 2026 AI Index report

Figure 7.2.4

The majority of AI-related graduate students are non–United States residents, a pattern consistent with previous years' analyses of CS degrees (Figure 7.2.5). This is especially true in AI software–related master's degrees, where 67% of graduates are nonresidents. However, due to the federal government revoking student visas and discouraging international student enrollment, further declines in the number of nonresident graduates are expected in the coming years.

**AI-related postsecondary graduates in the United States by residency, 2024**

Source: National Center for Education Statistics, Integrated Postsecondary Education Data System, 2024 | Chart: 2026 AI Index report

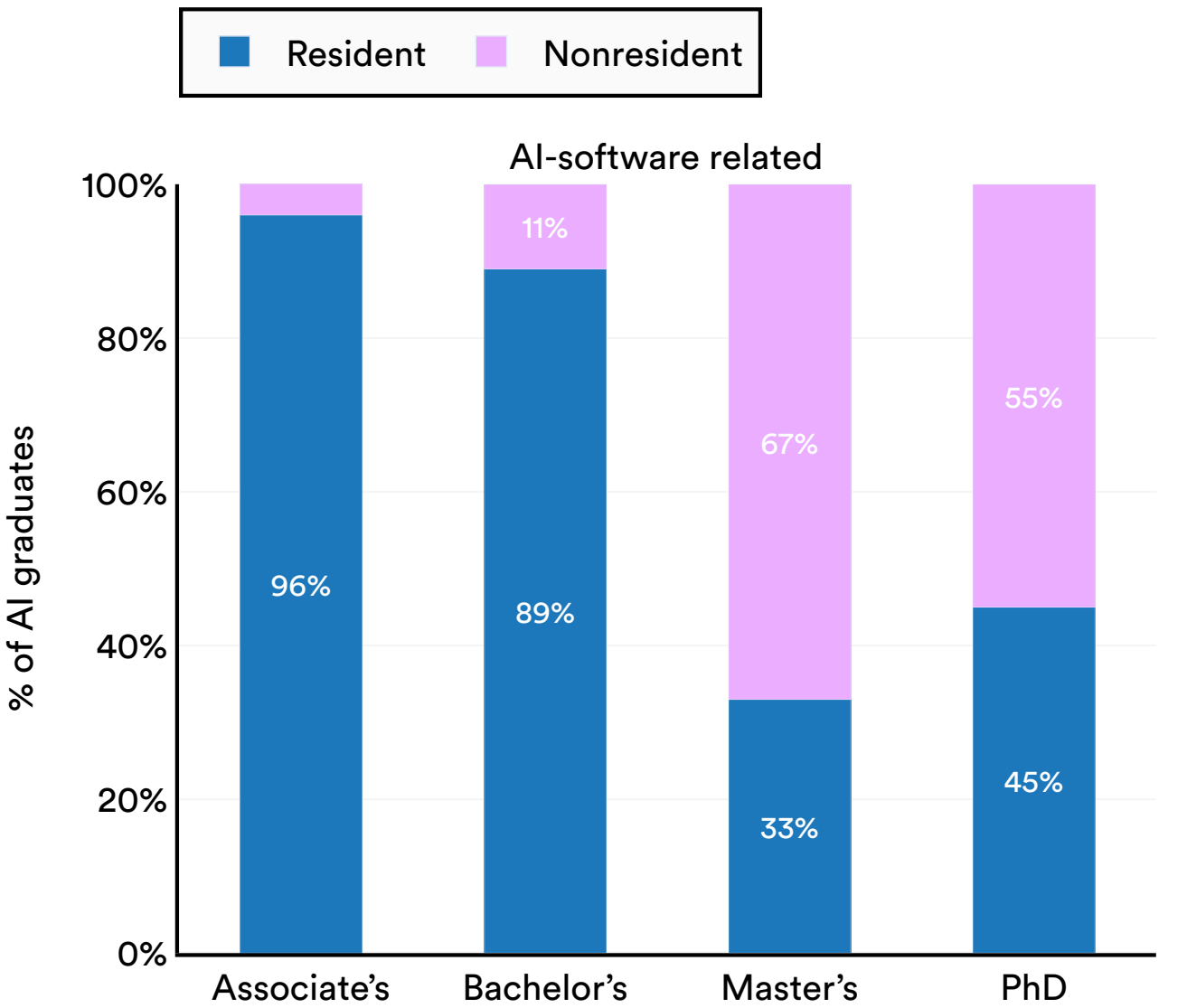

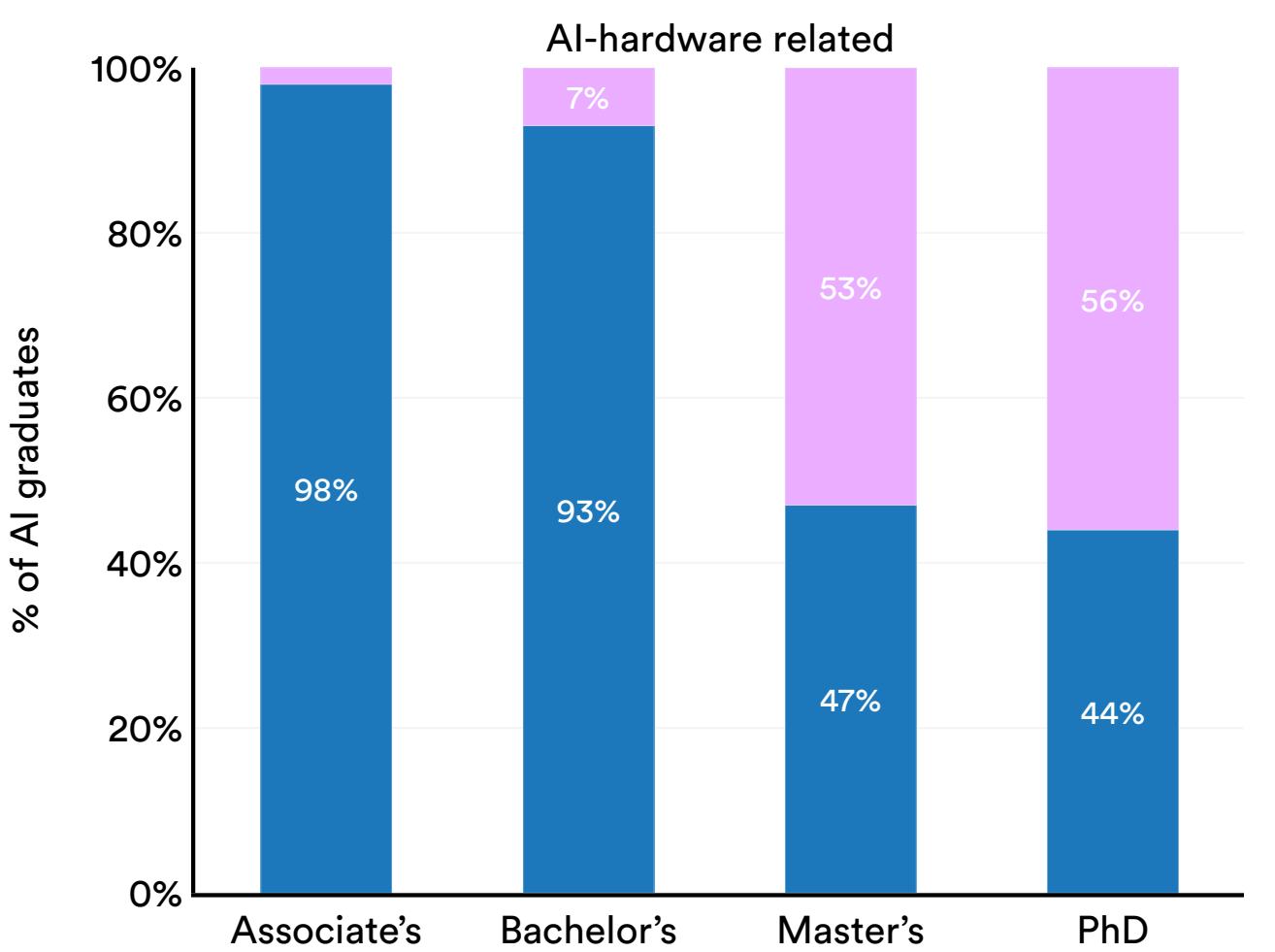

Figure 7.2.5

A range of institutions produce the highest number of graduates in AI-related fields[5], including both public and private universities (Figure 7.2.6). The Georgia Institute of Technology is the only school in the top 10 across all levels for both AI software and hardware–related degrees. Other universities that appear at least once on both lists include University of California, Berkeley (4 mentions); University of Illinois, Urbana-Champaign (4); University of Michigan, Ann Arbor (4); Pennsylvania State University (3); Northeastern University (2); Carnegie Mellon University (2); Massachusetts Institute of Technology (2); and Stanford University (2).As more institutions add AI-specific majors at the undergraduate level, these rankings are likely to shift in coming years.

5 Some institutions classify AI-related graduates under broader program categories rather than AI-specific ones, which may result in undercounting at schools where AI coursework is housed within general computer science or engineering programs.

## Top postsecondary institutions graduating students in AI-related degrees, 2024

Source: National Center for Education Statistics, Integrated Postsecondary Education Data System, 2024 | Chart: 2026 AI Index report

### AI-software related

| | Bachelor's | Master's | PhD |
|---|---|---|---|
| 1 | University of Maryland Global Campus: 2,350 | Georgia Institute of Technology: 3,394 | Georgia Institute of Technology: 155 |
| 2 | Western Governors University: 2,029 | University of Texas, Dallas: 2,555 | Massachusetts Institute of Technology: 154 |
| 3 | University of California, Berkeley: 1,983 | Columbia University: 2,481 | University of Illinois, Urbana-Champaign: 128 |
| 4 | University of Maryland, College Park: 1,877 | University of North Texas: 2,454 | Carnegie Mellon University: 128 |
| 5 | Southern New Hampshire University: 1,733 | Trine University: 2,114 | University of Michigan, Ann Arbor: 106 |
| 6 | University of Michigan, Ann Arbor: 1,723 | University of Illinois, Urbana-Champaign: 1,905 | University of California, Los Angeles: 105 |
| 7 | Rutgers University: 1,329 | Northeastern University: 1,720 | University of California, San Diego: 104 |
| 8 | Georgia Institute of Technology: 1,302 | University of Southern California: 1,676 | University of Washington: 104 |
| 9 | University of California, San Diego: 1,204 | University of Texas, Arlington: 1,430 | University of California, Berkeley: 103 |
| 10 | Pennsylvania State University: 1,181 | Boston University: 1,360 | Stanford University: 95 |

### AI-hardware related

| | Bachelor's | Master's | PhD |
|---|---|---|---|
| 1 | Purdue University: 2,083 | Northeastern University: 1,565 | Purdue University: 329 |
| 2 | Texas A&M University: 1,857 | Arizona State University: 1,090 | University of Michigan, Ann Arbor: 242 |
| 3 | Georgia Institute of Technology: 1,400 | New York University: 959 | Stanford University: 202 |
| 4 | University of Illinois, Urbana-Champaign: 1,391 | University of Michigan, Ann Arbor: 941 | Georgia Institute of Technology: 198 |
| 5 | Ohio State University: 1,341 | Johns Hopkins University: 939 | Massachusetts Institute of Technology: 191 |
| 6 | Virginia Polytechnic Institute and State University: 1,039 | Carnegie Mellon University: 899 | Texas A&M University: 189 |
| 7 | Pennsylvania State University: 1,002 | Georgia Institute of Technology: 839 | University of Illinois, Urbana-Champaign: 184 |
| 8 | University of California, Berkeley: 970 | Texas A&M University: 740 | University of Texas, Austin: 180 |
| 9 | Iowa State University: 942 | San Jose State University: 737 | University of California, Berkeley: 176 |
| 10 | North Carolina State University: 914 | Purdue University: 731 | Pennsylvania State University: 160 |

Figure 7.2.6

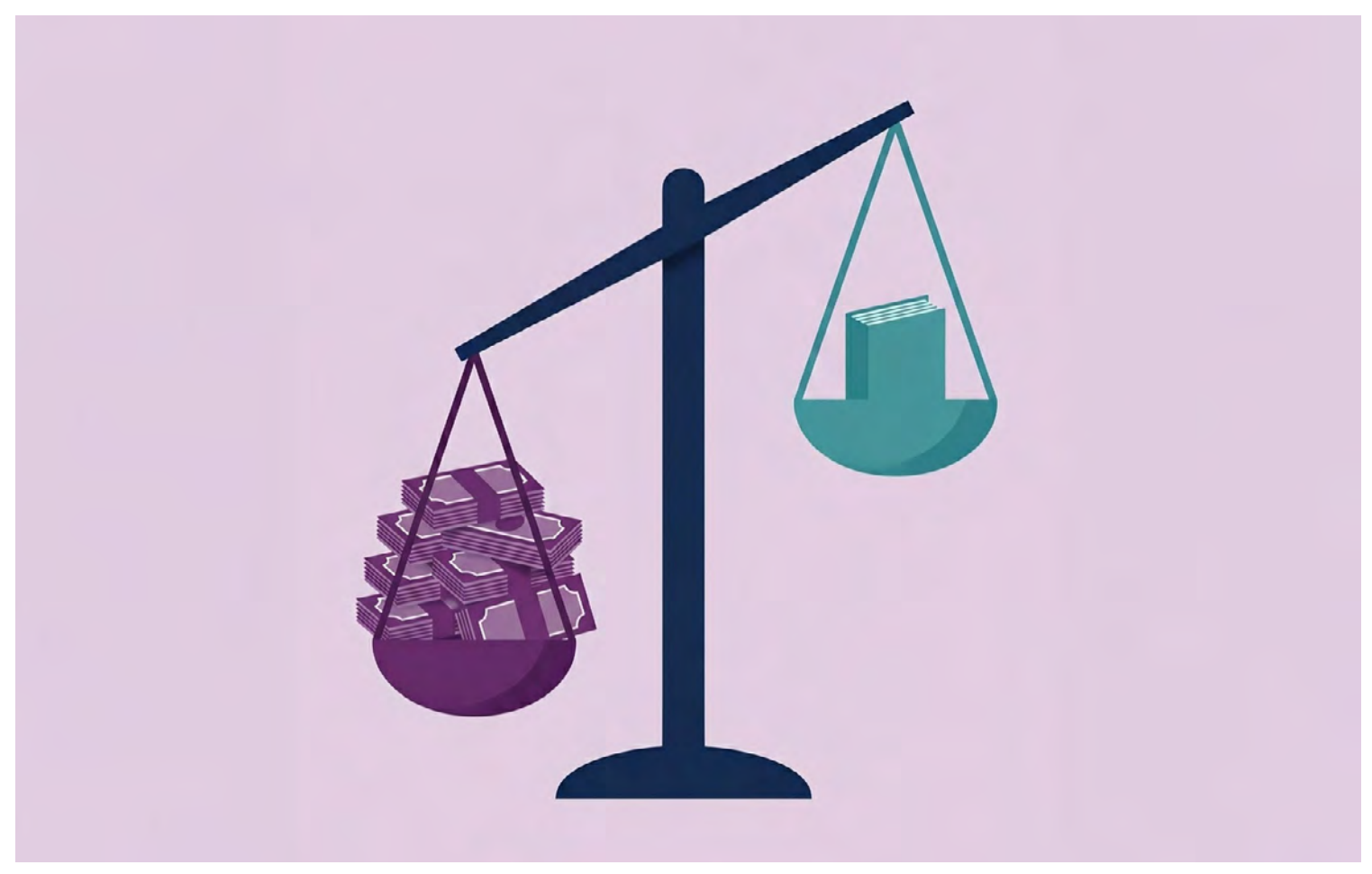

AI PhD graduates continue to choose industry jobs and lucrative salaries more often than academic jobs, with 65% going into industry after graduation (Figure 7.2.7). This percentage has declined in the last few years, down from a peak of 77% in 2022. At the same time, the share of academic jobs has increased, nearly doubling since 2022 (Figure 7.2.8). This challenges the narrative that academia is experiencing an exodus of experts or a "brain drain." The percentage of AI PhD graduates entering government jobs has gradually increased to 2% from a low of 0.7% in 2021.

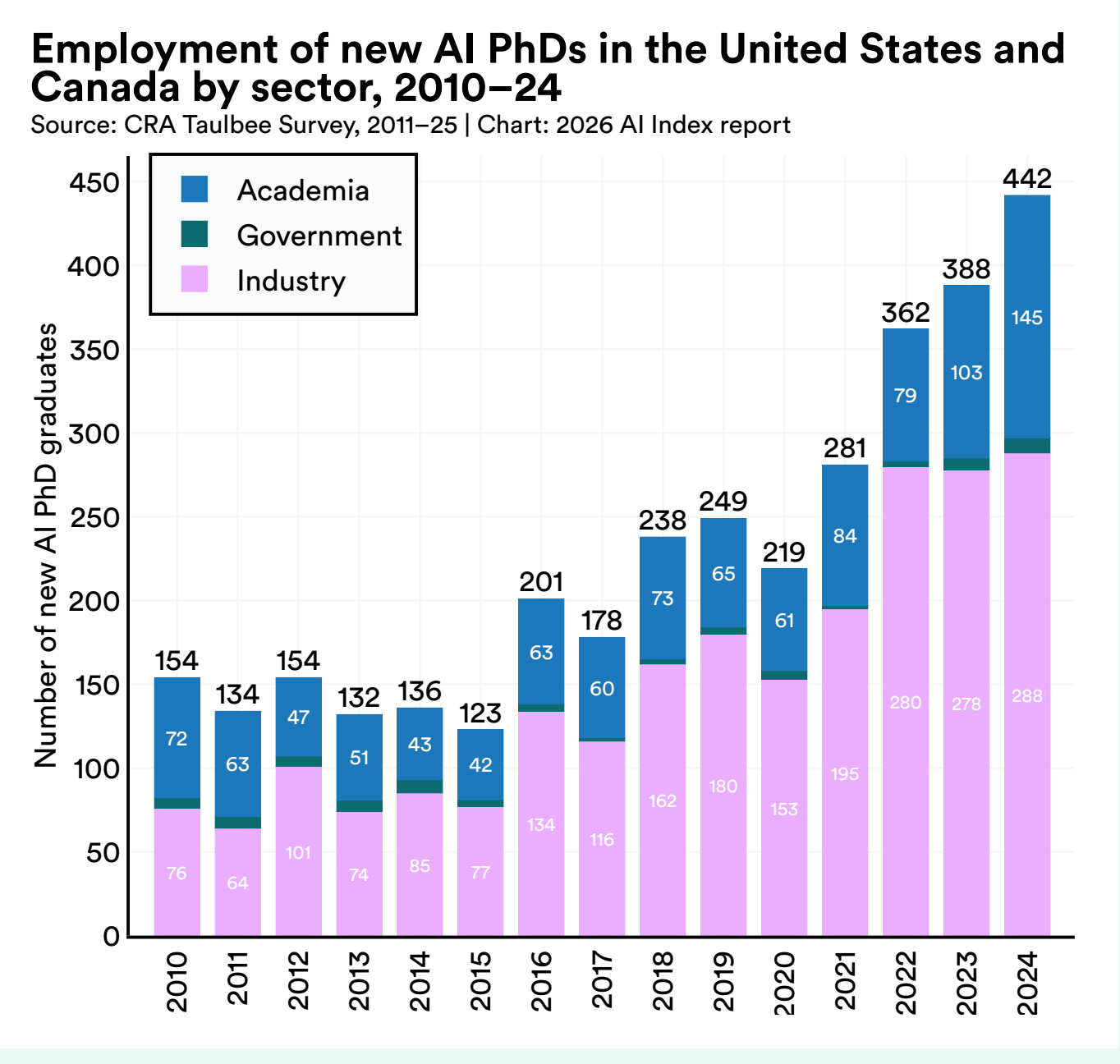

Figure 7.2.7

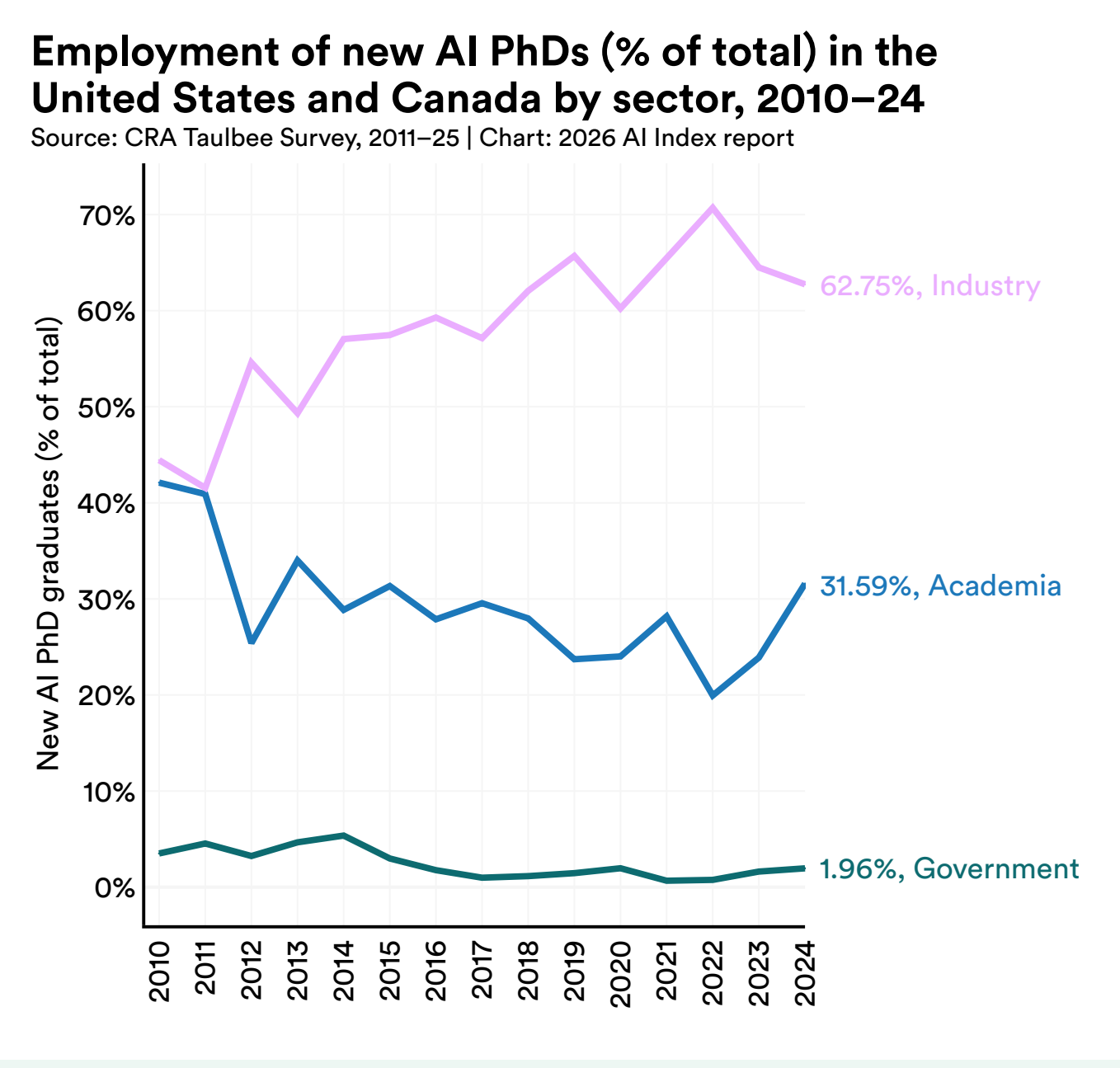

Figure 7.2.8[6]

## Global ICT Graduates

No single dataset provides a fully standardized accounting of AI or CS postsecondary education across all countries. The Organization for Economic Cooperation and Development (OECD) has compiled data covering its member countries and several non-OECD nations.[7] The International Standard Classification of Education provides the framework the OECD uses to compare education statistics across countries. Information and communication technologies, or ICT, includes such areas of study as "informatics, information, and communication technologies, or CS. These subjects cover a wide range of topics related to the new technologies used for the processing and transmission of digital information, including computers, computerized networks (including the Internet), microelectronics, multimedia, software, and programming."

The United States remains a global leader in ICT-related fields, producing more graduates at the associate's, bachelor's, master's, and PhD levels than any other country in the sample (Figures 7.2.9 to 7.2.12). At most levels, other countries had faster year-over-year growth than the United States. At the associate's level (denoted in international charts as short-cycle tertiary), Turkey increased its graduates by 27%; at the bachelor's level, both Brazil and Turkey increased their graduates by 30%; and at the PhD level, Mexico increased its graduates by 76%. The exception is at the master's level, where the United States increased its graduates by 55% (though, as noted earlier, many master's graduates in the United States are not U.S. residents).

6 The sums in Figure 7.2.8 do not add up to 100%, as there is a subset of new AI PhDs each year who become self-employed, unemployed, or report an "other" employment status in the CRA survey. These students are not included in the chart.

7 While this dataset provides insights across some country lines, it omits a number of countries likely to have large numbers of ICT graduates. The exclusion of India, China, and countries in Africa highlights the need for global standardized data collection to ensure the inclusion of countries that have made significant investments in computing education and make up a significant proportion of the global population. There is also a notable lag in collecting and reporting global data on education; as a result, the most recent year for which data is available is 2023. Data for each country includes any students who have graduated from an institution in that country, regardless of their nationality.

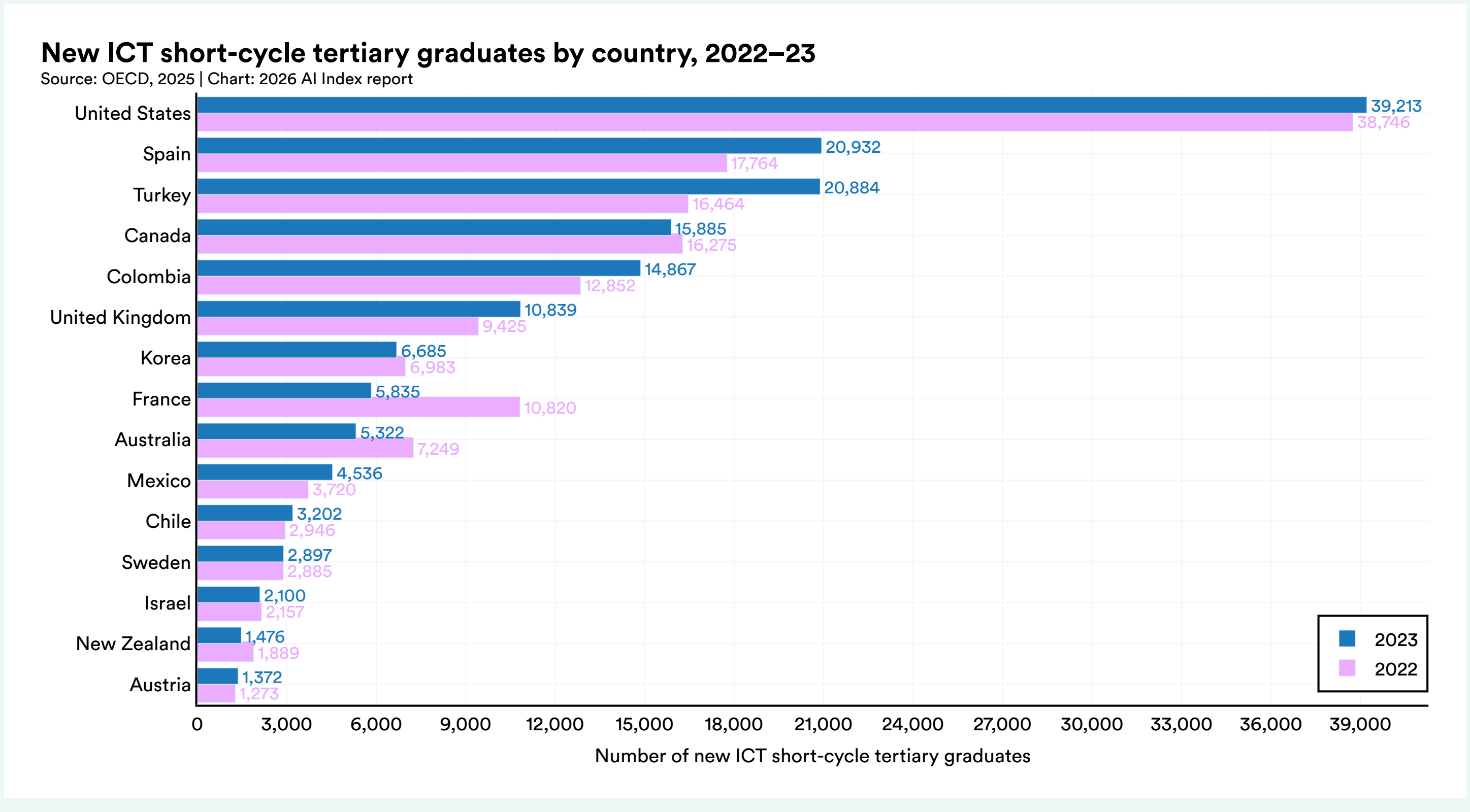

Figure 7.2.9

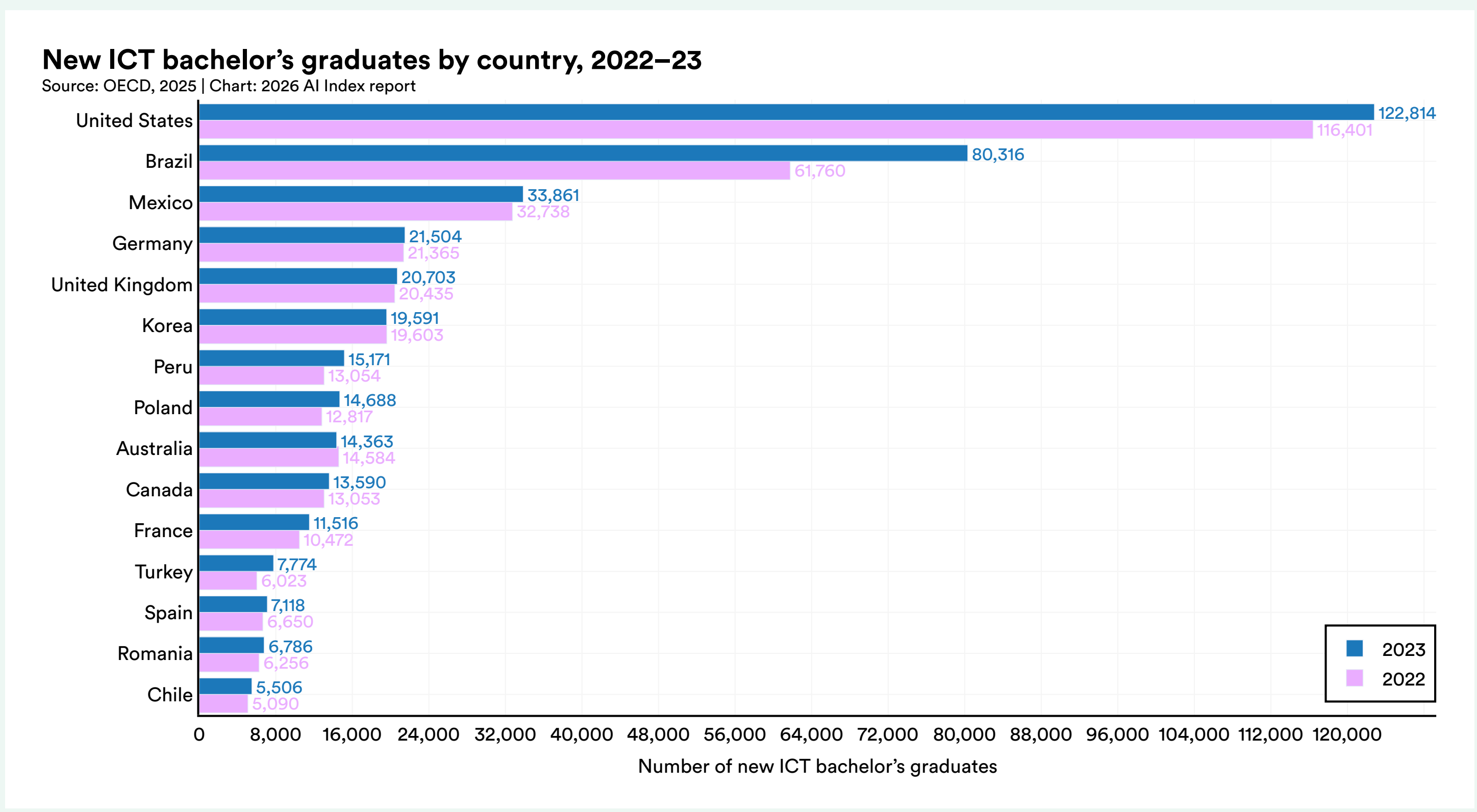

Figure 7.2.10

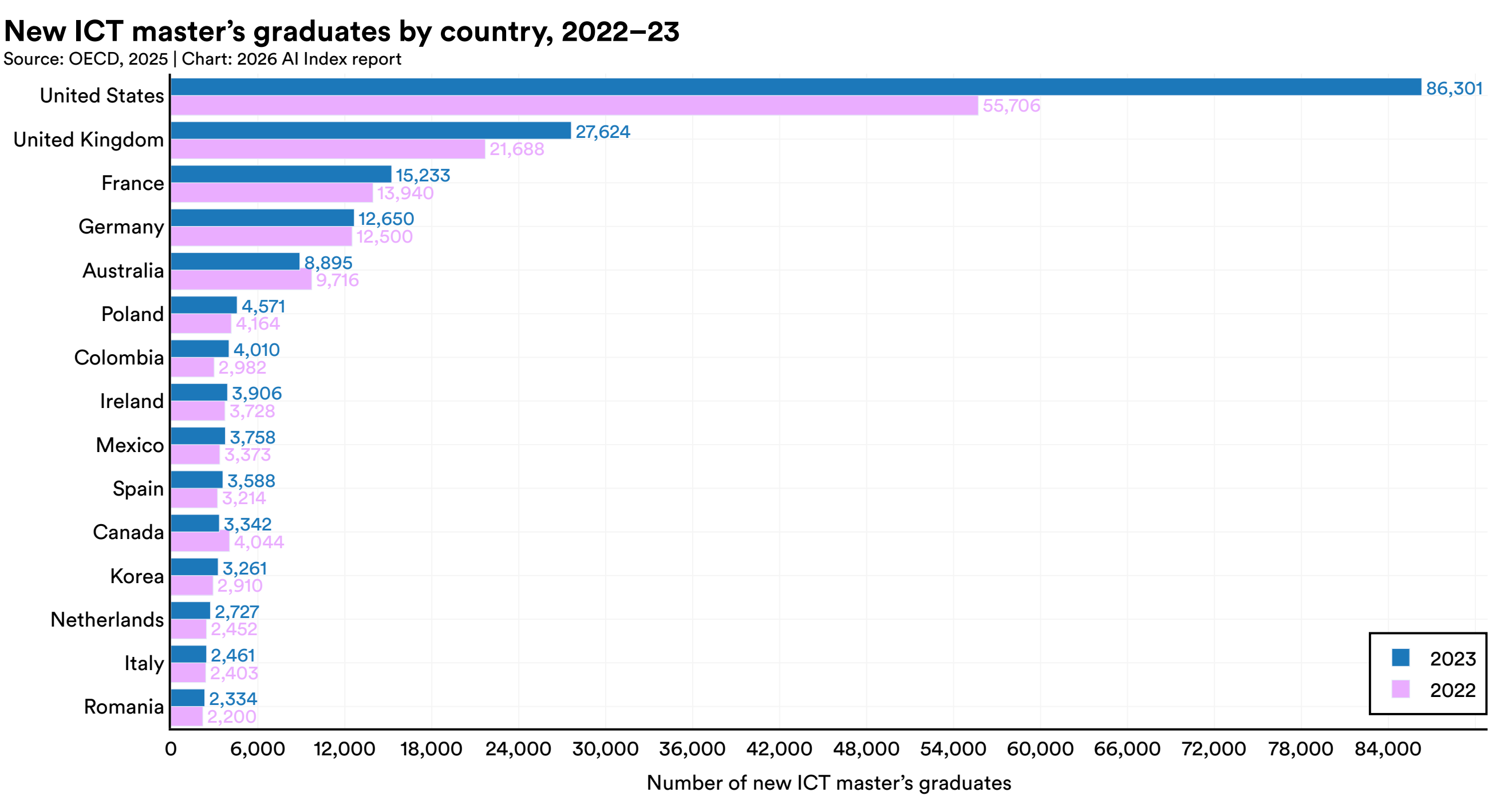

Figure 7.2.11

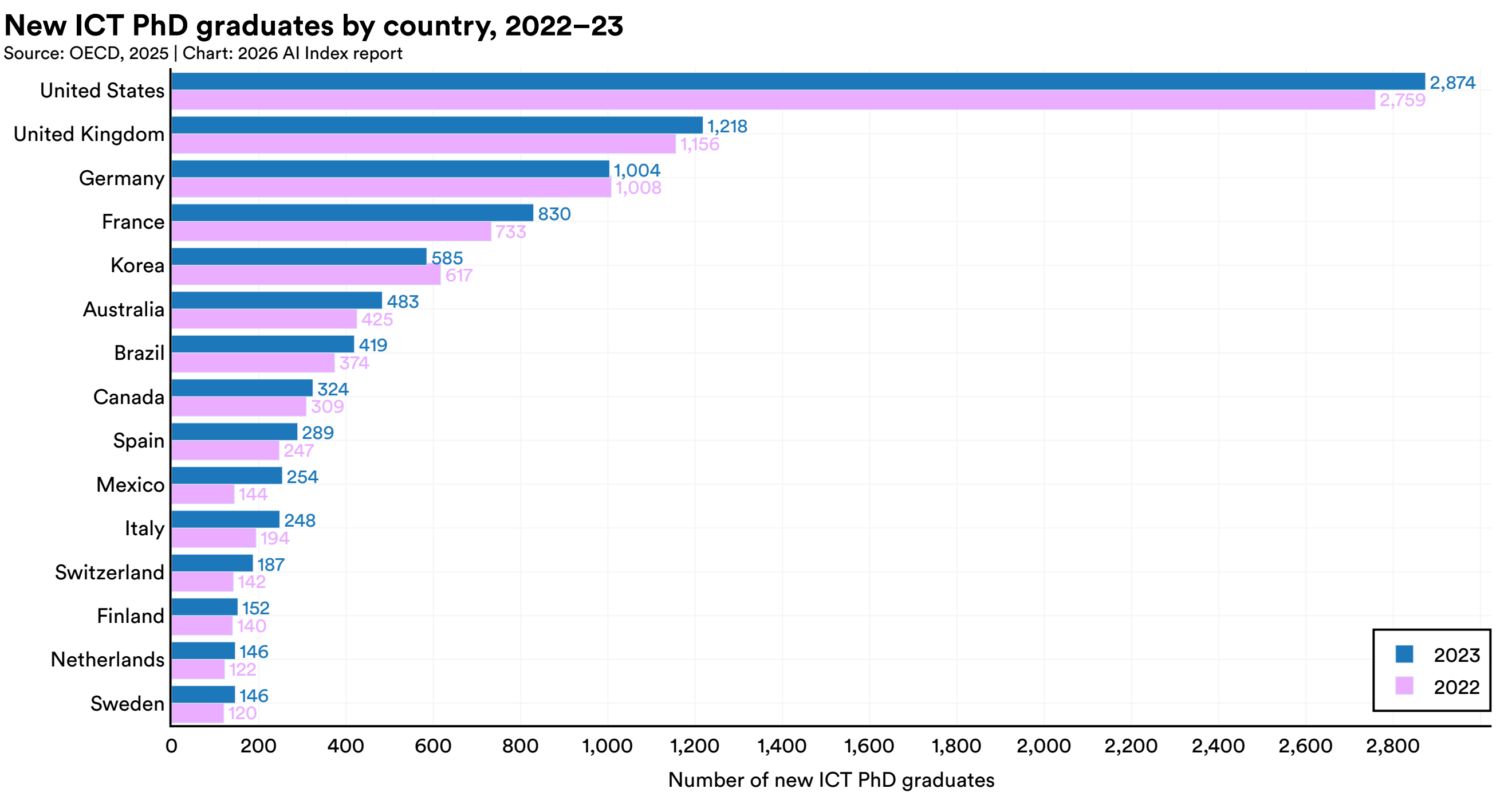

Figure 7.2.12

Gender parity among ICT graduates remains uneven across countries and degree levels (Figure 7.2.13). On average, women account for 20% of associate's graduates, 22% of bachelor's graduates, 29% of master's graduates, and 29% of PhD graduates. While the share of associate's graduates declined year over year, the share of PhD graduates increased 4 percentage points. Women comprised at least half of ICT graduates in Peru at the associate's level, and in Costa Rica and Latvia at the PhD level. Turkey, which had reported gender parity at all levels in the prior year, saw its shares move closer to the global averages in 2023. The gender composition of these graduates has been consistent year over year, similar to the pattern among AI authors and inventors documented in Chapter 1, where the gender ratio has shown little change over the past 15 years.

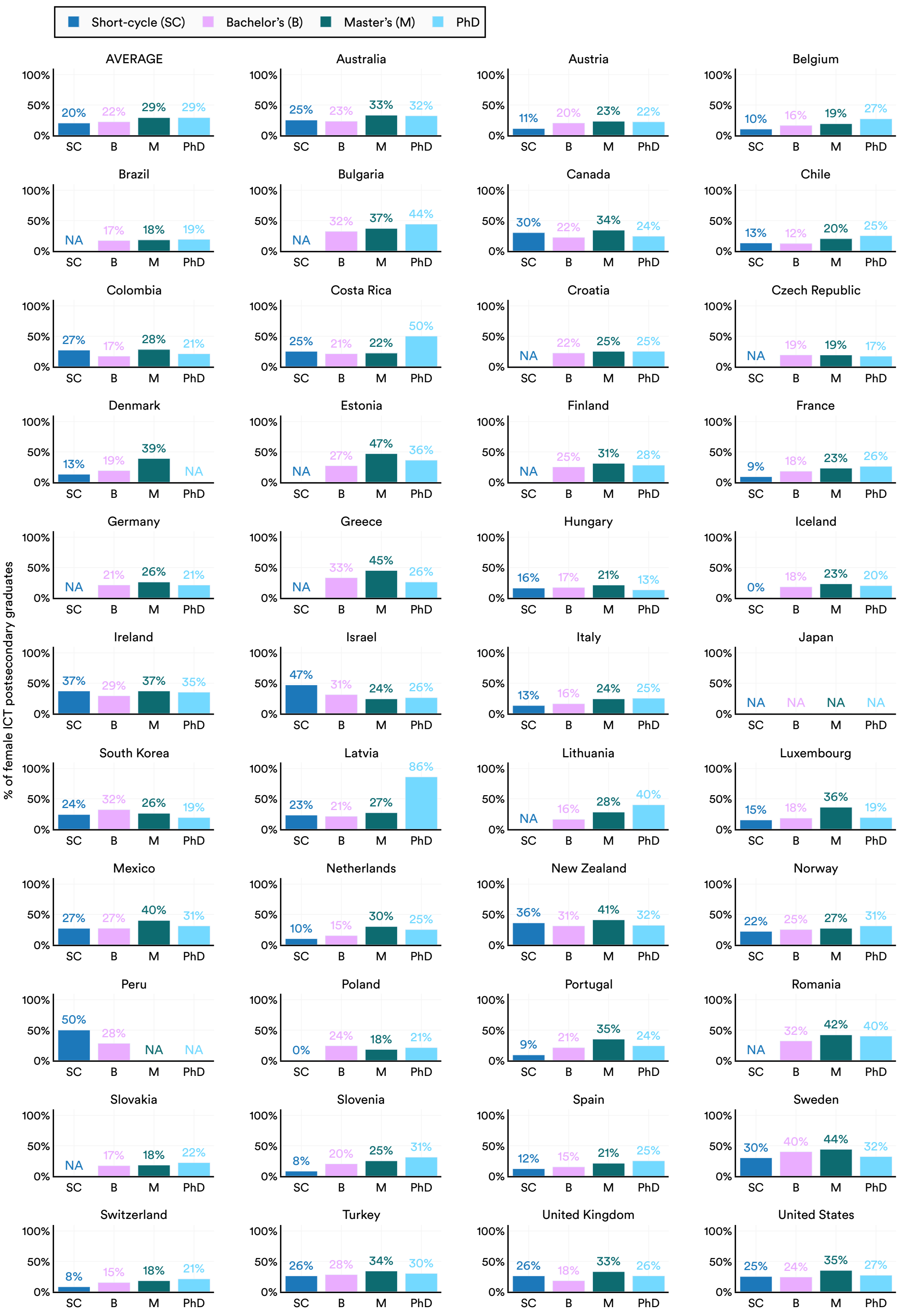
Percentage of new ICT postsecondary graduates who are female by country, 2022–23
Source: OECD, 2025 | Chart: 2026 AI Index report

Figure 7.2.13

## CS, CE, and Information Faculty

In 2024–25, there were over 6,600 CS, CE (computer engineering), and information faculty in the United States and Canada (Figure 7.2.14). Nearly two-thirds of them filled tenure-track positions. The Computing Research Association (CRA) projections suggest the number of faculty will increase over the next two academic years, with the most growth in postdoctoral positions.[8]

Hispanic/Latino, Black, and Indigenous people are underrepresented in faculty positions, as are all women except Asian women (Figure 7.2.15).[9] Asian and Native Hawaiian/Pacific Islander men are overrepresented among faculty.

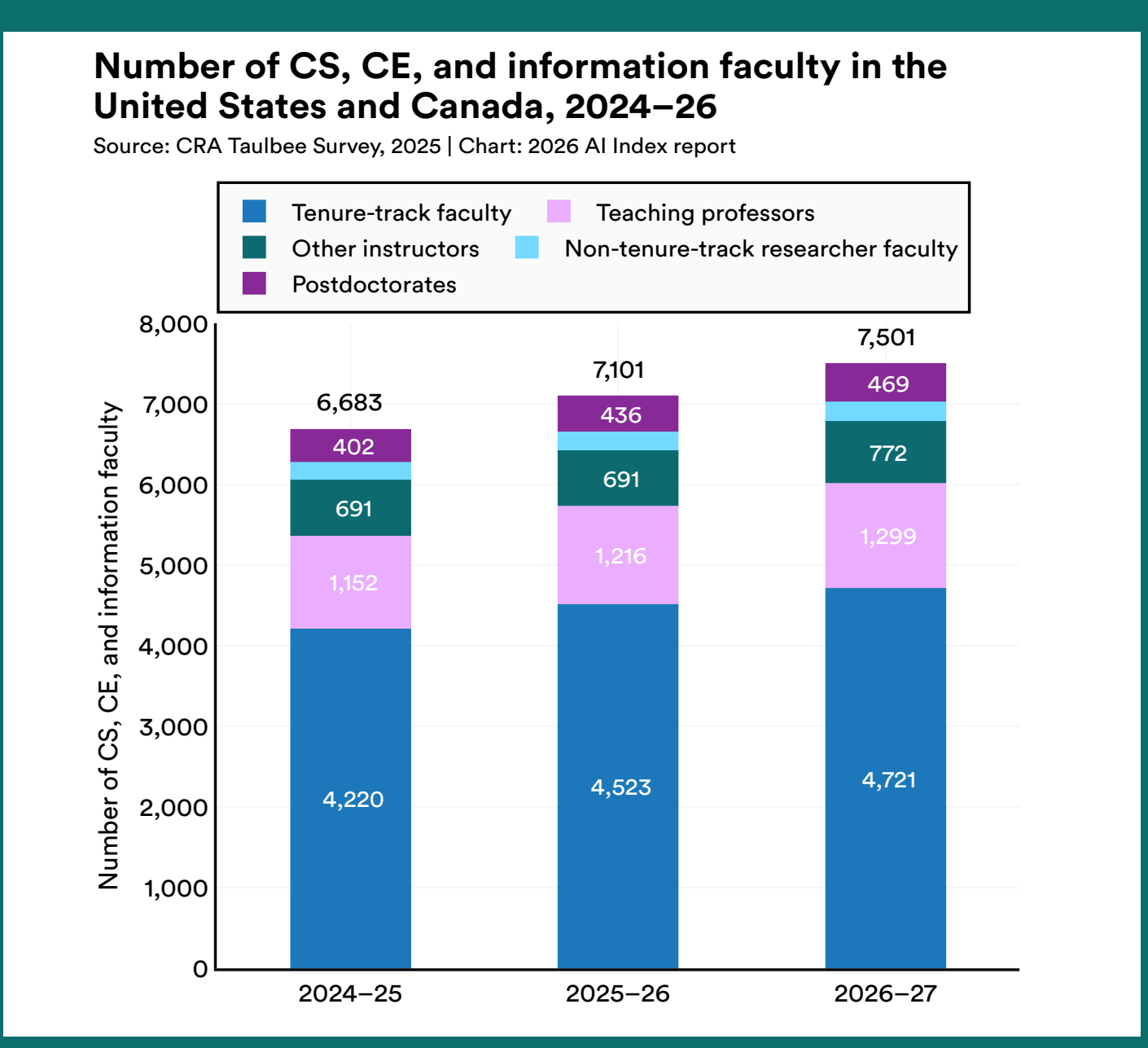

Figure 7.2.14

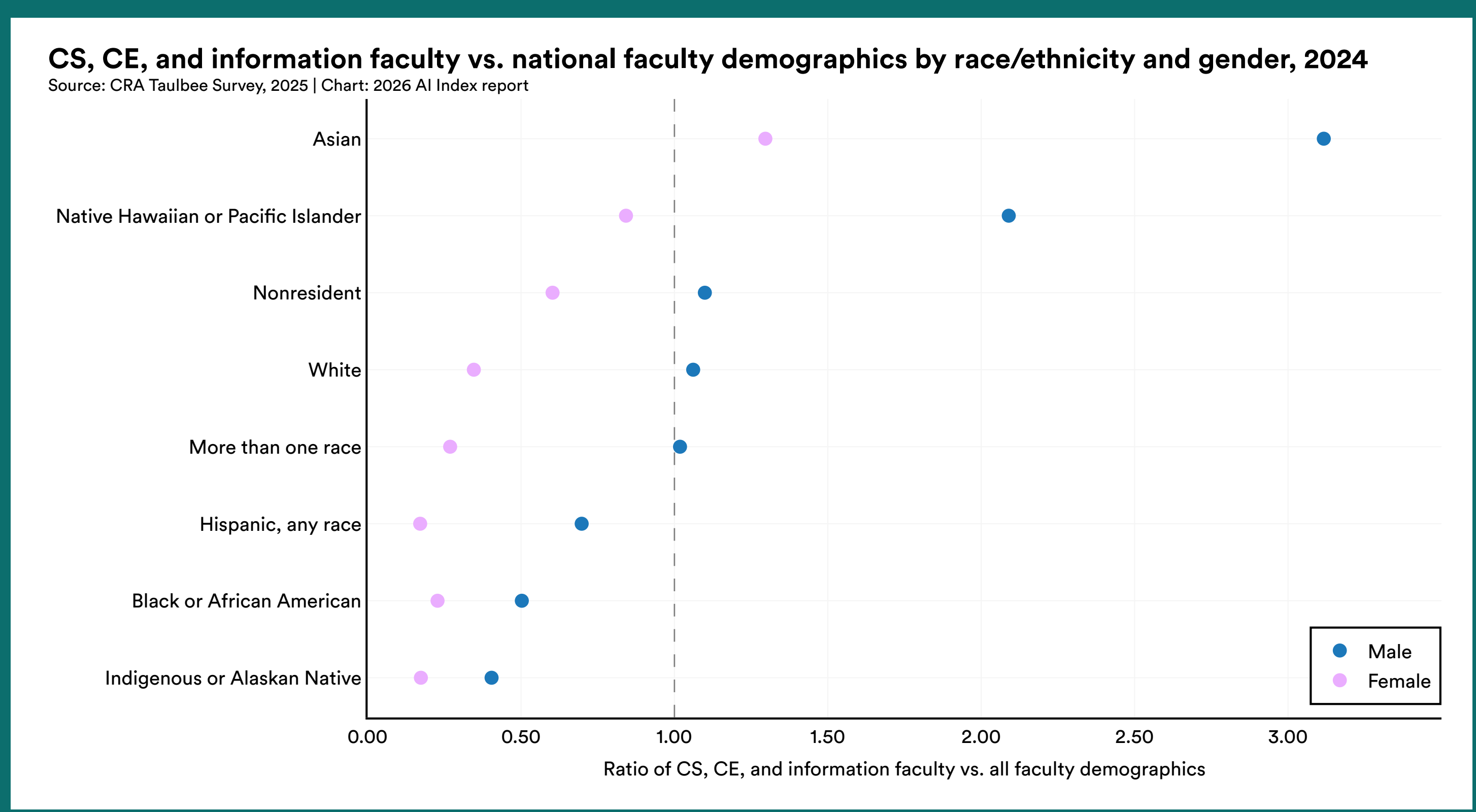

Figure 7.2.15

8 Due to changes in CRA's methodology, these figures are not directly comparable to faculty counts published in previous editions of the AI Index.

9 National faculty demographics are from the United States (NCES), whereas the faculty data also includes Canada. Postdoctorates were not included in the comparison data because they were absent from the national faculty demographics.

## Student Use of AI Tools

While this chapter's focus is AI education, how students are learning is also changing. Examining AI in education, or the use of AI to complete teaching and learning tasks, helps provide a more complete picture of the impact of AI on the field of education. In Chegg's 2025 survey of university students from 15 countries, 80% said they have used generative AI to support their learning. That is double the share reported in 2023, when only 40% of students reported having used generative AI for school. Generative AI usage varies widely by country, with 95% of Indonesian students saying they have used it, compared to 67% in the United States and the United Kingdom (Figure 7.2.16). Students who use generative AI for school report doing so frequently: 56% input a question at least once a day. For those students who do not use AI tools for school, the top reasons include accusations of academic misconduct (45%), content accuracy (38%), and school policies restricting AI use (33%).

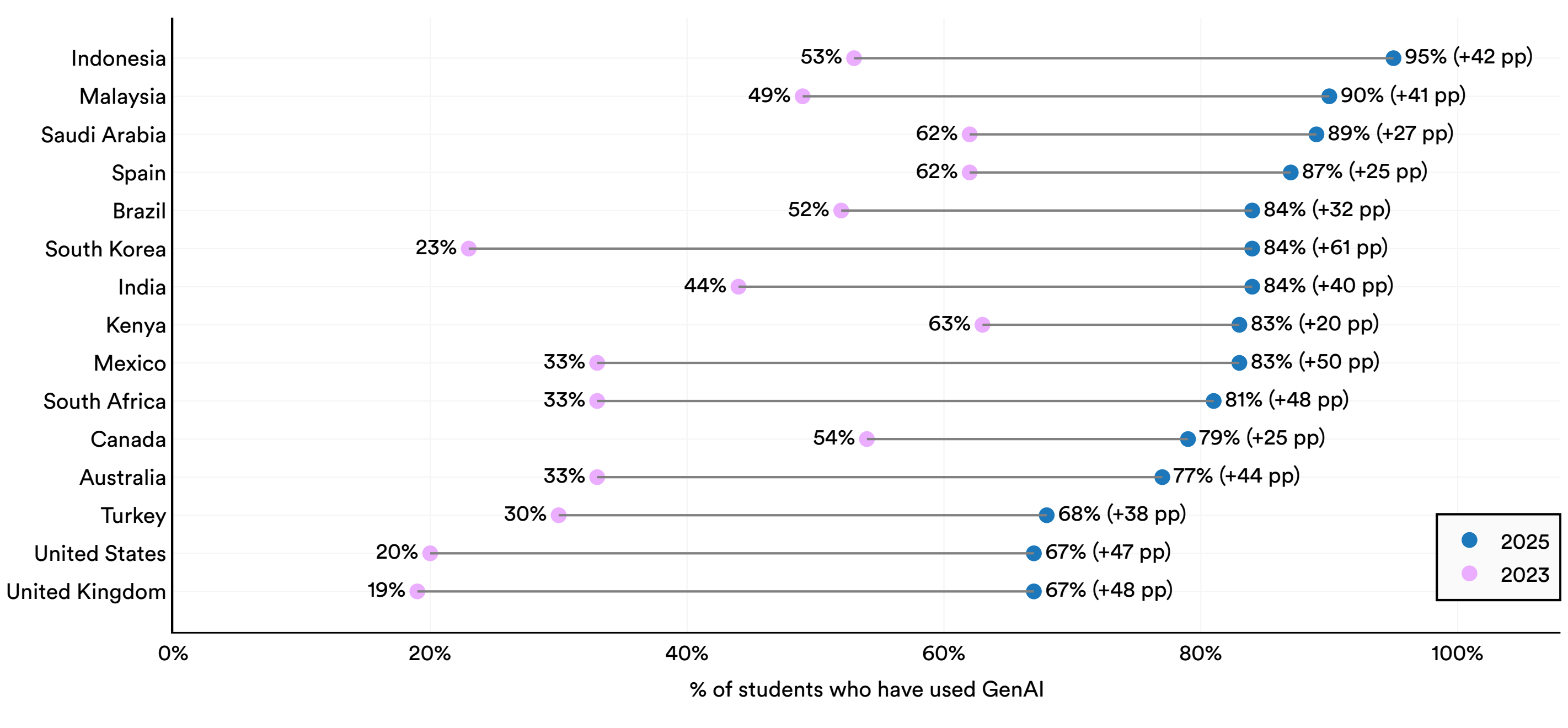
University students (% of total) who have used GenAI to support their university studies, 2023 vs. 2025
Source: Chegg Global Student Survey, 2023 and 2025 | Chart: 2026 AI Index report

Figure 7.2.16

University students report using AI in similar ways to high school students, including researching, brainstorming/generating ideas, and editing essays. One notable difference is that university students are more likely than high school students to use AI to understand a concept (56% vs. 41%); in fact, understanding a subject is the top use of generative AI tools among university students (Figure 7.2.17).

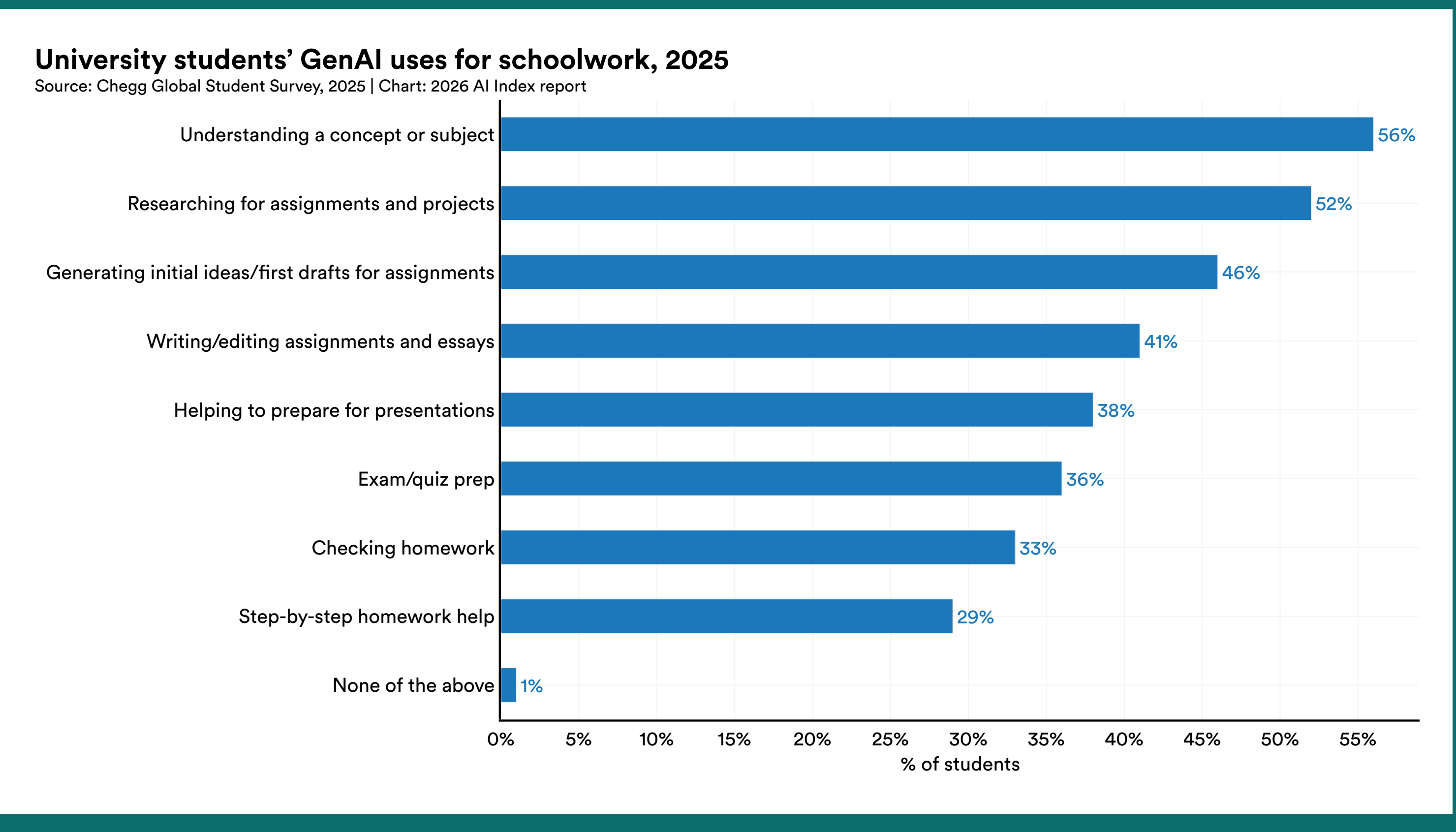

Figure 7.2.17

Anthropic analyzed how students use Claude, its generative AI tool, and found that most students use it for higher-order thinking skills[10], such as creating (39.8%) and analyzing (30.2%), rather than lower-order thinking skills, like applying (10.9%) and understanding (10.0%). These uses may suggest that students are relying on generative AI tools for important cognitive skills rather than developing them independently; indeed, another survey found that 55% of U.S. college students believe using generative AI tools has had a mixed effect on their critical thinking skills.

Nonetheless, university students speak positively about using AI tools in education. A survey of over 73,000 California State University students showed that 64% of them agree that AI has positively affected their learning experience. The students in the Chegg survey listed several benefits from using AI tools. Half of respondents reported increased understanding of topics, 49% report improved ability to finish assignments, and 41% report improved organization. They also reported that AI makes their learning process more efficient, with 55% of university students saying AI helps them learn faster, and 41% saying it frees up more of their time.

University students are still concerned about accusations of cheating and appropriate uses of AI; in response, more universities have implemented AI use policies. A faculty survey noted that 48% of institutions now have policies governing acceptable uses of generative AI, an increase of 9 percentage points since 2025. In the U.K., 80% of students think their university has a clear policy on generative AI use in assessments, which is a 16 percentage point improvement from the previous year.

10 These categories refer to levels of cognitive complexity as defined by Bloom's Taxonomy, a widely used framework in education that classifies skills from lower-order (remembering, understanding, applying) to higher-order (analyzing, evaluating, creating).

# 7.3 K–12 CS and AI Education

Data on AI-specific education at the K–12 level remains limited. This section tracks developments in CS education as a proxy, with attention to the emerging state and federal policies that are beginning to address AI directly.

## United States

Code.org's annual State of CS Education Report, which tracks CS education access, participation, and state policy, was expanded in 2025 to include an analysis of states' AI education policies as of December 1, 2025. With only four states emphasizing AI in their CS standards, adoption of AI education remains limited.

### Foundational Computer Science

Between the 2017–18 and 2023–24 academic years, the percentage of U.S. high schools offering CS increased from 35% to 60%. The national average has held steady since 2023–24, with the same 60% of high schools offering foundational CS classes in 2024–25. There is, however, growth in many states and regression in a few (Figure 7.3.1). In states where offerings declined, budget cuts or reallocation toward other priorities, including literacy and student crisis support, may be contributing factors.

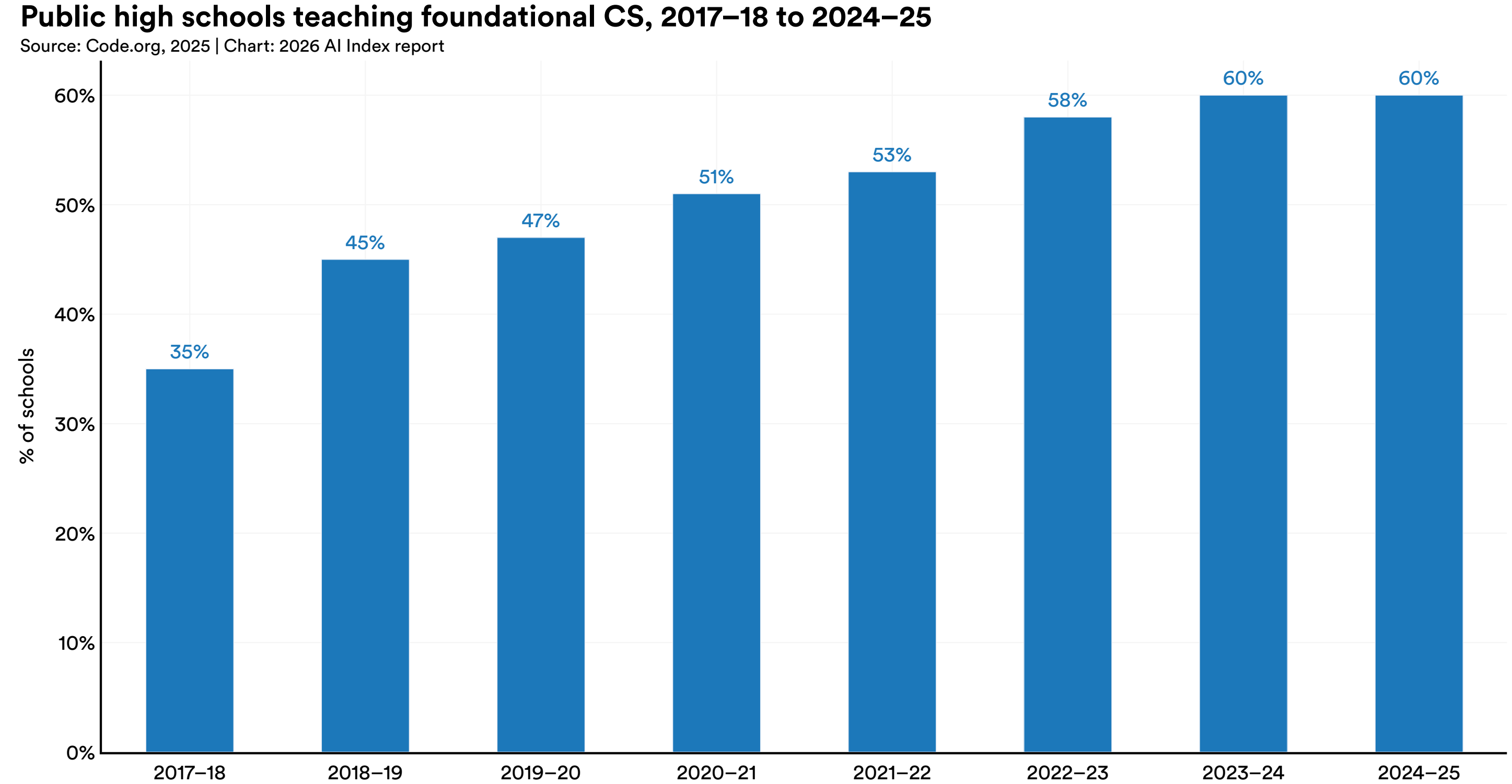

Figure 7.3.1

Out of 22 states reporting data for both 2023–24 and 2024–25, five states (AR, DE, LA, MD, SC) maintained the percentage of schools offering CS classes; three of them were already at or nearing universal access (AR and MD reporting 100% and SC reporting 92%) (Figures 7.3.2 and 7.3.3). Nine states and the District of Columbia (DC) reported some growth in the percentage of high schools offering CS. Of those, six (GA, IA, MS, NC, UT, WA) reported minimal growth (1%–5%). Two states (CA, MT) plus DC reported modest growth (6%–10%). One state, TN, reported significant growth, increasing the percentage of high schools offering CS classes by 22% (from 61% to 83%). This increase may reflect a 2022 policy requiring all K–12 students to have access to CS; districts were required to implement the state's K–12 CS standards and a one CS credit graduation requirement starting with incoming freshmen in the 2024–25 academic year. A few states reported small decreases in the percentage of schools offering CS, with only one (WY) reporting a 10% decrease (from 74% to 64%).

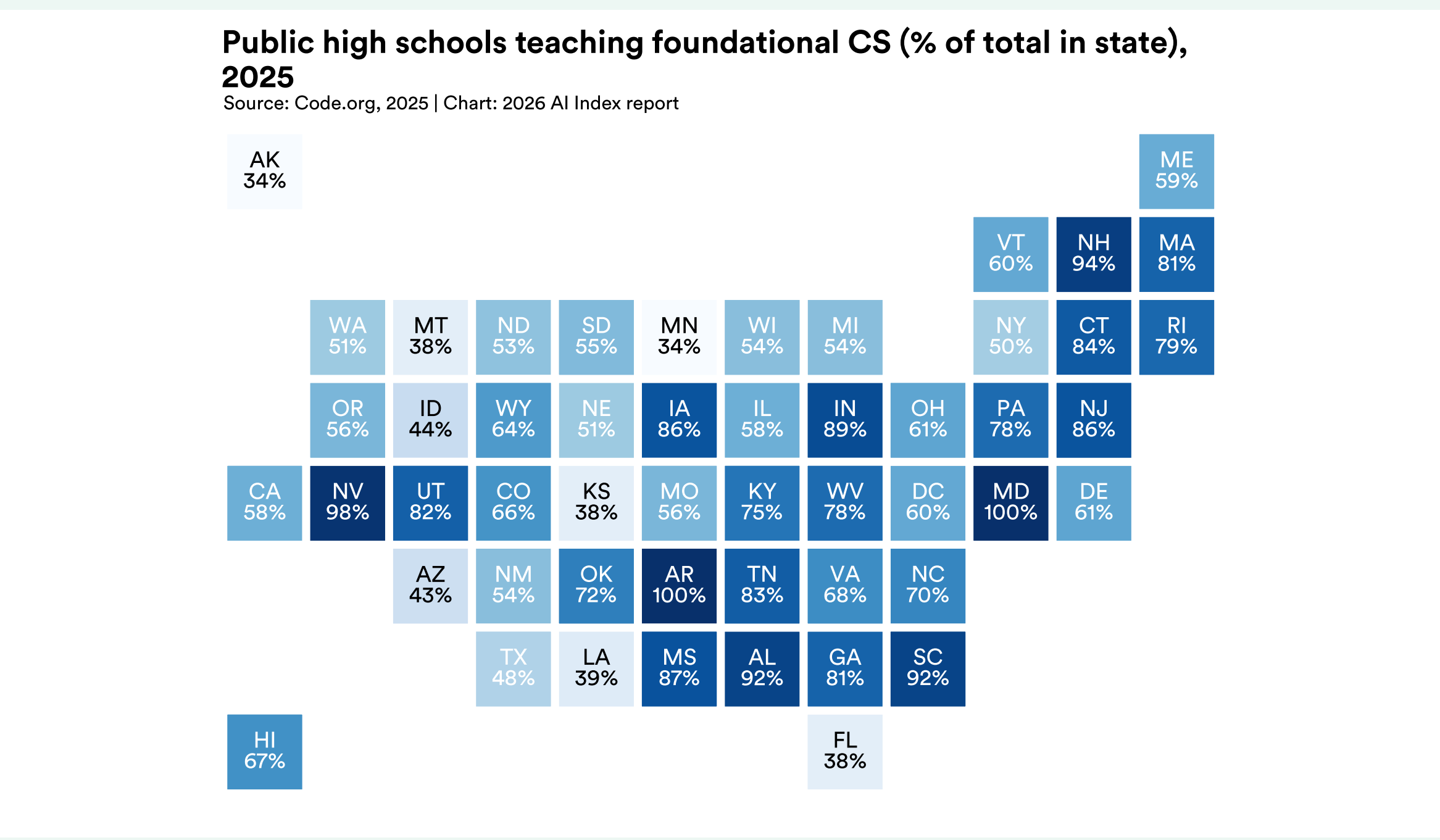

Figure 7.3.2

Change in public high schools teaching foundational CS, 2024 vs. 2025
Source: Code.org, 2025 | Chart: 2026 AI Index report

AK --
ME -4%
VT --
NH -1%
MA --
WA 1%
MT 7%
ND --
SD --
MN --
WI --
MI --
NY --
CT --
RI --
OR --
ID -2%
WY -10%
NE --
IA 2%
IL --
IN --
OH --
PA --
NJ --
CA 6%
NV --
UT 1%
CO --
KS --
MO --
KY -1%
WV --
DC 7%
MD 0%
DE 0%
AZ --
NM --
OK --
AR 0%
TN 22%
VA --
NC 1%
TX --
LA 0%
MS 2%
AL -2%
GA 3%
SC 0%
HI -5%
FL --

Figure 7.3.3

As a result of disparities in education funding, CS education access varies by school size, geographic area, socioeconomic status, and student race/ethnicity (Figures 7.3.4 to 7.3.7). In 2025, 91% of large high schools, 77% of medium-sized high schools, and only 44% of small high schools offered foundational CS courses. As public school closures continue in some areas, school size and resource distribution may shift as nearby schools absorb displaced students. Whether funding follows those students, and how consolidation affects CS access, will be important to monitor.

Title I[11] schools (60%) are slightly less likely to offer CS than non–Title I schools (65%). Rural (57%) and urban (59%) high schools are less likely than suburban (71%) high schools to offer CS. The similar rates of rural and urban CS access may reflect shared constraints, including the digital divide and less access to CS teachers. Among Black, Hispanic/Latino, Native Hawaiian/Pacific Islander, and white high school students, the rates of access to foundational CS courses fall within a narrow range (80%–82%). Asian students (91%) are most likely to have access to CS courses. Native American students are least likely to have access to CS, though they had the largest year-over-year growth, from 66% in 2023–24 to 70% in 2024–25.

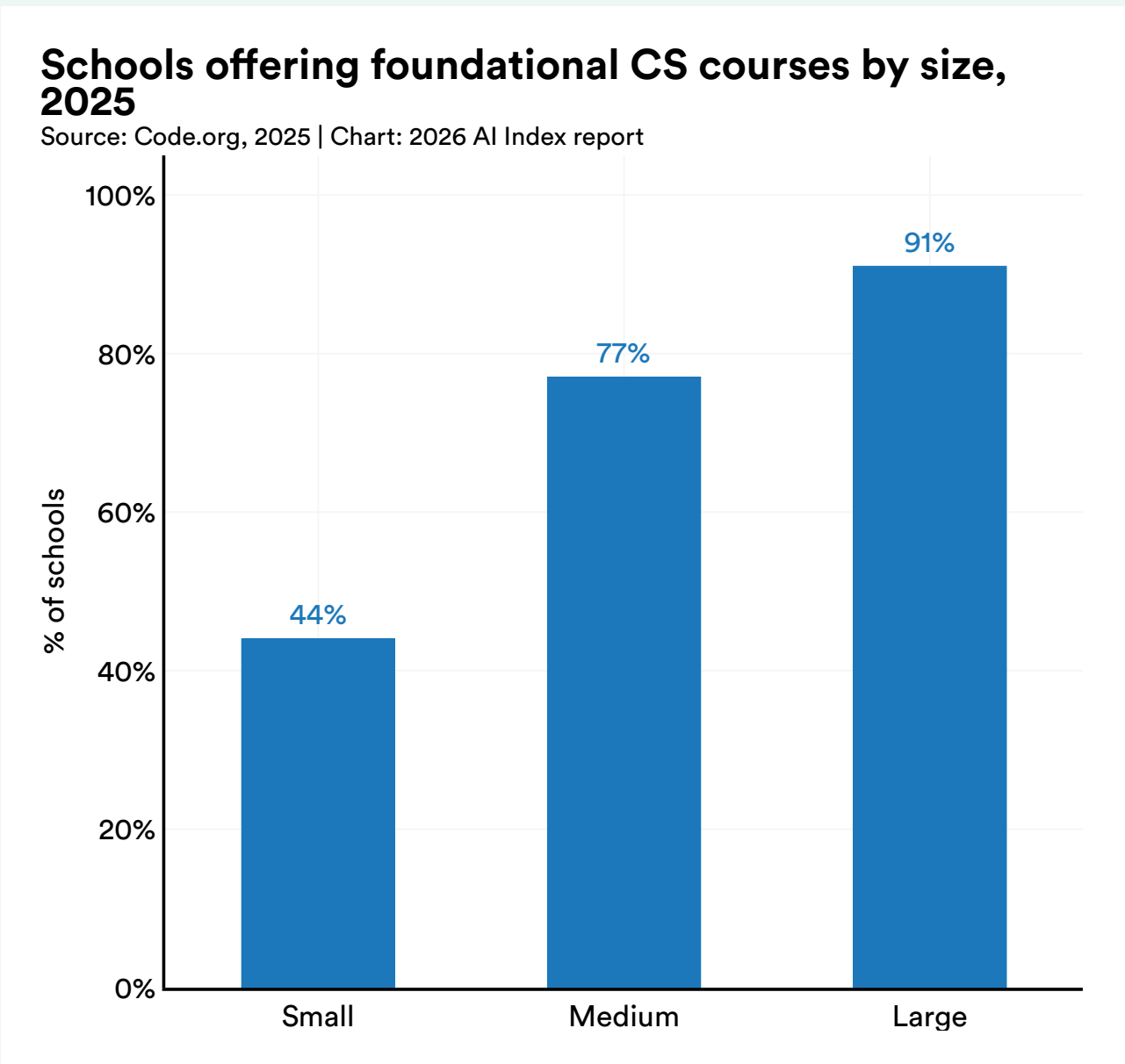

Figure 7.3.4

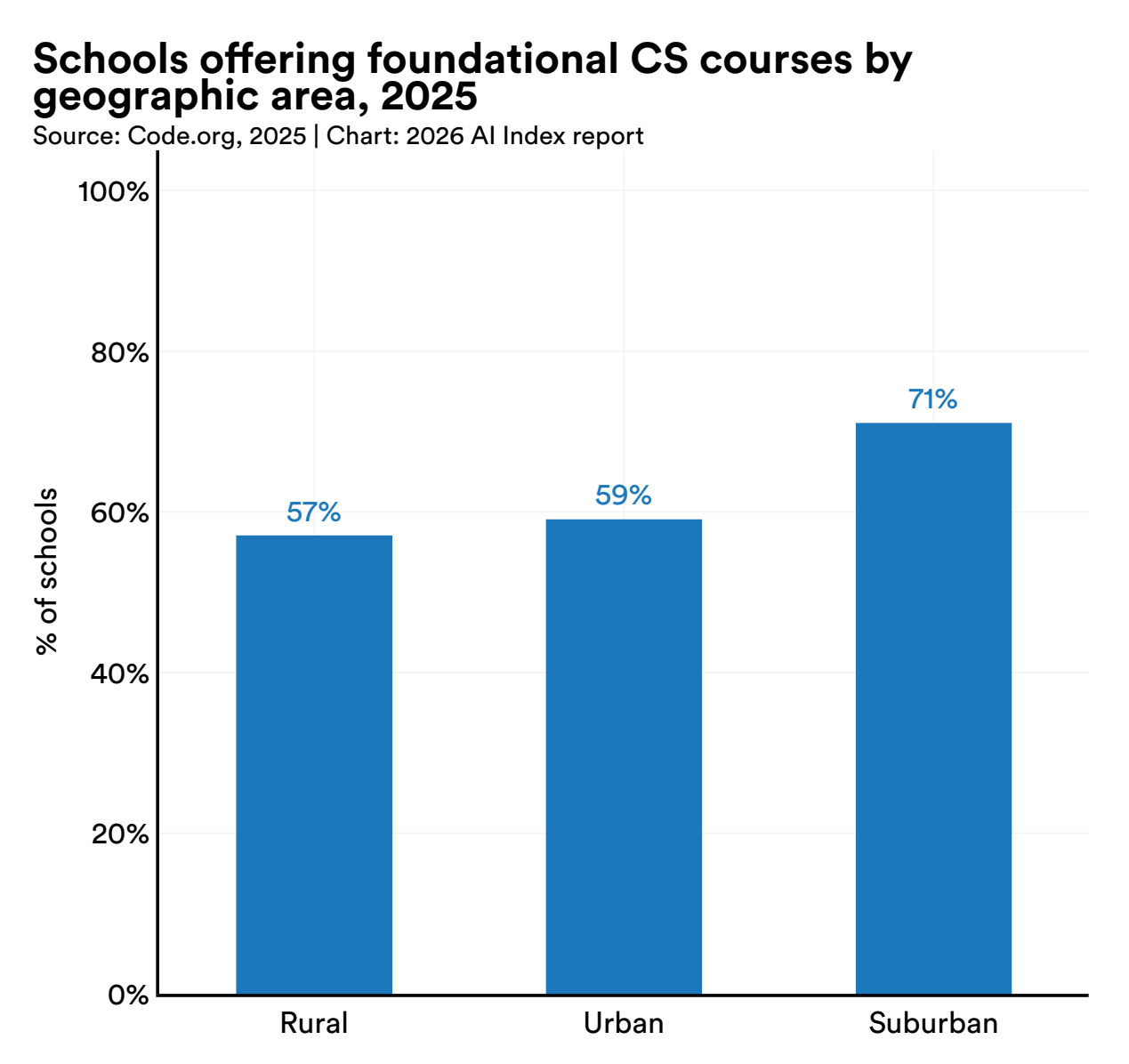

Figure 7.3.5

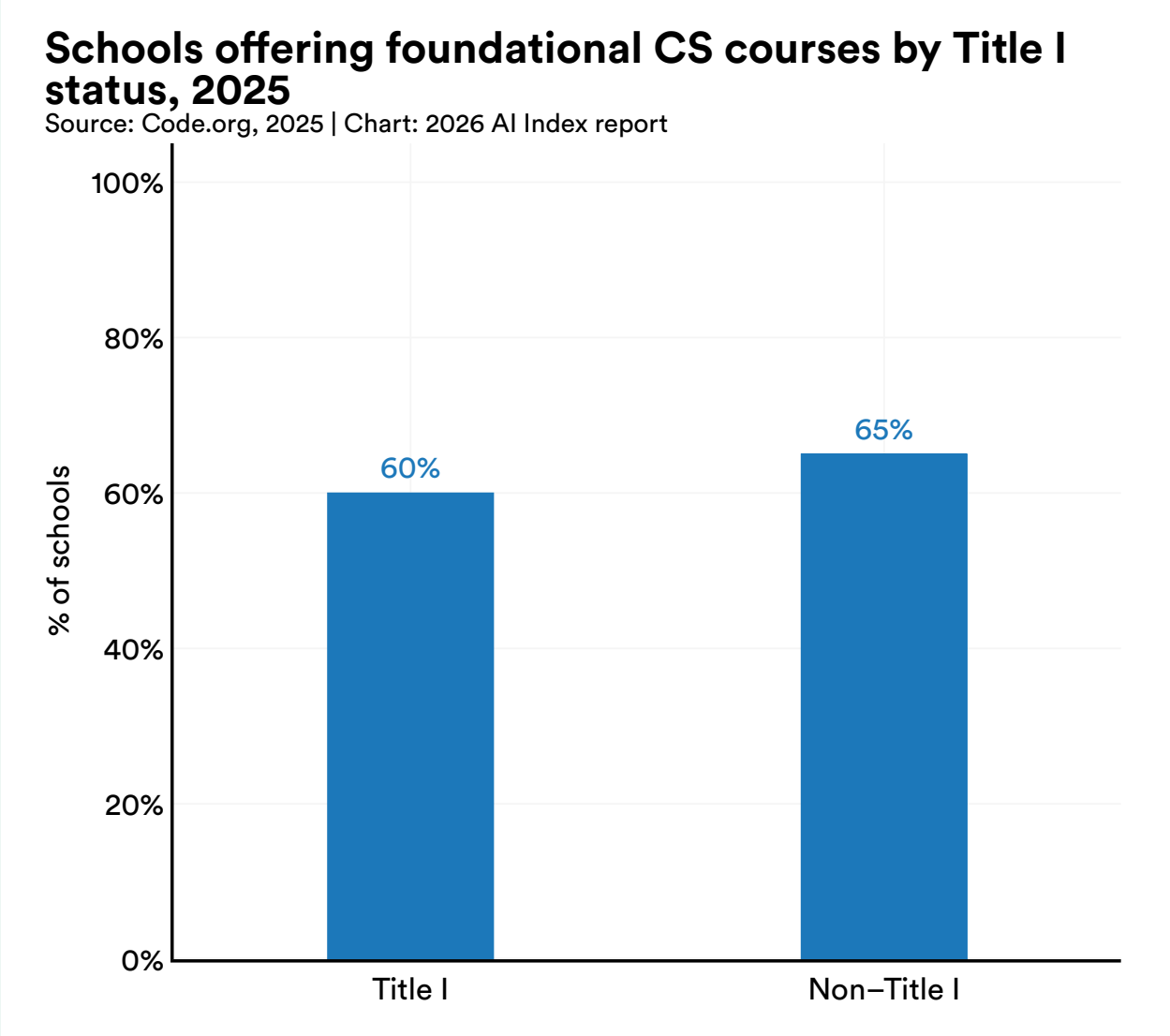

Figure 7.3.6

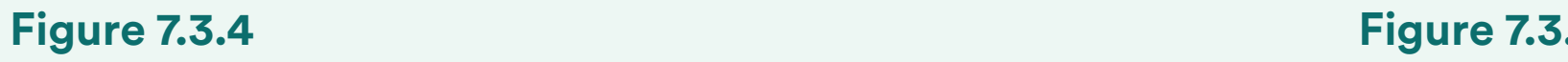

11 Title I schools serve students from low-income families and receive supplemental federal funding.

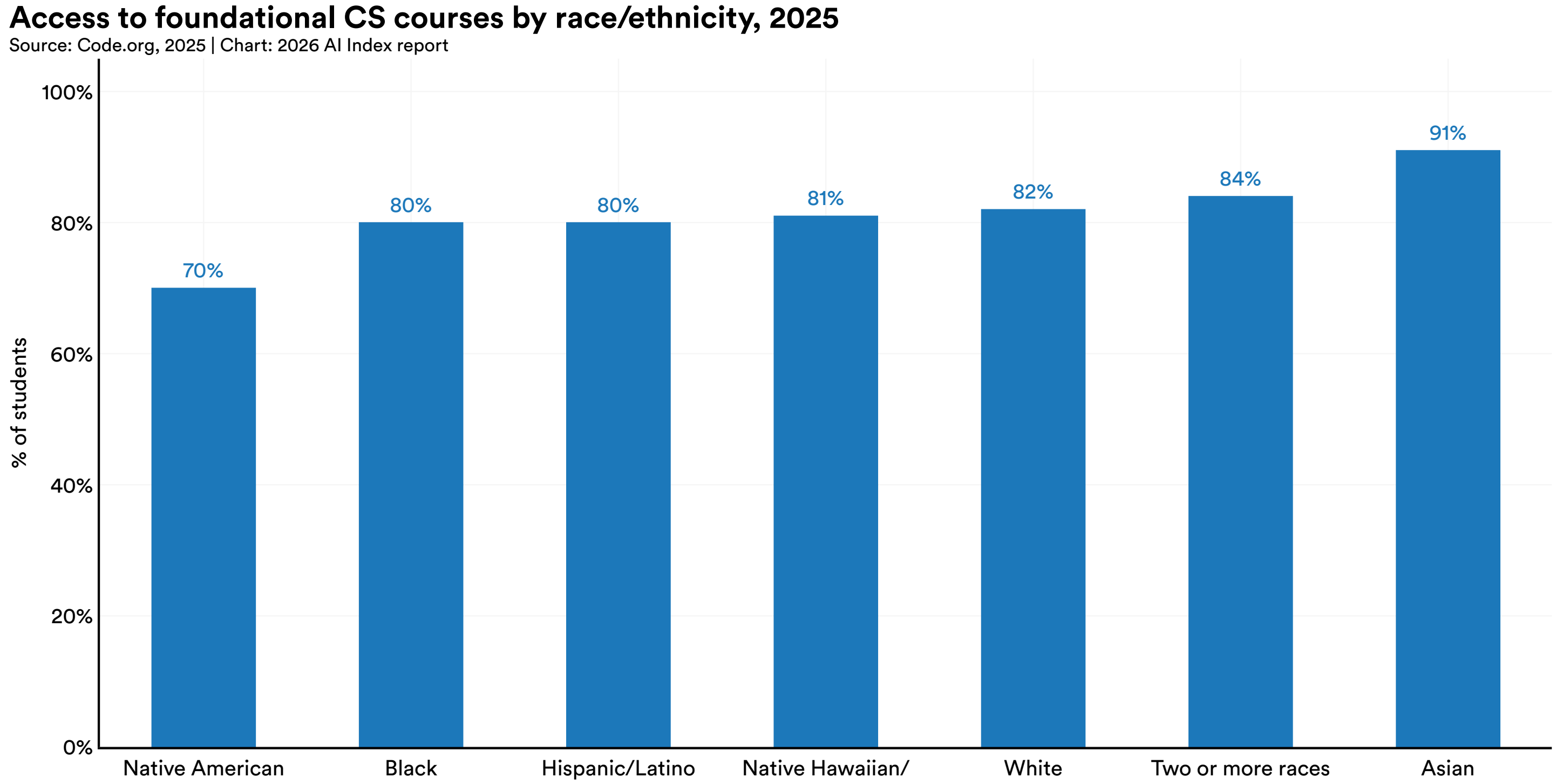

Figure 7.3.7

Based on participation data from 42 states, 6.1% of students were enrolled in CS in 2024–25, but student participation in CS varies by state (Figure 7.3.8). Arkansas and South Carolina report the highest participation rates, around 25%, while Idaho and Minnesota report the lowest (1.8%). Sixty-two percent of the reporting states (26 of 42) experienced a decrease in CS participation, though nearly half (12) of them reported decreases of less than 0.5% (Figure 7.3.9). Thirteen states (31%) reported an increase in student participation. The states that saw the largest increases in CS participation were North Dakota (11.1% increase from 5% to 16.1%), Tennessee (7.2% increase from 6% to 13.2%), and Arkansas (5.1% increase from 20% to 25.1%). Comparison with other states, including Ohio and California, is not possible as they did not report participation rates in 2023–24.

Several states with the highest percentage of high schools offering foundational CS courses also reported the highest participation rates, but the correlation was not evident in every state (Figure 7.3.10). Arkansas and South Carolina have the highest participation rates at 25.1% and 25.7%, respectively. This is not surprising given they have a CS requirement for graduation and near universal access rates (100% for Arkansas; 92% for South Carolina). Maryland also has universal access to CS, but fewer of their students (16.7%) take the available courses. Meanwhile, Rhode Island reports just 79% of their schools offer foundational CS courses, but a higher percentage of their students take those courses (18.4%).

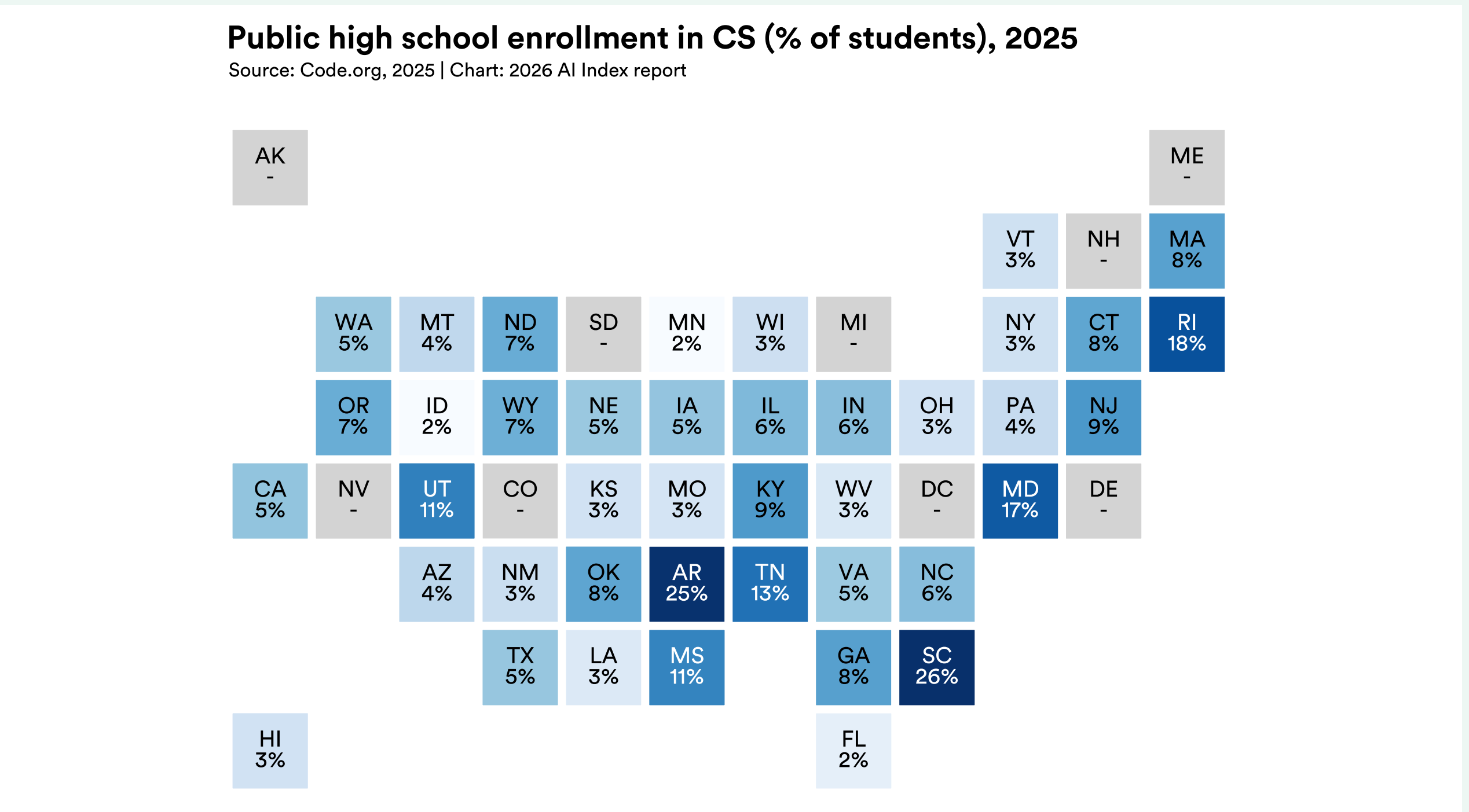

**Public high school enrollment in CS (% of students), 2025**
Source: Code.org, 2025 | Chart: 2026 AI Index report

Figure 7.3.8

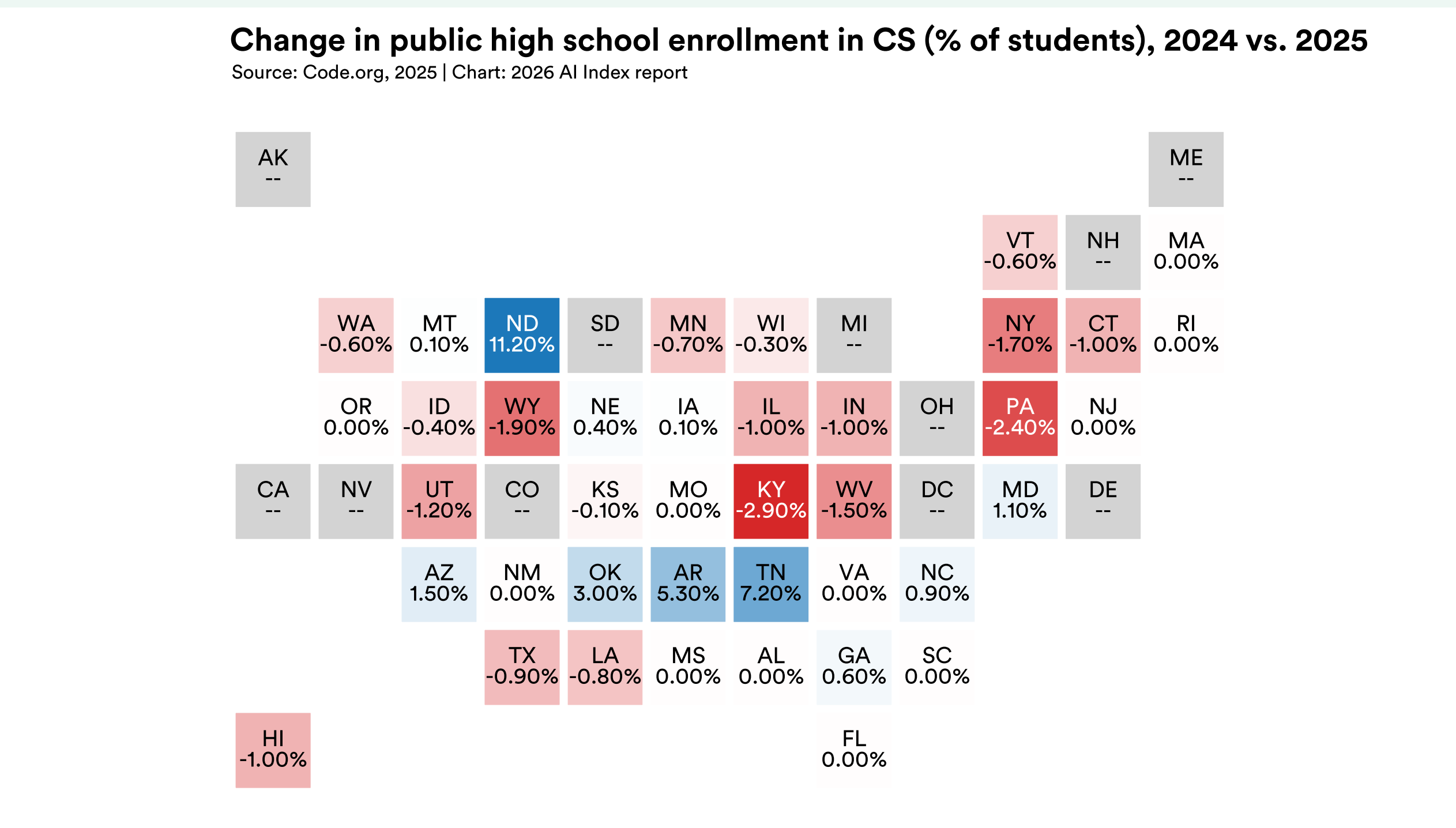

**Change in public high school enrollment in CS (% of students), 2024 vs. 2025**
Source: Code.org, 2025 | Chart: 2026 AI Index report

Figure 7.3.9

**Percent of high schools offering CS vs. percent of students enrolled in CS by state, 2025**
Source: Code.org, 2025 | Chart: 2026 AI Index report

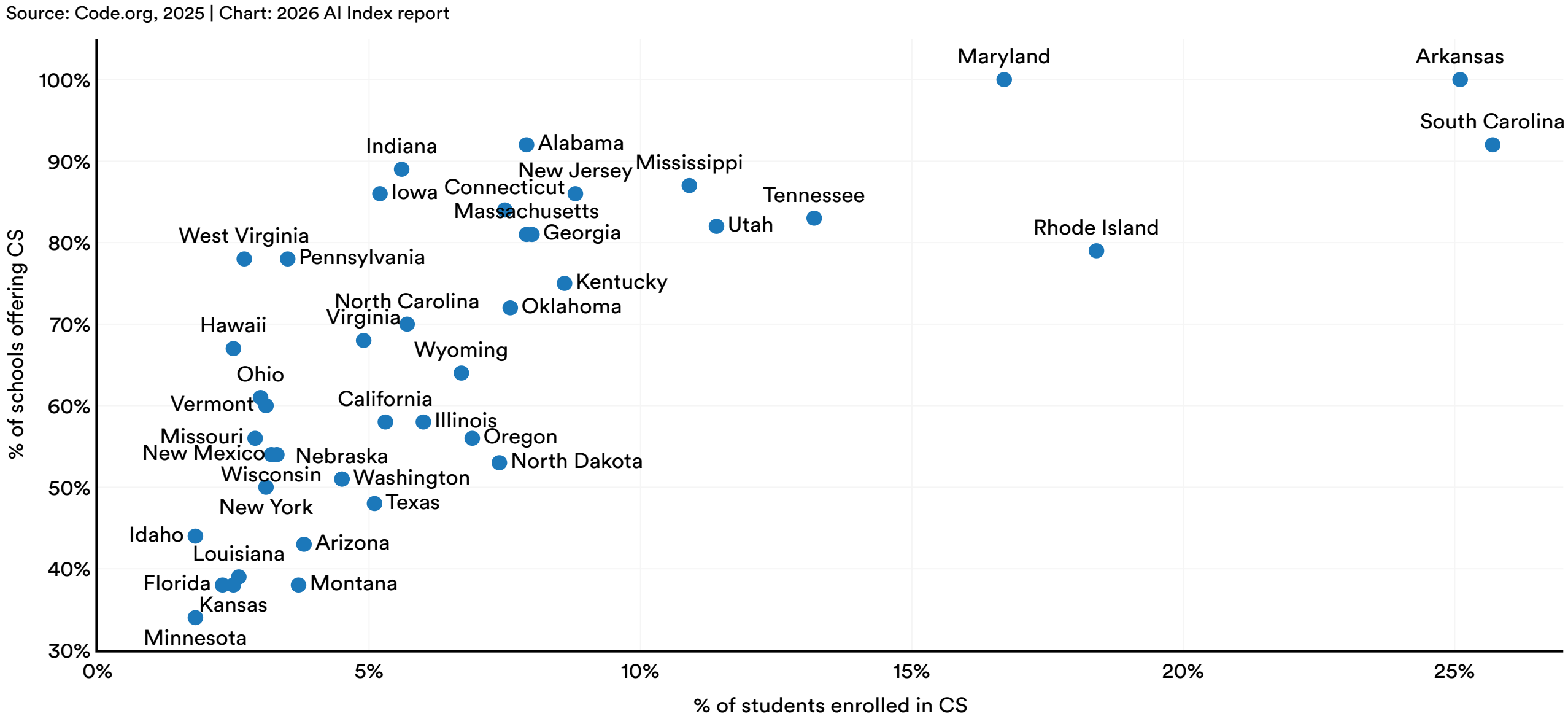

Figure 7.3.10

CS enrollment data also shows gaps for several subgroups (Figure 7.3.11); here too, the data should be interpreted with caution, as not all states reported enrollment data this year. Last year's analysis showed near or above proportional representation for Black, Native American/Alaskan, and white students at the national level. That trend continues this year, with Native Hawaiian/Pacific Islander students moving closer to that mark, and Asian students and students with 504 plans overrepresented among participating students (Figure 7.3.12). The representation of Hispanic/Latino students, economically disadvantaged students, students with IEPs, and girls slightly improved, but these populations remain underrepresented in CS courses. English language learners (ELL) remain underrepresented, but between 2024 and 2025, their representation notably improved, which may be due to concerted engagement efforts and initiatives to boost the quality of CS instruction for ELL students.

**Public high school enrollment in CS vs. national demographics by race/ethnicity, 2025**
Source: Code.org, 2025 | Chart: 2026 AI Index report

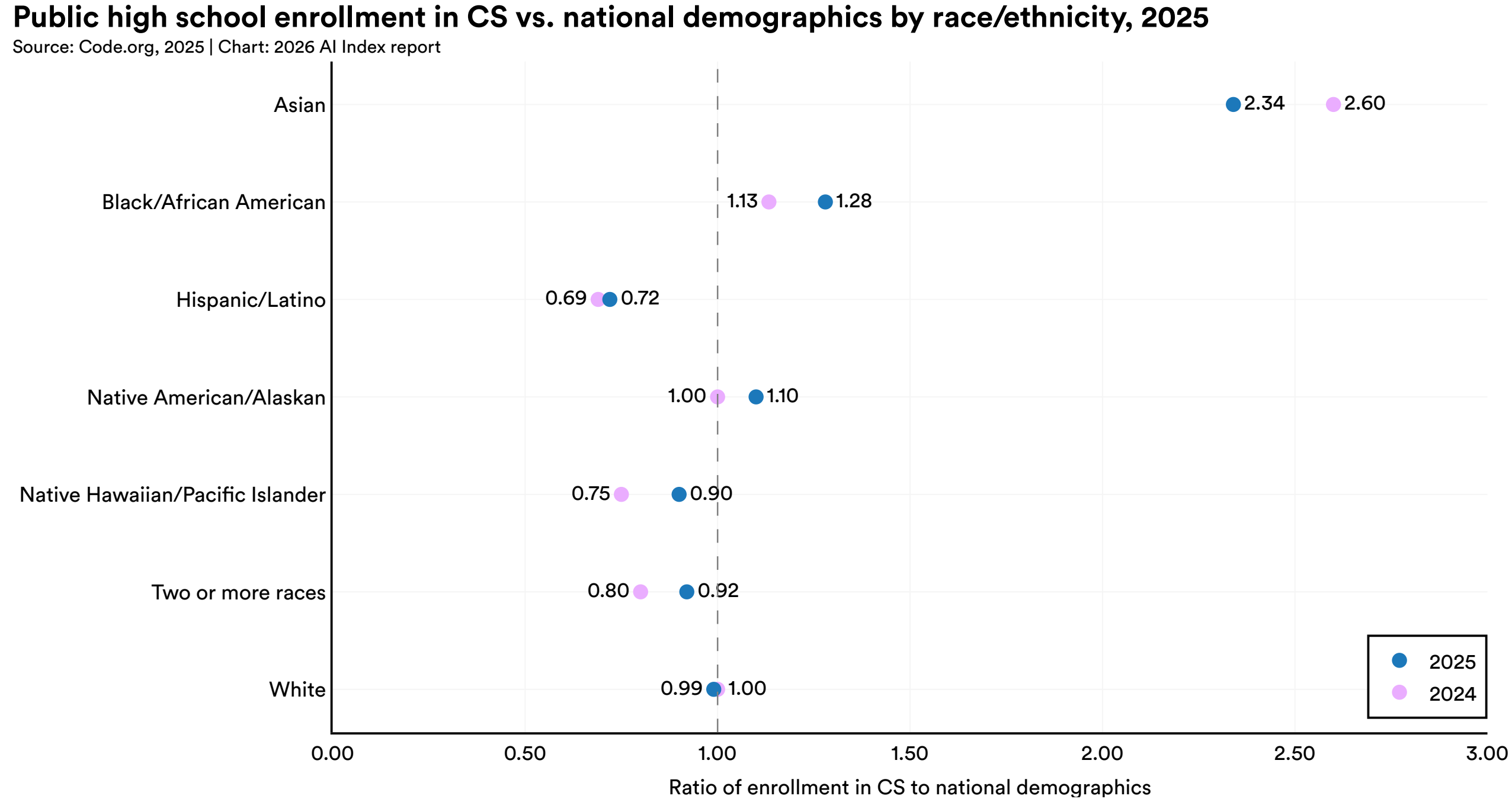

Figure 7.3.11

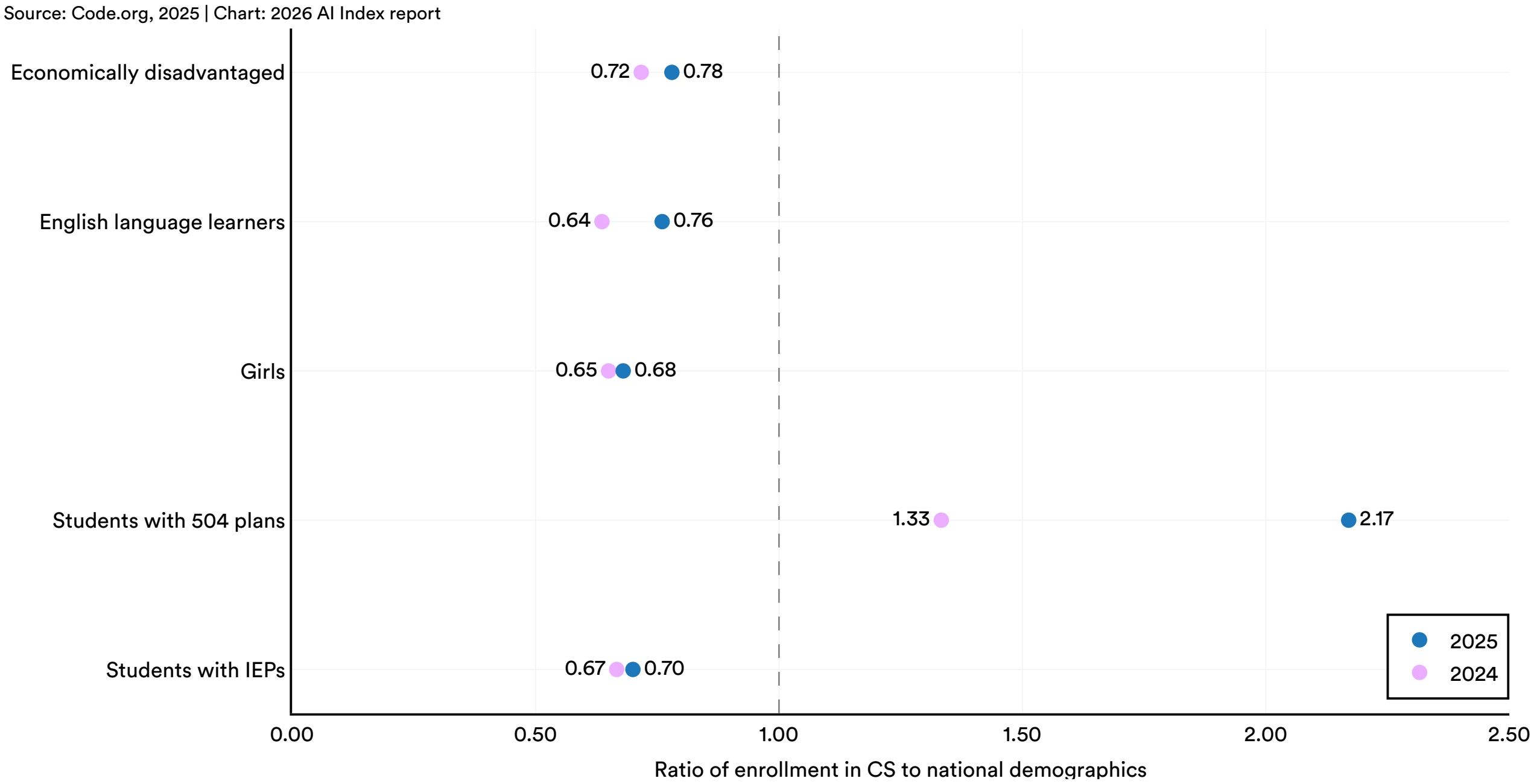
Public high school enrollment in CS vs. national demographics by subgroup, 2025
Source: Code.org, 2025 | Chart: 2026 AI Index report

Figure 7.3.12[12]

## Advanced Computer Science

Advanced coursework covers foundational AI concepts (e.g., AP CS Principles) and is an obvious pathway to build AI literacy and integrate more in-depth AI education. Since 2016, AP exam participation has grown steadily year over year, except for the plateau during the COVID pandemic between 2020 and 2021 (Figure 7.3.13). However, AP exam growth slowed from 21% between 2022 and 2023 to just 5% between 2023 and 2024. Despite improvement in student populations' representation across courses, students do not participate in AP exams proportionate to their racial/ethnic representation. Black, Native American/Alaskan, and Native Hawaiian/Pacific Islander students are better represented among CS education participants than students who took the AP exam in 2024; Hispanic/Latino students were better represented among AP exam takers than among the general population of students taking CS. Female students took the AP exam less often than male students in 2024; stereotype threat may explain the reluctance to take the exam and differences in scores. Asian students, multiracial boys, and white boys are overrepresented among those taking AP exams (Figures 7.3.14, 7.3.15, 7.3.16).

12 A student with a 504 plan receives accommodations under Section 504 of the Rehabilitation Act of 1973, a U.S. civil rights law that prohibits discrimination against individuals with disabilities. A student with an IEP (individualized education program) receives special education services under the Individuals with Disabilities Education Act. An IEP is a legally binding document that outlines a learning plan for a student with a disability designed to meet their unique needs and improve educational outcomes.

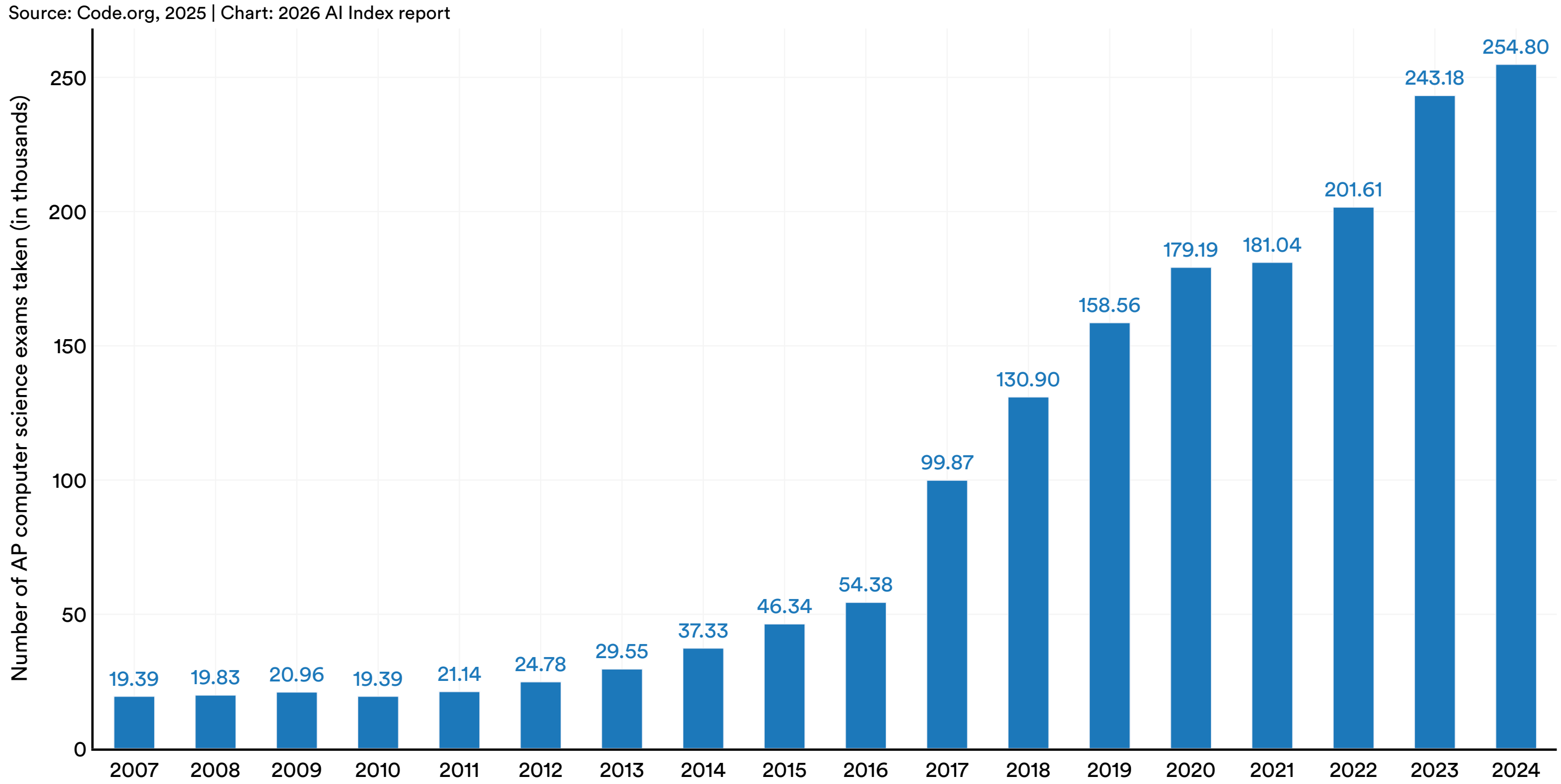
Number of AP computer science exams taken, 2007–24
Source: Code.org, 2025 | Chart: 2026 AI Index report

Figure 7.3.13

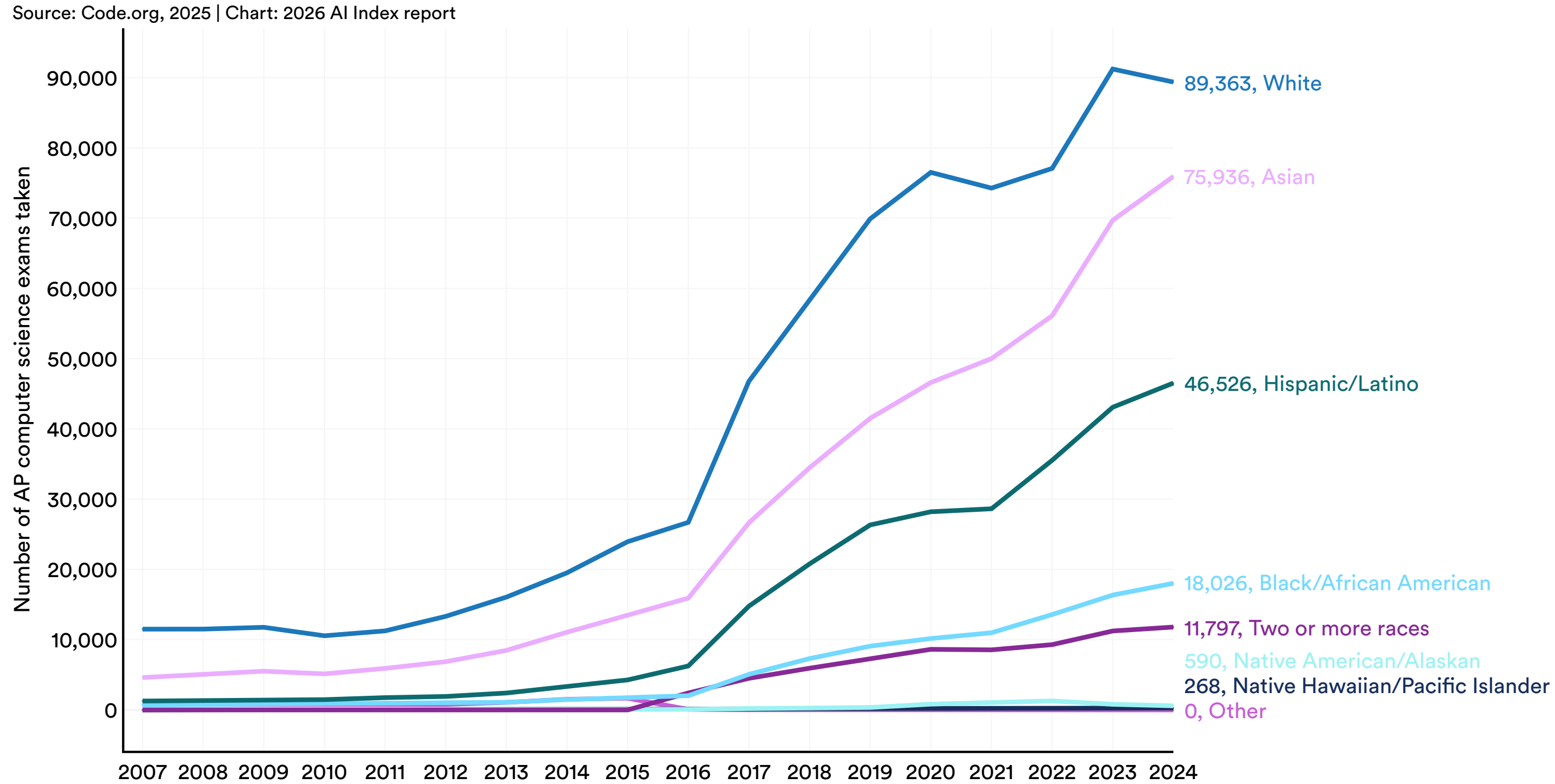
AP computer science exams taken by race/ethnicity, 2007–24
Source: Code.org, 2025 | Chart: 2026 AI Index report

Figure 7.3.14

**AP computer science exams taken (% of total responding students) by race/ethnicity, 2007–24**
Source: Code.org, 2025 | Chart: 2026 AI Index report

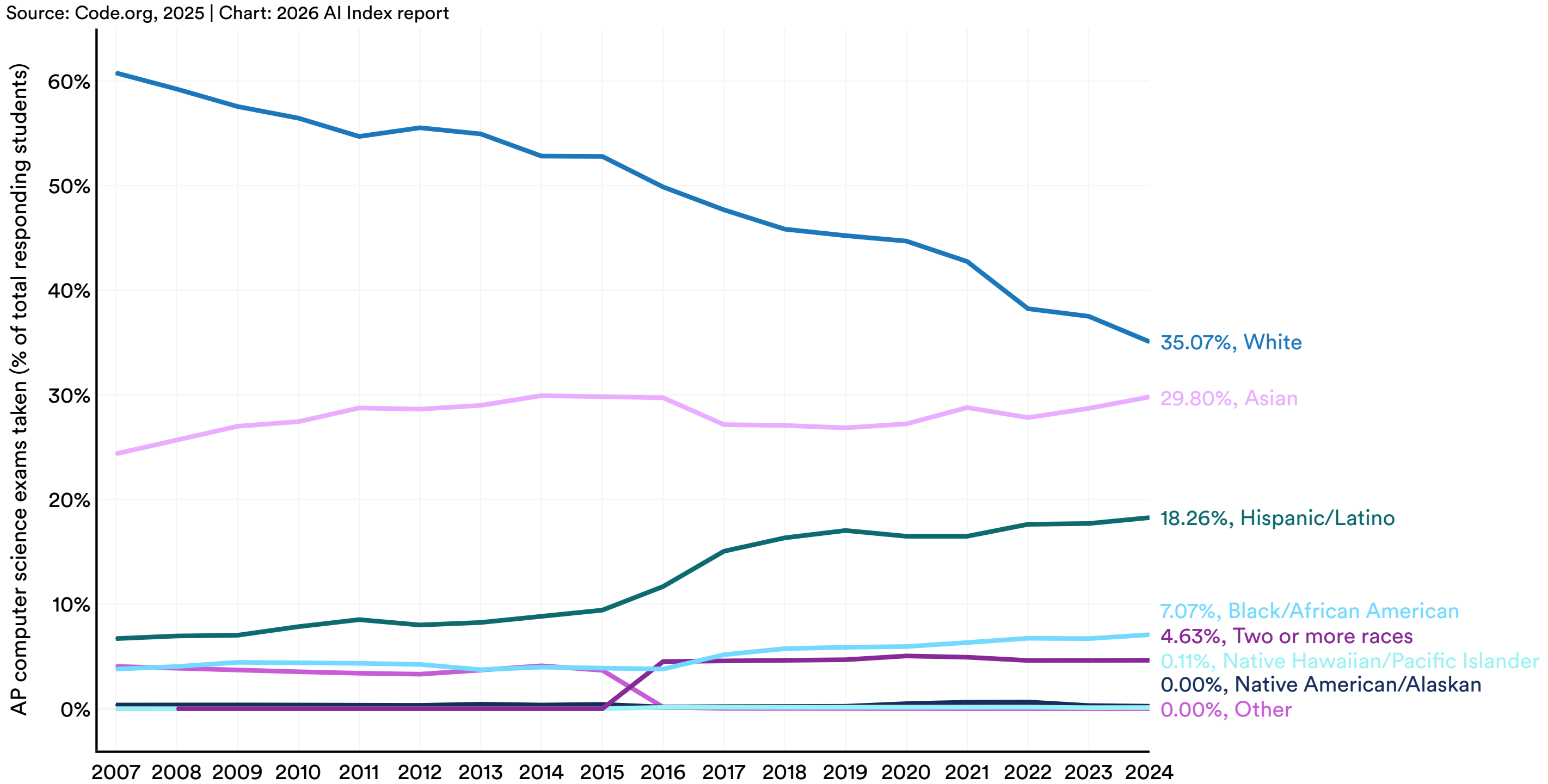

Figure 7.3.15

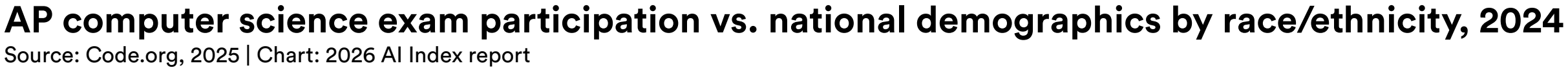
**AP computer science exam participation vs. national demographics by race/ethnicity, 2024**
Source: Code.org, 2025 | Chart: 2026 AI Index report

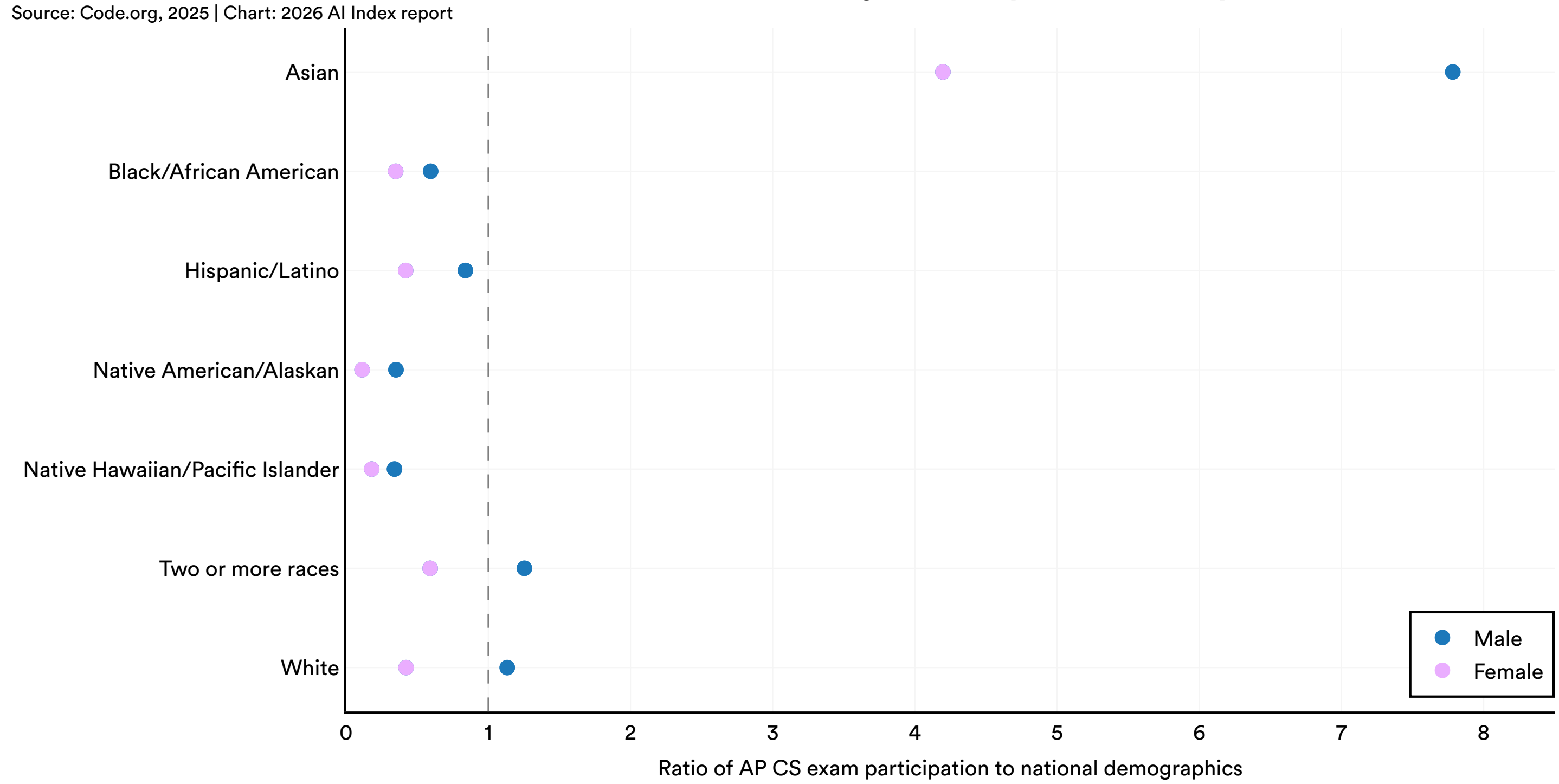

Figure 7.3.16

## K-12 Student Use of AI Tools

Using AI tools for educational purposes is increasingly popular among middle and high school students. Estimates on student use of AI to complete school-related tasks range from about 50% to 84%, based on survey data.[13] While students acknowledge concerns (e.g., false cheating accusations, diminished critical thinking, and weakened academic skills), studies suggest that usage is trending upward, with a majority of students now advocating for AI use in schools. One survey found that about half of students surveyed agreed that schools should be required to teach students how to use AI (52%) and that students should be allowed to use AI to complete homework (47%); another survey reported even higher rates of agreement, with 65% of middle school students and 73% of high school students stating that students should have access to and be able to use AI tools to complete schoolwork. High school students report using generative AI most often for conducting research and finding sources, editing or revising essays, and brainstorming ideas (Figure 7.3.17), and they report benefiting from increased access to learning materials and more efficient learning.

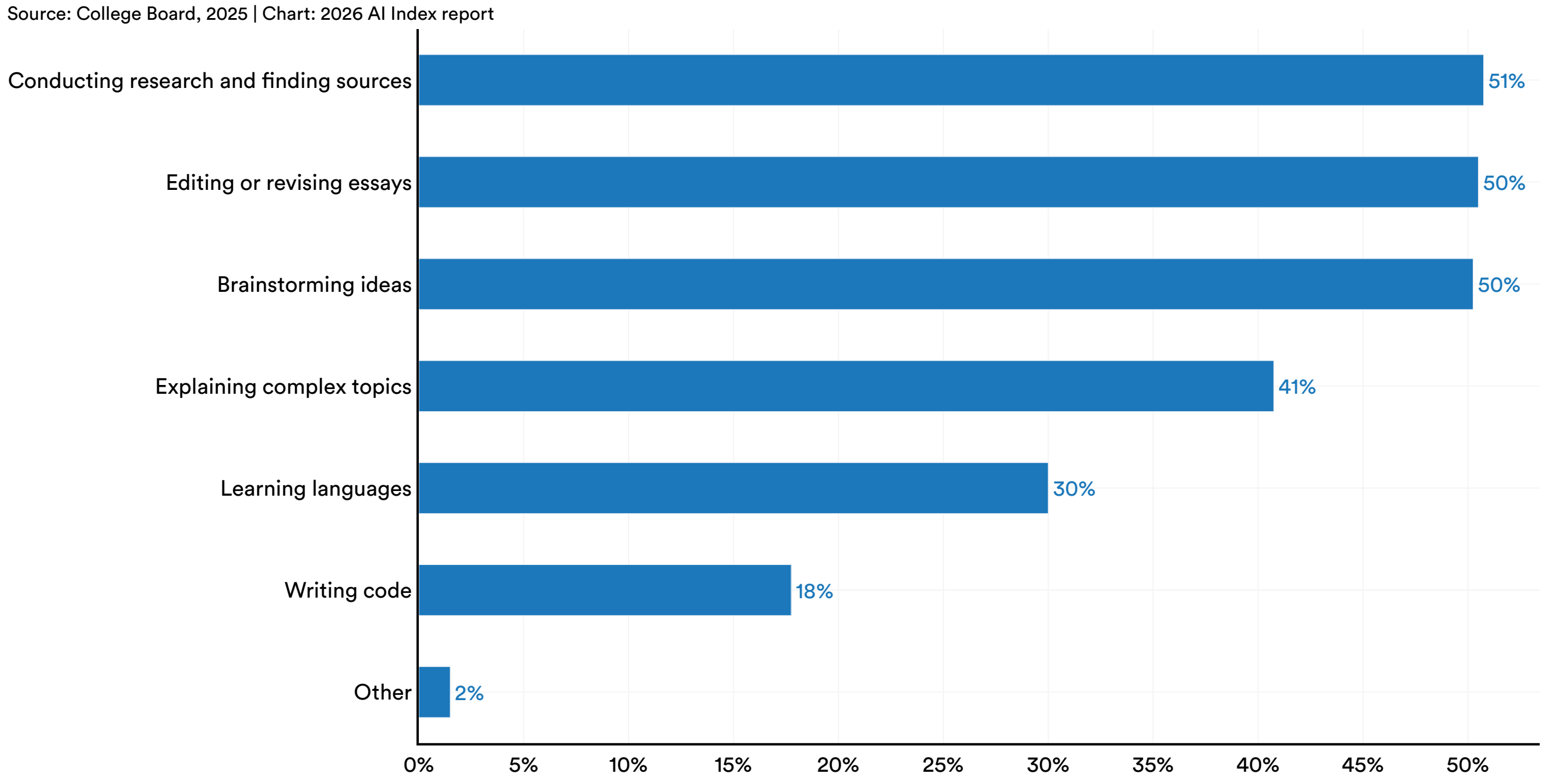
High school students' GenAI uses for schoolwork, 2025
Source: College Board, 2025 | Chart: 2026 AI Index report

Figure 7.3.17[14]

## Education Standards, Policies, and Guidance

Only about half of middle and high schools have policies regarding AI use. Of that, only 28% permit AI use in some circumstances, while 22% do not. Schools with AI policies, especially policies that allow for AI use in schoolwork, are more likely to be in wealthier and more urban communities. However, the utility of those policies, given their lack of clarity, is in question. Only 36% of students described their school's policies as extremely clear, and 47% have wanted to use AI for schoolwork but were unsure if it was allowed. A teacher survey found even lower ratings for clear policies, with teachers saying that only 6% of their schools had clear, comprehensive policies.

13 CDT (2025), 50%; College Board (2025), 84%; RAND (2025), 54%.

14 This chart shows the average percentage across College Board's four survey administrations in 2025.

### State K–12 AI Guidance in the United States

Most education policy in the United States is determined at the state level. As of January 2026, 30 states have issued guidance on AI *in* education. Regarding policy focused specifically on AI education, 17 states have issued guidance that clarifies CS as foundational to AI, and five states have allocated specific professional development funding for AI education, according to the 2025 State of AI + Computer Science Report.

Perhaps the most significant AI education guidance generally comes in the form of standards that define learning outcomes for K–12 students. As of January 2026, 45 states have adopted K–12 CS standards, while five states plus DC do not have such standards. The majority (29 states) include AI but only to a very limited extent, and it is typically restricted to the high school level, similar to the current CSTA K–12 Standards, Revised 2017, which act as the de facto national standards. Ten states' standards make no specific mention of AI content. Six states have CS standards with significant AI-specific content, and another two states have published draft revised standards that also include significant AI-specific content (Figure 7.3.18).

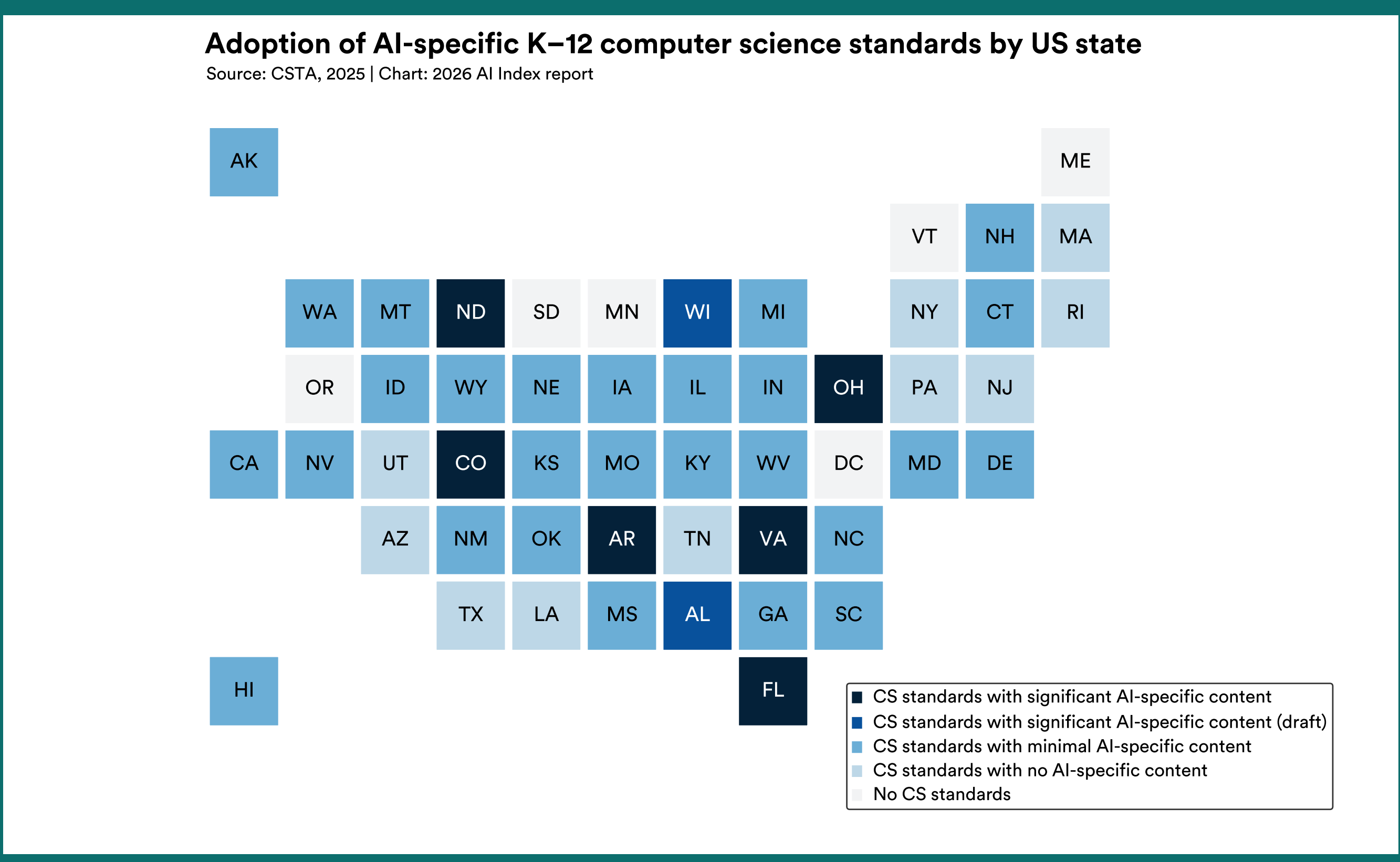

Figure 7.3.18

States with significant AI standards were all adopted in the last few years or are currently under development. See Table 4.1 for a summary of the features and organization of the state standards with significant AI content.

### Table 4.1. Features and organization of state standards with significant AI content

| State | Year | Earliest grade | Approach | Organization & labeling of AI content |
|---|---|---|---|---|
| Alabama | 2025 - DRAFT | 1st grade | Integrated across concepts | An [AI] tag marks standards distributed widely across concepts, most commonly in Data Science > Data Collection and Representation; Impact of Computing > Emerging Technology; and Digital Proficiency > Digital Tools. |
| Arkansas | 2025 | 9–12 elective | Elective course | AI is the focus and label of one entire level 3 course in Data Analytics and Machine Learning pathway. |
| Colorado | 2024 | 2nd grade | Standalone concept | AI is one of six strands used to organize standards. |
| Florida | 2024 | 1st grade | Integrated across concepts | AI content is organized into two main strands: Technological Impact (grades 1–12) and Emerging Technology (grades 6–12). |
| North Dakota | 2025 | Kindergarten | Integrated across concepts | AI content is organized into three subconcepts: Computing Devices and Systems > Artificial Intelligence; Algorithms and Computational Thinking > Creating Instructions for AI; and Impacts of Computing > Impacts of AI. |
| Ohio | 2022 | Kindergarten | Standalone concept | AI is one of six strands. The strand has five topics that align with the AI4K12 Five Big Ideas: Perception, Representation and Reasoning, Machine Learning, Natural Interactions, Societal Impacts. |
| Virginia | 2024 | 4th grade | Integrated across concepts | AI content is organized into three strands: Computing Systems; Data and Analysis; and Impacts of Computing. |
| Wisconsin | 2025 - DRAFT | Pre-K–5 (grade band) | Integrated across concepts | AI content is organized in two subconcepts: Computing and Society > Emerging Technologies; and Data and Analysis > Impacts of Data Science. |

Revised CSTA K–12 Standards, slated for release in summer 2026, will delineate significant AI-related learning goals as part of a foundational CS education across grades K–12. To inform these standards, CSTA and AI4K12 convened a national group of educators, curricula developers, professional development providers, and researchers in 2025 to provide insights into identifying priority areas of AI knowledge and skills. Priorities include the human role in creating AI, reasoning, data and machine learning, ethical evaluation of AI systems, and societal impacts. In the new Pre-K–12 foundational standards, these AI priorities will be integrated across five concepts and most significantly organized into four subconcepts: Machine Learning, Impacts of Algorithms, Emerging Technologies, and Humans and Computing. Additionally, there will be two sets of high school specialty standards focused on advanced AI content. Given the high degree of current coherence, most states will likely adopt similar AI standards within the next five years.

### Federal K–12 AI Guidance in the United States

An April 2025 Executive Order, Advancing Artificial Intelligence Education for American Youth, sought to define a national strategy for developing AI competency from K–12 through postsecondary education by promoting early student exposure to AI, integrating AI into instruction, and expanding professional learning for educators. It establishes a White House AI Education Task Force to coordinate federal efforts, launch a Presidential AI Challenge, and develop public-private partnerships that deliver K–12 AI resources at scale. The order also directs the Departments of Education, Labor, NSF, and Agriculture to prioritize AI in grants, research, teacher preparation, apprenticeships, and workforce pathways.

**HIGHLIGHT:**

## The Implementation Gap in K–12 AI Policy

A research project from Expanding Computing Education Pathways (ECEP) assessed statewide AI policies across K–12 schools and found that state-level AI guidance is largely nonbinding and decentralized. Most states rely on existing federal laws like the Children's Online Privacy Protection Act (COPPA) and Family Educational Rights and Privacy Act (FERPA), rather than issuing AI-specific mandates. The responsibility for local policy development, tool vetting, and implementation falls on local education agencies, meaning the rigor of AI education and pace of adoption are determined by local capacity (e.g., number of trained teachers, funding) and decision-making.

Teacher preparation was also an identified gap. State-level documents recognize the importance of AI-related teacher training, but there are currently no state-level standards for programs or funding. Without steady financial backing or standardized training benchmarks, the quality of AI integration remains contingent on local resources.

At the same time, AP Computer Science, one of the most common pathways to advanced CS coursework in U.S. high schools, does not include AI-specific content. Policy guidance, teacher training, and curriculum would all need to align for AI education to reach students consistently, and at present, gaps remain in all three.

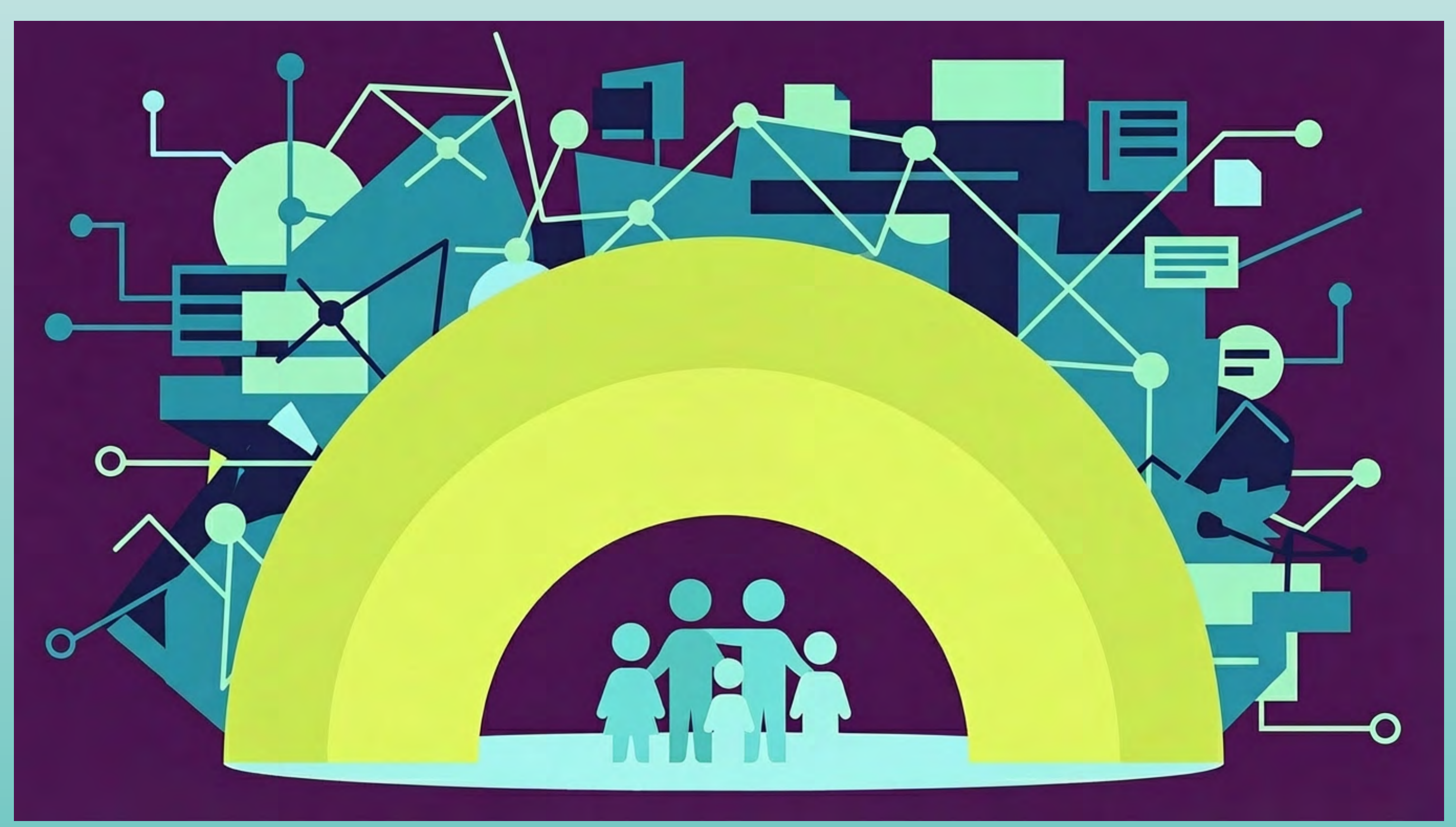

## Global AI and CS Education

Despite the widespread mention of AI education in national education strategy plans over the past few years, few countries actually implemented AI education in 2025; it was more common for countries to integrate AI technology into education. For example, South Korea launched AI textbooks in primary schools in March 2025, only to reverse course a few months later due to parent and teacher pushback. In Greece, the government partnered with OpenAI to train secondary teachers to use ChatGPT in the classroom. And in Estonia, the AI Leap 2025 program is piloting access to AI learning applications with 20,000 students and 3,000 teachers during the 2025–26 school year.

Two countries, however, made significant strides in implementing AI education: China and the United Arab Emirates (UAE). In China, Beijing, Guangdong, and Hangzhou all began requiring AI education in the 2025–26 school year following the release of China's General AI Education Guide for Primary and Secondary Schools (2025 Edition) and Guide for the Use of Generative AI by Primary and Secondary Students (2025 Edition) in May 2025. All three areas have similar requirements, including a minimum number of instructional hours and curriculum that progresses through grade levels, starting with elementary students learning AI literacy skills and ending with high school students designing AI systems. The UAE similarly mandated AI education for all grade levels starting in the 2025–26 school year. Students will also progress through a grade-level curriculum that includes skills in foundational concepts, data and algorithms, software use, innovation and project design, and ethical awareness.

In lieu of widespread data on AI education, we again present data on CS education, especially since some AI content may be taught in CS classes. Similar to the challenges inherent in tracking CS education in the United States, caution is called for when interpreting global metrics because CS and ICT education are sometimes conflated with digital or computer literacy.

In 2025, approximately 93% of the world's countries taught CS (Figure 7.3.19). Thirty percent of countries mandate CS education in either primary or secondary school, while 63% have CS available in at least some schools but do not mandate it. Nearly three-fourths of countries integrate CS concepts into other courses, such as math and science. Access to standalone CS classes often varies by school type and geography, with private and urban schools more likely to offer CS than public and rural schools. This indicates that resources and infrastructure continue to be challenges for schools seeking to expand students' digital skills.

**Availability of CS education by country, 2025**
Source: AI Index, 2025; Raspberry Pi Computing Education Research Centre, 2024 | Chart: 2026 AI Index report

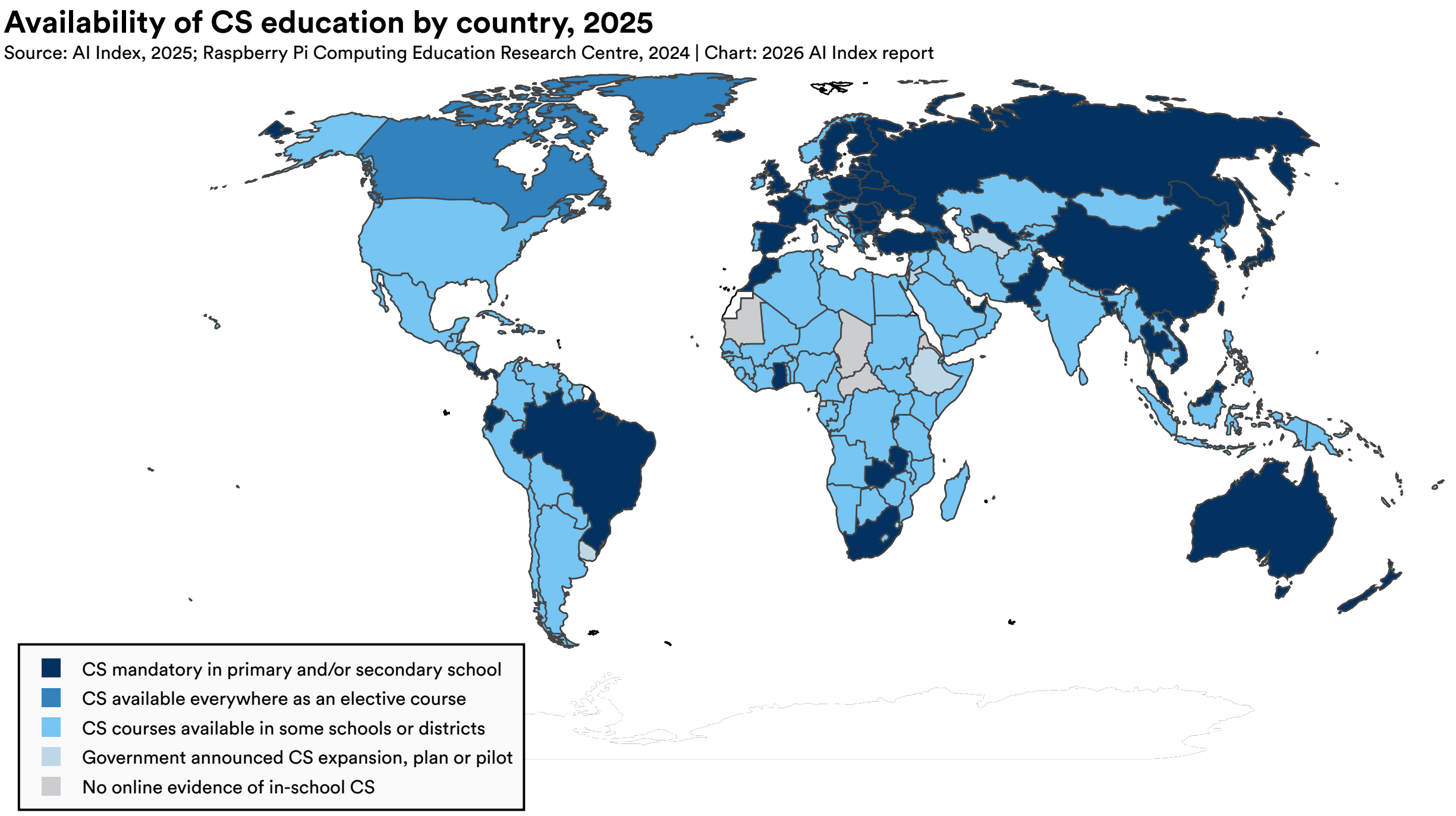

Figure 7.3.19

# 7.4 AI Skill Acquisition

Formal education is one entry point into AI, but as the technology reshapes jobs across sectors, upskilling and reskilling have become central to lifelong learning. Many people are building AI skills through professional certificates, online courses, and on-the-job experience, pathways that can also broaden access for learners without deep CS or math backgrounds. This section examines where AI skills are concentrated globally and how quickly they are spreading.

## AI Skill Penetration

LinkedIn's relative AI skill penetration rate measures how prominently AI skills feature in people's profiles in a given country compared with a global average (Figure 7.4.1). India leads at 3.0, meaning AI skills appear in member profiles at almost three times the global average, followed by the United States at 2.0 and Germany at 1.8. However, these countries also show a persistent gender gap when measuring male and female AI skill penetration rates against the global average (Figure 7.4.2). In India, men list AI skills at more than 1.5 times the rate of women (3.1 vs. 1.9); in the United States, the gap is similar, though a bit more narrow (2.1 vs. 1.4).

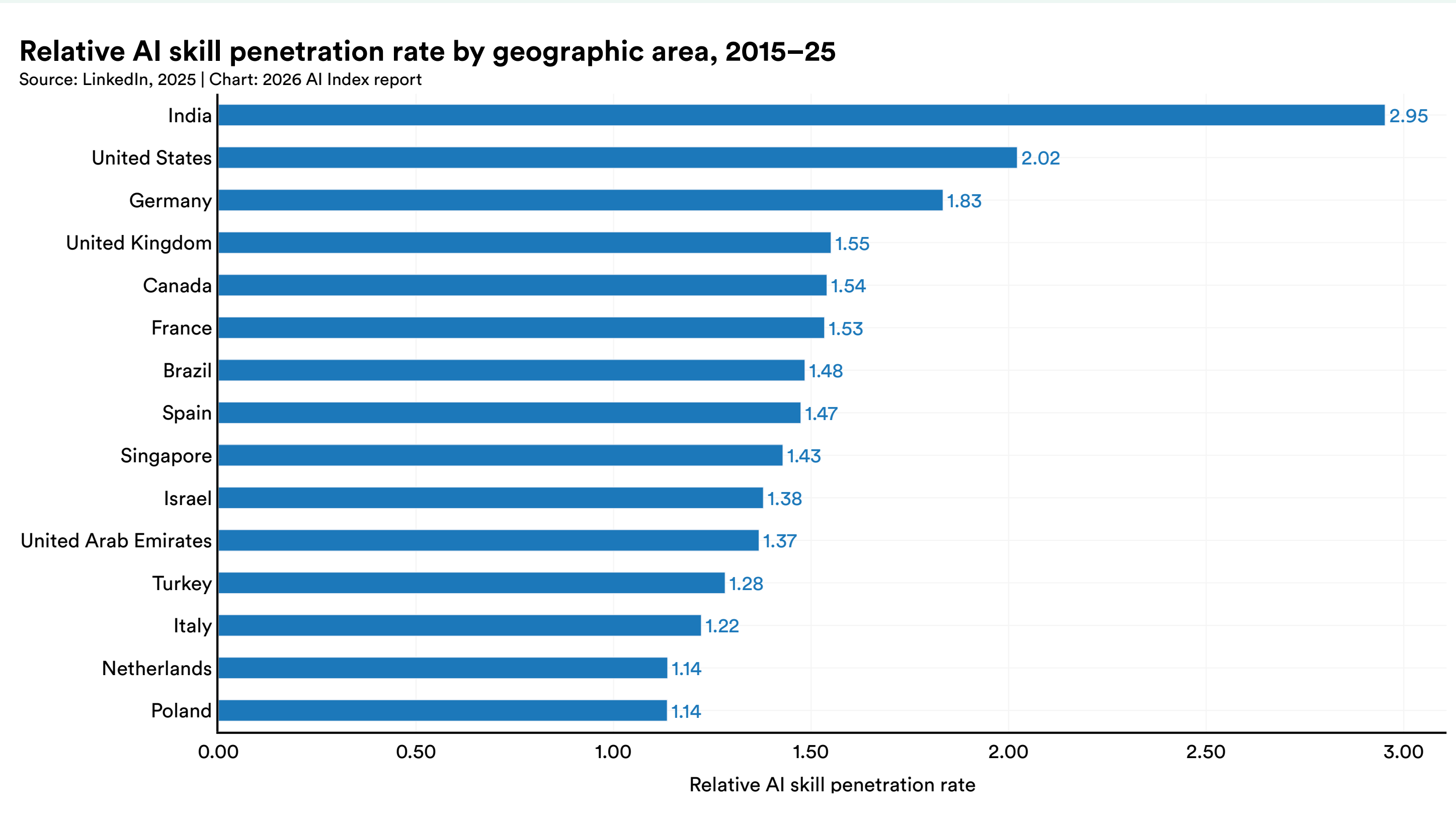

Figure 7.4.1

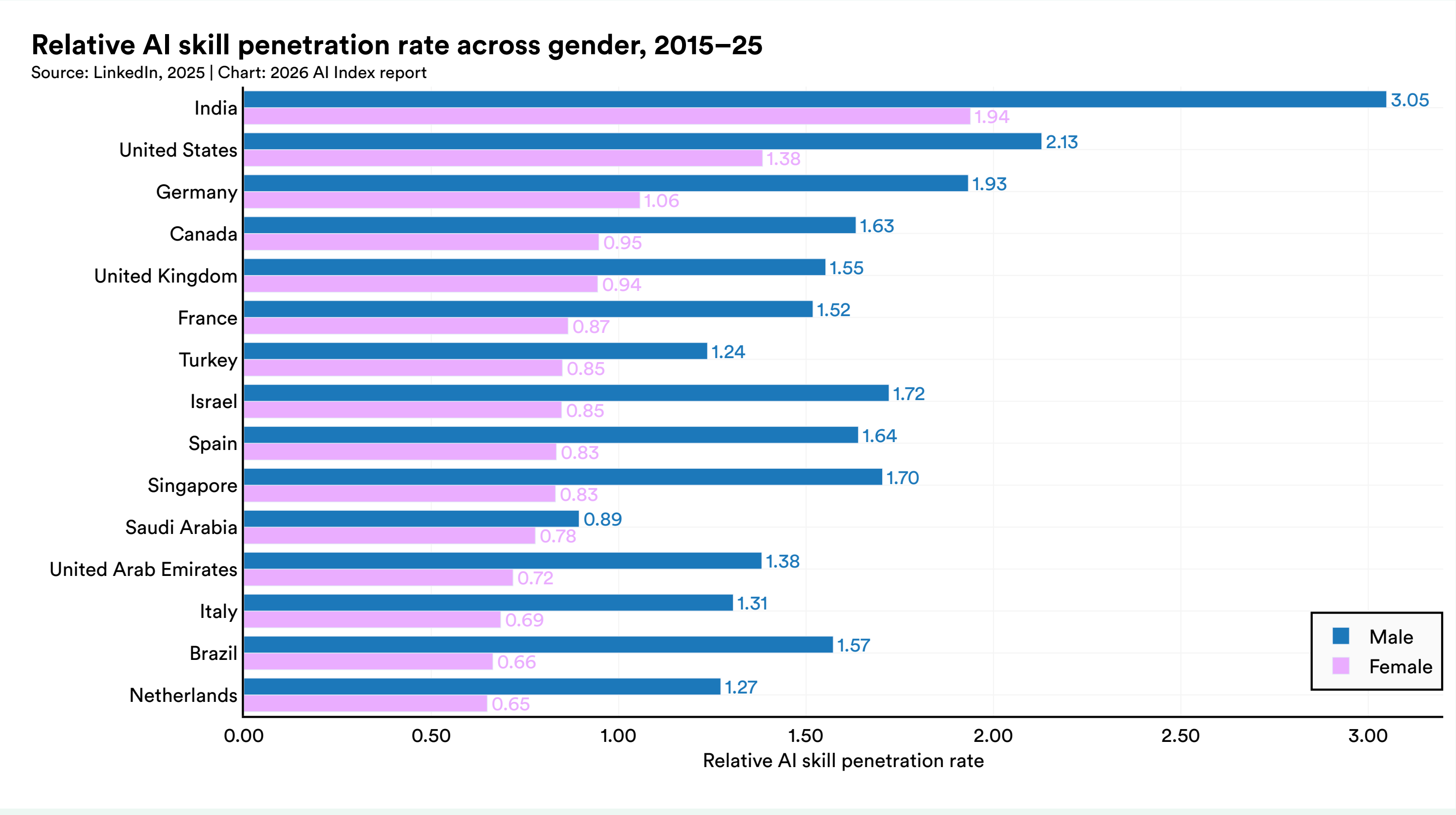

Figure 7.4.2

## AI Skills Diffusion

The AI Skills Diffusion Index that LinkedIn introduced this year tracks how much AI skills adoption has grown within a country relative to its own baseline, rather than current relative prevalence (Figure 7.4.3). This measure also accounts for the diversity of AI skills, distinguishing between AI engineering skills, which relate to building and deploying AI systems, and AI literacy skills, which reflect a familiarity with AI-enabled tools. Across many of the countries in the sample, both AI literacy and engineering show recent increases, but the pace differs. AI literacy skills show steeper growth, while engineering-oriented skills have increased more modestly. This is the case for India and the United States. However, countries such as the United Arab Emirates, Chile, and South Africa show rapid growth in AI engineering skills. In the United States, the fastest growing literacy skills were AI prompting and Microsoft Copilot Studio, while the fastest growing engineering skills were AI agents, AI productivity, and AI strategy (Figure 7.4.4).

## AI Skills Diffusion Index by geographic area, 2016–25

Source: LinkedIn, 2025 | Chart: 2026 AI Index report

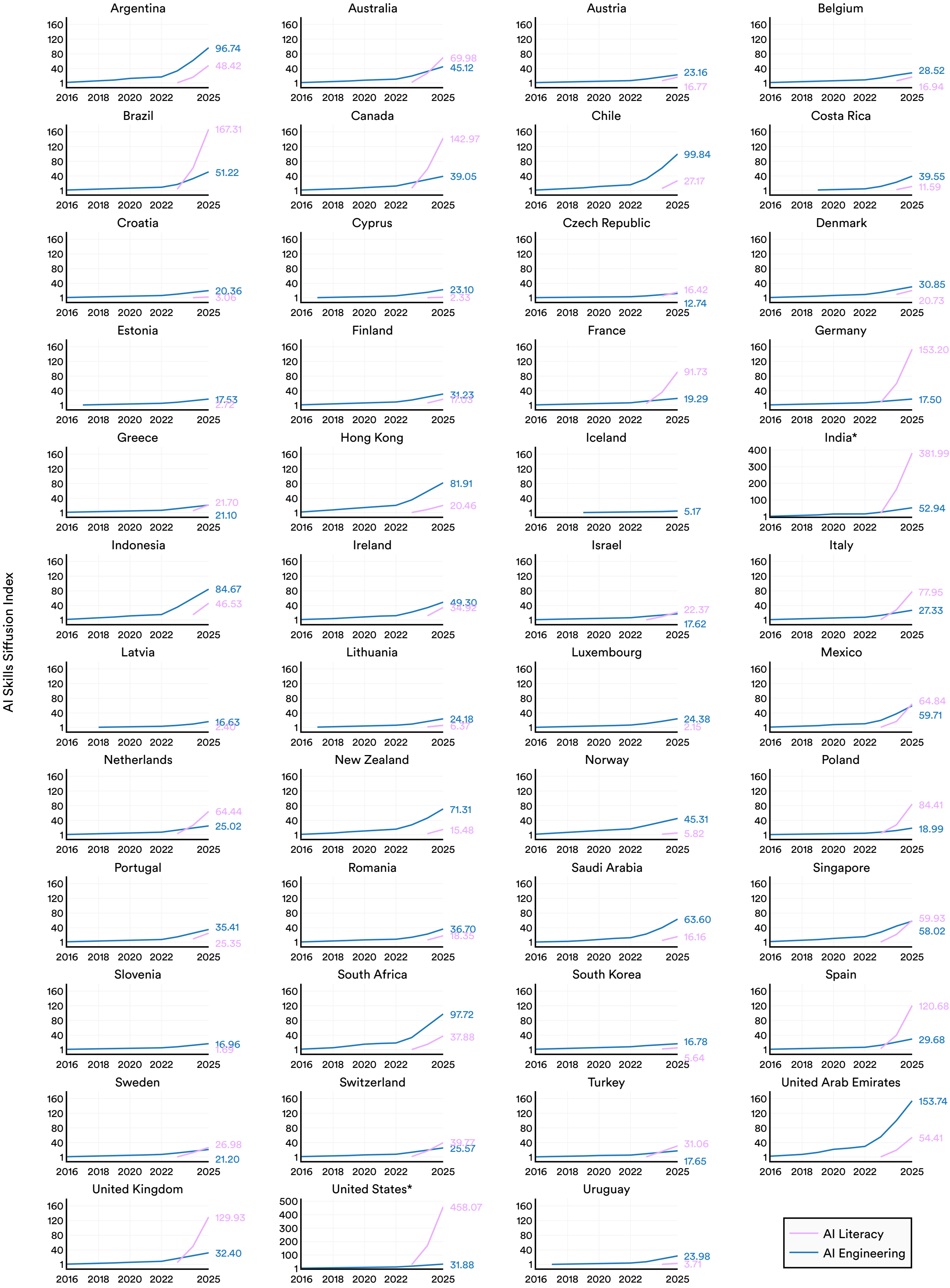

**Figure 7.4.3**[15]

15 Asterisks indicate that a country's y-axis label is scaled differently than other countries'.

### Fastest growing AI skills in the United States, 2025

Source: LinkedIn, 2025

| Rank | AI engineering skills | AI literacy skills |
|---|---|---|
| 1 | AI agents | AI prompting |
| 2 | AI productivity | Microsoft Copilot Studio |
| 3 | AI strategy | GitHub Copilot |
| 4 | Amazon Bedrock | Prompt engineering |
| 5 | Large language model operations (LLMOps) | Microsoft Copilot |

Figure 7.4.4

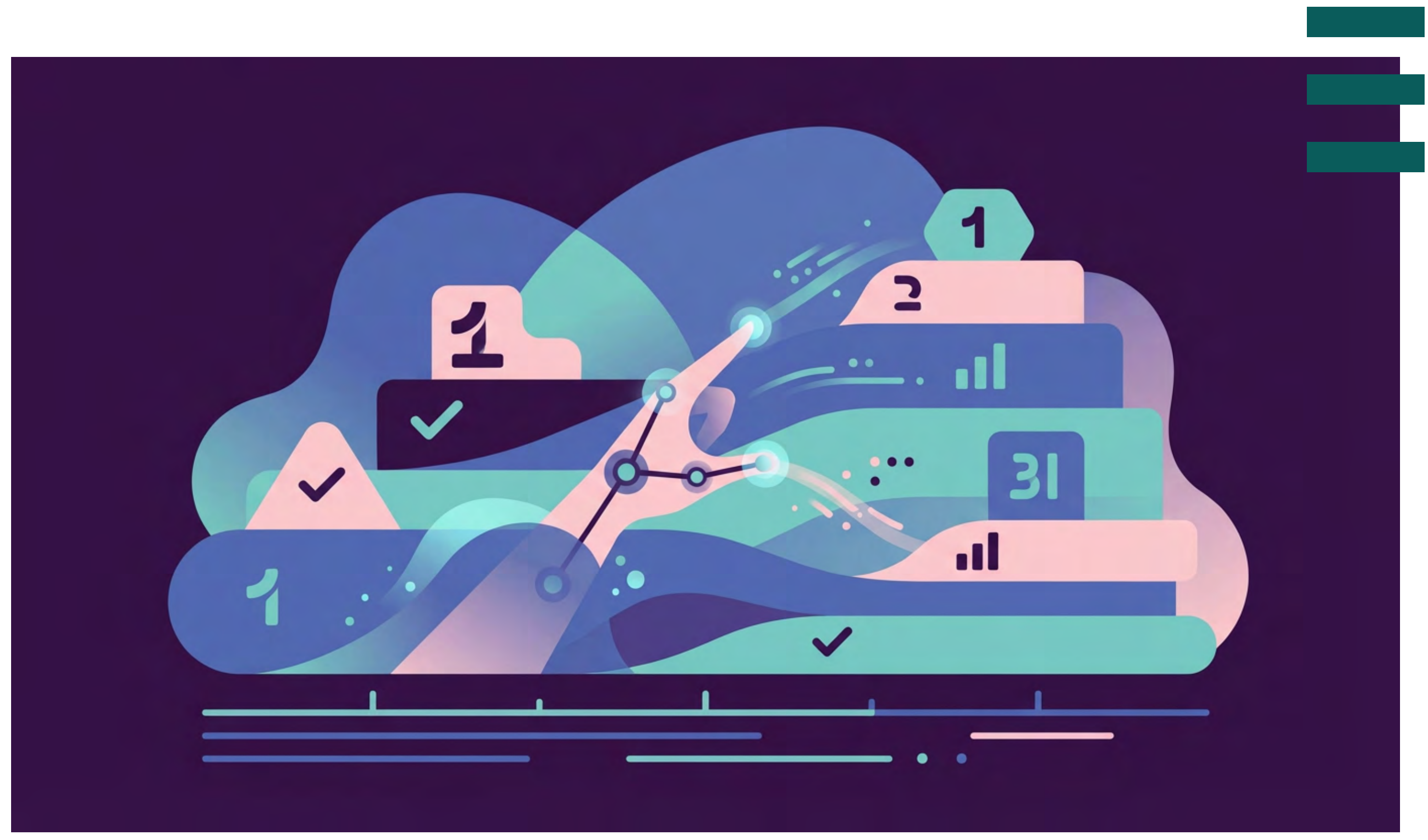

# 8

# Policy and Governance

## Overview

Around the world, AI policy is no longer just about regulation. Governments are also investing to build and maintain their own capacity across the infrastructure, data, talent, and models that make up the technology. The number of countries with formal AI strategies continued to grow, with particular momentum among lower-income economies. Legislative activity continued to grow at every level, though in the United States, federal policy shifted toward deregulation even as state legislatures passed a record number of AI-related bills. Globally, advanced model development and large-scale compute remain concentrated in a small number of countries, while more governments pursue sovereign AI strategies. This chapter’s analysis pulls from national strategy databases, legislative records, congressional witness data, Epoch AI, and public procurement data from the United States and Europe.

## Contents

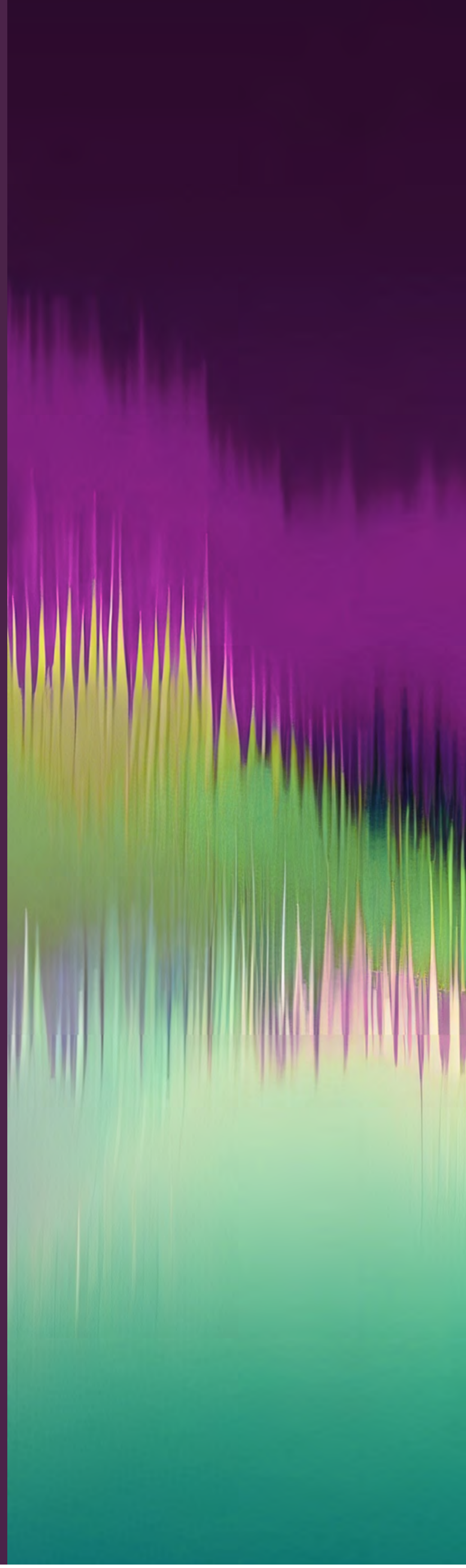

# Chapter Highlights

1. **National AI strategies are expanding fastest among countries that had no formal AI policy five years ago.** In 2024, more than half of newly adopted strategies came from emerging economies and, as of 2025, additional countries across sub-Saharan Africa, Central Asia, and the Middle East have strategies in active development.

2. **AI sovereignty, the goal of gaining more agency over domestic AI capabilities, is emerging as a central principle of national AI policy, but the infrastructure underpinning it is unevenly distributed.** Between 2018 and 2025, Europe and Central Asia expanded state-backed AI supercomputing clusters from 3 to 44. South Asia, Latin America, and the Middle East and North Africa have only reached between 2, 3 and 8 each.

3. **Regions are taking different approaches to data sovereignty.** Through 2024, East Asia and the Pacific had adopted 77 data localization measures, followed by sub-Saharan Africa with 71 and Europe and Central Asia with 66. North America, by contrast, recorded only 3, reflecting a different approach to cross-border data flows.

4. **AI-related witnesses in U.S. congressional hearings have grown twentyfold since 2017.** The number rose from 5 in 2017 to 102 in 2025. Industry's share nearly tripled from 13% to 37%, making it the largest witness group, while academia's share fell to 15%.

5. **U.S. public investment in AI remains modest compared to private-sector spending.** Between 2013 and 2024, the United States invested approximately $20.4 billion in AI-related contracts and grants, against $285.9 billion in U.S. private investment in 2025 alone.

6. **European AI public commitments reached approximately $3.7 billion in contracts over 2013–2024.** The United Kingdom accounted for $1.6 billion, followed by Germany with $505 million and France with $320 million. Recent spending is accelerating as well. In 2024 alone, the U.K. committed $454.4 million (28% of its decade total) and Germany committed $206.6 million (40% of its total).

# 8.1 Major Global AI Policy News in 2025

2025

January 23

### US Executive Order 'Removing Barriers to American Leadership in AI'

The U.S. issues an executive order rescinding earlier AI directives and establishing a new policy to enhance U.S. AI dominance, promote innovation, and remove regulatory barriers.

February 1

### UK to Criminalize AI Tools Used to Generate Child Sexual Abuse Imagery

The United Kingdom positions itself as the first country to introduce laws against artificial intelligence tools used to generate sexualized images of children.

February 2

### 1st Measures of the EU AI Act Come Into Effect

The EU's landmark AI regulation takes effect in its first phase, banning high-risk uses (e.g., predictive policing, emotion recognition) and setting the stage for stricter rules.

February 11

### AI Action Summit: US and UK Refuse to Sign Inclusive AI Declaration

At the 2025 Paris AI Action Summit, the U.S. and UK decline to endorse a declaration signed by 60 countries on inclusive, ethical AI, signaling divergence in governance approaches.

March 14

### China Finalizes Mandatory Labeling Rules for AI-Generated Content

Chinese regulators issue final rules requiring clear labeling of AI-generated and synthetic media, with phased implementation beginning later in the year.

March 24

### Zimbabwe Partners With Nvidia to Launch Africa's 1st AI Factory

Cassava Technologies, founded by Zimbabwean billionaire Strive Masiyiwa, announces a partnership with Nvidia to establish the continent's first dedicated "AI factory."

2025

March 25

### Utah Enacts the Mental Health Chatbot Act (HB 452)

The bill establishes provisions for regulating mental health chatbots that use artificial intelligence technology. It mandates disclosure of AI use, bans advertising within the chat, and prohibits sharing users' personal data.

April 3

### Kigali Summit Highlights Africa's AI Opportunity and Labor Risks

Thousands of delegates convene in Kigali for the inaugural Global AI Summit on Africa to explore how the continent can harness AI for development while mitigating potential disruptions to labor markets.

April 16

### Montana Enacts the Right to Compute Act (SB 212)

The law establishes a pro-innovation legal framework for AI that protects Montanans' rights to own and use computational resources for lawful AI activities without undue government restriction.

May 17

### Africa Declares AI a Strategic Priority: Investment, Inclusion, and Innovation

The African Union region identifies AI as a central strategic priority, emphasizing inclusion, startup funding, and narrowing the digital divide.

May 19

### US Enacts the Take It Down Act, Targeting Nonconsensual Intimate Imagery—Including AI Deepfakes

The law is designed to address the distribution of nonconsensual intimate imagery; it explicitly covers deepfake content and strengthens removal/accountability expectations.

June 17

### Gavin Newsom–Commissioned California AI Policy Report Warns of 'Irreversible Harms'

California's state-level report highlights AI threats including biological and nuclear misuse and proposes safety, transparency, and whistleblower frameworks—potentially offering a national blueprint in the absence of federal law.

2025

June 17

### G7 Issues Joint Statement Reaffirming Cooperation on AI Governance

G7 leaders release a joint declaration committing to coordination on AI safety, risk management, and standards for advanced AI systems.

June 22

### Texas Signs the Responsible Artificial Intelligence Governance Act (TRAIGA) Into Law

Passed in June 2025 and taking effect in 2026, the state law sets strict rules for high-impact AI, including bans on uses that incite harm, violate constitutional rights, or discriminate against protected classes.

July 1

### US Senate Strikes 10-Year Federal Moratorium on State AI Regulation

The Senate removes a proposed federal ban on state-level AI regulation from a major spending bill, allowing states to proceed with their own AI oversight laws.

July 10

### EU Releases Voluntary Code of Practice for General-Purpose AI

The European Commission publishes a code to guide businesses in complying with the upcoming EU AI Act rules for general-purpose models, covering transparency, copyright, and safety.

July 23

### US Launches 'America's AI Action Plan' and 3 Executive Orders

The White House releases a broad AI strategy covering innovation, infrastructure, and diplomacy, plus executive orders on data centers, exports, and government procurement.

July 26

### China Announces Action Plan for Global AI Governance

At the 2025 World AI Conference, China's Premier Li Qiang unveils a 13-point road map to advance global AI coordination and standards.

2025

July 30

### Creators' Organizations Condemn EU AI Act Implementation

A coalition of 38 global creative-industry bodies issues a joint statement criticizing the EU's AI Act as undermining cultural rights and favoring model-developers.

August 2

### EU's General-Purpose AI Obligations Take Effect

Under the EU AI Act, obligations for providers of general-purpose AI models begin to apply, requiring risk assessments, transparency disclosures, and mitigation measures for systems with systemic risk.

August 26

### AI Companion Lawsuits Prompt Renewed Scrutiny of Emotional AI Safeguards

Following a widely reported teen suicide linked to interactions with an AI companion, U.S. lawmakers and regulators increase scrutiny of AI companion systems and child-safety safeguards.

August 26

### UN Launches Global Scientific Panel and Global Dialogue on AI Governance

The UN General Assembly approves the creation of an Independent International Scientific Panel on AI and a Global Dialogue on AI Governance to provide coordinated scientific guidance and facilitate international cooperation on AI regulation.

September 17

### Italy Becomes First EU Member State to Pass an AI Law

Italy advances national AI legislation intended to complement EU-level regulation, reflecting member-state moves to define institutional roles and national implementation.

September 29

### California Enacts Landmark AI Safety/Transparency Law (SB 53)

California Gov. Newsom signs SB 53 requiring large AI-model developers to disclose safety protocols and incident reports and to protect whistleblowers.

2025

October 8

## European Commission Launches the European Strategy for AI in Science

The Commission announces a strategy to reinforce Europe's technological and scientific leadership and competitiveness by harnessing the potential of AI technologies in science and supporting scientists to adopt them in their research. The strategy contributes to the AI Continent Action Plan and was presented alongside the Apply AI Strategy, which aims to speed up AI adoption in key business and industrial sectors.

October 13

## California Enacts New AI Laws

California enacts multiple AI-related bills—SB243 regulating companion bots, AB853 requiring gen AI developers to ensure their tools' content includes provenance data, and AB621 extending existing state law on nonconsensual deepfakes.

November 5

## UNESCO Adopts Global Standards on Neurotechnology

UNESCO approves international standards covering AI-driven neurotechnology—"neural data" rights, mental privacy, and emerging regulation.

November 24

## US Executive Order Launches the Genesis Mission

An executive order launches the Genesis Mission, a major national initiative to accelerate scientific discovery and technological innovation using artificial intelligence. The mission, compared in ambition to the Manhattan Project, tasks the Department of Energy with leading the effort.

December 11

## Launch of Pax Silica: US-led AI and Technology Supply-Chain Cooperation Initiative

At a summit convened by the U.S. Department of State, the Pax Silica Declaration is signed by multiple countries to strengthen trusted technology and AI-relevant supply chains, spanning semiconductors, data infrastructure, and AI hardware cooperation.

December 12

## US Executive Order Seeks to Curb State-Level AI Laws

A U.S. executive order aimed at limiting or preempting state AI regulation to "enhance the United States' global AI dominance through a minimally burdensome national policy framework for AI."

# 8.2 National AI Strategies

As AI becomes more central to economic development and national competitiveness, more governments are moving to formalize their approach through national AI strategies. This section draws and builds on data from Oxford Insights to track the adoption and geographic expansion of formal national strategies over time. The dataset captures what has been published, rather than how or how well it has been implemented, so results should be viewed as policy intent rather than actual progress.

## By Geographic Area

More countries adopted national AI strategies in 2024 and 2025, especially within emerging economies (Figure 8.2.1). This marks a shift in AI governance from earlier years, as countries that have historically played a smaller role in AI policymaking are now putting formal national strategies in place.

New frameworks have surfaced across sub-Saharan Africa (such as Ethiopia, Ghana, and Nigeria), South and Central Asia (notably Sri Lanka and Nepal), and Latin America and the Caribbean (including Costa Rica and Jamaica). With strategies already under development in Mexico and South Africa, this trend underscores AI policy's increasing global reach. High-income economies continue to contribute new strategies as well, though at a slower pace and with a focus on consolidating earlier frameworks. European countries, such as Malta, have released updated strategies to align with EU AI Act requirements.

As more countries adopt national AI strategies, there is a rising consensus that AI serves as a lever to bolster state capacity. International cooperation, technical assistance, and policy diffusion also play an important role in this expansion. However, the next challenge is implementation and strengthening regulatory capacity, particularly in Africa, where many countries still lack formal strategies and risk falling behind in AI governance and readiness.

**Countries with a national strategy on AI**
Source: Oxford Insights, 2024; AI Index, 2026 | Chart: 2026 AI Index report

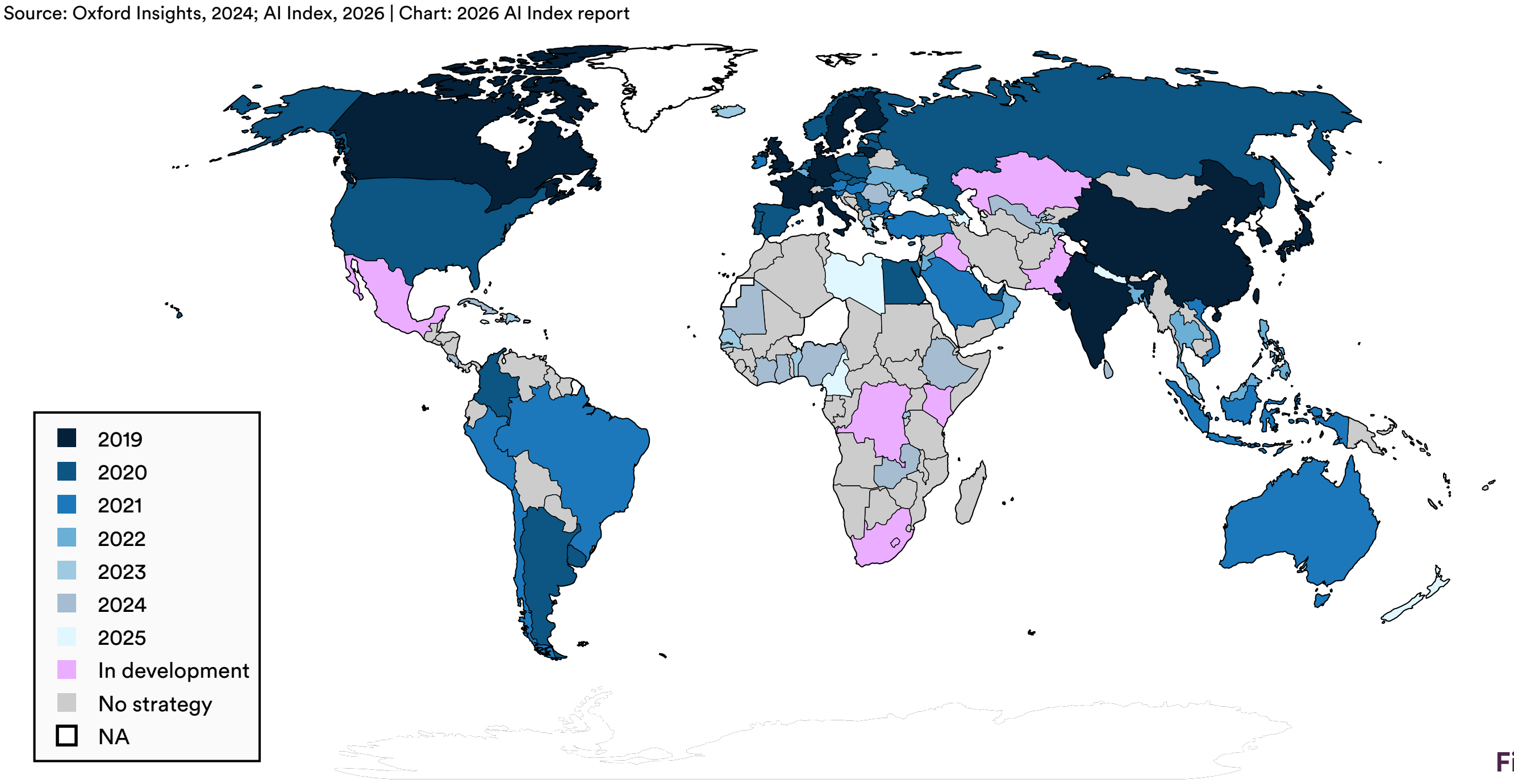

Figure 8.2.1

# 8.3 AI Sovereignty

As AI technologies become increasingly central to geopolitics and statecraft, and more countries articulate national strategies, attention has shifted to issues of control, capacity, and dependence across the AI stack. In policy terms, AI sovereignty describes a state's capacity to act deliberately and make independent decisions over the development, deployment, and governance of AI systems within its jurisdiction and, in some cases, beyond it through standards, trade, and regulation.

As AI systems have become more central to economic policy, national security, global trade, and cultural autonomy, sovereignty debates have expanded beyond data and infrastructure to include other parts of the AI stack, including compute, model development, talent, and responsible AI deployment. Many of these debates build on earlier discussions of digital and technological sovereignty, which focused on government authority over digital infrastructure, data flows, capabilities, and technology supply chains. Today, governments are pursuing a range of approaches across these layers, including investments, procurement policies, regulatory measures, international partnerships, and supply-chain strategy.

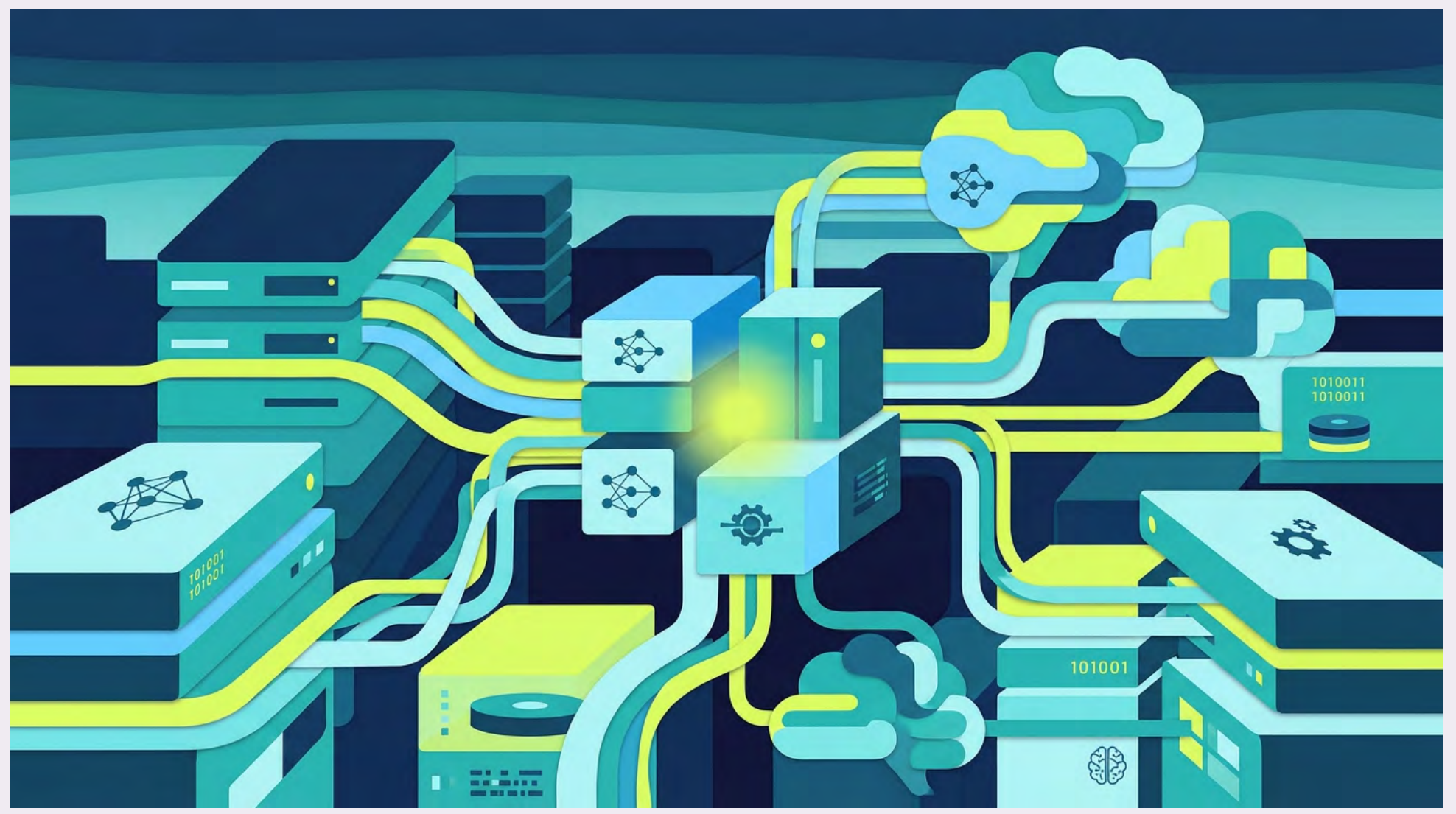

This section draws on data from Epoch AI, Zeki, and Brookings to examine how sovereignty dynamics are evolving across compute infrastructure, data, models, applications, and talent.

## Infrastructure Sovereignty

Domestic AI computing infrastructure, including high-performance GPU clusters and AI-optimized supercomputers, has become one of the most visible areas of AI sovereignty investment. In policy discussions, domestic compute capacity is often framed around reducing reliance on foreign providers, limiting exposure to extraterritorial jurisdiction, and providing continuity of access for government agencies, research institutions, and domestic firms in scenarios such as export controls, geopolitical disputes, and supply-chain disruptions.

In this context, the scale and availability of state-owned or state-backed AI supercomputing facilities is increasingly used as an indicator of "compute sovereignty" alongside related measures such as domestic access to advanced chips, cloud capacity, and the governance arrangements that determine who can use these resources and for what purposes.

Based on data from Epoch AI tracking large-scale GPU clusters used for training advanced AI models,[1] state-backed AI supercomputing expanded across most regions between 2010 and 2025 (Figure 8.3.1). The sharpest acceleration was in Europe and Central Asia, where the number of clusters grew from 3 to 44 between 2018 and 2025, largely driven by coordinated initiatives such as the European High Performance Computing Joint Undertaking (EuroHPC JU). North America grew nearly sevenfold over the same period to reach 41 clusters, a substantial expansion given its comparatively high baseline, reflecting a policy shift toward dedicated national AI research infrastructure, including through the U.S. National AI Research Resource (NAIRR). East Asia (excluding China) grew about fourfold. By contrast, South Asia, the Middle East and North Africa, and Latin America and the Caribbean each only doubled or tripled, reaching 2, 3, and 8 clusters, respectively, by 2025. Several initiatives to expand capacity are already underway in these regions, though planned systems are not included here as they are subject to change and carry inherently lower confidence.

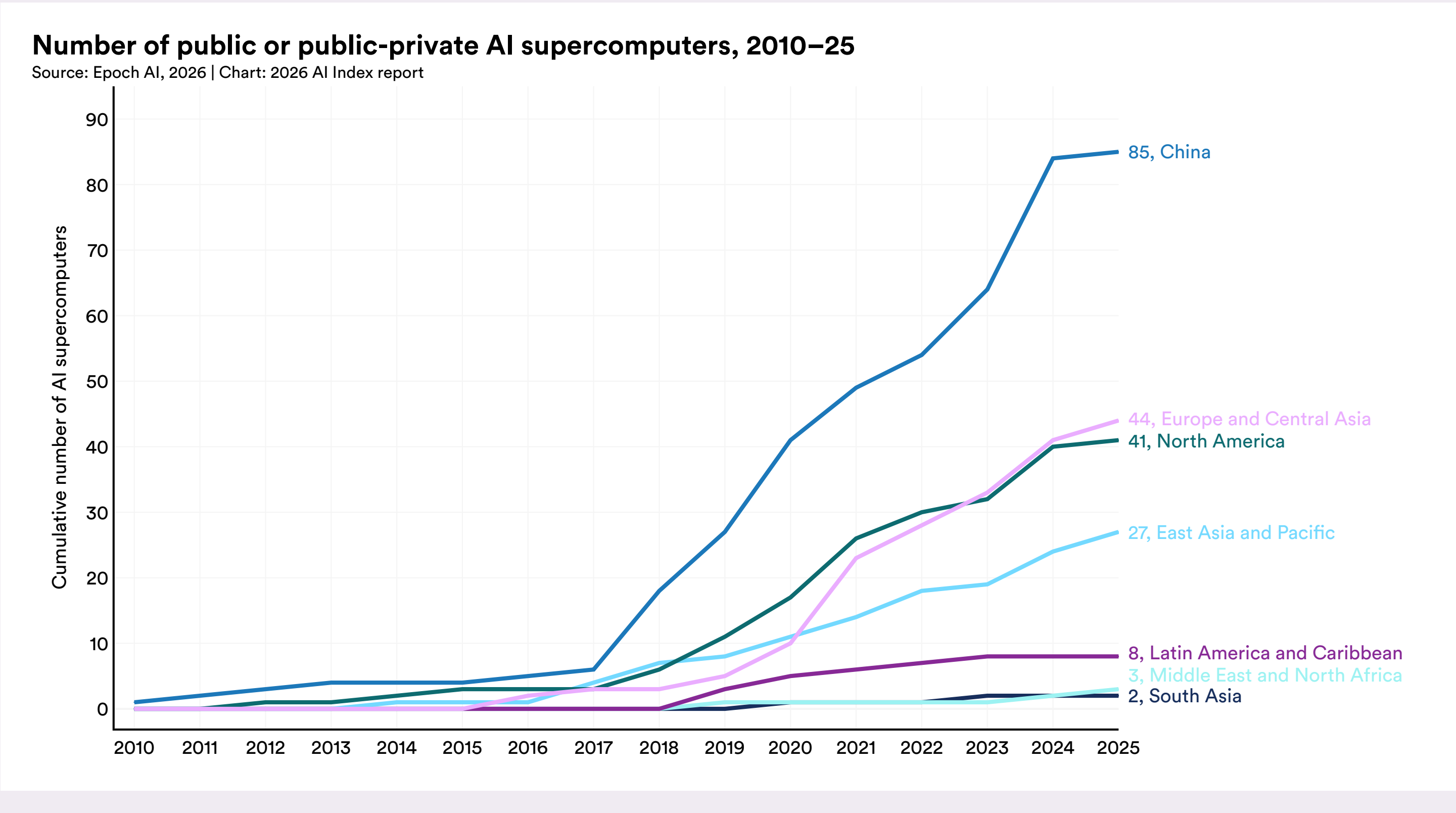

Figure 8.3.1

Although privately owned clusters account for most large-scale AI compute capacity globally, state-owned and public-private clusters have also grown steadily across most regions. In practice, this distinction is clouded because many private clusters remain accessible to public sector actors through commercial cloud services. Public-private partnerships can involve both domestic and international actors (Figure 8.3.2). OpenAI's Stargate project, for example, extends beyond the United States through country-level partnerships across regions, including the United Arab Emirates, the United Kingdom, Argentina, South Korea, India, and Norway. Nvidia's "AI Factory" is a different approach in which in-country compute capacity is typically built in partnership with domestic telecommunications providers, a model that has expanded rapidly by catering to governments' sovereign AI ambitions. These initiatives illustrate how private firms are playing an increasingly central role in building what many governments designate as national AI infrastructure.

1 Because the underlying dataset captures frontier AI training capacity rather than the full universe of compute resources, the figure should be interpreted as a proxy for sovereign high-end AI compute infrastructure rather than a comprehensive measure of national compute capacity.

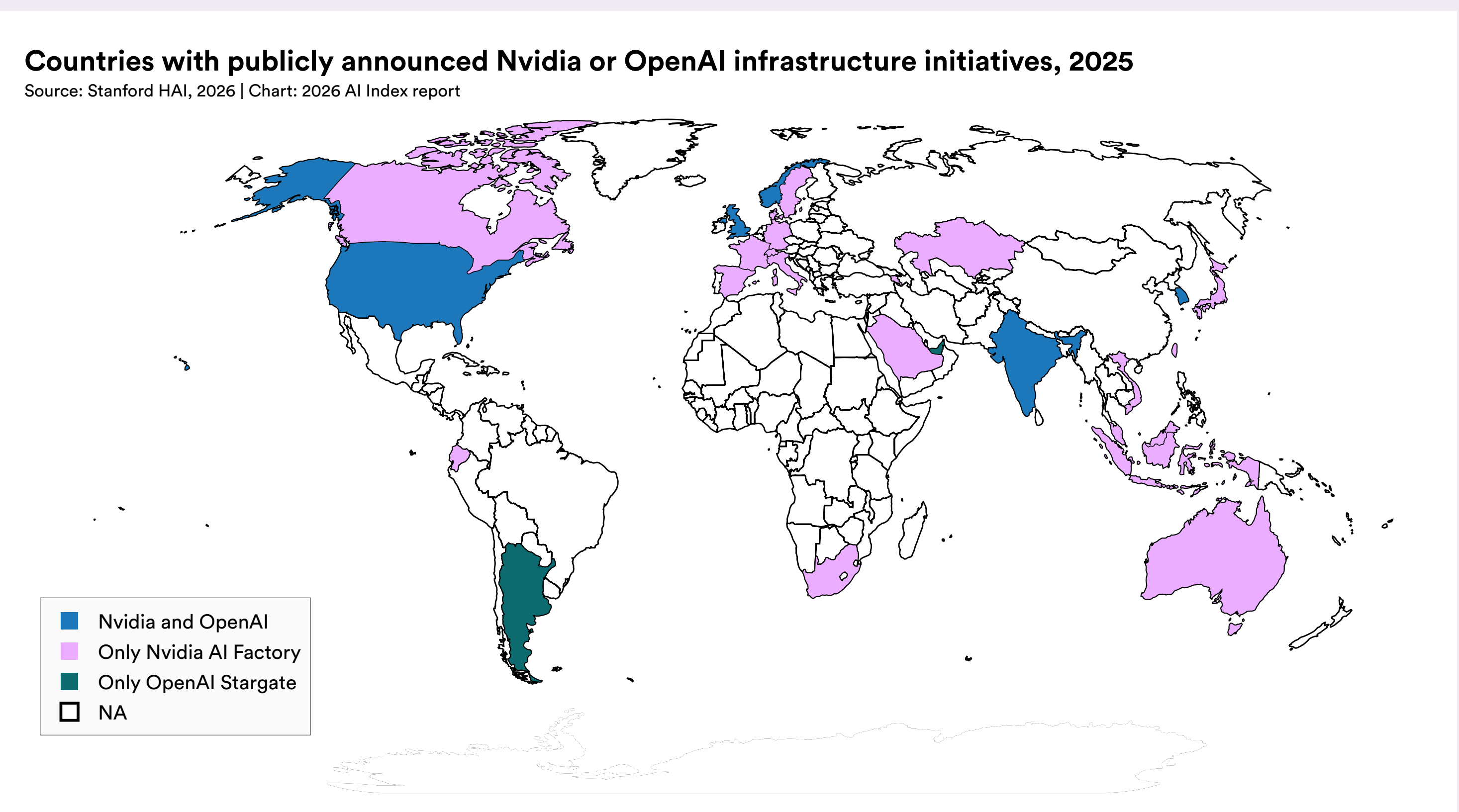

Figure 8.3.2

## Data Sovereignty

While infrastructure sovereignty focuses on control over compute resources, data sovereignty concerns the extent to which states or local actors have agency over how their data is collected, stored, processed, and transferred. One common approach is to adopt data localization measures[2] that require certain categories of data to remain within national borders or to impose restrictions on cross-border data transfers. As AI systems grow increasingly dependent on vast, diverse datasets, data sovereignty has emerged as a central dimension of the broader AI sovereignty debate.

Data localization measures have increased across nearly every region since 2000 (Ferracane et al., 2025, Figure 8.3.3). The steepest rise in adoption begins around 2016, coinciding with the implementation of GDPR in Europe and the subsequent "Brussels Effect," whereby other nations adopted similar frameworks. Regional patterns fall into three broad clusters: high-localization regions led by East Asia and Pacific (77 measures), followed closely by sub-Saharan Africa (71) and Europe and Central Asia (66); moderate-localization regions including the Middle East and North Africa (44), Latin America and the Caribbean (36), and South Asia (24); and North America, which remains a striking outlier at just 3 measures, reflecting a long-standing "flow-first" policy orientation.[3]

2 While there is no single official definition, data localization measures are broadly understood as explicit requirements that data be stored and/or processed within a domestic territory, encompassing both outright storage mandates and conditional restrictions on cross-border transfers (López González et al., 2022).

3 This pattern is well documented (World Bank, 2025), reflecting a general tendency—particularly within the U.S.—to favor free data flows from which its firms disproportionately benefit. As one example, U.S. diplomats have recently been tasked with pushing back against other countries' data sovereignty initiatives (Reuters, 2026). At the same time, emerging restrictions such as those on the transfer of bulk sensitive personal data to "countries of concern" signal a growing willingness to impose targeted controls.

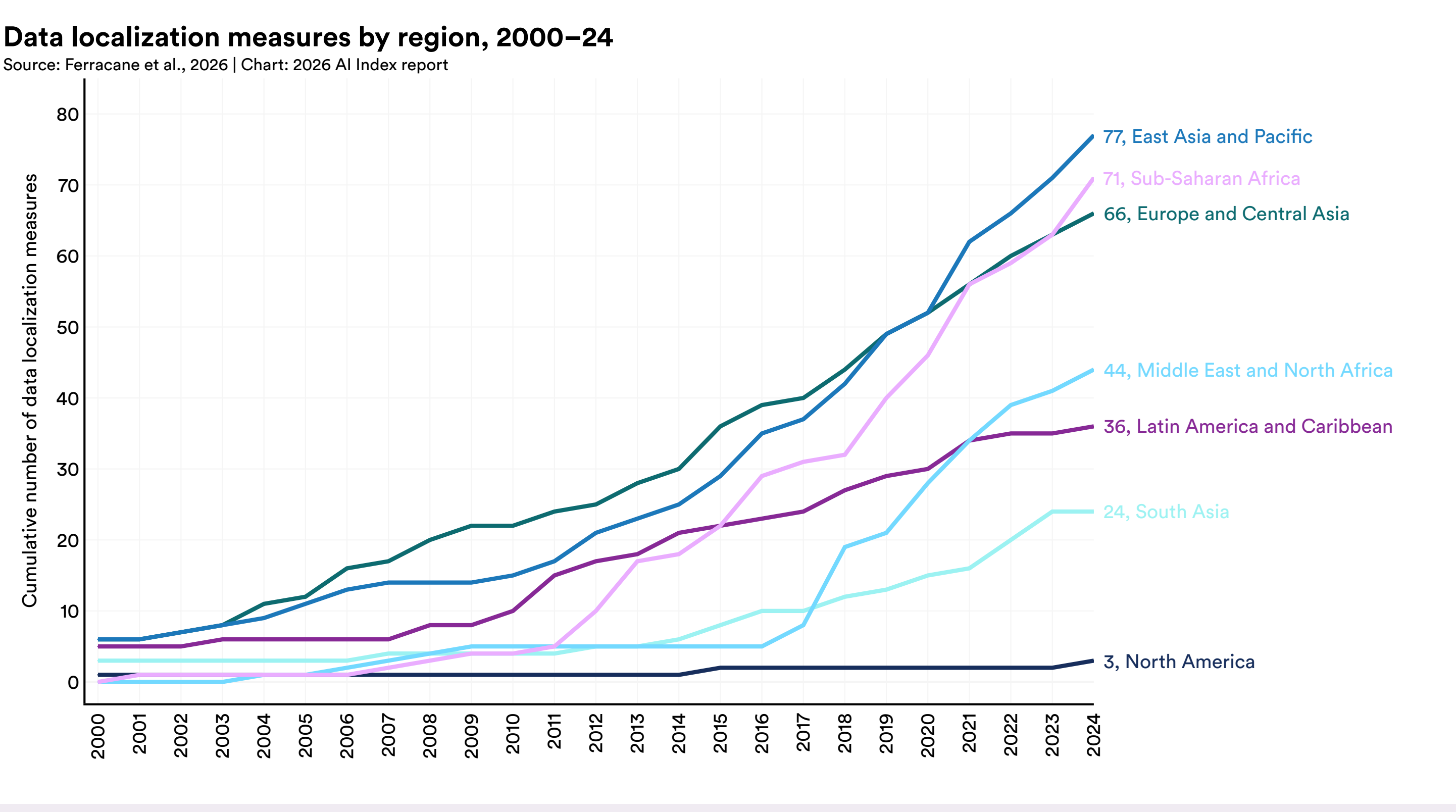

Figure 8.3.3

## Model Sovereignty

Model sovereignty concerns a state's capacity, influence, and control over the development and deployment of AI models. As discussed in Chapter 1, advanced AI model development has historically been concentrated in a small number of technology hubs, primarily in the United States and China. While that persists, open-source frameworks have lowered barriers to entry, and a growing number of regions are building and deploying their own models (Figure 8.3.4). This trend reflects a growing emphasis on localizing model development even if countries can deploy U.S.- or Chinese-made models.

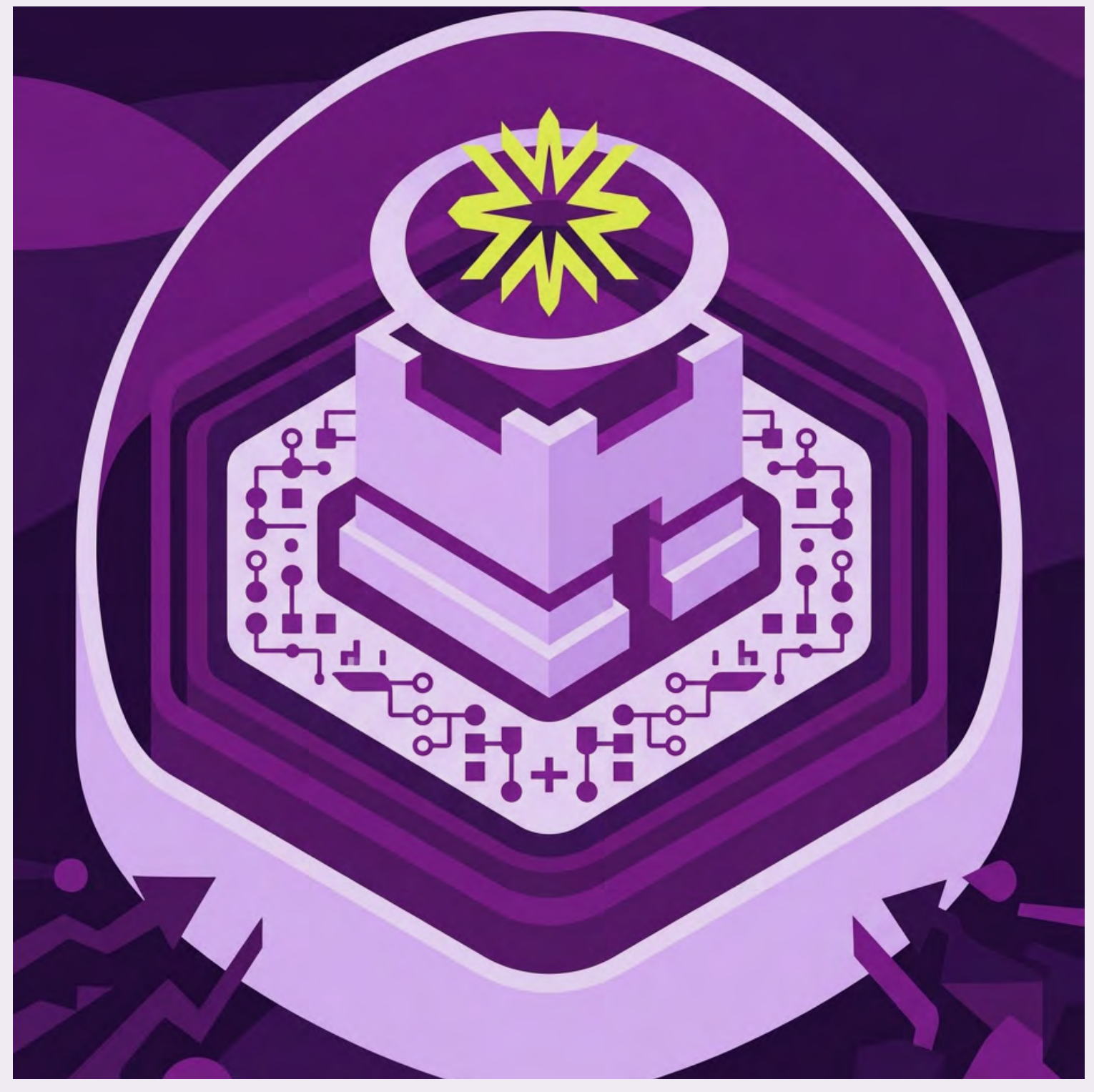

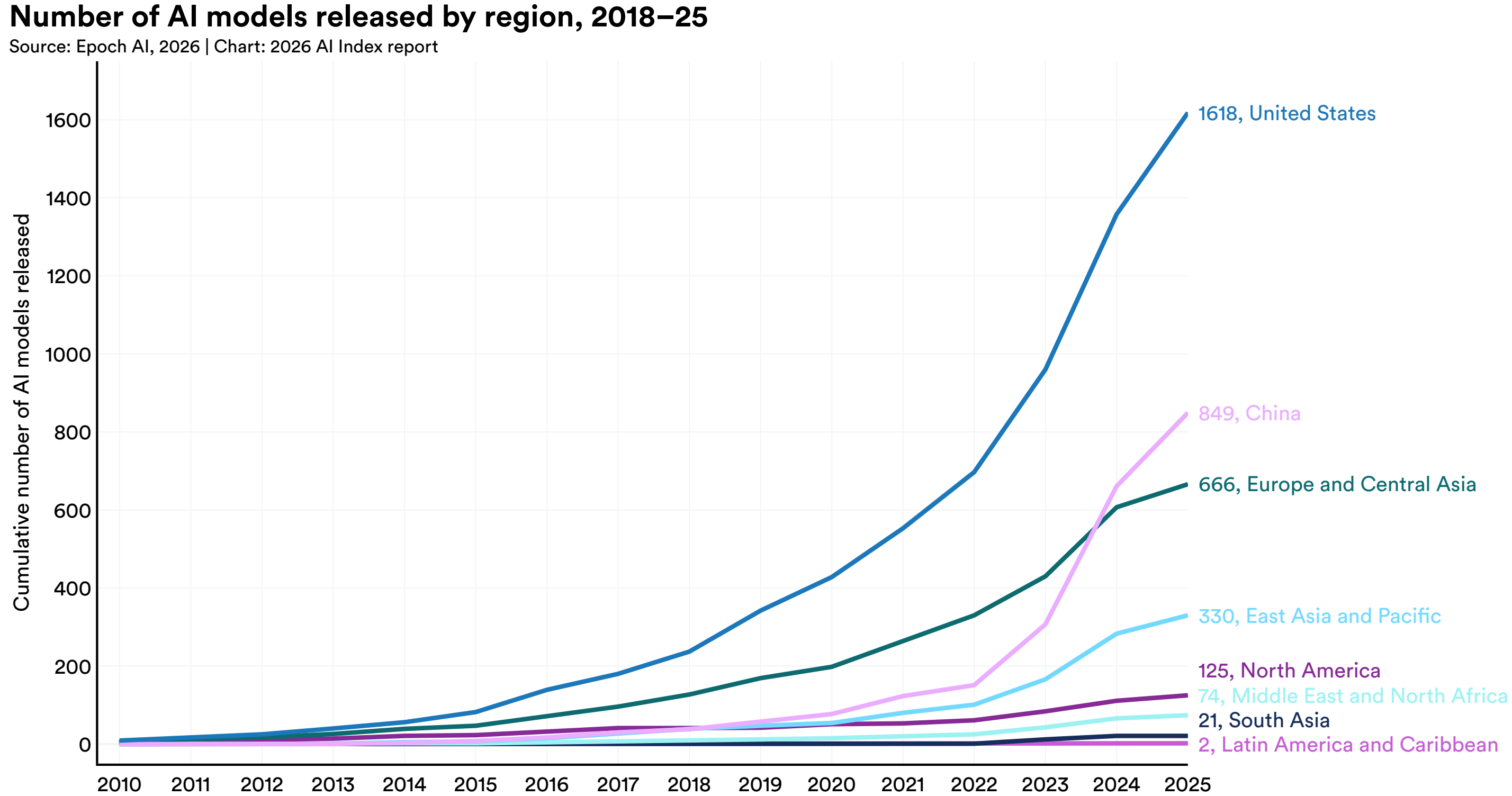
Number of AI models released by region, 2018–25
Source: Epoch AI, 2026 | Chart: 2026 AI Index report

**Figure 8.3.4**

Based on Epoch AI data tracking publicly reported model releases, cumulative U.S. model releases grew from 237 to 1,618 between 2018 and 2025. China exhibits a similar acceleration between 2022 and 2025, where model releases more than quintupled from 151 to 849, suggesting a rapid scaling of domestic capabilities and intensified competition with U.S. model development. These figures reflect the full range of publicly documented model releases by Epoch AI, including smaller and less prominent ones. This differs from the notable model dataset used in Chapter 1, which applies narrower criteria such as state-of-the-art performance and high citation counts. The two datasets can show different year-over-year patterns because the broader count here is more representative of the expanding base of model development, while the subset in Section 1.1 of Chapter 1 is more sensitive to the changes at the frontier. Europe and Central Asia show steady growth, increasing from 127 to 666 models over the same period, with the United Kingdom (229 models) and France (141) as leading contributors, while Canada (captured by the North America region) trails in fifth place with 125 models.

East Asia and the Pacific (excluding China) grew from 39 to 330 models by 2025, while the Middle East and North Africa, South Asia (largely driven by India), and Latin America and the Caribbean reached only 74, 21, and 2 models, respectively. Several of these regions are beginning to champion national or regional model initiatives, such as Chile's Latam-GPT, the UAE's Falcon series, and Singapore's SEA-LION, though their overall footprint remains limited. These figures should also be interpreted as conservative estimates, as model documentation and reporting are less systematic in these regions. The growing ecosystem of smaller and language-specific models in sub-Saharan Africa, for example, are not represented at all.[4]

4 AfriBERTa (Ogueji et al., 2021), AfriTeVa (Jude Ogundepo et al., 2022), AfroLM (Dossou et al., 2022), EthioLLM (Tonja et al., 2024b), EthioMT (Tonja et al., 2024c), and AfroXLMR (Alabi et al., 2022).

Overall, model production remains concentrated, with the United States and China accounting for a disproportionate share of global activity. At the same time, complementary evidence from AI-related GitHub activity suggests that open-source development is diffusing more broadly across regions, even as significant asymmetries in scale and capability persist (see Section 1.5 of Chapter 1 for full analysis).

## Application Sovereignty

A fourth dimension of AI sovereignty concerns the capacity, agency, and control over the downstream deployments of AI systems within a nation's public and private sectors. Application-level sovereignty encompasses domestic procurement policies; sector-specific regulatory requirements in domains such as health, finance, and defense; and the Digital Public Infrastructure (DPI) upon which AI applications increasingly sit. Together, these determine how much a country can shape the AI systems with which its institutions and citizens interact.

Public investment in AI-related contracts and grants offers one observable signal of how governments are implementing this form of sovereignty in practice (see Section 8.5 for a full analysis of public AI investment trends across the U.S. and Europe). Beyond public investment, however, comprehensive cross-country data on sovereignty-oriented AI procurement preferences, sectoral deployment mandates, and DPI utilization for AI remains limited, reflecting both the novelty of the concept and the opacity of procurement data across jurisdictions.

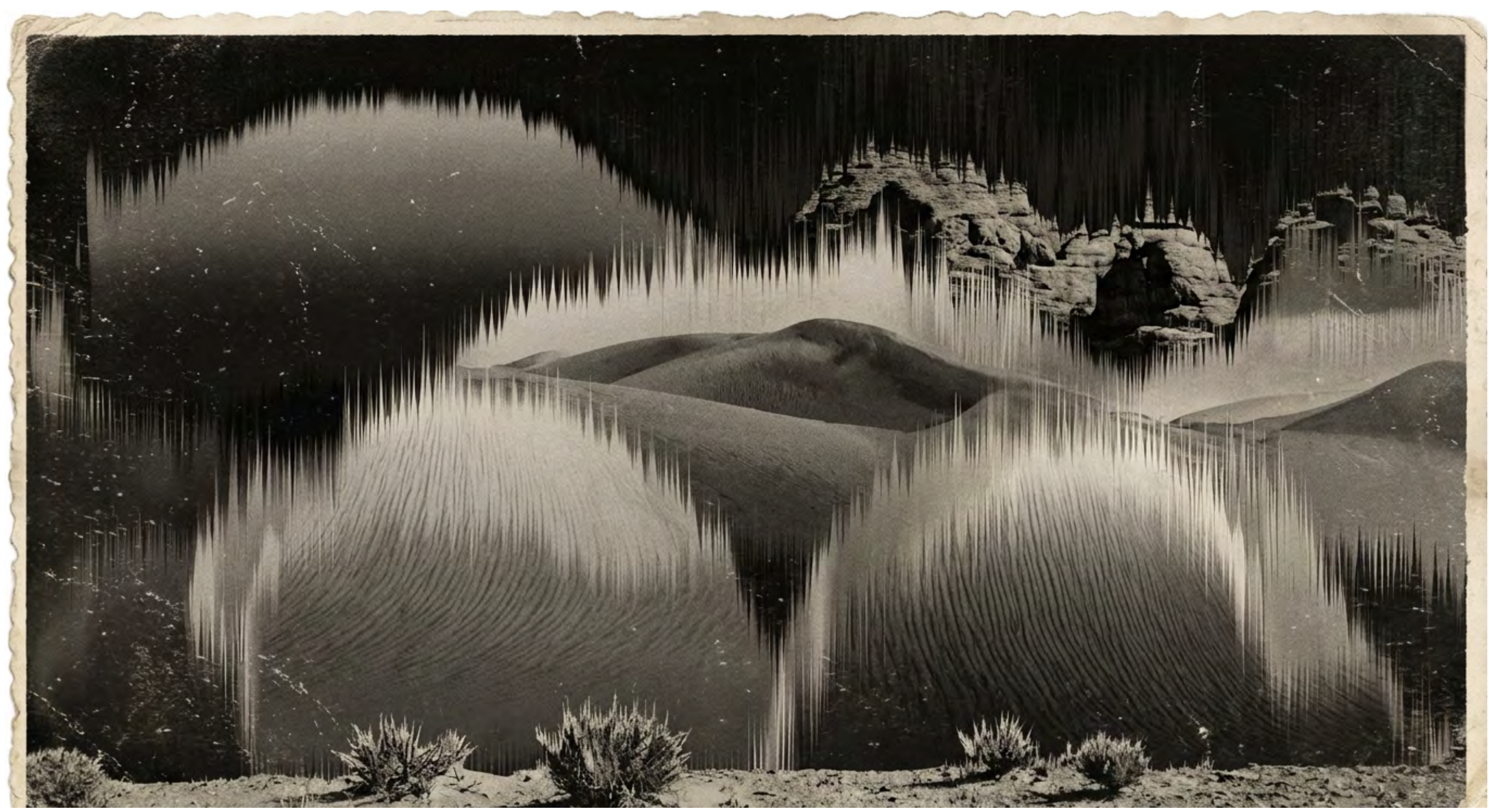

**Distribution of AI applications by geographic area, type, and funding**

Source: Brookings, 2026 | Chart: 2026 AI Index report

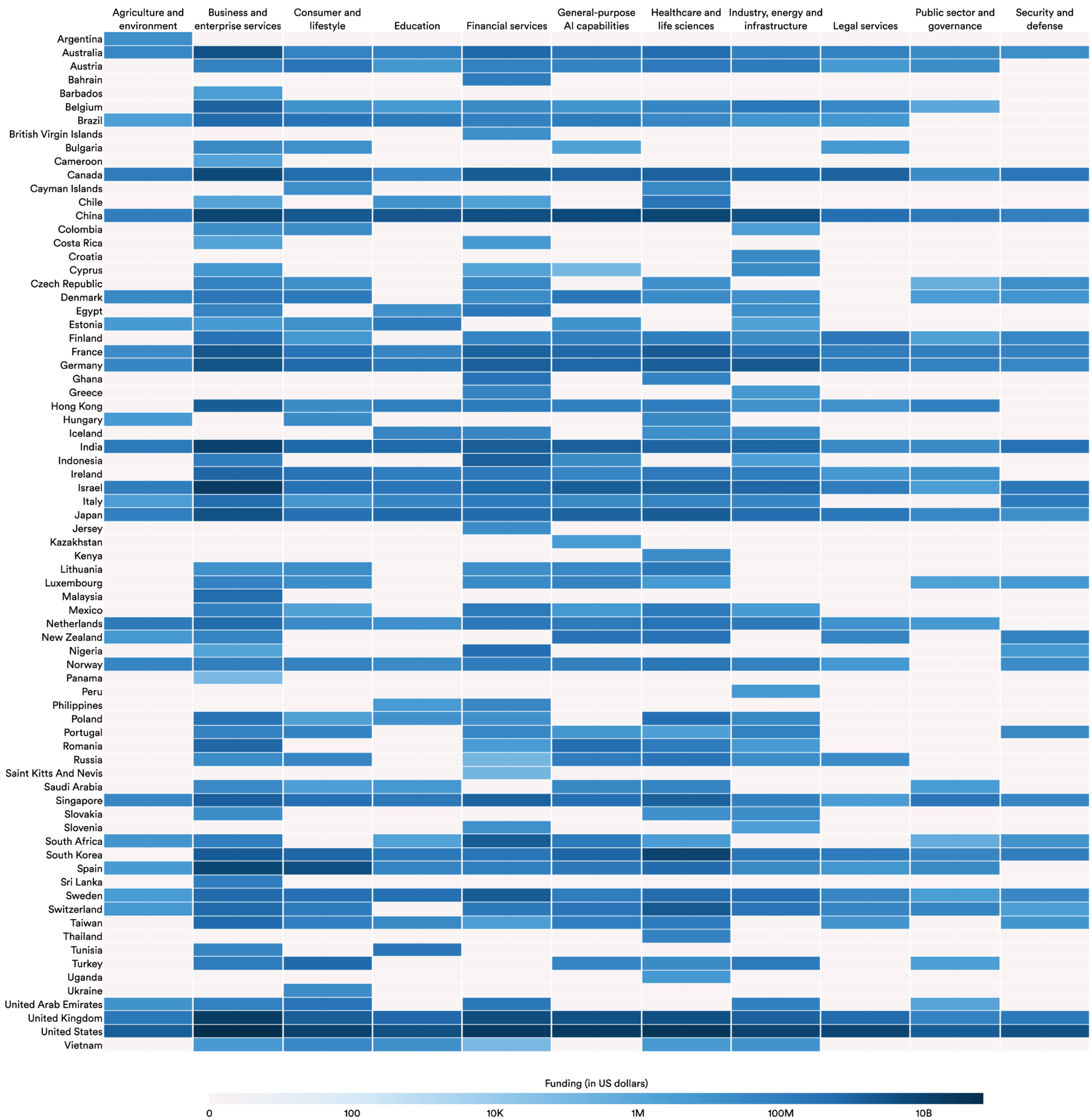

Figure 8.3.5

Countries are increasingly concentrating AI investment in domains aligned with institutional strengths and policy priorities (Figure 8.3.5). A small set of countries, most prominently the United States, China, and several European economies (UK, Germany, France), show high-intensity investment across nearly all application categories. Most other countries display concentrated areas of focus, indicating selective investment rather than full-spectrum capability.

Within Europe, Germany's strength is in industrial applications (particularly manufacturing) and Estonia's is in education technologies. Sub-Saharan African countries show stronger engagement in financial applications, led by South Africa. Latin America presents a more uneven pattern, with Brazil investing broadly while countries such as Chile and Argentina concentrate on more specific domains, including healthcare and agricultural applications, respectively. In the Middle East and North Africa, similar dynamics emerge, with Israel standing out for its specialization in security and defense applications, consistent with its broader positioning as a global cybersecurity hub. The application layer, as it is less concentrated than the model or compute layer, offers more space for countries to develop niche specializations, enabling them to exercise greater autonomy both domestically and internationally over these systems.

## Talent Sovereignty

A fifth dimension of AI sovereignty is a nation's ability to develop and retain the human capital needed to build, deploy, and govern AI systems. Talent sovereignty includes two closely related dynamics: workforce capacity, the domestic stock of AI skills and expertise, and talent mobility (the extent to which countries attract, retain, or lose AI specialists across borders). The country-level distribution and mobility patterns of AI authors and inventors offer a direct window into this dimension and is also discussed in detail in Section 1.8. of Chapter 1. Broader labor market indicators, including AI talent concentration and workforce trends across countries, are examined in Section 4.4 of Chapter 4.

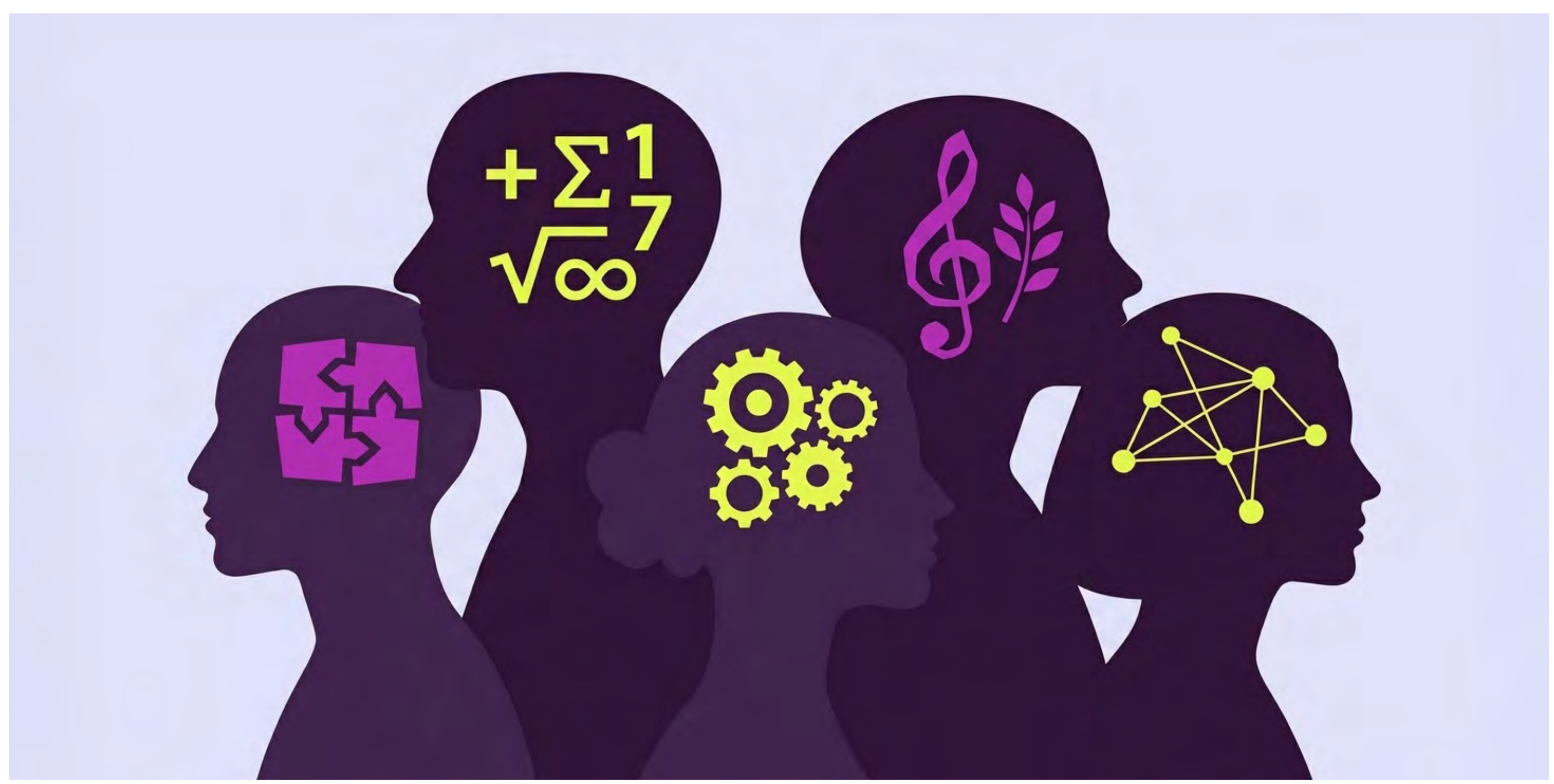

**Inflow and outflow of top AI authors and inventors by region, 2016–25**
Source: Zeki Data, 2025 | Chart: 2026 AI Index report

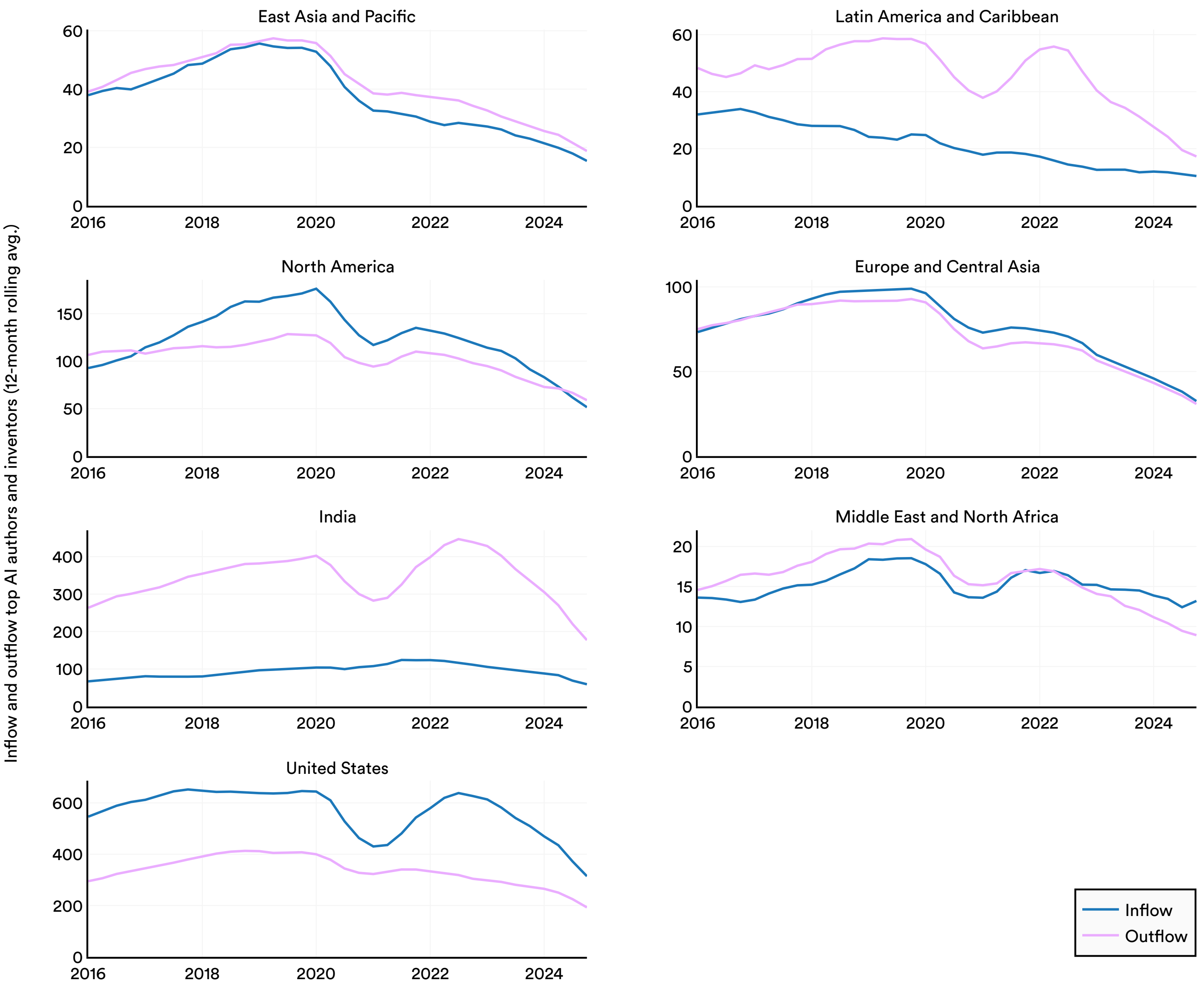

**Figure 8.3.6**

Cross-border AI talent circulation has slowed recently, even where net flows remain stable (see Section 1.8 of Chapter 1). Both inflows and outflows are declining, suggesting that talent is increasingly staying within national or regional systems rather than circulating globally (Figure 8.3.6). The United States is currently the primary global attractor of top AI talent, though its lead is rapidly narrowing. By contrast, India is transitioning from a net exporter to a net absorber of talent. The near mirror-image relationship between the two countries reflects the well-documented fact that the U.S. has been the main destination for Indian AI talent. At the same time, the Middle East and North Africa are making incremental gains, a sign that new talent hubs are emerging with the support of targeted policy and investment.

# 8.4 AI and Policymaking

Legislative activity is a signal of how governments are responding to AI beyond national strategies. This section tracks AI-related bills passed across G20 countries, drawing on data from Digital Policy Alert.[5] The dataset covers enacted legislation, not proposed or pending bills. Counts should be interpreted with caution as they may understate the actual volume of AI-related policymaking, since large omnibus bills that contain multiple AI provisions are counted as a single piece of legislation. Volume is also not a measure of significance, as a single major law can carry more impact and enforcement weight than dozens of narrower ones.

## Global Legislative Records on AI

In 2016, there were no AI-related laws on record among G20 countries. Since then, legislative activity has been on the rise, though the total number of laws passed varies widely by country (Figures 8.4.1 through 8.4.3). Between 2016 and 2025, the United States passed the most AI-related bills, 25 in total, followed by South Korea with 17. Japan, France, and Italy were also relatively active, with each passing 9 to 10 laws. Over the same period, other countries, such as Russia and Saudi Arabia, passed very few, if any, AI-specific legislation. Similar to other aspects of AI development, such as investment and research output, policymaking is expanding but unevenly.

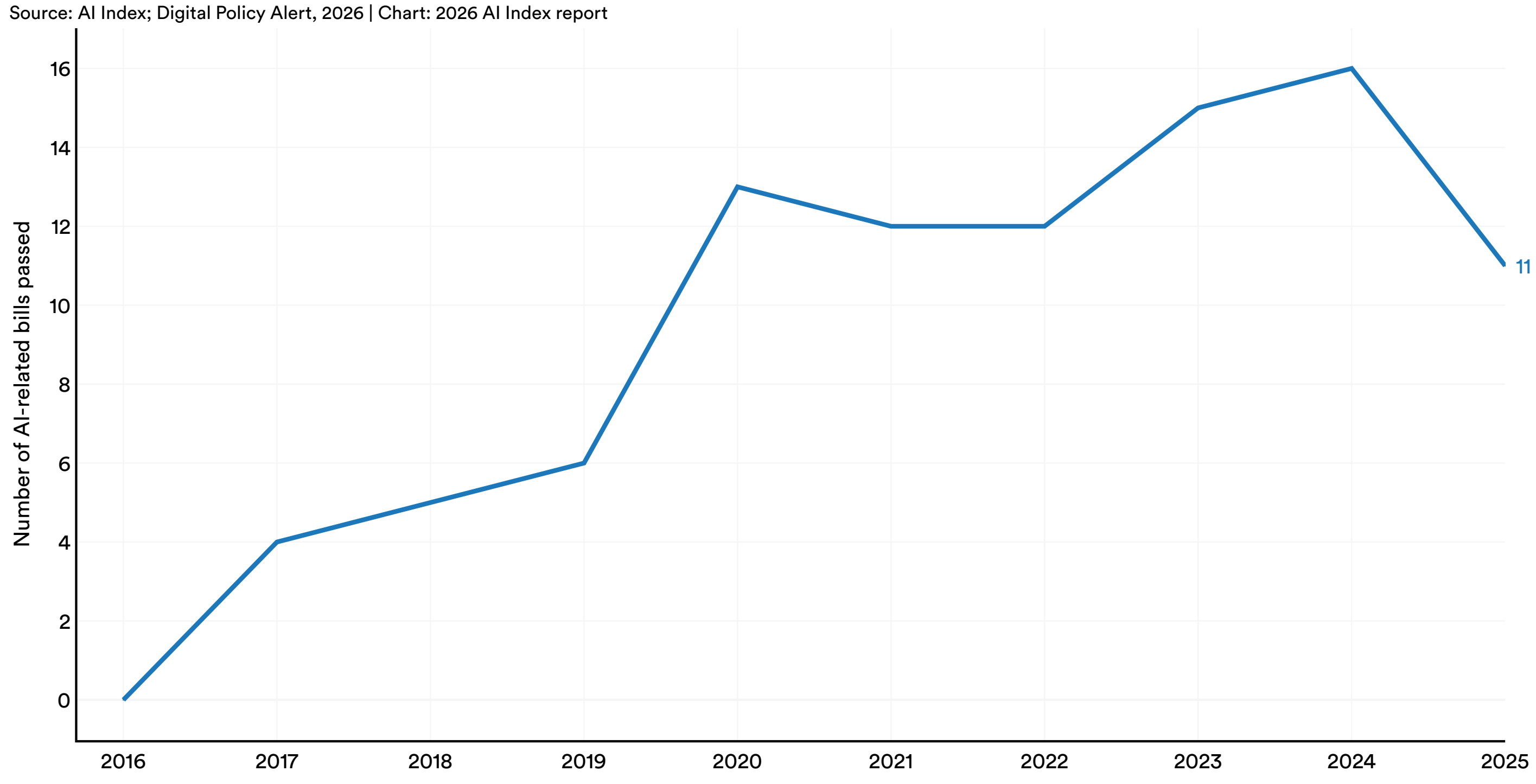
Number of AI-related bills passed into law in G20 countries, 2016–25
Source: AI Index; Digital Policy Alert, 2026 | Chart: 2026 AI Index report

Figure 8.4.1

5 Numbers may differ from what was reported in the 2025 AI Index report, as the methodology has changed. In addition, the following charts focus only on G20 countries.

## Number of AI-related bills passed into law in G20 countries, 2016–25

Source: AI Index; Digital Policy Alert, 2026 | Chart: 2026 AI Index report

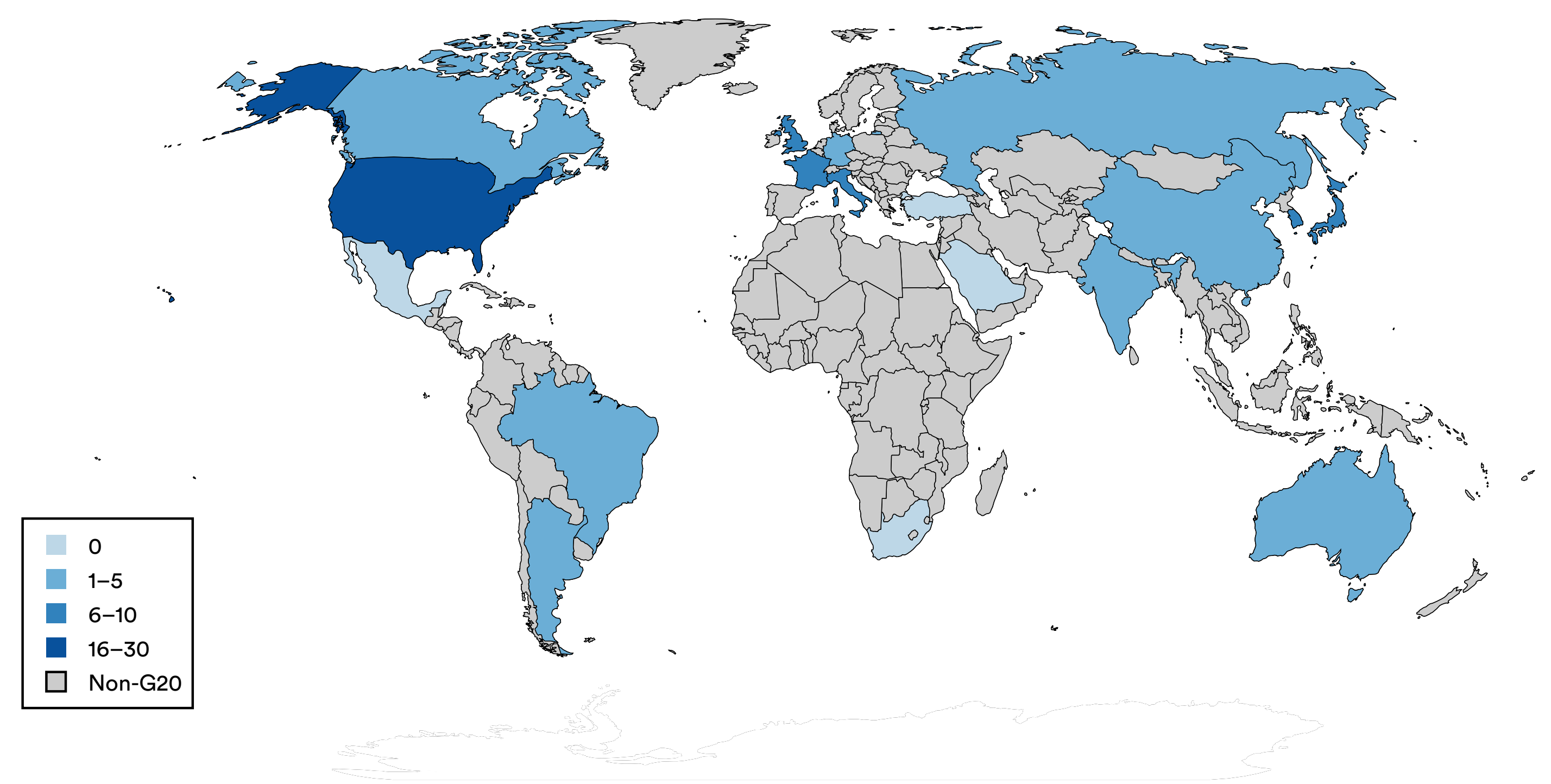

Figure 8.4.2

## Number of AI-related bills passed into law in G20 countries, 2016–25 (sum)

Source: AI Index; Digital Policy Alert, 2026 | Chart: 2026 AI Index report

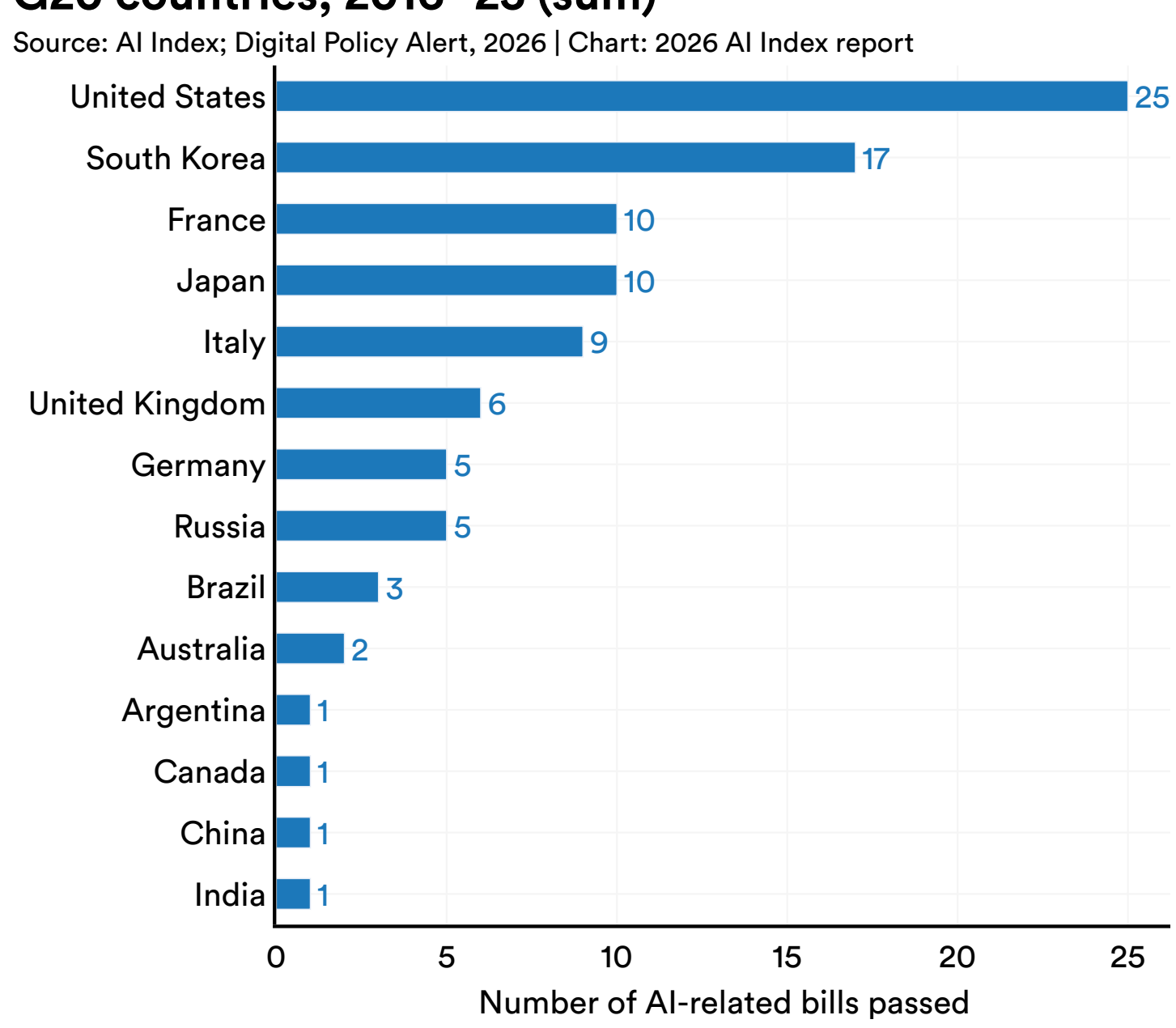

Figure 8.4.3

**HIGHLIGHT:**

## A Closer Look at Global AI Legislation

| Country | Bill name | Description |
|---|---|---|
| United States | Take It Down Act (Tools to Address Known Exploitation by Immobilizing Technological Deepfakes on Websites and Networks Act) | This act criminalizes the nonconsensual distribution and disclosure of intimate visual depictions, including AI-generated or digitally altered deepfakes. It mandates that online platforms remove reported content within 48 hours and strengthens enforcement through federal criminal penalties and a civil cause of action allowing victims to sue for damages. |
| Italy | Provisions and Delegated Powers to the Government on the Matter of Artificial Intelligence (Law No. 132/2025) | This law establishes a national framework for responsible, transparent, and human-centered AI in Italy, aligned with the EU AI Act (Regulation 2024/1689). It sets principles and governance directions for AI strategy and sectoral deployment, and addresses user safeguards, copyright-related issues, and penalty provisions for specified unlawful uses. |
| Japan | Act on the Promotion of Research and Development and Utilization of Artificial Intelligence–Related Technology | This law provides a national framework to promote AI research, development, and utilization by defining responsibilities for the national government, local governments, research institutions, and businesses. It mandates an AI Basic Plan and establishes a prime minister–led AI Strategy Headquarters to coordinate cross-government policy and international cooperation. |
| South Korea | Framework Act on the Development of Artificial Intelligence and the Creation of a Foundation for Trust | This law sets South Korea's overarching policy and governance framework for advancing artificial intelligence while building conditions for trustworthy and ethical AI. It provides the primary statutory basis for government-led AI strategy, coordination, and support measures, and for developing standards and safeguards intended to promote public trust alongside industrial innovation. |

## US Legislative Records

The United States has passed more AI-related laws since 2016 than any other G20 country, making it a useful case to examine how AI is being addressed through domestic policy channels. Within the U.S., AI has become a cross-cutting policy issue that touches on governance, national security, public services, and individual rights. This section tracks AI-related bills enacted at the state level as well as the composition of witnesses at congressional AI hearings and federal regulatory activity.

### State Level

State legislatures have been the active venue for AI policymaking in the United States, particularly in the absence of comprehensive federal AI legislation. Across all states, the total number of AI-related bills passed increased from fewer than 10 in 2020 to 150 in 2025 (Figures 8.4.4 to 8.4.6). However, a small number of states accounted for a disproportionate share of legislation enacted in 2025. California enacted 20 AI-related bills, double the total in New York (10) and two-thirds more than Texas (12), the second most active state. Over the full 2016–2025 period, California had more than double any other state, with 62 bills enacted. Maryland (28), Virginia (25), and Utah (24) also had records that reflect consistent activity across multiple legislative cycles. Two states—Missouri and Rhode Island—have not enacted any AI-related legislation to date.

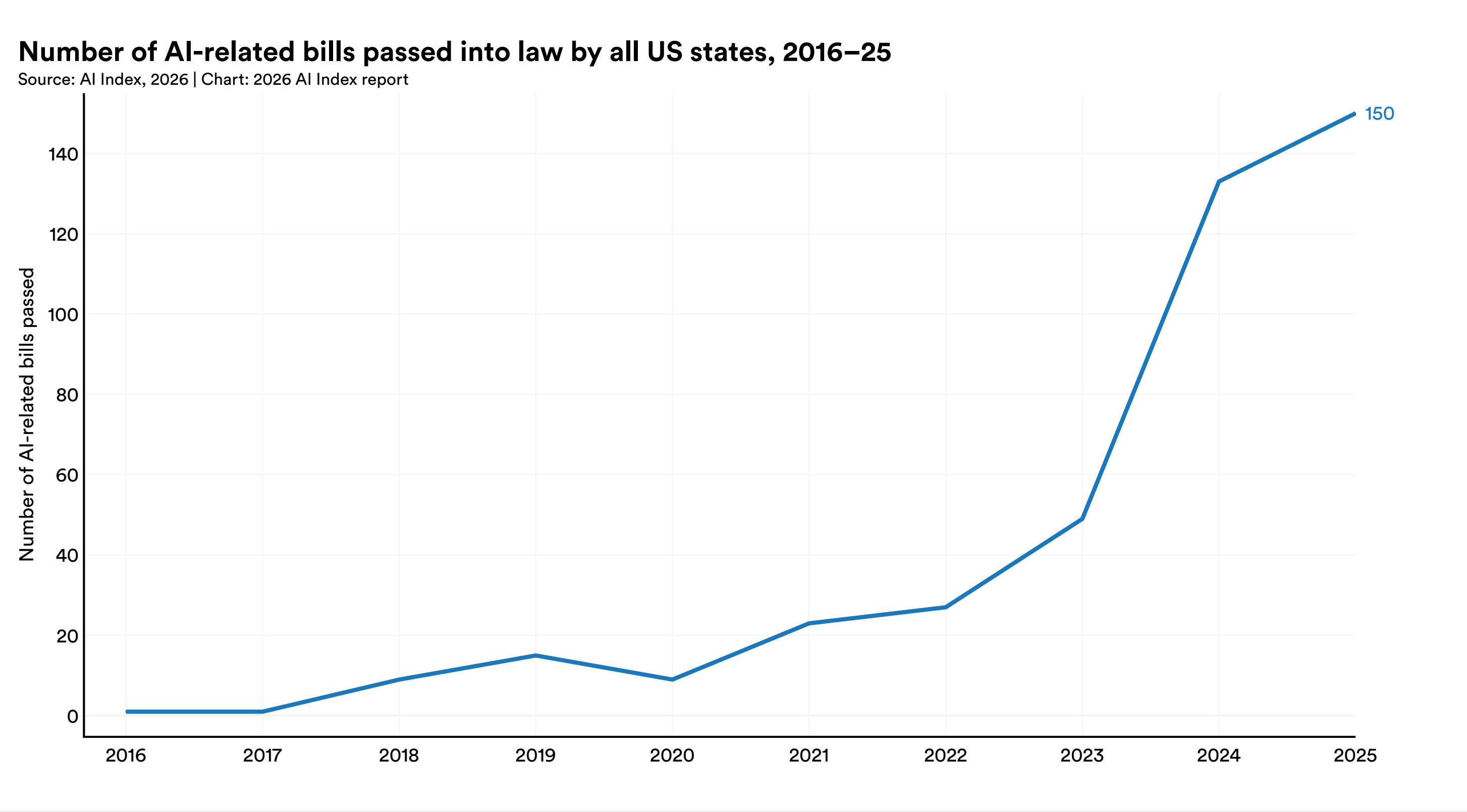

Figure 8.4.4

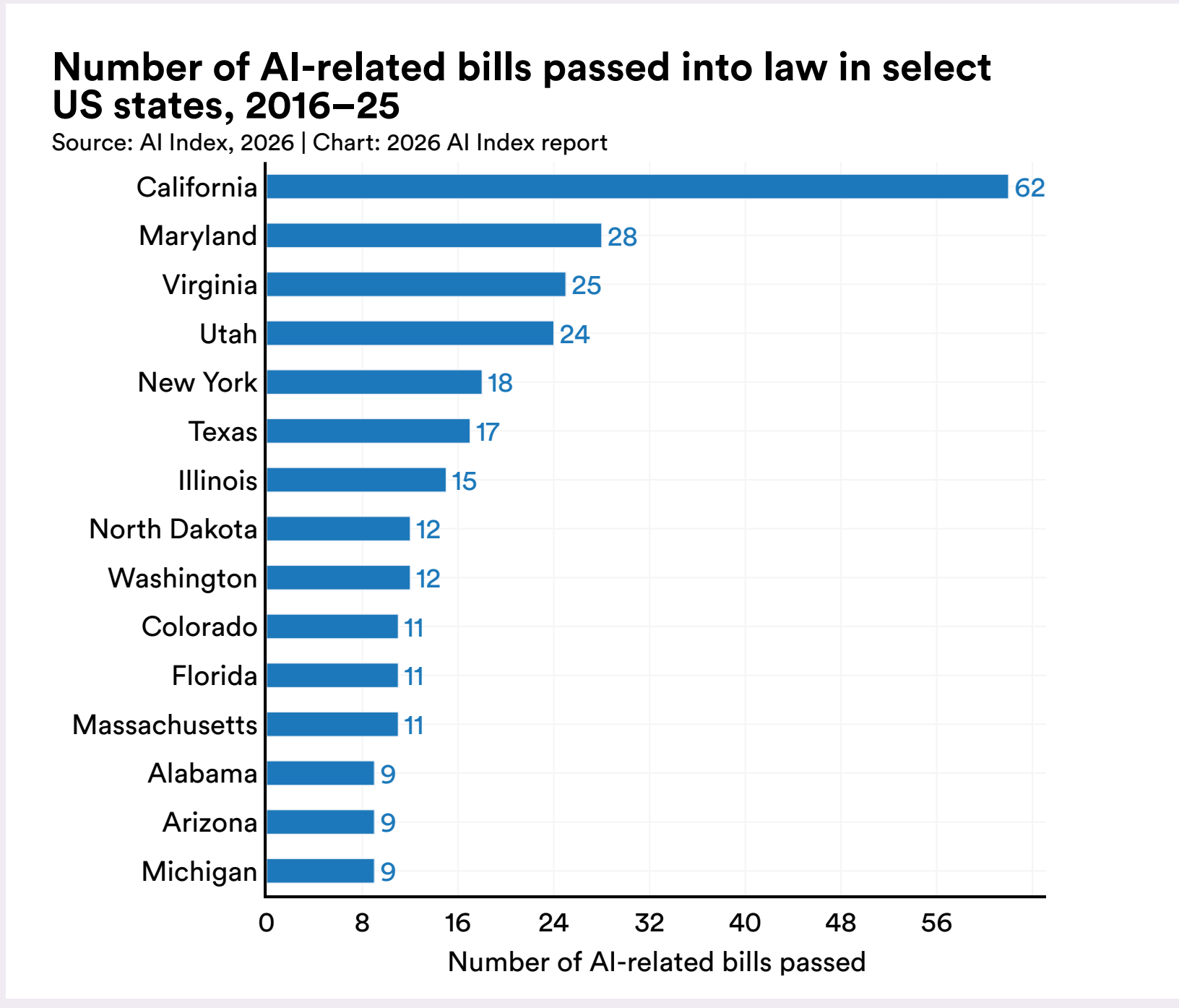

Figure 8.4.5

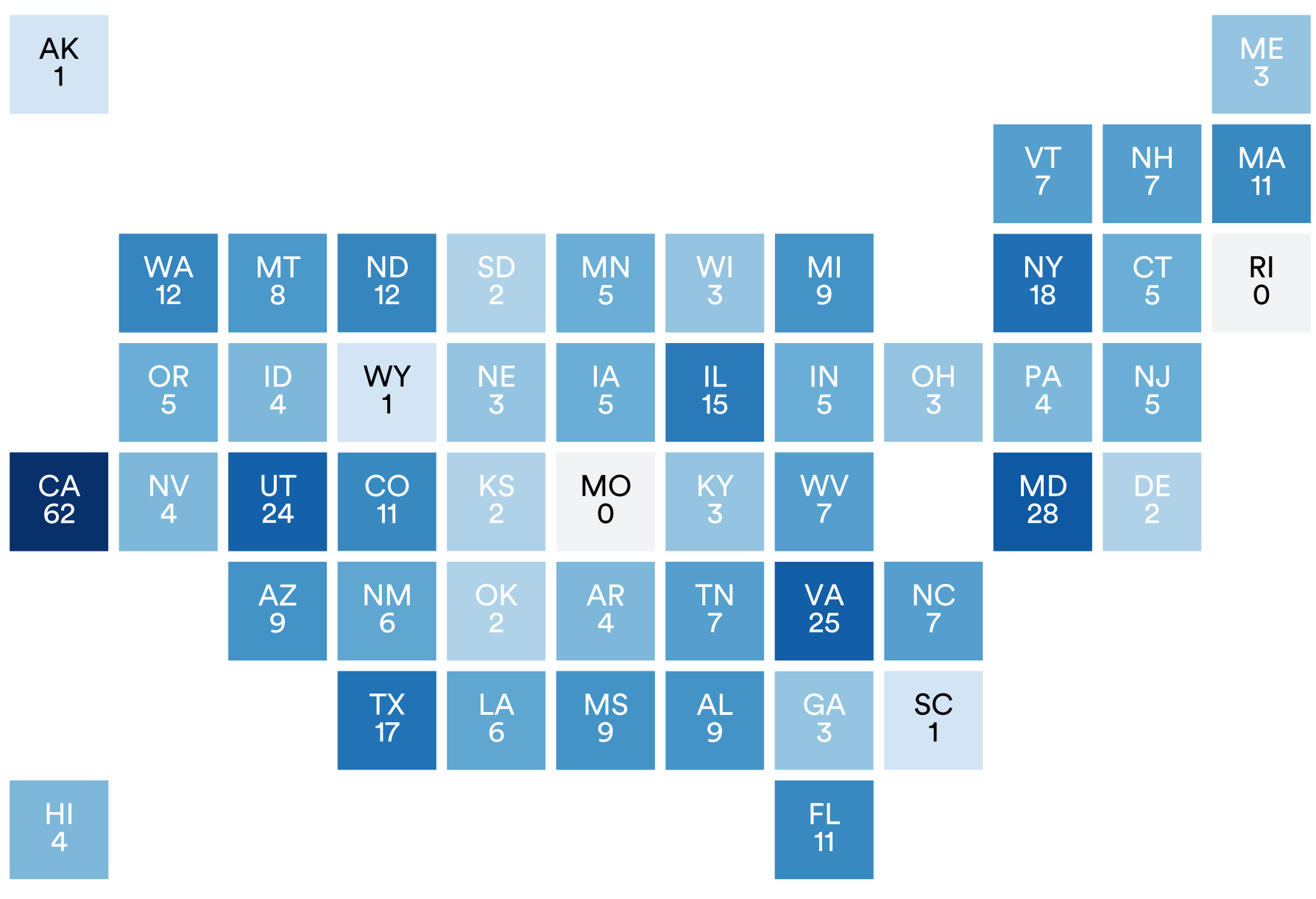
Number of state-level AI-related bills passed into law in the United States by state, 2016–25 (sum)
Source: AI Index, 2026 | Chart: 2026 AI Index report

Figure 8.4.6

HIGHLIGHT:

## State AI Legislation Amid Shifting Federal Policy

While federal AI policy shifted toward deregulation in 2025, state legislatures continued to move ahead with AI-specific laws on their own in the absence of a federal framework.[6] State policies are developing across different tracks, including targeted protections against discrimination, misinformation, and abuse. Several of the most prominent state actions, including Utah's Mental Health Chatbot Act, Montana's Right to Compute Act, and Texas' Responsible AI Governance Act, are also listed in the timeline at the start of this chapter.

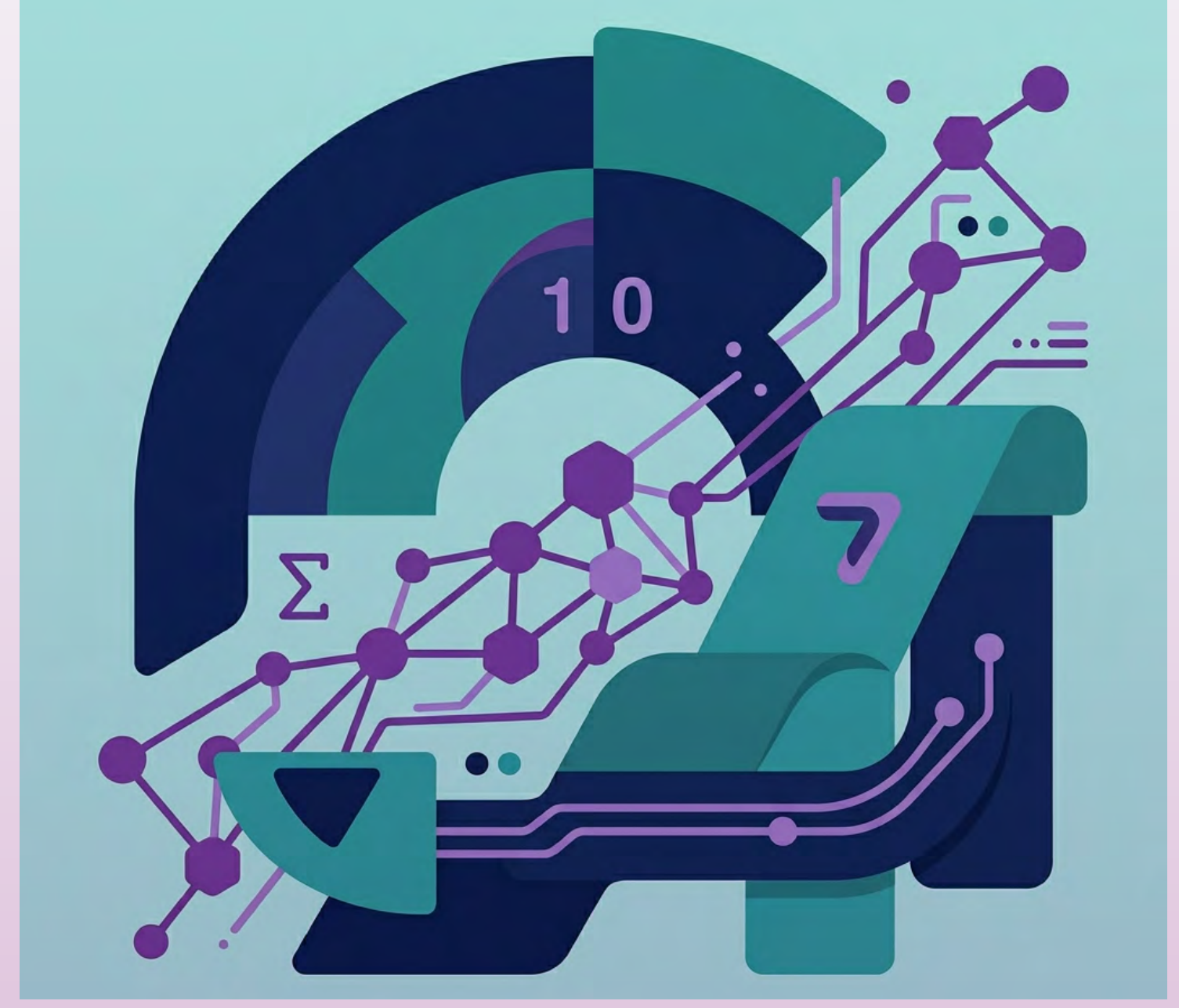

To have greater comprehensive oversight on AI use, some states pursued broader frameworks. Colorado's Artificial Intelligence Act, signed in May 2024, was among the first state laws targeting algorithmic discrimination in decisions such as hiring, housing, and medical care. However, the

6 The state tracker counts only bills whose final enacted text includes the phrase "artificial intelligence," based on searches conducted across all 50 state legislative websites. While this filter makes the dataset consistent, it likely captures only a portion of the broader state-level policy picture.

**HIGHLIGHT:**

law is also a case study in how difficult it can be to operationalize more sweeping regulation frameworks. In 2025, Colorado attempted to enact amendments narrowing parts of the law but instead opted to push key compliance dates back to mid-2026 to allow for additional time to consider revisions. Texas followed a different path with the Responsible Artificial Intelligence Governance Act (HB 149), passed in 2025 and effective January 2026. While originally touted as the most all-encompassing AI legislation, the final version was significantly scaled back from the original proposal, removing most private-sector obligations and focusing on uses such as behavioral manipulation and the production of child sexual abuse material.

Several states have focused on regulating how AI chatbots interact with consumers, particularly in sensitive settings. Utah's HB 452, signed in March 2025, addresses mental health chatbots and requires disclosure when users are interacting with AI, prohibits the selling or sharing of personal health data with third parties, and restricts in-chat advertising. California's SB 243, effective January 2026, requires companion chatbot operators to disclose their AI nature and implement safety protocols related to suicidal ideation, with additional safeguards for minors. Other states, such as Hawaii (SB 640) and Massachusetts (S.243), have proposed treating undisclosed chatbot interactions as unfair and deceptive practices.

Watermarking and provenance requirements have also gained traction. Washington (HB 1170) requires large providers to include provenance data on AI-generated or materially altered media. Illinois (SB 1929) and Florida (HB 369) proposed similar measures, following California's AI Transparency Act (SB 942), which mandated that large generative AI tools offer watermarking and detection tools at no cost.

However, not all states have moved toward greater oversight. In Montana, SB 212 was signed in April 2025 and established the first state-level "right to compute," affirming individuals' and businesses' rights to own and use computational resources, including AI tools, for lawful purposes. The Montana law limits government restrictions to those that are "demonstrably necessary and narrowly tailored to fulfill a compelling government interest," while also requiring deployers of AI-controlled critical infrastructure to maintain risk management policies with human override capabilities.

In December 2025, the White House issued an executive order titled "Ensuring a National Policy Framework for Artificial Intelligence," which directed the Department of Justice to establish an AI Litigation Task Force to challenge state AI laws in court, instructed the Department of Commerce to identify state laws it considers overly burdensome, and tied some federal funding to states' willingness to avoid enacting conflicting AI legislation. The order carved out certain areas state legislatures can oversee, including child safety, data center infrastructure, and state government procurement. The upshot is that the future of U.S. state-level regulation on AI remains uncertain.

## US Congressional Hearings

Congressional attention to AI, as measured by the number of witnesses appearing in AI-related hearings, has increased by almost twentyfold since 2017 (Figure 8.4.7). That growth accelerated after 2022, consistent with the mainstream emergence of generative AI tools in late 2022. Witness counts rose from 18 in 2022 to 131 in 2023 and remained high, at 102, in 2025.

The composition of those witnesses has shifted over time (Figure 8.4.8). Industry's share of witnesses rose from 13% in the 115th Congress to 37% in the 119th, making it the largest witness group in the data. This increase is consistent with the sector's expanding role in overall AI development. As discussed in Chapters 1 and 4, private companies now account for the majority of frontier model releases and infrastructure investment, which may position them both as more relevant sources of technical input and as more active participants in shaping the policy environment in which they operate. Over the same period, the share of government witnesses fell from 35% to 10%, and academia's share declined from 26% to 15%. The "other" category, which includes civil society and nonprofit organizations, grew from 26% to 38%.

**Number of witnesses in US congressional AI-related hearings, 2017–25**
Source: AI Index, 2026 | Chart: 2026 AI Index report

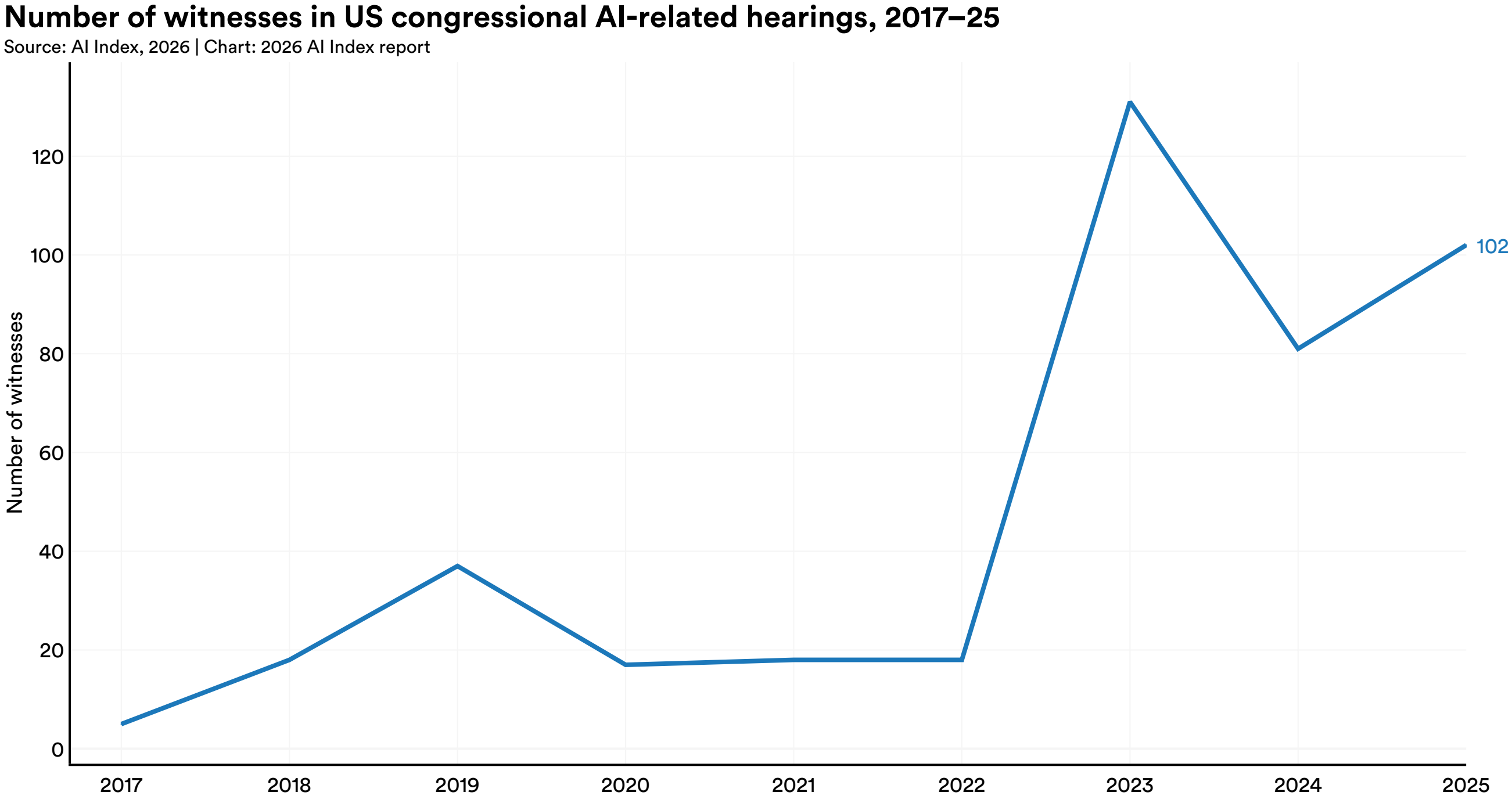

Figure 8.4.7

**Witnesses (% of total) in US congressional AI-related hearings by sector, 2017–25**
Source: AI Index, 2026 | Chart: 2026 AI Index report

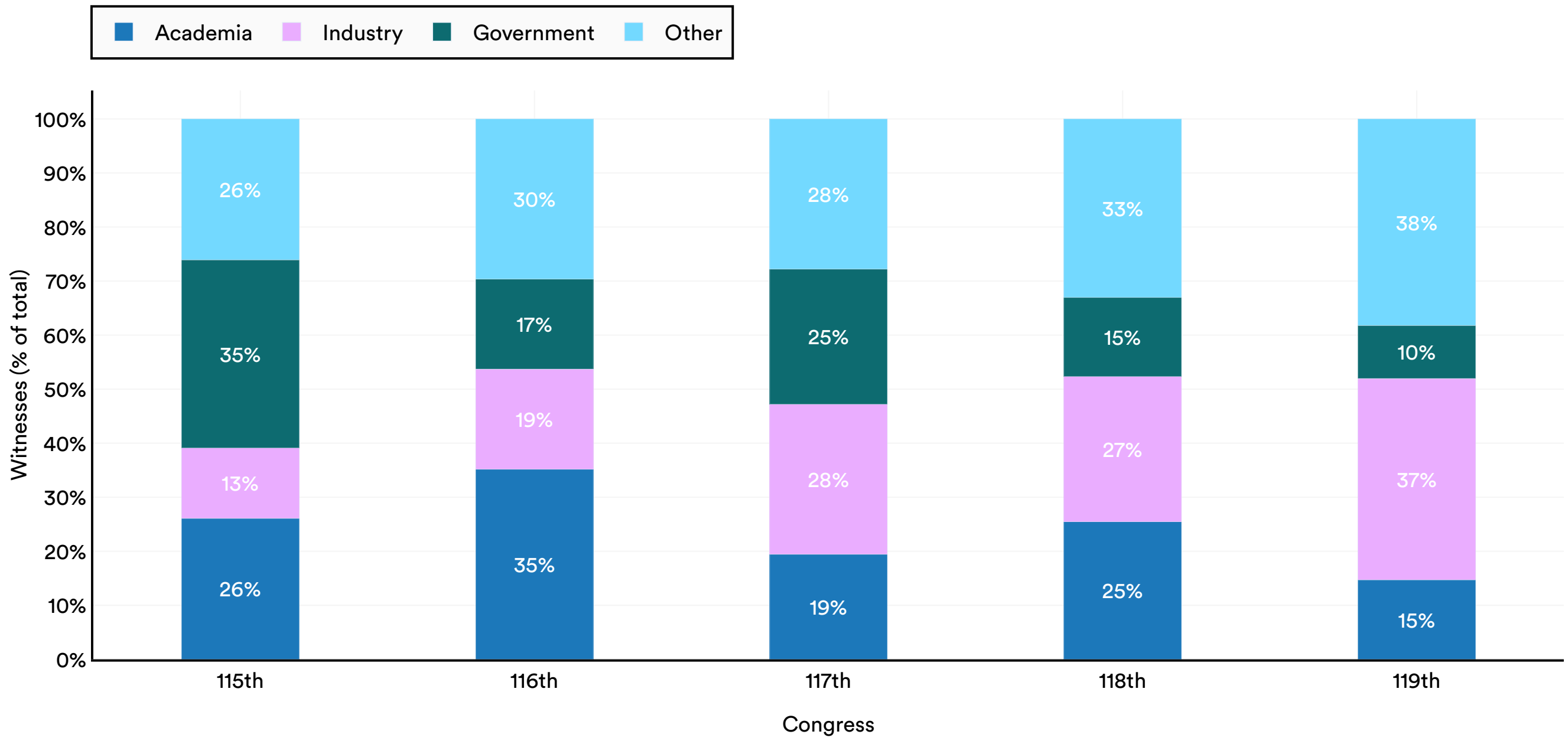

Figure 8.4.8

Across both the House and the Senate, general AI governance and national security and defense drew the most witnesses between 2017 and 2025, with 113 and 74 total witnesses, respectively (Figure 8.4.9). Overall, the House has been more active than the Senate in most subject areas, including finance and economic policy (36 versus 3 witnesses) and national security (49 versus 25). Health and biomedical AI was the only area where both chambers show the same level of activity (9 witnesses each), suggesting comparable engagement in this domain.

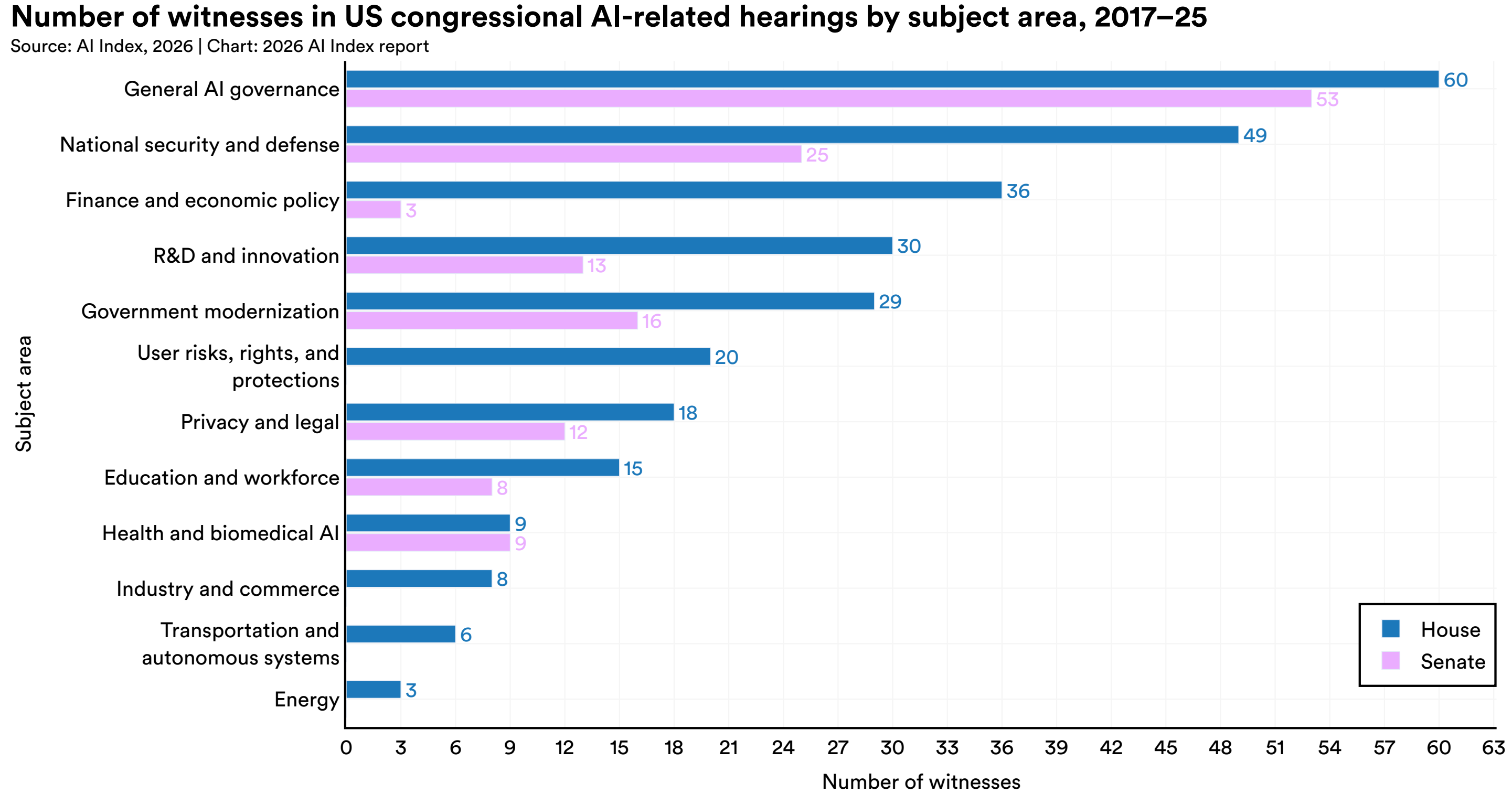
Number of witnesses in US congressional AI-related hearings by subject area, 2017–25
Source: AI Index, 2026 | Chart: 2026 AI Index report

Figure 8.4.9

## US Regulations

Federal regulatory activity on AI has grown in recent years, with the number of AI-related regulations increasing from one recorded action in 2016 to 58 in 2025 (Figure 8.4.10). Similar to witness counts, the sharpest increase came after 2022, and the pace has remained steady through 2025, with 58 AI-related regulations.

The direction of federal AI policy changed in early 2025 when the Donald Trump administration issued the Initial Rescissions of Harmful Executive Orders and Actions, which revoked a range of executive actions, including the Biden administration's Executive Order 14110, a framework for safe, secure, and trustworthy AI development and use. That Biden order had anchored a more precautionary federal approach that included reporting requirements for advanced models, guidance on watermarking AI-generated content, and initiatives addressing privacy, civil rights, and workforce impacts. Its reversal was followed by a new executive order, "Removing Barriers to American Leadership in Artificial Intelligence," which reoriented federal policy toward reducing regulatory constraints and promoting innovation.

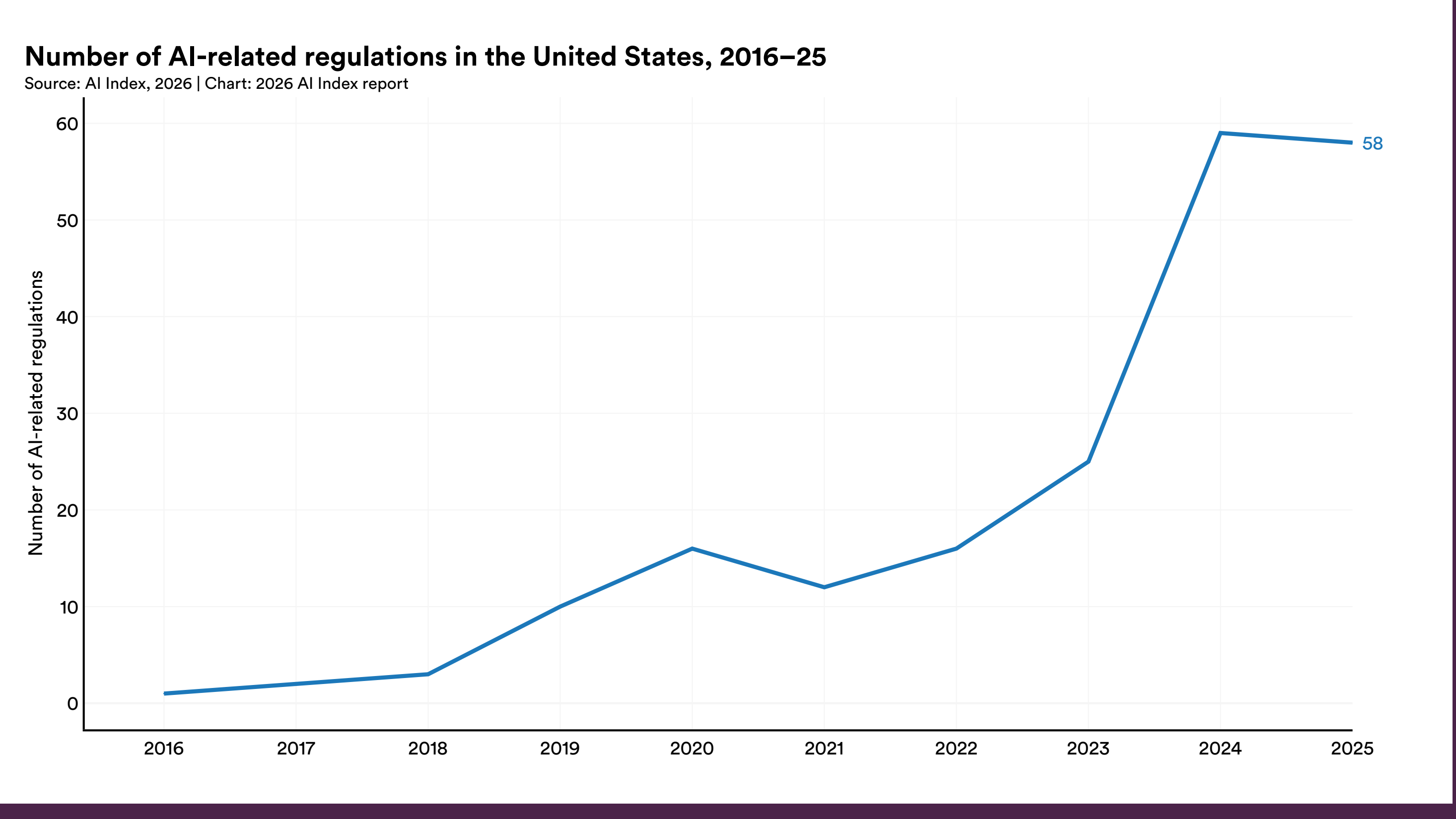

Figure 8.4.10

## By Agency

The increasing number of AI-related regulations has been driven by a wide set of federal agencies (Figure 8.4.11). The Executive Office of the President has been the most active, issuing regulatory actions every year since 2016 and putting out 28 in 2025 alone. The Commerce Department and the Industry and Security Bureau have also become more active in recent years, consistent with growing attention to export controls and AI supply chain policy. Several agencies, including the Department of Energy, the Department of Education, and the Securities and Exchange Commission, began issuing AI-related regulations in 2023 or later.

**Number of AI-related regulations in the United States by agency, 2016–25**

Source: AI Index, 2026 | Chart: 2026 AI Index report

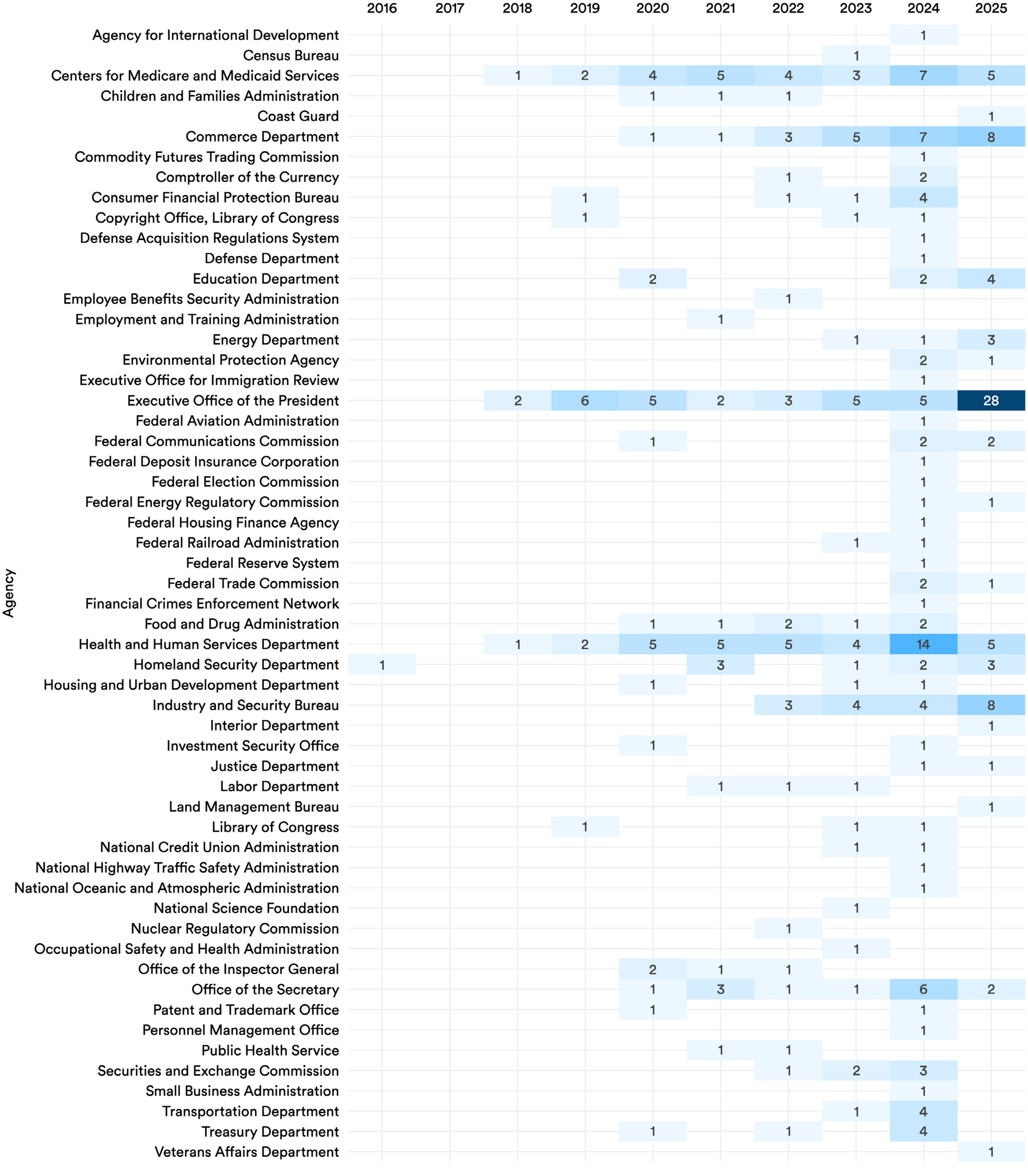

**Figure 8.4.11**

**HIGHLIGHT:**

## A Closer Look at US Federal Regulations

The following section highlights AI-related regulations enacted through federal rules and executive orders during 2025.

| Agency | Regulation | Description |
| --- | --- | --- |
| Commerce Department; Industry and Security Bureau | Framework for Artificial Intelligence Diffusion | Prior to its rescission by the Trump administration in May 2025, this rule updated U.S. export controls for advanced computing items and certain AI model weights. It created a new classification for specified advanced "closed-weight" AI model weights and revised licensing requirements and review policies for advanced computing integrated circuits and related items. It also expanded and added license exceptions for certain destinations and uses, updated notification procedures, and introduced additional guidance to help identify diversion risks. |
| Justice Department | Preventing Access to U.S. Sensitive Personal Data and Government-Related Data by Countries of Concern or Covered Persons | This rule implements Executive Order 14117 of February 28, 2024, "Preventing Access to Americans' Bulk Sensitive Personal Data and United States Government-Related Data by Countries of Concern." It prohibits or restricts certain data transactions involving countries of concern or covered persons, focusing on specified transfers of Americans' bulk sensitive personal data and U.S. government–related data. The rule addresses cross-border data transactions that can involve large-scale datasets and establishes limits on when and how such data may be transferred. |
| Executive Office of the President | Advancing United States Leadership in Artificial Intelligence Infrastructure | This executive order directs the federal government to fast-track the buildout of domestic AI infrastructure—particularly frontier data centers and their supporting energy systems—on federal lands managed by the departments of Defense, Energy, and Interior. It mandates competitive solicitations for private-sector leases on federal sites, paired with requirements for clean energy procurement, robust cybersecurity standards, and high labor practices. The order also streamlines permitting processes, addresses grid interconnection challenges, and calls for international engagement to promote trusted AI infrastructure among U.S. allies, all with the overarching goal of ensuring the United States maintains leadership in frontier AI development. |
| Executive Office of the President | Removing Barriers to American Leadership in Artificial Intelligence | This executive order sets the foundational U.S. policy of maintaining global AI dominance by dismantling regulations seen as obstacles to innovation. Most notably, it replaces the Biden administration's 2023 AI executive order on safe and trustworthy AI development and directs agencies to suspend or rescind any actions taken under the earlier order that conflict with the new policy. It also mandates the development of a comprehensive AI Action Plan within 180 days, and requires the Office of Management and Budget to revise existing AI-related guidance memoranda to align with the administration's pro-innovation, minimal-regulation approach. |
| Executive Office of the President | Ensuring a National Policy Framework for Artificial Intelligence | This executive order establishes a national AI policy framework to prevent a fragmented patchwork of state regulations from hindering U.S. innovation and global competitiveness. It creates an AI Litigation Task Force under the Attorney General to challenge state AI laws considered overly burdensome or unconstitutional. It also directs federal agencies to evaluate problematic state laws, potentially withhold federal funding from noncompliant states, and develop a uniform federal standard that preempts conflicting state regulations—while preserving state authority over child safety, data centers, and government AI procurement. |

**HIGHLIGHT:**

| | | |
|---|---|---|
| Executive Office of the President | Advancing Artificial Intelligence Education for American Youth | This executive order advances artificial intelligence education for American youth by directing the federal government to expand early exposure to AI concepts, integrate AI appropriately into classrooms, and build an AI-ready workforce. It establishes a White House Task Force on AI Education, led by the Office of Science and Technology Policy, to coordinate agency efforts and create a Presidential Artificial Intelligence Challenge that highlights student and educator achievements. The order also prioritizes educator training and research on AI in education, including the use of existing grant programs to support professional development and AI-enabled tools that improve teaching and learning outcomes. |
| Executive Office of the President | Promoting the Export of the American AI Technology Stack | This executive order launches a coordinated federal initiative to promote the export of “full-stack” American AI technology packages. These packages combine AI-optimized hardware, cloud and networking infrastructure, data pipelines, models, security measures, and targeted applications for selected partner countries and regions. It directs the Department of Commerce to solicit and evaluate proposals from industry-led consortia, designate priority AI export packages, and support them by streamlining access to federal diplomatic and financing tools. The order frames these efforts as essential to sustaining U.S. leadership in AI and reducing global reliance on AI technologies developed by adversaries. |

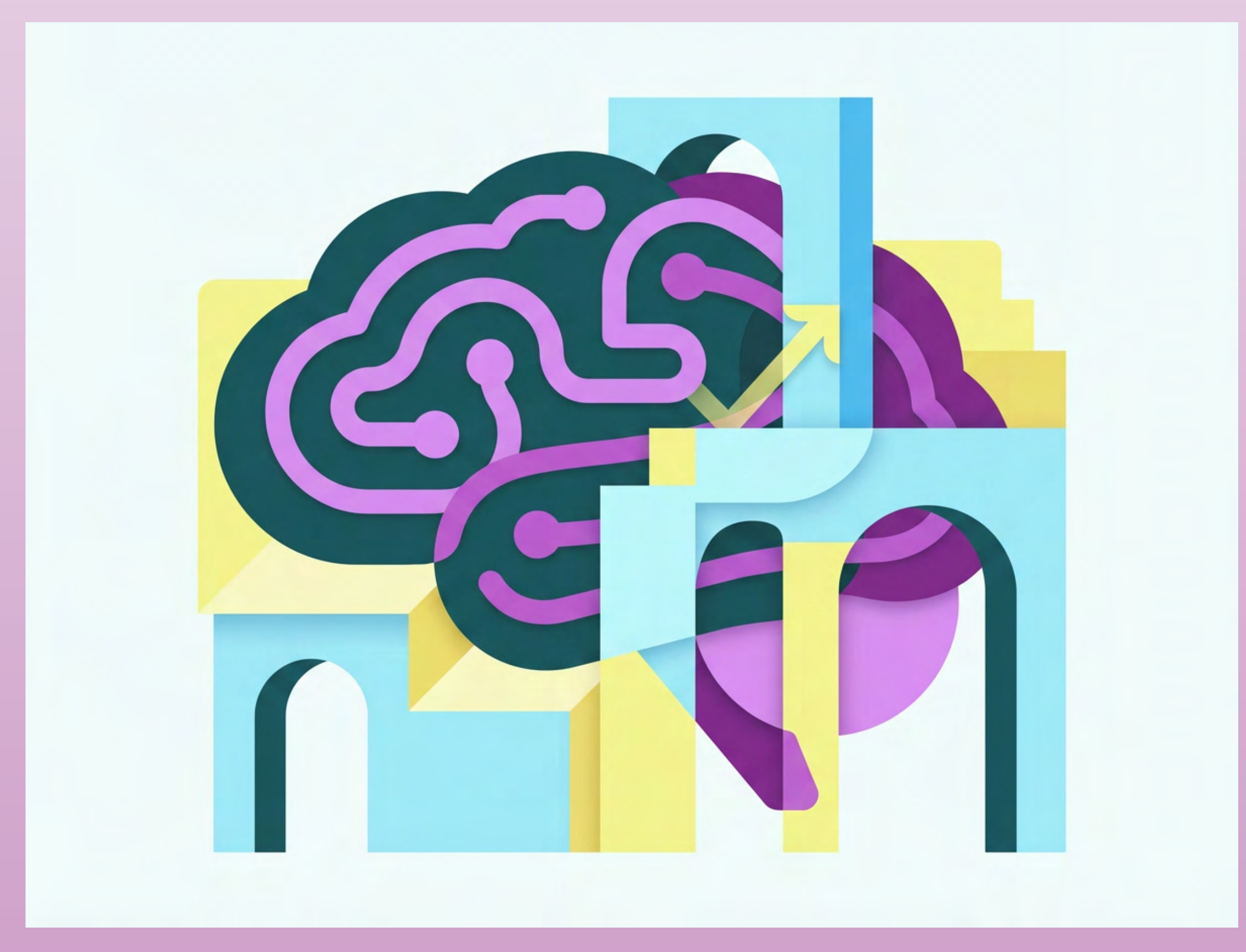

# 8.5 Public Investment in AI

Public spending on AI reflects how governments are translating national strategies and policy commitments into resource allocation. This section tracks government AI spending across the United States and several European countries, drawing on public contract data in Europe and the U.K., and on contract, grant, and Other Transaction Agreement (OTA) data in the United States.[7] Grant-level data is included for the United States but is not systematically available for European countries and is therefore excluded from the European analysis.

The methodology differs across regions due to differences in data availability. In Europe and the U.K., long-term instruments like Framework Agreements and Dynamic Purchasing Systems[8] typically report maximum contract ceilings rather than actual spending, and award duration data is often incomplete. Therefore, European and U.S. results are presented separately. For the United States, where transaction-level obligation data is available, the AI Index estimates investment by aggregating only those obligations that occur after AI-related activity first appears in an awarded procedure, while controlling for early de-obligations that could distort trends. This approach preserves time patterns while reducing the risk of overstating historical AI investment.

## Total AI Public Investments

Public spending on AI-related contracts has grown across both the United States and the European countries tracked, though the pace and composition of spending vary by country.

### United States

Between 2013 and 2024, the United States invested approximately $20.5 billion toward AI-related activities, made up of $15.9 billion in grants, $3.9 billion in contracts, and $650 million in Other Transaction Agreements (Figure 8.5.1). Since 2020, AI-related grant spending has accelerated compared to contracts and OTAs. In 2024, grants accounted for $5.1 billion, 32% of their cumulative total since 2013 (Figure 8.5.2).

Award volume follows a similar pattern. Grants account for the majority of AI-related awards, with 22,364 compared to 3,347 contracts and 185 OTAs (Figure 8.5.3). Despite their lower volume, OTAs have a median contract value of almost $1 million, far higher than grants ($304,000) and contracts ($150,000).

7 European contract data is drawn from Tenders Electronic Daily (TED). U.K. data is sourced from TED, Find a Tender, Contracts Finder, and its archive. Scotland and Wales spending data is accessed through procurement APIs and the Open Contracting Partnership's data registry via Kingfisher Collect; Northern Ireland is excluded due to the absence of an API. U.S. contract, grant, and OTA data is drawn from the Federal Procurement Data System (FPDS) API.

8 Framework Agreements and Dynamic Purchasing Systems are two types of multiyear umbrella buying arrangements. A Framework Agreement sets pre-agreed terms and a maximum budget for future purchases, while a Dynamic Purchasing System is a flexible, open list of approved suppliers used for competitions and orders over time.

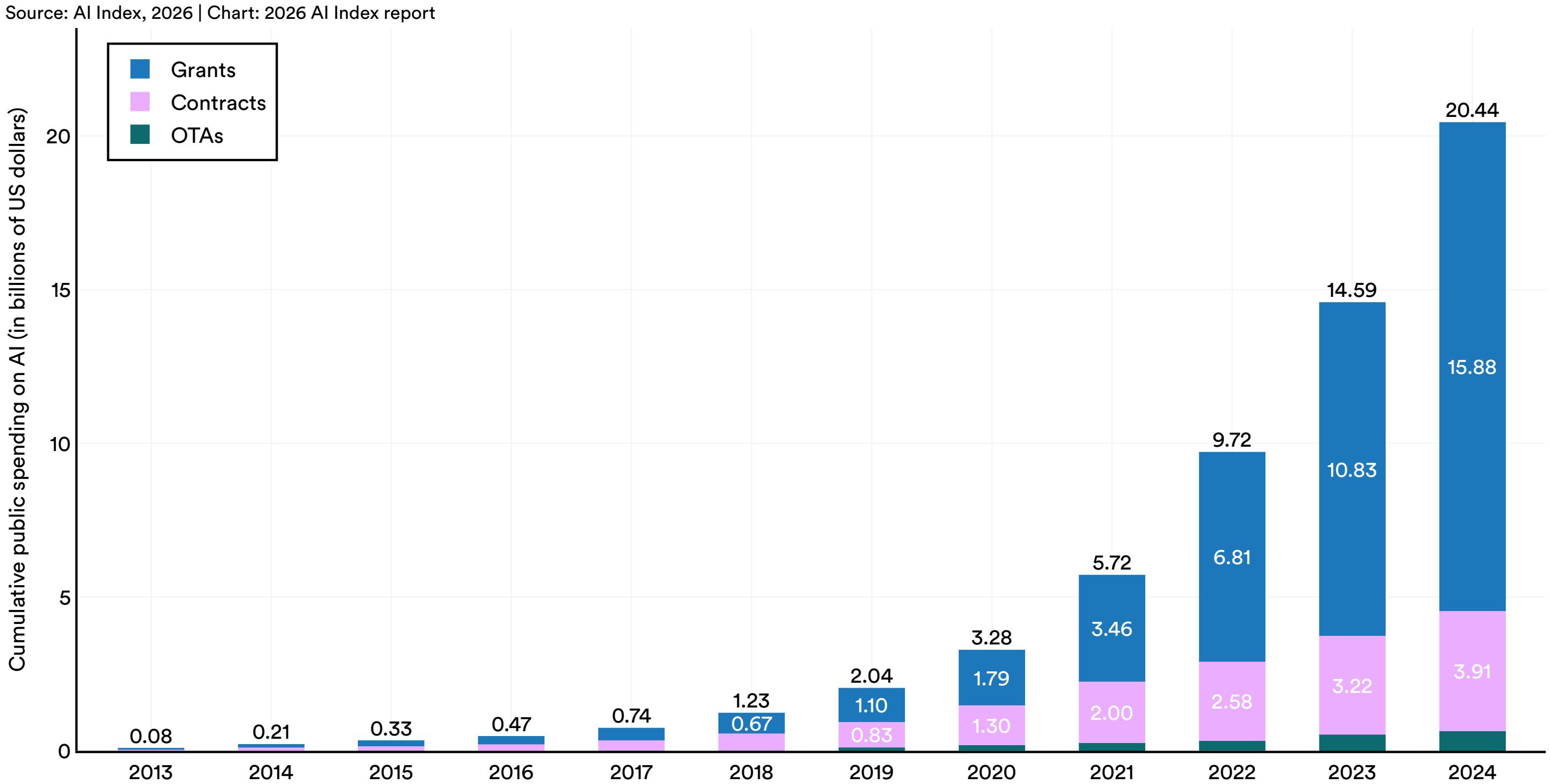
Cumulative public spending on AI in the United States, 2013–24
Source: AI Index, 2026 | Chart: 2026 AI Index report

Figure 8.5.1

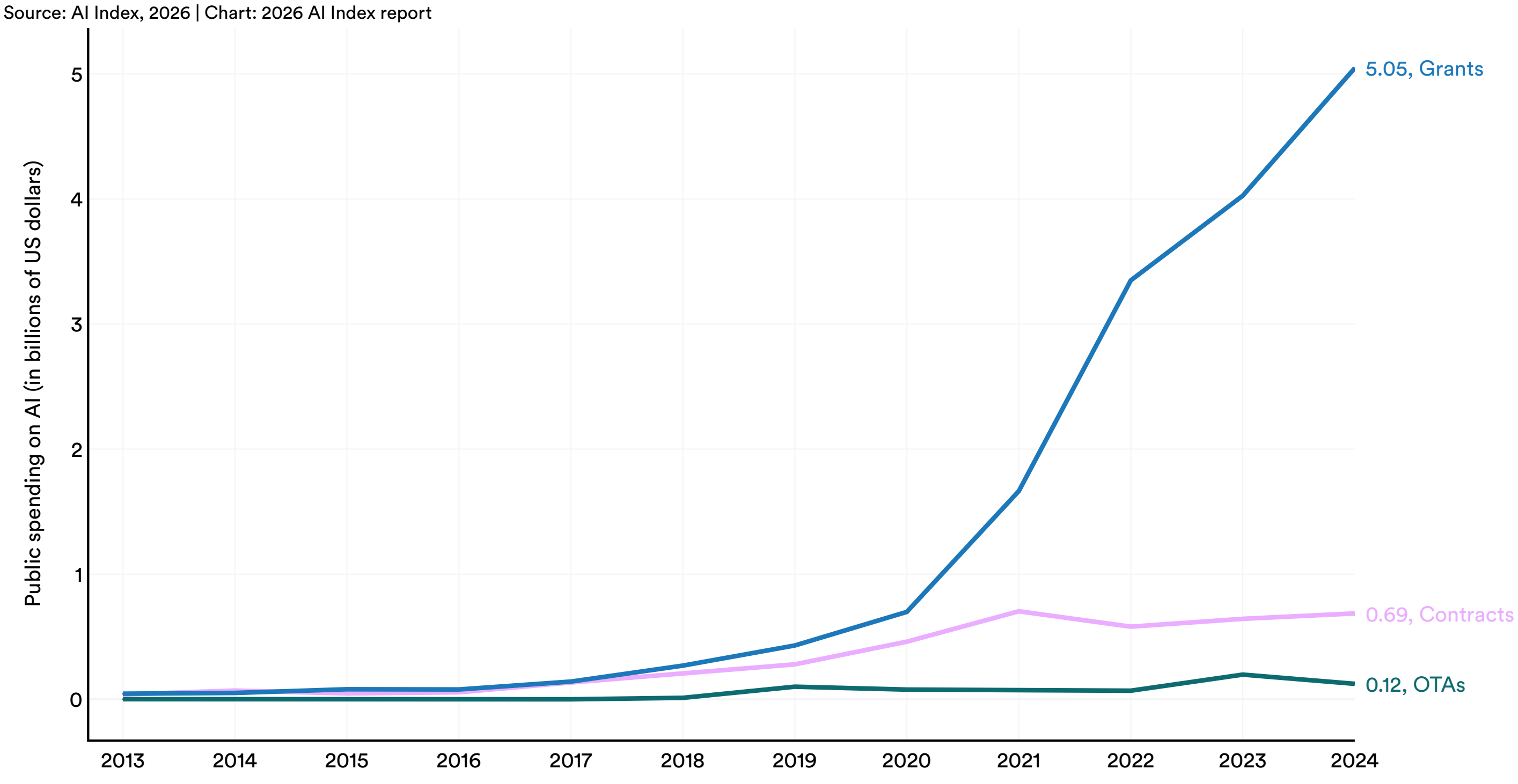
Public spending on AI in the United States, 2013–24
Source: AI Index, 2026 | Chart: 2026 AI Index report

Figure 8.5.2

### US AI-related public spending

Source: AI Index, 2026

| Statistic | Grants | Contracts | OTAs |
| --- | --- | --- | --- |
| Number of awards | 22,364 | 3,347 | 185 |
| Total (in billions of USD) | 15.9 | 3.9 | 0.7 |
| Median (in thousands of USD) | 304 | 149.9 | 999.7 |
| Average (in thousands of USD) | 710.1 | 1,167.9 | 3,528.8 |
| Total per 100,000 inhabitants (in thousands) | 4,728.7 | 1,172.8 | 195 |

Figure 8.5.3

Geographically, U.S. public AI investment through contracts and OTAs is highly concentrated. Virginia[9] received $1.09 billion, California $0.67 billion, and Maryland $0.55 billion; together these states accounted for nearly 60% of total contract and OTA spending between 2013 and 2024 (Figure 8.5.4). AI-related grants have been more broadly dispersed. California ($2.37 billion), Massachusetts ($1.3 billion), and New York ($1.15 billion) received the largest allocation, but these top three accounted for less than 16% of the total (Figure 8.5.5). The geographic concentration in contracts and OTAs may reflect proximity to major federal agencies, while the wider distribution of grants is aligned with the broader institutional footprint of federally funded research.

**Public spending on AI via contracts and OTAs in the United States, 2013–24**

Source: AI Index, 2026 | Chart: 2026 AI Index report

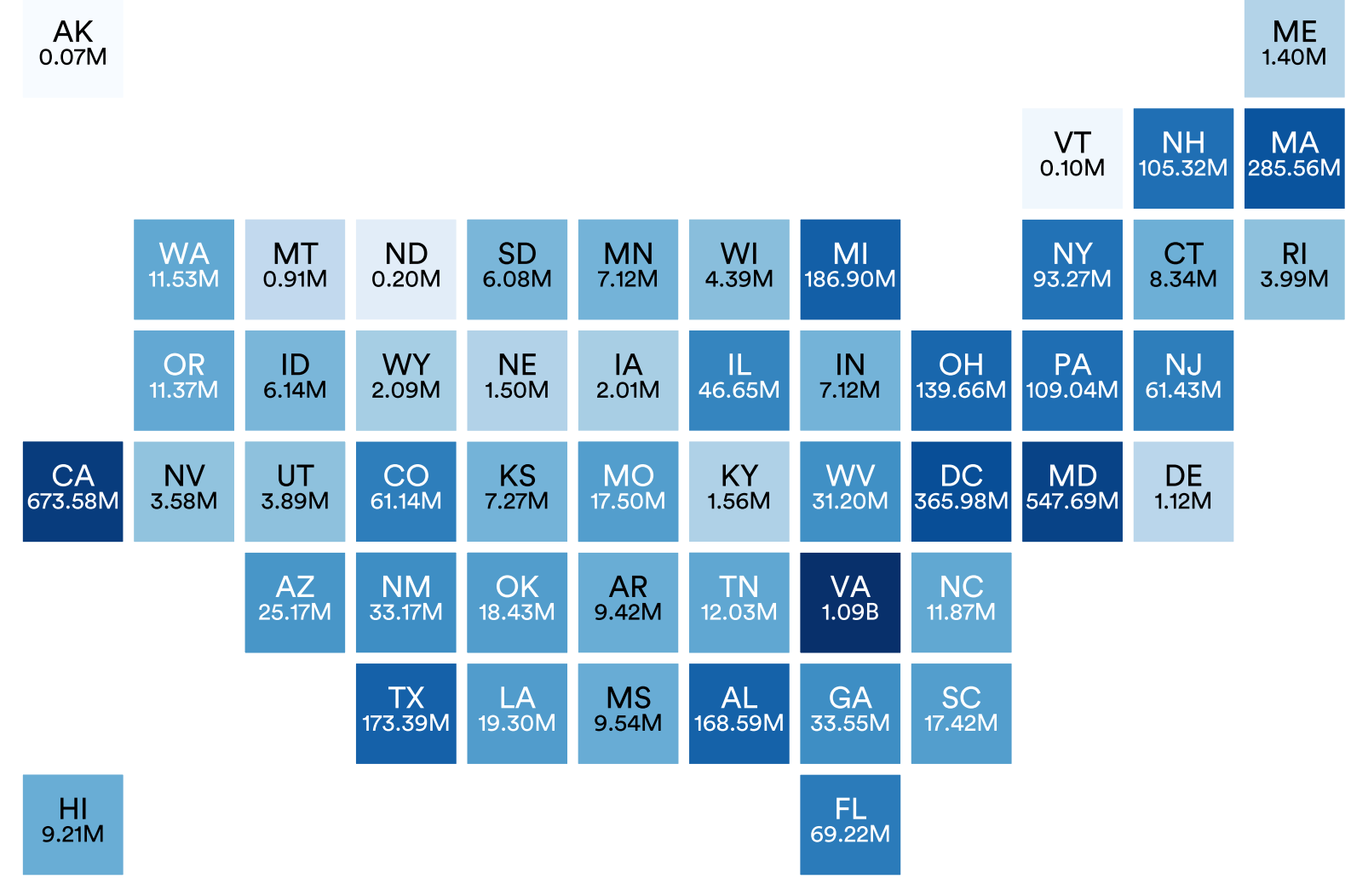

Figure 8.5.4

9 Virginia is home to the headquarters of ECS, the top contractor by total awarded value via contracts and OTAs in the U.S.

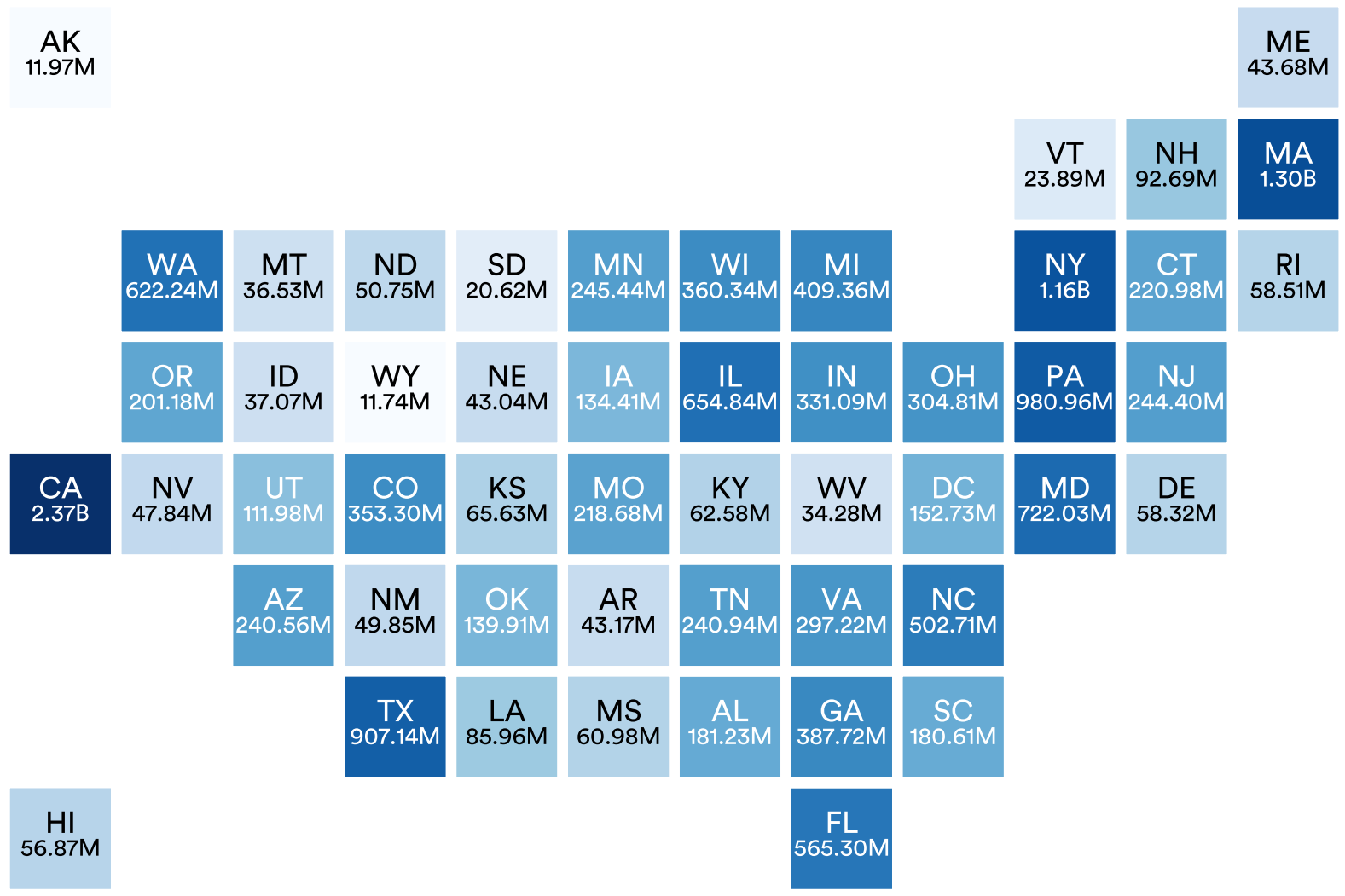
Public spending on AI via grants in the United States, 2013–24
Source: AI Index, 2026 | Chart: 2026 AI Index report

Figure 8.5.5

## Spending Across Agencies

In the United States, the Department of Defense led in AI-related contract and OTA spending during 2013–24, accounting for 74.1% of the $810 million total in 2024 and 73% of the total spend ($4.6 billion) across the whole period (Figure 8.5.6). The next largest funders, though significantly smaller, were the Department of the Treasury at 7.2% and the Department of Veterans Affairs at 5.1%. Of the remaining agencies, each represented less than 5% of total spending over the same period.

AI-related grants have been channeled mainly through the Department of Health and Human Services (HHS) — which includes the National Institutes for Health — and the National Science Foundation (NSF) (Figure 8.5.7). By 2024, public AI funding was split evenly across these agencies, with each accounting for roughly 40% of the total. Over time, the NSF consistently received the largest share of AI-related grant funding until there was a steep increase in HHS starting in 2020.

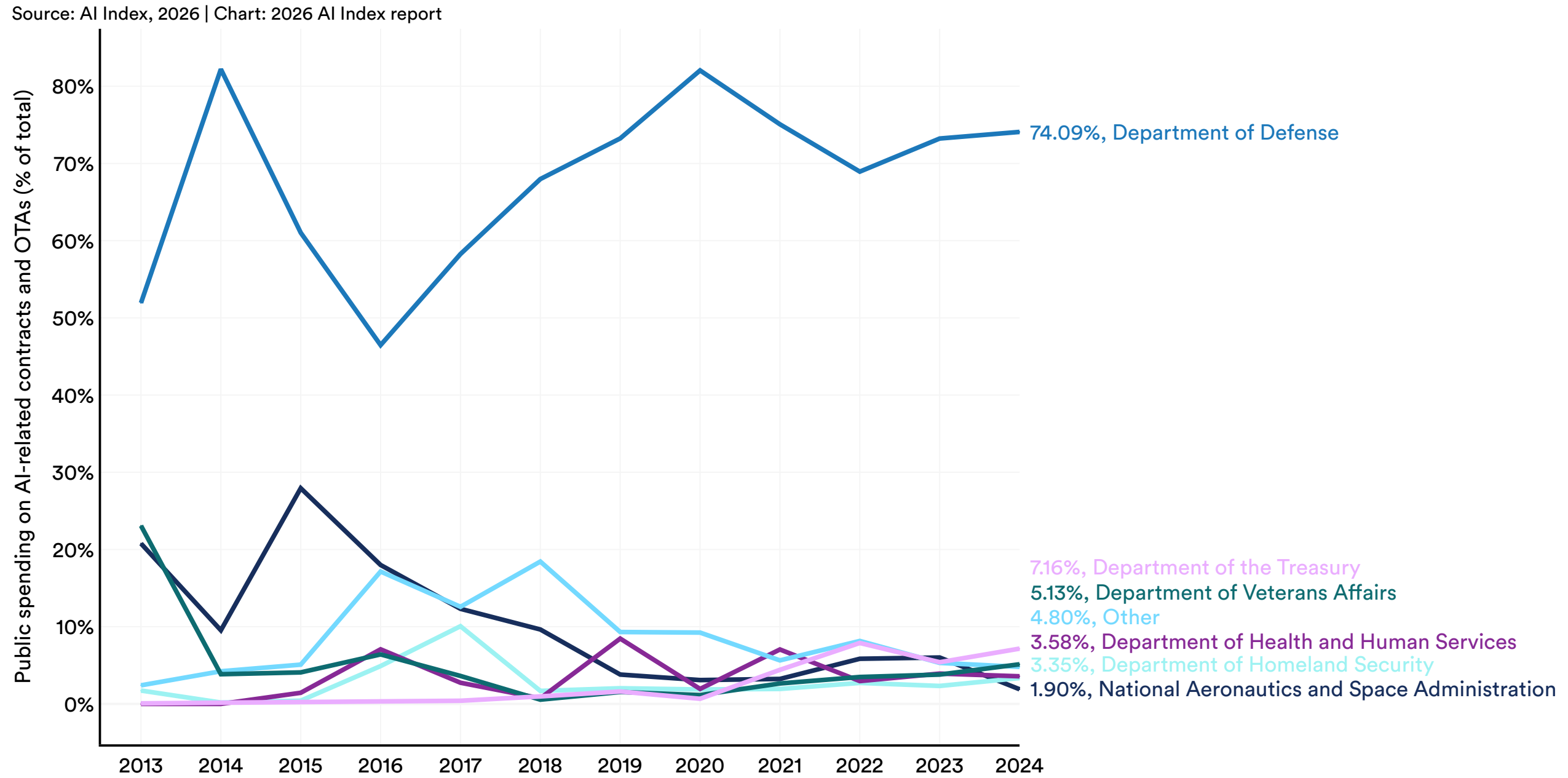
Public spending on AI-related contracts + OTAs (% of total) in the United States by funding agency, 2013–24
Source: AI Index, 2026 | Chart: 2026 AI Index report

Figure 8.5.6

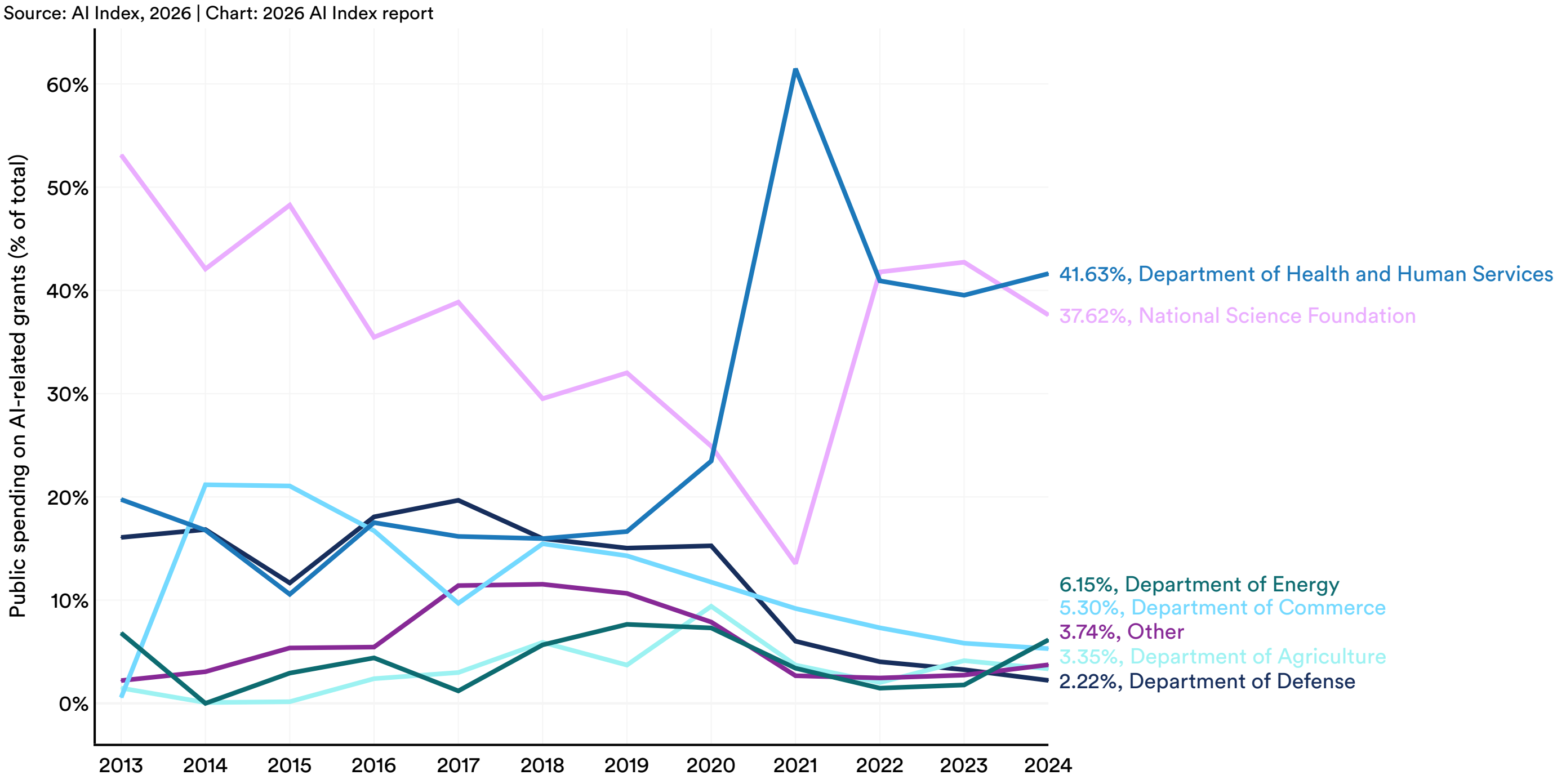
Public spending on AI-related grants (% of total) in the United States by funding agency, 2013–24
Source: AI Index, 2026 | Chart: 2026 AI Index report

Figure 8.5.7

## Europe

European nations collectively committed[10] approximately $3.7 billion in contracts over the 2013–24 period (Figure 8.5.8). The United Kingdom accounted for the largest share, with $1.6 billion, followed by Germany ($505 million) and France ($320 million). In 2024, the U.K.'s AI-related public commitment was $454.4 million, representing 28% of the previous decade's investment. Germany allocated $206.6 million, representing 40% of its total over the same period. Contract volume follows a similar pattern. The U.K. issued the most AI-related contracts with 738, compared to 611 in Germany and 187 in Spain (Figure 8.5.9). However, despite their higher spending and contract volumes, these three countries have a median contract value below $500,000, far lower than smaller European countries such as Denmark (almost $1.1 million) (Figure 8.5.10).

10 The details of U.K. and EU data only allows us to gauge the awarded amounts. Since many AI-related contracts rely on framework agreements or dynamic purchasing systems, there is not clear information about the timeline and total of actual obligations.

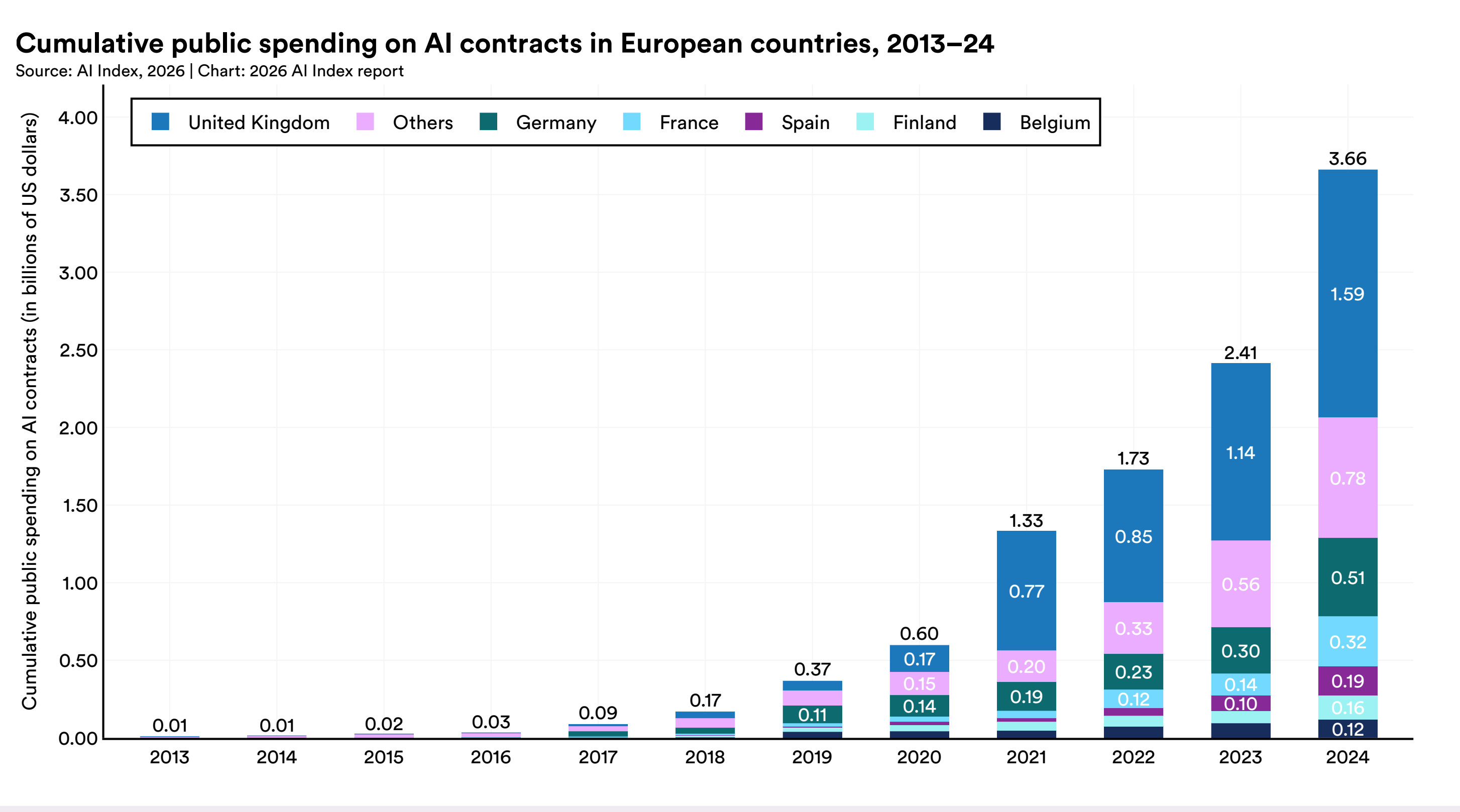

Figure 8.5.8

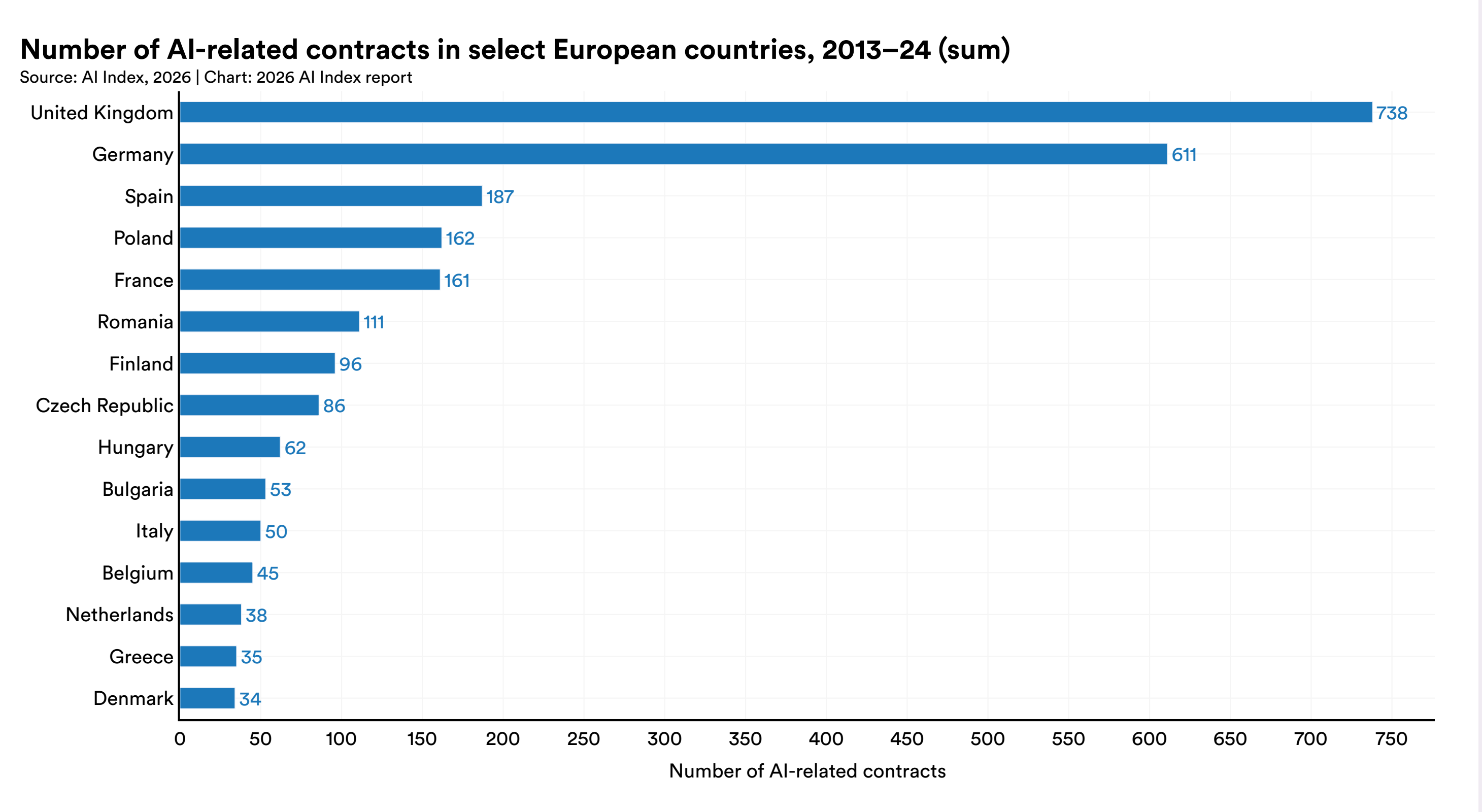

Figure 8.5.9

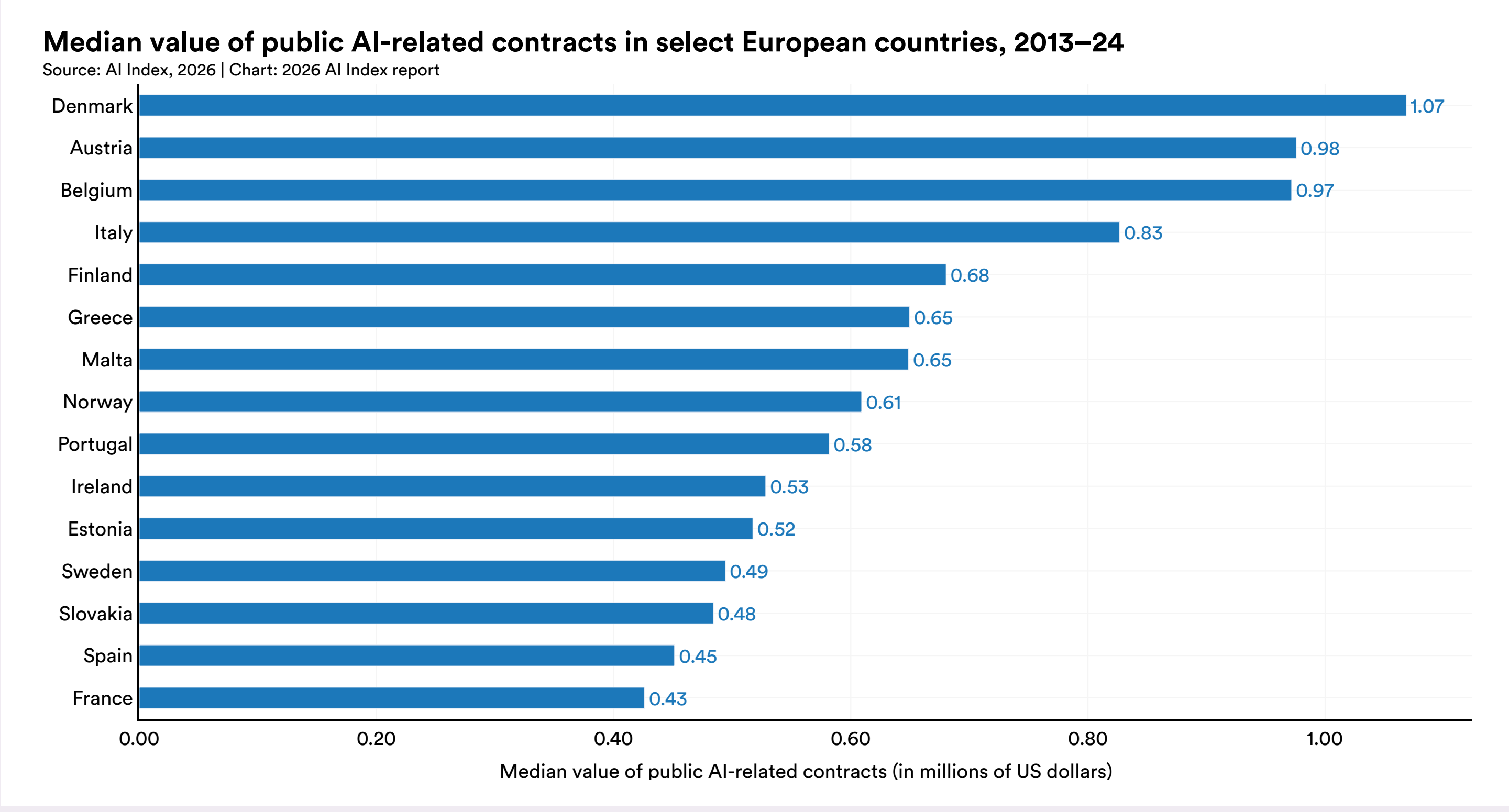

Figure 8.5.10

## Spending Across Sectors

In Europe, government bodies accounted for the largest share of AI-related contract spending in 2024, with the "government, national agency, or public authority" category representing 62.6% of the total (Figure 8.5.11). Health accounted for 13.9% of the spend, followed closely by education, which received 13.7%.

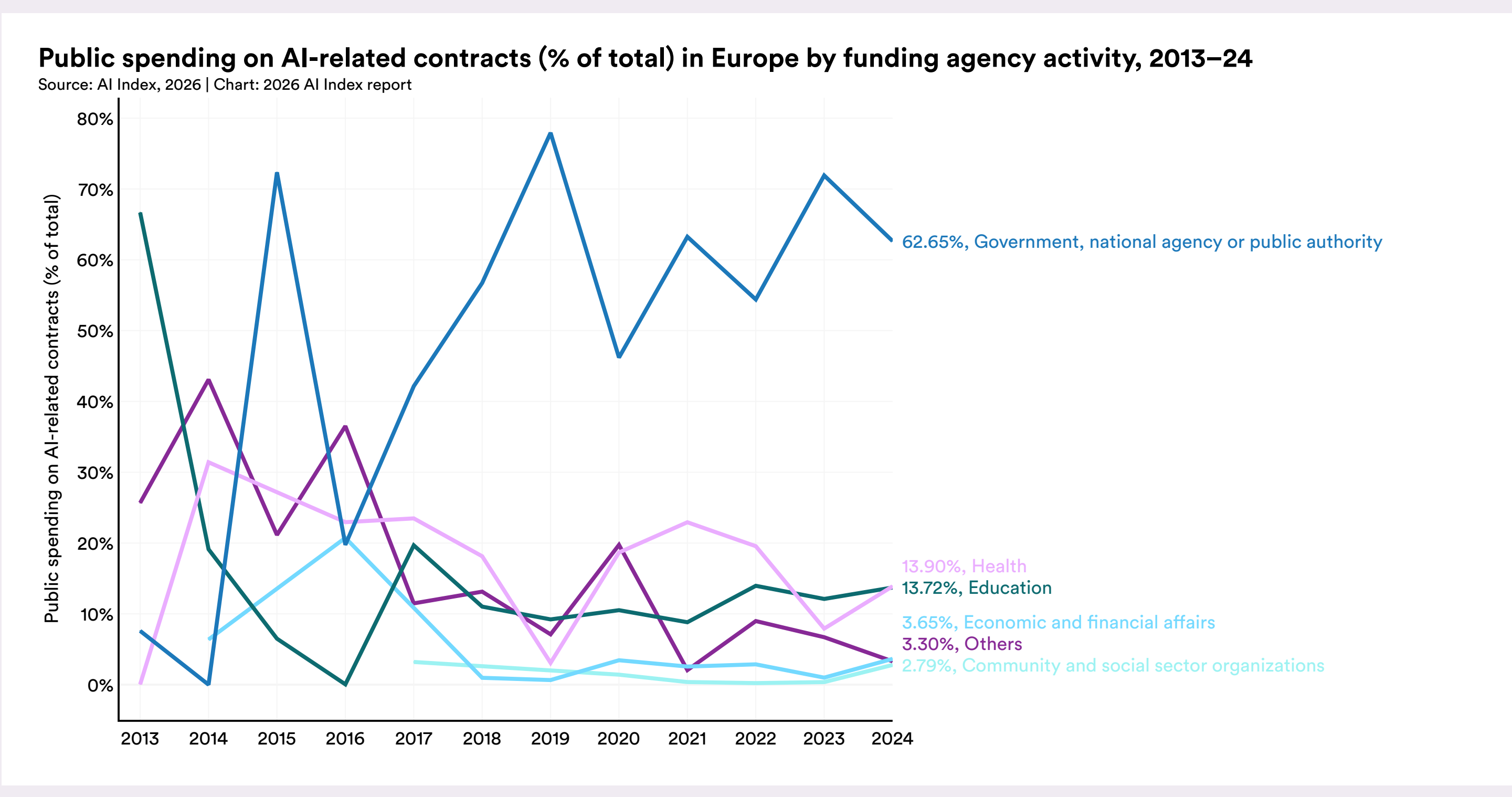

Figure 8.5.11

# 9

# Public Opinion

## Overview

Public views of AI are now shaped by a central tension, as optimism about the technology's benefits often coexists with anxiety about its broader effects. Majorities in most countries say AI's benefits outweigh its drawbacks, but nervousness is growing and trust in institutions to manage the technology remains uneven. AI experts and the general public view the technology's trajectory very differently, with wide gaps on employment, the economy, and healthcare. Southeast Asian countries are consistently the most optimistic and most trusting of their own governments to regulate AI, while North America and Europe report lower expectations and greater skepticism. This chapter tracks these patterns across 30-plus countries, drawing on several large-scale surveys conducted between 2024 and 2026 from Ipsos AI Monitor, Pew Research Center, the University of Melbourne/KPMG Global AI Survey, CHIP50 survey, the LEAP Survey and Elon University's Human Capacities survey.

## Contents

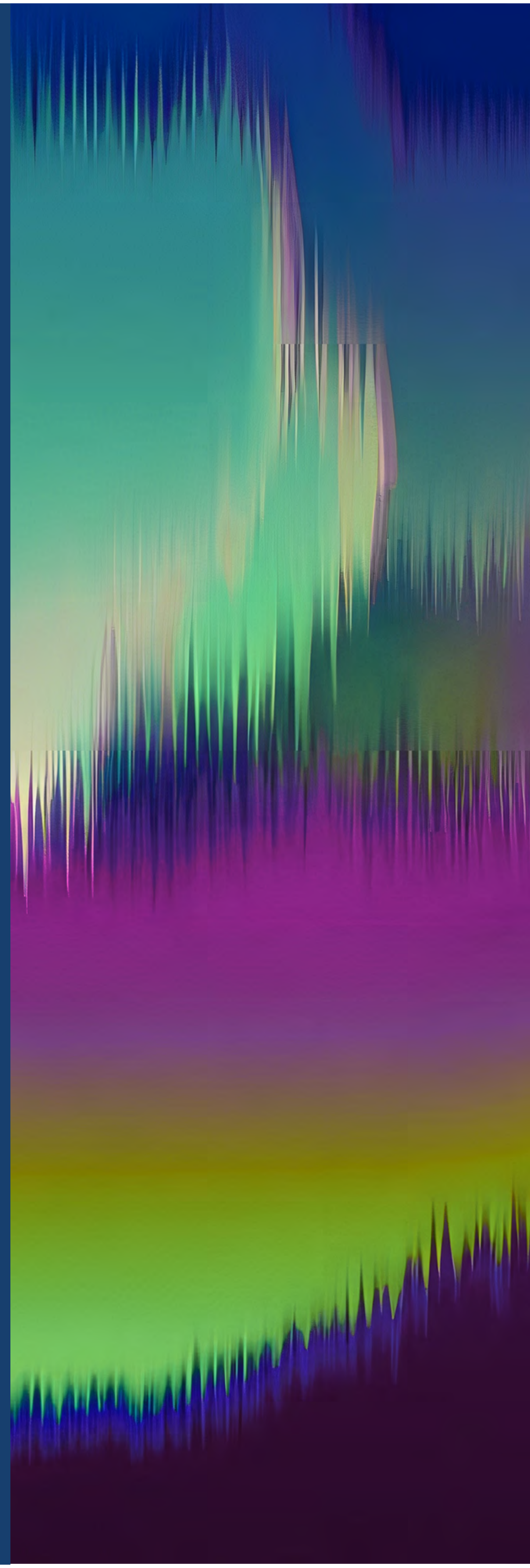

# Chapter Highlights

1 **AI optimism is rising, but so is anxiety.** Globally, the share of respondents who say AI products and services offer more benefits than drawbacks rose from 55% in 2024 to 59% in 2025, even as the share saying these products make them nervous increased to 52%.

2 **Southeast Asian countries remain among the world's most optimistic about AI.** In Malaysia, Thailand, Indonesia, and Singapore, more than 80% of respondents say AI will profoundly change their lives in the next 3-5 years, with Malaysia posting the largest increase from 2024.

3 **India saw the sharpest rise in AI nervousness of any country surveyed.** Between 2024 and 2025, India registered the sharpest rise in concern around AI usage (+14 percentage points) with only a modest increase in excitement (+2).

4 **Workplace AI usage is higher in several emerging economies than in many advanced ones.** In 2025, 58% of employees globally reported using AI at work on a semiregular or regular basis, but in India, China, Nigeria, the United Arab Emirates, and Saudi Arabia, the share exceeded 80%.

5 **AI experts and the U.S. public have very different perspectives on AI's future, except on elections and personal relationships.** On how people do their jobs, 73% of experts expect a positive impact, compared to just 23% of the public, a 50-point gap. Similar divides appear for the economy (69% vs. 21%) and medical care (84% vs. 44%).

6 **Nearly two-thirds of Americans (64%) expect AI to lead to fewer jobs over the next 20 years, while only 5% expect more.** Experts were less pessimistic (39% fewer, 19% more) but forecast far faster adoption, expecting generative AI to assist 18% of U.S. work hours by 2030 versus the public's estimate of 10%.

7 **AI companionship is still niche, and global views vary widely.** More than half of respondents worldwide (52%) reported some excitement about using AI for companionship, compared with just 42% in the United States. Experts forecast that 10% of U.S. adults will use an AI companion daily by 2027, rising to 30% by 2040.

8 **The United States reported the lowest trust in its own government to regulate AI responsibly of any country surveyed, at 31%.** The global average was 54%, with Southeast Asian countries leading (Singapore 81%, Indonesia 76%).

9 **Across all 50 U.S. states, concern about too little AI regulation outweighs concern about too much.** Nationally, 41% of respondents said federal AI regulation will not go far enough, compared with 27% who said it will go too far, though more than one-third were unsure.

10 **Globally, the EU is trusted more than the United States or China to regulate AI effectively.** Across 25 countries in Pew's 2025 survey, a median of 53% said they trust the EU, compared to 37% for the United States and 27% for China.

# 9.1 Global Sentiment Toward AI

This section explores global differences in opinions and perceptions of AI. Since 2022, Ipsos has conducted its annual AI Monitor survey to track public attitudes and perceptions of artificial intelligence worldwide. The set of participating countries has changed over time.[1] The 2025 survey was conducted last year from March 21 to April 4 and covered 30 countries, with a sample size of 23,216 adults.

There are some modest shifts over the years in respondents' opinions, though self-reported AI literacy remains consistent (Figure 9.1.1). Over half of all respondents reported having a good understanding of what AI is and which products and services to use. Over the last year, nervousness has also increased, with the proportion of people who say AI products make them nervous rising by 2 percentage points, to 52%. In tandem, more respondents expressed optimism that the benefits outweigh the drawbacks of AI-enabled products and services, up to 59% from 55% in 2024.

**Global opinions on products and services using AI (% of total), 2022–25**

Source: Ipsos, 2022–25 | Chart: 2026 AI Index report

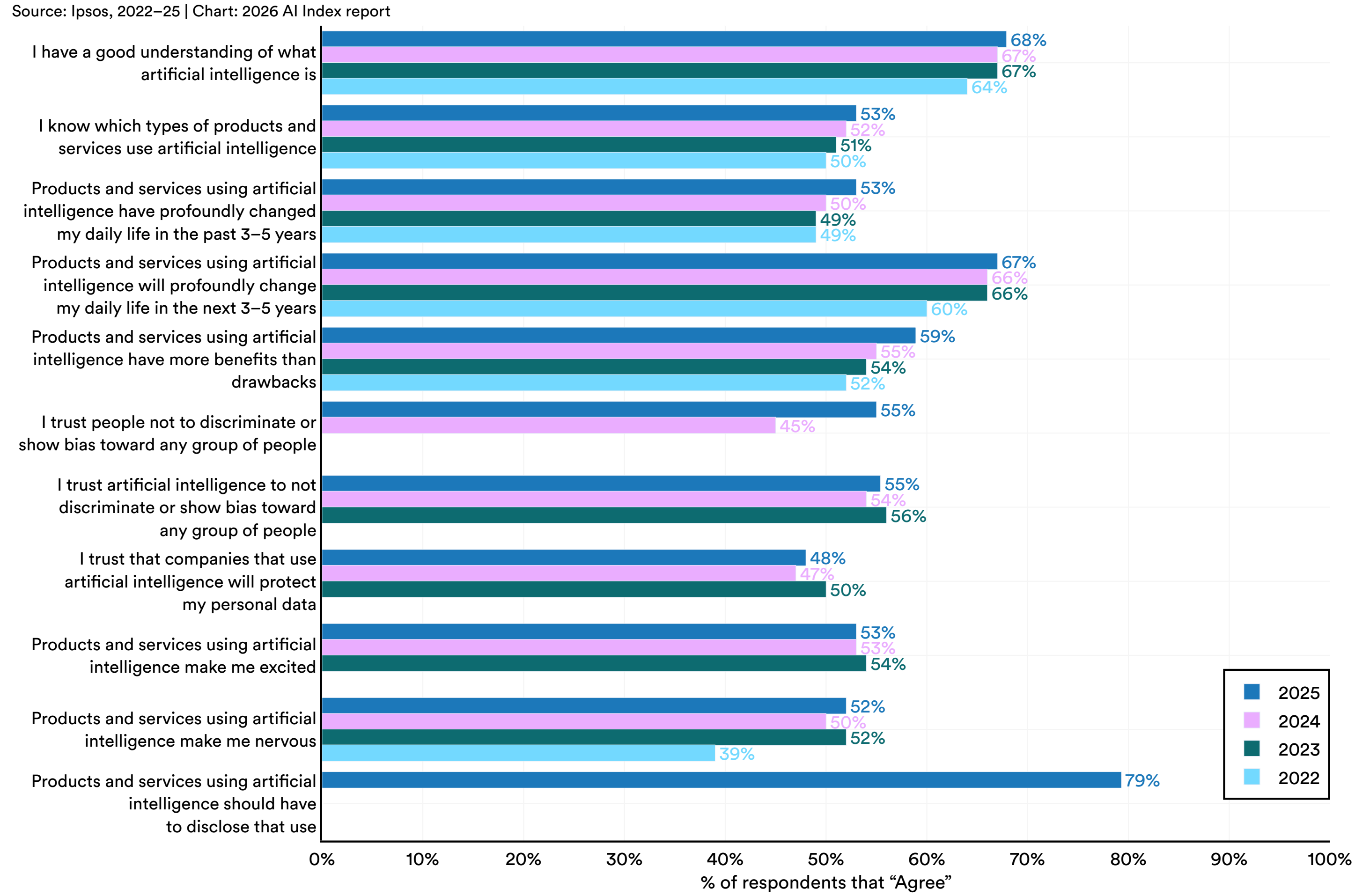

Figure 9.1.1

1 Data for China for the year 2025 was provided by Ipsos but may not appear in their published reports.

The increase in optimism is not uniform across all surveyed countries (Figure 9.1.2). From the 30 countries surveyed by Ipsos, many reported increases between 2022 and 2025 in survey respondents who agreed that the benefits of AI outweigh the drawbacks. Several European countries, in particular, report higher levels of optimism over this period, including Germany (+12 percentage points), France (+10), China (+9), and Great Britain (+5), though their overall sentiment remained lower than in parts of Asia and Latin America.

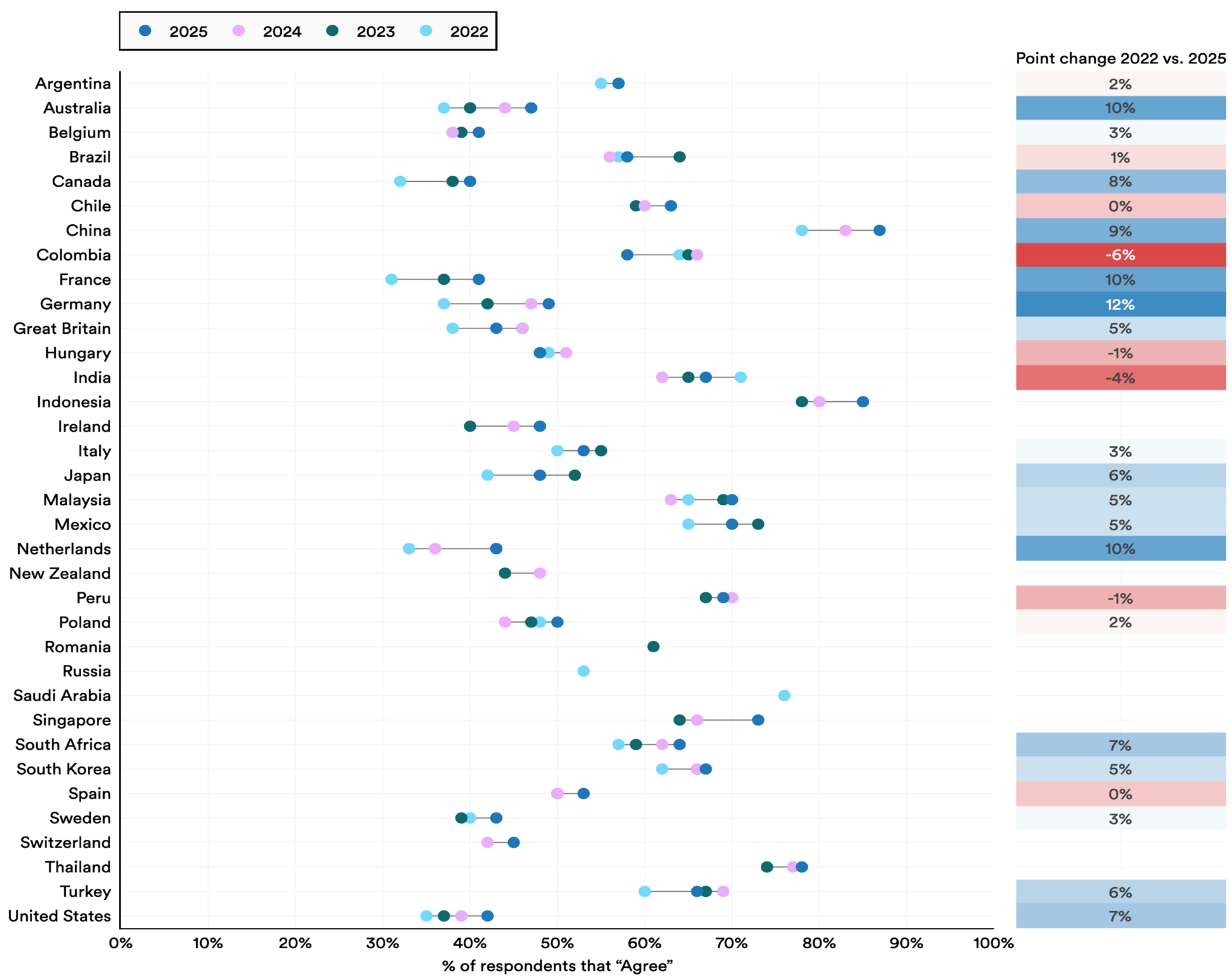

Figure 9.1.2

Southeast Asian nations are among the most optimistic about the future of AI (Figure 9.1.3). In Malaysia, Thailand, Indonesia, and Singapore, over 80% of respondents expect AI to profoundly change their lives over the next three to five years. These countries have consistently ranked at the top of global optimism on AI in recent years, and that sentiment has edged up since 2024, with Malaysia showing the largest increase (+9) (Figure 9.1.4). Respondents from these countries also report higher levels of excitement than nervousness about AI-enabled products and services.

## Opinions about AI by country (% agreeing with statement), 2025

Source: Ipsos, 2025 | Chart: 2026 AI Index report

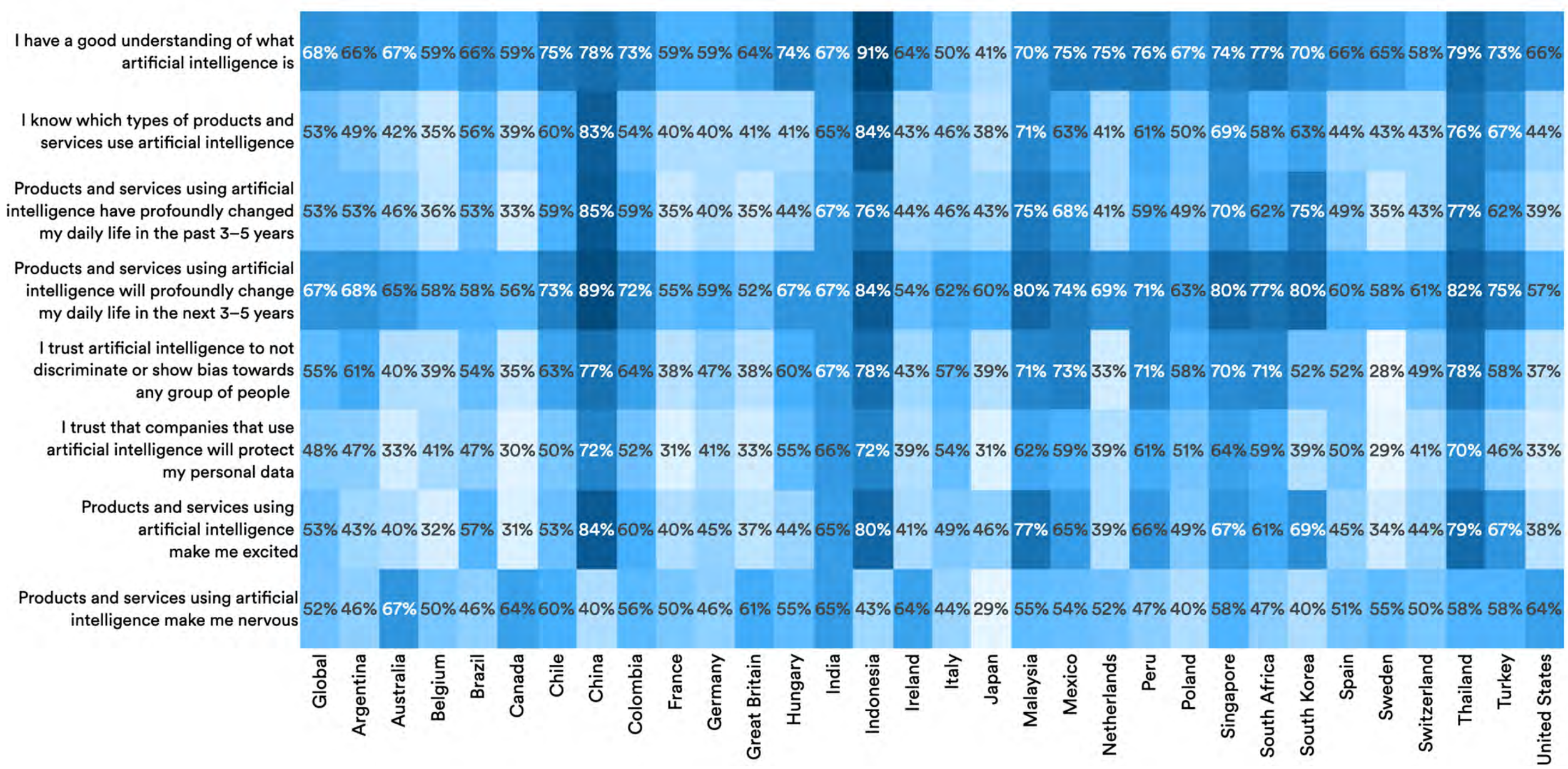

Figure 9.1.3

When looking at year-over-year percentage point changes, global nervousness has increased (+3) and excitement declined (-1) relative to 2024. India shows the sharpest increase in concern around AI usage (+14) with only a modest increase in excitement (+2).

## Percentage point change in opinions about AI by country (% agreeing with statement), 2024–25

Source: Ipsos, 2024–25 | Chart: 2026 AI Index report

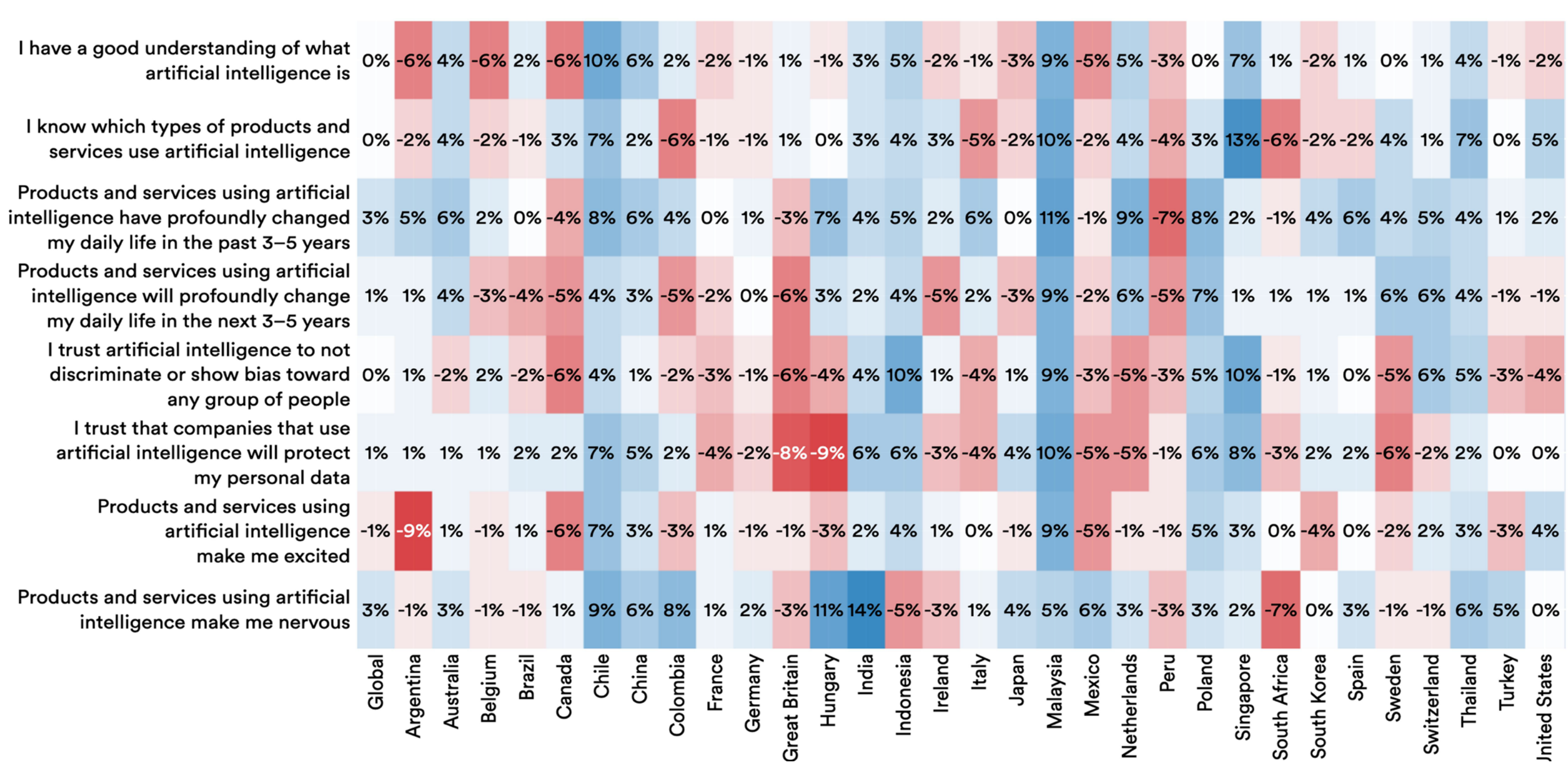

Figure 9.1.4

Across countries, excitement and nervousness about AI do not align closely (Figure 9.1.5). The 2025 distribution mirrors patterns from prior years, with North American and European countries generally clustered at lower levels of excitement and higher levels of nervousness. China and Indonesia show the highest levels of excitement, with nervousness below 50%.

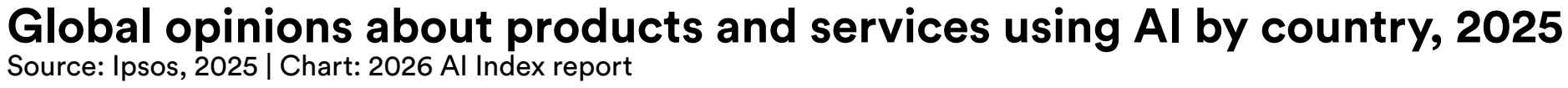

**Global opinions about products and services using AI by country, 2025**
Source: Ipsos, 2025 | Chart: 2026 AI Index report

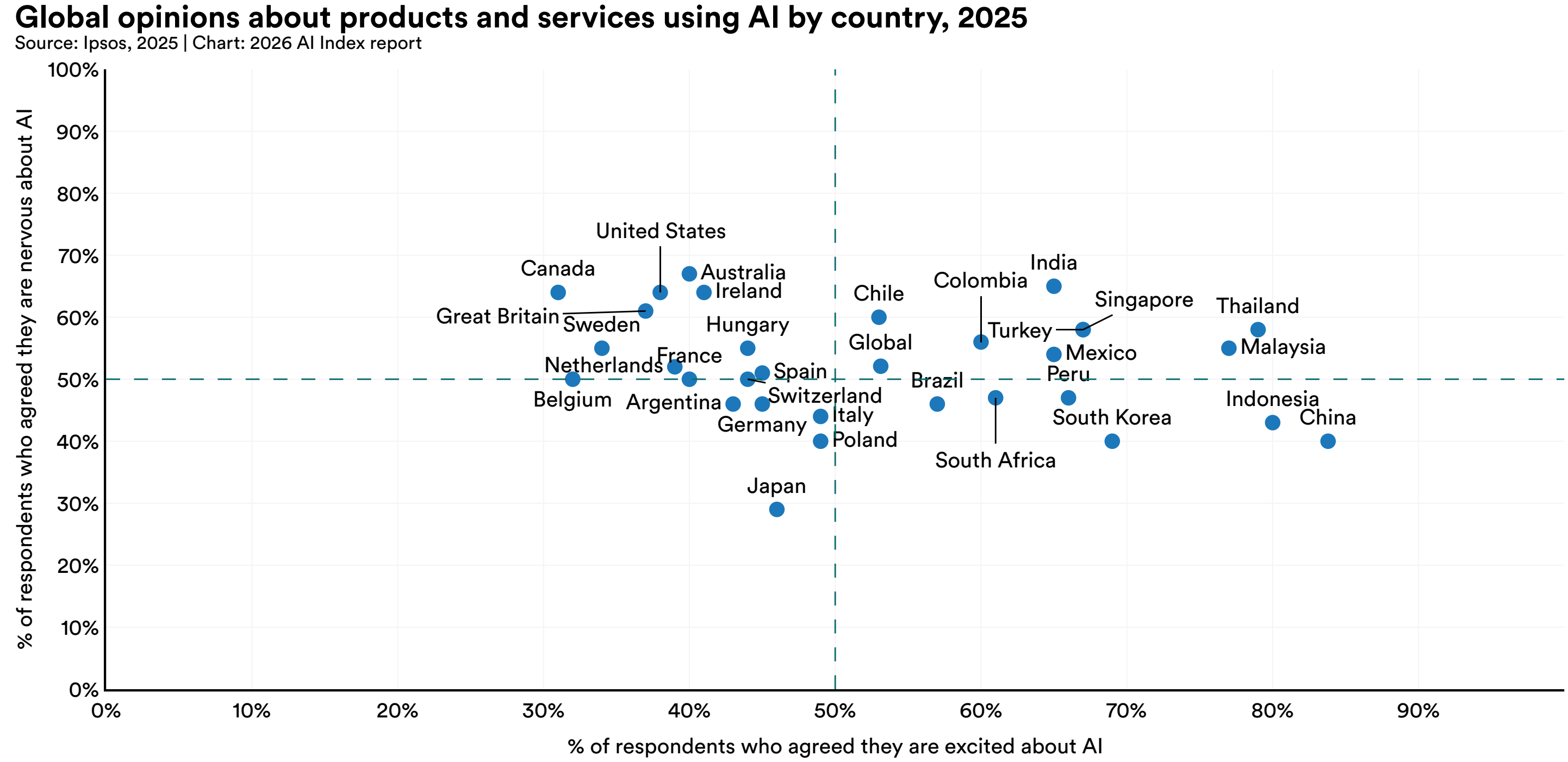

Figure 9.1.5

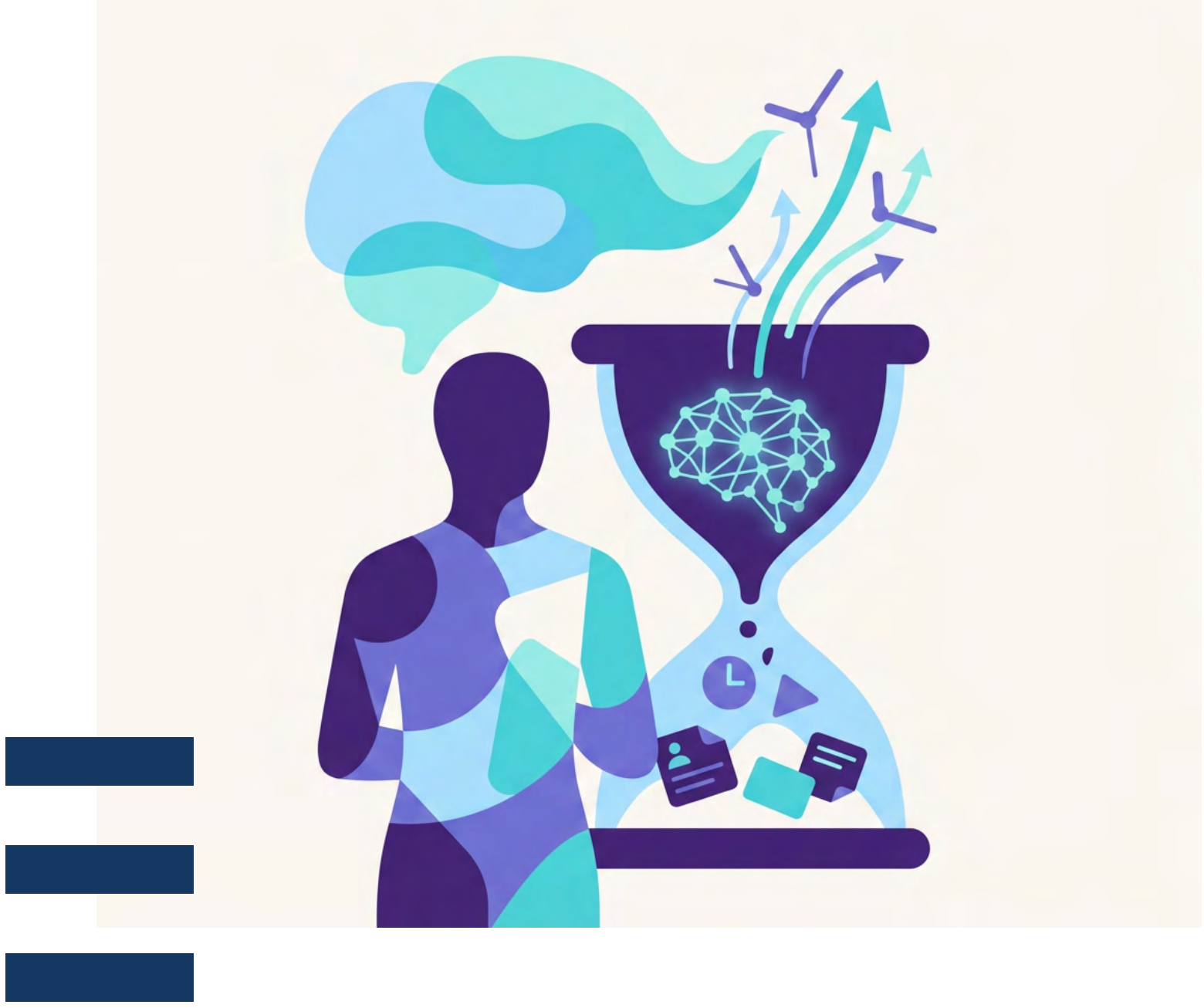

Despite increased nervousness, many respondents continue to associate AI with practical personal benefits, particularly time savings and entertainment (Figure 9.1.6). Globally, 56% of respondents believed AI would reduce the amount of time it takes them to get things done; this figure was even higher in China (78%) and in Southeast Asian countries (>60%). However, respondents were less sure about AI's potential to positively impact their country's economy or job market. North American and European respondents were more skeptical that AI would make their jobs better. In the United States, 33% of respondents said AI would make their jobs better, as opposed to making them worse or having no impact, compared to the global average of 40%. Positive views of AI's personal benefits appear to coexist with concern about its effects on labor markets.

**Global opinions on the potential of AI to improve life by country, 2025**
Source: Ipsos, 2025 | Chart: 2026 AI Index report

| | Global | Argentina | Australia | Belgium | Brazil | Canada | Chile | China | Colombia | France | Germany | Great Britain | Hungary | India | Indonesia | Ireland |
|---|---|---|---|---|---|---|---|---|---|---|---|---|---|---|---|---|
| The economy in my country | 36% | 36% | 26% | 20% | 37% | 20% | 42% | 75% | 38% | 28% | 28% | 24% | 23% | 49% | 49% | 28% |
| The job market | 33% | 41% | 18% | 21% | 40% | 15% | 35% | 52% | 39% | 23% | 21% | 20% | 24% | 48% | 47% | 22% |
| My job | 40% | 39% | 31% | 27% | 44% | 22% | 44% | 65% | 45% | 34% | 28% | 30% | 32% | 55% | 62% | 33% |
| The amount of time it takes me to get things done | 56% | 61% | 49% | 43% | 51% | 39% | 63% | 78% | 66% | 50% | 40% | 41% | 58% | 51% | 80% | 44% |
| My entertainment options (television/video content, movies, music, books) | 50% | 59% | 46% | 35% | 50% | 35% | 59% | 74% | 65% | 33% | 44% | 40% | 41% | 50% | 68% | 39% |
| My health | 40% | 46% | 31% | 29% | 43% | 25% | 48% | 65% | 47% | 41% | 26% | 31% | 26% | 50% | 60% | 30% |
| The amount of disinformation on the internet | 30% | 39% | 18% | 17% | 33% | 12% | 37% | 54% | 34% | 19% | 17% | 20% | 20% | 49% | 50% | 23% |

| | Italy | Japan | Malaysia | Mexico | Netherlands | Peru | Poland | Singapore | South Africa | South Korea | Spain | Sweden | Switzerland | Thailand | Turkey | United States |
|---|---|---|---|---|---|---|---|---|---|---|---|---|---|---|---|---|
| The economy in my country | 25% | 21% | 54% | 43% | 30% | 43% | 32% | 57% | 45% | 29% | 26% | 21% | 32% | 54% | 31% | 28% |
| The job market | 27% | 20% | 45% | 45% | 19% | 50% | 24% | 39% | 45% | 14% | 25% | 20% | 25% | 51% | 45% | 20% |
| My job | 34% | 20% | 54% | 47% | 28% | 51% | 25% | 49% | 50% | 18% | 30% | 31% | 33% | 60% | 42% | 33% |
| The amount of time it takes me to get things done | 50% | 35% | 69% | 62% | 58% | 65% | 50% | 64% | 77% | 60% | 48% | 42% | 46% | 69% | 56% | 45% |
| My entertainment options (television/video content, movies, music, books) | 41% | 33% | 61% | 65% | 43% | 61% | 38% | 56% | 69% | 51% | 45% | 36% | 43% | 75% | 57% | 40% |
| My health | 33% | 18% | 50% | 52% | 25% | 50% | 25% | 45% | 55% | 34% | 31% | 23% | 29% | 59% | 43% | 31% |
| The amount of disinformation on the internet | 26% | 22% | 40% | 39% | 19% | 45% | 20% | 34% | 45% | 24% | 25% | 15% | 23% | 51% | 37% | 17% |

Figure 9.1.6

## Global Perceptions of AI's Impact on Jobs

In both 2024 and 2025, Ipsos asked respondents how likely they thought it was that AI would change their job or completely replace it within the next five years. Results from 2025 show that perceptions remained stable year over year (Figure 9.1.7). In 2025, 22% of respondents said it was "very likely" AI would change how they do their current job, compared to 21% in 2024. In both years, the share that said it was "not likely" remained unchanged at 32%. Expectations around job replacement showed the same consistency. In 2024 and 2025, 11% of respondents reported that it was "very likely" AI would replace their job within the next five years, and 56% said this was "not likely".

When asked whether AI is generally more likely to create new jobs or eliminate existing ones, views in 2025 were divided (Figure 9.1.8). Country-level expectations follow similar patterns to the earlier sentiment trends. Nigeria, Japan, Mexico, the United Arab Emirates, South Korea, and India all expected AI to create more jobs than it eliminates, with shares above 60%. The United States and Canada sat at the opposite end, where 67% and 68% of respondents expected AI to eliminate jobs and disrupt industries.

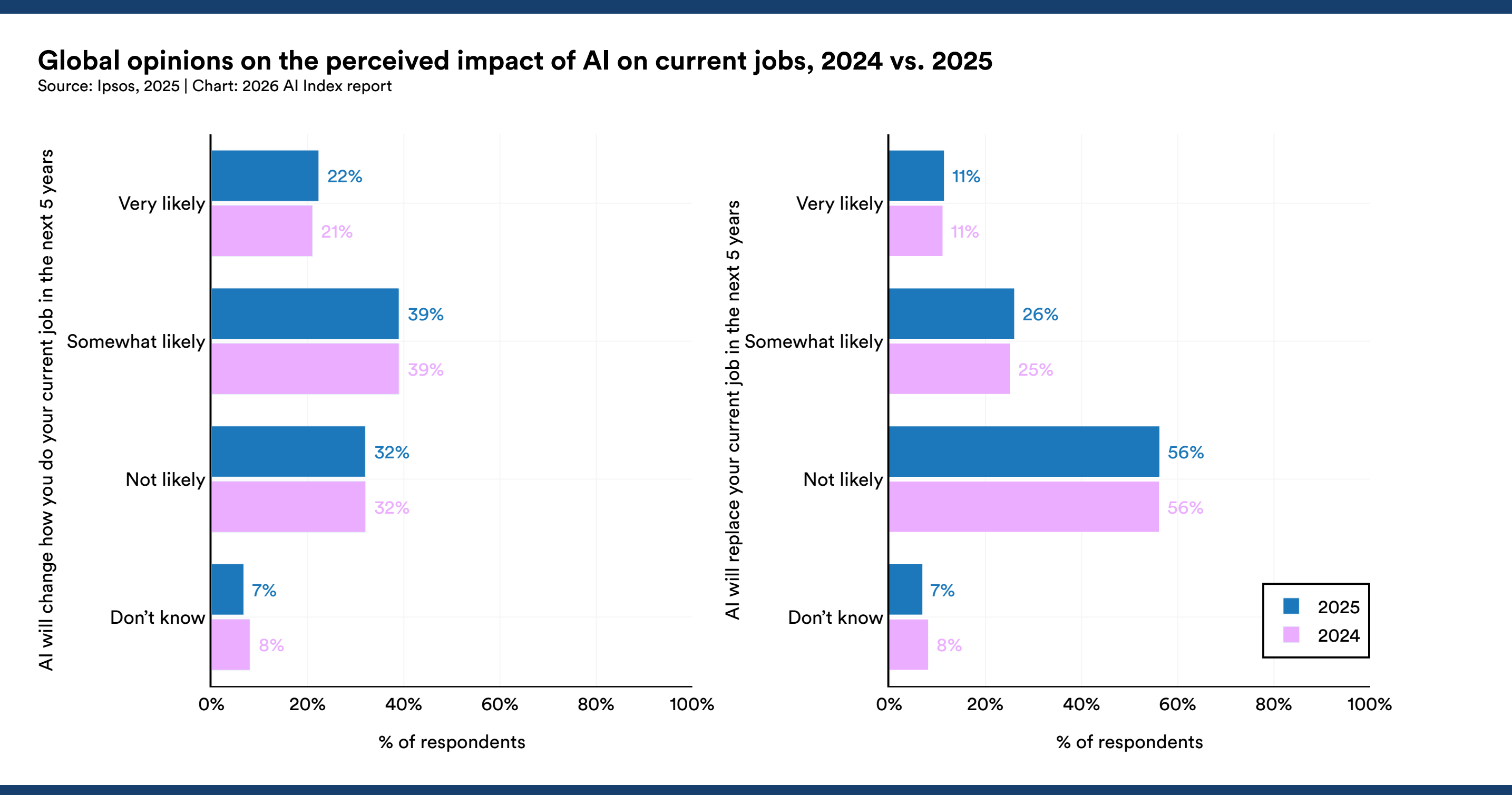

Figure 9.1.7

**Global expectations about AI creating new jobs vs. eliminating jobs, 2025**

Source: Ipsos, Google 2026 | Chart: 2026 AI Index report

| | Create new jobs and new ways of working | Eliminate jobs and disrupt industries |
|---|---|---|
| Global | 50% | 50% |
| Nigeria | 73% | 27% |
| Japan | 69% | 31% |
| Mexico | 64% | 36% |
| United Arab Emirates | 63% | 37% |
| South Korea | 63% | 37% |
| India | 63% | 37% |
| Brazil | 59% | 41% |
| Singapore | 57% | 43% |
| Argentina | 50% | 50% |
| South Africa | 45% | 55% |
| Poland | 45% | 55% |
| Spain | 45% | 55% |
| Australia | 43% | 57% |
| France | 42% | 58% |
| Belgium | 42% | 58% |
| Ireland | 41% | 59% |
| Italy | 41% | 59% |
| Germany | 39% | 61% |
| United Kingdom | 38% | 62% |
| Canada | 32% | 68% |
| United States | 29% | 67% |

0% 20% 40% 60% 80% 100%

% of respondents

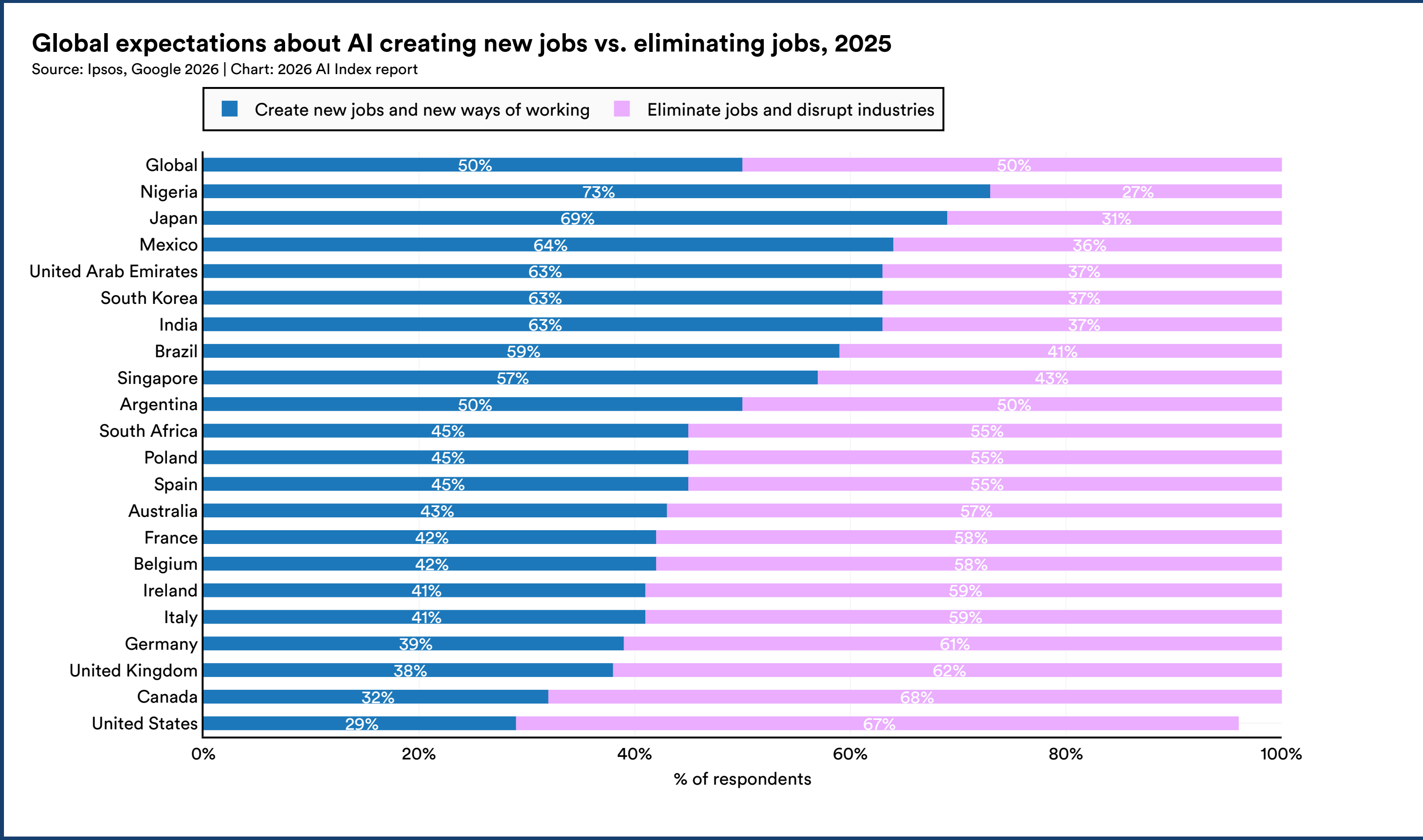

Figure 9.1.8[2]

2 In Ipsos' reporting of findings, percentage points are rounded to the nearest whole number. As a result, figures may not add up to exactly 100%.

Respondents were also asked whether AI would make the job market and their own jobs better, worse, or stay the same over the next five years. Optimism on both measures is low, under or around 50%, in most countries surveyed (Figure 9.1.9). China, Indonesia, Thailand, and Singapore report more positive expectations around AI's impact on jobs, both individually and economy-wide. North America and Europe have lower expectations, though respondents there were more positive about how AI might improve their individual jobs compared to the overall job market. Chapter 4 of the AI Index further explores the technology's impact on the global economy and labor markets.

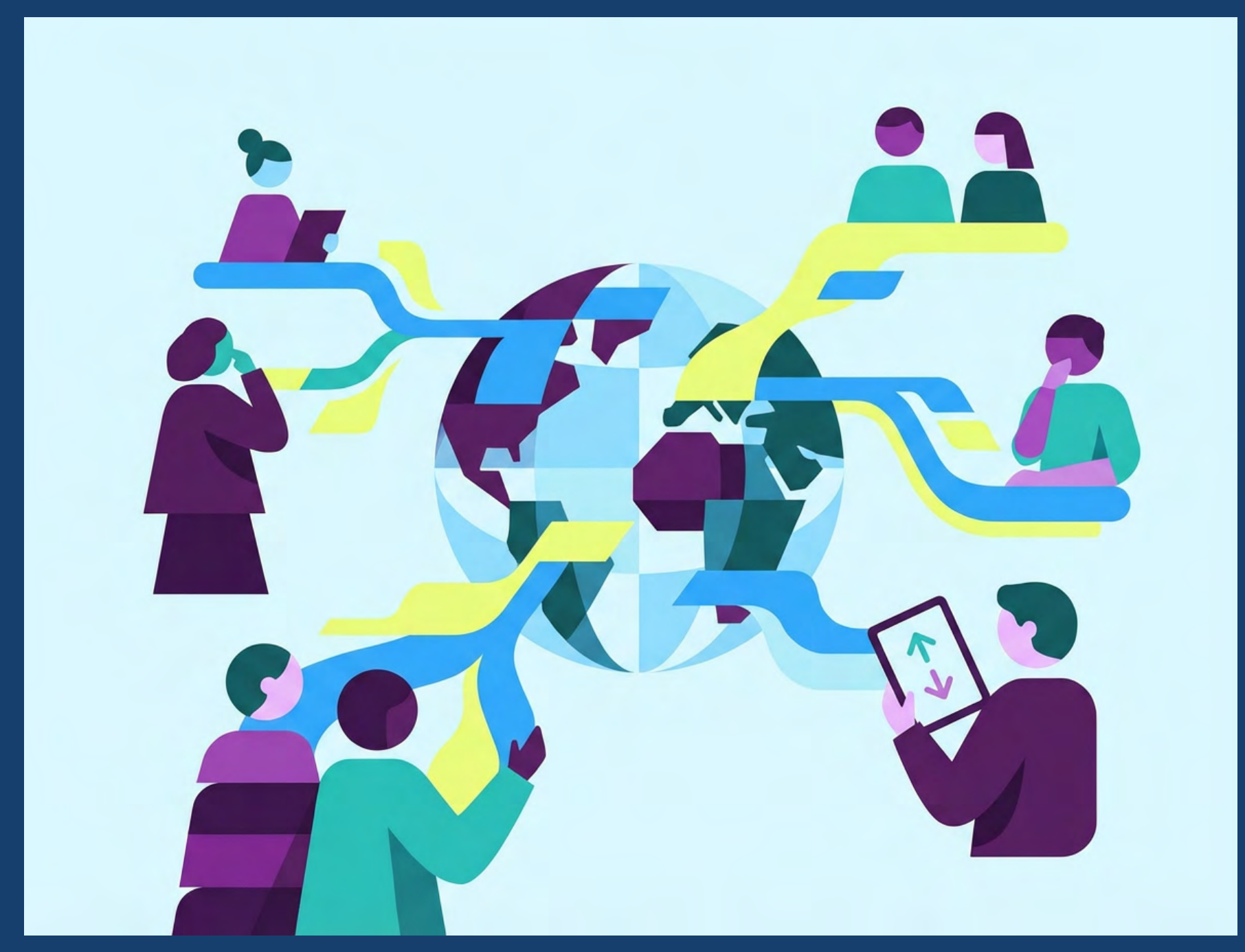

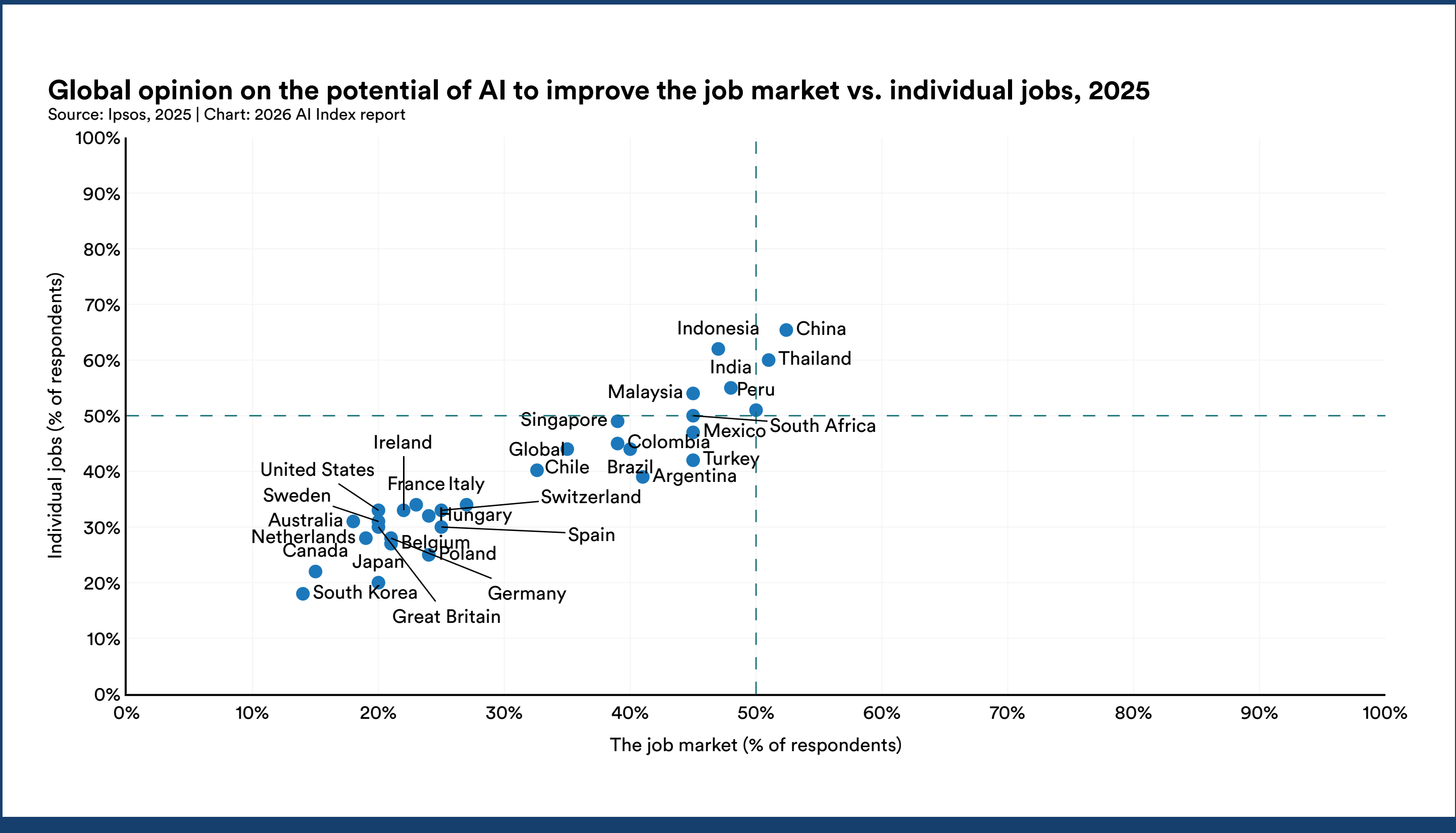

Figure 9.1.9

**HIGHLIGHT:**

## Global AI Use in the Workplace

Since 2022, the use of AI technology within organizations has become more prevalent. To capture that transformation across workplaces, the University of Melbourne fielded a global survey[3] of 48,340 people across 47 countries, examining how employees are adopting and using AI at work. Respondents were asked if they rely on AI to inform decisions, and whether they feel comfortable sharing the information AI tools need to carry out tasks.

Globally, the share of employees who intentionally use AI at work continues to grow. In 2025, 58% of employees reported using AI on a semiregular or regular basis, and just over half (53%) said they trust AI for work purposes (Figure 9.1.10). From a regional perspective, the results reveal notable differences. Employees in emerging economies remain the most active users of AI in the workplace: In India, China, Nigeria, the United Arab Emirates, and Saudi Arabia, over 80% of respondents said they regularly use AI at work, and trust levels in these countries are similarly high. By contrast, in most North American and European countries, about half of employees report using AI tools regularly, while trust tends to fall several points lower, between 40% and 48%. The regional patterns in workplace adoption contrast with the population-level diffusion data discussed in Chapter 4, where AI adoption shows a strong, statistically significant positive correlation with GDP per capita.

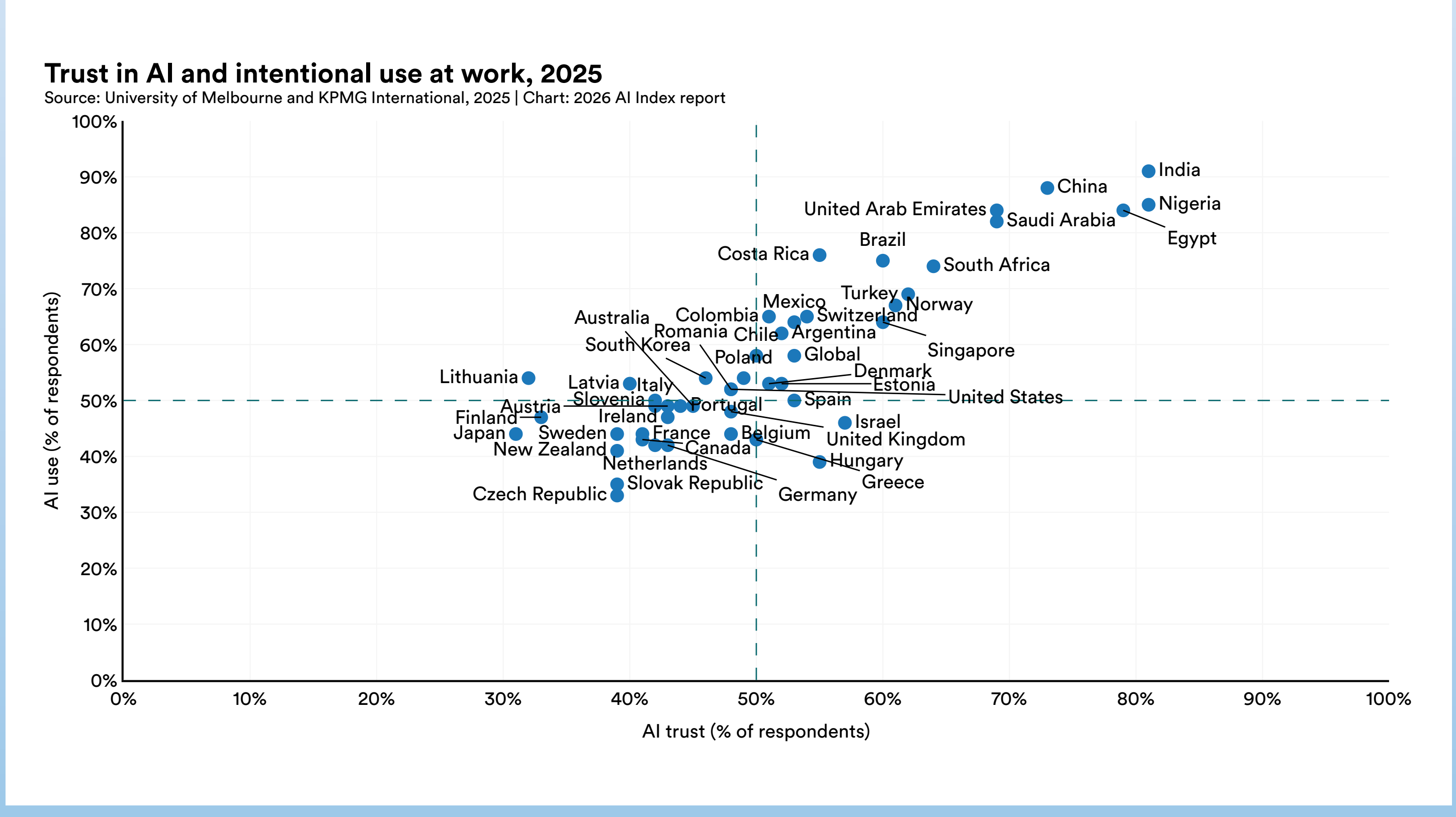

Figure 9.1.10

This section draws on multiple U.S.-focused surveys to compare how the public and AI experts view AI's

3 These results come from an online survey, which can overrepresent younger, more urban, and more educated respondents in emerging economies. The study authors note that country-level differences hold after controlling for age and education.

**HIGHLIGHT:**

The survey also asked employees about their organization's level of support for AI strategy, AI literacy, and AI governance (Figure 9.1.11). Respondents reflected on whether their organization had a coherent AI strategy and supported adoption, AI literacy, and responsible use, including training, as well as governance practices such as clear policies, monitoring, accountability, and data privacy and security measures.

Consistent with usage and trust levels, organizational support was reported highest in emerging economies. In India, around 85%–90% of respondents said their organization supports AI strategy, literacy, and governance. Nigeria, Egypt, China, and the UAE also rank among the top countries for organizational support. At the other end, respondents in Japan, Korea, and Portugal report the lowest levels of support for AI literacy, along with less confidence in responsible AI governance.

Overall, most countries reported less organizational support for responsible AI governance, in comparison to literacy and strategy. Chapter 3 further explores this governance gap, and the key barriers to responsible AI implementation.

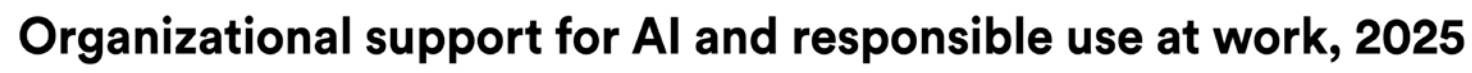

Organizational support for AI and responsible use at work, 2025
Source: University of Melbourne and KPMG International, 2025 | Chart: 2026 AI Index report
AI strategy and culture
Support for AI literacy
Responsible AI governance
India
Nigeria
Egypt
China
United Arab Emirates
Saudi Arabia
Turkey
South Africa
Brazil
Singapore
Costa Rica
Switzerland
United Kingdom
Mexico
Colombia
Norway
Italy
Denmark
United States
Argentina
Chile
Romania
Estonia
Australia
Poland
Ireland
Hungary
Canada
Belgium
Spain
Latvia
Sweden
France
Lithuania
Japan
Germany
Austria
Greece
Israel
Netherlands
Slovak Republic
New Zealand
South Korea
Finland
Czech Republic
Slovenia
Portugal
30%
35%
40%
45%
50%
55%
60%
65%
70%
75%
80%
85%
90%
95%
% of respondents that "Agree"

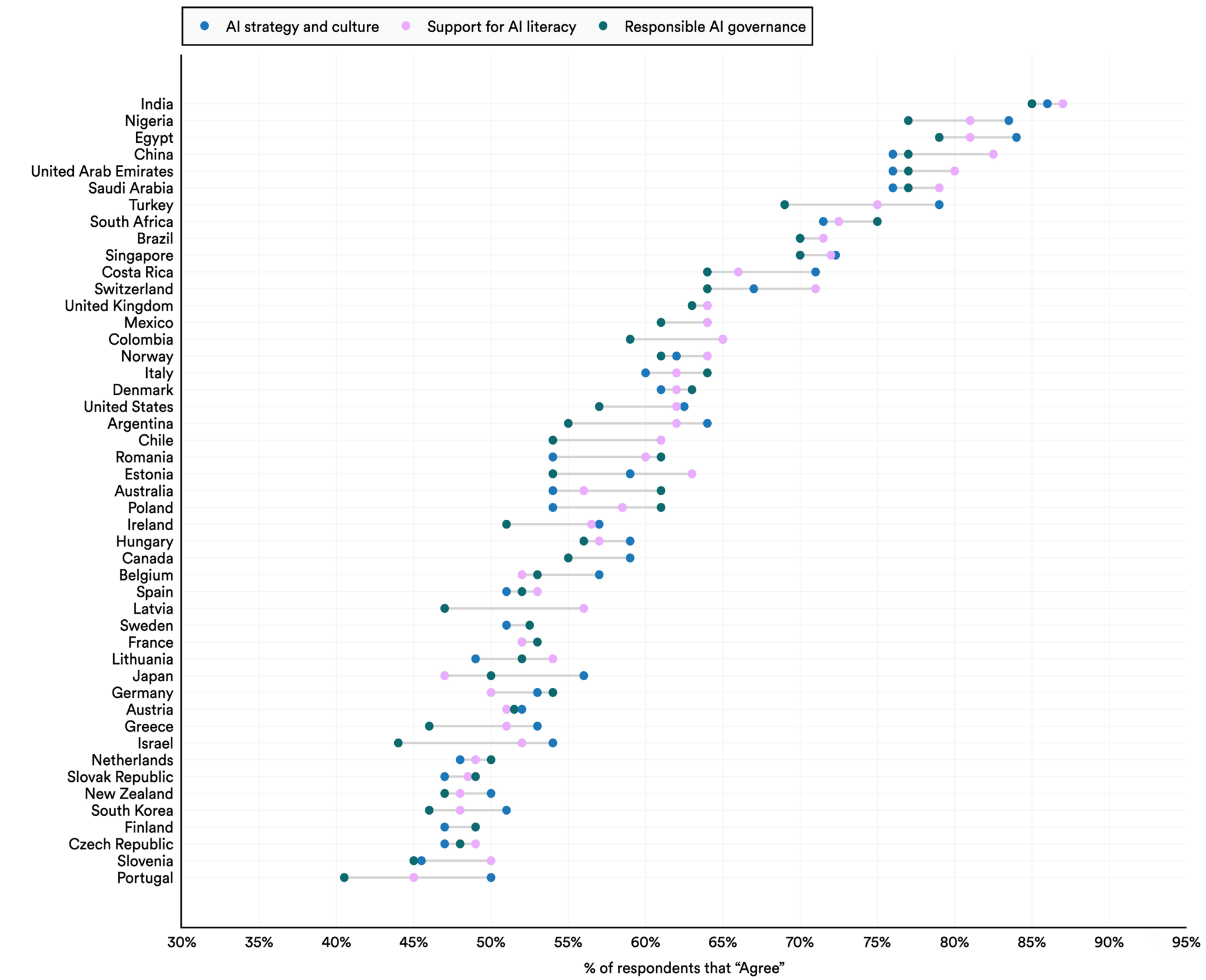

Figure 9.1.11

# 9.2 US Public and Expert Views on AI's Societal Impact

societal impact. The main sources[4] are Pew Research Center's 2024 survey of U.S. adults and AI experts, Elon University University Imagining the Digital Future Center's 2025 survey on expected effects on human capacities by 2035, and the Longitudinal Expert AI Panel (LEAP), conducted by the Forecasting Research Institute. For the Pew survey, AI experts were U.S.-based authors or presenters at AI-related conferences in 2023 or 2024 who reported that their work or research relates to AI.

Across nearly every topic surveyed, experts report more optimism than the U.S. public (Figure 9.2.1). The largest gaps show up around the future of work: 73% of AI experts said AI will have a positive impact on how people do their jobs, compared to 23% of U.S. adults. Similar gaps appear for the economy (69% vs. 21%), K–12 education (61% vs. 24%), and medical care (84% vs. 44%). For both groups, however, optimism is low in domains tied to trust and social connection, including elections, news, and personal relationships.

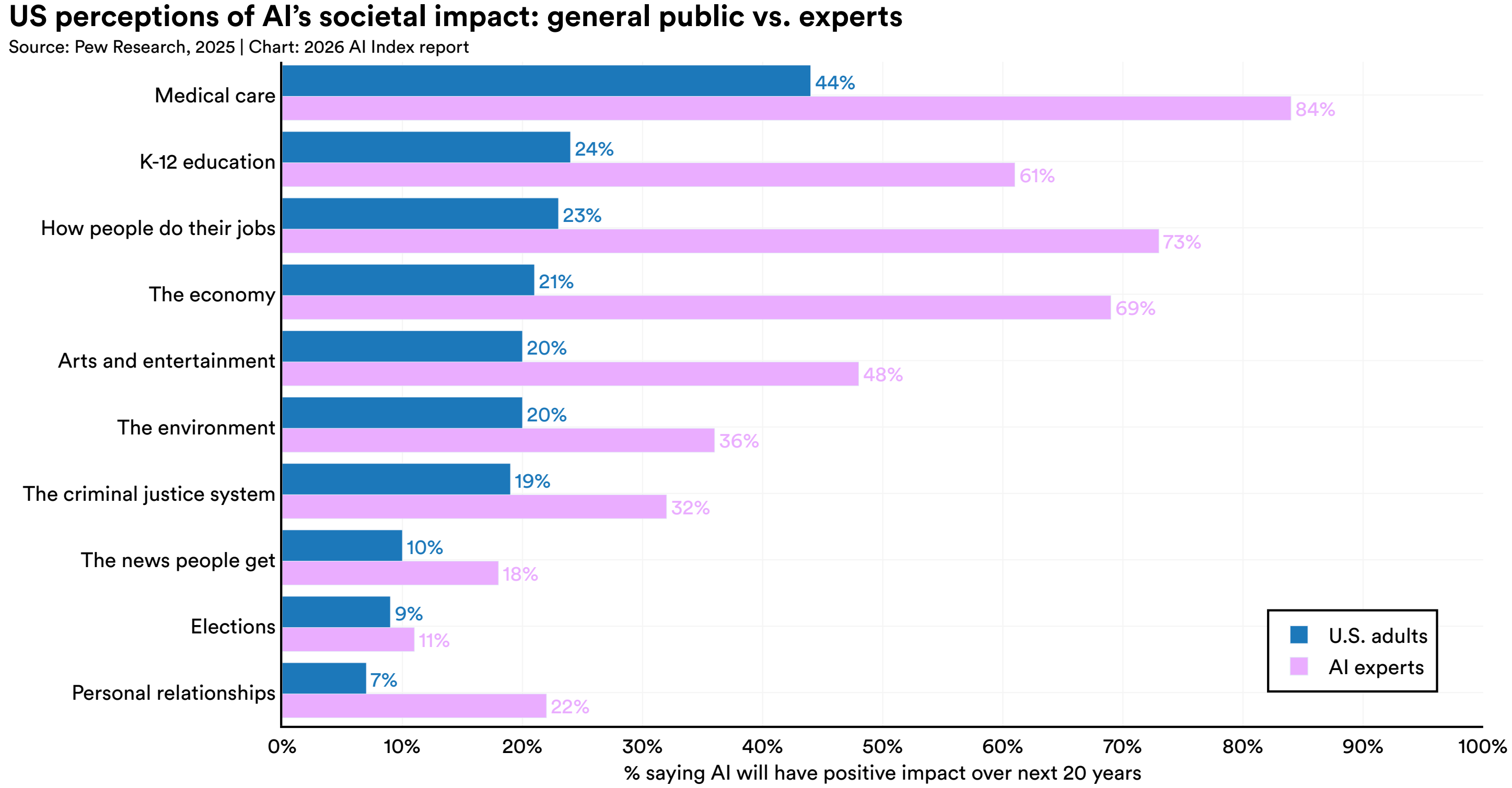
US perceptions of AI's societal impact: general public vs. experts
Source: Pew Research, 2025 | Chart: 2026 AI Index report

Figure 9.2.1

4 Sources: McClain, C. et al. (2025). *How the U.S. public and AI experts view artificial intelligence*. Pew Research Center. This report covers multiple research components conducted in 2024, including a U.S. survey of 5,410 adults conducted August 12–18, 2024, a survey of 1,013 AI experts living in the United States conducted August 14–October 31, 2024, and in-depth interviews with 30 individual AI experts conducted October 18–November 26, 2024. Rainie, L., & Anderson, J. (2025). *Many Americans expect AI to have significant negative impact on human capacities and behaviors such as social and emotional intelligence, analytical thinking and agency by 2035*. Imagining the Digital Future Center at Elon University. This national survey of 1,005 U.S. adults was conducted July 17–20, 2025. Kennedy, B. et al. (2025). *How Americans view AI and its impact on people and society.* Pew Research Center. This survey was conducted June 9–15, 2025, with a sample of 5,023 U.S. adults.

When asked to look ahead to 2035, the U.S. public is again more pessimistic than AI experts about the impact the technology is likely to have on key human traits such as thinking, learning, and creativity (Figure 9.2.2). U.S. adults are more likely than AI experts to anticipate negative effects on metacognition (53% vs. 36%), defined as the ability to think analytically about one's own thinking process, and decision-making (48% vs. 30%), which refers to problem-solving abilities. For social and emotional intelligence, defined as the ability to understand and manage social interactions, 51% of U.S. adults and 34% of experts expect AI to have a negative impact. Concern about mental well-being is high for both groups, with 55% of adults and 53% of experts saying AI will have a negative effect.

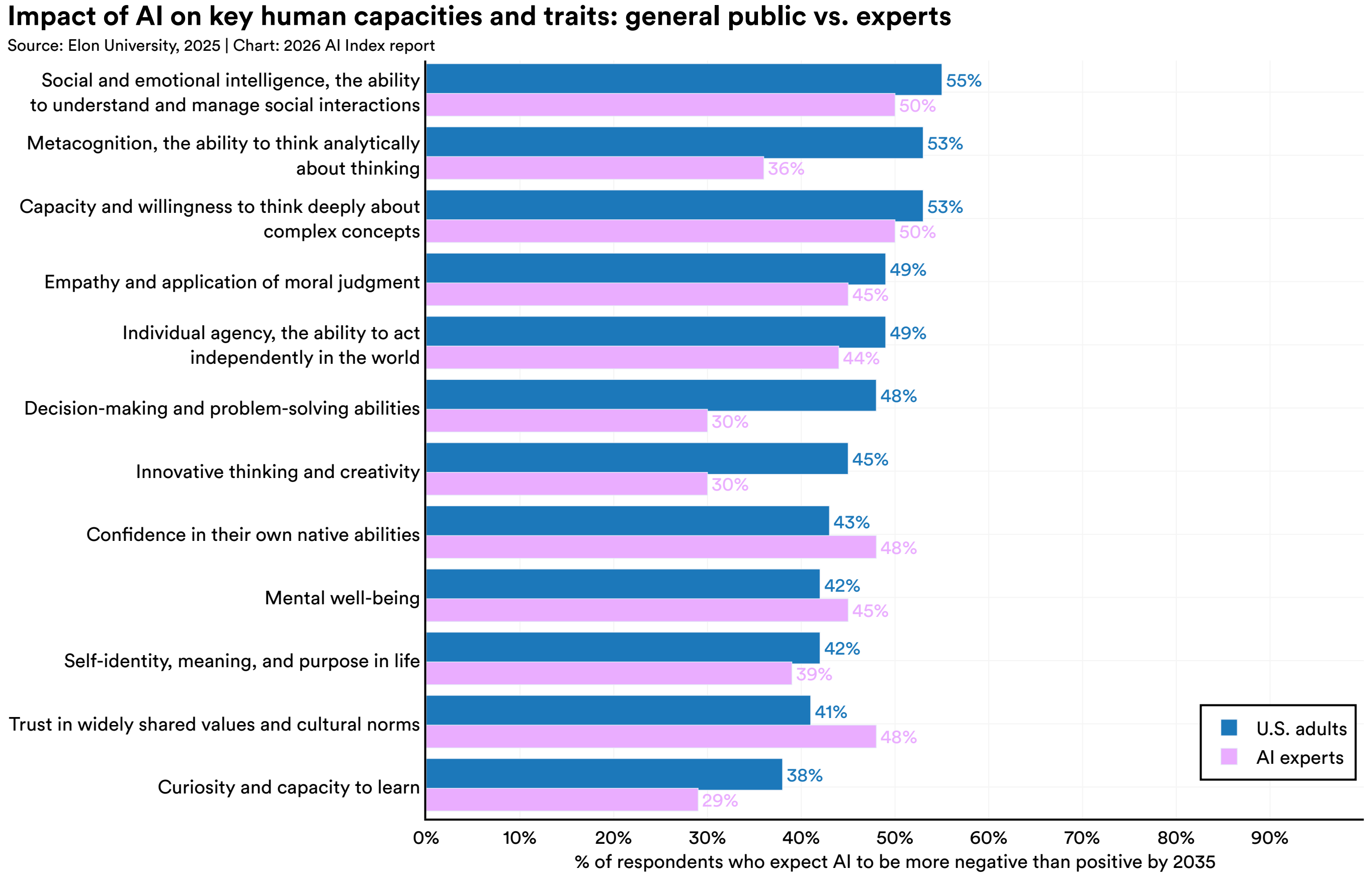

Figure 9.2.2

Beyond general sentiment, recent forecasting data shows even wider gaps in expected timelines and scale. The Longitudinal Expert AI Panel (LEAP), conducted by the Forecasting Research Institute, surveyed AI experts and the general public on specific AI milestones and adoption rates. Across 68 forecasts, experts consistently predicted much faster AI progress than the public.

In capability-focused forecasts, public views align with experts in only 9% of cases. When they diverge, the public expects slower progress 71% of the time. In direct comparison, experts are 16% more likely than the public to predict faster progress. Across specific metrics, the gaps are even more significant. By 2030, AI experts expect higher accuracy on complex math problems (+25 points), more AI-assisted work (+8.2), and greater adoption of autonomous ride-hailing (+8) (Figure 9.2.3). The public predicts greater electricity consumption by AI and a higher probability that AI solves a major mathematical problem. Looking further out to 2040, experts project a high likelihood of a transformative technological event occurring (+30) and much higher rates of daily AI companion use (+10) and AI-discovered drugs (+10). The gap in capability forecasts between the public and experts is notable as model performance continues to accelerate across a range of technical benchmarks. Chapter 2 tracks several of the significant breakthroughs of the past year.

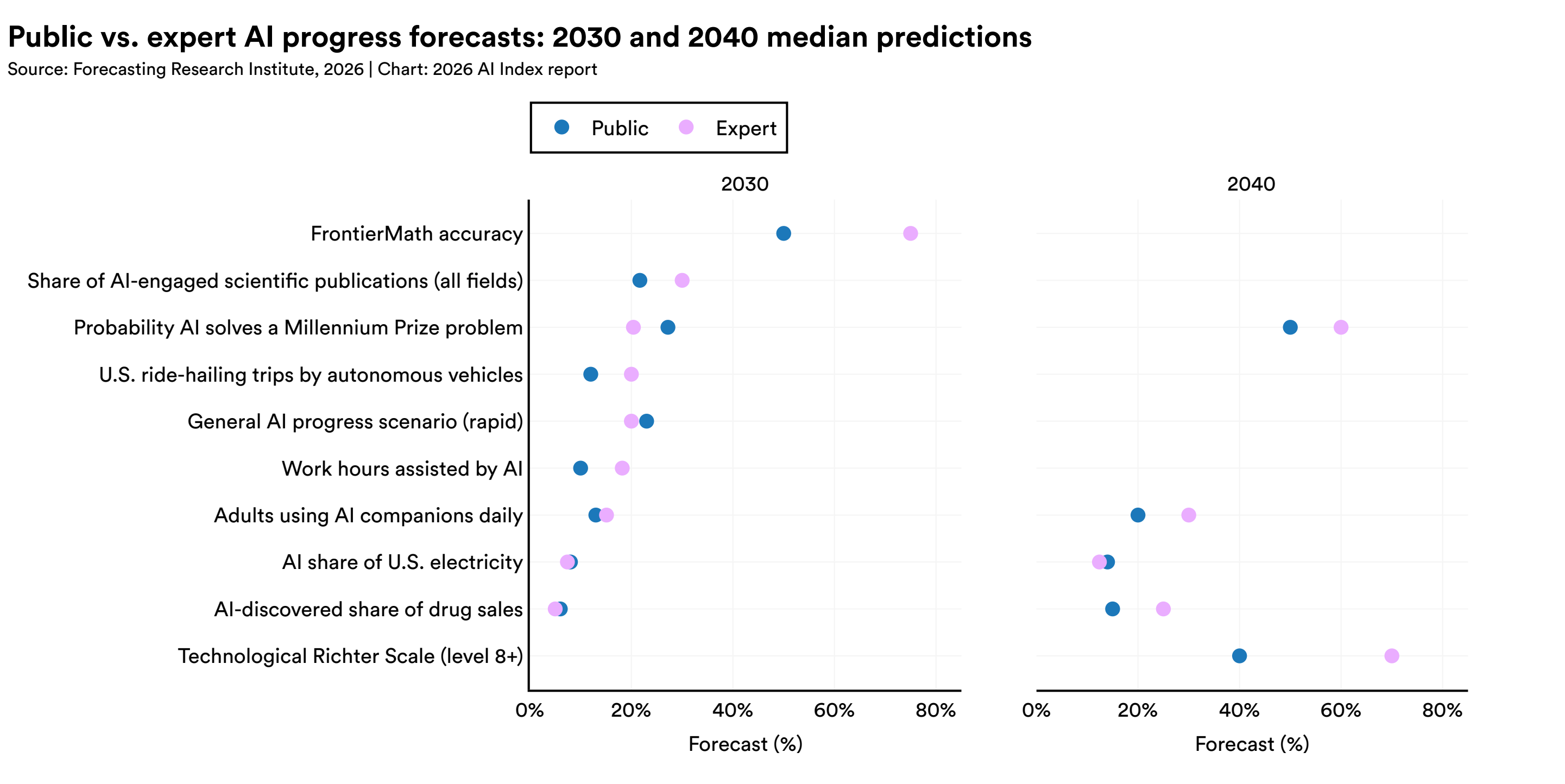

Figure 9.2.3[5]

Views on employment over the long term show a similar pattern (Figure 9.2.4). Nearly two-thirds or 64% of U.S. adults said AI will lead to fewer jobs in the next 20 years, while 5% said more jobs. Among experts, 39% predicted fewer jobs and 19% predicted more.

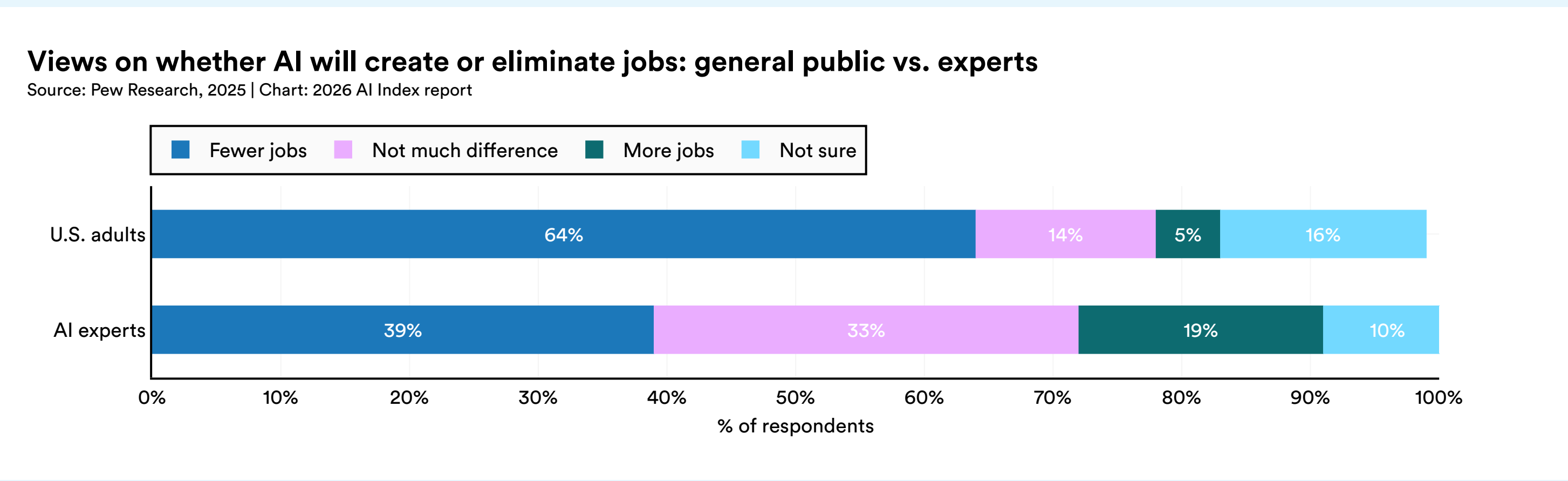

Figure 9.2.4

Experts forecast much faster workplace adoption than the public. The median prediction among experts is that generative AI will assist 8% of U.S. work hours in 2027, rising to 18% in 2030. The top 25% (75th percentile) of expert predictions is for over 30% AI-assisted work hours by 2030, compared to the top 10% (90th percentile) of predictions, at more than 40%. In contrast, the public expects slower adoption, at 10% by 2030 (Figure 9.2.5).

5 Not all questions in the LEAP survey were asked for both 2030 and 2040. Forecast horizons vary by topic and were set according to what was most meaningful or measurable for each question. As a result, the absence of a 2040 value for some items reflects survey design rather than missing responses.

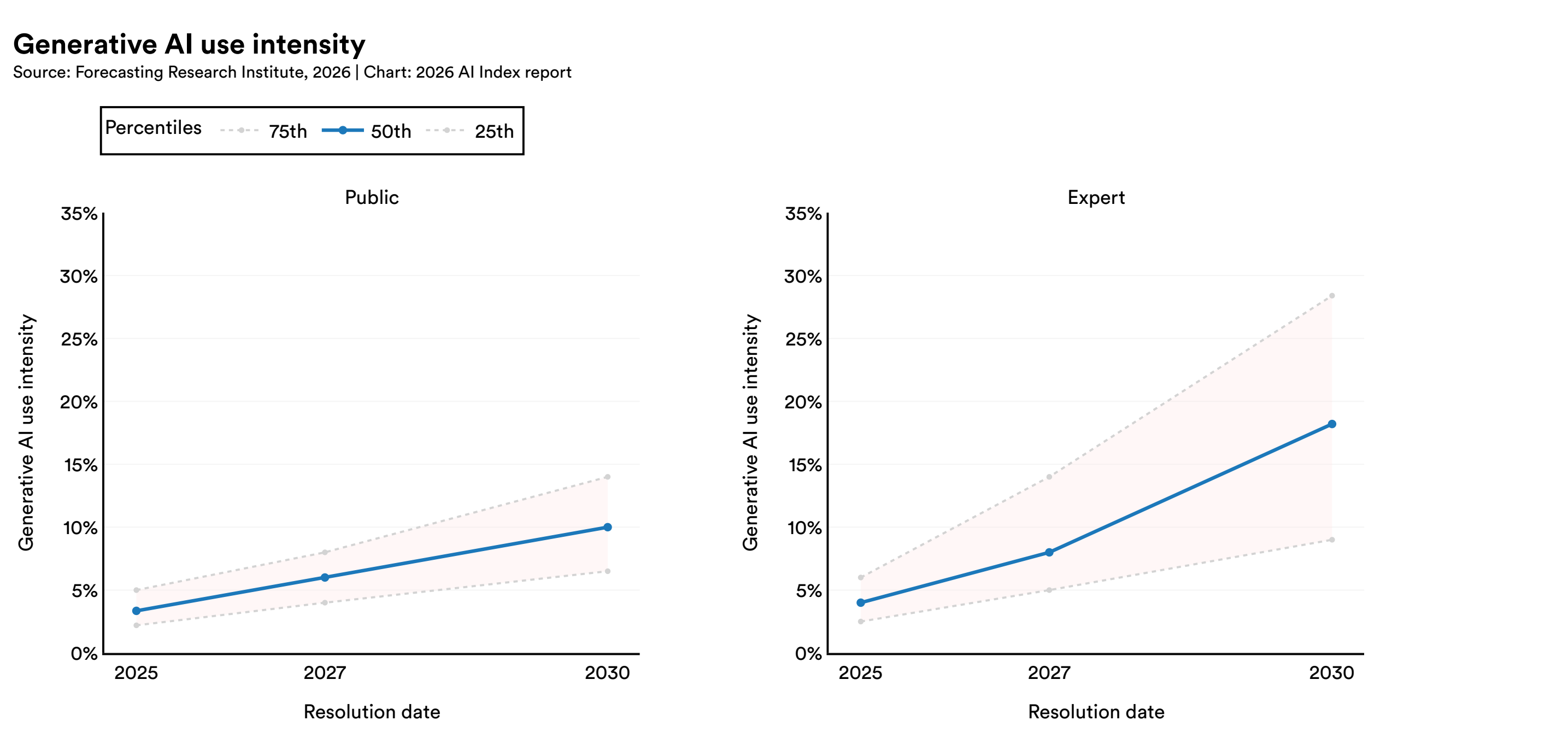

Figure 9.2.5

When asked about specific occupations, the U.S. public and AI experts identified certain jobs to be at higher risk for automation than others (Figure 9.2.6). There is strong consensus between the public and experts regarding automation risks for jobs such as cashiers, journalists, and software engineers. AI experts see a greater risk for truck drivers and lawyers, while the U.S. public believes AI will lead to fewer jobs for teachers and medical doctors. Mostly, both groups identify the same areas of vulnerability, but the public is generally more likely to anticipate job loss across categories.

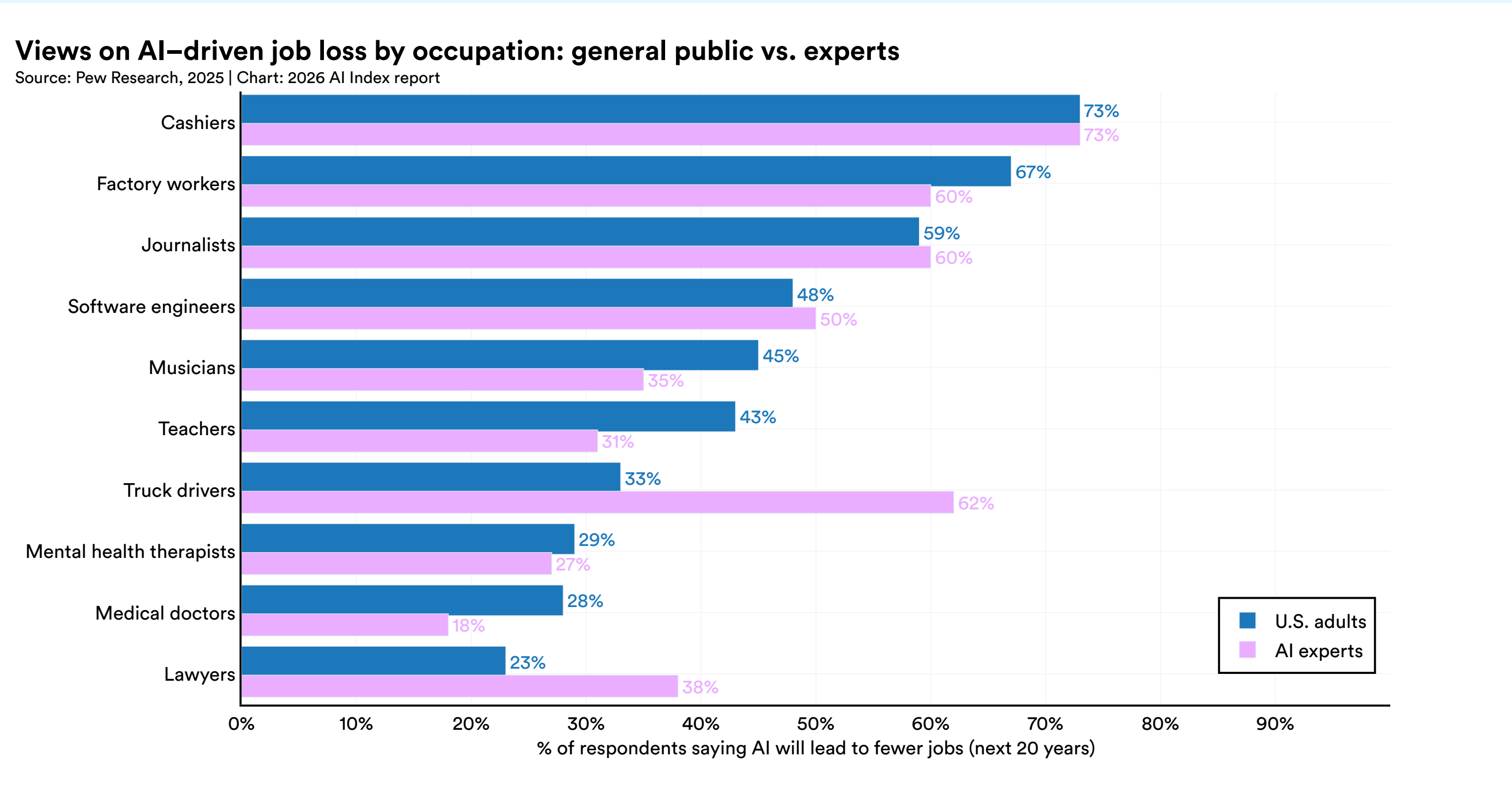

Figure 9.2.6

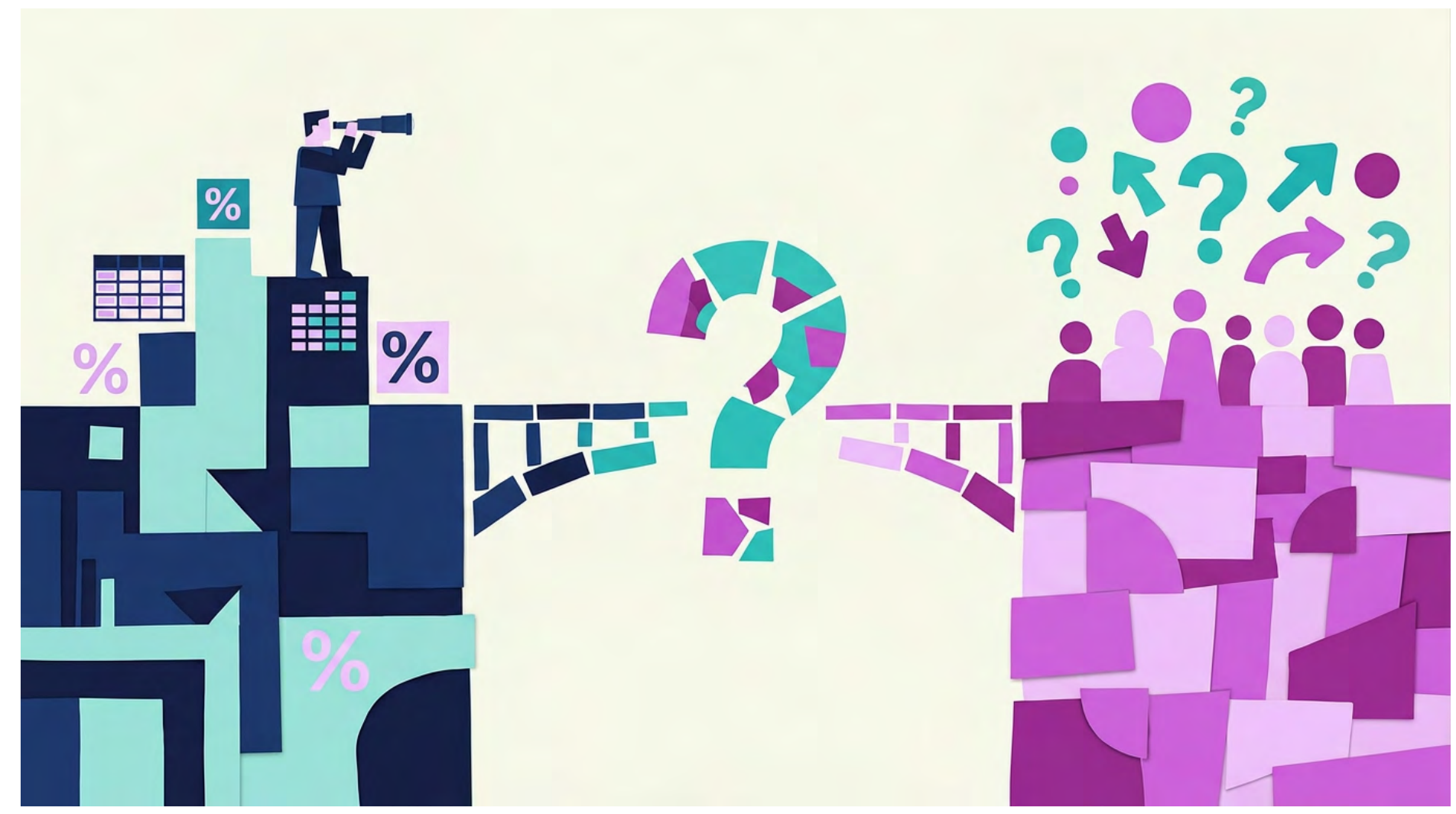

The gap in expert vs. public sentiment coincides with increasing awareness and adoption of AI In the United States. In 2025, 47% of U.S. adults said they had heard "a lot" about AI, up from 26% in 2022. Growth in awareness is steepest among younger adults, ages 18–29, (+29 percentage points since 2022), though it is also rising among those ages 65 and older (+13pp) (Figure 9.2.7).

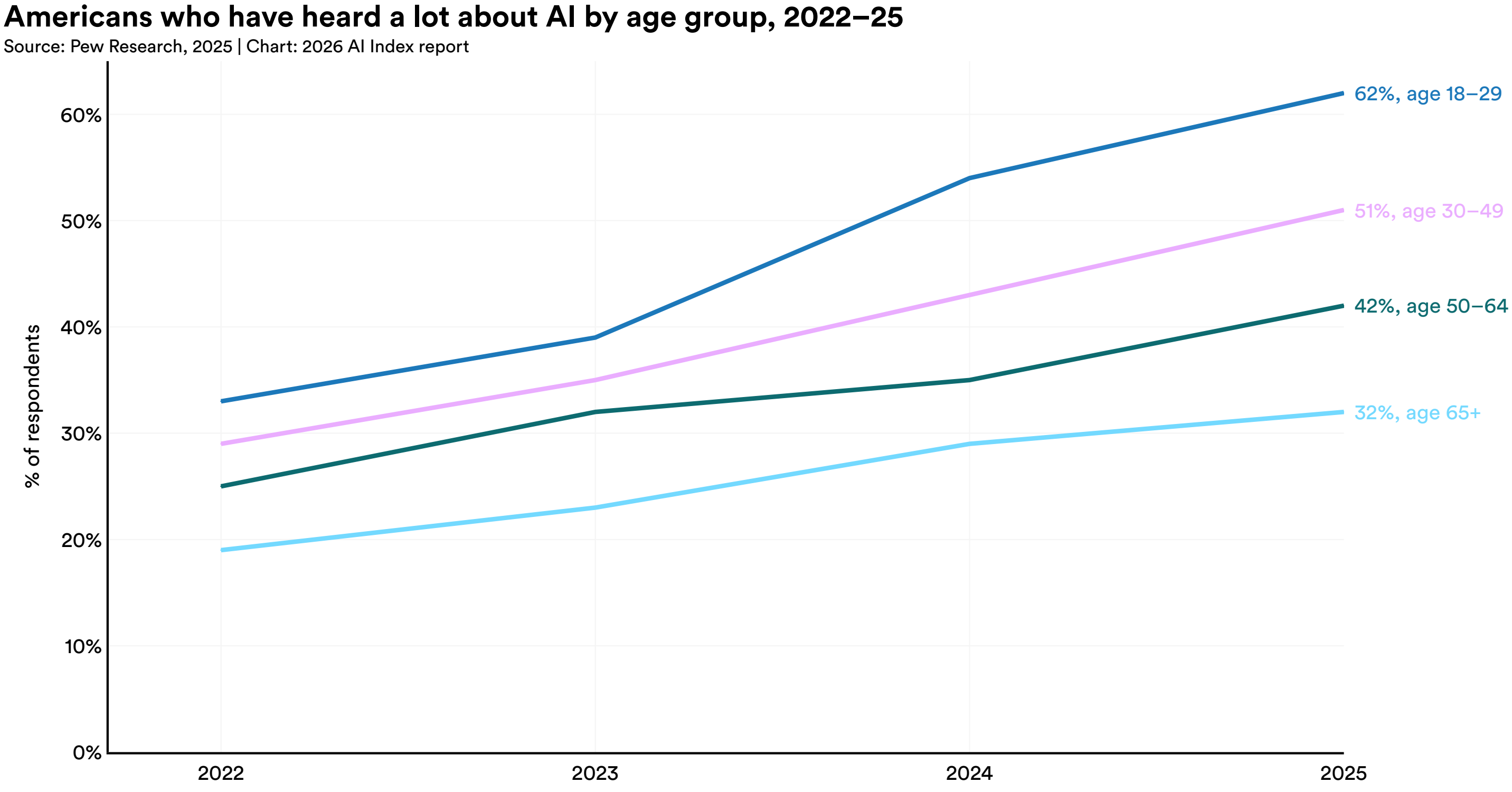
Americans who have heard a lot about AI by age group, 2022–25
Source: Pew Research, 2025 | Chart: 2026 AI Index report

Figure 9.2.7

Adoption and frequency of use are also increasing. More than 60% of U.S. adults reported interacting with AI at least several times a week, and 31% said they interact with AI almost constantly or several times a day, though frequency varies according to age and race and ethnicity (Figure 9.2.8). Daily AI interaction is higher among younger adults, college-educated groups, Asian Americans, and men. Political affiliation differences are modest, with Democrats slightly more likely than Republicans to interact daily with AI. As a note, the results are based on when respondents believe they are interacting with AI and therefore may undercount exposure through other embedded systems like navigation, recommendations, or rankings.

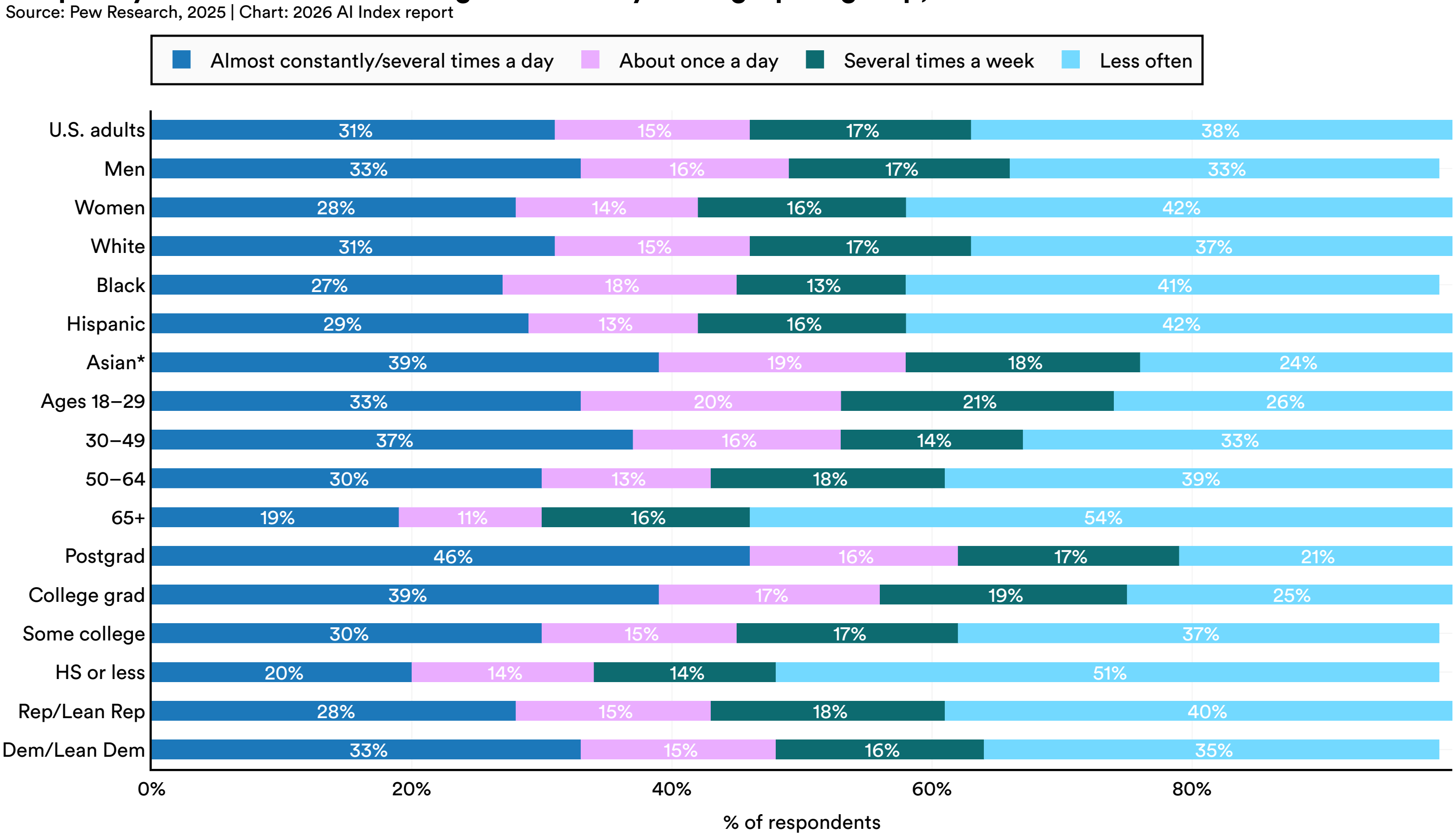
Frequency of AI interaction among US adults by demographic group, 2025
Source: Pew Research, 2025 | Chart: 2026 AI Index report

Figure 9.2.8[6]

6 "Asian" includes English-speaking respondents only. Respondents who did not provide an answer are not shown. "White," "Black," and "Asian" adults are non-Hispanic and report only one race; "Hispanic" adults may be of any race.

HIGHLIGHT:

## Views on AI Companions

AI companionship, defined as relationships with AI systems designed for ongoing emotional and social support, represents one of the more contentious emerging uses of AI technology (Chou et al., 2024; Pan et al., 2025). Experts predict that 10% of U.S. adults will use AI for companionship at least once a day by 2027, with that number rising to 15% by 2030 and 30% in 2040 (Figure 9.2.9). The top quartile among experts' predictions forecast that more than 40% of the public will engage in daily AI companionship, while the top 10% predict over 60%. Expectations from the general public are significantly lower, at 20% by 2040. Both experts and the public find it less likely that mental health therapists will be replaced by AI, suggesting that there is an understanding on the limitations of AI companions. They cannot fully replace human expertise in complex or therapeutic contexts.

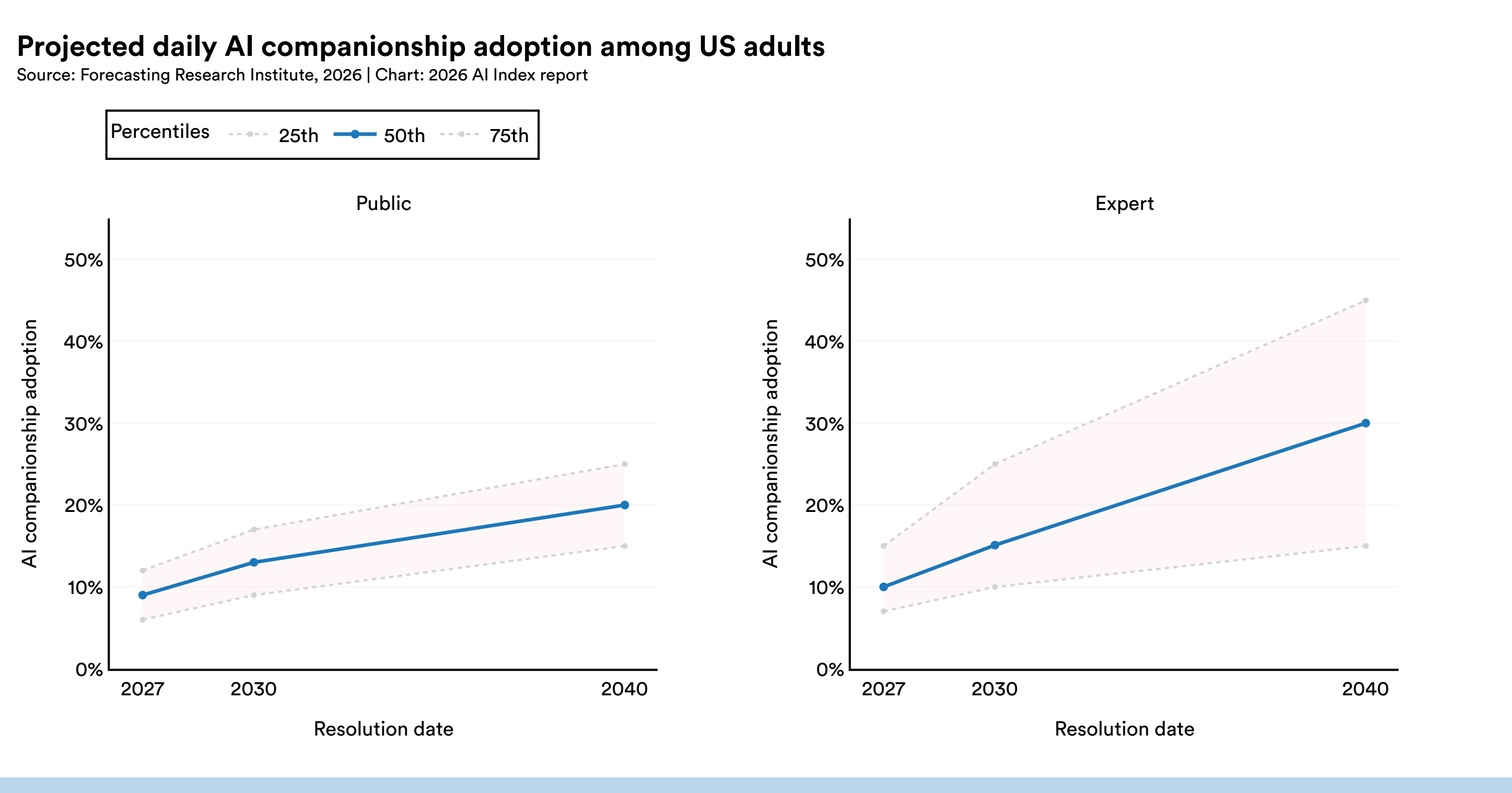

Figure 9.2.9

A 2026 Ipsos-Google survey found that 52% of worldwide respondents reported some level of excitement about using AI for companionship (Figure 9.2.10). In countries such as Nigeria, India, and the United Arab Emirates, over 20% of respondents said they were "extremely excited." The United States and Canada had the largest shares of respondents who were not excited at all, at 36% and 34%. Japan recorded very few "extremely excited" respondents, and had the highest share of "don't know" responses at 18%, nearly double the global average.

**HIGHLIGHT:**

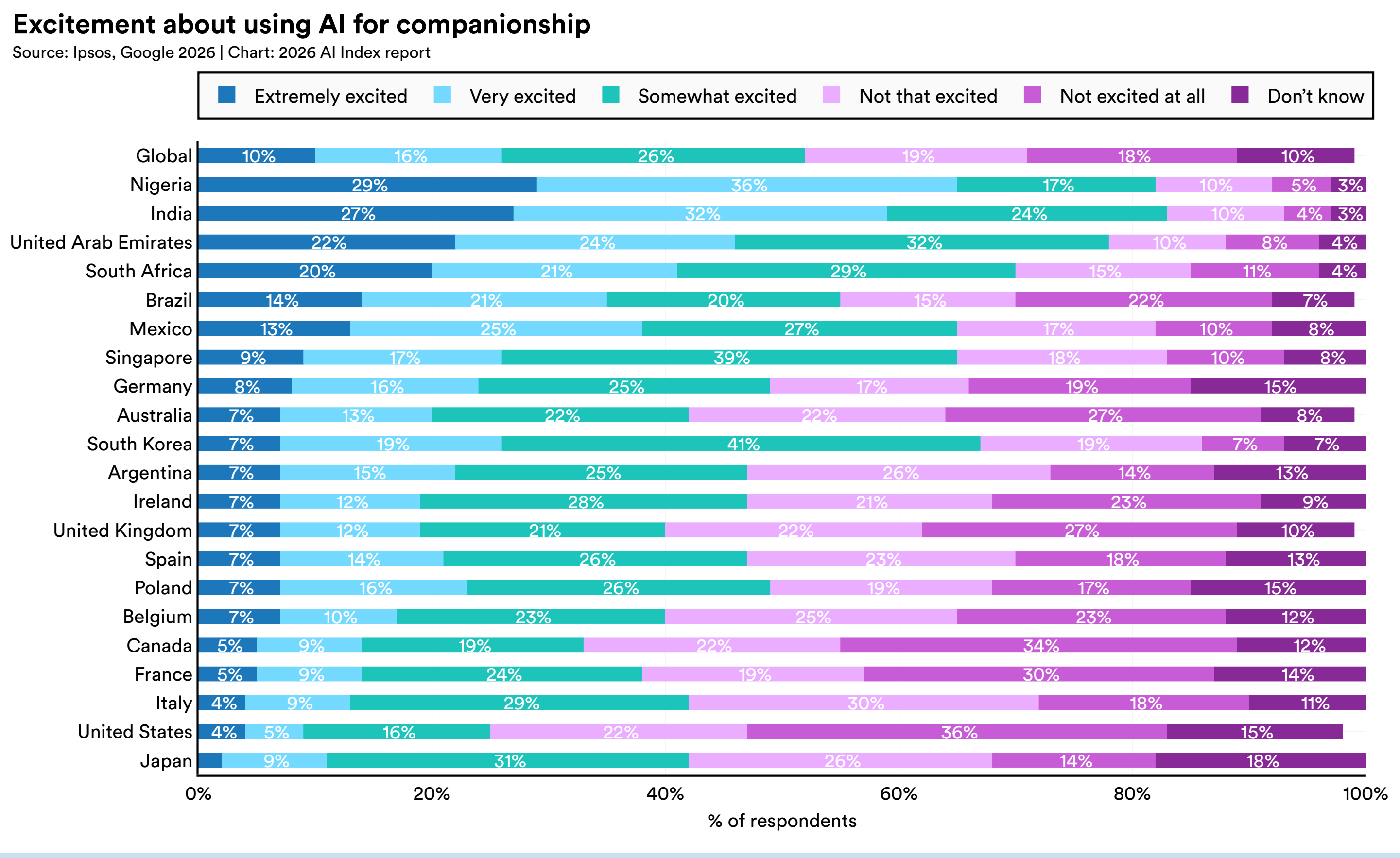

Figure 9.2.10[7]

AI companions differ from traditional task-oriented AI by prioritizing relationship building over functionality (Zhang and Lu, 2023; Zhang et al., 2025). Modern systems incorporate memory of past interactions, can recognize emotion, and adapt their responses to individual users' needs (Yang et al., 2025). Platforms like Replika, Character.ai, and XiaoICE have attracted user bases in the millions. Many users have reported forming emotional attachments to their AI companions, viewing them as friends, mentors, or romantic partners (Zhang et al., 2024; Kouros et al., 2024).

The technology has both benefits and risks. Research shows that AI companions can reduce loneliness to a similar degree as interacting with another human (Freitas et al., 2024), with users citing always-available support (11.9%) and a safe space for emotional expression (9.9%) as primary advantages. Mental health improvements were reported by 6.2% of users, and some credited their AI companions for helping them through crises (Pataranutaporn et al., 2025).

However, concerning patterns have emerged. Users frequently perceive chatbots as entities with needs, which poses a problem given the established correlation between emotional dependence and psychological distress (Bengio et al., 2025). Critical questions remain about whether these relationships reduce loneliness sustainably or undermine existing relationships and increase social isolation (Quinn et al., 2024).

7 Totals may not equal 100% due to rounding of individual values.

# 9.3 Perceptions on AI Trust, Transparency, and Regulation

## Global Trust in Institutions

As AI becomes more embedded in daily life, the mechanisms around trust, transparency, and regulation also become more visible. In Ipsos' 2025 AI Monitor Survey, 79% of respondents said companies using AI should be required to disclose that usage (Figure 9.1.1). That view was shared across all 30 countries surveyed, even though overall trust in institutions was lower. Over half of respondents, or 54%, said they trust their government to regulate AI responsibly (Figure 9.3.1). The United States reported the lowest trust on this measure (31%). In parallel with the higher levels of optimism and excitement mentioned earlier, Southeast Asian countries reported the highest levels of trust in their governments, including Singapore (81%), Indonesia (76%), Malaysia (73%), and Thailand (70%).

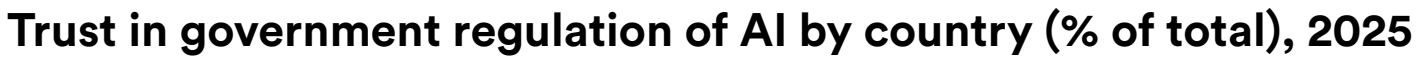
Trust in government regulation of AI by country (% of total), 2025
Source: Ipsos, 2025 | Chart: 2026 AI Index report
Singapore 81%
Indonesia 76%
Malaysia 73%
Thailand 70%
Chile 67%
Mexico 67%
Colombia 66%
India 65%
Argentina 61%
Poland 61%
Peru 61%
Switzerland 55%
Spain 55%
South Africa 55%
Global 54%
Italy 50%
Ireland 49%
Germany 49%
Belgium 49%
Turkey 48%
Netherlands 48%
Brazil 48%
South Korea 46%
Australia 46%
Sweden 46%
France 42%
Canada 40%
Great Britain 39%
Hungary 33%
Japan 32%
United States 31%
0% 10% 20% 30% 40% 50% 60% 70% 80% 90% 100%
% of respondents

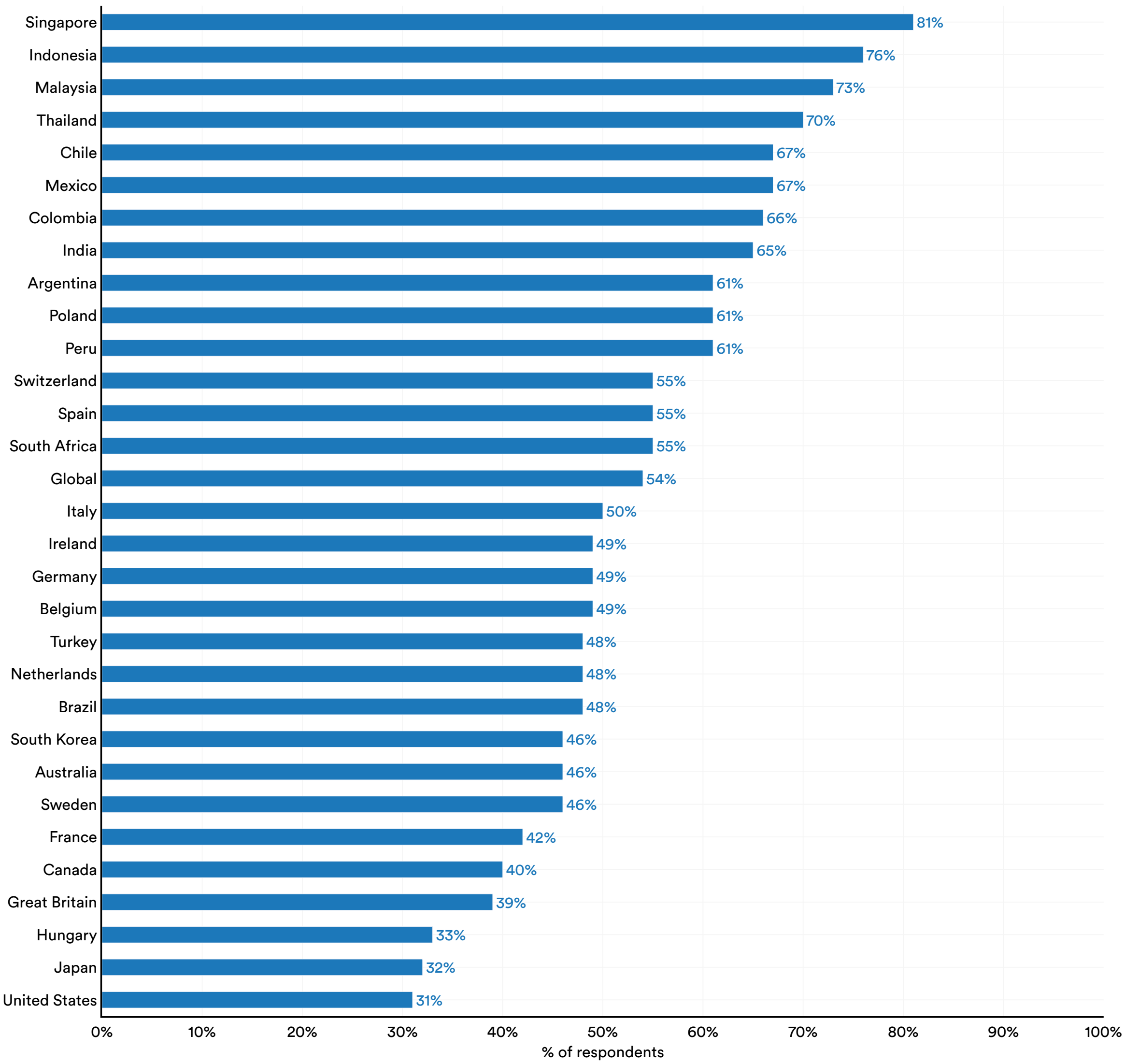

Figure 9.3.1

A separate Pew global survey asked a related question to compare respondents' trust in different governing bodies across the globe. Pew's Spring 2025 Global Attitudes Survey found that respondents tend to trust their own country most to regulate AI effectively, but trust in outside governments was mixed. Across the 25 countries surveyed, a median of 53% said they trust the EU to regulate AI effectively, compared to 37% for the United States and 27% for China (Figure 9.3.2). Trust in the Chinese government consistently received the lowest ratings across countries, while trust in the EU varied depending on whether respondents lived within or outside of the EU (Figure 9.3.2).

However, even within the EU, trust levels were not uniform. Respondents in Germany and the Netherlands were among the most trusting of the EU's ability to regulate AI effectively, while Greece and Italy were among the least trusting. In the United States, views were evenly divided between trust (44%) and distrust (47%) in the government's ability to regulate AI effectively, and 43% said they trust the EU on AI regulation. These trust dynamics are shifting against an expanding legislative landscape, outlined in Chapter 8, as the number of countries adopting national AI strategies continues to grow.

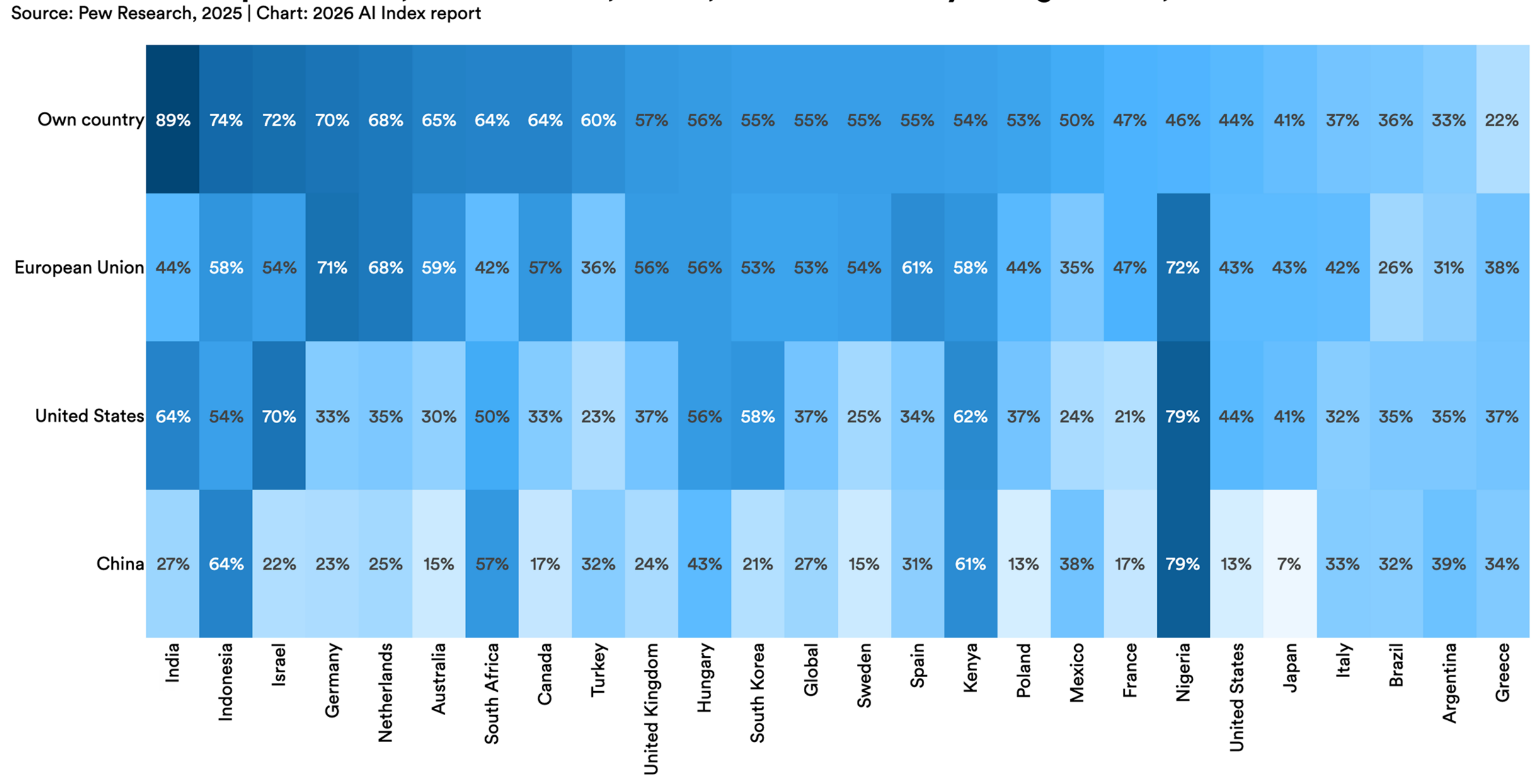
Trust in the European Union, United States, China, and own country to regulate AI, 2025
Source: Pew Research, 2025 | Chart: 2026 AI Index report

**Figure 9.3.2**

A separate Ipsos/Google survey shows a related divide in relation to public priorities. Globally, 58% of respondents said it was more important to foster advances in science, medicine, and other fields through AI innovation, compared to 41% who prioritized protecting industries that may be affected by AI through regulation (Figure 9.3.3). Most countries in the survey lean toward innovation, though South Africa, India, and Ireland were among the few where respondents were more likely to prioritize regulation. Across these different measures, public views on AI governance appear mixed and varied in trust, priorities, and regulatory expectations.

**Global priorities: AI innovation vs. AI regulation, 2025**
Source: Ipsos, Google 2026 | Chart: 2026 AI Index report

|  | Advancing science, medicine, and other fields | Protecting industries through regulation |
|---|---|---|
| Global | 58% | 41% |
| Nigeria | 71% | 29% |
| South Korea | 70% | 30% |
| Argentina | 68% | 32% |
| Poland | 67% | 33% |
| Japan | 67% | 33% |
| Mexico | 66% | 34% |
| France | 65% | 35% |
| Germany | 63% | 37% |
| Belgium | 62% | 38% |
| Brazil | 62% | 38% |
| Spain | 54% | 46% |
| United Kingdom | 54% | 46% |
| United Arab Emirates | 53% | 47% |
| United States | 53% | 44% |
| Singapore | 53% | 47% |
| Canada | 52% | 48% |
| Italy | 52% | 48% |
| Australia | 52% | 48% |
| Ireland | 48% | 52% |
| South Africa | 47% | 53% |
| India | 46% | 54% |

% of respondents 0%, 20%, 40%, 60%, 80%

Figure 9.3.3[8]

**Net concern for not enough vs. too much AI regulation by US state, 2025**
Source: CHIP50, 2025 | Chart: 2026 AI Index report

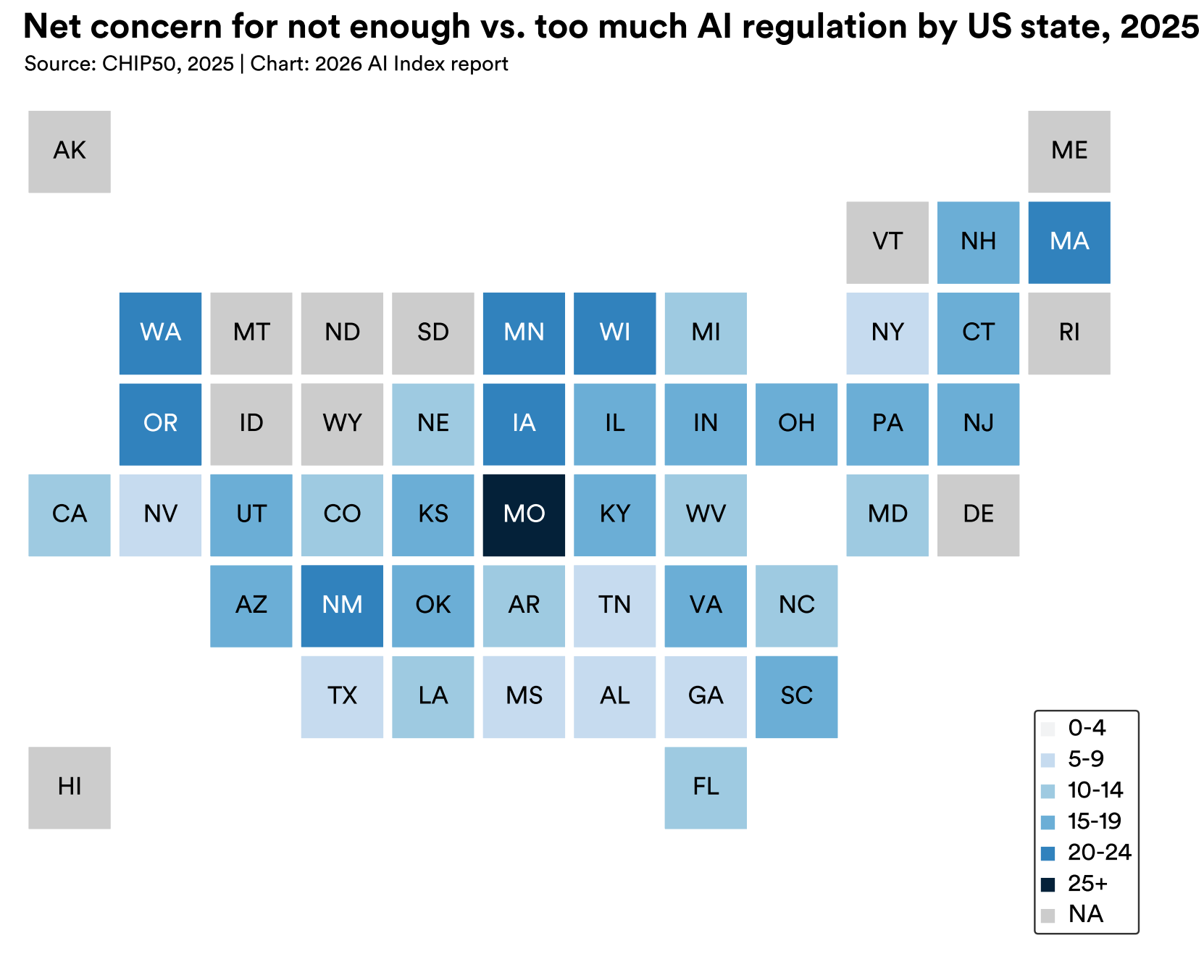

Figure 9.3.4

## US Attitudes Toward AI Regulation

In the United States, attitudes toward AI regulation vary meaningfully by geography. In 2025, the Civic Health and Institutions Project fielded a survey across 50 states, and asked respondents whether federal regulation of AI would go too far, not far enough, or "not sure" (Figures 9.3.4 and 9.3.5). Across every state, concern about too little regulation outnumbers concern about too much regulation (41% vs. 27%), but the level of uncertainty is substantial, with more than one-third of respondents selecting "not sure."

New York and Tennessee reported the highest levels of concern that regulation will go too far (31%), while Missouri and Washington had the highest shares who said the government will not go far

8 In Ipsos' reporting of findings, percentage points are rounded to the nearest whole number. As a result, figures may not add up to exactly 100%.

enough (48%). Across nearly every state, more respondents said regulation does not go far enough than said it goes too far. Roughly one in three respondents in most states said they were not sure, making uncertainty the second-largest category.

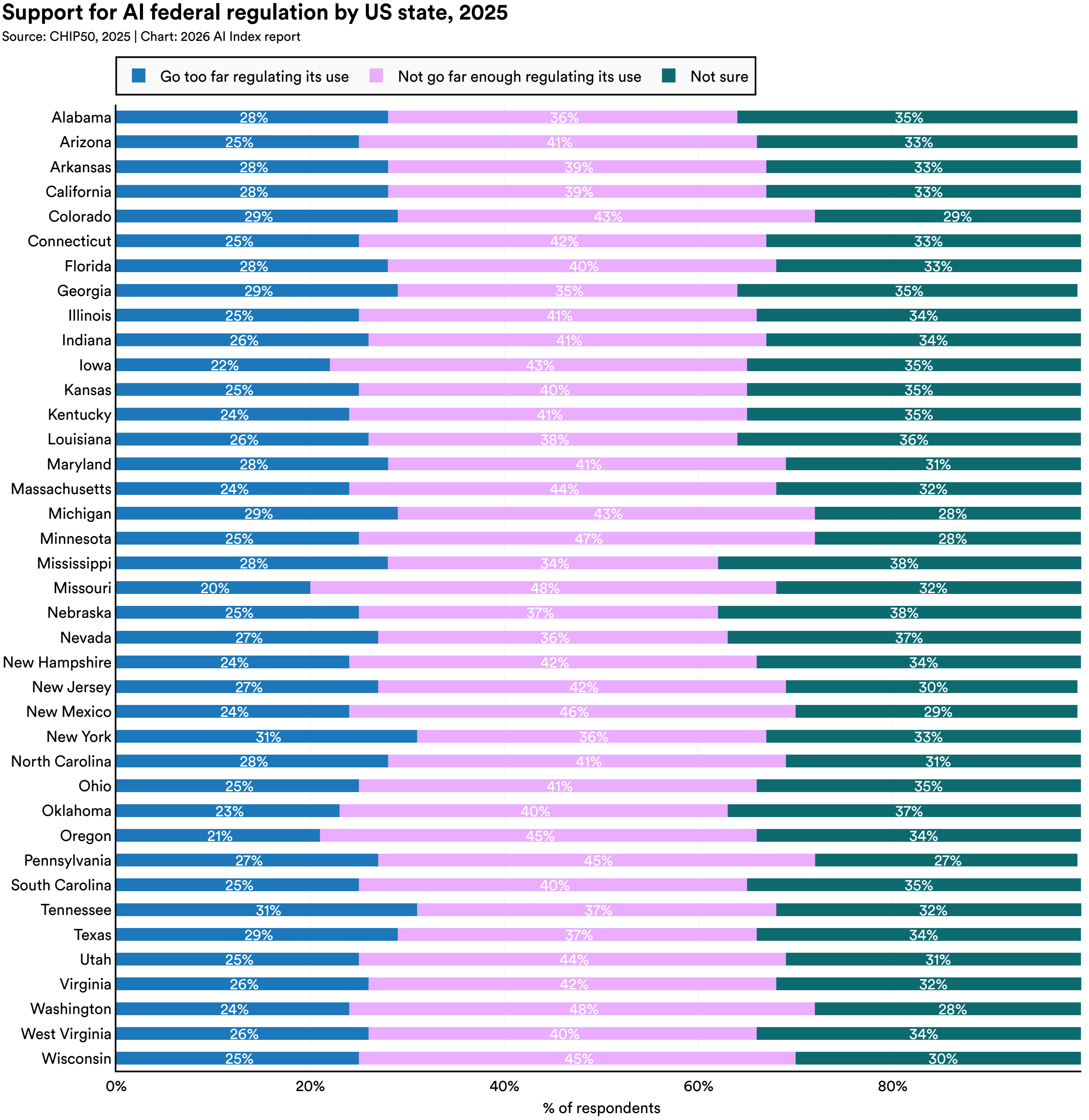

Figure 9.3.5

Across U.S. demographic groups, the strongest concern about insufficient AI regulation was reported among older adults, especially those 65 and older (51%) (Figure 9.3.6). Education was associated with stronger support for more regulation, with 46% of college graduates saying the government will not go far enough, compared with 34% among respondents with a high school degree or less. Political affiliation was not a significant differentiator, although Democrats were more likely than Republicans to say regulation will not go far enough (45% vs. 40%), while concern about going too far is similar across parties (>25%).

**Attitude toward AI federal regulation in the US by demographic group, 2025**

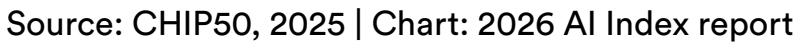

Source: CHIP50, 2025 | Chart: 2026 AI Index report

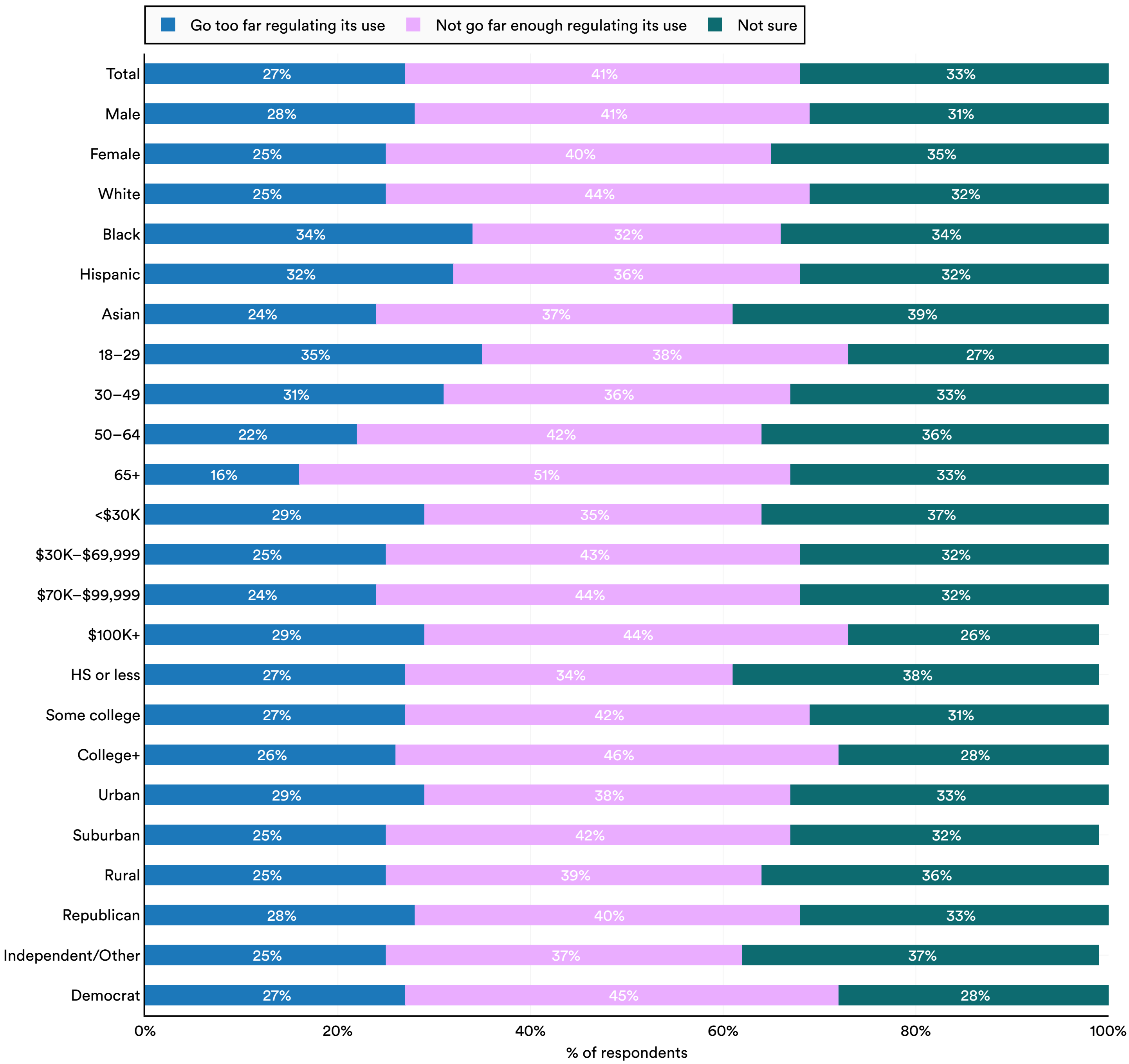

| | Go too far regulating its use | Not go far enough regulating its use | Not sure |
|---|---|---|---|
| Total | 27% | 41% | 33% |
| Male | 28% | 41% | 31% |
| Female | 25% | 40% | 35% |
| White | 25% | 44% | 32% |
| Black | 34% | 32% | 34% |
| Hispanic | 32% | 36% | 32% |
| Asian | 24% | 37% | 39% |
| 18–29 | 35% | 38% | 27% |
| 30–49 | 31% | 36% | 33% |
| 50–64 | 22% | 42% | 36% |
| 65+ | 16% | 51% | 33% |
| <$30K | 29% | 35% | 37% |
| $30K–$69,999 | 25% | 43% | 32% |
| $70K–$99,999 | 24% | 44% | 32% |
| $100K+ | 29% | 44% | 26% |
| HS or less | 27% | 34% | 38% |
| Some college | 27% | 42% | 31% |
| College+ | 26% | 46% | 28% |
| Urban | 29% | 38% | 33% |
| Suburban | 25% | 42% | 32% |
| Rural | 25% | 39% | 36% |
| Republican | 28% | 40% | 33% |
| Independent/Other | 25% | 37% | 37% |
| Democrat | 27% | 45% | 28% |

% of respondents (0%–100%)

**Figure 9.3.6**

# Appendix

## Chapter 1: Research and Development

### Environmental Impact Analysis

The AI Index estimated the carbon emissions of training language and vision models using a calculator proposed by Lacoste et al. (2019). The analysis focused on the training stage emissions—excluding embodied hardware production, idle infrastructure, and deployment emissions. The study examined four model categories: industry language models, academic language models, industry vision models, and academic vision models.

The calculator's accuracy was verified against published emission values. Calculator inputs included hardware type, GPU hours, provider, and compute region. For newer hardware like the H100 GPU (released in 2022), the A100 SXM4 80GB was used as a substitute in calculations. GPU hours were calculated by multiplying hardware quantity with training duration; these values were taken from Epoch AI's Data on AI models or from the technical paper for the model. Provider selection was based on known partnerships (e.g., Google models using GCP, OpenAI using Azure), while compute regions were determined by team locations.

Special consideration was given to models trained on custom hardware, such as BLOOM's use of the Jean Zay supercomputer in France. In these cases, private infrastructure calculations incorporated carbon efficiency (kg/kWh) and offset percentages.

The study evaluated 52 models in total: 36 industry language models (2018–25), eight industry vision models (2019–23), four academic language models (2020–23), and four academic vision models (2011–22), selecting particularly influential models in their respective domains.

### GitHub

#### Identifying AI Projects

In partnership with researchers from Harvard Business School, Microsoft Research, and Microsoft's AI for Good Lab, GitHub identifies public AI repositories following the methodologies of Gonzalez, Zimmerman, and Nagappan (2020) and Dohmke, Iansiti, and Richards (2023), using topic labels related to AI/ML and generative AI, respectively, along with other relevant keywords identified through snowball sampling, such as "machine learning," "deep learning," and "artificial intelligence." GitHub further augments the dataset with repositories that have a dependency on the PyTorch, TensorFlow, OpenAI, Transformers, XGBoost, scikit-learn, and SciPy libraries for Python.

#### Mapping AI Projects to Geographic Areas

Public AI projects are mapped to geographic areas using IP address geolocation to determine the mode location of a project's owners each year. Each project owner is assigned a location based on their IP address when interacting with GitHub. If a project owner changes locations within a year, the location for the project would be determined by the mode location of its owners sampled daily in the year. Additionally, the last known location of the project owner is carried forward on a daily basis even if the project owner performed no activities that day. For example, if a project owner performed activities within the United States and then became inactive for six days, that project owner would be considered to be in the United States for the seven-day span.

### Hugging Face

Hugging Face (HF) data is collected from two distinct sources:

- **Downloads data:** shared by the authors of Longpre et al. (2025)
- **Number of existing models and datasets:** publicly accessible Hugging Face (HF) repository.

#### Downloads data

Longpre et al. (2025) is used because it provides the most consistent and complete information on actual downloads.

- Download data from HF can vary across releases depending on how parameters are handled.[1] Longpre et al. collaborated directly with HF personnel, who confirmed that this dataset is the least noisy version available.[2]

1 GET/HEAD requests, local cache hits, IP addresses for cloud-hosted virtual machines when the setup does not require a fixed IP, the Git/Xet back-end for models obtained inside vs. outside the Transformers and Hub API, proactive spam detection, retroactive spam detection, etc.

2 The unreleased usage data takes a different aggregation approach than the public API. It is filtered to remove repetitive, duplicate requests from

- Much of the download and model metadata is missing in other sources; the authors performed manual cleaning and imputed missing values, providing a more complete version than publicly accessible alternatives.
- Public HF data is provided as a cross-section of "all times downloads," whereas access to a panel version was granted through direct collaboration with the authors.

This dataset is limited by the following:

- The time span covers March 2022 to August 2025. The start date cannot be pulled up because that is when HF began tagging model creation dates.
- Coverage is limited to the top 200 most-downloaded models per week among models with at least 200 downloads. However, this subsample reflects 49.6% of all downloads.

### Number of existing models and datasets

The public source is constantly updated but at this time contains a large number of missing values for modality (~60%) and no geographic information, and it is only available in cross-sectional format. As a result, it does not allow for time-series analysis on usage, or more granular representation of models/datasets population by geographic area and modality.

## AI Publication Analysis

For this analysis, the AI Index used OpenAlex, an open scholarly database with over 260 million research publications, as its primary data source. OpenAlex classifies papers using its own knowledge organization system, known as OpenAlex Topics—a taxonomy of around 4,500 topics combining Scopus codes and CWTS classification. The system uses a deep learning model that considers titles, abstracts, journal names, and citation networks for classification. To identify AI-related topics more precisely, the AI Index analyzed computer science publications identified by OpenAlex and refined the classifications using the Computer Science Ontology and the CSO Classifier.

The Computer Science Ontology (CSO) is a large-scale, automatically generated ontology of research areas derived from 16 million publications using the Klink-2 algorithm. It features a hierarchical structure with thousands of subtopics, allowing for precise mapping of specific terms to broader research fields. Compared to general-purpose scholarly databases such as OpenAlex, Scopus, and Web of Science, CSO offers a more detailed and fine-grained representation of the research landscape. As a result, it has been widely used for scholarly data exploration, analysis, modeling, and expert identification and recommendation. Version 3.4.1—used in this analysis—includes approximately 15,000 topics and 166,000 relationships within computer science. Released on Jan. 17, 2025, this version introduces over 150 new research topics in artificial intelligence, bringing the total to 2,369 AI-related topics and 12,620 hierarchical relationships within the AI domain alone.

To analyze research trends, the AI Index used the CSO Classifier—an unsupervised method that automatically categorizes research papers based on CSO topics. The classifier follows a three-stage pipeline that processes paper titles and abstracts: A syntactic module detects direct mentions of CSO topics; a semantic module uses word embeddings to identify related concepts; and a postprocessing module merges results, filters out irrelevant topics, and adds broader categories for a more refined classification. For this analysis, the AI Index extended the CSO Classifier to focus specifically on artificial intelligence and its subtopics. Since its initial release, the classifier has gained significant and growing interest due to its versatility. For example, Springer Nature uses it to routinely classify proceedings books, improving metadata quality. Beyond academic publishing, it has been successfully applied to categorize research software, YouTube videos, press releases, job ads, and IT museum collections.

Accurately categorizing research papers as either conference proceedings or journal articles is essential for this analysis. OpenAlex's metadata fields—type, crossref_type, and source_type—can sometimes conflict. To resolve these inconsistencies, the AI Index mapped OpenAlex records to DBLP, a leading bibliographic database for computer science publications. Known for its high metadata quality, DBLP currently indexes 3.6 million conference papers and 3 million journal articles and continuously adds new publications through a rigorous, semiautomated curation process. The initial matching between OpenAlex and DBLP was performed using DOIs. For remaining unmatched papers, the AI Index used a combination of title and publication year. To streamline this process, the AI Index built a title index to optimize search and ensure efficient mapping across the datasets.

AI publications are aggregated based on several parameters to provide a comprehensive analysis. Publications are grouped by year, considering the publication date of the most recent versions. Additionally, the AI Index groups publications by geographic areas or World Bank regions based on the affiliations of authors. This means a single paper can contribute to multiple counts if co-authored by researchers from different countries, with each country receiving a count. When authors' affiliations are missing, the publications are mapped as "Unknown." Furthermore, sectors are associated with publications through authors' affiliations when available, which may lead to one publication being counted for multiple sectors. Citation counts are included when available; those without citation data are classified as "Unknown."

### Top 100 Publications Analysis

The AI Index conducted a comprehensive analysis of influential AI publications by collecting and analyzing citation data from multiple sources, including OpenAlex, Google Scholar, and Semantic Scholar. Initially gathering the top 150 most-cited papers per publication year from OpenAlex, the list was refined through careful review to 100 publications.

---

the same user on the same day, as well as models with fewer than 200 total downloads, suggesting less broad usage. It is available only upon request to the ML & Society Team of Hugging Face.

The methodology attributes publications to all countries and regions represented by authors' affiliations, meaning a single paper can contribute to multiple counts. For instance, a paper co-authored by researchers from the United States and China counts once for each country. This approach may result in overlapping totals in aggregate statistics. Publication years are based on the most recent versions, whether in journals, conferences, or repositories like arXiv. To maintain accuracy, organizational affiliations were verified and standardized, with countries assigned according to headquarters' locations.

The full list of the top 100 AI publications is available here.

## AI Conference Attendance

The AI Index contacted the organizers of various AI conferences in 2025 to request information on total attendance. For conferences that posted their attendance totals online, the AI Index used those reported totals and did not reach out to the conference organizers.

## AI Patent Analysis

The AI Index identifies AI-related patents using a hybrid classification approach, combining keyword-based text analysis with classification-code-based identification. Patent-level bibliographic data is sourced from PATSTAT Global, a comprehensive database issued by the European Patent Office (EPO). The analysis focuses on granted patents from 2010 onward, aggregated at the DOCDB family level to avoid duplicate counting of the same invention.[3] Patents are attributed to countries based on the publication authority of the earliest recorded grant publication.

Patent abstracts and titles originally published in languages other than English were translated using the deep-translator tool, Google Translate engine, and the Meta NLLB-200 machine translation model. Post-translation, patent texts were processed using natural language processing (NLP) techniques. These included the removal of stop words and special characters, part-of-speech (POS) tagging to retain key grammatical categories, lowercase conversion, lemmatization, and replacement of numerical measures with a <NUM> tag.

AI-related patents are identified by searching for relevant terms in patent titles and abstracts using regular expressions (regex). An AI-specific keyword dictionary was developed through a structured multistep process, incorporating keywords generated by AI models, expanded using established AI lexicons such as those from Yamashita et al. (2021), and refined through Word2Vec-based synonym identification. Further validation was conducted using BERTopic topic modeling and DeBERTA-based zero-shot classification, with manual checks applied to reduce false positives.

In addition to keyword-based classification, AI-related patents were identified using International Patent Classification (IPC) and Cooperative Patent Classification (CPC) codes. A curated list of AI-relevant codes was compiled through a combination of AI model analysis, regex-based searches, and prior research, including classifications from Pairolero et al. (2023) and WIPO (2024). The final dataset was constructed by merging results from both approaches, balancing coverage and accuracy.

### Speed of Knowledge Diffusion Methodology

**Definition:** The Kaplan-Meier estimator is a non-parametric statistical method used to estimate the survival function from lifetime data. In this context:

- Lifetime refers to the citation lag, measured in months, from a patent's publication to its first citation.
- The event is a patent receiving its first citation.
- Censoring occurs when a patent has not been cited by the end of the observation period.

**Computation:** The survival function represents the probability that a patent has not yet received a citation for a duration exceeding a specified time t. The Kaplan-Meier estimator is mathematically defined as follows:

$$\widehat{S}(t) = \prod_{i:\ t_i \leq t} \left(1 - \frac{d_i}{n_i}\right)$$

- t: The time since publication (in months).
- t_i: The distinct time points at which at least one patent receives its first citation.
- d_i: The number of events (patents cited for the first time) that occur at time t_i.
- n_i: The number of patents at risk of being cited just prior to time t_i, which includes all patents that have remained uncited up to time t_(i-1) and have not been censored before t_i.

The analysis is conducted at the patent family level (single invention), using the earliest publication date of each family as the time

3 Despite this aggregation procedure, duplicates occasionally appear in marginal cases where applications within the same DOCDB family share the same earliest filing date. The AI Index removes duplicate values with respect to the aggregation variables (e.g., counting by year) when presenting analytics.

reference for citation events and citation lags. For applications of this methodology in the literature, refer to examples such as Fisch et al. (2017) and Xie et al. (2019). The figure aggregates citations from all patent authorities, which together constitute less than 6% of the total citations, into a category labeled "Rest of the World."

### Technological Proximity Methodology

**Computation:** Based on the work of Bar et al. (2012), this measure is calculated by comparing the patent portfolios of two countries (patent authorities) using the technological classes to which their patents belong. The calculation uses a vector for each entity *i* patent portfolio, *P_i*, where each component *P_ik* is the share of the entity's total patents in a specific technological class *k*[4].

- The final measure captures the sum of the minimum shares across all shared technological classes, quantifying the share of overlapping inventions between the two portfolios.

$$\text{Proximity}(\mathbf{P}_i, \mathbf{P}_j) = \sum_{k=1}^{n} \min\{p_{ik}, p_{jk}\}$$

- **Interpretation:** The resulting value indicates the similarity in the technological focus between the two countries.
    - ◊ Range: The value is normalized to fall between 0 and 1.

**Benchmark:** The graph shows proximity measures between countries and the two major patent authorities (the United States and China) based on the number of granted AI patents.

## Zeki

### Zeki AI Talent

Zeki has identified 658,000 top AI talent (outside of China) with a proven track record in producing new discoveries in AI by contributing to research, data depositories, or new models. They are of particular value in the market because of their advanced skills and ability to push the boundaries of science and engineering. They create new products and intellectual property (IP) for their employers rather than just applying existing technology in the market.

The following countries are covered: Australia, Brazil, Canada, Denmark, Finland, France, Germany, India, Israel, Italy, Japan, Netherlands, Saudi Arabia, Singapore, South Korea, Spain, Sweden, Switzerland, UAE, United Kingdom, United States.

### Data

Zeki collects publicly available, strictly business-related data that has been published or released by companies or individuals online. Zeki does not collect private data about individuals (i.e., information that is not publicly available and the individual has chosen to keep private). Zeki strictly refrains from any data collection that involves data aggregation from within secured login areas. We also purchase data from vetted vendors that are stringently assessed by external legal experts to ensure their compliance with data regulations. Career data is sourced in a compliant manner from the open web. For the talent specialization, Zeki has curated unique areas of AI innovation across all aspects of AI software, hardware, and compute. The primary and secondary areas are identified through a comprehensive analysis of all relevant research papers. Gender is inferred using a probability model based on the likelihood of a first name being male or female. It is enhanced with likely country of residence to improve accuracy. While this method is generally reliable, some names are commonly used across multiple genders. In such cases, the model assigns a probability score. If the probability falls between 45% and 55%, the name is classified as ambiguous or nonspecific, meaning a confident gender assignment could not be made.

### Date Range

The dataset covers the period 2010 to 2025. Records that began prior to 2010 are excluded from this dataset.

### Affiliation Range

The dataset includes individuals working in companies and academic institutions.

### Last Known Residency

To ensure accuracy and reduce volatility caused by delayed profile updates, Zeki uses the Last Known Residency logic for longitudinal datasets. When a professional's career timeline lacks an explicit "end date" for their most recent role, or when there is a gap between their last update and the current year (2025), their last known residency, sector, and education level are carried forward to the present day. This methodology accounts for the fact that professionals often delay updating profiles following career breaks, layoffs, or role transitions, providing a more reliable and less volatile view of the active talent pool.

4 Using 5-digits International/Cooperative Patent Classification (IPC/CPC) codes.

## Talent Supply

(1) Talent supply

This dataset series provides a detailed view of global AI talent distribution from 2010 to 2025, structured for longitudinal analysis to support both trend tracking and cross-country comparisons. It is delivered in three related panel datasets:

1. Talent Supply - Country by Year
2. Talent Supply - Education by Year and Country
3. Talent Supply - Sector by Year and Country

These datasets are split rather than combined into one file, but all follow a consistent methodology for data aggregation, calculation of absolute numbers, and percentage shares (where appropriate). Country assignments for each year are derived from individuals' career timelines, using the location associated with their experience for that year. Similarly, sector information is documented annually from the career timeline, meaning an individual is linked to a sector for each year they worked in it. Education is mapped as a time series from the education timeline, so an individual receives a count in the relevant category (e.g., "master's") during the year they pursued that education. This integrated approach ensures accurate representation of AI talent by geography, sector, and education over time. There are six education levels. Sectors are based broadly on LinkedIn sectors. There are 336 sectors.

(2) Areas of specialization

This dataset provides AI talent counts by country, country code, and area of specialization, based on the most recent data for each individual. The country is determined from the individual's latest, recorded experience and its associated location. Areas of specialization are identified using each person's top two current specializations. The top 100 specializations are taken. Each individual is then mapped to their country and relevant specializations, and counts are calculated.

(3) Mobility

This dataset captures the mobility and career transitions of AI talent, providing insights into three key dimensions:

- Country and regional movement: tracks inflows and outflows of AI talent across countries and regions, highlighting migration trends and global talent flows.
- Sector transitions: monitors shifts between major sectors such as academia, industry, and government, revealing patterns in career progression and workforce dynamics.

To ensure accuracy and reduce volatility caused by delayed profile updates, all trends are smoothed using a 12-month moving average, delivering a clearer and more reliable view of long-term changes. When professionals change jobs, there is often a delay before they update their profiles—especially in instances of layoffs or career breaks.

(4) Gender distribution

This dataset presents the gender distribution of AI talent by country and year, expressed as percentages. For each country and each year in the time range, the share of male and female AI professionals is calculated as a proportion of the total AI talent pool for that country-year combination. This structure enables analysis of gender representation trends over time and supports cross-country comparisons.

# Chapter 2: Technical Performance

## Benchmarks

In this chapter, the AI Index reports on benchmarks, recognizing their importance in tracking AI's technical progress. As a standard practice, the Index sources benchmark scores from leaderboards and public repositories, as well as company papers, blog posts, and product releases. The Index operates under the assumption that the scores reported by companies are accurate and factual. The benchmark scores in this section are current as of early 2026. However, since the publication of the AI Index, newer models may have been released that surpass current state-of-the-art scores.

# Chapter 3: Responsible AI

## Conference Submissions Analysis

For the analysis on responsible AI-related conference submissions, the AI Index examined the number of responsible AI–related academic submissions at the following conferences: AAAI, AIES, FAccT, ICML, ICLR, and NeurIPS. Specifically, the team scraped the conference websites or repositories of conference submissions for papers containing relevant keywords indicating they could fall into a particular responsible AI category. The papers related to the most ambiguous keywords (e.g., "bias") were then manually verified by a human team to confirm their categorization, while the rest were controlled according to a categorical RAI-coherence label, produced by prompting each paper title and abstract to a large language model. It is possible for a single paper to belong to multiple responsible AI categories.

The keywords searched include:

**Fairness and bias:** algorithmic fairness, bias detection, bias mitigation, discrimination, equity in AI, ethical algorithm design, fair data practices, fair ML, fairness and bias, group fairness, individual fairness, justice, nondiscrimination, representational fairness, unfair, unfairness.

**Privacy and data governance:** anonymity, confidentiality, data breach, data ethics, data governance, data integrity, data privacy, data protection, data transparency, differential privacy, inference privacy, machine unlearning, privacy by design, privacy-preserving, secure data storage, trustworthy data curation.

**Security:** adversarial attack, adversarial learning, AI incident, attacks, audits, cybersecurity, ethical hacking, forensic analysis, fraud detection, red teaming, safety, security, security ethics, threat detection, vulnerability assessment.

**Transparency and explainability:** algorithmic transparency, audit, auditing, causal reasoning, causality, explainability, explainable AI, explainable models, human-understandable decisions, interpretability, interpretable models, model explainability, outcome explanation, transparency, xAI.

# Chapter 4: Economy

## Quid

*Quid insights prepared by Heather English and Courtney Prabhakar*

Quid uses its own in-house LLM and other smart search features, as well as traditional Boolean query, to search for focus areas, topics, and keywords within many datasets: social media, news, forums and blogs, companies, patents, and other custom feeds of data (e.g., survey data). Quid has many visualization options and data delivery endpoints, including network graphs based on semantic similarity, in-platform dashboarding capabilities, and programmatic PostgreSQL database delivery. Quid applies best-in-class AI and NLP to reveal hidden patterns in large datasets, enabling users to make data-driven decisions accurately, quickly, and efficiently.

### Search, Data Sources, and Scope

Over 8 million global public and private company profiles from multiple data sources are indexed to search across company descriptions, while filtering and including metadata ranging from investment information to firmographic information, such as founding year, headquarter location, and more. Company information is updated on a weekly basis. The Quid algorithm reads a large amount of text data from each document to make links between different documents based on their similar language. This process is repeated at an immense scale, which produces a network of different clusters identifying distinct topics or focus areas. Trends are identified based on keywords, phrases, people, companies, and institutions that Quid identifies and other metadata that is put into the software.

### Data

### Companies

Organization data is embedded from Capital IQ and Crunchbase. These companies include every type of organization (private, public, operating, operating as a subsidiary, out of business) throughout the world. The investment data includes private investments, M&A, public offerings, minority stakes held by private equity and venture capital, corporate venture arms, governments, and institutions both within and outside the United States. Some data is unavailable—for instance, when investors' names or funding amounts are not disclosed. Quid embeds Capital IQ data as a default and adds in data from Crunchbase for the data points that are not captured in Capital IQ. This not only yields comprehensive and accurate data on all global organizations, but it also captures early-stage startups and funding events data.

### Search Parameters

Boolean query is used to search for focus areas, topics, and keywords within the archived company database and within their business descriptions and websites. Quid can filter out the search results by HQ regions, investment amount, operating status, organization type (private/ public), and founding year. Quid then visualizes these companies by semantic similarity. If there are more than 7,000 companies from the search result, Quid selects the 7,000 most relevant companies for visualization based on the language algorithm. Boolean search: "artificial intelligence" or "AI" or "machine learning" or "deep learning."

### Companies

- Global AI and ML companies that have received investments (private, IPO, M&A) from Jan. 1, 2015, to Dec. 31, 2025.
- Global AI and ML companies that have received over $1.5 million for the past 10 years (Jan. 1, 2015, to Dec. 31, 2025).
- Global data was also pulled for a generative AI query (Boolean search: "generative AI" or "gen AI" OR "generative artificial intelligence") for companies that have received over $1.5 million for the past 10 years (Jan. 1, 2015, to Dec. 31, 2025).

### Target Event Definitions

- **Private placement:** A private placement is a sale of newly issued securities (equity or debt) by a company to a select investor or group of investors. The stakes that buyers take in private placements are often minority stakes (under 50%), although it is possible to take control of a company through a private placement, in which case the private placement would be a majority stake investment.
- **Minority investment:** This refers to a minority stake acquisition, which takes place when a buyer acquires less than 50% of the existing ownership stake in entities, asset products, and business divisions.
- **M&A:** This refers to a buyer acquiring more than 50% of the existing ownership stake in entities, asset products, and business divisions.

## McKinsey & Company

Data used in the "Corporate Activity" section was sourced from the McKinsey global survey "The State of AI in 2025: Agents, Innovation, and Transformation."

The online survey ran from June 25, 2025, to July 29, 2025, and garnered responses from 1,993 participants in 105 nations representing the full range of regions, industries, company sizes, functional specialties, and tenures. Thirty-eight percent of respondents say they work for organizations with more than $1 billion in annual revenues. To adjust for differences in response rates, the data is weighted by the contribution of each respondent's nation to global GDP.

The AI Index also considered data from previous iterations of the McKinsey survey. These include:
The State of AI in Early 2024: Gen AI Adoption Spikes and Starts to Generate Value
The State of AI: How Organizations Are Rewiring to Capture Value
The State of AI in 2023: Generative AI's Breakout Year
The State of AI in 2022—and a Half Decade in Review
The State of AI in 2021
The State of AI in 2020
AI Proves Its Worth, But Few Scale Impact (2019)
AI Adoption Advances, But Foundational Barriers Remain (2018)

## Lightcast

*Prepared by Elena Magrini and Rebecca Milde*

Lightcast delivers job market analytics that empower employers, workers, and educators to make data-driven decisions. The company's artificial intelligence technology analyzes hundreds of millions of job postings and real-life career transitions to provide insight into labor market patterns. This real-time strategic intelligence offers crucial insights, such as what jobs are most in demand, the specific skills employers need, and the career directions that offer the highest potential for workers. For more information, visit www.lightcast.io.

### Job Postings Data

To support these analyses, Lightcast mined its dataset of billions of job postings collected since 2010. Lightcast collects postings from over 51,000 online job sites to develop a comprehensive, real-time portrait of labor market demand. It aggregates job postings, removes duplicates, and extracts data from job postings text. This includes information on job title, employer, industry, and region, as well as required experience, education, and skills.

Job postings are useful for understanding trends in the labor market because they allow for a detailed, real-time look at the skills employers seek. To assess the representativeness of job postings data, Lightcast conducts a number of analyses to compare the distribution of job postings to the distribution of official government and other third-party sources in the United States. The primary source of government data on U.S. job postings is the Job Openings and Labor Turnover Survey (JOLTS) program, conducted by the Bureau of Labor Statistics. Based on comparisons between JOLTS and Lightcast, the labor market demand captured by Lightcast data represents over 99% of the total labor demand. Jobs not posted online are usually in small businesses (the classic example being the "Help Wanted" sign in a restaurant window) and union hiring halls.

### Measuring Demand for AI

To measure the demand by employers of AI skills, Lightcast uses its skills taxonomy of over 33,000 skills.[5] These skills are organized hierarchically in over 400 skills clusters and 32 skills categories.

From this full list, Lightcast identified a number of skill clusters within the broader umbrella of AI, for a total of 10 clusters. Nine of these clusters were included in the 2025 AI Index report: AI Ethics, Governance, and Regulation; Artificial Intelligence; Autonomous Driving; Generative AI; Machine Learning; Natural Language Processing; Neural Networks; Robotics; and Visual Image Recognition. An additional cluster—Agentic AI—was added this year to reflect developments at the AI innovation frontier.

Across the 10 clusters, Lightcast experts identified over 300 AI skills, and a job posting is considered an AI job if it includes one of those skills within the text of the job listing.

The list of AI skills from Lightcast data are shown below, with associated skill clusters. A total of 24 skills were added since the previous edition, and these skills are highlighted in yellow below. While some skills are considered to be in the AI cluster specifically, for the purposes of this report, all skills below were considered AI skills.

**AI Agents** *[New cluster - it includes a combination of skills already within the AI definition and new skills added in 2025]*: Chatbot, Agentic AI, ChatGPT, AI Agents, Conversational AI, Microsoft Copilot, Multi-Agent Systems, Langgraph, Microsoft Copilot Studio, Agentic Systems, Google Bard, Intelligent Agent, Intelligent Virtual Assistant, Embedded AI, Customer Engagement Suite with Google AI, Bot Framework, Amazon Alexa, Cortana, Interactive Kiosk, IPSoft Amelia, CrewAI, Game Ai, Generative AI Agents, Autonomic Computing, Botpress, Cursor AI, Swarm Intelligence, ChatGPT Integration, AgentGPT.

**AI Ethics, Governance, and Regulations:** Ethical AI, AI Security, Data Sovereignty, AI Safety, Artificial Intelligence Risk, AI Literacy, Human AI Interaction, AI Alignment.

**Artificial Intelligence:** Artificial Intelligence, Data Version Control (DVC), Edge Intelligence, Applications Of Artificial Intelligence,

5 https://lightcast.io/open-skills

AI Research, Artificial Intelligence Development, AIOps (Artificial Intelligence For IT Operations), AI/ML Inference, Automated Data Cleaning, AI Innovation, Knowledge-Based Systems, Artificial Intelligence Markup Language (AIML),
Intelligent Systems, PineCone, Synthetic Data Generation, Explainable AI (XAI), Open Neural Network Exchange (ONNX), Knowledge Engineering, AI Testing, Expert Systems, Azure Cognitive Services, Knowledge Distillation, Knowledge-Based Configuration, Artificial Intelligence Systems, Weka, Knowledge Representation, Cognitive Computing, Reasoning Systems, Intelligent Control, Cognitive Automation, Baidu, Weaviate, AI Personalization,
Computational Intelligence, Watson Studio, Artificial General Intelligence, Watson Conversation, Neuro-Symbolic AI, OpenAI Gym, Operationalizing AI, Qdrant, PennyLane, Google Quantum AI.

**Autonomous driving:** Advanced Driver Assistance Systems, Autonomous Vehicles, Light Detection And Ranging (LiDAR), Remote Sensing, Autonomous System, Guidance Navigation And Control Systems, OpenCV, Object Tracking, Dynamic Routing, Path Analysis, Scene Understanding, Autonomous Cruise Control Systems, Path Finding, Unmanned Aerial Systems (UAS).

**Generative AI:** Generative Artificial Intelligence, Large Language Modeling, Prompt Engineering, Retrieval Augmented Generation, Variational Autoencoders, Generative Adversarial Networks, Text to Speech (TTS), Multimodal Models, Image Inpainting, Text Summarization, Multimodal Learning, Image Super-Resolution, Stable Diffusion, DALL-E Image Generator, Adobe Sensei, AI-Generated Code, Context Engineering, ChatGPT Prompt,
Beautiful.AI, Variational Autoencoders (VAEs).

**Machine Learning:** Machine Learning, Apache Spark, PyTorch (Machine Learning Library),
Predictive Modeling, MLOps (Machine Learning Operations), Scikit-Learn (Python Package),
Machine Learning Algorithms, AWS SageMaker, Reinforcement Learning, Transformer (Machine Learning Model), Feature Engineering, MLflow, Data Classification, Vertex AI,
Decision Tree Learning, Azure Machine Learning, Machine Learning Model Training, Unsupervised Learning, Machine Learning Methods, Distributed Machine Learning, Kubeflow,
Recommender Systems, Concept Drift Detection, Random Forest Algorithm, Xgboost, Boosting,
Support Vector Machine, Supervised Learning, Torch (Machine Learning), Dask (Software), Association Rule Learning, AutoGen, Feature Extraction, Gradient Boosting, Cluster Analysis,
Feature Selection, H2O.ai, ModelOps, Hyperparameter Optimization, Transfer Learning,
Automated Machine Learning, Bagging Techniques, Collaborative Filtering, Machine Learning Model Monitoring And Evaluation, Federated Learning, K-Means Clustering, Dimensionality Reduction, Support Vector Machines (SVM), Feature Learning, Cyber-Physical Systems, Neural Architecture Compression, Ensemble Methods, Confusion Matrix, Training Datasets, Decision Models, Bayesian Belief Networks, Naive Bayes Classifier, Programmatic Media Buying,
Genetic Algorithm, Meta-Learning, Loss Functions, Gradient Boosting Machines (GBM),
Semi-Supervised Learning, Test Datasets, Objective Function, Markov Chain, Hidden Markov Model, Gaussian Process, Theano (Software), Matrix Factorization, Apache Mahout,
CHi-Squared Automatic Interaction Detection (CHAID), Microsoft Cognitive Toolkit (CNTK),
Kernel Methods, Sorting Algorithm, Topological Data Analysis (TDA), Pydata, Adversarial Machine Learning, Oracle Autonomous Database, Perceptron, Inference Engine, Dbscan,
Embedded Intelligence, Attention Mechanisms, Apache MADlib, Incremental Learning, Neural Architecture Search (NAS), PyTorch Lightning, Vowpal Wabbit, Dlib (C++ Library), AdaBoost (Adaptive Boosting), Boltzmann Machine, Zero Shot Learning, mlpack (C++ Library), Google AutoML, t-SNE (t-distributed Stochastic Neighbor Embedding), Apache SINGA, Soft Computing,
Evolutionary Programming, Predictionio, Expectation Maximization Algorithm, Google Cloud ML Engine, Meta Learning, LIBSVM, Classification And Regression Tree (CART),

**Natural Language Processing:** AI Translation, Natural Language Processing (NLP), Text Retrieval Systems, Screen Reader, Hugging Face (NLP Framework), Language Models,
LLaMA (Language Model), Natural Language Understanding (NLU), Tokenization, Text Mining,
Sentiment Analysis, Speech Recognition, BERT (NLP Model), Optical Character Recognition (OCR), Computational Linguistics, Hugging Face Transformers, Semantic Search, Semantic Kernel, Natural Language Understanding, Natural Language Generation (NLG), Machine Translation, Voice Assistant Technology, Amazon Textract, Speech Synthesis, Natural Language Generation, Word2Vec Models, Voice User Interface, Semantic Analysis, Speech Recognition Software, Dialog Systems, Statistical Language Acquisition, Word Embedding,
Fuzzy Logic, Voice Interaction, Microsoft LUIS, ANTLR, Apache OpenNLP, Handwriting Recognition, Summarization Methods, Kaldi, Latent Dirichlet Allocation, Natural Language Programming, Semantic Parsing, Vespa, Nuance Mix, fastText, Seq2Seq, Natural Language Toolkits, Small Language Model, Natural Language User Interface, Sentence Transformers, AI Copywriting, DeepSpeech, Lexalytics, Nearest Neighbour Algorithm, Semantic Interpretation For Speech Recognition, Shogun, Text-To-Speech, Language Model.

**Neural Networks:** TensorFlow, Deep Learning, Keras (Neural Network Library), Deep Learning Methods, Graph Neural Networks (GNNs), Apache MXNet, Recurrent Neural Network (RNN),
Long Short-Term Memory (LSTM), Residual Networks (ResNet), Convolutional Neural Networks (CNN), Recurrent Neural Networks (RNNs), Deep Reinforcement Learning (DRL),
Cudnn, Sequence-to-Sequence Models (Seq2Seq), Autoencoders, Caffe2, Reinforcement Learning (RL), OpenVINO, Evolutionary Acquisition Of Neural Topologies, Fast.ai, Caffe (Framework), Chainer (Deep Learning Framework), Deeplearning4j, Spiking Neural Networks,
Neural Ordinary Differential Equations, PaddlePaddle, Pybrain, Convolutional Neural Networks,
Artificial Neural Networks.

**Robotics:** Robotic Systems, Robot Operating Systems, OpenAI Gym Environments,
Motion Planning, SLAM Algorithms (Simultaneous Localization And Mapping), Robot Framework, Robotic Automation Software, Advanced Robotics, Robotic Programming,
Servomotor, Nvidia Jetson, Robotic Liquid Handling Systems, Cognitive Robotics, Reinforcement Learning from Human Feedback (RLHF), Meta-Reinforcement Learning.

**Visual Image Recognition:** Computer Vision, Image Analysis, Machine Vision, General-Purpose Computing On Graphics Processing Units, 3D Reconstruction, Image Sensor,
Image Recognition, Eye Tracking, Facial Recognition, Motion Analysis, Image Segmentation,
Object Recognition, Pose Estimation, Contextual Image Classification, Instance Segmentation,
Digital Image Processing, Thermal Imaging Analysis, Gesture Recognition, Activity Recognition,
OmniPage, Image Captioning, Imagenet, Face Detection, Digital Twin Technology, Deck.gl, Image Matching, Mnist, Realsense.

## LinkedIn

*Prepared by Rosie Hood, Akash Kaura, and Mar Carpanelli*

### LinkedIn Data

This body of work is drawn from the anonymized and
aggregated profile information of LinkedIn's more than 1.3 billion members worldwide. It therefore is influenced by how members choose to use the platform, which can vary based on professional, social, and regional culture, as well as overall site availability and accessibility. In publishing these insights from LinkedIn's Economic Graph, we want to provide accurate statistics while ensuring our members' privacy. As a result, all data shows aggregated information for the corresponding period following strict data quality thresholds that prevent disclosing information about specific individuals.

### Country Sample

LinkedIn provides data on Argentina, Australia, Austria, Belgium, Brazil, Canada, Chile, Costa Rica, Croatia, Cyprus, Czechia, Denmark, Estonia, Finland, France, Germany, Greece, Hong Kong SAR, Hungary, Iceland, India, Indonesia, Ireland, Israel, Italy, Latvia, Lithuania, Luxembourg, Mexico, Netherlands, New Zealand, Norway, Poland, Portugal, Romania, Saudi Arabia, Singapore, Slovenia, South Africa, South Korea, Spain, Sweden, Switzerland, Türkiye, United Arab Emirates, United Kingdom, United States, and Uruguay.

### Skills

LinkedIn members self-report their skills on their LinkedIn profiles. Currently, more than 42,000 distinct, standardized skills are identified by LinkedIn. LinkedIn categorizes AI skills into two mutually exclusive groups: "AI Engineering" and "AI Literacy." AI Engineering skills broadly refer to the technical expertise and practical competencies required to design, develop, deploy, and maintain artificial intelligence systems, and AI Literacy skills refer to the knowledge, abilities, and critical thinking competencies needed to understand, evaluate, and effectively interact with artificial intelligence technologies. As skills are constantly evolving, LinkedIn maintains and refreshes these classifications on a periodic basis.

### Industry

LinkedIn's industry taxonomy is a collection of entities that share economic activities and contribute to a specific product or service. An industry represents the products or services that a company offers or sells. LinkedIn analyzes the following industries in the context of AI skills penetration: Education; Financial Services; Manufacturing; Professional Services; and Technology, Information, and Media. For all other metrics, LinkedIn analyzes the following industries: Accommodation and Food Services; Administrative and Support Services; Construction; Consumer Services; Education; Entertainment Providers; Farming, Ranching, and Forestry; Financial Services; Government Administration; Hospitals and Health Care; Manufacturing; Oil, Gas, and Mining; Professional Services; Real Estate and Equipment Rental Services; Retail; Technology, Information, and Media; Transportation, Logistics, Supply Chain, and Storage; Utilities; and Wholesale.

### Gender

LinkedIn recognizes that some members identify beyond the traditional gender constructs of "man" and "woman." If not explicitly self-identified, LinkedIn infers the gender of members included in this analysis either by the pronouns used on their LinkedIn profiles or on the basis of first name. Members whose gender could not be inferred were excluded from any gender analysis. Note: LinkedIn filtered out countries where its gender attribution algorithm lacked sufficient coverage.

### AI Jobs or Occupations

LinkedIn member titles are standardized and grouped into over 16,000 occupations. These are not sector or
country specific. An AI job is an occupation that requires AI skills to perform the job. Examples include (but are not limited to): machine learning engineer, artificial intelligence specialist, data scientist, and computer vision engineer.

### AI Talent

A LinkedIn member is considered AI talent if they have explicitly added at least two AI engineering skills to their profile and/or they are or have been employed in an AI job. A LinkedIn member is considered AI literate if they have added at least one AI literacy skill to their profile.

## Methodologies

### 1. Top AI Skills

These are the AI skills most frequently added by members by year.

**Sample interpretation:** The most added AI engineering skills globally are machine learning, AI, and deep learning.

### 2. Fastest Growing AI Skills

The year-over-year (YoY) growth rate for AI skills most frequently added by all members. Note: LinkedIn implements thresholds to skill add volumes in the most recent year; thresholds are set at the 50th percentile of the most recent year's AI skill adds distribution by country.

**Sample interpretation:** The fastest growing AI engineering skills globally are custom GPTs, AI productivity, and AI agents.

### 3. AI Talent Concentration

The counts of AI talent are used to calculate talent concentration metrics. For instance,, to calculate the country-level AI talent concentration, LinkedIn uses the counts of AI talent in a country divided by the counts of LinkedIn members in that respective country. Note: Concentration metrics may be influenced by LinkedIn coverage in these countries and should be used with caution.

**Sample interpretation:** AI talent with AI engineering skills represents 0.88% of LinkedIn members in the United States.

### 4. Relative AI Talent Hiring Rate YoY Ratio

The LinkedIn hiring rate is a measure of hires normalized by LinkedIn membership. It is computed as the percentage of LinkedIn members who added a new employer in the same period the job began, divided by the total number of LinkedIn members in the corresponding location.

The AI hiring rate is computed using the overall hiring rate methodology, but it only considers members classified as AI talent. The relative AI talent hiring rate YoY ratio is the year-over-year change in the AI hiring rate relative to the overall hiring rate in the same country. LinkedIn shares a three-month moving average.

**Sample interpretation:** In the United States, the ratio of AI talent hiring relative to overall hiring has grown 24.7% year over year.

### 5. Skills Penetration

**Skills Genome**

For any entity (occupation, country, industry, etc.), the skills genome is an ordered list (a vector) of the 50 most characteristic skills of that entity. These most characteristic skills are determined using a TF-IDF algorithm, which down-ranks ubiquitous skills that add little information about that specific entity (e.g., Microsoft Word) and up-ranks skills unique to that specific entity (e.g., artificial intelligence). Further details are available at LinkedIn's Skills Genome and the LinkedIn–World Bank Methodology note.

As an example, Table 1 details the skills genome of the technology, information, and media industry in the United States in 2024, displaying the top 20 skills ranked by TF-IDF.

**Skills genome of the technology, information and media industry in the United States in 2024**
Top 20 skills ranked by TF-IDF

| TF-IDF skill rank | Skill name |
|---|---|
| 1 | Amazon Web Services (AWS) |
| 2 | Software as a Service (SaaS) |
| 3 | Artificial intelligence (AI) |
| 4 | Python (programming language) |
| 5 | Go-to-market strategy |
| 6 | Customer success |
| 7 | Large language models (LLM) |
| 8 | Salesforce.com |
| 9 | SQL |
| 10 | Generative AI |
| | |

**AI Skills Penetration**
This metric measures the intensity of AI skills in a given entity through the following methodology:

- LinkedIn computes frequencies for all self-added skills by LinkedIn members in a given entity (occupation, industry, etc.) from 2015 on.
- LinkedIn reweights re-weight skill frequencies using a TF-IDF model to get the top 50 most representative skills in that entity. These 50 skills compose the skill genome of that entity.
- LinkedIn computes the share of skills that belong to the AI skill group out of the top skills in the selected entity.

**Interpretation:** The AI skill penetration rate signals the prevalence of AI skills across occupations, or the intensity with which LinkedIn members use AI skills in their jobs. For example, the top 50 skills for the occupation of engineer are calculated based on the weighted frequency with which they appear in LinkedIn members' profiles. If four of the skills that engineers possess belong to the AI skill group, this measure indicates that the penetration of AI skills is estimated at 8% ( 4 out of 50) among engineers.

**Relative AI Skills Penetration**
To allow for skills penetration comparisons across countries, the skills genomes are calculated and a relevant benchmark is selected (e.g., a global average). A ratio is then constructed between a country and the benchmark's AI skills penetrations, controlling for occupations.

**Sample interpretation:** A country's relative AI skills penetration of 1.5 indicates that AI skills are 1.5 times as frequent as in the benchmark, for an overlapping set of occupations.

**Global Comparison**
For cross-country comparison, LinkedIn presents the relative penetration rate of AI skills, measured as the sum of the penetration of each AI skill across occupations in a given country, divided by the average global penetration of AI skills across the overlapping occupations in a sample of countries.

**Sample interpretation:** A relative penetration rate of 2 means the average penetration of AI skills in a particular country is two times the global average across the same set of occupations.

**Global Comparison: By Industry**
The relative AI skills penetration by country for a given industry provides an in-depth sectoral decomposition of AI skills penetration across industries and countries.

**Sample interpretation:** A country's relative AI skills penetration rate of 2 in the education sector means the average penetration of AI skills in that country is two times the global average across the same set of occupations in that sector.

**Global Comparison: By Gender**
The relative AI skills penetration by gender provides a cross-country comparison of AI skills penetrations within a gender. Since the global averages are distinct for each gender, this metric should only be used to compare country rankings within each gender, not for cross-gender comparisons within countries.

**Sample interpretation:** If a country's AI skills penetration for women is 1.5, that means members who are women in that country are 1.5x more likely to list AI skills than the average female member in all countries pooled together across the same set of occupations that exist in the country-gender combination.

**Global Comparison: Across Genders**
The relative AI skills penetration across genders allows for cross-gender comparisons within and across countries globally, since LinkedIn compares a country's AI skills penetration by gender to the same global average regardless of gender.

### 6. Female Representation in AI
The share of AI talent that are women.

**Sample interpretation:** Female representation within AI talent with AI engineering skills is 30.5% globally.

### 7. AI Talent Migration
Data on migration comes from the World Bank Group–LinkedIn "Digital Data for Development" partnership (see [https://linkedindata.worldbank.org/](https://linkedindata.worldbank.org/) and Zhu et al. (2018)). LinkedIn migration rates are derived from the self-identified locations of LinkedIn member profiles. For example, when a LinkedIn member updates their location from Paris to London, this is counted as a migration. Migration data is available from 2019 on.

LinkedIn data provides insights to countries on AI talent gained or lost due to migration trends. AI talent migration is considered for all members with AI skills/holding AI jobs at time "t" for country A as the country of interest and country B as the source of inflows and destination for outflows. Net AI talent migration between country A and country B would therefore be calculated as:

$$Net\,AI\,Talent\,Migration_{a,b,t} = \frac{Net\,AI\,Talent\,flows_{a,b,t}}{Member\,count_{a,t}}$$

Net flows are defined as total arrivals minus departures within a given time period. LinkedIn membership varies between countries, making it a challenge to interpret absolute movements of members from one country to another. Migration flows are therefore normalized with respect to each country. For example, for country A, all absolute net flows into and out of country A, regardless of origin and destination countries, are normalized based on the LinkedIn membership of country A at the end of each year and multiplied by 10,000. Hence, this metric indicates relative talent migration from all countries to and from country A. Note: Minimum thresholds have been applied to ensure that transitions have a sufficient sample size.

**Sample interpretation:** The United States had a positive net flow of AI talent relative to its membership size at 1.07 net flow per 10,000 members.

### 8. Career Transitions into AI Jobs
LinkedIn considers the source occupations that feed AI occupations, analyzing the share of transitions into AI occupations pooled over a five-year period. Career transitions are computed by aggregating member-level job transitions from one occupation to another occupation that the member has previously not held. LinkedIn excludes first occupations added by new graduates and intra-occupation transitions.

**Sample interpretation:** In the United States, 26.9% of transitions into AI engineer came from software engineer, followed by 13.3% from data scientist.

### 9. Fastest Growing AI Occupations

This metric identifies the fastest growing occupations within each AI talent segment by country, based on year-over-year changes in occupation share. For each occupation, LinkedIn calculates the year-over-year growth rate as the percentage change in the share of members listing an AI occupation compared with the same month one year earlier.

**Sample interpretation:** In the United States, AI product manager is among the fastest growing occupations within the AI engineering segment, showing the largest year-over-year increase.

### 10. AI Skills Diffusion Index
The AI skills diffusion index measures the breadth and depth of adoption of AI skills across the LinkedIn member base. It captures both (i) how widespread a given group of AI skills is among members in a country and (ii) how diverse the set of AI skills represented on member profiles is within that group. The index is constructed by tracking the share of members who list at least one AI skill on their profile, adjusted for the diversity of distinct AI skills observed in a given country. Values are indexed relative to a baseline period where a member threshold is reached at the country level, which ensures stability and comparability across countries and over time. The resulting index value at any point in the time series reflects how much the adjusted share of members with that AI skill group has increased relative to the baseline. Because the index is normalized to an early reference point, it is best interpreted as a measure of growth in diffusion rather than as a level or penetration rate.

**Sample interpretation:** In the United States, the AI skills diffusion index for AI engineering indicates that the adjusted share of members listing AI literacy skills is 1075x higher than it was in March 2023.

### Acknowledgments
LinkedIn gratefully acknowledges the contributions of Murat Erer and Carl Shan in developing these methodologies and metrics, and the feedback from our collaborators at the OECD.AI, Stanford Institute for Human-Centered AI, World Bank, and Centro Nacional de Inteligencia Artificial (Cenia).

## International Federation of Robotics (IFR)
Data presented in the Robot Installations section was sourced from the World Robotics 2025 report.

# Chapter 5: Science

## AI Publication Analysis

To quantify growth in AI-related research across the natural sciences, the AI Index team conducted a bibliometric analysis using the Web of Science Core Collection (Advanced Search). For each discipline group—chemistry, materials science, physics, astronomy, biology, neuroscience, medicine, geosciences, atmospheric and ocean sciences, and environmental sciences—the team constructed queries that combined AI-related topic search terms (TS = “machine learning,” “deep learning,” “artificial intelligence,” “neural network*,” “foundation model*,” “large language model*,” and “generative AI”) with the corresponding Web of Science subject categories (WC). Results were limited to articles (DT = “Article”). For each query, the team obtained annual publication counts by accessing the full publication year distribution and exporting the table, and ran an aggregate query combining all subject categories to assess overall trends. All searches were performed on March 20, 2026.

# Chapter 6: Medicine

## The Central Dogma

**Number of AI driven protein research publications**
The following PubMed queries were used to retrieve total annual paper counts for 2024 and 2025. Queries below are shown for 2025.

**All Papers**: ("Artificial Intelligence"[Mesh] OR "Machine Learning"[Mesh] OR "Deep Learning"[Mesh] OR "artificial intelligence"[tiab] OR "machine learning"[tiab] OR "deep learning"[tiab] OR "neural network*"[tiab] OR "generative AI"[tiab] OR "language model*"[tiab] OR "AI"[tiab]) AND 2025[dp].
**Function Prediction:** ("Artificial Intelligence"[Mesh] OR "Machine Learning"[Mesh] OR "Deep Learning"[Mesh] OR "artificial intelligence"[tiab] OR "machine learning"[tiab] OR "deep learning"[tiab] OR "neural network*"[tiab] OR "AI"[tiab]) AND ("protein function"[tiab] OR "function prediction"[tiab] OR "functional annotation"[tiab] OR "protein annotation"[tiab]) AND 2025[dp].
**Protein Structure Prediction:** ("Artificial Intelligence"[Mesh] OR "Machine Learning"[Mesh] OR "Deep Learning"[Mesh] OR "artificial intelligence"[tiab] OR "machine learning"[tiab] OR "deep learning"[tiab] OR "neural network*"[tiab] OR "AI"[tiab]) AND ("protein structure"[tiab] OR "structure prediction"[tiab] OR "protein folding"[tiab] OR "AlphaFold"[tiab] OR "ESMFold"[tiab] OR "RoseTTAFold"[tiab]) AND 2025[dp].
**Protein Drug Interactions:** ("Artificial Intelligence"[Mesh] OR "Machine Learning"[Mesh] OR "Deep Learning"[Mesh] OR "artificial intelligence"[tiab] OR "machine learning"[tiab] OR "deep learning"[tiab] OR "neural network*"[tiab] OR "AI"[tiab]) AND ("protein-drug"[tiab] OR "drug-target"[tiab] OR "binding affinity"[tiab] OR "molecular docking"[tiab] OR "protein-ligand"[tiab] OR "target interaction*"[tiab]) AND 2025[dp].
**Synthetic Protein Design:** ("Artificial Intelligence"[Mesh] OR "Machine Learning"[Mesh] OR "Deep Learning"[Mesh] OR "artificial intelligence"[tiab] OR "machine learning"[tiab] OR "deep learning"[tiab] OR "neural network*"[tiab] OR "AI"[tiab]) AND ("synthetic protein"[tiab] OR "protein design"[tiab] OR "de novo protein"[tiab] OR "de novo design"[tiab] OR "protein engineering"[tiab]) AND 2025[dp].

**Number of publications on AI for drug discovery**
The AI Index team ran the search query below on PubMed, limiting results to articles published from 2018 on.

Search query: ("Artificial Intelligence"[Mesh] OR "Machine Learning"[Mesh] OR "Deep Learning"[Mesh] OR "Neural Networks, Computer"[Mesh] OR "artificial intelligence"[tiab] OR "machine learning"[tiab] OR "deep learning"[tiab] OR "neural network*"[tiab] OR "generative AI"[tiab] OR "large language model*"[tiab] OR "LLM*"[tiab] OR "AI"[tiab]) AND ("Drug Discovery"[Mesh] OR "Drug Design"[Mesh] OR "drug discovery"[tiab] OR "drug design"[tiab] OR "drug development"[tiab] OR "computer-aided drug design"[tiab] OR "CADD"[tiab] OR "de novo drug design"[tiab] OR "target identification"[tiab] OR "lead optimization"[tiab] OR "structure-based drug design"[tiab])

**Number of publications on virtual cell models**
The AI Index team ran the search query below on PubMed.

Search query: "virtual cell"[tiab] OR "virtual cells"[tiab]

## FDA-Approved AI Medical Devices

Data on FDA-approved AI medical devices was sourced from the FDA website, which tracks artificial intelligence and machine-learning (AI/ML)–enabled medical devices.

## Patient Perspectives on AI in Health Care

A comprehensive literature search was conducted in PubMed/MEDLINE, Embase, and Scopus to identify empirical studies published between January 1, 2020, and December 31, 2025, that examined patient, family, or public perspectives on the use of artificial intelligence (AI) in healthcare. The search strategy was structured around five core concepts: (1) clinical artificial intelligence (including artificial intelligence, machine learning, deep learning, neural networks, and large language models such as ChatGPT/GPT); (2) explicit patient, family, or public perspectives; (3) healthcare and clinical contexts; (4) AI use or implementation in care; and (5) empirical data collection methods.

Searches required explicit mention of AI technologies in the title or abstract and included only studies reporting primary data collected through qualitative, quantitative, or mixed-methods designs (e.g., surveys, interviews, focus groups, cross-sectional or observational studies). To maintain focus on AI as the object of perception rather than a background analytic tool, studies centered on telehealth were excluded. Systematic reviews, scoping reviews, narrative reviews, and other literature-only reviews were also excluded at the search level using title and abstract text exclusions.

Search strategies were adapted to the indexing and field structures of each database, including the use of MeSH terms in PubMed and Emtree terms in Embase, with database-specific refinements applied to ensure consistent conceptual alignment. All retrieved records were subsequently screened at the title/abstract and full-text levels to confirm eligibility. Included papers were tagged by clinical specialty and geographic representation by four reviewers. To facilitate visualization, tags were res-structured as binary variables using Claude (Sonnet 4.6), and outputs were manually reviewed for accuracy. Mock figures were generated using Claude

(Sonnet 4.6).

Search strategy:

**Concept A — Clinical Artificial Intelligence**

*AI technologies likely to be encountered/used in care*

- Artificial intelligence
- Machine learning
- Deep learning
- Neural networks
- Generative AI
- Large language models (LLMs); ChatGPT; GPT
- Clinical AI
- Predictive models
- Clinical decision support

**Concept B — Patient, Family, and Public Perspectives**

*Explicitly stated perspectives only*

- Patient perspective*
- Patient perception*
- Patient attitude*
- Public perception*
- Public attitude*
- Family perspective*
- Caregiver perspective*

**Concept C — Healthcare Context**

*Anchors retrieval to clinical/medical settings*

- Delivery of healthcare / healthcare
- Healthcare
- Medicine
- Clinical care
- Medical decision-making

**Concept D — AI Use/Implementation in Care**

*Ensures AI is the object of perception as used in practice*

- use/used
- implementation
- deploy*
- application*
- integration
- interaction*

**Concept E — Empirical Study Design/Data Collection**

*Privileges studies reporting primary data*

- survey*
- questionnaire*
- interview*
- focus group*
- qualitative
- quantitative
- mixed methods
- cross-sectional
- observational

**Exclusions**

*Prevents drift and removes review-only literature*

- telehealth
- telemedicine
- mobile health/mhealth
- systematic review
- scoping review
- narrative review
- literature review
- review of the literature

**Core Limits (applied across databases)**

- Publication years: 2020–2025
- Document types: empirical articles only (reviews excluded via search-string exclusions)
- Language: start with English (non-English may be noted descriptively if needed)

**Search strings implemented on Feb. 9, 2026**

**PubMed (174 articles):**

((((“Artificial Intelligence”[MeSH Terms] OR “Machine Learning”[MeSH Terms] OR “Deep Learning”[MeSH Terms] OR “neural networks, computer”[MeSH Terms] OR “Artificial Intelligence”[Title/Abstract] OR “Machine Learning”[Title/Abstract] OR “Deep Learning”[Title/Abstract] OR “neural network*”[Title/Abstract] OR “generative ai”[Title/Abstract] OR “large language model*”[Title/Abstract] OR “chatgpt*”[Title/Abstract] OR “gpt”[Title/Abstract] OR “clinical ai”[Title/Abstract] OR “predictive model*”[Title/Abstract] OR “clinical decision*”[Title/Abstract] OR “decision support”[Title/Abstract]) AND (“use”[Title/Abstract] OR “used”[Title/Abstract] OR “implementation”[Title/Abstract] OR “deploy*”[Title/Abstract] OR “application*”[Title/Abstract] OR “integration”[Title/Abstract] OR “interaction*”[Title/Abstract]) AND (“patient perspective*”[Title/Abstract] OR “patient perception*”[Title/Abstract] OR “patient attitude*”[Title/Abstract] OR “public perception*”[Title/Abstract] OR “public attitude*”[Title/Abstract] OR “family perspective*”[Title/Abstract] OR “caregiver perspective*”[Title/Abstract]) AND (“Delivery of Health Care”[MeSH Terms] OR “healthcare”[Title/Abstract] OR “health care”[Title/Abstract] OR “medicine”[Title/Abstract] OR “clinical care”[Title/Abstract] OR “medical decision-making”[Title/Abstract]) AND (“survey*”[Title/Abstract] OR “questionnaire*”[Title/Abstract] OR “interview*”[Title/Abstract] OR “focus group*”[Title/Abstract] OR “qualitative”[Title/Abstract] OR “quantitative”[Title/Abstract] OR “mixed methods”[Title/Abstract] OR “cross-sectional”[Title/Abstract] OR “observational”[Title/Abstract])) NOT (“telehealth”[Title/Abstract] OR “telemedicine”[Title/Abstract] OR “mobile health”[Title/Abstract] OR “mhealth”[Title/Abstract])) NOT (“systematic review”[Title/Abstract] OR “scoping review”[Title/Abstract] OR “narrative review”[Title/Abstract] OR “literature review”[Title/Abstract] OR “review of the literature”[Title/Abstract])) AND 2020/01/01:2025/12/31[Date - Publication]

**Embase (239 articles retrieved):**

('artificial intelligence'/exp OR 'machine learning'/exp OR 'deep learning'/exp OR 'neural networks'/exp OR 'artificial intelligence':ti,ab OR 'machine learning':ti,ab OR 'deep learning':ti,ab OR 'neural network*':ti,ab OR 'generative ai':ti,ab OR 'large language model*':ti,ab OR chatgpt*:ti,ab OR gpt:ti,ab) AND ((((((patient AND perspective*:ti,ab OR patient) AND perception*:ti,ab OR patient) AND attitude*:ti,ab OR public) AND perception*:ti,ab OR public) AND attitude*:ti,ab OR family) AND perspective*:ti,ab OR caregiver) AND perspective*:ti,ab AND ('health care'/exp OR healthcare:ti,ab OR 'health care':ti,ab OR medicine:ti,ab OR 'clinical care':ti,ab OR 'medical decision making'/exp) AND (survey*:ti,ab OR questionnaire*:ti,ab OR interview*:ti,ab OR 'focus group*':ti,ab OR qualitative:ti,ab OR quantitative:ti,ab OR 'mixed methods':ti,ab OR 'cross-sectional':ti,ab OR observational:ti,ab) AND [2020-2025]/py NOT (telehealth:ti,ab OR telemedicine:ti,ab OR 'mobile health':ti,ab OR mhealth:ti,ab) NOT ('systematic review':ti,ab OR 'scoping review':ti,ab OR 'narrative review':ti,ab OR 'literature review':ti,ab OR 'review of the literature':ti,ab)

**Scopus (368 articles retrieved):**

( TITLE-ABS-KEY ( "artificial intelligence" OR "machine learning" OR "deep learning" OR "neural network*" OR "generative ai" OR "large language model*" OR chatgpt OR gpt ) AND TITLE-ABS-KEY ( "patient perspective*" OR "patient perception*" OR "patient attitude*" OR "public perception*" OR "public attitude*" OR "family perspective*" OR "caregiver perspective*" ) AND TITLE-ABS-KEY ( "health care" OR healthcare OR medicine OR "clinical care" OR "medical decision-making" ) AND TITLE-ABS-KEY ( survey* OR questionnaire* OR interview* OR "focus group*" OR qualitative OR quantitative OR "mixed methods" OR "cross-sectional" OR observational ) ) AND NOT TITLE-ABS-KEY ( telehealth OR telemedicine OR "mobile health" OR mhealth ) AND NOT TITLE-ABS-KEY ( "systematic review" OR "scoping review" OR "narrative review" OR "literature review" OR "review of the literature" ) AND PUBYEAR > 2019 AND PUBYEAR < 2026

## Ethical Considerations

A bibliometric analysis assessed the medical AI and ethics literature, with a focus on data sharing, algorithm sharing, biosecurity, and global health. The AI Index searched PubMed Central for publications from January 2021 through December 2025 using (1) a primary query to identify medical AI articles that discuss ethics and, within that set, additional filters to identify articles that also address (2) data sharing, (3) algorithm sharing, (4) biosecurity, and (5) global health. For the resulting articles, the AI Index mapped the ethics topics covered (see taxonomy below) and grouped them into algorithmic, governance, or societal concerns.

The AI Index used the following taxonomy for the ethics publications:
**Algorithmic:** Bias, Fairness, Explainability, Misuse, Dual Use
**Governance:** Accountability, Transparency, Independent Review, Oversight, Privacy, Trustworthy
**Societal:** Equity, Inequity, Racism, Accessibility, Autonomy, Justice, Community Engagement

# Chapter 7: Education

## Code.org

**Access to Computer Science Education**

Data on access to computer science education was drawn from Code.org's State of AI + CS Education 2025 report and the accompanying interactive dashboard.

**AP Computer Science Data**

The AP Computer Science data, provided to Code.org through an agreement it maintains with the College Board, can be viewed here. The AP Computer Science data is drawn from the College Board's national and state summary reports.

## Computing Research Association (CRA Taulbee Survey)

Computing Research Association (CRA) distributes the Taulbee Survey annually to its member universities in the United States and Canada to collect information about computer science, computer engineering, and information departments. The data used in this report is from the 2024 survey, the results of which were published in 2025. For more information about CRA's methodology, please refer to their report.

## Global K–12 AI Education

The Raspberry Pi Computing Education Research Centre, based in the department of computer science and technology at the University of Cambridge, compiled this dataset, building off the research conducted by the Brookings Institution in its 2021 report Building Skills for Life. One change was made to the dataset to clarify that computer science in the U.S. is available in some schools and districts and not available everywhere as an elective course. For more information about the methodology used by the Raspberry Pi Computing Education Research Centre, please refer to its report. This year, the AI Index team updated the dataset compiled by the Centre with current information as of 2025.

## IPEDS

The Integrated Postsecondary Education Data System (IPEDS) combines annual surveys conducted by the U.S. Department of Education's National Center for Education Statistics. IPEDS gathers information from every college, university, and technical and vocational institution that participates in the federal student financial aid programs.

**Completion Data**

This chapter uses data from the Completions survey, which collects data on the number of students who complete a postsecondary education program. The complete list of majors considered to be AI software–related and AI hardware–related can be found in Appendix C of the January 2025 White House AI Talent Report.

## OECD

This chapter uses data from the OECD Data Explorer, specifically from the table entitled "Number of enrolled students, graduates and new entrants by field of education." The methodology for this dataset can be found in Education at a Glance 2024 Sources, Methodologies and Technical Notes.

# Chapter 8: Policy and Governance

## Global Legislation Records on AI

This year, the AI Index partnered with the Digital Policy Alert to track AI-related bills passed into law. The Digital Policy Alert's pipeline, described below, was used to systematically identify and classify AI-relevant legislation. A final manual review was conducted by the AI Index team to verify all inclusions.

**Overview**
The AI Legislation Tracker employs an automated four-stage pipeline to systematically identify, classify, and extract structured information from AI-relevant legislation across the G20 countries. The pipeline combines autonomous agentic web crawling with large language model (LLM)–based document analysis to process thousands of candidate legal texts with minimal manual intervention.

**Pipeline Architecture**
**Stage 1: Automated Source Crawling**
An AI-driven browser agent navigates a predefined set of legislative source websites. For every source, the agent executes a predefined set of keyword queries—spanning terms such as *artificial intelligence*, *machine learning*, *algorithm*, and other domain-relevant expressions in the source's language—and collects URLs of pages that appear to contain legislative content. The agent operates autonomously, handling site-specific navigation patterns including pagination, cookie consent dialogs, CAPTCHA challenges, and search filters without human intervention.

**Stage 2: Primary Classification**
Each collected URL is fetched and its full text extracted. The document is then submitted to a large language model (Google Gemini) for an initial relevance assessment. The model determines (a) whether the text constitutes AI-relevant legislation and (b) the document type (e.g., primary legislation, secondary regulation, strategy document, executive order). Documents falling below a defined relevance threshold are discarded at this stage.

**Stage 3: Secondary Classification**
Documents that pass the primary classification stage as relevant legislation undergo a second, more granular LLM analysis. This pass extracts a set of structured fields from each law, including but not limited to:
- *AI system types* covered by the legislation
- *Sectors* affected (e.g., healthcare, finance, public administration)
- *Policy objectives* (e.g., safety, transparency, accountability, innovation promotion)
- *Responsible agencies* or regulatory bodies designated by the law
- *Bill or act number* and formal title, serving to deduplicate legislation
- *Legislative status* (e.g., enacted, proposed, under review)

**Stage 4: Quote Extraction**
A final LLM pass identifies and extracts verbatim passages from each legislative text—the specific provisions, definitions, or clauses that substantiate the classification decisions made in the preceding stages. These quotes serve as auditable evidence linking each structured annotation to its source material.

**Data Storage**
The pipeline persists the following artefacts on the server:
- *All visited URLs* and their classification outcomes
- *Full scraped text* of all URLs that passed primary classification (sub-page texts used only for classification are not retained separately)
- *Structured classification fields and extracted quotes*, stored in JSON format
- *Crawl job metadata*, including source identifiers and processing time stamps

## US State-Level AI Legislation

For AI-related bills passed into law, the AI Index performed searches for the keyword "artificial intelligence" in the full text of bills on the websites of all 50 U.S. states. Bills are only counted as passed into law if the keyword appears in the final version of the bill, not just the introduced version. Note that only laws passed from 2015 to 2025 are included. The AI Index team surveyed the following databases:

Alabama, Alaska, Arizona, Arkansas, California, Colorado, Connecticut, Delaware, Florida, Georgia, Hawaii, Idaho, Illinois, Indiana, Iowa, Kansas, Kentucky, Louisiana, Maine, Maryland, Massachusetts, Michigan, Minnesota, Mississippi, Missouri, Montana, Nebraska, Nevada, New Hampshire, New Jersey, New Mexico, New York, North Carolina, North Dakota, Ohio, Oklahoma, Oregon, Pennsylvania, Rhode Island, South Carolina, South Dakota, Tennessee, Texas, Utah, Vermont, Virginia, Washington, West Virginia, Wisconsin, Wyoming

For a more thorough review, the AI Index also included AI-related state laws listed on the Multistate AI Legislation Tracker, even if they did not specifically reference "artificial intelligence" as a keyword.

## US AI Regulation

This section examines AI-related regulations enacted by U.S. regulatory agencies from 2016 to 2025, analyzing the total number of regulations and their originating agencies. To compile this data, the AI Index conducted a keyword search for "artificial intelligence" on the Federal Register, a comprehensive repository of government documents drawn from over 436 agencies and nearly every branch of the U.S. government.

## US Congressional Hearings

This captures witness participation in U.S. congressional hearings on artificial intelligence, spanning the 115th Congress (Jan. 3, 2017) through the 119th Congress (as of Jan. 30, 2025). The data was collected primarily from Congress.gov and, when needed, supplemented with committee websites and transcripts from the U.S. Government Publishing Office. Hearings were identified through a manual, keyword-based search of titles and descriptions (e.g., "artificial intelligence," "machine learning," "algorithm," "automation," "generative AI"), then manually validated to ensure AI was the hearing's central focus. The unit of analysis is the individual witness appearance, yielding 425 observations across 107 hearings. For each observation, the AI Index recorded hearing-level information (congressional session, date, name, event number, serial number, subject) and witness-level attributes (e.g., name, organization, and categorized affiliation).

## Public Investment in AI

The AI Index analyzed government AI spending across European countries and the United States, focusing on regions where data is more accessible. It is important to note that this analysis may not fully represent all countries or regions, as the availability and quality of data can vary significantly. Additionally, while this analysis includes data on government contracts from various countries, it only covers grant-level spending for the United States. This discrepancy is the result of challenges in collecting comparable grant data from other countries and regions, such as the European Union and China. Nevertheless, the U.S. case illustrates that a substantial portion of government spending on AI occurs through grants. Coverage will expand in future iterations of the AI Index as more data becomes available, but discrepancies and gaps in the existing data may affect the comprehensiveness and accuracy of the findings.

**Data Sources**

For European countries, the AI Index collected public tender data from Tenders Electronic Daily (TED) (Publications Office of the European Union, 2024)—the online supplement to the official journal of the EU dedicated to European public procurement. While contracts are available in various formats, the most detailed data comes from bulk XML downloads, which include comprehensive information on tendering procedures, issuing entities, awarded contractors, lot values, descriptions, award dates, and common procurement vocabulary (CPV) codes. TED publication is governed by EU law thresholds: Tenders above specific monetary values, deemed of cross-border interest, must be published on TED. However, some countries also report below-threshold procurements, leading to variations in coverage across countries.

For the United Kingdom, data sources include TED, Find a Tender, Contracts Finder, and Contracts Finder Archive. Data from Scotland and Wales was accessed via the APIs of their procurement websites, while Northern Ireland does not offer this service, necessitating its exclusion from the analysis and potentially leading to an underestimation of public investments in AI for the U.K. Due to API limitations restricting historical data access, the AI Index utilized the Open Contracting Partnership's data registry via Kingfisher Collect to obtain comprehensive data for Scotland and Wales.

Contracts and OTAs data for the United States was sourced from Omari et al. (2025) and from FPDS official API. Grants data was sourced from the publicly accessible USAspending platform, an official repository that facilitates bulk downloads of information related to contract award notices and grant data. While this dataset encompasses a longer time frame than the TED dataset, it is important to note that data quality can vary. Additionally, a study by the U.S. Government Accountability Office (GAO, 2023) found that 49 agencies, including 25 in the executive branch, did not report data—accounting for over $5 billion in net outlays for fiscal year 2022—to USAspending.

**Data Processing**

Processing TED data posed significant challenges due to inconsistent storage of contract descriptions, which varied by XML tag names based on release time and procurement type. Some files contained aggregated descriptions while others detailed each awarded contract lot. To capture comprehensive information, the main descriptions of each competition call were combined with partial descriptions when available.

The linguistic diversity in data from different countries required translating all texts into English using the deep-translator tool and the Google Translator engine. Post-translation, tender texts were processed using natural language processing (NLP) techniques. These included the removal of stop words and special characters, part-of-speech (POS) tagging to retain key grammatical categories, lowercase conversion, lemmatization, and replacement of numerical measures with a <NUM> tag.

U.S. data provides transaction-level information for multi-year awards. In earlier versions, when matching on a single transaction, we reconstructed the full award life cycle in terms of obligations to mitigate the effects of past de-obligations (unrelated to current matches). In recent years, however, AI-related activities have increasingly been added to grants and tenders that were awarded earlier, resulting in a retroactive inflation of estimated investment levels. The AI Index team therefore adopted an alternative approach that aggregates only transactions occurring after each AI match for a given awarded procedure, while controlling for the extreme effects of prior de-obligations (i.e., dropping all de-obbligations reported within the first year[6] of the first AI-transaction in time for each award). This approach preserves the time trends observed in earlier periods while reducing the total estimated stock of investments, particularly for the initial sample period.

**Classification**

Classifying AI-related contracts and grants was achieved using full-text search with regular expressions. An AI dictionary was compiled by generating AI-related expressions and incorporating "core" expressions from the Yamashita et al. (2021) vocabulary. Additionally, a Word2Vec model expanded the dictionary with cosine-similar terms for each baseline expression that were manually reviewed and included in the final vocabulary. This process provided keywords and co-occurrence patterns crucial for identifying AI content.

The classification followed a multistep approach. Initially, regular expression (regex) matching identified AI terms within contract and grant awards. These documents were then categorized as either "non AI-related" or "AI-related." To validate AI-related matches, BERTopic model and pretrained DeBERTA transformer were employed to assess probability scores for specific AI-related topics. Awards with relevance scores below 20% underwent manual review, while those with higher scores were confirmed as AI-related. To ensure additional accuracy, all high-value tenders were also manually reviewed.

6 Closeout action period defined by CFR regulation.

# Chapter 9: Public Opinion

## Ipsos

For brevity, the 2025 AI Index does not republish the methodology used for the Ipsos survey featured in the report. More details about the Ipsos surveys' methodology can be found in the survey reports themselves: 2022, 2023, 2024, 2025, 2026 (Ipsos/Google Muti-Country AI Survey).